%% file: critical.tex
\pdfoutput=1
\documentclass[11pt,twoside]{book} 
\usepackage[hyperfootnotes=false, linktoc=section]{hyperref}
\usepackage[a4paper]{geometry} %
\usepackage[numbers,sort&compress]{natbib}

\usepackage{color}
\usepackage{graphicx}
\usepackage{amsmath}
\usepackage{amssymb}
\usepackage{xspace}
\usepackage[small]{subfigure}
\usepackage[normalem]{ulem}

\usepackage{lmodern} %
\usepackage{mciteplus}
\usepackage{tikzsymbols} %
\usepackage[nottoc]{tocbibind} %
\newcommand{\strangechapter}[1]{\renewcommand{\thechapter}{#1}\chapter} %
\newcommand{\epiloguechapter}{
\renewcommand{\chaptername}{Epilogue}
\renewcommand{\thechapter}{$\varepsilon$}
\chapter
}

\usepackage{makeidx}
\makeindex

\usepackage{epigraph}
\usepackage{etoolbox}
\makeatletter
\newlength\epitextskip
\pretocmd{\@epitext}{\em}{}{}
\apptocmd{\@epitext}{\em}{}{}
\patchcmd{\epigraph}{\@epitext{#1}\\}{\@epitext{#1}\\[\epitextskip]}{}{}
\makeatother
\newcommand{\sbreak}{
    \begin{center}
        $\blacklozenge$$\blacklozenge$$\blacklozenge$$\blacklozenge$
    \end{center}
}
\newcommand{\term}[1]{\textbf{#1}\index{#1|textbf}}
\newcommand{\neo}[1]{\emph{#1}\index{#1}}
\newcommand{\terminate}[1]{#1\index{#1}}
\newcommand{\be}{\begin{equation}}
\newcommand{\ee}{\end{equation}}
\newcommand{\bea}{\begin{eqnarray}}
\newcommand{\eea}{\end{eqnarray}}
\newcommand{\bel}{\begin{align}}
\newcommand{\eel}{\end{align}}
\newcommand{\bi}{\begin{itemize}}
\newcommand{\ei}{\end{itemize}}
\renewcommand{\le}{\left}
\newcommand{\ri}{\right}

\definecolor{green}{rgb}{0.0, 0.64, 0.0}
\definecolor{purple}{rgb}{0.4,.2,0.7}
\definecolor{orange}{rgb}{1.0,0.65,0}
\definecolor{darkorange}{rgb}{1.0,0.40,0}

\renewcommand{\ker}{K} %
\newcommand{\kerm}{k} %
\newcommand{\mkerm}{\mu}%
\newcommand{\kermsub}{\widetilde{k}} %
\newcommand{\fea}{\phi} %
\newcommand{\featwo}{\psi} %
\newcommand{\feathree}{\Psi} %
\newcommand{\feaE}{\phi^{\text{E}}} %
\newcommand{\kermE}{k^{\text{E}}} %
\newcommand{\kermA}{\kerm^\sharp}%
\newcommand{\kermAsub}{\widetilde{\kermA}} %
\newcommand{\NTKMA}{\NTKM^\sharp}%
\newcommand{\NTKMAsub}{\widetilde{\NTKMA}} %
\newcommand{\NTKdd}{\mathbb{H}} %
\newcommand{\bra}{\le\langle}
\newcommand{\ket}{\ri\rangle}
\newcommand{\braket}[2]{\bra #1 \ket_{K^{(#2)}}} %
\newcommand{\brabra}{\le\langle\!\!\le\langle}
\newcommand{\ketket}{\ri\rangle\!\!\ri\rangle}

\newcommand{\E}[1]{\mathbb{E}\le[#1\ri]}
\newcommand{\Ec}[2]{\mathbb{E}\le[#1\ri]#2_{\text{connected}}}
\newcommand{\entropy}{\mathcal{S}} %
\newcommand{\surprise}{s} %
\newcommand{\outcomes}{\mathcal{X}} %
\newcommand{\cov}[2]{\text{Cov}\!\le[#1 , \, #2 \ri]} %
\newcommand{\dete}[1]{\le\vert#1\ri\vert} %

\newcommand{\ND}{N_{\D}} %
\newcommand{\NR}{N_{\A}} %
\newcommand{\tra}{\tilde{\alpha}}%
\newcommand{\tea}{\dot{\beta}} %
\newcommand{\kersub}{\widetilde{\ker}}%
\newcommand{\NTKIsub}{\widetilde{\NTKI}}%
\newcommand{\NTKMsub}{\widetilde{\NTKM}}%
\newcommand{\kerpos}{\mathbb{\ker}}%
\newcommand{\LLmax}{\mathbb{Y}} %
\newcommand{\posGPmean}{m^{\infty}} %
\newcommand{\posmean}{m} %
\newcommand{\GDGPmean}{m^{\infty}} %
\newcommand{\gsub}{\widetilde{g}}%
\newcommand{\gpos}{\mathbb{G}}%
\newcommand{\meanstring}{\Phi} %
\newcommand{\MI}{\mathcal{I}} %
\newcommand{\acfree}{\ac_{\text{F}}} %
\newcommand{\acvar}{\ac_{\text{I}}} %
\newcommand{\dO}{\overline{\dbar \O}}%
\newcommand{\Kdi}[1]{\Ti{\ker}{d}{#1}}%
\newcommand{\NTKIdi}[1]{\Ti{\NTKI}{d}{#1}}%
\newcommand{\Kdif}{\ker_{d}^{\star}}%
\newcommand{\dz}[3]{\Tia{\dbar z}{#1}{#2}{#3}} %
\newcommand{\dtheta}{\dbar \theta} %
\newcommand{\FPV}{V} %
\newcommand{\SPV}{U} %
\newcommand{\SPC}{u} %
\newcommand{\se}[2]{G_{#1}^{\le\{1\ri\}\le(#2\ri)}}
\newcommand{\PH}{G^{(0)}} %
\newcommand{\ratio}{r}%
\newcommand{\dimpre}{N} %
\newcommand{\tanhA}{\texttt{tanh}\index{activation function!tanh@$\texttt{tanh}$}} 
\newcommand{\sinA}{\texttt{sin}\index{activation function!sin@$\texttt{sin}$}}
\newcommand{\relu}{\texttt{ReLU}\index{activation function!ReLU@$\texttt{ReLU}$}} 
\newcommand{\lrelu}{\texttt{leaky ReLU}\index{activation function!leaky ReLU@$\texttt{leaky ReLU}$}} 
\newcommand{\perc}{\texttt{perceptron}\index{activation function!perceptron@$\texttt{perceptron}$}} 
\newcommand{\sigmoid}{\texttt{sigmoid}\index{activation function!sigmoid@$\texttt{sigmoid}$}} 
\newcommand{\linear}{\texttt{linear}\index{activation function!linear@$\texttt{linear}$}} 
\newcommand{\softplus}{\texttt{softplus}\index{activation function!softplus@$\texttt{softplus}$}} 
\newcommand{\gelu}{\texttt{GELU}\index{activation function!GELU@$\texttt{GELU}$}} 
\newcommand{\swish}{\texttt{SWISH}\index{activation function!SWISH@$\texttt{SWISH}$}} 
\newcommand{\block}{\texttt{R}} %
\newcommand{\layer}{\texttt{L}} %
\newcommand{\HS}{\Lambda} %
\newcommand{\GI}[1]{I_{#1}}
\newcommand{\PF}[1]{Z_{#1}}
\newcommand{\PW}{\mathcal{W}}%
\newcommand{\ZU}{\zeta}%
\newcommand{\Vdot}{\cdot}%
\newcommand{\Hypo}{\mathcal{H}} %
\newcommand{\Unit}{U}
\newcommand{\Iden}{I}

\newcommand{\zF}[2]{z^{\text{F}}_{#1;#2}}
\newcommand{\zI}[2]{z^{\text{I}}_{#1;#2}}
\newcommand{\eF}[2]{\epsilon^{\text{F}}_{#1;#2}}
\newcommand{\force}{\mathbb{F}}
\newcommand{\WF}{W^{\text{F}}}
\newcommand{\WI}{W^{\text{I}}}
\newcommand{\Ti}[3]{#1_{#2}^{(#3)}} %
\newcommand{\TI}[3]{#1^{#2}_{(#3)}} %
\newcommand{\tia}[3]{#1_{#2; #3}} %
\newcommand{\Tia}[4]{#1_{#2; #3}^{(#4)}} %
\newcommand{\dti}[2]{#1_{#2}^{\prime}} %
\newcommand{\dTi}[3]{#1_{#2}^{\prime\,(#3)}} %
\newcommand{\dTia}[4]{#1_{#2; #3}^{\prime\,(#4)}} %
\newcommand{\ddTia}[4]{#1_{#2; #3}^{\prime\prime\,(#4)}} %
\newcommand{\x}[2]{\tia{x}{#1}{#2}} %
\newcommand{\y}[2]{\tia{y}{#1}{#2}} %
\newcommand{\bias}[2]{\Ti{b}{#1}{#2}} %
\newcommand{\W}[2]{\Ti{W}{#1}{#2}} %
\newcommand{\Cb}[1]{\Ti{C}{b}{#1}} %
\newcommand{\CW}[1]{\Ti{C}{W}{#1}} %
\newcommand{\cn}[3]{c_{#1}^{(#2)\le\{#3\ri\}}} %
\newcommand{\cbn}[2]{\cn{b}{#1}{#2}} %
\newcommand{\cWn}[2]{\cn{W}{#1}{#2}} %
\newcommand{\lamWtil}[1]{\frac{\Ti{\lambda}{W}{#1}}{n_{#1-1}}} %
\newcommand{\lamW}[1]{\Ti{\lambda}{W}{#1}} %
\newcommand{\z}[3]{\Tia{z}{#1}{#2}{#3}} %
\newcommand{\zi}[2]{\Ti{z}{#1}{#2}} %
\newcommand{\zNL}[2]{\tia{z}{#1}{#2}} %
\newcommand{\s}[3]{\Tia{\sigma}{#1}{#2}{#3}} %
\newcommand{\dsNL}[1]{\dti{\sigma}{#1}} %
\newcommand{\ds}[3]{\dTia{\sigma}{#1}{#2}{#3}} %
\newcommand{\dds}[3]{\ddTia{\sigma}{#1}{#2}{#3}} %
\renewcommand{\o}[1]{O\!\le(#1\ri)} %
\newcommand{\oninv}{\o{\frac{1}{n}}} %
\newcommand{\M}{0}%
 
\newcommand{\KML}{K_{00}^{(\ell)}}

\newcommand{\V}[3]{\Ti{\FPV}{(#1)(#2)}{#3}} %
\newcommand{\VU}[3]{\TI{\FPV}{(#1)(#2)}{#3}} %
\newcommand{\Tif}[2]{#1_{#2}^{\star}} %
\renewcommand{\L}{\mathcal{L}} %
\newcommand{\Laux}[1]{\L_{#1,\, \kappa}} %

\newcommand{\NTH}{\widehat{H}} %
\newcommand{\NTKM}{H} %
\newcommand{\dbar}{d\hspace*{-0.1em}\bar{}\hspace*{0.15em}}
\newcommand{\dNTKM}{\text{d}H}%
\newcommand{\dNTK}{\widehat{\dNTKM}}%
\newcommand{\ddNTKM}{\text{dd}_{\text{I}}H}%
\newcommand{\ddNTK}{\widehat{\ddNTKM}}%
\newcommand{\ddNTKMII}{\text{dd}_{\text{II}}H}%
\newcommand{\ddNTKII}{\widehat{\ddNTKMII}}%

\newcommand{\DNTK}[3]{\Ti{\widehat{\Delta H}}{#1;#2}{#3}} %
\newcommand{\DNTKS}{\widehat{\Delta H}} %
\newcommand{\NTK}{\NTH} %
\newcommand{\NTHA}[3]{\Ti{A}{(#1)(#2)}{#3}} %
\newcommand{\NTHAwo}[1]{A^{(#1)}} %
\newcommand{\NTHB}[2]{\Ti{B}{#1}{#2}} %
\newcommand{\NTHBwo}[1]{B^{(#1)}} %
\newcommand{\NTHD}[2]{\Ti{D}{#1}{#2}} %
\newcommand{\NTHDwo}[1]{D^{(#1)}} %
\newcommand{\NTHF}[2]{\Ti{F}{#1}{#2}} %
\newcommand{\NTHFwo}[1]{F^{(#1)}} %
\newcommand{\dNTKP}[2]{\Ti{P}{#1}{#2}} %
\newcommand{\dNTKQ}[2]{\Ti{Q}{#1}{#2}} %
\newcommand{\ddNTKRS}{R} %
\newcommand{\ddNTKSS}{S} %
\newcommand{\ddNTKTS}{T} %
\newcommand{\ddNTKUS}{U} %
\newcommand{\ddNTKR}[2]{\Ti{\ddNTKRS}{#1}{#2}} %
\newcommand{\ddNTKS}[2]{\Ti{\ddNTKSS}{#1}{#2}} %
\newcommand{\ddNTKT}[2]{\Ti{\ddNTKTS}{#1}{#2}} %
\newcommand{\ddNTKU}[2]{\Ti{\ddNTKUS}{#1}{#2}} %
\newcommand{\reg}{\widehat{\mathcal{R}} } %
\newcommand{\Oi}[2]{\Ti{\widehat{\Omega}}{#1}{#2}} %
\renewcommand{\O}{\mathcal{O}} %
\newcommand{\D}{\mathcal{D}} %
\newcommand{\A}{\mathcal{A}} %
\newcommand{\B}{\mathcal{B}} %
\newcommand{\Lb}[1]{\Ti{\lambda}{b}{#1}} %
\newcommand{\LW}[1]{\Ti{\lambda}{W}{#1}} %
\newcommand{\NTKI}{\mathrm{\Theta}}
\newcommand{\gen}{\mathcal{E}} %
\newcommand{\geosumone}{X_\text{I}} %
\newcommand{\geosumtwo}{X_\text{II}} %
\newcommand{\geosumthree}{X_\text{III}} %
\newcommand{\dampsumNTKminus}{Y_2} %
\newcommand{\dampsumDNTK}{Y_1} %
\newcommand{\dampsumNTKone}{Y_3} %
\newcommand{\dampsumNTKtwo}{Y_4} %
\newcommand{\algodNTKone}{Z_\text{A}} %
\newcommand{\algodNTKtwo}{Z_\text{B}} %
\newcommand{\algoddNTKIone}{Z_\text{IA}} %
\newcommand{\algoddNTKItwo}{Z_\text{IB}} %
\newcommand{\algoddNTKIIone}{Z_\text{IIA}} %
\newcommand{\algoddNTKIItwo}{Z_\text{IIB}} %
\newcommand{\td}{d}%
\newcommand{\ac}{S} %
\newcommand{\SGP}{\ac_{\text{M}}} %
\newcommand{\SI}{\ac_{\text{I}}} %
\newcommand{\EFT}[1]{\ac\le(z^{(#1)}\ri)} %
\newcommand{\Kinv}[2]{\TI{G}{#1}{#2}}%
\newcommand{\SKinv}[2]{\TI{\widehat{G}}{#1}{#2}}%
\newcommand{\dK}{\widehat{\Delta G}}%
\newcommand{\dKi}[2]{\Ti{\dK}{#1}{#2}} %
\newcommand{\JW}[1]{\mathcal{J}_{#1}} %
\newcommand{\dthetaI}{\le(\theta^\star - \theta\ri)} %

\DeclareMathOperator*{\argmax}{arg\,max}
\DeclareMathOperator*{\argmin}{arg\,min}

\begin{document} 
\frontmatter
\include{preface/title}
\cleardoublepage %
\include{preface/contents}
\cleardoublepage

\include{preface/dirac}

\cleardoublepage
\mainmatter
\include{Chp0-intro/0_global}

\include{Chp1-prep/1_global}

\include{Chp2-MLP/2_global}

\include{Chp3-DLN/3_global}

\include{Chp4-NGP/4_global}

\include{Chp5-SignalProp/5_global}

\include{Chp6-Bayesian/6_global}

\include{Chp7-GD/7_global}

\include{Chp8-NTK/8_global}

\include{Chp9-NTKEFT/9_global}

\include{Chp10-kernel/10_global}

\include{Chp11-features/11_global}

\include{ChpInfinity-End/infinity_global}
\include{ChpEpsilon-epilogue/epsilon_global}

\appendix

\include{AppA-thermo/A_global}

\include{AppB-Residual/B_global}

\backmatter
\cleardoublepage
\renewcommand\bibname{References}
\mciteSetMidEndSepPunct{}{\ifmciteBstWouldAddEndPunct.\else\fi}{\relax}
\bibliographystyle{utphys}
\bibliography{critical.bib}{}
\cleardoublepage
\phantomsection %
\printindex
\end{document}

%% file: preface/title.tex
\title{\textbf{The Principles of Deep Learning Theory} \\~\\ \large{\emph{An Effective Theory Approach to Understanding Neural Networks}}}

\author{Daniel A. Roberts
and Sho Yaida
 \\~\\ \normalsize{\emph{based on research in collaboration with}} \\~\\ Boris Hanin  \\~\\~\\
\texttt{drob@mit.edu, shoyaida@fb.com}
}

\date{}

\maketitle

%% file: preface/contents.tex
\tableofcontents

%% file: preface/dirac.tex
\chapter{Preface}\label{sec:meta-learning}
\epigraph{This has necessitated a complete break from the historical line of development, but this break is an advantage through enabling the approach to the new ideas to be made as direct as possible.}{P.~A.~M.~Dirac in the 1930 preface of \emph{The Principles of Quantum Mechanics} \cite{dirac1930principles}.\index{Dirac, Paul Adrien Maurice}}

\noindent{}This is a research monograph in the style of a textbook about the theory of deep learning. While this book might look a little different from the other deep learning books that you've seen before, we assure you that it is appropriate for everyone with knowledge of linear algebra, multivariable calculus, and informal probability theory, and with a healthy interest in neural networks.
Practitioner and theorist alike, we want all of you to enjoy this book.
Now, let us tell you some things. %

First and foremost, in this book we've strived for pedagogy in every choice we've made, placing intuition above formality. 
This doesn't mean that calculations are incomplete or sloppy; quite the opposite, we've tried to provide full details of every calculation -- of which there are certainly very many -- and place a particular emphasis on the tools needed to carry out related calculations of interest. In fact, understanding how the calculations are done is as important as knowing their results, and thus often our pedagogical focus is on the details therein.

Second, while we present the details of all our calculations, we've kept the experimental confirmations to the privacy of our own computerized notebooks. 
Our reason for this is simple: while there's much to learn from explaining a derivation, there's  not much more to learn from printing a verification plot that shows two curves lying on top of each other.  Given the simplicity of modern deep-learning codes and the availability of compute, it's easy to verify any formula on your own; we certainly have thoroughly checked them all this way, so if knowledge of the existence of such plots are comforting to you, know at least that they do exist on our personal and cloud-based hard drives. %

Third, our main focus is on realistic models that are used  by the deep learning community in practice:
we want to study \emph{deep} neural networks. In particular, this means that
\emph{(i)}
a number of special results on single-hidden-layer networks will not be discussed and
\emph{(ii)}
the \emph{infinite-width limit} of a neural network 
-- which corresponds to a zero-hidden-layer network --
will be introduced only as a starting point. All such idealized models will eventually be \emph{perturbed} until they correspond to a real model.
We certainly acknowledge that there's a vibrant community of deep-learning theorists devoted to exploring different kinds of idealized theoretical limits.
However, our interests are fixed firmly on providing explanations for the tools and approaches used by practitioners, 
in an effort to shed light on what
makes them work so well.

Fourth, a large part of the book is focused on deep multilayer perceptrons. We made this choice in order to pedagogically illustrate the power of the effective theory framework --  not due to any technical obstruction -- and along the way we give pointers for how this formalism can be extended to other architectures of interest. In fact, we expect that many of our results have a broad applicability, and we've tried to focus on aspects that we expect to have lasting and universal value to the deep learning community.

Fifth, while much of the material is novel and appears for the first time in this book, and while much of our framing, notation, language, and emphasis breaks with the historical line of development, we're also very much indebted to the deep learning community. 
With that in mind, throughout the book 
we will try to reference important prior contributions, with an emphasis on recent seminal deep-learning results rather than on being completely comprehensive. Additional references for those interested can easily be found within the work that we cite.

Sixth, this book initially 
grew out of a research project in collaboration with Boris Hanin.
To account for his effort and then support,
we've accordingly commemorated him on the cover. More broadly, we've variously appreciated the
artwork,
discussions, 
encouragement,
epigraphs, %
feedback,
management,
refereeing,
reintroduction,
and support
from Rafael Araujo, L\'{e}on Bottou, Paul Dirac\index{Dirac, Paul Adrien Maurice}, Ethan Dyer, John Frank, Ross Girshick, 
Vince Higgs, Yoni Kahn, Yann LeCun, Kyle Mahowald, Eric Mintun, Xiaoliang Qi, Mike Rabbat, David Schwab, Stephen Shenker, Eva Silverstein, PJ Steiner, DJ Strouse, 
and Jesse Thaler. Organizationally, we're grateful to FAIR\index{Facebook AI Research} and Facebook, Diffeo and Salesforce, MIT and IAIFI, and Cambridge University Press and the arXiv. %

Seventh, given intense
(and variously uncertain) 
spacetime and energy-momentum
commitment that writing this book entailed,
Dan is grateful to Aya, Lumi, and Lisa Yaida; from the dual sample-space\index{sample space} perspective, Sho is grateful to Adrienne Rothschilds and would be
retroactively grateful to any
hypothetical future Mark or
Emily
that would have otherwise been thanked in this paragraph.

Eighth, %
we hope that this book spreads our
optimism that
it \emph{is} possible to have
a
general theory of deep learning, 
one that’s both derived from first principles and at the same time focused on describing 
how realistic models actually work: nearly-simple phenomena in practice should correspond to nearly-simple  effective theories.
We %
dream that
this type of thinking
will not only 
lead to  more
\emph{[redacted]} AI models
but also guide us towards a unifying framework for understanding universal aspects of intelligence.

As if that \terminate{eightfold way}\index{eightfold way|seealso{Gell-Mann, Murray}} of prefacing the book wasn't nearly-enough already, please note: 
this book has a website,\index{website|see{\texttt{deeplearningtheory.com}}}
\href{https://deeplearningtheory.com}{\texttt{deeplearningtheory.com}},
and you may want to visit it in order to determine whether the error that you just discovered is already common knowledge. If it's not, please let us know.
There may be pie.\\

\hfill\begin{minipage}{0.29\linewidth}
\emph{Dan Roberts \& Sho\index{simple harmonic oscillator|seealso{Sho}}\index{simple harmonic oscillator} Yaida}\\
\emph{Remotely Located}\\
\emph{June, 2021}
\end{minipage}

%% file: Chp0-intro/0_global.tex
\chapter{Initialization}\index{initialization (of you)}
\label{ch:introduction}

\epigraph{The simulation is such that [one] generally perceives the sum of many billions of elementary processes simultaneously, so that the leveling law of large numbers completely obscures the real nature of the individual processes.}{John von Neumann\index{von Neumann, John} \cite{von2018mathematical}}

\noindent{}Thanks to substantial investments into computer technology, modern \term{artificial intelligence} (AI) systems can now come equipped with many billions of elementary components. When these components are properly 
initialized 
and then trained, AI can accomplish tasks once considered so incredibly complex that philosophers have previously argued that only \emph{natural} intelligence systems -- i.e.~humans -- could perform them.

Behind much of this success in AI is \term{deep learning}. Deep learning uses artificial \textbf{neural networks}\index{neural network|textbf} as an underlying model for AI: while loosely based on \emph{biological} neural networks\index{biological neural network|seealso{brain}}\index{biological neural network} such as your \terminate{brain}, \emph{artificial} neural networks are probably best thought of as an especially nice way of specifying a flexible set of functions, built out of many basic computational blocks called \textbf{neurons}\index{neuron|textbf}. This model of computation is actually quite different from the one used to power the computer you're likely using to read this book. In particular, rather than \emph{programming} a specific set of instructions to solve a problem directly, deep learning models  are \emph{trained}
on data from the real world and learn how to solve problems.

The real power of the deep learning framework comes from \emph{deep} neural networks, with many neurons in parallel organized into sequential computational layers,  \emph{learning} useful representations of the world. 
Such \term{representation learning} transforms data into increasingly refined forms that are helpful for solving an underlying task, and is thought to be a hallmark of success in intelligence, both artificial and biological.

Despite these successes and the intense interest they created, deep learning \emph{theory} is still in its infancy. 
Indeed, there is a serious disconnect between theory and practice: while practitioners have reached amazing milestones, they have far outpaced the theorists,
whose analyses often involve assumptions
so unrealistic that they lead to conclusions that are irrelevant to understanding deep neural networks as they are typically used. 
More importantly,
very little theoretical work directly confronts
the \emph{deep} of deep learning, despite a mass of empirical evidence for its importance in the success of the framework.

\index{principle!InfoMax|see{InfoMax principle}}
\index{principle!of indifference|see{Laplace's principle of indifference}}
\index{principle!maximum entropy|see{maximum entropy, principle}}
\index{principle!learning-rate equivalence|see{equivalence principle}}
\index{principle!criticality|see{criticality}}
\index{principle!sparsity and near-sparsity|see{sparsity, principle of}}
\index{principle!typicality|see{typicality}}
\index{principle!variational|see{variational principle}}

The goal of this book is to put forth
a set of \textbf{principles}\index{principle|textbf} that enable us to theoretically analyze \emph{deep} neural networks of \emph{actual relevance}.  
To initialize you to this task, in the rest of this chapter
we'll explain at a very high-level both \emph{(i)} why such a goal is even attainable in theory 
and \emph{(ii)} how we are able to get there in practice.

\section{An Effective Theory Approach}\label{sec:ET-approach}
\epigraph{Steam navigation brings nearer together the most distant nations. \ldots their theory is very little understood, and the attempts to improve them are still directed almost by chance. \ldots We propose now to submit these questions to a deliberate examination.}{Sadi Carnot, commenting on the need for a theory of deep learning \cite{carnot1890reflections}.\index{Carnot, Sadi}}

\index{microscopic perspective}\index{microscopic perspective|seealso{parameter space}}
\noindent{}While modern deep learning models are built up from seemingly innumerable elementary computational components, a first-principles \emph{microscopic} description of \emph{how} a trained neural network computes a function from these low-level components is entirely manifest.
This microscopic description is just the set of instructions for transforming an input through the many layers of  components into an output. Importantly, during the training process, these components become very finely-tuned, and knowledge of the particular tunings is necessary for a system to produce useful output.

\index{macroscopic perspective}\index{macroscopic perspective|seealso{sample space}}
Unfortunately, the complexity of these tunings obscures any first-principles \emph{macroscopic} understanding of \emph{why} a deep neural network computes a particular function and not another.
With many
neurons performing different tasks as part of such a computation, it seems hopeless 
to think that we can use theory to understand these models at all, and 
silly to believe that
a small set of 
mathematical \emph{principles}
will be sufficient for that job.

Fortunately, \textbf{theoretical physics}\index{physics|textbf} has a long tradition of finding simple \textbf{effective theories}\index{effective theory|textbf} 
of complicated systems with a large number of components. 
The immense success of the program of physics in modeling our physical universe 
suggests that perhaps some of the same tools may be useful for theoretically understanding deep neural networks.
To motivate this connection, let's very briefly reflect on the successes of thermodynamics and statistical mechanics, physical theories that
together explain from microscopic first principles the macroscopic behavior of systems with many elementary constituents.

A scientific consequence of the 
\terminate{Industrial Age}, %
\textbf{thermodynamics}\index{thermodynamics} arose out of an effort to describe and innovate upon the \terminate{steam engine} -- a system consisting of many many particles and perhaps the original \neo{black box}. The laws of thermodynamics, derived from careful empirical observations, were used to codify the mechanics of steam, providing a high-level understanding of these macroscopic \emph{artificial} machines that were transforming society. While the advent of thermodynamics led to tremendous improvements in the efficiency of steam power, its laws were in no way
fundamental.

It wasn't until much later that Maxwell\index{Maxwell, James Clerk}, Boltzmann\index{Boltzmann, Ludwig}, and Gibbs\index{Gibbs, J. Willard} provided the missing link between experimentally-derived effective description on the one hand and a first-principles theory on the other hand.
Their \textbf{statistical mechanics}\index{statistical physics|textbf}\index{statistical mechanics|see{statistical physics}} explains how the macroscopic laws of thermodynamics describing human-scale machines could arise \emph{statistically} from the deterministic dynamics of many microscopic elementary constituents. 
From this perspective, the laws of thermodynamics were \emph{emergent} phenomena that only appear from the collective statistical behavior of a very large number of microscopic particles.
In fact, it was the detailed theoretical predictions derived from statistical mechanics that ultimately led to the general scientific acceptance 
that \terminate{matter} is really comprised of molecules\index{molecule} and atoms\index{atom}.
Relentless application of statistical mechanics
led to the discovery of \neo{quantum mechanics}, which 
is a precursor to the
invention of the \terminate{transistor} that powers the \terminate{Information Age}, and 
-- taking the long view -- 
is what has allowed us to 
begin to 
realize artificial machines that can 
think %
intelligently.

Notably, these physical theories originated from a desire to understand \emph{artificial} human-engineered objects, such as the steam engine. Despite a 
potential %
misconception, \terminate{physics}  doesn't make a distinction between natural and artificial phenomena. Most fundamentally, it's concerned with providing a unified set of principles that account for past empirical observations and predict the result of future experiments; the point of theoretical calculations is to connect measurable outcomes or \textbf{observables}\index{observable|textbf} directly to the fundamental underlying constants or \textbf{parameters}\index{model parameters!connection to observables} that define the theory. This perspective also implies a tradeoff between the predictive 
accuracy
of a model and its mathematical tractability, and the former must take precedence over the latter 
for any theory 
to be successful:  a short tether from theory to physical reality is essential.
When successful, such theories provide a comprehensive understanding of phenomena and empower practical advances in technology, as exemplified by the statistical-physics bridge from the Age of Steam to the Age of Information.\index{Information Age}

For our study of deep learning, the key takeaway from this discussion is that 
a theoretical \emph{matter} %
simplifies when it is made up of many elementary constituents. Moreover, unlike the molecules of water contained in a box of steam -- with their existence once being a controversial conjecture in need of experimental verification  -- the neurons comprising a deep neural network are put in
(the box)
by hand. Indeed, in this case we already understand the microscopic laws -- \emph{how} a network computes -- and so instead our task is to understand the new types of regularity that appear at the macroscopic scale -- \emph{why} it computes one particular function rather than another -- that emerge from the statistical properties of these gigantic deep learning models. %

\section{The Theoretical Minimum}\index{Landau, Lev}\label{sec:why-it-works}
\epigraph{The method is more important than the discovery, because the correct method of research will lead to new, even more valuable discoveries.}{Lev Landau\index{Landau, Lev} \cite{landau}.}

\noindent{}In this section, we'll give a high-level overview of our method, providing a minimal explanation for why we should expect a first-principles theoretical understanding of deep neural networks to be possible.
We'll then fill in all the details in the coming chapters.

\begin{figure}
\begin{center}
 \includegraphics[scale=.7]{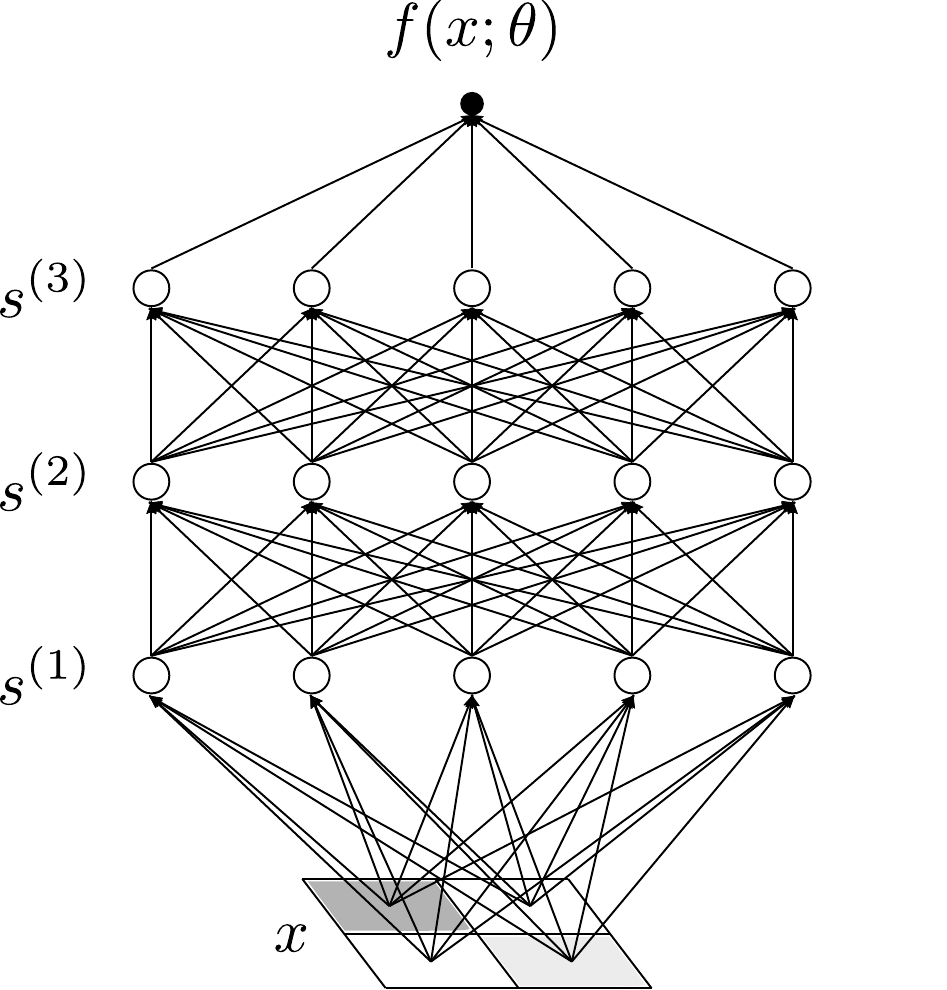}
\caption{A 
graph
of a simple multilayer neural network, depicting how the input $x$ is transformed through a sequence of intermediate signals, $s^{(1)}$, $s^{(2)}$, and $s^{(3)}$, into the output $f(x;\theta)$.  The white circles represent the neurons, the black dot at the top represents the network output, and the parameters $\theta$ are implicit; they weight the importance of the different arrows carrying the signals and bias the firing threshold of each neuron.
}
\label{fig:mlp-simple}
\end{center}
\end{figure}

In essence,
a \term{neural network} is a recipe for computing a function built out of many computational units called \textbf{neurons}\index{neuron}. Each neuron is itself a very simple function that 
considers %
a weighted sum of incoming signals and then \emph{fires} in a characteristic way by comparing the value of that sum against some threshold. 
Neurons are then organized in parallel into \textbf{layers}\index{layer}, and \emph{deep} neural networks are those composed of multiple layers in sequence.
The network  is parametrized by the firing thresholds and the weighted connections between the neurons, and, to give a sense of the potential scale, current state-of-the-art neural networks can have over 100 billion parameters. A graph depicting the structure of a much more reasonably-sized neural network is shown in Figure~\ref{fig:mlp-simple}.

For a moment, let's ignore all that structure and simply think of a neural network as a parameterized function
\be
f(x;\theta)\, ,
\ee
where $x$ is the input to the function and $\theta$ is a vector of a large number of \textbf{parameters}\index{model parameters}\index{parameters|see{model parameters}}\index{model parameters|seealso{biases}}\index{model parameters|seealso{weights}} controlling the shape of the function. 
For such a function to be useful,
we need to somehow tune the high-dimensional parameter vector $\theta$.
In practice, this is done in two steps: 
\bi
\item First, we \emph{initialize} the network by randomly sampling the parameter vector $\theta$ from a computationally simple probability distribution,
\be\label{eq:schematic-initialization}
p(\theta)\, .
\ee
We'll later discuss the theoretical reason why it is a good strategy to have an \term{initialization distribution} $p(\theta)$ but, more importantly, this corresponds to what is done in practice, and our approach in this book is to have our theoretical analysis correspond to realistic deep learning scenarios.
\item Second, we adjust the parameter vector as $\theta \to \theta^\star$, such that the resulting \emph{network function} $f(x;\theta^\star)$ is as close as possible to a desired
\emph{target function} $f(x)$: 
\be\label{eq:function-approximation-def}
f(x;\theta^\star) \approx f(x) \, .
\ee
This is called \term{function approximation}\index{function approximation|seealso{machine learning}}.
To find these tunings $\theta^\star$, we 
fit the network function $f(x;\theta)$ to \textbf{training data}\index{training data|see{training set}}\index{training set|textbf}, consisting of many pairs of the form $\big(x, f(x)\big)$ observed from the desired -- but only partially observable -- target function $f(x)$. Overall, making these adjustments to the parameters is called \term{training}\index{training|seealso{gradient descent}}\index{training|seealso{model fitting}}, and the particular procedure used to tune them is called a \textbf{learning algorithm}.\index{learning algorithm}\index{learning algorithm|seealso{Bayesian inference}}\index{learning algorithm|seealso{gradient descent}}
\ei

Our goal is to understand this \emph{trained} network function:
\be\label{eq:lofty-goal}
f(x;\theta^\star)\, .
\ee
In particular, we'd like to understand the macroscopic behavior of this function
from a first-principles microscopic description of the network in terms of these trained parameters $\theta^\star$. We'd also like to understand how the function approximation \eqref{eq:function-approximation-def} works and evaluate how $f(x;\theta^\star)$ uses the training data $\big(x, f(x)\big)$ in its approximation of $f(x)$.
Given the high dimensionality of the parameters $\theta$ and the degree of fine-tuning required for the approximation \eqref{eq:function-approximation-def}, this goal might seem naive and beyond the reach of any realistic theoretical approach. 

One way to more directly see the kinds of technical problems that we'll encounter 
is to \emph{Taylor expand}\index{Taylor series} our trained network function $f(x;\theta^\star)$ around the initialized value of the parameters $\theta$. Being schematic and ignoring for a moment that $\theta$ is a vector and that the derivatives of $f(x;\theta)$ are tensors\index{tensor}, we see
\begin{align}\label{eq:proto-dynamics}
f(x;\theta^\star) =&f(x;\theta) +\dthetaI \frac{df}{d\theta } +\frac{1}{2} \dthetaI^2  \frac{d^2f }{d\theta^2}+\ldots\,,  
\end{align}
where $f(x;\theta)$ and its derivatives on the right-hand side are all evaluated at initialized value of the parameters. This Taylor representation illustrates our three main problems:
\begin{description}
\item[Problem 1]~\\ 
In general, the series \eqref{eq:proto-dynamics} contains an infinite number of terms
\be\label{eq:grand-probelm-1}
f\, ,\quad \frac{df}{d\theta }\, ,\quad \frac{d^2f }{d\theta^2}\, ,\quad \frac{d^3f }{d\theta^3}\,,\quad \frac{d^4f }{d\theta^4}\,,\quad \dots \, , %
\ee
and to use this Taylor representation of the function \eqref{eq:proto-dynamics}, in principle we need to compute them all. 
More specifically, as the difference between the trained and initialized parameters, $\dthetaI$, becomes large, so too does the number of terms 
 needed to get a good approximation of the trained network function $f(x;\theta^\star)$.
\item[Problem 2]~\\ 
Since the parameters $\theta$ are randomly sampled from the initialization distribution, 
$p(\theta)$,
each time we initialize our network 
we get a different function $f(x;\theta)$.
This means that each term $f$, $df/d\theta$, $d^2f/d\theta^2$, \ldots\,, from \eqref{eq:grand-probelm-1} is really a \emph{random function} of the input $x$.
Thus, 
the initialization
induces a distribution over the network function and its derivatives, and we need to determine the mapping,
\be\label{eq:grand-probelm-2}
p(\theta)\to p\!\le(f, \frac{df}{d\theta }, \frac{d^2f }{d\theta^2},\, \dots  \ri) \, ,
\ee
that takes us from the distribution of initial parameters $\theta$ to the \emph{joint} distribution of the network function, $f(x;\theta)$, its \term{gradient}, $df/d\theta$, its \term{Hessian},  $d^2f/d\theta^2$, and so on. This is a joint distribution comprised of an infinite number of random functions,
and in general such functions will have an intricate statistical dependence.
Even if we set aside this infinity of functions for a moment and consider 
just
the
marginal distribution of the network function \emph{only}, $p(f)$, there's still no reason to expect that it's analytically tractable.
\item [Problem 3]~\\
The learned value of the parameters, $\theta^{\star}$, is the result of a complicated training process. In general, $\theta^{\star}$ is not unique and can depend on
\emph{everything}:
\be\label{eq:grand-probelm-3}
\theta^\star \equiv\le[\theta^{\star}\ri]\!\le(\theta,\, f,\, \frac{df}{d\theta},\, \frac{d^2f}{d\theta^2},\, \ldots;\, \text{learning algorithm};\, \text{training data}\ri)\, .
\ee
In practice, the learning algorithm is \emph{iterative}, accumulating changes over many steps, and the dynamics are nonlinear. Thus, the trained parameters $\theta^{\star}$ will depend in a very complicated way on all the quantities at initialization -- such as the specific random sample of the  parameters $\theta$, the network function $f(x;\theta)$ and all of its derivatives, $df/d\theta$, $d^2f/d\theta^2$, \ldots\, -- as well as on the details of the 
learning algorithm and also on the particular pairs, $\big(x, f(x)\big)$, that comprise the training data. Determining an \emph{analytical} expression for $\theta^\star$ must involve taking all of this into account.
\end{description}
If we could solve all three of these problems, then we could in principle use the Taylor-series representation  \eqref{eq:proto-dynamics} to study the trained network function. More specifically, we'd find a \emph{distribution} over trained network functions
\be\label{eq:lofty-goal-refined}
p(f^{\star})\equiv p\Big(f(x;\theta^\star) \Big\vert\, \text{learning algorithm};\, \text{training data} \Big)\, ,
\ee
now conditioned in a simple way on the learning algorithm and the data we used for training. Here, by \emph{simple} we mean that it is easy to evaluate this distribution for different algorithms or choices of training data without having to solve a version of  \textbf{Problem 3} each time. 
The development of a method for the
analytical computation of \eqref{eq:lofty-goal-refined} is a principle goal of this book.

Of course, solving our three problems for a general parameterized function $f(x;\theta)$ is not tractable. However, we are not trying to solve these problems in general; we only care about the functions that are deep neural networks. Necessarily, any solution to the above problems will thus have to make use of the particular \emph{structure} of neural-network function. %
While specifics of how this works form the basis of the book, in the rest of this section we'll try to give intuition for how these complications can be resolved.

\subsubsection{A Principle of Sparsity}
To elaborate on the structure of neural networks, please scroll back a bit and look at Figure~\ref{fig:mlp-simple}.
Note that for the network depicted in this figure, each intermediate or \emph{hidden} layer consists of five neurons, and the input $x$ passes through three such hidden layers before the output is produced at the top after the final layer. In general, two essential aspects of a neural network \neo{architecture} are its \term{width}, $n$, and its \term{depth}, $L$.

As we foreshadowed in \S\ref{sec:ET-approach}, there are often simplifications to be found in the limit of a large number of components. %
However, it's not enough to consider any massive macroscopic system, and taking the right limit often requires some care.
Regarding the neurons as the components of the network, there are essentially two primal ways that we can make a network grow in size: we can increase its width $n$ holding its depth $L$ fixed, or we can increase its depth $L$ holding its width $n$ fixed.
In this case, it will actually turn out that the former limit will make everything really simple, while the latter limit will be hopelessly complicated and useless in practice.

So let's begin by 
formally taking the limit
\be\label{eq:infinite-width-limit-distribution}
\lim_{n \to \infty} p(f^\star) \, ,
\ee
and studying an \emph{idealized} neural network in this limit.
This is known as the \term{infinite-width limit}\index{infinite-width limit|seealso{not really deep}} of the network, and
as a strict limit
it's rather \emph{unphysical} for a
network: obviously you cannot directly program a function to have an infinite number of components on a finite computer.
However, this extreme limit does massively simplify the distribution over trained networks $p(f^\star)$, rendering each of our three problems completely benign: 
\bi
\item Addressing \textbf{Problem 1}, all the higher derivative terms $d^k f/d\theta^k$ for $k\geq 2$ will effectively vanish, meaning we only need to keep track of two terms,
\be\label{eq:example-of-sparsity-1}
f\, , \quad \frac{df}{d\theta}\, .
\ee
\item Addressing \textbf{Problem 2}, the distributions of these random functions will be independent, 
\be\label{eq:example-of-sparsity-2}
\lim_{n \to \infty}  p\!\le(f, \frac{df}{d\theta}, \frac{d^2f}{d\theta^2},\, \dots  \ri)=p\!\le(f\ri) p\!\le(\frac{df}{d\theta} \ri) \, ,
\ee 
with each marginal distribution factor taking a very simple form. 
\item Addressing \textbf{Problem 3}, the training dynamics become linear and completely independent of the details of the learning algorithm, letting us find a complete analytical solution for $\theta^\star$ in a \emph{closed form}
\be\label{eq:example-of-sparsity-3}
\lim_{n \to \infty}\theta^\star = \le[\theta^{\star}\ri]\!\le(\theta,\, f,\, \frac{df}{d\theta};\, \text{training data}\ri)\, .
\ee
\ei
As a result, the trained distribution \eqref{eq:infinite-width-limit-distribution} is a simple \term{Gaussian distribution} with a nonzero mean, and we can easily analyze the functions that such networks are computing.

These simplifications are the consequence  of a \textbf{principle of sparsity}\index{sparsity, principle of}.
Even though it seems like we've made the network more complicated 
by growing  it to have an infinite number of components, from the perspective of any particular neuron the 
input of an infinite number of signals is such that the leveling law of large numbers completely obscures much of the details in the signals. The result is that the \emph{effective theory} of many such infinite-width networks leads to extreme sparsity 
in their description,
e.g.~enabling the truncation \eqref{eq:example-of-sparsity-1}.

Unfortunately, the formal infinite-width limit, $n \to \infty$, leads to a poor model of \emph{deep} neural networks: not only is infinite width an unphysical property for a network to possess, but the resulting trained distribution \eqref{eq:infinite-width-limit-distribution} also leads to a mismatch between theoretical description and practical observation for networks of more than one layer. In particular, it's empirically known that the distribution over such trained networks \emph{does} depend on the properties of the learning algorithm used to train them. Additionally, we will show in detail that such infinite-width networks 
cannot learn representations of their inputs:
for any input $x$, its transformations in the hidden layers, $s^{(1)}$, $s^{(2)}$, \ldots, will remain
unchanged from 
initialization, leading to \emph{random representations}\index{feature function!random}\index{feature function!random|seealso{random feature model}}\index{random feature function|see{feature function}} and thus severely restricting the class of functions that such networks are capable of learning. Since nontrivial \term{representation learning} is an empirically demonstrated essential property of multilayer networks, this really underscores the breakdown of the correspondence between theory and reality in this strict infinite-width limit.

From the theoretical perspective, the problem with this limit is the washing out of the fine details at each neuron due to the consideration of an infinite number of incoming signals. In particular, such an infinite accumulation  completely eliminates the subtle correlations between neurons that get amplified over the course of training for representation learning. To make progress, we'll need to find a way to restore and then study the \term{interactions} between neurons that are present in realistic \emph{finite-width} networks.

With that in mind, perhaps the infinite-width limit can be corrected in a way such that the corrections become small when the width $n$ is large. To do so, we can use \textbf{perturbation theory}\index{perturbation theory} -- just as we do in \terminate{physics} to analyze interacting systems -- and study deep learning using a \textbf{1/\emph{n} expansion}\index{$1/n$ expansion|textbf}, treating the inverse layer width, $\epsilon \equiv 1/n$, as our small parameter of expansion: $\epsilon \ll 1$. 
In other words, we're going to back off the strict infinite-width limit and compute the trained distribution \eqref{eq:lofty-goal-refined} with the following expansion:
\be\label{eq:finite-width-limit-distribution}
p(f^\star) \equiv p^{\{0\}}(f^\star) + \frac{p^{\{1\}}(f^\star) }{n} + \frac{p^{\{2\}}(f^\star) }{n^2} + \dots \, ,
\ee
where $p^{\{0\}}(f^\star) \equiv \lim_{n\to \infty} p (f^\star)$ is the infinite-width limit we discussed above, \eqref{eq:infinite-width-limit-distribution}, and the $p^{\{k\}}(f^\star)$ for $k \geq 1 $ give a series of corrections to this limit. 

In this book, we'll in particular compute the first such correction, truncating the expansion as
\be\label{eq:finite-width-limit-distribution-truncated}
p(f^\star) \equiv p^{\{0\}}(f^\star) + \frac{p^{\{1\}}(f^\star) }{n} + \o{\frac{1}{n^2}} \, .
\ee
This \neo{interacting theory} is still simple enough to make our three problems tractable: 
\bi
\item Addressing \textbf{Problem~1}, now all the higher derivative terms $d^k f/d\theta^k$ for $k\geq 4$ will effectively give contributions  of the order $1/n^2$ or smaller, meaning that to capture the leading contributions of order $1/n$, we only need to keep track of four terms:
\be\label{eq:example-of-sparsity-finite-n-1}
f\, , \quad \frac{df}{d\theta}\, , \quad \frac{d^2f}{d\theta^2}\,, \quad \frac{d^3f}{d\theta^3}\, .
\ee
Thus, we see that the \emph{principle of sparsity}\index{sparsity, principle of} will still limit the dual\index{duality} effective theory description, though not quite as
extensively
as in the infinite-width limit.
\item Addressing \textbf{Problem~2}, the distribution of these random functions at initialization,
\be\label{eq:example-of-sparsity-finite-n-2}
p\!\le(f, \frac{df}{d\theta }, \frac{d^2f}{d\theta^2 },\frac{d^3f}{d\theta^3}  \ri)\, ,
\ee
will be \emph{nearly} simple at order $1/n$, and we'll be able to work it out in full detail using perturbation theory. 
\item Addressing \textbf{Problem~3}, we'll be able to use a dynamical perturbation theory to tame the nonlinear training  dynamics and find an analytic solution for $\theta^\star$ in a \emph{closed form}:
\be\label{eq:example-of-sparsity-finite-n-3}
\theta^\star=\le[\theta^{\star}\ri]\le(\theta, f,\, \frac{df}{d\theta}\, ,\frac{d^2f}{d\theta^2}\, \frac{d^3f}{d\theta^3};\, \text{learning algorithm};\, \text{training data}\ri)\, .
\ee
In particular, this will make the dependence of the solution on the details of the learning algorithm transparent and manifest.
\ei
As a result, our description of the trained distribution at order $1/n$, \eqref{eq:finite-width-limit-distribution-truncated},  will be a \term{nearly-Gaussian distribution}.

In addition to being analytically tractable, this truncated description at order $1/n$ will satisfy our goal of computing and understanding the distribution over trained network functions
$p(f^\star)$. As a consequence of incorporating the interactions between neurons, this description has a dependence on the details of the learning algorithm and, as we'll see, includes nontrivial representation learning. Thus, \emph{qualitatively}, this effective theory at order $1/n$ corresponds much more closely to realistic neural networks than the infinite-width description, making it far more useful as a theoretically \term{minimal model} for understanding deep learning. 

How about the quantitative correspondence? As there is a sequence of finer descriptions that we can get by computing higher-order terms in the expansion \eqref{eq:finite-width-limit-distribution}, do these terms also need to be included?
 
While the formalism we introduce in the book makes computing these additional terms in the \terminate{$1/n$ expansion} completely systematic -- though perhaps somewhat tedious -- an important byproduct of studying the leading correction is actually a deeper understanding of this truncation error. In particular, what we'll find is that the correct scale to compare with width $n$ is the depth $L$. That is, we'll see that the relative magnitudes of the terms in the expansion \eqref{eq:finite-width-limit-distribution} are given by the depth-to-width aspect ratio: 
\be
r \equiv L/n\, .
\ee
This lets us recast our understanding of infinite-width vs.~finite-width and shallow vs.~deep in the following way:
\bi
\item In the strict limit $r \to 0$, the interactions between neurons turn off: the infinite-width limit \eqref{eq:infinite-width-limit-distribution} is actually a decent description. However, these networks are \neo{not really deep}\index{not really deep|seealso{infinite-width limit}}, as their relative depth is zero: $L/n =0$. 
\item
In the regime $0 < r \ll 1$, there are nontrivial interactions between neurons: the finite-width effective theory truncated at order $1/n$, \eqref{eq:finite-width-limit-distribution-truncated}, gives an accurate accounting of the trained network output.
These networks are \neo{effectively deep}.\index{effectively deep|seealso{optimal aspect ratio}}
\item In the regime $ r \gg 1$, the neurons are strongly coupled: networks will behave chaotically,
and there is no effective description due to large fluctuations from instantiation to instantiation. These networks are \neo{overly deep}.\index{overly deep|seealso{degradation problem}}\index{overly deep|seealso{chaos}}
\ei 
As such, most networks of practical use actually have reasonably small depth-to-width ratios, 
and so our
truncated description at order $1/n$, \eqref{eq:finite-width-limit-distribution-truncated}, will provide a great \emph{quantitative} correspondence as well.\footnote{More precisely, there is an \neo{optimal aspect ratio}\index{optimal aspect ratio|seealso{effectively deep}}, $r^{\star}$, that divides the effective regime $r \leq r^{\star}$ and the ineffective regime $r > r^{\star}$. In Appendix~\ref{app:mi-stuff}, we'll estimate this \terminate{optimal aspect ratio} from an information-theoretic\index{information theory}\index{information theory|seealso{statistical mechanics}} perspective. In Appendix \ref{app:residual}, we'll further show how \emph{residual connections}\index{residual connection} can be introduced to shift the optimal aspect ratio $r^{\star}$ to larger values, making the formerly overly-deep networks 
more practically trainable as well as quantitatively describable by our 
effective theory approach.}

From this, we see that to really describe the properties of \emph{multilayer} neural networks, i.e.~to understand \neo{deep learning}, we need to study large-but-finite-width networks. In this way, we'll be able to find a macroscopic \emph{effective theory} description of realistic deep neural networks.

%% file: Chp1-prep/1_global.tex
\chapter{Pretraining}
\label{ch:tools}

\epigraph{My strongest memory of the class is the very beginning, when he started, not with some deep principle of nature, or some experiment, but with a review of Gaussian integrals. Clearly, there was some calculating to be done.}{Joe Polchinski, reminiscing about Richard Feynman's\index{Feynman, Richard} quantum mechanics class \cite{Polchinski:2017vik}.\index{Polchinski, Joseph}}

\noindent{}The goal of this book is to develop principles that enable a theoretical understanding of deep learning.
Perhaps the most important principle is that wide and deep neural networks are governed by nearly-Gaussian distributions.\index{nearly-Gaussian distribution}\index{perturbation theory}
Thus, to make it through the book, you will need to achieve mastery of Gaussian integration and perturbation theory.
Our \neo{pretraining} in this chapter consists of whirlwind introductions to these  toolkits as well as a brief overview of some key concepts in statistics that we'll need.
The only prerequisite is fluency in linear algebra, multivariable calculus, and rudimentary probability theory.\index{probability (branch of mathematics)}\index{probability (branch of mathematics)|seealso{frequentist probability}}\index{probability (branch of mathematics)|seealso{Bayesian probability}}

With that in mind, we begin in \S\ref{sec:Gauss} with an extended discussion of
Gaussian integrals. Our emphasis will be on calculational tools for computing averages of monomials against Gaussian distributions, culminating in a derivation of \neo{Wick's theorem}.

Next, in~\S\ref{sec:not-Gauss}, we begin by giving a general discussion of \emph{expectation values}\index{expectation value} and \emph{observables}\index{observable}. 
Thinking of observables as a way of learning about a \terminate{probability distribution} through repeated experiments, we're led to the statistical concepts of moment and cumulant and the corresponding physicists' concepts of 
full $M$-point correlator and connected $M$-point correlator. 
A particular emphasis is placed on the connected correlators as they directly characterize a distribution's deviation from Gaussianity.

In~\S\ref{sec:perturbation}, we introduce the negative log probability or \neo{action} representation of a \terminate{probability distribution} and explain how the action lets us systematically deform Gaussian distributions in order to give a compact representation of non-Gaussian distributions. In particular, we specialize to nearly-Gaussian distributions, for which deviations from Gaussianity are implemented by small \emph{couplings}\index{coupling} in the action, and show how perturbation theory can be used to connect the non-Gaussian couplings to observables such as the connected correlators.
By treating such couplings perturbatively, we can transform any correlator of a nearly-Gaussian distribution into a sum of
Gaussian integrals; each integral can then be evaluated by the tools we developed in~\S\ref{sec:Gauss}.
This will be one of our most important tricks, as the neural networks we'll study are all governed by nearly-Gaussian distributions, with non-Gaussian couplings that become perturbatively small as the networks become wide.

Since all these manipulations need to be on our fingertips, in this first chapter we've erred on the side of being verbose -- in words and equations and examples -- 
with the goal of making these materials as transparent and comprehensible as possible.

\section{Gaussian Integrals}\label{sec:Gauss}
\index{probability distribution}\index{Gaussian distribution}
The goal of this section is to introduce Gaussian integrals and Gaussian probability distributions, and ultimately derive Wick's theorem~\eqref{eq:Wick-multi}. This theorem provides an operational formula for computing any moment of a multivariable Gaussian distribution, and will be used throughout the book.

\subsubsection{Single-variable Gaussian integrals}
Let's take it slow and start with the simplest single-variable \terminate{Gaussian function}, %
\be\label{eq:Gaussian-function-single-variable}
e^{-\frac{z^2}{2}}\, .
\ee
The graph of this function depicts the famous \neo{bell curve}\index{bell curve|seealso{Gaussian function}}, symmetric around the peak at $z=0$ and quickly tapering off for large $\vert z \vert\gg1$. By itself, \eqref{eq:Gaussian-function-single-variable} cannot serve as a \terminate{probability distribution} since it's not normalized. In order to find out the proper normalization, we need to perform the \terminate{Gaussian integral}
\be\label{eq:single-variable-gaussian}
\GI{1} \equiv \int_{-\infty}^{\infty} d z\ e^{-\frac{z^2}{2}}\, .
\ee

As an ancient object, there exists a neat trick to evaluate such an integral. To begin, consider its square
\be
\GI{1}^2 = \le(\int_{-\infty}^{\infty}  d z\ e^{-\frac{z^2}{2}}\ri)^2=\int_{-\infty}^{\infty}  d x\ e^{-\frac{x^2}{2}}\int_{-\infty}^{\infty}  d y\ e^{-\frac{y^2}{2}}=\int_{-\infty}^{\infty}  \int_{-\infty}^{\infty}  d x d y\ e^{-\frac{1}{2}\le(x^2+y^2\ri)}\,  ,
\ee
where in the middle we just changed the names of the dummy integration variables. Next, we change variables to polar coordinates $(x,y)= (r \cos \phi,r \sin \phi )$, 
which transforms the integral measure as $dx dy = r d r d\phi$ and gives us two elementary integrals to compute:
\begin{align}
\GI{1}^2 = \int_{-\infty}^{\infty}  \int_{-\infty}^{\infty}  d x d y\ e^{-\frac{1}{2}\le(x^2+y^2\ri)}=&\int_{0}^{\infty} r d r \int_{0}^{2\pi} d \phi\ e^{-\frac{r^2}{2}}\, \\
=&2\pi \int_{0}^{\infty} d r \ r e^{-\frac{r^2}{2}}=2\pi \le\vert -e^{-\frac{r^2}{2}}\ri\vert_{r=0}^{r=\infty}=2\pi \,. \nonumber
\end{align}
Finally, by taking a square root we can evaluate the Gaussian integral \eqref{eq:single-variable-gaussian} as
\be
\GI{1} = \int_{-\infty}^{\infty} d z\ e^{-\frac{z^2}{2}}=\sqrt{2\pi}\, .
\ee
Dividing the Gaussian function with this normalization factor, we define the \textbf{Gaussian probability distribution}\index{Gaussian distribution!single-variable} with unit variance as
\be
p\!\le(z\ri)\equiv\frac{1}{\sqrt{2\pi}}e^{-\frac{z^2}{2}}\, ,
\ee
which is now properly normalized, i.e., $\int_{-\infty}^{\infty} d z\, p\!\le(z\ri)=1$. Such a distribution with zero mean and unit variance is sometimes called the \emph{standard normal distribution}.\index{Gaussian distribution!normal distribution, standard}\index{normal distribution|see{Gaussian distribution}}

Extending this result to a Gaussian distribution with \term{variance}\index{variance|seealso{cumulant}} $\ker>0$ is super-easy. The corresponding \terminate{normalization factor}\index{normalization factor|seealso{partition function}} is given by 
\be\label{eq:single_Gauss}
\GI{\ker}\equiv  \int_{-\infty}^{\infty} d z\ e^{-\frac{z^2}{2\ker}}=\sqrt{\ker}\int_{-\infty}^{\infty} d u\ e^{-\frac{u^2}{2}}=\sqrt{2\pi \ker}\, ,
\ee
where in the middle we rescaled the integration variable as $u=z/\sqrt{\ker}$. We can then define the Gaussian distribution with variance $\ker$ as
\be\label{eq:single_Gauss_with_mean}
p\!\le(z\ri)\equiv\frac{1}{\sqrt{2\pi\ker}}e^{-\frac{z^2}{2\ker}}\, .
\ee
The graph of this distribution again depicts a \terminate{bell curve} symmetric around $z=0$, but it's now equipped with a scale $\ker$ characterizing its broadness, tapering off for $\vert z \vert\gg\sqrt{\ker}$.
More generally, we can shift the center of the bell curve as
\be\label{eq:Gaussian-with-mean}
p\!\le(z\ri)\equiv\frac{1}{\sqrt{2\pi\ker}}e^{-\frac{\le(z-s\ri)^2}{2\ker}}\, ,
\ee
so that it is now symmetric around $z=s$.
This center value $s$ is called the \textbf{mean}\index{mean}\index{mean|seealso{moment}}\index{mean|seealso{cumulant}} of the distribution, because it is:
\begin{align}
\E{z}\equiv \int_{-\infty}^{\infty} d z\ p\!\le(z\ri) z=&\frac{1}{\sqrt{2\pi\ker}} \int_{-\infty}^{\infty}d z\  e^{-\frac{\le(z-s\ri)^2}{2\ker}} z\, \\
=&\frac{1}{\GI{\ker}} \int_{-\infty}^{\infty}d w\ e^{-\frac{w^2}{2\ker}} \le(s+w\ri)\, \nonumber\\
=&\frac{s\GI{\ker}}{\GI{\ker}}+\frac{1}{\GI{\ker}}  \int_{-\infty}^{\infty}d w \le(e^{-\frac{w^2}{2\ker}}w\ri)\, \nonumber\\
=&s\, ,\nonumber
\end{align}
where in the middle we shifted the variable as $w=z-s$ and in the very last step noticed that the integrand of the second term is odd with respect to the sign flip  of the integration variable $w\leftrightarrow -w$ and hence integrates to zero.

Focusing on Gaussian distributions with zero mean, let's consider other \textbf{expectation values}\index{expectation value|textbf} for general functions $\mathcal{O}\!\le(z\ri)$, i.e.,
\be
\E{\mathcal{O}\!\le(z\ri)}\equiv \int_{-\infty}^{\infty} d z\ p\!\le(z\ri) \mathcal{O}\!\le(z\ri)= \frac{1}{\sqrt{2\pi\ker}}\int_{-\infty}^{\infty}d z\  e^{-\frac{z^2}{2\ker}} \mathcal{O}\!\le(z\ri)\, .
\ee
We'll often refer to such functions $\mathcal{O}\!\le(z\ri)$ as \textbf{observables}\index{observable|textbf}, since they can correspond to measurement outcomes of experiments.
A special class of expectation values are called \textbf{moments}\index{moment|textbf}\index{moment|seealso{full correlator}} 
and correspond to the insertion of $z^M$ into the integrand for any integer $M$:
\be
\E{z^{M}}=\frac{1}{\sqrt{2\pi\ker}} \int_{-\infty}^{\infty}d z\  e^{-\frac{z^2}{2\ker}} z^{M}\, .
\ee
Note that the integral vanishes for any odd exponent $M$, because then the integrand is odd with respect to the sign flip $z\leftrightarrow -z$. As for the even number $M=2m$ of $z$ insertions, we will need to evaluate integrals of the form
\be\label{eq:single-variable-z-insertions}
\GI{\ker,m} \equiv \int_{-\infty}^{\infty} d z\ e^{-\frac{z^2}{2\ker}} z^{2m}\, .
\ee
As objects almost as ancient as \eqref{eq:single-variable-gaussian}, again there exists a neat trick to evaluate them:
\begin{align}
\GI{\ker,m}=&\int_{-\infty}^{\infty} d z\ e^{-\frac{z^2}{2\ker}} z^{2m}=\le(2\ker^2\frac{d}{d\ker}\ri)^m \int_{-\infty}^{\infty} d z\ e^{-\frac{z^2}{2\ker}}=\le(2\ker^2\frac{d}{d\ker}\ri)^m \GI{\ker}\, \\
=&\le(2\ker^2\frac{d}{d\ker}\ri)^m \sqrt{2\pi} \ker^{\frac{1}{2}}=\sqrt{2\pi}K^{\frac{2m+1}{2}} (2m-1)(2m-3)\cdots 1   \, ,\nonumber
\end{align}
where in going to the second line we substituted in our expression \eqref{eq:single_Gauss} for $\GI{\ker}$. Therefore, we see that the even moments are given by the simple formula\footnote{This equation with $2m=2$ makes clear why we called $K$ the variance, since for zero-mean Gaussian distributions with variance $K$ we have $\text{var}(z)\equiv\E{\le(z-\E{z}\ri)^2}=\E{z^2} - \E{z}^2 = \E{z^2} =K$.}
\be\label{eq:single_Wick}
\E{z^{2m}}=\frac{\GI{\ker,m}}{\sqrt{2\pi\ker}} = K^m \le(2m-1\ri)!!\, ,
\ee
where we have introduced the \terminate{double factorial}
\be
\le(2m-1\ri)!!\equiv(2m-1)(2m-3)\cdots 1 = \frac{\le(2m\ri) !}{2^m m!}\, .
\ee 
The result~\eqref{eq:single_Wick} is \terminate{Wick's theorem} for single-variable Gaussian distributions.

There's actually another nice way to derive~\eqref{eq:single_Wick}, which can much more naturally be extended to multivariable Gaussian distributions.
This derivation starts with the consideration of a Gaussian integral with a \term{source term}\index{source term|seealso{generating function}} $J$, which we define as
\be\label{eq:partition-function-single-variable-definition}
\PF{\ker,J} \equiv \int_{-\infty}^{\infty} d z\ e^{-\frac{z^2}{2\ker}+J z}\, .
\ee
Note that when setting the source to zero we recover the normalization of the Gaussian integral, giving the relationship $\PF{K,J=0}=\GI{K}$.
In the physics literature $\PF{K,J}$ is sometimes called a \textbf{partition function with source}\index{partition function!with source}\index{partition function!with source|seealso{generating function}} and, as we will soon see, this integral serves as a \term{generating function} for the moments\index{moment}. 
We can evaluate $\PF{K,J}$ by completing the square in the exponent
\be
-\frac{z^2}{2\ker}+J z=-\frac{\le(z-J\ker\ri)^2}{2\ker}+\frac{\ker J^2}{2} \, ,
\ee
which lets us rewrite the integral \eqref{eq:partition-function-single-variable-definition} as
\be
\PF{\ker,J} = e^{\frac{K J^2}{2}}\int_{-\infty}^{\infty} d z\ e^{-\frac{\le(z-JK\ri)^2}{2K}}
 =e^{\frac{K J^2}{2}} \GI{K} 
 =e^{\frac{K J^2}{2}} \sqrt{2\pi\ker} \, ,
\ee
where in the middle equality we noticed that the integrand is just a shifted Gaussian function with variance $\ker$.

We can now relate the Gaussian integral with a source $\PF{\ker, J}$ to the Gaussian integral with insertions $\GI{K,m}$. By differentiating $\PF{\ker, J}$ with respect to the source $J$ and \emph{then} setting the source to zero, we observe that
\be
\GI{\ker,m}= \int_{-\infty}^{\infty} d z\ e^{-\frac{z^2}{2\ker}} z^{2m}=\le[\le(\frac{d}{dJ}\ri)^{2m} \int_{-\infty}^{\infty} d z\ e^{-\frac{z^2}{2\ker}+J z}\ri]\Bigg\vert_{J=0}=\le[\le(\frac{d}{dJ}\ri)^{2m} \PF{\ker,J} \ri]\Bigg\vert_{J=0}\, .
\ee
In other words, the integrals $\GI{\ker,m}$ are simply related to the even Taylor coefficients of the \terminate{partition function}\index{partition function|seealso{normalization factor}} $\PF{\ker,J}$ around $J=0$.
For instance, for $2m=2$ we have
\be
\E{z^{2}}=\frac{\GI{K,1}}{\sqrt{2\pi K}} = \le[\le(\frac{d}{dJ}\ri)^{2} e^{\frac{K J^2}{2}}\ri]\Bigg\vert_{J=0}=\le[e^{\frac{K J^2}{2}}\le(K+K^2J^2\ri)\ri]\Bigg\vert_{J=0}=K\, ,
\ee
and for $2m=4$ we have
\be
\E{z^{4}}=\frac{\GI{K,2}}{\sqrt{2\pi K}} = \le[\le(\frac{d}{dJ}\ri)^{4} e^{\frac{K J^2}{2}}\ri]\Bigg\vert_{J=0}=\le[e^{\frac{K J^2}{2}}\le(3K^2+6K^3J^2+K^4 J^4\ri)\ri]\Bigg\vert_{J=0}=3K^2\, .
\ee
Notice that any terms with dangling sources $J$ vanish upon setting $J=0$. This observation gives a simple way to evaluate correlators for general $m$: Taylor-expand the exponential $\PF{\ker,J}/\GI{\ker} = \exp\!\le(\frac{K J^2}{2}\ri)$ and keep the term with the right amount of sources such that the expression doesn't vanish. Doing exactly that, we get
\begin{align}\label{eq:gaussian-single-variable-insertions}
\E{z^{2m}}=&\frac{\GI{K,m}}{\sqrt{2\pi K}}  =\le[\le(\frac{d}{dJ}\ri)^{2m} e^{\frac{K J^2}{2}}\ri]\Bigg\vert_{J=0}=\le\{\le(\frac{d}{dJ}\ri)^{2m}\le[ \sum_{k=0}^{\infty}\frac{1}{k!} \le(\frac{K}{2}\ri)^{k} J^{2k}\ri]\ri\}\Bigg\vert_{J=0}\, \\
=&\le(\frac{d}{dJ}\ri)^{2m}\le[ \frac{1}{m!} \le(\frac{K}{2}\ri)^m J^{2m}\ri]= K^m\frac{\le(2m\ri) !}{2^m m!}=K^m (2m-1)!!\, ,\nonumber
\end{align}
which completes our second derivation of \terminate{Wick's theorem}~\eqref{eq:single_Wick} for the single-variable Gaussian distribution. This derivation was much longer than the first neat derivation, but can be very naturally extended to the multivariable Gaussian distribution, which we turn to next.

\index{indices!vectorial|see{vectorial indices}}
\index{indices!sample|see{sample indices}}
\index{indices!neural|see{neural indices}}
\index{indices!layer|see{layer indices}}
\index{indices!feature|see{feature indices}}

\subsubsection{Multivariable Gaussian integrals}
Picking up speed, we are now ready to handle multivariable Gaussian integrals for an $\dimpre$-dimensional variable $z_{\mu}$ with $\mu=1,\ldots,\dimpre$.\footnote{Throughout this book, we will explicitly write out the component indices of vectors, matrices, and tensors\index{tensor} as much as possible, except on some occasions when it is clear enough from context.}
The multivariable \terminate{Gaussian function} is defined as
\be\label{eq:multi-gauss-fn}
  \exp\!\le[-\frac{1}{2}\sum_{\mu,\nu=1}^{\dimpre} z_{\mu} (\ker^{-1})_{\mu\nu}\,z_{\nu}\ri]\, ,
\ee
where the \terminate{variance} or \textbf{covariance matrix}\index{covariance}\index{covariance|seealso{cumulant}} $K_{\mu\nu}$ is an $\dimpre$-by-$\dimpre$ symmetric positive definite matrix\index{positive semidefinite matrix!positive definite matrix}, and its inverse $(\ker^{-1})_{\mu\nu}$
is defined so that their matrix product gives the $\dimpre$-by-$\dimpre$ identity matrix
\be\label{eq:inverse-kernel}
\sum_{\rho=1}^{\dimpre}(\ker^{-1})_{\mu\rho}\,\ker_{\rho\nu}=\delta_{\mu\nu}\, .
\ee
Here we have also introduced the \term{Kronecker delta} $\delta_{\mu\nu}$,
which satisfies
\be\label{eq:Kronecker-delta}
\delta_{\mu \nu} \equiv 
    \begin{cases}
  1 \, , & \mu = \nu  \, , \\
    0 \, , & \mu \neq \nu \, .
    \end{cases}
\ee
The Kronecker delta
is just a convenient representation of the \terminate{identity matrix}. 

Now, to construct a \terminate{probability distribution} from the Gaussian function~\eqref{eq:multi-gauss-fn}, we again need to evaluate the \terminate{normalization factor}
\begin{align}\label{eq:multivariable-gaussian-integral}
\GI{\ker} \equiv& \int d^\dimpre\! z\, \exp\!\le[-\frac{1}{2}\sum_{\mu,\nu=1}^{\dimpre} z_{\mu} (\ker^{-1})_{\mu\nu}\, z_{\nu}\ri]\, \\
=& \int_{-\infty}^{\infty} dz_1  \int_{-\infty}^{\infty} dz_2 \cdots \int_{-\infty}^{\infty} dz_\dimpre \, \exp\!\le[-\frac{1}{2}\sum_{\mu,\nu=1}^{\dimpre} z_{\mu} (\ker^{-1})_{\mu\nu}\, z_{\nu}\ri]\, .\nonumber
\end{align}
To compute this integral, first recall from linear algebra that, given an $\dimpre$-by-$\dimpre$ symmetric matrix $K_{\mu\nu}$, there is always an orthogonal matrix\footnote{An \neo{orthogonal matrix} $O_{\mu\nu}$  is a matrix whose transpose $\le(O^T\ri)_{\mu\nu}$ equals its inverse, i.e.,
$(O^T O)_{\mu\nu}=\delta_{\mu\nu}$.
}
$O_{\mu\nu}$ that diagonalizes $K_{\mu\nu}$ as $(OK O^T)_{\mu\nu}=\lambda_{\mu} \delta_{\mu\nu}$ with eigenvalues $\lambda_{\mu=1,\ldots,\dimpre}$ and diagonalizes its inverse as $(OK^{-1} O^T)_{\mu\nu}=\le(1/\lambda_{\mu}\ri) \delta_{\mu\nu}$.
With this in mind, after twice inserting the identity matrix as $\delta_{\mu\nu} = (O^T O)_{\mu\nu}$,
the sum in the exponent of the integral can be expressed in terms of the eigenvalues as
\begin{align}
\sum_{\mu,\nu=1}^{\dimpre} z_{\mu} (\ker^{-1})_{\mu\nu}z_{\nu}
&=\sum_{\mu,\rho,\sigma,\nu=1}^{\dimpre} z_{\mu} \, (O^T O)_{\mu\rho} (\ker^{-1})_{\rho\sigma}(O^T O)_{\sigma\nu}\, z_{\nu}  \\ \notag
&=\sum_{\mu,\nu=1}^{\dimpre}(Oz)_{\mu} (O \ker^{-1} O^T)_{\mu\nu} (Oz)_{\nu}\\ \notag
&=\sum_{\mu=1}^{\dimpre} \frac{1}{\lambda_{\mu}}(Oz)_{\mu}^2\, ,
\end{align}
where to reach the final line we used the \terminate{diagonalization} property of the inverse covariance matrix.
Remembering that for a positive definite matrix $K_{\mu\nu}$ the eigenvalues\index{eigenvalue} are all positive $\lambda_\mu > 0$, we see that the $\lambda_\mu$ sets the scale of the falloff of the Gaussian function in each of the eigendirections.
Next, recall from multivariable calculus that a change of variables $u_{\mu}\equiv (O z)_{\mu}$ with an orthogonal matrix $O$ leaves the integration measure invariant, i.e., $d^{\dimpre}\! z=d^{\dimpre}\!u$. All together, this lets us factorize the multivariable Gaussian integral \eqref{eq:multivariable-gaussian-integral} into a product of single-variable Gaussian integrals~\eqref{eq:single_Gauss}, yielding
\begin{align}
\GI{\ker} =&\int_{-\infty}^{\infty} du_1\int_{-\infty}^{\infty} du_2\cdots \int_{-\infty}^{\infty} du_{\dimpre}\ \exp\!\le(-\frac{u_1^2}{2\lambda_1}-\frac{u_2^2}{2\lambda_2}-\ldots-\frac{u_\dimpre^2}{2\lambda_\dimpre}\ri)\, \\
=&\prod_{\mu=1}^{\dimpre}\le[ \int_{-\infty}^{\infty} du_{\mu}\ \exp\!\le(-\frac{u_{\mu}^2}{2\lambda_{\mu}}\ri)\ri]
=\prod_{\mu=1}^{\dimpre} \sqrt{2\pi \lambda_{\mu}}= \sqrt{\prod_{\mu=1}^{\dimpre}\le(2\pi \lambda_{\mu}\ri)}\, . \nonumber
\end{align}
Finally, recall one last fact from linear algebra that the product of the eigenvalues of a matrix is equal to the matrix \terminate{determinant}. Thus, compactly, we can express the value of the multivariable Gaussian integral as
\be\label{eq:det_formula}
\GI{\ker} = \int d^\dimpre\! z\ \exp\!\le[-\frac{1}{2}\sum_{\mu,\nu=1}^{\dimpre} z_{\mu} (K^{-1})_{\mu\nu}z_{\nu}\ri]=  \sqrt{\dete{2\pi K}}\, ,
\ee
where $\dete{A}$ denotes the determinant of a square matrix $A$.

Having figured out the normalization factor, we can define the zero-mean \textbf{multivariable Gaussian probability distribution}\index{Gaussian distribution!multivariable} with variance $K_{\mu\nu}$ as
\be\label{eq:multi-gauss-dis}
p\!\le(z\ri)=\frac{1}{ \sqrt{\dete{2\pi \ker}}}\exp\!\le[-\frac{1}{2}\sum_{\mu,\nu=1}^{\dimpre} z_{\mu} (\ker^{-1})_{\mu\nu}\, z_{\nu}\ri] \, .
\ee 
While we're at it, let us also introduce the conventions of suppressing the superscript ``$-1$'' for the inverse covariance $(K^{-1})_{\mu\nu}$, instead placing the component indices upstairs as
\be\label{eq:Einstein}
K^{\mu\nu}\equiv (K^{-1})_{\mu\nu}\, .
\ee
This way, we 
distinguish the covariance $K_{\mu\nu}$ and the inverse covariance $K^{\mu\nu}$ by whether or not component indices are lowered or raised. With this notation, inherited from \neo{general relativity}, the defining equation for the inverse covariance~\eqref{eq:inverse-kernel} is written instead as
\be\label{eq:inverse-kernel-compacter}
\sum_{\rho=1}^{\dimpre}\ker^{\mu\rho}\ker_{\rho\nu}=\delta^{\mu}_{\ \nu}\, ,
\ee
and the multivariable Gaussian distribution~\eqref{eq:multi-gauss-dis} is written as
\be\label{eq:multi-gauss-dis-compacter}
p\!\le(z\ri)=\frac{1}{ \sqrt{\dete{2\pi \ker}}}\exp\!\le(-\frac{1}{2}\sum_{\mu,\nu=1}^{\dimpre} z_{\mu} \ker^{\mu\nu}z_{\nu}\ri)\, .
\ee
Although it might take some getting used to, this notation saves us some space and saves you some handwriting pain.\footnote{If you like, in your notes you can also go full general-relativistic mode and adopt \neo{Einstein summation convention}, suppressing the summation symbol any time indices are repeated in upstair-downstair pairs. For instance, if we adopted this convention we would write the defining equation for inverse simply as $\ker^{\mu\rho}\ker_{\rho\nu}=\delta^{\mu}_{\ \nu}$ and the Gaussian function as $\exp\!\le(-\frac{1}{2}z_{\mu} \ker^{\mu\nu}z_{\nu}\ri)$. 

Specifically for neural networks, you might find the Einstein summation convention helpful for \emph{sample} indices, but sometimes confusing for \emph{neural} indices.  For extra clarity, we won't adopt this convention in the text of the book, but we mention it now since we do often use such a convention to simplify our own calculations in private.}
Regardless of how it's written, the zero-mean multivariable Gaussian \terminate{probability distribution}~\eqref{eq:multi-gauss-dis-compacter} peaks at $z=0$, and its falloff is direction-dependent, determined by the covariance matrix $K_{\mu\nu}$.
More generally, we can shift the peak of the Gaussian distribution to $s_{\mu}$
\be\label{eq:multi-gauss-dis-compacter-with-mean}
p\!\le(z\ri)= \frac{1}{ \sqrt{\dete{2\pi \ker}}}\exp\!\le[-\frac{1}{2}\sum_{\mu,\nu=1}^{\dimpre} \le(z-s\ri)_{\mu} \ker^{\mu\nu}\le(z-s\ri)_{\nu}\ri]\, ,
\ee
which defines a general multivariable Gaussian distribution with mean $\E{z_{\mu}}=s_{\mu}$ and covariance $\ker_{\mu\nu}$. This is the most general version of the Gaussian distribution.

Next, let's consider the moments\index{moment} of the mean-zero multivariable Gaussian distribution
\begin{align}
\E{z_{\mu_1}\cdots z_{\mu_M}}\equiv&  \int d^\dimpre\! z\ p\!\le(z\ri) z_{\mu_1}\cdots z_{\mu_M}\, \\
=& \frac{1}{ \sqrt{\dete{2\pi \ker}}} \int d^\dimpre\! z\  \exp\!\le(-\frac{1}{2}\sum_{\mu,\nu=1}^{\dimpre} z_{\mu} \ker^{\mu\nu}z_{\nu}\ri)\, z_{\mu_1}\cdots z_{\mu_M}=\frac{\GI{K, \le(\mu_1,\ldots,\mu_{M}\ri)}}{\GI{K}}\, , \nonumber
\end{align}
where we introduced multivariable Gaussian integrals with insertions
\be\label{eq:multi_insertions}
\GI{K, \le(\mu_1,\ldots,\mu_{M}\ri)}\equiv \int d^\dimpre\! z\  \exp\!\le(-\frac{1}{2}\sum_{\mu,\nu=1}^{\dimpre} z_{\mu} \ker^{\mu\nu}z_{\nu}\ri)\, z_{\mu_1} \cdots  z_{\mu_{M}} \, .
\ee
Following our approach in the single-variable case, let's construct the \terminate{generating function} for the integrals $\GI{K, \le(\mu_1,\ldots,\mu_{M}\ri)}$ by including a source term $J^{\mu}$ as %
\be\label{eq:multi-variate-gaussian-partition}
\PF{K,J} \equiv \int d^\dimpre\! z\ \exp\!\le(-\frac{1}{2}\sum_{\mu,\nu=1}^{\dimpre} z_{\mu} K^{\mu\nu}z_{\nu}+\sum_{\mu=1}^{\dimpre}J^{\mu} z_{\mu}\ri)\, .
\ee
As the name suggests, differentiating the \terminate{generating function} $\PF{K,J} $ with respect to the source\index{source term} $J^\mu$ brings down a power of $z_\mu$ such that after $M$ such differentiations we have
 \begin{align}\label{eq:multi-gaussian-insertions-and-partition-function}
&\le[ \frac{d}{d J^{\mu_1}}  \frac{d}{d J^{\mu_2}}\cdots  \frac{d}{d J^{\mu_{M}}} \PF{\ker,J}\ri]\Bigg\vert_{J=0}\, \\
=& \int d^\dimpre\! z\  \exp\!\le(-\frac{1}{2}\sum_{\mu,\nu=1}^{\dimpre} z_{\mu} K^{\mu\nu}z_{\nu}\ri)\, z_{\mu_1} \cdots  z_{\mu_{M}} = \GI{K, \le(\mu_1,\ldots,\mu_{M}\ri)}\, .\nonumber
 \end{align}
So, as in the single-variable case, the Taylor coefficients of the \terminate{partition function} $\PF{K,J}$ expanded around $J^\mu=0$ are simply related to the integrals with insertions $\GI{K, \le(\mu_1,\ldots,\mu_{M}\ri)}$. Therefore, if we knew a closed-form expression for  $\PF{K,J}$, we could easily compute the values of the integrals $\GI{K, \le(\mu_1,\ldots,\mu_{M}\ri)}$.

\index{complete the square}
To evaluate the \terminate{generating function} $\PF{K,J}$ in a closed form, again we follow the lead of the single-variable case and complete the square in the exponent of the integrand in \eqref{eq:multi-variate-gaussian-partition} as
\begin{align}
&-\frac{1}{2}\sum_{\mu,\nu=1}^{\dimpre} z_{\mu} K^{\mu\nu}z_{\nu}+\sum_{\mu=1}^{\dimpre}J^{\mu} z_{\mu}\, \\
=&-\frac{1}{2}\sum_{\mu,\nu=1}^{\dimpre} \le(z_{\mu}-\sum_{\rho=1}^{\dimpre}K_{\mu\rho}J^{\rho}\ri) K^{\mu\nu} \le(z_{\nu}-\sum_{\lambda=1}^{\dimpre}K_{\nu\lambda}J^{\lambda}\ri)+\frac{1}{2}\sum_{\mu,\nu=1}^{\dimpre}J^{\mu} K_{\mu\nu}J^{\nu}\, \nonumber\\
=&-\frac{1}{2}\sum_{\mu,\nu=1}^{\dimpre} w_{\mu}K^{\mu\nu}w_{\nu}+\frac{1}{2}\sum_{\mu,\nu=1}^{\dimpre}J^{\mu} K_{\mu\nu}J^{\nu}\, , \nonumber
\end{align}
where we have introduced the shifted variable $w_{\mu} \equiv z_{\mu}-\sum_{\rho=1}^{\dimpre}K_{\mu\rho}J^{\rho}$. Using this substitution, the generating function can be evaluated explicitly
\begin{align}\label{eq:multi-gaussian-Z-J}
\PF{\ker,J} =& \exp\!\le(\frac{1}{2}\sum_{\mu,\nu=1}^{\dimpre}J^{\mu} K_{\mu\nu}J^{\nu}\ri)  \int d^\dimpre\! w\ \exp\!\le[-\frac{1}{2}\sum_{\mu,\nu=1}^{\dimpre} w_{\mu}\ker^{\mu\nu}w_{\nu}\ri]\, \\
 =& \sqrt{\dete{2\pi K}} \exp\!\le(\frac{1}{2}\sum_{\mu,\nu=1}^{\dimpre}J^{\mu} K_{\mu\nu}J^{\nu}\ri) \, ,\nonumber
\end{align}
where at the end we used our formula for the multivariable integral $\GI{\ker}$,  \eqref{eq:det_formula}.
With our closed-form expression~\eqref{eq:multi-gaussian-Z-J} for the generating function $\PF{K,J}$, we can compute the Gaussian integrals with insertions $\GI{K, \le(\mu_1,\ldots,\mu_{M}\ri)}$ by differentiating it, using~\eqref{eq:multi-gaussian-insertions-and-partition-function}. For an even number $M=2m$ of insertions, we find a really nice formula
\begin{align}\label{eq:nice-formula-for-multivariable-moments}
\E{z_{\mu_1}\cdots z_{\mu_{2m}}}=&\frac{\GI{\ker, \le(\mu_1,\ldots,\mu_{2m}\ri)}}{\GI{\ker}}=\frac{1}{\GI{\ker}}\le[ \frac{d}{d J^{\mu_1}}\cdots  \frac{d}{d J^{\mu_{2m}}} \PF{K,J}\ri]\Bigg\vert_{J=0}\, \\
=&\frac{1}{2^m m!} \frac{d}{d J^{\mu_1}}  \frac{d}{d J^{\mu_2}}\cdots  \frac{d}{d J^{\mu_{2m}}}\le(\sum_{\mu,\nu=1}^{\dimpre}J^{\mu} K_{\mu\nu}J^{\nu}\ri)^m\, .\nonumber
\end{align}
For an odd number $M=2m+1$ of insertions, there is dangling source upon setting $J=0$, and so those integrals vanish. You can also see this by looking at the integrand for any odd moment and noticing that it is odd with respect to the sign flip of the integration variables $z_\mu \leftrightarrow -z_\mu$.

Now,
let's
take a few moments to
evaluate a few moments\index{moment} using this formula.
For $2m=2$, we have
\be\label{eq:Wick-second-moment}
\E{z_{\mu_1}z_{\mu_{2}}} = \frac{1}{2} \frac{d}{d J^{\mu_1}}  \frac{d}{d J^{\mu_2}}\le(\sum_{\mu,\nu=1}^{\dimpre}J^{\mu} K_{\mu\nu}J^{\nu}\ri)=K_{\mu_1\mu_2} .
 \ee
  Here, there are $2!=2$ ways to apply the product rule for derivatives and differentiate the two $J$'s, both of which evaluate to the same expression due to the symmetry of the covariance, $K_{\mu_1 \mu_2}=K_{\mu_2 \mu_1}$. This expression \eqref{eq:Wick-second-moment} validates in the multivariable setting why we have been calling $\ker_{\mu\nu}$ the covariance, because we see explicitly that it is the covariance.

Next, for $2m=4$ we get a more complicated expression
\begin{align}\label{eq:Wick-fourth-moment}
\E{z_{\mu_1}z_{\mu_2}z_{\mu_{3}} z_{\mu_{4}}}=& \frac{1}{2^2 2!} \frac{d}{d J^{\mu_1}}  \frac{d}{d J^{\mu_2}} \frac{d}{d J^{\mu_3}}  \frac{d}{d J^{\mu_4}}\le(\sum_{\mu,\nu=1}^{\dimpre}J^{\mu} K_{\mu\nu}J^{\nu}\ri)\le(\sum_{\rho,\lambda=1}^{\dimpre}J^{\rho} K_{\rho\lambda}J^{\lambda}\ri)\, \notag\\
=& K_{\mu_1\mu_2}K_{\mu_3\mu_4}+K_{\mu_1\mu_3}K_{\mu_2\mu_4}+K_{\mu_1\mu_4}K_{\mu_2\mu_3}\, .
 \end{align}
Here we note
that there are now $4!=24$ ways to differentiate the four $J$'s, though only three distinct ways to pair the four auxiliary indices $1,2,3,4$ that sit under $\mu$. This gives $24/3=8=2^2 2!$ equivalent terms for each of the three pairings, which cancels against the overall factor $1/(2^2 2!)$.

For general $2m$, there are $(2m)!$ ways to differentiate the sources, of which $2^m m!$ of those ways are equivalent. This gives $(2m)!/(2^m m!) = (2m-1)!!$ distinct terms, corresponding to the $(2m-1)!!$ distinct pairings of $2m$ auxiliary indices $1,\ldots,2m$ that sit under $\mu$. The factor of $1/(2^m m!)$ in the denominator of \eqref{eq:nice-formula-for-multivariable-moments} ensures that the coefficient of each of these terms is normalized to unity.
Thus, most generally, we can express the moments\index{moment} of the multivariable Gaussian with the following formula
 \begin{align}\label{eq:Wick-multi}
\E{z_{\mu_1}\cdots z_{\mu_{2m}}}=\sum_{\text{all pairing}}K_{\mu_{k_1}\mu_{k_2}}\cdots K_{\mu_{k_{2m-1}}\mu_{k_{2m}}} \, ,
 \end{align}
where, to reiterate, the sum is over all the possible distinct pairings of the $2m$ auxiliary indices under $\mu$ such that the result has the $(2m-1)!!$ terms that we described above.
Each factor of the covariance $K_{\mu\nu}$ in a term in sum is called a \term{Wick contraction}, corresponding to a particular pairing of auxiliary indices. Each term then is composed of $m$ different Wick contractions, representing a distinct way of pairing up all the auxiliary indices.
To make sure you understand how this pairing works, look back at the $2m=2$ case~\eqref{eq:Wick-second-moment} -- with a single Wick contraction -- and the $2m=4$ case~\eqref{eq:Wick-fourth-moment} -- with three distinct ways of making two Wick contractions -- and try to work out the $2m=6$ case, which yields $(6-1)!!=15$ distinct ways of making three Wick contractions:
\begin{align}
\E{z_{\mu_1}z_{\mu_2}z_{\mu_{3}} z_{\mu_{4}}z_{\mu_5} z_{\mu_{6}}}=&\ker_{\mu_1 \mu_2}\ker_{\mu_3 \mu_4}\ker_{\mu_5 \mu_6}+\ker_{\mu_1 \mu_3}\ker_{\mu_2 \mu_4}\ker_{\mu_5 \mu_6}+\ker_{\mu_1 \mu_4}\ker_{\mu_2 \mu_3}\ker_{\mu_5 \mu_6}\, \nonumber\\
+&\ker_{\mu_1 \mu_2}\ker_{\mu_3 \mu_5}\ker_{\mu_4 \mu_6}+\ker_{\mu_1 \mu_3}\ker_{\mu_2 \mu_5}\ker_{\mu_4 \mu_6}+\ker_{\mu_1 \mu_5}\ker_{\mu_2 \mu_3}\ker_{\mu_4 \mu_6}\, \nonumber\\
+&\ker_{\mu_1 \mu_2}\ker_{\mu_5 \mu_4}\ker_{\mu_3 \mu_6}+\ker_{\mu_1 \mu_5}\ker_{\mu_2 \mu_4}\ker_{\mu_3 \mu_6}+\ker_{\mu_1 \mu_4}\ker_{\mu_2 \mu_5}\ker_{\mu_3 \mu_6}\, \nonumber\\
+&\ker_{\mu_1 \mu_5}\ker_{\mu_3 \mu_4}\ker_{\mu_2 \mu_6}+\ker_{\mu_1 \mu_3}\ker_{\mu_5 \mu_4}\ker_{\mu_2 \mu_6}+\ker_{\mu_1 \mu_4}\ker_{\mu_5 \mu_3}\ker_{\mu_2 \mu_6}\, \nonumber\\
+&\ker_{\mu_5 \mu_2}\ker_{\mu_3 \mu_4}\ker_{\mu_1 \mu_6}+\ker_{\mu_5 \mu_3}\ker_{\mu_2 \mu_4}\ker_{\mu_1 \mu_6}+\ker_{\mu_5 \mu_4}\ker_{\mu_2 \mu_3}\ker_{\mu_1 \mu_6}\, .\nonumber
\end{align}

The formula~\eqref{eq:Wick-multi} is \term{Wick's theorem}. Put a box around it.
Take a few moments for reflection. 

\begin{center}
\ldots\\

\ldots\\

\ldots\\
\end{center}

\noindent Good.
You are now a Gaussian sensei. %
Exhale, and then say as Neo\index{Anderson, Thomas A. ``Neo''} would say,
``I know Gaussian integrals." 

Now that the moments have passed, it is an appropriate time to transition to the next section where you will learn about more general probability distributions.

\section{Probability, Correlation and Statistics, and All That}\label{sec:not-Gauss}

\index{expectation value}\index{moment}
In introducing the \terminate{Gaussian distribution} in the last section we briefly touched upon the concepts of 
expectation and moments.
These are defined for non-Gaussian
probability distributions too, so now let us reintroduce these concepts and expand on their definitions, with an eye towards understanding the nearly-Gaussian distributions that describe wide neural networks. %

Given a \term{probability distribution} $p(z)$ of an $\dimpre$-dimensional random variable $z_\mu$, we can learn about its statistics\index{statistics (of a random variable)}\index{statistics (of a random variable)|seealso{probability distribution}} by measuring functions of $z_{\mu}$. We'll refer to such measurable functions in a generic sense as \textbf{observables}\index{observable|textbf} and denote them as $\mathcal{O}(z)$. The \term{expectation value} of an observable
\be\label{eq:expectation-value-definition}
\E{\mathcal{O}(z)}\equiv \int d^\dimpre\! z\ p(z) \, \mathcal{O}(z)\, %
\ee
characterizes the 
mean value of the random function $\mathcal{O}(z)$. %
Note that the observable $\mathcal{O}(z)$ needs not be a scalar-valued function, e.g.~the second moment of a distribution is a matrix-valued observable given by $\mathcal{O}(z) = z_\mu z_\nu$.

Operationally, an observable is a quantity that we measure by conducting experiments in order to connect to a theoretical model for the underlying probability distribution describing $z_\mu$. In particular, we repeatedly measure the observables that are naturally accessible to us as experimenters, collect their statistics, and then compare them with predictions for the expectation values of those observables computed from some theoretical model of $p(z)$.

With that in mind, it's very natural to ask: what kind of information can we learn about an underlying distribution $p(z)$ by measuring an observable $\mathcal{O}(z)$? For an a priori unknown distribution, is there a set of observables that can serve as a sufficient probe of $p(z)$ such that we could use that information to predict the result of all future experiments involving $z_\mu$?

Consider a
class of observables that we've already encountered, the  \textbf{moments}\index{moment|textbf} or \textbf{\emph{M}-point correlators}\index{correlator!$M$-point} of $z_\mu$, given by the expectation\footnote{In the rest of this book, we'll often use the physics term \emph{$M$-point correlator} rather than the statistics term \emph{moment}, though they mean the same thing and can be used interchangeably.}
\be
\E{z_{\mu_{1}}z_{\mu_{2}}\cdots z_{\mu_{M}}} = \int d^\dimpre\! z\ p(z)\, z_{\mu_{1}}z_{\mu_{2}}\cdots z_{\mu_{M}} \, .
\ee
In principle, knowing the $M$-point correlators of a distribution lets us compute the expectation value of any analytic observable $\mathcal{O}(z)$ via Taylor expansion
\begin{align}\label{eq:observables-and-moments}
\E{\mathcal{O}(z)}=&\E{\sum_{M=0}^{\infty}\frac{1}{M!}\sum_{\mu_1,\ldots,\mu_M=1}^{\dimpre}\frac{\partial^M \mathcal{O}}{\partial z_{\mu_1}\cdots\partial z_{\mu_M}}\Bigg\vert_{z=0} z_{\mu_{1}}z_{\mu_{2}}\cdots z_{\mu_{M}}}\, \\
=&\sum_{M=0}^{\infty}\frac{1}{M!}\sum_{\mu_1,\ldots,\mu_M=1}^{\dimpre}\frac{\partial^M \mathcal{O}}{\partial z_{\mu_1}\cdots\partial z_{\mu_M}}\Bigg\vert_{z=0}\E{ z_{\mu_{1}}z_{\mu_{2}}\cdots z_{\mu_{M}}}\, , \notag
\end{align}
where on the last line we took the Taylor coefficients out of the expectation by using the linearity property of the expectation, inherited from the linearity property of the integral in \eqref{eq:expectation-value-definition}. As such, it's clear that the collection of all the $M$-point correlators completely characterizes a \terminate{probability distribution} for all intents and purposes.\footnote{In fact, the moments offer a dual description of the probability distribution through either the \terminate{Laplace transform} or the \terminate{Fourier transform}. For instance, the Laplace transform of the \terminate{probability distribution} $p(z)$ is given by
\be
Z_{J}\equiv\E{\exp\!\le(\sum_\mu J^\mu z_\mu\ri)}=\int \le[\prod_{\mu}dz_{\mu}\ri]\ p(z)\exp\!\le(\sum_\mu J^\mu z_\mu\ri). 
\ee
As in the Gaussian case, this integral gives a \terminate{generating function} for the $M$-point correlators of $p(z)$, %
which means that $Z_{J}$ can be reconstructed from these correlators.
The probability distribution can then be obtained through the inverse Laplace transform. %
}

\index{correlator!$M$-point}
However, this description in terms of all the correlators is somewhat cumbersome and operationally infeasible. To get a reliable estimate of the $M$-point correlator, we must simultaneously measure $M$ components of a random variable for each draw and repeat such measurements many times.
As $M$ grows, this task quickly becomes impractical.
In fact, if we could easily perform such measurements for all $M$, then our theoretical model of $p(z)$ would no longer be a useful abstraction; from \eqref{eq:observables-and-moments} we would already know the outcome of all possible experiments that we could perform, leaving nothing for us to predict.

\index{Gaussian distribution!zero-mean, defined by variance}
To that point, essentially all useful distributions can be effectively described in terms of a finite number of quantities, giving them a parsimonious representation.
For instance, consider the zero-mean $n$-dimensional Gaussian distribution with the variance $K_{\mu\nu}$. %
The nonzero $2m$-point correlators are given by Wick's theorem~\eqref{eq:Wick-multi} as
\be\label{eq:Wick_compact}
\E{z_{\mu_{1}}z_{\mu_{2}}\cdots z_{\mu_{2m}}}=\sum_{\text{all pairing}}K_{\mu_{k_1}\mu_{k_2}}\cdots K_{\mu_{k_{2m-1}}\mu_{k_{2m}}}\, ,
\ee
and are determined entirely by
the $\dimpre(\dimpre+1)/2$ independent components of the variance $K_{\mu\nu}$. The variance itself can be estimated by measuring the two-point correlator
\be
\E{z_{\mu}z_{\nu}}=K_{\mu\nu}\, .
\ee
This is consistent with our description of the distribution itself as ``the zero-mean $\dimpre$-dimensional Gaussian distribution with the variance $K_{\mu\nu}$'' in which we only had to specify these same set of numbers, $K_{\mu\nu}$, to pick out the particular distribution we had in mind.
For zero-mean Gaussian distributions, there's no reason to measure or keep track of any of the higher-point correlators as they are completely constrained by the variance through \eqref{eq:Wick_compact}.

More generally, it would be nice if there were a systematic way for learning about non-Gaussian \terminate{probability distribution}s without performing an infinite number of experiments. 
For nearly-Gaussian distributions\index{nearly-Gaussian distribution}, a useful set of observables is given by
what statisticians call \textbf{cumulants}\index{cumulant|textbf}\index{cumulant|seealso{connected correlator}} and physicists call \textbf{connected correlators}\index{connected correlator|textbf}.\footnote{Outside of this chapter, just as we'll often use the term $M$-point correlator rather than the term moment, we'll use the term $M$-point connected correlator rather than the term cumulant. When we want to refer to the moment and not the cumulant, we might sometimes say \emph{full correlator}\index{full correlator|see{correlator}} to contrast with \emph{connected correlator}.\index{correlator!full}\index{correlator!full|seealso{moment}}} As the formal definition of these quantities is somewhat cumbersome and unintuitive, let's start with a few simple examples.

\index{cumulant!first (mean)}\index{cumulant!first (mean)|seealso{mean}}\index{connected correlator!one-point}\index{connected correlator!one-point|seealso{mean}}\index{correlator!connected|see{connected correlator}}
The first cumulant or the connected one-point correlator is the same as the full one-point correlator
\be\label{eq:C1}
\Ec{z_\mu}{\big|} \equiv  \E{z_\mu}\, .
\ee
This is just the \neo{mean} of the distribution. The second cumulant or the connected two-point correlator is given by
\begin{align}\label{eq:C2}
\Ec{z_\mu z_\nu}{\big|} \equiv&  \E{z_\mu z_\nu} - \E{z_\mu}\E{z_\nu}\, \\
=&\E{\le(z_{\mu}-\E{z_{\mu}}\ri)\le(z_{\nu}-\E{z_{\nu}}\ri)}\equiv \cov{z_\mu}{z_\nu} \, ,\nonumber
\end{align}
which is also known as the \neo{covariance}\index{cumulant!second (covariance)}\index{cumulant!second (covariance)|seealso{covariance}}\index{connected correlator!two-point}\index{connected correlator!two-point|seealso{covariance}}\index{connected correlator!two-point|seealso{metric}} of the distribution. Note how the mean is subtracted from the random variable $z_\mu$ before taking the square in the connected version. The quantity $\widehat{\Delta z}_\mu \equiv z_{\mu}-\E{z_{\mu}}$ represents a \textbf{fluctuation}\index{fluctuations|textbf} of the random variable around its mean. Intuitively, such fluctuations are equally likely to contribute positively as they are likely to contribute negatively, $\E{\widehat{\Delta z}_\mu} = \E{z_{\mu}}-\E{z_{\mu}}=0$, so it's necessary to take the square in order to get an estimate of the magnitude of such fluctuations.

\index{nearly-Gaussian distribution}\index{connected correlator!odd-point vanish with parity}
At this point, let us restrict our focus to distributions that are invariant under a sign-flip symmetry $z_\mu \to - z_\mu$, which holds for the zero-mean Gaussian distribution \eqref{eq:multi-gauss-dis-compacter}.
Importantly, this \neo{parity symmetry} will also hold for the nearly-Gaussian distributions that we will study in order to describe neural networks. For all such even distributions with this symmetry, all odd moments\index{moment} and all odd-point connected correlators vanish.

With this restriction, the next simplest observable is the fourth cumulant or the connected four-point correlator, given by the formula\index{connected correlator!four-point|textbf}\index{connected correlator!four-point|seealso{kurtosis, excess}}\index{connected correlator!four-point|seealso{four-point vertex}}
\begin{align}\label{eq:C4}
&\E{z_{\mu_1} z_{\mu_2}z_{\mu_3}z_{\mu_4}}\big|_{\text{connected}}\, \\
=&\E{z_{\mu_1} z_{\mu_2}z_{\mu_3}z_{\mu_4}}\,\nonumber\\
&-\E{z_{\mu_1} z_{\mu_2}}\E{z_{\mu_3} z_{\mu_4}}-\E{z_{\mu_1} z_{\mu_3}}\E{z_{\mu_2} z_{\mu_4}}-\E{z_{\mu_1} z_{\mu_4}}\E{z_{\mu_2} z_{\mu_3}}\, .\nonumber
\end{align}
For the Gaussian distribution, recalling the Wick theorem~\eqref{eq:Wick_compact}, the last three terms precisely subtract off the three pairs of Wick contractions used to evaluate the first term, meaning
\be\label{eq:C4-gaussian}
\E{z_{\mu_1} z_{\mu_2}z_{\mu_3}z_{\mu_4}}\big|_{\text{connected}} = 0. 
\ee
Essentially by design, the connected four-point correlator vanishes for the Gaussian distribution, and a nonzero value signifies a deviation from Gaussian statistics.\footnote{In statistics, the connected four-point correlator for a single random variable $z$ is called the \emph{excess kurtosis}\index{kurtosis, excess}\index{kurtosis, excess|seealso{connected correlator}} when normalized by the square of the variance. It is a natural measure of the tails of the distribution, as compared to a Gaussian distribution, and also serves as a measure of the potential for outliers. In particular, a positive value indicates fatter tails while a negative value indicates thinner tails.\label{footnote-kurtosis}} 
In fact, the connected four-point correlator is perhaps the simplest measure of non-Gaussianity.

\index{cumulant!general definition}\index{connected correlator!general definition}
Now that we have a little intuition, we are as ready as we'll ever be to discuss the definition for the $M$-th cumulant or the $M$-point connected  correlator. For completeness, we'll give the general definition, before restricting again to distributions that are symmetric under parity $z_\mu \to - z_\mu$. 
The definition is \emph{inductive} and somewhat counterintuitive, expressing the $M$-th moment in terms of connected correlators from degree $1$ to $M$: 
\begin{align}\label{eq:cumu}
&\E{z_{\mu_1} z_{\mu_2}\cdots z_{\mu_{M} }}\, \\
\equiv&\E{z_{\mu_1} z_{\mu_2}\cdots z_{\mu_{M}}}\big|_{\text{connected}}\, \nonumber\\
&+\sum_{\text{all\ subdivisions}}\E{z_{\mu_{k^{[1]}_1}}\cdots z_{\mu_{k^{[1]}_{\nu_1}}}}\Bigg|_{\text{connected}}\cdots\E{z_{\mu_{k^{[s]}_1}} \cdots z_{\mu_{k^{[s]}_{\nu_s}}}}\Bigg|_{\text{connected}}\, , \nonumber
\end{align}
where the sum is over all the possible subdivisions of $M$ variables into $s>1$ clusters of sizes $(\nu_1,\ldots,\nu_s)$ as $(k^{[1]}_1,\ldots,k^{[1]}_{\nu_1}),\ldots,(k^{[s]}_1,\ldots,k^{[s]}_{\nu_s})$. 
By decomposing the $M$-th moment into a sum of products of connected correlators of degree $M$ and lower, we see that the connected $M$-point correlator corresponds to a \emph{new} type of correlation that cannot be expressed by the connected correlators of a lower degree. We saw an example of this above when discussing the connected four-point correlator as a simple measure of non-Gaussianity.

To see how this abstract definition actually works, let's revisit the examples. First, we trivially recover the relation between the mean and the one-point connected correlator
\be
\Ec{z_\mu}{\big|}=\E{z_\mu}\, ,
\ee
as there is no subdivision of a $M=1$ variable into any smaller pieces. For $M=2$, the definition~\eqref{eq:cumu} gives
\begin{align}
\E{z_{\mu_1}z_{\mu_2}}=&\Ec{z_{\mu_1} z_{\mu_2}}{\big|}+\Ec{z_{\mu_1}}{\big|}\Ec{z_{\mu_2}}{\big|}\, \\
=&\Ec{z_{\mu_1} z_{\mu_2}}{\big|}+\E{z_{\mu_1}}\E{z_{\mu_2}}\, .\nonumber
\end{align}
Rearranging to solve for the connected two-point function in terms of the moments, we see that this
is equivalent to our previous definition for the covariance~\eqref{eq:C2}. %

\index{parity symmetry}
At this point, let us again restrict to parity-symmetric distributions invariant under $z_\mu \to - z_\mu$, remembering that this means that all the odd-point connected correlators will vanish. For such distributions, evaluating the definition~\eqref{eq:cumu} for $M=4$ gives
\begin{align}\label{eq:C4-reversed}
\E{z_{\mu_1} z_{\mu_2}z_{\mu_3}z_{\mu_4}}=&\E{z_{\mu_1} z_{\mu_2}z_{\mu_3}z_{\mu_4}}\big|_{\text{connected}}\, \\
&+\E{z_{\mu_1} z_{\mu_2}}\big|_{\text{connected}}\E{z_{\mu_3} z_{\mu_4}}\big|_{\text{connected}}\, \nonumber\\
&+\E{z_{\mu_1} z_{\mu_3}}\big|_{\text{connected}}\E{z_{\mu_2} z_{\mu_4}}\big|_{\text{connected}}\, \nonumber\\
&+\E{z_{\mu_1} z_{\mu_4}}\big|_{\text{connected}}\E{z_{\mu_2} z_{\mu_3}}\big|_{\text{connected}}\, .\nonumber
\end{align}
Since $\E{z_{\mu_1} z_{\mu_2}}=\E{z_{\mu_1} z_{\mu_2}}\big|_{\text{connected}}$ when the mean vanishes, this is also just a rearrangement of our previous expression \eqref{eq:C4} for the connected four-point correlator for such zero-mean distributions. 

In order to see something new, let us carry on for $M=6$:
\begin{align}\label{eq:six-point-moment-in-terms-of-connected}
\E{z_{\mu_1} z_{\mu_2}z_{\mu_3}z_{\mu_4}z_{\mu_5}z_{\mu_6}}=&\E{z_{\mu_1} z_{\mu_2}z_{\mu_3}z_{\mu_4}z_{\mu_5}z_{\mu_6}}\big|_{\text{connected}}\, \\
&+\E{z_{\mu_1} z_{\mu_2}}\big|_{\text{connected}}\E{z_{\mu_3} z_{\mu_4}}\big|_{\text{connected}}\E{z_{\mu_5} z_{\mu_6}}\big|_{\text{connected}}\, \nonumber\\
&+\le[14 \ \text{other}\ (2,2,2)\ \text{subdivisions}\ri]\, \nonumber\\
&+\E{z_{\mu_1} z_{\mu_2}z_{\mu_3} z_{\mu_4}}\big|_{\text{connected}}\E{z_{\mu_5} z_{\mu_6}}\big|_{\text{connected}}\, \nonumber\\
&+\le[14 \ \text{other}\ (4,2)\ \text{subdivisions}\ri]\, ,\nonumber
\end{align}
in which we have expressed the full six-point correlator in terms of a sum of products of connected two-point, four-point, and six-point correlators.
Rearranging the above expression and expressing the two-point and four-point connected correlators in terms of their definitions,  \eqref{eq:C2} and \eqref{eq:C4}, we obtain an expression for the connected six-point correlator:
\begin{align}\label{eq:C6}
&\E{z_{\mu_1} z_{\mu_2}z_{\mu_3}z_{\mu_4}z_{\mu_5}z_{\mu_6}}\big|_{\text{connected}}\, \\
=&\E{z_{\mu_1} z_{\mu_2}z_{\mu_3}z_{\mu_4}z_{\mu_5}z_{\mu_6}}\,\nonumber\\
&-\le\{\E{z_{\mu_1} z_{\mu_2}z_{\mu_3} z_{\mu_4}}\E{z_{\mu_5}z_{\mu_6}}+\le[14 \ \text{other}\ (4,2)\ \text{subdivisions}\ri]\ri\}\, \nonumber\\
&+2\le\{\E{z_{\mu_1} z_{\mu_2}}\E{z_{\mu_3} z_{\mu_4}}\E{z_{\mu_5} z_{\mu_6}}+\le[14 \ \text{other}\ (2,2,2)\ \text{subdivisions}\ri]\ri\}\, .\nonumber
\end{align}
The rearrangement is useful for computational purposes, in that it's simple to first compute the moments of a distribution and then organize the resulting expressions in order to evaluate the connected correlators.

Focusing back on \eqref{eq:six-point-moment-in-terms-of-connected}, it's easy to see that the connected six-point correlator vanishes for Gaussian distributions.
Remembering that the connected four-point correlator also vanishes for Gaussian distributions, we see that the fifteen $(2,2,2)$ subdivision terms are exactly equal to the fifteen terms generated by the Wick contractions resulting from evaluating the full correlator on the left-hand side of the equation. In fact, applying the general definition of connected correlators \eqref{eq:cumu} to the zero-mean Gaussian distribution, we see inductively that all $M$-point connected correlators for $M > 2$ will vanish.\footnote{To see this, 
note that if all the higher-point connected correlators vanish, then the definition~\eqref{eq:cumu} is equivalent to Wick's theorem~\eqref{eq:Wick_compact}, with nonzero terms in \eqref{eq:cumu} -- the subdivisions into clusters of sizes (2, \dots, 2) --  corresponding exactly to the different pairings in \eqref{eq:Wick_compact}.} %
Thus, the connected correlators are a very natural measure of how a distribution deviates from  Gaussianity.

With this in mind, we can finally define a \term{nearly-Gaussian distribution} as a distribution for which all the connected correlators for $M>2$ are \emph{small}.\footnote{
As we discussed in \S\ref{sec:Gauss}, the variance sets the scale of the Gaussian distribution. For nearly-Gaussian distributions, we require that all $2m$-point connected correlators be parametrically small when compared to an appropriate power of the variance, i.e., $\vert \E{z_{\mu_1} \cdots z_{\mu_{2m}}}|_{\text{connected}}\vert \ll \vert K_{\mu\nu}\vert^m$, schematically.}
In fact, the non-Gaussian distributions that describe neural networks generally have the 
property that, as the network becomes wide, the connected four-point correlator becomes small and the
higher-point connected correlators become even smaller.
For these nearly-Gaussian distributions, a few leading connected correlators give a concise and accurate description of the distribution, just as a few leading Taylor coefficients can give a good description of a function near the point of expansion.

\section{Nearly-Gaussian Distributions}\label{sec:perturbation}

\index{nearly-Gaussian distribution!connected correlators as observables}\index{connected correlator!relation to nearly-Gaussian distributions}
Now that we have defined nearly-Gaussian distributions in terms of measurable deviations from Gaussian statistics, i.e.~via small but nonzero connected correlators, it's natural to ask how we can link these observables to the actual functional form of the distribution, $p(z)$.
We can make this connection through the action.

The \term{action} $\ac(z)$ is a function that defines a \terminate{probability distribution} $p(z)$ through the relation
\be\label{eq:action-representation-of-distribution}
p(z)\propto e^{-\ac(z)}\, .
\ee
In the
statistics literature, the action $\ac(z)$ is sometimes called the \emph{negative log probability}\index{negative log probability|see{action}}, but we will again follow the physics literature and call it the action.
In order for \eqref{eq:action-representation-of-distribution} to make sense as a \terminate{probability distribution}, $p(z)$ needs be normalizable so that we can satisfy
\be
\int d^\dimpre\! z\ p(z)=1\, .
\ee
That's where the \neo{normalization factor}or \term{partition function}
\be
Z\equiv \int d^\dimpre\! z\ e^{-\ac(z)}\, 
\ee
comes in.
After computing the partition function, we can define a \terminate{probability distribution} for a particular action $S(z)$ as
\be\label{eq:general-prob-ac-map}
p(z)\equiv \frac{e^{-\ac(z)}}{Z}\, .
\ee
Conversely, given a \terminate{probability distribution} we can associate an action, $\ac(z)=-\log \le[p(z)\ri]$, up to an additive ambiguity: the ambiguity arises because a constant shift in the action can be offset by the multiplicative factor in the partition function.\footnote{One convention is to pick the constant such that the \terminate{action} vanishes when evaluated at its global minimum.}

The action is a very convenient way to approximate certain types of statistical processes, particularly those with nearly-Gaussian statistics. To demonstrate this, we'll first start with the simplest action, which describes the Gaussian distribution, and then we'll show how to systematically perturb it in order to include various non-Gaussianities.

\subsubsection{Quadratic action and the Gaussian distribution}
\index{partition function!quadratic action}\index{Gaussian distribution!action}
Since we already know the functional form of the Gaussian distribution, it's simple to identify the action by reading it off from the exponent in \eqref{eq:multi-gauss-dis-compacter}
\be\label{eq:intro-quadratic-action-reprint}
\ac(z)=\frac{1}{2}\sum_{\mu,\nu=1}^\dimpre \ker^{\mu\nu}z_{\mu}z_{\nu}\, ,
\ee
where, as a reminder, the matrix $K^{\mu\nu}$ is the inverse of the variance matrix $K_{\mu\nu}$. The partition function is given by the normalization integral~\eqref{eq:det_formula} that we computed in~\S\ref{sec:Gauss}
\be
Z=\int d^\dimpre\! z\ e^{-\ac(z)} =\GI{\ker}= \sqrt{\dete{2\pi \ker}}\, .
\ee
This \textbf{quadratic action}\index{action!quadratic|textbf}\index{action!quadratic|seealso{Gaussian distribution}} is the simplest normalizable action and serves as a starting point for defining other distributions.

\index{bra-ket notation}\index{bra-ket notation|seealso{Gaussian expectation}}
As we will show next, integrals against the Gaussian distribution are a primitive for evaluating expectations against nearly-Gaussian distributions. Therefore, in order to differentiate between a general expectation and an integral against the Gaussian distribution, let us introduce a special \emph{bra-ket}, or $\bra \Vdot \ket$ notation for computing \emph{Gaussian} expectation values.
For an observable $\O(z)$, define a \textbf{Gaussian expectation}\index{Gaussian expectation|textbf}\index{Gaussian expectation|seealso{bra-ket notation}} as
\be\label{eq:gauss-braket}
\bra \O(z)\ket_{\ker}\equiv\frac{1}{ \sqrt{\dete{2\pi \ker}}}\int\le[ \prod_{\mu=1}^{\dimpre}dz_{\mu}\ri] \exp\!\le(-\frac{1}{2}\sum_{\mu,\nu=1}^{\dimpre}K^{\mu\nu} z_{\mu}z_{\nu}\ri)\O(z)\, .
\ee
In particular, with this notation we can write \terminate{Wick's theorem} as
\be\label{eq:Wick_compacter}
\bra z_{\mu_{1}}z_{\mu_{2}}\cdots z_{\mu_{2m}}\ket_{K}=\sum_{\text{all pairing}}K_{\mu_{k_1}\mu_{k_2}}\cdots K_{\mu_{k_{2m-1}}\mu_{k_{2m}}}\, .
\ee
If we're talking about a Gaussian distribution with variance $K_{\mu\nu}$, then we can use the notation $\E{\,\Vdot\,}$ and $\bra \Vdot \ket_K$ interchangeably.
If instead we're talking about a \terminate{nearly-Gaussian distribution} $p(z)$, then $\E{\,\Vdot\,}$ indicates expectation with respect to $p(z)$, \eqref{eq:expectation-value-definition}. However, in the evaluation of such an expectation, we'll often encounter Gaussian integrals, for which we'll use this bra-ket notation $\bra\Vdot \ket_K$ to simplify expressions.

\subsubsection{Quartic action and perturbation theory}
Now, let's find an action that represents a nearly-Gaussian distribution with a connected four-point correlator that is \emph{small} but non-vanishing
\be\label{eq:C4epsilon}
\Ec{z_{\mu_1} z_{\mu_2}z_{\mu_3}z_{\mu_4}}{\big|} = \o{\epsilon}\,  .
\ee
Here we have introduced a small parameter  $\epsilon\ll 1$ and indicated that the correlator should be of order $\epsilon$. For %
neural networks, we will later find that the role of the small parameter $\epsilon$ is played by $1/\text{width}$.

We should be able to generate a small connected four-point correlator  by \emph{deforming}\index{deformation!Gaussian distribution} the Gaussian distribution through the addition of a small quartic term to the quadratic action~\eqref{eq:intro-quadratic-action-reprint}, giving us a \textbf{quartic action}\index{action!quartic|textbf}\index{action!quartic|seealso{nearly-Gaussian distribution}}
\be\label{eq:quartic-action-intro}
\ac(z)=\frac{1}{2}\sum_{\mu,\nu=1}^\dimpre K^{\mu\nu}z_{\mu}z_{\nu} + \frac{\epsilon}{4!}\sum_{\mu,\nu,\rho,\lambda=1}^\dimpre V^{\mu\nu\rho\lambda}z_{\mu}z_{\nu}z_{\rho}z_{\lambda} \, ,
\ee
where the \textbf{quartic coupling}\index{coupling!quartic} $\epsilon V^{\mu\nu\rho\lambda}$ is an $(\dimpre \times \dimpre \times \dimpre \times \dimpre)$-dimensional \textbf{tensor}\index{tensor} that is completely symmetric in all of its four indices.
The factor of $1/4!$ is conventional
in order to compensate for the overcounting in the sum due to the symmetry of the indices. While it's not a proof of the connection, note that the coupling $\epsilon V^{\mu\nu\rho\lambda}$ has the right number of components to faithfully reproduce the four-point connected correlator~\eqref{eq:C4epsilon}, which is also an $(\dimpre \times \dimpre \times \dimpre \times \dimpre)$-dimensional symmetric  tensor. At least from this perspective we're off to a good start.

\index{connected correlator!four-point}\index{coupling!quartic} 
Let us now establish this correspondence between the quartic coupling and connected four-point correlator. 
Note that in general it is impossible to compute any expectation value in closed form with a non-Gaussian action -- this includes even the \terminate{partition function}. Instead, in order to compute the connected four-point correlator we'll need to employ \term{perturbation theory} to expand everything to first order in the small parameter $\epsilon$, each term of which can then be evaluated in a closed form. As this is easier done than said, let's get to the computations.

To start, let's evaluate the \terminate{partition function}:
\begin{align}
Z=& \int \le[\prod_{\mu}d z_{\mu}\ri]\ e^{-\ac(z)}\, \\
=&\int \le[\prod_{\mu}d z_{\mu}\ri] \exp\!\le(-\frac{1}{2}\sum_{\mu,\nu}K^{\mu\nu}z_{\mu}z_{\nu}%
-\frac{\epsilon}{24}\sum_{\rho_1,\ldots,\rho_4}V^{\rho_1\rho_2\rho_3\rho_4}z_{\rho_1}z_{\rho_2}z_{\rho_3}z_{\rho_4}\ri)\, \nonumber\\
=&\sqrt{\dete{2\pi\ker}}\bra\exp\!\le(-\frac{\epsilon}{24}\sum_{\rho_1,\ldots,\rho_4}V^{\rho_1\rho_2\rho_3\rho_4}z_{\rho_1}z_{\rho_2}z_{\rho_3}z_{\rho_4}\ri)\ket_{\ker}\, .\nonumber
\end{align}
In the second line we inserted our expression for the quartic action\index{action!quartic} \eqref{eq:quartic-action-intro}, and in the last line we used our \terminate{bra-ket notation}~\eqref{eq:gauss-braket} for a Gaussian expectation with variance $K_{\mu\nu}$. As advertised, the Gaussian expectation in the final line cannot be evaluated in closed form.
However, since our parameter $\epsilon$ is small, we can Taylor-expand the exponential to express the partition function as a sum of simple Gaussian expectations that can be evaluated using \terminate{Wick's theorem}~\eqref{eq:Wick_compacter}:
\begin{align}\label{eq:first-order-perturbation-theory-partition-function}
Z=&\sqrt{\dete{2\pi\ker}}\bra 1-\frac{\epsilon}{24}\sum_{\rho_1,\ldots,\rho_4}V^{\rho_1\rho_2\rho_3\rho_4}z_{\rho_1}z_{\rho_2}z_{\rho_3}z_{\rho_4}+\o{\epsilon^2}\ket_{\ker}\, \\
=&\sqrt{\dete{2\pi\ker}}\le[1-\frac{\epsilon}{24}\sum_{\rho_1,\ldots,\rho_4}V^{\rho_1\rho_2\rho_3\rho_4}\bra z_{\rho_1}z_{\rho_2}z_{\rho_3}z_{\rho_4}\ket_{\ker}+\o{\epsilon^2}\ri]\, \nonumber\\
=&\sqrt{\dete{2\pi\ker}}\le[1-\frac{\epsilon}{24}\sum_{\rho_1,\ldots,\rho_4}V^{\rho_1\rho_2\rho_3\rho_4}\le(K_{\rho_1\rho_2}K_{\rho_3\rho_4} +K_{\rho_1\rho_3}K_{\rho_2\rho_4}+ K_{\rho_1\rho_4}K_{\rho_2\rho_3} \ri)+\o{\epsilon^2}\ri]\, \nonumber\\
=&\sqrt{\dete{2\pi K}}\le[1-\frac{1}{8}\epsilon  \sum_{\rho_1,\ldots,\rho_4}V^{\rho_1\rho_2\rho_3\rho_4} K_{\rho_1\rho_2}K_{\rho_3\rho_4}+\o{\epsilon^2}\ri]\, .\nonumber
\end{align}
In the final line, we 
 were able to combine the three $K^2$ terms together by using the total symmetry of the quartic coupling and then relabeling some of the summed-over dummy indices.\index{coupling!quartic}

Similarly, let's evaluate the two-point correlator:
\begin{align}\label{eq:second-moment-interaction}
&\E{z_{\mu_1}z_{\mu_2}}=\frac{1}{Z}\int \le[\prod_{\mu}d z_{\mu}\ri]\ e^{-\ac(z)}\,  z_{\mu_1}z_{\mu_2}\,  \\
=&\frac{\sqrt{\dete{2\pi\ker}}}{Z}\bra z_{\mu_1}z_{\mu_2}\exp\!\le(-\frac{\epsilon}{24}\sum_{\rho_1,\ldots,\rho_4}V^{\rho_1\rho_2\rho_3\rho_4}z_{\rho_1}z_{\rho_2}z_{\rho_3}z_{\rho_4}\ri)\ket_{\ker}\, \notag \\
=&\frac{\sqrt{\dete{2\pi\ker}}}{Z}\le[\bra z_{\mu_1}z_{\mu_2}\ket_{\ker}-\frac{\epsilon}{24}\sum_{\rho_1,\ldots,\rho_4}V^{\rho_1\rho_2\rho_3\rho_4}\bra z_{\mu_1}z_{\mu_2}z_{\rho_1}z_{\rho_2}z_{\rho_3}z_{\rho_4}\ket_{\ker}+\o{\epsilon^2}\ri]\, \nonumber\\
=&\le[1+\frac{1}{8}\epsilon  \sum_{\rho_1,\ldots,\rho_4}V^{\rho_1\rho_2\rho_3\rho_4} K_{\rho_1\rho_2}K_{\rho_3\rho_4}\ri]K_{\mu_1\mu_2}\, \nonumber\\
&-\frac{\epsilon}{24}\sum_{\rho_1,\ldots,\rho_4}V^{\rho_1\rho_2\rho_3\rho_4}\le(3K_{\mu_1\mu_2}K_{\rho_1\rho_2}K_{\rho_3\rho_4}+12K_{\mu_1\rho_1}K_{\mu_2\rho_2}K_{\rho_3\rho_4} \ri)+\o{\epsilon^2}  \, \nonumber\\
=&K_{\mu_1\mu_2}-\frac{\epsilon}{2}\sum_{\rho_1,\ldots,\rho_4}V^{\rho_1\rho_2\rho_3\rho_4}K_{\mu_1\rho_1}K_{\mu_2\rho_2}K_{\rho_3\rho_4}+\o{\epsilon^2}\, .\nonumber
\end{align}
Here, to go from the first line to the second line we inserted our expression for the quartic action\index{action!quartic} \eqref{eq:quartic-action-intro} and rewrote the integral as a Gaussian expectation. Then, after expanding in $\epsilon$ to first order, in the next step
we substituted \eqref{eq:first-order-perturbation-theory-partition-function} in for the partition function  $Z$ in the denominator and expanded $1/Z$ to the first order in $\epsilon$ using the expansion $1/(1-x)=1+x+\o{x^2}$. In that same step, we also noted that, of the fifteen terms coming from the Gaussian expectation $\bra z_{\mu_1}z_{\mu_2}z_{\rho_1}z_{\rho_2}z_{\rho_3}z_{\rho_4}\ket_{\ker}$,
there are three ways in which $z_{\mu_1}$ and $z_{\mu_2}$ contract with each other but twelve ways in which they don't. Given again the symmetry of $V^{\rho_1\rho_2\rho_3\rho_4}$, this is the only distinction that matters.

At last,
let's compute the full four-point correlator: 
\begin{align}\label{eq:full-four-point-intro}
&\E{z_{\mu_1}z_{\mu_2}z_{\mu_3}z_{\mu_4}} =\frac{1}{Z}\int \le[\prod_{\mu}d z_{\mu}\ri]\ e^{-\ac(z)}\,  z_{\mu_1}z_{\mu_2}z_{\mu_3}z_{\mu_4}\, \\
=&\frac{\sqrt{\dete{2\pi\ker}}}{Z}\le[\bra z_{\mu_1}z_{\mu_2}z_{\mu_3}z_{\mu_4}\ket_{\ker}-\frac{\epsilon}{24}\sum_{\rho_1,\ldots,\rho_4}V^{\rho_1\rho_2\rho_3\rho_4}\bra z_{\mu_1}z_{\mu_2}z_{\mu_3}z_{\mu_4}z_{\rho_1}z_{\rho_2}z_{\rho_3}z_{\rho_4}\ket_{\ker}+\o{\epsilon^2}\ri]\, \nonumber\\
=&\le[1+\frac{1}{8}\epsilon  \sum_{\rho_1,\ldots,\rho_4}V^{\rho_1\rho_2\rho_3\rho_4} K_{\rho_1\rho_2}K_{\rho_3\rho_4}\ri]\le[K_{\mu_1\mu_2}K_{\mu_3\mu_4}+K_{\mu_1\mu_3}K_{\mu_2\mu_4}+K_{\mu_1\mu_4}K_{\mu_2\mu_3}\ri]\, \nonumber\\
&-\frac{\epsilon}{24}\sum_{\rho_1,\ldots,\rho_4}V^{\rho_1\rho_2\rho_3\rho_4} \nonumber\\
 &\times\Big(3K_{\mu_1\mu_2}K_{\mu_3\mu_4}K_{\rho_1\rho_2}K_{\rho_3\rho_4}+12K_{\mu_1\rho_1}K_{\mu_2\rho_2}K_{\mu_3\mu_4}K_{\rho_3\rho_4}+12K_{\mu_3\rho_1}K_{\mu_4\rho_2}K_{\mu_1\mu_2}K_{\rho_3\rho_4}\, \nonumber\\
 &\ \ +3K_{\mu_1\mu_3}K_{\mu_2\mu_4}K_{\rho_1\rho_2}K_{\rho_3\rho_4}+12K_{\mu_1\rho_1}K_{\mu_3\rho_2}K_{\mu_2\mu_4}K_{\rho_3\rho_4}+12K_{\mu_2\rho_1}K_{\mu_4\rho_2}K_{\mu_1\mu_3}K_{\rho_3\rho_4}\, \nonumber\\
  &\ \ +3K_{\mu_1\mu_4}K_{\mu_2\mu_3}K_{\rho_1\rho_2}K_{\rho_3\rho_4}+12K_{\mu_1\rho_1}K_{\mu_4\rho_2}K_{\mu_2\mu_3}K_{\rho_3\rho_4}+12K_{\mu_2\rho_1}K_{\mu_3\rho_2}K_{\mu_1\mu_4}K_{\rho_3\rho_4}\, \nonumber\\
 &\quad +24K_{\mu_1\rho_1}K_{\mu_2\rho_2}K_{\mu_3\rho_3}K_{\mu_4\rho_4}\Big)+\o{\epsilon^2}  \, .\nonumber
\end{align}
To go from the first line to the second line we inserted our expression for the quartic action\index{action!quartic} \eqref{eq:quartic-action-intro}, expanded to first order in $\epsilon$, and rewrote in the \terminate{bra-ket notation} \eqref{eq:gauss-braket}. On the third line, we again substituted in the expression~\eqref{eq:first-order-perturbation-theory-partition-function} for the partition function $Z$, expanded $1/Z$ to first order in $\epsilon$, and then used \terminate{Wick's theorem} \eqref{eq:Wick_compacter} to evaluate the fourth and eighth Gaussian moments. (Yes, we know that the evaluation of $\bra z_{\mu_1}z_{\mu_2}z_{\mu_3}z_{\mu_4}z_{\rho_1}z_{\rho_2}z_{\rho_3}z_{\rho_4}\ket_{\ker}$ is not fun. The breakdown of the terms depends again on whether or not the $\mu$-type indices are contracted with the $\rho$-type indices or not.)
We can simplify this expression by noticing that some terms cancel due to $\frac{1}{8}-\frac{3}{24}=0$ and some other terms can be nicely regrouped once we notice  through the expression for the two-point correlator~\eqref{eq:second-moment-interaction} that
\begin{align}
&K_{\mu_1\mu_2}K_{\mu_3\mu_4}-\frac{\epsilon}{24}\sum_{\rho_1,\ldots,\rho_4}V^{\rho_1\rho_2\rho_3\rho_4}\le(12K_{\mu_1\rho_1}K_{\mu_2\rho_2}K_{\mu_3\mu_4}K_{\rho_3\rho_4}+12K_{\mu_3\rho_1}K_{\mu_4\rho_2}K_{\mu_1\mu_2}K_{\rho_3\rho_4}\ri)\, \nonumber\\
&=\E{z_{\mu_1}z_{\mu_2}}\E{z_{\mu_3}z_{\mu_4}}+\o{\epsilon^2}\, ,
\end{align}
yielding in the end
\begin{align}
&\E{z_{\mu_1}z_{\mu_2}z_{\mu_3}z_{\mu_4}}\, \\
=&\E{z_{\mu_1}z_{\mu_2}}\E{z_{\mu_3}z_{\mu_4}}+\E{z_{\mu_1}z_{\mu_3}}\E{z_{\mu_2}z_{\mu_4}}+\E{z_{\mu_1}z_{\mu_4}}\E{z_{\mu_2}z_{\mu_3}}\, \nonumber\\
&-\epsilon\sum_{\rho_1,\ldots,\rho_4}V^{\rho_1\rho_2\rho_3\rho_4} K_{\mu_1\rho_1}K_{\mu_2\rho_2}K_{\mu_3\rho_3}K_{\mu_4\rho_4}+\o{\epsilon^2}\, .\nonumber
\end{align}

\index{connected correlator!four-point}\index{action!quartic}\index{coupling!quartic}\index{nearly-Gaussian distribution}
Given the full four-point correlator \eqref{eq:full-four-point-intro} and the two-point correlator \eqref{eq:second-moment-interaction},
we can finally evaluate the connected four-point correlator~\eqref{eq:C4} 
as
\begin{align}\label{eq:single-variable-connected-four-point}
\Ec{z_{\mu_1} z_{\mu_2}z_{\mu_3}z_{\mu_4}}{\big|} = -\epsilon\sum_{\rho_1,\ldots,\rho_4}V^{\rho_1\rho_2\rho_3\rho_4} K_{\mu_1\rho_1}K_{\mu_2\rho_2}K_{\mu_3\rho_3}K_{\mu_4\rho_4}+\o{\epsilon^2}\, .
\end{align}
This makes explicit the relationship between the connected four-point correlator and the quartic coupling in the action, when both are small.
We see that for the nearly-Gaussian distribution realized by the quartic action \eqref{eq:quartic-action-intro}, the distribution is -- as promised -- \emph{nearly} Gaussian: the strength of the coupling $\epsilon V^{\rho_1\rho_2\rho_3\rho_4}$ directly controls the distribution's deviation from Gaussian statistics, as measured by the connected four-point correlator. This also shows that the four-index tensor $V^{\rho_1\rho_2\rho_3\rho_4}$ creates nontrivial correlations between the components $z_{\rho_1}z_{\rho_2}z_{\rho_3}z_{\rho_4}$ that cannot otherwise be built up by the  correlation $K_{\mu\nu}$ in any pair of random variables $z_{\mu}z_{\nu}$.

\index{coupling!quartic}
Finally, note that the connected two-point correlator \eqref{eq:second-moment-interaction} -- i.e.~the covariance of this \terminate{nearly-Gaussian distribution} -- is also shifted from its Gaussian value of $K_{\mu_1\mu_2}$ by the quartic coupling $\epsilon V^{\rho_1\rho_2\rho_3\rho_4}$. Thus, the nearly-Gaussian deformation not only creates complicated patterns of four-point correlation as measured by the connected four-point correlator \eqref{eq:single-variable-connected-four-point}, it also can modify the details of the Gaussian two-point correlation.

Now that we see how to compute the statistics of a \terminate{nearly-Gaussian distribution}, let's take a step back and think about what made this possible.
We can perform these perturbative calculations any time there exists in the problem a dimensionless parameter~$\epsilon$ that is small $\epsilon \ll 1$, but nonzero $\epsilon > 0$.
This makes \terminate{perturbation theory} an extremely powerful tool for theoretical analysis any time a problem has any extreme scales, small \emph{or} large.

\index{$1/n$ expansion}\index{large-$n$ expansion|see{$1/n$ expansion}}\index{nearly-Gaussian distribution}\index{non-Gaussian distribution}\index{non-Gaussian distribution|seealso{nearly-Gaussian distribution}}
Importantly, this is directly relevant to theoretically understanding neural networks in practice. 
As we will explain in the following chapters, real networks have a parameter $n$ -- the number of neurons in a layer -- that is typically large $n \gg 1$, but certainly not infinite $n < \infty$.  This means that we can expand the distributions that describe such networks in the inverse of the large parameter as $\epsilon =1/n$.
Indeed, when the parameter $n$ is large -- as is typical in practice -- the distributions that describe neural networks become nearly-Gaussian and thus theoretically tractable.
This type of expansion is known as the \textbf{1/\emph{n} expansion} or \textbf{large-\emph{n} expansion} and will be one of our main tools for learning the principles of deep learning theory.

\subsubsection{\emph{Aside}: statistical independence and interactions}
\index{interactions!connection to statistical (in)dependence}\index{statistical independence|textbf}\index{action!quartic}\index{coupling!quartic}\index{statistical independence!absence of interactions and connection to Gaussian distribution}

The quartic action~\eqref{eq:quartic-action-intro} is one of the simplest models of an \term{interacting theory}\index{interacting theory|seealso{non-Gaussian distribution}}. We showed this explicitly by connecting the quartic coupling to the non-Gaussian statistics of the non-vanishing connected four-point correlator. Here, let us try to offer an intuitive meaning of \emph{interaction} by appealing to the notion of \emph{statistical independence}.

Recall from the probability\index{probability (branch of mathematics)} theory that two random variables $x$ and $y$ are statistically independent if their joint distribution factorizes as
\be\label{eq:independence-random-variables}
p(x, y) = p(x)p(y) \, .
\ee
For the \terminate{Gaussian distribution}, if the variance matrix $K_{\mu\nu}$ is diagonal, there is no correlation at all between different components of $z_{\mu}$; they are manifestly statistically independent from each other. 

Even if $K_{\mu\nu}$ is not diagonal, we can still unwind the correlation of a Gaussian distribution by rotating to the right basis. As discussed in~\S\ref{sec:Gauss}, there always exists an \terminate{orthogonal matrix} $O$ that diagonalizes\index{diagonalization} the covariance as $(OK O^T)_{\mu\nu}=\lambda_{\mu} \delta_{\mu\nu}$. In terms of the variables $u_{\mu}\equiv (O z)_{\mu}$, the distribution looks like
\begin{align}\label{eq:gaussian-statistical-independence-day}
p(z) &= \frac{1}{\sqrt{\dete{2\pi K}}} \exp\!\le(-\sum_{\mu=1}^\dimpre\frac{u_\mu^2}{2\lambda_\mu}\ri)=\prod_{\mu=1}^\dimpre  \le( \frac{e^{-\frac{u_\mu^2}{2\lambda_\mu}}}{\sqrt{2\pi \lambda_\mu}}  \ri) =p\!\le(u_1\ri)\cdots p(u_\dimpre) \, .
\end{align}
Thus, we see that in the $u$-coordinate basis the original multivariable Gaussian distribution factorizes into $\dimpre$ single-variable Gaussians that are statistically independent.\index{statistical independence}

\index{coupling!non-Gaussian}\index{statistical independence}\index{interactions}
We also see that in terms of the action, statistical independence is characterized by the action breaking into a sum over separate terms.
This unwinding of interaction between variables is generically impossible when there are nonzero non-Gaussian couplings.
For instance, there are $\sim \dimpre^2$ components of an orthogonal matrix $O_{\mu\nu}$ to change basis, while there are $\sim \dimpre^4$ components of the quartic coupling $\epsilon V^{\mu\nu\rho\lambda}$ that correlate random variables, so it is generically impossible to re-express  the quartic action as a sum of functions of $\dimpre$ different variables. Since the action cannot be put into
a sum over $\dimpre$ separate terms,
the joint distribution cannot factorize, and the components will not be independent from each other.
Thus, it is impossible to factor the nearly-Gaussian distribution into the product of $\dimpre$ statistically independent distributions.
In this sense, what is meant by \emph{interaction} is the breakdown of \neo{statistical independence}.\footnote{
An astute reader might wonder if there is any interaction when we consider a single-variable distribution with $\dimpre = 1$, since there's no other variables to interact with. For nearly-Gaussian distributions, even if $\dimpre=1$, we saw in~\eqref{eq:second-moment-interaction} that the variance of the distribution is shifted from its Gaussian value, $K$, and depends on the quartic coupling $\epsilon V$. In physics, we say that
this shift is due to the \emph{self-interaction}\index{self-interaction|see{interactions}}\index{interactions!self-interactions} induced by the quartic coupling $\epsilon V$,
since it modifies the value of observables from the \emph{free} Gaussian theory that we are comparing to, even though there's no notion of statistical independence to appeal to here.\index{nearly-Gaussian distribution}

Said another way, even though the action just involves one term, such a non-Gaussian distribution does not have a closed-form solution for its partition function or correlators; i.e.~there's no trick that lets us compute integrals of the form $e^{-\ac(z)}$ exactly, when $\ac(z) = \frac{z^2}{2K} + \frac{1}{4!} \epsilon Vz^4$. This means that we still have to make use of \terminate{perturbation theory} to analyze the self-interaction in such distributions.
\label{footnote:self-interaction}
}

\subsubsection{Nearly-Gaussian actions}
\index{nearly-Gaussian distribution}\index{nearly-Gaussian distribution!action}\index{action}
Having given a concrete example in which we illustrated how to deform the quadratic action to realize the simplest nearly-Gaussian distribution, we now give a more general perspective on nearly-Gaussian distributions. 
In what follows, we will continue to require that our distributions are invariant under the \terminate{parity symmetry} that takes $z_\mu \to -z_\mu$. In the action representation, this corresponds to including only terms of even degree.\footnote{The imposition of such a \terminate{parity symmetry}, and thus the absence of odd-degree terms in the action, means that all of the odd moments and hence all of the odd-point connected correlators will vanish.}

With that caveat in mind, though otherwise very generally, we can express a \term{non-Gaussian distribution} by \emph{deforming}\index{deformation!Gaussian distribution} the Gaussian action as
\be\label{eq:schematic-action-decomposition}
\ac(z) = \frac{1}{2}\sum_{\mu,\nu=1}^\dimpre \ker^{\mu\nu}z_{\mu}z_{\nu} +  \sum_{m=2}^{k} \frac{1}{(2m)!} \sum_{\mu_1, \ldots, \mu_{2m}=1}^\dimpre s^{\mu_1 \cdots \mu_{2m}} z_{\mu_1} \cdots z_{\mu_{2m}} \, ,
\ee
where the factor of $1/(2m)!$ is conventional in order to compensate for the overcounting in the sum due to the implied symmetry of the indices $\mu_1, \ldots, \mu_{2m}$ in the coefficients $s^{\mu_1 \cdots \mu_{2m}}$, given the permutation symmetry of the product of variables $z_{\mu_1} \cdots z_{\mu_{2m}}$.
The number of terms in the non-Gaussian part of the action is controlled by the integer $k$.
If $k$ were unbounded, then $S(z)$ would be an arbitrary even function, and $p(z)$ could be any parity-symmetric distribution. %
The action is most useful when the expanded polynomial  $S(z)$ truncated to reasonably small degree $k$ -- like $k=2$ for the quartic action -- yields a good representation for the statistical process of interest.

\index{coupling!non-Gaussian}
The coefficients $s^{\mu_1 \cdots \mu_{2m}}$ are generally known as \textbf{non-Gaussian couplings}, and they control the \term{interactions} of the $z_\mu$.\footnote{
    In the similar vein, the coefficient $K^{\mu\nu}$ in the \terminate{action} is sometimes called a \emph{quadratic coupling}\index{coupling!quadratic} since the \emph{coupling} of the component $z_\mu$ with the component $z_\nu$ in the quadratic action leads to a nontrivial \emph{correlation}, i.e.~$\cov{z_\mu}{z_\nu} = K_{\mu\nu}$.
} In particular, there is a direct correspondence between the product of the specific components $z_\mu$ that appear together in the action and the presence of connected correlation between those variables, with the degree of the term in \eqref{eq:schematic-action-decomposition} directly contributing to connected correlators of that degree.
We saw an example of this in~\eqref{eq:single-variable-connected-four-point}, which connected the quartic term to the connected four-point correlator.
In this way, the couplings give a very direct way of controlling the degree and pattern of non-Gaussian correlation,
and the overall degree of the action offers a way of systematically including more and more complicated patterns of such correlations.

\index{connected correlator!}

If you recall from \S\ref{sec:not-Gauss}, we defined nearly-Gaussian distributions as ones for which all these connected correlators are small. Equivalently, from the action perspective, a \terminate{nearly-Gaussian distribution} is a non-Gaussian distribution with an \terminate{action} of the form \eqref{eq:schematic-action-decomposition} for which all the couplings $s^{\mu_1 \cdots \mu_{2m}}$ are parametrically small for all $1 \leq m \leq k$:
\be\label{eq:parametrical-small-world}
\vert s^{\mu_1 \cdots \mu_{2m}}\vert \ll |K^{\mu\nu}|^{m} \, ,
\ee
where this equation is somewhat schematic given the mismatch of the indices.\footnote{
This schematic equation is, nonetheless, dimensionally consistent.\label{foot:dimensional-analysis} To support that remark, let us give a brief introduction to \emph{dimensional analysis}\index{dimensional analysis|textbf}: let the random variable $z_\mu$ have dimension $\ZU$, which we denote as $[z_{\mu}]=\ZU^1$. By \emph{dimension}, you should have in mind something like a unit of length, so e.g.~we read the expression $[z_{\mu}]=\ZU^1$ as ``a component of $z$ is measured in units of $\ZU$.'' The particular units are arbitrary: e.g.~for length, we can choose between \texttt{meters} or \texttt{inches} or \texttt{parsecs} as long as we use a unit of length but \emph{not}, say, \texttt{meters}$^2$, which instead would be a unit of area. Importantly, we cannot add or equate quantities that have different units: it doesn't make any logical sense to add a length to an area. This is similar to the concept of \neo{type safety}\index{type safety|seealso{dimensional analysis}} in computer science, e.g.~we should not add a type \texttt{str}\index{str@\texttt{str}|see{type (data)}}\index{type (data)!string} variable to a type \texttt{int}\index{int@\texttt{int}|see{type (data)}}\index{type (data)!integer} variable.

Now, since the action $S(z)$ is the argument of an exponential $p(z) \propto e^{-S(z)}$, it must be \emph{dimensionless}; otherwise, the exponential $e^{-S}=1-S+\frac{S^2}{2}+\ldots$ would violate the addition rule that we just described. From this dimensionless requirement for the action, we surmise that the inverse of the covariance matrix has dimension $[K^{\mu\nu}]=\ZU^{-2}$, and that the covariance itself has dimension $[K_{\mu\nu}]=\ZU^{2}$. Similarly, all the non-Gaussian couplings in \eqref{eq:schematic-action-decomposition} have dimensions $[s^{\mu_1 \cdots \mu_{2m}}]=\ZU^{-2m}$. Thus, both sides of~\eqref{eq:parametrical-small-world} have the same dimension, making this equation dimensionally consistent.  

Even more concretely, consider the quartic action \eqref{eq:quartic-action-intro}. If we let the tensorial part of the quartic coupling have dimensions $[V^{\mu\nu\rho\lambda}]=\ZU^{-4}$, then the parameter $\epsilon$ is dimensionless, as claimed. This means that we can consistently compare $\epsilon$ to unity, and its parametric smallness $\epsilon\ll 1$ means that the full quartic coupling $\epsilon V^{\mu\nu\rho\lambda}$ is much smaller than the square of the quadratic coupling, and that the connected four-point correlator \eqref{eq:single-variable-connected-four-point} is much smaller than the square of the connected two-point correlator \eqref{eq:second-moment-interaction}. 
\index{action!quartic}\index{coupling!quartic}\index{dimensional analysis}\index{coupling!quadratic}
}
Importantly the comparison is with an appropriate power of the inverse variance or quadratic coupling $K^{\mu\nu}$ since, as we already explained, the variance sets the scale of the Gaussian distribution to which we are comparing these nearly-Gaussian distributions.

\index{connected correlator!higher-point}\index{nearly-Gaussian distribution}\index{Gaussian distribution}\index{statistical independence}\index{hierarchical scaling}
As we will see in \S\ref{ch:ngp}, wide neural networks are described by nearly-Gaussian distributions. In particular, we will find that such networks are described by a special type of nearly-Gaussian distribution where the connected correlators are \emph{hierarchically} small, scaling as
\be\label{eq:connected-correlator-hierarchy}
\Ec{z_{\mu_1} \cdots z_{\mu_{2m}}}{\big|} = O(\epsilon^{m-1})\, ,
\ee
with the same parameter $\epsilon$ controlling the different scalings for each of the $2m$-point connected correlators.
Importantly, the non-Gaussianities coming from higher-point connected correlators become parametrically less important as $\epsilon$ becomes smaller.

\index{action!truncation}\index{nearly-Gaussian distribution}\index{action!truncation|seealso{hierarchical scaling}}
This means that for a \terminate{nearly-Gaussian distribution} with hierarchical scalings~\eqref{eq:connected-correlator-hierarchy}, we can consistently approximate the distribution by \emph{truncating} the action at some fixed order in $\epsilon$. To be concrete, we can use an action of the form~\eqref{eq:schematic-action-decomposition} to faithfully represent all the correlations up to order $O(\epsilon^{k-1})$, neglecting connected correlations of order $O(\epsilon^k)$ and higher. The resulting action offers a useful and effective description for the statistical process of interest, as long as $\epsilon$ is small enough and $k$ is high enough that $O(\epsilon^k)$ is negligible.

\index{Gaussian distribution}\index{statistical dependence}\index{statistical dependence|seealso{interactions}}\index{statistical dependence|seealso{nearly-Gaussian distribution}}
In practice, a quartic action~\eqref{eq:quartic-action-intro} truncated to $k=2$  will let us model realistic finite-width neural networks.
This quartic action captures the important \emph{qualitative} difference between nearly-Gaussian distributions and the Gaussian distribution, incorporating nontrivial \terminate{interactions} between the different components of the random variable.
In addition, the difference between the statistics \eqref{eq:connected-correlator-hierarchy} of a nearly-Gaussian distribution truncated to $\o{\epsilon}$ versus one truncated to $\o{\epsilon^2}$ is mostly \emph{quantitative}: in both cases there are nontrivial non-Gaussian correlations, but the pattern of higher-order correlation differs only in a small way, with the difference suppressed as $\o{\epsilon^2}$.
In this way, the distribution represented by the quartic action is complex enough to capture the most salient non-Gaussian effects in neural networks while still being simple enough to be analytically tractable.

%% file: Chp2-MLP/2_global.tex
\chapter{Neural Networks}
\label{ch:NN}

\index{Perceptron architecture}
\epigraph{On being asked, ``How is Perceptron performing today?'' I am often tempted to respond, ``Very well, thank you, and how are Neutron and Electron behaving?''}{Frank Rosenblatt, inventor of the perceptron and also the Perceptron  \cite{rosenblatt1961principles}.\index{Rosenblatt, Frank}}

\noindent{}With our mathematical lessons concluded, we turn to an introductory overview of deep learning.

In~\S\ref{sec:MLP_intro}, we introduce the basic components of neural network architectures -- neurons, activations, biases, weights, and layers -- in order to define the \emph{multilayer perceptron} (MLP),
a simple model that is iteratively composed of these basic components.  Given that all deep networks are by definition iteratively composed of many structurally identical layers, MLPs will play the role of archetype network architecture for illustrating the principles of deep learning throughout the book.
This class of neural-network models is rich enough to capture all the essential aspects of deep learning theory, while simple enough to maintain the pedagogical focus of the book. Nevertheless,
we'll also briefly comment on how one could work out an effective theory for other network architectures.

In~\S\ref{sec:activations} we list some common activation functions that are 
often used in practice.

Finally, we discuss in~\S\ref{sec:MLP_distribution} how MLPs are initialized. %
Here, we make a key conceptual shift from thinking about the weights and biases as the random variables
to thinking about the induced distribution over the neural activities and network outputs.
The expressions we derive here will provide a natural starting point for our analysis in~\S\ref{ch:ngp} when we start developing our effective theory of MLPs with general activation functions.

\section{Function Approximation}\label{sec:MLP_intro}

\index{neural network}
The subject of \emph{artificial neural networks}\index{artificial neural network|see{neural network}} has a rich history as \terminate{cognitive science} and \terminate{neuroscience}--inspired
\terminate{artificial intelligence}.\footnote{The \neo{artificial neuron} was invented by McCulloch and Pitts in 1943 \cite{mcculloch1943logical} as a model of the \neo{biological neuron}. Their neuron was essentially a perceptron with a bias, but did not have learnable weights. The perceptron model\index{Perceptron architecture}, with learnable weights, was invented by Rosenblatt in 1958 \cite{rosenblatt1958perceptron}.
Deep learning 
really came into its own
in 2012~\cite{ImageNet2012} after the realization that the \terminate{graphical processing unit}\index{GPU|see{graphical processing unit}} (GPU) is well-suited for the parallel computations required to train and run neural networks.\index{deep learning!history}\index{neural network!history}} Here, our starting point will be a discussion of the function, $f(x)$.

\index{Taylor series}
Some functions are really simple, %
easily described in terms of the elementary operations\index{elementary operation}: addition, subtraction, multiplication, and division.
For instance, consider either the identity function $f(x)=x$ or the exponential function $f(x)=e^{x}$. The former is the definition of trivial, involving no operations. The latter is a special function and can be defined in many ways, e.g.~through its Taylor series
\be\label{eq:exponential-Taylor-expansion}
e^{x}\equiv\sum_{k=0}^\infty \frac{x^k}{k!}\, .
\ee
This definition constructs the exponential function in terms of elementary operations of addition, multiplication, and division: the numerator $x^k$ represents the repeated multiplication of the variable $x$ for $k$ times, and the factorial $k!$ in the denominator represents the repeated multiplication of integers $k!=1 \times 2 \times \cdots \times (k-1) \times k$. Although this description of the exponential function involves a sum of an infinite number of terms, the actual instructions~\eqref{eq:exponential-Taylor-expansion} for computing this function in terms of these simple operations are so compact that they takes up only about one seventh of a line and, for many purposes, it only takes the first few terms in the sum
to get a
useful approximation of $e^x$.

Some functions are really complicated and their description in terms of elementary operations is unlikely to fit in the confines of any printed book. For instance, imagine a function $f(x)$ that takes as input an image $x_i$ 
-- represented as a vector of numbers corresponding to a black-and-white pixelated image -- 
and outputs $1$ if the image $x_i$ depicts a cat and $0$ otherwise. While such a classification function should exist since humans can recognize images of cats, it's not at all clear how to describe such a function in terms of simple operations like addition and multiplication. The subject of \term{artificial intelligence} (AI) is mostly concerned with functions of this sort: easy for humans to compute, but difficult for humans to describe in terms of elementary operations.

The conceptual leap needed to represent such hard-to-describe functions is to start with a flexible \emph{set} of functions $\le\{f(x;\theta)\ri\}$, constructed from simple components parametrized by a vector of adjustable \term{model parameters} $\theta_{\mu}$. One then tries to tune these model parameters $\theta_{\mu}$ judiciously in order to approximate the original complicated function such that $f(x;\theta^\star) \approx f(x)$. The description of the set of functions $\le\{ f(x;\theta)\ri\}$ as well as the settings of the model parameters $\theta^\star_{\mu}$ then serve as a useful approximate description of a desired function $f(x)$. This is called \term{function approximation} and the procedure for adjusting the model parameters $\theta_{\mu}$ is called a \term{learning algorithm}.%

\index{dataset|see{input data}}\index{input data|seealso{training set}}\index{input data|seealso{test set}}\index{input data|seealso{validation set}}
To be more concrete, let us represent the collection of 
inputs to our function $f(x)$ as a set $\D$ of $n_0$-dimensional vectors
\be
\D=\le\{\x{i}{\alpha}\ri\}_{\alpha=1,\ldots,\ND}\, ,
\ee
called \textbf{input data}.
Here, the \textbf{sample index} $\alpha$ labels each sample in the dataset of $\ND$ elements, and the vector index $i=1,\ldots,n_0$ labels the component of the input vector. 
In our motivating example above, each number $\x{i}{\alpha}$ refers to the $i$-th pixel
of the $\alpha$-th  image in the dataset\index{input data|textbf} $\D$ of $\ND$ images, each of which might or might not 
depict a cat. By adjusting the model parameters $\theta_{\mu}$ so that the function $f(x;\theta^\star)$ outputs the correct answer for as much input data as possible, we can try to approximate the elusive cat-or-not %
function in a way that no longer defies description. The overall idea of \neo{training} such functions using a dataset $\D$ -- rather than \neo{programming} them --  goes by the name \term{machine learning}\index{machine learning|seealso{statistics (branch of mathematics)}}
and stands in contrast to the conventional von Neumann\index{von Neumann, John} model of the digital computer. \index{sample indices}

\index{function approximation}\index{artificial intelligence}\index{deep learning}
While any set of parameterized functions can be used for function approximation,\footnote{\index{Gaussian function}E.g.~consider a sum of  Gaussian functions, where the mean and variance of each Gaussian play the role of the adjustable parameters.} our focus will be on a particular set of composable functions originally derived from a simplified model of the \terminate{brain}. Such functions were originally termed \textbf{artificial neural networks} and are now just referred to as \textbf{neural networks}\index{neural network|textbf}. \textbf{Deep learning} is a branch of machine learning that uses neural networks as function approximators, with a particular emphasis on stacking \emph{many layers} of structurally similar components. Let's see how this works in more detail.

\index{biological neuron}\index{artificial neuron}
The most basic component of the neural network is the \term{neuron}. Loosely inspired by the behavior of biological neurons, the artificial neuron essentially consists of two simple operations:
\bi
\item The \term{preactivation} $z_i$ of a neuron is a linear aggregation of 
incoming signals $s_j$ where each signal is weighted by $W_{ij}$ and
biased by $b_i$
\be\label{eq:preactivation}
z_{i}(s) = b_i+\sum_{j=1}^{n_{\text{in}}}W_{ij}\, s_{j}\,  \quad \text{for} \quad i=1,\ldots,n_{\text{out}}\, .
\ee
\item Each neuron then \emph{fires} or not according to the weighted and biased evidence, i.e.~according to the value of the preactivation $z_i$, and produces an \term{activation}
\be
\sigma_i \equiv \sigma\!\le(z_{i}\ri) \, .
\ee
The scalar-valued function $\sigma(z)$ is called the \term{activation function} and acts independently on each component of the preactivation vector.
\ei
Taken together, these  $n_{\text{out}}$ neurons form a \term{layer}, which takes in the $n_{\text{in}}$-dimensional vector of signals $s_j$ and outputs the $n_{\text{out}}$-dimensional vector of  activations $\sigma_i$.   With this collective perspective, a layer is parameterized by a vector of \term{biases} $b_i$ and a matrix of \term{weights} $W_{ij}$, where $i=1,\ldots,n_{\text{out}}$ and $j=1,\ldots,n_{\text{in}}$, together with a fixed activation function $\sigma(z)$.\index{weights|seealso{model parameters}}\index{biases|seealso{model parameters}}

With these components, we can make an increasingly flexible set of functions by organizing many neurons into a layer and then iteratively stacking many such layers, so that the outgoing activations of the neurons in one layer become the input signals to the neurons in some other layer.
The organization of the neurons and their pattern of connections is known as the neural network \term{architecture}.
The archetypical neural network architecture based on this principle of stacking 
layers of many neurons is called the \term{multilayer perceptron} (MLP).\footnote{Here, the name ``perceptron''\index{perceptron|see{Perceptron architecture}} was inherited from Rosenblatt's Perceptron architecture\index{Perceptron architecture} \cite{rosenblatt1958perceptron}, which was originally envisioned for emulating \neo{human perception}. The name perceptron is also used to refer to the original step-function activation function, cf.~the first entry of \S\ref{sec:activations}. 
    \index{activation function!perceptron@$\texttt{perceptron}$}
} 

The activation function is usually chosen to be a nonlinear function in order to increase the \terminate{expressivity}
of the neural-network function $f(x;\theta)$.
The simplest -- and historically first -- activation function either fires or does not fire: $\sigma(z) =1$ for $z\geq0$ and $\sigma(z) =0$ for $z<0$. In other words, each neuron fires if and only if the weighted accumulated evidence $\sum_jW_{ij}\, x_{j}$ exceeds the firing threshold $-b_i$. More generally, activation functions are not binary and can incorporate
the strength of the evidence into their output. In~\S\ref{sec:activations} we'll describe many of the commonly-used activation functions in deep learning. 

\begin{figure}
\begin{center}
 \includegraphics[width=.9\linewidth]{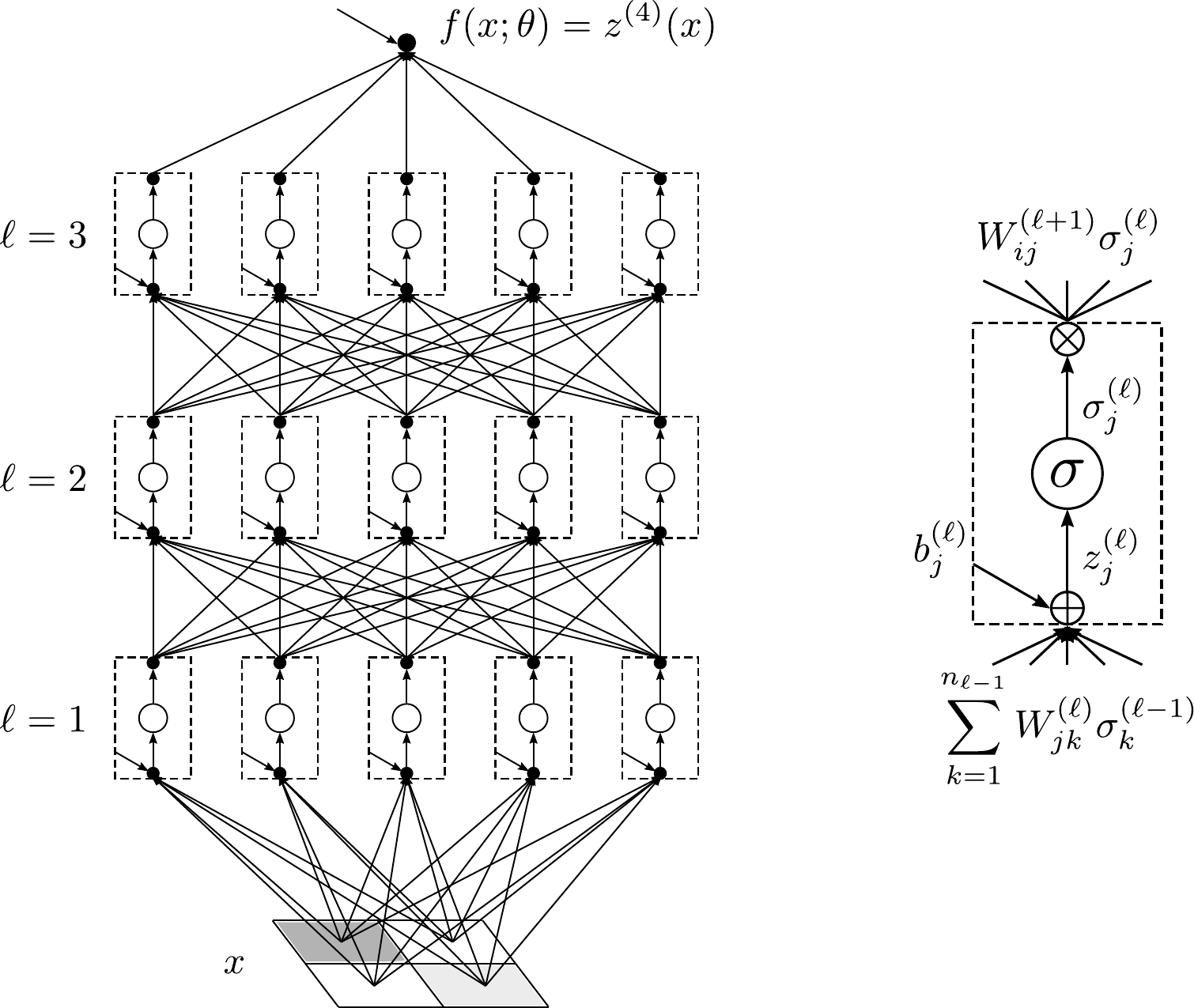}
\caption{
\textbf{Left:} depiction of the neurons and connections for an example \terminate{multilayer perceptron} (MLP) architecture.  This particular MLP has $L=4$ layers, defining a set of functions $f(x; \theta)$ with input dimension $n_0=4$ and output dimension $n_{4}=1$. The three hidden layers have five neurons each $n_1, n_2, n_3 = 5$, implying $P=91$ total model parameters. 
The graph describing the connections between neurons is a \neo{directed acyclic graph}, meaning that signals only propagate in one direction and do not loop inside the network.
For this reason, MLPs are also sometimes called \emph{feedforward} networks\index{feedforward network}.
\textbf{Right:} the detailed structure of each \terminate{neuron} that \emph{(i)} adds the bias 
and the weighted signals 
to produce the preactivation, 
\emph{(ii)} generates the activation 
from the preactivation, and \emph{(iii)} multiplies the activation by the next-layer weight.
}
\label{fig:ff-example}
\end{center}
\end{figure}

\index{function approximation}\index{model parameters}\index{FCN|see{fully-connected network}}\index{fully-connected network|seealso{multilayer perceptron}}
The MLP is recursively defined through the following iteration equations
\begin{align}\label{eq:mlp-definition}
&z_i^{(1)}(x_\alpha)\equiv\bias{i}{1}+\sum_{j=1}^{n_{0}}\W{ij}{1}\x{j}{\alpha}\,, \quad \text{for} \quad i=1,\ldots,n_1\, ,\, \\
&z_i^{(\ell+1)}(x_\alpha)\equiv\bias{i}{\ell+1}+\sum_{j=1}^{n_{\ell}}\W{ij}{\ell+1}\sigma\!\le(z_j^{(\ell)}(x_\alpha)\ri)\, , \quad \text{for} \quad i=1,\ldots,n_{\ell+1}\, ;~ \ell=1,\ldots,L-1\, , \notag
\end{align}
which describes a network with $L$ layers of neurons, with each layer $\ell$ composed of $n_\ell$ neurons.\footnote{
 A more modern name for the MLP is the \neo{fully-connected network} (FCN),
highlighting the fact that each neuron in a given layer $\ell$ has a connection to every neuron in layer $\ell+1$, as Figure~\ref{fig:ff-example} makes clear. Such a dense pattern of connections is computationally expensive in terms of the number of parameters required for the architecture and should be contrasted with the sparser architectures described at the end of this section.
To place an emphasis on the \emph{deepness} of networks rather than on the \emph{density} of the connections, we'll mainly stick with the name \emph{multilayer} perceptron over the name \emph{fully-connected} network in this book.\index{multilayer perceptron!a.k.a.~a fully-connected network}
}
We depict an example MLP architecture in Figure~\ref{fig:ff-example}. The number of layers $L$ defines the \term{depth} of the network and the different number of neurons in each layer $n_{\ell=1,\ldots,L-1}$ define the \textbf{widths}\index{width|textbf} of the layers.
The depth and hidden-layer widths are variable \term{architecture hyperparameters} that define the shape of the network, while the values of $n_0$ and $n_{L}$ are set by input and output dimensions of the function-approximation task, respectively.
In particular, the final-layer preactivations computed by the network
\be
f\!\le(x; \theta\ri) = z^{(L)}(x),
\ee
serves as the function approximator, with its model parameters $\theta_{\mu}$ being the union of the biases and weights from all the layers. Sometimes it will be convenient to think of this collection of model parameters as an explicit vector $\theta_\mu$ whose components cover all the model parameters.
In that case, the dimension of $\theta_\mu$ and therefore the total number of the model parameters is given
by
\be
P = \sum_{\ell=1}^{L} \le(n_{\ell} + n_{\ell}n_{\ell-1} \ri) \, ,
\ee
which scales quadratically with the widths of the network and linearly with the depth.

The intermediate layers $\ell= 1, \dots, L-1$ are referred to as \textbf{hidden layers}\index{hidden layer|textbf}, since preactivations and activations of the neurons from those layers are not part of the network's output. On the one hand, the variables $z^{(\ell)}(x)$ for $\ell < L$ are simply temporary variables introduced to construct
an increasingly flexible set of functions, expressive enough to have a chance of approximating hard-to-describe functions.
On the other hand, in analogy to the physical \terminate{brain}, these variables are thought to encode 
useful information about \emph{how} the \terminate{neural network} is approximating; for example, a particular neuron might fire if it recognizes a tail, a whisker, or a pattern representing fur -- all potentially useful \emph{features}\index{feature}\index{feature|seealso{representation}}
for determining whether an image contains a cat or not.

\index{architecture}\index{inductive bias!of model architectures}\index{convolutional neural network|textbf}
Moving beyond MLPs, the choice of neural network architecture is often motivated by the nature of the function we are trying to approximate. For instance, the properties of the dataset\index{input data} $\D$, when known and articulated, can be used to build \textbf{inductive biases} into the architecture so that the resulting set of functions may better represent the underlying function.\footnote{We'll discuss the inductive bias of MLPs from various different perspectives in 
\S\ref{ch:bayesian-inference}, \S\ref{ch:features}, and Epilogue~\ref{epi:overparameterization}.} 
Let's look at a few examples.
\bi
\item For \term{computer vision} (CV) applications, \textbf{convolutional neural networks} (CNN) or conv-nets~\cite{fukushima1980neocognitron,lecun89conv,LeCunZipCode1989,lecun1998gradient,ImageNet2012} are used to take advantage of the fact that information in images is organized in a spatially local manner, often respecting \neo{translational invariance}.\footnote{
For a two-dimensional convolutional layer, the iteration equation~\eqref{eq:mlp-definition} for MLPs is replaced by
\be
\label{eq:conv-layer}
z_{i,(c,d)}^{(\ell+1)}(x_\alpha)\equiv\bias{i}{\ell+1}+\sum_{j=1}^{n_{\ell}}\sum_{c'=-k}^{k}\sum_{d'=-k}^{k}\W{ij}{\ell+1}\sigma\!\le(z_{j,(c+c',d+d')}^{(\ell)}(x_\alpha)\ri)\, ,
\ee
where in $z_{i,(c,d)}^{(\ell)}$, the first index $i$ is an auxiliary \neo{channel}\index{channel|seealso{convolutional neural network}} index and the paired index $(c,d)$ is a two-dimensional spatial index, and the number $k$ is a fixed constant for each layer, determining the size of the convolutional window. In particular, the same weights are used on different spatial locations of the input, which promotes the inductive bias that image data are often translationally invariant. In other words,
a cat is still a cat regardless of its location in an image.
At the time of writing, the convolutional layer is an essential part of many modern deep learning architectures, but this situation may change in the future. Please pay \neo{attention}.
}
\item For \term{natural language processing} (NLP) applications, the \term{transformer} architecture (no acronym yet) is used to process sequential input -- such as a paragraph of text or an amino acid sequence coding a protein -- in a way that encourages correlations to develop between any of the elements in the sequence \cite{attention2017}. This property of the model is aptly called \neo{attention}.\index{language model}\index{attention|seealso{transformer}}
\ei
An important property of these inductive biases is that they induce constraints or relationships between the weights. For instance, we can think of the convolutional layer as a particular type of MLP layer, where many weights are set to zero and the values of remaining weights are further shared among several different neurons. This property is known as \neo{weight tying}\index{weight tying|seealso{convolutional neural network}}. That means that convolutional layers are actually within the class of functions describable by using MLP layers, but they are very unlikely to be found via training unless the constraints are explicitly enforced. As long as the inductive bias of spatial locality and translational invariance is well founded, the convolution layer
has obvious computational advantages by heavily curtailing the number of weights to be trained and stored.

Regardless of these specific inductive biases ingrained into modern neural network architectures used in deep learning, the common thread to all is the idea of constructing a flexible set of functions by organizing neural components into many iterated layers. MLPs are the simplest of these neural network architectures that hinges on this stacking idea, and thus provide a \emph{minimal model}\index{minimal model!of deep learning} for an \term{effective theory of deep learning}.
Specifically, we expect that (\textit{a}) the \neo{principles of deep learning theory} that we uncover to be general and valid across the large variety of architectures that are based on the idea of stacking many layers of neural components and (\textit{b}) the resulting effective theory formalism can be specialized to specific architectures of interest as needed, using this book as a guide for how to work out such a theory. In particular, one can study other architectures in our formalism simply by swapping out the MLP iteration equation~\eqref{eq:mlp-definition} -- e.g.~for the convolution layer iteration equation~\eqref{eq:conv-layer} --  in the appropriate place. We'll provide pointers on where to make such substitutions when we begin working out our effective theory in~\S\ref{ch:ngp}. 

Finally, in Appendix~\ref{app:residual} we'll study neural networks with \emph{residual connections}\index{residual connection}, known as \textbf{residual networks}\index{residual network|textbf}. These architectures are specially modified to enable the training of deeper and deeper networks. In the final section of that appendix, we'll also explain how our effective theory approach\index{effective theory} can be extended to \emph{general} residual networks, including the \emph{residual convolutional network}\index{convolutional neural network!with residual connections|see{ResNet}} or \term{ResNet} and the \neo{transformer} architecture.

\index{RNN|see{recurrent neural network}}\index{CNN|see{convolutional neural network}}\index{MLP|see{multilayer perceptron}}\index{ConvNet|see{convolutional neural network}}

\section{Activation Functions}\label{sec:activations}
\begin{figure}\label{fig:activations}
\begin{center}
\includegraphics[width=.9\linewidth]{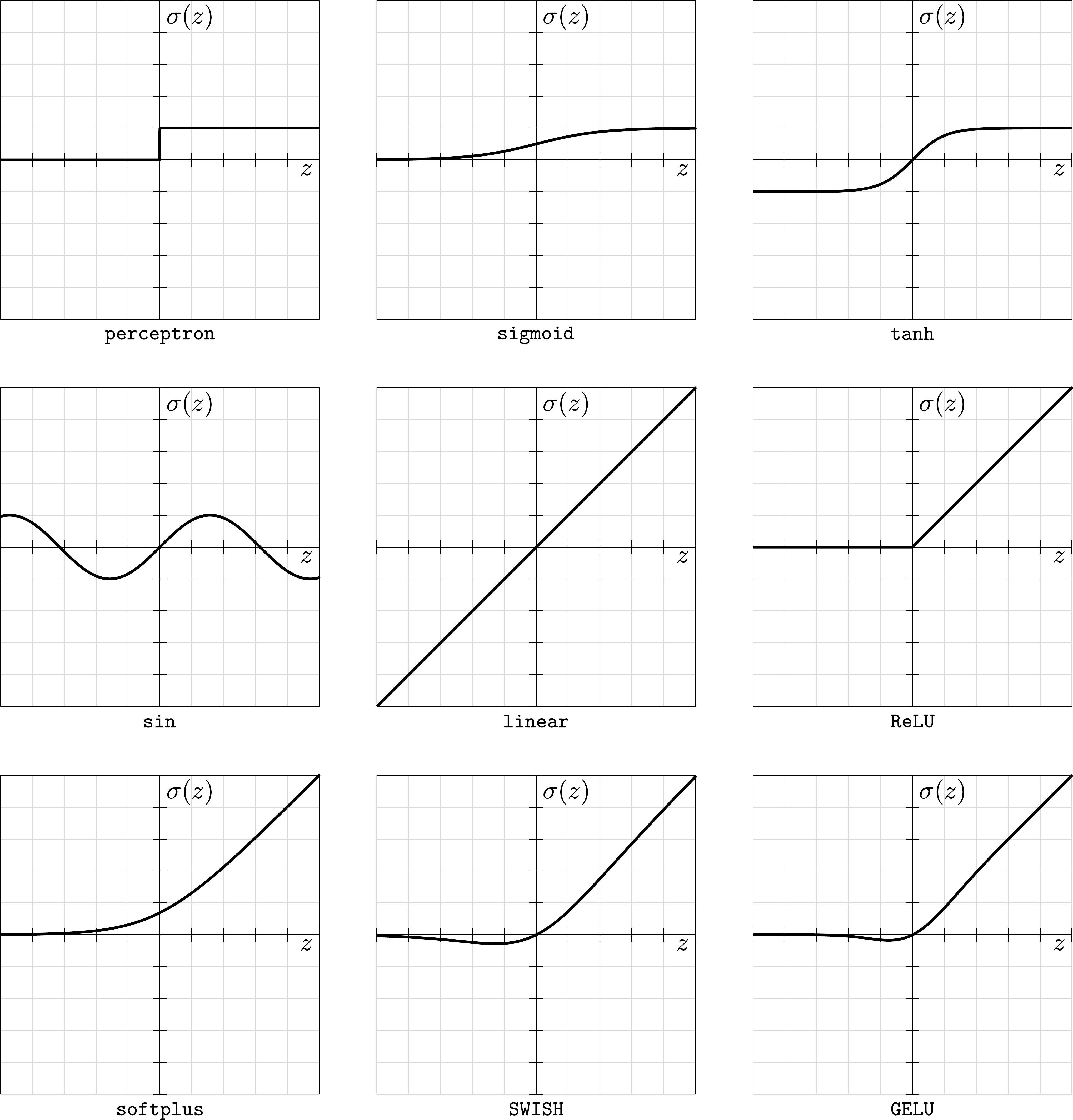}
\caption{Commonly-used activation functions $\sigma(z)$. Grids are in units of one for both the preactivation $z$ and activation $\sigma$. 
(The $\lrelu$ is not shown.)}
\end{center}
\end{figure}
\index{activation function}
In this section, we discuss some of the most common activation functions. This list is non-exhaustive, so hopefully you won't find this section exhausting.
To make it easier for you, we've plotted all these activation functions together in Figure~\ref{fig:activations}. In \S\ref{ch:signalprop}, we'll use our effective theory to evaluate the relative usefulness of these activation functions in allowing input signals to effectively pass through a deep network.

\subsubsection{Perceptron}
The $\perc$ was the original activation function \cite{mcculloch1943logical}.\index{perceptron@$\texttt{perceptron}$|see{activation function}}
It is just a step function
\be
\sigma(z) = 
    \begin{cases}
    1\, , & z \ge 0  \, , \\
    0 \, , & z < 0 \, ,
    \end{cases}
\ee
corresponding to a computer scientist's notion of simplicity: the neuron either \emph{fires} and outputs $1$ or \emph{doesn't fire} and outputs $0$.\footnote{Alternatively, the $\perc$ may be shifted and scaled such that $\sigma(z)=\text{sign}(z)$.
}

Despite the logical simplicity, this turns out to be a poor choice. As we will see, in order to both effectively pass signals through networks (\S\ref{ch:signalprop} and \S\ref{ch:eft-ntk}) and train them~(\S\ref{ch:NTHb}), 
it's helpful to propagate more than one bit of information about the preactivation $z$. The $\perc$ has historical significance, but is never used in deep neural networks.

\subsubsection{Sigmoid}\index{sigmoid@$\texttt{sigmoid}$|see{activation function}}
The $\sigmoid$ activation function is a logistic function\index{logistic function}\index{logistic function|seealso{softmax distribution}}
\be\label{eq:sigmoid}
\sigma(z)=\frac{1}{1+e^{-z}} = \frac{1}{2}+\frac{1}{2}\tanh\!\le( \frac{z}{2}\ri)\, ,
\ee
which is a smoothed version of the $\perc$. Not only is it continuous, but also it preserves information about the magnitude of the preactivation, albeit mostly in the range near $z=0$ where the function is nearly linear. Outside of this range, the $\sigmoid$ heavily compresses such information as it becomes more and more $\perc$-like, \emph{saturating}\index{saturation (of an activation)} as $\sigma(z) = 1$ when $z \to \infty$ and as $\sigma(z) = 0$ when $z \to -\infty$.

As a mapping from the domain of $(-\infty,\infty)$ to the range $[0,1]$, the $\sigmoid$ also has a natural interpretation of converting log-odds to a probability, which is its main application in \terminate{machine learning}.
For \terminate{deep learning}, the differentiability of the $\sigmoid$ was essential in the development of a learning algorithm -- \terminate{backpropagation} -- for training neural networks with hidden layers \cite{rumelhart1985learning}. 
Nevertheless, the $\sigmoid$ activation function is still a poor choice %
in \emph{deep} neural networks:
as we'll see in~\S\ref{ch:signalprop}, a problem arises from the fact that it doesn't pass through the origin.

\subsubsection{Tanh}\index{tanh@$\texttt{tanh}$|see{activation function}}
The hyperbolic tangent or $\tanhA$ activation function
\be
\sigma(z) = \tanh(z) = \frac{e^z - e^{-z}}{e^z + e^{-z}} = \frac{e^{2z} - 1}{e^{2z} + 1} \, ,
\ee
is a scaled (both before and after the activation) and shifted $\sigmoid$, as is clear from~\eqref{eq:sigmoid}. Of particular importance is the fact that it's shifted such that $\sigma(0)=0$ \cite{lecun2012efficient}.

The $\tanhA$ is probably the most popular choice of activation function aside from the $\relu$ or $\relu$-like activation functions to be discussed shortly, and arguably $\tanhA$ is the most popular smooth activation function.
As an exemplary smooth activation function, the $\tanhA$ will be of significant interest for us in this book.

\subsubsection{Sin}\index{sin@$\texttt{sin}$|see{activation function}}
The $\sinA$ activation function is just what it sounds like:
\be
\sigma(z) = \sin(z) \, ,
\ee
i.e.~one of the three standard trigonometric function.
Periodic nonlinearities have been cycling in and out of popularity for a long while now, see e.g.~\cite{periodic-activations}, though they have never really achieved true popularity.

\subsubsection{Scale-invariant: linear, ReLU, and leaky ReLU}\index{scale invariance|textbf}
A scale-invariant activation function is any activation function that satisfies
\be\label{eq:scale-invariant-def-first}
\sigma(\lambda z) = \lambda \sigma(z) \, , %
\ee
for any positive rescaling $\lambda$. We call these activation functions scale-invariant because any scaling of the preactivation $z\to \lambda z$ can be undone by an inverse scaling of the activation $\sigma(z) \to \lambda^{-1} \sigma(z)$. 
This condition is met by -- and only by\footnote{In order to prove this necessity statement, first take the derivative of the scale-invariance equation~\eqref{eq:scale-invariant-def-first} with respect to $z$, which gives $\sigma'(\lambda z) = \sigma'(z)$ for any $\lambda>0$. Then note that this enforces a constant derivative, $a_{+}$, for $z>0$ and another constant derivative, $a_{-}$, for $z<0$. Finally, to satisfy~\eqref{eq:scale-invariant-def-first} we also must have  $\lim_{z\rightarrow \pm 0} \sigma(z)=0$. %
Quantum Electrodynamics.\index{QED}\index{quantum electrodynamics|see{QED}}} -- activation functions of the form
\be\label{eq:scale-invariant-def}
\sigma(z) = 
    \begin{cases}
   a_+ z \, , & z \ge 0  \, , \\
    a_- z \, , & z < 0 \, .
    \end{cases}
\ee

\index{relu@$\texttt{ReLU}$|see{activation function}}
The class of scale-invariant activation functions includes $\linear$\index{linear@$\texttt{linear}$|see{activation function}} ($a_+=a_-=a$),
 Rectified Linear Unit or $\relu$ ($a_+=1$, $a_-=0$) 
 \cite{nair2010rectified,glorot2011deep},
and $\lrelu$\index{leaky relu@$\texttt{leaky ReLU}$|see{activation function}} ($a_+=1$, $a_-=a$) \cite{maas2013rectifier} activation functions.
The $\relu$ is the most popular of the activation functions used in deep neural networks and therefore will be of substantial interest for us in this book.

\index{saturation (of an activation)}
In order to deepen our understanding of scale invariance, let's consider how other activation functions can break it. For instance, consider the $\tanhA$ activation function $\sigma(z) = \tanh(z)$. Mathematically, $\tanhA$ violates scale invariance because $\tanh(\lambda z) \ne\lambda \tanh \sigma(z)$ unless $\lambda=1$. In particular, while the activation function is approximately linear for small preactivations, i.e.~$\tanh(z)\approx z$ for  $\vert z\vert\ll 1$, it saturates for large preactivations, i.e.~$\vert\tanh(z)\vert\approx 1$ for $|z| \gg 1$. Thus, $\tanhA$ comes with an intrinsic crossover scale $\vert z\vert\sim1$ that separates the two regimes.
We can see this visually in Figure~\ref{fig:activations}: if we zoom out, all the non-scale-invariant activation functions -- e.g. $\perc$, $\sigmoid$, and $\tanhA$ -- will look squashed, while the scale-invariant activation functions -- e.g. $\relu$ and $\linear$ -- will look the same at any scale.

\index{kink|seealso{$\texttt{ReLU}$}}\index{kink}\index{kink|seealso{$\texttt{leaky ReLU}$}}
Finally, note that all the scale-invariant activation functions -- except the aptly-named $\linear$ activation -- create a nonlinear relationship between the network inputs and outputs due to the kink at the origin $z=0$. Stacking up many layers of neurons with these nonlinear activation functions accumulates the nonlinearity, allowing such deep neural networks to express highly nonlinear functions.

\subsubsection{ReLU-like: softplus, SWISH, and GELU}
Despite the popularity of the $\relu$, there's an uneasiness about the fact that it's not smooth. In an attempt to rectify the situation, a variety of smoothed-out $\relu$-like activations have been proposed and achieved semi-popularity, of which we will consider the following three:
\bi
\item  The $\softplus$ activation function \cite{dugas2000incorporating}\index{softplus@$\texttt{softplus}$|see{activation function}}
\be\label{eq:softplus}
\sigma(z)=\log\!\le(1+e^{z}\ri)\, ,
\ee
behaves linearly $\sigma(z) \approx z$  for a large argument $z\gg 1$ and vanishes exponentially for a negative argument, $\sigma(z) \approx e^{-\vert z\vert}$ for $z < 0$. Importantly
the $\softplus$ does not pass through the origin: $\sigma(0) = \log(2)$.

\item The $\swish$ activation function \cite{ramachandran2017searching} is defined as  \index{swish@$\texttt{SWISH}$|see{activation function}}
\be
\sigma(z)=\frac{z}{1+e^{-z}}\, ,
\ee
which is a logistic function \eqref{eq:sigmoid} multiplied by the preactivation $z$. The \terminate{logistic function} behaves as a continuous on/off switch, and so the $\swish$ approximates the $\relu$, which we recall was defined as a discrete on/off switch multiplied by the preactivation $z$. In particular, for $z>0$ the $\swish$ behaves as $\sigma(z) \approx z$, but for $z<0$ it behaves as $\sigma(z) \approx 0$. Also, the multiplication by $z$ ensures that the $\swish$ passes through the origin, $\sigma(0) = 0$.
\item The Gaussian Error Linear Unit ($\gelu$) \index{gelu@$\texttt{GELU}$|see{activation function}}
activation function \cite{hendrycks2016gaussian} is a lot like  the $\swish$. It's given by the expression
\be
\sigma(z)=\le[\frac{1}{2}+\frac{1}{2}\text{erf}\!\le(\frac{z}{\sqrt{2}}\ri)\ri]\times z\, ,
\ee
where the error function\index{error function} $\text{erf}(z)$ is given by
\be
\text{erf}(z) \equiv \frac{2}{\sqrt{\pi}} \int_0^z dt\, e^{-t^2}\, ,%
\ee
which is a partial integration of the Gaussian function.
In particular, the graph of $\text{erf}(z)$ looks very similar to graph of $\tanh(z)$, and so the graph of the scaled and shifted version used in the definition of the $\gelu$, $\frac{1}{2}+\frac{1}{2}\text{erf}\le(\frac{z}{\sqrt{2}}\ri)$, looks very similar to the graph of the logistic function\index{logistic function} \eqref{eq:sigmoid}. 
Like the $\swish$, it crosses the origin and behaves more like the $\relu$ the further we go away from $0$ in either direction.
\ei
In smoothing the $\relu$, all three of these activation functions introduce an intrinsic scale and violate the scale-invariance\index{scale invariance} condition~\eqref{eq:scale-invariant-def-first}.

\section{Ensembles}\label{sec:MLP_distribution}

\index{training}\index{programming}\index{function approximation}
As we discussed in \S\ref{sec:MLP_intro}, neural networks are \emph{trained} rather than \emph{programmed}.
Practically speaking, to begin
training a neural network for \terminate{function approximation}, we need to set initial values of the biases $\bias{i}{\ell}$ and weights $\W{ij}{\ell}$. 
Since the learned values of these model parameters are almost always iteratively built up from their initial values, the initialization strategy can
have a major impact on the success or failure of the function approximation.

\index{permutation symmetry}\index{zero initialization}\index{zero initialization|seealso{initialization distribution}}
Perhaps the simplest strategy would be to set all the biases and weights to zero, $\bias{i}{\ell}=\W{ij}{\ell}=0$.
However, this initialization fails to break the permutation symmetry among the $n_\ell$ different neurons in a hidden layer $\ell$. If this symmetry isn't broken,
then
we
cannot distinguish between the different neurons in a layer as all these neurons perform exactly the same computation. In effect,
the network would behave as if it only had single neuron $n_{\ell}=1$ in each hidden layer.
Thus, in order to leverage all the different components of the biases and weights in a wider network, we need to somehow break the permutation symmetry.

\index{permutation symmetry}\index{function approximation}
Perhaps the simplest strategy that breaks this permutation symmetry is to 
sample each bias and weight independently from some \neo{probability distribution}.
Theoretically speaking, we should pick this \term{initialization distribution} so that the resulting \term{ensemble}\index{ensemble|seealso{probability distribution}} of networks are well behaved with respect to the function-approximation task.
This section initializes ourselves for analyzing such an ensemble.

\subsubsection{Initialization distribution of biases and weights}

Among the many potential reasonable choices for the initialization distribution, the obvious choice is the \terminate{Gaussian distribution}.\footnote{ 
Two other choices seen in the wild for the initialization distribution are the \terminate{uniform distribution} and the truncated normal distribution\index{truncated normal distribution}.
For the weights, the difference between the Gaussian distribution and any other distribution -- when the means are set zero and the variances are set equal -- turns out to be suppressed by $1/\text{width}$ for wide networks. 
That is, due to the \terminate{central limit theorem}, ultimately only the first and second moment -- i.e.~the mean and variance -- for the weight initialization distribution is of any real consequence. Thus, we might as well just use a Gaussian distribution.
For the biases, the difference between the Gaussian distribution and any other distribution is mostly moot in practice because we shall find that the bias variance $\Cb{\ell}$ should be set to zero for all good activation functions.
}
As we discussed, Gaussian distributions are defined solely in terms of their mean and variance, so they're easy to specify and work with in theory. They're also extremely easy to sample from, which is also an essential consideration when picking a sampling distribution in practice.

In particular, to initialize MLPs, we'll
independently sample each bias and each weight from zero-mean Gaussian distributions with variances given by
\begin{align}
\label{eq:bias-variance-def-naive}
\mathbb{E}\le[b^{(\ell)}_{i_1}b^{(\ell)}_{i_2}\ri]&=\delta_{i_1 i_2} C_{b}^{(\ell)}\, ,\\ %
\label{eq:weight-variance-def-naive}
\mathbb{E}\le[W^{(\ell)}_{i_1 j_1}W^{(\ell)}_{i_2 j_2}\ri]&=\delta_{i_1 i_2} \delta_{j_1 j_2}\frac{C_{W}^{(\ell)}}{n_{\ell-1}}\, ,%
\end{align}
respectively. Here the \terminate{Kronecker delta}s indicate that each bias and each weight are all drawn independently from the others.
Explicitly, the functional forms of these Gaussian distributions \index{Gaussian distribution} are given by
\begin{align}
\label{eq:full-bias-initialization}
p\!\le(b_i^{(\ell)}\ri) &= \frac{1 }{ \sqrt{2\pi \Cb{\ell}} }\exp\!\le[-\frac{1}{2\Cb{\ell} }\le(\bias{i}{\ell}\ri)^2\ri] \, , \\
\label{eq:full-weights-initialization}
p\!\le(W_{ij}^{(\ell)}\ri) &=\sqrt{\frac{n_{\ell-1} }{ 2\pi \CW{\ell}}} \exp\!\le[-\frac{n_{\ell-1}}{2\CW{\ell} } \le(\W{ij}{\ell}\ri)^2\ri] \, .
\end{align}
Here the normalization of weight variances by $1/n_{\ell-1}$ is purely conventional but, as we will show explicitly in~\S\ref{ch:deep-linear-eft} and~\S\ref{ch:ngp}, it is necessary for wide neural networks and natural for comparing the behavior of networks with different widths.\footnote{We can trace this convention to the MLP iteration equation \eqref{eq:mlp-definition}. To compute the preactivation $\z{i}{\alpha}{\ell}=\bias{i}{\ell}+\sum_{j=1}^{n_{\ell-1}}\W{ij}{\ell}\s{j}{\alpha}{\ell-1}$, we essentially add together $n_{\ell-1}$ random weights. For large $n_{\ell-1}$, the normalization factor of $1/n_{\ell-1}$ in the variance -- which is tantamount to normalizing each weight by $1/\sqrt{n_{\ell-1}}$ -- essentially counteracts this summation of many zero-mean random numbers.
Since there is no such summation for the biases, there is no need for such a normalization factor.
}
Also note that we allow the bias variance $\Cb{\ell}$ and rescaled weight variance $\CW{\ell}$ to potentially vary from layer to layer.
Together, the set of bias variances $\le\{\Cb{1}, \ldots, \Cb{L}\ri\}$ and the set of rescaled weight variances $\le\{\CW{1}, \ldots, \CW{L}\ri\}$ are called \textbf{initialization hyperparameters}\index{initialization hyperparameters|textbf}. 
One practical result of our effective theory approach will be prescriptions for setting these initialization hyperparameters so that the output of the neural network is well behaved.\index{hyperparameters!initialization|see{initialization hyperparameters}}\index{hyperparameters!architecture|see{architecture hyperparameters}}\index{hyperparameters!residual|see{residual hyperparameters}}\index{hyperparameters!training|see{training hyperparameters}}

\subsubsection{Induced distributions}
\index{induced distribution}%
Given a \index{input data}dataset $\D=\le\{\x{i}{\alpha}\ri\}$ consisting of $\ND$ input data, an MLP with model parameters $\theta_{\mu}=\le\{\bias{i}{\ell},\W{ij}{\ell}\ri\}$ 
evaluated on $\D$ 
outputs an array of $n_L\times \ND $ numbers
\be
f_i\!\le(x_{\alpha}; \theta\ri) = z_i^{(L)}(x_{\alpha})\equiv \z{i}{\alpha}{L} \, ,
\ee 
indexed by both \textbf{neural indices}\index{neural indices|textbf} $i=1,\ldots,n_L$ and \textbf{sample indices}\index{sample indices|textbf} $\alpha=1,\ldots,\ND$. 
Each time we instantiate MLPs by drawing \terminate{model parameters} $\theta_{\mu}$ from the initialization distribution $p(\theta)$, we get a different initial set of outputs $\z{i}{\alpha}{L}$. 
It follows that since the biases $\bias{i}{\ell}$ and weights $\W{ij}{\ell}$ are random variables at initialization, then so must be the network outputs $\z{i}{\alpha}{L}$. 
In this way, the initialization distribution \emph{induces} a distribution on the network outputs.

This \term{output distribution} $p\!\le(z^{(L)}\Big\vert \D\ri)$
controls the statistics of network outputs at the point of initialization. In practice, 
the properties of this distribution are directly related to how hard it is
for a network to approximate its target function through iterated adjustments of its model parameters. As such, having control over this distribution is of significant interest from a practitioner's perspective.
From a theorist's perspective, even though the initialization distribution for model parameters is simple by design, the induced output distribution is not.
In theory, we need to calculate the following gigantic integral over all the model parameters
\begin{align}\label{eq:gigantic-beast-that-we-tame}
p\!\le(z^{(L)} \Big| \D \ri) = \int \le[  \prod_{\mu=1}^{P} d \theta_{\mu}\ri]    p\!\le(z^{(L)} \Big|\theta, \D \ri) p(\theta)\, .
\end{align}

Before performing this heroic integration,
notice that the conditional distribution $p\!\le(z^{(L)} \Big|\theta, \D \ri)$ in the integrand~\eqref{eq:gigantic-beast-that-we-tame} is actually \emph{deterministic}. In other words, if we know the set of inputs $\D$ and the settings of all the model parameters $\theta_{\mu}$, then we know how to compute the network outputs: we just use the iteration equation~\eqref{eq:mlp-definition} that defines the MLP.
What we don't yet know
is how to express this determinism as a distribution.

\subsubsection{Deterministic distributions and the Dirac delta function}
What kind of a probability distribution is \emph{deterministic}?
Let's abstractly denote such a distribution as $p(z|s) = \delta(z|s)$, which intend to encode the deterministic relationship $z=s$.
What properties should this distribution have?
First, the mean of $z$ should be $s$
\be\label{eq:dirac-mean}
\E{z} =\int dz\ \delta(z| s)\, z \equiv s\, .
\ee
Second, the variance should vanish, since this is a deterministic relationship. In other words,
\be\label{eq:dirac-variance-zero}
\mathbb{E}[z^2]-\le(\E{z}\ri)^2 = \le[\int dz\ \delta(z| s)\, z^2\ri] -s^2 \equiv 0\, ,
\ee
or, equivalently,
\be\label{eq:dirac-variance}
\int dz\ \delta(z| s)\, z^2\equiv s^2.
\ee
In fact, this determinism implies an even stronger condition. %
In particular, 
the expectation of any function $f(z)$ of $z$, should evaluate to $f(s)$:
\be\label{eq:dirac-defining-property}
\E{f(z)} = \int dz\ \delta(z| s)\, f(z) \equiv f(s)\, ,
\ee
which includes the properties \eqref{eq:dirac-mean} and~\eqref{eq:dirac-variance} as special cases when $f(z) = z$ and $f(z) = z^2$, respectively, as well as the probability normalization condition
\be \label{eq:dirac-delta-normalization}
\int dz\ \delta(z| s) = 1\, ,
\ee
when $f(z) = 1$.\footnote{A random variable that obeys $\E{f(z)} = f\!\le(\E{z}\ri)$ is said to \emph{self-average}\index{self-averaging}\index{self-averaging|seealso{Dirac delta function}},  meaning that we can exchange the order of the expectation with the function evaluation. The condition~\eqref{eq:dirac-defining-property} is equivalent to saying that the distribution $\delta(z|s)$ is self-averaging.\label{footnote:self-average}} 
In fact, \eqref{eq:dirac-defining-property} is the defining property of the \term{Dirac delta function}.\footnote{The Dirac delta function is really a generalization of the \terminate{Kronecker delta}~\eqref{eq:Kronecker-delta} for continuous variables.
In this footnote we also include the obligatory disclaimer that -- despite its name -- the Dirac delta function is a \emph{distribution} and not a \emph{function}, as should have been clear from our discussion. Despite this, we will stick with common convention and continue to refer to it as the \emph{Dirac delta function}.} 
 
 \index{Gaussian distribution!relationship to Dirac delta function}
As a representation though, \eqref{eq:dirac-defining-property} is
a little too abstract, even for us. However, our discussion above paves the way for a much more concrete representation. 
Since the Dirac delta function is a normalized distribution~\eqref{eq:dirac-delta-normalization} with mean $s$~\eqref{eq:dirac-mean} and zero variance~\eqref{eq:dirac-variance-zero}, 
let's
consider a normalized Gaussian distribution with mean $s$~\eqref{eq:Gaussian-with-mean}
and take the limit as the variance $K$ goes to zero:
\be\label{eq:gaussian-limit-delta-function}
\delta(z|s)\equiv\lim_{K\rightarrow +0} \frac{1}{\sqrt{2\pi K}}e^{-\frac{(z-s)^2}{2K}}\, .
\ee
This distribution is infinitely peaked at $z=s$ while vanishing everywhere else, so any function $f(z)$ integrated against \eqref{eq:gaussian-limit-delta-function} will 
give $f(s)$ after taking the limit.
In other words, it satisfies the defining property of the Dirac delta function~\eqref{eq:dirac-defining-property}.

\index{Dirac delta function!integral representation}
The limit in \eqref{eq:gaussian-limit-delta-function} should always be taken after integrating the distribution against some function. Having said that, perhaps this representation still makes you a little bit uncomfortable as
it is still a very singular limit.
Let's try to fix this and find a yet even better representation.
Here's a \terminate{magic trick}: starting from \eqref{eq:gaussian-limit-delta-function}, let's insert ``1'' on the right hand side as
\begin{align}
\delta(z|s)=&\lim_{K\rightarrow +0} \frac{1}{\sqrt{2\pi K}}e^{-\frac{(z-s)^2}{2K}} \le\{\frac{1}{\sqrt{2\pi/K}} \int_{-\infty}^\infty d\Lambda\ \exp\!\le[ -\frac{K}{2}\le(\Lambda-\frac{i(z-s)}{K}\ri)^2  \ri]\ri\} \,  \\
=& \lim_{K\rightarrow +0} \frac{1}{2\pi}  \int_{-\infty}^\infty d\Lambda\ \exp\!\le[-\frac{1}{2}K \Lambda^2 + i \Lambda(z-s) \ri] \, , \notag
\end{align}
where in the curly brackets we inserted an integral over a dummy variable $\Lambda$ of a normalized Gaussian with variance $1/K$ and imaginary mean $i(z-s)/K$, and on the second line we simply combined the exponentials.  Now we can easily take the limit $K\to+0$ to find an \emph{integral representation} of the Dirac delta function
\be\label{eq:integral-form-delta-function}
\delta(z|s)= \frac{1}{2\pi}\int_{-\infty}^{\infty} d\Lambda\ e^{i \Lambda(z-s)}\equiv \delta(z-s) \, .
\ee
In this final expression, we noted that the function depends only on the difference $z-s$.
This \terminate{integral representation} will come in handy in~\S\ref{ch:ngp}.

\subsubsection{Induced distributions, redux}\index{induced distribution}\index{output distribution}

Now that we are familiar with the \terminate{Dirac delta function}, we can use it to express the output distribution~\eqref{eq:gigantic-beast-that-we-tame} more concretely.
To start, for a one-layer network of depth $L=1$, the distribution of the first layer output~\eqref{eq:mlp-definition} is given by
\begin{align}\label{eq:first-layer-formal-expression-first-encounter}
p\!\le(z^{(1)} \Big| \D \ri)=&\int \le[  \prod_{i=1}^{n_1} d b^{(1)}_{i}\ p\!\le(b_{i}^{(1)}\ri) \ri]  \le[\prod_{i=1}^{n_1}\prod_{j=1}^{n_0} d W^{(1)}_{ij}\ p\!\le(W^{(1)}_{ij}\ri) \ri]\, \\
&\quad\times\le[ \prod_{i=1}^{n_{1}}\prod_{\alpha\in\D}\delta\!\le(\z{i}{\alpha}{1}-b^{(1)}_i-\sum_{j=1}^{n_0}W^{(1)}_{ij}\x{j}{\alpha}\ri)\ri]\, .\nonumber
\end{align}
Here, we needed $n_1 \times \ND$ Dirac delta functions, one for each component of $\z{i}{\alpha}{1}$. In~\S\ref{sec:first-layer-gaussian} we will explicitly evaluate the above integrals, though you should feel free to do so now on your own, if you're impatient.
In passing, let us also introduce a cousin of \eqref{eq:first-layer-formal-expression-first-encounter} 
\begin{align}\label{eq:deeper-layer-formal-expression-first-encounter}
p\!\le(z^{(\ell+1)} \Big| z^{(\ell)} \ri)=&\int \le[  \prod_{i=1}^{n_{\ell+1}} d b^{(\ell+1)}_{i}\ p\!\le(b_{i}^{(\ell+1)}\ri) \ri]  \le[\prod_{i=1}^{n_{\ell+1}}\prod_{j=1}^{n_{\ell}} d W^{(\ell+1)}_{ij}\ p\!\le(W^{(\ell+1)}_{ij}\ri) \ri]\, \\
&\quad\times\le[ \prod_{i=1}^{n_{\ell+1}}\prod_{\alpha\in\D}\delta\!\le(\z{i}{\alpha}{\ell+1}-b^{(\ell+1)}_i-\sum_{j=1}^{n_\ell}W^{(\ell+1)}_{ij}\sigma\!\le(\z{j}{\alpha}{\ell}\ri)\ri)\ri]\, ,\nonumber
\end{align}
which determines the distribution of the preactivations in the $(\ell+1)$-th layer, conditioned on the preactivations in the $\ell$-th layer, after integrating out the \terminate{model parameters}.%

More generally, for any parameterized model with output $z^{\text{out}}_{i;\alpha}\equiv f_i(x_{\alpha};\theta)$ for $i=1,\ldots,n_{\text{out}}$ and with the model parameters $\theta_{\mu}$ distributed according to $p(\theta)$, the \terminate{output distribution}~\eqref{eq:gigantic-beast-that-we-tame} can be written using the \terminate{Dirac delta function} as%

\be\label{eq:gigantic-beast-that-we-tame-with-Dirac}
p\!\le(z^{\text{out}} \Big| \D \ri)=\int \le[  \prod_{\mu=1}^{P} d \theta_{\mu}\ri]  p(\theta) \le[ \prod_{i=1}^{n_{\text{out}}}\prod_{\alpha\in\D}\delta\Big(z^{\text{out}}_{i;\alpha}-f_i(x_{\alpha};\theta)\Big)\ri] \, .
\ee
Our pretraining was designed precisely to prepare ourselves for performing this integral for MLPs.

%% file: Chp3-DLN/3_global.tex
\chapter{Effective Theory of Deep Linear Networks at Initialization}\label{ch:deep-linear-eft}
\epigraph{\dots a system which has spherical symmetry \dots certainly cannot result in an organism such as a horse, which is not spherically symmetrical. %
}{Alan Turing, on %
the limitations of toy models \cite{turing1952chemical}.\index{Turing, Alan}}

\index{deep linear network|textbf}\index{function approximation}\index{artificial intelligence}\index{brain}\index{deep linear network|seealso{$\texttt{linear}$}}
\noindent{}In this final warm-up chapter, we introduce and then solve a toy model of deep learning, the \textbf{deep linear network}.\footnote{For physicists, we give an analogy: the deep linear network is to deep learning as the \terminate{simple harmonic oscillator} is to \terminate{quantum mechanics}.} As will be explained in~\S\ref{sec:DLN}, the deep linear network is simply an MLP with $\linear$ activation functions. 
In particular, such a network can only 
compute
linear transformations of its inputs 
and certainly cannot result in a function such as a human, which is empirically known to be nonlinear.
Nonetheless, the study of deep linear networks\index{deep linear network} will serve as a useful blueprint for an \neo{effective theory of deep learning} that we will develop more generally over the subsequent chapters. %
Specifically, the exercises in this chapter illustrate how layer-to-layer recursions control the statistics of deep neural networks in a very intuitive way,
without getting bogged down by all the technical details.

To that end, in~\S\ref{sec:criticality_DLN} we obtain and then exactly solve a layer-to-layer recursion for the two-point correlator of preactivations in deep linear networks.
The result highlights that the statistics of the network sensitively depend on the setting of the \neo{initialization hyperparameters}, with the sensitivity increasing exponentially with depth.
This leads to the important concept of \neo{criticality}, which we will explore in \S\ref{ch:signalprop} in greater depth and sensitivity. In short, we learn that for networks to be well behaved, these hyperparameters need to be finely tuned. 

\index{connected correlator!four-point}\index{architecture hyperparameters}
Next, in~\S\ref{sec:fluctuations_DLN} we obtain and then 
solve a layer-to-layer recursion for the four-point correlator, 
albeit for a single input to further simplify the algebra. 
This showcases the way in which the behavior of the network can depend on the \emph{architecture hyperparameters}, particularly the width and depth of the network. 
In addition, we interpret the four-point \emph{connected} correlator as a measurement of the \emph{fluctuation}\index{fluctuations} of the network function from
draw to
draw of the model parameters. 
Such fluctuations can interfere with the tuning of the initialization hyperparameters
and need to be controlled so that networks
behave reliably for typical draws. 
The scale of the fluctuations are set by the depth-to-width ratio of the network, highlighting this important \neo{emergent scale} in the analysis of MLPs, and we'll see that 
the fluctuations can be kept under control by
keeping the depth-to-width ratio of the network sufficiently small.

\index{emergent scale}
Finally, in~\S\ref{sec:solution_DLN} we obtain a recursion for 
an arbitrary $M$-point
correlator for a deep linear network evaluated on a single input. Such recursions are all exactly solvable at any width 
$n$ and depth $L$, meaning we can fully determine the statistics of these networks at initialization.\footnote{This notion of \emph{solve} should not be confused with the solving of the training dynamics for a particular \terminate{learning algorithm}. In the context of deep linear networks\index{deep linear network}, the dynamics of \terminate{gradient descent} were analyzed in~\cite{saxe2013exact}.  In \S\ref{ch:NTHb} and \S\ref{ch:eot}, we will solve the training dynamics of gradient descent for MLPs with general activation functions in the context of our effective theory formalism.
}
Given these nonperturbative solutions, we
take the limit of large width, with fixed depth, and the limit of large depth, with fixed width, and show explicitly that these two limits do not commute. 
We also construct an interpolating solution with both large width and large depth, but fixed depth-to-width ratio $L/n$, and see how this 
\emph{scale}
serves as a perturbative parameter that controls 
all the 
interactions 
in the network
and 
controls
the validity of the perturbative analysis.

\section{Deep Linear Networks}
\label{sec:DLN}
\index{deep linear network}\index{linear transformations}\index{Gaussian distribution}
A deep linear network iteratively transforms an input $\x{i}{\alpha}$ through a sequence of simple linear transformations
\be\label{eq:deep-linear-foward-pass}
\z{i}{\alpha}{\ell+1} = \bias{i}{\ell+1} + \sum_{j=1}^{n_\ell}\W{ij}{\ell+1} \z{j}{\alpha}{\ell} \, ,
\ee
with $\z{i}{\alpha}{0}\equiv\x{i}{\alpha}$ and $\z{i}{\alpha}{\ell}\equiv z_i^{(\ell)}(x_{\alpha})$.
Since the $\linear$ activation function is the identity function, $\sigma(z) = z$, there's no distinction here between preactivations and activations.

\index{linear transformations}
In this chapter, we'll simplify matters a bit by turning off all the biases, $\bias{i}{\ell}=0$, so that the preactivations in layer $\ell$ are simply given by a repeated matrix multiplication of weight matrices as
\begin{align}\label{eq:deep-linear-preactivation-compact}
\z{i}{\alpha}{\ell}=\sum_{j_0=1}^{n_0} \sum_{j_1=1}^{n_1}\cdots \sum_{j_{\ell-1}=1}^{n_{\ell-1}} W_{i j_{\ell-1}}^{(\ell)} W_{j_{\ell-1}j_{\ell-2}}^{(\ell-1)} \cdots W_{j_{1} j_{0}}^{(1)} \x{j_{0}}{\alpha}\equiv \sum_{j=1}^{n_0}\PW_{ij}^{(\ell)} \x{j}{\alpha}\, .
\end{align}
Here we have introduced an $n_{\ell}$-by-$n_{0}$ matrix  
\be
\PW_{ij}^{(\ell)}= \sum_{j_1=1}^{n_1}\cdots \sum_{j_{\ell-1}=1}^{n_{\ell-1}}  W_{i j_{\ell-1}}^{(\ell)} W_{j_{\ell-1}j_{\ell-2}}^{(\ell-1)} \cdots W_{j_{1} j}^{(1)} \, ,
\ee
which highlights the fact that the preactivation at the $\ell$-th layer is simply a linear transformation of the input.
Additionally, let us set $\CW{\ell} \equiv C_W$ so that the order-one part of the weight variance is layer independent. All together, this means that the \terminate{initialization distribution} over the weights is characterized by the following expectations
\be\label{eq:deep-linear-weight-init}
\E{W^{(\ell)}_{i j}} = 0\,, \qquad \E{W^{(\ell)}_{i_1 j_1}W^{(\ell)}_{i_2 j_2}}=\delta_{i_1 i_2} \delta_{j_1 j_2}\frac{C_{W}}{n_{\ell-1}}\, .
\ee

\index{linear transformations}\index{deep linear network}
Somewhat counterintuitively, deep linear networks generically represent a smaller set of functions than fully general linear transformations, a.k.a.~one-layer networks of the same input-output dimensions.\footnote{This is not necessarily a bad thing, since there are often both computational and representational advantages to focusing on a specialized class of functions. For instance, we saw that convolutional networks represent a much smaller set of functions than MLPs, and yet they are known to perform better on \terminate{computer vision} tasks due to their translational-invariance-respecting \emph{inductive bias} %
as well as the fact that they require significantly less computation due to their sparse pattern of connections.\index{convolutional neural network}\index{translational invariance}
Having said that, it's not obvious if deep linear networks have a useful inductive bias when compared to general linear transformations.
} As an extreme example, let's take a two-layer deep linear network in which the first hidden layer consists of a single neuron $n_1=1$ and consider the network output in the second layer $\ell=2$. 
In this case, all the information in the input is compressed through a \neo{bottleneck} into a single number in the first layer before being converted into an $n_2$-dimensional vector in the output layer. Surely, 
such a deep linear network represents a
tinier subspace of linear transformations than those given by all the possible $n_2$-by-$n_0$ matrices, so long as $n_0, n_2 > 1$.

\index{deep linear network}
More importantly, we will show that the statistics of deep linear networks at initialization are also very different from those of one-layer networks.
In particular, while the statistics of each $W_{i j}^{(\ell)}$ are given by a simple \terminate{Gaussian distribution}, the statistics of their product $\PW_{ij}^{(\ell)}$ are non-Gaussian, depending in a complicated way on the depth $\ell$ and widths $n_{1},\ldots,n_{\ell}$ of the network.

\index{correlator!$M$-point}
The goal of the rest of this chapter is to exactly work out this dependence. Concretely, we are going to compute the nontrivial distribution
\be
p\!\le(z^{(\ell)} \Big\vert \D\ri)\,  \equiv p\!\le(z^{(\ell)}\le(x_1\ri), \ldots, z^{(\ell)}\le(x_{\ND}\ri) \ri)\, ,%
\ee
of the preactivations $\z{i}{\alpha}{\ell}\equiv z_{i}^{(\ell)}\!\le(x_\alpha \ri)$
implied by the iterated multiplication \eqref{eq:deep-linear-preactivation-compact}
when evaluated on the entire dataset\index{input data} $\D$.
As mentioned in \S\ref{sec:not-Gauss}, a distribution is completely determined by the set of all its $M$-point correlators, and so our method for determining $p\!\le(z^{(\ell)} \Big\vert \D\ri)$ will be to directly compute these correlators.

Before moving onto the next section, let's consider the simplest observable, the mean of the preactivation $\z{i}{\alpha}{\ell}$.
Taking an expectation of the defining equation~\eqref{eq:deep-linear-preactivation-compact},
it's easy to see that the mean preactivation must vanish at any layer:
\begin{align}
\E{\z{i}{\alpha}{\ell}}&=\sum_{j_0=1}^{n_0} \sum_{j_1=1}^{n_1}\cdots \sum_{j_{\ell-1}=1}^{n_{\ell-1}} \E{W_{i j_{\ell-1}}^{(\ell)} W_{j_{\ell-1}j_{\ell-2}}^{(\ell-1)} \cdots W_{j_{1} j_{0}}^{(1)} \x{j_{0}}{\alpha}}  \, \\
&=\sum_{j_0=1}^{n_0} \sum_{j_1=1}^{n_1}\cdots \sum_{j_{\ell-1}=1}^{n_{\ell-1}}  \E{W_{i j_{\ell-1}}^{(\ell)}} \E{ W_{j_{\ell-1}j_{\ell-2}}^{(\ell-1)}} \cdots \E{W_{j_{1} j_{0}}^{(1)}} \x{j_{0}}{\alpha}= 0\, ,\notag
\end{align}
since the weight matrices are mutually independent -- and independent of the input -- and have zero mean \eqref{eq:deep-linear-weight-init}. By a similar argument, it's easy to see that any odd-point correlator of preactivations will vanish as well. Thus, going forward, we will only have to concern ourselves with the even-point correlators.

\section{Criticality}\label{sec:criticality_DLN}

Since the mean is trivial, 
the next simplest candidate for an interesting observable 
is the two-point correlator $\E{\z{i_1}{\alpha_1}{\ell}\z{i_2}{\alpha_2}{\ell}}$, which quantifies the typical magnitudes of the preactivations. We'll first go through the math, and then we'll discuss the physics.

\index{correlator!two-point}\index{Wick contraction}
\subsubsection{Math: recursion for the two-point correlator}
Let's start slowly by first considering the two-point correlator 
in the first layer. 
Using the defining equation~\eqref{eq:deep-linear-preactivation-compact} to express the first-layer preactivations in terms of the inputs as 
\be\label{eq:deep-linear-preactivations-first-layer}
\z{i}{\alpha}{1} = \sum_j^{n_0} \Ti{W}{ij}{1}\x{j}{\alpha} \, ,
\ee
we can express the two-point correlator as
\begin{align}\label{eq:deep-linear-two-point-first-layer}
\E{\z{i_1}{\alpha_1}{1}\z{i_2}{\alpha_2}{1}}&=\sum_{j_1,j_2=1}^{n_0}\E{\Ti{W}{i_1j_1}{1}\x{j_1}{\alpha_1}\Ti{W}{i_2j_2}{1}\x{j_2}{\alpha_2}}\\
&=\sum_{j_1,j_2=1}^{n_0}\E{\Ti{W}{i_1j_1}{1}\Ti{W}{i_2j_2}{1}} \x{j_1}{\alpha_1}\x{j_2}{\alpha_2} \notag \\
&=\sum_{j_1,j_2=1}^{n_0}\frac{C_W}{n_0}\delta_{i_1 i_2}\delta_{j_1 j_2}\x{j_1}{\alpha_1}\x{j_2}{\alpha_2}=\delta_{i_1 i_2}C_W\frac{1}{n_0}\sum_{j=1}^{n_0}\x{j}{\alpha_1}\x{j}{\alpha_2}\, , \notag
\end{align}
where
to go from the second line to the third line we Wick-contracted the two weights and inserted the variance~\eqref{eq:deep-linear-weight-init}.
Additionally, let us introduce the notation
\be\label{eq:deep-linear-normalized-inner-product-inputs}
\PH_{\alpha_1\alpha_2}\equiv \frac{1}{n_0}\sum_{i=1}^{n_0}\x{i}{\alpha_1}\x{i}{\alpha_2}\, ,
\ee
for the inner product of the two inputs, normalized by the input dimension $n_0$. In terms of this object, we can rewrite the first-layer two-point correlator \eqref{eq:deep-linear-two-point-first-layer} as
\be\label{eq:deep-linear-two-point-first-layer-simple}
\E{\z{i_1}{\alpha_1}{1}\z{i_2}{\alpha_2}{1}}=\delta_{i_1i_2} C_W \PH_{\alpha_1\alpha_2}\, .
\ee

\index{Wick contraction}\index{correlator!two-point}
Next, we could mindlessly repeat the same exercise to get the two-point correlator in any arbitrary layer, using the defining equation~\eqref{eq:deep-linear-preactivation-compact} to express $\z{i}{\alpha}{\ell}$ in terms of the input. Instead,
in order to practice our recursive approach,
let's 
evaluate the two-point correlator recursively.
To do so, we inductively assume that the two-point correlator at the $\ell$-th layer is known and then derive the two-point correlator at the $(\ell+1)$-th layer.
Using the iteration equation~\eqref{eq:deep-linear-foward-pass} with the bias set to zero, we find
\begin{align}
\label{eq:two-point-function-deep-linear-layer-ell}
\E{\z{i_1}{\alpha_1}{\ell+1}\z{i_2}{\alpha_2}{\ell+1}}=&\sum_{j_1,j_2=1}^{n_{\ell}}\E{\Ti{W}{i_1j_1}{\ell+1}\Ti{W}{i_2j_2}{\ell+1}\z{j_1}{\alpha_1}{\ell}\z{j_2}{\alpha_2}{\ell}}\, \\
=&\sum_{j_1,j_2=1}^{n_{\ell}}\E{\Ti{W}{i_1j_1}{\ell+1}\Ti{W}{i_2j_2}{\ell+1}}\E{\z{j_1}{\alpha_1}{\ell}\z{j_2}{\alpha_2}{\ell}}\, \notag \\
=&\delta_{i_1 i_2}C_W \frac{1}{n_{\ell}}\sum_{j=1}^{n_{\ell}}\E{\z{j}{\alpha_1}{\ell}\z{j}{\alpha_2}{\ell}}\, ,\nonumber%
\end{align}
where to go from the first line to the second line we used the fact that the weights $W^{(\ell+1)}$ of the $(\ell+1)$-th layer are statistically independent from the preactivations $z^{(\ell)}$ in the $\ell$-th layer, and to go from the second line to the third line we Wick-contracted the two weights and substituted in the variance \eqref{eq:deep-linear-weight-init}. Notice that at \emph{any} layer, the two-point correlator is proportional to the \terminate{Kronecker delta} $\delta_{i_1 i_2}$, vanishing unless the \terminate{neural indices} $i_1$ and $i_2$ are the same. 
With that in mind, let us decompose the two-point correlator as
\be\label{eq:deep-linear-covariance-notation}
\E{\z{i_1}{\alpha_1}{\ell}\z{i_2}{\alpha_2}{\ell}}\equiv  \delta_{i_1 i_2 }\Ti{G}{\alpha_1\alpha_2}{\ell}\, ,
\ee
and introduce a generalization of the above notation \eqref{eq:deep-linear-normalized-inner-product-inputs} for an arbitrary layer $\ell$.
Multiplying this equation by $\delta_{i_1i_2}$, summing over $i_1,i_2=1,\ldots,n_{\ell}$ and dividing it by $n_{\ell}$, the quantity $\Ti{G}{\alpha_1\alpha_2}{\ell}$ can also be expressed as 
\be\label{eq:deep-linear-covariance-notation-2}
\Ti{G}{\alpha_1\alpha_2}{\ell}= \frac{1}{n_{\ell}}\sum_{j=1}^{n_{\ell}}\E{\z{j}{\alpha_1}{\ell}\z{j}{\alpha_2}{\ell}}\, ,
\ee
and can thus be thought of as the average inner-product of preactivations in the $\ell$-th layer, divided by the number of neurons in the layer $n_{\ell}$. 
This inner product
depends on sample indices \emph{only} and lets us interpret $\Ti{G}{\alpha_1\alpha_2}{\ell} \equiv G^{(\ell)}\!\le(x_{\alpha_1},x_{\alpha_2}\ri)$ as the covariance of the two inputs, $x_{\alpha_1}$ and $x_{\alpha_2}$, after passing through an $\ell$-layer \terminate{deep linear network}.

With all this notation introduced and fully interpreted, it's easy to see that the above recursion~\eqref{eq:two-point-function-deep-linear-layer-ell} can be compactly summarized by %
\be\label{eq:deep-linear-kernel-recursion}
 \Ti{G}{\alpha_1\alpha_2}{\ell+1}=C_W  \Ti{G}{\alpha_1\alpha_2}{\ell}\, ,
\ee
which describes how the covariance $\Ti{G}{\alpha_1\alpha_2}{\ell}$ evolves from layer to layer.
Apparently, to transform the covariance from layer $\ell$ to layer $\ell+1$, we simply multiply by the constant $C_W$.  The initial condition $\Ti{G}{\alpha_1\alpha_2}{0}$ is given by the inner product of the two inputs \eqref{eq:deep-linear-normalized-inner-product-inputs}, and the solution is an exponential
\be\label{eq:deep-linear-exponential-solution}
\Ti{G}{\alpha_1\alpha_2}{\ell} = \le(C_W\ri)^{\ell}\Ti{G}{\alpha_1\alpha_2}{0}\, ,
\ee
as is typical for a repeated application of 
matrix multiplication. 
Note that the factor of the width $n_{\ell}$ in the variance of the weights~\eqref{eq:deep-linear-weight-init} nicely dropped out, indicating that this was in fact the proper way to scale the variance.

\subsubsection{Physics: criticality}
\index{criticality}
Already at this point our analysis illustrates an interesting and very general phenomenon. 
Considering the solution \eqref{eq:deep-linear-exponential-solution}, generically one of two things happens.
If $C_W> 1$, the covariance blows up exponentially, quickly being driven to a fixed point $\Tif{G}{\alpha_1\alpha_2} = \infty$ for all pairs of inputs and leading to a divergent network output.
If $C_W< 1$, the covariance exponentially decays to a fixed point $\Tif{G}{\alpha_1\alpha_2} = 0$ for all pairs of inputs, quickly curtailing any data dependence in the network output. 
Any time an observable approaches a value exponentially quickly, we'll refer to the limiting value as a \textbf{trivial fixed point}\index{fixed point!trivial|textbf}\index{trivial fixed point|see{fixed point}}. The value $\Tif{G}{\alpha_1\alpha_2} = \infty$ associated with $C_W>1$ and the value $\Tif{G}{\alpha_1\alpha_2} = 0$ associated with $C_W<1$ are prime examples of a trivial fixed point.

Exploring this further, first note that
the diagonal part of the covariance at the output layer $L$ estimates the typical magnitude of the output for a given input $\x{i}{\alpha}$
\be
\Ti{G}{\alpha\alpha}{L} =  \E{\frac{1}{n_{L}}\sum_{j=1}^{n_{L}}\le(\z{j}{\alpha}{L}\ri)^2} \, .
\ee
With this observable in mind, the aforementioned exponential behavior should immediately set off alarm bells, signaling either some sort of numerical instability ($C_W>1$) or loss of information ($C_W<1$). 
In addition, note that the target values for the different components of the network output are typically $\o{1}$ numbers, neither exponentially large nor small. Such exponential behavior of the network should thus make it extremely difficult to learn to approximate the desired function.
In this way, this \emph{exploding and vanishing covariance problem}\index{exploding and vanishing kernel problem} is a baby version of the infamous \terminate{exploding and vanishing gradient problem} -- a well-known obstacle to gradient-based training of deep networks -- which we shall make more precise in~\S\ref{ch:eft-ntk}. %

However, we were actually a little too quick in our analysis before: what happens if we tune the weight variance $C_W$ so that it's precisely equal to $1$? This is clearly a special point in the hyperparameter space of initialization, separating the exponentially growing solution from the exponentially decaying solution. Going back to the recursion \eqref{eq:deep-linear-kernel-recursion}, we see that if $C_W=1$ then the covariance is fixed $\Ti{G}{\alpha_1\alpha_2}{\ell}=\Ti{G}{\alpha_1\alpha_2}{0}\equiv\Tif{G}{\alpha_1\alpha_2}$, manifestly preserving the full covariance of the input data even after passing through many layers of the \terminate{deep linear network}.
This is a bona fide \textbf{nontrivial fixed point}\index{fixed point!nontrivial|textbf}\index{fixed point!nontrivial|seealso{criticality}}, as it doesn't exponentially trivialize the structure of input data.
Thus, at least at this heuristic level of analysis, choosing $C_W=1$  appears to be essential for preserving the structure of the input data in a numerically stable manner. More generally, flowing to a nontrivial fixed point seems to be a necessary condition for deep networks to do anything useful.

\index{spin|seealso{\texttt{bit} (unit of entropy)}}\index{critical initialization hyperparameters|see{initialization hyperparameters}}\index{initialization hyperparameters!critical|textbf}\index{criticality}\index{statistical physics}\index{critical phenomena}\index{magnetism}\index{spin}\index{temperature}\index{iron}\index{atom}\index{paramagnetism}\index{ferromagnetism}\index{phase transition}\index{magnetic field}
When we fine-tune\index{fine tuning} the \terminate{initialization hyperparameters} of a network so that the covariance avoids exponential behavior, we'll call them \textbf{critical initialization hyperparameters}.\footnote{This word choice is motivated by the analogy to critical phenomena in statistical physics. For instance, consider the prototypical example: a magnet made of iron. At high temperature, the magnetic moments -- or spins -- of the iron atoms point in random directions, leading to a paramagnetic phase without any coherent magnetic field. By contrast, at low temperature, the spins instead try to collectively orient in the same direction, leading to a ferromagnetic phase with coherent magnetic field -- think of the $\cap$-shaped cartoon magnet that children play with. A \terminate{critical temperature} separates these two phases of magnetism, and the magnet set to the critical temperature will exhibit very special behavior that is neither paramagnetism nor ferromagnetism but known as self-similarity.}
For deep linear networks\index{deep linear network}, the critical initialization hyperparameter $C_W=1$ separates two regimes, one with an exponentially growing covariance for $C_W>1$, and the other with an exponentially decaying covariance for $C_W<1$.
When the weight variance is tuned to criticality $C_W=1$, the network has a perfect self-similarity of the covariance, preserving it exactly through the evolution from layer to layer.

In \S\ref{ch:signalprop}, we will extend our analysis of \term{criticality} to MLPs that use any 
particular
activation function. And, as shall be seen further on
in~\S\ref{ch:NTHb} and~\S\ref{ch:eot},
tuning a network to criticality is \emph{critical} for any \emph{deep} network to be well behaved and perform useful tasks -- at least without otherwise employing ad-hoc
tricks to ensure that signals can propagate stably.

\section{Fluctuations}\label{sec:fluctuations_DLN}
\index{correlator!higher-point} \index{connected correlator!four-point}
Recall from~\S\ref{ch:tools} that if a distribution 
is Gaussian and has a zero mean, then the covariance 
completely specifies the distribution. 
If the preactivation distribution $p\!\le(z^{(\ell)}\Big\vert\D \ri)$ were Gaussian, this would
mean that the critical tuning of the one initialization hyperparameter $C_W = 1$ would be sufficient to ensure that any observable is well behaved. However, if the distribution $p\!\le(z^{(\ell)}\Big\vert\D \ri)$ is not Gaussian, then it's not clear a priori whether observables depending on higher-point connected correlators will be well behaved with the same tuning. In principle, such observables could require other tunings of $C_W$ that are incompatible with the critical setting $C_W = 1$ for the covariance $\Ti{G}{\alpha_1\alpha_2}{\ell}$. To settle this question, let's look at the next simplest observable, the connected four-point correlator. As before, we'll go through the math first and discuss the physics second.

In this section and the next, to simplify the algebra we'll focus on correlators of preactivations that are evaluated only on a single input $x_{\alpha}=x$. This is sufficient to qualitatively highlight the importance of the higher-point correlators while letting us avoid the interference of some annoying technical manipulations.
Accordingly, in these sections we will drop the \terminate{sample indices} on preactivations and  denote the covariance as
\be
G_{2}^{(\ell)}\equiv\Ti{G}{\alpha\alpha}{\ell}=G^{(\ell)}(x,x) \, .
\ee
In the next chapter, we'll consider the fully general case.

\subsubsection{Math: recursion for the four-point correlator}\index{correlator!four-point}
As we did for the two-point correlator in the previous section, we'll begin by working out the four-point correlator in the first layer and 
then next derive and solve a recursion for the correlator in the deeper layers.
First for the first layer, using the defining equation~\eqref{eq:deep-linear-preactivations-first-layer} with the sample index omitted,
we have
for the \emph{full} four-point correlator
\begin{align}\label{eq:deep-linear-four-point-first-layer}
&\E{z_{i_1}^{(1)}z_{i_2}^{(1)}z_{i_3}^{(1)}z_{i_4}^{(1)}}\, \\
=&\sum_{j_1,j_2,j_3,j_4=1}^{n_0}\E{\Ti{W}{i_1j_1}{1}\Ti{W}{i_2j_2}{1}\Ti{W}{i_3j_3}{1}\Ti{W}{i_4j_4}{1}}x_{j_1}x_{j_2}x_{j_3}x_{j_4}\, \nonumber\\
=&\frac{C_W^2}{n_0^2}\sum_{j_1,j_2,j_3,j_4=1}^{n_0}\le(\delta_{i_1i_2}\delta_{j_1j_2}\delta_{i_3i_4}\delta_{j_3j_4}+\delta_{i_1i_3}\delta_{j_1j_3}\delta_{i_2i_4}\delta_{j_2j_4}+\delta_{i_1i_4}\delta_{j_1j_4}\delta_{i_2i_3}\delta_{j_2j_3}\ri)x_{j_1}x_{j_2}x_{j_3}x_{j_4}\, \nonumber\\
=&C_W^2  \le(\delta_{i_1i_2}\delta_{i_3i_4}+\delta_{i_1i_3}\delta_{i_2i_4}+\delta_{i_1i_4}\delta_{i_2i_3}\ri)\le(\PH_2\ri)^2\, \notag.
\end{align}
where to go from line two to line three, we made three distinct pairings for the two Wick contractions\index{Wick contraction} of the four weights, 
and then used the weight variance~\eqref{eq:deep-linear-weight-init} to evaluate each contraction.
To get to the final line, we evaluated the sums over the $j$ indices and then substituted using our definition of the inner product~\eqref{eq:deep-linear-normalized-inner-product-inputs}, which 
for a single input simply reads
\be
\PH_2=\frac{1}{n_0}\sum_{j=1}^{n_0}x_j x_j\, .
\ee
Comparing this result~\eqref{eq:deep-linear-four-point-first-layer} with the two-point correlator in the first layer~\eqref{eq:deep-linear-two-point-first-layer-simple}, we note that this answer is precisely what we'd expect for the full four-point correlator if the preactivation distribution were exactly Gaussian. Thus, deep linear networks appear to be simply Gaussian after a single layer, at least at the four-point correlator level of analysis.\footnote{In the next chapter, we'll show very generally that the preactivation distribution is always Gaussian in the first layer.}
\index{deep linear network}

This Gaussianity does \emph{not} hold in deeper layers.  To see that, let's derive and solve a recursion for the four-point correlator. 
Beginning with the iteration equation~\eqref{eq:deep-linear-foward-pass} with zero bias, we find
\begin{align}\label{eq:deep-linear-four-point-a}
&\E{z_{i_1}^{(\ell+1)}z_{i_2}^{(\ell+1)}z_{i_3}^{(\ell+1)}z_{i_4}^{(\ell+1)}}\, \\
=&\sum_{j_1,j_2,j_3,j_4=1}^{n_{\ell}}\E{\Ti{W}{i_1j_1}{\ell+1}\Ti{W}{i_2j_2}{\ell+1}\Ti{W}{i_3j_3}{\ell+1}\Ti{W}{i_4j_4}{\ell+1}}\E{z_{j_1}^{(\ell)}z_{j_2}^{(\ell)}z_{j_3}^{(\ell)}z_{j_4}^{(\ell)}}\, \nonumber\\
=&\frac{C_W^2}{n_{\ell}^2}\sum_{j_1,j_2,j_3,j_4=1}^{n_{\ell}}\le(\delta_{i_1i_2}\delta_{j_1j_2}\delta_{i_3i_4}\delta_{j_3j_4}+\delta_{i_1i_3}\delta_{j_1j_3}\delta_{i_2i_4}\delta_{j_2j_4}+\delta_{i_1i_4}\delta_{j_1j_4}\delta_{i_2i_3}\delta_{j_2j_3}\ri)\, \nonumber\\
&\quad \quad \quad \quad \quad \quad \times \E{z_{j_1}^{(\ell)}z_{j_2}^{(\ell)}z_{j_3}^{(\ell)}z_{j_4}^{(\ell)}}\,  \nonumber \\
=&C_W^2\le(\delta_{i_1i_2}\delta_{i_3i_4}+\delta_{i_1i_3}\delta_{i_2i_4}+\delta_{i_1i_4}\delta_{i_2i_3}\ri)\frac{1}{n_{\ell}^2}\sum_{j,k=1}^{n_{\ell}}\E{z_{j}^{(\ell)}z_{j}^{(\ell)}z_{k}^{(\ell)}z_{k}^{(\ell)}}\, ,\nonumber
\end{align}
where on the second line we used the independence of the $(\ell+1)$-th-layer weights from the $\ell$-th-layer preactivations, 
on the third line we again made three distinct pairings for the two pairs of Wick contractions\index{Wick contraction} of 
the four weights, and on the last line we made judicious use of the Kronecker deltas to collapse the sums.  
\index{Kronecker delta} 

\index{tensor decomposition!four-point correlator}
Now, we see from this recursion that at \emph{any} layer the full four-point correlator is proportional to the factor $\le(\delta_{i_1i_2}\delta_{i_3i_4}+\delta_{i_1i_3}\delta_{i_2i_4}+\delta_{i_1i_4}\delta_{i_2i_3}\ri)$, a fixed tensor structure that specifies the neural-index dependence of the correlator. Thus by decomposing the full four-point correlator as
\be\label{eq:four-point-decompose}
\E{z_{i_1}^{(\ell)}z_{i_2}^{(\ell)}z_{i_3}^{(\ell)}z_{i_4}^{(\ell)}}\equiv \le(\delta_{i_1i_2}\delta_{i_3 i_4}+ \delta_{i_1i_3}\delta_{i_2 i_4}+ \delta_{i_1i_4}\delta_{i_2 i_3}\ri)G_4^{(\ell)} \, ,
\ee
we can put all of the layer dependence into this simpler object $\Ti{G}{4}{\ell}$ and not worry about \terminate{neural indices} in our recursion.
In terms of this decomposition, the result~\eqref{eq:deep-linear-four-point-first-layer} for the correlator in the first layer becomes
\be\label{eq:deep-linear-four-point-first-layer-after-decompose}
\Ti{G}{4}{1} = C_W^2 \le( \Ti{G}{2}{0}\ri)^2 \, ,
\ee
and the final factor in the above recursion~\eqref{eq:deep-linear-four-point-a} can be rewritten as
\be
\frac{1}{n_{\ell}^2}\sum_{j,k=1}^{n_{\ell}}\E{z_{j}^{(\ell)}z_{j}^{(\ell)}z_{k}^{(\ell)}z_{k}^{(\ell)}}=\frac{1}{n_{\ell}^2}\sum_{j,k=1}^{n_{\ell}} \le(\delta_{jj}\delta_{kk}+ \delta_{jk}\delta_{jk}+ \delta_{jk}\delta_{kj}\ri)G_4^{(\ell)}=\le(1+\frac{2}{n_{\ell}}\ri)G_4^{(\ell)}\, .
\ee
Using this, the entire recursion above~\eqref{eq:deep-linear-four-point-a} can be rewritten simply as a recursion for $\Ti{G}{4}{\ell}$ as
\be
G_4^{(\ell+1)}=C_W^2\le(1+\frac{2}{n_{\ell}}\ri)G_4^{(\ell)}\, .
\ee
This recursion, with the initial condition set by~\eqref{eq:deep-linear-four-point-first-layer-after-decompose}, has a simple solution
\begin{align}\label{eq:deep-linear-four-point-solution}
G_4^{(\ell)} =& C_W^{2\ell}\le[\prod_{\ell'=1}^{\ell-1}\le(1+\frac{2}{n_{\ell'}}\ri)\ri]\le(\Ti{G}{2}{0}\ri)^2 \, \\
=&\le[\prod_{\ell'=1}^{\ell-1}\le(1+\frac{2}{n_{\ell'}}\ri)\ri] \le(\Ti{G}{2}{\ell}\ri)^2\, ,\nonumber
\end{align}
where in the final line we substituted in the solution~\eqref{eq:deep-linear-exponential-solution} for the covariance.
Now let's 
extract some physics from this compact formula.

\subsubsection{Physics: large-$n$ expansion, non-Gaussianities, interactions, and fluctuations}
\index{$1/n$ expansion}\index{infinite-width limit!of deep linear networks}
To start, we note that the four-point correlator \eqref{eq:deep-linear-four-point-solution} drastically simplifies in the limit of an infinite number of neurons per hidden layer $(n_{\ell}\to\infty)$. In such a limit, the solution~\eqref{eq:deep-linear-four-point-solution} degenerates to
\be
G_4^{(\ell)}= \le(\Ti{G}{2}{\ell}\ri)^2 \, ,
\ee
and the full four-point correlator~\eqref{eq:four-point-decompose} becomes
\be
\E{z_{i_1}^{(\ell)}z_{i_2}^{(\ell)}z_{i_3}^{(\ell)}z_{i_4}^{(\ell)}} %
= \le(\delta_{i_1i_2}\delta_{i_3 i_4}+ \delta_{i_1i_3}\delta_{i_2 i_4}+ \delta_{i_1i_4}\delta_{i_2 i_3}\ri) \le(\Ti{G}{2}{\ell}\ri)^2 \, .%
\ee
This is exactly what we'd find if the preactivation distribution were Gaussian: the four-point correlator is determined entirely by the two-point correlator, with the tensor structure determined by \terminate{Wick's theorem}. In fact, as we will show
in the next chapter, for any MLP with any particular choice of a nonlinear activation function, the preactivation distribution is governed by Gaussian statistics in this infinite-width limit, implying no interactions between the neurons in such a limit.
\index{correlator!four-point}\index{infinite-width limit!of deep linear networks}
However, despite the rather large computational resources that \neo{big tech} can throw at machine-learning problems, realistic MLPs simply do not have an infinite number of neurons per layer. To understand such realistic MLPs, we'll have to back off this infinite-width limit.

To illustrate this most clearly, let's set all the hidden layer widths to be equal $n_1=n_2=\ldots=n_{L-1}\equiv n$. Then, evaluating \eqref{eq:deep-linear-four-point-solution}, the deviation from the infinite-width limit at the level of four-point correlator statistics is encoded by the difference
\begin{align}\label{eq:leading-four-point-correction}
G_4^{(\ell)}-\le(\Ti{G}{2}{\ell}\ri)^2=&\le[\le(1+\frac{2}{n}\ri)^{\ell-1} -1\ri]\le(\Ti{G}{2}{\ell}\ri)^2\, \\
=&\frac{2(\ell-1)}{n}\le(\Ti{G}{2}{\ell}\ri)^2+\o{\frac{1}{n^2}}\, ,\nonumber
\end{align}
where in the last line we expanded in $1/n$ and kept
the leading correction to the infinite-width limit.\footnote{This approximation is valid so long as the depth of the network doesn't grow too large.
Stay tuned for the analysis in the next section where we will discuss how this limit breaks down.} In particular, at criticality where $\Ti{G}{2}{\ell}$ is constant, this leading correction \eqref{eq:leading-four-point-correction} scales inversely proportionally with the width and proportionally with the depth.
Thus, the deviation from infinite width is proportional to the depth-to-width ratio of the network, our first encounter with this important \term{emergent scale}.
There are multiple ways to think about this finite-width correction.

\index{connected correlator!four-point}\index{nearly-Gaussian distribution}\index{coupling!quartic}
First, the connected four-point correlator~\eqref{eq:C4} is given by
\be\label{eq:four-point-decompose-connected}
\Ec{z_{i_1}^{(\ell)}z_{i_2}^{(\ell)}z_{i_3}^{(\ell)}z_{i_4}^{(\ell)}}{\Big|} = \le(\delta_{i_1i_2}\delta_{i_3 i_4}+ \delta_{i_1i_3}\delta_{i_2 i_4}+ \delta_{i_1i_4}\delta_{i_2 i_3}\ri)\le[G_4^{(\ell)}-\le(\Ti{G}{2}{\ell}\ri)^2\ri]  \, ,
\ee
which directly connects the difference~\eqref{eq:leading-four-point-correction} to our measure of non-Gaussianity for the distribution. We see that the non-Gaussianity grows as the network deepens, and the preactivation statistics in layer $\ell$ are \emph{nearly-Gaussian} so long as the emergent scale, the depth-to-width-ratio, remains perturbatively small.
From the action perspective, this means that the quartic coupling changes -- or \textbf{runs} -- as the layer at which we consider the preactivation distribution changes, with the coupling growing in proportion with layer $\ell$. \index{running coupling}

\index{connected correlator!four-point} 
Second,  in~\S\ref{sec:perturbation} we gave another interpretation for a nonzero connected four-point correlator as measuring \terminate{interactions} -- i.e.~the breakdown of \terminate{statistical independence} -- between the different components of the random vector.
To be very specific, let us look at a particular entry of the connected four-point correlator tensor with $i_1=i_2=j$ and $i_3=i_4=k$ for $j\ne k$. This entry can be expressed as
\be\label{eq:deep-linear-scale}
\E{\le(z_{j}^{(\ell)}z_{j}^{(\ell)}  -\Ti{G}{2}{\ell}\ri)\le(z_{k}^{(\ell)}z_{k}^{(\ell)}  -\Ti{G}{2}{\ell}\ri)} = \Ti{G}{4}{\ell} - \le(\Ti{G}{2}{\ell}\ri)^2\, , \quad \text{for}\ j\ne k\, .
\ee
This shows that the deviation of $z_{j}^{(\ell)}z_{j}^{(\ell)}$ from its mean value $\E{z_{j}^{(\ell)}z_{j}^{(\ell)}}=\Ti{G}{2}{\ell}$ on a particular neuron $j$ is correlated with the same deviation from the mean on a different neuron $k$.
We can thus interpret the finite-width difference~\eqref{eq:leading-four-point-correction} as controlling intralayer interactions between distinct neurons, with the strength of the interactions growing with depth.

\index{fluctuations!in deep linear networks}\index{typicality}
Third, we can see that some observables that are deterministic in the \terminate{infinite-width limit} start to fluctuate at finite width.
To this end, let us consider the simple observable
\be
\O^{(\ell)}\equiv \O\!\le(z^{(\ell)}\ri)\equiv \frac{1}{n}\sum_{j=1}^{n} z_{j}^{(\ell)}z_{j}^{(\ell)}\, , \quad \text{for}\ \ell< L\, ,
\ee
which captures the average magnitude of the preactivations over all the different neurons in a hidden layer $\ell$ for a given instantiation of the network weights. Its mean over different realizations of the weights is given by the expectation
\be
\E{\O^{(\ell)}}=\frac{1}{n}\sum_{j=1}^{n}\E{z_{j}^{(\ell)}z_{j}^{(\ell)}}=\Ti{G}{2}{\ell}\, ,
\ee
and the magnitude of this observable's fluctuation from instantiation to instantiation is measured by its variance
\begin{align}
\E{\le(\O^{(\ell)}-\E{\O^{(\ell)}}\ri)^2}=&\frac{1}{n^2}\sum_{j,k=1}^{n}\E{z_{j}^{(\ell)}z_{j}^{(\ell)}z_{k}^{(\ell)}z_{k}^{(\ell)}}-\le(\Ti{G}{2}{\ell}\ri)^2\, \\
=&\frac{1}{n^2}\sum_{j,k=1}^{n} \le(\delta_{jj}\delta_{kk}+ \delta_{jk}\delta_{jk}+ \delta_{jk}\delta_{kj}\ri)G_4^{(\ell)}-\le(\Ti{G}{2}{\ell}\ri)^2\, \nonumber\\
=&\frac{2}{n}G_4^{(\ell)}+ \le[G_4^{(\ell)}-\le(\Ti{G}{2}{\ell}\ri)^2\ri]\, \nonumber\\
=&\frac{2\ell}{n}\le(\Ti{G}{2}{\ell}\ri)^2+\o{\frac{1}{n^2}}\, ,\nonumber
\end{align}
where in the last step we recalled the expansion~\eqref{eq:leading-four-point-correction} for the finite-width difference.
As promised, $\O^{(\ell)}$ is deterministic at infinite width, since this variance is suppressed by $1/n$ and vanishes identically in the \terminate{infinite-width limit}.
However, as we back off the \terminate{infinite-width limit}, the variance grows linearly with depth at criticality due to the finite-width correction~\eqref{eq:leading-four-point-correction}.
As such depth increases, the fluctuation becomes larger, meaning that 
 the \emph{typical} magnitude of the preactivations $\O^{(\ell)}$ measured on any given realization of the deep linear network may deviate more from the mean value $\E{\O^{(\ell)}}=\Ti{G}{2}{\ell}$.

All these finite-width effects -- be they non-Gaussianities, intralayer interactions, or finite-width fluctuations --  are proportional to the depth-to-width ratio of the network.
This is perhaps the most important recurring theme of the book: the leading finite-width contributions at \terminate{criticality} grow linearly with depth, despite being suppressed by the inverse of the layer widths.
Since the depths of a real networks with at least one hidden layer are bounded from below as $L \ge 2$ -- that is, at minimum such networks have one hidden layer and one output layer -- in practice, networks of any finite size will express some amount of finite-width effects in their \terminate{output distribution} proportional to their aspect ratio $L/n$.
As we will see later in \S\ref{ch:signalprop}, this emergent scaling will hold very generally
at \terminate{criticality} for networks with any particular activation function.

\index{representation learning}
Thus, the deeper a network is, the less the infinite-width Gaussian description will apply, due to accumulation of finite-width \terminate{fluctuations}. This is actually a good thing because, as we shall emphasize more in~\S\ref{ch:features}, 
infinite-width networks do not have correlations among neurons within a layer 
and cannot learn %
nontrivial
representations from input data.
Real useful deep learning systems that are used in practice do both of these things, and our later analysis will show that
deeper networks have the capacity to do more of these things.

Depth, however, is a double-edged sword.
As the overall depth $L$ of a network becomes comparable to its hidden-layer width, fluctuations can begin to dominate. %
In particular, such extremely deep networks will have a huge variation in observables from instantiation to instantiation. Thus, even if we choose the critical initialization hyperparameter $C_W=1$, in some instantiations signals blow up, in other instantiations signals decay, and rarely do they stay tamed to be of order one. From a practical point of view, these networks are pretty useless.

\index{architecture hyperparameters}\index{initialization hyperparameters}
This set of circumstances is actually very fortuitous from a theorist's vantage point: our \neo{effective theory of deep learning} is most accurate when the 
aspect ratio
$L/n$ of the network is small but nonzero -- due to the applicability of the perturbative large-width expansion -- and this is exactly the setting of these architecture hyperparameters where networks work best
in practice.
In fact, one could expect that balancing the utility of nonzero depth for learning features\index{feature} against the cost of growing fluctuations 
could result in some optimal
aspect ratio  $L/n$ for MLPs of a particular activation, just as we saw that there is a correct tuning for the initialization hyperparameter $C_W$ for deep linear networks. We will return to this question of tuning $L/n$ when we discuss \terminate{inductive bias} in~\S\ref{ch:bayesian-inference} after first redoing our analysis of 
\terminate{criticality} and \terminate{fluctuations} %
for arbitrary activation functions
in the following two chapters, \S\ref{ch:ngp} and \S\ref{ch:signalprop}. In particular, in these chapters we will understand how the statistics of the preactivations run\index{running coupling} with depth, and see the emergence of the depth-to-width ratio as a scale that controls the validity of 
the perturbative \terminate{$1/n$ expansion}, as was the case here for deep linear networks\index{deep linear network}.

Quite generally, in the regime where perturbation theory works, the finite-width corrections grow linearly -- \emph{not} exponentially -- with depth and the network remains well behaved. By contrast, when the depth-to-width ratio becomes large, perturbation theory breaks down, making it very difficult to analyze such networks. However, in the special case of deep linear networks\index{deep linear network} a nonperturbative analysis is possible. In the next section we'll illustrate explicitly what happens to deep linear networks\index{deep linear network} when the depth-to-width ratio grows very large in order to paint an intuitive picture of the way networks behave in this regime.

\section{Chaos}\label{sec:solution_DLN}
\index{chaos|seealso{overly deep}}\index{chaos}
\index{correlator!higher-point}
In the last two sections we used the method of Wick contractions\index{Wick contraction} to derive recursions for the two-point and four-point correlators of deep linear networks\index{deep linear network}, which we then easily solved. Now, we will use this same method 
to compute all the higher-point correlators
in order to complete our goal 
of determining the full distribution $p\!\le(z^{(\ell)}\Big| \D \ri)$. 
So that we may first simplify the algebra and then focus on the interesting properties of this distribution,
we'll again evaluate the correlators only on a single input $x_\alpha=x$
and drop the \terminate{sample indices} in all of the following equations.
Math then physics.

\subsubsection{Math: recursions for six-point and higher-point correlators}
Starting with the first layer, let's compute a general $2m$-point \emph{full} correlator. As this involves many Wick contractions\index{Wick contraction}, it might be helpful to remind yourself of the formal statement of \terminate{Wick's theorem} by flipping back to \S\ref{sec:Gauss} and consulting \eqref{eq:Wick-multi}\ldots. Good.

Now, using the defining equation~\eqref{eq:deep-linear-preactivations-first-layer} to express the first-layer preactivations in terms of the input, we get
\begin{align}\label{eq:deep-linear-gaussian-first-layer}
&\E{\Ti{z}{i_1}{1}\Ti{z}{i_2}{1}\cdots \Ti{z}{i_{2m-1}}{1}\Ti{z}{i_{2m}}{1}} \\
\notag
=&\sum_{j_1,\ldots,j_{2m}=1}^{n_0}\E{\Ti{W}{i_1j_1}{1}\Ti{W}{i_2j_2}{1}\cdots\Ti{W}{i_{2m-1}j_{2m-1}}{1}\Ti{W}{i_{2m}j_{2m}}{1}}x_{j_1}x_{j_2}\cdots x_{j_{2m-1}}x_{j_{2m}} \\
\notag
=&\le( \sum_{\text{all parings}}\delta_{i_{k_1} i_{k_2}} \cdots \delta_{i_{k_{2m-1}}i_{k_{2m}}}\ri)C_W^m\le(\Ti{G}{2}{0}\ri)^m \,  \notag\\
=&\le( \sum_{\text{all parings}}\delta_{i_{k_1} i_{k_2}} \cdots \delta_{i_{k_{2m-1}}i_{k_{2m}}}\ri)\le(\Ti{G}{2}{1}\ri)^m \, , \notag
\end{align}
where, as before we used \terminate{Wick's theorem} to determine the Wick contractions\index{Wick contraction} and then evaluated each contraction by substituting in \eqref{eq:deep-linear-weight-init} for the variance. Here, the sum is over all the possible pairing of the $2m$ auxiliary indices, $k_1,\ldots, k_{2m}$, resulting in $(2m-1)!!$ distinct terms, and on the final line  we substituted in the solution~\eqref{eq:deep-linear-exponential-solution} for the first-layer covariance.

The result \eqref{eq:deep-linear-gaussian-first-layer} confirms what we suspected in the last section,
that the preactivation distribution for the first layer is completely Gaussian. If this isn't clear by inspection, 
it's easy to check directly
-- via basically the same application of \terminate{Wick's theorem} -- 
that the correlators \eqref{eq:deep-linear-gaussian-first-layer} are precisely the $2m$-point correlators of a \terminate{Gaussian distribution} with zero mean and variance $\delta_{i_1i_2}G_2^{(1)}$.
In other words, the preactivation distribution in the first layer is governed by the quadratic action \index{action!quadratic}
\be
S\!\le(z^{(1)} \ri) = \frac{1}{2 G_2^{(1)}} \sum_{i=1}^{n_1} z_i^{(1)} z_i^{(1)}   \, .
\ee

\index{correlator!six-point}
Before presenting a recursion for general $2m$-point correlators, let us work out the recursion for the six-point correlator in detail. Beginning with the iteration equation~\eqref{eq:deep-linear-foward-pass} with the bias set to zero,
we find
\begin{align}
&\E{\Ti{z}{i_1}{\ell+1}\Ti{z}{i_2}{\ell+1} \Ti{z}{i_{3}}{\ell+1}\Ti{z}{i_{4}}{\ell+1}\Ti{z}{i_{5}}{\ell+1}\Ti{z}{i_{6}}{\ell+1}}\, \\
=&\sum_{j_1,j_2,j_3,j_4,j_5,j_6 = 1}^{n_{\ell}}\E{\Ti{W}{i_1j_1}{\ell+1}\Ti{W}{i_2j_2}{\ell+1} \Ti{W}{i_3j_3}{\ell+1}\Ti{W}{i_4j_4}{\ell+1}\Ti{W}{i_5j_5}{\ell+1}\Ti{W}{i_6j_6}{\ell+1}}\E{\Ti{z}{j_1}{\ell}\Ti{z}{j_2}{\ell} \Ti{z}{j_{3}}{\ell}\Ti{z}{j_{4}}{\ell}\Ti{z}{j_{5}}{\ell}\Ti{z}{j_{6}}{\ell}}\, \nonumber\\
=& C_W^3\Big(\delta_{i_1 i_2}\delta_{i_3 i_4}\delta_{i_5 i_6}+\delta_{i_1 i_3}\delta_{i_2 i_4}\delta_{i_5 i_6}+\delta_{i_1 i_4}\delta_{i_2 i_3}\delta_{i_5 i_6}\, \nonumber\\
&\quad +\delta_{i_1 i_2}\delta_{i_3 i_5}\delta_{i_4 i_6}+\delta_{i_1 i_3}\delta_{i_2 i_5}\delta_{i_4 i_6}+\delta_{i_1 i_5}\delta_{i_2 i_3}\delta_{i_4 i_6}\, \nonumber\\
&\quad +\delta_{i_1 i_2}\delta_{i_5 i_4}\delta_{i_3 i_6}+\delta_{i_1 i_5}\delta_{i_2 i_4}\delta_{i_3 i_6}+\delta_{i_1 i_4}\delta_{i_2 i_5}\delta_{i_3 i_6}\, \nonumber\\
&\quad +\delta_{i_1 i_5}\delta_{i_3 i_4}\delta_{i_2 i_6}+\delta_{i_1 i_3}\delta_{i_5 i_4}\delta_{i_2 i_6}+\delta_{i_1 i_4}\delta_{i_5 i_3}\delta_{i_2 i_6}\, \nonumber\\
&\quad +\delta_{i_5 i_2}\delta_{i_3 i_4}\delta_{i_1 i_6}+\delta_{i_5 i_3}\delta_{i_2 i_4}\delta_{i_1 i_6}+\delta_{i_5 i_4}\delta_{i_2 i_3}\delta_{i_1 i_6} \Big)\frac{1}{n_{\ell}^3}\sum_{i,j,k=1}^{n_{\ell}}\E{\Ti{z}{i}{\ell}\Ti{z}{i}{\ell} \Ti{z}{j}{\ell}\Ti{z}{j}{\ell}\Ti{z}{k}{\ell}\Ti{z}{k}{\ell}}\, ,\nonumber
\end{align}
noting again the independence of the $(\ell+1)$-th layer weights from the $\ell$-th layer preactivations.
On the final line, we see that there were fifteen distinct ways to make the three Wick contractions of six weights.

\index{Kronecker delta}
As we saw for the four-point correlator, the structure of \terminate{neural indices} for the full six-point correlator is the same for \emph{any} layer and proportional to a constant tensor, given by the object in the parenthesis above with all those Kronecker deltas. This suggests a decomposition of the six-point correlator as %
\be\label{eq:deep-linear-six-point-decomposition}\index{tensor decomposition!six-point correlator}
\E{\Ti{z}{i_1}{\ell}\Ti{z}{i_2}{\ell} \Ti{z}{i_{3}}{\ell}\Ti{z}{i_{4}}{\ell}\Ti{z}{i_{5}}{\ell}\Ti{z}{i_{6}}{\ell}}\equiv \le(\delta_{i_1 i_2}\delta_{i_3 i_4}\delta_{i_5 i_6}+\ldots+\delta_{i_5 i_4}\delta_{i_2 i_3}\delta_{i_1 i_6}\ri) \Ti{G}{6}{\ell}\, ,
\ee
with the neural-dependence encapsulated by that complicated sum-over-products of Kronecker deltas and the layer dependence captured solely by $ \Ti{G}{6}{\ell}$.

Now, to find a recursion for $\Ti{G}{6}{\ell}$, we need to perform the sum
\be
\frac{1}{n_{\ell}^3}\sum_{i,j,k=1}^{n_{\ell}}\E{\Ti{z}{i}{\ell}\Ti{z}{i}{\ell} \Ti{z}{j}{\ell}\Ti{z}{j}{\ell}\Ti{z}{k}{\ell}\Ti{z}{k}{\ell}}\, 
\ee
after substituting in the decomposition \eqref{eq:deep-linear-six-point-decomposition}.
With the given pattern of \terminate{neural indices}, there are 
really only three types of terms in the sum. In particular, there is one term that looks like this
\be
\frac{1}{n_{\ell}^3}\sum_{i,j,k=1}^{n_{\ell}} \delta_{ii}\delta_{jj}\delta_{kk}=1\, ,
\ee
six terms that look like this
\be
\frac{1}{n_{\ell}^3}\sum_{i,j,k=1}^{n_{\ell}} \delta_{ij}\delta_{ji}\delta_{kk}=\frac{1}{n_{\ell}}\, ,
\ee
and eight terms that look like this
\be
\frac{1}{n_{\ell}^3}\sum_{i,j,k=1}^{n_{\ell}} \delta_{i j}\delta_{jk}\delta_{ki}=\frac{1}{n_{\ell}^2}\, .
\ee
Putting all these terms together, we find a recursion for the layer-dependence of the full six-point correlator
\be
G_{6}^{(\ell+1)}=C_W^3 \le(1+\frac{6}{n_{\ell}}+\frac{8}{n_{\ell}^2}\ri) G_{6}^{(\ell)}\, ,
\ee
which has a simple solution
\begin{align}
G_{6}^{(\ell)}=&C_W^{3\ell} \le[\prod_{\ell'=1}^{\ell-1}\le(1+\frac{6}{n_{\ell'}}+\frac{8}{n_{\ell'}^2}\ri)\ri]\le(G_{2}^{(0)} \ri)^3\, \\
=&\le[ \prod_{\ell'=1}^{\ell-1}\le(1+\frac{6}{n_{\ell'}}+\frac{8}{n_{\ell'}^2}\ri)\ri]\le(G_{2}^{(\ell)} \ri)^3 \nonumber\, .
\end{align}
Here, we used the initial condition \eqref{eq:deep-linear-gaussian-first-layer} $\Ti{G}{6}{0}=\le(\Ti{G}{2}{0} \ri)^3$, and on the final line we substituted in our solution for the variance of a single input~\eqref{eq:deep-linear-exponential-solution}.

\index{correlator!$2m$-point}
Similarly, we can decompose an arbitrary $2m$-point correlator as
\be\label{eq:deep-linear-inductive-ansatz}
\E{\Ti{z}{i_1}{\ell}\Ti{z}{i_2}{\ell}\cdots \Ti{z}{i_{2m-1}}{\ell}\Ti{z}{i_{2m}}{\ell}}=\le(\sum_{\text{all parings}}\delta_{i_{k_1} i_{k_2}} \cdots \delta_{i_{k_{2m-1}}i_{k_{2m}}}\ri)G^{(\ell)}_{2m} \, ,
\ee 
and use a similar set of manipulations to show that the layer dependence $G^{(\ell)}_{2m}$ obeys a recursion
\be
G_{2m}^{(\ell+1)}=c_{2m}(n_{\ell})\, C_W^{m}  G_{2m}^{(\ell)}\, ,
\ee
with the combinatorial factor $c_{2m}(n)$ given by
\be\label{eq:combinatorial-2m}
 c_{2m}(n)=\le(1+\frac{2}{n}\ri)\le(1+\frac{4}{n}\ri)\cdots\le(1+\frac{2m-2}{n}\ri)=\frac{\le(\frac{n}{2}-1 + m\ri )!}{\le( \frac{n}{2} -1 \ri)!} \le( \frac{2}{n} \ri) ^{m}\, .
\ee
We included the explicit form of this factor only for completeness. If you insist on checking this factor, note that it reproduces the right combinatorial factors for $2m=2,4,6$,
though
we strongly suggest that you do not explicitly write out all of the terms for any other particular value of $m$.
Overall, this recursion is still just a simple sequence of multiplications, with a simple solution
\be\label{eq:2m-full-solution}
G_{2m}^{(\ell)}=\le[\prod_{\ell'=1}^{\ell-1} c_{2m}(n_{\ell}') \ri]  \le( G_{2}^{(\ell)}\ri)^m \ .
\ee
Enough with the math, time for the physics.\footnote{
If you \emph{do} want more math, 
check out \cite{ZavatonePehlevanFinite} for an alternative derivation of these $2m$-point correlators and a nonperturbative expression for the distribution $p\big(z^{(\ell)}\big| x \big)$.
}

\subsubsection{Physics: breakdown of perturbation theory and the emergence of chaos}
Let's play with this formula \eqref{eq:2m-full-solution} a bit by taking various limits. For simplicity, let's set all the hidden layer widths to be equal $n_1=n_2=\ldots=n_{L-1}\equiv n$,
and also focus only on \terminate{output distribution} $p\!\le(z^{(L)} \Big| x \ri)$.

\bi
\item On the one hand, if we send the network width to infinity, $n\rightarrow\infty$, while keeping the depth $L$ fixed, then all the combinatorial factors~\eqref{eq:combinatorial-2m} become unity:
\be
\lim_{n\rightarrow\infty} c_{2m}(n)=1\, .
\ee
In this \terminate{infinite-width limit}, all the correlators~\eqref{eq:2m-full-solution} are given by their Gaussian values
\be\label{eq:deep-linear-higher-order-gaussian-limit}
G_{2m}^{(L)}=\le( G_{2}^{(L)}\ri)^m\, ,
\ee
and the \terminate{output distribution} $p\!\le(z^{(L)} \Big| x \ri)$ is precisely Gaussian.
More generally, even for multiple inputs the \terminate{output distribution} $p\!\le(z^{(L)} \Big| \D \ri)$ remains Gaussian,\index{Gaussian distribution} with covariance $\Ti{G}{\alpha_1\alpha_2}{L}= C_W^L\le(\frac{1}{n_0}\sum_{i=1}^{n_0}\x{i}{\alpha_1}\x{i}{\alpha_2}\ri)$.
As this distribution is equivalent to that of one-layer networks initialized with weight variance $C_W^L$, we see that such networks are not really deep after all.
\index{non-Gaussian distribution}\index{correlator!higher-point}
\item  On the other hand, if we send the depth to infinity, $L\rightarrow\infty$, while keeping the width $n$ fixed, then all the combinatorial factors are fixed and greater than one, $c_{2m}> 1$. This means that the higher-point correlators for $2m >2$ will all blow up exponentially as
\be\label{eq:deep-linear-higher-order-chaotic-limit}
G_{2m}^{(L)} = \Big[c_{2m}(n)\Big]^{L-1} \le(G_2^{(L)} \ri)^m \, .
\ee
Note that this behavior persists \emph{even if} we tune the network to \terminate{criticality} by setting $C_W=1$ so that the two-point correlator is fixed $G_2^{(\ell)}  = G_2^{(0)}$. This shows explicitly how our large-width analysis from the last section can break down if the network depth becomes too large.
Furthermore, the distribution implied by these correlators is extremely non-Gaussian, to say the least, and in practice 
the outputs of these networks will fluctuate chaotically from instantiation to instantiation. Such networks are entirely unusable. %
\item Clearly these limits do not commute, i.e., 
\be
\lim_{n \to \infty}  \lim_{L \to \infty} G_{2m}^{(L)}\neq \lim_{L \to \infty}  \lim_{n \to \infty}G_{2m}^{(L)}\, .
\ee
However, we can construct an \emph{interpolating solution} by sending both the width and depth to infinity, $n,L\rightarrow\infty$, while keeping 
their ratio fixed:
\be
\ratio\equiv \frac{L}{n}\, .
\ee
Noting that we can expand the combinatorial factors as
\be
c_{2m}(n)=1+\frac{1}{n}\le(\sum_{s=1}^{m-1}2s\ri)+\o{\frac{1}{n^2}}=1+\frac{m(m-1)}{n}+\o{\frac{1}{n^2}}\, ,
\ee
and then using the well-known formula for the exponential
\be
\lim_{L\rightarrow\infty}\le[1+\frac{a}{L}+\o{\frac{1}{L^2}}\ri]^{L}=e^{a}\, , 
\ee
we can construct a limiting value for any correlator at a given value of $m$ and fixed aspect ratio $r$:
\be\label{eq:interpolating}
G_{2m}^{(L)}\rightarrow e^{m(m-1)\ratio}\le( G_{2}^{(L)}\ri)^m\, .
\ee
This solution interpolates between the two extreme limits: by sending $r\to 0$ we recover the Gaussian limit \eqref{eq:deep-linear-higher-order-gaussian-limit}, and by sending $r\to \infty$ we recover the chaotic limit \eqref{eq:deep-linear-higher-order-chaotic-limit} that demonstrates the breakdown of \terminate{criticality}.\footnote{This \emph{double-scaling limit} corresponds to neglecting terms that scale like $\frac{L}{n^2}$, $\frac{L^3}{n^5}$, $\frac{L^{120}}{n^{157}}$, etc., which are all subleading when the depth and the width are large, $n,L\rightarrow\infty$, but their ratio $r$ is fixed.

Furthermore, there is a very subtle point in using this interpolating solution -- albeit a theoretical subtlety -- when we consider not just a particular correlator at a given $2m$, but the set of \emph{all} the correlators.
Namely, for any \emph{finite} $n,L$ -- no matter how big -- there always exist higher-point correlators
for which the exponential approximation~\eqref{eq:interpolating} is invalid because the factor of $m(m-1)$ becomes too big. %
That is, since we constructed this interpolating solution assuming fixed $m$, such a solution can break down if $m$ is large enough.
}
\ei

\index{connected correlator!four-point}\index{running coupling}\index{coupling!quartic}\index{connected correlator!six-point}\index{nearly-Gaussian distribution}
Let us play a tiny bit more with the last interpolating formula~\eqref{eq:interpolating} at \terminate{criticality}
where $G_{2}^{(L)}=G_{2}^{(0)}$.
Here, the finite-width difference~\eqref{eq:leading-four-point-correction} that governs the connected four-point correlator~\eqref{eq:four-point-decompose-connected} becomes
\begin{align}
G_{4}^{(L)}-\le(G_{2}^{(L)}\ri)^2&=\le(e^{2r}-1\ri)\le(G_{2}^{(0)}\ri)^2\, \\
&=2r \le(G_{2}^{(0)}\ri)^2+\o{r^2}\, . \notag
\end{align}
This reproduces the \emph{running} of the quartic coupling with the depth-to-width ratio~\eqref{eq:leading-four-point-correction}. Similarly,
the corresponding quantity governing the layer dependence of the connected six-point correlator~\eqref{eq:C6} is given by
\begin{align}
G_{6}^{(L)}-3G_{2}^{(L)}G_{4}^{(L)}+2\le(G_{2}^{(L)}\ri)^3&=\le(e^{6r}-3e^{2r}+2\ri)\le(G_{2}^{(0)}\ri)^3\\
&=12r^2 \le(G_{2}^{(0)}\ri)^3 +\o{r^3}\, , \notag
\end{align}
which scales like the depth-to-width ratio \emph{squared}. Therefore, the connected six-point correlator is even more suppressed than the connected four-point correlator for large networks with sufficiently small depth-to-width ratio $r$. This is in accord with the comments we made in~\S\ref{sec:perturbation}: neural networks obey \emph{nearly-Gaussian} statistics, and
the connected correlators 
have a hierarchical structure. In particular, we see here that the scaling of the correlators is controlled by the same small parameter $r$,  with the higher-point connected correlators suppressed by a higher power of that parameter. This means that for small $r$, we should be able to consistently truncate our distribution and only compute up to a fixed order in $r$.

%% file: Chp4-NGP/4_global.tex
\chapter{RG Flow of Preactivations}\index{representation group flow!of preactivations}
\label{ch:ngp}

\epigraph{``You can hide a lot in a large-$N$ matrix.'' -- Steve Shenker\index{Shenker, Stephen} \emph{-- John McGreevy~\cite{mcgreevy2010holographic}.\index{McGreevy, John}} }{}

\noindent{}At the end of the last chapter,
we computed the statistics of preactivations for deep linear networks at initialization
and saw them \emph{run} as a function of the network depth. For that toy model, using a handful of
Wick contractions
and the recursive structure of the network architecture, we were able to fully understand the effects of the network's hyperparameters -- its initialization scheme, width, and depth -- on preactivation correlators. This exercise in particular highlighted the importance of \emph{critical initialization hyperparameters} and sufficiently small \emph{depth-to-width ratio}
in order for the network outputs to be well-behaved, theoretically and practically. 
To extend these insights beyond deep linear networks, we need to develop an \terminate{effective theory of deep learning} for networks with any activation function.

While ultimately the goal of our effective theory is to explain how a \emph{particular} neural network learns from a given dataset, our immediate goal in~\S\ref{ch:ngp} and~\S\ref{ch:signalprop}
will be to understand how an \emph{ensemble} of neural networks at initialization behaves as a function of data.
In
\S\ref{ch:NTHb}, \S\ref{ch:features}, and \S\ref{ch:eot},
we'll find that these goals are closely tied together:
through the judicious study of the ensemble,
we can systematically evaluate the \emph{typical}\index{typicality} behavior of trained
networks
as well as
how any particular network may \emph{fluctuate} away from typicality.
Our starting point will thus be a study of the statistics of neural-network preactivations with Gaussian-initialized biases and weights. 
All in all, the formalism developed in this chapter for analyzing the ensemble of networks at initialization will be the key to a principled understanding of deep learning.

As stressed in the introduction,~\S\ref{ch:introduction}, our focus will always be on describing real finite-width networks, since a lot is lost in idealized infinite-width networks.
One salient phenomenon lost in the infinite-width limit is the increasing non-Gaussianity in the preactivation distributions of deeper layers.
Such non-Gaussianity makes the behavior of finite-width networks much richer but more complicated to analyze.
In order to tame these complications, we'll
need to
borrow
some tools from theoretical physics\index{physics}. 
In particular, physicists have a long tradition of finding simple descriptions of complicated systems in the limit of a large number of degrees of freedom, while keeping in mind the true goal of modeling real systems. In our context, this hints at tractability and simplification in the regime where networks become very wide, though not infinitely so.
To make this precise, in this chapter we introduce the \emph{large-}$n$ expansion or  \terminate{$1/n$ expansion} in order to perform perturbative expansions when hidden-layer width $n$ becomes parametrically big.
With this tool, we'll be able to systematically study the preactivation distributions of finite neural networks to arbitrary precision.\footnote{
Back in 1996, Neal introduced the \neo{infinite-width limit} in a seminal work \cite{neal1996priors}, focusing on single-hidden-layer networks.  Much later, this program was continued in \cite{lee2018deep,matthews2018gaussian}, extending the infinite-width limit to deeper networks, and then was extended further by Yaida   
in \cite{Yaida2019} to \emph{finite-width networks}. A large part of this chapter is focused on reproducing the recursions first derived in \cite{Yaida2019}.

However, our perspective here is different than the one taken in this prior work. In particular, our main motivation is in computing the  distribution of preactivations at initialization, with an eye towards ultimately understanding gradient-based training (\S\ref{ch:NTHb}, \S\ref{ch:features}, \S\ref{ch:eot}), rather than providing a starting point for Bayesian inference. (We will give our own perspective on Bayesian learning for deep learning in \S\ref{ch:bayesian-inference}.) Additionally, in contrast to \cite{Yaida2019}, our results here are derived by first focusing on the couplings in the \neo{action}, rather than directly on the correlators of the distribution. This method is more intuitive and can be more easily extended.
}

As we did for deep linear networks, we will proceed recursively, investigating how the
distribution of preactivations changes from layer to layer by following the transformation of inputs
via the iterative MLP forward-pass equation.
We start in \S\ref{sec:first-layer-gaussian} by computing the distribution of preactivations in the first layer, integrating out the first set of weights and biases. This procedure recovers a well-known result that the distribution of the first-layer preactivations is Gaussian.
Since this calculation is so central to the rest of the chapter, we'll present two different derivations: a combinatorial derivation in terms of Wick contractions and an algebraic derivation using the Hubbard-Stratonovich transformation.

Next, in \S\ref{sec:second-layer-non-gaussian}, we'll consider the
distribution of preactivations in the second layer and see the emergence of non-Gaussianity in four-point and higher-point connected correlators.
The magnitude of these correlators is
suppressed when the network is very wide, vanishing in the strict infinite-width limit.
This suppression for wide networks in turn enables us to write down an action describing the preactivation distribution, building on the correspondence explored in~\S\ref{ch:tools} between such connected correlators and the couplings in the action.
In particular, the large-$n$ expansion lets us start with the quadratic action describing the Gaussian distribution in the infinite-with limit and then perturbatively expand around it in a series of the inverse width, $1/n$, to arbitrary desired precision.
Given the importance of this result, we again provide two derivations, one based on Wick contractions and the other based on expanding the stochastic metric.

Finally, in~\S\ref{sec:deeper-layer-accumulation}, we'll analyze the
distribution of preactivations at any depth.
At this point we can simply repurpose the calculations from the preceding sections to see how the distribution of preactivations recursively transforms from the $\ell$-th layer to the $(\ell+1)$-th layer.
In particular, keeping the leading finite-width $1/n$ corrections, we'll obtain recursion equations for the two-point and 
four-point correlators,
encoding how these observables evolve with increasing depth.
We'll see that the preactivation distribution of the $(\ell+1)$-th layer contains a nearly-Gaussian piece inherited from the $\ell$-th layer as well as an additional near-Gaussianity generated in the transition from the $\ell$-th to $(\ell+1)$-th layer.
In the next chapter, \S\ref{ch:signalprop}, we'll see in detail how the near-Gaussianity accumulates with depth by explicitly solving these recursions and analyzing their solutions, which extends the notion of criticality and emergence of the depth-to-width ratio to networks with general activation functions.

After a short clarifying section on some implications of marginalization (\S\ref{sec:sum-rule}) and a section on subleading corrections (\S\ref{sec:loop-correction}),
we take a step back in \S\ref{sec:marginalization-group-flow}  in order to
draw a parallel between our formalism and 
 the \emph{renormalization group} in theoretical physics\index{physics}. 
Renormalization group is a powerful recursive method for understanding complicated interacting systems, capturing how the effective interactions between the constituents of a system change when the scale at which they are measured changes from microscopic to macroscopic.
Specifically, renormalization marginalizes over the microscopic degrees of freedom in the system to yield an effective \emph{coarse-grained}\index{coarse-graining} description at long distances.\index{coarse-graining|seealso{representation group flow}}\index{coarse-graining|seealso{renormalization group flow}}
This is analogous to the way we recursively marginalize over preactivations in previous layers to obtain an effective description of a \neo{representation}\index{representation|seealso{feature}} at the current layer,
in our case capturing how the interactions between neurons change with depth.
In both cases the flow of the distributions is created by the marginalization of fine-grained information. Given the complete parallel, we will call our flow \emph{representation group (RG) flow}.\index{representation group flow}

If this sounds like a popular heuristic explanation for what deep neural networks do -- transforming fine-grained information at the input level into coarser information at the feature levels and finally into fully coarse-grained representation at the output level
-- that's because our formalism makes this heuristic picture of representation coarse-graining concrete.\footnote{There have been many formal and informal comments on the connection between renormalization and deep learning, but the relationship has never before been made precise.}
Our formalism will further let us directly probe the effect of the \emph{deep} in \emph{deep learning} by tracking the change in preactivation distributions as we increase the number of layers.
Thus, it is the starting point for an \terminate{effective theory of deep learning}, which we will continue to develop throughout the book.

\section{First Layer: Good-Old Gaussian}
\label{sec:first-layer-gaussian}

Given a dataset
\be
\D=\le\{\x{i}{\alpha}\ri\}_{i=1,\ldots,n_0;\, \alpha=1,\ldots,\ND}
\ee
containing $\ND$ inputs of $n_0$-dimensional vectors, the preactivations in the first layer are given by
\be\label{eq:first-layer-preactivation-def}
\z{i}{\alpha}{1} \equiv z_i^{(1)}(x_\alpha)=\bias{i}{1}+\sum_{j=1}^{n_{0}}\W{ij}{1}\x{j}{\alpha}\,,  \quad \text{for} \quad i=1,\ldots,n_1\, .
\ee
At initialization the biases $b^{(1)}$ and weights $W^{(1)}$ are independently distributed according to mean-zero Gaussian distributions with variances
\begin{align}
\label{eq:bias-variance-def-first}
\mathbb{E}\le[b^{(1)}_{i}b^{(1)}_{j}\ri]&=\delta_{ij} C_{b}^{(1)}\, ,\\ %
\label{eq:weight-variance-def-first}
\mathbb{E}\le[W^{(1)}_{i_1 j_1}W^{(1)}_{i_2 j_2}\ri]&=\delta_{i_1 i_2} \delta_{j_1 j_2}\frac{C_{W}^{(1)}}{n_{0}}\, .%
\end{align}
The first-layer preactivations $z^{(1)}=\z{i}{\alpha}{1}$ form an $(n_1\ND)$-dimensional vector, and we are interested in its distribution at initialization, 
\be\label{eq:first-layer-distribution}
p\!\le(z^{(1)}\Big\vert\D\ri)= p\!\le(z^{(1)}\le(x_1\ri), \ldots, z^{(1)}\le(x_{\ND}\ri) \ri)\, .%
\ee
Note how this distribution depends conditionally on the input data, representing the fact that the preactivations are functions of the input.

Now, let us compute the distribution of the first-layer preactivations at initialization.
Since this will be so important, we give two derivations, one combinatorial and one algebraic.

\subsubsection{Wick this way: combinatorial derivation via correlators}
The first derivation involves direct application of Wick contractions\index{Wick contraction} to compute correlators of the first-layer distribution~\eqref{eq:first-layer-distribution}.
Starting with the one-point correlator, simply inserting the definition of the first-layer preactivations~\eqref{eq:first-layer-preactivation-def} gives
\be
\E{\z{i}{\alpha}{1}} = \E{\bias{i}{1}+\sum_{j=1}^{n_{0}}\W{ij}{1}\x{j}{\alpha_1}} = 0\,,
\ee
since $\E{\bias{i}{1}}=\E{\W{ij}{1}}=0$. In fact, it's easy to see that all the odd-point correlators of $p\!\le(z^{(1)}\Big\vert\D\ri)$ vanish because there always is an odd number of either biases $b^{(1)}$ or weights $W^{(1)}$ left unpaired under Wick contractions.

Next for the two-point correlator, again inserting the definition~\eqref{eq:first-layer-preactivation-def}, we see
\begin{align}\label{eq:first-layer-second-moment}
\E{\z{i_1}{\alpha_1}{1}\z{i_2}{\alpha_2}{1}}&=\E{\le(\bias{i_1}{1}+\sum_{j_1=1}^{n_{0}}\W{i_1j_1}{1}\x{j_1}{\alpha_1}\ri)\le(\bias{i_2}{1}+\sum_{j_2=1}^{n_{0}}\W{i_2 j_2}{1}\x{j_2}{\alpha_2}\ri)}\,  \\
&=\delta_{i_1 i_2}\le(\Cb{1}+\CW{1}\frac{1}{n_0}\sum_{j=1}^{n_0}\x{j}{\alpha_1}\x{j}{\alpha_2} \ri)=\delta_{i_1 i_2} \Ti{G}{\alpha_1 \alpha_2}{1}\, , \nonumber
\end{align}
where to get to the second line we Wick-contracted the biases and weights using \eqref{eq:bias-variance-def-first} and \eqref{eq:weight-variance-def-first}. We also introduced the first-layer \term{metric}\index{metric|seealso{kernel}}\index{metric|seealso{data-dependent coupling}}\index{metric!first-layer}
\be\label{eq:first-layer-metric}
\Ti{G}{\alpha_1 \alpha_2}{1}\equiv\Cb{1}+\CW{1}\frac{1}{n_0}\sum_{j=1}^{n_0}\x{j}{\alpha_1}\x{j}{\alpha_2}\,  ,
\ee
which is a function of the two samples, $\Ti{G}{\alpha_1\alpha_2}{1}=G^{(1)}(x_{\alpha_1}, x_{\alpha_2})$, and represents the two-point correlation of preactivations in the first layer between different samples. 

The higher-point correlators can be obtained similarly. For instance, the full four-point correlation can be obtained by inserting the definition~\eqref{eq:first-layer-preactivation-def} four times and Wick-contracting the biases and weights, yielding
\begin{align}\label{eq:first-layer-fourh-moment}
&\E{\z{i_1}{\alpha_1}{1}\z{i_2}{\alpha_2}{1}\z{i_3}{\alpha_3}{1}\z{i_4}{\alpha_4}{1}}\, \\
=&\delta_{i_1 i_2} \delta_{i_3 i_4} \Ti{G}{\alpha_1 \alpha_2}{1}\Ti{G}{\alpha_3 \alpha_4}{1}+\delta_{i_1 i_3} \delta_{i_2 i_4}\Ti{G}{\alpha_1 \alpha_3}{1}\Ti{G}{\alpha_2 \alpha_4}{1}+\delta_{i_1 i_4}\delta_{i_2 i_3} \Ti{G}{\alpha_1 \alpha_4}{1}\Ti{G}{\alpha_2 \alpha_3}{1}\,   \nonumber\\
=&\E{\z{i_1}{\alpha_1}{1}\z{i_2}{\alpha_2}{1}}\E{\z{i_3}{\alpha_3}{1}\z{i_4}{\alpha_4}{1}}+\E{\z{i_1}{\alpha_1}{1}\z{i_3}{\alpha_3}{1}}\E{\z{i_2}{\alpha_2}{1}\z{i_4}{\alpha_4}{1}}\, \nonumber\\
&+\E{\z{i_1}{\alpha_1}{1}\z{i_4}{\alpha_4}{1}}\E{\z{i_2}{\alpha_2}{1}\z{i_3}{\alpha_3}{1}}\,  .\nonumber
\end{align}
Note that the end result is same as Wick-contracting $z^{(1)}$'s with the variance given by~\eqref{eq:first-layer-second-moment}.  As we recall from \S\ref{ch:tools}, this can compactly be summarized by saying that the \emph{connected} four-point correlator vanishes,
\be
\E{\z{i_1}{\alpha_1}{1}\z{i_2}{\alpha_2}{1}\z{i_3}{\alpha_3}{1}\z{i_4}{\alpha_4}{1}}\Big\vert_{\text{connected}} = 0 \, .
\ee
Similar Wick combinatorics shows that all the full higher-point correlators can be obtained simply by Wick-contracting $z^{(1)}$'s with the variance given by~\eqref{eq:first-layer-second-moment}, and hence all the connected higher-point correlators vanish. 
This means that all correlators can be generated from a Gaussian distribution with zero mean and the variance \eqref{eq:first-layer-second-moment}.%

Then, in order to write down the first-layer action, all we need is the inverse of this variance, given by a matrix $\delta_{i_1i_2}  \Kinv{\alpha_1\alpha_2}{1}$ that satisfies
\be
\sum_{j =1}^{n_1}\sum_{\beta\in\D}\le(\delta_{i_1 j}  \Kinv{\alpha_1\beta}{1} \ri)\le(\delta_{j i_2} \Ti{G}{\beta\alpha_2}{1}\ri)=\delta_{i_1 i_2}\delta^{\alpha_1}_{\ \alpha_2}\, ,
\ee
with the inverse of the first-layer metric\index{metric!inverse} $\Ti{G}{\alpha_1\alpha_2}{1}$ denoted as $ \Kinv{\alpha_1\alpha_2}{1}$ and defined by
\be
\sum_{\beta\in \D}\Kinv{\alpha_1\beta}{1} \Ti{G}{\beta\alpha_2}{1}=\delta^{\alpha_1}_{\ \alpha_2}\, .
\ee
Just as in~\S\ref{ch:tools}, we
follow the conventions of \emph{general relativity} and suppress the superscript ``$-1$'' for the inverse metric, distinguishing the metric $\Ti{G}{\alpha_1\alpha_2}{1}$ and the inverse metric $\Kinv{\alpha_1\alpha_2}{1}$  by whether sample indices are lowered or raised.
With this notation, the Gaussian distribution for the first-layer preactivations is expressed as
\be
p\!\le(z^{(1)}\Big\vert\D\ri)=\frac{1}{Z} e^{-\EFT{1}} \, ,
\ee
with the quadratic action\index{action!quadratic}\index{Gaussian distribution!action}
\be\label{eq:Gauss-action}
\ac\!\le(z^{(1)}\ri)=\frac{1}{2} \sum_{i=1}^{n_1}\sum_{\alpha_1,\alpha_2\in\D}\Kinv{\alpha_1\alpha_2}{1} \z{i}{\alpha_1}{1} \z{i}{\alpha_2}{1} \, ,
\ee
and the \terminate{partition function}
\be
Z=\int \le[\prod_{i,\alpha}d\z{i}{\alpha}{1}\ri] e^{-\EFT{1}}=\dete{2\pi G^{(1)}}^{\frac{n_1}{2}}\, ,
\ee
where $\dete{2\pi G^{(1)}}$ is the determinant of the  $\ND$-by-$\ND$ matrix $2\pi G^{(1)}_{\alpha_1 \alpha_2}$ and, whenever we write out a determinant involving the metric, it will always be that of the metric and \emph{not} of the inverse metric.\footnote{N.B.~compared to the generic quadratic action introduced in \eqref{eq:intro-quadratic-action-reprint} where the random variable $z_\mu$ was a \emph{vector} with a general index $\mu$, here in \eqref{eq:Gauss-action} we've subdivided the general index into a pair of indices, $\mu \to (i, \alpha)$, so that the first-layer preactivation $\z{i}{\alpha}{1}$ is a \neo{tensor} with a neural index\index{neural indices} $i$ and a sample index\index{sample indices} $\alpha$.}

\subsubsection{Hubbard-Stratonovich this way: algebraic derivation via action}
Rather than first computing correlators and then backing out the distribution that generates them, we can instead work with the distribution directly. Let's start with the formal expression for the preactivation distribution~\eqref{eq:first-layer-formal-expression-first-encounter} worked out in the last chapter\footnote{For architectures other than MLPs, the expression inside the \terminate{Dirac delta function} would be different, but we expect much of the following to hold so long as the parameters are sampled from simple distributions.\index{multilayer perceptron}}
\be\label{eq:first-layer-formal-expression}
p\!\le(z\big\vert\D\ri)=\int \le[  \prod_{i} d b_{i}\ p\!\le(b_{i}\ri) \ri]  \le[\prod_{i,j} d W_{ij}\ p\!\le(W_{ij}\ri) \ri] \prod_{i,\alpha} \delta\!\le(z_{i;\alpha}-b_i-\sum_{j}W_{ij}\x{j}{\alpha}\ri)\, ,
\ee 
where we have momentarily suppressed the layer superscripts ``${(1)}$'' because it is distracting. At this point, we could try to eliminate some of the integrals over the model parameters against the constraints imposed by the Dirac delta functions, but it's easy to get confused by the different numbers of model-parameter integrals and delta-function constraints.\index{Dirac delta function}

\index{Dirac delta function!integral representation}
To clarify matters, we import a neat trick from theoretical physics\index{physics} called
the \term{Hubbard-Stratonovich transformation}.
Specifically, using the following \terminate{integral representation} of the Dirac delta function~\eqref{eq:integral-form-delta-function}
\be\label{eq:integral-form-delta-function-reprint}
\delta(z-a)=\int \frac{d\Lambda}{2\pi} e^{i \Lambda (z-a)}\,   
\ee
for each constraint and also plugging in explicit expressions for the Gaussian distributions over the parameters,  we obtain
\begin{align}
p\!\le(z\big\vert\D\ri)&= \int \le[\prod_{i} \frac{d b_{i}}{\sqrt{2\pi C_b}}\ri]  \le[\prod_{i,j} \frac{d W_{ij}}{\sqrt{2\pi C_W/n_0}} \ri]\le[\prod_{i,\alpha} \frac{d \HS_{i}^{\ \alpha}}{2\pi}\ri]\, \\
&\quad \quad \times\exp\!\le[-\sum_{i} \frac{b_i^2}{2C_b}-n_0\sum_{i,j} \frac{W_{ij}^2}{2C_W}+i\sum_{i,\alpha} \HS_{i}^{\ \alpha}\le(z_{i;\alpha}-b_i-\sum_{j}W_{ij}\x{j}{\alpha}\ri)\ri]\, .  \nonumber
\end{align}
Completing the square in the exponential for both the biases $b$ and weights $W$, we see that the action is quadratic in the model parameters
\begin{align}
&- \sum_{i}\frac{b_i^2}{2C_b}-n_0\sum_{i,j} \frac{W_{ij}^2}{2C_W}+i\sum_{i,\alpha} \HS_{i}^{\ \alpha}\le(z_{i;\alpha}-b_i-\sum_{j}W_{ij}\x{j}{\alpha}\ri)\, \\
=&-\frac{1}{2C_b}\sum_{i}\le(b_i+iC_b \sum_{\alpha}\HS_{i}^{\ \alpha}\ri)^2-\frac{C_b}{2}\sum_{i}\le(\sum_{\alpha}\HS_{i}^{\ \alpha}\ri)^2\, \nonumber\\
&- \frac{n_0}{2C_W}\sum_{i,j}\le(W_{ij}+i\frac{C_W}{n_0} \sum_{\alpha}\HS_{i}^{\ \alpha}\x{j}{\alpha}\ri)^2-\frac{C_W}{2n_0}\sum_{i,j}\le(\sum_{\alpha}\HS_{i}^{\ \alpha}\x{j}{\alpha}\ri)^2+i\sum_{i,\alpha} \HS_{i}^{\ \alpha} z_{i;\alpha}\, .\nonumber
\end{align}
The biases and weights can then be integrated out, yielding
an integral representation for the first-layer distribution $p(z)$ as
\be\label{eq:to-be-referenced-in-interlayer-part-far-ahead-in-the-future}
\int \le[\prod_{i,\alpha} \frac{d \HS_{i}^{\ \alpha}}{2\pi}\ri]\ \exp\!\le[-\frac{1}{2}\sum_{i,\alpha_1,\alpha_2}\HS_{i}^{\ \alpha_1}\HS_{i}^{\ \alpha_2} \le(C_b+C_W\sum_{j}\frac{\x{j}{\alpha_1}\x{j}{\alpha_2}}{n_0} \ri)+i\sum_{i,\alpha} \HS_{i}^{\ \alpha} z_{i;\alpha}\ri]\, .
\ee
In essence, we've so far traded the delta-function constraints and the model parameters for the auxiliary Hubbard-Stratonovich variables $\HS_{i}^{\ \alpha}$, which have quadratic action and a simple linear interaction with the preactivations $z_{i;\alpha}$.

Note that  the inverse variance for the Hubbard-Stratonovich variables $\HS_{i}^{\ \alpha}$ is just the first-layer metric~\eqref{eq:first-layer-metric} we introduced in the Wick-contraction derivation,
\be
\Cb{1}+\CW{1}\sum_{j}\frac{\x{j}{\alpha_1}\x{j}{\alpha_2}}{n_0}=\Ti{G}{\alpha_1\alpha_2}{1}\, ,
\ee
where by now enough dust has settled that layer superscripts  ``$(1)$'' have been restored.
Once again completing the square,
the argument of the exponential becomes
\be
-\frac{1}{2}\sum_{i,\alpha_1,\alpha_2}\le[\Ti{G}{\alpha_1\alpha_2}{1}\le(\HS_{i}^{\ \alpha_1}-i \sum_{\beta_1} \Kinv{\alpha_1 \beta_1}{1}\z{i}{\beta_1}{1}\ri)\le(\HS_{i}^{\ \alpha_2}-i \sum_{\beta_2} \Kinv{\alpha_2 \beta_2}{1}\z{i}{\beta_2}{1}\ri)+\Kinv{\alpha_1 \alpha_2}{1}\z{i}{\alpha_1}{1}\z{i}{\alpha_2}{1}\ri]\, ,
\ee
which finally lets us integrate out the Hubbard-Stratonovich variables $\HS_{i}^{\ \alpha}$ and recover our previous result
\be\label{eq:first-layer-distribution-HS-derivation}
p\!\le(z^{(1)}\Big\vert\D\ri)= \frac{1}{\dete{2\pi G^{(1)}}^{\frac{n_1}{2}}}\exp\!\le(-\frac{1}{2}\sum_{i=1}^{n_1}\sum_{\alpha_1,\alpha_2\in\D}\Kinv{\alpha_1 \alpha_2}{1}\z{i}{\alpha_1}{1}\z{i}{\alpha_2}{1}\ri)\, .
\ee
As before, $\dete{2\pi G^{(1)}}$ represents the determinant of the  matrix $2\pi G^{(1)}_{\alpha_1 \alpha_2}$.
The first-layer distribution is Gaussian with each neuron independent, and correlations between preactivations for different samples are encoded entirely in the metric $\Ti{G}{\alpha_1\alpha_2}{1}$. 

\subsubsection{Gaussian action in action}
Now that we've obtained an action representation for the distribution of the first-layer preactivations in two different ways, let's get a feel for how to compute with it. We'll start by computing the expectation of some quantities that will be needed in~\S\ref{sec:second-layer-non-gaussian}: the expectation of two activations on the same neuron, $\E{\sigma\!\le(\z{i_1}{\alpha_1}{1}\ri)\sigma\!\le(\z{i_1}{\alpha_2}{1}\ri)}$,  and the expectation of four activations, $\E{\sigma\!\le(\z{i_1}{\alpha_1}{1}\ri)\sigma\!\le(\z{i_1}{\alpha_2}{1}\ri)\sigma\!\le(\z{i_2}{\alpha_3}{1}\ri)\sigma\!\le(\z{i_2}{\alpha_4}{1}\ri)}$, either with all four on the same neuron $i_1=i_2$ or with each pair on two separate neurons $i_1\ne i_2$.

Let's start with the two-point correlator of activations. Using the definition of the expectation and inserting the action representation of the distribution~\eqref{eq:first-layer-distribution-HS-derivation}, we get
\begin{align}
&\E{\sigma\!\le(\z{i_1}{\alpha_1}{1}\ri)\sigma\!\le(\z{i_1}{\alpha_2}{1}\ri)}\, \\
=&\int \le[ \prod_{i=1}^{n_1} \frac{\prod_{\alpha\in\D} d z_{i;\alpha}}{\sqrt{\dete{2\pi G^{(1)}}}}\ri]\exp\!\le(-\frac{1}{2}\sum_{j=1}^{n_1}\sum_{\beta_1,\beta_2\in\D}\Kinv{\beta_1 \beta_2}{1}z_{j;\beta_1}z_{j;\beta_2}\ri)\sigma\!\le(z_{i_1;\alpha_1}\ri)\sigma\!\le(z_{i_1;\alpha_2}\ri)\,  \nonumber\\
=&\le\{\prod_{i\ne i_1}\int \le[ \frac{ \prod_{\alpha\in\D} d z_{i;\alpha}}{\sqrt{\dete{2\pi G^{(1)}}}}\ri]\exp\!\le(-\frac{1}{2}\sum_{\beta_1,\beta_2\in\D}\Kinv{\beta_1 \beta_2}{1}z_{i;\beta_1}z_{i;\beta_2}\ri)\ri\}\, \nonumber\\
&\times\int \le[ \frac{\prod_{\alpha\in\D} d z_{i_1;\alpha}}{\sqrt{\dete{2\pi G^{(1)}}}}\ri]\exp\!\le(-\frac{1}{2}\sum_{\beta_1,\beta_2\in\D}\Kinv{\beta_1 \beta_2}{1}z_{i_1;\beta_1}z_{i_1;\beta_2}\ri)\sigma\!\le(z_{i_1;\alpha_1}\ri)\sigma\!\le(z_{i_1;\alpha_2}\ri)\, \nonumber\\
=&\{1\}\times \le[\int \frac{\prod_{\alpha\in\D}  d z_{\alpha}}{\sqrt{\dete{2\pi G^{(1)}}}}\ri]\exp\!\le(-\frac{1}{2}\sum_{\beta_1,\beta_2\in\D}\Kinv{\beta_1 \beta_2}{1}z_{\beta_1}z_{\beta_2}\ri)\sigma\!\le(z_{\alpha_1}\ri)\sigma\!\le(z_{\alpha_2}\ri)\, \nonumber\\
\equiv& \bra \sigma\!\le(z_{\alpha_1}\ri)\sigma\!\le(z_{\alpha_2}\ri)\ket_{G^{(1)}}\, .\nonumber
\end{align}
The second equality states that the probability distribution factorizes for each neuron due to the relation $e^{x+y}=e^{x}e^{y}$. To go from the second equality to the third, we compute the integrals for the neurons with $i \neq i_1$, which are all trivial, and we also rename the dummy integral variable $z_{i_1;\alpha}$ to $z_{\alpha}$. The final equality reintroduces the notation \eqref{eq:gauss-braket}
\be\label{eq:gaussian-integration-metric-g}
\bra F\le(z_{\alpha_1},\ldots, z_{\alpha_m}\ri)\ket_{g}\equiv \int \le[\frac{\prod_{\alpha\in \D} dz_{\alpha}}{\sqrt{\dete{2\pi g}}}\ri]\exp\!\le(-\frac{1}{2}\sum_{\beta_1,\beta_2\in\D}g^{\beta_1\beta_2}z_{\beta_1}z_{\beta_2}\ri) F\!\le(z_{\alpha_1},\ldots, z_{\alpha_m}\ri)\, 
\ee
to describe a Gaussian expectation with variance $g$ and an arbitrary function $F\!\le(z_{\alpha_1},\ldots, z_{\alpha_m}\ri)$ over variables with sample indices \emph{only}. In other parts of this book we'll explicitly evaluate this type of Gaussian expectation in various setups for concrete choices of activation functions, but for the purpose of this chapter we will view computations as complete when they are reduced to such Gaussian expectations\index{Gaussian expectation} without any neural indices. %
Introducing further the simplifying notation
\be
\sigma_{\alpha}\equiv \sigma\!\le(z_{\alpha}\ri)\, ,
\ee
the result of the computation above can be succinctly summarized as
\be\label{eq:two-activations-Gauss}
\E{\sigma\!\le(\z{i_1}{\alpha_1}{1}\ri)\sigma\!\le(\z{i_1}{\alpha_2}{1}\ri)}= \bra \sigma_{\alpha_1}\sigma_{\alpha_2}\ket_{G^{(1)}}\, .
\ee

It's easy to generalize this to correlators of more than two activations. For instance, for four activations on the same neuron $i_1=i_2$, we have by the exact same manipulations
\be\label{eq:four-activations-one-neuron-Gauss} 
\E{\sigma\!\le(\z{i_1}{\alpha_1}{1}\ri)\sigma\!\le(\z{i_1}{\alpha_2}{1}\ri)\sigma\!\le(\z{i_1}{\alpha_3}{1}\ri)\sigma\!\le(\z{i_1}{\alpha_4}{1}\ri)}=\bra \sigma_{\alpha_1}\sigma_{\alpha_2}\sigma_{\alpha_3}\sigma_{\alpha_4}\ket_{G^{(1)}}\, ,
\ee
and for each pair on two different neurons $i_1\ne i_2$, we have
\begin{align}\label{eq:four-activations-two-neurons-Gauss} 
&\E{\sigma\!\le(\z{i_1}{\alpha_1}{1}\ri)\sigma\!\le(\z{i_1}{\alpha_2}{1}\ri)\sigma\!\le(\z{i_2}{\alpha_3}{1}\ri)\sigma\!\le(\z{i_2}{\alpha_4}{1}\ri)}\, \\
=&\le\{\prod_{  i\notin \{i_1,i_2\} }\int \le[ \frac{ \prod_{\alpha\in\D} d z_{i;\alpha}}{\sqrt{\dete{2\pi G^{(1)}}}}\ri]\exp\!\le(-\frac{1}{2}\sum_{\beta_1,\beta_2\in\D}\Kinv{\beta_1 \beta_2}{1}z_{i;\beta_1}z_{i;\beta_2}\ri)\ri\}\, \nonumber\\
&\times\int \le[ \frac{\prod_{\alpha\in\D} d z_{i_1;\alpha}}{\sqrt{\dete{2\pi G^{(1)}}}}\ri]\exp\!\le(-\frac{1}{2}\sum_{\beta_1,\beta_2\in\D}\Kinv{\beta_1 \beta_2}{1}z_{i_1;\beta_1}z_{i_1;\beta_2}\ri)\sigma\!\le(z_{i_1;\alpha_1}\ri)\sigma\!\le(z_{i_1;\alpha_2}\ri)\, \nonumber\\
&\times\int \le[ \frac{\prod_{\alpha\in\D} d z_{i_2;\alpha}}{\sqrt{\dete{2\pi G^{(1)}}}}\ri]\exp\!\le(-\frac{1}{2}\sum_{\beta_1,\beta_2\in\D}\Kinv{\beta_1 \beta_2}{1}z_{i_2;\beta_1}z_{i_2;\beta_2}\ri)\sigma\!\le(z_{i_2;\alpha_3}\ri)\sigma\!\le(z_{i_2;\alpha_4}\ri)\,  \nonumber\\
&=\bra \sigma_{\alpha_1}\sigma_{\alpha_2}\ket_{G^{(1)}}\bra \sigma_{\alpha_3}\sigma_{\alpha_4}\ket_{G^{(1)}}\, , \nonumber
\end{align}
where it's clear each neuron factorizes and gives separate Gaussian integrals.
This illustrates the fact that neurons are independent, and thus there is no interaction among different neurons in the first layer.
In deeper layers, the preactivation distributions are nearly-Gaussian\index{nearly-Gaussian distribution} and things will be a bit more complicated. %

\section{Second Layer: Genesis of Non-Gaussianity}
\label{sec:second-layer-non-gaussian}
In this section, we'll move onto evaluating the distribution of preactivations in the second layer of an MLP.
The second-layer preactivations are defined via
\be\label{eq:second-layer-preactivations}
\z{i}{\alpha}{2} \equiv z_i^{(2)}(x_\alpha)=
\bias{i}{2}+\sum_{j=1}^{n_{1}}\W{ij}{2}\s{j}{\alpha}{1}\,,  \quad \text{for} \quad i=1,\ldots,n_2\, ,
\ee
with the first-layer activations denoted as
\be
\s{i}{\alpha}{1}\equiv \sigma\!\le(\z{i}{\alpha}{1}\ri)\, ,
\ee
and the biases $b^{(2)}$ and weights $W^{(2)}$ sampled from Gaussian distributions.

The joint distribution of preactivations in the first and second layers can be factorized as
\be
p\!\le(z^{(2)},z^{(1)}\Big\vert\D\ri)=p\!\le(z^{(2)}\Big\vert z^{(1)}\ri) p\!\le( z^{(1)}\Big\vert\D\ri)\, .
\ee
Here the first-layer marginal distribution $p\!\le( z^{(1)}\Big\vert\D\ri)$ was evaluated in the last section, \S\ref{sec:first-layer-gaussian}, to be a Gaussian distribution~\eqref{eq:first-layer-distribution-HS-derivation} with the variance given in terms of the first-layer metric $\Ti{G}{\alpha_1\alpha_2}{1}$. As for the conditional distribution, we know that it can be expressed as\footnote{Again, the expression in the \terminate{Dirac delta function} is specific to \terminate{multilayer perceptron} architectures, but this formalism can easily be adapted for other architectures.}
\begin{align}\label{eq:second-conditioned-first}
&p\!\le(z^{(2)}\Big\vert z^{(1)}\ri)\, \\
=&\int \le[\prod_{i} d b_{i}^{(2)}\ p\!\le(b_{i}^{(2)}\ri)  \ri]\le[ \prod_{i,j} d W_{ij}^{(2)}\ p\!\le(W_{ij}^{(2)}\ri) \ri]\prod_{i,\alpha} \delta\!\le(\z{i}{\alpha}{2}-b_i^{(2)}-\sum_{j}W_{ij}^{(2)}\s{j}{\alpha}{1}\ri)\, ,\nonumber
\end{align}
from the formal expression~\eqref{eq:deeper-layer-formal-expression-first-encounter} for the preactivation distribution conditioned on the activations in the previous layer.
The marginal distribution of the second-layer preactivations can then be obtained by \term{marginalizing over}\index{marginalizing over|seealso{integrating out}} or \term{integrating out}\index{integrating out|seealso{marginalizing over}} the first-layer preactivations as
\be\label{eq:marginalization-first-layer}
p\!\le(z^{(2)}\Big\vert\D\ri)=\int \le[\prod_{i,\alpha}  d\z{i}{\alpha}{1}\ri]\ p\!\le(z^{(2)}\Big\vert z^{(1)}\ri)  p\!\le( z^{(1)}\Big\vert\D\ri)\, .
\ee
To evaluate this expression for the marginal distribution $p\!\le( z^{(2)}\Big\vert\D\ri)$, first we'll discuss how to treat the conditional distribution $p\!\le(z^{(2)}\Big\vert z^{(1)}\ri)$, and then we'll explain how to integrate over the first-layer preactivations $z^{(1)}$ governed by the Gaussian distribution~\eqref{eq:first-layer-distribution-HS-derivation}.

\subsubsection{Second-layer conditional distribution}
The conditional distribution~\eqref{eq:second-conditioned-first} can be evaluated exactly in the same way as we evaluated the first-layer distribution~\eqref{eq:first-layer-formal-expression} conditioned on the inputs, with the simple replacement of the layer indices $\ell$ as $1\to 2$ and exchanging the network input for the first-layer preactivation as $\x{j}{\alpha}\to \s{j}{\alpha}{1}$. Giving you a moment to flip back to \eqref{eq:first-layer-formal-expression} to make these substitutions and then remind yourself of the answer $\eqref{eq:first-layer-distribution-HS-derivation}$, it's easy to see that this evaluation yields
\be\label{eq:second-layer-conditional}
p\!\le(z^{(2)}\Big\vert z^{(1)}\ri)= \frac{1}{\sqrt{\dete{2\pi \widehat{G}^{(2)}}^{n_2}}} \exp\!\le(-\frac{1}{2}\sum_{i=1}^{n_2}\sum_{\alpha_1,\alpha_2\in\D}\SKinv{\alpha_1 \alpha_2}{2}\z{i}{\alpha_1}{2}\z{i}{\alpha_2}{2}\ri)\, ,
\ee
where we have defined the \emph{stochastic} second-layer metric\index{metric!stochastic}\index{metric!second-layer}
\be
\Ti{\widehat{G}}{\alpha_1 \alpha_2}{2}\equiv  \Cb{2}+\CW{2}\frac{1}{n_1}\sum_{j=1}^{n_1}\s{j}{\alpha_1}{1}\s{j}{\alpha_2}{1}\, ,
\ee
with a hat to emphasize that it is a random variable that depends on the stochastic variable $z^{(1)}$ through $\sigma^{(1)} \equiv \sigma\!\le(z^{(1)}\ri)$. Thus, we see that the second-layer conditional distribution \eqref{eq:second-layer-conditional} is a Gaussian whose variance itself is a random variable.
In particular, the stochastic second-layer metric fluctuates around the \emph{mean} second-layer metric\index{metric!mean}
\begin{align}\label{eq:second-layer-metric-mean}
\Ti{G}{\alpha_1 \alpha_2}{2}\equiv\E{\Ti{\widehat{G}}{\alpha_1 \alpha_2}{2}}&=\Cb{2}+\CW{2}\frac{1}{n_1}\sum_{j=1}^{n_1}\E{\s{j}{\alpha_1}{1}\s{j}{\alpha_2}{1}}\,   \\
&=\Cb{2}+\CW{2}\bra \sigma_{\alpha_1} \sigma_{\alpha_2} \ket_{G^{(1)}}\, , \nonumber
\end{align}
where in the last step we recalled the result~\eqref{eq:two-activations-Gauss} for evaluating the two-point correlator of the first-layer activations on the same neuron.

Around this mean, we define the fluctuation of the second-layer metric as
\be\label{eq:second-layer-mean-fluctuation}
\dKi{\alpha_1\alpha_2}{2}\equiv \Ti{\widehat{G}}{\alpha_1 \alpha_2}{2}-\Ti{G}{\alpha_1 \alpha_2}{2}=\CW{2}\frac{1}{n_1}\sum_{j=1}^{n_1}\le(\s{j}{\alpha_1}{1}\s{j}{\alpha_2}{1}-\bra \sigma_{\alpha_1} \sigma_{\alpha_2} \ket_{G^{(1)}}\ri)\, ,
\ee
which by construction has the mean zero when averaged over the first-layer preactivations,
\be\label{eq:meanzero}
\E{\dKi{\alpha_1\alpha_2}{2}}=0 \, .
\ee
The typical size of the fluctuations is given by its two-point correlator. Recalling the expressions we derived for Gaussian integrals~\eqref{eq:two-activations-Gauss} and~\eqref{eq:four-activations-one-neuron-Gauss} of two and four activations on the same neuron and their factorization property on separate neurons~\eqref{eq:four-activations-two-neurons-Gauss}, we obtain
\begin{align}\label{eq:second-layer-metric-fluctuation-two-point-function}
&\E{\dKi{\alpha_1\alpha_2}{2}\dKi{\alpha_3\alpha_4}{2}}\, \\
=&\le(\frac{\CW{2}}{n_1}\ri)^2  \sum_{j,k=1}^{n_1} \E{\le(\s{j}{\alpha_1}{1}\s{j}{\alpha_2}{1}-  \E{ \s{j}{\alpha_1}{1}\s{j}{\alpha_2}{1}}\ri) \le(\s{k}{\alpha_3}{1}\s{k}{\alpha_4}{1}-  \E{ \s{k}{\alpha_3}{1}\s{k}{\alpha_4}{1}}\ri) } \,  \nonumber\\
=&\le(\frac{\CW{2}}{n_1}\ri)^2  \sum_{j=1}^{n_1}  \le\{\E{ \s{j}{\alpha_1}{1}\s{j}{\alpha_2}{1}\s{j}{\alpha_3}{1}\s{j}{\alpha_4}{1} }  - \E{ \s{j}{\alpha_1}{1}\s{j}{\alpha_2}{1}}\E{\s{j}{\alpha_3}{1}\s{j}{\alpha_4}{1} }\ri\} \,  \nonumber\\
=& \frac{1}{n_1} \le(\CW{2}\ri)^2  \le[ \bra\sigma_{\alpha_1} \sigma_{\alpha_2} \sigma_{\alpha_3} \sigma_{\alpha_4}\ket_{G^{(1)}}  - \bra\sigma_{\alpha_1} \sigma_{\alpha_2}\ket_{G^{(1)}}\bra\sigma_{\alpha_3} \sigma_{\alpha_4}\ket_{G^{(1)}} \ri]\,   \notag \\
\equiv&\frac{1}{n_1}V^{(2)}_{(\alpha_1\alpha_2)(\alpha_3\alpha_4)}\,, \nonumber
\end{align}
where at the end we introduced the second-layer \term{four-point vertex}\index{four-point vertex|seealso{data-dependent coupling}} $V^{(2)}_{(\alpha_1\alpha_2)(\alpha_3\alpha_4)}=V\le(x_{\alpha_1},x_{\alpha_2}; x_{\alpha_3},x_{\alpha_4}\ri)$, which depends on four input data points and is symmetric under the exchanges of sample indices $\alpha_1\leftrightarrow\alpha_2$, $\alpha_3\leftrightarrow\alpha_4$, and $(\alpha_1,\alpha_2)\leftrightarrow(\alpha_3,\alpha_4)$. We will understand the significance of this quantity soon in a future equation, ~\eqref{eq:C4_MLP2}.

Here, we also see our first hint of simplification in the wide regime $n_1\gg 1$: since the four-point vertex here is manifestly of order one, we see that the metric fluctuation will be suppressed in that regime.
Essentially, as the number of neurons in the first layer grows, the metric fluctuation becomes more and more Gaussian due to the central limit theorem. In the strict limit of infinite $n_1$, the metric would \emph{self-average}\index{self-averaging}, meaning that the fluctuation would vanish.

Now that we have a feel for the distribution of metric fluctuations, we are only too ready to actually integrate out the first-layer preactivations $z^{(1)}$ and obtain the marginal distribution of the second-layer preactivations $p\!\le(z^{(2)}\Big\vert\D\ri)$. We again provide two derivations, one brute-force and the other clever.

\subsubsection{Wick Wick Wick: combinatorial derivation}
The correlators of the second-layer preactivations can be written nicely in terms of the expectations of the stochastic metric that we just computed. In order to compute the correlators, first we use the fact that the conditional distribution $p\!\le(z^{(2)}\Big\vert z^{(1)}\ri)$ is Gaussian \eqref{eq:second-layer-conditional} to Wick contract the second-layer preactivations $z^{(2)}$, resulting in expressions involving expectations of the stochastic metric $\Ti{\widehat{G}}{\alpha_1\alpha_2}{2}$; we then insert expressions  for the expectations of the stochastic metric obtained above.

With this in mind, the two-point correlator of the second-layer preactivations is given by
\be\label{eq:C2_MLP2}
\E{\z{i_1}{\alpha_1}{2}\z{i_2}{\alpha_2}{2}}=\delta_{i_1i_2}\E{\Ti{\widehat{G}}{\alpha_1 \alpha_2}{2}}=\delta_{i_1i_2}\Ti{G}{\alpha_1 \alpha_2}{2}=\delta_{i_1i_2}\le(\Cb{2}+\CW{2} \bra\sigma_{\alpha_1} \sigma_{\alpha_2}\ket_{G^{(1)}}\ri)\, ,
\ee
where to be clear we first used \eqref{eq:second-layer-conditional} to do the single Wick contraction and then inserted the expression \eqref{eq:second-layer-metric-mean} for the mean of the stochastic metric.

Similarly, the full four-point function can be evaluated as
\begin{align}\label{eq:M4_MLP2}
&\E{\z{i_1}{\alpha_1}{2}\z{i_2}{\alpha_2}{2}\z{i_3}{\alpha_3}{2}\z{i_4}{\alpha_4}{2}}\, \\
=&\delta_{i_1i_2}\delta_{i_3 i_4} \E{\Ti{\widehat{G}}{\alpha_1 \alpha_2}{2}\Ti{\widehat{G}}{\alpha_3 \alpha_4}{2}}+\delta_{i_1i_3}\delta_{i_2 i_4} \E{\Ti{\widehat{G}}{\alpha_1 \alpha_3}{2}\Ti{\widehat{G}}{\alpha_2 \alpha_4}{2}}+\delta_{i_1i_4}\delta_{i_2 i_3} \E{\Ti{\widehat{G}}{\alpha_1 \alpha_4}{2}\Ti{\widehat{G}}{\alpha_2 \alpha_3}{2}}\,, \nonumber\\
=&\delta_{i_1i_2}\delta_{i_3 i_4} \Ti{G}{\alpha_1 \alpha_2}{2}\Ti{G}{\alpha_3 \alpha_4}{2}+\delta_{i_1i_3}\delta_{i_2 i_4} \Ti{G}{\alpha_1 \alpha_3}{2}\Ti{G}{\alpha_2 \alpha_4}{2}+\delta_{i_1i_4}\delta_{i_2 i_3} \Ti{G}{\alpha_1 \alpha_4}{2}\Ti{G}{\alpha_2 \alpha_3}{2}\, \nonumber\\
&+\frac{1}{n_1}\le[\delta_{i_1i_2}\delta_{i_3 i_4}V^{(2)}_{(\alpha_1\alpha_2)(\alpha_3\alpha_4)}+\delta_{i_1i_3}\delta_{i_2 i_4}V^{(2)}_{(\alpha_1\alpha_3)(\alpha_2\alpha_4)}+\delta_{i_1i_4}\delta_{i_2 i_3}V^{(2)}_{(\alpha_1\alpha_4)(\alpha_2\alpha_3)} \ri]\, ,\nonumber
\end{align}
where in the first line we made three Wick contractions of the four second-layer preactivations $z^{(2)}$'s using the Gaussian distribution~\eqref{eq:second-layer-conditional}, and then in the second line we recalled~\eqref{eq:meanzero} and~\eqref{eq:second-layer-metric-fluctuation-two-point-function} for the expectations of the stochastic metric $ \Ti{\widehat{G}}{\alpha_1 \alpha_2}{2}=\Ti{G}{\alpha_1 \alpha_2}{2}+\dKi{\alpha_1\alpha_2}{2}$ over the first-layer preactivations $z^{(1)}$.
This means that the \emph{connected} four-point correlator -- recall~\eqref{eq:C4} -- after subtracting the contributions from the two-point correlators of the second-layer preactivations is given by
\begin{align}\label{eq:C4_MLP2}
&\E{\z{i_1}{\alpha_1}{2}\z{i_2}{\alpha_2}{2}\z{i_3}{\alpha_3}{2}\z{i_4}{\alpha_4}{2}}\Big\vert_{\text{connected}}\, \\
=&\frac{1}{n_1}\le[\delta_{i_1i_2}\delta_{i_3 i_4}V^{(2)}_{(\alpha_1\alpha_2)(\alpha_3\alpha_4)}+\delta_{i_1i_3}\delta_{i_2 i_4}V^{(2)}_{(\alpha_1\alpha_3)(\alpha_2\alpha_4)}+\delta_{i_1i_4}\delta_{i_2 i_3}V^{(2)}_{(\alpha_1\alpha_4)(\alpha_2\alpha_3)} \ri]\, . \nonumber
\end{align}
Here we see the true importance of the four-point vertex we introduced in~\eqref{eq:second-layer-metric-fluctuation-two-point-function}; it gives the connected second-layer four-point correlator and controls the near-Gaussianity of the second-layer preactivation distribution. Thus, we see that this connected correlator is suppressed in the wide regime of $n_1\gg1$, suggesting that the preactivation distribution will become more and more Gaussian 
as the network gets wider and wider. 
Given this, we see that the second-layer preactivation distribution $p\!\le(z^{(2)}\Big\vert\D\ri)$ is in general \emph{non-Gaussian}
but also simplifies significantly in the large-$n_1$ regime, becoming Gaussian in the strict $n_1=\infty$ limit and with the four-point vertex $V^{(2)}_{(\alpha_1\alpha_3)(\alpha_2\alpha_4)}$ measuring the leading deviation from Gaussianity.

To complete our combinatorial derivation, we need to find an action that generates correlations~\eqref{eq:C2_MLP2} and~\eqref{eq:C4_MLP2}.
As we know, a quadratic action cannot generate non-Gaussian distributions with nontrivial connected four-point correlators, so we need a different action that's appropriate for a \terminate{nearly-Gaussian distribution}. Intuition from single-variable non-Gaussian integrals in~\S\ref{sec:not-Gauss} suggests that we could perhaps generate the requisite correlations by including a quartic term in the action.

With that in mind, let's start with a quartic action for an $(n \ND)$-dimensional random variable $z$
\begin{align}\label{eq:generic-quartic}
S\!\le[z\ri]&=\frac{1}{2}\sum_{\alpha_1,\alpha_{2}\in\D}g^{\alpha_1\alpha_2}\sum_{i=1}^{n} z_{i;\alpha_1}z_{i;\alpha_2}\, \nonumber\\
&-\frac{1}{8}\sum_{\alpha_1,\ldots,\alpha_4\in\D}v^{(\alpha_1\alpha_2)(\alpha_3\alpha_4)}\sum_{i_1,i_2=1}^{n} z_{i_1;\alpha_1}z_{i_1;\alpha_2} \, z_{i_2;\alpha_3}z_{i_2;\alpha_4} \, ,
\end{align}
with undetermined couplings $g$ and $v$.
We will treat the quartic coupling $v$ perturbatively, an assumption that we will justify later by relating the quartic coupling $v$ to the $1/n_1$-suppressed connected four-point correlator.
Note that by construction the quartic coupling $v^{(\alpha_1\alpha_2)(\alpha_3\alpha_4)}$ has the same symmetric structure as the four-point vertex $V^{(2)}_{(\alpha_1\alpha_2)(\alpha_3\alpha_4)}$ with respect to the sample indices.\footnote{The conventional factor of $1/8$ in \eqref{eq:generic-quartic} is to account for this symmetry.}
Using this action, we can compute to the first order in $v$ the two-point and four-point correlators. Then, by matching with the expressions~\eqref{eq:C2_MLP2} and~\eqref{eq:C4_MLP2} for these quantities, we'll learn how to adjust the couplings $g$ and $v$ to reproduce the right statistics of second-layer preactivations in the wide regime.

Before proceeding further, it is convenient to introduce some notation. 
In~\eqref{eq:gaussian-integration-metric-g}, we defined $\bra F\!\le(z_{\alpha_1},\ldots, z_{\alpha_m}\ri) \ket_{g}$
for the average of an arbitrary function $F$ over a Gaussian distribution with variance $g$, where preactivation variables $z_{\alpha}$ have sample indices \emph{only}. 
In addition, we here define
\begin{align}\label{eq:many-neuron-gaussian-notation}
&\bra\!\bra F\!\le(z_{i_1;\alpha_1},\ldots, z_{i_{m};\alpha_m}\ri)\ket\!\ket_{g}\, \\
\equiv& \int \le[\prod_{i=1}^{n}\frac{\prod_{\alpha\in\D} dz_{i;\alpha}}{\sqrt{\dete{2\pi g}}}\ri]\exp\!\le(-\frac{1}{2}\sum_{j=1}^n\sum_{\beta_1,\beta_2\in\D}g^{\beta_1 \beta_2}z_{j;\beta_1}z_{j;\beta_2}\ri) F\!\le(z_{i_1;\alpha_1},\ldots, z_{i_m;\alpha_m}\ri)\,, \nonumber
\end{align}
which now includes neural indices. As we saw while working through~\eqref{eq:two-activations-Gauss} and~\eqref{eq:four-activations-two-neurons-Gauss}, this type of average factorizes into integrals of the form~\eqref{eq:gaussian-integration-metric-g} for each neuron.

With this notation in hand, the expectation of an arbitrary function $F\!\le(z_{i_1;\alpha_1},\ldots, z_{i_{m};\alpha_m}\ri)$ against a distribution with the quartic action \eqref{eq:generic-quartic} can be rewritten in terms of Gaussian expectations, enabling the perturbative expansion in the coupling $v$ as
\begin{align}\label{eq:arbitrary-function-quartic-expectation}
&\E{F\!\le(z_{i_1;\alpha_1},\ldots, z_{i_{m};\alpha_m}\ri)}\, \\
=&\frac{\int \le[\prod_{i,\alpha} d z_{i;\alpha}\ri] e^{-\ac\le(z\ri)}F\!\le(z_{i_1;\alpha_1},\ldots, z_{i_{m};\alpha_m}\ri)}{\int \le[\prod_{i,\alpha} d z_{i;\alpha}\ri] e^{-\ac\le(z\ri)}}\,  \nonumber \\
=&\frac{\bra\!\!\bra \exp\!\le\{\frac{1}{8}\sum_{\beta_1,\ldots,\beta_4\in\D}v^{(\beta_1\beta_2)(\beta_3\beta_4)}\sum_{j_1,j_2=1}^{n} z_{j_1;\beta_1}z_{j_1;\beta_2}\, z_{j_2;\beta_3}z_{j_2;\beta_4} \ri\}\,F\!\le(z_{i_1;\alpha_1},\ldots, z_{i_{m};\alpha_m}\ri)\ket\!\!\ket_{g}}{\bra\!\!\bra \exp\!\le\{\frac{1}{8}\sum_{\beta_1,\ldots,\beta_4\in\D}v^{(\beta_1\beta_2)(\beta_3\beta_4)}\sum_{j_1,j_2=1}^{n} z_{j_1;\beta_1}z_{j_1;\beta_2}\, z_{j_2;\beta_3}z_{j_2;\beta_4} \ri\}\ket\!\!\ket_{g}}\,  \nonumber\\
=&\bra\!\bra F\!\le(z_{i_1;\alpha_1},\ldots, z_{i_{m};\alpha_m}\ri)\ket\!\ket_{g}\, \nonumber\\
&+\frac{1}{8}\sum_{\beta_1,\ldots,\beta_4\in\D}v^{(\beta_1\beta_2)(\beta_3\beta_4)}\sum_{j_1,j_2=1}^{n}\Big[\bra\!\bra z_{j_1;\beta_1}z_{j_1;\beta_2}\, z_{j_2;\beta_3}z_{j_2;\beta_4} F\!\le(z_{i_1;\alpha_1},\ldots, z_{i_{m};\alpha_m}\ri)  \ket\!\ket_{g}\, \nonumber\\
&\quad \quad \quad \quad \quad \quad \quad \quad \quad \quad \quad \quad \quad \ -\bra\!\bra z_{j_1;\beta_1}z_{j_1;\beta_2}\, z_{j_2;\beta_3}z_{j_2;\beta_4} \ket\!\ket_{g}\bra\!\bra F\!\le(z_{i_1;\alpha_1},\ldots, z_{i_{m};\alpha_m}\ri)\ket\!\ket_{g}\Big]\, \nonumber\\
&+\o{v^2}\, ,\nonumber
\end{align}
where in the first line we used the definition of the expectation, in the second line we rewrote the numerator and denominator using the notation~\eqref{eq:many-neuron-gaussian-notation} that we just introduced, and in the third line we expanded the exponential in the coupling $v$, both in the denominator and numerator. %
In short, this tells us how to perturbatively express an expectation against the full distribution with the quartic action~\eqref{eq:generic-quartic}  in terms of the leading Gaussian expectation and perturbative corrections; these perturbative contributions nonetheless involve only Gaussian expectations and hence are easy to evaluate.

With this in mind, let's consider some particular choices for $F$. 
Starting with the two-point correlator, we get
\begin{align}\label{eq:second-moment-from-action}
&\E{z_{i_1;\alpha_1} z_{i_2;\alpha_2}}\, \\
=&\delta_{i_1 i_2}\le[g_{\alpha_1\alpha_2}+\frac{1}{2}\sum_{\beta_1,\ldots,\beta_4\in\D}v^{(\beta_1\beta_2)(\beta_3\beta_4)}\le(n g_{\alpha_1\beta_1}g_{\alpha_2\beta_2}g_{\beta_3\beta_4}+2g_{\alpha_1\beta_1}g_{\alpha_2\beta_3}g_{\beta_2\beta_4}\ri)\ri]+\o{v^2}\, .\nonumber
\end{align}
Here the variance $g_{\alpha_1\alpha_2}$ is the inverse of the quadratic coupling, with $\sum_{\beta }g_{\alpha_1\beta}g^{\beta\alpha_2}=\delta_{\alpha_1}^{\ \alpha_2}$.
Similarly, we find that the connected four-point correlator evaluates to
\begin{align}\label{eq:fourth-cumulant-from-action}
&\E{z_{i_1;\alpha_1} z_{i_2;\alpha_2}z_{i_3;\alpha_3} z_{i_4;\alpha_4}}\Big\vert_{\text{connected}}\, \\
\equiv&\E{z_{i_1;\alpha_1} z_{i_2;\alpha_2}z_{i_3;\alpha_3} z_{i_4;\alpha_4}}-\E{z_{i_1;\alpha_1} z_{i_2;\alpha_2}}\E{z_{i_3;\alpha_3} z_{i_4;\alpha_4}}\, \nonumber\\
&-\E{z_{i_1;\alpha_1} z_{i_3;\alpha_3}}\E{z_{i_2;\alpha_2} z_{i_4;\alpha_4}}-\E{z_{i_1;\alpha_1} z_{i_4;\alpha_4}}\E{z_{i_2;\alpha_2} z_{i_3;\alpha_3}}\,  \nonumber\\
=&\delta_{i_1 i_2}\delta_{i_3 i_4}\sum_{\beta_1,\ldots,\beta_4\in\D}v^{(\beta_1\beta_2)(\beta_3\beta_4)}g_{\alpha_1\beta_1}g_{\alpha_2\beta_2}g_{\alpha_3\beta_3}g_{\alpha_4\beta_4}\, \nonumber\\
&+\delta_{i_1 i_3}\delta_{i_2 i_4}\sum_{\beta_1,\ldots,\beta_4\in\D}v^{(\beta_1\beta_3)(\beta_2\beta_4)}g_{\alpha_1\beta_1}g_{\alpha_3\beta_3}g_{\alpha_2\beta_2}g_{\alpha_4\beta_4}\, \nonumber\\
&+\delta_{i_1 i_4}\delta_{i_2 i_3}\sum_{\beta_1,\ldots,\beta_4\in\D}v^{(\beta_1\beta_4)(\beta_2\beta_3)}g_{\alpha_1\beta_1}g_{\alpha_4\beta_4}g_{\alpha_2\beta_2}g_{\alpha_3\beta_3}+\o{v^2}\, .\nonumber
\end{align}
Comparing these expressions,~\eqref{eq:second-moment-from-action} and \eqref{eq:fourth-cumulant-from-action}, with correlators in the second layer,~\eqref{eq:C2_MLP2} and~\eqref{eq:C4_MLP2}, it's easy to see that setting the couplings as
\begin{align}
g^{\alpha_1\alpha_2}&=\TI{G}{\alpha_1\alpha_2}{2}+\o{\frac{1}{n_1}}\, ,\label{eq:second-layer-quadratic-coupling}\\
v^{(\alpha_1\alpha_2)(\alpha_3\alpha_4)}&=\frac{1}{n_1}V_{(2)}^{(\alpha_1\alpha_2)(\alpha_3\alpha_4)}+\o{\frac{1}{n_1^2}}\, ,\label{eq:second-layer-quartic-coupling}
\end{align}
reproduces the second-layer preactivation correlators to the leading order in $1/n_1$, with the marginal distribution
\be
p\!\le(z^{(2)}\Big\vert\D\ri)=\frac{1}{Z} e^{-\EFT{2}} \, 
\ee
and quartic action \eqref{eq:generic-quartic}.
Here for convenience we have defined a version of the four-point vertex with indices \emph{raised} by the inverse of the second-layer mean metric
\be
V_{(2)}^{(\alpha_1\alpha_2)(\alpha_3\alpha_4)}\equiv\sum_{\beta_1,\ldots,\beta_4} \TI{G}{\alpha_1\beta_1}{2}\TI{G}{\alpha_2\beta_2}{2}\TI{G}{\alpha_3\beta_3}{2}\TI{G}{\alpha_4\beta_4}{2}V^{(2)}_{(\beta_1\beta_2)(\beta_3\beta_4)} \, .
\ee
Note that the quartic coupling $v$ is $\o{1/n_1}$, justifying our earlier perturbative treatment of the coupling for wide networks.
Note also that these couplings -- the inverse metric $\TI{G}{\alpha_1\alpha_2}{2}$ and the quartic coupling $V_{(2)}^{(\alpha_1\alpha_2)(\alpha_3\alpha_4)}$ -- are input-dependent.
In particular, the effective strength of interaction between neurons is set by the particular set of inputs to the network. 

This completes our first combinatorial derivation of the second-layer preactivation distribution.

\subsubsection{Schwinger-Dyson this way: algebraic derivation}
Here is a neat way to derive the action for the second-layer preactivation distribution. Plugging the conditional distribution~\eqref{eq:second-layer-conditional} into the marginalization equation~\eqref{eq:marginalization-first-layer}, the second-layer marginal distribution becomes
\be\label{eq:stochastic-marginalization-second}
p\!\le(z^{(2)}\Big\vert\D\ri)=\int \le[\prod_{i,\alpha}  d\z{i}{\alpha}{1}\ri] p\!\le(z^{(1)}\Big\vert\D\ri)\frac{ \exp\!\le(-\frac{1}{2}\sum_{j=1}^{n_2}\sum_{\alpha_1,\alpha_2\in\D}\SKinv{\alpha_1 \alpha_2}{2}\z{j}{\alpha_1}{2}\z{j}{\alpha_2}{2}\ri)}{\sqrt{\dete{2\pi \widehat{G}^{(2)}}^{n_2}}}\, .\\
\ee
We saw that the stochastic metric has a natural decomposition into mean and fluctuating parts as\index{tensor decomposition!metric mean and fluctuation}
\be
\Ti{\widehat{G}}{\alpha_1 \alpha_2}{2}=\Ti{G}{\alpha_1 \alpha_2}{2}+\dKi{\alpha_1\alpha_2}{2}\, .
\ee
Inverting this matrix to the second order in the fluctuation around the mean, we get the inverse stochastic metric\footnote{This together with the defining equation for the metric fluctuation \eqref{eq:second-layer-mean-fluctuation} are sometimes called the \terminate{Schwinger-Dyson equations} \cite{DysonEq,Schwinger452} from which this subsubsection takes its title.}
\begin{align}\label{eq:stochastic-metric-inversion}
\SKinv{\alpha_1\alpha_2}{2}=&\Kinv{\alpha_1\alpha_2}{2}-\sum_{\beta_1,\beta_{2}\in\D}\Kinv{\alpha_1\beta_1}{2}\dKi{\beta_1\beta_2}{2}\Kinv{\beta_2\alpha_2}{2}\, \\
&+\sum_{\beta_1,\ldots,\beta_{4}\in\D}\Kinv{\alpha_1\beta_1}{2}\dKi{\beta_1\beta_2}{2}\Kinv{\beta_2\beta_3}{2} \dKi{\beta_{3}\beta_{4}}{2} \Kinv{\beta_{4}\alpha_2}{2}+\o{\Delta^3}\, .\nonumber
\end{align}
Putting this into the exponential that appears in the integrand of the marginal distribution~\eqref{eq:stochastic-marginalization-second} and Taylor-expanding in the fluctuation $\dKi{\alpha_1\alpha_2}{2}$, we find
\begin{align}\label{eq:second-layer-stochastic-exponential}
&\exp\!\le(-\frac{1}{2}\sum_{j=1}^{n_2}\sum_{\alpha_1,\alpha_2\in\D}\SKinv{\alpha_1 \alpha_2}{2}\z{j}{\alpha_1}{2}\z{j}{\alpha_2}{2}\ri)\, \\
=&\exp\!\le(-\frac{1}{2}\sum_{j=1}^{n_2}\sum_{\alpha_1,\alpha_2\in\D}\Kinv{\alpha_1 \alpha_2}{2}\z{j}{\alpha_1}{2}\z{j}{\alpha_2}{2}\ri)\, \nonumber\\
&\times\Bigg\{ 1+\frac{1}{2}\sum_{i=1}^{n_2}\sum_{\alpha_1,\alpha_{2}\in\D} \le(\sum_{\beta_1,\beta_{2}\in\D}\Kinv{\alpha_1\beta_1}{2}\dKi{\beta_1\beta_2}{2}\Kinv{\beta_2\alpha_2}{2}\ri)\z{i}{\alpha_1}{2}\z{i}{\alpha_2}{2}\, \nonumber\\
&\quad -\frac{1}{2}\sum_{i=1}^{n_2}\sum_{\alpha_1,\alpha_{2}\in\D} \le(\sum_{\beta_1,\ldots,\beta_{4}\in\D}\Kinv{\alpha_1\beta_1}{2}\dKi{\beta_1\beta_2}{2}\Kinv{\beta_2\beta_3}{2} \dKi{\beta_{3}\beta_{4}}{2} \Kinv{\beta_{4}\alpha_2}{2}\ri)\z{i}{\alpha_1}{2}\z{i}{\alpha_2}{2}\, \nonumber\\
&\quad + \frac{1}{2!}\le(\frac{1}{2}\ri)^2\sum_{i_1,i_2=1}^{n_2}\sum_{\alpha_1,\ldots,\beta_{4}\in\D}\Kinv{\alpha_1\beta_1}{2}\cdots\Kinv{\alpha_4\beta_4}{2} \dKi{\beta_1\beta_2}{2} \dKi{\beta_3\beta_4}{2} \z{i_1}{\alpha_1}{2}\z{i_1}{\alpha_2}{2}\z{i_2}{\alpha_{3}}{2}\z{i_2}{\alpha_{4}}{2}+\o{\Delta^3}\Bigg\}\, .\nonumber
\end{align}
Using this expression, the determinant in the denominator becomes
\begin{align}\label{eq:second-layer-stochastic-determinant}
&\sqrt{\dete{2\pi \widehat{G}^{(2)}}^{n_2}}=\int \le[\prod_{i,\alpha}  d\z{i}{\alpha}{2}\ri]\exp\!\le(-\frac{1}{2}\sum_{j=1}^{n_2}\sum_{\alpha_1,\alpha_2\in\D}\SKinv{\alpha_1 \alpha_2}{2}\z{j}{\alpha_1}{2}\z{j}{\alpha_2}{2}\ri)\, \\
=&\sqrt{\dete{2\pi G^{(2)}}^{n_2}}\Bigg[1+\frac{n_2}{2}\sum_{\beta_1,\beta_2\in\D}\dKi{\beta_1\beta_2}{2}\Kinv{\beta_1\beta_2}{2}\, \nonumber\\
&\quad+\sum_{\beta_1,\ldots,\beta_4\in\D}\dKi{\beta_1\beta_2}{2}\dKi{\beta_3\beta_4}{2}\le(\frac{n_2^2}{8}\Kinv{\beta_1\beta_2}{2}\Kinv{\beta_3\beta_4}{2}-\frac{n_2}{4}\Kinv{\beta_1\beta_3}{2}\Kinv{\beta_2\beta_4}{2}\ri)+\o{\Delta^3}\Bigg]\, ,\nonumber
\end{align}
where on the first line we re-expressed the determinant as a Gaussian integral, and on the subsequent line we plugged in \eqref{eq:second-layer-stochastic-exponential} and integrated over the second-layer preactivations $z^{(2)}$.

Next, plugging these two expressions~\eqref{eq:second-layer-stochastic-exponential} and~\eqref{eq:second-layer-stochastic-determinant} back into our expression for the second-layer distribution~\eqref{eq:stochastic-marginalization-second}, we can now integrate out the first-layer preactivations, giving
\begin{align}
p\!\le(z^{(2)}\Big\vert\D\ri)=&\frac{1}{\sqrt{\dete{2\pi G^{(2)}}^{n_2}}}\exp\!\le(-\frac{1}{2}\sum_{j=1}^{n_2}\sum_{\alpha_1,\alpha_2\in\D}\Kinv{\alpha_1 \alpha_2}{2}\z{j}{\alpha_1}{2}\z{j}{\alpha_2}{2}\ri)\, \\
&\times\Bigg\{ \le[1+\o{\frac{1}{n_1}}\ri]+\sum_{i=1}^{n_2}\sum_{\alpha_1,\alpha_{2}\in\D}\le[\o{\frac{1}{n_1}}\ri]\z{i_1}{\alpha_1}{2}\z{i_1}{\alpha_2}{2}\,\nonumber\\
&\quad +\frac{1}{8n_1}\sum_{i_1,i_2=1}^{n_2}\sum_{\alpha_1,\ldots,\alpha_{4}\in\D}\TI{V}{(\alpha_1\alpha_2)(\alpha_3\alpha_4)}{2} \z{i_1}{\alpha_1}{2}\z{i_1}{\alpha_2}{2}\z{i_2}{\alpha_{3}}{2}\z{i_2}{\alpha_{4}}{2}\Bigg\}+\o{\frac{1}{n_1^2}}\,, \nonumber
\end{align}
where we have used the fact that expectations of the metric fluctuation are given by $\E{\dKi{\beta_1\beta_2}{2}}=0$ and $\E{\dKi{\beta_1\beta_2}{2}\dKi{\beta_3\beta_4}{2}}=\frac{1}{n_1}\Ti{V}{(\beta_1\beta_2)(\beta_3\beta_4)}{2}$.\footnote{We tacitly assumed that the expectation of $\widehat{\Delta G}^{m\geq3}$ are of order $\o{1/n_1^2}$ or greater.\label{foot:second-layer-hierarchy}
For instance, you can follow exactly the same steps as in \eqref{eq:second-layer-metric-fluctuation-two-point-function} and compute
\begin{align}
&\E{\dKi{\beta_1\beta_2}{2}\dKi{\beta_3\beta_4}{2}\dKi{\beta_5\beta_6}{2}}\, \\
=&\frac{1}{n_1^2}\le(\CW{2}\ri)^3\Big[\bra\sigma_{\alpha_1} \sigma_{\alpha_2} \sigma_{\alpha_3} \sigma_{\alpha_4}\sigma_{\alpha_5} \sigma_{\alpha_6}\ket_{G^{(1)}}  - \bra\sigma_{\alpha_1} \sigma_{\alpha_2}\ket_{G^{(1)}}\bra\sigma_{\alpha_3} \sigma_{\alpha_4}\sigma_{\alpha_5} \sigma_{\alpha_6}\ket_{G^{(1)}}\, \notag\\
&\qquad\qquad\qquad- \bra\sigma_{\alpha_3} \sigma_{\alpha_4}\ket_{G^{(1)}}\bra\sigma_{\alpha_5} \sigma_{\alpha_6}\sigma_{\alpha_1} \sigma_{\alpha_2}\ket_{G^{(1)}}- \bra\sigma_{\alpha_5} \sigma_{\alpha_6}\ket_{G^{(1)}}\bra\sigma_{\alpha_1} \sigma_{\alpha_2}\sigma_{\alpha_3} \sigma_{\alpha_4}\ket_{G^{(1)}}\, \notag\\
&\qquad\qquad\qquad+2\bra\sigma_{\alpha_1} \sigma_{\alpha_2}\ket_{G^{(1)}}\bra\sigma_{\alpha_3} \sigma_{\alpha_4}\ket_{G^{(1)}}\bra\sigma_{\alpha_5} \sigma_{\alpha_6}\ket_{G^{(1)}}\Big]\, .\notag
\end{align}
Just as in the middle step of \eqref{eq:second-layer-metric-fluctuation-two-point-function}, here again you've likely noticed that nonzero contributions arise only when all the neural indices coincide. You can further use that same insight to show that $\E{\le(\widehat{\Delta G}^{(2)}\ri)^{m}}=\o{1/n^{m-1}_1}$.
}
Taking the logarithm to isolate the action and absorbing the irrelevant constant terms into the partition function, we arrive at the correct expression for the second-layer quartic action to leading order in the first layer width
\begin{align}\label{eq:second-layer-quartic-action-in-SD-subsubsection}
\ac\!\le(z\ri)=&\frac{1}{2}\sum_{\alpha_1,\alpha_{2}\in\D}\le[G_{(2)}^{\alpha_1\alpha_2}+\o{\frac{1}{n_1}}\ri]\sum_{i=1}^{n_2} z_{i;\alpha_1}z_{i;\alpha_2}\, \\
&-\frac{1}{8}\sum_{\alpha_1,\ldots,\alpha_4\in\D}\frac{1}{n_1}V_{(2)}^{(\alpha_1\alpha_2)(\alpha_3\alpha_4)}\sum_{i_1,i_2=1}^{n_2} z_{i_1;\alpha_1}z_{i_1;\alpha_2}z_{i_2;\alpha_3}z_{i_2;\alpha_4}+\o{\frac{1}{n_1^2}}\, .\nonumber
\end{align}

Here, a prudent reader might wonder about our dropping of the $1/n_1$ correction to the quadratic coupling, while keeping the quartic coupling despite being of the same order. The main reason for this is that such a correction is a \emph{subleading} contribution to the two-point correlator, while the quartic coupling gives the \emph{leading} contribution to the connected 
four-point correlator. Indeed, we shall encounter various observables whose leading %
contributions stem solely from the nontrivial neuron-neuron interaction induced by the quartic coupling. By contrast, the correction to the quadratic coupling at finite-width is just a small quantitative effect. Nevertheless, we will compute this subleading correction in \S\ref{sec:loop-correction} for completeness.\footnote{It will also turn out (\S\ref{sec:signal_prop_finite_width}) that by fine-tuning the initialization hyperparameters such subleading corrections are suppressed with depth in comparison to nearly-Gaussian corrections, so in a sense this subleading correction to the quadratic coupling can be doubly ignored.}

\subsubsection{Nearly-Gaussian action in action}\index{nearly-Gaussian distribution!action}
Having completed the two derivations, before moving on to the next section, let's use this opportunity to get a bit more of a feel for how to compute with a \terminate{nearly-Gaussian distribution}. Paralleling what we did with the Gaussian action in the last section, let's evaluate the expectation of two activations on the same neuron
and four activations,
with all four on the same neuron or pairs on separate neurons.
The resulting expressions will enable us to obtain the distributions of the preactivations in deeper layers.

In the following, we are just applying the formula~\eqref{eq:arbitrary-function-quartic-expectation} for the expectation of a general function.
These expressions will be valid for any layer $\ell > 1$.
First, for two activations on the same neuron, we find %
\begin{align}\label{eq:two-activations-deep}
&\E{\sigma\!\le(z_{i_1;\alpha_1}\ri)\sigma\!\le(z_{i_1;\alpha_2}\ri)}\, \\
=&\bra \sigma_{\alpha_1}\sigma_{\alpha_2}\ket_{g}+\frac{1}{8}\sum_{\beta_1,\ldots,\beta_4\in\D}v^{(\beta_1\beta_2)(\beta_3\beta_4)}\, \nonumber\\
&\quad \quad \quad \quad \quad \times\Big[\bra \sigma_{\alpha_1}\sigma_{\alpha_2} \le(z_{\beta_1} z_{\beta_2}-g_{\beta_1\beta_2}\ri)\le(z_{\beta_3} z_{\beta_4}-g_{\beta_3\beta_4}\ri)\ket_{g}\, \nonumber\\
&\quad \quad \quad \quad \quad \quad +2n \bra \sigma_{\alpha_1}\sigma_{\alpha_2} \le(z_{\beta_1} z_{\beta_2}-g_{\beta_1\beta_2}\ri)\ket_{g} g_{\beta_3\beta_4}-2\bra \sigma_{\alpha_1}\sigma_{\alpha_2}\ket_{g}g_{\beta_1\beta_3}g_{\beta_2\beta_4}\Big]+\o{v^2}\, , \nonumber
\end{align}
where we assume the reader is by now familiar enough with Gaussian integrals and factorization into separate neurons so as not to include the middle steps. This result
highlights that the addition of the quartic coupling $v$ has a nontrivial effect even on the two-point correlator of same-neuron activations. 
We can similarly compute the expectation of four activations on the same neuron, but we'll need only the leading Gaussian contribution, namely
\begin{align}\label{eq:four-activations-same-deep}
&\E{\sigma\!\le(z_{i_1;\alpha_1}\ri)\sigma\!\le(z_{i_1;\alpha_2}\ri)\sigma\!\le(z_{i_1;\alpha_3}\ri)\sigma\!\le(z_{i_1;\alpha_4}\ri)}-\E{\sigma\!\le(z_{i_1;\alpha_1}\ri)\sigma\!\le(z_{i_1;\alpha_2}\ri)}\E{\sigma\!\le(z_{i_1;\alpha_3}\ri)\sigma\!\le(z_{i_1;\alpha_4}\ri)}\, \\
=&\bra \sigma_{\alpha_1}\sigma_{\alpha_2}\sigma_{\alpha_3}\sigma_{\alpha_4}\ket_{g}-\bra \sigma_{\alpha_1}\sigma_{\alpha_2}\ket_{g}\bra\sigma_{\alpha_3}\sigma_{\alpha_4}\ket_{g}+\o{v}\, ,\nonumber
\end{align}
where we subtracted off the contribution from the two-point correlators as that's what'll appear in the next section.
Finally, the similar expectation of four activations on two different pairs of neurons $i_1 \neq i_2$ can be evaluated by the application of the formula~\eqref{eq:arbitrary-function-quartic-expectation} and neuron factorizations in Gaussian expectations, yielding 
\begin{align}\label{eq:four-activations-different-deep-connected}
&\E{\sigma\!\le(z_{i_1;\alpha_1}\ri)\sigma\!\le(z_{i_1;\alpha_2}\ri)\sigma\!\le(z_{i_2;\alpha_3}\ri)\sigma\!\le(z_{i_2;\alpha_4}\ri)}-\E{\sigma\!\le(z_{i_1;\alpha_1}\ri)\sigma\!\le(z_{i_1;\alpha_2}\ri)}\E{\sigma\!\le(z_{i_2;\alpha_3}\ri)\sigma\!\le(z_{i_2;\alpha_4}\ri)}\, \nonumber\\
=&\frac{1}{8}\sum_{\beta_1,\ldots,\beta_4\in\D}v^{(\beta_1\beta_2)(\beta_3\beta_4)}\sum_{j_1,j_2=1}^{n}\, \\
&\times\Big[\bra\!\bra z_{j_1;\beta_1}z_{j_1;\beta_2}\, z_{j_2;\beta_3}z_{j_2;\beta_4} \sigma_{i_1;\alpha_1} \sigma_{i_1;\alpha_2} \sigma_{i_2;\alpha_3}\sigma_{i_2;\alpha_4}   \ket\!\ket_{g}\, \nonumber\\
&\quad -\bra\!\bra z_{j_1;\beta_1}z_{j_1;\beta_2}\, z_{j_2;\beta_3}z_{j_2;\beta_4} \sigma_{i_1;\alpha_1} \sigma_{i_1;\alpha_2} \ket\!\ket_{g}\bra\!\bra\sigma_{i_2;\alpha_3}\sigma_{i_2;\alpha_4}   \ket\!\ket_{g}\, \nonumber\\
&\quad -\bra\!\bra z_{j_1;\beta_1}z_{j_1;\beta_2}\, z_{j_2;\beta_3}z_{j_2;\beta_4} \sigma_{i_2;\alpha_3} \sigma_{i_2;\alpha_4} \ket\!\ket_{g}\bra\!\bra\sigma_{i_1;\alpha_1}\sigma_{i_1;\alpha_2}   \ket\!\ket_{g}\, \nonumber\\
&\quad +\bra\!\bra z_{j_1;\beta_1}z_{j_1;\beta_2}\, z_{j_2;\beta_3}z_{j_2;\beta_4} \ket\!\ket_{g}\bra\!\bra\sigma_{i_1;\alpha_1} \sigma_{i_1;\alpha_2} \ket\!\ket_{g}\bra\!\bra\sigma_{i_2;\alpha_3}\sigma_{i_2;\alpha_4}   \ket\!\ket_{g}\Big]\, \nonumber\\
=&\frac{1}{4}\sum_{\beta_1,\ldots,\beta_4\in\D}v^{(\beta_1\beta_2)(\beta_3\beta_4)}\bra \sigma_{\alpha_1}\sigma_{\alpha_2} \le(z_{\beta_1} z_{\beta_2}-g_{\beta_1\beta_2}\ri)\ket_{g}\bra \sigma_{\alpha_3}\sigma_{\alpha_4} \le(z_{\beta_3} z_{\beta_4}-g_{\beta_3\beta_4}\ri)\ket_{g}+\o{v^2}\, ,\nonumber 
\end{align}
where we get nonzero contributions only when $j_1=i_1$ and $j_2=i_2$ or when $j_1=i_2$ and $j_2=i_1$.
This shows that pairs of activations can only correlate with the addition of the quartic coupling to the action, hinting at the role of finite width for features learning.
More generally, consider functions $\mathcal{F}\!\le(z_{i_1;\A_1}\ri)$
and $\mathcal{G}\!\le(z_{i_2; \A_2}\ri)$
of preactivations that depend on subsamples $\A_1$ and $\A_2\subset\D$, respectively, where with a slight abuse of notation we put the set dependences into the subscripts. For distinct neurons $i_1\ne i_2$, the calculation identical to the one just above shows that their covariance is given by
\begin{align}\label{eq:general-covariance}
&\text{Cov}\Big[\mathcal{F}\!\le(z_{i_1; \A_1}\ri)\!, \, \mathcal{G}\!\le(z_{i_2; \A_2}\ri)\Big] \\
\equiv&\mathbb{E}\Big[ \mathcal{F}\!\le(z_{i_1;\A_1}\ri)\mathcal{G}\!\le(z_{i_2;\A_2}\ri)\Big]-\mathbb{E}\Big[\mathcal{F}\!\le(z_{i_1;\A_1}\ri)\Big]\, \mathbb{E}\Big[\mathcal{G}\!\le(z_{i_2;\A_2}\ri)\Big]\, \notag \\
=&\frac{1}{4}\sum_{\beta_1,\ldots,\beta_4\in\D}v^{(\beta_1\beta_2)(\beta_3\beta_4)}\Big\langle \le(z_{\beta_1} z_{\beta_2}-g_{\beta_1\beta_2}\ri)  \mathcal{F}\!\le(z_{\A_1}\ri) \Big\rangle_{g}\Big\langle  \le(z_{\beta_3} z_{\beta_4}-g_{\beta_3\beta_4}\ri)\mathcal{G}\!\le(z_{\A_2}\ri)\Big\rangle_{g}+\o{v^2}\, .\nonumber 
\end{align}
This formula will be very useful in the future.

\section{Deeper Layers: Accumulation of Non-Gaussianity}
\label{sec:deeper-layer-accumulation}
The preactivations in the deeper layers are recursively given by
\be
\z{i}{\alpha}{\ell+1} =\bias{i}{\ell+1}+\sum_{j=1}^{n_{\ell}}\W{ij}{\ell+1}\s{j}{\alpha}{\ell}\, , \quad \text{for} \quad i=1,\ldots,n_{\ell+1}\, ,
\ee
with the activations in the previous layer abbreviated as
\be
\s{i}{\alpha}{\ell}\equiv \sigma\!\le(\z{i}{\alpha}{\ell}\ri)\, .
\ee
We can obtain the marginal distributions of the preactivations in these deeper layers -- including the output distribution $p\!\le(z^{(L)}\Big\vert\D\ri)$ -- by following the procedure that we implemented for the second-layer distribution.
The only complication is that the preactivation distribution in the previous layer is no longer Gaussian, like it was for the first layer.

The three key concepts of the derivation are: recursion, action, and $1/n$-expansion. Let's walk through them one by one.

\subsubsection{Recursion}
The idea of recursion is to start with information contained in the marginal distribution $p\!\le( z^{(\ell)}\Big\vert\D\ri)$ in the $\ell$-th layer and obtain the marginal distribution for the $(\ell+1)$-th layer.
The change of the marginal preactivation distribution from layer to layer can be captured by first writing out the joint probability distribution of preactivations in adjacent layers $\ell$ and $\ell+1$,
\be\label{eq:joint-distribution-adjacent-layers}
p\!\le(z^{(\ell+1)},z^{(\ell)}\Big\vert\D\ri)=p\!\le(z^{(\ell+1)}\Big\vert z^{(\ell)}\ri) p\!\le( z^{(\ell)}\Big\vert\D\ri)\, ,
\ee
then calculating the conditional probability distribution $p\!\le(z^{(\ell+1)}\Big\vert z^{(\ell)}\ri)$, and finally marginalizing over the preactivations at the $\ell$-th layer as
\be\label{eq:marginal-integrated-out-ell-layer}
p\!\le(z^{(\ell+1)}\Big\vert\D\ri)=\int \le[\prod_{i,\alpha}  d\z{i}{\alpha}{\ell}\ri] p\!\le(z^{(\ell+1)}\Big\vert z^{(\ell)}\ri)  p\!\le( z^{(\ell)}\Big\vert\D\ri)\, .
\ee
In particular, the conditional probability distribution $p\!\le(z^{(\ell+1)}\Big\vert z^{(\ell)}\ri)$ serves as a \term{transition matrix}, bridging preactivation distributions in adjacent layers.

The calculation of this conditional distribution proceeds identically to the one we performed for the first layer \eqref{eq:first-layer-formal-expression} and then repurposed for computing the second-layer conditional distribution \eqref{eq:second-conditioned-first}. 
If you'd like, you can again follow along with~\S\ref{sec:first-layer-gaussian}, replacing $z^{(1)}$ by $z^{(\ell+1)}$ and $\x{j}{\alpha}$ by $\s{j}{\alpha}{\ell}$, and obtain
\be\label{eq:general-layer-conditional}
p\!\le(z^{(\ell+1)}\Big\vert z^{(\ell)}\ri) = \frac{1}{\sqrt{\dete{2\pi \widehat{G}^{(\ell+1)}}^{n_{\ell+1}}}} \exp\!\le(-\frac{1}{2}\sum_{i=1}^{n_{\ell+1}}\sum_{\alpha_1,\alpha_2\in\D}\SKinv{\alpha_1 \alpha_2}{\ell+1}\z{i}{\alpha_1}{\ell+1}\z{i}{\alpha_2}{\ell+1}\ri)\,,
\ee
with the $(\ell+1)$-th-layer stochastic metric\index{metric!l-th-layer@$\ell$-th-layer} %
\be\label{eq:general-stochastic-metric}
\Ti{\widehat{G}}{\alpha_1 \alpha_2}{\ell+1}\equiv  \Cb{\ell+1}+\CW{\ell+1}\frac{1}{n_{\ell}}\sum_{j=1}^{n_{\ell}}\s{j}{\alpha_1}{\ell}\s{j}{\alpha_2}{\ell}\, ,
\ee
depending on the random variables $z^{(\ell)}$ in the previous layer $\ell$ through the activations $\sigma^{(\ell)}$.  Note that all the correlators with odd numbers of the $(\ell+1)$-th-layer preactivations vanish while even-point correlators are obtained through Wick's contractions, yielding
\be\label{eq:general-even-moment}
\E{\z{i_1}{\alpha_1}{\ell+1}\cdots\z{i_{2m}}{\alpha_{2m}}{\ell+1}}=\sum_{\text{all pairings}}\delta_{i_{k_1} i_{k_2}}\cdots \delta_{i_{k_{2m-1}} i_{k_{2m}}} \E{\Ti{\widehat{G}}{\alpha_{k_1} \alpha_{k_2}}{\ell+1}\cdots \Ti{\widehat{G}}{\alpha_{k_{2m-1}} \alpha_{k_{2m}}}{\ell+1}},
\ee
where the sum runs over all the $(2m-1)!!$ parings of auxiliary indices $(k_1,\ldots,k_{2m})$.
On the left hand, the expectation value characterizes the $(\ell+1)$-th-layer preactivation distribution;
on the right hand, the expectation value becomes a correlator of $\ell$-th-layer activations upon plugging in the stochastic metric~\eqref{eq:general-stochastic-metric}, which can be evaluated with the $\ell$-th-layer distribution.

The mean of the stochastic metric is given by
\be\label{eq:mean-metric-any-layer}
\Ti{G}{\alpha_1 \alpha_2}{\ell+1}\equiv\E{\Ti{\widehat{G}}{\alpha_1 \alpha_2}{\ell+1}}=\Cb{\ell+1}+\CW{\ell+1}\frac{1}{n_{\ell}}\sum_{j=1}^{n_{\ell}}\E{\s{j}{\alpha_1}{\ell}\s{j}{\alpha_2}{\ell}}\, ,
\ee
and this mean metric governs the two-point correlator in the $(\ell+1)$-th layer through
\be\label{eq:C2_MLPH}
\E{\z{i_1}{\alpha_1}{\ell+1}\z{i_2}{\alpha_2}{\ell+1}}=\delta_{i_1i_2}\E{\Ti{\widehat{G}}{\alpha_1 \alpha_2}{\ell+1}}=\delta_{i_1i_2}\Ti{G}{\alpha_1 \alpha_2}{\ell+1}\, ,
\ee
as we saw for the second layer~\eqref{eq:C2_MLP2} as a special case of the equation~\eqref{eq:general-even-moment}.
Meanwhile, the fluctuation around the mean
\be\label{eq:metric-fluctuation-general-layer}
\dKi{\alpha_1\alpha_2}{\ell+1}\equiv \Ti{\widehat{G}}{\alpha_1 \alpha_2}{\ell+1}-\Ti{G}{\alpha_1 \alpha_2}{\ell+1}=\CW{\ell+1}\frac{1}{n_{\ell}}\sum_{j=1}^{n_{\ell}}\le(\s{j}{\alpha_1}{\ell}\s{j}{\alpha_2}{\ell}-\E{\s{j}{\alpha_1}{\ell}\s{j}{\alpha_2}{\ell}}\ri)\, ,
\ee
obviously has zero mean,
\be
\E{\dKi{\alpha_1\alpha_2}{\ell+1}}=0\, ,
\ee
and has a magnitude 
\be\label{eq:vertex-in-terms-of-metric-fluctuation}
\frac{1}{n_{\ell}}V^{(\ell+1)}_{(\alpha_1\alpha_2)(\alpha_3\alpha_4)}\equiv \E{\dKi{\alpha_1\alpha_2}{\ell+1}\dKi{\alpha_3\alpha_4}{\ell+1}}=\E{ \Ti{\widehat{G}}{\alpha_1 \alpha_2}{\ell+1} \Ti{\widehat{G}}{\alpha_3 \alpha_4}{\ell+1}}-\Ti{G}{\alpha_1 \alpha_2}{\ell+1}\Ti{G}{\alpha_3 \alpha_4}{\ell+1}\, .
\ee
Here we have introduced the $(\ell+1)$-th-layer four-point vertex $V^{(\ell+1)}_{(\alpha_1\alpha_2)(\alpha_3\alpha_4)}$, generalizing the second-layer four-point vertex~\eqref{eq:second-layer-metric-fluctuation-two-point-function},
which governs the connected four-point correlator in the $(\ell+1)$-th layer. Specifically, following along with the manipulations for the second layer -- cf.~\eqref{eq:M4_MLP2} and~\eqref{eq:C4_MLP2} -- or simply applying the general expression~\eqref{eq:general-even-moment}, we see
\begin{align}\label{eq:C4_MLPH}
&\E{\z{i_1}{\alpha_1}{\ell+1}\z{i_2}{\alpha_2}{\ell+1}\z{i_3}{\alpha_3}{\ell+1}\z{i_4}{\alpha_4}{\ell+1}}\Big\vert_{\text{connected}}\, \\
=&\frac{1}{n_{\ell}}\le[\delta_{i_1i_2}\delta_{i_3 i_4}V^{(\ell+1)}_{(\alpha_1\alpha_2)(\alpha_3\alpha_4)}+\delta_{i_1i_3}\delta_{i_2 i_4}V^{(\ell+1)}_{(\alpha_1\alpha_3)(\alpha_2\alpha_4)}+\delta_{i_1i_4}\delta_{i_2 i_3}V^{(\ell+1)}_{(\alpha_1\alpha_4)(\alpha_2\alpha_3)} \ri]\, .\nonumber
\end{align}

In summary, what we have so far are the expressions for the two-point correlator~\eqref{eq:C2_MLPH} and the connected four-point correlator~\eqref{eq:C4_MLPH} of the $(\ell+1)$-th-layer preactivations in terms of the correlators of the $\ell$-th-layer activations, and related expressions for higher-point correlators~\eqref{eq:general-even-moment} if the need arises. The strategy of our recursive approach is to first evaluate these $\ell$-th-layer activation correlators given the $\ell$-th-layer distribution $p\!\le(z^{(\ell)}\Big\vert\D\ri)$ and from them obtain the $(\ell+1)$-th-layer preactivation correlators. 
Using these correlators, we can then reconstruct the $(\ell+1)$-th layer marginal distribution $p\!\le(z^{(\ell+1)}\Big\vert\D\ri)$. %
Both the evaluation of the $\ell$-th-layer activation correlators and the reconstruction of the distribution at the $(\ell+1)$-th layer can be efficiently implemented through the use of the action.

\subsubsection{Action}
The preactivation distribution $p\!\le(z^{(\ell)}\Big\vert\D\ri)$ can be written in terms of an action as
\be\label{eq:marginal-distribution-action-ansatz}
p\!\le(z^{(\ell)}\Big\vert\D\ri)=\frac{e^{-\EFT{\ell}}}{Z(\ell)} \, ,
\ee
with the $\ell$-th layer \terminate{partition function} given by
\be\label{eq:chapter-ngp-partition-function}
Z(\ell) \equiv \int \le[\prod_{i,\alpha} d\z{i}{\alpha}{\ell}\ri] \, e^{-\EFT{\ell}} \, ,
\ee
and our ansatz for the action given by the following expansion:
\begin{align}\label{eq:general-ell-action}
\ac\!\le(z^{(\ell)}\ri)
\equiv&\frac{1}{2}\sum_{i=1}^{n_{\ell}}\sum_{\alpha_1,\alpha_2\in\D} g^{\alpha_1\alpha_2}_{(\ell)} \z{i}{\alpha_1}{\ell}\z{i}{\alpha_2}{\ell}\, \\
&-\frac{1}{8}\sum_{i_1,i_2=1}^{n_{\ell}}\sum_{\alpha_1,\ldots,\alpha_4\in\D}v^{(\alpha_1\alpha_2)(\alpha_3\alpha_4)}_{(\ell)} \z{i_1}{\alpha_1}{\ell}\z{i_1}{\alpha_2}{\ell}\, \z{i_2}{\alpha_3}{\ell}\z{i_2}{\alpha_4}{\ell}+\ldots\, .\nonumber%
\end{align}
This ansatz encompasses both the actions we had in~\S\ref{sec:first-layer-gaussian} for the first-layer preactivations -- with $\TI{g}{\alpha_1\alpha_2}{1}=\TI{G}{\alpha_1\alpha_2}{1}$ and $v_{(1)}=0$ -- and for the second-layer preactivations in~\S\ref{sec:second-layer-non-gaussian} -- with the couplings $g_{(2)}$ and $v_{(2)}$ given by~\eqref{eq:second-layer-quadratic-coupling} and~\eqref{eq:second-layer-quartic-coupling}, respectively. In fact, this represents the most general expansion around the Gaussian action, given the symmetries of preactivation correlators~\eqref{eq:general-even-moment}. In particular, only even powers of preactivations show up in the action since we know that correlators with odd numbers of preactivations vanish.

Here, the coefficients $g^{\alpha_1\alpha_2}_{(\ell)}$, $v^{(\alpha_1\alpha_2)(\alpha_3\alpha_4)}_{(\ell)}$,
and the implied additional terms in the expansion are \textbf{data-dependent couplings}\index{data-dependent coupling|textbf}\index{coupling!data-dependent|see{data-dependent coupling}} that together govern the interactions of the neural preactivations and are simply related to the correlators of preactivations $z^{(\ell)}$. In particular, in~\S\ref{sec:second-layer-non-gaussian} we gave two derivations  for the relations between quadratic and quartic couplings on the one hand and two-point and four-point correlators on the other hand. The same argument applies for an arbitrary layer $\ell$, and so we have
\begin{align}
\TI{g}{\alpha_1\alpha_2}{\ell}&=\TI{G}{\alpha_1\alpha_2}{\ell}+\o{v,\ldots}\, ,\label{eq:two-point-match-general}\\
\TI{v}{(\alpha_1\alpha_2)(\alpha_3\alpha_4)}{\ell}&=\frac{1}{n_{\ell-1}}\TI{V}{(\alpha_1\alpha_2)(\alpha_3\alpha_4)}{\ell}+\o{v^2, \ldots}\, , \label{eq:four-point-match-general} 
\end{align}
with the understanding that the raised indices of the four-point vertex are shorthand for contraction with the $\ell$-th-layer inverse metric
\be\label{eq:vertex-UUUU-dddd}
\TI{V}{(\alpha_1\alpha_2)(\alpha_3\alpha_4)}{\ell}\equiv \sum_{\beta_1,\ldots,\beta_4\in\D}\TI{G}{\alpha_1\beta_1}{\ell}\TI{G}{\alpha_2\beta_2}{\ell}\TI{G}{\alpha_3\beta_3}{\ell}\TI{G}{\alpha_4\beta_4}{\ell}\Ti{V}{(\beta_1\beta_2)(\beta_3\beta_4)}{\ell}\, .
\ee
Note that the higher-order terms $\o{...}$ in~\eqref{eq:two-point-match-general} and~\eqref{eq:four-point-match-general} can be neglected self-consistently if and only if the quartic coupling $v$ and higher-order couplings are perturbatively small. This is indeed the case when networks are sufficiently wide,
as we will show next.

\subsubsection{Large-width expansion}
\index{$1/n$ expansion}
Now we have our work cut out for us. 
First, note that these mappings,~\eqref{eq:two-point-match-general} and~\eqref{eq:four-point-match-general},
between the correlators and couplings already accomplish one task mentioned in our recursive strategy. Namely, when applied to the $(\ell+1)$-th layer, they reconstruct the $(\ell+1)$-th-layer distribution out of the $(\ell+1)$-th-layer preactivation correlators. The only remaining task then is to use the $\ell$-th-layer action~\eqref{eq:general-ell-action} to compute
the expectations of the $\ell$-th-layer activations $\sigma^{(\ell)}$ that appear in the expressions for the two-point correlator~\eqref{eq:C2_MLPH} and four-point correlator~\eqref{eq:C4_MLPH} of the $(\ell+1)$-th-layer preactivations $z^{(\ell+1)}$. 

\index{$1/n$ expansion}
These calculations simplify in the wide regime with a large number of neurons per layer
\be\label{eq:wide-regime}
n_1,n_2,\ldots,n_{L-1}\sim n \gg1\, .
\ee 
As has been advertised, this large-but-finite-width regime is where networks become both practically usable and theoretically tractable.
Specifically, the relations~\eqref{eq:two-point-match-general} and~\eqref{eq:four-point-match-general} between correlators and couplings simplify in this regime and higher-order non-Gaussian corrections can be self-consistently truncated in a series in $1/n$.\footnote{In the language of~\S\ref{sec:marginalization-group-flow}, such a truncation is preserved under the RG flow.}
To be precise, we inductively assume that the mean metric $G^{(\ell)}=\o{1}$ and the four-point vertex $V^{(\ell)}=\o{1}$ are both of order one at the $\ell$-th layer  -- as was the case for the first and second layers -- and show that the same holds true at the $(\ell+1)$-th layer.
This inductive assumption in particular implies through~\eqref{eq:two-point-match-general} and~\eqref{eq:four-point-match-general} that the quartic coupling $v_{(\ell)}=\o{1/n}$ is perturbatively small at the $\ell$-th layer and that the quadratic coupling is given by $g_{(\ell)}=G_{(\ell)}+\o{1/n}$. In carrying out this inductive proof, we obtain the recursion relations that govern the change in the preactivation distributions from the $\ell$-th layer to the $(\ell+1)$-th layer.%

To begin, we see that the two-point correlator in the $(\ell+1)$-th layer~\eqref{eq:C2_MLPH} is given simply in terms of the metric
\be\label{eq:C2_MLP_exact}
\Ti{G}{\alpha_1 \alpha_2}{\ell+1}=\Cb{\ell+1}+\CW{\ell+1}\frac{1}{n_{\ell}}\sum_{j=1}^{n_{\ell}}\E{\s{j}{\alpha_1}{\ell}\s{j}{\alpha_2}{\ell}}\, .
\ee
With foresight, we already evaluated this particular two-point correlator of activations \eqref{eq:two-activations-deep} in the last section. Inserting this result, along with the quadratic coupling $g_{(\ell)}=G_{(\ell)}+\o{1/n}$ and quartic coupling $v_{(\ell)}=\o{1/n}$, we find  
\be\label{eq:G-recursion-tree}
\Ti{G}{\alpha_1 \alpha_2}{\ell+1}=\Cb{\ell+1}+\CW{\ell+1}\bra \sigma_{\alpha_1}\sigma_{\alpha_2}\ket_{G^{(\ell)}}+\o{\frac{1}{n}}\, ,
\ee
which is the leading recursion for the two-point correlator of preactivations.\footnote{Note that the difference from the second-layer calculation in \S\ref{sec:second-layer-non-gaussian} is just that the expectation in \eqref{eq:C2_MLP_exact} is not exactly Gaussian, but has a $1/n$ correction. This highlights the main difference with that section, which is that the distribution in the prior layer is nearly-Gaussian.} We see  that this is self-consistent; any metric $\Ti{G}{}{\ell}$ that is of order one will give an order-one metric $\Ti{G}{}{\ell+1}$ in the next layer as well. The correction is suppressed by $\o{1/n}$, which affects only the subleading term  in the  quadratic coupling $g_{(\ell+1)}=G_{(\ell+1)}+\o{1/n}$. Note that, neglecting the subleading $1/n$ correction and replacing $G$ by $g$,  the recursion~\eqref{eq:G-recursion-tree} for the two-point correlator can also be thought of as the leading recursion for the quadratic coupling.

Next, let's evaluate the four-point correlator~\eqref{eq:C4_MLPH},
which involves computing the magnitude of the metric fluctuation~\eqref{eq:vertex-in-terms-of-metric-fluctuation}. Substituting in our general expression for the $(\ell+1)$-th-layer metric fluctuation \eqref{eq:metric-fluctuation-general-layer}, we get
\begin{align}\label{eq:midpoint-vertex-recursion}
&\frac{1}{n_{\ell}}V^{(\ell+1)}_{(\alpha_1\alpha_2)(\alpha_3\alpha_4)}\, \\
=&\le(\frac{\CW{\ell+1}}{n_{\ell}}\ri)^2  \sum_{j,k=1}^{n_{\ell}}  \le\{ \E{ \s{j}{\alpha_1}{\ell}\s{j}{\alpha_2}{\ell}\s{k}{\alpha_3}{\ell}\s{k}{\alpha_4}{\ell} }  -\E{\s{j}{\alpha_1}{\ell}\s{j}{\alpha_2}{\ell}}\E{\s{k}{\alpha_3}{\ell}\s{k}{\alpha_4}{\ell}} \ri\} \, .\nonumber%
\end{align}
Here, there are two types of the contributions: from coincident neurons and from separate pairs of neurons. Again, with foresight, we have already evaluated both types of four-point activation correlators in the last section. When all four are coincident $j=k$, substituting in~\eqref{eq:four-activations-same-deep} we find
\begin{align}\label{eq:activation-four-point-same-neurons}
 &\E{ \s{j}{\alpha_1}{\ell}\s{j}{\alpha_2}{\ell}\s{j}{\alpha_3}{\ell}\s{j}{\alpha_4}{\ell} }  -\E{\s{j}{\alpha_1}{\ell}\s{j}{\alpha_2}{\ell}}\E{\s{j}{\alpha_3}{\ell}\s{j}{\alpha_4}{\ell}} \, \\
 =&\bra\sigma_{\alpha_1} \sigma_{\alpha_2} \sigma_{\alpha_3} \sigma_{\alpha_4}\ket_{G^{(\ell)}}  - \bra \sigma_{\alpha_1} \sigma_{\alpha_2}\ket_{G^{(\ell)}} \bra \sigma_{\alpha_3} \sigma_{\alpha_4}\ket_{G^{(\ell)}} +\o{\frac{1}{n}}\, , \nonumber
\end{align}
where we have truncated to leading order in $1/n$ as a consequence of the inductive assumption at the $\ell$-th layer.\index{$1/n$ expansion} Meanwhile, when $j \neq k$ and the correlation is between two neurons, we substitute in our expression~\eqref{eq:four-activations-different-deep-connected}, finding
\begin{align}\label{eq:activation-four-point-different-neurons}
&\E{ \s{j}{\alpha_1}{\ell}\s{j}{\alpha_2}{\ell}\s{k}{\alpha_3}{\ell}\s{k}{\alpha_4}{\ell} }  -\E{\s{j}{\alpha_1}{\ell}\s{j}{\alpha_2}{\ell}}\E{\s{k}{\alpha_3}{\ell}\s{k}{\alpha_4}{\ell}}\, \\
=&\frac{1}{4n_{\ell-1}}\sum_{\beta_1,\ldots,\beta_4\in\D}\TI{V}{(\beta_1\beta_2)(\beta_3\beta_4)}{\ell}\bra \sigma_{\alpha_1}\sigma_{\alpha_2} \le(z_{\beta_1} z_{\beta_2}-g_{\beta_1\beta_2}\ri)\ket_{G^{(\ell)}}\bra \sigma_{\alpha_3}\sigma_{\alpha_4} \le(z_{\beta_3} z_{\beta_4}-g_{\beta_3\beta_4}\ri)\ket_{G^{(\ell)}}\, \nonumber\\
&+\o{\frac{1}{n^2}} \, ,\nonumber
\end{align}
where again we have truncated to leading order in the large-width expansion using the inductive assumption.\footnote{Again, the difference with the second-layer calculation is that in \S\ref{sec:second-layer-non-gaussian} these expectations are over the exactly Gaussian first-layer distribution. In that case, there was a contribution of the form \eqref{eq:activation-four-point-same-neurons} from the case with all neurons coincident, but \emph{not} of the form \eqref{eq:activation-four-point-different-neurons} from the two neurons -- cf.~\eqref{eq:second-layer-metric-fluctuation-two-point-function}.}
Inserting both of these expressions back into \eqref{eq:midpoint-vertex-recursion} and performing the sums, we get a recursion for the four-point vertex
\begin{align}\label{eq:V-recursion-tree}
&\frac{1}{n_{\ell}}V^{(\ell+1)}_{(\alpha_1\alpha_2)(\alpha_3\alpha_4)}\, \\
=&\frac{1}{n_{\ell}} \le(\CW{\ell+1}\ri)^2\le[\bra\sigma_{\alpha_1} \sigma_{\alpha_2} \sigma_{\alpha_3} \sigma_{\alpha_4}\ket_{G^{(\ell)}}  - \bra \sigma_{\alpha_1} \sigma_{\alpha_2}\ket_{G^{(\ell)}} \bra \sigma_{\alpha_3} \sigma_{\alpha_4}\ket_{G^{(\ell)}} \ri]\, \nonumber\\
&+\frac{1}{n_{\ell-1}}\frac{\le(\CW{\ell+1}\ri)^2}{4}\sum_{\beta_1,\ldots,\beta_4\in\D}\TI{V}{(\beta_1\beta_2)(\beta_3\beta_4)}{\ell}\bra\sigma_{\alpha_1}\sigma_{\alpha_2} \le(z_{\beta_1} z_{\beta_2}-g_{\beta_1\beta_2}\ri)\ket_{G^{(\ell)}}\, \nonumber\\
&\quad \quad \quad \quad \quad \quad \quad \quad \quad \quad \quad \quad \quad \quad \quad \quad \times\bra \sigma_{\alpha_3}\sigma_{\alpha_4} \le(z_{\beta_3} z_{\beta_4}-g_{\beta_3\beta_4}\ri)\ket_{G^{(\ell)}}+\o{\frac{1}{n^2}} \, . \nonumber
\end{align}
Importantly, we see that
\be\label{eq:V-recursion-tree-redux}
\frac{1}{n_{\ell}}V^{(\ell+1)}=\o{\frac{1}{n}}\, ,
\ee
and $V^{(\ell+1)}=\o{1}$, thus completing our inductive proof and concluding our derivations of the recursion relations~\eqref{eq:G-recursion-tree} and~\eqref{eq:V-recursion-tree} for the two-point and four-point correlators.
As was the case for the quadratic coupling,  if we neglect the subleading $1/n^2$ correction and replace $G$ by $g$ and $V$ by $v$, the recursion~\eqref{eq:V-recursion-tree} for the connected four-point correlator can also be thought of as the recursion for the quartic coupling. \index{$1/n$ expansion}

\index{$1/n$ expansion}
Note that in the strict $n\to \infty$ limit, the quartic coupling vanishes, and the marginal distribution of preactivations $p\!\le(z^{(\ell)}\Big\vert\D\ri)$ is Gaussian for all layers $\ell$. The first nontrivial correction to this \terminate{infinite-width limit} is captured by studying the quartic action with couplings $v_{(\ell)}$. 
In what follows, we will mostly focus on the \terminate{effective theory} with this quartic action, as we expect significant qualitative differences in the behavior of networks described by the quadratic action vs.~the quartic action. The additional finite-width corrections given by the higher-order terms in the action can change quantitative results but should not really exhibit qualitative differences.

\section{Marginalization Rules} 
\label{sec:sum-rule}
In the past sections, at each step in the recursions we marginalized over all the preactivations in a given layer.
This section collects two remarks on other sorts of \emph{partial} marginalizations we can perform, rather than integrating out an \emph{entire} layer. In particular, we'll discuss marginalization over a subset of the $\ND$ samples in the dataset\index{input data} $\D$ and marginalization over a subset of neurons in a layer. 

Loosely speaking, these marginalizations let us focus on specific input data and neurons of interest. Tightly speaking, let's consider evaluating the expectation of a function $F\!\le(z_{I;\A}\ri)=F\!\le(\le\{z_{i;\alpha}\ri\}_{i\in I; \alpha\in\A}\ri)$ that depends on a subsample $\A\subset\D$ and a subset of neurons $I\subset\le\{1,\ldots,n_\ell\ri\}\equiv \mathcal{N}$ in a layer $\ell$, where with a slight abuse of notation we put the set dependences into the subscripts. We then have
\begin{align}\label{eq:marginalization-rule}
&\E{F\!\le(z_{I;\A}\ri)}\, \\
=&\int \le[\prod_{i\in \mathcal{N}}\prod_{\alpha\in\D} dz_{i;\alpha}\ri]F\!\le(z_{I;\A}\ri)\, p\!\le(z_{\mathcal{N};\D}\Big\vert\D\ri) \, \nonumber\\
=&\int \le[\prod_{i\in I}\prod_{\alpha\in\A} dz_{i;\alpha}\ri] F\!\le(z_{I;\A}\ri)\le\{\int \le[ \prod_{(j;\beta)\in\le[ \mathcal{N}\times\D-I\times\A\ri]}dz_{j;\beta}\ri] p\!\le(z_{\mathcal{N};\D}\Big\vert\D\ri)\ri\}\, \nonumber\\
=&\int \le[\prod_{i\in I}\prod_{\alpha\in\A} dz_{i;\alpha}\ri] F\!\le(z_{I;\A}\ri)\, p\!\le(z_{I;\A}\Big\vert\A\ri) \notag
\end{align}
where the last equality is just the marginalization over the spectator variables that do not enter into the observable of interest and, in a sense, defines the subsampled and subneuroned distribution as
\be\label{eq:sum-rule-mlp}
p\!\le(z_{I;\A}\Big\vert\A\ri) \equiv \int \le[ \prod_{(j;\beta)\in\le[ \mathcal{N}\times\D-I\times\A\ri]}dz_{j;\beta}\ri]\ p\!\le(z_{\mathcal{N};\D}\Big\vert\D\ri) \, .
\ee
In words, in evaluating the expectation of the function $F\!\le(z_{I;\A}\ri)$, the full distribution $p\!\le(z_{\mathcal{N};\D}\Big\vert\D\ri)$ can simply be restricted to that of the subsample $\A$ and subneurons $I$, i.e., $p\!\le(z_{I;\A}\Big\vert\A\ri)$.
We call this property a \textbf{marginalization rule}\index{marginalization rule|textbf}.
Yes, this is somewhat trivial -- we're just restating the consistency of probability distributions with respect to marginalization -- but it has two rather useful consequences for us.

\subsubsection{Marginalization over samples}
The first corollary of the marginalization rule is that we can use it to reduce a gigantic integral over all the samples in the dataset to a compact integral over only a handful of samples. For example,
in recursively obtaining the two-point correlator through
\be
\E{\z{i_1}{\alpha_1}{\ell+1}\z{i_2}{\alpha_2}{\ell+1}}
=\delta_{i_1i_2} \le[\Cb{\ell+1}+\CW{\ell+1}\bra \sigma_{\alpha_1}\sigma_{\alpha_2}\ket_{G^{(\ell)}}+\o{\frac{1}{n}}\, \ri]\, ,
\ee
we can reduce the $\ND$-dimensional Gaussian integrals
$\bra \sigma_{\alpha_1} \sigma_{\alpha_2}\ket_{G^{(\ell)}}$ with the $\ND$-by-$\ND$ variance matrix $G^{(\ell)}$
to a manageable two-dimensional integral with a two-by-two submatrix spanned by $\alpha_1$ and $\alpha_2$ (or a one-dimensional integral if $\alpha_1=\alpha_2$). Similarly, a Gaussian integral for four activations $\bra\sigma_{\alpha_1} \sigma_{\alpha_2} \sigma_{\alpha_3} \sigma_{\alpha_4}\ket_{G^{(\ell)}}$ that appears in the recursion for four-point vertex involves integrals over four variables \emph{at most}.
Generally, in using the action~\eqref{eq:general-ell-action} to evaluate a specific expectation, the summation over the whole dataset\index{input data} $\D$ in the action can be restricted to the subset of input data that actually appears in the expectation.
By the same token, in recursively evaluating the four-point vertex $V^{(\ell+1)}_{(\alpha_1\alpha_2)(\alpha_3\alpha_4)}$ via the recursion \eqref{eq:V-recursion-tree}, the summation on the right-hand side over the dataset $\D$  can be restricted to the set of samples being correlated, $\{\alpha_1,\alpha_2,\alpha_3,\alpha_4\}$.
However, please keep in mind that the inverse metrics used to construct $\TI{V}{(\beta_1\beta_2)(\beta_3\beta_4)}{\ell}$ in \eqref{eq:vertex-UUUU-dddd} must then be taken to be the inverse of the metric submatrix on this restricted subspace.\footnote{A similar restriction of the summation can be applied to any of our other recursions and will prove especially useful when you try to evaluate them numerically or analytically.}

\subsubsection{Marginalization over neurons}
\index{marginalization rule}
The second corollary involves integrating out a subset of neurons in a layer.
Prudent readers might have worried that the quartic term in the $\ell$-th-layer action,
\be\label{eq:marginalization-rules-quartic}
-\frac{1}{8}\sum_{i_1,i_2=1}^{n_{\ell}}\sum_{\alpha_1,\ldots,\alpha_4\in\D}v^{(\alpha_1\alpha_2)(\alpha_3\alpha_4)}_{(\ell)} \z{i_1}{\alpha_1}{\ell}\z{i_1}{\alpha_2}{\ell}\, \z{i_2}{\alpha_3}{\ell}\z{i_2}{\alpha_4}{\ell} \, ,
\ee
seems to naively scale like $\sim n_{\ell}^2 / n_{\ell-1} = \o{n}$, since there are two sums over $n_{\ell}$, and we know from \eqref{eq:four-point-match-general}  that the coupling $v_{(\ell)}$ scales like $\sim 1/n_{\ell-1}$. Similarly, the quadratic term,
\be\label{eq:marginalization-rules-quadratic}
\frac{1}{2}\sum_{i=1}^{n_{\ell}}\sum_{\alpha_1,\alpha_2\in\D} g^{\alpha_1\alpha_2}_{(\ell)} \z{i}{\alpha_1}{\ell}\z{i}{\alpha_2}{\ell} \, ,
\ee
has a single sum over $n_{\ell}$ and so seems naively $\o{n}$ as well. This would imply that the quartic term isn't perturbatively suppressed in comparison to the quadratic term, naively calling our perturbative approach into question.

We first observe that this problem never arises for the final layer $\ell=L$, since the output dimension $n_{L}$ is never parametrically large: the quadratic term scales as $\sim n_{L}=\o{1}$ while the quartic term scales as $\sim n_{L}^2/n_{L-1}=\o{1/n}$, which is perturbatively suppressed.

\index{marginalization rule}
This observation, combined with the marginalization rule, points at a resolution to the naive scale-counting problem above for the hidden layers.
Indeed, all the expectation we evaluated so far
-- both preactivation and activation correlators -- each individually involves only a few neurons $m_{\ell}$ in any given layer $\ell$, with $m_{\ell} \ll n_{\ell}$. This will always be true; we can't actually correlate an infinite number of neurons at once!
 Thus, when using the action representation~\eqref{eq:general-ell-action} of the probability distribution to compute these correlators at the $\ell$-th layer, we can first use the marginalization rule~\eqref{eq:marginalization-rule}\index{marginalization rule} to integrate out the $(n_{\ell} - m_{\ell})$ spectator neurons that do not participate in the computation, letting us focus on those $m_{\ell}$ relevant neurons that actually appear in the expectation. This in turn lets us replace the summations over $n_{\ell}$ neurons by ones over the $m_{\ell}$ neurons.\footnote{In evaluating generic expectation value such as~\eqref{eq:arbitrary-function-quartic-expectation}, one can always check that the contributions from the $(n_{\ell}-m_{\ell})$ spectator neurons consistently cancel out at each order in $1/n_{\ell-1}$ expansion. If you go back to your personal note that fills in the small gaps between lines in our computations, you will surely notice this cancellation due to Gaussian factorization.}

 All the while, the numbers of neurons in the previous layers $n_{1},\ldots,n_{\ell-1}$, having been integrated out to get the action representation at the $\ell$-th layer, \emph{are} parametrically large.
 This means that the quadratic term in the $\ell$-th-layer action, reduced to the $m_{\ell}$ relevant neurons, scales as $\sim m_{\ell}=\o{1}$, while the quartic term scales as $\sim m_{\ell}^2/n_{\ell-1}=\o{1/n}$. Thus, this ensures a perturbative treatment of the non-Gaussianity.

\subsubsection{Running couplings with partial marginalizations}\index{running coupling}\index{coupling!running|see{running coupling}}
In focusing our attention on only a subset of samples or neurons, the data-dependent couplings\index{data-dependent coupling} of the action need to be adjusted. Since this running of the couplings is instructive and will be necessary for later computations, let us illustrate here how the quadratic coupling\index{running coupling!quadratic} $g_{(\ell),m_{\ell}}^{\alpha_1\alpha_2}$ depends on the number of neurons $m_{\ell}$ in the action.

For simplicity in our illustration, let us specialize to a single input $x$ and drop all the \terminate{sample indices}. Then, denote the distribution over $m_{\ell}$ neurons as 
\begin{align}\label{eq:m-neuron-action}
p\!\le(z_1^{(\ell)},\ldots, z_{m_{\ell}}^{(\ell)}\ri)&\propto e^{-\ac\big(z_1^{(\ell)}\!,\,\ldots\,, \, z_{m_{\ell}}^{(\ell)}\big)}\, \\
&=\exp\!\le[-\frac{g_{(\ell),m_{\ell}}}{2}\sum_{j=1}^{m_{\ell}}z_{j}^{(\ell)}z_{j}^{(\ell)}+\frac{v_{(\ell)}}{8}\sum_{j_1,j_2=1}^{m_{\ell}}z_{j_1}^{(\ell)}z_{j_1}^{(\ell)}z_{j_2}^{(\ell)}z_{j_2}^{(\ell)}\ri]\, ,\notag
\end{align}
which is expressed by the same action we've already been using~\eqref{eq:general-ell-action}, though now the dependence of the quadratic coupling\index{coupling!quadratic} on $m_{\ell}$ is made explicit.\footnote{Note that in principle the quartic coupling\index{coupling!quartic} should also depend on $m_{\ell}$: $v_{(\ell)} \to v_{(\ell),m_{\ell}}$. However, since such a dependence only shows up at higher order in $v$, we will suppress it.}
We'll now see in two ways how the quadratic coupling\index{coupling!quadratic} $g_{(\ell),m_{\ell}}$ \emph{runs} with $m_{\ell}$.\index{running coupling}\index{running coupling}

\index{integrating out}
The first way is to begin with the action for $n_{\ell}$ neurons and formally integrate out $(n_{\ell}-m_{\ell})$ neurons. Without loss of generality, let's integrate out the \emph{last} $(n_{\ell}-m_{\ell})$ neurons, leaving the \emph{first} $m_\ell$ neurons labeled as $1, \ldots, m_\ell$. 
Using the marginalization rule\index{marginalization rule}~\eqref{eq:sum-rule-mlp}, we see that
\begin{align}
e^{-\ac\big(z_1^{(\ell)}\!,\,\ldots\,, \, z_{m_{\ell}}^{(\ell)}\big)}\propto&\ p\!\le(z_1^{(\ell)},\ldots, z_{m_{\ell}}^{(\ell)}\ri)=\int d{ z_{m_{\ell}+1}^{(\ell)}}\cdots d{ z_{n_{\ell}}^{(\ell)}}\ p\!\le(z_1^{(\ell)},\ldots, z_{n_{\ell}}^{(\ell)}\ri)\\
\propto&\int d{ z_{m_{\ell}+1}^{(\ell)}}\cdots d{ z_{n_{\ell}}^{(\ell)}}\exp\!\!\le[-\frac{g_{(\ell),n_{\ell}}}{2}\sum_{i=1}^{n_{\ell}}z_{i}^{(\ell)}z_{i}^{(\ell)}+\frac{v_{(\ell)}}{8}\!\!\sum_{i_1,i_2=1}^{n_{\ell}}z_{i_1}^{(\ell)}z_{i_1}^{(\ell)}z_{i_2}^{(\ell)}z_{i_2}^{(\ell)}\ri]\, ,\notag
\end{align}
throughout which we neglected \terminate{normalization factor}s that are irrelevant if we're just interested in the running of the coupling.
Next, we can separate out the dependence on the $m_{\ell}$ neurons, perturbatively expand the integrand in quartic coupling\index{coupling!quartic}, and finally integrate out the last $(n_{\ell}-m_{\ell})$ neurons by computing a few simple Gaussian integrals:
\begin{align}
&p\!\le(z_1^{(\ell)},\ldots, z_{m_{\ell}}^{(\ell)}\ri)\, \\
\propto&\exp\!\!\le[-\frac{g_{(\ell),n_{\ell}}}{2}\sum_{j=1}^{m_{\ell}}z_{j}^{(\ell)}z_{j}^{(\ell)}+\frac{v_{(\ell)}}{8}\!\!\sum_{j_1,j_2=1}^{m_{\ell}}z_{j_1}^{(\ell)}z_{j_1}^{(\ell)}z_{j_2}^{(\ell)}z_{j_2}^{(\ell)}\ri]\, \notag\\
&\times\int d{ z_{m_{\ell}+1}^{(\ell)}}\cdots d{z_{n_{\ell}}^{(\ell)}}\exp\!\!\le[-\frac{g_{(\ell),n_{\ell}}}{2}\sum_{k=m_{\ell}+1}^{n_{\ell}}z_{k}^{(\ell)}z_{k}^{(\ell)}\ri]\, \notag\\
&\quad \quad \quad \times\le[1+\frac{2v_{(\ell)}}{8}\sum_{j=1}^{m_{\ell}}\sum_{k=m_{\ell}+1}^{n_{\ell}}z_{j}^{(\ell)}z_{j}^{(\ell)}z_{k}^{(\ell)}z_{k}^{(\ell)}+\frac{v_{(\ell)}}{8}\!\!\sum_{k_1,k_2=m_{\ell}+1}^{n_{\ell}}z_{k_1}^{(\ell)}z_{k_1}^{(\ell)}z_{k_2}^{(\ell)}z_{k_2}^{(\ell)}+\o{v^2}\ri]\, \notag\\
=&\exp\!\!\le[-\frac{g_{(\ell),n_{\ell}}}{2}\sum_{j=1}^{m_{\ell}}z_{j}^{(\ell)}z_{j}^{(\ell)}+\frac{v_{(\ell)}}{8}\!\!\sum_{j_1,j_2=1}^{m_{\ell}}z_{j_1}^{(\ell)}z_{j_1}^{(\ell)}z_{j_2}^{(\ell)}z_{j_2}^{(\ell)}\ri]\, \notag\\
&\times\le\{1+\frac{(n_{\ell}-m_{\ell})}{4} \frac{v_{(\ell)}}{g_{(\ell),n_{\ell}}}\!\le(\sum_{i=1}^{m_{\ell}}z_{i}^{(\ell)}z_{i}^{(\ell)}\ri)+\frac{v_{(\ell)}}{8g_{(\ell),n_{\ell}}^2}\!\le[(n_{\ell}-m_{\ell})^2+2(n_{\ell}-m_{\ell})\ri]\!+\!\o{v^2}\!\ri\}\, .\notag%
\end{align}
Finally,
resumming the correction arising from the quartic coupling proportional to $\sum_{i=1}^{m_{\ell}}z_{i}^{(\ell)}z_{i}^{(\ell)}$ back into the exponential, ignoring the proportionality factor, and comparing with the action for $m_{\ell}$ neurons~\eqref{eq:m-neuron-action}, we find
\be\label{eq:quadratic-partial-running}
g_{(\ell),m_{\ell}}=g_{(\ell),n_{\ell}}-\frac{(n_{\ell}-m_{\ell})}{2} \frac{v_{(\ell)}}{g_{(\ell),n_{\ell}}}+\o{v^2}\, 
\ee
as the running equation for the quadratic coupling.

\index{running coupling}
The second way to see the coupling run -- and find a solution to the running equation~\eqref{eq:quadratic-partial-running} -- is to compute the single-input metric $\Ti{G}{}{\ell}\equiv\E{z_{i}^{(\ell)}z_{i}^{(\ell)}}$ and compute it directly using the $m_{\ell}$-neuron action~\eqref{eq:m-neuron-action}.
We've already computed this in~\eqref{eq:second-moment-from-action} using the quartic action for multiple inputs. Specializing to a single input, considering an action of $m_\ell$ neurons, and being explicit about the dependence of the quadratic coupling\index{running coupling!quadratic} on the number of neurons, we get
\be\label{eq:second-moment-from-partial-action}
\Ti{G}{}{\ell}=\le[\frac{1}{g_{(\ell),m_{\ell}}}+\frac{(m_{\ell}+2)}{2}\frac{v^{(\ell)}}{g_{(\ell),m_{\ell}}^3}\ri]+\o{v^2}\, .
\ee
Solving this equation for $g_{(\ell),m_{\ell}}$ by perturbatively expanding in $v^{(\ell)}$, we find 
\be\label{eq:quadratic-reprint-m-emphasis}
\frac{1}{g_{(\ell),m_{\ell}}}=G^{(\ell)}-\frac{(m_{\ell}+2)}{2}\frac{ V^{(\ell)}}{n_{\ell-1} G^{(\ell)}}+\o{\frac{1}{n^2}}\, ,
\ee
where we have also plugged in
\be\label{eq:quartic-single-input-coupling-for-vertex}
v_{(\ell)}= \frac{V^{(\ell)}}{n_{\ell-1} \le(G^{(\ell)}\ri)^4 }+\o{\frac{1}{n^2}} \, ,
\ee
using \eqref{eq:four-point-match-general} and \eqref{eq:vertex-UUUU-dddd} to relate the quartic coupling to the \terminate{four-point vertex} and again specializing to a single input.
Now, it's easy to check that this expression~\eqref{eq:quadratic-reprint-m-emphasis} solves the running equation~\eqref{eq:quadratic-partial-running}.\footnote{Note that the coupling $g_{(\ell),m_{\ell}}$ depends on $m_{\ell}$ -- and also on the other hidden-layer widths $n_{1},n_{2},\ldots,n_{\ell-1}$ -- but does \emph{not} depend on the overall width of the current layer $n_{\ell}$. This implies that the quadratic coupling\index{coupling!quadratic} $g_{(\ell),m_{\ell}}$ is the same coupling we would have used if instead there were actually only $m_{\ell}$ neurons in the $\ell$-th layer. 
} 

The key step in this alternative derivation is realizing that observables without any \terminate{neural indices} such as $\Ti{G}{}{\ell}$ should \emph{not} depend on which version of the $m_{\ell}$ action we use in computing them.  Interpreted another way, what this running of the coupling\index{running coupling} means is that for different numbers of neurons in a layer $\ell$ -- e.g.~$m_\ell$ and $n_\ell$ -- we need different quadratic couplings -- in this case $g_{(\ell),m_{\ell}}$ and $g_{(\ell),n_{\ell}}$ -- in order to give the correct value for an $\ell$-th-layer observable such as $\Ti{G}{}{\ell}$.
If you're ever in doubt, it's always safest to express an observable of interest in terms of the metric $\Ti{G}{}{\ell}$ and the \terminate{four-point vertex} $\Ti{V}{}{\ell}$ rather than the couplings. %
\index{marginalizing over}

\section{Subleading Corrections}
\label{sec:loop-correction}
\index{$1/n$ expansion}\index{subleading corrections|textbf}\index{subleading corrections|seealso{$1/n$ expansion}}
At finite width, all of the correlators receive an infinite series of subleading corrections. Concretely, the metric governing two-point correlator and the four-point vertex governing the connected four-point correlator have $1/n$ series expansions of the form
\begin{align}\label{eq:self-energy-decomposition}
\Ti{G}{\alpha_1\alpha_2}{\ell}=&G_{\alpha_1\alpha_2}^{\le\{0\ri\}(\ell)}+\frac{1}{n_{\ell-1}}\se{\alpha_1\alpha_2}{\ell}+\frac{1}{n_{\ell-1}^2}G_{\alpha_1\alpha_2}^{\le\{2\ri\}(\ell)}+\o{\frac{1}{n^3}}\, ,\\
\label{eq:vertex-decomposition}
\Ti{V}{(\alpha_1\alpha_2)(\alpha_3\alpha_4)}{\ell}=&V_{(\alpha_1\alpha_2)(\alpha_3\alpha_4)}^{\le\{0\ri\}(\ell)}+\frac{1}{n_{\ell-1}}V_{(\alpha_1\alpha_2)(\alpha_3\alpha_4)}^{\le\{1\ri\}(\ell)}+\o{\frac{1}{n^2}}\, .
\end{align}
While so far we have focused on the leading contributions $G_{\alpha_1\alpha_2}^{\le\{0\ri\}(\ell)}$ and $V_{(\alpha_1\alpha_2)(\alpha_3\alpha_4)}^{\le\{0\ri\}(\ell)}$, the subleading corrections can be systematically calculated as well. Let us illustrate the procedure by deriving the recursion for the next-to-leading-order (NLO) correction to the metric, $\se{\alpha_1\alpha_2}{\ell}$.

Before proceeding, let us remark that the leading contribution of the mean metric fully describes the infinite-width limit of the preactivation distributions and so is given a symbol
\be\label{eq:definition-of-kernel-first}
 \Ti{\ker}{\alpha_1\alpha_2}{\ell}\equiv G_{\alpha_1\alpha_2}^{\le\{0\ri\}(\ell)}\, ,
\ee 
and name, the \textbf{kernel}\index{kernel|textbf}\index{kernel|seealso{metric}}\index{kernel!infinite-width limit of the metric}\index{metric!infinite-width limit}. Since the kernel captures the leading-order correlation between any pair of samples, it will be a central object of study for us in the following chapters.
In a similar vein, we will call $\se{\alpha_1\alpha_2}{\ell}$ the \textbf{NLO metric}\index{NLO metric|see{metric}}\index{metric!next-to-leading-order correction|textbf}.

\index{$1/n$ expansion}
Our first step will be to express the layer-$\ell$ quadratic coupling $\TI{g}{\beta_1\beta_2}{\ell}$ to order $1/n$ in terms of the $1/n$ correlator data in \eqref{eq:self-energy-decomposition} and \eqref{eq:vertex-decomposition}.
Let's begin by recalling the expression \eqref{eq:second-moment-from-action} for the two-point correlator that we derived from the quartic action, reprinted here for layer $\ell$
\begin{align}\label{eq:second-moment-from-action-reprinted}
&\E{\z{i_1}{\alpha_1}{\ell} \z{i_2}{\alpha_2}{\ell}} = \delta_{i_1 i_2}\Ti{G}{\alpha_1\alpha_2}{\ell} \\
=&\delta_{i_1 i_2}\le[\Ti{g}{\alpha_1\alpha_2}{\ell}+\frac{1}{2}\sum_{\beta_1,\ldots,\beta_4\in\D}\TI{v}{(\beta_1\beta_2)(\beta_3\beta_4)}{\ell}\le(n_\ell\, \Ti{g}{\alpha_1\beta_1}{\ell}\Ti{g}{\alpha_2\beta_2}{\ell}\Ti{g}{\beta_3\beta_4}{\ell}+2\Ti{g}{\alpha_1\beta_1}{\ell}\Ti{g}{\alpha_2\beta_3}{\ell}\Ti{g}{\beta_2\beta_4}{\ell}\ri)\ri]+\o{v^2}\, .\nonumber
\end{align}
As a reminder $\Ti{g}{\alpha_1\alpha_2}{\ell}$ is the matrix inverse of the quadratic coupling $\TI{g}{\alpha_1\alpha_2}{\ell}$ .
Substituting in the expansion \eqref{eq:self-energy-decomposition} into \eqref{eq:second-moment-from-action-reprinted}, substituting for the quartic coupling $\TI{v}{}{\ell} = \TI{V}{}{\ell}/n_{\ell-1}$~\eqref{eq:four-point-match-general}, and rearranging to solve for $\Ti{g}{\alpha_1\alpha_2}{\ell}$ to the subleading order, we get
\be\label{eq:matching-two-point-subleading}
\Ti{g}{\alpha_1\alpha_2}{\ell}=\Ti{\ker}{\alpha_1\alpha_2}{\ell}+\frac{1}{n_{\ell-1}}\le[\se{\alpha_1\alpha_2}{\ell}-\sum_{\beta_1,\beta_2\in\D}\TI{\ker}{\beta_1\beta_2}{\ell}\le(\frac{n_{\ell}}{2} \Ti{V}{(\alpha_1\alpha_2)(\beta_1\beta_2)}{\ell}+\Ti{V}{(\alpha_1\beta_1)(\alpha_2\beta_2)}{\ell}\ri)\ri]+\o{\frac{1}{n^2}}\, .
\ee
Note that in obtaining the above, we have self-consistently replaced $g^{(\ell)}$ by $K^{(\ell)}$ in the subleading term, which in turn let us lower the indices of the four-point vertices.
Inverting this expression~\eqref{eq:matching-two-point-subleading} yields the subleading correction to the quadratic coupling in terms of the correlators
\begin{align}\label{eq:inverse-difference}
&\TI{g}{\beta_1\beta_2}{\ell}-\TI{\ker}{\beta_1\beta_2}{\ell}\, \\
=&\frac{1}{n_{\ell-1}}\sum_{\beta_3,\beta_4\in\D}\le[-\TI{\ker}{\beta_1\beta_3}{\ell}\TI{\ker}{\beta_2\beta_4}{\ell}\se{\beta_3\beta_4}{\ell}+\Ti{\ker}{\beta_3\beta_4}{\ell}\le(\frac{n_{\ell}}{2} \TI{V}{(\beta_1\beta_2)(\beta_3\beta_4)}{\ell}+\TI{V}{(\beta_1\beta_3)(\beta_2\beta_4)}{\ell}\ri)\ri]+\o{\frac{1}{n^2}}\, .\nonumber
\end{align}

\index{$1/n$ expansion}\index{subleading corrections}
Note that one term in this correction scales as $n_\ell / n_{\ell-1}$. As discussed in the previous section, the marginalization rule for the $\ell$-th-layer action guarantees that we can treat this quantity as small, $n_\ell / n_{\ell-1} \ll 1$, ensuring that $\Ti{g}{\alpha_1\alpha_2}{\ell}-\Ti{\ker}{\alpha_1\alpha_2}{\ell}$ is a subleading-in-$1/n$ correction to the quadratic coupling. In line with this statement, we'll soon see the cancellation for the factor of $n_{\ell}$ when computing the recursion for this subleading correction to the metric $\se{}{\ell}$.

\index{$1/n$ expansion}\index{subleading corrections}
Having finished working out the $1/n$-corrected $\ell$-th-layer action, we turn to computing the $(\ell+1)$-th-layer two-point correlator.\footnote{We already knew the $1/n$ contribution to the quartic coupling, namely the relation $\TI{v}{}{\ell} = \TI{V}{}{\ell}/n_{\ell-1}$.} This will let us express the $(\ell+1)$-th-layer two-point correlator in terms of the $\ell$-th-layer statistics, ultimately yielding a recursion for $\se{\alpha_1\alpha_2}{\ell}$.
Starting with the expansion \eqref{eq:self-energy-decomposition} in the $(\ell+1)$-th layer  and substituting in the expression~\eqref{eq:C2_MLP_exact} for the two-point correlator, we obtain
\be\label{eq:recursion-two-point-subleading}
\Ti{\ker}{\alpha_1\alpha_2}{\ell+1}+\frac{1}{n_{\ell}}\se{\alpha_1\alpha_2}{\ell+1}+\o{\frac{1}{n^2}}=\Ti{G}{\alpha_1\alpha_2}{\ell+1}=\Cb{\ell+1}+\CW{\ell+1}\frac{1}{n_{\ell}}\sum_{j=1}^{n_{\ell}}\E{\s{j}{\alpha_1}{\ell}\s{j}{\alpha_2}{\ell}}\, .
\ee
Thus, we need the expectation of two activations in the $\ell$-th layer up to the order $\o{1/n}$, which we evaluated before in expression~\eqref{eq:two-activations-deep} in terms of the $\ell$-th-layer couplings.

\index{$1/n$ expansion}\index{subleading corrections}
Looking at \eqref{eq:two-activations-deep}, there are two types of contributions at the subleading order, one arising from the $1/n$ correction to the quadratic coupling $g_{(\ell)}$ in~\eqref{eq:matching-two-point-subleading} and the other from the near-Gaussianity of the distribution due to the quartic coupling $v_{(\ell)}$.
The latter contribution is easy to handle: since the quartic coupling is already suppressed by $1/n$, we can just make the replacement $g^{(\ell)} \to \ker^{(\ell)}$ in the second term in~\eqref{eq:two-activations-deep}, yielding
\begin{align}\label{eq:quartic-easy}
&\frac{1}{8n_{\ell-1}}\sum_{\beta_1,\ldots,\beta_4\in\D}\TI{V}{(\beta_1\beta_2)(\beta_3\beta_4)}{\ell}\, \\
&\times\Big[\bra \sigma_{\alpha_1}\sigma_{\alpha_2} \le(z_{\beta_1} z_{\beta_2}-\Ti{\ker}{\beta_1\beta_2}{\ell}\ri)\le(z_{\beta_3} z_{\beta_4}-\Ti{\ker}{\beta_3\beta_4}{\ell}\ri)\ket_{\ker^{(\ell)}}\, \nonumber\\
&\quad +2n_{\ell} \bra \sigma_{\alpha_1}\sigma_{\alpha_2} \le(z_{\beta_1} z_{\beta_2}-\Ti{\ker}{\beta_1\beta_2}{\ell}\ri)\ket_{\ker^{(\ell)}} \Ti{\ker}{\beta_3\beta_4}{\ell}-2\bra \sigma_{\alpha_1}\sigma_{\alpha_2}\ket_{\ker^{(\ell)}}\Ti{\ker}{\beta_1\beta_3}{\ell}\Ti{\ker}{\beta_2\beta_4}{\ell}\Big]+\o{\frac{1}{n^2}}\, .\nonumber
\end{align}
 However, for the former contribution, the Gaussian term $\bra \sigma_{\alpha_1}\sigma_{\alpha_2}\ket_{g^{(\ell)}}$ needs be carefully separated into the leading and subleading pieces. To that end, we can trade the Gaussian expectation with $g^{(\ell)}$ for one in terms of the leading kernel $K^{(\ell)}$
\begin{align}\label{eq:subleading-sigma-sigma}
&\bra \sigma_{\alpha_1}\sigma_{\alpha_2}\ket_{g^{(\ell)}}\, \\
=&\frac{\bra \sigma_{\alpha_1}\sigma_{\alpha_2} \exp\!\le[-\frac{1}{2}\sum_{\beta_1,\beta_2}\le(\TI{g}{\beta_1\beta_2}{\ell}-\TI{\ker}{\beta_1\beta_2}{\ell}\ri)z_{\beta_1}z_{\beta_2} \ri]\ket_{\ker^{(\ell)}}}{\bra \exp\!\le[-\frac{1}{2}\sum_{\beta_1,\beta_2}\le(\TI{g}{\beta_1\beta_2}{\ell}-\TI{\ker}{\beta_1\beta_2}{\ell}\ri)z_{\beta_1}z_{\beta_2} \ri]\ket_{\ker^{(\ell)}}}\, \nonumber\\
=&\bra \sigma_{\alpha_1}\sigma_{\alpha_2}\ket_{\ker^{(\ell)}}-\frac{1}{2}\sum_{\beta_1,\beta_2}\le(\TI{g}{\beta_1\beta_2}{\ell}-\TI{\ker}{\beta_1\beta_2}{\ell}\ri)\bra \sigma_{\alpha_1}\sigma_{\alpha_2} \le(z_{\beta_1}z_{\beta_2}-\Ti{\ker}{\beta_1\beta_2}{\ell}\ri)\ket_{\ker^{(\ell)}}+\o{\frac{1}{n^2}}\, .\nonumber
\end{align}
Plugging~\eqref{eq:inverse-difference} into~\eqref{eq:subleading-sigma-sigma}, we obtain the subleading contribution due to the change in the quadratic coupling, giving
\begin{align}\label{eq:subleading-sigma-sigma-more-explicit}
&\bra \sigma_{\alpha_1}\sigma_{\alpha_2}\ket_{g^{(\ell)}}\, \\
=&\bra \sigma_{\alpha_1}\sigma_{\alpha_2}\ket_{\ker^{(\ell)}}+\frac{1}{2n_{\ell-1}}\TI{\ker}{\beta_1\beta_3}{\ell}\TI{\ker}{\beta_2\beta_4}{\ell}\se{\beta_3\beta_4}{\ell}\bra \sigma_{\alpha_1}\sigma_{\alpha_2} \le(z_{\beta_1}z_{\beta_2}-\Ti{\ker}{\beta_1\beta_2}{\ell}\ri)\ket_{\ker^{(\ell)}}\, \nonumber\\
&-\frac{1}{n_{\ell-1}}\sum_{\beta_1,\ldots,\beta_4}\le(\frac{n_{\ell}}{4} \TI{V}{(\beta_1\beta_2)(\beta_3\beta_4)}{\ell}+\frac{1}{2}\TI{V}{(\beta_1\beta_3)(\beta_2\beta_4)}{\ell}\ri)\Ti{\ker}{\beta_3\beta_4}{\ell}\bra \sigma_{\alpha_1}\sigma_{\alpha_2} \le(z_{\beta_1}z_{\beta_2}-\Ti{\ker}{\beta_1\beta_2}{\ell}\ri)\ket_{\ker^{(\ell)}}\, \nonumber\\
&+\o{\frac{1}{n^2}}\, .\nonumber
\end{align}\index{subleading corrections}

Now that we've computed everything, we can add the two contributions to $\E{\s{j}{\alpha_1}{\ell}\s{j}{\alpha_2}{\ell}}$, \eqref{eq:quartic-easy} and~\eqref{eq:subleading-sigma-sigma-more-explicit}, and plug them into the expression for the preactivation correlator~\eqref{eq:recursion-two-point-subleading}. Collecting terms, we recover the leading contribution, the recursion for the kernel
\be\label{eq:K-recursion}
\Ti{\ker}{\alpha_1\alpha_2}{\ell+1}=\Cb{\ell+1}+\CW{\ell+1}\bra \sigma_{\alpha_1}\sigma_{\alpha_2}\ket_{\ker^{(\ell)}}\, ,
\ee
and also find a recursion for the NLO metric\index{metric!next-to-leading-order correction} as promised\index{subleading corrections}\index{metric!next-to-leading-order correction}
\begin{align}\label{eq:NLO-metric-recursion}
&\frac{1}{n_{\ell}}\se{\alpha_1\alpha_2}{\ell+1}\, \\
=&\CW{\ell+1}\frac{1}{n_{\ell-1}}\sum_{\beta_1,\ldots,\beta_4\in\D}\Bigg[\frac{1}{2}\TI{\ker}{\beta_1\beta_3}{\ell}\TI{\ker}{\beta_2\beta_4}{\ell}\se{\beta_3\beta_4}{\ell}\bra \sigma_{\alpha_1}\sigma_{\alpha_2} \le(z_{\beta_1}z_{\beta_2}-\Ti{\ker}{\beta_1\beta_2}{\ell}\ri)\ket_{\ker^{(\ell)}}\, \nonumber\\
&\quad \quad \quad \quad \quad \quad \quad +\frac{1}{8} \TI{V}{(\beta_1\beta_2)(\beta_3\beta_4)}{\ell}\bra \sigma_{\alpha_1}\sigma_{\alpha_2} \le(z_{\beta_1}z_{\beta_2}-\Ti{\ker}{\beta_1\beta_2}{\ell}\ri) \le(z_{\beta_3}z_{\beta_4}-\Ti{\ker}{\beta_3\beta_4}{\ell}\ri)\ket_{\ker^{(\ell)}}\, \nonumber\\
&\quad \quad \quad \quad \quad \quad \quad +\frac{1}{4}\TI{V}{(\beta_1\beta_3)(\beta_2\beta_4)}{\ell}\Ti{\ker}{\beta_3\beta_4}{\ell}\bra \sigma_{\alpha_1}\sigma_{\alpha_2} \le(-2z_{\beta_1}z_{\beta_2}+\Ti{\ker}{\beta_1\beta_2}{\ell}\ri) \ket_{\ker^{(\ell)}}\Bigg]\, . \nonumber
\end{align}
In going through this calculation in your personal notes or on the margins of this book, you can explicitly see the cancellation of contributions from $n_{\ell}-1$ spectator neurons that does not participate in the expectation $\E{\s{j}{\alpha_1}{\ell}\s{j}{\alpha_2}{\ell}}$ 
as required by the marginalization rule for the $\ell$-th-layer action. Indeed, every term in the square bracket on the right-hand side of the equation~\eqref{eq:NLO-metric-recursion} is manifestly of order one.

\index{$1/n$ expansion}\index{subleading corrections}
This process can be systematically pushed to higher orders. Just as the computation of the NLO metric\index{metric!next-to-leading-order correction} $G_{\alpha_1\alpha_2}^{\le\{1\ri\}(\ell)}$ involved the leading quartic coupling, the computation of the subleading correction to the four-point vertex, $V_{(\alpha_1\alpha_2)(\alpha_3\alpha_4)}^{\le\{1\ri\}(\ell)}$, and the computation of the order $1/n^2$ correction to the two-point correlator, $G_{\alpha_1\alpha_2}^{\le\{2\ri\}(\ell)}$, would involve the leading sextic coupling. Such a sextic coupling appears at order $1/n^2$ in the action and contributes to the connected six-point function, which also vanishes as
$\o{1/n^2}$.\footnote{
For those familiar with field theory, the leading part of the couplings in the action are \emph{tree-level} contributions to correlators.
They are to be contrasted with \terminate{subleading corrections} to the two-point correlator discussed in this section, which included both \emph{loop-level} contributions from quartic interaction and tree-level contributions from the NLO correction to the bare quadratic coupling.\index{coupling!quadratic}\index{coupling!quartic}\index{interactions}}

\section{RG Flow and RG Flow}\index{RG flow|see{representation group flow}}\index{RG flow|see{renormalization group flow}}\index{RG flow and RG flow|textbf}
\label{sec:marginalization-group-flow}
Since the past five sections have been a whirlwind of equations, algebra, and integration, let's take a moment to recap and assemble the main results.

The goal of this chapter was to find the marginal distribution of preactivations $p\!\le(z^{(\ell)}\Big\vert\D\ri)$ in a given layer $\ell$  in terms of an \term{effective action}\index{action!effective|see{effective action}} with data-dependent couplings\index{data-dependent coupling}. These couplings change -- or \textbf{run}\index{running coupling|textbf} -- from layer to layer, and the running is determined via recursions, which in turn determine how the distribution of preactivations changes with depth.
Equivalently, these recursions tell us how correlators of preactivations evolve with layer.
In this language, starting with independent neurons in the first layer (\S\ref{sec:first-layer-gaussian}),
we saw how interactions among neurons are induced in the second layer (\S\ref{sec:second-layer-non-gaussian})
and then amplified in deeper layers  (\S\ref{sec:deeper-layer-accumulation}).

Concretely, let's summarize the behavior of finite-width networks to leading order in the wide-network expansion. Expressing the two-point correlator of preactivations in terms of the \textbf{kernel}\index{kernel} $\Ti{\ker}{\alpha_1\alpha_2}{\ell}$ as
\be\label{eq:leading-kernel-reminder}
\E{\z{i_1}{\alpha_1}{\ell}\z{i_2}{\alpha_2}{\ell}}=\delta_{i_1i_2}\Ti{G}{\alpha_1\alpha_2}{\ell}=\delta_{i_1i_2}\le[\Ti{\ker}{\alpha_1\alpha_2}{\ell}+\o{\frac{1}{n}}\ri] \, ,
\ee
and expressing the four-point connected correlator in terms of the \term{four-point vertex} $\Ti{V}{(\alpha_1\alpha_2)(\alpha_3\alpha_4)}{\ell}$  as
\begin{align}\label{eq:C4_MLPH_reprint}
&\E{\z{i_1}{\alpha_1}{\ell}\z{i_2}{\alpha_2}{\ell}\z{i_3}{\alpha_3}{\ell}\z{i_4}{\alpha_4}{\ell}}\Big\vert_{\text{connected}}\, \\
=&\frac{1}{n_{\ell-1}}\le[\delta_{i_1i_2}\delta_{i_3 i_4}\Ti{V}{(\alpha_1\alpha_2)(\alpha_3\alpha_4)}{\ell}+\delta_{i_1i_3}\delta_{i_2 i_4}\Ti{V}{(\alpha_1\alpha_3)(\alpha_2\alpha_4)}{\ell}+\delta_{i_1i_4}\delta_{i_2 i_3}\Ti{V}{(\alpha_1\alpha_4)(\alpha_2\alpha_3)}{\ell} \ri]\, , \nonumber
\end{align}
the running of these correlators is given by the recursions
\begin{align}
\label{eq:K-recursion-reprint}
\Ti{\ker}{\alpha_1\alpha_2}{\ell+1}=&\Cb{\ell+1}+\CW{\ell+1}\bra \sigma_{\alpha_1}\sigma_{\alpha_2}\ket_{\ker^{(\ell)}}\, ,\\
\label{eq:V-recursion-reprint}
V^{(\ell+1)}_{(\alpha_1\alpha_2)(\alpha_3\alpha_4)}=&\le(\CW{\ell+1}\ri)^2\Big[\bra\sigma_{\alpha_1} \sigma_{\alpha_2} \sigma_{\alpha_3} \sigma_{\alpha_4}\ket_{\ker^{(\ell)}}  - \bra \sigma_{\alpha_1} \sigma_{\alpha_2}\ket_{\ker^{(\ell)}} \bra \sigma_{\alpha_3} \sigma_{\alpha_4}\ket_{\ker^{(\ell)}} \Big]\, \\
&+\frac{1}{4}\le(\CW{\ell+1}\ri)^2\frac{n_{\ell}}{n_{\ell-1}}\sum_{\beta_1,\ldots,\beta_4\in\D}\TI{V}{(\beta_1\beta_2)(\beta_3\beta_4)}{\ell}\bra\sigma_{\alpha_1}\sigma_{\alpha_2} \le(z_{\beta_1} z_{\beta_2}-\Ti{\ker}{\beta_1\beta_2}{\ell}\ri)\ket_{\ker^{(\ell)}}\, \nonumber\\
&\quad \quad \quad \quad \quad \quad \quad \quad \quad \quad \quad \times\bra \sigma_{\alpha_3}\sigma_{\alpha_4} \le(z_{\beta_3} z_{\beta_4}-\Ti{\ker}{\beta_3\beta_4}{\ell}\ri)\ket_{\ker^{(\ell)}}+\o{\frac{1}{n}} \, ,\nonumber
\end{align}
where the indices on the four-point vertex are raised by the inverse metric $G_{(\ell)}$
\begin{align}
\TI{V}{(\alpha_1\alpha_2)(\alpha_3\alpha_4)}{\ell}\equiv& \sum_{\beta_1,\ldots,\beta_4\in\D}\TI{G}{\alpha_1\beta_1}{\ell}\TI{G}{\alpha_2\beta_2}{\ell}\TI{G}{\alpha_3\beta_3}{\ell}\TI{G}{\alpha_4\beta_4}{\ell}\Ti{V}{(\beta_1\beta_2)(\beta_3\beta_4)}{\ell}\, \\
=&\sum_{\beta_1,\ldots,\beta_4\in\D}\TI{\ker}{\alpha_1\beta_1}{\ell}\TI{\ker}{\alpha_2\beta_2}{\ell}\TI{\ker}{\alpha_3\beta_3}{\ell}\TI{\ker}{\alpha_4\beta_4}{\ell}\Ti{V}{(\beta_1\beta_2)(\beta_3\beta_4)}{\ell}+\o{\frac{1}{n}}\, .\nonumber
\end{align}
These recursions dictate how the statistics of preactivations  flow with depth.

This flow is very reminiscent of the following heuristic picture, which is offered as an explanation for how neural networks are supposed to work: given an input, such as the image of a cat,
the first few layers identify low-level features from the pixels -- such as the \texttt{edges} between areas of low and high intensity -- and then the middle layers assemble these low-level features into
mid-level features -- such as the texture and pattern of \texttt{fur} -- which are further aggregated in deeper layers into higher-level representations
-- such as \texttt{tails} and \texttt{ears} -- which the last layer combines into an estimate of the probability that original pixels represents a \texttt{cat}.
Indeed, some studies support this hierarchically-ordered arrangement of feature representation
in trained networks~\cite{Rob}.\footnote{It has been suggested that even untrained networks have features that can act as types of filters, effectively allowing for primitive edge detecting in untrained networks. For a related set of ideas, see~\cite{NIPS2007_3182}.} The desirability of such an arrangement emphasizes both the role and importance of depth in deep learning.

Some of the terms we used in discussing this heuristic picture can actually be given more precise definitions. For instance, each neuron in the network -- including not only those in the output layer but also those in the hidden layers -- is a scalar function of the input and called a \term{feature}. The neurons of a given layer can be organized into a vector-valued function of the input, which we'll refer to as a \term{representation}.\footnote{While the main focus of our study of supervised learning (\S\ref{ch:training}) will be understanding how the representation $z^{(L)}$ in the output layer is learned via gradient-based training,
it is also important to understand how representations 
are learned in hidden layers (\S\ref{ch:features}). In addition to being necessary components of determining the coarse-grained representation at the output,
in some applications of deep learning learned representations in the hidden layers can be used as inputs themselves for other learning tasks. This occurs quite often in \terminate{unsupervised learning}, for example with the word embeddings of natural language processing tasks. In these scenarios, the embeddings -- representations for an input word in the larger context of a full sentence -- typically are taken not just from the final layer, but from the concatenation of the final few layers. See e.g.~\cite{BERT2018}.}
In terms of these concepts, our formalism tracks the transformation of representations from one layer to the next. It is this flow of representations that we term \term{representation group flow} or \textbf{RG flow} for short.\footnote{
Two apologies are in order for the name \emph{representation group flow}\index{representation group flow!name}: \emph{(i)} it is confusingly close to the notion of \neo{group representation theory} in mathematics; and \emph{(ii)} the flow is technically a \neo{semigroup}\index{semigroup|seealso{RG flow}}, rather than a group. (Both group theory and semigroup theory are the studies of transformations but a group requires inverses while a semigroup does not; and the flow has no inverse.) 
This is just to repeat a historic mistake in physics as we'll explain further in a footnote below.} 
RG flow is induced via the repeated marginalization of fine-grained features in the shallow layers to give a coarse-grained representation in the output layer. Our notion of RG flow makes the heuristic picture given above concrete. %

This pattern of coarse-graining has a parallel in theoretical physics\index{physics}, known as \term{renormalization group flow} or \textbf{RG flow} for short. In this case, the RG flow is generated by the repeated marginalization of the microscopic fine-grained degrees of freedom in the system in order to obtain an \neo{effective theory} of the system in terms of macroscopic coarse-grained variables. Analogously, the physical couplings controlling the interactions of these effective degrees of freedom \emph{run}\index{running coupling} with the length scale at which they are probed -- e.g.~the effective charge of the electron will change when interrogated at different scales. Similar to the recursion equations describing the running couplings of the network representations, one can derive differential equations -- historically called beta functions -- that govern the running of the physical couplings with scale.\footnote{\emph{A brief history of renormalization in physics.}

Renormalization was originally developed in the 1930s and 1940s to deal with divergences -- infinities -- that plagued the calculations of experimental observables in quantum field theory. At first, these infinities were simply subtracted off -- swept under the rug, if you will -- yielding answers that, despite these shenanigans, matched extremely well with experiments. This whole state of affairs was considered embarrassing, leading to near abandonment of the theory. 

These divergences arose essentially due to a failure to properly take into account 
that couplings can be scale-dependent. The idea of running couplings was first put forth by Gell-Mann and Low~\cite{gellmanlow} in 1954,
however a full conceptualization of renormalization
wasn't available until Wilson developed the modern notion of RG flow~\cite{PhysRevB.4.3174,PhysRevB.4.3184} in 1971, offering a theoretical explanation for critical phenomena in statistical physics as well as giving a
sound grounding for the understanding of
divergences in quantum field theory.

At this point, all historical accounts of RG are contractually obligated to mention the following: the renormalization group is not a group; it's a semigroup. (The mistake was made in an early paper by Stueckelberg and Petermann, referring to the flow as a ``group of normalization''~\cite{Petermann:1953wpa}.)  Mathematically, this is because there are no
inverse elements; the marginalization of variables out of a joint distribution deletes information and cannot be undone. In particular, two different joint distributions can sometimes flow to the same distribution after marginalization. Intuitively, this is because these flows go from fine-grained descriptions to coarse-grained descriptions. (Such convergent flows lead to the notion of \neo{universality}, which we will explain in \S\ref{ch:signalprop} in the context of neural networks with different activations that flow to the same marginal distributions under RG.)

Clearly, RG flow in physics is a very rich subject. If you're interested in learning more, we recommend both~\cite{goldenfeld2018lectures,cardy_1996}.%
}

To make the connection between this RG flow and that RG flow abundantly clear, let's peek into how it is implemented in field theory in physics. In this scenario, the \terminate{degrees of freedom} are represented by a field $\phi\!\le(x\ri)$ that may take different values as a function of spacetime coordinate $x$. First, one divides $\phi\!\le(x\ri)$ into fine-grained variables $\phi^+$ consisting of high-frequency modes and coarse-grained variables $ \phi^-$ consisting of low-frequency modes, such that the field decomposes as $\phi\!\le(x\ri)=\phi^{+}\!\le(x\ri) + \phi^{-}\!\le(x\ri)$. The full distribution is governed by the full action
\be
\ac_{\text{full}}(\phi) = \ac(\phi^+) + \ac(\phi^-) + \SI(\phi^+,\phi^-)\, ,
\ee
where in particular the last term describes the interactions between these two sets of modes.

\index{coarse-graining}\index{effective action!as an effective theory}
Now, if all we care about are observables that depend only on the coarse-grained modes $\phi^{-}$ at macroscopic scales -- and such long-range scales are usually the relevant ones for experiments -- then this full description is too cumbersome to usefully describe the outcome of such experiments. In order to obtain an effective description in terms of only these coarse-grained variable $\phi^{-}$, we can integrate out (i.e.~marginalizes over) the fine-grained variables $\phi^{+}$ as
\be\label{eq:field-theory-marginalization}
e^{-\ac_{\text{eff}}(\phi^-)}=\int d\phi^{+}\ e^{-\ac_{\text{full}}(\phi)}\, ,
\ee
and obtain an \textbf{effective action}\index{effective action!in physics}\index{effective action!connection to RG flow} $\ac_{\text{eff}}(\phi^-)$, providing an \terminate{effective theory} for the observables of experimental interest. 
In practice, this marginalization is carried out scale by scale, dividing up the field as $\phi=\phi^{(1)}+\ldots+\phi^{(L)}$ from microscopic modes $\phi^{(1)}$ all the way to macroscopic modes $\phi^{(L)}=\phi^{-}$, and then integrating out the variables $\phi^{(1)}$, \ldots, $\phi^{(L-1)}$ in sequence. Tracking the flow of couplings in the effective action through this marginalization results in the aforementioned beta functions, and
in solving these differential equations up to the scale of interest, we get an effective description of observables at that scale.

This is precisely what we have been doing in this chapter for neural networks. The full field $\phi$ is analogous to a collection of all the preactivations $\le\{z^{(1)}, \ldots, z^{(L)}\ri\}$.
Their distribution is governed by the full joint distribution of preactivations
\be\label{eq:full-distribution-factorization}
p\!\le(z^{(1)}, \dots,  z^{(L)}\Big\vert \D\ri) = p\!\le(z^{(L)} \Big\vert z^{(L-1)} \ri) \cdots p\!\le(z^{(2)} \Big\vert z^{(1)} \ri) p\!\le(z^{(1)}\Big\vert \D \ri) \, ,
\ee
with the full action
\be\label{eq:full-action-decomposition}
 \ac_{\text{full}}\!\le(z^{(1)}, \dots,  z^{(L)}\ri)\equiv \sum_{\ell=1}^{L} \SGP\!\le(z^{(\ell)}\ri) + \sum_{\ell=1}^{L-1} \SI\!\le(z^{(\ell+1)}\Big\vert z^{(\ell)} \ri) \, .
\ee
Here, the full action is decomposed into the mean quadratic action for variables $z^{(\ell)}$
\be
\SGP\!\le(z^{(\ell)}\ri)=\frac{1}{2}\sum_{i=1}^{n_{\ell}}\sum_{\alpha_1,\alpha_2\in\D}\TI{G}{\alpha_1 \alpha_2}{\ell}\z{i}{\alpha_1}{\ell}\z{i}{\alpha_2}{\ell} \, 
\ee
in terms of the mean metric
$\Ti{G}{}{\ell}$, \eqref{eq:mean-metric-any-layer}, and the interaction between neighboring layers
\be
\SI\!\le(z^{(\ell+1)}\Big\vert z^{(\ell)} \ri)=\frac{1}{2}\sum_{i=1}^{n_{\ell+1}}\sum_{\alpha_1,\alpha_2\in\D}\le[\TI{\widehat{G}}{\alpha_1\alpha_2}{\ell+1}\le(z^{(\ell)}\ri)-\TI{G}{\alpha_1 \alpha_2}{\ell+1}\ri]\z{i}{\alpha_1}{\ell+1}\z{i}{\alpha_2}{\ell+1} \, .
\ee
Here we emphasized that the stochastic metric $\Ti{\widehat{G}}{\alpha_1\alpha_2}{\ell+1}$ is a function of $z^{(\ell)}$, and the induced coupling of $z^{(\ell)}$ with $z^{(\ell+1)}$ is what leads to the interlayer interactions.

Now, if all we care about are observables that depend only on the outputs of the network -- which includes a very important observable \ldots the output! -- then this full description is too cumbersome.
In order to obtain an effective (i.e.~useful) description of the distribution of outputs $z^{(L)}$, we can marginalizes over all the features $\le\{z^{(1)},\ldots,z^{(L-1)}\ri\}$ as
\be\label{eq:effective-action-full}
e^{-\ac_{\text{eff}}\le(z^{(L)}\ri)} = \int \le[ \prod_{\ell=1}^{L-1}dz^{(\ell)}\ri] e^{-\ac_{\text{full}}\le(z^{(1)}, \dots,  z^{(L)}\ri)}\, ,
\ee
just as we integrated out the fine-grained modes $\phi^{+}$ in~\eqref{eq:field-theory-marginalization} to get the effective description in terms of coarse-grained modes $\phi^{-}$.
And, just like in the field theory example, rather than carrying out this marginalization all at once, we proceeded sequentially, integrating out the preactivations layer by layer.
This resulted in the recursion relations \eqref{eq:K-recursion-reprint} and \eqref{eq:V-recursion-reprint}, and
in solving these recursion relations up to the depth of interest, we get an effective description of neural network output at that depth.\footnote{Note to physicists: the flow in networks from input to output is a flow from the ultraviolet\index{ultraviolet (RG flow)} to the infrared\index{infrared (RG flow)}.}

Now, this last sentence suggests a subtle but interesting shift of perspective, so let us elaborate.
So far in this chapter, we have implicitly assumed a fixed network depth $L$ and described how the preactivation distribution changes as an input $x$ propagates through the intermediate layers, yielding recursion relations for correlators and couplings for the evolution from layer $\ell$ to layer $\ell+1$, for $\ell=0,\ldots,L-1$.
However, it is also valid to view the resulting recursion equations as governing the change in output distributions as the overall network depth changes from $L$ to $L+1$.\footnote{To be precise, the output dimension $n_{\text{out}}$ is fixed. So, as the depth changes from $L$ to $L+1$, we imagine holding fixed the widths for $\ell<L$, inserting a new layer $L$ with $n_L \sim n\gg1$, and then setting the final layer $L+1$ to have width $n_{\text{out}}$.}
In other words, these recursion relations describe the effect of adding an additional layer to the neural network by comparing distributions $p\!\le(z^{(L)}\Big\vert\D\ri)$ and $p\!\le(z^{(L+1)}\Big\vert\D\ri)$.

\index{representation group flow}
Given this perspective, our RG flow can address head-on the effect of the \emph{deep} in \emph{deep learning}.
For instance, as a network get deeper, do the interactions between neurons -- encoded in the finite-width corrections such as the four-point vertex $V^{(\ell)}$ -- get amplified or attenuated?
In the language of RG flow,  couplings that grow with the flow are called \textbf{relevant}\index{relevant (RG flow)|textbf} and those that shrink
are called \textbf{irrelevant}\index{irrelevant (RG flow)|textbf}.\footnote{Couplings that neither grow nor shrink are called \textbf{marginal}\index{marginal (RG flow)|textbf}.}
These names are evocative of whether the interaction matters or not for the effective theory, and so we'll employ the same terminology.
Thus, to explore the effect of depth on the neuron-neuron interactions, we are simply asking whether the four-point vertex  $V^{(\ell)}$ is relevant or irrelevant.

This question has important implications for deep learning. If all the finite-width couplings were irrelevant, then finite-width networks would asymptote to infinite-width architectures under RG flow.
This would then mean that these networks behave more like infinite-width models as they get deeper, and so deep learning would really be the study of these much simpler Gaussian models.
Fortunately
we'll soon find that the couplings \emph{are} relevant, making our life richer, albeit more complicated. In the next chapter, we'll show that finite networks deviate more and more from their infinite-width counterparts as they get deeper. This has important practical consequences in controlling the instantiation-to-instantiation fluctuations in supervised training 
and also in allowing networks to learn nontrivial representations of their input (\S\ref{ch:features}).

The next chapter explores these relevant questions by explicitly solving recursion equations such as~\eqref{eq:K-recursion-reprint} and~\eqref{eq:V-recursion-reprint}.

%% file: Chp5-SignalProp/5_global.tex
\chapter{Effective Theory of Preactivations at Initialization}\label{ch:signalprop}
\label{ch:eft-mlp}

\epigraph{We believe this realm of work to be immensely important and rich, but we expect its growth to require a degree of critical analysis that its more romantic advocates have always been reluctant to pursue \dots.}{
Minsky\index{Minsky, Marvin} and 
Papert\index{Papert, Seymour} in the prologue to their 1988 expanded edition of \emph{Perceptrons} \cite{miskypapertperceptron}.}

\noindent{}The key defining feature of \terminate{deep learning} is the stacking of components on top of each other in order to get a \emph{deep} \terminate{neural network} \terminate{architecture}.
Despite the empirical preference for deeper networks, it's not at all obvious \emph{why} deep is good for learning. For a fixed number of neurons per layer, deep implies many more parameters, and often in deep learning more parameters lead to better performance. But there are other ways to include more parameters. For instance, why not just have a single hidden layer that is very wide?
In fact, in the strict infinite-width limit, such a single-layer model has the same number of parameters as any deeper MLP: infinity.

The proper way to think about the effects of depth is not to just count the number of model parameters, but instead to ask what happens when we add an additional layer to our MLP. In~\S\ref{ch:ngp}, we developed a formalism to address exactly this question through recursions for observable quantities of interest, enabling us to compute how the distributions of initial network outputs change upon adding a layer.
What we need, then, is a tool to effectively extract the explicit depth dependence from these recursions.

Building on the effective theory formalism developed in~\S\ref{ch:ngp}, in this chapter we'll extend the criticality and fluctuation analyses performed in~\S\ref{ch:deep-linear-eft} to MLPs with any nonlinear activation function.
Enlightened by the success of the previous chapter in finding simplification in the wide regime ($n \gg 1$), we now seek additional simplicity in the limit of large depth ($L \gg 1$).\footnote{What this means -- in light of the discussion in \S\ref{sec:solution_DLN} -- is that we take the limit of large width \emph{first} and then look at the limit of large depth.}
We'll first analyze the limit of infinite number of neurons per layer,
and then back off this limit to consider networks of large but finite width and depth. The result will be explicit expressions for the two-point and four-point correlators of preactivations in these asymptotic limits.\footnote{Despite the asymptotic nature of these solutions, we note many of these tools were developed to study the strong interactions, where a parameter that is $3$ in practice is taken to be infinity. Thus, sometimes even $3 \sim \infty$, since $1/3 \ll 1$ can be made to work as perturbative parameter \cite{Coleman:1985rnk}.
}

This will let us address the question of what happens to input signals as they propagate through the many layers of deep neural networks at initialization (\S\ref{sec:bootstrapping}).
We'll come to understand that the order-one values of the \terminate{initialization hyperparameters} -- i.e.~the bias variances $\Cb{\ell}$ and the rescaled weight variances $\CW{\ell}$ -- have pronounced qualitative effects on the behavior of the observables, just as we saw in \S\ref{ch:deep-linear-eft} for deep linear networks.
In particular, we'll explain how neural-network behavior becomes increasingly sensitive to these \terminate{initialization hyperparameters} with increasing depth.\footnote{The initial part of this analysis  was first carried out in a series of papers, \cite{poole2016exponential,raghu2017expressive,schoenholz2016deep}, using a different set of techniques that are ultimately equivalent to ours in the infinite-width limit.  Extending this analysis, we'll identify two very general  conditions according to the \emph{principle of criticality}\index{criticality!principle of} that let us determine the correct order-one values for the initialization hyperparameters.
In \S\ref{sec:generalization-at-infinity}, we'll see that the need for these conditions can also be understood by demanding that fully-trained networks generalize well. 
}

Such a tuning brings a network to \neo{criticality}, a term we borrow from statistical physics used to describe self-similar systems.
To this end, we give a general prescription for tuning \terminate{initialization hyperparameters} to their critical values for a given activation function and network architecture (\S\ref{sec:scale-invariant-eft} and~\S\ref{sec:non-scale-invariant-eft}). In the process, we also identify some activation functions that don't allow for criticality.
We will also see that certain activation functions behave very similarly to each other when tuned to criticality, highlighting an important connection to the notion of \neo{universality} in statistical physics.\index{statistical physics}

\index{effective theory of deep learning}
The study of finite-width corrections at criticality then leads us to one of the main results of this chapter, an \neo{emergent scale} given by the aspect ratio of the network depth to the network width, $L/n$ (\S\ref{sec:signal_prop_finite_width}).
This aspect ratio ultimately
serves as the 
\emph{cutoff}\index{cutoff, effective theory} of our effective theory, %
controlling the region of validity of our effective theory as well as determining the strength and importance of the finite-width corrections to the infinite-width description.
On the one end of the spectrum, we find that the shorter and fatter networks are, the more and more that they behave like their infinite-width counterparts. 
On the other end, skinny and tall networks become increasingly dominated by non-Gaussian \neo{fluctuations} due to \neo{interactions} between neurons. Overall, this serves to generalize our fluctuation analysis for deep linear networks\index{deep linear network} in \S\ref{sec:fluctuations_DLN}.

Lastly, we'll  conclude the chapter by addressing and resolving a subtlety that arises in our criticality analysis for non-smooth activation functions such as the $\relu$ (\S\ref{sec:finite_angle}).

\section{Criticality Analysis of the Kernel}\label{sec:bootstrapping}
For the bulk of this chapter, our goal is to extend the notion of criticality discussed in~\S\ref{ch:deep-linear-eft} to deep MLPs with general activation functions $\sigma(z)$. Our starting point is the kernel recursion
\be\label{eq:kernel-recursion-reminder}
\Ti{\ker}{\alpha\beta}{\ell+1} = C_b + C_W \braket{\sigma_{\alpha}\sigma_{\beta}}{\ell} \, ,
\ee
derived in the previous chapter. %
As a reminder,
the \neo{kernel} $\Ti{\ker}{\alpha\beta}{\ell}$, defined by~\eqref{eq:leading-kernel-reminder}, is the infinite-width limit of the mean metric $\Ti{G}{\alpha\beta}{\ell}$. Here, in order to restrict the number of distinct hyperparameters, we have set the bias variance $C_b^{(\ell)}=C_b$ and the rescaled weight variance $C_W^{(\ell)}=C_W$ to be layer-independent.
The initial condition for this recursion is given by the first-layer kernel %
\be\label{eq:kernel-initial-condition}
\Ti{\ker}{\alpha\beta}{1}=C_b+C_W\!\le(\frac{1}{n_0}\sum_{i=1}^{n_0}\x{i}{\alpha}\x{i}{\beta}\ri)\, ,
\ee 
set by the inner products of inputs $\sum_{i=1}^{n_0} \x{i}{\alpha}\x{i}{\beta}$. Our goal is to analyze how the kernel changes as a function of layer and, as we'll see, this analysis at the level of the kernel is sufficient to pin down the critical initialization hyperparameters to leading order in $1/\text{width}$.\index{initialization hyperparameters!critical}

For deep linear networks\index{deep linear network}, the Gaussian expectation of the activations  with respect to the kernel is just given by the kernel, $\braket{\sigma_{\alpha}\sigma_{\beta}}{\ell}=\braket{z_{\alpha}z_{\beta}}{\ell}=\Ti{\ker}{\alpha\beta}{\ell}$, and the recursion equation was simple enough for us to obtain a full solution. Stepping outside the realm of the $\linear$ activation function, however, the kernel recursion acquires two new complications that require some care: (i) the expectation value $\braket{\sigma_{\alpha}\sigma_{\beta}}{\ell}$ will be a nonlinear function of the kernel, and (ii) for two distinct inputs $\alpha\ne\beta$ the recursion mixes off-diagonal components $\Ti{\ker}{\alpha\beta}{\ell}$ with diagonal components $\Ti{\ker}{\alpha\alpha}{\ell}$ and $\Ti{\ker}{\beta\beta}{\ell}$.

\index{activation function!quadratic@\texttt{quadratic}}
Let's illustrate this with a $\texttt{quadratic}$ activation function $\sigma(z)=z^2$. 
As should be second nature by now, evaluating \eqref{eq:kernel-recursion-reminder} requires two pairs of Wick contractions\index{Wick contraction}, giving
\begin{align}
\Ti{\ker}{\alpha\beta}{\ell+1} &= C_b + C_W \braket{z_{\alpha}^2 z_{\beta}^2}{\ell} \,  \\
&=C_b+C_W \le(\Ti{\ker}{\alpha\alpha}{\ell}\Ti{\ker}{\beta\beta}{\ell}+2\Ti{\ker}{\alpha\beta}{\ell}\Ti{\ker}{\alpha\beta}{\ell}\ri)\, .\nonumber
\end{align}
Thus, unlike deep linear networks\index{deep linear network}, for quadratic activations $\Ti{\ker}{\alpha\beta}{\ell+1}$ depends not only on $\Ti{\ker}{\alpha\beta}{\ell}$ but also on $\Ti{\ker}{\alpha\alpha}{\ell}$ and $\Ti{\ker}{\beta\beta}{\ell}$, requiring us to solve three coupled nonlinear recursion equations for $\alpha\ne\beta$.
This mixing is generic for nonlinear activation functions.

For practitioners, this is good news: the off-diagonal elements of the kernel are related to the generalization ability of the network. For deep linear networks\index{deep linear network} the lack of mixing via nonlinearity suggests an \neo{inductive bias} that limits such networks' ability to develop nontrivial correlations for pairs of samples. %
While mixing is a benefit in practice, it's an obstacle in theory, albeit a surmountable one. Since the kernel recursion mixes at most two inputs, it is sufficient to analyze the case with a single input and the case with two distinct inputs. We shall now perform these analyses in turn, with an eye towards deriving the general conditions for criticality.

\subsubsection{A single input}
Each diagonal component of the kernel can be solved self-consistently by itself. Specifically, labeling a single input by $\alpha=\M$,
\begin{align}\label{eq:kernel-diagonal}
\Ti{\ker}{\M\M}{\ell+1}& = C_b + C_W\braket{\sigma(z_{\M})\,\sigma(z_{\M})}{\ell} \,  \\
&=C_b + C_W g\!\le(\Ti{\ker}{\M\M}{\ell}\ri) \, . \nonumber
\end{align}
Here, we introduced a helper function
\be\label{eq:helper_first}
g\!\le(K\ri)\equiv \le\langle \sigma(z)\, \sigma(z)\ri\rangle_{K} \equiv \frac{1}{\sqrt{2\pi\ker}} \int_{-\infty}^{\infty}d z\  e^{-\frac{z^2}{2\ker}}\, \sigma(z)\, \sigma(z)  \, ,
\ee
to emphasize that the expectation $\braket{\sigma(z_{\M})\sigma(z_{\M})}{\ell}$ is a function only of a single component of the kernel, $\Ti{\ker}{\M\M}{\ell}$.
By focusing our attention first on the single-input kernel, we can deal with the nonlinearity before confronting the mixing of kernel components.

\index{fixed point!trivial}
In particular, the single-input recursion~\eqref{eq:kernel-diagonal} is really telling us how the average magnitude of the preactivations for the input
\be\label{eq:midpoint-kernel-def}
\Ti{\ker}{\M\M}{\ell} = \mathbb{E}\le[\frac{1}{n_{\ell}}\sum_{i=1}^{n_{\ell}}\le(\z{i}{\M}{\ell}\ri)^2\ri]\, ,
\ee
changes as a function of layer $\ell$, with the initial condition for the recursion set by \eqref{eq:kernel-initial-condition}.
For the same reasons we considered in~\S\ref{sec:criticality_DLN}, we would like that the kernel $\Ti{\ker}{\M\M}{\ell}$ neither exponentially explode nor exponentially vanish.
However, such exponential behavior is generic, and so for most choices of initialization hyperparameters $\le(C_b,C_W\ri)$, the kernel will either explode exponentially towards a \emph{trivial} fixed point at infinity or collapse exponentially onto a trivial fixed point at a finite value $\Tif{\ker}{\M\M}$. Thus, our first \terminate{criticality} condition is to mitigate this exploding or collapsing kernel problem for the single input.\index{exploding and vanishing kernel problem}

There is an ancient technique used to analyze nonlinear recursions such as the single-input kernel recursion~\eqref{eq:kernel-diagonal}: linearization around a fixed point.
Namely,
we first identify a fixed point of the recursion, i.e.~a value $\Tif{\ker}{\M\M}$ that satisfies
\be\label{eq:fixed_single}
\Tif{\ker}{\M\M}=C_b + C_W g\!\le(\Tif{\ker}{\M\M}\ri)\, ,
\ee
and then expand the kernel around it as 
\be\label{eq:fixed-point-expansion-Delta-K}
\KML=\Tif{\ker}{\M\M}+\Delta \KML. 
\ee
This expansion for the single-input recursion~\eqref{eq:kernel-diagonal} results in the linearized recursion
\be\label{eq:single_expand}\index{kernel!linearized recursion}
\Ti{\Delta \ker}{\M\M}{\ell+1}=\chi_{\parallel}\!\le(\Tif{\ker}{\M\M}\ri)\Ti{\Delta \ker}{\M\M}{\ell}+O\!\le(\Delta^2\ri)\, , %
\ee
where we introduced the \term{parallel susceptibility}\index{susceptibility!parallel|see{parallel susceptibility}}\index{susceptibility!perpendicular|see{perpendicular susceptibility}}
\begin{align}\label{eq:chi_parallel_first}
\chi_{\parallel}(\ker)&\equiv C_W g'\!\le(\ker\ri)\, \\
&=C_W \frac{d}{d \ker}\le[\frac{1}{\sqrt{2\pi\ker}}\int_{-\infty}^{\infty} dz\,  e^{-\frac{z^2}{2K}} \sigma(z)\, \sigma(z)\ri]\, \notag \\
&=\frac{C_W}{2\ker^2} \bra \sigma(z)\,\sigma(z)\le(z^2-K\ri)\ket_{\ker}\, .\nonumber
\end{align}
The susceptibility $\chi_{\parallel}(\ker)$ characterizes how susceptible the kernel is to perturbations around the fixed point, hence the name: the kernel value exponentially expands away from or contracts towards the fixed-point value, according to whether $ \chi_{\parallel}\!\le(\Tif{\ker}{\M\M}\ri)>1$ or $\chi_{\parallel}\!\le(\Tif{\ker}{\M\M}\ri)<1$. (The label \emph{parallel} will be explained at the very end of this section.) 

Thus, we see that in order to mitigate this \neo{exploding and vanishing kernel problem} for a single input at linear order, we require the tuning of \terminate{initialization hyperparameters} $\le(C_b, C_W\ri)$ such that
\be\label{eq:parallel_criticality}
 \chi_{\parallel}\!\le(\Tif{\ker}{\M\M}\ri)=1\, ,
\ee
with the fixed-point value $\Tif{\ker}{\M\M}$ 
defined implicitly through the fixed-point equation~\eqref{eq:fixed_single}.
\index{scale invariance}
As we shall detail later, \terminate{criticality} can happen in three ways depending on the choice of activation functions. 
\bi
\item First, as we saw for deep linear networks\index{deep linear network}, the single-input kernel can be perfectly preserved as $\Ti{\ker}{\M\M}{\ell} =\Ti{\ker}{\M\M}{1}=C_b+C_W\le(\sum_i \x{i}{\M}\x{i}{\M}/n_{0}\ri)$,
resulting in a line of fixed points parametrized by input norms $\sum_i \x{i}{\M}\x{i}{\M}$. We will see this happening in~\S\ref{sec:scale-invariant-eft} for \emph{scale-invariant} activation functions due to the absence of higher-order corrections $O\!\le(\Delta^{p>1}\ri)$ in~\eqref{eq:single_expand}. 
\item Second, the kernel can slowly decay toward a fixed point $\Tif{\ker}{\M\M}=0$ for all input norms, with a power law $\Ti{\ker}{\M\M}{\ell}\sim1/\ell^q$ with $0<q\leq 1$. We will see this happening in~\S\ref{subsec:tanh_univ} for a class of activation functions that include $\tanhA$ and $\sinA$, due to the presence of $O\!\le(\Delta^{p>1}\ri)$ corrections in~\eqref{eq:single_expand}. 
\item Third, \terminate{criticality} can happen with a power-law decay towards a nonzero fixed-point value $\Tif{\ker}{\M\M}\ne0$. We will see this happening in~\S\ref{subsec:half_stability} for the $\swish$ and $\gelu$ activation functions.
\ei
In all these cases, when \terminate{initialization hyperparameters} are tuned to \terminate{criticality}, we call $\Tif{\ker}{\M\M}=\Tif{\ker}{\M\M}\!\le(C_b^{\text{critical}}, C_W^{\text{critical}}\ri)$ a \textbf{nontrivial fixed point}\index{fixed point!nontrivial|textbf}\index{nontrivial fixed point|see{fixed point}}, distinguishing it from trivial fixed points for generic hyperparameters around which perturbations behave exponentially.\index{fixed point!trivial}

\subsubsection{Two inputs}
Now, given two distinct inputs, let's label them with sample indices $\alpha=\pm$. For such a pair of inputs, we have three distinct kernel components to consider: $\Ti{\ker}{++}{\ell}$,  $\Ti{\ker}{--}{\ell}$, and $\Ti{\ker}{+-}{\ell}=\Ti{\ker}{-+}{\ell}$.
The single-input analysis can be directly applied to determine the 
layer dependence of the diagonal components $\Ti{\ker}{++}{\ell}$ and $\Ti{\ker}{--}{\ell}$, so to complete our analysis we need to extract the layer dependence of the off-diagonal component  $\Ti{\ker}{+-}{\ell}$, given solutions for the diagonal pieces.
Such an analysis will yield a second \terminate{criticality} condition that, together with~\eqref{eq:parallel_criticality}, will pin down the critical initialization hyperparameters $\le(C_b, C_W\ri)^{\text{critical}}$ for a given activation function. \index{initialization hyperparameters!critical}

Ultimately, our approach will be to linearize around the degenerate limit where both inputs coincide identically, i.e., $\x{i}{+},\x{i}{-}\to\x{i}{\M}$.
In such a limit, all the ways of pairing up the two inputs are the same, and so all the components of the full kernel matrix must take the same value,
i.e., $\Ti{\ker}{++}{\ell}, \Ti{\ker}{--}{\ell}, \Ti{\ker}{+-}{\ell} \to \Ti{\ker}{\M\M}{\ell}$. Thus, each recursion degenerates to the same single-input recursion \eqref{eq:kernel-diagonal}, which we know has a fixed-point value $\Tif{\ker}{\M\M}$. 
This means that the coincident-limit solution,
\be\label{eq:coincidental-solution}
\begin{pmatrix}
\Ti{\ker}{++}{\ell} & \Ti{\ker}{+-}{\ell} \\
\Ti{\ker}{-+}{\ell}  & \Ti{\ker}{--}{\ell} 
\end{pmatrix}=\begin{pmatrix}
\Tif{\ker}{\M\M} & \Tif{\ker}{\M\M} \\
\Tif{\ker}{\M\M}  & \Tif{\ker}{\M\M}
\end{pmatrix}
=\Tif{\ker}{\M\M}\begin{pmatrix}
1 & 1 \\
1  & 1
\end{pmatrix}\, ,
\ee
must also be a fixed point of the full kernel recursion for the two inputs.

There are three different kinds of perturbations that we need to consider in order to understand the approach of the full kernel matrix to this degenerate fixed point. 
The first kind corresponds to the perturbation $\Delta K^{(\ell)}_{00}$ that appeared in our single-input analysis, which controls how the average of the kernel's four components approaches the fixed-point value $\Tif{\ker}{\M\M}$. Next, in backing off the coincident limit, the single input splits into two distinct inputs, $\x{i}{\M} \to \x{i}{+},\x{i}{-}$. Then, we could imagine separating these two inputs so that they become endowed with different magnitudes, i.e.~$\sum_i \x{i}{+}^2 \neq \sum_i \x{i}{-}^2$, and follow the expected evolution of this difference through the network:
\be\label{eq:radial-distance-def}
R^{(\ell)}  \equiv \E{\frac{1}{n_{\ell}}\sum_{i=1}^{n_{\ell}}\le(\z{i}{+}{\ell}\ri)^2} -\E{\frac{1}{n_{\ell}}\sum_{i=1}^{n_{\ell}}\le(\z{i}{-}{\ell}\ri)^2} = \Ti{\ker}{++}{\ell}-\Ti{\ker}{--}{\ell} \, .
\ee
Such a perturbation is actually still covered by the single-input analysis, since the evolutions of the diagonal components $\Ti{\ker}{++}{\ell}$ and $\Ti{\ker}{--}{\ell}$ are mutually independent of each other, with their approach to the fixed point simply controlled by the single-input recursion.

Finally, rather than considering the difference of the squares, we could consider the square of the difference
\be\label{eq:distance-def}
D^{(\ell)}  \equiv \E{\frac{1}{n_{\ell}}\sum_{i=1}^{n_{\ell}}\le(\z{i}{+}{\ell}-\z{i}{-}{\ell}\ri)^2} = \Ti{\ker}{++}{\ell}+\Ti{\ker}{--}{\ell}-2\Ti{\ker}{+-}{\ell} \, ,
\ee
where to get the expression on the right-hand side we expanded the binomial and then used the definition of the kernel components.
This quantity measures the magnitude of the difference between the two inputs after being passed through $\ell$ layers of the network and can be non-vanishing even when such inputs themselves have the same magnitudes, i.e.~$\sum_i \x{i}{+}^2 = \sum_i \x{i}{-}^2$.
Importantly, this distance measure $D^{(\ell)}$ depends on the 
off-diagonal component of the kernel, $\Ti{\ker}{+-}{\ell}$, and so we expect that analyzing this perturbation will give something new. As we will see, the approach of this third perturbation $D^{(\ell)}$ to the coincident fixed point with $D^{(\ell)}=0$ will yield a second \terminate{criticality} condition.  Together with the single-input \terminate{criticality} condition~\eqref{eq:parallel_criticality}, this will be sufficient to completely determine the critical initialization hyperparameters.\index{initialization hyperparameters!critical}

Let's translate the above discussion into math. To do so, we will find it convenient to
project
the full kernel matrix into the following basis\index{$\gamma^{[a]}$ basis|textbf}\index{kernel!$\gamma^{[a]}$ basis|see{$\gamma^{[a]}$ basis}}
\be\label{eq:kernel-expand-gamma}
\Ti{\ker}{\alpha_1\alpha_2}{\ell}=\begin{pmatrix}
\Ti{\ker}{++}{\ell} & \Ti{\ker}{+-}{\ell} \\
\Ti{\ker}{-+}{\ell}  & \Ti{\ker}{--}{\ell} 
\end{pmatrix}=\Ti{\ker}{[0]}{\ell}\gamma^{[0]}_{\alpha_1\alpha_2}+\Ti{\ker}{[1]}{\ell}\gamma^{[1]}_{\alpha_1\alpha_2}+\Ti{\ker}{[2]}{\ell}\gamma^{[2]}_{\alpha_1\alpha_2} \, ,
\ee
where we've introduced symmetric matrices
\be\label{eq:gamma_def}
\gamma^{[0]}_{\alpha_1\alpha_2}\equiv\begin{pmatrix}
1 & 1 \\
1  & 1
\end{pmatrix}
\, , \ \ \ 
\gamma^{[1]}_{\alpha_1\alpha_2}\equiv\begin{pmatrix}
1 & 0 \\
0  & -1
\end{pmatrix}
\, , \ \ \ 
\gamma^{[2]}_{\alpha_1\alpha_2}\equiv\begin{pmatrix}
1 & -1 \\
-1  & 1
\end{pmatrix}
\, .
\ee
In this basis, the components of the kernel are
\begin{align}
\Ti{\ker}{[0]}{\ell} &= \frac{1}{4}\le[\Ti{\ker}{++}{\ell}+\Ti{\ker}{--}{\ell} + 2\Ti{\ker}{+-}{\ell}  \ri] = \E{\frac{1}{n_{\ell}}\sum_{i=1}^{n_{\ell}}\le(\frac{\z{i}{+}{\ell}+\z{i}{-}{\ell}}{2}\ri)^2} \, , \label{eq:K0-decomposition}\\
\Ti{\ker}{[1]}{\ell} &= \frac{1}{2}\le[\Ti{\ker}{++}{\ell}-\Ti{\ker}{--}{\ell} \ri] = \frac{1}{2}R^{(\ell)} \, ,\label{eq:K1-decomposition}\\
\Ti{\ker}{[2]}{\ell} &= \frac{1}{4}\le[\Ti{\ker}{++}{\ell}+\Ti{\ker}{--}{\ell} - 2\Ti{\ker}{+-}{\ell}  \ri] = \frac{1}{4} D^{(\ell)} \, . \label{eq:K2-decomposition}
\end{align}
This basis was strategically chosen so that both $\Ti{\ker}{[1]}{\ell} $ and $\Ti{\ker}{[2]}{\ell}$ correspond to our two natural distance measures of two distinct inputs -- the difference in the magnitudes $R^{(\ell)}$ and the magnitude of the difference $D^{(\ell)}$ -- and both vanish in the coincident limit. The remaining component, $\Ti{\ker}{[0]}{\ell}$, measures the overall average magnitude or the magnitude of the center of mass of the two $\ell$-th-layer preactivations.

This basis has two additional nice properties that will later prove useful: (i) both $\Ti{\ker}{[0]}{\ell}$ and $\Ti{\ker}{[2]}{\ell}$ are even (invariant) under the parity swap of two inputs $+\leftrightarrow -$, while $\Ti{\ker}{[1]}{\ell}$ is odd, changing its sign $\Ti{\ker}{[1]}{\ell}\rightarrow -\Ti{\ker}{[1]}{\ell}$ as $+\leftrightarrow -$, and (ii) the $\gamma^{[a]}$ matrices are orthogonal.\footnote{This symmetry decomposition mirrors tensorial decomposition\index{tensor decomposition!$\gamma^{[a]}$ basis|see{$\gamma^{[a]}$ basis}} used by physicists to organize particles by their spin\index{spin}.
Operationally, we can project out the components of any $2 \times 2$ matrix $M_{\alpha\beta}$ into components $M_{[a]}$ in the $\gamma^{[a]}$ basis by tracing over the \terminate{sample indices} and normalizing
\be\label{eq:trace-projection}
M_{[a]} = \frac{
    \sum_{\alpha, \beta} M_{\alpha \beta} \gamma^{[a]}_{\beta \alpha}
}{
    \sum_{\alpha, \beta} \gamma^{[a]}_{\alpha \beta} \gamma^{[a]}_{\beta \alpha} 
} \, ,
\ee
with $\alpha, \beta \in \{+, - \}$.  One can easily check that the $\gamma^{[a]}_{\alpha\beta}$ matrices themselves are orthogonal under this inner product \eqref{eq:trace-projection}.}

Now, let's discuss perturbations around the coincident fixed point~\eqref{eq:coincidental-solution} in terms of this new basis.
These perturbations have a very natural interpretation in terms of two infinitesimally-separated input points, $\x{i}{+}$ and $\x{i}{-}$, perturbed around a \term{midpoint input} $\x{i}{\M} \equiv (\x{i}{+} + \x{i}{-})/2$ as
\be\label{eq:x-pm}
\x{i}{\pm}=\x{i}{\M}\pm\frac{1}{2}\delta x_i\, .
\ee
The dynamics of such preactivations $\z{i}{\pm}{\ell}\equiv z_i^{(\ell)}(x_\pm)$ then
encode the evolution of these perturbed signals through the network as a function of layer depth $\ell$.
The coincident limit corresponds to $\delta x_i\rightarrow0$ and, as we back off from this limit, we should be able to expand the $\Ti{\ker}{[a]}{\ell}$ components around their coincident-limit values $\Ti{\ker}{[0]}{\ell}=\Ti{\ker}{\M\M}{\ell}$ and $\Ti{\ker}{[1]}{\ell}=\Ti{\ker}{[2]}{\ell}=0$.
This results in an expansion\index{kernel!$\delta$ expansion|see{$\delta$ expansion}}\index{$\delta$ expansion|textbf}
\begin{align}\label{eq:kernel-expand-1}
\Ti{\ker}{[0]}{\ell}&=\KML+\Ti{\delta\delta\ker}{[0]}{\ell}+\o{\delta^4}\, ,\\
\Ti{\ker}{[1]}{\ell}&=\Ti{\delta\ker}{[1]}{\ell}+\Ti{\delta\delta\delta\ker}{[1]}{\ell}+\o{\delta^5}\, , \label{eq:kernel-expand-2}\\
\Ti{\ker}{[2]}{\ell}&=\Ti{\delta\delta\ker}{[2]}{\ell}+\Ti{\delta\delta\delta\delta\ker}{[2]}{\ell}+\o{\delta^6}\, , \label{eq:kernel-expand-3}
\end{align}
where the order of the kernel perturbation in $\delta x$ is denoted by a preceding $\delta^p$, and we used the even/odd behavior of the components $\Ti{\ker}{[a]}{\ell}$ under the parity symmetry $+\leftrightarrow -$ to limit which terms appear in each expansion.\footnote{\label{foot:kink1}These expansions are valid when the activation function $\sigma(z)$ is sufficiently smooth. For non-smooth activation functions such as $\relu$, these expansions are more complicated, though still analyzable. We will consider these subtleties in more detail in \S\ref{sec:finite_angle}.
}
Next, we will use these expansions to determine whether the behavior of the leading perturbations $\Ti{\delta\ker}{[1]}{\ell}$ and $\Ti{\delta\delta\ker}{[2]}{\ell}$ around their fixed point values are exponential, power-law, or constant. 

To do so, we'll need to expand the original kernel recursion~\eqref{eq:kernel-recursion-reminder} order by order in $\delta$.
Before embarking on such an algebraic journey, let's think about what we should expect.
At the zeroth order in $\delta$, we'll just recover the recursion for the single-input kernel~\eqref{eq:kernel-diagonal}
\be
\Ti{\ker}{\M\M}{\ell+1}=C_b + C_W g\!\le(\Ti{\ker}{\M\M}{\ell}\ri) \, .
\ee
This should demystify our notational choice in the single-input analysis, as %
$\Ti{\ker}{\M\M}{\ell}$ simply represents the kernel for a single input $x_0$ corresponding to the midpoint of a pair of inputs $x_+, x_-$, as per~\eqref{eq:x-pm}. Going forward, we will call $\Ti{\ker}{\M\M}{\ell}$ the \textbf{midpoint kernel}\index{kernel!midpoint|textbf}\index{midpoint kernel|see{kernel}}.\footnote{We note that $\Ti{\ker}{[0]}{\ell}$ is the kernel for the midpoint of the layer-$\ell$ preactivations, $(\z{i}{+}{\ell}+\z{i}{-}{\ell})/2$, which is not quite the same as the midpoint kernel $\Ti{\ker}{\M\M}{\ell}$ for the preactivations of the \terminate{midpoint input} $x_0$ propagated to layer $\ell$. The difference is expressed in \eqref{eq:kernel-expand-1} and will turn out negligible for quantities at leading order in the $\delta$ expansion.}

Next, at first order in $\delta$, we will get the recursion
\be\label{eq:K1_naive}
\Ti{\delta \ker}{[1]}{\ell+1}=\chi_{\parallel}\!\le(\Ti{\ker}{\M\M}{\ell}\ri)\Ti{\delta \ker}{[1]}{\ell}\, .
\ee
This is to be expected. On account of the parity symmetry, $\Ti{\delta \ker}{[1]}{\ell+1}$ can only be proportional to $\Ti{\delta \ker}{[1]}{\ell}$ at this order, and the proportionality factor must be none other than the parallel susceptibility, because $\Ti{ \ker}{[1]}{\ell}=\frac{1}{2}\le[\Ti{\ker}{++}{\ell}-\Ti{\ker}{--}{\ell} \ri]$ behaves in the same way as the single-input kernels: if the single-input kernels behave exponentially, then this difference should as well.

Lastly, at the second order in $\delta$, we expect a recursion of the form
\be\label{eq:K2_naive}
\Ti{\delta\delta \ker}{[2]}{\ell+1}=\le[\text{something}\ri]\Ti{\delta\delta \ker}{[2]}{\ell}+\le[\text{something}'\ri]\le(\Ti{\delta \ker}{[1]}{\ell}\ri)^2+\le[\text{something}''\ri]\Ti{\delta\delta\ker}{[0]}{\ell}\, ,
\ee
which is the most general form it can take given the even parity symmetry of $\Ti{\delta\delta \ker}{[2]}{\ell}$, and where the $\le[\text{somethings}\ri]$ can be functions of the single-input kernel $\Ti{\ker}{\M\M}{\ell}$.
In the rest of this subsection we will derive the form of $\le[\text{something}\ri]$ and $\le[\text{something}'\ri]$, while also showing that $\le[\text{something}''\ri]$ vanishes due to the orthogonality of the $\gamma_{\alpha\beta}^{[a]}$ matrices.

Already at this heuristic level of the analysis, the bootstrapping nature of the system of equations should be clear. First, we find a solution for the midpoint kernel\index{kernel!midpoint} $\Ti{\ker}{\M\M}{\ell}$, which then bootstraps the layer dependence of $\Ti{\delta \ker}{[1]}{\ell}$ through~\eqref{eq:K1_naive}, the solution of which in turn feeds into~\eqref{eq:K2_naive} and together with $\Ti{\ker}{\M\M}{\ell}$ bootstraps the layer dependence of $\Ti{\delta\delta \ker}{[2]}{\ell}$. In other words, rather than confronting three coupled nonlinear recursions, we can solve decoupled recursions one by one.

\subsubsection{Deriving bootstrapped recursions}
Now let's embark on our algebraic journey. Our goal is to expand the kernel recursion~\eqref{eq:kernel-recursion-reminder} up to the second order in $\delta$. This requires us to evaluate $\braket{\sigma(z_+) \sigma(z_+)}{\ell}$, $\braket{\sigma(z_-) \sigma(z_-)}{\ell}$, and $\braket{\sigma(z_+) \sigma(z_-)}{\ell}$ to that order, all of which are two-dimensional Gaussian integrals. Rather than treating all of these Gaussian integrals separately, we instead will evaluate the Gaussian expectation of an arbitrary function $\braket{F\le(z_{+},z_{-}\ri)}{\ell}$ and then  will plug in $F\le(z_{+},z_{-}\ri)=\sigma(z_+) \sigma(z_+)$, $\sigma(z_-) \sigma(z_-)$, or $\sigma(z_+) \sigma(z_-)$.  Moreover, we will find that this general expression $\braket{F\le(z_{+},z_{-}\ri)}{\ell}$ will come in handy
in later chapters.

In order to evaluate this Gaussian expectation, 
it is natural to write the integral in the eigenbasis of the kernel rather than in the $\le(z_{+},z_{-}\ri)$ coordinates. Denote such orthonormal eigenvectors by $\le\{\hat{e}^{u},\hat{e}^{w}\ri\}$, which satisfy the eigenvalue equations
\be\label{eq:eigen-equation-you-know}
\sum_{\beta=\pm}\Ti{\ker}{\alpha\beta}{\ell}\hat{e}^{u}_{\beta}=\lambda_u \hat{e}^{u}_{\alpha}\, , %
\qquad \sum_{\beta=\pm}\Ti{\ker}{\alpha\beta}{\ell}\hat{e}^{w}_{\beta}=\lambda_w \hat{e}^{w}_{\alpha}\, ,
\ee
with eigenvalues $\lambda_u$ and $\lambda_w$, respectively. Transforming to coordinates $(u,w)$ defined via
\be\label{eq:coordinate_transformation}
z_{\alpha}(u,w)=u\hat{e}^{u}_{\alpha}+w\hat{e}^{w}_{\alpha}\, ,
\ee
the Gaussian expectation becomes
\be\label{eq:diagonalized-gaussian-integral-in-abstract}
\braket{F\!\le(z_{+},z_{-}\ri)}{\ell}=\frac{\int d u d w\ \exp\!\le(-\frac{u^2}{2\lambda_u}-\frac{w^2}{2\lambda_w}\ri) F\Big(z_{+}\le(u,w\ri),~z_{-}\le(u,w\ri)\Big)}{\int d u d w\ \exp\!\le(-\frac{u^2}{2\lambda_u}-\frac{w^2}{2\lambda_w}\ri)}\, .
\ee
As we discussed in \S\ref{sec:perturbation}, this equation expresses the idea that the $(u,w)$-coordinate basis diagonalizes\index{diagonalization} the kernel such that the distribution factorizes as $p(z_+,z_-) = p(u) p(w)$. The integral in the denominator represents the \terminate{normalization factor} of this factorized Gaussian expectation.

Now, we need to actually determine 
the eigenvalues $\lambda_u$ and $\lambda_w$ and eigenvectors $\le\{\hat{e}^{u},\hat{e}^{w}\ri\}$. %
We'll start our perturbative eigen-analysis by taking the by-now-familiar coincidental limit, $\delta\rightarrow 0$. As we discussed around \eqref{eq:kernel-expand-1}, in this limit the kernel is degenerate:
\be
\Ti{\ker}{\alpha\beta}{\ell}=\Ti{\ker}{\M\M}{\ell}\gamma^{[0]}_{\alpha\beta}=\Ti{\ker}{\M\M}{\ell}\begin{pmatrix}
1 & 1 \\
1  & 1
\end{pmatrix}\,  ,
\ee
with the $\gamma^{[0]}_{\alpha\beta}$ component equal to the midpoint kernel\index{kernel!midpoint} $\Ti{\ker}{[0]}{\ell}=\Ti{\ker}{\M\M}{\ell}$, and the other components vanishing.
Such a matrix has the normalized eigenvectors
\be\label{eq:eigen-degenerate}
\hat{e}^{u}_{\alpha}=\frac{1}{\sqrt{2}}\begin{pmatrix}
1 \\
1
\end{pmatrix}\, , \qquad \text{and} \qquad \hat{e}^{w}_{\alpha}=\frac{1}{\sqrt{2}}\begin{pmatrix}
1 \\
-1
\end{pmatrix}\, ,
\ee
with eigenvalues $\lambda_u=2\Ti{\ker}{\M\M}{\ell}$ and $\lambda_w=0$, respectively.\footnote{Here, the zero eigenvalue for $w$ signifies that the matrix is degenerate. This implies that the distribution for the $w$ coordinate is given by a \terminate{Dirac delta function}, $p(w)=\delta(w)$, indicating that there's really only one input in this limit.
}

Next, let's back off from the coincidental limit and look back at the $\delta$ expansions~\eqref{eq:kernel-expand-1}--\eqref{eq:kernel-expand-3} for $\Ti{\ker}{[0,1,2]}{\ell}$ around the midpoint kernel\index{kernel!midpoint}. With similar expansions for the eigenvectors $\hat{e}^{u,w}_\pm$ and eigenvalues $\lambda_{u,w}$, we can solve the eigenvalue equations~\eqref{eq:eigen-equation-you-know} order by order.\footnote{
    For physicists, note that this is second-order time-independent \terminate{perturbation theory} from \terminate{quantum mechanics}.
} 
Carrying out such expansions (in the margins or -- if this isn't your personal copy of our book -- in a private notebook) and solving \eqref{eq:eigen-equation-you-know} to second order, we find normalized eigenvectors 
\begin{align}\label{eq:kernel-eigenvectors}
\hat{e}^{u}_{\alpha}&= \begin{pmatrix} 
\hat{e}^{u}_{+} \\
\hat{e}^{u}_{-}
\end{pmatrix}
= \frac{1}{\sqrt{2}}\begin{pmatrix} 
1 + \frac{\Ti{\delta\ker}{[1]}{\ell}}{2\Ti{\ker}{\M\M}{\ell} }- \frac{1}{8}\le(\frac{\Ti{\delta\ker}{[1]}{\ell}}{\Ti{\ker}{\M\M}{\ell}}\ri)^2 \\
1 - \frac{\Ti{\delta\ker}{[1]}{\ell}}{2\Ti{\ker}{\M\M}{\ell} }- \frac{1}{8}\le(\frac{\Ti{\delta\ker}{[1]}{\ell}}{\Ti{\ker}{\M\M}{\ell}}\ri)^2 
\end{pmatrix} + \o{\delta^3} \, , \\ \notag
\hat{e}^{w}_{\alpha}&= \begin{pmatrix} 
\hat{e}^{w}_{+} \\
\hat{e}^{w}_{-}
\end{pmatrix}
=  \frac{1}{\sqrt{2}}\begin{pmatrix} 
1 - \frac{\Ti{\delta\ker}{[1]}{\ell}}{2\Ti{\ker}{\M\M}{\ell} }- \frac{1}{8}\le(\frac{\Ti{\delta\ker}{[1]}{\ell}}{\Ti{\ker}{\M\M}{\ell}}\ri)^2 \\
-1 - \frac{\Ti{\delta\ker}{[1]}{\ell}}{2\Ti{\ker}{\M\M}{\ell} }+ \frac{1}{8}\le(\frac{\Ti{\delta\ker}{[1]}{\ell}}{\Ti{\ker}{\M\M}{\ell}}\ri)^2  
\end{pmatrix} + \o{\delta^3} \, ,
\end{align}
and corresponding eigenvalues
\begin{align}\label{eq:kernel-eigs}
\lambda_u&= 2 \Ti{\ker}{\M\M}{\ell} + 2 \Ti{\delta\delta\ker}{[0]}{\ell} + \frac{\le(\Ti{\delta\ker}{[1]}{\ell}\ri)^2}{2\Ti{\ker}{\M\M}{\ell} } + \o{\delta^4} \, , \\ 
\lambda_w &= 2 \Ti{\delta\delta\ker}{[2]}{\ell} - \frac{\le(\Ti{\delta\ker}{[1]}{\ell}\ri)^2}{2\Ti{\ker}{\M\M}{\ell} } + \o{\delta^4} \,  .\notag
\end{align}
Even if you don't have a private notebook, it's easy to check on a scrap of paper that \eqref{eq:kernel-eigenvectors} and \eqref{eq:kernel-eigs} solve \eqref{eq:eigen-equation-you-know} to $\o{\delta^2}$.

Now, having solved the eigenproblem, we can implement the change of coordinates. Before doing so, notice that the $u$ coordinate is closely related to the coordinate $z_0$, the preactivation corresponding to the \terminate{midpoint input}. This makes it very natural to use $z_0$ as a coordinate instead of $u$.  We can implement this by rescaling $u$ as
\be
\frac{u^2}{2\lambda_{u}}=\frac{z_0^2}{2\KML} \, ,
\ee
changing variables in the integral \eqref{eq:diagonalized-gaussian-integral-in-abstract}
so that the Gaussian integral over $u$ becomes a Gaussian integral over $z_0$ with a variance given by the midpoint kernel\index{kernel!midpoint} $\KML$. With this rescaling, the full coordinate transformation becomes
\be\label{eq:new-coordinate-transformation}
z_{\pm}(z_0,w)=z_0 \!\le[1 \pm \le(\frac{\Ti{\delta\ker}{[1]}{\ell}}{2\Ti{\ker}{\M\M}{\ell} }\ri) +\le(\frac{\Ti{\delta\delta\ker}{[0]}{\ell}}{2\Ti{\ker}{\M\M}{\ell} }\ri)  +\o{\delta^3}\ \ri] +\frac{w}{\sqrt{2}}\le[\pm 1 +\o{\delta} \ri]\, .
\ee
Here, we can truncate the term in the square brackets multiplying $w$ at $\o{1}$, since the $w$ coordinate has zero mean and a variance $\lambda_w = \o{\delta^2}$. This means that, when performing the $w$ integration, terms proportional to $w^0$ will be $\o{1}$, terms proportional to $w^2$ will be $\o{\delta^2}$, higher-order terms will be subleading, and, of course, all the odd terms will vanish.
By contrast, the $z_0$ coordinate has zero mean and a variance $\KML=\o{1}$,
so we actually need keep terms up to $\o{\delta^2}$.

Next, we need to plug this expression \eqref{eq:new-coordinate-transformation} into our arbitrary function 
\be\label{eq:function-in-new-coordinate}
F\!\le(z_{+},z_{-}\ri)=F\Big(z_{+}\!\le(z_0,w\ri)\!,~z_{-}\!\le(z_0,w\ri)\Big)\, ,
\ee
now viewed as a function of the two independent Gaussian variables $z_0$ and $w$, and perform the integration over them.
To do so, first we need to Taylor expand the function in both $\delta$ and $w$ around $F\!\le(z_0,z_0\ri)$.
 This gives
\begin{align}\label{eq:expectation-of-general-F}
&F\!\le(z_{+},z_{-}\ri)\, \\
=&F\!\le(z_0, z_0\ri) +z_0 \le(\frac{\Ti{\delta\ker}{[1]}{\ell}}{2\Ti{\ker}{\M\M}{\ell} }\ri)\le(\partial_{+}-\partial_{-}\ri) F+ z_0\le(\frac{\Ti{\delta\delta\ker}{[0]}{\ell}}{2\Ti{\ker}{\M\M}{\ell} }\ri)\le(\partial_{+}+\partial_{-}\ri) F\, \nonumber\\
&+z_0^2\le(\frac{\Ti{\delta\ker}{[1]}{\ell}}{2\Ti{\ker}{\M\M}{\ell}}\ri)^2\frac{\le(\partial_{+}-\partial_{-}\ri)^2 F}{2}+\frac{w^2}{2}\frac{\le(\partial_{+}-\partial_{-}\ri)^2 F}{2}\, \nonumber\\
&+(\text{odd in }w)+\o{\delta^3, w^2 \delta, w^4}\, ,\nonumber
\end{align}
with the abbreviation $\partial_{+}^p\partial_{-}^qF\equiv\partial_{+}^p\partial_{-}^q F(z_{+},z_{-}) \vert_{z_{+}=z_{-}=z_0}$.
The Gaussian integral over $w$ is simple to perform, we just replace $w^2$ with its variance $\lambda_w$ \eqref{eq:kernel-eigs}. Finally, we will express our final answer in terms of single-variable Gaussian expectations over the variable $z_0$ -- which, as you should recall, has a variance given by the scalar midpoint kernel\index{kernel!midpoint} $K_{\M\M}^{(\ell)}$ -- giving
\begin{align}\label{eq:F-expectaiton-final}
&\braket{F(z_{+},z_{-})}{\ell}\, \\
=&\le\langle F(z_0,z_0)\ri\rangle_{\KML}+\le(\frac{\Ti{\delta\ker}{[1]}{\ell}}{2\KML}\ri)\bra z_0 \le(\partial_{+}-\partial_{-}\ri)F\ket_{\KML}+\le(\frac{\Ti{\delta\delta\ker}{[0]}{\ell}}{2\KML}\ri)\le\langle z_0 \le(\partial_{+}+\partial_{-}\ri)F\ri\rangle_{\KML}\, \nonumber\\
&+\frac{1}{2}\bra \le[\Ti{\delta\delta\ker}{[2]}{\ell}+\le(\frac{\Ti{\delta \ker}{[1]}{\ell}}{2\KML}\ri)^2\le(z_0^2-\KML\ri)\ri] \le(\partial_{+}-\partial_{-}\ri)^2F\ket_{\KML}+\o{\delta^3}\, .\nonumber
\end{align}
This completes our computation of this general expectation.

In order to apply this formula to evaluate the expectations $\braket{\sigma(z_{\alpha}) \sigma(z_{\beta})}{\ell}$ in the kernel recursion \eqref{eq:kernel-recursion-reminder}, recall definitions of gamma matrices in~\eqref{eq:gamma_def} and note
\begin{align}
\le[\sigma(z_{\alpha}) \sigma(z_{\beta})\ri] \vert_{z_{+}=z_{-}=z_0}&=\sigma(z_0)\sigma(z_0) \gamma^{[0]}_{\alpha\beta}\, ,\\
\le\{\le(\partial_{+}-\partial_{-}\ri)\le[\sigma(z_{\alpha}) \sigma(z_{\beta})\ri] \ri\}\vert_{z_{+}=z_{-}=z_0}&=2\sigma'(z_0)\sigma(z_0) \gamma^{[1]}_{\alpha\beta}\, ,\\
\le\{\le(\partial_{+}+\partial_{-}\ri)\le[\sigma(z_{\alpha}) \sigma(z_{\beta})\ri] \ri\}\vert_{z_{+}=z_{-}=z_0}&=2\sigma'(z_0)\sigma(z_0) \gamma^{[0]}_{\alpha\beta}\, ,\\
\le\{\le(\partial_{+}-\partial_{-}\ri)^2\le[\sigma(z_{\alpha}) \sigma(z_{\beta})\ri] \ri\}\vert_{z_{+}=z_{-}=z_0}&=2\sigma''(z_0)\sigma(z_0)\gamma^{[0]}_{\alpha\beta}+2\sigma'(z_0)\sigma'(z_0)\gamma^{[2]}_{\alpha\beta}\, .
\end{align}
Plugging these individually into our general expression~\eqref{eq:F-expectaiton-final}, we get\index{$\gamma^{[a]}$ basis!$\sigma \sigma$}
\begin{align}
&\braket{\sigma(z_{\alpha}) \sigma(z_{\beta})}{\ell} \, \label{eq:useful-much-later-too}\\
=&\le[\le\langle \sigma(z_0)\sigma(z_0)\ri\rangle_{\KML}+\o{\delta^2}\ri] \gamma^{[0]}_{\alpha\beta}\, \nonumber\\
+&\le[\le(\frac{\Ti{\delta\ker}{[1]}{\ell}}{\KML}\ri)\bra z_0\sigma'(z_0)\sigma(z_0)\ket_{\KML}\ri] \gamma^{[1]}_{\alpha\beta}\,  \nonumber\\
+&\le[\Ti{\delta\delta\ker}{[2]}{\ell}\bra \sigma'(z_0)\sigma'(z_0)\ket_{\KML}+\le(\frac{\Ti{\delta \ker}{[1]}{\ell}}{2\KML}\ri)^2\bra\le(z_0^2-\KML\ri)\sigma'(z_0)\sigma'(z_0)\ket_{\KML}\ri] \gamma^{[2]}_{\alpha\beta}\, .\nonumber
\end{align}
The coefficients of the matrix $\braket{\sigma(z_{\alpha}) \sigma(z_{\beta})}{\ell}$ in the $\gamma^{[a]}_{\alpha\beta}$ basis can be simply read off from the above expression. Therefore, plugging this into the right-hand side of the full kernel recursion~\eqref{eq:kernel-recursion-reminder}, we can expand the left-hand side of that equation in this basis as
\be
\Ti{\ker}{\alpha\beta}{\ell+1} = \Ti{\ker}{[0]}{\ell+1}\gamma^{[0]}_{\alpha\beta}+\Ti{\ker}{[1]}{\ell+1}\gamma^{[1]}_{\alpha\beta}+\Ti{\ker}{[2]}{\ell+1}\gamma^{[2]}_{\alpha\beta} \, ,
\ee
and equate both sides to find recursions for each component in this basis. These are given just below.

\subsubsection{Summary}
Just above, we explained how to derive the recursions
\begin{align}
\Ti{\ker}{\M\M}{\ell+1}&=C_b+C_W\, g\!\le(\KML\ri)\,  ,\label{K0}\\
\Ti{\delta \ker}{[1]}{\ell+1}&=\chi_{\parallel}\!\le(\Ti{\ker}{\M\M}{\ell}\ri)\Ti{\delta \ker}{[1]}{\ell}\, ,\label{K1}\\
\Ti{\delta\delta \ker}{[2]}{\ell+1}&=\chi_{\perp}\!\le(\Ti{\ker}{\M\M}{\ell}\ri)\Ti{\delta\delta \ker}{[2]}{\ell}+h\!\le(\KML\ri)\le(\Ti{\delta \ker}{[1]}{\ell}\ri)^2\, . \label{K2}
\end{align}
Here the by-now familiar helper function~\eqref{eq:helper_first} is defined as
\be\label{eq:g-function}
g\!\le(K\ri)= \le\langle \sigma(z) \, \sigma(z)\ri\rangle_{K}\, ,
\ee
 the \terminate{parallel susceptibility} that we already encountered in~\eqref{eq:chi_parallel_first} is given by
\be\label{eq:chi-parallel}
\chi_{\parallel}(\ker)=C_W g'(K)=\frac{C_W}{2\ker^2} \bra \sigma(z)\, \sigma(z)\le(z^2-K\ri)\ket_{\ker}=\frac{C_W}{\ker} \bra z\, \sigma'(z)\, \sigma(z)\ket_{\ker}\, ,
\ee
the \term{perpendicular susceptibility} is newly introduced as
\be\label{eq:chi-perp}
\chi_{\perp}(\ker)\equiv C_W \bra\sigma^\prime(z)\, \sigma^\prime(z) \ket_{\ker} \, ,
\ee
and the helper function that generates perturbations $\Ti{\delta\delta \ker}{[2]}{\ell+1}$ from perturbations $\Ti{\delta \ker}{[1]}{\ell}$ is given by
\be\label{eq:h-function}
h\!\le(K\ri)\equiv \frac{C_W}{4K^2}\le\langle \sigma'(z)\, \sigma'(z)\le(z^2-K\ri)\ri\rangle_{K}=\frac{1}{2}\frac{\td }{\td K}\chi_{\perp}(\ker)\, .
\ee
In the last steps of~\eqref{eq:chi-parallel} and~\eqref{eq:h-function} we made use of the following identity for the single-variable Gaussian expectation
\begin{align}\label{eq:gaussian-integration-by-parts-formula}
\frac{d}{d \ker}\le[\frac{1}{\sqrt{2\pi\ker}}\int_{-\infty}^{\infty} dz\,  e^{-\frac{z^2}{2K}}F(z)\ri]=&\frac{1}{2 \ker^2}\le[\frac{1}{\sqrt{2\pi\ker}}\int_{-\infty}^{\infty} dz\,  e^{-\frac{z^2}{2K}}F(z) (z^2-\ker)\ri]\, \\
 =& \frac{1}{2 \ker}\le[\frac{1}{\sqrt{2\pi\ker}}\int_{-\infty}^{\infty} dz\,  e^{-\frac{z^2}{2K}}z\frac{\td}{\td z} F(z)\ri]\, ,\notag
\end{align}
where to go from the first line to the second line we integrated by parts.\index{integration by parts}
These three recursions~\eqref{K0}--\eqref{K2} are sufficient to completely fix the initialization hyperparameters and tune the network to criticality.

The first equation~\eqref{K0} is a recursion for the midpoint kernel\index{kernel!midpoint} $\KML$. %
To analyze this equation, we look for a fixed-point value $\Tif{\ker}{\M\M}$ satisfying $\Tif{\ker}{\M\M}=C_b + C_W g\!\le(\Tif{\ker}{\M\M}\ri)$ and then linearize around such a fixed point as $\KML=\Tif{\ker}{\M\M}+\Delta \KML$. Doing so, we see that $\Ti{\Delta \ker}{\M\M}{\ell+1}=\chi_{\parallel}\!\le(\Tif{\ker}{\M\M}\ri)\Ti{\Delta \ker}{\M\M}{\ell}+\o{\Delta^2}$ and realize that the parallel susceptibility $\chi_{\parallel}\!\le(\Tif{\ker}{\M\M}\ri)$ governs the growth/decay of deviations $\Delta \KML$ from the fixed-point value $\Tif{\ker}{\M\M}$.

The second equation~\eqref{K1} is the first equation~\eqref{K0} in disguise, since the $\Ti{\delta \ker}{[1]}{\ell}$ component is the leading difference in magnitude $R^{(\ell)}=\le(\Ti{\ker}{++}{\ell}-\Ti{\ker}{--}{\ell}\ri)/2$ of preactivations for two inputs. As such, the same susceptibility $\chi_{\parallel}\!\le(\KML\ri)$ governs its growth/decay. Another perspective is that the $\Ti{\delta \ker}{[1]}{\ell}$ component can be generated by considering a perturbation $\delta x_i \propto \x{i}{\M}$ that is parallel to the original input $\x{i}{\M}$, creating a difference in the norm of the two inputs. This deviation is naturally measured by $R^{(\ell)}$, and setting $\chi_\parallel\!\le(\Tif{\ker}{\M\M} \ri) = 1$ ensures that such a perturbation neither exponentially explodes nor exponentially vanishes. And that, after a long-winded journey, explains why we called this susceptibility \emph{parallel}.

This third recursion~\eqref{K2} is something new, controlling the layer dependence of 
the magnitude of the difference of the two inputs
$D^{(\ell)}=4\Ti{\delta\delta \ker}{[2]}{\ell}+\o{\delta^4}$.
Such a perturbation in layer $\ell+1$ is sourced by two types of perturbations in layer $\ell$, as exhibited by the two terms on right-hand side of \eqref{K2}. One term $\propto\!\le(\Ti{\delta \ker}{[1]}{\ell}\ri)^2$ is generated by preactivations in the $\ell$-th layer with different norms. The other term $\propto\Ti{\delta\delta \ker}{[2]}{\ell}$ is generated by preactivations in the $\ell$-th layer with a nonzero difference $D^{(\ell)}$ and is present even if the preactivations have the same norm.
Such same-norm perturbations in the infinitesimal regime correspond to perturbations of the input that are \emph{perpendicular} to the \terminate{midpoint input}, i.e.~$\sum_{i=1}^{n_0} \x{i}{\M} \, \delta x_i  = 0$. 
The \terminate{perpendicular susceptibility} $\chi_{\perp}\!\le(\Tif{\ker}{\M\M}\ri)$ determines the dynamics of such perpendicular perturbations.\footnote{An alternative view is that, for a given instantiation of the network, this perpendicular susceptibility $\chi_\perp\!\le(K^{(\ell)}_{\M\M}\ri)$ controls changes of the preactivations with respect to changes in the input. To see that, note that the distance $D^{(\ell)}$ can be rewritten to leading order in the perturbation as
\be
D^{(\ell)}=\frac{1}{n_{\ell}}\sum_{i=1}^{n_{\ell}}\E{\le(\z{i}{+}{\ell}-\z{i}{-}{\ell}\ri)^2} = \frac{1}{n_{\ell}}\sum_{i=1}^{n_{\ell}}\E{ \le(\sum_{j=1}^{n_0} \frac{dz^{(\ell)}_{i;0}}{d\x{j}{\M} } \delta x_{j} \ri)^2~} + \o{\delta^4} \, .
\ee 
This makes quantity $\chi_{\perp}\!\le(\Tif{\ker}{\M\M}\ri)$  of interest for controlling the infamous \neo{exploding and vanishing gradient problem},
a perspective that we will make more concrete in~\S\ref{ch:eft-ntk}. \label{footnote:Jacobian}
}
As a nonzero distance $D^{(\ell)}$ is essential for being able to compare and contrast the two inputs $\x{i}{\pm}$ after being propagated to layer $\ell$, we need to ensure that this quantity is well behaved. To avoid exponential behavior, we will demand $\chi_{\perp}\!\le(\Tif{\ker}{\M\M}\ri) = 1$.

Taken all together, our general notion of \terminate{criticality} requires the following two conditions to hold\footnote{
Note that this further underscores the need for an \terminate{ensemble}. In \S\ref{sec:MLP_distribution}, we motivated the \terminate{initialization distribution} by pointing out that the \neo{zero initialization} $\bias{i}{\ell}=\W{ij}{\ell}=0$ doesn't break the \terminate{permutation symmetry} among the $n_\ell$ neurons of a layer. Here we see more generally that any zero-mean deterministic (i.e.~$C_W=0$) distribution for the weights -- which includes the zero initialization -- cannot satisfy the \terminate{criticality} conditions $\chi_{\parallel}=\chi_{\perp}=1$, since both susceptibilities~\eqref{eq:chi-parallel} and~\eqref{eq:chi-perp} are proportional to $C_W$. Such a zero-weight initialization will always suffer from an exponential decay towards a trivial fixed point\index{fixed point!trivial} at $\Tif{\ker}{\M\M}=C_b$.
\label{footnote:zero-init-and-criticality}
}
\be\label{eq:criticality-conditions}
\chi_{\parallel}\!\le(\Tif{\ker}{\M\M}\ri)=1\, ,  \qquad \chi_{\perp}\!\le(\Tif{\ker}{\M\M}\ri)=1\ ,
\ee
with the fixed-point value of the midpoint kernel\index{kernel!midpoint} $\Tif{\ker}{\M\M}$ implicitly defined via
\be
\Tif{\ker}{\M\M}=C_b + C_W g\!\le(\Tif{\ker}{\M\M}\ri) \, .
\ee
These conditions are sufficient to ensure that the entire kernel matrix is preserved to leading order, namely that
\be
\Ti{\Delta\ker}{\M\M}{\ell+1} = \Ti{\Delta\ker}{\M\M}{\ell}+\o{\Delta^2}  \,,  \qquad \Ti{\ker}{[1]}{\ell+1} = \Ti{\ker}{[1]}{\ell}+\o{\delta^3} \,,  \qquad  \Ti{\ker}{[2]}{\ell+1} = \Ti{\ker}{[2]}{\ell}+\o{\delta^4}\, .
\ee
This generalizes the notion of \terminate{criticality} that we discussed for deep linear networks in $\S\ref{ch:deep-linear-eft}$.
Over two sections we will give a prescription for finding these critical initialization hyperparameters $(C_b,C_W)^{\text{critical}}$ for any nonlinear activation function. \index{initialization hyperparameters!critical}

\section{Criticality for Scale-Invariant Activations}\label{sec:scale-invariant-eft}
\index{scale invariance}\index{initialization hyperparameters!critical!for scale-invariant universality}
Now, let's extend our \terminate{criticality} analysis to \emph{scale-invariant}\index{scale invariance} activation functions by applying the formalism that we just developed. %
Recall from~\S\ref{sec:activations} that a scale-invariant activation function satisfies
\be\label{eq:scale-invariant-def-eft-chapter}
\sigma(\lambda z) = \lambda \sigma(z) \, , %
\ee
for any positive rescaling $\lambda>0$, and always takes the form
\be\label{eq:scale-invariant-one-kink}
\sigma(z) = 
    \begin{cases}
   a_+ z \, , & z \ge 0  \, , \\
    a_- z \, , & z < 0 \, .
    \end{cases}
\ee
As a reminder, this class of activations includes the $\linear$ activation -- by setting $a_+=a_-=1$ -- and the $\relu$ -- by setting $a_+=1$ and $a_-=0$.

These activation functions are particularly simple in that the criticality conditions \eqref{eq:criticality-conditions},  $\chi_\parallel\!\le(K^{(\ell)}_{\M\M}\ri)=\chi_\perp\!\le(K^{(\ell)}_{\M\M}\ri)=1$, can be solved exactly.
To start, 
we can easily compute $g(K)$,~\eqref{eq:g-function}, which reduces to two Gaussian integrals on half the real line times an even polynomial, yielding %
\be\label{eq:needed-to-recall-in-kernel-learning-chapter-I}
g(K)=A_2 K \, , 
\ee
where we have introduced an activation-dependent constant
\be
A_2\equiv \frac{a_+^2+a_-^2}{2} \, .
\ee
From \eqref{eq:chi-parallel}, we see that we can find $\chi_\parallel(K)$ by differentiating this expression with respect to $K$ and multiplying by $C_W$. Inspecting \eqref{eq:chi-perp}, we see that to get  $\chi_\perp(K)$, we can perform two more simple Gaussian integrals on half the real line.
Together, we find that both susceptibilities are equal and independent of $\KML$
\be\label{eq:needed-to-recall-in-kernel-learning-chapter-II}
\chi_\parallel\!\le(K^{(\ell)}_{\M\M}\ri)=\chi_\perp\!\le(K^{(\ell)}_{\M\M}\ri)=A_2 C_W \equiv\chi \, .
\ee
Lastly $h(K)$,~\eqref{eq:h-function}, identically vanishes because it is a derivative of $\chi_\perp(K)$.

\index{initialization hyperparameters}
With all that, we can write the general kernel recursions \eqref{K0}, \eqref{K1}, and \eqref{K2} for scale-invariant activations as
\begin{align}
\Ti{ \ker}{\M\M}{\ell+1}&=C_b+\chi\KML\, ,\label{eq:recursion-scale-invariant-spin-0}\\
\Ti{\delta \ker}{[1]}{\ell+1}&=\chi\Ti{\delta \ker}{[1]}{\ell}\, ,\\
\Ti{ \delta\delta\ker}{[2]}{\ell+1}&=\chi\Ti{ \delta\delta\ker}{[2]}{\ell}\, .
\label{eq:scale-invariant-perpendicular-naive}
\end{align}
These are quite simple to solve. Just as the initialization hyperparameter $C_W$ governed the \terminate{exploding and vanishing kernel problem} in~\S\ref{sec:criticality_DLN}, the constant susceptibility $\chi=A_2 C_W$ governs the same problem here:
\begin{itemize}
\item If $\chi>1$,  all quantities explode exponentially in $\ell$ towards a trivial fixed point\index{fixed point!trivial} at infinity.

\item If $\chi <1$, the fixed-point value of the kernel is given by $\Tif{\ker}{\M\M}=\frac{C_b}{1-\chi}$ and all perturbations around the fixed point vanish exponentially with $\ell$.
\item If $C_W =1/A_2$ and $C_b=0$, then the network is at \terminate{criticality}. %
Not only does every perturbation stays constant,\footnote{\label{foot:kink2}
One caveat is in order. While the constancy of the preactivation norm $\Ti{\ker}{[0]}{\ell}$ and the parallel perturbation $\Ti{\ker}{[1]}{\ell}$ is exact, the constancy of $\Ti{\ker}{[2]}{\ell}$ is an artifact of our infinitesimal perturbation analysis. In fact, the finite-angle analysis of nonlinear scale-invariant activation functions in \S\ref{sec:finite_angle} describes how $\Ti{\ker}{[2]}{\ell}$ crosses over from near constancy for small $\ell$ to a power-law decay $\sim1/\ell^2$ for large $\ell$. In short, the preservation of the whole kernel matrix seen in~\S\ref{sec:criticality_DLN} is a special property of the $\linear$ activation, and for nonlinear scale-invariant activation functions there is a slow power-law decay of some observables. This power-law behavior is quite benign compared to exponential behavior and is typical at \terminate{criticality}.
} but also any value of $\Tif{\ker}{\M\M}$ serves as a nontrivial fixed point, i.e., there is a \emph{line of nontrivial fixed points}.\footnote{For physicists, note that a similar line of fixed points often appears in scale-invariant field theories with exactly marginal deformations.} In particular, the value of the fixed point is given by
\be\label{eq:chapter-5-scale-invariant-kernel-fixed}
\Tif{\ker}{\M\M}=\frac{1}{A_2} \le(\frac{1}{n_0}\sum_{i=1}^{n_0}\x{i}{\M}^2\ri) \, .
\ee
\item If $C_W =1/A_2$ and $C_b>0$, then $\Ti{\delta \ker}{[1]}{\ell}$ and $\Ti{\delta\delta \ker}{[2]}{\ell}$ stay constant at this infinitesimal level of analysis.
However, $\KML$ grows linearly towards a nontrivial fixed point\index{fixed point!nontrivial} at infinity, with the rate set by $C_b$. 
Since the kernel does not exhibit any exponential behavior, such a network is at \terminate{criticality} in a broad sense.
This \textbf{semi-criticality}\index{semi-criticality|see{criticality}}\index{criticality!semi-criticality|textbf} results in a \emph{line of semi-critical initialization hyperparameters} parameterized by $C_b$ in the hyperparameter plane spanned by $(C_b, C_W)$. 
\end{itemize}
In conclusion, this study generalizes the analysis carried out for deep linear networks in~\S\ref{sec:criticality_DLN} and identifies
\be\label{eq:criticality_scale_invariant}
\le(C_b, C_W\ri)^{\text{critical}}=\le(0, \frac{1}{A_2}\ri)\,  ,
\ee
with $A_2= (a_+^2+a_-^2)/2$ as the critical \terminate{initialization hyperparameters} for scale-invariant activation functions.\footnote{We see here that our simplification of $C_b =0$ for deep linear networks\index{deep linear network} in \S\ref{ch:deep-linear-eft} was completely warranted, \emph{ex post facto}.} For the $\relu$ activation function, this reproduces the \neo{Kaiming initialization}\index{Kaiming initialization|seealso{initialization hyperparameters}} $\le(C_b, C_W\ri)^{\text{critical}}= (0,2)$ \cite{he2015delving}.

\section{Universality beyond Scale-Invariant Activations}\label{sec:non-scale-invariant-eft}
All of the activation functions treated in the last section shared a rather special property: \terminate{scale invariance}~\eqref{eq:scale-invariant-def-eft-chapter}.
This property 
gave rise to equal and kernel-independent parallel and perpendicular susceptibilities, $\chi_\parallel(K)=\chi$ and  $\chi_\perp(K)=\chi$, all together enabling us to drastically simplify the \terminate{criticality} analysis for these activation functions.\footnote{The kernel-independence property follows directly from the scale-invariance, as any dependence would have introduced a scale into the problem.}\index{parallel susceptibility}\index{perpendicular susceptibility}
Such an analysis showed that networks equipped with a scale-invariant activation function will behave similarly to each other under \neo{representation group flow} at \terminate{criticality}.

In theoretical physics\index{physics}, systems at criticality that behave similarly under \neo{renormalization group flow} are said to fall into the same \term{universality class}. The \terminate{effective action} describing such systems converge under the iterative coarse-graining procedure, such that at long-range scales these systems share the same underlying mathematical model or \neo{effective theory},  independent of the microscopic details of the particular system. This phenomenon is known as \term{universality} \cite{Kadanoff:1971pc}.

This motivates the use of the same term, universality class, to characterize activation functions that share the same limiting behavior under \terminate{representation group flow}, thus furthering the connection between \neo{RG flow and RG flow} that we began developing in \S\ref{sec:marginalization-group-flow}. Activation functions that form a universality class will have an identical effective description after flowing through many layers, meaning that the \terminate{effective theory} describing the preactivation distribution becomes independent of the fine details of the particular activation function. The power of universality is that a \emph{single} effective theory enables us to understand \terminate{criticality} for the many different activation functions within the same universality class.

\index{initialization hyperparameters!critical}\index{universality class!scale-invariant|textbf}
Clearly, all the scale-invariant activation functions form a universality class. However, the simplifications that enabled us to easily analyze this \textbf{scale-invariant universality class}, e.g.~the kernel-independence of the susceptibilities, do not hold  for
other activation functions.
For activation functions such as the $\sigmoid$, $\tanhA$, or $\swish$, we'll need to develop a much more general algorithm to find critical initialization hyperparameters. 
In~\S\ref{subsec:strategize}, we'll illustrate how this algorithm works, and then we'll analyze specific activation functions in~\S\ref{subsec:no_thinking_no_money},~\S\ref{subsec:tanh_univ}, and~\S\ref{subsec:half_stability}.

\subsection{General Strategy}\label{subsec:strategize}
Let's start with some recollections. As discussed most recently in~\S\ref{sec:bootstrapping}, for a generic choice of initialization hyperparameters $C_b$ and $C_W$, the kernel recursion for a single-input $x_\M$,
\be\label{eq:kernel_rec}
\Ti{\ker}{\M\M}{\ell+1} =C_b+ C_Wg\!\le(\Ti{\ker}{\M\M}{\ell}\ri) \, ,
\ee
admits a fixed-point solution satisfying
\be\label{eq:kernel_fixed_eq}
\Tif{\ker}{\M\M}=C_b+C_W g\!\le(\Tif{\ker}{\M\M}\ri)\, ,
\ee
where the helper function
\be\label{eq:helper-function-general-strategy}
 g\!\le(\ker\ri)\equiv \bra\sigma(z)\sigma(z) \ket_{\ker}\, ,
\ee
is understood as a function of the kernel value $K$. Our goal is to find critical initialization hyperparameters whose associated fixed-point value $\Tif{\ker}{\M\M}=\Tif{\ker}{\M\M}\!\le(C_b,C_W\ri)$ gives rise to $\chi_{\parallel}\!\le(\Tif{\ker}{\M\M}\ri)=\chi_{\perp}\!\le(\Tif{\ker}{\M\M}\ri)=1$.\index{initialization hyperparameters!critical}

\begin{figure}[h]
\begin{center}
 \includegraphics[width=0.85\linewidth]{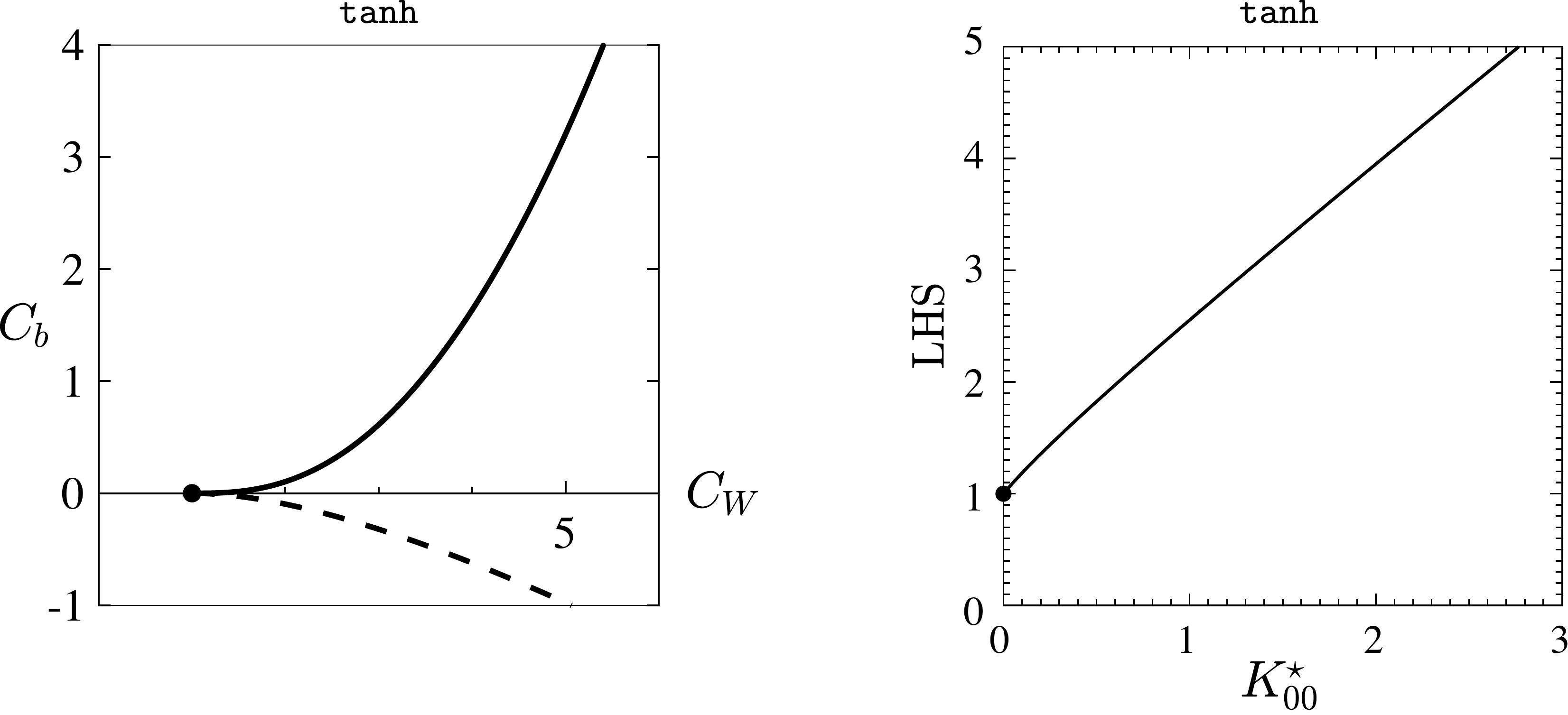}
\end{center}
\caption{Two algorithms to pin down a nontrivial fixed point\index{fixed point!nontrivial}, illustrated here for the $\tanhA$ activation function. \textbf{Left:} the lines defined by the conditions $\Tif{\chi}{\perp}=1$ (solid) and $\Tif{\chi}{\parallel}=1$ (dashed) are shown in the hyperparameter plane $(C_W,C_b)$ for the $\tanhA$ activation function.
The intersection of these two lines gives the critical initialization hyperparameters $(C_W,C_b)=(1,0)$.
\textbf{Right:} the left-hand side of the condition~\eqref{easycriticality} is plotted as a function of $\Tif{\ker}{\M\M}$. The plotted line hits unity as $\Tif{\ker}{\M\M}\rightarrow 0$.
\index{initialization hyperparameters!critical}
}
\label{fig:tanh_critical}
\end{figure}

How do we actually find these critical values? %
Conceptually, the most obvious route -- illustrated in Figure \ref{fig:tanh_critical} for the $\tanhA$ activation function -- is the following procedure:
\begin{enumerate}
\item For each value of $C_b$ and $C_W$, with $C_b\geq 0$ and $C_W\geq 0$, find a fixed-point value of the kernel $\Tif{\ker}{\M\M}=\Tif{\ker}{\M\M}\!\le(C_b,C_W\ri)$, implicitly defined via $\Tif{\ker}{\M\M}=C_b+C_W g_0\!\le(\Tif{\ker}{\M\M}\ri)$ with the constraint $\Tif{\ker}{\M\M}\geq 0$.
\item %
With $\Tif{\ker}{\M\M}\!\le(C_b,C_W\ri)$, evaluate both $\chi_{\parallel}\!\le(\Tif{\ker}{\M\M}\ri)$ and $\chi_{\perp}\!\le(\Tif{\ker}{\M\M}\ri)$, scanning over values in the $(C_b, C_W)$ plane until the criticality conditions $\chi_{\parallel}=1$ and  $\chi_{\perp}=1$ are both met.%
\end{enumerate}

This algorithm, however, is practically cumbersome to carry out for general activation functions, both numerically and analytically. In order to obtain a more implementation-friendly algorithm, let's reshuffle the logic a bit.
First, note that for a candidate fixed-point value $\Tif{\ker}{\M\M}$, setting
\begin{align}
\label{CWcri}
C_W&=\le[ \bra\sigma'(z)\sigma'(z) \ket_{\Tif{\ker}{\M\M}}\ri]^{-1} \, ,\\ %
\label{Cbcri}
C_b&=\Tif{\ker}{\M\M}-\frac{\bra\sigma(z)\sigma(z) \ket_{\Tif{\ker}{\M\M}}}{\bra\sigma'(z)\sigma'(z) \ket_{\Tif{\ker}{\M\M}}} \, ,
\end{align}
satisfies both the fixed-point equation $\Tif{\ker}{\M\M}=C_b+C_W g_0\!\le(\Tif{\ker}{\M\M}\ri)$ as well as the first \terminate{criticality} condition $\chi_{\perp}\!\le(\Tif{\ker}{\M\M}\ri)=1$.
The second \terminate{criticality} condition $\chi_{\parallel}\!\le(\Tif{\ker}{\M\M}\ri)=1$ then is tantamount to $\chi_{\perp}\!\le(\Tif{\ker}{\M\M}\ri)/\chi_{\parallel}\!\le(\Tif{\ker}{\M\M}\ri)=1$, which is simply the following ratio of expectations
\be\label{easycriticality}
\le[\frac{2\ker^2 \bra\sigma'(z)\sigma'(z)\ket_{\ker}}{\bra \sigma(z)\sigma(z)\le(z^2-K\ri)\ket_{\ker}}\ri]\Bigg\vert_{\ker=\Tif{\ker}{\M\M}}=1\, ,
\ee
independent of the \terminate{initialization hyperparameters} $C_W$ and $C_b$.
Therefore, we can use the following simpler algorithm:
\begin{enumerate}
\item Scan over values of $\Tif{\ker}{\M\M}\geq 0$ until \eqref{easycriticality} is satisfied.
\item Plug the resulting value of $\Tif{\ker}{\M\M}$ into~\eqref{CWcri} and~\eqref{Cbcri} to evaluate the critical initialization hyperparameters (and also make sure $C_b\geq0$).
\end{enumerate}
In Figure \ref{fig:tanh_critical}, the left-hand side of~\eqref{easycriticality} is plotted as a function of $\Tif{\ker}{\M\M}$ for the $\tanhA$ activation function, which we see hits unity at $\Tif{\ker}{\M\M}=0$. \index{initialization hyperparameters!critical} Then, evaluating equations~\eqref{CWcri} and~\eqref{Cbcri} in the limit $\Tif{\ker}{\M\M}\rightarrow0$ efficiently gives the critical initialization hyperparameters for $\tanhA$: $(C_W,C_b)=(1,0)$\index{universality class!K@$K^\star=0$}.\footnote{
Even though the fixed-point value of the midpoint kernel\index{kernel!midpoint} is zero, this is a \emph{nontrivial} fixed point.\index{fixed point!nontrivial} In particular, we will see in~\S\ref{subsec:tanh_univ} that kernels with a nontrivial fixed point\index{fixed point!nontrivial} at $\Tif{\ker}{\M\M}=0$ form a \emph{universality class}, characterized by a benign power-law decay in $\ell$. In practice, the power-law behavior means that for any finite depth the kernel will remain finite.}
\index{universality class!K@$K^\star=0$}

\index{initialization hyperparameters!critical!for scale-invariant universality}\index{universality class!scale-invariant}
In passing, we note that scale-invariant activation functions trivially satisfy the condition~\eqref{easycriticality} for any fixed-point value $\Tif{\ker}{\M\M}$, since the susceptibilities are equal to the same kernel-independent constant, $\chi_{\parallel}\!\le(\ker\ri)=\chi_{\perp}\!\le(\ker\ri)=\chi$. It's easy to check that for this universality class, the above algorithm recovers the critical initialization hyperparameters~\eqref{eq:criticality_scale_invariant} given in~\S\ref{sec:scale-invariant-eft}.

\subsection{No Criticality: sigmoid, softplus, nonlinear monomials, etc.}\label{subsec:no_thinking_no_money}
\begin{figure}[h]
\begin{center}
 \includegraphics[width=0.85\linewidth]{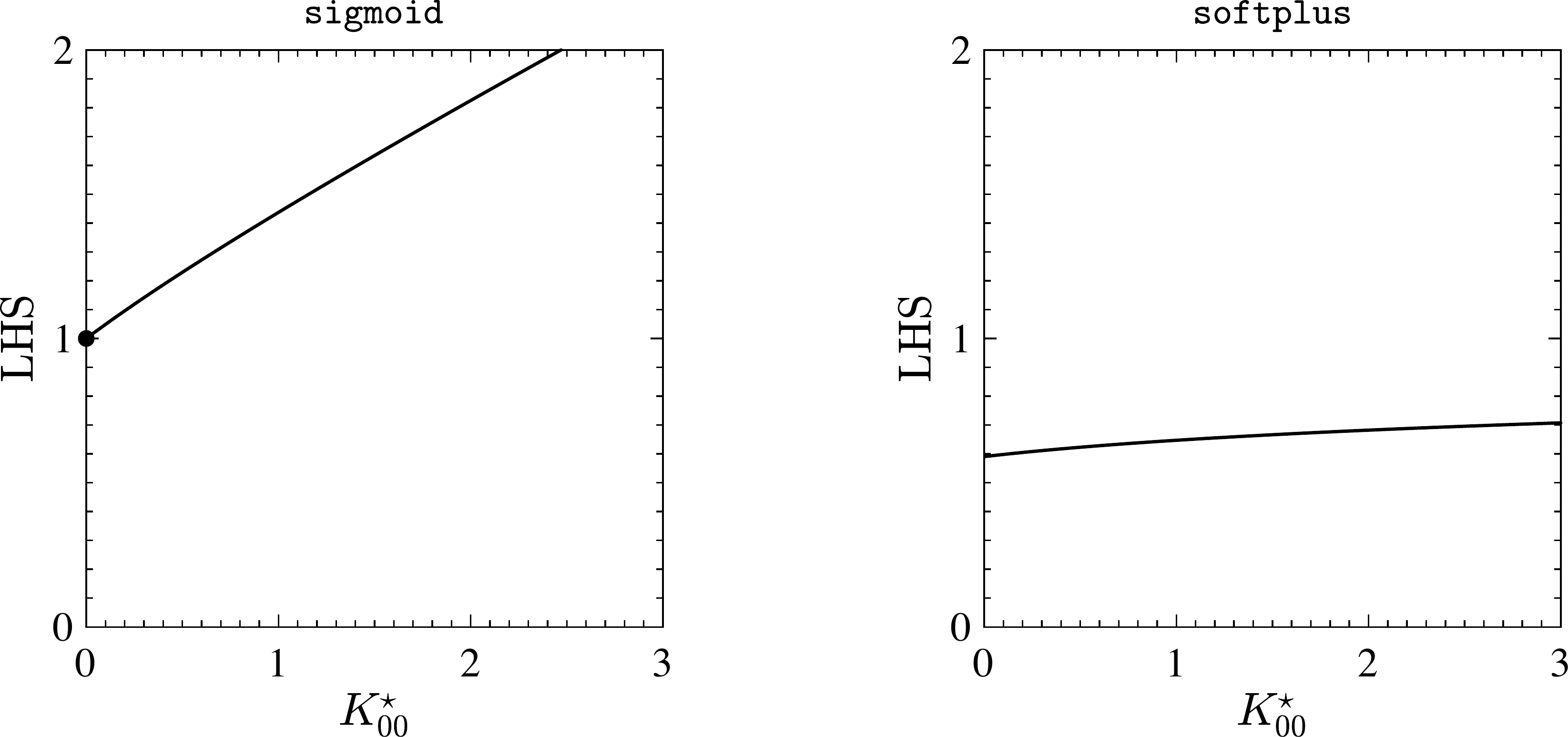}
\end{center}
\caption{The left-hand side of the condition~\eqref{easycriticality} is plotted as a function of $\Tif{\ker}{\M\M}$ for the $\sigmoid$ activation function (left) and the $\softplus$ activation function (right).
For the $\sigmoid$, the plotted line hits unity as $\Tif{\ker}{\M\M}\rightarrow 0$, but the associated critical initialization hyperparameters $(C_b,C_W)$ are unphysical because $C_b<0$.
For the $\softplus$, the plotted line does not hit unity.
These activation functions cannot be tuned to \terminate{criticality}.}
\label{fig:no-crit}
\end{figure}

For some activation functions, a nontrivial fixed point\index{fixed point!nontrivial} for the kernel does not exist. For example, consider the $\sigmoid$ activation function
\be
\sigma(z)=\frac{1}{1+e^{-z}}\, .
\ee
The condition~\eqref{easycriticality} is plotted for this activation in Figure~\ref{fig:no-crit}. While this condition is satisfied at  $\Tif{\ker}{\M\M}=0$, evaluating \eqref{Cbcri} in this limit yields $C_b=-\le(\frac{\sigma(0)}{\sigma'(0)}\ri)^2<0$. Since the variance of the bias cannot be negative, this is unphysical.\footnote{The limiting value of $C_b=-\le(\frac{\sigma(0)}{\sigma'(0)}\ri)^2$ hints that the conditions $\sigma(0)=0$ and $\sigma'(0) \neq 0$ may be necessary constraints for an activation function to have a nontrivial fixed point.}
Thus, the $\sigmoid$ cannot be tuned to \terminate{criticality} and should not be used.\footnote{Similarly, as a non-smooth limit of a \terminate{logistic function}, the $\perc$ activation function is even worse and doesn't merit discussion.}

Next let's consider the $\softplus$ activation function
\be
\sigma(z)=\log\!\le(1+e^{z}\ri)\, ,
\ee
which, as a reminder, is a smooth approximation of the $\relu$.
Plotting the condition~\eqref{easycriticality} in Figure~\ref{fig:no-crit}, we see that it cannot be satisfied for any $\Tif{\ker}{\M\M}\geq0$.
Thus, in contrast to the $\relu$, the $\softplus$ cannot be tuned to \terminate{criticality}. This supports the lore in the community that the $\relu$ is superior to the $\softplus$, despite their similarity and the $\softplus$' smoothness.

As we will see in the next subsection, the real problem with these activation functions is that they do not cross zero at $z=0$. There is an easy fix, namely, setting
\be
\sigma(0)=0\, ,
\ee
by an appropriate constant shift for each activation. With such a shift the $\sigmoid$ turns into the $\tanhA$, albeit with the preactivation and activation each scaled by a half. Such a scaled $\tanhA$ indeed admits a critical initialization, which is easy to check after reading the discussion in the next subsection.

With that in mind, let's see what happens for activation functions that cross zero nonlinearly.
For simplicity, take any nonlinear $\texttt{monomial}$ activation function \index{activation function!monomial@\texttt{monomial}} 
\be
\sigma(z)=z^p\, ,  \quad p=2,3,4,\ldots\, .
\ee
In this case, direct Gaussian integration translates the condition~\eqref{easycriticality}  into the constraint
\be
\frac{p}{2p-1}=1 \, ,
\ee
which cannot be satisfied for nonlinear monomials, since $p\ne 1$. Thus, such nonlinear monomials also shouldn't be used in deep networks.
More importantly, in addition to $\sigma(0)=0$, \terminate{criticality} seems to require
the condition
\be
\sigma'(0)\ne 0\, ,
\ee
which we will investigate more generally in the next subsection.

The impossibility of \terminate{criticality} for all of the activation functions discussed in this subsection means that their use should be discouraged. While the problem is somewhat mitigated for shallow networks -- since there are fewer layers for the exponential behavior to damage the signals -- as networks become deeper and deeper, \terminate{criticality} becomes more and more essential.

\subsection{\texorpdfstring{$\Tif{\ker}{}=0$}{K*=0} Universality Class: tanh, sin, etc.}\label{subsec:tanh_univ}
\index{universality class!K@$K^\star=0$}

In \S\ref{subsec:strategize},  we learned through a numerical investigation that $\tanhA$ has a nontrivial fixed point\index{fixed point!nontrivial} at $\Tif{\ker}{\M\M}=0$. In addition, in the last subsection~\S\ref{subsec:no_thinking_no_money}, 
our analysis suggested
that the conditions $\sigma(0)=0$ and $\sigma'(0)\ne 0$ are %
important %
for any smooth activation to have a nontrivial fixed point\index{fixed point!nontrivial}. 

 \index{universality class!K@$K^\star=0$}
In this subsection, we will connect these two observations.
In particular,  in the vicinity of $\Tif{\ker}{\M\M}=0$, we can analytically analyze the kernel recursions \eqref{K0}--\eqref{K2} by Taylor expanding around $\Tif{\ker}{\M\M}=0$ and directly integrating the Gaussian expectations. This analysis will show that the conditions $\sigma(0)=0$ and $\sigma'(0)\ne 0$ are both necessary and sufficient for a smooth activation function to have a nontrivial fixed point at $\Tif{\ker}{\M\M}=0$, leading to the definition of our second universality class.

Let's use the following notation for the Taylor coefficients of any analytic activation function:
\be \label{eq:taylor-expansion-activation}
\sigma(z)=\sum_{p=0}^{\infty}\frac{\sigma_{p}}{p!}z^p\, .
\ee
Plugging this expansion into the definition of the helper function \eqref{eq:helper-function-general-strategy}  and performing the Gaussian integral, we find 
\be
g\!\le(\ker\ri)=\bra\sigma(z)\sigma(z) \ket_{\ker}=\sigma_0^2+\le(\sigma_1^2+2\sigma_0\sigma_2\ri)\ker+\o{\ker^2}\, .
\ee
From this we see that the fixed point of the recursion for the midpoint kernel\index{kernel!midpoint}
\be
\Tif{\ker}{\M\M}=C_b+C_W g\!\le(\Tif{\ker}{\M\M}\ri)\, ,
\ee 
has a solution at $\Tif{\ker}{\M\M}=0$  if and only if $C_b=C_W\sigma_0^2=0$.
Recalling that $C_W=0$ violates the \terminate{criticality} conditions, we must pick $\sigma_0=0$. Henceforth we will assume that this choice has been made.

Continuing on with $\sigma_0=0$ and $C_b =0$ in mind, inserting the expansion \eqref{eq:taylor-expansion-activation} into our expressions for the susceptibilities, \eqref{eq:chi-parallel} and \eqref{eq:chi-perp}, and performing the Gaussian integrals we find %
\begin{align}\label{eq:g0}
C_W g(\ker)
&=\le(C_W \sigma_1^2\ri)\le[K+a_1 K^2+a_2 K^3+\o{K^4}\ri]\, , \\
\label{eq:chi-parallel-expansion-K-star-equals-zero}
\chi_{\parallel}(\ker)%
&=\le(C_W \sigma_1^2\ri)\le[1+2 a_1K+3 a_2 K^2+\o{K^3}\ri]\, ,\\
\label{eq:chi-perp-expansion-K-star-equals-zero}
\chi_{\perp}(\ker)%
&=\le(C_W\sigma_1^2\ri)\le[1+b_1 K+\o{K^2}\ri]\, ,
\end{align}
where here we have also expanded $g(K)$ to higher order in the kernel, and the coefficients $a_1$, $a_2$, and $b_1$ are given by the following combinations of Taylor coefficients of the activation function
\begin{align}
\label{eq:a1}
a_1&\equiv \le(\frac{\sigma_3}{\sigma_1}\ri)+\frac{3}{4}\le(\frac{\sigma_2}{\sigma_1}\ri)^2\ ,\\
a_2&\equiv \frac{1}{4}\le(\frac{\sigma_5}{\sigma_1}\ri)+\frac{5}{8}\le(\frac{\sigma_4}{\sigma_1}\ri)\le(\frac{\sigma_2}{\sigma_1}\ri)+\frac{5}{12}\le(\frac{\sigma_3}{\sigma_1}\ri)^2\, ,\\
\label{eq:b1}
b_1&\equiv \le(\frac{\sigma_3}{\sigma_1}\ri)+\le(\frac{\sigma_2}{\sigma_1}\ri)^2\, .%
\end{align}
It's easy to check that, e.g., for $\tanhA$ these coefficients take the following values
$a_1=-2$, $a_2=17/3$, $b_1=-2$.
Now, examining expansions \eqref{eq:chi-parallel-expansion-K-star-equals-zero} and \eqref{eq:chi-perp-expansion-K-star-equals-zero}, we see that to satisfy the \terminate{criticality} conditions $\chi_{\parallel}\!\le(\Tif{\ker}{\M\M}=0\ri)=1$ and $\chi_{\perp}\!\le(\Tif{\ker}{\M\M}=0\ri)=1$ we must set $C_W= 1/\sigma_1^2$. To ensure a finite variance, we also see that the activation function must have $\sigma_1\ne 0$.

Thus, for any smooth activation function to have a nontrivial fixed point\index{fixed point!nontrivial} at $\Tif{\ker}{\M\M}=0$, it is necessary and sufficient that $\sigma(z)$ satisfy
\be\label{eq:tanh_cri_condition}
\sigma_0=0\, , \qquad  \sigma_1\ne0\, .
\ee
For such an activation, the critical initialization hyperparameters are then given by \index{initialization hyperparameters!critical!for K@for $K^\star=0$ universality}
\be\label{eq:k-star-equals-zero-critical-initialization}
\le(C_b, C_W\ri)^{\text{critical}}=\le(0, \frac{1}{\sigma_1^2}\ri)\, .
\ee
Just to emphasize this a bit, any activation with these conditions \eqref{eq:tanh_cri_condition} initialized with \eqref{eq:k-star-equals-zero-critical-initialization} will have a nontrivial fixed point at $\Tif{\ker}{\M\M}=0$. The set of activation functions that vanish at the origin with a nonzero first derivative make up the
\textbf{\boldmath$K^\star=0$ universality class}.\index{universality class!K@$K^\star=0$|textbf}
The canonical class member is the $\tanhA$ activation function, though there are obviously a very large number of members in this class, e.g.~the $\sinA$ activation function is a member too.

\index{initialization hyperparameters!critical}
Having determined the critical initialization hyperparameters, let's now try to understand the behavior of the kernel 
for the $K^\star=0$ universality class.
We will see that when tuned to \terminate{criticality} the activations satisfying \eqref{eq:tanh_cri_condition} all behave similarly under RG flow, with the large-depth behavior of the kernel depending only on the first few Taylor coefficients of $\sigma(z)$.
\index{representation group flow}

\subsubsection{Deep asymptotic analysis for the midpoint kernel}

Recalling the expansion  $\Ti{ \ker}{\M\M}{\ell}=\Tif{\ker}{\M\M}+\Delta\Ti{ \ker}{\M\M}{\ell}$ around the fixed point and considering the expansion \eqref{eq:g0} for $g(K)$, the  midpoint kernel\index{kernel!midpoint} recursion at $\Tif{ \ker}{\M\M}=0$ \terminate{criticality} becomes
\be\label{eq:K-star-equals-zero-single-input-imprecise}
\Ti{\Delta \ker}{\M\M}{\ell+1}=\Ti{ \Delta\ker}{\M\M}{\ell}+a_1\le(\Ti{ \Delta\ker}{\M\M}{\ell}\ri)^2+a_2\le(\Ti{\Delta \ker}{\M\M}{\ell}\ri)^3+O\!\le(\le(\Ti{ \Delta\ker}{\M\M}{\ell}\ri)^4\ri)\, .
\ee
Since the whole point of \terminate{criticality} is to alleviate exponential behavior, we expect a gentler decay back to the $\Ti{ \ker}{\M\M}{\ell}=0$ fixed point. 
With that in mind, let's plug a power-law ansatz $\Ti{\Delta\ker}{\M\M}{\ell}\sim\le(\frac{1}{\ell}\ri)^{p_\M}$ into \eqref{eq:K-star-equals-zero-single-input-imprecise}. Noting that $\le(\frac{1}{\ell+1}\ri)^{p_\M}=\frac{1}{\ell^{p_\M}}\le[1-\frac{p_\M}{\ell}+O\!\le(\frac{1}{\ell^2}\ri)\ri]$ and matching the leading terms on both sides, we get a solution 
\be\label{eq:initial-tanh-wthout-subleading}
\Ti{\Delta\ker}{\M\M}{\ell}=\le[\frac{1}{(-a_1)}\ri]\frac{1}{\ell}+ \ldots \, .%
\ee
Thus, the behavior at \terminate{criticality} is a mild power law decay, with a \term{critical exponent} $p_\M=1$. Such an exponent is said to be \emph{universal} for the $K^\star=0$ universality class, since it is completely independent of the details of the particular activation function.

Importantly, for this asymptotic solution to be consistent, we must have $(-a_1)>0$ to ensure the positivity of the kernel. If instead we had $(-a_1) <0$, then the asymptotic solution~\eqref{eq:initial-tanh-wthout-subleading} would be negative, making it invalid. In this case the fixed point would be unstable, exponentially repelling the kernel away from $\Tif{ \ker}{\M\M}=0$\index{universality class!scale-invariant}.\footnote{Generically, $(-a_1)<0$ implies that $\chi_{\parallel}>1$ away from $\Tif{ \ker}{\M\M}=0$,
which repels the midpoint kernel\index{kernel!midpoint} first with a power law and then exponentially. However, the semi-criticality\index{criticality!semi-criticality} that we discussed in \S\ref{sec:scale-invariant-eft} for scale-invariant activations was exceptional. For this universality class, $a_1=0$ and hence growth towards the fixed point at infinity is governed by a power law.} 
We will see in the next subsection that 
$\swish$ and $\gelu$ activation functions exhibit such an instability near $\Tif{ \ker}{\M\M}=0$.

Moreover, in the last subsection we suggested that an activation function that doesn't satisfy $\sigma(0)=0$ could be potentially salvaged with a constant shift. In particular, perhaps the $\softplus$ could be saved by subtracting a constant $\log(2)$ so that $\sigma(0)=0$? %
However, in this case we'd have $(-a_1)<0$, and the kernel will get repelled from the only candidate nontrivial fixed point at $\Tif{ \ker}{\M\M}=0$.
And since $\chi_{\parallel}(K)>1$ away from $K=0$, the midpoint kernel\index{kernel!midpoint} will diverge exponentially.
Thus, despite this attempt, we see that the $\softplus$ cannot be saved.

Returning to our solution~\eqref{eq:initial-tanh-wthout-subleading}, we can actually do quite a bit better than ``$\ldots$'' for the subleading asymptotic analysis. As a first guess to improve our ansatz, let's include a subleading $1/\ell^2$ term in $\Ti{\Delta\ker}{\M\M}{\ell}$. However, if we try to match terms on both sides of \eqref{eq:K-star-equals-zero-single-input-imprecise}, we'd find that there's no way of canceling the $1/\ell^3$ terms.
What we can do instead is to also add $\log(\ell) /\ell^2$ with an independent coefficient to our ansatz. This generates an additional $1/\ell^3$ term, allowing for a consistent solution.
Generally for any of the observables $\O^{(\ell)}$ that we will consider, the correct \term{scaling ansatz} for the large-$\ell$ asymptotic expansion  is of the form
\begin{align}\label{eq:master-scaling-ansatz}
\Ti{\O}{}{\ell} &= \le( \frac{1}{\ell} \ri)^{p_\O} \le[c_{0,0}+  c_{1,1} \le( \frac{\log \ell}{\ell} \ri) + c_{1,0}\le( \frac{ 1}{\ell}\ri) +  c_{2,2} \le(  \frac{\log^2 \ell}{\ell^2} \ri)+  \dots \ri]%
\notag \\
&= \le( \frac{1}{\ell} \ri)^{p_\O} \le[\sum_{s = 0}^\infty \sum_{q=0}^{s} c_{s,q} \le( \frac{\log^q \ell}{\ell^s}  \ri)\ri],
\end{align}
where the \terminate{critical exponent} $p_\O$ is expected to be universal for a given class, while the constants $c_{s,q}$ will depend on the details of a particular activation function.
Carrying this process forward for $\O^{(\ell)} = \Ti{\Delta\ker}{\M\M}{\ell}$, we can systematically determine the subleading behavior of the kernel perturbation as
\begin{align}\label{eq:tanh_asymptotic}
\Ti{\Delta\ker}{\M\M}{\ell}&=\le[\frac{1}{(-a_1)}\ri]\frac{1}{\ell}+\le[\frac{-(a_2-a_1^2)}{a_1^3}\ri]\frac{\log\!\le(\frac{\ell}{\ell_0}\ri)}{\ell^2}\, \\
&+\le[\frac{-\le(a_2-a_1^2\ri)^2}{a_1^5}\ri]\frac{\le[\log\!\le(\frac{\ell}{\ell_0}\ri)\ri]^2}{\ell^3}+\le[\frac{\le(a_2-a_1^2\ri)^2}{a_1^5}\ri]\frac{\log\!\le(\frac{\ell}{\ell_0}\ri)}{\ell^3}+O\!\le(\frac{1}{\ell^3}\ri)\, ,\nonumber
\end{align}
and with enough effort this asymptotic expansion can be refined to arbitrary degree by %
including the higher-order corrections according to the scaling ansatz \eqref{eq:master-scaling-ansatz} described above. 

Here, the constant $\ell_0$ is undetermined by this large-$\ell$ asymptotic analysis and nontrivially depends on the input norm through
\be
\Ti{\ker}{\M\M}{1}=\frac{1}{\sigma_1^2}\frac{1}{n_0}\sum_{i=1}^{n_0}\x{i}{0}^2\, ,
\ee
which sets the initial condition~\eqref{eq:kernel-initial-condition} for the kernel recursion~\eqref{eq:kernel-recursion-reminder} when the rescaled weight variance is set to \terminate{criticality}, $C_W=1/\sigma_1^2$.
To get a sense of what this means, let's assume that $\chi_{\parallel}(\ker)$ is monotonically decreasing for $\ker\geq0$ with $\chi_{\parallel}(0)=1$ -- as is true for $\tanhA$ --  and consider what happens when an input $\x{i}{0}$ has a very large magnitude. Such a large-norm input will lead to a large value for the first-layer midpoint kernel\index{kernel!midpoint}, $\Ti{\ker}{\M\M}{1} \gg 1$. 
In the range $0<k_{\sharp}<\Ti{\ker}{\M\M}{\ell}$, for some constant $k_{\sharp}$, the kernel $\Ti{\ker}{\M\M}{\ell}$ will decay quicker than $\chi_{\parallel}(k_{\sharp})^{\ell}$, with $\chi_\parallel(k_{\sharp})<1$, until it enters the power-law regime near $\Tif{\ker}{\M\M}=0$. The undetermined constant $\ell_0$ is
a remnant of this complicated crossover behavior, capturing the leading \neo{data dependence}\index{data dependence|seealso{data-dependent coupling}}\index{data dependence|seealso{connected correlator}} of the midpoint kernel. 

\index{representation group flow}
Additionally, the asymptotic expansion for the midpoint kernel\index{kernel!midpoint} \eqref{eq:tanh_asymptotic} has a nice interpretation under RG flow. While the critical exponent of the falloff $p_\M=1$ is generic for the universality class, we see that the coefficients of the terms do depend on the details of the activation function, albeit only the first few Taylor coefficients. In fact, for larger and larger $\ell$, the dependence is on fewer and fewer of the coefficients, with the leading term only depending on $a_1$, \eqref{eq:a1}. In this asymptotic limit, any activation function in the $K^\star=0$ universality class with the same first three Taylor coefficients around zero will be completely indistinguishable. Thus, from the representation group flow perspective, one of the results of having a deeper network is to make the particular details of the activation function %
more and more irrelevant.

\index{critical exponent}
Lastly, let us note for all aspiring ``activation designers'' out there that we can engineer critical exponents other than $p_\M = 1$ by fine-tuning the Taylor coefficients of the activation function. For example, if we set $a_1=0$ by balancing $\sigma_3$ and $\sigma_2$, then the kernel approaches a $\Tif{\ker}{\M\M}=0$ nontrivial fixed point\index{fixed point!nontrivial} with a $ 1/\sqrt{\ell}$ power law decay so long as $(-a_2)>0$.
The need for such tuning indicates that the $\sim 1/\ell$ behavior is generic for activation functions in the $K^\star=0$ universality class.\footnote{
    More precisely, we should have defined the $K^\star=0$ universality class with the requirement $a_1\neq0$. This in turn would lead us to define a whole family of universality classes labeled by the degree of fine tuning of the $a_1, a_2,$ etc., or equivalently labeled by the value of the critical exponent $p_\M$. %
}  %

\subsubsection{Deep asymptotic analysis for parallel perturbations}
Next, let's solve the $\Ti{\delta\ker}{[1]}{\ell}$ recursion for parallel perturbations. 
Plugging the expansion \eqref{eq:chi-parallel-expansion-K-star-equals-zero} for $\chi_\parallel(K)$ into the recursion \eqref{K1}, we get an algebraic equation
\be\label{eq:parallel-agebraic}
\Ti{\delta\ker}{[1]}{\ell+1}=\le[1+2a_1\Ti{ \Delta\ker}{\M\M}{\ell}+3a_2\le(\Ti{ \Delta\ker}{\M\M}{\ell}\ri)^2+O\!\le(\le(\Ti{ \Delta\ker}{\M\M}{\ell}\ri)^3\ri)\ri]\Ti{\delta \ker}{[1]}{\ell}\, .
\ee
Then, plugging in the large-$\ell$ solution for $\Ti{\Delta\ker}{\M\M}{\ell}$ \eqref{eq:tanh_asymptotic}
and a large-$\ell$ asymptotic expansion for $\delta \ker_{[1]}^{(\ell)}$ based on our \terminate{scaling ansatz}~\eqref{eq:master-scaling-ansatz}, we can solve
the resulting equation by matching the terms on both sides:
\be
\delta \ker_{[1]}^{(\ell)}= \frac{\delta_{\parallel}}{\ell^2}\le[1+\frac{2a_1\le(a_2-a_1^2\ri)}{a_1^3}\frac{\log\!\le(\frac{\ell}{\ell_0}\ri)}{\ell}+O\!\le(\frac{1}{\ell}\ri)\ri]\, .
\ee

\index{universality class!K@$K^\star=0$} Inspecting our solution, we identify our second \terminate{critical exponent} for the $K^\star=0$ universality class: $p_\parallel = 2$ corresponding to the $1/\ell^2$ falloff of $\delta \ker_{[1]}^{(\ell)}$. The particular value of this exponent is to be expected.
As noted before, the parallel perturbation is just a difference of single-input kernels for two inputs with differing norms, $\Ti{\ker}{[1]}{\ell}=\le(\Ti{\ker}{++}{\ell}-\Ti{\ker}{--}{\ell}\ri)/2$. The leading $1/\ell^2$ scaling occurs because the diagonal components $\Ti{\ker}{++}{\ell}$ and $\Ti{\ker}{--}{\ell}$ are governed by the same asymptotic behavior up to order $\log(\ell)/\ell^2$, including the same coefficients. Thus, the leading difference appears at order $1/\ell^2$, due to different input-dependent constants $\ell_+$ and $\ell_-$ in expansions analogous to \eqref{eq:tanh_asymptotic} for $\Ti{\ker}{++}{\ell}$ and $\Ti{\ker}{--}{\ell}$, with the undetermined constant $\delta_{\parallel}\propto \log\!\le(\ell_{+}/\ell_{-}\ri)$.
In this way, this constant explicitly carries the \terminate{data dependence} of the parallel perturbation.

\subsubsection{Deep asymptotic analysis for perpendicular perturbations}
Finally, let's conclude our analysis by solving the $\Ti{\delta \delta\ker}{[2]}{\ell}$ recursion for perpendicular perturbations. 
Let's begin by plugging the expansion \eqref{eq:chi-perp-expansion-K-star-equals-zero} for $\chi_\perp(K)$ into the recursion \eqref{K2}. Since we want to focus on perpendicular perturbations with $\sum_{i=1}^{n_0}  \x{i}{\M}\, \delta x_i = 0$, we will also turn off parallel perturbations by setting $\delta \ker_{[1]}^{(\ell)}=0$. Putting this all together gives an algebraic equation
\be
\Ti{\delta\delta \ker}{[2]}{\ell+1}=\le[1+b_1\Ti{ \Delta\ker}{\M\M}{\ell}+O\!\le(\le(\Ti{\Delta \ker}{\M\M}{\ell}\ri)^2\ri)\ri]\Ti{\delta\delta \ker}{[2]}{\ell}\, .
\ee
Plugging in the large-$\ell$ asymptotic solution for $\Ti{\Delta\ker}{\M\M}{\ell}$ and solving the resulting equation with another large-$\ell$  asymptotic expansion  for $\delta\delta \ker_{[2]}^{(\ell)}$ based on our \terminate{scaling ansatz} \eqref{eq:master-scaling-ansatz},
we get
\be\label{eq:perp-asymptotic-solution}
\delta\delta \ker_{[2]}^{(\ell)}= \frac{\delta^2}{\ell^{\frac{b_1}{a_1}}}\le[1+\frac{b_1\le(a_2-a_1^2\ri)}{a_1^3}\frac{\log\!\le(\frac{\ell}{\ell_0}\ri)}{\ell}+O\!\le(\frac{1}{\ell}\ri)\ri]\,  ,
\ee
where $\delta^2$ is another unfixed constant undetermined by the large-$\ell$ solution, in this case related nontrivially to the magnitude of the difference of the inputs: $\sum_{i=1}^{n_0}\le(\x{i}{+} -  \x{i}{-}\ri)^2$. Here we see that the presumptive \terminate{critical exponent}, $p_\perp \equiv b_1 / a_1$, depends mildly on the details of the activation function.

\index{universality class!K@$K^\star=0$}
However, note that something nice happens for odd activation functions such as $\tanhA$ and $\sinA$. In this case, we see from \eqref{eq:a1} and \eqref{eq:b1} that $a_1 = b_1$, giving us a bona fide critical exponent, $p_\perp = 1$, when restricting the universality class to odd activations. This means that perpendicular perturbations decay with the same power in $\ell$ as the midpoint kernel\index{kernel!midpoint} decays to the fixed point, $\sim 1/\ell$. Thus, at \terminate{criticality} the ratio $ \ker_{[2]}^{(\ell)} / \ker_{[0]}^{(\ell)}$ is fixed at the leading order, preserving the angles between nearby perpendicular inputs. Importantly, this ensures that the relationship between input points is conserved under the RG flow, even if the signals propagate through a very deep network.\index{representation group flow}

Furthermore, the milder falloff of the perpendicular perturbations suggests that they are in some sense more important than the parallel ones. 
This is because with enough depth the $\Ti{\ker}{[1]}{\ell}$ component will become subleading to the $\Ti{\ker}{[0]}{\ell}$ and the $\Ti{\ker}{[2]}{\ell}$ components, due to the $1/\ell^2$ scaling of the former compared to the $1/\ell$ scaling of the latter two.
For this reason, we are going to
ignore these parallel perturbations of the kernel going forward.

\subsection{Half-Stable Universality Classes: SWISH, etc.~and GELU, etc.}\label{subsec:half_stability}\index{fixed point!nontrivial!half-stable}\index{fixed point!nontrivial!half-stable|seealso{\texttt{SWISH}}}\index{fixed point!nontrivial!half-stable|seealso{\texttt{GELU}}}
In this final subsection, we consider two other semi-popular activation functions in order to explore nontrivial fixed points away from zero, $\ker_{\M\M}^{\star}\ne 0$.
 \begin{figure}[ht]
\begin{center}
 \includegraphics[width=0.85\linewidth]{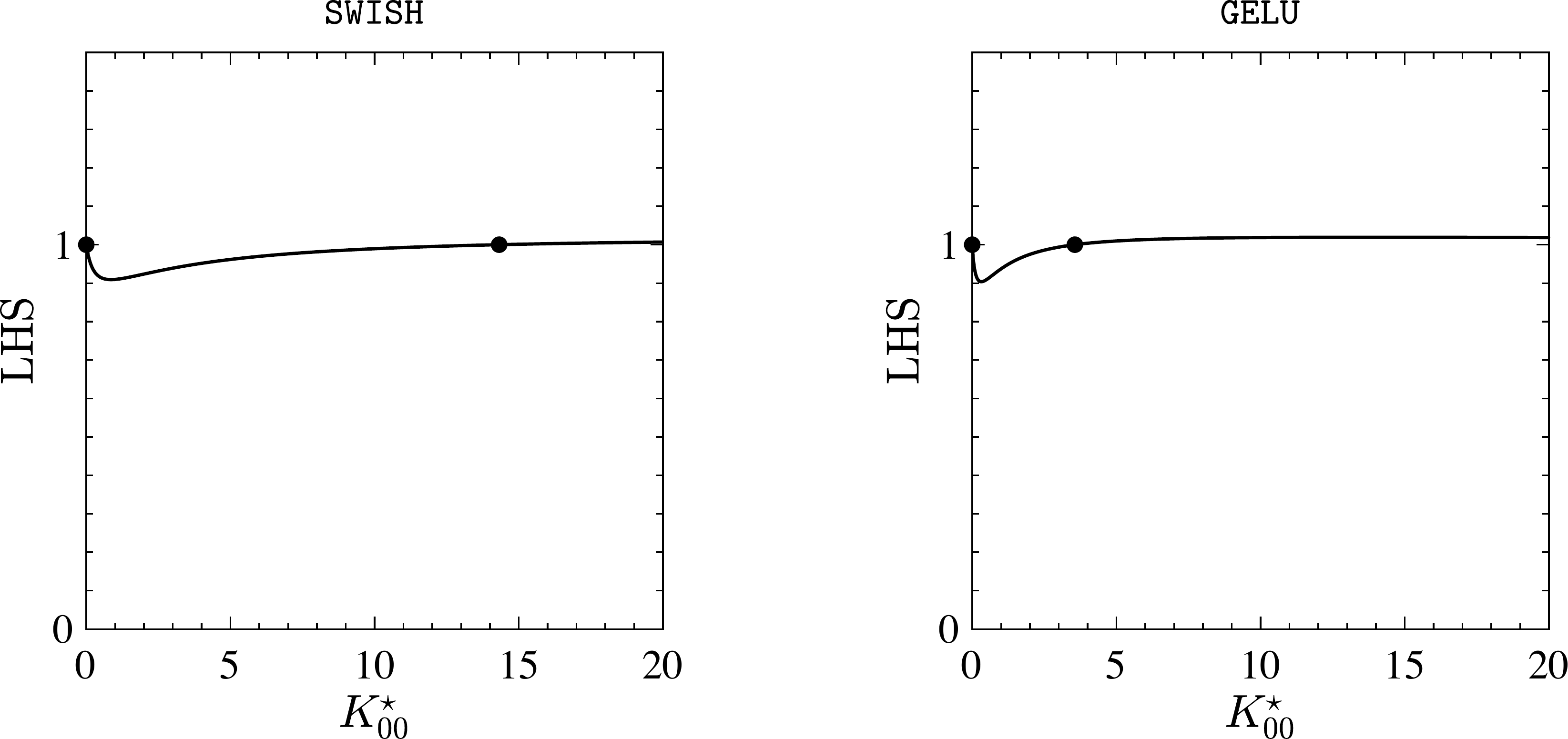}
\end{center}
\caption{The left-hand side of the condition~\eqref{easycriticality} is plotted as a function of $\Tif{\ker}{\M\M}$ for the $\swish$ activation function (left) and the $\gelu$ activation function (right).
For both activation functions, the plotted line hits unity (black dots) at $\Tif{\ker}{\M\M}=0$ as well as at a nonzero half-stable nontrivial fixed point\index{fixed point!nontrivial!half-stable} $\Tif{\ker}{\M\M}\ne0$.}
\label{fig:half-crit}
\end{figure}
\begin{itemize}
\item The $\swish$ activation function is defined as
\be
\sigma(z)=\frac{z}{1+e^{-z}}\, .
\ee
Similar to the intuition for the $\softplus$, the $\swish$ is intended as a smooth version of the $\relu$.
Following our general algorithm in~\S\ref{subsec:strategize} for finding the critical initialization hyperparameters, we actually find two nontrivial fixed points for the kernel, see Figure~\ref{fig:half-crit}.  In particular, the condition~\eqref{easycriticality} is met at $\Tif{\ker}{\M\M}=0$ with $\le(C_b,C_W\ri)=\le(0,4\ri)$ and at
$\Tif{\ker}{\M\M}\approx 14.32017362$ with
\be\label{eq:swish-half-criticality}
\le(C_b,C_W\ri)\approx\le(0.55514317, 1.98800468\ri)\, .
\ee 
 For the $\Tif{\ker}{\M\M}=0$ nontrivial fixed point, one can check that $(-a_1)<0$, and hence it's unstable. For the $\Tif{\ker}{\M\M}\approx 14.3$ nontrivial fixed point, we expand the midpoint kernel\index{kernel!midpoint} recursion as $\KML=\Tif{\ker}{\M\M}+\Delta \KML$, yielding
 \be
\Delta \Ti{\ker}{\M\M}{\ell+1}=\Delta \Ti{\ker}{\M\M}{\ell}+\tilde{a}_1\le(\Delta \Ti{\ker}{\M\M}{\ell}\ri)^2+O\!\le(\le(\Delta \Ti{\ker}{\M\M}{\ell}\ri)^3\ri)\, ,
\ee
with 
$(-\tilde{a}_1)\approx-2.84979219\cdot 10^{-6}$. \index{initialization hyperparameters!critical} 

Here, the large-$\ell$ asymptotic analysis around the finite fixed point is identical to the case of $\Tif{\ker}{\M\M}=0$, resulting in
\be
\Delta \KML\sim\le[\frac{1}{\le(-\tilde{a}_1\ri)}\ri]\frac{1}{\ell} \, .
\ee
However, the interpretation is slightly different, because the fixed-point value $\Tif{\ker}{\M\M}\approx 14.3$ is non-vanishing. In particular, this implies that when $\KML<\Tif{\ker}{\M\M}$ the kernel is attracted to the fixed point, while when $\KML>\Tif{\ker}{\M\M}$ the kernel is repelled.\footnote{With the half-critical initialization hyperparameters for the $\swish$~\eqref{eq:swish-half-criticality}, there is a \emph{trivial} fixed point at $\Tif{\ker}{\M\M}\approx 14.5$ that exponentially attracts the midpoint kernel when $\KML>14.3$.} Hence, this fixed point is \textbf{half-stable}\index{fixed point!nontrivial!half-stable}, and so the activation function is perhaps half-useful. In practice, however,  $\vert \tilde{a}_1\vert $ is small enough that the $\swish$ behaves in an almost scale-invariant manner around $\KML\sim\Tif{\ker}{\M\M}\approx 14.3$.\index{scale invariance}
\index{fixed point!trivial}
 
\item The $\gelu$ activation is defined as 
\be
\sigma(z)=\frac{z}{2}\le[1+\text{erf}\le(\frac{z}{\sqrt{2}}\ri)\ri]\, ,
\ee
and as a reminder is another smoothed $\relu$.
Following our recipe for criticality, the condition~\eqref{easycriticality} is again met twice, at $\Tif{\ker}{\M\M}=0$ with $\le(C_b,C_W\ri)=\le(0,4\ri)$ and at
 $\Tif{\ker}{\M\M}=\frac{3+\sqrt{17}}{2}$ with
 \be\label{eq:gelu-half-criticality}
 \le(C_b,C_W\ri)\approx\le(0.17292239, 1.98305826\ri)\, ,
\ee
see Figure~\ref{fig:half-crit}.
Similar to the $\swish$, the fixed point at $\Tif{\ker}{\M\M}=0$ is unstable with $(-a_1)=-6/\pi<0$, and the fixed point at $\Tif{\ker}{\M\M}=\frac{3+\sqrt{17}}{2}$ is half-stable\index{fixed point!nontrivial!half-stable}, in this case with
$(-\tilde{a}_1)\approx (1.43626419)\cdot 10^{-4}$.
Note that the sign of $\tilde{a}_1$ here differs from the sign for the $\swish$.
Thus, this time, when $\KML>\Tif{\ker}{\M\M}$ the midpoint kernel is attracted to the fixed point, while when $\KML<\Tif{\ker}{\M\M}$ it is repelled.\footnote{With the half-critical initialization hyperparameters for the $\gelu$~\eqref{eq:gelu-half-criticality}, there is a \emph{trivial} fixed point at $\Tif{\ker}{\M\M}\approx 3.2$ that exponentially attracts the midpoint kernel when $\KML<\frac{3+\sqrt{17}}{2}\approx 3.6$.} \index{fixed point!trivial}
Note that the absolute value $\vert \tilde{a}_1\vert $ is bigger for the $\gelu$ than for the $\swish$, meaning that it behaves less scale-invariantly and looks less like the $\relu$. \index{scale invariance}
\end{itemize} \index{initialization hyperparameters!critical}\index{fixed point!nontrivial!half-stable}
Unlike the shifted $\softplus$ which admits only an unstable nontrivial fixed point at $\ker_{\M\M}^{\star}=0$, here the non-monotonicity of the $\gelu$ and $\swish$ activation functions gave rise to half-stable nontrivial fixed points at $\ker_{\M\M}^{\star}\ne 0$.
They are both representatives of \emph{half-stable universality classes}.\index{universality class!half-stable}
For both of these $\relu$-like activations functions, the critical initialization hyperparameters for the $\ker_{\M\M}^{\star}\ne 0$ half-stable nontrivial fixed points are very similar to the critical $\relu$ initialization $\le(C_b,C_W\ri) = (0,2)$;
the activations in each of these classes really are just small perturbations of the $\relu$. At the same time, the fact that there's a fixed point at a particular kernel value $\ker_{\M\M}^{\star}\ne 0$ indicates -- however weakly -- the introduction of a particular scale. This is one way to see that these universality classes break \terminate{scale invariance}.

In summary, despite being $\relu$-like and also smooth, both of the $\swish$ and $\gelu$ are inferior to the $\relu$ itself. If you want to use a smooth activation function, use $\tanhA$.

\section{Fluctuations}\label{sec:signal_prop_finite_width}

\index{$1/n$ expansion}\index{deep linear network}\index{representation group flow}
Now that we fully understand how to tune infinite-width networks
to \terminate{criticality}, let's back off this large-$n$ limit to analyze the behavior of realistic networks. Specifically, we're going to extend the finite-width analysis that we performed for deep linear networks in~\S\ref{sec:fluctuations_DLN} to MLPs with nonlinear activation functions.
Before diving in, let's review the motivation for carrying out such an analysis. %

\index{typicality}
First, note that practitioners only use a single network rather than an \terminate{ensemble} of networks.\footnote{Actually in some cases practitioners can use ensembles of networks, though the computational cost of such models grows in proportion to the number of networks in the ensemble.} 
As we have discussed, sometimes a single instantiation will generically deviate from the mean.
Therefore, in order to understand what \emph{typically} happens in a single instantiation for an observable of interest, we have to compute not only the mean but also instantiation-to-instantiation \terminate{fluctuations} around the mean. As we explained in \S\ref{sec:fluctuations_DLN}, such fluctuations are generically finite-width effects, controlled by the $1/n$-suppressed \terminate{four-point vertex} $\Ti{\FPV}{(\alpha_1\alpha_2)(\alpha_3\alpha_4)}{\ell}$.
If fluctuations are large, then a single instantiation can behave poorly, despite being sampled from an \terminate{initialization distribution} tune to \terminate{criticality}.

\index{statistical independence}\index{Gaussian distribution}\index{infinite-width limit}
Second, we saw in \S\ref{sec:deeper-layer-accumulation} that the infinite-width $\ell$-th-layer preactivation distribution factorizes as
\be\label{eq:infinite-distribution-factorization}
p\!\le(\zi{1}{\ell}, \dots, \zi{n_\ell}{\ell} \Big| \D \ri) = p\!\le(\zi{1}{\ell} \Big| \D \ri) \cdots p\!\le(\zi{n_\ell}{\ell} \Big| \D \ri) + \o{\frac{1}{n_\ell}} \, ,
\ee
where the distributions $p\!\le(\zi{i}{\ell} \Big| \D \ri)$ on each neuron are given by statistically independent Gaussian distributions.
(To emphasize the neural dependence here, we have included \terminate{neural indices} while suppressing \terminate{sample indices}.)
Recalling our discussion of \terminate{interactions} and statistical independence in \S\ref{sec:perturbation}, this means that intralayer correlations among neurons are entirely finite-width phenomenon.
Later, we will show how this lack of interactions connects to the fact that the representations\index{representation} of an infinite-width network cannot evolve during gradient-based learning. Thus, understanding these finite-width effects is a prerequisite to understanding how  
practical networks actually learn 
from input data\index{input data}.\footnote{We'll go into more detail about the role that these correlations play in the \terminate{inductive bias} of MLPs in~\S\ref{ch:bayesian-inference} and then connect these interactions to representation learning in~\S\ref{ch:features}.}

\index{metric!next-to-leading-order correction}\index{initialization hyperparameters!critical}\index{subleading corrections}
Third, finite-width corrections can modify the mean value of observables.
As we saw in~\S\ref{sec:loop-correction}, at finite width all observables in principle receive an infinite series of subleading corrections. For instance, a possible finite-width NLO correction to the metric, $\se{\alpha_1\alpha_2}{\ell}$, can shift the infinite-width metric, $G_{\alpha_1\alpha_2}^{\le\{0\ri\}(\ell)} \equiv \Ti{\ker}{\alpha_1\alpha_2}{\ell}$, a.k.a.~the \terminate{kernel}. 
Such a finite-width correction could potentially
ruin \terminate{criticality}, since our derivation of the critical initialization hyperparameters depended explicitly on the infinite-width fixed-point value of the kernel.\footnote{
    In \S\ref{subsec:relu_univ_revisit} we will show that the NLO metric $\se{\alpha_1\alpha_2}{\ell}=0$ vanishes for the scale-invariant universality class, which is why we didn't discuss this type of correction for deep linear networks\index{deep linear network} in \S\ref{ch:deep-linear-eft}.\index{universality class!scale-invariant}
} %

\index{metric!next-to-leading-order correction}\index{representation group flow}
There will be two main takeaways from this section.
\bi
\item First, we will find that the leading finite-width fluctuations scale with the depth-to-width ratio of the network, $L/n$. We saw the importance of this \neo{emergent scale} for the $\linear$ activation function in~\S\ref{sec:fluctuations_DLN}; here, we see that it persists very generally for nonlinear activation functions.
In the language of \S\ref{sec:marginalization-group-flow}, this means that finite-width corrections are \emph{relevant}\index{relevant (RG flow)} under representation group flow and that deeper networks deviate more and more from the simple \terminate{infinite-width limit}. This emphasizes the importance of including such corrections when analyzing such networks and -- taking into account the fact that overly deep networks suffer from overwhelming fluctuations -- suggests
that our perturbative \terminate{effective theory} works best in the regime where practical networks also work best.

\index{subleading corrections}\index{metric!next-to-leading-order correction}
\item Second,
the NLO metric $\se{\alpha_1\alpha_2}{\ell}$ is subdominant to the kernel $\Ti{\ker}{\alpha_1\alpha_2}{\ell}$ as long as an appropriate $\o{1/n}$ correction is made to 
$C_W$. This means that the NLO metric vanishes in the interpolating limit -- $n,L\rightarrow\infty$, with $L/n$ fixed -- and thus can safely be neglected
 for most wide networks of reasonable depths.
\ei

\subsubsection{A single input, reloaded}
In order to illustrate the important qualitative effects of finite width, we will again specialize to just a single input. 
The reason for this choice can be best understood by progressing through another twofold list: 
\bi
\item[\emph{(i)}] Once the two initialization hyperparameters, $C_b$ and $C_W$, are tuned to criticality at leading order by the one- and two-input analysis of the kernel, the only additional tuning comes from the single-input analysis of the NLO metric\index{metric!next-to-leading-order correction} $\se{\alpha_1\alpha_2}{\ell}$. Therefore, the multi-input solutions for the vertex and NLO metric\index{metric!next-to-leading-order correction} do not add anything to the \terminate{criticality} analysis.
\item[\emph{(ii)}]
The most interesting part of the two-input vertex is a component that gives variance of the input-output Jacobian\index{Jacobian!input-output} of the network. (As we described in footnote~\ref{footnote:Jacobian}, the mean value of this Jacobian is captured by 
the $\Ti{\ker}{[2]}{\ell}$ component of the kernel.) However, the would-be analysis of this input-output variance will be subsumed by our analysis of the variance of the \terminate{neural tangent kernel} in~\S\ref{ch:NTKa}, which more directly gives the variance of gradients relevant for training.
\ei

In the rest of this section we'll omit the $\alpha=0$ \terminate{sample indices}, since such notation is unnecessarily cumbersome when considering only a single input. We'll also simplify things further by picking all the hidden-layer widths to be equal
\be\label{eq:chapter-eft-hidden-layer-widths-equal}
n_1=n_2=\cdots=n_{L-1}\equiv n \, .
\ee 
In addition to being a sensible choice, this means notationally that we don't have to carry around factors of $n_{\ell} / n_{\ell-1}$ everywhere.
With these decisions in mind, the relevant recursions from~\S\ref{ch:ngp} become
\begin{align}\label{eq:finite-width-reprinted-kernel}
\Ti{\ker}{}{\ell+1}&=C_b+C_W g\!\le(\Ti{\ker}{}{\ell}\ri)\, ,\\
\label{eq:finite-width-reprinted-vertex}
\Ti{\FPV}{}{\ell+1}&=\chi^2_{\parallel}\!\le(\Ti{\ker}{}{\ell}\ri)\Ti{\FPV}{}{\ell}+C_W^2\le[\bra \sigma^4(z) \ket_{\Ti{\ker}{}{\ell}}-\bra \sigma^2(z) \ket_{\Ti{\ker}{}{\ell}}^2\ri]\, ,\\
\label{eq:finite-width-reprinted-self-energy}
\se{}{\ell+1}&=\chi_{\parallel}\!\le(\Ti{\ker}{}{\ell}\ri)\se{}{\ell}+\frac{1}{8}\, j\!\le(\Ti{\ker}{}{\ell}\ri)\frac{\Ti{\FPV}{}{\ell}}{\le(\Ti{\ker}{}{\ell}\ri)^2}\, ,
\end{align}
where the helper function $g(K)$ and the parallel susceptibility $\chi_{\parallel}(K)$ were defined in~\eqref{eq:helper_first} and~\eqref{eq:chi-parallel}, and we have defined another
helper function
\be\label{eq:g4}
j(K)\equiv C_W\le\langle \sigma(z)\, \sigma(z)\!\le[\le(\frac{z^2}{K}\ri)^2-6\le(\frac{z^2}{K}\ri)+3\ri]\ri\rangle_{K}\, .
\ee
These three recursions can be solved for each universality class by mirroring our bootstrap analysis of $\Ti{\ker}{00}{\ell}$, $\Ti{\delta \ker}{[1]}{\ell}$, $\Ti{\delta\delta\ker}{[2]}{\ell}$ in \S\ref{sec:scale-invariant-eft} and \S\ref{sec:non-scale-invariant-eft}.

\subsection{Fluctuations for the Scale-Invariant Universality Class}\label{subsec:relu_univ_revisit}
Recall from~\S\ref{sec:scale-invariant-eft} that the scale-invariant universality class contains any activation function of the form \index{universality class!scale-invariant}
\be\label{eq:scale-invariant-reprinted}
\sigma(z) = 
    \begin{cases}
    a_+ z \, , & z \ge 0\, , \\
     a_- z \, , & z < 0  \, ,
    \end{cases}
\ee
with the $\relu$ $(a_+=1, a_-=0)$ as the exemplar member to keep in mind. Also recall that for this class we evaluated the helper function as $g(K)=A_2 K$ and the parallel susceptibility as $\chi_{\parallel}=A_2 C_W\equiv\chi$, with the activation-dependent constant given by $A_2\equiv (a_+^2+a_-^2)/2$. The other terms in the new recursions \eqref{eq:finite-width-reprinted-vertex} and \eqref{eq:finite-width-reprinted-self-energy} can similarly be evaluated by computing Gaussian integrals on the half-line, yielding
\be\label{eq:acctivation-function-gaussian-cov}
C_W^2\le[\bra \sigma^4(z) \ket_{\ker}-\bra \sigma^2(z) \ket_{\ker}^2\ri]= C_W^2\le(3A_4-A_2^2\ri)\ker^2\, , \qquad
j(K)=0\, ,
\ee
with a new activation-dependent constant 
\be
A_4\equiv\frac{a_+^4+a_-^4}{2} \, ,
\ee
to pair with our other constant, $A_2$.
With these expressions, the three recursions can be simplified as
\begin{align}\label{eq:kernel-scale-invariant-after-sub}
\Ti{\ker}{}{\ell+1}&=C_b+\chi \Ti{\ker}{}{\ell}\, ,\\
\label{eq:vertex-scale-invariant-after-sub}
\Ti{\FPV}{}{\ell+1}&=\chi^2\le(\frac{3A_4}{A_2^2}-1\ri)\le(\Ti{\ker}{}{\ell}\ri)^2+ \chi^2\,\Ti{\FPV}{}{\ell}\, ,\\
\se{}{\ell+1}&=\chi\, \se{}{\ell}\, . \label{eq:self-energy-scale-invariant-after-sub}
\end{align}
As a reminder, we already solved the kernel recursion in \S\ref{sec:scale-invariant-eft}.

\index{metric!next-to-leading-order correction}
Things are now quite simple. \index{universality class!scale-invariant}
\bi
\item First, remember from \S\ref{sec:first-layer-gaussian} that the first layer preactivation distribution is always exactly Gaussian, implying that the first-layer two-point correlator
is simply given in terms of the first-layer kernel $\Ti{\ker}{}{1}$ to all orders in $n$
\be\label{eq:two-point-first-layer-reprinted}
\E{z^{(1)}_i z^{(1)}_j}=\delta_{ij} \ker^{(1)} \, .
\ee
This means that the first-layer NLO metric\index{metric!next-to-leading-order correction} must vanish $\se{}{1}=0$, and recursion \eqref{eq:self-energy-scale-invariant-after-sub} then tell us that the NLO metric will vanish in any subsequent layer. Thus, for activations in the scale-invariant universality class, we learn that the single-input metric does not get corrected at $\o{1/n}$. 
\item Second, let's focus on \terminate{criticality} by setting $C_b=0$ and $C_W = 1/A_2$. As discussed in \S\ref{sec:scale-invariant-eft}, this setting of hyperparameters fixes the kernel to be an input-dependent layer-independent constant 
\be
\Ti{\ker}{}{\ell}=\Tif{\ker}{}\equiv\frac{1}{A_2} \le(\frac{1}{n_0}\sum_{i=1}^{n_0}x_i^2\ri) \, .
\ee
In particular, this means that the \terminate{critical exponent} for the single-input kernel is given by $p_0=0$. Setting $\chi=1$ and substituting this expression into \eqref{eq:vertex-scale-invariant-after-sub}, we find a linearly growing  solution for the \terminate{four-point vertex}
\be\label{eq:scale-invariant-vertex-solution}
\Ti{\FPV}{}{\ell}=\le(\ell-1\ri)\le(\frac{3A_4}{A_2^2}-1\ri) \le(\Tif{\ker}{}\ri)^2 \, .
\ee
By inspection, we identify another \terminate{critical exponent} for the scale-invariant universality class: assuming $V^{(\ell)}\sim (1/\ell)^{p_V}$, then $p_V=-1$. This exponent encodes the linear growth of the vertex under RG flow.
\index{universality class!scale-invariant}
Of particular note, the coefficient in front of \eqref{eq:scale-invariant-vertex-solution} evaluates to $\le(\frac{3 A_4}{A_2^2} -1\ri)= 2$ for $\linear$ activations in contrast to =$5$ for $\relu$ activations. Apparently the fluctuations in $\relu$ networks are significantly stronger than in deep linear networks\index{deep linear network}. More generally, we conclude that the strength of such fluctuations is \emph{not} universal.

\index{universality class!scale-invariant}
\item
Third, let's revisit semi-criticality\index{criticality!semi-criticality}  by setting $C_W = 1/A_2$, but setting the bias variance to an arbitrary positive constant, $C_b>0$. As we saw in~\S\ref{sec:scale-invariant-eft},   
in this case the kernel grows linearly towards a nontrivial fixed point\index{fixed point!nontrivial} at infinity, $\Ti{\ker}{}{\ell}\sim \ell$, i.e., $p_0=-1$. Plugging such a solution into the vertex recursion \eqref{eq:vertex-scale-invariant-after-sub}, we see that the four-point vertex grows cubicly $\Ti{\FPV}{}{\ell} \sim \ell^3$, i.e., $p_V=-3$.  However, the appropriate dimensionless quantity
-- normalizing the vertex by the square of the kernel --
still grows linearly in $\ell$, i.e., $p_V-2p_0=-1$.\footnote{
To elaborate a bit more, first please reread~footnote~\ref{foot:dimensional-analysis} in \S\ref{sec:perturbation} on \neo{dimensional analysis}. Now, if we give the preactivations a dimension $[z]=\zeta$, then we have for the kernel $[\ker]=\zeta^2$, while for the four-point vertex $[\FPV]=\zeta^4$. Thus, the ratio $V/K^2$ is dimensionless.
}
Thus, even for semi-criticality\index{criticality!semi-criticality} the universal $\ell/n$-scaling of the finite-width corrections is preserved.\index{universality class!scale-invariant}
\ei

\subsection{Fluctuations for the \texorpdfstring{$\Tif{\ker}{}=0$}{K*=0} Universality Class}\label{subsec:tanh_univ_revisit}
Let's now consider the $K^{\star}=0$ universality class. As a reminder, this class contains all smooth activation functions that satisfy $\sigma(0)=0$ and $\sigma'(0)\ne0$, %
with $\tanhA$ as the exemplar member to keep in mind. 
In \S\ref{subsec:tanh_univ}, we determined that activations in this class have a nontrivial fixed point\index{fixed point!nontrivial} at $\Tif{\ker}{}=0$ and found that the associated critical initialization hyperparameters are given by $C_b=0$ and $C_W=1/\sigma_1^2$.
For the rest of this subsection
we will focus on such networks at \terminate{criticality}. %
\index{universality class!K@$K^\star=0$}\index{initialization hyperparameters!critical}

Mirroring our approach in \S\ref{subsec:tanh_univ} to solve the kernel recursions, we can evaluate the Gaussian expectations in the vertex recursion~\eqref{eq:finite-width-reprinted-vertex} and the NLO-metric recursion~\eqref{eq:finite-width-reprinted-self-energy} by Taylor expanding the activation around $z=0$ and explicitly computing the Gaussian integrals. Keeping in mind the \terminate{criticality} condition $C_W=1/\sigma_1^2$, this gives the following expressions 
\begin{align}\label{eq:chi-parallel-agebriac}
\chi_{\parallel}(K)&=1+2a_1\ker+3a_2\ker^2+\o{\ker^3}\, ,\\
\label{eq:first-part-of-vertex-algebriac}
C_W^2\le[\bra \sigma^4(z) \ket_{\ker}-\bra \sigma^2(z) \ket_{\ker}^2\ri]&= 2 \ker^2+\le(-52a_1+60b_1\ri)K^3+\o{\ker^4}\, , \\
\label{eq:g4-algebriac-k-star}
\frac{j(K)}{8 \ker^2}&= a_1 +3 a_2 \ker+\o{\ker^2}\, .
\end{align}
Here, the expression for $\chi_{\parallel}\!\le(\ker\ri)$ is simply reprinted from~\S\ref{subsec:tanh_univ}. Similarly, to limit the amount of time you have to flip back and forth, let us also reprint the large-$\ell$ asymptotic expansion of the kernel perturbation originally given by~\eqref{eq:tanh_asymptotic}:
\begin{align}\label{k-star-equals-zero-kernel-solution-reprinted}
\Ti{\Delta \ker}{}{\ell}&=\le[\frac{1}{(-a_1)}\ri]\frac{1}{\ell}+\le[\frac{-(a_2-a_1^2)}{a_1^3}\ri]\frac{\log\!\le(\frac{\ell}{\ell_0}\ri)}{\ell^2}\, \\
&+\le[\frac{-\le(a_2-a_1^2\ri)^2}{a_1^5}\ri]\frac{\le[\log\!\le(\frac{\ell}{\ell_0}\ri)\ri]^2}{\ell^3}+\le[\frac{\le(a_2-a_1^2\ri)^2}{a_1^5}\ri]\frac{\log\!\le(\frac{\ell}{\ell_0}\ri)}{\ell^3}+\o{\frac{1}{\ell^3}}\, .\nonumber
\end{align}

\subsubsection{Four-Point Vertex}
Now, let's find a solution for the \terminate{four-point vertex}.
Substituting in \eqref{eq:chi-parallel-agebriac} and \eqref{eq:first-part-of-vertex-algebriac} into the single-input vertex recursion \eqref{eq:finite-width-reprinted-vertex} gives an algebraic equation
\begin{align}
\Ti{\FPV}{}{\ell+1}&=\Ti{\FPV}{}{\ell}\le[1+4a_1 \Ti{\Delta\ker}{}{\ell}+\le(6a_2+4a_1^2 \ri) \le(\Ti{\Delta\ker}{}{\ell}\ri)^2+\ldots\ri]\,  \\
&+2 \le(\Ti{\Delta\ker}{}{\ell}\ri)^2+\le(-52a_1+60b_1\ri) \le(\Ti{\Delta\ker}{}{\ell}\ri)^3+\ldots\, .\nonumber
\end{align}
Using our \terminate{scaling ansatz}~\eqref{eq:master-scaling-ansatz} for the large-$\ell$ asymptotic expansion
\be\label{eq:vertex-scaling-ansatz}
\Ti{\FPV}{}{\ell} =\le( \frac{1}{\ell} \ri)^{p_V} \le[\# +  \#' \frac{\log \ell}{\ell}  +   \frac{\#''}{\ell} +  \ldots \ri]\, ,
\ee
and \eqref{k-star-equals-zero-kernel-solution-reprinted} for $\Ti{\Delta \ker}{}{\ell}$ and then
matching terms, we find 
\begin{align}\label{eq:Kstar-equals-zero-vertex-solution}
\Ti{\FPV}{}{\ell}&=\le[\frac{2}{3a_1^2}\ri]\frac{1}{\ell}+\le[\frac{2(a_2-a_1^2)}{3 a_1^4}\ri]\frac{\log\!\le(\frac{\ell}{\ell_0}\ri)}{\ell^2}\, \\
&+\le[\frac{5a_2+a_1(82 a_1-90 b_1)}{3 a_1^4}\ri]\frac{1}{\ell^2}+\o{\frac{\log^2(\ell)}{\ell^3}}\, , \nonumber
\end{align}
where the constant scale $\ell_0$ is same as the one in the $\Ti{\Delta\ker}{}{\ell}$ expansion
just above, again carrying the \terminate{data dependence} of the solution.
We can also read off the \terminate{critical exponent} controlling the asymptotic falloff of the vertex for the $K^\star=0$ universality class:  
 $p_V=1$.%

\index{universality class!scale-invariant}\index{universality class!K@$K^\star=0$}
Note that the value of the exponent $p_V=1$ and the behavior of the four-point vertex $V^{(\ell)}\sim  1/\ell$ here is different from the value of the exponent $p_V=-1$ and the associated behavior $V^{(\ell)}\sim \ell$  that we found for the scale-invariant universality class. Also note that we saw this difference in the behavior of the kernel, $p_0=1$ vs.~$p_0=0$, for the $K^\star=0$ and scale-invariant classes, respectively. However, when instead considering the dimensionless quantity
\be\label{eq:k-star-equals-zero-normalized-four-point-scaling-law}
\frac{\Ti{\FPV}{}{\ell}}{n \le(\Ti{ \ker}{}{\ell} \ri)^2} \sim  \frac{1}{n}\le(\frac{1}{\ell} \ri)^{p_V-2p_0} + \dots \, ,%
\ee
we see that its scaling is consistent across both classes of activations:
\be\label{eq:vertex-scaling-law}
p_V-2p_0 = -1\, .
\ee
Thus, this \term{scaling law} holds across different universality classes. \index{universality class!transcended by scaling laws}
As the normalized quantity \eqref{eq:k-star-equals-zero-normalized-four-point-scaling-law} controls leading finite-width corrections to observables  -- this was discussed in detail in 
\S\ref{sec:fluctuations_DLN} --  such a law means that these corrections are always \emph{relevant}\index{relevant (RG flow)} under \terminate{representation group flow}.

Concretely, the normalized vertex function is given by 
\be\label{eq:k-star-equals-zero-normalized-four-point-relu-univ}
\frac{\Ti{\FPV}{}{\ell}}{n \le(\Ti{ \ker}{}{\ell} \ri)^2} = \le(\frac{3 A_4}{A_2^2} -1\ri)  \frac{\ell}{n} + \o{\frac{1}{n}} \, ,
\ee
for the scale-invariant universality class
and 
\be\label{eq:k-star-equals-zero-normalized-four-point-tanh-univ}
\frac{\Ti{\FPV}{}{\ell}}{n \le(\Ti{ \ker}{}{\ell} \ri)^2} = \le(\frac{2}{3}\ri)\frac{\ell}{n} + \o{\frac{\log \le(\ell\ri)}{n}} \, ,
\ee
for the $\ker^{\star}=0$ universality class. 
Of practical relevance, this means that $\relu$ networks and $\tanhA$ networks of the same depth and width will have a mostly similar sensitivity to such corrections. However, the $\o{1}$ coefficient of this quantity \emph{does} depend on the particular activation function: =$5$ for $\relu$ and =$2/3$ for $\tanhA$. In Appendix~\ref{app:mi-stuff}, we'll analyze this a bit more using tools from \neo{information theory} and see how it can lead to a preferred aspect ratio, $L/n$, that is different for specific choices of activation functions.
\index{universality class!scale-invariant}\index{universality class!K@$K^\star=0$}

\subsubsection{NLO metric, bare}
\index{metric!next-to-leading-order correction}\index{subleading corrections}\index{metric!next-to-leading-order correction}
Next, let's solve the NLO-metric recursion \eqref{eq:finite-width-reprinted-self-energy}. 
Substituting in \eqref{eq:chi-parallel-agebriac} for $\chi_{\parallel}(K)$ and \eqref{eq:g4-algebriac-k-star} for $j(K)$, we get
\be\label{eq:nlo-metric-recursion-bare}
\se{}{\ell+1}=\se{}{\ell}\le[1+2a_1 \Ti{\Delta\ker}{}{\ell}+\ldots\ri] +\Ti{\FPV}{}{\ell}\le[a_1+3a_2  \Ti{\Delta\ker}{}{\ell}+\ldots\ri]\, .
\ee
As should now be familiar, let's assume a large-$\ell$ \terminate{scaling ansatz}
\be\label{eq:nlo-metric-scaling-ansatz}
\se{}{\ell}= \# \le( \frac{1}{\ell} \ri)^{p_1}+ \ldots \, ,
\ee
with $p_1$ as the associated \terminate{critical exponent}.
Bootstrapping \eqref{eq:nlo-metric-recursion-bare} by substituting in our previous solutions -- \eqref{k-star-equals-zero-kernel-solution-reprinted} for $\Delta\ker^{(\ell)}$  and \eqref{eq:Kstar-equals-zero-vertex-solution} for $\Ti{\FPV}{}{\ell}$ -- we then insert our ansatz  for $\se{}{\ell}$ \eqref{eq:nlo-metric-scaling-ansatz} and match terms to find 
\be\label{eq:k-star-equal-zero-self-energy}
\se{}{\ell}= -\le[\frac{1}{3(-a_1)}\ri]+ \o{\frac{\log \le(\ell\ri)}{\ell}} \, .
\ee
This solution required us to set $p_1 = 0$ and gave a constant-in-$\ell$ leading contribution.
Combining this with the kernel, we see that the finite-width-corrected two-point correlator 
\be\label{eq:two-point-with-leading-finite-width-reprinted}
\E{z^{(\ell)}_i z^{(\ell)}_j}=\delta_{ij} \le[\ker^{(\ell)} + \frac{1}{n}\se{}{\ell}+\o{1/n^2}\ri] \, ,
\ee
is given by
\be\label{eq:k-star-full-two-point-with-self-energy}
\ker^{(\ell)} + \frac{1}{n}\se{}{\ell} = \le[\frac{1}{(-a_1)}\ri]\le(\frac{1}{\ell} -\frac{1}{3n}\ri)+ \ldots\, . %
\ee
This result is to be contrasted with the scale-invariant universality class, where the NLO metric\index{metric!next-to-leading-order correction} vanished identically. 

\index{critical exponent}\index{subleading corrections}\index{metric!next-to-leading-order correction}
For the NLO metric\index{metric!next-to-leading-order correction}, the appropriate dimensionless quantity to consider is the ratio between the correction term and the infinite-width term in the two-point correlator \eqref{eq:two-point-with-leading-finite-width-reprinted}
\be\label{eq:nlo-dimensionless-ratio}
\frac{1}{n}\frac{\se{}{\ell}}{\ker^{(\ell)}} \sim  \frac{1}{n}\le(\frac{1}{\ell}\ri)^{p_1-p_0}+ \dots \, ,
\ee
with the exponent $p_1 - p_0$ controlling the relative importance of this NLO correction. In this case we see that $p_1-p_0=-1$, meaning that the above ratio scales with the depth-to-width ratio $\ell/n$.
This again illustrates the perturbative cutoff\index{cutoff, effective theory} of our \terminate{effective theory}, $\ell \lesssim n$.
However, in this particular case such a scaling turns out to be an artifact of not properly tuning the initialization hyperparameters $C_W$ at finite width, as we will see next.
\index{universality class!scale-invariant}

\subsubsection{NLO metric, renormalized}
\index{renormalization group flow}\index{metric!next-to-leading-order correction}\index{$1/n$ expansion}\index{initialization hyperparameters!critical!at finite width}\index{universality class!K@$K^\star=0$}\index{subleading corrections}\index{metric!next-to-leading-order correction}

In \S\ref{subsec:tanh_univ}, we learned how to find the critical initialization hyperparameters for the $K^\star=0$ universality class, fixing the hyperparameters $C_b$ and $C_W$ using the infinite-width recursions for the kernel components. However, in \S\ref{sec:loop-correction} we explained that all of the observables computed in a large-$n$ expansion receive an infinite series of subleading corrections in $1/n$.  This suggests that we should have allowed further fine-tuning of the initialization hyperparameters at \terminate{criticality} by considering large-$n$ expansions
\begin{align}
\label{eq:Cb-expansion}
\Cb{\ell} &= \cbn{\ell}{0} + \frac{\cbn{\ell}{1}}{n_{\ell-1}} +\frac{\cbn{\ell}{2}}{n_{\ell-1}^2} + \dots \, , \\
\label{eq:CW-expansion}
\CW{\ell} &= \cWn{\ell}{0} + \frac{\cWn{\ell}{1}}{n_{\ell-1}} +\frac{\cWn{\ell}{2}}{n_{\ell-1}^2} + \dots \, ,
\end{align}
allowing us to adjust such hyperparameters order by order in $1/n$. Such an expansion could potentially give additional criticality conditions at each order in perturbation theory.

\index{subleading corrections}\index{metric!next-to-leading-order correction}
Considering the finite-width recursions \eqref{eq:finite-width-reprinted-vertex} and \eqref{eq:finite-width-reprinted-self-energy}, we see that such subleading tunings will not affect the leading order result for observables that depend on the \terminate{four-point vertex}, since the leading contributions to such observables are already at $\o{1/n}$. However, these tunings do affect the solution for the NLO metric\index{metric!next-to-leading-order correction}, because the NLO metric is itself subleading.

Concretely, there is an additional contribution to the NLO-metric recursion \eqref{eq:finite-width-reprinted-self-energy} coming from inserting the expansions \eqref{eq:Cb-expansion} and \eqref{eq:CW-expansion} into the kernel recursion \eqref{eq:finite-width-reprinted-kernel}. The terms proportional to $\cbn{\ell}{1}$ or $\cWn{\ell}{1}$ are now subleading and thus contribute to the NLO metric\index{metric!next-to-leading-order correction} recursion: 
\be\label{eq:nlo-metric-recursion-renormalized}
\se{}{\ell+1}= \le[\cbn{\ell}{1} + \cWn{\ell}{1} g\!\le(K^{(\ell)}\ri) \ri]+ \chi_{\parallel}^{(\ell)}\se{}{\ell}+ \frac{1}{8}\, j\!\le(\Ti{\ker}{}{\ell}\ri) \frac{\Ti{\FPV}{}{\ell}}{\le(\Ti{\ker}{}{\ell}\ri)^2}   \, .
\ee 
With this new ``renormalized'' perspective, we see that the analysis we did in the ``bare'' subsubsection before was just a particular choice of subleading corrections, $\cbn{\ell}{1}=\cWn{\ell}{1}=0$. More generally, we really do have additional knobs to turn at this subleading order.\index{subleading corrections}\index{metric!next-to-leading-order correction}

\index{subleading corrections}\index{metric!next-to-leading-order correction}
Substituting in \eqref{eq:chi-parallel-agebriac} for $\chi_{\parallel}(K)$, \eqref{eq:g4-algebriac-k-star} for $j(K)$, and \eqref{eq:g0} for $g(K)$, we find an algebraic equation
\begin{align}
\se{}{\ell+1}= &\cbn{\ell}{1} + \cWn{\ell}{1} \sigma_1^2\le[\ker^{(\ell)}+a_1 \le(\ker^{(\ell)}\ri)^2+\ldots\ri] \\
&+\se{}{\ell}\le[1+2a_1 \Ti{\ker}{}{\ell}+\ldots\ri] +\Ti{\FPV}{}{\ell}\le[a_1+3a_2  \Ti{\ker}{}{\ell}+\ldots\ri] \, . \notag 
\end{align}
Plugging in the solution for the kernel \eqref{k-star-equals-zero-kernel-solution-reprinted} and vertex \eqref{eq:Kstar-equals-zero-vertex-solution} -- making sure to include the subleading-in-$\ell$ terms in both -- inserting our large-$\ell$ \terminate{scaling ansatz} for $\se{}{\ell}$ \eqref{eq:nlo-metric-scaling-ansatz} and matching terms, we find that the tunings
\be\label{eq:fine-critical-tuning}
\cbn{\ell}{1} = 0\, , \qquad \cWn{\ell}{1} = \frac{2}{3}\cWn{\ell}{0} = \frac{2}{3\sigma_1^2} \, ,
\ee
result in an asymptotically suppressed solution for the NLO metric\index{metric!next-to-leading-order correction}
\be\label{eq:nlo-metric-subleading}
\se{}{\ell}= \frac{2}{3}\le[ \frac{3 a_2 - a_1^2 }{ (-a_1)^3} \ri]\frac{1}{\ell} + \o{ \frac{\log(\ell)}{\ell^2}} \, ,
\ee 
with a \terminate{critical exponent} $p_1=1$.
Specifically, the tuning of $\cbn{\ell}{1}$ was required to suppress a linear growing $\sim \ell$ contribution, while the tuning of $\cWn{\ell}{1}$ cancels the constant $\o{1}$ piece we found before in \eqref{eq:k-star-equal-zero-self-energy}. 
\bi
\item In a sense, we got lucky before in our bare analysis: redoing this analysis without a $\cbn{\ell}{1}=0$ tuning, 
the dimensionless ratio \eqref{eq:nlo-dimensionless-ratio} grows quadratically with depth and
implies that the NLO metric\index{metric!next-to-leading-order correction} dominates the kernel at $\ell \sim \sqrt{n}$.
The fact that this subleading correction becomes parametrically large before reaching the $\ell/n$ perturbative cutoff\index{cutoff, effective theory} of the \terminate{effective theory} really means that it's growing exponentially; $\cbn{\ell}{1}\neq0$ eventually spoils \terminate{criticality}.
\item In another sense, we got unlucky before: without the $\cWn{\ell}{1} = \frac{2}{3}\cWn{\ell}{0}$ tuning, the NLO metric\index{metric!next-to-leading-order correction} is a leading $\ell/n$ correction. We see now that when properly handled, $p_1 - p_0 = 0$ and the dimensionless ratio \eqref{eq:nlo-dimensionless-ratio} is $\o{1}$ in depth at leading order. Such a correction is said to be \emph{marginal} under the RG flow. \index{representation group flow}\index{marginal (RG flow)}
This means that, while we'll always have to take into account the \emph{relevant}\index{relevant (RG flow)} \terminate{four-point vertex} corrections, we should be able to neglect NLO metric\index{metric!next-to-leading-order correction} corrections as long as we respect the finite-width tunings~\eqref{eq:fine-critical-tuning}.\index{subleading corrections}\index{metric!next-to-leading-order correction}
\ei

\index{initialization hyperparameters!critical!at finite width}\index{ensemble}\index{subleading corrections}\index{metric!next-to-leading-order correction}
Finally, the necessity of including such perturbative corrections to the critical initialization hyperparameters 
gives an alternate perspective on what can go wrong in practice when the network depth $L$ approaches the network width $n$. Even for ensembles of such networks, the averaged quantities will require finer and finer tunings -- e.g.~\eqref{eq:Cb-expansion} and \eqref{eq:CW-expansion} -- in order for the \terminate{effective theory} describing the ensemble to reach \terminate{criticality}. For any reasonable value of $n$, such corrections will quickly become finer than the floating-point precision limit used to represent the hyperparameters. Thus, in practice it becomes essentially impossible to tune
such large square networks to criticality.\footnote{Note that this is an entirely different problem than the chaotic behavior at large depth that we described in \S\ref{sec:solution_DLN} for deep linear networks\index{deep linear network}. For the scale-invariant universality class, the NLO metric correction vanishes and therefore $\cWn{\ell}{1}=0$.
\index{universality class!scale-invariant}
}

\section{Finite-Angle Analysis for the Scale-Invariant Universality Class}\label{sec:finite_angle}
In this section, we'll confront an important subtlety for activation functions in the scale-invariant universality class\index{universality class!scale-invariant}.

Recall that activation functions in this class take the form
\be\label{eq:scale-invariant-one-kink-reprint-finite-angle}
\sigma(z) = 
    \begin{cases}
   a_+ z \, , & z \ge 0  \, , \\
    a_- z \, , & z < 0 \, ,
    \end{cases}
\ee
and generally have a kink at the origin $z=0$ (except for the \emph{degenerate} member, the $\linear$ activation function, which has $a_{+}=a_{-}$). In footnote~\ref{foot:kink1} we first mentioned the existence of a subtlety after giving our $\delta$ expansions for the kernel~\eqref{eq:kernel-expand-1}--\eqref{eq:kernel-expand-3}, lightly questioning the validity of our expansions for non-smooth $\sigma(z)$.
In footnote~\ref{foot:kink2}, we then described the main consequence of this subtlety. In particular, we claimed that for nonlinear scale-invariant activation functions the constant value -- as a function of layer -- of the perpendicular perturbation $\Ti{ \delta\delta\ker}{[2]}{\ell}$ at \terminate{criticality} is an artifact of the perturbative $\delta$ expansion. To understand this claim properly, we'll need to work out the full nonperturbative kernel recursion for activation functions in this class. This in turn will let us see the aforementioned correction to the asymptotic large-$\ell$ behavior of the kernel component $\Ti{ \delta\delta\ker}{[2]}{\ell}$.

For this analysis, it will be sufficient to focus on two
inputs $\x{i}{\pm}$ of the same norm. In our previous setup, we assumed that both inputs were nearby such that their difference $\delta x_i \equiv (\x{i}{+} - \x{i}{-})$ was perturbatively small, $\delta x_i \ll 1$; here, we will make no assumptions at all about their difference.
Given the symmetries of the network evolution, the individual norms of the two preactivations corresponding to these inputs will also be equal:
\be\label{eq:our-points-ghave-equal-norms}
\Kdi{\ell}\equiv\E{\frac{1}{n_{\ell}}\sum_{i=1}^{n_{\ell}}\le(\z{i}{+}{\ell}\ri)^2}=\E{\frac{1}{n_{\ell}}\sum_{i=1}^{n_{\ell}}\le(\z{i}{-}{\ell}\ri)^2}\, .
\ee
Geometrically this means that our preactivations live together on an $n_\ell$-dimensional sphere with radius $\sqrt{n_{\ell}\Kdi{\ell}}$, and algebraically this means that the parallel component vanishes $K_{[1]}^{(\ell)}=0$, cf.~\eqref{eq:K1-decomposition}.
Going forward, we will call $\Kdi{\ell}$ the \textbf{diagonal kernel}\index{kernel!kernel matrix!diagonal|textbf}.\footnote{
    Perturbatively, the \emph{diagonal kernel} $\Kdi{\ell}$ is equal to the \emph{midpoint kernel}\index{kernel!midpoint} $K_{\M\M}^{(\ell)}$ -- the kernel for the \terminate{midpoint input} $\x{i}{\M} \equiv (\x{i}{+} + \x{i}{-})/2$ -- at leading order in the $\delta$ expansion, cf.~\eqref{eq:kernel-expand-1}--\eqref{eq:kernel-expand-3}. %
    Nonperturbatively, these two kernels are very different. To see this most vividly, consider two antipodal inputs $\x{i}{+}=-\x{i}{-}$. Then, the midpoint input is the zero vector $\x{i}{\M}=0$, and the midpoint kernel in the first layer is given by $K_{\M\M}^{(1)}=\Cb{1}$. In contrast, the diagonal kernel is given by either of $\Kdi{1}=\Cb{1}+(\CW{1}/n_0)\sum_{i=1}^{n_0}\x{i}{\pm}^2$. %
} 

The remaining dynamical variable is the polar angle between the preactivations. Therefore, we can decompose the two-input kernel matrix with the following parameterization:\index{kernel!kernel matrix!polar angle parameterization}
\be\label{eq:angle-parametrization}
\Ti{\ker}{\alpha_1\alpha_2}{\ell}=\begin{pmatrix}
\Ti{\ker}{++}{\ell} & \Ti{\ker}{+-}{\ell} \\
\Ti{\ker}{-+}{\ell}  & \Ti{\ker}{--}{\ell} 
\end{pmatrix}=\Kdi{\ell}\begin{pmatrix}
1 & \cos\!\le(\psi^{(\ell)}\ri)\\
\cos\!\le(\psi^{(\ell)}\ri)  & 1 
\end{pmatrix} \, , \qquad \psi^{(\ell)}\in\le[0,\pi\ri]\, .
\ee
The polar angle $\psi^{(\ell)}$ ranges from $0$ -- where the preactivations are coincident as $z_{i;+}=z_{i;-}$, making the kernel matrix degenerate -- to $\pi$ -- where they're anti-correlated as $z_{i;+}=-z_{i;-}$.
So far all we've done is fixed the norm of our two inputs to be equal and decomposed the kernel into a particular choice of coordinates;  
such a choice and parameterization can be applied to the analysis of any activation function. We'll now specialize to scale-invariant activation functions for which class it's possible to derive a nonperturbative recursion for the polar angle.

\subsubsection{RG flow of the polar angle}
The diagonal kernel follows the by-now familiar recursion for the single-input kernel~\eqref{eq:kernel-diagonal}
\be
\Kdi{\ell+1}=C_b+C_W\, g\!\le(\Kdi{\ell}\ri)=C_b+A_2C_W \Kdi{\ell}\, ,
\ee
where on the right-hand side we plugged in the explicit details for the scale-invariant universality class\index{universality class!scale-invariant}~\eqref{eq:recursion-scale-invariant-spin-0} and recalled $A_2 \equiv \le(a_{+}^2+a_{-}^2\ri)/2$.
This part of the analysis carries over from \S\ref{sec:scale-invariant-eft}. We recall here that we can readily solve the recursion for any choice of \terminate{initialization hyperparameters}, and in particular \terminate{criticality} is attained by setting $C_b=0$ and $A_2 C_W=1$, where the diagonal kernel stays exactly constant:  $\Kdi{\ell} = \Kdi{1} \equiv \Kdif$.

With the evolution of the magnitude determined, we now need to find a recursion for the polar angle $\psi^{(\ell)}$. Plugging our new decomposition~\eqref{eq:angle-parametrization} into the full kernel recursion~\eqref{eq:kernel-recursion-reminder}, the off-diagonal component of the recursion becomes
\be\label{eq:off-diagonal-recursion}
\Kdi{\ell+1}\cos\!\le(\psi^{(\ell+1)}\ri)=C_b+C_W \bra\sigma\!\le(z_{+}\ri)\sigma\!\le(z_{-}\ri)\ket_{\ker^{(\ell)}}\, .
\ee
In this parameterization, the \terminate{Gaussian expectation} reads 
\be\label{eq:painful-integral}
 \bra\sigma\!\le(z_{+}\ri)\sigma\!\le(z_{-}\ri)\ket_{\ker^{(\ell)}}\equiv \frac{\int dz_{+} dz_{-}\ \sigma\!\le(z_{+}\ri)\sigma\!\le(z_{-}\ri) e^{-\frac{1}{2}\sum_{\alpha_1,\alpha_2=\pm}\ker^{\alpha_1\alpha_2}_{(\ell)}z_{\alpha_1}z_{\alpha_2}}}{2\pi\Kdi{\ell}\sin\!\le(\psi^{(\ell)}\ri)}\, ,
 \ee
 where the denominator comes from evaluating the determinant $\sqrt{\dete{2\pi K^{(\ell)}}}$.
 To make further progress, we need to evaluate this painful integral.

Before working out the general case, let's focus on the $\relu$.
Setting $a_{+}=1$ and $a_{-}=0$, we see that the argument of the \terminate{Gaussian expectation} is given by $\sigma\!\le(z_{+}\ri)\sigma\!\le(z_{-}\ri)=z_{+}z_{-}$ when $z_{+}>0$ and $z_{-}>0$ and vanishes otherwise. This means that the \terminate{Gaussian expectation}~\eqref{eq:painful-integral} is concentrated entirely in the first quadrant. In addition, noting that the integrand is invariant under parity $\le(z_{+},z_{-}\ri)\to\le(-z_{+},-z_{-}\ri)$, we can niftily substitute the integral over the first quadrant for half the integral over the first and third quadrants. This lets us rewrite the above Gaussian expectation as
\begin{align}\label{eq:nifty-step} 
\bra\sigma\!\le(z_{+}\ri)\sigma\!\le(z_{-}\ri)\ket_{\ker^{(\ell)}}
=&\frac{\frac{1}{2}\int dz_{+} dz_{-}\big\vert_{z_{+}z_{-}>0}\ z_{+}z_{-}\, e^{-\frac{1}{2}\sum_{\alpha_1,\alpha_2=\pm}\ker^{\alpha_1\alpha_2}_{(\ell)}z_{\alpha_1}z_{\alpha_2}}}{2\pi\Kdi{\ell}\sin\!\le(\psi^{(\ell)}\ri)}\, .
\end{align}
The above actually turns out to be the only nifty step of the derivation; everything else is just a \terminate{Herculean sequence} of coordinate changes.

There are three coordinate changes in said sequence:
\begin{align}\label{eq:Herculean-0}
z_{\pm}=&\frac{u\pm w}{\sqrt{2}} \\ 
=&\sqrt{\frac{\Kdi{\ell}\le[1+\cos\!\le(\psi^{(\ell)}\ri)\ri]}{2}}\ x \pm \sqrt{\frac{\Kdi{\ell}\le[1-\cos\!\le(\psi^{(\ell)}\ri)\ri]}{2}}\ y\, \notag \\
=&\sqrt{\frac{\Kdi{\ell}\le[1+\cos\!\le(\psi^{(\ell)}\ri)\ri]}{2}}\ r\cos(\phi) \pm \sqrt{\frac{\Kdi{\ell}\le[1-\cos\!\le(\psi^{(\ell)}\ri)\ri]}{2}}\ r\sin(\phi)\, .\notag
\end{align}
The first one diagonalizes\index{diagonalization} the kernel so that the distribution factorizes $p(z_+,z_-) = p(u) p(w)$, the second one normalizes the coordinates with the kernel's eigenvalues, and the last one exchanges Cartesian coordinates for polar coordinates.\footnote{Unlike the perturbative calculations in \eqref{eq:kernel-eigenvectors} and \eqref{eq:kernel-eigs}, the \terminate{diagonalization} and normalization here are nonperturbatively exact.
To reflect more on this, while we can always change coordinates as \eqref{eq:Herculean-0}, we used the details of the $\relu$ in going from \eqref{eq:painful-integral} to \eqref{eq:nifty-step}, establishing both the restricted domain of integration and the simplified form of the integrand,  $\sigma\!\le(z_{+}\ri)\sigma\!\le(z_{-}\ri) \to z_{+}z_{-}$, within that domain. For a general activation function, the resulting integral in the new coordinates \eqref{eq:Herculean-0} would still be difficult to evaluate, and we would have to resort to a perturbative expansion in $\psi^{(\ell)}$, ultimately analogous to the $\delta$ expansion, in order to make progress.}
Accordingly, this lets us rewrite the sum in the exponential in~\eqref{eq:nifty-step} as
\begin{align}\label{eq:Herculean-1}
\sum_{\alpha_1,\alpha_2=\pm}\ker^{\alpha_1\alpha_2}_{(\ell)}z_{\alpha_1}z_{\alpha_2}= \frac{u^2}{\Kdi{\ell}\le[1+\cos\!\le(\psi^{(\ell)}\ri)\ri]}+\frac{w^2}{\Kdi{\ell}\le[1-\cos\!\le(\psi^{(\ell)}\ri)\ri]}=x^2+y^2=r^2\, ,
\end{align}
while the product in the integrand becomes
\be\label{eq:Herculean-2}
z_{+}z_{-}=\frac{\Kdi{\ell} r^2}{2}\le[\cos(2\phi)+\cos\!\le(\psi^{(\ell)}\ri)\ri]\, ,
\ee
and the integral measure transforms as
\be\label{eq:Herculean-3}
dz_{+}dz_{-}=\Kdi{\ell}\sin\!\le(\psi^{(\ell)}\ri) r\, dr \,d\phi\, .
\ee
Substituting \eqref{eq:Herculean-1}--\eqref{eq:Herculean-3} back into the \terminate{Gaussian expectation}~\eqref{eq:nifty-step}, we get
\begin{align}\label{eq:relu-non-perturbative-mid}
\bra\sigma\!\le(z_{+}\ri)\sigma\!\le(z_{-}\ri)\ket_{\ker^{(\ell)}}=&\frac{\Kdi{\ell}}{8\pi}\!\le[\int_{0}^{\infty}\!\!\!dr\ r^3 e^{-\frac{r^2}{2}}\ri]\! \int_{0}^{2\pi} d\phi \Big\vert_{\cos(2\phi)+\cos\le(\psi^{(\ell)}\ri)>0}\!\!\le[\cos(2\phi)+\cos\!\le(\psi^{(\ell)}\ri)\ri]\, .
\end{align}

The rest is now relatively straightforward. The radial integral can be evaluated
by another change of the coordinate $s=r^2/2$:
\be\label{eq:relu-radial-integral}
\int_{0}^{\infty}dr\ r^3 e^{-\frac{r^2}{2}}=\int_{0}^{\infty}ds\ 2s\, e^{-s}= \Big[ -2e^{-s}- 2s\, e^{-s}\Big]\Big\vert_{0}^{\infty}=2\, .
\ee
For the angle integral, note that any function of $\cos(2\phi)$ gives the same contribution from the four intervals $\widetilde{\phi}\equiv2\phi\in[0,\pi], [\pi, 2\pi], [2\pi, 3\pi], [3\pi, 4\pi]$. 
Further, within that first interval the constraint $\cos\!\le(\widetilde{\phi}\ri) >- \cos\!\le(\psi^{(\ell)}\ri)$  can  be simply expressed as $\widetilde{\phi} <\pi -\psi^{(\ell)}$. Together, this lets us write 
\begin{align}\label{eq:relu-angle-integral}
&\int_{0}^{2\pi} d\phi \Big\vert_{\cos(2\phi)+\cos\le(\psi^{(\ell)}\ri)>0}\le[\cos(2\phi)+\cos\!\le(\psi^{(\ell)}\ri)\ri]\, \\
=&4\int_{0}^{\pi} \frac{d\widetilde{\phi}}{2} \bigg\vert_{\cos(\widetilde{\phi})+\cos\le(\psi^{(\ell)}\ri)>0}\le[\cos(\widetilde{\phi})+\cos\!\le(\psi^{(\ell)}\ri)\ri]\, \notag\\
=&2\int_{0}^{\pi-\psi^{(\ell)}}\!\!\!\! d\widetilde{\phi}\le[\cos\!\le(\widetilde{\phi}\ri)+\cos\!\le(\psi^{(\ell)}\ri)\ri]=2\sin\!\le(\psi^{(\ell)}\ri)+2\le(\pi-\psi^{(\ell)}\ri)\cos\!\le(\psi^{(\ell)}\ri)\, .\notag
\end{align}
Inserting \eqref{eq:relu-radial-integral} and \eqref{eq:relu-angle-integral} into \eqref{eq:relu-non-perturbative-mid}, we finally arrive at an expression for the Gaussian expectation of $\relu$ activations:
\be
\bra\sigma\!\le(z_{+}\ri)\sigma\!\le(z_{-}\ri)\ket_{\ker^{(\ell)}}=\frac{\Kdi{\ell}}{2\pi}\le[\sin\!\le(\psi^{(\ell)}\ri)+\le(\pi-\psi^{(\ell)}\ri)\cos\!\le(\psi^{(\ell)}\ri)\ri]\, .
\ee

Now, let's work out the painful integral~\eqref{eq:painful-integral} for an arbitrary scale-invariant activation function~\eqref{eq:scale-invariant-one-kink-reprint-finite-angle}.
In general, there are contributions from the first quadrant proportional to $a_+^2$ and similar contributions from the third quadrant proportional to $a_{-}^2$, in both cases with the constraint $z_{+}z_{-}>0$ after our nifty trick.
Then, there are also contributions from the second and fourth quadrants, both proportional to $a_{+}a_{-}$ and with the constraint $z_{+}z_{-}<0$. Following a very similar sequence of steps as we did before for the $\relu$, we can evaluate the Gaussian expectation~\eqref{eq:painful-integral} as
\begin{align}\label{eq:evaluated-gaussian-expectation}
\bra\sigma\!\le(z_{+}\ri)\sigma\!\le(z_{-}\ri)\ket_{\ker^{(\ell)}}=&\ \frac{\Kdi{\ell}}{2\pi}(a_{+}^2+a_{-}^2)\int_{0}^{\pi-\psi^{(\ell)} } d\widetilde{\phi} \le[\cos\!\le(\widetilde{\phi}\ri)+\cos\!\le(\psi^{(\ell)}\ri)\ri]\, \\
&+\frac{\Kdi{\ell}}{2\pi}(2a_{+}a_{-})\int_{\pi-\psi^{(\ell)} }^{\pi} d\widetilde{\phi} \le[\cos\!\le(\widetilde{\phi}\ri)+\cos\!\le(\psi^{(\ell)}\ri)\ri]\, \notag\\
=&\ \frac{\Kdi{\ell}}{2\pi}\le(a_{+}-a_{-}\ri)^2\le[\sin\!\le(\psi^{(\ell)}\ri)-\psi^{(\ell)}\cos\!\le(\psi^{(\ell)}\ri)\ri]\, \notag\\
&+\le(\frac{a_{+}^2+a_{-}^2}{2}\ri)\Kdi{\ell}\cos\!\le(\psi^{(\ell)}\ri) \, .\notag
\end{align}
The full nonperturbative recursion for the off-diagonal part of the kernel~\eqref{eq:off-diagonal-recursion} thus evaluates to
\begin{align}\label{eq:off-diagonal-recursion-integrated}
&\Kdi{\ell+1}\cos\!\le(\psi^{(\ell+1)}\ri)\\
=&C_b+C_W\! \le\{ \frac{\Kdi{\ell}}{2\pi}\le(a_{+}-a_{-}\ri)^2\!\le[\sin\!\le(\psi^{(\ell)}\ri)-\psi^{(\ell)}\cos\!\le(\psi^{(\ell)}\ri)\ri]\! +\!\le(\frac{a_{+}^2+a_{-}^2}{2}\ri)\Kdi{\ell}\cos\!\le(\psi^{(\ell)}\ri) \ri\} \, . \notag
\end{align}
One thing we notice here is that even though we evaluated the \terminate{Gaussian expectation}, we'll still have to deal with the fact the recursion is highly nonlinear in $\psi^{(\ell+1)}$.

While you're here and we have your attention, let's record the result for one additional nonperturbative \terminate{Gaussian expectation} for the scale-invariant universality class\index{universality class!scale-invariant}:  $\bra\sigma'\!\le(z_{+}\ri)\sigma'\!\le(z_{-}\ri)\ket_{\ker^{(\ell)}}$. The integral here is much simpler to evaluate than the undifferentiated one above since in each quadrant the argument of the expectation, $\sigma'\!\le(z_{+}\ri)\sigma'\!\le(z_{-}\ri)$, is constant. Following otherwise the same set of steps as above, in this case we find
\begin{align}
\bra\sigma'\!\le(z_{+}\ri)\sigma'\!\le(z_{-}\ri)\ket_{\ker^{(\ell)}}=&\frac{(a_{+}^2+a_{-}^2)}{4\pi}\le[\int_{0}^{\infty}dr\ r e^{-\frac{r^2}{2}}\ri]\int_{0}^{2\pi} d\phi \Big\vert_{\cos(2\phi)+\cos\le(\psi^{(\ell)}\ri)>0} \, \label{eq:johnny-b-goode}\\
&+\frac{2a_{+}a_{-}}{4\pi}\le[\int_{0}^{\infty}dr\ r e^{-\frac{r^2}{2}}\ri]\int_{0}^{2\pi} d\phi \Big\vert_{\cos(2\phi)+\cos\le(\psi^{(\ell)}\ri)<0}\, \notag\\
=&\le(\frac{a_{+}^2+a_{-}^2}{2}\ri)-\frac{\psi^{(\ell)}}{2\pi}\le(a_{+}-a_{-}\ri)^2\, .\notag
\end{align}
We guess you guys aren't ready for that yet. But your future-selves are gonna love it.\footnote{This result will turn out to be really useful in \S\ref{sec:generalization-at-infinity} when we investigate
generalization error for the scale-invariant universality class\index{universality class!scale-invariant} at infinite width.
}

\subsubsection{Criticality analysis of the polar angle}
Having evaluated the recursion, let's now tune to \terminate{criticality} and work out the correct large-$\ell$ asymptotic behavior of the polar angle $\psi^{(\ell)}$. Working at scale-invariant \terminate{criticality}, with $C_b=0$ and $A_2 C_W=1$, and where the diagonal kernel\index{kernel!kernel matrix!diagonal} is constant as $\Kdi{\ell} =\Kdif$,
the off-diagonal recursion~\eqref{eq:off-diagonal-recursion-integrated} simplifies to a decoupled recursion for the polar angle,
\begin{align}\label{eq:off-diagonal-recursion-at-criticality}
\cos\!\le(\psi^{(\ell+1)}\ri)=&\cos\!\le(\psi^{(\ell)}\ri)+\rho\le[\sin\!\le(\psi^{(\ell)}\ri)-\psi^{(\ell)}\cos\!\le(\psi^{(\ell)}\ri)\ri]\, .
\end{align}
Here, it was convenient to define a new constant,
\be\label{eq:rho-definition}
\rho\equiv \frac{1}{\pi} \frac{\le(a_{+}-a_{-}\ri)^2}{\le(a_{+}^2+a_{-}^2\ri)}\, ,
\ee
that encapsulates all of the details of the specific scale-invariant activation function. Roughly, $\rho$ is a dimensionless measure of the kinkiness of the activation function at the origin, equal to zero for the $\linear$ activation function and $1/\pi$ for the $\relu$. We see right away that the polar angle is exactly preserved for, and only for, $\rho =0$.
In particular, the preservation of the full two-input kernel matrix that we saw for the $\linear$ activation function in~\S\ref{sec:criticality_DLN} doesn't extend to any other
member of the universality class.

In order to determine the large-$\ell$ behavior of the polar angle $\psi^{(\ell)}$, we need a way to analyze the recursion \eqref{eq:off-diagonal-recursion-at-criticality}. As we've been emphasizing, our main tool for analyzing such a nonlinear recursion is to find a fixed point and then linearize around it.\footnote{
    Since we already nonperturbatively evaluated the \terminate{Gaussian expectation}~\eqref{eq:evaluated-gaussian-expectation} and fully took into account the lack of smoothness of the activation function -- with the constant $\rho$ \eqref{eq:rho-definition} characterizing its kinkiness -- at this point it's completely safe to employ a perturbative expansion.
} By inspection of the recursion, it's clear that $\psi^{\star}=0$ is a fixed point.
Thus, we should focus in on the small-angle regime:  $\psi^{(\ell)}\ll 1$.

Taylor expanding the trigonometric functions in the recursion~\eqref{eq:off-diagonal-recursion-at-criticality} around a vanishing polar angle, the linearized recursion becomes
\begin{align}\label{eq:off-diagonal-recursion-at-criticality-Taylor}
\psi^{(\ell+1)}=\psi^{(\ell)}\sqrt{1-\frac{2\rho}{3}\psi^{(\ell)}+\o{\psi^2}}=\psi^{(\ell)}-\frac{\rho}{3}\le(\psi^{(\ell)}\ri)^2+\o{\psi^3}\, .
\end{align}
To solve this recursion, we can use our \terminate{scaling ansatz}~\eqref{eq:master-scaling-ansatz}, which here reads
\begin{align}
\psi^{(\ell)} &= \le( \frac{1}{\ell} \ri)^{p_{\psi}} \le[c_{0,0}+\o{\frac{\log \ell}{\ell}}\ri]\, ,
\end{align} 
with the \terminate{critical exponent} $p_{\psi}$ governing the decay of the polar angle. Plugging this ansatz into our recursion~\eqref{eq:off-diagonal-recursion-at-criticality-Taylor} and matching the terms on both sides of the equation, we find a solution:
\be\label{eq:scale-invariant-polar-angle}
\psi^{(\ell)} = \le(\frac{3}{\rho}\ri) \frac{1}{\ell} +\o{\frac{\log \ell}{\ell^2}}\, .
\ee
From this we can read off the critical exponent, $p_{\psi}=1$, which is universal excepting the degenerate $\linear$ limit of $\rho=0$, for which we instead have $p_{\psi}=0$.

In order to recast this result in the language of the rest of this chapter, let's project the two-input kernel \eqref{eq:angle-parametrization} into the $\gamma^{[a]}_{\alpha\beta}$ representation using \eqref{eq:trace-projection} and then insert \eqref{eq:scale-invariant-polar-angle}:
\begin{align}\label{eq:finite-angle-spin-0-solution}
K_{[0]}^{(\ell)}&=\Kdi{\ell}\le[\frac{1+\cos\!\le(\psi^{(\ell)}\ri)}{2}\ri]=\Kdif+\o{\frac{1}{\ell^2}}\, ,\\
K_{[2]}^{(\ell)}&=\Kdi{\ell}\le[\frac{1-\cos\!\le(\psi^{(\ell)}\ri)}{2}\ri]=\Kdif\le(\frac{9}{4\rho^2}\ri)\frac{1}{\ell^2}+\o{\frac{\log \ell}{\ell^3}}\, .\label{eq:finite-angle-spin-2-solution}
\end{align}
These solutions form the basis of what we claimed earlier in footnote \ref{foot:kink2}.
In particular, the perpendicular perturbation $K_{[2]}^{(\ell)}$ crosses over from being nearly constant for small depth $\ell \ll \ell_{\text{cross}}$ to power-law decaying $\sim1/\ell^2$ for large depth $\ell \gg \ell_{\text{cross}}$.\footnote{
    For deep linear networks\index{deep linear network} where $\rho=0$, the solution~\eqref{eq:finite-angle-spin-2-solution} is degenerate and doesn't apply. However, from our discussion just before we know that for such networks the polar angle remains constant at any depth.
}
This implies that the true \terminate{critical exponent} for scale-invariant perpendicular perturbations is $p_\perp=2$.

Here, the crossover scale $\ell_{\text{cross}}$ is approximately given by
\be\label{eq:chrono-cross}
 \ell_{\text{cross}}\sim\frac{3}{\rho \psi^{(\ell=1)}}\sim\frac{3}{2\rho}\sqrt{\frac{K_{[0]}^{(\ell=1)}}{K_{[2]}^{(\ell=1)}}}\, .
\ee 
We get this 
by equating the small-depth constant answer, set by the first-layer condition, with the large-$\ell$ asymptotic answer given by \eqref{eq:scale-invariant-polar-angle}; on the right-hand side of \eqref{eq:chrono-cross} we further wrote the polar angle $\psi^{(\ell=1)}$ in terms of the kernel components using \eqref{eq:finite-angle-spin-0-solution} and \eqref{eq:finite-angle-spin-2-solution}.
What we see is that the smaller this initial angle $\psi^{(\ell=1)}$ is -- meaning that the closer the two inputs are to each other -- the longer our original constant solution to the naive perpendicular recursion~\eqref{eq:scale-invariant-perpendicular-naive} is valid, and the longer it takes for the power-law regime to kick in.

The discussion above explains why our $\delta$ expansion failed to see the crossover: in such an analysis, by construction, $K_{[2]}^{(\ell=1)}$ is infinitesimally small. This means that the crossover scale \eqref{eq:chrono-cross} is pushed to infinity, invisible to perturbation theory. Here is another way to see it. For small separation of two inputs, we can rewrite the angle as
\be
\psi^{(\ell)}\approx 2\sqrt{\frac{\delta\delta K_{[2]}^{(\ell)}}{\Kdif}}+\ldots \, ,
\ee
and hence the angle recursion \eqref{eq:off-diagonal-recursion-at-criticality-Taylor} can be recast -- upon a squaring and a rearrangement of terms -- as
\be
\delta\delta K_{[2]}^{(\ell+1)}=\delta\delta K_{[2]}^{(\ell)}-\frac{4\rho}{3\sqrt{\Kdif}}\le(\delta\delta K_{[2]}^{(\ell)}\ri)^{\frac{3}{2}}+\ldots\, .
\ee
Unfortunately, it's impossible to generate such a non-integer power, $3/2$, via a Taylor expansion.  Given our our ansatz for the perpendicular perturbation $K_{[2]}^{(\ell)}$~\eqref{eq:kernel-expand-3}, this explains why the correction term was invisible before. (There is no such issue for smooth activation functions; our Taylor expansion and subsequent analysis can be completely trusted for the $K^\star=0$ universality class\index{universality class!K@$K^\star=0$}.)

The overall lesson here is that we should
be very careful whenever encountering singular \terminate{Gaussian expectation}s.
In the future when we need to consider multiple inputs for nonlinear scale-invariant activation functions, we'll make sure to recall the results here.

%% file: Chp6-Bayesian/6_global.tex
\chapter{Bayesian Learning}
\label{ch:bayesian-inference}

\epigraph{\dots the mathematical rules of probability theory are not merely rules for calculating frequencies of `random variables'; they are also the unique rules for conducting inference (i.e.~plausible reasoning) of any kind, and we shall apply them in full generality to that end.}{E. T. Jaynes, explaining the theme of his book~\cite{jaynes2003probability}.\index{Jaynes, Edwin T.}}

\noindent{}In the previous three chapters, we've spent a considerable amount of \terminate{spacetime} analyzing the ensemble of wide neural networks at initialization.
In particular, through the $1/n$ expansion\index{$1/n$ expansion} and
deep asymptotic analysis, we've obtained a rather thorough understanding of the interplay between the architecture, width, depth, and \terminate{initialization hyperparameters}
that together
define
the effective
distribution of preactivations.

\index{deep learning!deep but not yet learning}
In this study, we've paid very careful attention to the \emph{deep} 
of \neo{deep learning} to the total neglect of the \emph{learning}. 
But this is a \emph{deep learning book}, not just a \emph{deep book}. Thus, in this chapter we will begin to learn about learning 
and -- if the titles of our chapters are any real guide to their contents --  will continue learning about learning for
the rest of the book.

We'll begin on our learning quest with a discussion of \neo{Bayesian inference}, as it provides a natural framework for thinking about learning in general.
We'll first explain in \S\ref{sec:PP-Bayes} the Bayesian approach to probability, in which probabilities are reinterpreted to represent the strength of our beliefs about the world according to different hypotheses. There, we'll learn 
that
the rules of \terminate{Bayesian inference} 
-- really the rules of logic
extended to
probabilistic
reasoning --
pick out a logically consistent way of incorporating newly observed information into the
probabilistic models representing our hypotheses.

From \S\ref{sec:PP-Bayes-2} on out, we'll see why this simple yet powerful framework enables us to analyze and then understand how deep neural networks learn from observed data.

In \S\ref{subsec:ForIO}, we'll detail how Bayesian model fitting\index{Bayesian inference!model fitting} works for neural networks.
First, we'll reinterpret our well-studied effective preactivation distribution as a \emph{prior} distribution, encoding our initial beliefs about the model outputs before observing any data.
With this as a starting point, the rules of \terminate{Bayesian inference} then imply a \terminate{learning algorithm} for sharpening our beliefs
so as to best fit our
observations.
The result of inference -- the \emph{posterior} distribution\index{posterior} -- further lets us make Bayesian predictions\index{Bayesian inference!prediction}
on novel inputs whose outputs we haven't observed but need to infer.
This naturally segues into a discussion of practical implementations: first we'll discuss approximation methods -- giving a Bayesian
interpretation to the gradient-based learning methods that we'll explore in the epochs following this chapter
-- and then we'll discuss an exact method on which the rest of the current chapter will be based.

In \S\ref{subsec:bayesian-model-comparison}, we'll expand our horizons by contemplating 
the ultimate question of
Life, the Universe, and Everything\index{Life, the Universe, \& Everything, HHGTTG}: Bayesian model comparison.\index{Bayesian inference!model comparison} 
We'll explain how to use Bayesian \emph{evidence}\index{Bayesian inference!evidence} to select between different plausible hypotheses, organized according to different choices of hyperparameters and network architectures, in order to pick the best ones.
Bayesian model comparison also gives us a quantitative means to address \emph{inductive biases}\index{inductive bias}, the often hidden assumptions built into deep learning models. As a bonus, we'll further see how \neo{Occam's razor}\index{Occam's razor|seealso{sparsity, principle of}} is automatically incorporated in the rules of \terminate{Bayesian inference} applied to such model comparison.
With these tools, we can really begin to address one of the fundamental questions we posed at the beginning of the book:
why do some neural network models perform so well while others fail?

These abstract discussions are then followed by an onslaught of concrete calculations for infinite- and finite-width neural networks in \S\ref{sec:infinite-posterior} and \S\ref{sec:finite-posterior}, respectively.

Some of these calculations reinforce the themes of the previous chapter.
We'll first show that Bayesian model comparison prefers critical initialization hyperparameters, giving additional evidence for the \emph{principle of criticality}\index{criticality!principle of} (\S\ref{subsec:Occam-criticality}). 
We'll also illustrate another role of finite-width interactions. Specifically, the accumulation of correlated \neo{fluctuations} induces an \terminate{inductive bias} for \neo{neural association}\index{neural association|seealso{Hebbian learning}}, leading to a propensity for \terminate{Hebbian learning}\index{Hebbian learning|seealso{neural association}} -- a learning principle inspired by biological neurons\index{biological neuron} (\S\ref{subsec:Hebbian}).

Some of these calculations
contrast qualitatively different characteristics of infinite- and finite-width models that are trained with exact Bayesian learning.
Analyzing the posterior distribution of network outputs, we'll see that correlations among different components of the output are nonzero at finite width only (\S\ref{subsec:absence-FF-Bayes}~$\!\perp$\S\ref{subsec:presence-FF-Bayes}). 
The resulting expressions will also make it clear why -- while theoretically quite tractable --  exact Bayesian learning is impractical for any dataset of reasonable size.
Next, analyzing the posterior distribution of hidden-layer representations, we'll see the absence/presence of representation learning at infinite/finite width (\S\ref{subsec:absence-RL-Bayes}~$\!\perp$\S\ref{subsec:presence-RL-Bayes}).
Overall, this contrasting will provide a valuable blueprint for when we later consider infinite- and finite-width models trained with gradient-based learning (\S\ref{ch:NTHb}~$\!\perp$\S\ref{ch:eot}).

\section{Bayesian Probability}\label{sec:PP-Bayes}
\index{Bayesian inference}\index{Bayesian probability}
A Bayesian
always starts with a \textbf{hypothesis}\index{hypothesis (Bayesian inference)|textbf}\index{Bayesian inference!hypothesis|see{hypothesis (Bayesian inference)}}\index{Bayesian probability!hypothesis|see{hypothesis (Bayesian inference)}} $\Hypo$. Mathematically, a hypothesis is a mechanism for assigning numbers $p(A\vert\Hypo)$ to \emph{statements}\index{Bayesian probability!statements}\index{statement|see{Bayesian probability}} $A$ about the world. These statements are logical propositions -- such as ``it will rain tomorrow" or ``this image $x$ contains a cat" or ``the output value for this function $f(x)$ evaluated on an input $x$ is $z$'' -- and these numbers $p(A\vert\Hypo)$ represent the relative plausibilities of those statements according to the assumptions or model of the world summarized by the hypothesis $\Hypo$.
In the context of \terminate{machine learning}, $p(A\vert\Hypo)$ is often called a \term{probabilistic model}.

As this notation and discussion 
should make clear, these beliefs $p(A\vert\Hypo)$ are expressed in the language of probability\index{probability (branch of mathematics)}. However, the \emph{Bayesian} interpretation of the probability $p(A\vert\Hypo)$ subtlety differs from the \neo{ensemble} interpretation that we gave in \S\ref{sec:MLP_distribution}. Namely, rather than representing the statistics\index{statistics (of a random variable)!Bayesian interpretation} of a random variable -- the relative frequency or chance observing $A$, given the conditions $\Hypo$ -- this probability instead 
constitutes
the strength of our belief in the proposition $A$ according to the assumptions $\Hypo$.\footnote{
The 
\neo{ensemble} interpretation is often called \term{frequentist probability} 
when contrasted with \term{Bayesian probability}.
In this book, we use the interpretation
that is most appropriate for the particular problem under consideration: if we're instantiating models by randomly drawing parameters from an \terminate{initialization distribution}, it makes sense to analyze an ensemble; if we're
making inferences based on a fixed hypothesis or comparing different hypotheses, it makes sense to adopt the Bayesian perspective.
}
\index{Bayesian inference}
Further, with such a Bayesian perspective all of probability theory and statistical inference can be uniquely derived as a consequence of logical constraints on these beliefs $p(A\vert\Hypo)$.\footnote{
    See Jaynes'\index{Jaynes, Edwin T.} book \cite{jaynes2003probability} for 
    an
    extended development of this perspective
    for which our brief summary does not give 
    justice.
}
We'll next brief you through these constraints as
they form the foundation of this chapter
but, as we have been using probabilities for quite a while now in this book, let us be brief.

Formally, the first logical constraint is known as the \textbf{product rule}\index{Bayesian probability!product rule},
\be\label{eq:logical-product}
p(A,B\vert\Hypo)=p(A\vert B,\Hypo)\,p(B\vert\Hypo)=p(B\vert A,\Hypo)\,p(A\vert\Hypo)\, ,
\ee
where $p(A,B\vert\Hypo)$ represents a \emph{joint} belief in both $A$ \emph{and} $B$ 
according to the hypothesis\index{hypothesis (Bayesian inference)} $\Hypo$, while $p(A\vert B,\Hypo)$ represents a \emph{conditional} belief 
in $A$
according to $\Hypo$ \emph{given} that $B$ 
has been observed.
The second logical constraint is known as the \textbf{sum rule}\index{Bayesian probability!sum rule},
\be\label{eq:logical-sum}
p(A\vert\Hypo) = \sum_{B} p(A,B\vert\Hypo) \, ,
\ee
and relates the joint belief in $A$ and $B$ to a marginal belief in just $A$.\footnote{We essentially discussed this sum rule as \eqref{eq:sum-rule-mlp}
under \S\ref{sec:sum-rule} \emph{Marginalization Rules}. \index{marginalization rule}
}
Here, the symbol $\sum_{B}$ represents a sum over all the logically possible values of a discrete variable $B$,
or for a continuous variable it represents an integral.\footnote{Though (Bayesian) probably it's already clear if you've made it this deep in the book, as we cannot be (Bayesian) certain, let us clarify the meaning of the \emph{statement}\index{Bayesian probability!statements} $A$ inside the belief system $p(A\vert\Hypo)$. Sometimes a statement represents a \emph{fixed} logical proposition, such as $A=$ ``Schr\"{o}dinger's cat is alive'' with $p(A\vert\Hypo)$ encoding the plausibility of cat's aliveness.\index{Schr\"{o}dinger's cat} Sometimes a statement represents a binary \emph{variable}, such as $B=$ ``the livelihood of Schr\"{o}dinger's cat'' which takes values in $\le\{\text{dead}, \text{alive}\ri\}$ with $p(B\vert\Hypo)$ giving the distribution over the two binary outcomes.
More generally, the statement can represent observable outcomes $\O$ of experiments -- a.k.a.~\emph{observables}\index{observable} -- with $p(\O\vert \Hypo)$ encoding our relative belief in the plausibilities of the different outcomes, where such observables can take on a discrete or continuous spectrum of values. Prominent examples of such general observables for us include the model parameters $\theta$ and preactivations $z^{(\ell)}$.\index{Bayesian probability!statements}
}
This sum rule in particular implies the normalization condition if we assign $p(C\vert\Hypo) \equiv 1$ for the statement $C$ that holds with \neo{absolute certainty} according to $\Hypo$: 
\be
\sum_{B} p(B\vert\Hypo) = \sum_{B} p(C,B\vert\Hypo)= p(C\vert\Hypo) = 1 \, .
\ee

With these rules in mind, after fixing a hypothesis\index{hypothesis (Bayesian inference)} a Bayesian then gathers information in order to refine the plausibilities of different beliefs.
For instance, after \emph{observing} $A$, we may want to \emph{update} our beliefs about $B$.
Such \term{Bayesian inference} can be accomplished by noting that an algebraic rearrangement of the product rule~\eqref{eq:logical-product} tells us how our beliefs should change as we condition on additional information $A$:
\be\label{eq:Bayes-rule}
p(B\vert A,\Hypo)=\frac{p(A\vert B,\Hypo)\, p(B\vert\Hypo)}{p(A\vert\Hypo)}\, . %
\ee
This rearrangement is so important that it's given its own name, \textbf{Bayes' rule}\index{Bayesian probability!Bayes' rule|textbf}, and even each individual factor of the equation
is named as well:
\bi
\item The factor $p(B\vert\Hypo)$ is called the \term{prior}\index{Bayesian inference!prior|see{prior}} of $B$, thusly named because it quantifies our belief in $B$ \emph{a priori}; that is, it encodes our belief in $B$  based entirely on our model $\Hypo$ before we observe any additional information.
\item The factor $p(B\vert A,\Hypo)$ is called the \term{posterior}\index{Bayesian inference!posterior|see{posterior}} of $B$ given $A$, thusly named because it quantifies our belief in $B$ \emph{a posteriori} upon learning $A$; that is, it encodes how our model $\Hypo$ updates its belief in $B$ after
 observing $A$.
\item The factor $p(A\vert B,\Hypo)$ is called the \textbf{likelihood}\index{likelihood|see{Bayesian inference}}\index{Bayesian inference!likelihood|textbf}.
We'll elaborate more on its name and interpretation later in \S\ref{subsec:ForIO} where we talk about model \emph{fitting}.\index{Bayesian inference!model fitting}
\item The factor $p(A\vert\Hypo)$ is called the \textbf{evidence}\index{evidence|see{Bayesian inference}}\index{Bayesian inference!evidence|textbf} for $\Hypo$.
We'll elaborate more on its name and interpretation later in \S\ref{subsec:bayesian-model-comparison} where we talk about model \emph{comparison}.\index{Bayesian inference!model comparison}
\ei
Note that the posterior is automatically normalized:
\be\label{eq:Bayes-consistency}
\sum_{B} p(B\vert A,\Hypo)=\sum_{B}\frac{p(A\vert B,\Hypo)\, p(B\vert\Hypo)}{p(A\vert\Hypo)}=\sum_{B}\frac{p(A, B\vert\Hypo)}{p(A\vert\Hypo)}=\frac{p(A\vert\Hypo)}{p(A\vert\Hypo)}=1\, .
\ee
More importantly, Bayes' rule %
 is the only logically consistent way to update a set of beliefs after making observations.

\section{Bayesian Inference and Neural Networks}\label{sec:PP-Bayes-2}
The Bayesian framework for inference can be used for building, updating, and reasoning with powerful probabilistic models\index{probabilistic model} of the world.
Let's now see how we can apply the Bayesian framework to \terminate{deep learning}, first 
for 
model fitting (\S\ref{subsec:ForIO}) and then 
for 
model comparison (\S\ref{subsec:bayesian-model-comparison}).\index{Bayesian inference!model fitting}\index{Bayesian inference!model comparison}

\subsection{Bayesian Model Fitting}\label{subsec:ForIO}
For neural networks, it's most natural to begin by discussing the  prior distribution $p(\theta \vert\Hypo)$ of the model parameters $\theta_{\mu}=\le\{\bias{i}{\ell},\W{ij}{\ell}\ri\}$.
This prior lets us quantify our initial beliefs about the particular values of the model parameters that determine our neural-network function approximator $f(x; \theta)$. The most common choice  is to simply reinterpret the \terminate{initialization distribution} of the ensemble,
\be\label{eq:parameter-prior}
p(\theta | \Hypo) \equiv \prod_{\ell=1}^{L}\le\{ \le[\prod_{i=1}^{n_
\ell} p\!\le(\bias{i}{\ell}\ri) \ri] \le[ \prod_{i=1}^{n_\ell}\prod_{j=1}^{n_{\ell-1} } p\!\le(\W{ij}{\ell}\ri) \ri] \ri\} \, ,
\ee
as our Bayesian prior distribution\index{prior}.
Here we recall that $p\!\le(\bias{i}{\ell}\ri)$ and $p\!\le(\W{ij}{\ell}\ri)$ -- given by \eqref{eq:full-bias-initialization} and \eqref{eq:full-weights-initialization} -- are zero-mean Gaussian distributions with bias variance $\Cb{\ell}$ and weight variance $\CW{\ell}/n_{\ell-1}$, respectively.

From the Bayesian perspective, these \terminate{initialization hyperparameters} are part of the hypothesis\index{hypothesis (Bayesian inference)} $\Hypo$.
This hypothesis\index{hypothesis (Bayesian inference)} $\Hypo$ also contains
our choice of architecture -- MLP\index{multilayer perceptron}, CNN\index{convolutional neural network}, \terminate{transformer}, etc.~-- as well as all the \terminate{architecture hyperparameters} within that architecture class -- e.g.~for MLPs we need to further select the depth $L$, the hidden-layer widths $n_\ell$, and the activation function $\sigma(z)$.
In short, $\Hypo$ is for $\Hypo$yperparameters\index{hyperparameters@$\Hypo$yperparameters}\index{hyperparameters@$\Hypo$yperparameters|seealso{hypothesis (Bay\-es\-ian inference)}}.\footnote{
To be strict, we should have always conditioned on 
$\Cb{\ell}$, $\CW{\ell}$, and $n_{\ell}$
whenever we discussed the initialization distribution: $p(\theta) \to p\!\le(\theta\Big\vert  n_0, \Cb{1}, \CW{1}, \ldots, n_{L-1}, \Cb{L}, \CW{L} \ri)$. 
Thankfully we've so far left, and will continue to leave, this type of detailed dependence implicit for notational simplicity. However, to underscore the importance of the hypothesis for \terminate{Bayesian inference}, in this chapter we \emph{(i)} will leave the conditioning on the overall hypothesis $\Hypo$ explicit until the end of \S\ref{subsec:Occam-criticality} 
and at the same time 
\emph{(ii)} will move the dependence of a dataset\index{input data} $\D$ to an overall subscript of the preactivations. 
As a particular example, the prior distribution\index{prior} of the $\ell$-layer preactivations $p\big(z^{(\ell)}_{\D} \big| \Hypo \big)$, defined next paragraph in~\eqref{eq:gigantic-beast-that-we-tame-reprint}, is equivalent to
what we've been denoting as $p\big(z^{(\ell)} \big| \D \big)$ outside of this chapter.
}

Here, we've taken familiar objects -- the hyperparameters and the initialization distribution characterizing the frequency of potential network realizations -- and interpreted them in the way of Bayes -- as the hypothesis $\Hypo$ and as the prior distribution $p(\theta \vert\Hypo)$ characterizing our initial beliefs
about the value of the model parameters.
Another familiar object, of course, is the distribution of
$\ell$-th-layer
preactivations that we've spent last three chapters evaluating explicitly.
To give that a Bayesian interpretation, let us first denote by $z^{(\ell)}_{\D}\equiv\le\{\z{i}{\delta}{\ell}\ri\}$ the set of $\ell$-th-layer preactivations
evaluated on inputs $\x{j}{\delta}\in\D$ in some dataset\index{input data} $\D$.
Then, the prior distribution\index{prior} over these $\ell$-th-layer preactivations can be related to the prior distribution over the model parameters by
\begin{align}\label{eq:gigantic-beast-that-we-tame-reprint}
p\!\le(z^{(\ell)}_{\D} \Big| \Hypo \ri) &= \int \Bigg[  \prod_{\mu=1}^{P} d \theta_{\mu}\Bigg] p\!\le(z^{(\ell)}_{\D}, \theta\Big\vert \Hypo \ri)=\int \Bigg[\prod_{\mu=1}^{P} d \theta_{\mu}\Bigg] p\!\le(z^{(\ell)}_{\D} \Big|\theta, \Hypo \ri)\, p(\theta| \Hypo)\, ,
\end{align}
where we've applied the sum rule\index{Bayesian probability!sum rule}~\eqref{eq:logical-sum} in the first equality and the product rule\index{Bayesian probability!product rule}~\eqref{eq:logical-product} in the second. 
This \terminate{prior}
quantifies our initial beliefs about the different neural-network variables.
More specifically, for a hidden layer $\ell$, this distribution represents our beliefs about a particular \neo{feature} \neo{representation} of the input
and, for the output layer $L$, this represents our initial beliefs about 
the behavior of 
the 
\neo{function approximation} $f(x; \theta)$.
More generally, for any neural-network observable $\O=\O(\theta)$, our prior beliefs are determined by
\begin{align}\label{eq:prior-observable-predition}
p\!\le(\O \big| \Hypo \ri) &=\int \Bigg[\prod_{\mu=1}^{P} d \theta_{\mu}\Bigg] p\!\le(\O \big|\theta, \Hypo \ri)\, p(\theta| \Hypo)\, .
\end{align}

To better illustrate what these formal expressions represent, let us take the network output $z^{(L)}_{\D}$ as an observable.
Then, the prior distribution for the output layer $p\!\le(z^{(L)}_{\D} \Big| \Hypo \ri)$,~\eqref{eq:gigantic-beast-that-we-tame-reprint}, is the same distribution as the \terminate{output distribution} induced by the initialization ensemble,~\eqref{eq:gigantic-beast-that-we-tame-with-Dirac},
\emph{if and only if} we also pick the conditional distribution of the outputs given the parameters to be deterministic:
\be\label{eq:deterministic-conditional-Bayes}
p\!\le(z^{(L)}_{\D} \Big|\theta, \Hypo \ri)=\prod_{i=1}^{n_{L}}\prod_{\delta\in\D}\delta\Big(z^{(L)}_{i;\delta}-f_i(x_{\delta};\theta)\Big)\, .
\ee
Here, $f_i(x_{\delta};\theta)$ is an expression for the network output given in terms of the iteration equation that defines the MLP \eqref{eq:mlp-definition}, while $z^{(L)}_{i;\delta}$ is interpreted as a random variable.
The resulting prior distribution\index{prior} for the network outputs $p\!\le(z^{(L)}_{\D} \Big| \Hypo \ri)$ then characterizes our overall initial belief
about the joint set of output values for a given set of inputs $\D$ according to the hypothesis $\Hypo$,
instead of characterizing the relative frequency of such output values at initialization across different realizations of the model parameters. That said, operationally, the formalism developed in the previous chapters can be directly brought to bear on calculating 
with
these  
beliefs.

Importantly, note that the deterministic conditional distribution for the output~\eqref{eq:deterministic-conditional-Bayes} is a part of our hypothesis within the Bayesian framework: according to the hypothesis $\Hypo$, given the model parameters $\theta$, the outputs are \emph{definitely} the ones computed by the function $f(x;\theta)$.\index{hypothesis (Bayesian inference)!deterministic}\index{hypothesis (Bayesian inference)!deterministic|seealso{Dirac delta function}}
Another common hypothesis is the
\emph{uncertain hypothesis}\index{hypothesis (Bayesian inference)!uncertain}\index{hypothesis (Bayesian inference)!uncertain|seealso{mean squared error}}\index{MSE loss|see{loss}}\index{hypothesis (Bayesian inference)!uncertain|seealso{Gaussian distribution}}
\be\label{eq:noisy-conditional-Bayes}
p\!\le(z^{(L)}_{\D} \Big|\theta, \Hypo \ri)=\prod_{i=1}^{n_{L}}\prod_{\delta\in\D}\le\{\frac{1}{\sqrt{2\pi \sigma_{\varepsilon}^2}}\exp\!\le[-\frac{1}{2\sigma_{\varepsilon}^2}\le(z^{(L)}_{i;\delta}-f_i(x_{\delta};\theta)\ri)^2\ri]\ri\}\, ,
\ee
which reduces to the deterministic hypothesis~\eqref{eq:deterministic-conditional-Bayes} in the
limit of zero variance and \terminate{absolute certainty}: $\sigma_{\varepsilon}^2\rightarrow 0$.\footnote{
This hypothesis is equivalent to injecting random noise $\varepsilon_i$ with mean zero and variance $\sigma_{\varepsilon}^2$ into the network output.
This in turn is tantamount to shifting the last-layer biases as $\bias{i}{L}\to\bias{i}{L}+\varepsilon_i$, and hence we can easily incorporate this in our analysis by shifting the final bias variance as $\Cb{L}\to\Cb{L}+\sigma_{\varepsilon}^2$.
You should keep in mind, however, that $\varepsilon_i$ is separate from the bias $\bias{i}{L}$ and is \emph{not} a part of the adjustable model parameters $\theta_{\mu}$; instead, this noise is intended to embody an intrinsic uncertainty present in our observation of the model's output.

Before moving on, let us also mention one other common 
hypothesis for the network output, the \emph{categorical hypothesis}\index{hypothesis (Bayesian inference)!categorical}\index{hypothesis (Bayesian inference)!categorical|seealso{softmax distribution}}\index{hypothesis (Bayesian inference)!categorical|seealso{cross-entropy loss}}, defined for each input $x$ by
\be\label{eq:softmax}
p(i \vert\theta, \Hypo)\equiv \frac{\exp\!\le[f_i(x;\theta)\ri]}{\sum_{j=1}^{n_L}\exp\!\le[f_j(x;\theta)\ri]}\, .
\ee
This distribution is also sometimes known as the \emph{softmax}\index{softmax distribution}\index{softmax distribution|seealso{logistic function}}.
Here, instead of considering a continuous distribution over the $n_{L}$ output \emph{values} $z_i^{(L)}$, we consider a discrete distribution over output \emph{classes} $i$, such as \texttt{dog} or \texttt{cat} or \texttt{car};
then, for such classification tasks, each number $p(i \vert\theta, \Hypo)$ quantifies our belief about how likely the input $x$ represents the class $i$. Functionally, the softmax\index{softmax distribution} can be thought of as a generalization of the \terminate{logistic function} \eqref{eq:sigmoid} in the sense that it maps a vector of real numbers to 
a discrete probability distribution.
}

Having now thoroughly discussed the \terminate{prior}, let's next consider the \terminate{posterior}. As we gather more information $A$ about the true behavior of our desired function $f(x)$, we should update our beliefs about our \terminate{probabilistic model} for $f(x;\theta)$.
In order to incorporate this information in a logically consistent manner, we should use  Bayes' rule\index{Bayesian probability!Bayes' rule}.
Specifically, to update our belief about the model parameters, Bayes' rule \eqref{eq:Bayes-rule} instructs us to use
\be\label{eq:Bayes-rule-model-fitting}
p(\theta \vert A,\Hypo)=\frac{p(A \vert \theta,\Hypo) \, p(\theta\vert\Hypo)}{p(A\vert\Hypo)}\, .%
\ee
Here, to find the posterior distribution\index{posterior} $p(\theta \vert A,\Hypo)$,  the  prior distribution\index{prior} $p(\theta\vert\Hypo)$ gets multiplied by the likelihood\index{Bayesian inference!likelihood} $p(A \vert \theta,\Hypo)$ of the model parameters $\theta$ for the observation of $A$,
and divided by
the evidence\index{Bayesian inference!evidence} $p(A\vert\Hypo)$.
Consequently, with such a posterior distribution\index{posterior} of the model parameters, our beliefs about any neural-network observable $\O$ shifts from our \terminate{prior} $p\!\le(\O \big|\Hypo \ri)$ \eqref{eq:prior-observable-predition} to a \terminate{posterior} with the insertion of $A$,
\begin{align}\label{eq:posterior-observable-predition}
p\!\le(\O \big|A, \Hypo \ri) &=\int \Bigg[\prod_{\mu=1}^{P} d \theta_{\mu}\Bigg] p\!\le(\O \big|\theta, \Hypo \ri)\, p(\theta|A, \Hypo)\, .
\end{align}
These two equations~\eqref{eq:Bayes-rule-model-fitting} and~\eqref{eq:posterior-observable-predition} uniquely determine how new information $A$ can be incorporated to change our beliefs about the value of any neural-network observable.

For function approximation tasks, such information often
comes in the form of some dataset $\A$ containing observed input-output pairs:
\be\label{eq:input-output-information}
A\equiv \le\{(\x{j}{\tra},\y{i}{\tra})\ri\}\vert_{\tra\in\A} \,.
\ee 
Here, each input $\x{j}{\tra}\in{\A}$ is paired with its corresponding true output $\y{i}{\tra}\equiv f_i(x_{\tra})$ recorded from our desired function $f(x)$.\footnote{For maximal disambiguation, in this chapter we'll use \terminate{sample indices} of the form $\tra$ -- the Greek letter alpha with a tilde on top --  for elements of the dataset $\A$ corresponding to input-output pairs for which the \emph{true} output values from $f(x)$ are observed.} 
With our observation of the true values $y_{\A}\equiv\le\{\y{i}{\tra}\ri\}$, the likelihood\index{Bayesian inference!likelihood} and evidence\index{Bayesian inference!evidence} are then given by the conditional belief $p\!\le(y_{\A} \big|\theta, \Hypo \ri)$ and the belief $p\!\le(y_{\A} \big| \Hypo \ri)$ for outputs, respectively. Such beliefs appeared before when considering the prior distribution of the outputs, \eqref{eq:gigantic-beast-that-we-tame-reprint} with $\ell= L$, but are now evaluated on the \emph{fixed} values $y_{\A}$ associated with the given inputs $x_{\A}$. 

To develop some intuition for what this means, let's again take the deterministic hypothesis\index{hypothesis (Bayesian inference)!deterministic}~\eqref{eq:deterministic-conditional-Bayes}. In this case, the likelihood\index{Bayesian inference!likelihood} is given by
\be\label{eq:posterior-deterministic-hypothesis}
p(A \vert \theta,\Hypo)\equiv p(y_{\A} \vert \theta,\Hypo)=\prod_{\tra\in\A}\prod_{i=1}^{n_{L}}\delta\Big(\y{i}{\tra}-f_i(x_{\tra};\theta)\Big)\, .
\ee
This likelihood\index{Bayesian inference!likelihood} quite explicitly restricts
the model parameters to those \emph{exactly} satisfying the constraints $f_i(x_{\tra};\theta)=\y{i}{\tra}$ \emph{fitting}\index{Bayesian inference!model fitting} our observations.
Vice versa, the functions in our set that do \emph{not} satisfy these constraints are completely thrown away from the posterior distribution\index{posterior}, deemed \emph{unlikely}.
Note what has just happened. Naively, $p(y_{\A}  \vert \theta,\Hypo)$ represents our beliefs about the output values $y_{\A}$, given that we set the parameters of our model to $\theta$. However, here we \emph{first} observed the true output values $y_{\A}$ and \emph{then} interpreted $p(y_{\A}  \vert \theta,\Hypo)$ in terms of how likely the model parameters $\theta$ fit the observation $A$.
This is the origin of the name ``likelihood\index{Bayesian inference!likelihood}'' and why the proper way to refer to it is ``the likelihood\index{Bayesian inference!likelihood} of the model parameters $\theta$ for the observation $A$.''

To develop even more intuition, it's customary to introduce the \term{negative log-likelihood}\index{negative log-likelihood|seealso{loss}}  $\L_{\A}\!\le(\theta\ri)$
 -- or \neo{loss} -- representation of the likelihood\index{Bayesian inference!likelihood}:
 \be
p(y_{\A} \vert \theta,\Hypo)\equiv  \exp\!\le[-\L_{\A}\!\le(\theta\ri)\ri]\, .
\ee
Here, by parameterizing the loss as a function of the parameters $\theta$, we are emphasizing that it's the (negative log-)likelihood \emph{of} the parameters.\footnote{While the likelihood\index{Bayesian inference!likelihood} function -- and therefore the loss\index{loss!auxiliary} -- is considered auxiliary from the perspective of \terminate{function approximation}, 
from the perspective of \terminate{Bayesian inference} the form of the likelihood is considered to be part of the hypothesis, cf.~the deterministic hypothesis\index{hypothesis (Bayesian inference)!deterministic} \eqref{eq:deterministic-conditional-Bayes} vs.~the uncertain hypothesis\index{hypothesis (Bayesian inference)!uncertain}~\eqref{eq:noisy-conditional-Bayes}.
}
For the uncertain hypothesis\index{hypothesis (Bayesian inference)!deterministic}~\eqref{eq:noisy-conditional-Bayes}, the negative log-likelihood takes the form of the famous mean-squared-error or \textbf{MSE loss}\index{loss!MSE}:
\be\label{eq:MSE-loss-preview}
\L_{\text{MSE}}(\theta)= \sum_{\tra\in \A}\le\{\frac{1}{2\sigma_{\varepsilon}^2}\big[f_i(x_{\tra}; \theta)-\y{i}{\tra}\big]^2+\frac{1}{2}\log\!\le(2\pi \sigma_{\varepsilon}^2\ri)\ri\}\, .
\ee
In particular, as the network outputs $f_i(x_{\tra}; \theta)$ get closer to their target values $\y{i}{\tra}$,
the MSE loss decreases
and the likelihood increases.\footnote{
In the deterministic limit $\sigma_{\varepsilon}^2\rightarrow 0$, the \terminate{loss} $\L_{\A}\!\le(\theta\ri)$ would be infinite for functions that don't \emph{exactly} fit all the constraints $f_i(x_{\tra};\theta)=\y{i}{\tra}$ and negative infinite for those that do.
Thus, the uncertain hypothesis\index{hypothesis (Bayesian inference)!uncertain} softens these hard-fitting constraints of the deterministic hypothesis\index{hypothesis (Bayesian inference)!deterministic} by relaxing the \terminate{Dirac delta function} distribution
to a Gaussian distribution with a finite variance $\sigma_{\varepsilon}^2$.

When we consider the categorical hypothesis\index{hypothesis (Bayesian inference)!categorical}~\eqref{eq:softmax}, the negative log-likelihood of the softmax\index{softmax distribution} distribution gives the \emph{cross-entropy loss}\index{loss!cross-entropy}\index{cross-entropy loss|see{loss}}. 
We'll more systematically address the consequences of these different choices of \terminate{loss} functions in \S\ref{ch:NTHb}.
} 
As such, the loss is a natural measure of how well our model is approximating the true behavior of the function.
Additionally, since the loss \eqref{eq:MSE-loss-preview} involves an explicit sum over observations,
as the number of observed input-output pairs $\NR$  increases, the likelihood can dominate the prior; that is, if we gather enough information, eventually our prior beliefs  can become entirely replaced by what we learned from our observations.

This is \textbf{Bayesian model fitting}\index{Bayesian inference!model fitting|textbf}: %
\terminate{Bayesian inference} \eqref{eq:Bayes-rule-model-fitting} is used
as a \neo{learning algorithm} to increase the accuracy of a \terminate{function approximation}. 
It gives greater preference to the functions that better fit  the constraints $f_i(x_{\tra};\theta)=\y{i}{\tra}$ and penalize the ones that don't.
The posterior~\eqref{eq:Bayes-rule-model-fitting} is then updated to reflect a balance between this preference for fitting our observations and an adherence to our prior beliefs about the values the model parameters should take.

Ultimately, we want to use our fit Bayesian model to make \textbf{Bayesian predictions}\index{Bayesian inference!prediction|textbf}. This is generically and abstractly embodied in~\eqref{eq:posterior-observable-predition}. Specifically and concretely, for function approximation tasks we are most often interested in posterior beliefs about the network outputs $\O = z^{(L)}$, for which~\eqref{eq:posterior-observable-predition} reads
\begin{align}\label{eq:posterior-observable-predition-but-actually}
p\!\le(z^{(L)} \Big|A, \Hypo \ri)  &=\int \Bigg[\prod_{\mu=1}^{P} d \theta_{\mu}\Bigg] p\!\le(z^{(L)} \Big|\theta, \Hypo \ri)\, p(\theta|A, \Hypo) \, .
\end{align}
Once we have this distribution, then we can in particular use its mean as our prediction and its variance as our level of confidence.
To compute any of these quantities, one way or another we need to perform  a gigantic integral over the model parameters $\theta$ in order to properly weight our different beliefs.
With that in mind, we'll now present two kinds of methods to tackle this model marginalization: \emph{(i)} approximate methods based on saddle-point approximations\index{saddle-point approximation} and \emph{(ii)} an exact method based on our \terminate{effective theory} approach.

\subsubsection{Approximation methods for model marginalization: MAP and MLE}\index{model fitting!Bayesian|see{Bayesian inference}}\index{Bayesian inference!model fitting!approximation methods}

One way to tackle such a gigantic integral is to presume that the integral measure, given by the posterior distribution $p(\theta \vert A,\Hypo)$ \eqref{eq:Bayes-rule-model-fitting}, is very concentrated around its \emph{mode}\index{mode}\index{mode|seealso{maximum a posteriori}}:
\be\label{eq:map-estimate}
\theta^\star_{\text{MAP}} \equiv \argmax_\theta p(\theta \vert A,\Hypo)= \argmax_\theta \le[p(y_{\A} \vert \theta,\Hypo) \, p(\theta\vert\Hypo)\ri] \, .
\ee
This maximum is known as the \term{maximum a posteriori} (MAP) estimate.
After such a maximization, we can use the function $f(x;\theta^\star_{\text{MAP}})$ for tasks and more generally approximate the full posterior distribution\index{posterior} $p\!\le(\O \big|A, \Hypo \ri)$  \eqref{eq:Bayes-rule-model-fitting} by the point estimate $\O\!\le(\theta^\star_{\text{MAP}}\ri)$.
This notion of approximating a probability distribution with single value of the random variable is known in statistics\index{statistics (branch of mathematics)}\index{statistics (branch of mathematics)|seealso{machine learning}} as a \neo{point estimate} and in \terminate{physics} as a \neo{saddle-point approximation}.\index{saddle-point approximation|seealso{point estimate}}\index{point estimate|seealso{mode}}
Another commonly-used saddle is given by the maximum of the likelihood\index{Bayesian inference!likelihood},
\be\label{eq:mle-estimate}
\theta^\star_{\text{MLE}} \equiv \argmax_\theta p(y_{\A} \vert \theta,\Hypo)  \, ,
\ee
known as the  \term{maximum likelihood estimation} (MLE)\index{MLE|see{maximum likelihood estimation}} of the model parameters.

In terms of the \terminate{negative log-likelihood}  $\L_{\A}\!\le(\theta\ri)$,
MLE is equivalent to the minimization of the \terminate{loss}
\be\label{eq:mle-estimate-loss}
\theta^\star_{\text{MLE}} = \argmin_\theta  \L_{\A}\!\le(\theta\ri) \, ,
\ee
while MAP\index{MAP|see{maximum a posteriori}}  estimate \eqref{eq:map-estimate} is a joint minimization of the loss and the negative log of the \terminate{prior},
\be\label{eq:map-estimate-loss}
\theta^\star_{\text{MAP}} = \argmin_\theta \, \le[\L_{\A}\!\le(\theta\ri)-\log p(\theta\vert\Hypo) \ri]\, .
\ee
In particular for a generic Gaussian prior of the form $p(\theta\vert\Hypo)\propto\exp\!\big(\!-\!\sum_{\mu=1}^P\! a_{\mu}\theta_\mu^2\big)$, the negative-log prior acts as a \neo{regularization} term of the form $\sum_{\mu=1}^P a_{\mu}\theta_\mu^2$ that has an effect of penalizing large parameter magnitudes.
Since the loss grows extensively\index{extensivity!of loss} with the size of the dataset $\A$ while this \terminate{regularization} term stays constant, when we've made sufficiently many observations, we naively expect that the prior will be eventually overwhelmed by the likelihood and that the MAP and MLE estimates will become similar.

If we are to apply these approximation methods to wide neural networks, there are certain things we need to keep in mind.\footnote{
In \S\ref{ch:NTHb}, we'll go through how all of this works in detail.
}
First of all,
there is actually no single optimal value for the \terminate{maximum likelihood estimation} $\theta^\star_{\text{MLE}}$. Instead, there are continuum of such optima, and we still have to consider a distribution over them. Importantly, such a distribution over maxima depends critically on how the maxima are obtained. For instance, it depends on the way you initialize model parameters $\theta_{\text{init}}$, the \terminate{learning algorithm} used to estimate these maxima -- such as \terminate{gradient descent} vs.~\emph{stochastic} gradient descent\index{stochastic gradient descent} -- and the \neo{training hyperparameters} controlling the learning algorithm.\index{learning algorithm} 
The study of this ensemble over optima and its dependence on the initialization\index{initialization hyperparameters} and \terminate{training hyperparameters} will more or less be the focus of the following chapters \S\ref{ch:training}--\S\ref{ch:eot}.\footnote{Since those following chapters will unsentimentally drop our Bayesian lens, let's interpret these different methods with fresh Bayesian eyes here.\label{foot:foretelling-GD}

In the \emph{impure} Bayesian approach -- that is MLE\index{maximum likelihood estimation} -- we have an \terminate{initialization distribution} $p(\theta_{\text{init}})$, but no prior distribution\index{prior} $p(\theta\vert\Hypo)$. By construction, the prior distribution\index{prior} does not enter into the estimate of the impure Bayesian~\eqref{eq:mle-estimate}, but the \terminate{initialization distribution}s \eqref{eq:full-bias-initialization} and \eqref{eq:full-weights-initialization} enters into their code to give particular realizations of networks acting as the starting points for optimization and \terminate{training}. Thus, such an initialization distribution induces a distribution over the resulting MLE estimates.

In the \emph{less impure} Bayesian approach -- that is MAP\index{maximum a posteriori} -- we have \emph{both} an \terminate{initialization distribution} $p(\theta_{\text{init}})$ \emph{and}  a prior distribution\index{prior} $p(\theta\vert\Hypo)$. For the former, we again use the initialization distributions \eqref{eq:full-bias-initialization} and \eqref{eq:full-weights-initialization} to provide starting points for optimization; for the latter, we typically use a Gaussian prior $p(\theta\vert\Hypo)\propto\exp\!\big(\!-\!\sum_{\mu=1}^P\! a_{\mu}\theta_\mu^2\big)$ which, as we said, serves as a \terminate{regularization} term when added to the optimization objective -- the loss -- as per \eqref{eq:map-estimate-loss}.

In the \emph{pure} Bayesian approach -- which is the focus of the rest of this chapter -- there is a prior distribution\index{prior} $p(\theta\vert\Hypo)$ but the \terminate{initialization distribution} $p(\theta_{\text{init}})$ isn't needed.
Pure Bayesians always integrate. What we really did with~\eqref{eq:parameter-prior} was pick a Gaussian prior over the parameters and then adopt the same conventions for the variances as we've been using for the \terminate{initialization distribution} \eqref{eq:full-bias-initialization} and \eqref{eq:full-weights-initialization}. %
We'll see in the rest of the chapter why this is sensible.
}

\subsubsection{Exact method for model marginalization: effective theory}\index{Bayesian inference!model fitting!exact marginalization}\index{integrating out}
For the \terminate{prior}~\eqref{eq:gigantic-beast-that-we-tame-reprint}, we know very well that it's possible to directly integrate out\index{integrating out} the model parameters through the use of a \terminate{$1/n$ expansion}. 
Such a gigantic marginalization was the focus of \S\ref{ch:ngp}, and in writing \eqref{eq:gigantic-beast-that-we-tame-reprint} we already reinterpreted our effective preactivation distribution at initialization as our prior beliefs about the preactivations.
For the \terminate{posterior}, the only hypothetical worry would be that we'd need to carry out entirely different sets of integrals. We'll show next that there is no such need.
Thus, in a very real sense the most painstaking theoretical part of \terminate{Bayesian inference} has already been taken care of for us!

Let's continue to suppose that we've made some observations $A$ of the true outputs $\y{i}{\tra}\equiv f_i(x_{\tra})$ of our function $f(x)$ for a given set of inputs $x_\A$ in a subsample $\A$ as defined by~\eqref{eq:input-output-information}.
We now want to incorporate what we've learned from these observations in order to update our beliefs about the output values
$z^{(L)}_{\B}\equiv\le\{\z{i}{\tea}{L}\ri\}$ for a potentially different set of inputs $\x{j}{\tea}\in\B$ in another subsample $\B$.\footnote{For maximal disambiguation, in this chapter we'll use \terminate{sample indices} of the form $\tea$ -- the Greek letter beta with a dot on top --  for elements of the dataset $\B$ corresponding to input-output pairs for which outputs values from $f(x)$ are \emph{not} observed but instead to be \emph{inferred}. 
}
Beginning with the joint \terminate{prior} for the network outputs over the union of both subsamples $\D\equiv\A\cup\B$, 
\be
p\!\le(z_{\D}^{(L)}\Big\vert  \Hypo \ri) \equiv p\!\le(z_{\A}^{(L)}\!,z_{\B}^{(L)}\Big\vert  \Hypo \ri)\, ,
\ee
we can set $z_{\A}^{(L)} \to y_\A$ and use the product rule~\eqref{eq:logical-product} to condition our beliefs about $z_{\B}^{(L)}$ on the observed \emph{true} values $y_\A$:
\be
p\!\le(y_{\A},z_{\B}^{(L)}\Big\vert  \Hypo \ri)=p\!\le(z_{\B}^{(L)}\Big\vert y_{\A}, \Hypo \ri)p\!\le(y_{\A}\Big\vert  \Hypo \ri)\, .
\ee
Then, rearranging terms like we are Reverend Thomas Bayes\index{Bayes, Reverend Thomas}, we get
\be\label{eq:BB-NN}
p\!\le(z_{\B}^{(L)}\Big\vert y_{\A}, \Hypo \ri)=\frac{p\!\le(y_{\A},z_{\B}^{(L)}\Big\vert  \Hypo \ri)}{p\!\le(y_{\A}\Big\vert  \Hypo \ri)}\, .
\ee
Since this iteration of Bayes' rule\index{Bayesian probability!Bayes' rule} is so important, let us be verbose and crystal clear about its interpretation: the denominator $p\!\le(y_{\A}\big\vert  \Hypo \ri)$ is the \terminate{prior} for the network outputs given the inputs $x_\A$ in the subsample $\A$, evaluated on the \emph{fixed} observed values $y_{\A}$, hence it is just a \emph{number}; the numerator $p\!\le(y_{\A},z_{\B}^{(L)}\Big\vert  \Hypo \ri)$ is the \terminate{prior} for the network outputs given the inputs $x_\D$ in
the joint dataset $\D\equiv\A\cup\B$, evaluated on the \emph{fixed} observed values $y_{\A}$ but with the network outputs $z_{\B}^{(L)}$ still \emph{variable}, hence it is a \emph{function} of the $z_{\B}^{(L)}$.\footnote{
    The reason we say \emph{given the inputs} here is that technically we should also be conditioning on $x_\A$ and $x_\B$ as well. In particular, while $y_{\A}$ is fixed and $z_{\B}^{(L)}$ is completely variable in the expression for the joint prior
    \be
    p\!\le(y_{\A},z_{\B}^{(L)}\Big\vert  \Hypo \ri)  \equiv p\!\le(y_{\A},z_{\B}^{(L)}\Big\vert x_\D, \Hypo \ri)  \,, 
    \ee
    the full set of inputs
    $x_\D \equiv x_\A \cup x_\B$ determines the \emph{data-dependent couplings}\index{data-dependent coupling} $g_{(L)}$ and $v_{(L)}$ -- or equivalently the metric  $G^{(L)}$ and the \terminate{four-point vertex} $V^{(L)}$ -- that parameterize the output distribution. We will see how this works in more detail in the following sections.
}
The numerator and denominator combine to make the \terminate{posterior} on the left-hand side, which is thus a function of the random variable $z_{\B}^{(L)}$ encoding our \terminate{posterior} beliefs about the plausible values of the network outputs $z_{\B}^{(L)}$ for the inputs $x_\B$ in $\B$, updated with our observations about the true values $y_{\A}$ of the outputs for the inputs $x_\A$ in $\A$.
In this way, rather than performing \terminate{Bayesian inference} to learn about the model parameters as way of maintaining different beliefs about the different functions $f(x; \theta)$ in our flexible set, here we simply update our beliefs about the behavior of the function $f(x)$ directly.

In this presentation of Bayes' rule \eqref{eq:BB-NN}, the marginalization over all the model parameters\index{marginalizing over} already occurred in our transition from \eqref{eq:parameter-prior}, the prior over the parameters, to \eqref{eq:gigantic-beast-that-we-tame-reprint}, the prior over the preactivations. The resulting posterior~\eqref{eq:BB-NN} is in fact exactly equivalent to what you'd get by explicitly doing a marginalization over a posterior distribution of the model parameters, e.g.~as in \eqref{eq:posterior-observable-predition}. %
To see why, consider the following set of manipulations:
\begin{align}
p\!\le(z_{\B}^{(L)}\Big\vert y_{\A}, \Hypo \ri)=&\int \le[  \prod_{\mu=1}^{P} d \theta_{\mu}\ri] p\!\le(z_{\B}^{(L)}, \theta \Big\vert y_{\A},\Hypo\ri)=\int \le[  \prod_{\mu=1}^{P} d \theta_{\mu}\ri] p\!\le(z_{\B}^{(L)}\Big\vert \theta, \Hypo \ri)\ p(\theta \vert y_{\A},\Hypo)\, \notag\\
=&\int \le[  \prod_{\mu=1}^{P} d \theta_{\mu}\ri] p\!\le(z_{\B}^{(L)}\Big\vert \theta, \Hypo \ri) \le[ \frac{p(y_{\A}\vert \theta,\Hypo) \, p(\theta\vert\Hypo)}{p(y_{\A}\vert\Hypo)}\ri] \, \notag\\
=&\frac{1}{p(y_{\A}\vert\Hypo)}\int \le[  \prod_{\mu=1}^{P} d \theta_{\mu}\ri] p\!\le(y_{\A}, z_{\B}^{(L)}\Big\vert \theta, \Hypo \ri) \, p(\theta\vert\Hypo) \notag\\
=&\frac{1}{p(y_{\A}\vert\Hypo)}\int \le[  \prod_{\mu=1}^{P} d \theta_{\mu}\ri] p\!\le(y_{\A}, z_{\B}^{(L)}, \theta\Big\vert \Hypo \ri) =\frac{p\!\le(y_{\A},z_{\B}^{(L)}\Big\vert  \Hypo \ri)}{p\!\le(y_{\A}\Big\vert  \Hypo \ri)}\,  .\label{eq:BB-NN-long}
\end{align}
The only nontrivial step is 
in the third line, where we reversed the factorization,
\be\label{eq:factorization-joint-network-outputs}
p\!\le(z_{\A}^{(L)}\!, z_{\B}^{(L)}\Big\vert \theta, \Hypo \ri)=p\!\le(z_{\A}^{(L)}\Big\vert \theta, \Hypo \ri)\, p\!\le(z_{\B}^{(L)}\Big\vert \theta, \Hypo \ri)\, ,
\ee
and evaluated at $z_{\A}^{(L)} \to y_\A$. 
This factorization \eqref{eq:factorization-joint-network-outputs} says that the network outputs are \emph{conditionally independent}, given the parameters.
This is a consequence of the fact that -- for a fixed set of network parameters -- the output on an example $x_{\tilde{\alpha}}$ is entirely independent from the output evaluated on any other example $x_{\dot{\beta}}$, which is manifestly true for all the hypotheses that we mentioned. (If it were not, neural networks would be pretty useless in practice.) 
The use of Bayes' rule\index{Bayesian probability!Bayes' rule} for the model parameters in the square brackets in the second line also makes manifest the connection between \emph{Bayesian model fitting}\index{Bayesian inference!model fitting}~\eqref{eq:Bayes-rule-model-fitting} on the one hand and \emph{Bayesian prediction}\index{Bayesian inference!prediction}~\eqref{eq:posterior-observable-predition} on the other hand.

\index{integrating out}
As we already alluded to, this exact method for model marginalization is closely connected with our \terminate{effective theory} approach to understanding neural networks. 
In particular, while the model parameters are always part of the definition of our neural networks, we've always had to integrate them out in the process of determining the distribution over the network outputs. In this way, our \terminate{effective theory of deep learning} has always worked directly with the entire \terminate{ensemble} of network functions implied by the \terminate{initialization distribution} of the parameters~\eqref{eq:parameter-prior} rather than with any \emph{particular} network. Up until now, we've motivated this ensemble approach via the \emph{principle of typicality}\index{typicality!principle of}, in which we use the ensemble to analyze how a typical realization is likely to behave.\footnote{
    In \S\ref{ch:NTKa} and onwards, we'll see how this principle is manifested in neural networks \emph{trained} via gradient-based learning.
}
\index{marginalizing over}
Here we have a slightly different interpretation: rather than trying to make the ensemble describe a typical network, we actually want to consider the posterior predictions across the full set of potential networks, each weighted according to our posterior beliefs about how plausible those predictions are.

Now, after a brief detour into Bayesian model comparison\index{Bayesian inference!model comparison}, much of the focus of \S\ref{sec:infinite-posterior} and \S\ref{sec:finite-posterior} will be the explicit evaluation of these Bayesian predictions\index{Bayesian inference!prediction}~\eqref{eq:BB-NN} for infinite- and finite-width MLPs, respectively.

\subsection{Bayesian Model Comparison}\label{subsec:bayesian-model-comparison}\index{model comparison!Bayesian|see{Bayesian inference}}\index{marginalizing over}\index{integrating out}
In the context of Bayesian model fitting\index{Bayesian inference!model fitting} and Bayesian prediction\index{Bayesian inference!prediction}, the \emph{evidence}\index{Bayesian inference!evidence} $p\big(y_{\A}\big\vert  \Hypo\big)$ has thus far played essentially no role. In the context of our approximation methods, MAP and MLE\index{maximum a posteriori}\index{maximum likelihood estimation} and their respective maximizations \eqref{eq:map-estimate} and \eqref{eq:mle-estimate}, the value of the argument maximization is strictly independent of the evidence, since it doesn't depend on the \terminate{model parameters}. In the context of our exact method for Bayesian prediction\index{Bayesian inference!prediction}, the evidence is simply the \terminate{normalization factor} of the posterior, which is trivial for us to compute.

To actually see the role of the evidence\index{Bayesian inference!evidence} in action, %
\emph{you mustn't be afraid to dream a little bigger, darling}.\index{Eames (\emph{Inception} meme)}
That is, rather than being fixated on a single hypothesis $\Hypo$, we instead consider a multitude of different hypotheses $\Hypo_{a}$ as possible explanations for our data. This is the essence of \textbf{Bayesian model comparison}\index{Bayesian inference!model comparison|textbf}: using the \emph{evidence}\index{Bayesian inference!evidence} to weigh the plausibility of different probabilistic models\index{probabilistic model} as explanations for all of our observations. In the context of \terminate{deep learning}, this corresponds to comparing our relative beliefs in the different modeling choices encapsulated in each $\Hypo_{a}$ -- i.e.~comparing different hyperparameter settings -- and determining which modeling choice provides the best description of our observations $y_\A$.

To begin, let us again use Bayes' rule\index{Bayesian probability!Bayes' rule} -- this time on the evidence -- to invert the conditioning as
\be\label{eq:bayes-rule-model-comparison}
p\!\le( \Hypo_{a}\big\vert y_{\A}  \ri) = \frac{p\!\le(y_{\A}\big\vert \Hypo_{a}\ri) p\!\le(\Hypo_{a}\ri) }{p\!\le( y_{\A}\ri)  } \, .
\ee
In this form, the \terminate{posterior} $p\!\le( \Hypo_{a}\big\vert y_{\A}  \ri) $ on the left-hand side encodes our updated beliefs in the plausibility of the different hypotheses  $\Hypo_{a}$ -- the different hyperparameters settings -- given our observation $y_{\A}$, while the \terminate{prior} $p\!\le( \Hypo_a \ri)$ on the right-hand side encodes our initial beliefs about 
these hypotheses. 
Amusingly,  the \emph{old} \emph{evidence}\index{Bayesian inference!evidence} $p\!\le(y_{\A}\big\vert \Hypo_{a}\ri)$ for the hypothesis $\Hypo_{a}$ from our Bayesian model fitting\index{Bayesian inference!model fitting}
now appears as the \emph{new} \emph{likelihood}\index{Bayesian inference!likelihood} $p\!\le(y_{\A}\big\vert \Hypo_{a}\ri)$ of the hypothesis $\Hypo_{a}$ for the observation $y_{\A}$ in the context of Bayesian model comparison. Lastly, the \emph{new} \emph{evidence}\index{Bayesian inference!evidence} $p\!\le( y_{\A}\ri)$  
is just a \terminate{normalization factor} that we can safely ignore.\footnote{
    Unless, of course, we aren't afraid to dream even bigger.
    If we did -- \emph{narrator:} they won't\index{Narrator (\emph{Arrested Development})} -- 
    we'd need to introduce a \neo{meta hypothesis}\index{hypothesis (Bayesian inference)!meta hypothesis|see{meta hypothesis}},  $\mathcal{G}$, that encodes our \terminate{prior} beliefs about different hyperparameter configurations $p(\Hypo_{a}  \vert\mathcal{G})$. This is sometimes called \emph{Bayesian hierarchical modeling}\index{Bayesian inference!hierarchical modeling}.
    In this case, Bayesian model comparison in terms of this even grander evidence $p( y_{\A}) \to p( y_{\A}\vert\mathcal{G})$  in principle involves integrating overall \emph{all} the probabilistic models\index{probabilistic model} as $p( y_{\A}\vert\mathcal{G})=\sum_{a}p(y_{\A}\vert \Hypo_{a})\ p(\Hypo_{a}\vert\mathcal{G})$,  i.e.~any and all hypotheses $\Hypo_{a}$ that are encoded by $\mathcal{G}$. 
    The distinction between the \terminate{meta hypothesis} $\mathcal{G}$ and hypotheses $\Hypo_{a}$ is somewhat arbitrary, however; for instance, we could put into $\mathcal{G}$ our overall choice of architecture -- e.g.~MLP\index{multilayer perceptron}, CNN\index{convolutional neural network}, \terminate{transformer} -- and then let $\Hypo_{a}$ index the different settings of $\Hypo$yperparameters. Then, recursing again, a Bayesian model comparison over $\mathcal{G}$ would be a weighted evaluation of the best architecture for the data, taking into account all possible settings of the hyperparameters for those architectures.
}

To see how the model comparison works, let's use \eqref{eq:bayes-rule-model-comparison} to compare two different hypothesis, $\Hypo_1$ and $\Hypo_2$, in order to determine which is a better fit for our observations. 
Since our \emph{relative beliefs} are all that matter, let's take the ratio of the two  \terminate{posterior}s,
\be\label{eq:bayes-factor}
\frac{p\!\le( \Hypo_{1}\big\vert y_{\A}  \ri)}{p\!\le( \Hypo_{2}\big\vert y_{\A}  \ri)} = \le[\frac{p\!\le(y_{\A}\big\vert \Hypo_{1}\ri)}{p\!\le(y_{\A}\big\vert \Hypo_{2}\ri)}\ri] \frac{p\!\le(\Hypo_{1}\ri) }{p\!\le(\Hypo_{2}\ri) }\, ,
\ee
from which we see that the irrelevant \terminate{normalization factor} $p( y_{\A})$ simply drops out.
Here, the ratio in the square brackets is sometimes given the name the \textbf{Bayes' factor}\index{Bayesian inference!model comparison!Bayes' factor|textbf}, which in turn multiplies the ratio of our prior beliefs.
In particular, the Bayes' factor contains all of the observation dependence and characterizes how we should update our relative prior beliefs in each hypothesis given the new data $y_{\A}$.
A ratio 
greater than one indicates that the model specified by hypothesis $\Hypo_1$ is favored, while a ratio less than one indicates that the model specified by hypothesis $\Hypo_2$ is favored. In this way, the old evidence -- i.e.~the new likelihood -- $p\!\le(y_{\A}\big\vert \Hypo_{a}\ri)$ can be very useful, indeed.

\subsubsection{Occam's razor}
In order to further elaborate on the mechanism behind Bayesian model comparison\index{Bayesian inference!model comparison}~\eqref{eq:bayes-factor}, let us pick up \term{Occam's razor} \cite{occam},
which is the famous \emph{principle of sparsity}\index{sparsity, principle of}. It says that we should favor the simplest hypothesis that fits all the observations. In the context of \terminate{machine learning} and parameterized probabilistic modeling\index{probabilistic model}, this principle is often intended as a heuristic that guides us to favor models with fewer parameters, all else being equal. The intuitive explanation for this heuristic is that models with more parameters have greater flexibility to fit the observed data, making them more likely to \emph{overfit}\index{overfitting}\index{overfitting|seealso{generalization}} and less likely to \emph{generalize}\index{generalization} to explain new observations.\footnote{It's natural to wonder here how to interpret this overfitting in light of the fact that we've actually integrated out all our parameters! (In the \terminate{machine learning} literature, such ensembles are sometimes called \emph{non-parametric models}\index{non-parametric model|seealso{Gaussian process}}\index{non-parametric model}, though we really do not like such terminology, given the following explanation.) The culprit for this potential confusion is the overloaded usage of the word \emph{parameter}.
To illustrate this with the extreme, let's consider the \terminate{infinite-width limit}.
Despite formally starting with an infinite number of model parameters 
-- giving a model that is naively very \emph{overparameterized}\index{overparameterization}, to say the least -- 
the \terminate{effective theory} of the output distribution is completely characterized by the kernel $\ker^{(L)}$, which can be described by a finite number of \emph{data-dependent couplings}\index{data-dependent coupling} $\sim N_{\D}^2$. 
Thus, from the \terminate{macroscopic perspective} of Bayesian model comparison\index{Bayesian inference!model comparison}, it's these couplings that control the \terminate{model complexity} and not what we usually call the parameters, the tunable weights and biases. We will discuss this further and in greater detail in Epilogue~\ref{epi:overparameterization}, and in particular we'll highlight how the \terminate{$1/n$ expansion} for finite-width networks leads to a sequence of effective theories with increasing complexity.\label{footnote:occam-non-parametric}
}

Naively, Bayesian model comparison\index{Bayesian inference!model comparison}~\eqref{eq:bayes-rule-model-comparison} seems to give us a very natural way to implement this razor: we can \emph{subjectively} adjust the ratio of our prior beliefs $p(\Hypo_1) / p(\Hypo_2)$ to explicitly favor the simpler hypothesis, a priori penalizing more complicated models. However, as MacKay\index{MacKay, David} \cite{mackay1995probable} points out:

\vspace{-.30\baselineskip}
\setlength\epigraphwidth{.93\textwidth}
\epigraph{Coherent [Bayesian] inference embodies Occam's Razor automatically and quantitatively.}{}
\setlength\epigraphwidth{.9\textwidth}

\vspace{-1.80\baselineskip}
\noindent{}That is, Occam's razor is \emph{objectively} built into Bayesian model comparison\index{Bayesian inference!model comparison}~\eqref{eq:bayes-factor} through the Bayes' factor.\footnote{
See MacKay's\index{MacKay, David} excellent exposition \cite{mackay1995probable} for further details and examples,
with a particular emphasis on (pre-deep-learning-era) neural networks.
}

To understand why,
note that the prior distribution\index{prior} $p\!\le(z_\A^{(L)} \Big\vert \Hypo_{a}\ri)$
needs to be normalized. This means that for a given hypothesis\index{hypothesis (Bayesian inference)} $\Hypo_{a}$
to be complicated enough to explain an overwhelmingly wide variety of \emph{potential} observations $z_\A^{(L)}$, it must have small support on any \emph{particular} observation $y_\A$.
Hence the evidence $p(y_\A \vert \Hypo_{a})$  for such a hypothesis
will be small regardless of which actual observation we make.
In contrast, if the hypothesis is very simple, the \terminate{prior} $p\!\le(z_\A^{(L)} \Big\vert \Hypo_{a}\ri)$ will make a constrained set of predictions, but make them strongly, by concentrating its support on only a few plausible outcomes. Thus, the simplest models that still correctly predict the observation $y_{\A}$ are naturally preferred by the Bayes' factor\index{Bayesian inference!model comparison!Bayes' factor} $p(y_\A \vert \Hypo_{1}) / p(y_\A \vert \Hypo_{2})$ alone. In addition, the more observations we make that are correctly predicted, the more the Bayes' factor\index{Bayesian inference!model comparison!Bayes' factor} will amplify this preference for simpler models that still fit.\footnote{This is analogous to the way the likelihood\index{Bayesian inference!likelihood} factor will dominate the \terminate{prior} as observations accumulate when Bayesian model fitting\index{Bayesian inference!model fitting}.}

Since the Bayes' factor\index{Bayesian inference!model comparison!Bayes' factor} automatically and objectively implements \terminate{Occam's razor}, there's no need to subjectively express a preference for simpler models using the prior over hypotheses $p(\Hypo_a)$. This means that for a discrete set of hypothesis $\{\Hypo_a \}$, we can choose the prior distribution to be uniform, giving equal a priori preference to any particular hypothesis  $\Hypo_a$ regardless of their complexity.
With this choice our Bayesian model comparison\index{Bayesian inference!model comparison} is completely characterized by the  Bayes' factor\index{Bayesian inference!model comparison!Bayes' factor}:
\be\label{eq:bayes-factor-equal-hypo}
\frac{p\!\le( \Hypo_{1}\big\vert y_{\A}  \ri)}{p\!\le( \Hypo_{2}\big\vert y_{\A}  \ri)} = \frac{p\!\le(y_{\A}\big\vert \Hypo_{1}\ri)}{p\!\le(y_{\A}\big\vert \Hypo_{2}\ri)}\, .
\ee
Thus, we should really think of \terminate{Occam's razor} as the \emph{inductive bias} of \terminate{Bayesian inference} applied to model comparison.\index{Bayesian inference!model comparison}

\subsubsection{Inductive Bias}
Given our last statement, we should clarify about something that we've been informally referring to since \S\ref{sec:MLP_intro} but now are finally ready to formally address:
\term{inductive bias}.

Way back in \S\ref{sec:MLP_intro}, inductive biases were introduced as something implicit that is 
built into a neural network architecture in order that the set of functions $\{ f(x;\theta)\}$ may better represent the properties of a particular dataset\index{input data} $\D$ and the
\terminate{function approximation} task at hand. From the Bayesian perspective, inductive biases represent the a priori assumptions made about the desired function $f(x)$ before any observations are made. More broadly, both hypotheses and \terminate{learning algorithm}s may have their own set of inductive biases; e.g.~we've just pointed out that \terminate{Occam's razor} is an inductive bias of \terminate{Bayesian inference}.

Throughout \S\ref{sec:infinite-posterior} and \S\ref{sec:finite-posterior}, we'll encounter various 
inductive biases  while performing concrete calculations for infinite- and finite-width MLPs.
Here, let's consider a very simple example for illustration: suppose that a Bayesian firmly believes with \neo{absolute certainty} that a statement $B$ is false such that their hypothesis $\Hypo_{\overline{B}}$ assigns an a priori probability of zero to this belief as $p(B\vert\Hypo_{\overline{B}})=0$; then, via Bayes' rule\index{Bayesian probability!Bayes' rule}~\eqref{eq:Bayes-rule}, there's no way that the \terminate{posterior} on $B$ can be updated to be anything other than zero, even if the Bayesian gathers some new information $A$ that would serve as positive evidence for $B$. In this case,
$\Hypo_{\overline{B}}$ is clearly a bad hypothesis; its inductive bias is leading to an absurdly stubborn set of beliefs. Alternatively, if $B$ turns out to be actually false, $\Hypo_{\overline{B}}$ is a good hypothesis because it can then assign more probability to other plausibly true statements. As this \neo{gedanken inference} illustrates, the advantage and disadvantage of an inductive bias depends on the ground truth.

Returning to our initial example in \S\ref{sec:MLP_intro} of the \terminate{inductive bias} of different neural-network architectures, the advantage of one architecture over another is a highly data- and task-dependent question.
In principle, we could use Bayesian model comparison\index{Bayesian inference!model comparison}~\eqref{eq:bayes-factor} to directly compare these different architectures -- MLPs\index{multilayer perceptron}, CNNs\index{convolutional neural network}, and transformers\index{transformer} -- for different sets of observations $y_{\A}$ if only we knew how to compute the evidence\index{Bayesian inference!evidence} $p\!\le(y_{\A}  \big\vert \Hypo \ri)$ for those architectures.\footnote{
Recall from \S\ref{sec:MLP_intro} that CNNs~\eqref{eq:conv-layer} are designed to capitalize on the fact that \terminate{computer vision} data organizes useful information in a spatially-local translationally-invariant manner.\index{translational invariance}
Incorporating this property into the architecture design is an \terminate{inductive bias} of the CNN; in particular, the assumption is that a cat is still a $\texttt{cat}$, even if it's shifted \emph{up up down down left right left right} $BA$\index{Konami Code}. The advantage of such an \terminate{inductive bias} as compared to MLPs should be directly encoded in a Bayes' factor\index{Bayesian inference!model comparison!Bayes' factor} $p\!\le( y_\A  \big\vert \Hypo_{\text{CNN}} \ri)/p\!\le( y_\A  \big\vert \Hypo_{\text{MLP}} \ri)$. This ratio should presumably be greater than one for any dataset with desired outputs $y_\A$ for which 
the assumption of spatial locality is a useful inductive bias.
}
The formalism of our \terminate{effective theory of deep learning} as laid out in the earlier chapters is precisely a blueprint for computing such factors for different architectures as a function of a particular dataset. We encourage you to give it a try.

\section{Bayesian Inference at Infinite Width}\label{sec:infinite-posterior}
In this section, we'll give three lessons on Bayesian learning in the \terminate{infinite-width limit}. %
First, we'll calculate the evidence\index{Bayesian inference!evidence} $p(y_{\A}\vert\Hypo)$ 
and see 
that Bayesian model comparison\index{Bayesian inference!model comparison} prefers \terminate{criticality} for sufficiently deep networks (\S\ref{subsec:Occam-criticality}). 
Then, we'll calculate the posterior distribution for the network outputs $p\!\le(z_{\B}^{(L)}\Big\vert y_{\A},\Hypo\ri)$ and see that different output components are completely independent 
in this limit 
(\S\ref{subsec:absence-FF-Bayes}). 
Finally, we'll calculate the posterior distribution\index{posterior} of preactivations in the \emph{penultimate} layer $p\!\le(z_{\D}^{(L-1)}\Big\vert y_{\A},\Hypo\ri)$ 
and show that it's identical to the penultimate prior distribution $p\!\le(z_{\D}^{(L-1)}\Big\vert \Hypo\ri)$, %
thus implying that such infinitely-wide networks lack \terminate{representation learning} (\S\ref{subsec:absence-RL-Bayes}).

Before we begin, let's start with some reminiscence, recast through the lens of our new Bayesian glasses.\index{glasses (Bayesian)}
In the infinite-width limit, the prior distribution\index{prior} over the network outputs is given by a simple zero-mean Gaussian distribution
\be\label{eq:GP-for-once}
p\!\le(z^{(L)}_{\D}\Big\vert \Hypo\ri) = \frac{1}{\sqrt{\dete{2\pi \ker}^{n_{L}}}} \exp\!\le(-\frac{1}{2}\sum_{i=1}^{n_{L}}\sum_{\delta_1,\delta_2\in\D}\ker^{\delta_1 \delta_2}\z{i}{\delta_1}{L}\z{i}{\delta_2}{L}\ri)\, ,
\ee
with the variance $\ker_{\delta_1\delta_2}\equiv\ker_{\delta_1\delta_2}^{(L)}=\ker^{(L)}(x_{\delta_1},x_{\delta_2})$ given by the 
kernel at the output layer -- here with the layer index dropped -- 
depending
explicitly on pairs of inputs $x_{\delta_1}$ and $x_{\delta_2}$ from the dataset\index{input data} $\D$ and implicitly on the $\Hypo$yperparameters $C_b$ and $C_W$. %
Also recall that, as per our \emph{general relativistic}\index{general relativity} conventions, the matrix $\ker^{\delta_1 \delta_2}$ is the \emph{inverse} of the covariance matrix $\ker_{\delta_1\delta_2}$
\be\label{eq:whole-inverse}
\sum_{\delta_2\in\D}\ker^{\delta_1\delta_2}\ker_{\delta_2\delta_3}=\delta^{\delta_1}_{\ \delta_3}\, , %
\ee
where we are entertained by -- but also apologize for --  the collision of \terminate{sample indices} $\delta_1, \delta_2$ with the overall Kronecker delta\index{Kronecker delta},
and further recall that $\dete{2\pi \ker}$ is the determinant of the $\ND$-by-$\ND$ matrix $(2\pi \ker)_{\delta_1\delta_2}$. 

\subsection{The Evidence for Criticality}\label{subsec:Occam-criticality} %
As we elaborated on in the last section, the evidence\index{Bayesian inference!evidence} is just the prior distribution\index{prior} for the network outputs evaluated on the observed true output values $\y{i}{\tra}$ given the inputs $\x{i}{\tra}$ in the subsample $\A$:
\begin{align}\label{eq:Gaussian-evidence-general}
p(y_{\A}\vert\Hypo)= \frac{1}{\sqrt{\dete{2\pi \kersub}^{n_{L}}}} \exp\!\le(-\frac{1}{2}\sum_{i=1}^{n_{L}}\sum_{\tra_1,\tra_2\in\A}\kersub^{\tra_1 \tra_2}\y{i}{\tra_1}\y{i}{\tra_2}\ri)\, .
\end{align}
Here we've put tildes both on the \terminate{sample indices} $\tra$ and on the kernel as well, $\kersub_{\tra_1\tra_2}$, %
in order to indicate that it's an
$\NR$-by-$\NR$ submatrix built from the pairs of inputs $(x_{\tra_1},x_{\tra_2})$ in the subsample $\A$ of size $\NR$. Importantly, this means that the inverse $\kersub^{\tra_1 \tra_2}$ is taken with respect to the samples in the set $\A$ \emph{only},
\be\label{eq:inverse-subA}
\sum_{\tra_2\in\A}\kersub^{\tra_1\tra_2}\kersub_{\tra_2\tra_3}=\delta^{\tra_1}_{\ \tra_3}\, ,
\ee
and in particular that $\kersub^{\tra_1 \tra_2}\ne\ker^{\tra_1 \tra_2}$. In other words, $\kersub^{\tra_1 \tra_2}$ is \emph{not} the same as the inverse of the kernel $\ker^{\delta_1\delta_2}$ on the whole dataset\index{input data} $\D$~\eqref{eq:whole-inverse} evaluated on the sample indices $\delta_{1}=\tra_{1}$ and $\delta_{2}=\tra_{2}$; if you'd like, flip ahead and cf.~\eqref{eq:Kinv-tratra}.
Accordingly, the determinant $\dete{2\pi \kersub}$ is also computed from this
$N_{\A}$-by-$N_{\A}$ submatrix.
The usefulness of this notation and the essentialness of this distinction will become clearer when we consider the posterior in \S\ref{subsec:absence-FF-Bayes}.

Before we analyze the evidence \eqref{eq:Gaussian-evidence-general} in detail, we should establish our space of hypotheses.
Considering MLP architectures in the infinite-width limit, there's only three hyperparameters of relevance, the bias variance and rescaled weight variance $C_b$ and $C_W$, and the depth $L$. In principle, each combination of these three hyperparameters is a different hypothesis. However, in the asymptotic limit of large depth $L \gg 1$, we know from our discussion in \S\ref{sec:criticality_DLN} and our analysis in \S\ref{ch:eft-mlp} that generically the kernel recursion will either exponentially lead to a \emph{trivial fixed point}\index{fixed point!trivial} at zero $K^\star=0$ or at infinity $K^\star =\infty$, or slowly approach a \emph{nontrivial fixed point}\index{fixed point!nontrivial} at \terminate{criticality}.\footnote{Yes, we know, for some activation functions there exist hyperparameter settings that lead to trivial fixed point\index{fixed point!trivial}s at nonzero values of the kernel $K^{\star}\ne0$. We'll eventually consider -- and make a case against -- such hypotheses as well, though only in a future footnote,~\ref{foot:parallel-criticality}, and only after first considering the details of the two-input evidence\index{Bayesian inference!evidence}.}
Thus, for deep networks Bayesian model comparison\index{Bayesian inference!model comparison} essentially reduces to the comparison of three different hypotheses, $\Hypo_{0}$, $\Hypo_{\infty}$ and $\Hypo_{\text{critical}}$, corresponding to the two trivial fixed points and the one nontrivial fixed point, respectively.

Having established our space of hypotheses, let's first see how Bayesian model comparison\index{Bayesian inference!model comparison} works when we have only a single input $x$.
In this case the kernel is just a scalar, and the evidence is simply given by
\begin{align}\label{eq:Gaussian-evidence-single}
p(y\vert\Hypo) = \frac{1}{\le(2\pi \kersub\ri)^{\frac{n_{L}}{2}}} \exp\!\le(-\frac{1}{2\kersub}\sum_{i=1}^{n_{L}}y_{i}^2\ri)\, .
\end{align}
Here, the output norm $\sum_{i=1}^{n_{L}}y_{i}^2$ is fixed by a given function approximation task.\footnote{
Many common datasets for \neo{classification} tasks employ ``one-hot\index{one-hot encoding}'' true outputs in which all but one component $y_i$ of a particular output are zero, and the remaining single component -- corresponding to the \emph{correct} class -- is equal to one. For such datasets, the output norm is trivial $\sum_{i=1}^{n_{L}}y_{i}^2 = 1$.
}
Thus all the dependence on the hyperparameters $\Hypo$ is encoded in a single number: $\kersub$.

Let's start with $\Hypo_\infty$, for which $\kersub\to\infty$. In this case, the argument of the exponential in~\eqref{eq:Gaussian-evidence-single} vanishes and thus the exponential evaluates to unity, while the normalization factor in front 
vanishes. Therefore, the evidence will vanish polynomially:%
\be\label{eq:exploding-kernel-evidence}
p(y\vert\Hypo_\infty) 
= \lim_{\kersub\to\infty} \frac{1}{\le(2\pi \kersub\ri)^{\frac{n_{L}}{2}}} = 0\, .
\ee
In fact, in this limit the output distribution becomes an (unnormalizable) uniform distribution over all possible output norms.
Next, let's consider $\Hypo_0$ with $\kersub\to0$. In this case, while the normalization factor grows polynomially, the argument in the exponent approaches negative infinity. Thus, the evidence approaches zero exponentially quickly:
\be
p(y\vert\Hypo_0) 
= \lim_{\kersub\to 0} \exp\!\le[ 
    -\frac{1}{2\kersub}\sum_{i=1}^{n_{L}}y_{i}^2 + \o{\log \kersub}
\ri]  = 0\, .
\ee
Indeed, recalling \eqref{eq:gaussian-limit-delta-function}, the evidence~\eqref{eq:Gaussian-evidence-single} in this limit becomes a \terminate{Dirac delta function},
\be\label{eq:vanishing-kernel-evidence}
p(y\vert\Hypo_0) 
= \prod_{i=1}^{n_L}\delta\!\le( y_{i} \ri) \, ,
\ee
which is a fairly useless hypothesis unless all of the true outputs are the zero vector.
Therefore, for generic nonzero and finite output values, the maximal evidence should lie between these two extrema. 
Specifically, seen as a function of $\kersub$, the evidence~\eqref{eq:Gaussian-evidence-single} peaks at
$\kersub=\kersub^{(L)}(x,x)\equiv \sum_{i=1}^{n_{L}}y_{i}^2/n_{L}$.
Our remaining hypothesis,
\terminate{criticality} $\Hypo_{\text{critical}}$, comes the closest to realizing this maximum.

To reiterate, for a single input we just need the kernel $\kersub$ to be of order one. For deep neural networks, this is precisely the condition that we imposed in order to avoid the \neo{exploding and vanishing kernel problem} for a single input, which we satisfied with the \terminate{parallel susceptibility} condition $\chi_{\parallel}\!\le(\Tif{\ker}{}{}\ri)=1$. Physically, the exploding kernel gives a very flat distribution spread over a big range of output norms, yielding insubstantial evidence for any particular output norm; the vanishing kernel gives sharp support for the zero norm \eqref{eq:vanishing-kernel-evidence} and no support anywhere else. Clearly the Bayes' factor\index{Bayesian inference!model comparison!Bayes' factor}~\eqref{eq:bayes-factor-equal-hypo} will prefer any hypothesis that gives more focused support over reasonable output norms.
In the language of our \terminate{Occam's razor} discussion, $\Hypo_{\infty}$ is too complex, predicting every possible norm, while $\Hypo_{0}$ is too simple, predicting only one particular norm.
The only hypothesis that gives a finite and nonzero $\kersub$ in the deep asymptotic regime is $\Hypo_{\text{critical}}$, whereat the \terminate{initialization hyperparameters} are tuned to satisfy $\chi_{\parallel}\!\le(\Tif{\ker}{}{}\ri)=1$.\footnote{N.B.~polynomially vanishing kernels give finite evidence\index{Bayesian inference!evidence} for all practical depths. To be very pedantic about this, for such kernels  -- for instance, for the $\tanhA$ -- for absurdly deep networks the truly Bayesian-optimal $C_W$ would be ever so slightly above its critical value.}

Now that we see how this works, let's  extend our analysis of the evidence\index{Bayesian inference!evidence} to two inputs, with $\tra=\pm$.
Intuitively, we expect to find the \terminate{perpendicular susceptibility} condition $\chi_{\perp}\!\le(\Tif{\ker}{}{}\ri)=1$ and thus demonstrate a conclusive preference for the criticality hypothesis $\Hypo_{\text{critical}}$.
To rediscover $\chi_{\perp}\!\le(\Tif{\ker}{}{}\ri)=1$, it will be sufficient to consider the case where both inputs have the same norm 
\be\label{eq:same-norm-dah}
\sum_{i=1}^{n_0}\x{i}{+}^2=\sum_{i=1}^{n_0}\x{i}{-}^2\, .
\ee
Then, recalling our\index{$\gamma^{[a]}$ basis!kernel}
decomposition into the $\gamma^{[a]}_{\tra_1\tra_2}$ basis~\eqref{eq:kernel-expand-gamma}, we can write the kernel as
\be
\kersub_{\tra_1\tra_2}=\begin{pmatrix}
\kersub_{[0]}+\kersub_{[2]} & \kersub_{[0]}-\kersub_{[2]}  \\
\kersub_{[0]}-\kersub_{[2]}   & \kersub_{[0]}+\kersub_{[2]}
\end{pmatrix}\, ,%
\ee
where we've used the fact that $\kersub_{[1]}=0$ when both inputs have the same norm~\eqref{eq:same-norm-dah}.

In this basis, the determinant is given by $\dete{2\pi \kersub}=16\pi^2\kersub_{[0]}\kersub_{[2]}$, and the inverse of the kernel is given by
\be
\kersub^{\tra_1\tra_2}=\frac{1}{4\kersub_{[0]}\kersub_{[2]}}\begin{pmatrix}
\kersub_{[0]}+\kersub_{[2]} & -\kersub_{[0]}+\kersub_{[2]}  \\
-\kersub_{[0]}+\kersub_{[2]}   & \kersub_{[0]}+\kersub_{[2]}
\end{pmatrix}\, ,%
\ee
which in turn lets us evaluate the argument of the exponential in~\eqref{eq:Gaussian-evidence-general} as
\begin{align}
\sum_{i=1}^{n_{L}}\sum_{\tra_1,\tra_2=\pm}\kersub^{\tra_1 \tra_2}\y{i}{\tra_1}\y{i}{\tra_2}\, 
=&\sum_{i=1}^{n_{L}}\frac{1}{4\kersub_{[0]}\kersub_{[2]}}\le[\kersub_{[2]}\le(\y{i}{+}+\y{i}{-}\ri)^2+\kersub_{[0]}\le(\y{i}{+}-\y{i}{-}\ri)^2\ri] \notag \\
=&\frac{\LLmax_{[0]}}{\kersub_{[0]}}+\frac{\LLmax_{[2]}}{\kersub_{[2]}}\, ,
\end{align}
where in the last equality we introduced the components\index{output matrix!$\gamma^{[a]}$ basis|see{$\gamma^{[a]}$ basis}}
\be\label{eq:output-matrix-decomposition}
\LLmax_{[0]}\equiv\sum_{i}^{n_L}\le(\frac{\y{i}{+}+\y{i}{-}}{2}\ri)^2\, , \qquad \LLmax_{[2]}\equiv\sum_{i}^{n_L}\le(\frac{\y{i}{+}-\y{i}{-}}{2}\ri)^2 \, .
\ee
All together, this gives a simple expression for the two-input evidence\index{Bayesian inference!evidence},
\begin{align}\label{eq:Gaussian-evidence-double}
p\!\le(y_{+},y_{-} \vert\Hypo\ri)=&\le(16\pi^2 \kersub_{[0]}\kersub_{[2]}\ri)^{-\frac{n_{L}}{2}} \exp\!\le(-\frac{\LLmax_{[0]}}{2\kersub_{[0]}}-\frac{\LLmax_{[2]}}{2\kersub_{[2]}}\ri)\, \\
=&\le[\le(4\pi \kersub_{[0]}\ri)^{-\frac{n_{L}}{2}} \exp\!\le(-\frac{\LLmax_{[0]}}{2\kersub_{[0]}}\ri)\ri]\times\le[\le(4\pi \kersub_{[2]}\ri)^{-\frac{n_{L}}{2}}  \exp\!\le(-\frac{\LLmax_{[2]}}{2\kersub_{[2]}}\ri)\ri]\, .\notag
\end{align}

Now, let's consider a generic pair of input-output pairs $(x_{+}, y_{+})$ and $(x_{-}, y_{-})$ for which both the average and the difference of the true outputs, $\LLmax_{[0]}$ and $\LLmax_{[2]}$~\eqref{eq:output-matrix-decomposition}, are nonzero and of order one.  Then, running the same argument as we did for the single-input evidence\index{Bayesian inference!evidence}, we prefer a hypothesis that comes as close as possible to having both $\kersub_{[0]}\approx\LLmax_{[0]}/n_{L}=\o{1}$ -- from maximizing the object in the first square brackets of~\eqref{eq:Gaussian-evidence-double} -- and  $\kersub_{[2]}\approx\LLmax_{[2]}/n_{L}=\o{1}$ -- from maximizing the object in the second square brackets of~\eqref{eq:Gaussian-evidence-double}. And, as we learned in~\S\ref{ch:eft-mlp}, to keep both $\kersub_{[0]}$ \emph{and} $\kersub_{[2]}$ of order one, we need to set both the critical \terminate{parallel susceptibility} condition $\chi_{\parallel}\!\le(\Tif{\ker}{}{}\ri)=1$ \emph{and}  the critical \terminate{perpendicular susceptibility} condition $\chi_{\perp}\!\le(\Tif{\ker}{}{}\ri)=1$\index{$\gamma^{[a]}$ basis!output matrix}.\footnote{Finally, let's consider the trivial fixed point\index{fixed point!trivial}s with nonzero kernel values $K^{\star}\ne0$.\label{foot:parallel-criticality} 
(This can occur, e.g., for the $K^\star=0$ universality class, for which there exists fixed points $K^\star$ that have $\chi_\perp(K^\star)=1$ but $\chi_\parallel(K^\star)<1$.)%
For this analysis, we need to relax the same-norm condition~\eqref{eq:same-norm-dah} and consider the most general form of the two-input kernel. Projecting the kernel into the $\gamma^{[a]}_{\tra_1\tra_2}$ basis~\eqref{eq:kernel-expand-gamma} as
\be
\kersub_{\tra_1\tra_2}=\begin{pmatrix}
\kersub_{[0]}+\kersub_{[1]}+\kersub_{[2]} & \kersub_{[0]}-\kersub_{[2]}  \\
\kersub_{[0]}-\kersub_{[2]}   & \kersub_{[0]}-\kersub_{[1]}+\kersub_{[2]}
\end{pmatrix}\, ,%
\ee
we can similarly use~\eqref{eq:trace-projection} to decompose the \neo{output matrix}, $\LLmax_{\tra_1\tra_2}\equiv \sum_{i=1}^{n_L} \y{i}{\tra_1}\y{i}{\tra_2}$, into
components
\be
\LLmax_{[0]}=\sum_{i}\le(\frac{\y{i}{+}+\y{i}{-}}{2}\ri)^2\, ,\quad \LLmax_{[1]}=\frac{\sum_{i}\y{i}{+}^2-\sum_{i}\y{i}{-}^2}{2}\, ,\quad \LLmax_{[2]}=\sum_{i}\le(\frac{\y{i}{+}-\y{i}{-}}{2}\ri)^2\, .
\ee
Then, a quick calculations shows that the evidence\index{Bayesian inference!evidence} evaluates to
\begin{align}\label{eq:Gaussian-evidence-double-more}
p\!\le(y_{+},y_{-} \vert\Hypo\ri)=&\le[4\pi^2 (4\kersub_{[0]}\kersub_{[2]}-\kersub_{[1]}^2)\ri]^{-\frac{n_{L}}{2}} \exp\!\le[-\frac{(4\kersub_{[0]}\LLmax_{[2]}+4\kersub_{[2]}\LLmax_{[0]}-2\kersub_{[1]}\LLmax_{[1]})}{2(4\kersub_{[0]}\kersub_{[2]}-\kersub_{[1]}^2)}\ri]\, .
\end{align}
Now, we see from this expression that a hypothesis with $\kersub_{[1]}\LLmax_{[1]}>0$ has improved evidence compared to the one with non-positive $\kersub_{[1]}\LLmax_{[1]}$. In particular, if a fixed point is trivial then the parallel perturbation $\kersub_{[1]}$ always vanishes exponentially, even if the fixed-point value of the kernel is non-vanishing $\ker^{\star}\ne0$. Thus, such a hypothesis will be disfavored compared to $\Hypo_{\text{critical}}$,  completing our argument.

It should be noted that for this distinction to matter, we must have a nonzero $\LLmax_{[1]}$, meaning $\sum_{i}\y{i}{+}^2\ne \sum_{i}\y{i}{-}^2$. For networks used as generic function approximators -- or for tasks where the network outputs are general and used downstream for other tasks -- this may matter. For deep-learning tasks where all the true outputs have the same norm, this may not matter.}
Therefore, with this \emph{evidence for criticality}, Bayesian model comparison\index{Bayesian inference!model comparison} demonstrates a full preference for $\Hypo_{\text{critical}}$.\footnote{Technically, what we've shown here is a preference for criticality in the Bayesian prior distribution. In~\S\ref{sec:EVGP-WEP}, we'll also find a natural preference for \terminate{criticality} in the \terminate{initialization distribution}, by showing that such a tuning is necessary for controlling the exploding and vanishing \emph{gradient} problem\index{exploding and vanishing gradient problem} that arises with gradient-based learning.} \\

\noindent{}\emph{Programming note}\index{programming note}: since conditioning on $\Hypo$ is so deeply ingrained in our minds by now, for notational simplicity we'll re-start the suppression of this conditioning from here on out.

\subsection{Let's Not Wire Together}\label{subsec:absence-FF-Bayes}
Now, let's work out the full posterior distribution\index{posterior}~\eqref{eq:BB-NN} at infinite width.\footnote{
    The form of this distribution was first worked out by Williams in \cite{williams-infinite} for one-hidden-layer networks.
} As we already have an expression for the evidence $p\!\le(y_{\A}\ri)$~\eqref{eq:Gaussian-evidence-general} in the denominator, let's focus on the joint distribution $p\!\le(y_{\A},z_{\B}^{(L)}\ri)$ in the numerator. Recall also that to discuss the posterior we need to partition the data into two subsamples, $\D\equiv\A\cup\B$, one for which we have observed the true output values $y_{\A}$ and the other for which we are going to infer the output values. 

With such a data partitioning in mind, we can write out the joint distribution as
\begin{align}\label{eq:joint-numerator}
&p\!\le(y_{\A},z_{\B}^{(L)}\ri)\, \\
=&\frac{1}{\sqrt{\dete{2\pi \ker}^{n_{L}}}}\exp\!\Bigg[-\frac{1}{2}\sum_{i=1}^{n_{L}}\Bigg(\sum_{\tra_1,\tra_2\in\A}\!\!\!\ker^{\tra_1 \tra_2}\y{i}{\tra_1}\y{i}{\tra_2}+\sum_{\tra_1\in\A,\tea_2\in\B}\!\!\!\!\!\ker^{\tra_1 \tea_2}\y{i}{\tra_1}\z{i}{\tea_2}{L}\, \notag\\
&\quad \quad \quad \quad \quad \quad \quad \quad \quad \quad \quad +\sum_{\tea_1\in\B, \tra_2\in\A}\!\!\!\!\!\ker^{\tea_1 \tra_2}\z{i}{\tea_1}{L}\y{i}{\tra_2}+\sum_{\tea_1,\tea_2\in\B}\!\!\!\ker^{\tea_1 \tea_2}\z{i}{\tea_1}{L}\z{i}{\tea_2}{L}\Bigg)\Bigg]\, ,\notag
\end{align}
where $\ker^{\tra_1 \tra_2}$, $\ker^{\tra_1 \tea_2}$, $\ker^{\tea_1 \tra_2}$, and $\ker^{\tea_1 \tea_2}$ are the blocks of %
\be\label{eq:kernel-inverse-joint-decompose}
\ker^{\delta_1\delta_2}\equiv\begin{pmatrix}
\ker^{\tra_1\tra_2} & \ker^{\tra_1\tea_2} \\
\ker^{\tea_1\tra_2}  &\ker^{\tea_1\tea_2} 
\end{pmatrix}\, ,
\ee
which is the inverse of the whole $\ND$-by-$\ND$ kernel matrix,
\be\label{eq:kernel-submatrix-decomposition}
\ker_{\delta_1\delta_2}
=\begin{pmatrix}
\kersub_{\tra_1\tra_2} & \ker_{\tra_1\tea_2} \\
\ker_{\tea_1\tra_2}  &\ker_{\tea_1\tea_2} 
\end{pmatrix}\, .
\ee
To make progress, we need to relate the submatrices in the inverse~\eqref{eq:kernel-inverse-joint-decompose} to the submatrices in the kernel decomposition \eqref{eq:kernel-submatrix-decomposition}, since, recalling 
\be
\ker_{\delta_1\delta_2} \equiv \frac{1}{n_L} \sum_i^{n_L} \E{z^{(L)}_i\!\le(x_{\delta_1}\ri) z^{(L)}_i\!\le(x_{\delta_2}\ri)} + \oninv \, ,
\ee
it's these blocks that are naturally defined in terms of the data.\footnote{
As we explained before, the over-tilde on $\kersub_{\tra_1\tra_2}$ indicates that it's a submatrix of the kernel evaluated on samples in the set $\A$, only. The inverse of that block was defined explicitly in~\eqref{eq:inverse-subA} and is symbolized as $\kersub^{\tra_1 \tra_2}$. Also note that the symmetry of the full kernel, $\ker_{\delta_1\delta_2} = \ker_{\delta_2\delta_1}$, endows a similar set of symmetries on the submatrices: $\kersub_{\tra_1\tra_2}=\kersub_{\tra_2\tra_1}$, $\ker_{\tea_1\tea_2} =\ker_{\tea_2\tea_1} $, and $\ker_{\tea\tra}=\ker_{\tra\tea}$. 
}

Explicitly inverting $\ker_{\delta_1\delta_2}$ according to the inverse formula~\eqref{eq:whole-inverse}, we find that the submatrices of \eqref{eq:kernel-inverse-joint-decompose} can be defined in terms of the blocks of the kernel~\eqref{eq:kernel-submatrix-decomposition} and the inverse submatrix $\kersub^{\tra_1 \tra_2}$ on $\A$ as
\begin{align}
\ker^{\tra_1\tra_2}&\equiv\kersub^{\tra_1\tra_2}+\sum_{\tra_3,\tra_4\in\A}\sum_{\tea_3,\tea_4\in\B} \kersub^{\tra_1\tra_3}\ker_{\tra_3\tea_3}\kerpos^{\tea_3\tea_4}\ker_{\tea_4\tra_4}\kersub^{\tra_4\tra_2}\, ,\label{eq:Kinv-tratra}\\
\ker^{\tra_1\tea_2}&\equiv-\sum_{\tra_3\in\A}\sum_{\tea_3\in\B}\kersub^{\tra_1\tra_3}\ker_{\tra_3\tea_3}\kerpos^{\tea_3\tea_2}\, ,\label{eq:Kinv-tratea}\\
\ker^{\tea_1\tra_2}&\equiv-\sum_{\tra_3\in\A}\sum_{\tea_3\in\B}\kerpos^{\tea_1\tea_3}\ker_{\tea_3\tra_3}\kersub^{\tra_3\tra_2}\, ,\label{eq:Kinv-teatra}\\
\ker^{\tea_1\tea_2}&\equiv\kerpos^{\tea_1\tea_2} \, ,\label{eq:Kinv-teatea}
\end{align}
where we've had to introduce (and name a posteori) the \emph{posterior covariance}\index{posterior!posterior covariance},
\be\label{eq:GP-posterior-variance}
\kerpos_{\tea_1\tea_2}\equiv \ker_{\tea_1\tea_2}-\sum_{\tra_3,\tra_4\in\A}\ker_{\tea_1\tra_3}\kersub^{\tra_3\tra_4}\ker_{\tra_4\tea_2}\, .
\ee
The expression for~\eqref{eq:Kinv-teatea} is defined implicitly by taking the inverse of \eqref{eq:GP-posterior-variance}:
\be\label{eq:inverse-posB}
\sum_{\tea_2\in\B}\kerpos^{\tea_1\tea_2}\, \kerpos_{\tea_2\tea_3}=\delta^{\tea_1}_{\ \tea_3}\, .
\ee

Since these are essential relations, let us check all the components of the inverse formula~\eqref{eq:whole-inverse}, one by one.
Firstly, considering the $\delta^{\delta_1}_{\ \delta_3} \to \delta^{\tra_1}_{\ \tra_3}$ component, we see
\begin{align}
\sum_{\delta_2\in\D}\ker^{\tra_1\delta_2}\ker_{\delta_2 \tra_3}&=\sum_{\tra_2\in\A}\ker^{\tra_1\tra_2}\ker_{\tra_2 \tra_3}+\sum_{\tea_2\in\B}\ker^{\tra_1\tea_2}\ker_{\tea_2 \tra_3}\, \notag \\
&=\sum_{\tra_2\in\A}\kersub^{\tra_1\tra_2}\kersub_{\tra_2 \tra_3}=\delta^{\tra_1}_{\ \tra_3}\, ,
\end{align}
where in the first line we decomposed the sum over $\delta_2 \in \D$ into separate sums over $\tra_2 \in \A$ and over $\tea_2 \in \B$ according to our partitioning $\D = \A \cup \B$, then in going to the second line we plugged in our expressions for the inverse blocks~\eqref{eq:Kinv-tratra} and~\eqref{eq:Kinv-tratea}, 
and finally in the last step we used the fact that $\kersub^{\tra_1\tra_2}$ is the inverse of the submatrix $\kersub_{\tra_1\tra_2}$~\eqref{eq:inverse-subA}.
Secondly, considering the $\delta^{\delta_1}_{\ \delta_3} \to \delta^{\tea_1}_{\ \tea_3}$ component, we see
\begin{align}
&\sum_{\delta_2\in\D}\ker^{\tea_1\delta_2}\ker_{\delta_2 \tea_3}=\sum_{\tra_2\in\A}\ker^{\tea_1\tra_2}\ker_{\tra_2 \tea_3}+\sum_{\tea_2\in\B}\ker^{\tea_1\tea_2}\ker_{\tea_2 \tea_3}\, \\
=&\sum_{\tea_2\in\B}\kerpos^{\tea_1\tea_2}\le(\ker_{\tea_2 \tea_3}-\sum_{\tra_3,\tra_2\in\A} \ker_{\tea_2\tra_3}\kersub^{\tra_3\tra_2}\ker_{\tra_2\tea_3}\ri)=\sum_{\tea_2\in\B}\kerpos^{\tea_1\tea_2}\kerpos_{\tea_2\tea_3}=\delta^{\tea_1}_{\ \tea_3}\, ,\notag
\end{align}
where as  before in the first line we decomposed the sum over $\delta_2 \in \D$ into separate sums over $\tra_2 \in \A$ and over $\tea_2 \in \B$ according to our partitioning $\D = \A \cup \B$, then in going to the second line we plugged in our expressions for the inverse blocks~\eqref{eq:Kinv-teatra} and~\eqref{eq:Kinv-teatea}, and finally, identifying the expression in the parenthesis as the definition of the posterior covariance\index{posterior!posterior covariance} $\kerpos_{\tea_1\tea_2}$~\eqref{eq:GP-posterior-variance}, we get the final result since $\kerpos^{\tea_1\tea_2}$ is the inverse of the posterior covariance~\eqref{eq:inverse-posB}.
Lastly, we consider the off-diagonal block:
\begin{align}
&\sum_{\delta_2\in\D}\ker^{\tra_1\delta_2}\ker_{\delta_2 \tea_3}=\sum_{\tra_2\in\A}\ker^{\tra_1\tra_2}\ker_{\tra_2 \tea_3}+\sum_{\tea_2\in\B}\ker^{\tra_1\tea_2}\ker_{\tea_2 \tea_3}\, \\
=&\sum_{\tra_2\in\A,\tea_2\in\B}\!\!\!\!\!\kersub^{\tra_1\tra_2}\ker_{\tra_2\tea_2}\le(\delta^{\tea_2}_{\ \tea_3}+\sum_{\tra_3,\tra_4\in\A}\sum_{\tea_4\in\B}\kerpos^{\tea_2\tea_4}\ker_{\tea_4\tra_4}\kersub^{\tra_4\tra_3}\ker_{\tra_3\tea_3}-\sum_{\tea_4\in\B}\kerpos^{\tea_2\tea_4}\ker_{\tea_4\tea_3}\ri)\, \notag\\
=&\sum_{\tra_2\in\A,\tea_2\in\B}\!\!\!\!\!\kersub^{\tra_1\tra_2}\ker_{\tra_2\tea_2}\le(\delta^{\tea_2}_{\ \tea_3}-\sum_{\tea_4\in\B}\kerpos^{\tea_2\tea_4}\kerpos_{\tea_4\tea_3}\ri)=0\, ,\notag
\end{align}
Here,  we follow the same pattern as before, \emph{(i)} decomposing the sum according to the partitioning $\D = \A \cup \B$, \emph{(ii)} plugging in expressions for inverse blocks \eqref{eq:Kinv-tratra} and \eqref{eq:Kinv-tratea}, and \emph{(iii)} using the posterior covariance\index{posterior!posterior covariance} \eqref{eq:GP-posterior-variance} and the inverse equation \eqref{eq:inverse-posB}. Everything checks out.

Now that we have some confidence in our inversions, let's plug our expressions for these submatrices~\eqref{eq:Kinv-tratra}--\eqref{eq:Kinv-teatea} into the joint prior~\eqref{eq:joint-numerator}. 
Since the posterior~\eqref{eq:BB-NN} is only a function of the outputs $z_{\B}^{(L)}$,
we can make things easier by limiting our focus to the $z_{\B}^{(L)}$ dependence only, ignoring the $y_{\A}$ terms independent of $z_{\B}^{(L)}$  and ignoring the normalization factor:
\begin{align}
&p\!\le(y_{\A},z_{\B}^{(L)}\ri)\propto \exp\Bigg[-\frac{1}{2}\sum_{i=1}^{n_{L}}\sum_{\tea_1,\tea_2\in\B}\kerpos^{\tea_1\tea_2}\z{i}{\tea_1}{L}\z{i}{\tea_2}{L}\, \\
&\quad \quad \quad \quad \quad \quad \quad \quad \quad \ +\sum_{i=1}^{n_{L}}\sum_{\tea_1\in\B,\tra_1\in\A}\z{i}{\tea_1}{L}\le(\sum_{\tra_2\in\A, \tea_2\in\B}\kerpos^{\tea_1 \tea_2}\ker_{\tea_2\tra_2}\kersub^{\tra_2\tra_1}\ri)\y{i}{\tra_1}\Bigg]\, .\notag
\end{align}
At this point you know what to do: completing the square\index{complete the square} -- as should be your second nature by now -- and ignoring the new $z_{\B}^{(L)}$-independent additive constant in the exponential, you get
\begin{align}\label{eq:mid-point-in-infinite-posterior}
p\!\le(y_{\A},z_{\B}^{(L)}\ri)\propto& \exp\Bigg[-\frac{1}{2}\sum_{i=1}^{n_{L}}\sum_{\tea_1,\tea_2\in\B}\kerpos^{\tea_1\tea_2}\Bigg(\z{i}{\tea_1}{L}-\sum_{\tra_3,\tra_4\in\A}\ker_{\tea_1\tra_3}\kersub^{\tra_3\tra_4}\y{i}{\tra_4}\Bigg)\, \\
&\quad \quad \quad \quad \quad \quad \quad \quad \quad \quad\  \times\Bigg(\z{i}{\tea_2}{L}-\sum_{\tra_5,\tra_6\in\A}\ker_{\tea_2\tra_5}\kersub^{\tra_5\tra_6}\y{i}{\tra_6}\Bigg)\Bigg]\, .\notag
\end{align}
This distribution \eqref{eq:mid-point-in-infinite-posterior} is still Gaussian, with a variance given by the posterior covariance\index{posterior!posterior covariance} $\kerpos_{\tea_1\tea_2}$ and a \emph{nonzero} posterior mean\index{posterior!posterior mean}:
\be\label{eq:GP-mean}
\posGPmean_{i;\tea}\equiv \sum_{\tra_1,\tra_2\in\A} \ker_{\tea \tra_1}\kersub^{\tra_1\tra_2}\y{i}{\tra_2}\, .
\ee
Here, the
superscript $\infty$ is used to remind us that we're in the infinite-width
limit.
Finally, we realize that the posterior distribution\index{posterior}~\eqref{eq:BB-NN} is proportional to the joint prior~\eqref{eq:mid-point-in-infinite-posterior},
\be 
p\!\le(z_{\B}^{(L)}\Big\vert y_{\A}\ri)\propto p\!\le(y_{\A},z_{\B}^{(L)}\ri) \, ,
\ee
and that the posterior distribution is automatically normalized~\eqref{eq:Bayes-consistency} as a function of the variable $z_{\B}^{(L)}$.
Thus, computing the normalization factor for \eqref{eq:mid-point-in-infinite-posterior} -- or really just writing it down, since at this point you know by heart how to normalize any Gaussian distribution -- we get the posterior at infinite width\index{posterior!infinite-width distribution}:
\begin{align}\label{eq:posterior-at-infinite-width}
&p\!\le(z_{\B}^{(L)}\Big\vert y_{\A}\ri)= \frac{1}{\sqrt{\dete{2\pi \kerpos}^{n_{L}}}} \exp\!\le[-\frac{1}{2}\sum_{i=1}^{n_{L}}\sum_{\tea_1,\tea_2\in\B}\kerpos^{\tea_1 \tea_2}\le(\z{i}{\tea_1}{L}-\posGPmean_{i;\tea_1}\ri)\le(\z{i}{\tea_2}{L}-\posGPmean_{i;\tea_2}\ri)\ri]\, .
\end{align}

The \emph{posterior mean}\index{posterior!posterior mean} $\posGPmean_{i;\tea}$ represents our updated belief about the expected network output for the input $\x{j}{\tea}\in\B$ after incorporating information about the true outputs $y_{\A}$ for all the inputs $\x{j}{\tra}\in\A$; as such, it is explicitly a function of the true input-output pairs $x_{\A}$ and $y_{\A}$ in the subsample $\A$, as we see in~\eqref{eq:GP-mean}. Importantly, our expected predictions were a priori zero -- indicating an \terminate{inductive bias} towards vanishing outputs on average -- and now a posteriori our predictions are shifted to something nonzero. Such a nonzero posterior mean\index{posterior!posterior mean} is a signature that learning is (finally!) happening.
In addition, the posterior covariance\index{posterior!posterior covariance} $\kerpos_{\tea_1\tea_2}$ encodes the confidence interval: the smaller the covariance is, the more sharply peaked the posterior is around its mean, and the more confident the model is about its predictions.

Practically speaking, note that in order to compute the mean prediction $\posGPmean_{i;\tea_1}$ according to its definition~\eqref{eq:GP-mean}, we'd in principle need to invert -- and then represent -- the $\NR$-by-$\NR$ submatrix $\kersub_{\tra_1\tra_2}$. As the size of our observations $\NR$ grows, the computational cost of such an inversion grows very fast.\footnote{For instance,~the computational cost of \neo{Gauss-Jordan elimination} scales as $\sim \NR^3$ and requires us to represent the $\NR \times \NR$-dimensional inverse in memory. Things can be improved a bit by realizing that to compute the posterior mean we only really require the \neo{matrix-vector product} of the inverse with the observations: $\sum_{\tra_2 \in \A} \kersub^{\tra_1\tra_2} y_{i;\tra_2}$. However, such an improvement is still not really sufficient for Bayesian learning to compete practically with gradient-based learning for large datasets $\A$.} This hidden catch is why -- though theoretically quite elegant -- (at least any naive implementation of) Bayesian learning is not practical for large datasets. Instead, for this reason we will essentially need to rely on approximation methods for model fitting, such as MLE \eqref{eq:mle-estimate}\index{maximum likelihood estimation}.\index{Bayesian inference!practicalities}
We'll comment more on this next chapter (\S\ref{ch:training}).

Theoretically \emph{and} practically speaking, there is another serious issue with the infinite-width posterior mean. Looking at its expression~\eqref{eq:GP-mean},
we see that the mean prediction on the output component $i$ is entirely independent from the observations $\y{j}{\alpha}$ that we made on the other components with $j\ne i$. Thus, our updated best estimate of these different output components are entirely uncorrelated, though in principle observations of different components $j$ may contain very useful information about a given component $i$.\footnote{\label{foot:distillation}The concept of \neo{knowledge distillation}\index{distillation|see{knowledge distillation}} \cite{hinton2015distilling} is predicated on this principle of correlations among the output components. For example, if a network is trying to classify\index{classification} images of hand-written digits, a certain example of a ``$2$'' may be more ``$7$''-like or more ``$3$''-like. Such feature information is quite useful, especially if the output of the network is used downstream for some other task.
}
In fact, we see from \eqref{eq:posterior-at-infinite-width} that the posterior distribution actually factorizes as
\be\label{eq:stat-independence-as-bad-posterior}
p\!\le(z_{i; \B}^{(L)}, z_{j; \B}^{(L)} \Big\vert y_{i;\A}, y_{j;\A}\ri)=p\!\le(z_{i; \B}^{(L)} \Big\vert y_{i;\A}\ri)\, p\!\le(z_{j; \B}^{(L)} \Big\vert y_{j;\A}\ri)\, , \qquad (i \neq j) \, ,
\ee
meaning that the different output components are entirely statistically independent.\index{statistical independence}\footnote{To be FAIR\index{FAIR|see{Facebook AI Research}}\index{Facebook AI Research}, the issue is with the \terminate{infinite-width limit} itself, as different output components are also decorrelated for infinite-width networks trained with gradient-based learning (\S\ref{ch:NTHb}).
}

We can trace this independence back to a similar property of the infinite-width prior distribution
\be\label{eq:stat-independence-as-bad-prior}
p\!\le(z_{i;\A}^{(L)},z_{j;\A}^{(L)} \ri)=p\!\le(z_{i;\A}^{(L)} \ri)p\!\le(z_{j;\A}^{(L)} \ri)\, , \qquad (i\neq j),
\ee
a property that we've recognized for a while now, see e.g.~\eqref{eq:infinite-distribution-factorization}.
Thus, with Bayesian learning output features do not \emph{wire together}: recalling our discussion of \terminate{inductive bias} before (\S\ref{subsec:bayesian-model-comparison}), we see that the prior endows on the posterior an absurdly stubborn set of beliefs, namely that the components of the output are completely independent with \neo{absolute certainty}. Such an \terminate{inductive bias} is incurable by any amount of learning, irregardless of how large the set of observations $\A$ are; the inductive bias of this prior can never be overwhelmed in the infinite width limit.

Luckily, this state of affiars is completely curable --  for both \terminate{learning algorithm}s, Bayesian learning and gradient-based learning -- by backing off of the \terminate{infinite-width limit} and working with finite-width networks \ldots the actual kind of networks that are used in practice.

\subsection{Absence of Representation Learning}\label{subsec:absence-RL-Bayes}
Considering the independence of the different components of the output in the posterior, a natural follow-up question is whether or not Bayesian learning at infinite width enables \terminate{representation learning}. Here, we will show decisively that it does \emph{not}.

As a representative avatar of this question, let's compute the \emph{posterior} distribution of preactivations in the penultimate layer $\ell=L-1$ on the full set of samples $\D$, given observations $y_\A$:
\be\label{eq:Bayes-posterior-general}
p\!\le(z_{\D}^{(L-1)}\Big\vert y_\A \ri)=\frac{p\!\le(y_\A \Big\vert z_{\D}^{(L-1)}\ri)p\!\le(z_{\D}^{(L-1)} \ri)}{p\!\le(y_\A \ri)}\, .
\ee
This is an application of Bayes' rule\index{Bayesian probability!Bayes' rule}~\eqref{eq:Bayes-rule}, following from applying the product rule~\eqref{eq:logical-product} to the joint distribution $p\!\le(y_\A, z_{\D}^{(L-1)} \ri)$
between the observations $y_\A$ and the penultimate preactivations $z_{\D}^{(L-1)}$.
Here, the likelihood\index{Bayesian inference!likelihood} $p\!\le(y_\A \Big\vert z_{\D}^{(L-1)} \ri)$ is the conditional distribution $p\!\le(z_\A^{(L)} \Big\vert z_{\D}^{(L-1)} \ri)$ evaluated on our set of observations $z_{\A}^{(L)} \to y_\A$. 

We already know the form of this conditional distribution, as it is the same object~\eqref{eq:general-layer-conditional} that we needed in order to work out the layer-to-layer RG flow\index{representation group flow} of the preactivations. In general, this distribution involves the \emph{stochastic} metric  $\Ti{\widehat{G}}{\tra_1 \tra_2}{L}= \Ti{\widehat{G}}{\tra_1 \tra_2}{L}\!\le( z_{\D}^{(L-1)}\ri)$. However, in the \terminate{infinite-width limit} the metric  is entirely \emph{deterministic} $\Ti{\widehat{G}}{\tra_1 \tra_2}{L} \to \Ti{G}{\tra_1 \tra_2}{L}$, with no dependence at all on the penultimate-layer preactivations $z_{\D}^{(L-1)}$.
Thus, the likelihood\index{Bayesian inference!likelihood} at infinite width -- swapping the deterministic metric for the kernel -- is given by
\be
p\!\le(y_\A \Big\vert z_{\D}^{(L-1)} \ri)= \frac{1}{\sqrt{\dete{2\pi \kersub^{(L)}}^{n_{L}}}} \exp\!\le(-\frac{1}{2}\sum_{i=1}^{n_{L}}\sum_{\tra_1,\tra_2\in\A}\TI{\kersub}{\tra_1 \tra_2}{L}\y{i}{\tra_1}\y{i}{\tra_2}\ri)=p\!\le(y_\A  \ri)\, ,
\ee
and our expression for the posterior of the penultimate layer~\eqref{eq:Bayes-posterior-general} reduces to the prior:
\be\label{eq:infinite-width-penultimate-posterior-equals-prior}
p\!\le(z_{\D}^{(L-1)}\Big\vert y_\A \ri)=p\!\le(z_{\D}^{(L-1)}\ri)\, .
\ee
Since the posterior equals the prior, our observation of $y_\A$ had no consequence on the penultimate-layer \terminate{representation}; thus, we conclude that there is no \terminate{representation learning} at infinite width. 

This lack of \terminate{representation learning} stems from the lack of  interlayer correlation in the joint distribution $p\!\le(z_{\D}^{(\ell)}, z_{\D}^{(\ell+1)} \ri)$ at infinite width, and thus it persists for all hidden layers with $\ell<L$. This is another bad \terminate{inductive bias} of the infinite-width hypotheses: regardless of the set of observations $y_\A$ that we make, there's no amount of new information that will allow the network to update its \terminate{representation}s in the hidden layers $\ell<L$. 

\index{wiring!in Bayesian inference|see{Bayesian inference}}
This state of affairs is somewhat tragic as the whole point of having many layers -- in fact, the main motivation given for \terminate{deep learning} on the whole -- is the learning of complex representations in those hidden layers.
As we will see next, we can solve this lack of \terminate{representation learning} -- as well as the lack of \emph{wiring together}\index{Bayesian inference!wiring!infinite width} in the output -- by backing off the \terminate{infinite-width limit} and looking at finite-width effects.\footnote{
    In \S\ref{ch:NTHb}, will also show the same lack of \terminate{representation learning} occurs for the ensemble of infinite-width networks that are (theoretically) trained with gradient-based learning. This issue is also resolved (practically) in \S\ref{ch:features} by going to finite width.
}

\section{Bayesian Inference at Finite Width}\label{sec:finite-posterior}
In this section, we'll give three lessons on Bayesian learning at finite width. 
To begin, we'll show that finite-width neural networks are automatically endowed with an \terminate{inductive bias} for \terminate{neural association} due to non-Gaussian \terminate{interactions} between neurons,
leading to a natural predisposition towards \terminate{Hebbian learning} (\S\ref{subsec:Hebbian}).
With that in mind, we'll in turn demonstrate how such learning works by first
calculating the mean of the posterior distribution for the network outputs  $p\!\le(z_{\B}^{(L)}\Big\vert y_{\A} \ri)$ -- showing how  \emph{intralayer} neural interactions in the prior give rise to nontrivial correlations among the components of the output (\S\ref{subsec:presence-FF-Bayes}) -- and then
calculating the posterior distribution\index{posterior} of preactivations in the penultimate layer $p\!\le(z_{\B}^{(L-1)}\Big\vert y_{\A}\ri)$ 
-- showing how \emph{interlayer} interactions
give rise to a nonzero shift between prior and posterior, thus signaling the presence of \terminate{representation learning} at finite width (\S\ref{subsec:presence-RL-Bayes}).

\subsection{Hebbian Learning, Inc.}\label{subsec:Hebbian}
In this subsection, we'll see that finite-width neural networks have an \terminate{inductive bias} that facilitates \neo{neural association}.
To explain \term{Hebbian learning}, let's begin first with a few words from our
honorary
guest speaker, Donald Hebb:

\epigraph{The general idea is an old one, that any two cells or systems of cells that are repeatedly active at the same time will tend to become ``associated,'' so that activity in one facilitates activity in the other.}{Donald Hebb\index{Hebb, Donald}, in his 1949 classic \emph{The Organization of Behavior} \cite{hebb2005organization}.}

\noindent{}(\emph{Applause}\index{applause}.) 

\epigraph{Thank you very much.}{Donald Hebb, apocryphal.}

\noindent{}While Hebb was originally thinking about biological neurons\index{biological neuron}, Hebbian learning has become a popular guiding principle for systems of artificial neurons\index{artificial neuron} as well. We've actually already seen this \terminate{inductive bias} for \terminate{neural association} any of the numerous times we've discussed the presence of neural interactions in the finite-width prior distribution. 
To make this manifest, we're now going to explicitly determine the neural influence of one preactivation on another in our effective preactivation distribution at initialization.

Concretely, let's suppose that a single input $x$ is fed into a network, and we've 
checked
that at layer $\ell$ the value of the first preactivation $z_1^{(\ell)}  = \check{z}_1^{(\ell)}$ is larger than typical; given this atypical value $\check{z}_1^{(\ell)}$, we can then ask whether the second preactivation $z_2^{(\ell)}$ is likely to be atypically large. This kind of \terminate{neural association} or influence is encoded in the conditional distribution
\be\label{eq:condition-from-one-to-two}
p\Big(z_2^{(\ell)}\Big\vert \check{z}_1^{(\ell)}\Big)=\frac{p\Big(\check{z}_1^{(\ell)},z_2^{(\ell)}\Big)}{p\Big(\check{z}_1^{(\ell)}\Big)}\, .
\ee
Note that at infinite width $p\Big(z_2^{(\ell)}\Big\vert \check{z}_1^{(\ell)}\Big)=p\Big(z_2^{(\ell)}\Big)$ due to the factorization of the prior on neurons~\eqref{eq:infinite-distribution-factorization},
 and so we see right away that there is a complete absence of neural association in such a limit.

To compute this association for finite-width networks, recall from~\S\ref{sec:sum-rule} the action representation~\eqref{eq:m-neuron-action} for a distribution over $m$ neurons
\be\label{eq:m-neuron-action-reprint-bayes}
p(z_1,\ldots, z_{m})\propto\exp\!\le(-\frac{g_{m}}{2}\sum_{i=1}^{m}z_{i}^2+\frac{v}{8}\sum_{i,j=1}^{m}z_{i}^2z_{j}^2\ri)\, ,
\ee
where we have temporarily dropped \terminate{layer indices} from the variables and couplings.
Here, the quadratic coupling\index{coupling!quadratic} $g_{m}$ is given implicitly by the expression~\eqref{eq:quadratic-reprint-m-emphasis},
\be\label{eq:quadratic-reprint-m-emphasis-reprinted}
\frac{1}{g_m}=G^{(\ell)}-\frac{(m+2)}{2n_{\ell-1}}\frac{ V^{(\ell)}}{G^{(\ell)}}+\o{\frac{1}{n^2}}\, ,
\ee
and we have emphasized the dependence of the coupling on $m$; similarly, the quartic coupling\index{coupling!quartic} is given by~\eqref{eq:quartic-single-input-coupling-for-vertex},
\be\label{eq:quartic-reprint-m-emphasis}
v=\frac{1}{n_{\ell-1}} \frac{V^{(\ell)}}{\le(G^{(\ell)}\ri)^4}+\o{\frac{1}{n^2}}\, ,
\ee 
which is independent of $m$ to this order in $1/n$.
Evaluating the action representation~\eqref{eq:m-neuron-action-reprint-bayes} on $m=1$ and $m=2$ neurons and plugging the resulting distributions into our expression for the conditional distribution~\eqref{eq:condition-from-one-to-two}, we get 
\begin{align}\label{eq:conditional-one-two-neuron}
p(z_2\vert \check{z}_1)&\propto \exp\!\le[-\frac{g_2}{2}z_2^2+\frac{v}{8}\le(z_2^4+2z_2^2\check{z}_1^2\ri)\ri] \, ,
\end{align}
where, similar to the last section, for such a conditional distribution we only need to keep track of the terms in the action that depend on $z_2$.

Now that we have a conditional distribution, let's evaluate some conditional expectations. Since this distribution is manifestly even in $z_2$, i.e.~invariant under a sign flip $z_2\leftrightarrow-z_2$, all the odd-point correlators vanish, including the conditional mean. This means that the first nontrivial observable is the two-point correlator or conditional variance:
\begin{align}
\int dz_2\ p(z_2\vert \check{z}_1) \, z_2^2=&\frac{\int dz_2\ \exp\!\le[-\frac{g_2}{2}z_2^2+\frac{v}{8}\le(z_2^4+2z_2^2\check{z}_1^2\ri)\ri]z_2^2}{\int dz_2\ \exp\!\le[-\frac{g_2}{2}z_2^2+\frac{v}{8}\le(z_2^4+2z_2^2\check{z}_1^2\ri)\ri]}\, \label{eq:Hebbian-calc-middle}\\
=&\frac{\int dz_2\ e^{-\frac{g_2 z_2^2}{2}}\le[z_2^2+\frac{v}{8}\le(z_2^6+2z_2^4\check{z}_1^2\ri)+\o{v^2}\ri]}{\int dz_2\ e^{-\frac{g_2 z_2^2}{2}}\le[1+\frac{v}{8}\le(z_2^4+2z_2^2\check{z}_1^2\ri)+\o{v^2}\ri]}\, \notag\\
=&\frac{g_2^{-1}+\frac{v}{8}\le(15g_2^{-3}+6g_2^{-2}\check{z}_1^2\ri)}{1+\frac{v}{8}\le(3 g_2^{-2}+2g_2^{-1} \check{z}_1^2\ri)}+\o{v^2}\, \notag\\
=&g_2^{-1}+\frac{v}{2}g_2^{-2}\le(3 g_2^{-1}+\check{z}_1^2\ri)+\o{v^2}\, .\notag
\end{align}
Above, on the first line we used \eqref{eq:conditional-one-two-neuron} in the numerator and at the same time computed its normalization in the denominator,
on the second line we expanded both the numerator and denominator in $v$, on the third line we  
computed the single-variable Gaussian integrals, and on the final line we expanded the denominator in $v$.
Plugging in our expressions for the quadratic coupling\index{coupling!quadratic}~\eqref{eq:quadratic-reprint-m-emphasis-reprinted} and the quartic coupling\index{coupling!quartic}~\eqref{eq:quartic-reprint-m-emphasis} and reimplementing \terminate{layer indices}, we find
\be\label{eq:conditional-variance}
\int dz_2^{(\ell)}\ p\!\le(z_2^{(\ell)}\Big\vert \check{z}_1^{(\ell)}\ri) \le(z_2^{(\ell)}\ri)^2=G^{(\ell)}+\frac{1}{2}\le[\le(\check{z}_1^{(\ell)}\ri)^2-G^{(\ell)}\ri]\le[\frac{ V^{(\ell)}}{n_{\ell-1} \le(G^{(\ell)}\ri)^2}\ri]+\o{\frac{1}{n^2}}\, .
\ee
In passing, note for later that
this result holds for any distinct pair of neurons by replacing \terminate{neural indices} as $1,2\to i_1,i_2$, with $i_1\ne i_2$.

This conditional variance \eqref{eq:conditional-variance} embodies some really interesting \terminate{physics}. 
If the observed value $\le(\check{z}_1^{(\ell)}\ri)^2$ is larger/smaller than its expected value $\E{\le(z_1^{(\ell)}\ri)^2}=G^{(\ell)}$, then
the variance of $z_2^{(\ell)}$ will itself be larger/smaller than is typical. Thus, $z_1^{(\ell)}$ and $z_2^{(\ell)}$ correlate their atypical firing.\footnote{You may or may not recall from footnote~\ref{footnote-kurtosis} in \S\ref{sec:not-Gauss} that having a nontrivial connected four-point correlator\index{connected correlator!four-point} serves as a measure of the potential for outliers. In statistics\index{statistics (branch of mathematics)}, for single-variable distributions this is called the \emph{excess kurtosis}\index{kurtosis, excess}; here, we see a multi-neuron generalization (which apparently can be called the \emph{cokurtosis}\index{cokurtosis|see{kurtosis, excess}}\index{kurtosis, excess!cokurtosis}). In particular, observing an outlying value $z_1^{(\ell)} = \check{z}_1^{(\ell)}$ implies that we are more likely to see outlying values for $z_2^{(\ell)}$ as well.
At the end of Appendix \ref{app:mi-stuff}, we'll provide an information-theoretic reformulation of this phenomenon that will also shed further light on how deep a network should be in order to best take advantage of it.
} This effect is proportional to the normalized \terminate{four-point vertex} in the second square brackets of \eqref{eq:conditional-variance}, which  as we know from \eqref{eq:k-star-equals-zero-normalized-four-point-scaling-law} and~\eqref{eq:vertex-scaling-law} is proportional to $\ell/n$ across our universality classes when at \terminate{criticality}.
In other words, deeper layers have an \terminate{inductive bias} to build more \terminate{neural association}s. Moreover, the presence of these associations is mediated by the interaction\index{interactions}s in the effective action induced at finite width \emph{only}. As we will soon show, nontrivial \terminate{representation learning} is a direct descendant of such associations.

Note that this result should be interpreted as a \emph{propensity} for atypicality rather than a \emph{guarantee}. Since the conditional variance~\eqref{eq:conditional-variance} applies to any pair of neurons, conditioned on a particular neuron $i_*$ having a larger/smaller norm than expected, then all of the other neurons with $i \neq i_*$ are more likely to have a larger/smaller norm, though not all will. In a given realization of a network in practice, the ones that happen to have a larger/smaller norm are the ones that are more likely to develop a correlation with $i_*$ as learning progresses.

\index{wiring|seealso{Hebbian learning}}
\terminate{Hebbian learning} is often summarized by the following slogan: \emph{neurons that fire together, wire together}. What we see here is that conditioned on an atypical firing $\check{z}_1$, another preactivation, e.g.~$z_2$, is much more likely to have an atypical firing itself. This propensity of finite-width networks to \emph{fire together} is an \terminate{inductive bias} of our \terminate{prior} beliefs before Bayesian learning as well as of our \terminate{initialization distribution} before gradient-based learning.
To understand the \emph{wire together} part, let's now consider the Bayesian \terminate{posterior}.\footnote{
For some models of \terminate{artificial neuron}s -- such as the \terminate{Hopfield network} -- \terminate{Hebbian learning} is often added in \emph{by hand}. For instance,  one learning rule for such networks that explicitly implements the Hebbian principle is updating the weights connecting two neurons $i$ and $j$ as $W_{ij} \propto z_i(x) z_j(x)$ when observing activities $z_i(x)$ and $z_j(x)$ for a given input $x$. 

In contrast, any finite-width feedforward neural network\index{feedforward network} 
should automatically incorporate \terminate{Hebbian learning} \emph{by nature}.
To underscore this point further, in~\S\ref{ch:eot} we'll perform an analogous computation for a gradient-descent update.\index{gradient descent} Since the prior has the same form as the initialization distribution, we expect that all learned finite-width networks will 
inc.~the \terminate{Hebbian learning} principle automatically, regardless of whether that learning is Bayesian or gradient-based.}

\subsection{Let's Wire Together}\label{subsec:presence-FF-Bayes}\index{Bayesian inference!wiring!finite width}
Let's start with some more reminiscing 
through our now well-adjusted Bayesian lens. Recall from~\eqref{eq:general-ell-action} that the prior distribution\index{prior} over peractivations is nearly-Gaussian\index{nearly-Gaussian distribution} at large-but-finite width:
\begin{align}\label{eq:general-ell-action-reprint}
p\!\le(z^{(L)}_{\D}\ri)
&\propto\exp\!\Bigg[-\frac{1}{2}\sum_{j=1}^{n_{L}}\sum_{\delta_1,\delta_2\in\D} g^{\delta_1\delta_2} \z{j}{\delta_1}{L}\z{j}{\delta_2}{L}\, \\
&\quad \quad \quad \ \ +\frac{1}{8}\sum_{j,k=1}^{n_{L}}\sum_{\delta_1,\ldots,\delta_4\in\D}v^{(\delta_1\delta_2)(\delta_3\delta_4)} \z{j}{\delta_1}{L}\z{j}{\delta_2}{L}\, \z{k}{\delta_3}{L}\z{k}{\delta_4}{L}+\ldots\Bigg]\, .\nonumber%
\end{align}
As a reminder, the quadratic coupling\index{coupling!quadratic} $g^{\delta_1\delta_2}\equiv g^{\delta_1\delta_2}_{(L)}$ \eqref{eq:two-point-match-general} and the quartic coupling\index{coupling!quartic}  $v^{(\delta_1\delta_2)(\delta_3\delta_4)}\equiv v^{(\delta_1\delta_2)(\delta_3\delta_4)}_{(L)}$ \eqref{eq:four-point-match-general} 
depend explicitly on groups of inputs
from the dataset\index{input data} $\D$ and implicitly on the $\Hypo$yperparameters $C_b$ and $C_W$, the widths $n_{\ell}$, and the depth $L$.
As a consequence of the nonzero \emph{intralayer} interaction between different output preactivations in the prior, there will be non-vanishing correlations between the components of the network outputs in the posterior.

As we did at infinite width, we'll start with the prior distribution~\eqref{eq:general-ell-action-reprint} and then obtain the posterior distribution $p\!\le(z_{\B}^{(L)}\Big\vert y_{\A}\ri)\propto p\!\le(y_{\A},z_{\B}^{(L)}\ri)$ by plugging in our observations $\z{i}{\tra}{L}\to \y{i}{\tra}$ and keeping track of the dependence on the remaining variables $\z{i}{\tea}{L}$. For the quadratic term in the action, with exactly the same set of manipulations as we did in the infinite-width limit (\S\ref{subsec:absence-FF-Bayes}), replacing the inverse kernel with the quadratic coupling at finite width $\ker^{\delta_1\delta_2}\to g^{\delta_1\delta_2}$, we find
\begin{align}\label{eq:quardratic-posterior}
&\frac{1}{2}\sum_{j=1}^{n_{L}}\sum_{\delta_1,\delta_2\in\D} g^{\delta_1\delta_2} \z{j}{\delta_1}{L}\z{j}{\delta_2}{L}\Big\vert_{\z{i}{\tra}{L}=\y{i}{\tra}}\, \\
=&\ \text{constant}+\frac{1}{2}\sum_{j=1}^{n_{L}}\sum_{\tea_1,\tea_2\in\B} \gpos^{\tea_1\tea_2} \le(\z{j}{\tea_1}{L}-\posmean_{j;\tea_1}\ri)\le(\z{j}{\tea_2}{L}-\posmean_{j;\tea_2}\ri)\, ,\notag
\end{align}
with the \emph{naive} posterior mean\index{posterior!posterior mean!finite width}
\be\label{eq:naive-mean}
\posmean_{i;\tea}\equiv \sum_{\tra_1,\tra_2\in\A} g_{\tea \tra_1}\gsub^{\tra_1\tra_2}\y{i}{\tra_2}\, ,
\ee
and the \emph{naive} posterior covariance\index{posterior!posterior covariance}\index{posterior!posterior covariance!finite width}
\be\label{eq:naive-posterior-variance}
\gpos_{\tea_1\tea_2}\equiv g_{\tea_1\tea_2}-\sum_{\tra_3,\tra_4\in\A}g_{\tea_1\tra_3}\gsub^{\tra_3\tra_4}g_{\tra_4\tea_2}\, .
\ee
We say \emph{naive} here because there are additional corrections we need to consider coming from the 
the quartic term\index{coupling!quartic} in the action. Let see explicitly how this works for the posterior mean\index{posterior!posterior mean}.

Given the observed true outputs $\y{i}{\tra}$ and the quadratic term~\eqref{eq:quardratic-posterior} centered at the naive posterior mean\index{posterior!posterior mean} $\posmean_{i;\tea}$, it is natural to center ourselves at 
\be\label{eq:posmean-definition}
\meanstring_{i;\delta}\equiv\le(\y{i}{\tra}\ ,\ \posmean_{i;\tea}\ri)=\Big(\y{i}{\tra}\ ,\ \sum_{\tra_1,\tra_2\in\A} g_{\tea \tra_1}\gsub^{\tra_1\tra_2}\y{i}{\tra_2}\Big)\, ,
\ee
and define a fluctuating variable $w_{i;\tea}\equiv\z{i}{\tea}{L}-\posmean_{i;\tea}$ so that we can plug the decomposition
\be\label{eq:posterior-mean-finite-width-decomposition}
\z{i}{\delta}{L}=\le(\z{i}{\tra}{L}\ ,\ \z{i}{\tea}{L}\ri) \to \le(\y{i}{\tra}\ ,\ \posmean_{i;\tea}+w_{i;\tea}\ri)=\meanstring_{i;\delta}+\le(0,\ w_{i;\tea}\ri)\, ,
\ee
into the action~\eqref{eq:general-ell-action-reprint}, thus making the partitioning into subsamples $\D = \A \cup \B$ manifest. In terms of this fluctuation, the quadratic term~\eqref{eq:quardratic-posterior} 
takes the form
\be
\text{constant}+\frac{1}{2}\sum_{j=1}^{n_{L}}\sum_{\tea_1,\tea_2\in\B} \gpos^{\tea_1\tea_2} w_{j;\tea_1} w_{j;\tea_2}\, ,
\ee
and the quartic term can be evaluated as
\begin{align}
&\mathcal{Q}\!\le(w\ri)\equiv \Bigg[\frac{1}{8}\sum_{j,k=1}^{n_{L}}\sum_{\delta_1,\ldots,\delta_4\in\D}v^{(\delta_1\delta_2)(\delta_3\delta_4)} \z{j}{\delta_1}{L}\z{j}{\delta_2}{L}\ \z{k}{\delta_3}{L}\z{k}{\delta_4}{L}\Bigg]\Bigg\vert_{\z{i}{\tra}{L}=\Phi_{i;\tra};\ \z{i}{\tea}{L}=\Phi_{i;\tea}+w_{i;\tea}}\, \notag \\
=&\ \text{constant}+\frac{4}{8}\sum_{j}\sum_{\tea_1\in\B}w_{j;\tea_1}\le(\sum_{k}\sum_{\delta_1,\delta_2,\delta_3\in\D}v^{(\tea_1\delta_1)(\delta_2\delta_3)} \meanstring_{j;\delta_1}\meanstring_{k;\delta_2}\meanstring_{k;\delta_3}\ri)\, \notag\\
&+\frac{2}{8}\sum_{j}\sum_{\tea_1,\tea_2\in\B}w_{j;\tea_1}w_{j;\tea_2}\le(\sum_{k}\sum_{\delta_1,\delta_2\in\D}v^{(\tea_1\tea_2)(\delta_1\delta_2)} \meanstring_{k;\delta_1}\meanstring_{k;\delta_2}\ri)\, \notag\\
&+\frac{4}{8}\sum_{j,k}\sum_{\tea_1,\tea_2\in\B}w_{j;\tea_1}w_{k;\tea_2}\le(\sum_{\delta_1,\delta_2\in\D}v^{(\tea_1\delta_1)(\tea_2\delta_2)} \meanstring_{j;\delta_1}\meanstring_{k;\delta_2}\ri)\, \notag\\
&+\frac{4}{8}\sum_{j,k}\sum_{\tea_1,\tea_2,\tea_3\in\B}w_{j;\tea_1}w_{j;\tea_2}w_{k;\tea_3}\le(\sum_{\delta_1\in\D}v^{(\tea_1\tea_2)(\tea_3\delta)} \meanstring_{k;\delta_1}\ri)\, \notag\\
&+\frac{1}{8}\sum_{j,k}\sum_{\tea_1,\ldots,\tea_4\in\B}w_{j;\tea_1}w_{j;\tea_2}\ w_{k;\tea_3}w_{k;\tea_4}\le(v^{(\tea_1\tea_2)(\tea_3\tea_4)}\ri)\, .
\label{eq:posterior-quartic}
\end{align}

\index{finite-width prediction!Bayesian inference}
Given all these expressions, we can finally determine the true posterior mean\index{posterior!posterior mean} by computing the following expectation:
\begin{align}\notag
\int dz^{(L)}_{\B} p\!\le(z_{\B}^{(L)} \Big\vert y_{\A} \ri)  \z{i}{\tea}{L}=& \int dz^{(L)}_{\B} \frac{p\!\le(y_{\A},z_{\B}^{(L)} \ri)}{p\big(y_{\A} \big)} \z{i}{\tea}{L}\\
=&\posmean_{i;\tea}+\int dw_{\B}\, \frac{p\!\le(y_{\A},\ \posmean_{\B}+w_{\B} \ri)}{p(y_{\A} )} w_{i;\tea}\, \notag \\
=&\posmean_{i;\tea}+\frac{ \brabra w_{i;\tea} e^{\mathcal{Q}\le(w\ri)}\ketket_{\gpos}}{\bra\!\bra e^{\mathcal{Q}\le(w\ri)}\ket\!\ket_{\gpos}}\, \notag\\
=&\posmean_{i;\tea}+\frac{ \brabra w_{i;\tea} \le[1+\mathcal{Q}\!\le(w\ri)\ri]\ketket_{\gpos}}{\bra\!\bra 1+\mathcal{Q}\!\le(w\ri)\ket\!\ket_{\gpos}}+\o{v^2}\, \notag\\
=&\posmean_{i;\tea}+\brabra w_{i;\tea}\mathcal{Q}\!\le(w\ri)\ketket_{\gpos}+\o{v^2}\, ,
\label{eq:posterior-mean-at-finite-width}
\end{align}
where on the first line we used Bayes' rule for the posterior~\eqref{eq:BB-NN}, on the second line we inserted our decomposition \eqref{eq:posterior-mean-finite-width-decomposition} in two places, on the third line we separated out the quartic term in order to rewrite the posterior expectation as a Gaussian expectation\index{Gaussian expectation} with respect to the naive posterior covariance\index{posterior!posterior covariance} $\gpos$ divided by the distribution's normalization,
on the fourth line we expanded the exponential, and on the final line we used the fact that the fluctuation has zero mean $\brabra w_{i;\tea} \ketket_{\gpos}=0$ in Gaussian expectation\index{Gaussian expectation}. We can now evaluate the remaining Gaussian expectation\index{Gaussian expectation} by plugging in our expression for the quartic term~\eqref{eq:posterior-quartic} and making Wick contractions\index{Wick contraction}:
\begin{align}\label{eq:mean-posterior-prediction-by-exact-Bayesian-at-finite-width}
&\posmean_{i;\tea}+\brabra w_{i;\tea}\mathcal{Q}\le(w\ri)\ketket_{\gpos}\, \\
=&\posmean_{i;\tea}+\frac{1}{2}\sum_{\tea_1\in\B}\gpos_{\tea \tea_1}\le(\sum_{k}\sum_{\delta_1,\delta_2,\delta_3\in\D}v^{(\tea_1\delta_1)(\delta_2\delta_3)} \meanstring_{i;\delta_1}\meanstring_{k;\delta_2}\meanstring_{k;\delta_3}\ri)\, \notag\\
&+\frac{1}{2}\sum_{\tea_1,\tea_2,\tea_3\in\B}\le(n_{L}\gpos_{\tea_1 \tea_2}\gpos_{\tea \tea_3}+2\gpos_{\tea \tea_1}\gpos_{\tea_2 \tea_3} \ri)\le(\sum_{\delta_1\in\D}v^{(\tea_1\tea_2)(\tea_3\delta_1)} \meanstring_{i;\delta_1}\ri)\, .\notag
\end{align}
Thus, we see that the naive posterior mean \eqref{eq:naive-mean} is further corrected by a number of $v$-dependent terms.

To extract some \terminate{physics} from this complicated expression, note from the definition of $\meanstring$~\eqref{eq:posmean-definition} that the $i$-th component of $\meanstring_{i;\delta}$ depends on
the $i$-th component of our observation $\y{i}{\tra}$. This in particular means that the term above  $\propto \sum_{k}\meanstring_{i;\delta_1}\meanstring_{k;\delta_2}\meanstring_{k;\delta_3}$ does incorporate information from all of the components of the observed true outputs. In other words, information from the $k$-th component of the observed outputs successfully influences the posterior mean\index{posterior!posterior mean} prediction on the $i$-th component for $i\ne k$. This means that at finite width we have a dependence among the components of the posterior outputs
\be\label{eq:stat-dependence-as-good-posterior}
p\!\le(z_{i; \B}^{(L)}, z_{k; \B}^{(L)} \Big\vert y_{i;\A}, y_{k;\A}\ri) \neq p\!\le(z_{i; \B}^{(L)} \Big\vert y_{i;\A}\ri)\, p\!\le(z_{k; \B}^{(L)} \Big\vert y_{k;\A}\ri)\, .
\ee
This property of the posterior distribution\index{posterior} descends from the nontrivial \emph{fire-together} \terminate{inductive bias} $p\Big(z_i^{(L)}\Big\vert z_k^{(L)}\Big)$ present  in the finite-width prior as discussed in~\S\ref{subsec:Hebbian}. The dependence among the components of the posterior outputs~\eqref{eq:stat-dependence-as-good-posterior} is a signature of our posterior beliefs' learning to \emph{wire together}, and we will see a further manifestation of this when we again consider \terminate{representation learning} in the next section.

Before we move on, we should address practical matters.
Practically speaking, it is even more computationally infeasible 
to evaluate the finite-width predictions of Bayesian learning\index{Bayesian inference!practicalities}~\eqref{eq:posterior-mean-at-finite-width} than it was at infinite width. In particular, evaluating the quartic coupling\index{coupling!quartic} involves first \emph{representing} the \terminate{four-point vertex} -- a $\NR\times\NR\times\NR\times\NR$-dimensional tensor -- and then multiply contracting it with inverse kernels. Thus, both the cost of computation and the memory requirements of Bayesian learning grow terrifyingly quickly with 
our observations, i.e.~with size of our dataset $\A$. However, please \emph{Don't Panic}\index{Don't Panic, HHGTTG}: we are getting ever closer to the point where we can show you how gradient-based learning resolves all these practical difficulties at finite width.

\subsection{Presence of Representation Learning}\label{subsec:presence-RL-Bayes}
The fact that the individual components of the finite-width posterior mean\index{posterior!posterior mean} prediction can incorporate information from our observations of the other components is suggestive of the idea that these observations might also be used to build up representations\index{representation} in the hidden layers. Here we will show that such \neo{representation learning} 
actually does occur at finite width as a direct consequence of the nonzero \emph{interlayer} interaction\index{interactions}s. %

Analogous to our parallel subsection  at infinite width (\S\ref{subsec:absence-RL-Bayes}),  we can investigate representation learning by considering the posterior distribution in the penultimate layer $\ell = L-1$ on the full set of samples $\D$, given observations $y_\A$. In particular, to show how the \emph{features}\index{feature} of the penultimate-layer representation evolve, our goal will be to compute the change in the expectation of a penultimate-layer observable $\O\!\le(z^{(L-1)}_{\D}\ri)$ taken with respect to the posterior as compared to the expectation taken with respect to the prior
\be\label{eq:posterior-minus-prior}
\dO\equiv \int dz^{(L-1)}_{\D} p\!\le(z^{(L-1)}_{\D}\Big\vert y_\A\ri) \O\!\le(z^{(L-1)}_{\D}\ri)-\int dz^{(L-1)}_{\D} p\!\le(z^{(L-1)}_{\D} \ri) \O\!\le(z^{(L-1)}_{\D}\ri)\, ,
\ee
where $p\!\le(z^{(L-1)}_{\D} \ri)$ and $p\!\le(z^{(L-1)}_{\D}\Big\vert y_\A \ri)$ are the prior and posterior distributions, respectively. This expectation difference was strictly zero in the infinite-width limit 
since the penultimate-layer posterior was exactly equal to the penultimate-layer prior \eqref{eq:infinite-width-penultimate-posterior-equals-prior}. A non-vanishing difference in contrast will mean that the penultimate-layer preactivations are being updated after making  observations $y_{\A}$. Such an \emph{update} is a direct avatar of \terminate{representation learning}.

As before, by Bayes' rule\index{Bayesian probability!Bayes' rule} we can write the posterior distribution of the penultimate preactivations $z_\D^{(L-1)}$ given our observations $y_\A$ as
\be\label{eq:Bayes-posterior-general-reprint}
p\!\le(z_{\D}^{(L-1)}\Big\vert y_\A \ri)=\frac{p\!\le(y_\A \Big\vert z_{\D}^{(L-1)}\ri)p\!\le(z_{\D}^{(L-1)} \ri)}{p\!\le(y_\A \ri)}\, .
\ee
Just as before, the likelihood\index{Bayesian inference!likelihood} $p\!\le(y_\A \Big\vert z_{\D}^{(L-1)} \ri)$ is the conditional distribution $p\!\le(z_\A^{(L)} \Big\vert z_{\D}^{(L-1)} \ri)$ evaluated on our set of observations $z_{\A}^{(L)} \to y_\A$. With this expression for the posterior \eqref{eq:Bayes-posterior-general-reprint}, 
we can express the update $\dO$ after Bayesian learning as
\be\label{eq:different-of-expectations}
\dO = \E{\frac{p\!\le(y_\A \Big\vert z_{\D}^{(L-1)}\ri)}{p\!\le(y_\A \ri)} \O\!\le(z^{(L-1)}_{\D}\ri)}- \E{\O\!\le(z^{(L-1)}_{\D}\ri)}\, .
\ee
As always, the full expectation $\E{\,\Vdot\,}$ is to be evaluated with respect to the \emph{prior} or initialization distribution $p\!\le(z_{\D}^{(L-1)} \ri)$; all learning will always be represented explicitly with the insertion of other factors as we did above.

Let's now determine how this insertion, the likelihood-to-evidence ratio 
\be\label{eq:likelihood-to-evidence-ratio}
\frac{p\!\le(y_\A \Big\vert z_{\D}^{(L-1)} \ri)}{p\!\le(y_\A \ri)}\, ,
\ee 
depends on the preactivations $z^{(L-1)}_{\D}$. As we pointed out when working through the infinite-width example, we already worked out the form of this likelihood in~\eqref{eq:general-layer-conditional} as the conditional distribution between layers. In our current context and notation, the likelihood\index{Bayesian inference!likelihood} reads
\be\label{eq:general-layer-conditional-reprint}
p\!\le(y_\A \Big\vert z_{\D}^{(L-1)} \ri)= \frac{1}{\sqrt{\dete{2\pi \widehat{G}^{(L)}}^{n_{L}}}} \exp\!\le(-\frac{1}{2}\sum_{i=1}^{n_{L}}\sum_{\tra_1,\tra_2\in\A}\TI{\widehat{G}}{\tra_1 \tra_2}{L}\y{i}{\tra_1}\y{i}{\tra_2}\ri)\, ,
\ee
where as a reminder the \emph{stochastic metric} $\Ti{\widehat{G}}{\tra_1 \tra_2}{L}= \Ti{\widehat{G}}{\tra_1 \tra_2}{L}\!\le( z_{\A}^{(L-1)}\ri)$ depends explicitly on the preactivations in the penultimate layer $\z{i}{\A}{L-1}$.\footnote{
    \label{foot:gross-tilde-hat}Strictly speaking, we should really denote the stochastic metric here as $\Ti{\widehat{\widetilde{G}}}{\tra_1 \tra_2}{L}$ to indicate that we're focusing on the $\NR$-by-$\NR$ submatrix of the full stochastic metric on $\D$, $\Ti{\widehat{G}}{\delta_1 \delta_2}{L}$. It's the matrix inverse of this submatrix $\Ti{\widehat{\widetilde{G}}}{\tra_1 \tra_2}{L}$ -- and not the $(\tra_1, \tra_2)$ block of the inverse of the full matrix $\Ti{\widehat{G}}{\delta_1 \delta_2}{L}$ -- that appears in \eqref{eq:general-layer-conditional-reprint}.
    Since  this tilde-with-a-hat looks ridiculous -- and since we are already heavily overburdened on the notational front -- if you promise to keep this caveat in mind, we'll do everyone a favor and temporarily suppress this tilde.
} 
Thus, the stochastic metric acts as a coupling here, inducing interlayer interactions between the $(L-1)$-th-layer preactivations and the observations $y_\A$. As we will see, this endows the updated distribution over $z_{\D}^{(L-1)}$ with a dependence on $y_\A$.

\index{Schwinger-Dyson equations}
As should be fairly familiar at this point, we can decompose the stochastic metric into a mean and a fluctuation,\index{tensor decomposition!metric mean and fluctuation} 
\be
\Ti{\widehat{G}}{\tra_1 \tra_2}{L}\equiv\Ti{G}{\tra_1 \tra_2}{L}+\Ti{\widehat{\Delta G}}{\tra_1 \tra_2}{L} \, ,
\ee 
in terms of which the likelihood \eqref{eq:general-layer-conditional-reprint} can be Taylor-expanded \`{a} la Schwinger-Dyson as we did before in~\eqref{eq:second-layer-stochastic-exponential} and~\eqref{eq:second-layer-stochastic-determinant}. At first nontrivial order, we find for the likelihood-to-evidence ratio \eqref{eq:likelihood-to-evidence-ratio}
\begin{align}\label{eq:SD-for-likelihood-to-evidence-ratio}
&\frac{p\!\le(y_\A \Big\vert z_{\D}^{(L-1)}\ri)}{p\!\le(y_\A \ri)} \\
=& \frac{1}{p\!\le(y_\A \ri)\sqrt{\dete{2\pi G^{(L)}}^{n_{L}}}} \le[1\!+\frac{1}{2}\!\sum_{\tra_1,\ldots,\tra_4\in\A}\!\!\!\!\!\!\!\!\Ti{\widehat{\Delta G}}{\tra_1\tra_2}{L}\TI{G}{\tra_1\tra_3}{L}\TI{G}{\tra_2\tra_4}{L}\!\sum_{i=1}^{n_{L}}\!\le(\y{i}{\tra_3}\y{i}{\tra_4}\!\!-\!\Ti{G}{\tra_3\tra_4}{L}\ri)\!+\!\o{\Delta^2}\!\ri]\,\! . \notag
\end{align}
Here, the prefactor 
before the square brackets
is constant with respect to the variables $z^{(L-1)}_{\D}$, and so all of the relevant dependence needed to evaluate update $\dO$~\eqref{eq:different-of-expectations} is contained implicitly in the metric fluctuation $\Ti{\widehat{\Delta G}}{\tra_1\tra_2}{L}$.
We can thus compute a posterior expectation -- i.e.~the first expectation in \eqref{eq:different-of-expectations} -- of any observable by integrating against the quantity in the square bracket, so long as we also divide by an integral of ``$1$'' against the same quantity in order to properly normalize.
With this by-now familiar trick in mind, we can rewrite the posterior expectation  as
\begin{align}
&\frac{\mathbb{E}\le\{\O\!\le(z^{(L-1)}_{\D}\ri)\!\le[1+\frac{1}{2}\sum_{\tra_1,\ldots,\tra_4}\!\Ti{\widehat{\Delta G}}{\tra_1\tra_2}{L}\TI{G}{\tra_1\tra_3}{L}\TI{G}{\tra_2\tra_4}{L}\sum_{i}\le(\y{i}{\tra_3}\y{i}{\tra_4}\!-\!\Ti{G}{\tra_3\tra_4}{L}\ri)\!+\!\o{\Delta^2}\ri]\ri\}}{\E{1+\frac{1}{2}\sum_{\tra_1,\ldots,\tra_4}\!\Ti{\widehat{\Delta G}}{\tra_1\tra_2}{L}\TI{G}{\tra_1\tra_3}{L}\TI{G}{\tra_2\tra_4}{L}\sum_{i}\le(\y{i}{\tra_3}\y{i}{\tra_4}\!-\!\Ti{G}{\tra_3\tra_4}{L}\ri)\!+\!\o{\Delta^2}}}\, \\
=&\E{\O\!\le(z^{(L-1)}_{\D}\ri)}\, \notag\\
&+\frac{1}{2}\sum_{\tra_1,\ldots,\tra_4\in\A}\E{\Ti{\widehat{\Delta G}}{\tra_1\tra_2}{L}\O\!\le(z^{(L-1)}_{\D}\ri)}\TI{G}{\tra_1\tra_3}{L}\TI{G}{\tra_2\tra_4}{L}\sum_{i}\le(\y{i}{\tra_3}\y{i}{\tra_4}\!-\!\Ti{G}{\tra_3\tra_4}{L}\ri)+\o{\frac{1}{n^2}}\, ,\notag
\end{align}
where the details of what we actually did are hidden in this here footnote.\footnote{The reason that we treated the additional $\o{\Delta^2}$ pieces as $\o{1/n^2}$ is hidden under the rug in the main body. To peak under that rug, first let us schematically express the likelihood-to-evidence ratio~\eqref{eq:SD-for-likelihood-to-evidence-ratio} as $\text{constant}\times \le[1+\sharp_1\Delta G+\sharp_2 (\Delta G)^2+\o{\Delta^3}\ri]$. Then, the posterior expectation becomes
\begin{align}
&\frac{\mathbb{E}\le\{\O \le[1+\sharp_1\Delta G+\sharp_2 (\Delta G)^2+\o{\Delta^3}\ri]\ri\}}{\E{1+\sharp_1\Delta G+\sharp_2 (\Delta G)^2+\o{\Delta^3}}}=\frac{\E{\O}+\sharp_1\E{\O\Delta G}+\sharp_2\E{\le(\Delta G\ri)^2\O}+\o{1/n^2}}{1+\sharp_2 \E{(\Delta G)^2}+\o{1/n^2}}\, \\
=&\E{\O}+\sharp_1\E{\O \Delta G}+\sharp_2 \le\{\E{\le(\Delta G\ri)^2\O}-\E{\le(\Delta G\ri)^2}\E{\O}\ri\}+\o{1/n^2}\, .\notag
\end{align}
Decomposing the observable into a mean and a fluctuation as $\O=\E{\O}+\Delta \O$, we see that the term proportional to the coefficient $\sharp_2$ is $\E{\o{\Delta^3}}=\o{1/n^2}$ and thus can be neglected, while the leading finite-width correction cannot be neglected: $\sharp_1\E{\O \Delta G}=\sharp_1\E{\Delta\O \Delta G}=\o{1/n}$.
} We see that the first term is just the prior expectation, while the second term expresses the update $\dO$ \eqref{eq:posterior-minus-prior}.
Finally, 
taking only the leading finite-width corrections at order $1/n$ and 
restoring the tildes to correctly represent the submatrices on $\A$ alone, we can write down a very general expression for the update to any penultimate-layer observable at leading nontrivial order in $1/n$:
\be\label{eq:update-bayes-penultimate-formula}
\dO = \frac{1}{2}\sum_{\tra_1,\ldots,\tra_4\in\A}\E{\Ti{\widehat{\Delta G}}{\tra_1\tra_2}{L}\O\!\le(z^{(L-1)}_{\D}\ri)} \TI{\kersub}{\tra_1\tra_3}{L}\TI{\kersub}{\tra_2\tra_4}{L}\sum_{i}^{n_L}\le(\y{i}{\tra_3}\y{i}{\tra_4}\!-\!\Ti{\kersub}{\tra_3\tra_4}{L}\ri)\, . %
\ee
Again, please be careful and remember that the $\E{\,\Vdot\,}$ in \eqref{eq:update-bayes-penultimate-formula} is to be evaluated with respect to the prior distribution $p\!\le(z_{\D}^{(L-1)} \ri)$.
Note also that the lone expectation in the update \eqref{eq:update-bayes-penultimate-formula} is just the covariance of the stochastic metric with the observable:
\be
\E{\Ti{\widehat{\Delta G}}{\tra_1\tra_2}{L}\O\!\le(z^{(L-1)}_{\D}\ri)}=\E{\Ti{\widehat{G}}{\tra_1\tra_2}{L}\O\!\le(z^{(L-1)}_{\D}\ri)}-\E{\Ti{\widehat{G}}{\tra_1\tra_2}{L}}\E{\O\!\le(z^{(L-1)}_{\D}\ri)}\, .
\ee
As we addressed in that rugly footnote, for a general order-one observable this covariance is $1/n$-suppressed but nonzero. Thus, we see that at large-but-finite width $(1 \ll n < \infty)$, such observables get updated: representations are learned.\index{representation learning}

In order to see how this works, let's consider a concrete example. The simplest observable turns out to be the average norm of the activations
\be
\O\!\le(z^{(L-1)}_{\D}\ri) \equiv \frac{1}{n_{L-1}}\sum_{j=1}^{n_{L-1}}\s{j}{\delta_1}{L-1}\s{j}{\delta_2}{L-1} \, ,
\ee
which we can decompose in terms of a mean and a fluctuation as
\be
\O\!\le(z^{(L-1)}_{\D}\ri)  = \E{\O\!\le(z^{(L-1)}_{\D}\ri)}+\frac{1}{\CW{L}}\Ti{\widehat{\Delta G}}{\delta_1\delta_2}{L}\, ,
\ee
if we also recall the explicit form of the metric fluctuation~\eqref{eq:metric-fluctuation-general-layer}
\be
\Ti{\widehat{\Delta G}}{\tra_1\tra_2}{L}=\CW{L}\frac{1}{n_{L-1}}\sum_{j=1}^{n_{L-1}}\le(\s{j}{\tra_1}{L-1}\s{j}{\tra_2}{L-1}-\E{\s{j}{\tra_1}{L-1}\s{j}{\tra_2}{L-1}}\ri)\, .
\ee
Then, plugging into our expression for the leading-order finite-width update \eqref{eq:update-bayes-penultimate-formula}, we find
\begin{align}\notag
\dO=&\frac{1}{2 \CW{L}  }\sum_{\tra_1,\ldots,\tra_4\in\A}\E{\Ti{\widehat{\Delta G}}{\delta_1\delta_2}{L}\Ti{\widehat{\Delta G}}{\tra_1\tra_2}{L}  } \TI{\kersub}{\tra_1\tra_3}{L}\TI{\kersub}{\tra_2\tra_4}{L}\sum_{i}\le(\y{i}{\tra_3}\y{i}{\tra_4}\!-\!\Ti{\kersub}{\tra_3\tra_4}{L}\ri)\, \\
=& \frac{1}{2n_{L-1}\CW{L}}\sum_{\tra_1,\ldots,\tra_4\in\A}V^{(L)}_{(\delta_1\delta_2)(\tra_1\tra_2)}\TI{\kersub}{\tra_1\tra_3}{L}\TI{\kersub}{\tra_2\tra_4}{L}\sum_{i}\le(\y{i}{\tra_3}\y{i}{\tra_4}\!-\!\Ti{\kersub}{\tra_3\tra_4}{L}\ri)\, ,
\end{align}
where to go to the second line we used the definition of the four-point vertex in terms of the two-point function of the metric fluctuation \eqref{eq:vertex-in-terms-of-metric-fluctuation}. As this vertex characterizes the non-Gaussianity of the output distribution, we see explicitly here how interactions are mediating updates to the penultimate-layer activations. In addition, the leading factor of $1/n_{L-1}$ makes it clear that this update is a finite-width effect. Further,  the term in the last parenthesis shows that the update depends explicitly on the difference between our observations of the outputs, $\y{i}{\tra_3}\y{i}{\tra_4}$, and our prior expectations of them, $\E{z^{(L)}_{i;\tra_3} z^{(L)}_{i;\tra_4} } \equiv \Ti{\kersub}{\tra_3\tra_4}{L} + \o{1/n}$. This means that the observations are in fact propagating \emph{backward} to induce changes in the hidden-layer representations.\footnote{This kind of backward-propagation or \neo{backpropagation}, if you will, persists further into the shallower hidden layers as well.
However, in the $(L-2)$-th layer, the posterior update turns out to be of order $\o{1/n^2}$. Intuitively this makes sense because the change in the representation in the penultimate layer $(L-1)$ is already down by a factor of $1/n$, and it gets further suppressed due to the $1/n$-suppression of the interlayer interaction in going back to the $(L-2)$-th layer.

Mathematically, we can consider 
the update to 
an $(L-2)$-th-layer observable $\O\!\le(z^{(L-2)}_{\D}\ri)$ 
as
\be\label{eq:posterior-minus-prior-two-layers-back}
\dO\equiv \int dz^{(L-2)}_{\D} p\!\le(z^{(L-2)}_{\D}\Big\vert y_\A\ri) \O\!\le(z^{(L-2)}_{\D}\ri)-\int dz^{(L-2)}_{\D} p\!\le(z^{(L-2)}_{\D} \ri) \O\!\le(z^{(L-2)}_{\D}\ri)\, .
\ee
Through the chain of Bayes', sum, and product rules, the posterior insertion in this formula is given in terms of the following marginalization:
\be
p\!\le(z^{(L-2)}_{\D}\Big\vert y_\A\ri)=\frac{p\!\le(y_\A\Big\vert z^{(L-2)}_{\D}\ri)p\!\le(z^{(L-2)}_{\D}\ri)}{p\!\le(y_\A \ri)}=\int dz^{(L-1)}_{\D} \frac{p\!\le(y_\A\Big\vert z^{(L-1)}_{\D}\ri)}{p\!\le(y_\A \ri)}p\!\le(z^{(L-1)}_{\D}, z^{(L-2)}_{\D}\ri)\, .
\ee
From here, through the same set of manipulations that led to the update equation for the penultimate layer \eqref{eq:update-bayes-penultimate-formula}, we get
\be\label{eq:update-bayes-two-layers-back-formula}
\dO = \frac{1}{2}\sum_{\tra_1,\ldots,\tra_4\in\A}\E{\Ti{\widehat{\Delta G}}{\tra_1\tra_2}{L}\!\le(z^{(L-1)}_{\D}\ri)\O\!\le(z^{(L-2)}_{\D}\ri)} \TI{\kersub}{\tra_1\tra_3}{L}\TI{\kersub}{\tra_2\tra_4}{L}\sum_{i}^{n_L}\le(\y{i}{\tra_3}\y{i}{\tra_4}\!-\!\Ti{\kersub}{\tra_3\tra_4}{L}\ri)+\o{\frac{1}{n^2}}\, . %
\ee
Thus, to show that this change is of order $\o{1/n^2}$, we need to show that the \neo{interlayer correlation},
\be\label{eq:interlayer-next-layer}
\E{\Ti{\widehat{\Delta G}}{\tra_1\tra_2}{L}\!\le(z^{(L-1)}_{\D}\ri)\O\!\le(z^{(L-2)}_{\D}\ri)} \, ,
\ee
is of order $\o{1/n^2}$. This is most swiftly carried out in the future, first by the application of the formula~\eqref{eq:no-weight-insertion-general} with $\ell=L-2$ and then with the associated trickery~\eqref{eq:SD-again}. If you are up for a challenge, please flip forward and write a note next to \eqref{eq:SD-again} reminding yourself to come back to footnote~\ref{foot:come-back-to-me} in \S\ref{subsec:presence-RL-Bayes}.\label{foot:come-back-to-me}
\emph{Spoiler alert:}\index{spoiler alert} you should in fact find that~\eqref{eq:interlayer-next-layer}
is of order $\o{1/n^2}$.
}

Although perhaps not practically useful, this Bayesian analysis of representation learning at finite width 
will serve as a theoretically useful blueprint for studying a similar type of \terminate{representation learning} that occurs with gradient-based learning at finite width in \S\ref{ch:features}. Now, with all these allusions to gradient-based learning having accrued with interest, you must be really excited to flip the page to the next chapter!

%% file: Chp7-GD/7_global.tex
\chapter{Gradient-Based Learning}\index{gradient-based learning|see{gradient descent}}
\label{ch:training}

\epigraph{%
Of course, that’s like saying \terminate{Newton's second law} $F = ma$, as it appears in textbooks on mechanics\index{mechanics (physics)}, is just a definition of what you mean by ``force''.  That’s true, strictly speaking, but we live in a landscape where there is an implicit promise that when someone writes that down \dots
that they will give laws for the force\index{force|see{Newton's second law}}, and not, say, for some quantity involving the 17th time derivative of the position.}{Sidney Coleman, in his ``Quantum Mechanics\index{quantum mechanics} in Your Face'' Dirac\index{Dirac, Paul Adrien Maurice} Lecture \cite{Coleman:2020put}.\index{Coleman, Sidney}}

\index{optimization|see{gradient descent}}\index{optimization|see{training}}

\noindent{}In the last chapter, we discussed \terminate{Bayesian inference} as a learning algorithm, which followed naturally from our study of networks at initialization. Starting from a description of a neural network architecture with parameters -- weights and biases -- we integrated out these parameters to find a distribution over preactivations $z^{(\ell)}(x)$ as a function of layer and input sample, which in particular includes the \terminate{output distribution} $p\!\le(z^{(L)}(x) \ri)$. %
This was interpreted as a \emph{prior distribution}\index{prior} over an \terminate{ensemble} of such models, and then we explained how the logic of Bayes' rule\index{Bayesian probability!Bayes' rule} lets us evolve the prior into a \emph{posterior distribution}\index{posterior} conditioned on observed data. Despite the theoretical elegance of \terminate{Bayesian inference}, the naive implementation quickly became computationally intractable as the number of conditioned data samples grew large.

Stepping back, there's actually something a little bit odd about this setup. Once we worked out the \terminate{output distribution}, the actual network itself was discarded, with the parameters long since integrated out. Since \terminate{Bayesian inference} only cares about the output distribution of a model, the starting point for inference can really be any ensemble of models as it isn't specifically tailored to neural networks\index{neural network} at all. So why go through all the trouble of starting with neural-network models? How did we even know that these models are a good abstraction to begin with?

\index{training}\index{machine learning}
Deep neural networks are exciting because they work surprisingly well. We know this because in practice such networks are explicitly \emph{trained} and used to perform useful tasks. Most commonly, learning occurs by repeatedly updating the \terminate{model parameters} via a gradient-based optimization procedure such as \neo{gradient descent}. 

\index{learning algorithm}
In particular, gradient-based learning algorithms can efficiently process a large amount of training data by optimizing an auxiliary \neo{loss} function that directly compares the network output $f(x;\theta) \equiv z^{(L)}(x)$ to some desired result or \emph{label}. 
This optimization procedure involves sampling only a single set of network parameters from the \terminate{initialization distribution}, yielding just a single network trained for the task of interest rather than a full \terminate{ensemble} of networks. 
In this way, gradient-based learning methods offset their inability to express confidence in their predictions -- due to the absence of an \terminate{ensemble} -- with data efficiency and easy scalability.

Since \terminate{gradient descent} involves making explicit updates to the \terminate{model parameters}, the first step is to bring them back (from whatever place that variables go when they are integrated out). 
In supervised learning, the adjustments of \terminate{model parameters} are directly proportional to the function-approximation\index{function approximation} error times the gradient of the model output with respect to the parameters.
This decomposition motivates the study of the \neo{neural tangent kernel} (NTK).\footnote{The NTK was first identified in the seminal work of 
Jacot \emph{et al.}~\cite{jacot2018neural} in the context of infinite-width networks.
}
In short, the NTK is a type of \terminate{Hamiltonian}\index{Hamiltonian|seealso{neural tangent kernel}}  that controls the training dynamics\index{training dynamics!controlled by the NTK} of observables\index{observable} whenever gradient descent is used to optimize an auxiliary loss that scores a \terminate{function approximation}. As we detail 
in~\S\ref{ch:NTHb}, \S\ref{ch:features}, and~\S\ref{ch:eot},
understanding the NTK for a given neural-network architecture will enable us to effectively describe gradient-based learning for that model.

In this chapter, we give a short introduction to supervised learning in \S\ref{sec:supervised-learning}, followed by a discussion of \terminate{gradient descent} in \S\ref{sec:gd} with a very general focus on how the NTK arises in supervised learning. In the next chapter, we'll incorporate the NTK into our \terminate{effective theory of deep learning} by exploiting the same layer-to-layer RG flow technique we used in \S\ref{ch:ngp}. %
\index{representation group flow}

\section{Supervised Learning}\label{sec:supervised-learning}
One of the most basic modeling tasks at which neural networks\index{neural network} excel is known as \term{supervised learning}.
Given a \term{data distribution}\index{data distribution|seealso{input data}} $p(x,y)=p(y|x)p(x)$, the goal is to predict a \term{label} $y$ given an input $x$, for any pair that is jointly sampled from the distribution.\footnote{In this section, we suppress \neo{vectorial indices}  on the inputs $x_\delta$, labels $y_\delta$, and model outputs $z\le(x_\delta;\theta\ri)$, while often retaining \neo{sample indices} $\delta\in\D$.} To be precise, the model tries to learn the conditional distribution $p(y|x)$, and the resulting model is sometimes called a \textbf{discriminative model}.\index{discriminative model}\index{discriminative model|seealso{probabilistic model}}
In one canonical example from \terminate{computer vision}, we might want to \index{classification|textbf}\textbf{classify} an image $x_{\delta}$ of a hand-written digit ``$3$'' according to its literal value $y_{\delta}=$ \texttt{3}. Or, for a \terminate{natural language processing} example, given a sentence containing the word $x_{\delta} =$ \texttt{cat} we might want to identify the part of the speech as $y_{\delta} =$ \texttt{noun}.  The better the \neo{probabilistic model} learns the distribution $p(y|x)$,  the more accurately it can predict a true label $y$ for a novel input example $x$. Generating these datasets generally requires human annotators to label the inputs, hence the name supervised learning.

In this setup, the supervised-learning model outputs a prediction $z(x_\delta; \theta)$.  This notation emphasizes that the model output is both a function of the input $x_\delta$ as well as some adjustable parameters $\theta$. This should already be familiar in the context of neural-network \terminate{function approximation}, where the \terminate{model parameters} consist of
the \terminate{biases} and \terminate{weights}.\index{neural network}

As discussed in \S\ref{sec:MLP_distribution}, the \terminate{model parameters} are drawn from an easy-to-sample prior distribution\index{prior} over the parameters, which is also known as the \neo{initialization distribution} in the context of gradient-based learning. Importantly, this parameter distribution knows nothing about the \terminate{data distribution}. Thus, in order for the model to make good predictions, its parameters will need to be adjusted somehow. Really, this is just a specific application of the \terminate{function approximation} that we discussed in \S\ref{sec:MLP_intro} where the function to be approximated is a conditional distribution $p(y|x)$.

Before we understand how to adjust or \emph{fit} the model parameters\index{gradient descent!model fitting}\index{model fitting!gradient-based optimization|see{gradient descent}}\index{model fitting|seealso{training}},  we need to understand what we mean by making good predictions. What we want is, for a typical input $x_\delta$ and a label $y_\delta$ sampled from the \terminate{data distribution} $p(x,y)$, that the model output $z(x_\delta; \theta)$ is as close to the label $y_\delta$ as possible on average. In order to measure this proximity, for a prediction-label pair we need to define an auxiliary \neo{objective function}\index{objective function|seealso{loss}} or \term{loss},
\be\label{eq:loss-per-example}
\L\Big(z(x_\delta; \theta),\, y_\delta\Big) \,,
\ee
with the property that the closer $z(x_\delta; \theta)$ is to $y_\delta$, the lower the value of the function is. One very intuitive choice for the loss is 
\emph{MSE loss}\index{loss!MSE}\index{mean squared error|see{loss}}\index{MSE|see{mean squared error}}~\eqref{eq:MSE-loss-preview},
\be\label{eq:MSE-loss}
\L_{\text{MSE}}\Big(z(x_\delta; \theta),\, y_{\delta}\Big)\equiv \frac{1}{2}\Big[z(x_\delta; \theta)-y_{\delta}\Big]^2 \,,
\ee
which clearly has the required property,
though this is not the most common choice in deep learning. The specific form of the loss will not matter for the rest of the chapter. \index{loss!MSE}

With the loss function in hand, the goal of \terminate{training} is to adjust \terminate{model parameters} so as to minimize the loss for as many input-label pairs as possible. Ideally, we would like to minimize the loss averaged over the entire \terminate{data distribution},
\be\label{eq:entireloss}
\E{\L(\theta)} = \int dx dy\, p(x,y)\, \L\Big(z(x; \theta),\, y\Big) \, .
\ee
But since we almost never have access to the analytical form of the data distribution $p(x,y)$, in practice this would require the sampling of an infinite number of input-label pairs. Instead, as a proxy of the entire loss~\eqref{eq:entireloss},
we sample a large-but-finite number of pairs $(x_{\tra}, y_{\tra})_{\tra\in \A}$ and try to minimize
\be\label{eq:training-loss}
\L_\A(\theta) \equiv \sum_{\tra \in \A} \L\Big(z(x_{\tra}; \theta),\, y_{\tra}\Big)\, .
\ee
This set of examples $\A$ is referred to as the \term{training set}, and the estimate of the loss \eqref{eq:training-loss} is called the \textbf{training loss}\index{loss!training loss|textbf}\index{training loss|see{loss}}; 
here we've also inherited from \S\ref{ch:bayesian-inference} our sample-index notation of alpha-with-tilde for the inputs in the training set $\tra \in\A$, while denoting generic inputs as delta-with-no-decoration $\delta\in\D$, and soon we'll  use beta-with-dot for inputs in the \emph{test set} $\tea\in\B$.\footnote{Note that our definition of the training loss \eqref{eq:training-loss} is a bit at odds with our definition of the expected loss \eqref{eq:entireloss}. In particular, the expected loss is \emph{intensive}\index{intensivity (of loss)}, while the training loss
is \emph{extensive}\index{extensivity!of loss}, scaling linearly with the size of the training set $\NR \equiv \vert\A\vert$.
This latter choice is consistent with our first definition of this loss, \eqref{eq:MSE-loss-preview}, in the context of MLE\index{maximum likelihood estimation} as an approximate method for Bayesian model fitting\index{Bayesian inference!model fitting} in \S\ref{subsec:ForIO}. There, the extensivity of the loss was natural according to the Bayesian framework: as the number of observed input-output pairs $\NR$ increases, we want the likelihood to dominate the prior. %
As such, we will find it natural to follow that convention. (You also might more accurately call the extensive loss \eqref{eq:MSE-loss-preview} as the \emph{\sout{mean} squared error}.\index{loss!MSE!name}\index{loss!SE})
However, from a non-Bayesian perspective, it is often customary to define a training loss as
\be\label{eq:training-loss-intense}
\L_\A(\theta) \equiv \frac{1}{\vert\A\vert}\sum_{\tra \in \A} \L\Big(z(x_{\tra}; \theta),\, y_{\tra}\Big)\, ,
\ee
which better corresponds to the expected loss \eqref{eq:entireloss}. Since in the context of gradient-based learning the overall normalization can always be absorbed in a redefinition of the global learning rate\index{learning rate!global} $\eta$, to be introduced next section, the only advantage we see of this latter definition \eqref{eq:training-loss-intense} is the better correspondence of the loss with its name.
}
To train our model, we try to find a configuration of the model parameters that minimizes the training loss\index{loss!training loss} 
\be\label{eq:supervised-learning-criterion}
\theta^{\star} = \argmin_\theta \L_\A(\theta)=\argmin_\theta \le[\sum_{\tra \in \A} \L\Big(z(x_{\tra}; \theta),\, y_{\tra}\Big) \ri]\, .
\ee
In the next section, 
we will present the \terminate{gradient descent} algorithm as a way to accomplish this goal.

\index{bias-variance tradeoff}
Having set the minimization of the training loss\index{loss!training loss}~\eqref{eq:training-loss} as our optimization problem, it is important to keep in mind that the true goal of \terminate{supervised learning} is the minimization of the \terminate{loss} over the entire \terminate{data distribution} in the sense of \eqref{eq:entireloss}.
Said another way, the question is not whether the model is able to memorize all the input-label pairs in the training set, but rather whether it's able to generalize its predictions to additional input-label pairs not seen during \terminate{training}.
One might then worry about whether a \terminate{training set} is \emph{biased} in its sampling of the \terminate{data distribution} or
whether
there is high \emph{variance} in a particular set of samples.

To explicitly assess this \term{generalization} property of a model, a separate set of input-label samples $(x_{\tea}, y_{\tea})_{\tea \in \B}$ -- known as the \term{test set} -- is typically set aside and only used to evaluate a model after \terminate{training} is complete.  To the extent that the \terminate{training set} $\A$ is representative of the full \terminate{data distribution} $p(x,y)$, 
decreasing the training loss\index{loss!training loss} will often decrease the entire loss~\eqref{eq:entireloss}, as estimated by the test loss\index{loss!test loss|textbf}\index{test loss|see{loss}} $\L_\B$. We will address this question directly in~\S\ref{ch:NTHb}.

\section{Gradient Descent and Function Approximation}\label{sec:gd}
Considering the training loss\index{loss!training loss} minimization~\eqref{eq:supervised-learning-criterion}, we see that learning is a complicated optimization problem.
Being entirely naive about it,  in order to find extrema of a function, calculus instructs us to differentiate the training loss\index{loss!training loss} and find the value of the argument for which the resulting expression vanishes:
\be\label{eq:gradient-vanishing-mind}
0 = \frac{\td\L_\A}{\td\theta_{\mu}}\Bigg|_{\theta=\theta^{\star}}\, .
\ee
Unfortunately this equation is exactly solvable only in special cases, for instance when the \terminate{loss} is quadratic in the \terminate{model parameters}. Rather than trying to find minima analytically, practitioners typically employ an iterative procedure to bring the loss closer and closer to a minimum.

\index{loss!training loss}\index{gradient descent!model fitting}
\textbf{Gradient descent}\index{gradient descent|textbf} is one such method that can be used to minimize nontrivial functions like the training loss~\eqref{eq:training-loss}, and so it's a natural candidate for model fitting. The algorithm involves the  computation of the gradient of the \terminate{loss}
and iteratively updates the \terminate{model parameters} in the (negative) direction of the gradient
\be\label{eq:gd-update}
\theta_\mu(t+1) = \theta_\mu(t) - \eta \frac{\td \L_\A}{\td \theta_\mu}\Bigg|_{\theta_\mu = \theta_\mu(t)} \,,
\ee
where  $t$ keeps track of the number of steps in the iterative training process, with $t=0$ conventionally being the point of initialization.
Here, $\eta$ is a positive \textbf{training hyperparameter}\index{training hyperparameters|textbf} called the \term{learning rate},
which controls how large of a step is taken in \neo{parameter space}\index{parameter space|seealso{microscopic perspective}}. Note that the computational cost of gradient descent scales linearly with the size of the dataset $\A$, as one just needs to compute the gradient for each sample and then add them up. 

\index{optimization|seealso{Newton's method}}\index{optimization|seealso{direct optimization}}

For sufficiently small learning rates, the updates \eqref{eq:gd-update} are guaranteed to decrease the training loss $\L_\A$. In order to see this, let us Taylor-expand the training loss\index{loss!training loss} around the current value of the parameters $\theta(t)$ and compute the change in the loss after making an update
\be\label{eq:gd-decreases-loss}
\Delta\L_\A \equiv \L_\A\Big(\theta(t+1)\Big) -\L_\A\Big(\theta(t)\Big) = - \eta  \sum_{\mu}\le(\frac{\td \L_\A}{\td \theta_\mu}\ri)^2\Bigg\vert_{\theta=\theta(t)} + O(\eta^2) \, .
\ee 
As minus a sum of squares, this is strictly negative. Pretty typically, iterating these updates will eventually lead to (at least) a local minimum of the training loss\index{loss!training loss}. In practice, small variants of the gradient descent algorithm are responsible for almost all \terminate{training} and optimization in \terminate{deep learning}.\footnote{In particular, the most popular \terminate{learning algorithm} is \term{stochastic gradient descent} (SGD). SGD uses updates of the form\index{SGD|see{stochastic gradient descent}}\index{gradient descent!stochastic|see{stochastic gradient descent}}
\be\label{eq:sgd-update}
\theta_\mu(t+1) = \theta_\mu(t) - \eta \frac{\td \L_{\mathcal{S}_t}}{\td \theta_\mu}\Bigg|_{\theta_\mu = \theta_\mu(t)} \,,
\ee
where $\mathcal{S}_t$ is a subset of the \terminate{training set}, $\mathcal{S}_t \subset \A$. 
Each subset $\mathcal{S}_t$ is called a \emph{mini-batch}\index{mini-batch|see{batch}} or \term{batch}\index{batch|seealso{stochastic gradient descent}}. Training\index{training} is then organized by \term{epoch}\index{epoch|seealso{stochastic gradient descent}}, which is a complete passes through the training set. Typically, for each epoch the training set is \emph{stochastically} partitioned into subsets of equal size, which are then sequentially used to estimate the gradient. 

The advantage of this algorithm is twofold: \emph{(i)} the computational cost of training now scales with the fixed size of the sets $\mathcal{S}_t$ rather than with the size of the whole training set $\mathcal{A}$; and \emph{(ii)} SGD is thought to have better \terminate{generalization} properties than gradient descent. Nevertheless, essentially everything we will say about gradient descent
will apply to stochastic gradient descent
as well.
}

\subsubsection{Tensorial Gradient Descent}\index{tensor!learning-rate tensor|see{learning rate}}\index{tensorial gradient descent|see{gradient descent}}
In one such variant, we can define a more general family of learning algorithms by modifying the update \eqref{eq:gd-update} as \index{learning algorithm}\index{gradient descent!tensorial}\index{learning rate!global}
\be\label{eq:gd-update-lambda}
\theta_\mu(t+1) =\theta_\mu(t) - \eta \sum_{\nu}\lambda_{\mu\nu} \frac{\td \L_\A}{\td \theta_\nu}\Bigg\vert_{\theta=\theta(t)} \, ,
\ee
where the \terminate{tensor} $\lambda_{\mu\nu}$ is a \textbf{learning-rate tensor}\index{learning rate!learning-rate tensor|textbf} on \terminate{parameter space}; the original gradient-descent update~\eqref{eq:gd-update} is a special case with the \terminate{Kronecker delta} as the tensor $\lambda_{\mu\nu}=\delta_{\mu\nu}$. While in  the original gradient descent~\eqref{eq:gd-update} we have one global learning rate $\eta$, in the tensorial gradient descent~\eqref{eq:gd-update-lambda}  we have the freedom to separately specify how the $\nu$-th component of the gradient $\td \L_{\A} / \td \theta_\nu$ 
contributes to the update of the $\mu$-th parameter $\theta_\mu$ via the  tensor $\lambda_{\mu\nu}$.
Repeating the same Taylor-expansion in $\eta$ \eqref{eq:gd-decreases-loss} with the generalized update \eqref{eq:gd-update-lambda}, we find
\be\label{eq:change-in-loss-lambda}
 \Delta\L_\A = - \eta \sum_{\mu,\nu} \lambda_{\mu\nu}  \frac{\td \L_\A}{\td \theta_\mu} \frac{\td  \L_\A}{\td \theta_\nu}  + O(\eta^2)\, ,
\ee
indicating that the training loss\index{loss!training loss} again is almost surely decreasing for sufficiently small learning rates, so long as the learning-rate tensor $\lambda_{\mu\nu}$ is a \terminate{positive semidefinite matrix}. \index{learning rate}

\subsubsection{Neural Tangent Kernel}\label{sec:nth}
Everything we have said so far about \terminate{gradient descent} could be applied equally to the optimization of any
function.
However, in the context of \terminate{function approximation}
there is additional structure: the optimization objective is a function of the model output.

To take advantage of this structure,  first note that by the \terminate{chain rule} the gradient of the \terminate{loss} can be expressed as
\be\label{eq:gd-update-decomposed}
\frac{\td \L_\A}{\td \theta_\mu}=\sum_{i=1}^{n_{\text{out}}} \sum_{\tra \in \A} \frac{\partial \L_\A}{\partial z_{i;\tra}}\frac{\td z_{i;\tra}}{\td \theta_\mu}\, ,
\ee
which means that the change in the loss~\eqref{eq:change-in-loss-lambda} after an update  can be nicely decomposed as
\be\label{eq:change-in-loss-NTH}
\Delta\L_\A= - \eta\sum_{i_1,i_2=1}^{n_{\text{out}}} \sum_{\tra_1,\tra_2 \in \A} \le[\frac{\partial \L_\A}{\partial z_{i_1;\tra_1}}\frac{\partial  \L_\A}{\partial z_{i_2;\tra_2}}\ri]  \le[ \sum_{\mu,\nu} \lambda_{\mu\nu} \frac{\td z_{i_1;\tra_1} }{\td \theta_\mu} \frac{\td z_{i_2;\tra_2} }{\td \theta_\nu}  \ri]  + O(\eta^2)\,  .
\ee
The quantity in the first square bracket is a measure of the \terminate{function approximation} error. For instance, for the MSE loss~\eqref{eq:MSE-loss} we see that the gradient of the loss with respect to the model output is exactly the prediction error, \index{loss!MSE}\index{error factor!MSE loss}
\be\label{eq:mse-function-approximation-error}
\frac{\partial \L_\A}{\partial z_{i;\tra}}=z_i(x_{\tra}; \theta)-\y{i}{\tra} \, .
\ee
More generally for other losses, the gradient of the loss or \term{error factor}
\be\label{eq:grad-loss-def}
\epsilon_{i;\tra}\equiv \frac{\partial \L_\A}{\partial z_{i;\tra}}\, ,
\ee
is small when the model output is close to the label.
Sensibly, the greater the 
error factor,
the larger the update \eqref{eq:gd-update-decomposed}, and the greater the change in the loss \eqref{eq:change-in-loss-NTH}.
The quantity in the second square bracket is called the \term{neural tangent kernel} (NTK)\index{NTK|see{neural tangent kernel}}
\be\label{eq:NTH-definition}
\NTKM_{i_1i_2;\tra_1 \tra_2} \equiv \sum_{\mu,\nu} \lambda_{\mu\nu} \frac{\td z_{i_1;\tra_1} }{\td \theta_\mu} \frac{\td z_{i_2;\tra_2} }{\td \theta_\nu}  \, .
\ee
As is clear from \eqref{eq:NTH-definition}, the NTK is independent of the auxiliary \terminate{loss} function.

Importantly, the NTK is the main driver of the function-approximation\index{function approximation} dynamics.\index{supervised learning}
To the point, it governs the evolution of a much more general set of observables than the training loss\index{loss!training loss}.
Consider any \terminate{observable} that depends on the model's outputs
\be\label{eq:observable-last-layer}
\O\!\le(\theta \ri) \equiv \O\Big( z\!\le(x_{\delta_1} ; \theta \ri)\!,\,\ldots\,,z\!\le(x_{\delta_M} ; \theta \ri) \Big) \, , 
\ee
where $x_{\delta_1}, \ldots, x_{\delta_M} \in \D$  for some dataset\index{input data} $\D$.
For example,
if $\D$ is the test set $\B$ and $\O$  is the \terminate{loss} function, then this \terminate{observable} would be the test loss\index{loss!test loss} $\L_{\B}$.
In addition to the test loss\index{loss!test loss}, one might want to observe the change in a particular component of the output $\O = z_i(x)$
or perhaps track correlations among different vectorial components of the output $\O = z_i(x)\, z_j(x)$
for a given input $x$.
For any such \terminate{observable} \eqref{eq:observable-last-layer}, its change after an update is given by the expression
\be\label{eq:change-in-observable-NTH}
\O\Big(\theta(t+1)\Big) - \O\Big(\theta(t)\Big) = - \eta\sum_{i_1,i_2=1}^{n_{\text{out}}} \sum_{\tra \in \A} \sum_{\delta \in \D} \le[\frac{\partial \L_\A}{\partial z_{i_1;\tra}}\frac{\partial \O}{\partial z_{i_2;\delta}}\ri]  \NTKM_{i_1 i_2;\tra\delta}  + O(\eta^2)\, .
\ee
As we see, the square bracket contains the function-approximation error as well as the particulars about how the \terminate{observable} depends on the model output.
In contrast, the NTK contains all the dynamical information pertaining to the particular model, depending only on the model \terminate{architecture} and parameters.\footnote{As our discussion makes clear, the NTK can generally be defined for any function approximator. This means that its name masks its true generality. In addition to objecting to the ``neural'' part of the name, one could object to the ``kernel'' part. In particular, the NTK is more akin to a \terminate{Hamiltonian} than a kernel as it generates the evolution of observables; we'll fully justify this claim in \S\ref{subsec:real-GD-at-finite-width}.\label{footnote:ntk-name}}
\index{neural tangent kernel!name}\index{kernel!NTK|see{neural tangent kernel}}\index{function approximation}\index{model parameters}

\index{neural indices}\index{sample indices}\index{loss!MSE}
We can further understand the function-approximation\index{function approximation} dynamics under \terminate{gradient descent} by considering a particular vectorial component of the output for a particular sample as an \terminate{observable}, i.e.~$\O=z_{i}(x_{\delta})$. In this case, the derivative of $\O$ in \eqref{eq:change-in-observable-NTH} is a \terminate{Kronecker delta} on both the \terminate{vectorial indices} and \terminate{sample indices}, and the evolution reduces to
\be\label{eq:NTK-change-in-output}
z_{i}\Big(x_{\delta}; \theta(t+1)\Big) - z_{i}\Big(x_{\delta}; \theta(t)\Big)= - \eta\sum_{j=1}^{n_{\text{out}}} \sum_{\tra \in \A}     \NTKM_{i j;\delta\tra}\epsilon_{j;\tra}  + O(\eta^2)\, .
\ee
This equation shows how the model output changes after a training update. Importantly, we see how the 
\terminate{error factor}
$\epsilon_{j;\tra}$~\eqref{eq:grad-loss-def} from example $x_{\tra}$ on the model output component $j$ affects the updated behavior of the model output component $i$ on a different example $x_{\delta}$: it's mediated by the NTK component $\NTKM_{i j;\delta\tra}$. This is what makes \terminate{function approximation} possible; the ability to learn something about one example, $x_{\delta}$, by observing another, $x_{\tra}$. We see that the off-diagonal components of the NTK in the \terminate{sample indices} determine the \terminate{generalization} behavior of the model, while the off-diagonal components in \terminate{vectorial indices} allow for one \terminate{feature} to affect the training of another feature. We will have more to say about the former property in \S\ref{ch:NTHb} and the latter property in \S\ref{ch:eot}.

\index{loss!training loss}\index{learning rate}\index{loss!test loss}\index{generalization}\index{overfitting}
Finally, let us note in passing that unlike the case of the training loss~\eqref{eq:change-in-loss-NTH}, for general observables\index{observable} \eqref{eq:change-in-observable-NTH} the term in the square bracket is not necessarily positive. While the training loss\index{loss!training loss} $\L_\A$ will always decrease for small enough learning rates, a given observable may not.
In particular, nothing guarantees that the test loss\index{loss!test loss} will decrease and -- for models that overfit their \terminate{training set} -- the test loss\index{loss!test loss} may even increase.

%% file: Chp8-NTK/8_global.tex
\chapter{RG Flow of the Neural Tangent Kernel}\index{representation group flow!of the NTK}
\label{ch:NTKa}

\epigraph{People get things backwards and they shouldn't---it has been said, and wisely said, that every successful physical theory swallows its predecessors alive.}{Sidney Coleman,
more forward and a little bit deeper in that same \\
``Quantum Mechanics\index{quantum mechanics} in Your Face'' Dirac\index{Dirac, Paul Adrien Maurice} Lecture  \cite{Coleman:2020put}.\index{Coleman, Sidney}}

\noindent{}In the last chapter, we introduced gradient-based learning as an alternative to Bayesian learning and specifically focused on the \terminate{gradient descent} algorithm.
In short, the gradient descent algorithm involved instantiating a network from the prior distribution and then repeatedly updating the model parameters by running training data through the network. 
This algorithm is straightforward to implement and very efficient to run for any particular network.
In practice, it makes things very easy.

In theory, it makes things a little more difficult. %
For the Bayesian prior, we were able to integrate out the model parameters layer by layer in deriving the output distribution because the initialization distribution\index{initialization distribution} of the biases and weights was extremely simple; in addition,
the large-width expansion made 
it possible to derive analytic expressions for the Bayesian posterior for finite-width networks.
By contrast, the model parameters and the outputs of any particular
network trained by gradient descent are a complicated correlated mess.

\index{typicality}
To make progress, we first need to shift the perspective back to a statistical one. Rather than focusing on how any particular network learns from the data, we instead ask how a \emph{typical} network behaves when being trained. 
If we understand the typical behavior (i.e.~the mean) under gradient descent and have control of the fluctuations from network instantiation to instantiation (i.e.~the variance), then we can describe gradient-based learning as used in practice.

With that statistical perspective in mind, recall from the last chapter that the gradient-descent 
updates decompose 
into an error factor times a function-approximation factor. The latter factor was dubbed the \neo{neural tangent kernel} (NTK) and conveniently summarizes the effect of the model parameters' changes on the behavior of the network.  
This means that the statistics of changes in network observables
in the initial stage of training are governed by the statistics of the NTKs at initialization.
To proceed forward, the core of the current chapter and the next will involve explicitly computing such NTK statistics for deep MLPs; we will postpone the actual analysis of neural network training -- enabled by these computations of the NTK statistics -- until~\S\ref{ch:NTHb} and~\S\ref{ch:eot}.

\index{forward equation!MLP preactivations}\index{forward equation!NTK}
In~\S\ref{sec:NTH-recursions}, we will lay the groundwork for the recursive computation of the NTK statistics. Namely, starting from the MLP iteration equation, or the \emph{forward} equation for the preactivations, we'll derive a corresponding forward equation for the NTK.
This equation is a layer-to-layer iteration equation that holds for each distinct instantiation of the model parameters.
(Here we'll also remark on how the learning-rate tensor\index{learning rate!learning-rate tensor} should be scaled with network width, an important point that is often neglected in practice.) %

\index{representation group flow}
By averaging over different instantiations, we can then use the forward equation to recursively compute the joint statistics of the NTK and the preactivations.  %
The approach taken here completely mirrors the \emph{RG-flow} approach taken in~\S\ref{ch:ngp} for the preactivations. In~\S\ref{sec:first-layer-deterministic-NTK},~\S\ref{sec:second-layer-fluctuating-NTK}, and~\S\ref{sec:deeper-layer-accumulation-NTK}, we will progressively determine the sequence of joint NTK-preactivation distributions in the first, second, and deeper layers, respectively.

\setcounter{section}{-1}
\section{Forward Equation for the NTK}\label{sec:NTH-recursions}
As we saw in the previous chapter, 
the evolution of observables $\mathcal{O}(z)$ under \terminate{gradient descent} is governed by the NTK,
\be\label{eq:NTH-definition-reprint}
H_{i_1i_2;\alpha_1\alpha_2} \equiv \sum_{\mu,\nu} \lambda_{\mu\nu} \frac{\td z_{i_1;\alpha_1}}{\td \theta_\mu} \frac{\td z_{i_2;\alpha_2}}{\td \theta_\nu}\,  ,
\ee
where $\lambda_{\mu\nu}$ is the learning-rate tensor\index{learning rate!learning-rate tensor}. %

\index{feature}\index{representation}
Specializing to MLPs, observables can depend not only  on the network's output $z_{i;\alpha}=\z{i}{\alpha}{L}$, but also on the preactivations $z_{i;\alpha}=\z{i}{\alpha}{\ell}$ in any layer. Such $\ell$-th-layer observables for $\ell <L$ tell us about the \emph{hidden-layer representations} of the network. For instance, the neural component $\O = z_i^{(\ell)}(x)$ tells us about an $\ell$-th-layer \terminate{feature} evaluated on an input $x$, while $\O = z_i^{(\ell)}(x)\, z_j^{(\ell)}(x)$ with \terminate{neural indices} $i\neq j$ tracks correlations among different features given $x$.

With similar manipulations as before, we find that an observable $\O$ that depends only on the $\ell$-th-layer preactivations 
\be\label{eq:observable-ell-layer}
\O\!\le(\theta \ri) \equiv \O\Big( z^{(\ell)}\!\le(x_{\delta_1} ; \theta \ri)\!,\,\ldots\,,z^{(\ell)}\!\le(x_{\delta_M} ; \theta \ri) \Big)\, , 
\ee
evolves after a \terminate{gradient descent} update as
\be\label{eq:obsevable-evolution-layer-ell}
\O\Big(\theta(t+1)\Big) - \O\Big(\theta(t)\Big) = - \eta\sum_{i_1,i_2=1}^{n_{\ell}} \sum_{\alpha \in \A} \sum_{\delta \in \D} \le[\frac{\td \L_\A}{\td z_{i_1;\alpha}^{(\ell)}}\frac{\partial \O}{\partial z_{i_2;\delta}^{(\ell)}}\ri]  \NTKM_{i_1 i_2;\alpha\delta}^{(\ell)}  + O(\eta^2)\, ,
\ee
where $x_{\delta_1}, \ldots, x_{\delta_M} \in \D$  for some dataset\index{input data} $\D$.\footnote{
However, note that this is not quite as simple as the expression for the evolution of the network output that we gave in the last chapter~\eqref{eq:change-in-observable-NTH}. In particular, the derivative of the loss with respect to the $\ell$-th-layer preactivations needs to be computed by the \terminate{chain rule} as
\be
\frac{\td \L_\A}{\td z_{i_1;\alpha}^{(\ell)}}=\sum_{j=1}^{n_{L}}\frac{\partial \L_\A}{\partial z_{j;\alpha}^{(L)}}\frac{\td z_{j;\alpha}^{(L)}}{\td z_{i_1;\alpha}^{(\ell)}} \, ,
\ee
with the \terminate{error factor} $\partial \L_\A / \partial z_{j;\alpha}^{(L)}$ now multiplied by the \terminate{chain-rule factor} $\td \z{j}{\alpha}{L}/\td \z{i_1}{\alpha}{\ell}$.
For observables that depend on preactivations from multiple layers, the generalization of \eqref{eq:obsevable-evolution-layer-ell} further involves additional chain-rule factors as well as a sum over NTKs from different layers.
\index{chain rule}
} 
Here, we have defined the
\textbf{$\ell$-th-layer NTK}\index{neural tangent kernel!l-th-layer@$\ell$-th-layer|textbf} as
\be\label{eq:midNTH-definition}
\Tia{H}{i_1i_2}{\alpha_1\alpha_2}{\ell} \equiv \sum_{\mu,\nu} \lambda_{\mu\nu} \frac{\td \z{i_1}{\alpha_1}{\ell}}{\td \theta_\mu} \frac{\td \z{i_2}{\alpha_2}{\ell}}{\td \theta_\nu}  \, ,
\ee
which governs the evolution of the $\ell$-th-layer observables; in terms of this notation, the output NTK is simply $\Tia{H}{i_1i_2}{\alpha_1\alpha_2}{\ell=L}$. Note that whenever we write the $\ell$-th-layer NTK as above, we will always assume that the learning-rate tensor\index{learning rate!learning-rate tensor!layer-diagonal} $\lambda_{\mu\nu}$ does not mix network parameters from different layers, though in general it can still mix the biases and weights within a layer. We will place further restrictions on this in another paragraph.

At initialization, the model parameters are sampled from their initialization distributions, and the $\ell$-th-layer NTK is a stochastic object. In order to emphasize this stochasticity, in what follows we'll decorate the NTK \emph{at initialization} with a hat: $\Tia{\NTK}{i_1i_2}{\alpha_1\alpha_2}{\ell}$. Our goal is to evaluate its statistics.

\index{initialization distribution}\index{learning rate!global}
Before we go any further, it is convenient to make a specialized choice for the learning-rate tensor\index{learning rate!learning-rate tensor} $\lambda_{\mu\nu}$.
In practice, typically $\lambda_{\mu\nu}=\delta_{\mu\nu}$, and there is only the \emph{global} learning rate $\eta$ for the entire model. %
Even in a more general setup,
a learning rate is often shared among each group of parameters that are sampled from the same distribution. Recalling that the same distribution was shared among the biases in a given layer with the same variance $\Cb{\ell}$~\eqref{eq:bias-variance-def-naive} and similarly for the weights with the \emph{rescaled} weight variance $\CW{\ell}$~\eqref{eq:weight-variance-def-naive}, this suggests 
an ansatz
for our \term{training hyperparameters}: we should decompose the learning-rate tensor\index{learning rate!learning-rate tensor} $\lambda_{\mu\nu}$ into a diagonal matrix
\be\label{eq:diag_LR}
\lambda_{\bias{i_1}{\ell} \bias{i_2}{\ell}}=\delta_{i_1i_2}\Lb{\ell}\, ,\quad \lambda_{\W{i_1j_1}{\ell} \W{i_2 j_2}{\ell}}=\delta_{i_1 i_2}\delta_{j_1 j_2}\lamWtil{\ell}\, ,
\ee
giving each group of biases in a layer the same learning rate and each group of weights in a layer the same learning rate, and allowing such learning rates to vary from layer to layer. 

Importantly, we have normalized the learning rate for a given weight $\W{i_1j_1}{\ell}$ by the width of the previous layer $n_{\ell-1}$, just as we did for the variance of the weight's \terminate{initialization distribution}. This normalization is there for much the same reason: the freedom to tune the weight learning rates separately from the bias learning rates will prove necessary for having a sensible large-width expansion.
Going forward, our \terminate{training hyperparameters} will consist of the global learning rate $\eta$
and the individual $\ell$-th-layer learning rates for the biases and weights, $\Lb{\ell}$ and $\lamW{\ell}$.

\index{neural tangent kernel!l-th-layer@$\ell$-th-layer}
Substituting our choice for $\lambda_{\mu\nu}$ back into the definition of the $\ell$-th-layer NTK~\eqref{eq:midNTH-definition}, this expression decomposes as
\be\label{eq:nth-layer-sum-definition}
\Tia{\NTH}{i_1i_2}{\alpha_1\alpha_2}{\ell} = \sum_{\ell'=1}^{\ell} \le[ \sum_{j=1}^{n_{\ell'}} \le( \Lb{\ell'}  \frac{\td \z{i_1}{\alpha_1}{\ell}}{\td \bias{j}{\ell'}} \frac{\td \z{i_2}{\alpha_2}{\ell}}{\td \bias{j}{\ell'}} + \lamWtil{\ell'}  \sum_{k=1}^{n_{\ell'-1}} \frac{\td \z{i_1}{\alpha_1}{\ell}}{\td \W{j k}{\ell'}} \frac{\td \z{i_2}{\alpha_2}{\ell}}{\td \W{j k}{\ell'}}\ri)\ri] \, .
\ee
Here, the part in the square brackets is the per-layer contribution of the model parameters to the $\ell$-th-layer NTK, treating the biases and weights separately.
We also see that our intuition above in \eqref{eq:diag_LR} was correct: the $\ell'$-th-layer weight learning rate $\LW{\ell'}$ needs to be accompanied by a factor of $1/n_{\ell'-1}$ in order to compensate for the additional summation over the $(\ell'-1)$-th layer \terminate{neural indices} in the second term as compared to the first.
Even so, the layer sum in \eqref{eq:nth-layer-sum-definition} makes this expression somewhat unwieldy and suggests that we should
search for an alternate representation.

\index{neural tangent kernel!l-th-layer@$\ell$-th-layer}
Following our analysis of the preactivations, let's try to find a recursive expression.
To that end, 
consider the $(\ell+1)$-th-layer NTK, $\Tia{\NTH}{i_1i_2}{\alpha_1\alpha_2}{\ell+1}$, and decompose the sum over layers in its definition by separating the $(\ell+1)$-th-layer term from all of the rest, giving
\begin{align}\label{eq:NTHchain}
\Tia{\NTH}{i_1i_2}{\alpha_1\alpha_2}{\ell+1} =& \sum_{j=1}^{n_{\ell+1}} \le(\Lb{\ell+1}\frac{\td \z{i_1}{\alpha_1}{\ell+1}}{\td \bias{j}{\ell+1}} \frac{\td \z{i_2}{\alpha_2}{\ell+1}}{\td \bias{j}{\ell+1}} + \frac{\lamW{\ell+1}}{n_{\ell}} \sum_{k=1}^{n_{\ell}} \frac{\td \z{i_1}{\alpha_1}{\ell+1}}{\td \W{jk}{\ell+1}} \frac{\td \z{i_2}{\alpha_2}{\ell+1}}{\td \W{jk}{\ell+1}}\ri)\, \\
&+  \sum_{j_1,j_2=1}^{n_{\ell}} \frac{\td \z{i_1}{\alpha_1}{\ell+1}}{\td \z{j_1}{\alpha_1}{\ell}} \frac{\td \z{i_2}{\alpha_2}{\ell+1}}{\td  \z{j_2}{\alpha_2}{\ell}}\Tia{\NTH}{j_1j_2}{\alpha_1\alpha_2}{\ell} \, .\nonumber
\end{align}
Here, the first line is the $(\ell+1)$-th-layer term that we left alone, while the second line gives the terms from all the other layers after applying the \terminate{chain rule} and then recalling the definition \eqref{eq:nth-layer-sum-definition}.
In this way, the $\ell$-th-layer NTK appears naturally.
This 
means that we can find a simple iterative expression for the NTK,
similar in spirit to the forward equation for the preactivations that defines the MLP.

\index{forward equation!MLP preactivations}\index{forward equation!NTK}
To finish our derivation, we need to evaluate the derivatives in \eqref{eq:NTHchain}. To do so,
recall the preactivation forward iteration equation
\be\label{eq:forward-pass}
\z{i}{\alpha}{\ell+1} = \bias{i}{\ell+1}+\sum_{j=1}^{n_{\ell}}\W{ij}{\ell+1}\s{j}{\alpha}{\ell} \, ,
\ee
and remember that the activations are explicit functions of the preactivations $\s{i}{\alpha}{\ell}\equiv\sigma\!\le(\z{i}{\alpha}{\ell}\ri)$.
The factors in the second line of \eqref{eq:NTHchain} 
coming from the \terminate{chain rule} evaluate to
\be\label{eq:chain-rule-factor}
 \frac{\td \z{i}{\alpha}{\ell+1}}{\td \z{j}{\alpha}{\ell}}= \W{ij}{\ell+1}\ds{j}{\alpha}{\ell} \, ,
\ee
while the derivatives with respect to the $(\ell+1)$-th-layer parameters evaluate to
\be\label{eq:same-layer-derivatives}
\frac{\td \z{i}{\alpha}{\ell+1}}{\td \bias{j}{\ell+1}} = \delta_{ij}\, , \qquad \frac{\td \z{i}{\alpha}{\ell+1}}{\td \W{jk}{\ell+1}} = \delta_{ij} \, \s{k}{\alpha}{\ell} \, .
\ee
All together, we can rewrite \eqref{eq:NTHchain} as
\begin{align}\label{eq:NTH-recursion-without-expectation}
\Tia{\NTH}{i_1i_2}{\alpha_1\alpha_2}{\ell+1}=&\delta_{i_1i_2} \le[\Lb{\ell+1} +  \lamW{\ell+1}\!\le(\frac{1}{n_{\ell}}\sum_{j=1}^{n_{\ell}} \s{j}{\alpha_1}{\ell} \s{j}{\alpha_2}{\ell} \ri) \ri]\, \\
&+  \sum_{j_1,j_2=1}^{n_{\ell}} \W{i_1j_1}{\ell+1} \W{i_2j_2}{\ell+1}  \ds{j_1}{\alpha_1}{\ell} \ds{j_2}{\alpha_2}{\ell} \Tia{\NTH}{j_1j_2}{\alpha_1\alpha_2}{\ell}\, .\nonumber
\end{align}
This is the \textbf{forward equation for the NTK}\index{forward equation!NTK}, which is an iteration equation that computes the NTK  layer by layer  for any realization of the biases and weights. This is analogous to the way in which \eqref{eq:forward-pass} computes the network output -- as well as all the hidden-layer preactivations -- via a layer-to-layer iteration for a given realization of model parameters.

\subsubsection{Scaling in the effective theory}
\index{forward equation!NTK}\index{learning rate!global}\index{effective theory}
The forward equation \eqref{eq:NTH-recursion-without-expectation} further clarifies our decomposition \eqref{eq:diag_LR} in which we made a distinction between the learning rates for the biases and those for the weights, giving each a different scaling with respect to the layer widths $n_\ell$ of the network.\footnote{You'll have to wait until \S\ref{ch:eft-ntk} to understand why it is advantageous to give a layer dependence to $\Lb{\ell}$ and $\lamW{\ell}$ and to learn how they should be scaled with depth.}

To see why, first recall from \S\ref{ch:training} that the change in the training loss after a step of gradient descent is proportional to the product of the global learning rate $\eta$ and the final-layer NTK $\Tia{\NTH}{i_1i_2}{\alpha_1\alpha_2}{L}$:
\be\label{eq:change-in-loss-NTH-reprint}
\Delta\L_\A= - \eta\sum_{i_1,i_2=1}^{n_{L}} \sum_{\alpha_1,\alpha_2 \in \A}   \epsilon_{i_1;\alpha_1}  \epsilon_{i_2;\alpha_2}\, \Tia{\NTH}{i_1i_2}{\alpha_1\alpha_2}{L} + O(\eta^2)\,  ,
\ee
where here we also recall the definition of the \terminate{error factor}%
\be\label{eq:error-factor-ntk}
\epsilon_{i;\alpha} \equiv \frac{\partial \L_\A}{\partial z^{(L)}_{i;\alpha}}\, .
\ee
Note that this \terminate{error factor} generally stays of order one in the large-width limit, cf.~the explicit expression when using the MSE loss~\eqref{eq:mse-function-approximation-error}. Thus, it's essential that the product of the global learning rate and the NTK, $ \eta \NTK^{(L)}$, also stays of order one for large-width networks: if it diverged as the width increases, then the higher-order terms in~\eqref{eq:change-in-loss-NTH-reprint} would dominate and the loss would no longer be guaranteed to decrease; if instead it vanished in this limit, then no training would take place. Either way, training would fail.\index{loss!MSE} 

\index{learning rate!global}\index{forward equation!NTK}
With that in mind, we chose the width scaling of our learning-rate tensor\index{learning rate!learning-rate tensor} so that the NTK naturally stays of order one in the large-width limit and hence a (sufficiently small but not parametrically small) order-one global learning $\eta$ ensures the success of training. In particular, the $(\ell+1)$-th-layer contribution in the first line of the forward equation~\eqref{eq:NTH-recursion-without-expectation} stays of order one if we take
$\Lb{\ell+1}, \LW{\ell+1}=\o{1}$,
with the $1/n_{\ell}$ normalization of the $(\ell+1)$-th-layer weight learning rate playing an essential role in compensating for the summation over the $n_{\ell}$ terms.\footnote{
With this choice, the recursive term in the second line of the forward equation~\eqref{eq:NTH-recursion-without-expectation} also stays of order one. To see this, let's evaluate its expectation:
\begin{align}
\E{\sum_{j_1,j_2=1}^{n_{\ell}} \W{i_1j_1}{\ell+1} \W{i_2j_2}{\ell+1}  \ds{j_1}{\alpha_1}{\ell} \ds{j_2}{\alpha_2}{\ell} \Tia{\NTH}{j_1j_2}{\alpha_1\alpha_2}{\ell}}&=\sum_{j_1,j_2=1}^{n_{\ell}} \E{\W{i_1j_1}{\ell+1} \W{i_2j_2}{\ell+1}}\E{ \ds{j_1}{\alpha_1}{\ell} \ds{j_2}{\alpha_2}{\ell} \Tia{\NTH}{j_1j_2}{\alpha_1\alpha_2}{\ell}}\, \notag\\
&=\delta_{i_1i_2}\,\CW{\ell+1}\!\le( \frac{1}{n_{\ell}}\sum_{j=1}^{n_{\ell}}\E{\ds{j}{\alpha_1}{\ell} \ds{j}{\alpha_2}{\ell} \Tia{\NTH}{jj}{\alpha_1\alpha_2}{\ell}}\ri)\, .
\end{align}
In particular, we see that the $1/n_{\ell}$ scaling of the initialization weight variance $\CW{\ell+1}$ is important for ensuring principled behavior of not only the network output, but also the NTK.\index{neural tangent kernel}
}

If instead we 
had
considered the original version of gradient descent with $\lambda_{\mu\nu} = \delta_{\mu\nu}$ rather than \emph{tensorial gradient descent}, \index{gradient descent!tensorial}
we would have been in trouble. In the language of our \terminate{effective theory}, the original gradient descent corresponds to setting $\Lb{\ell}=1$ and $\LW{\ell}=n_{\ell-1}$, which means that the NTK itself would be $O(n)$. We'd then have to scale the global learning rate as $\eta=O(1/n)$ %
to compensate for this $O(n)$ scaling of the NTK. However, since in this case $\eta\Lb{\ell}=O(1/n)$, the order-one contribution from the weights to the NTK would completely overwhelm the %
$1/n$-suppressed contribution from the biases. This would lead to both a lack of appropriate contribution of the biases to the updates of the %
weights
 as well as an extreme under-training of the biases themselves.

Finally,
let's make 
a general point: in any \terminate{effective theory},
it's really essential to make all large or small scales explicit -- and rescale hyperparameters\index{hyperparameters!scaling in an effective theory} accordingly -- as we did earlier for the variance of the weight \terminate{initialization distribution}, did here for the weight \terminate{learning rate}, and will do later for the depth scaling of both the bias and weight learning rates.
For the effective theorist this ensures that the asymptotic $1/n$ and $1/\ell$ expansions are sound and nontrivial,
and for the practical practitioner\index{practical practitioners} this enables comparisons of hyperparameter values across architectures with different widths and depths. In particular, we expect very generally that this should help mitigate expensive hyperparameter tuning, remove the need for heuristic fixes, and increase the robustness of optimal hyperparameter settings when scaling a model up. %
\index{$1/n$ expansion} %

\subsubsection{Getting things backwards}\index{chain rule}\index{chain-rule factor}
N.B.~the chain-rule factors \eqref{eq:chain-rule-factor} also appear when evaluating the derivative of the network outputs with respect to model parameters
\be
\frac{\td \z{i}{\alpha}{L}}{\td \bias{j}{\ell}}=\frac{\td \z{i}{\alpha}{L}}{\td \z{j}{\alpha}{\ell}}\, , \qquad  \frac{\td \z{i}{\alpha}{L}}{\td  \W{jk}{\ell}}=\sum_{m}\frac{\td \z{i}{\alpha}{L}}{\td \z{m}{\alpha}{\ell}}\frac{\td \z{m}{\alpha}{\ell}}{\td  \W{jk}{\ell}}=\frac{\td \z{i}{\alpha}{L}}{\td \z{j}{\alpha}{\ell}}\s{k}{\alpha}{\ell-1} \, . 
\ee
Evaluating these derivatives gives another neural-network iteration equation,
\be\label{eq:backward-pass}
\frac{\td \z{i}{\alpha}{L}}{\td \z{j}{\alpha}{\ell}}=\sum_{k=1}^{n_{\ell+1}}\frac{\td \z{i}{\alpha}{L}}{\td \z{k}{\alpha}{\ell+1}}\frac{\td \z{k}{\alpha}{\ell+1}}{\td \z{j}{\alpha}{\ell}}=\sum_{k=1}^{n_{\ell+1}}\frac{\td \z{i}{\alpha}{L}}{\td \z{k}{\alpha}{\ell+1}}\W{kj}{\ell+1}\ds{j}{\alpha}{\ell} \, \ \ \ \text{for}\ \ \ \ell < L\, ,
\ee
but in this case for the derivative of the output. In particular,
\eqref{eq:backward-pass} is a \emph{backward} equation: starting from the \emph{final} condition \index{backward equation!MLP}
\be
\frac{\td \z{i}{\alpha}{L}}{\td \z{j}{\alpha}{L}}=\delta_{ij}\, ,
\ee 
we iterate layer-to-layer 
backwards, $\ell=L-1, L-2, \ldots, 1$, by sequential multiplications of the chain-rule factors~\eqref{eq:chain-rule-factor}.\index{chain rule}\index{chain-rule factor}

\index{forward equation!MLP preactivations}
An algorithm based on this backward equation can be efficiently implemented to compute derivatives with respect to the model parameters and, for that reason, is used by most deep-learning packages to compute the gradient as part of any neural network gradient-based learning algorithm. Such a package typically lets practitioners specify a deep learning model by defining a \term{forward pass} -- for MLPs a practitioner would implement the forward equation \eqref{eq:forward-pass} -- and then the package will automatically work out the \term{backward pass} -- i.e.~for MLPs it would implement \eqref{eq:backward-pass}.
The computational algorithm based on \eqref{eq:backward-pass} is termed \term{backpropagation},
which was discovered and rediscovered numerous times in the history of \terminate{deep learning}.
Among them, a particular rediscovery \cite{rumelhart1985learning} was essential in convincing the \terminate{machine learning} community that multilayer neural networks can be trained efficiently.

All that said, when evaluating the NTK in the \terminate{effective theory} it's essential that we use the forward equation \eqref{eq:NTH-recursion-without-expectation}  rather than getting things backwards.
\index{forward equation!NTK}
In the next three-plus-one sections, we'll indeed use the forward equation to recursively compute the \emph{joint} \terminate{initialization distribution} for the $\ell$-th-layer preactivations \emph{and} the $\ell$-th-layer NTK:  %
\begin{align}\label{eq:joint-preactivation-NTK}
&p\!\le(z^{(\ell)}, \NTK^{(\ell)}\Big\vert \D\ri)  \equiv p
\begin{pmatrix}
z^{(\ell)}(x_1) & z^{(\ell)}(x_2) & \ldots &  z^{(\ell)}(x_{N_\D})\\
\NTK^{(\ell) }(x_1, x_1) & \NTK^{(\ell) }(x_1, x_2)& \ldots & \NTK^{(\ell) }(x_1, x_{N_\D})\\
\NTK^{(\ell) }(x_2, x_1) & \NTK^{(\ell) }(x_2, x_2) & \ldots & \NTK^{(\ell) }(x_2, x_{N_\D})\\
\vdots  & \vdots  & \ddots & \vdots  \\
\NTK^{(\ell) }(x_{N_\D}, x_1) & \NTK^{(\ell) }(x_{N_\D}, x_2) & \ldots & \NTK^{(\ell) }(x_{N_\D}, x_{N_\D}) \\ 
\end{pmatrix} \, .  \notag \\
\end{align}
(On the right-hand side, we've
suppressed \terminate{neural indices} while explicitly writing out the input dependence.
This emphasizes that the preactivations are each functions of a single input and that the NTK components are each functions of a pair of inputs.)

\section{First Layer: Deterministic NTK}
\label{sec:first-layer-deterministic-NTK}\index{neural tangent kernel!first-layer}
Recall from~\S\ref{sec:first-layer-gaussian} that at initialization the first-layer preactivations,
\be\label{eq:first-layer-preactivation-def-reprint}\index{forward equation!MLP preactivations}
\z{i}{\alpha}{1} \equiv \bias{i}{1}+\sum_{j=1}^{n_{0}}\W{ij}{1}\x{j}{\alpha}\, , %
\ee
are distributed according to a zero-mean \terminate{Gaussian distribution},
\be\label{eq:first-layer-distribution-reprint}
p\!\le(z^{(1)}\Big\vert \D\ri)= \frac{1}{\dete{2\pi G^{(1)}}^{\frac{n_1}{2}}}\exp\!\le(-\frac{1}{2}\sum_{i=1}^{n_1}\sum_{\alpha_1,\alpha_2\in\D}\Kinv{\alpha_1 \alpha_2}{1}\z{i}{\alpha_1}{1}\z{i}{\alpha_2}{1}\ri)\, ,
\ee
with the first-layer deterministic metric -- a function of the inputs -- given by \index{action!quadratic}
\be\label{eq:first-layer-metric-reprint}
\Ti{G}{\alpha_1 \alpha_2}{1}\equiv\Cb{1}+\CW{1}\frac{1}{n_0}\sum_{j=1}^{n_0}\x{j}{\alpha_1}\x{j}{\alpha_2}\,  .
\ee
In particular, the quadratic action in the exponent of \eqref{eq:first-layer-distribution-reprint} indicates the absence of \terminate{interactions} between neurons.
This enables us to factor expectation values of first-layer observables into separate Gaussian integrals for each neuron.

\index{neural tangent kernel!first-layer}
The first-layer NTK at initialization is even more trivial and can be read off from the original definition of the NTK~\eqref{eq:nth-layer-sum-definition} by plugging in the derivatives \eqref{eq:same-layer-derivatives} and remembering the identification $\s{i}{\alpha}{0}=\x{i}{\alpha}$:
\be\label{eq:NTHinitial}
\Tia{\NTK}{i_1i_2}{\alpha_1\alpha_2}{1}=\delta_{i_1i_2} \le[\Lb{1} + \lamW{1} \le(\frac{1}{n_0}\sum_{j=1}^{n_{0}}\x{j}{\alpha_1} \x{j}{\alpha_2}\ri)  \ri]\equiv\delta_{i_1i_2}\Ti{\NTKM}{\alpha_1\alpha_2}{1}\, .
\ee
Like the first-layer metric,
the first-layer NTK is completely deterministic -- hence no hat on the right-hand side of the equation -- 
and is diagonal in its \terminate{neural indices}. Remembering our exposition on the off-diagonal components of the NTK in \S\ref{sec:gd}, this in particular means that, for single-layer networks, a feature captured by a particular neuron cannot affect the gradient-descent update for another feature on any other neuron.
\index{neural tangent kernel!first-layer}\index{gradient descent}

\index{neural tangent kernel!first-layer}
Finally, recalling our discussion of deterministic distributions in \S\ref{sec:MLP_distribution}, the joint distribution of the first-layer preactivations and the first-layer NTK can be written as
\be
p\!\le(z^{(1)},\NTK^{(1)}\Big\vert \D\ri)=p\!\le(z^{(1)}\Big\vert \D\ri) \prod_{(i_1i_2),(\alpha_1\alpha_2)}\delta\!\le(\Tia{\NTK}{i_1i_2}{\alpha_1\alpha_2}{1}-\delta_{i_1i_2}\Ti{\NTKM}{\alpha_1\alpha_2}{1}\ri)\, ,
\ee
where the product of the \terminate{Dirac delta function}s runs over all pairs of \terminate{neural indices} and \terminate{sample indices}.  Just as the first-layer preactivation distribution was representative of deeper layers in the \terminate{infinite-width limit}, this first-layer joint distribution is also representative of deeper-layer joint distributions in the infinite-width limit: the preactivation distribution is exactly Gaussian, the NTK distribution is completely deterministic, and there is no correlation between the two, i.e., they are statistically independent from each other once the dataset is fixed.
\index{statistical independence}

\section{Second Layer: Fluctuating NTK}
\label{sec:second-layer-fluctuating-NTK}\index{forward equation!MLP preactivations}\index{neural tangent kernel!second-layer}
Now, let us see how finite-width corrections can modify this picture in the second layer.

Recall from~\S\ref{sec:second-layer-non-gaussian} that the second-layer preactivations are given by
\be\label{eq:second-layer-preactivations-reprint}\index{forward equation!MLP preactivations}
\z{i}{\alpha}{2}=\bias{i}{2}+\sum_{j=1}^{n_{1}}\W{ij}{2}\s{j}{\alpha}{1}\, .%
\ee
After \terminate{marginalizing over} the first-layer preactivations $z^{(1)}$, the correlated fluctuations of the preactivations in the first layer resulted in nontrivial interaction\index{interactions} between different neurons in the second layer.
At the leading nontrivial order in $1/n_1$, this led to a \terminate{nearly-Gaussian distribution} with a quartic action~\eqref{eq:second-layer-quartic-action-in-SD-subsubsection} for the second-layer preactivations, with the leading non-Gaussianity captured by a nonzero connected four-point correlator.\index{connected correlator!four-point} %

\index{forward equation!NTK}\index{neural tangent kernel!second-layer}\index{neural tangent kernel!first-layer}
As for the NTK, looking at its forward equation~\eqref{eq:NTH-recursion-without-expectation} and 
recalling that the first-layer NTK is deterministic~\eqref{eq:NTHinitial}, we see that
the second-layer NTK is given by
\be\label{eq:NTK-second-stochastic}
\Tia{\NTH}{i_1i_2}{\alpha_1\alpha_2}{2}=\delta_{i_1i_2} \!\le[\Lb{2} + \lamW{2}\le(\frac{1}{n_{1}}\sum_{j=1}^{n_{1}} \s{j}{\alpha_1}{1} \s{j}{\alpha_2}{1}\ri)  \ri] +  \sum_{j=1}^{n_{1}} \W{i_1 j }{2} \W{i_2 j}{2} \, \ds{j}{\alpha_1}{1} \ds{j}{\alpha_2}{1} \,\Ti{\NTKM}{\alpha_1\alpha_2}{1}\, .
\ee
This second-layer NTK depends on two sets of stochastic variables, the weights $W^{(2)}_{ij}$ and the first-layer preactivations $\z{i}{\alpha}{1}$, and hence it fluctuates.

To compute its mean we take an expectation of \eqref{eq:NTK-second-stochastic}, finding
\begin{align}\label{eq:NTK-second-mean}
&\E{\Tia{\NTK}{i_1i_2}{\alpha_1\alpha_2}{2}} \\
=&\delta_{i_1i_2} \!\le[\Lb{2} + \lamW{2}\le(\frac{1}{n_1}\sum_{j=1}^{n_{1}} \E{\s{j}{\alpha_1}{1} \s{j}{\alpha_2}{1}}\ri)\ri]\, \notag +\sum_{j=1}^{n_{1}} \E{\W{i_1j}{2} \W{i_2j}{2}} \E{\ds{j}{\alpha_1}{1} \ds{j}{\alpha_2}{1}}\Ti{\NTKM}{\alpha_1\alpha_2}{1}\, \nonumber\\
=&\delta_{i_1i_2} \le[\Lb{2} + \lamW{2}\bra\sigma_{\alpha_1} \sigma_{\alpha_2}\ket_{G^{(1)}}  +  \CW{2}\bra\sigma^{\prime}_{\alpha_1} \sigma^{\prime}_{\alpha_2}\ket_{G^{(1)}}\Ti{\NTKM}{\alpha_1\alpha_2}{1}\ri] \notag\\
\equiv& \delta_{i_1i_2}\, \Ti{\NTKM}{\alpha_1\alpha_2}{2}\, .\nonumber
\end{align}
Here, in the second line, the expectation of the recursive term
factorized because the second-layer weights $W^{(2)}_{ij}$ are 
statistically independent from the first-layer preactivations. Additionally, in the third line we recalled \eqref{eq:two-activations-Gauss}, in which we showed that the two-point correlators can be expressed as a separate Gaussian expectations\index{Gaussian expectation} 
for each neuron, with the variance given by the first-layer metric $G^{(1)}$.\footnote{Note that the logic around \eqref{eq:two-activations-Gauss} is the same whether or not the Gaussian expectation is of activations or derivatives of the activation. In other words, for the first-layer preactivations we also have $\E{\ds{j}{\alpha_1}{1} \ds{j}{\alpha_2}{1}}= \bra \sigma^{\prime}_{\alpha_1}\sigma^{\prime}_{\alpha_2}\ket_{G^{(1)}}$.}
Further, inspecting our answer \eqref{eq:NTK-second-mean}, we see that the mean of the second-layer NTK is diagonal in its \terminate{neural indices}. Furthermore, we separated the part that encodes the sample dependence and symbolized it by taking off its hat because it is a mean, not a stochastic variable.
\index{Gaussian expectation}\index{statistical independence}\index{neural tangent kernel!second-layer}

\index{neural tangent kernel!second-layer}\index{fluctuations}\index{correlator!two-point}\index{correlator!four-point}\index{neural tangent kernel!variance}
Now, let's compute the variance.
First, define the second-layer NTK fluctuation through our usual decomposition,\index{tensor decomposition!NTK mean and fluctuation}
\be
\Tia{\NTK}{i_1i_2}{\alpha_1\alpha_2}{2}\equiv \delta_{i_1i_2} \Ti{\NTKM}{\alpha_1\alpha_2}{2}+\DNTK{i_1i_2}{\alpha_1\alpha_2}{2}\, ,
\ee
so that the expectation of the magnitude of this fluctuation determines the covariance:
 \be
\E{\DNTK{i_1i_2}{\alpha_1\alpha_2}{2}\DNTK{i_3i_4}{\alpha_3\alpha_4}{2}}=\E{\Tia{\NTK}{i_1i_2}{\alpha_1\alpha_2}{2}\Tia{\NTK}{i_3i_4}{\alpha_3\alpha_4}{2}}-\E{\Tia{\NTK}{i_1i_2}{\alpha_1\alpha_2}{2}}\E{\Tia{\NTK}{i_3i_4}{\alpha_3\alpha_4}{2}}\, .
\ee
Substituting in our expression \eqref{eq:NTK-second-stochastic} for the second-layer stochastic NTK and using the independence of the second-layer weights from the first-layer preactivations, we find a complicated-looking result
\begin{align}\label{eq:second-layer-ntk-pre-omega}
&\E{\DNTK{i_1i_2}{\alpha_1\alpha_2}{2}\DNTK{i_3i_4}{\alpha_3\alpha_4}{2}}\, \\
=&\frac{1}{n_1}\delta_{i_1i_2}\delta_{i_3i_4}\Bigg\{\le(\lamW{2}\ri)^2\Big[\bra\sigma_{\alpha_1}\sigma_{\alpha_2}\sigma_{\alpha_3}\sigma_{\alpha_4}\ket_{G^{(1)}}-\bra\sigma_{\alpha_1}\sigma_{\alpha_2}\ket_{G^{(1)}}\bra\sigma_{\alpha_3}\sigma_{\alpha_4}\ket_{G^{(1)}}\Big]\, \nonumber\\
&\quad \quad \quad \quad \quad +\CW{2}\Ti{\NTKM}{\alpha_1\alpha_2}{1}\lamW{2}\le[\bra\sigma^{\prime}_{\alpha_1}\sigma^{\prime}_{\alpha_2}\sigma_{\alpha_3}\sigma_{\alpha_4}\ket_{G^{(1)}}-\bra\sigma^{\prime}_{\alpha_1}\sigma^{\prime}_{\alpha_2}\ket_{G^{(1)}}\bra\sigma_{\alpha_3}\sigma_{\alpha_4}\ket_{G^{(1)}}\ri]\, \nonumber\\
&\quad \quad \quad \quad \quad +\lamW{2}\CW{2}\Ti{\NTKM}{\alpha_3\alpha_4}{1}\le[\bra\sigma_{\alpha_1}\sigma_{\alpha_2}\sigma^{\prime}_{\alpha_3}\sigma^{\prime}_{\alpha_4}\ket_{G^{(1)}}-\bra\sigma_{\alpha_1}\sigma_{\alpha_2}\ket_{G^{(1)}}\bra\sigma^{\prime}_{\alpha_3}\sigma^{\prime}_{\alpha_4}\ket_{G^{(1)}}\ri]\, \nonumber\\
&\quad \quad \quad \quad \quad +\le(\CW{2}\ri)^2\Ti{\NTKM}{\alpha_1\alpha_2}{1}\Ti{\NTKM}{\alpha_3\alpha_4}{1}\le[\bra\sigma^{\prime}_{\alpha_1}\sigma^{\prime}_{\alpha_2}\sigma^{\prime}_{\alpha_3}\sigma^{\prime}_{\alpha_4}\ket_{G^{(1)}}-\bra\sigma^{\prime}_{\alpha_1}\sigma^{\prime}_{\alpha_2}\ket_{G^{(1)}}\bra\sigma^{\prime}_{\alpha_3}\sigma^{\prime}_{\alpha_4}\ket_{G^{(1)}}\ri]\Bigg\}\, \nonumber\\
&+\frac{1}{n_1}\le(\delta_{i_1i_3}\delta_{i_2i_4}+\delta_{i_1i_4}\delta_{i_2i_3}\ri)\le(\CW{2}\ri)^2\Ti{\NTKM}{\alpha_1\alpha_2}{1}\Ti{\NTKM}{\alpha_3\alpha_4}{1}\bra\sigma^{\prime}_{\alpha_1}\sigma^{\prime}_{\alpha_2}\sigma^{\prime}_{\alpha_3}\sigma^{\prime}_{\alpha_4}\ket_{G^{(1)}}\, .\nonumber
\end{align}
To get this expression, we recalled not only \eqref{eq:two-activations-Gauss} for the two-point correlators, 
but also both \eqref{eq:four-activations-one-neuron-Gauss} and \eqref{eq:four-activations-two-neurons-Gauss} for the different pairings of the four-point correlators, with the pairings depending on whether all activations are on the same neuron or are on two different neurons, respectively.
As with our computation of the mean above,
the computations of these four-point correlators proceed similarly regardless of whether an activation has a derivative or not.

\index{neural tangent kernel!second-layer}
To help make sense of this rather ugly expression \eqref{eq:second-layer-ntk-pre-omega}, let's first decompose the second-layer NTK variance\index{neural tangent kernel!variance} into a sum of two different types of tensors
\begin{align}\label{eq:NTH-variance-decomposition-second}\index{tensor decomposition!NTK variance $A$/$B$}
&\E{\DNTK{i_1i_2}{\alpha_1\alpha_2}{2}\DNTK{i_3i_4}{\alpha_3\alpha_4}{2}}\, \\
\equiv&\frac{1}{n_1}\le[\delta_{i_1i_2}\delta_{i_3i_4} \NTHA{\alpha_1\alpha_2}{\alpha_3\alpha_4}{2}+\delta_{i_1i_3}\delta_{i_2i_4}\NTHB{\alpha_1\alpha_3\alpha_2\alpha_4}{2}+\delta_{i_1i_4}\delta_{i_2i_3}\NTHB{\alpha_1\alpha_4\alpha_2\alpha_3}{2}\ri]\, . \nonumber
\end{align}
This decomposition was motivated by the pattern of \terminate{Kronecker delta}s that appear in \eqref{eq:second-layer-ntk-pre-omega}. Next, by comparing this 
to our original expression \eqref{eq:second-layer-ntk-pre-omega}, we 
see that these tensors are given by 
\begin{align}
\NTHA{\alpha_1\alpha_2}{\alpha_3\alpha_4}{2}=& \bra \Oi{\alpha_1\alpha_2}{2} \Oi{\alpha_3\alpha_4}{2} \ket_{G^{(1)}}-\bra \Oi{\alpha_1\alpha_2}{2}\ket_{G^{(1)}}\bra \Oi{\alpha_3\alpha_4}{2} \ket_{G^{(1)}}\, ,\label{eq:A-recursion-second}\\
\NTHB{\alpha_1\alpha_3\alpha_2\alpha_4}{2}=& \le(\CW{2}\ri)^2\Ti{\NTKM}{\alpha_1\alpha_2}{1}\Ti{\NTKM}{\alpha_3\alpha_4}{1}\bra\sigma^{\prime}_{\alpha_1}\sigma^{\prime}_{\alpha_2}\sigma^{\prime}_{\alpha_3}\sigma^{\prime}_{\alpha_4}\ket_{G^{(1)}}\, ,\label{eq:B-recursion-second}
\end{align}
where on the first line we've introduced an auxiliary stochastic variable,
\be\label{eq:def-omega-without-neural-second}
\Oi{\alpha_1\alpha_2}{2} \equiv \lamW{2} \, \Ti{\sigma}{\alpha_1}{1} \Ti{\sigma}{\alpha_2}{1} + \CW{2}\Ti{\NTKM}{\alpha_1\alpha_2}{1}\, \dTi{\sigma}{\alpha_1}{1} \dTi{\sigma}{\alpha_2}{1}  \, ,
\ee
in order to remedy the ugliness of what would have otherwise been a very long expression.\footnote{Note that $\NTHA{\alpha_1\alpha_2}{\alpha_3\alpha_4}{2}$~\eqref{eq:A-recursion-second} has the same symmetries as the \terminate{four-point vertex} $\V{\alpha_1\alpha_2}{\alpha_3\alpha_4}{\ell}$. In particular, it's symmetric under exchanges of \terminate{sample indices} $\alpha_1\leftrightarrow \alpha_2$, $\alpha_3\leftrightarrow \alpha_4$, and $(\alpha_1\alpha_2) \leftrightarrow (\alpha_3\alpha_4)$, and this symmetry persists to deeper layers, cf.~\eqref{eq:A-recursion}.

Now, you might raise your hand and say that $\NTHB{\alpha_1\alpha_3\alpha_2\alpha_4}{2}$~\eqref{eq:B-recursion-second} also respects the same symmetry. That's correct here, but this symmetry will be broken in deeper layers. In general -- cf.~\eqref{eq:B-recursion} -- $\NTHB{\alpha_1\alpha_3\alpha_2\alpha_4}{\ell}$ will be symmetric under  $(\alpha_1\alpha_2) \leftrightarrow (\alpha_3\alpha_4)$ and $(\alpha_1\alpha_3) \leftrightarrow (\alpha_2\alpha_4)$ but \emph{not} under $\alpha_1\leftrightarrow \alpha_2$ or $\alpha_3\leftrightarrow \alpha_4$ individually.}
From \eqref{eq:A-recursion-second} and \eqref{eq:B-recursion-second} we see clearly that these tensors are of order one. Given \eqref{eq:NTH-variance-decomposition-second}, this in turn means that the second-layer NTK variance\index{neural tangent kernel!variance} is suppressed by $1/n_1$ in the large-width limit. In other words, the second-layer NTK is deterministic in the strict \terminate{infinite-width limit}, but in backing off that limit it fluctuates according to \eqref{eq:NTH-variance-decomposition-second}, \eqref{eq:A-recursion-second}, and \eqref{eq:B-recursion-second}.

\index{correlator!two-point}\index{correlator!four-point}
Moreover, at finite width the second-layer NTK not only fluctuates, but also has nontrivial cross correlation\index{cross correlation!NTK-preactivation} with the second-layer preactivations. 
This can ultimately be traced to the fact that the second-layer preactivations~\eqref{eq:second-layer-preactivations-reprint} and the second-layer NTK~\eqref{eq:NTK-second-stochastic} 
are both functions of the same stochastic variables: the second-layer weights $W^{(2)}_{ij}$ and the first-layer preactivations $\z{i}{\alpha}{1}$.

This cross correlation can be computed analogously to the way we computed the NTK mean and variance. 
Substituting in the definition of the second-layer preactivations~\eqref{eq:second-layer-preactivations-reprint} and the second-layer NTK~\eqref{eq:NTK-second-stochastic}, and again using the \terminate{statistical independence} of the second-layer weights $W^{(2)}_{ij}$ from the first-layer preactivations $\z{i}{\alpha}{1}$, we find
\begin{align}\label{eq:unrolling-cross-first}
&\E{\z{i_1}{\alpha_1}{2}\DNTK{i_2i_3}{\alpha_2\alpha_3}{2}} = 0 \, ,\\
\label{eq:unrolling-cross-second}
&\E{\z{i_1}{\alpha_1}{2}\z{i_2}{\alpha_2}{2}\DNTK{i_3i_4}{\alpha_3\alpha_4}{2}}=\E{\z{i_1}{\alpha_1}{2}\z{i_2}{\alpha_2}{2}\Tia{\NTK}{i_3i_4}{\alpha_3\alpha_4}{2}}-\E{\z{i_1}{\alpha_1}{2}\z{i_2}{\alpha_2}{2}}\E{\Tia{\NTK}{i_3i_4}{\alpha_3\alpha_4}{2}}\, \notag\\
=&\frac{1}{n_1}\delta_{i_1i_2}\delta_{i_3i_4}\Bigg\{\lamW{2}\CW{2}\Big[\bra\sigma_{\alpha_1}\sigma_{\alpha_2}\sigma_{\alpha_3}\sigma_{\alpha_4}\ket_{G^{(1)}}-\bra\sigma_{\alpha_1}\sigma_{\alpha_2}\ket_{G^{(1)}}\bra\sigma_{\alpha_3}\sigma_{\alpha_4}\ket_{G^{(1)}}\Big]\, \nonumber\\
&\quad \quad \quad \quad \quad +\le(\CW{2}\ri)^2\Ti{\NTKM}{\alpha_3\alpha_4}{1}\Big[\bra\sigma_{\alpha_1}\sigma_{\alpha_2}\sigma^{\prime}_{\alpha_3}\sigma^{\prime}_{\alpha_4}\ket_{G^{(1)}}-\bra\sigma_{\alpha_1}\sigma_{\alpha_2}\ket_{G^{(1)}}\bra\sigma^{\prime}_{\alpha_3}\sigma^{\prime}_{\alpha_4}\ket_{G^{(1)}}\Big]\Bigg\}\, \nonumber\\
&+\frac{1}{n_1}\le(\delta_{i_1i_3}\delta_{i_2i_4}+\delta_{i_1i_4}\delta_{i_2i_3}\ri)\le(\CW{2}\ri)^2\Ti{\NTKM}{\alpha_3\alpha_4}{1}\bra\sigma_{\alpha_1}\sigma_{\alpha_2}\sigma^{\prime}_{\alpha_3}\sigma^{\prime}_{\alpha_4}\ket_{G^{(1)}}\, . 
\end{align}
Here, as for the variance \eqref{eq:second-layer-ntk-pre-omega}, we recalled the suitably generalized versions of
\eqref{eq:two-activations-Gauss}, \eqref{eq:four-activations-one-neuron-Gauss}, and \eqref{eq:four-activations-two-neurons-Gauss} for the two- and four-point correlators. %
Thus,
we see that the first measure of cross correlation\index{cross correlation!NTK-preactivation} between the second-layer preactivations and the second-layer NTK~\eqref{eq:unrolling-cross-first} vanishes, but the second one~\eqref{eq:unrolling-cross-second} is nonzero at finite width.

To aid us in our deep-layer analysis, it will be convenient to 
 decompose this cross correlation\index{cross correlation!NTK-preactivation} \eqref{eq:unrolling-cross-second} into two tensors with \terminate{sample indices} only, just as we did for the variance in \eqref{eq:NTH-variance-decomposition-second}:
\begin{align}\label{eq:F-and-D-decomposition}\index{tensor decomposition!NTK-preactivation $D$/$F$}
&\E{\z{i_1}{\alpha_1}{2}\z{i_2}{\alpha_2}{2}\DNTK{i_3i_4}{\alpha_3\alpha_4}{2}} \\
=&\frac{1}{n_1}\le[\delta_{i_1i_2}\delta_{i_3i_4} \NTHD{\alpha_1\alpha_2\alpha_3\alpha_4}{2}+\delta_{i_1i_3}\delta_{i_2i_4}\NTHF{\alpha_1\alpha_3\alpha_2\alpha_4}{2}+\delta_{i_1i_4}\delta_{i_2i_3}\NTHF{\alpha_1\alpha_4\alpha_2\alpha_3}{2}\ri]\, \notag .
\end{align}
Comparing this decomposition with our explicit formula for the correlator \eqref{eq:unrolling-cross-second}, we can identify expressions for these tensors
\begin{align}
\NTHD{\alpha_1\alpha_2\alpha_3\alpha_4}{2}=&\CW{2}\le[ \bra \sigma_{\alpha_1}\sigma_{\alpha_2} \Oi{\alpha_3\alpha_4}{2} \ket_{G^{(1)}}-\bra\sigma_{\alpha_1}\sigma_{\alpha_2}\ket_{G^{(1)}}\bra \Oi{\alpha_3\alpha_4}{2} \ket_{G^{(1)}}\ri]\, ,\label{eq:D-recursion-second}\\
\NTHF{\alpha_1\alpha_3\alpha_2\alpha_4}{2}=&\le(\CW{2}\ri)^2\Ti{\NTKM}{\alpha_3\alpha_4}{1}\bra\sigma_{\alpha_1}\sigma_{\alpha_2}\sigma^{\prime}_{\alpha_3}\sigma^{\prime}_{\alpha_4}\ket_{G^{(1)}}\, ,\label{eq:F-recursion-second}
\end{align}
where we've also recalled the stochastic tensor $\Oi{\alpha_1\alpha_2}{2}$ defined in~\eqref{eq:def-omega-without-neural-second}.\footnote{The cross-correlation tensor $\NTHD{\alpha_1\alpha_2\alpha_3\alpha_4}{2}$~\eqref{eq:D-recursion-second} -- and more generally $\NTHD{\alpha_1\alpha_2\alpha_3\alpha_4}{\ell}$ in deeper layers, cf.~\eqref{eq:D-recursion} -- is symmetric under exchanges of \terminate{sample indices} $\alpha_1\leftrightarrow \alpha_2$ and $\alpha_3\leftrightarrow \alpha_4$, but of course \emph{not} under $(\alpha_1\alpha_2) \leftrightarrow (\alpha_3\alpha_4)$.
The other tensor $\NTHF{\alpha_1\alpha_3\alpha_2\alpha_4}{2}$~\eqref{eq:F-recursion-second} respects this same symmetry in the second layer, but has no symmetry at all in deeper layers, cf.~\eqref{eq:F-recursion}.}
Just as for $\NTHAwo{2}$ and  $\NTHBwo{2}$ above, both $\NTHDwo{2}$ and $\NTHFwo{2}$ are manifestly of order one. Similar to the second-layer NTK variance\index{neural tangent kernel!variance}, this means that the cross correlator~\eqref{eq:F-and-D-decomposition} is suppressed by $1/n_1$ in the large-width limit, vanishing in the strict \terminate{infinite-width limit}.

In summary, the joint distribution of the second-layer preactivations and second-layer NTK,
\be
p\!\le(z^{(2)},\NTK^{(2)}\Big\vert \D\ri) \, ,
\ee
at leading nontrivial order in the \terminate{$1/n$ expansion} is \neo{nearly-Gaussian distribution} with  \emph{(i)} a quartic interaction among preactivations on different neurons, \emph{(ii)} a fluctuating NTK, and \emph{(iii)} cross correlation\index{cross correlation!NTK-preactivation} between the preactivations and NTK. %
All of these finite-width effects become more complicated for deeper layers.

\section{Deeper Layers: Accumulation of NTK Fluctuations}
\label{sec:deeper-layer-accumulation-NTK}
As before with~\S\ref{sec:first-layer-gaussian}$\,\parallel\,$\S\ref{sec:first-layer-deterministic-NTK} and~\S\ref{sec:second-layer-non-gaussian}$\,\parallel\,$\S\ref{sec:second-layer-fluctuating-NTK}, this section parallels~\S\ref{sec:deeper-layer-accumulation}. 
In \S\ref{sec:deeper-layer-accumulation}, we investigated the \terminate{nearly-Gaussian distribution} of preactivations $p\!\le(z^{(\ell+1)}\Big\vert\D\ri)$
at finite width by considering an interlayer joint distribution $p\!\le(z^{(\ell+1)}, z^{(\ell)}\Big\vert\D\ri)$ and then integrating out the $\ell$-th-layer preactivations.
In particular, due to correlated dependence on the preactivations in previous layers, the non-Gaussianity in the preactivation distribution accumulated as depth increased, manifesting itself in the running \terminate{four-point vertex} $V^{(\ell)}$.
\index{nearly-Gaussian distribution}\index{Gaussian distribution}

\index{neural tangent kernel!l-th-layer@$\ell$-th-layer}\index{neural tangent kernel!variance}\index{neural tangent kernel!mean}
This same mechanism makes the NTK fluctuations accumulate, amplifying the NTK variance as well as the cross correlation\index{cross correlation!NTK-preactivation} between the NTK and preactivations.
In this section, we will derive recursions for the NTK mean,  the NTK-preactivation cross correlation\index{cross correlation!NTK-preactivation}, and the NTK variance that together determine the $\ell$-th-layer joint distribution at leading nontrivial order in $1/n$.\index{$1/n$ expansion}
What follows is a \neo{goode olde calculation}, so please sharpen your quills, unfurl your parchment, and inform your majordomo that you require a cleared schedule for the rest of the day. %

\setcounter{subsection}{-1}
\subsection{\emph{Inter}lude: \emph{Inter}layer Correlations}\label{subsec:interlayer-interlude}\index{interlayer correlation}
If you have an eidetic memory, then perhaps you recall that the main complication with our derivation of the general $(\ell+1)$-th-layer preactivation statistics -- as compared to the second layer statistics -- was that the $\ell$-th-layer preactivation distribution $p\!\le(z^{(\ell)}\Big\vert\D\ri)$ was also non-Gaussian, unlike the Gaussian preactivation distribution in the first layer.
For such a \terminate{nearly-Gaussian distribution} $p\!\le(z^{(\ell)}\Big\vert\D\ri)$, \neo{interactions} imply a nontrivial \term{intralayer correlation} between observables of the preactivations across different neurons $i_1\ne i_2$. Specifically, the covariance of two arbitrary single-neuron functions $\mathcal{F}\!\le(\z{i_1}{\A_1}{\ell}\ri)$ and $\mathcal{G}\!\le(\z{i_2}{\A_2}{\ell}\ri)$ depending on data subsamples $\A_1, \A_2\subset\D$, respectively, is given by~\eqref{eq:general-covariance} and reprinted here:
\begin{align}\label{eq:general-covariance-reprint}
&\cov{\mathcal{F}\!\le(\z{i_1}{\A_1}{\ell}\ri)\!}{\mathcal{G}\!\le(\z{i_2}{\A_2}{\ell}\ri)} \equiv \E{\mathcal{F}\!\le(\z{i_1}{\A_1}{\ell}\ri)\mathcal{G}\!\le(\z{i_2}{\A_2}{\ell}\ri)}-\E{\mathcal{F}\!\le(\z{i_1}{\A_1}{\ell}\ri)}\E{\mathcal{G}\!\le(\z{i_2}{\A_2}{\ell}\ri)}\, \notag \\
=&\!\!\!\!\!\sum_{\beta_1,\ldots,\beta_4\in\D}\!\frac{1}{4n_{\ell-1}}\TI{V}{(\beta_1\beta_2)(\beta_3\beta_4)}{\ell}\bra \le(z_{\beta_1} z_{\beta_2}-G^{(\ell)}_{\beta_1\beta_2}\ri) \mathcal{F}\!\le(z_{\A_1}\ri) \ket_{G^{(\ell)}}\!\!\bra  \le(z_{\beta_3} z_{\beta_4}-G^{(\ell)}_{\beta_3\beta_4}\ri)\mathcal{G}\!\le(z_{\A_2}\ri) \ket_{G^{(\ell)}}\, \notag\\
&+\o{\frac{1}{n^2}}\,  .
\end{align}
In this reprinting, we implicitly substituted in our leading large-width expressions~\eqref{eq:two-point-match-general} and~\eqref{eq:four-point-match-general} for the quadratic coupling $g_{(\ell)}$ and quartic coupling $v_{(\ell)}$, respectively.
We have also recalled our long-forgotten shorthand notation for the covariance of random variables~\eqref{eq:C2}, which we will use judiciously throughout this section.
This \emph{intralayer} formula will soon prove itself useful.\index{coupling!quadratic}\index{coupling!quartic}

Enlarging
our view to the preactivation-NTK joint distribution~\eqref{eq:joint-preactivation-NTK}, we'll encounter another complication due to \textbf{\emph{inter}layer correlation}\index{interlayer correlation} of the form
\be\label{eq:abstract-interlayer-goal}
\E{\mathcal{O}\!\le(z^{(\ell+1)}\ri)\mathcal{P}\!\le(W^{(\ell+1)}\ri)\mathcal{Q}\!\le(z^{(\ell)},  \NTK^{(\ell)} \ri)}\, ,
\ee
where $\O$ is some
function of $(\ell+1)$-th-layer preactivations,
$\mathcal{P}$ is a polynomial of $(\ell+1)$-th-layer weights,
and $\mathcal{Q}$ is a function of $\ell$-th-layer preactivations
and the $\ell$-th-layer NTK.\index{neural tangent kernel!l-th-layer@$\ell$-th-layer}
For instance, taking the NTK-preactivation cross correlation\index{cross correlation!NTK-preactivation}
\be\label{eq:cross-general-ell-first-appearance}
\E{\z{i_1}{\alpha_1}{\ell+1}\z{i_2}{\alpha_2}{\ell+1} \Tia{\NTK}{i_3i_4}{\alpha_3\alpha_4}{\ell+1}}\, 
\ee
and unraveling the NTK through its forward equation~\eqref{eq:NTH-recursion-without-expectation}\index{forward equation!NTK}, we get an interlayer correlation of the form~\eqref{eq:abstract-interlayer-goal} with $\mathcal{P}=1$ from the additive term in the square brackets and an interlayer correlation of the same form with $\mathcal{P}=\W{i_3j_3}{\ell+1}\W{i_4j_4}{\ell+1}$ from the recursive term.
While it was simple enough to evaluate such an expectation for the second layer,
it's somewhat subtle for a general layer.

\index{interlayer correlation}
That said, there's actually a pretty neat trick that lets us reduce such interlayer correlations~\eqref{eq:abstract-interlayer-goal} to expectations of solely $\ell$-th-layer variables. Such expectations can subsequently be evaluated with the \emph{intra}layer formula~\eqref{eq:general-covariance-reprint} above.
Let us now teach you this \terminate{magic trick} before diving deep into learning the deeper-layer analysis.\footnote{
    As similar interlayer correlations appear in~\S\ref{ch:features}, we'll keep our exposition completely general rather than specializing to the NTK-preactivation cross correlation\index{cross correlation!NTK-preactivation} \eqref{eq:cross-general-ell-first-appearance}.
}

First, using the definition of the expectation and the conditional structure of the distribution, the \terminate{interlayer correlation}~\eqref{eq:abstract-interlayer-goal} can be expressed as (suppressing all indices)
\begin{align}\label{eq:decompose-interlayer}
&\E{\mathcal{O}\!\le(z^{(\ell+1)}\ri)\mathcal{P}\!\le(W^{(\ell+1)}\ri)\mathcal{Q}\!\le(z^{(\ell)},  \NTK^{(\ell)} \ri)}\, \\
=&\int dz^{(\ell)} d\NTK^{(\ell)} p\!\le(z^{(\ell)}, \NTK^{(\ell)}\Big\vert \D\ri)\mathcal{Q}\!\le(z^{(\ell)},  \NTK^{(\ell)} \ri)\, \notag\\
&\quad \times\Bigg[\int db^{(\ell+1)}dW^{(\ell+1)} p\!\le(b^{(\ell+1)}\ri)p\!\le(W^{(\ell+1)}\ri)\mathcal{P}\!\le(W^{(\ell+1)}\ri)\, \notag\\
&\quad \quad\quad \times\int dz^{(\ell+1)} p\!\le(z^{(\ell+1)}\Big\vert b^{(\ell+1)},W^{(\ell+1)},z^{(\ell)}\ri) \mathcal{O}\!\le(z^{(\ell+1)}\ri)\Bigg]\, .\notag
\end{align}
Our strategy will be to \emph{integrate out}\index{integrating out} or marginalize over\index{marginalizing over} the $(\ell+1)$-th-layer parameters in order to express the object inside the square bracket as a function of the $\ell$-th-layer variables \emph{only}. In this way, 
the entire object will become an $\ell$-th-layer expectation that we already know how to handle.

\index{interlayer correlation}
Second, rather than working with an abstract polynomial $\mathcal{P}$, let's construct a \neo{generating function} for these interlayer correlations through the use of a \neo{source term}: 
\be\label{eq:weighty-source}
\mathcal{P}\!\le(W^{(\ell+1)}\ri)=e^{\sum_{i,j} \JW{ij}W^{(\ell+1)}_{ij}}\, .
\ee
Recall from your pretraining days (\S\ref{sec:Gauss}) that a generating function such as \eqref{eq:weighty-source} could be used to evaluate expectations such as~\eqref{eq:abstract-interlayer-goal} with any polynomial insertions of weights $W^{(\ell+1)}_{ij}$. To do this, we differentiate the evaluated generating function some number of times with respect to the source $\JW{ij}$ and then set the source to zero.

\index{initialization distribution}
Now with our choice~\eqref{eq:weighty-source} 
in mind for $\mathcal{P}$, we can explicitly evaluate the expression in the square brackets in~\eqref{eq:decompose-interlayer} as follows: \emph{(i)} recall the initialization distributions for the biases~\eqref{eq:full-bias-initialization} and weights~\eqref{eq:full-weights-initialization}, \emph{(ii)} recall from  \eqref{eq:deeper-layer-formal-expression-first-encounter} that the conditional distribution $p\!\le(z^{(\ell+1)}\Big\vert b^{(\ell+1)},W^{(\ell+1)},z^{(\ell)}\ri)$
encodes the MLP forward equation~\eqref{eq:forward-pass} as a \terminate{Dirac delta function}, and finally \emph{(iii)} recall the integral representation of the \terminate{Dirac delta function} \eqref{eq:integral-form-delta-function}.
All together, this gives the following set of integrals
\begin{align}
& \int \le[\prod_{i} \frac{d b_{i}}{\sqrt{2\pi C_b^{(\ell+1)}}}\ri]  \le[\prod_{i,j} \frac{d W_{ij}}{\sqrt{2\pi C_W^{(\ell+1)}/n_{\ell}}} \ri]\le[\prod_{i,\alpha} \frac{d \HS_{i}^{\ \alpha} \,d\z{i}{\alpha}{\ell+1}}{2\pi}\ri] \mathcal{O}\!\le(z^{(\ell+1)}\ri)\, \\
&\times\exp\!\le[-\sum_{i} \frac{b_i^2}{2C_b^{(\ell+1)}}-\sum_{i,j} \frac{n_{\ell} W_{ij}^2}{2C_W^{(\ell+1)}}+i\sum_{i,\alpha} \HS_{i}^{\ \alpha}\le(\z{i}{\alpha}{\ell+1}-b_i-\sum_{j}W_{ij}\,\x{j}{\alpha}\ri)+\sum_{i,j}\JW{ij}W_{ij}\ri]\, ,  \nonumber
\end{align}
which we recognize as the good-old \neo{Hubbard-Stratonovich transformation} that we first encountered in~\S\ref{sec:first-layer-gaussian}.

Next, as we did in~\S\ref{sec:first-layer-gaussian} and in high school, we can \emph{complete the squares}\index{complete the square} with respect to the biases and weights and integrate them out.\index{integrating out} The only substantial deviation here from the presentation in~\S\ref{sec:first-layer-gaussian} is that the source term shifts the linear coupling of 
the weights as
\be
 -iW_{ij}\sum_{\alpha} \HS_{i}^{\ \alpha}\s{j}{\alpha}{\ell}\to -i W_{ij}\le(\sum_{\alpha} \HS_{i}^{\ \alpha}\s{j}{\alpha}{\ell}+i\JW{ij}\ri)\, .
\ee
Performing these Gaussian integrals, we find
\begin{align}
&\int \!\! \le[\prod_{i,\alpha} \frac{d \HS_{i}^{\ \alpha}\,d\z{i}{\alpha}{\ell+1}}{2\pi}\ri]\!\mathcal{O}\!\le(z^{(\ell+1)}\ri)\exp\!\Bigg[\!-\!\sum_{i,\alpha_1,\alpha_2}\HS_{i}^{\ \alpha_1}\HS_{i}^{\ \alpha_2} \le(\frac{C_b^{(\ell+1)}}{2}+\frac{C_W^{(\ell+1)}}{2n_{\ell}}\sum_{j} \s{\alpha_1}{j}{\ell} \s{\alpha_2}{j}{\ell} \ri) \, \notag \\
&\quad \quad \quad \quad \quad \quad \quad \quad + i\sum_{i,\alpha} \HS_{i}^{\ \alpha}\le(\z{i}{\alpha}{\ell+1}-\frac{\CW{\ell+1}}{n_{\ell}}\sum_{j}\JW{ij}\s{j}{\alpha}{\ell}\ri)+\frac{\CW{\ell+1}}{2n_\ell}\sum_{i,j}\JW{ij}^2
\Bigg]\, .
\end{align}
Just as in our previous Hubbard-Stratonoviching~\eqref{eq:to-be-referenced-in-interlayer-part-far-ahead-in-the-future}, the \emph{stochastic metric}~\eqref{eq:general-stochastic-metric},
\be\label{eq:general-stochastic-metric-reprint}
\Ti{\widehat{G}}{\alpha_1 \alpha_2}{\ell+1}\equiv  \Cb{\ell+1}+\CW{\ell+1}\frac{1}{n_{\ell}}\sum_{j=1}^{n_{\ell}}\s{j}{\alpha_1}{\ell}\s{j}{\alpha_2}{\ell}\, ,
\ee
appears in the quadratic term of the Hubbard-Stratonovich variables $\HS_{i}^{\ \alpha}$, while the linear term is slightly modified by a shifting of the preactivations with the subtraction of the quantity
\be\label{eq:stochastic-mean}
\widehat{\mathcal{M}}_{i;\alpha}\equiv\CW{\ell+1}\le(\frac{1}{n_{\ell}}\sum_{j=1}^{n_{\ell}}\JW{ij}\s{j}{\alpha}{\ell}\ri)\, .
\ee
Completing the squares\index{complete the square} with the Hubbard-Stratonovich variables and integrating them out\index{integrating out}, we get
\begin{align}
&\frac{\exp\!\le(\frac{\CW{\ell+1}}{2n_\ell}\sum_{i,j}\JW{ij}^2\ri)}{\sqrt{\dete{2\pi \widehat{G}^{(\ell+1)}}^{n_{\ell+1}}}} \int  \le[\prod_{i,\alpha}d\z{i}{\alpha}{\ell+1}\ri] \mathcal{O}\!\le(z^{(\ell+1)}\ri) \, \\
&\qquad\qquad\quad\times\exp\!\le[-\frac{1}{2}\sum_{i}\sum_{\alpha_1,\alpha_2}\SKinv{\alpha_1 \alpha_2}{\ell+1}\le(\z{i}{\alpha_1}{\ell+1}-\widehat{\mathcal{M}}_{i;\alpha_1}\ri)\le(\z{i}{\alpha_2}{\ell+1}-\widehat{\mathcal{M}}_{i;\alpha_2}\ri)\ri]\, .  \nonumber
\end{align}
Ignoring the quadratic source factor $\JW{ij}^2$ outside the integral, this is just a \emph{Gaussian expectation}\index{Gaussian expectation} 
of $\mathcal{O}$ against the $(\ell+1)$-th-layer preactivation distribution
with a mean $\widehat{\mathcal{M}}_{i;\alpha}$ and a variance $\Ti{\widehat{G}}{\alpha_1 \alpha_2}{\ell+1}$. (Make sure you remember our \terminate{general relativity} convention: $\SKinv{\alpha_1 \alpha_2}{\ell+1}$ is the inverse of $\Ti{\widehat{G}}{\alpha_1 \alpha_2}{\ell+1}$.)

Now, let's compensate for this mean by shifting the dummy integration variable as $\z{i}{\alpha}{\ell+1} \to \z{i}{\alpha}{\ell+1} + \widehat{\mathcal{M}}_{i;\alpha_1}$, which yields a compact expression in terms of our \emph{zero-mean} Gaussian expectation notation~\eqref{eq:many-neuron-gaussian-notation}:
\begin{align}\label{eq:compact-stochastic-with-source}
\exp\!\le(\frac{\CW{\ell+1}}{2n_\ell}\sum_{i,j}\JW{ij}^2\ri) \brabra\mathcal{O}\!\le(z^{(\ell+1)}+\widehat{\mathcal{M}}\ri)\ketket_{\widehat{G}^{(\ell+1)}} \, .
\end{align}
Plugging this result~\eqref{eq:compact-stochastic-with-source} back into our \terminate{interlayer correlation}~\eqref{eq:abstract-interlayer-goal} 
and substituting back in for the mean shift~\eqref{eq:stochastic-mean}, we arrive at a simple formula for our \terminate{generating function}:
\begin{align}\label{eq:master-weight-insertion}
&\E{\mathcal{O}\!\le(z^{(\ell+1)}\ri)e^{\sum_{i,j} \JW{ij}W^{(\ell+1)}_{ij}}\mathcal{Q}\!\le(z^{(\ell)},  \NTK^{(\ell)} \ri)}\, \\
=&\exp\!\le(\frac{\CW{\ell+1}}{2n_\ell}\sum_{i,j}\JW{ij}^2\ri) \E{\bra\!\!\!\bra\mathcal{O}\Big(z^{(\ell+1)}_{i;\alpha}+\CW{\ell+1}\frac{1}{n_{\ell}}\sum_{j=1}^{n_{\ell}}\JW{ij}\s{j}{\alpha}{\ell}\Big)\ket\!\!\!\ket_{\widehat{G}^{(\ell+1)}}\mathcal{Q}\!\le(z^{(\ell)},  \NTK^{(\ell)}\ri) }\, .\notag 
\end{align}
After performing the \terminate{Gaussian expectation} over the $(\ell+1)$-th-layer preactivations $z^{(\ell+1)}_{i;\alpha}$ -- which is typically trivial in all the concrete applications that we'll encounter -- the expectation in \eqref{eq:master-weight-insertion} is only with respect to $\ell$-th-layer variables.\footnote{Recall that the stochastic metric $\widehat{G}^{(\ell+1)}$~\eqref{eq:general-stochastic-metric-reprint} depends only on the $\ell$-th-layer preactivations.} This was our desired result.

To see how to use the \terminate{generating function}~\eqref{eq:master-weight-insertion}, let's work out some explicit examples. First, consider the case with no weight insertions. Setting the source to zero, $\JW{ij}=0$, we find
\begin{align}\label{eq:no-weight-insertion-general}
&\E{\mathcal{O}\!\le(z^{(\ell+1)}\ri)\mathcal{Q}\!\le(z^{(\ell)},  \NTK^{(\ell)} \ri)}=\E{\brabra\mathcal{O}\Big(z^{(\ell+1)}\Big)\ketket_{\widehat{G}^{(\ell+1)}}\mathcal{Q}\!\le(z^{(\ell)},  \NTK^{(\ell)}\ri) }\, .%
\end{align}
This formula is not trivial and is not something we knew before: here we see that the correlation between preactivations in neighboring layers\index{interlayer correlation} is given by first computing a Gaussian expectation\index{Gaussian expectation} of the $(\ell+1)$-th-layer function against the stochastic metric and then taking the full expectation of the resulting $\ell$-th-layer quantity.

Next, let's consider two weight insertions. Twice-differentiating the \terminate{generating function}~\eqref{eq:master-weight-insertion} by the source as $\frac{d}{d\JW{i_3j_3}}\frac{d}{d\JW{i_4j_4}}$ and then setting the source to zero, we get
\begin{align}\label{eq:two-weight-insertions-general}
&\E{\mathcal{O}\!\le(z^{(\ell+1)}\ri)\W{i_3j_3}{\ell+1}\W{i_4j_4}{\ell+1}\mathcal{Q}\!\le(z^{(\ell)},  \NTK^{(\ell)} \ri)}\, \\
=&\delta_{i_3i_4}\delta_{j_3j_4}\frac{\CW{\ell+1}}{n_{\ell}}\E{\bra\!\bra\mathcal{O}\ket\!\ket_{\widehat{G}^{(\ell+1)}}\mathcal{Q}\!\le(z^{(\ell)},  \NTK^{(\ell)}\ri) }\, \notag\\
&\quad\quad+\le(\frac{\CW{\ell+1}}{n_{\ell}}\ri)^2\sum_{\beta_3,\beta_4,\gamma_3,\gamma_4}\mathbb{E}\Bigg[\brabra\le(\z{i_3}{\beta_3}{\ell+1}\z{i_4}{\beta_4}{\ell+1}-\delta_{i_3i_4}\Ti{\widehat{G}}{\beta_3\beta_4}{\ell+1}\ri)\mathcal{O}\ketket_{\widehat{G}^{(\ell+1)}}\, \nonumber\\
&\quad \quad \quad \quad \quad \quad \quad \quad \quad \quad \quad \quad \times\SKinv{\beta_3\gamma_3}{\ell+1}\SKinv{\beta_4\gamma_4}{\ell+1}\s{j_3}{\gamma_3}{\ell}\s{j_4}{\gamma_4}{\ell}\mathcal{Q}\!\le(z^{(\ell)},  \NTK^{(\ell)}\ri) \Bigg]\, .\notag
\end{align}
Here, we used \terminate{integration by parts} to exchange the derivatives for a projection as
\be
\bra\!\!\!\bra\frac{\partial^2\mathcal{O}}{\partial \z{i_3}{\gamma_3}{\ell+1}\partial \z{i_4}{\gamma_4}{\ell+1}}\ket\!\!\!\ket_{\widehat{G}^{(\ell+1)}}=\sum_{\beta_3,\beta_4}\SKinv{\beta_3\gamma_3}{\ell+1}\SKinv{\beta_4\gamma_4}{\ell+1}\brabra\le(\z{i_3}{\beta_3}{\ell+1}\z{i_4}{\beta_4}{\ell+1}-\delta_{i_3i_4}\Ti{\widehat{G}}{\beta_3\beta_4}{\ell+1}\ri)\mathcal{O}\ketket_{\widehat{G}^{(\ell+1)}}\, .
\ee
Intuitively, the first term in~\eqref{eq:two-weight-insertions-general} comes from forming a Wick contraction\index{Wick contraction} with the two weight insertions, while the second term comes from two pairs of Wick contractions\index{Wick contraction}, each between an inserted weight and a weight hidden inside the $z^{(\ell+1)}$'s in $\mathcal{O}$.

Thusly, with the \emph{intra}layer formula~\eqref{eq:general-covariance-reprint} recalled and the \emph{inter}layer formulae~\eqref{eq:no-weight-insertion-general} and~\eqref{eq:two-weight-insertions-general} derived, we are as ready as we'll ever be to recursively analyze the joint statistics of the NTK and preactivations in deeper layers. This concludes our \emph{inter}lude.

\subsection{NTK Mean}\label{subsec:NTKmean}\index{forward equation!NTK}\index{neural tangent kernel!l-th-layer@$\ell$-th-layer}
Taking the expectation of the stochastic NTK forward equation~\eqref{eq:NTH-recursion-without-expectation}, we get
\begin{align}\label{eq:NTK-mean-pre-recursion}
\E{\Tia{\NTH}{i_1i_2}{\alpha_1\alpha_2}{\ell+1}}=&\delta_{i_1i_2} \le[\Lb{\ell+1} + \lamW{\ell+1}\le(\frac{1}{n_{\ell}}\sum_{j=1}^{n_{\ell}}\E{\s{j}{\alpha_1}{\ell} \s{j}{\alpha_2}{\ell}}\ri)\ri]\, \\
&+ \delta_{i_1i_2} \CW{\ell+1} \frac{1}{n_{\ell}}\sum_{j=1}^{n_{\ell}} \E{ \ds{j}{\alpha_1}{\ell} \ds{j}{\alpha_2}{\ell} \Tia{\NTK}{jj}{\alpha_1\alpha_2}{\ell}}\, ,\nonumber
\end{align}
where, as is now familiar, on the second line we used the independence of the $(\ell+1)$-th-layer weights from the $\ell$-th-layer preactivations, and then immediately evaluated the weight expectation. Immediately, we see that NTK mean is diagonal in \terminate{neural indices} at any network depth.

Given that, let's decompose the $\ell$-th-layer NTK into a mean and fluctuation as\index{neural tangent kernel!mean}
\be\label{eq:NTK-fluc-decomposition}\index{tensor decomposition!NTK mean and fluctuation}
\Tia{\NTK}{i_1i_2}{\alpha_1\alpha_2}{\ell}\equiv \delta_{i_1i_2} \Ti{\NTKM}{\alpha_1\alpha_2}{\ell}+\DNTK{i_1i_2}{\alpha_1\alpha_2}{\ell}\, ,
\ee
where we have denoted the $\ell$-th-layer NTK mean as $ \delta_{i_1i_2} \Ti{\NTKM}{\alpha_1\alpha_2}{\ell}$.
As before, we separated the part of the mean that encodes the sample dependence and symbolized it without a hat. 
Substituting this decomposition into our expression \eqref{eq:NTK-mean-pre-recursion} for the NTK mean, we see that the $(\ell+1)$-th-layer mean obeys a recursion
\begin{align}\label{eq:NTK-mean-recursion-with-E}
\Ti{\NTKM}{\alpha_1\alpha_2}{\ell+1}=&\Lb{\ell+1} + \lamW{\ell+1}\!\le(\frac{1}{n_{\ell}}\sum_{j=1}^{n_{\ell}}\E{\s{j}{\alpha_1}{\ell} \s{j}{\alpha_2}{\ell}}\ri)+\CW{\ell+1}\Ti{\NTKM}{\alpha_1\alpha_2}{\ell} \!\le(\frac{1}{n_{\ell}}\sum_{j=1}^{n_{\ell}} \E{\ds{j}{\alpha_1}{\ell} \ds{j}{\alpha_2}{\ell}}\ri)\, \nonumber\\
&+\CW{\ell+1} \!\le(\frac{1}{n_{\ell}}\sum_{j=1}^{n_{\ell}} \E{ \ds{j}{\alpha_1}{\ell} \ds{j}{\alpha_2}{\ell} \DNTK{jj}{\alpha_1\alpha_2}{\ell}}\ri)\, ,
\end{align}
depending on both the mean and fluctuation in the previous layer $\ell$.

To the leading order in $1/n$, the first two expectation values on the right-hand side of~\eqref{eq:NTK-mean-recursion-with-E} %
are given by Gaussian expectations\index{Gaussian expectation}
\begin{align}\label{eq:ntk-mean-gaussian-1}
\E{\s{j}{\alpha_1}{\ell} \s{j}{\alpha_2}{\ell}}=&\bra\sigma_{\alpha_1}\sigma_{\alpha_2}\ket_{G^{(\ell)}}+\o{\frac{1}{n}}\, ,\\
\E{\ds{j}{\alpha_1}{\ell} \ds{j}{\alpha_2}{\ell}}=&\bra\sigma^{\prime}_{\alpha_1}\sigma^{\prime}_{\alpha_2}\ket_{G^{(\ell)}}+\o{\frac{1}{n}}\, .\label{eq:ntk-mean-gaussian-2}
\end{align}
To see this, note that the first expectation is just the leading Gaussian contribution~\eqref{eq:two-activations-deep} with the non-Gaussian coupling $v$ suppressed as $\sim1/n$ as per \eqref{eq:four-point-match-general}, and that the evaluation of the second expectation proceeds identically to the first regardless of whether the activation has a derivative or not. Meanwhile, the final expectation on the second line of \eqref{eq:NTK-mean-recursion-with-E} involves an NTK-preactivation cross correlation, which is also suppressed in the large-width limit:
\be\label{eq:ntk-mean-cross-leading}
 \E{ \ds{j}{\alpha_1}{\ell} \ds{j}{\alpha_2}{\ell} \DNTK{jj}{\alpha_1\alpha_2}{\ell}}=\o{\frac{1}{n}}\, .
\ee
We will prove this rather shortly in the next subsection in~\eqref{eq:most-general}.

Assembling these leading contributions, the NTK mean recursion simplifies to 
\begin{align}\label{eq:NTHmean_final}\index{neural tangent kernel!mean|textbf}
\Ti{\NTKM}{\alpha_1\alpha_2}{\ell+1}=&\Lb{\ell+1} + \lamW{\ell+1}\bra\sigma_{\alpha_1}\sigma_{\alpha_2}\ket_{G^{(\ell)}}+\CW{\ell+1}\bra\sigma^{\prime}_{\alpha_1}\sigma^{\prime}_{\alpha_2}\ket_{G^{(\ell)}}\Ti{\NTKM}{\alpha_1\alpha_2}{\ell}+\o{\frac{1}{n}}\, .
\end{align}
If you're in the habit of marking up your book, feel free to draw a box around this formula.

\subsection{NTK-Preactivation Cross Correlations}\label{subsec:NTK-cross}\index{cross correlation!NTK-preactivation|textbf}\index{neural tangent kernel!NTK-preactivation cross correlation|see{cross correlation}}
Next, let's evaluate a cross-correlation expectation of a very general form
\be\label{eq:cross-general}
\E{\O\!\le(z^{(\ell+1)}\ri)\,\DNTK{i_3i_4}{\alpha_3\alpha_4}{\ell+1}}\, .
\ee
For instance, setting $\O=\z{i_1}{\alpha_1}{\ell+1}\z{i_2}{\alpha_2}{\ell+1}$ gives the elementary cross correlation~\eqref{eq:cross-general-ell-first-appearance}, while setting $\O=\ds{i_1}{\alpha_1}{\ell} \ds{i_2}{\alpha_2}{\ell}$ gives the subleading cross correlation~\eqref{eq:ntk-mean-cross-leading} that just appeared in (and then immediately disappeared from) our recursion for the NTK mean.\index{neural tangent kernel!mean}

\index{forward equation!NTK}\index{cross correlation!NTK-preactivation}
To begin, simply substitute the NTK forward equation~\eqref{eq:NTH-recursion-without-expectation} into the cross correlator~\eqref{eq:cross-general}, which yields
\begin{align}\label{eq:cross-general-reexpress}
&\E{\O\!\le(z^{(\ell+1)}\ri)\DNTK{i_3i_4}{\alpha_3\alpha_4}{\ell+1}}=\cov{\O\!\le(z^{(\ell+1)}\ri)}{\Tia{\NTK}{i_3i_4}{\alpha_3\alpha_4}{\ell+1}}\, \\
=&\delta_{i_3i_4}\,\lamW{\ell+1}\frac{1}{n_{\ell}}\sum_{j=1}^{n_{\ell}}\le\{\E{\O\!\le(z^{(\ell+1)}\ri)\s{j}{\alpha_3}{\ell}\s{j}{\alpha_4}{\ell}}-\E{\O\!\le(z^{(\ell+1)}\ri)}\E{\s{j}{\alpha_3}{\ell}\s{j}{\alpha_4}{\ell}}\ri\}\, \nonumber\\
&+\sum_{j_3,j_4=1}^{n_{\ell}}\Big\{\E{\O\!\le(z^{(\ell+1)}\ri)\W{i_3j_3}{\ell+1}\W{i_4j_4}{\ell+1}\ds{j_3}{\alpha_3}{\ell}\ds{j_4}{\alpha_4}{\ell}\Tia{\NTK}{j_3j_4}{\alpha_3\alpha_4}{\ell}}\, \nonumber\\
&\quad \quad \quad \quad -\E{\O\!\le(z^{(\ell+1)}\ri)}\E{\W{i_3j_3}{\ell+1}\W{i_4j_4}{\ell+1}}\E{\ds{j_3}{\alpha_3}{\ell}\ds{j_4}{\alpha_4}{\ell}\Tia{\NTK}{j_3j_4}{\alpha_3\alpha_4}{\ell}}\Big\}\, .\nonumber
\end{align}
Now putting the freshly-derived interlayer formulae~\eqref{eq:no-weight-insertion-general} and~\eqref{eq:two-weight-insertions-general} to use,
this cross correlator becomes
\begin{align}\label{eq:cross-general-reexpress-II}
&\E{\O\!\le(z^{(\ell+1)}\ri)\DNTK{i_3i_4}{\alpha_3\alpha_4}{\ell+1}}\, \\
=&\delta_{i_3i_4}\frac{\lamW{\ell+1}}{\CW{\ell+1}}\,\E{\brabra\O\!\le(z^{(\ell+1)}\ri)\ketket_{\widehat{G}^{(\ell+1)}} \dKi{\alpha_3\alpha_4}{\ell+1}}\, \nonumber\\
&+\delta_{i_3i_4}\frac{\CW{\ell+1}}{n_{\ell}}\sum_{j=1}^{n_{\ell}}\Big\{\E{\brabra\O\!\le(z^{(\ell+1)}\ri)\ketket_{\widehat{G}^{(\ell+1)}}\ds{j}{\alpha_3}{\ell}\ds{j}{\alpha_4}{\ell}\Tia{\NTK}{jj}{\alpha_3\alpha_4}{\ell}}\, \nonumber\\
&\quad \quad \quad \quad \quad \quad \quad \quad -\E{\brabra\O\!\le(z^{(\ell+1)}\ri)\ketket_{\widehat{G}^{(\ell+1)}}}\E{\ds{j}{\alpha_3}{\ell}\ds{j}{\alpha_4}{\ell}\Tia{\NTK}{jj}{\alpha_3\alpha_4}{\ell}}\Big\}\, \nonumber\\
&+\le(\frac{\CW{\ell+1}}{n_{\ell}}\ri)^2\sum_{j_3,j_4=1}^{n_{\ell}}\sum_{\beta_3,\beta_4,\gamma_3,\gamma_4}\mathbb{E}\Bigg[\brabra\le(\z{i_3}{\beta_3}{\ell+1}\z{i_4}{\beta_4}{\ell+1}-\delta_{i_3i_4}\Ti{\widehat{G}}{\beta_3\beta_4}{\ell+1}\ri)\O\!\le(z^{(\ell+1)}\ri)\ketket_{\widehat{G}^{(\ell+1)}}\, \nonumber\\
&\quad \quad \quad \quad \quad \quad \quad \quad \quad \quad \quad \quad \quad \quad \times\SKinv{\beta_3\gamma_3}{\ell+1}\SKinv{\beta_4\gamma_4}{\ell+1}\s{j_3}{\gamma_3}{\ell}\s{j_4}{\gamma_4}{\ell}\ds{j_3}{\alpha_3}{\ell}\ds{j_4}{\alpha_4}{\ell}\Tia{\NTK}{j_3j_4}{\alpha_3\alpha_4}{\ell} \Bigg]\, .\nonumber
\end{align}
Here, for the first term, we also recalled the definition of the metric fluctuation~\eqref{eq:metric-fluctuation-general-layer}.
From this general expression, we can already learn two important lessons.

First, setting $\O=\z{i_1}{\alpha_1}{\ell+1}\z{i_2}{\alpha_2}{\ell+1}$, we get an expression for the  elementary $(\ell+1)$-th-layer cross correlation\index{cross correlation!NTK-preactivation} in terms of $\ell$-th-layer variables %
\begin{align}\label{eq:D-F-decomposition-general-layer}\index{tensor decomposition!NTK-preactivation $D$/$F$}
&\E{\Tia{z}{i_1}{\alpha_1}{\ell+1}\Tia{z}{i_2}{\alpha_2}{\ell+1}\DNTK{i_3i_4}{\alpha_3\alpha_4}{\ell+1}}\, \notag\\
=&\delta_{i_1i_2}\delta_{i_3i_4}\le\{\frac{\lamW{\ell+1}}{\CW{\ell+1}}\E{\dKi{\alpha_1\alpha_2}{\ell+1}\dKi{\alpha_3\alpha_4}{\ell+1}}+\CW{\ell+1}\frac{1}{n_{\ell}}\sum_{j=1}^{n_{\ell}}\E{\dKi{\alpha_1\alpha_2}{\ell+1}\ds{j}{\alpha_3}{\ell}\ds{j}{\alpha_4}{\ell}\Tia{\NTK}{jj}{\alpha_3\alpha_4}{\ell}}\ri\}\, \nonumber\\
&+\delta_{i_1i_3}\delta_{i_2i_4}\le(\frac{\CW{\ell+1}}{n_{\ell}}\ri)^2\sum_{j,k=1}^{n_{\ell}}\mathbb{E}\Bigg[\s{j}{\alpha_1}{\ell}\s{k}{\alpha_2}{\ell}\ds{j}{\alpha_3}{\ell}\ds{k}{\alpha_4}{\ell}\Tia{\NTK}{jk}{\alpha_3\alpha_4}{\ell} \Bigg]\, \nonumber\\
&+\delta_{i_1i_4}\delta_{i_2i_3}\le(\frac{\CW{\ell+1}}{n_{\ell}}\ri)^2\sum_{j,k=1}^{n_{\ell}}\mathbb{E}\Bigg[\s{j}{\alpha_2}{\ell}\s{k}{\alpha_1}{\ell}\ds{j}{\alpha_3}{\ell}\ds{k}{\alpha_4}{\ell}\Tia{\NTK}{jk}{\alpha_3\alpha_4}{\ell} \Bigg]\, \nonumber\\
\equiv&\frac{1}{n_{\ell}}\le[\delta_{i_1i_2}\delta_{i_3i_4} \NTHD{\alpha_1\alpha_2\alpha_3\alpha_4}{\ell+1}+\delta_{i_1i_3}\delta_{i_2i_4}\NTHF{\alpha_1\alpha_3\alpha_2\alpha_4}{\ell+1}+\delta_{i_1i_4}\delta_{i_2i_3}\NTHF{\alpha_1\alpha_4\alpha_2\alpha_3}{\ell+1}\ri]\, ,  
\end{align}
where on the final line we decomposed the cross correlation\index{cross correlation!NTK-preactivation} into two tensors with \terminate{sample indices} only,
just as we did for the second layer in~\eqref{eq:F-and-D-decomposition}.
Equating the first expression with the second, we see that these tensors are defined by the following $\ell$-th-layer expectations:
\begin{align}
\frac{1}{n_{\ell}}\NTHD{\alpha_1\alpha_2\alpha_3\alpha_4}{\ell+1}\equiv&\frac{\lamW{\ell+1}}{\CW{\ell+1}}\,\E{\dKi{\alpha_1\alpha_2}{\ell+1}\dKi{\alpha_3\alpha_4}{\ell+1}}\, \label{eq:D-recursion-almost-there}\\
&+\CW{\ell+1}\le(\frac{1}{n_{\ell}}\sum_{j=1}^{n_{\ell}}\E{\dKi{\alpha_1\alpha_2}{\ell+1}\ds{j}{\alpha_3}{\ell}\ds{j}{\alpha_4}{\ell}\Tia{\NTK}{jj}{\alpha_3\alpha_4}{\ell}}\ri)\, ,\notag\\
\frac{1}{n_{\ell}}\NTHF{\alpha_1\alpha_3\alpha_2\alpha_4}{\ell+1}\equiv&\le(\CW{\ell+1}\ri)^2\le(\frac{1}{n_{\ell}^2}\sum_{j,k=1}^{n_{\ell}}\mathbb{E}\Bigg[\s{j}{\alpha_1}{\ell}\s{k}{\alpha_2}{\ell}\ds{j}{\alpha_3}{\ell}\ds{k}{\alpha_4}{\ell}\Tia{\NTK}{jk}{\alpha_3\alpha_4}{\ell} \Bigg]\ri)\, .\label{eq:F-recursion-almost-there}
\end{align}
We'll come back to evaluate these last two expressions -- and thereby derive recursions for both cross correlation tensors -- after we reveal the second lesson.

Second, we can start again from the general cross correlator~\eqref{eq:cross-general-reexpress-II} and push our calculation a little bit further to leading order in $1/n$.
The key step is perturbatively expanding the Gaussian expectation -- just as we expanded the stochastic Gaussian distribution~\eqref{eq:second-layer-stochastic-exponential} before using the Schwinger-Dyson equation~\eqref{eq:stochastic-metric-inversion} -- to get
\begin{align}\label{eq:SD-again}
&\brabra\O\!\le(z^{(\ell+1)}\ri)\ketket_{\widehat{G}^{(\ell+1)}}\, \\
=&\brabra\O\!\le(z^{(\ell+1)}\ri)\ketket_{G^{(\ell+1)}}\, \notag\\
&+\frac{1}{2}\!\sum_{\beta_1,\beta_2,\gamma_1,\gamma_2}\!\!\!\bra\!\!\!\bra\sum_{m}\le(z_{m;\beta_1}^{(\ell+1)}z_{m;\beta_2}^{(\ell+1)}-\Ti{G}{\beta_1\beta_2}{\ell+1}\ri)\O\!\le(z^{(\ell+1)}\ri)\ket\!\!\!\ket_{G^{(\ell+1)}}\!\!\!\!\!\!\!\TI{G}{\beta_1\gamma_1}{\ell+1}\TI{G}{\beta_2\gamma_2}{\ell+1}\Ti{\widehat{\Delta G}}{\gamma_1 \gamma_2}{\ell+1}\!\!\!+\o{\Delta^2}\, .\notag
\end{align}
Plugging this back into~\eqref{eq:cross-general-reexpress-II}, picking up the (leading-order) pieces, and using the definitions~\eqref{eq:D-recursion-almost-there} and~\eqref{eq:F-recursion-almost-there}, we get
\begin{align}\label{eq:most-general}
&\E{\O\!\le(z^{(\ell)}\ri)\DNTK{i_3i_4}{\alpha_3\alpha_4}{\ell}}\, \\
=&\delta_{i_3i_4}\frac{1}{n_{\ell-1}}\le[\frac{1}{2}\!\sum_{\beta_1,\beta_2,\gamma_1,\gamma_2}\!\!\bra\!\!\!\bra\sum_{m=1}^{n_{\ell}}\le(z_{m;\beta_1}^{(\ell)}z_{m;\beta_2}^{(\ell)}-\Ti{G}{\beta_1\beta_2}{\ell}\ri)\O\!\le(z^{(\ell)}\ri)\ket\!\!\!\ket_{G^{(\ell)}}\!\!\!\!\!\!\!\!\!\TI{G}{\beta_1\gamma_1}{\ell}\TI{G}{\beta_2\gamma_2}{\ell}\ri]\NTHD{\gamma_1\gamma_2\alpha_3\alpha_4}{\ell}\, \notag\\
&\,\,\,+\frac{1}{n_{\ell-1}}\sum_{\beta_1,\beta_2,\gamma_1,\gamma_2}\brabra\le(\z{i_3}{\beta_1}{\ell}\z{i_4}{\beta_2}{\ell}-\delta_{i_3i_4}\Ti{G}{\beta_1\beta_2}{\ell}\ri)\O\!\le(z^{(\ell)}\ri)\ketket_{G^{(\ell)}}\TI{G}{\beta_1\gamma_1}{\ell}\TI{G}{\beta_2\gamma_2}{\ell}\NTHF{\gamma_1\alpha_3\gamma_2\alpha_4}{\ell}\, \notag\\
&\,\,\,+\o{\frac{1}{n^2}}\, \, ,\nonumber
\end{align}
where we have also relabeled \neo{layer indices} as $(\ell+1)\to\ell$ everywhere for the ease of later substitutions.\footnote{You might worry about the summation $\sum_{m=1}^{n_{\ell}}$ inside the first Gaussian expectation in~\eqref{eq:most-general}. However, due to Gaussian factorization, this  expectation stays of order one so long as the observable $\O$ depends on only a finite number of neurons.}
This result illustrates that these more general cross correlations
are governed by the same tensors, $D^{(\ell)}$ and $F^{(\ell)}$, as the elementary cross correlation~\eqref{eq:D-F-decomposition-general-layer}.
We can indeed compute all the cross correlators
if we find and solve recursions for $D^{(\ell)}$ and $F^{(\ell)}$. %
It is this task that we turn to next.

\subsubsection{$D$-recursion}\index{cross correlation!NTK-preactivation!D-recursion@$D$-recursion}
Starting from our expression for $D^{(\ell+1)}$~\eqref{eq:D-recursion-almost-there} and substituting in the definition of the stochastic metric~\eqref{eq:general-stochastic-metric-reprint}, we get
\begin{align}\label{eq:D-all-pieces}
\NTHD{\alpha_1\alpha_2\alpha_3\alpha_4}{\ell+1}=&\CW{\ell+1}\frac{1}{n_{\ell}}\sum_{j,k=1}^{n_{\ell}}\cov{\s{j}{\alpha_1}{\ell}\s{j}{\alpha_2}{\ell}}{\lamW{\ell+1}\s{k}{\alpha_3}{\ell}\s{k}{\alpha_4}{\ell}+ \CW{\ell+1}\Ti{\NTKM}{\alpha_3\alpha_4}{\ell}\ds{k}{\alpha_3}{\ell}\ds{k}{\alpha_4}{\ell}}\, \nonumber\\
&+\le(\CW{\ell+1}\ri)^2\frac{1}{n_{\ell}}\sum_{j,k=1}^{n_{\ell}}\cov{\s{j}{\alpha_1}{\ell}\s{j}{\alpha_2}{\ell}}{\ds{k}{\alpha_3}{\ell}\ds{k}{\alpha_4}{\ell}\DNTK{k k}{\alpha_3\alpha_4}{\ell}}+\o{\frac{1}{n}}\, ,
\end{align}
where we have again decomposed the $\ell$-th-layer NTK\index{neural tangent kernel!l-th-layer@$\ell$-th-layer} into a mean and fluctuation piece.
We see that there are two types of terms here: covariances on a single neuron $j=k$ and covariances between pairs of neurons $j\ne k$.

For the single-neuron contribution with $j=k$, at the leading order, we find the same contribution that we found for the second layer~\eqref{eq:D-recursion-second}\footnote{The subleading $\o{1/n}$ piece includes both a contribution from the non-Gaussian part of the distribution as well as a cross correlation contribution from the previous layer.
}
\be\label{eq:D-piece1}
\CW{\ell+1} \le[\bra \sigma_{\alpha_1}\sigma_{\alpha_2}\Oi{\alpha_3\alpha_4}{\ell+1} \ket_{G^{(\ell)}}-\bra \sigma_{\alpha_1}\sigma_{\alpha_2}\ket_{G^{(\ell)}}\bra\Oi{\alpha_3\alpha_4}{\ell+1} \ket_{G^{(\ell)}}\ri]+\o{\frac{1}{n}}\, .
\ee
Here, the auxiliary stochastic matrix $\Oi{\alpha_1\alpha_2}{\ell+1}$ is defined as
\be\label{eq:def-omega-without-neural}
\Oi{\alpha_1\alpha_2}{\ell+1} \equiv \lamW{\ell+1} \Ti{\sigma}{\alpha_1}{\ell} \Ti{\sigma}{\alpha_2}{\ell} + \CW{\ell+1}\Ti{\NTKM}{\alpha_1\alpha_2}{\ell}  \dTi{\sigma}{\alpha_1}{\ell} \dTi{\sigma}{\alpha_2}{\ell}  \, ,
\ee
which simply generalizes the second-layer definition~\eqref{eq:def-omega-without-neural-second}.
As a reminder the unhatted matrix $\Ti{\NTKM}{\alpha_1\alpha_2}{\ell}$ is the %
NTK mean\index{neural tangent kernel!mean}, which is not a random variable and can safely be taken outside the Gaussian expectation\index{Gaussian expectation} $\bra\cdot\ket_{G^{(\ell)}}$. This means that, as a stochastic variable, $\Oi{\alpha_1\alpha_2}{\ell+1}$ depends only on the $\ell$-th-layer preactivations.

Next, for the pairs-of-neurons contribution to~\eqref{eq:D-all-pieces} with $j\ne k$, the first term can be evaluated by the intralayer formula~\eqref{eq:general-covariance-reprint} and yields
\be
\frac{n_{\ell}}{4n_{\ell-1}} \CW{\ell+1} \!\!\!\!\!\sum_{\gamma_1, \gamma_2, \gamma_3, \gamma_4} \!\!\!\!\!
\VU{\gamma_1 \gamma_2}{\gamma_3 \gamma_4}{\ell}\!
\bra\le(\!z_{\gamma_1} z_{\gamma_2} \!-\!\Ti{G}{\gamma_1 \gamma_2}{\ell} \!\ri)\sigma_{\alpha_1}\sigma_{\alpha_2}\!\ket_{G^{(\ell)}}\!\bra\le(\!z_{\gamma_3} z_{\gamma_4} \!-\!\Ti{G}{\gamma_3 \gamma_4}{\ell} \!\ri)\Oi{\alpha_3\alpha_4}{\ell+1}\!\ket_{G^{(\ell)}}\, .
\ee
Meanwhile, the covariance in the second term can be unrolled as
\begin{align}\label{eq:how-to-unroll}
&\cov{\s{j}{\alpha_1}{\ell}\s{j}{\alpha_2}{\ell}}{\ds{k}{\alpha_3}{\ell}\ds{k}{\alpha_4}{\ell}\DNTK{k k}{\alpha_3\alpha_4}{\ell}}\, \\
=&\E{\s{j}{\alpha_1}{\ell}\s{j}{\alpha_2}{\ell}\ds{k}{\alpha_3}{\ell}\ds{k}{\alpha_4}{\ell}\DNTK{k k}{\alpha_3\alpha_4}{\ell}}-\E{\s{j}{\alpha_1}{\ell}\s{j}{\alpha_2}{\ell}}\E{\ds{k}{\alpha_3}{\ell}\ds{k}{\alpha_4}{\ell}\DNTK{k k}{\alpha_3\alpha_4}{\ell}}\, \notag\\
=&\frac{1}{2n_{\ell-1}}\!\sum_{\beta_1,\beta_2,\gamma_1,\gamma_2}\!\!\!\!\!\!\bra\le(\!z_{\beta_1}z_{\beta_2}\!-\!\Ti{G}{\beta_1\beta_2}{\ell}\!\ri) \sigma_{\alpha_1}\sigma_{\alpha_2}\!\ket_{G^{(\ell)}}\!\bra \sigma^{\prime}_{\alpha_3}\sigma^{\prime}_{\alpha_4}\ket_{G^{(\ell)}}\!\!\TI{G}{\beta_1\gamma_1}{\ell}\TI{G}{\beta_2\gamma_2}{\ell}\NTHD{\gamma_1\gamma_2\alpha_3\alpha_4}{\ell}\!+\o{\frac{1}{n^2}}\,\! ,\notag
\end{align}
where in the last line we used the cross-correlation formula~\eqref{eq:most-general} with the observables
$\O = \s{j}{\alpha_1}{\ell}\s{j}{\alpha_2}{\ell}\ds{k}{\alpha_3}{\ell}\ds{k}{\alpha_4}{\ell}$ and $\O = \ds{k}{\alpha_3}{\ell}\ds{k}{\alpha_4}{\ell}$, respectively.
Combining these contributions with the first piece~\eqref{eq:D-piece1}, we get our desired recursion:
\begin{align}\label{eq:D-recursion}
&\NTHD{\alpha_1\alpha_2\alpha_3\alpha_4}{\ell+1}\, \\
=&\CW{\ell+1} \le(\bra \sigma_{\alpha_1}\sigma_{\alpha_2}\Oi{\alpha_3\alpha_4}{\ell+1} \ket_{G^{(\ell)}}-\bra \sigma_{\alpha_1}\sigma_{\alpha_2}\ket_{G^{(\ell)}}\bra\Oi{\alpha_3\alpha_4}{\ell+1} \ket_{G^{(\ell)}}\ri)\, \nonumber\\
&+\frac{n_{\ell}}{4n_{\ell-1}} \CW{\ell+1} \!\!\!\!\!\sum_{\gamma_1, \gamma_2, \gamma_3, \gamma_4} \!\!\!\!\!
\VU{\gamma_1 \gamma_2}{\gamma_3 \gamma_4}{\ell}\!
\bra\le(\!z_{\gamma_1} z_{\gamma_2} \!-\!\Ti{G}{\gamma_1 \gamma_2}{\ell} \!\ri)\sigma_{\alpha_1}\sigma_{\alpha_2}\!\ket_{G^{(\ell)}}\!\bra\le(\!z_{\gamma_3} z_{\gamma_4} \!-\!\Ti{G}{\gamma_3 \gamma_4}{\ell} \!\ri)\Oi{\alpha_3\alpha_4}{\ell+1}\!\ket_{G^{(\ell)}}\, \nonumber\\
&+\frac{n_{\ell}}{2n_{\ell-1}}\!\le(\CW{\ell+1}\ri)^2\!\!\!\!\!\sum_{\beta_1,\beta_2,\gamma_1,\gamma_2}\!\!\!\!\!\NTHD{\gamma_1\gamma_2\alpha_3\alpha_4}{\ell}\!\bra\le(\!z_{\beta_1}z_{\beta_2}\!-\!\Ti{G}{\beta_1\beta_2}{\ell}\!\ri) \sigma_{\alpha_1}\sigma_{\alpha_2}\!\ket_{G^{(\ell)}}\!\TI{G}{\beta_1\gamma_1}{\ell}\TI{G}{\beta_2\gamma_2}{\ell}\!\bra \sigma^{\prime}_{\alpha_3}\sigma^{\prime}_{\alpha_4}\ket_{G^{(\ell)}}\, \notag \\ &+\o{\frac{1}{n}}\, .\nonumber
\end{align}
Interestingly, we see that at leading order the $D$-type cross correlation\index{cross correlation!NTK-preactivation} in layer $(\ell+1)$ mixes $D$-type correlations from layer $\ell$ with the \terminate{four-point vertex} $V^{(\ell)}$, but does not mix with the $F$-type cross correlations\index{cross correlation!NTK-preactivation} or any part of the NTK variance\index{neural tangent kernel!variance}.

\subsubsection{$F$-recursion}\index{cross correlation!NTK-preactivation!F-recursion@$F$-recursion}
Starting from our expression for $F^{(\ell+1)}$~\eqref{eq:F-recursion-almost-there} and decomposing the NTK into a mean and fluctuation, we get
\begin{align}
\NTHF{\alpha_1\alpha_3\alpha_2\alpha_4}{\ell+1}=&\le(\CW{\ell+1}\ri)^2\frac{1}{n_{\ell}}\sum_{j=1}^{n_{\ell}}\E{\s{j}{\alpha_1}{\ell}\ds{j}{\alpha_3}{\ell}\s{j}{\alpha_2}{\ell}\ds{j}{\alpha_4}{\ell}}\Ti{\NTKM}{\alpha_3\alpha_4}{\ell}\, \\
&+\le(\CW{\ell+1}\ri)^2\frac{1}{n_{\ell}}\sum_{j,k=1}^{n_{\ell}}\E{\s{j}{\alpha_1}{\ell}\ds{j}{\alpha_3}{\ell}\s{k}{\alpha_2}{\ell}\ds{k}{\alpha_4}{\ell}\DNTK{jk}{\alpha_3\alpha_4}{\ell}}\, .\nonumber
\end{align}
At leading order, the first term simply becomes a single-neuron Gaussian expectation\index{Gaussian expectation}; the second term can be evaluated with the cross-correlation formula~\eqref{eq:most-general}, where the diagonal sum with $j=k$ is of order $\o{1/n}$ and can be neglected while the off-diagonal sum with $j\ne k$ yields the term involving $F^{(\ell)}$. All together, this gives
\begin{align}\label{eq:F-recursion}
\NTHF{\alpha_1\alpha_3\alpha_2\alpha_4}{\ell+1}
=&\le(\CW{\ell+1}\ri)^2\bra\sigma_{\alpha_1}\sigma_{\alpha_2}\sigma^{\prime}_{\alpha_3}\sigma^{\prime}_{\alpha_4}\ket_{G^{(\ell)}}\Ti{\NTKM}{\alpha_3\alpha_4}{\ell}\, \\
&+\frac{n_{\ell}}{n_{\ell-1}}\le(\CW{\ell+1}\ri)^2\!\!\!\!\!\sum_{\beta_1,\beta_2,\gamma_1,\gamma_2}\!\!\!\!\!\bra \sigma_{\alpha_1}\sigma^{\prime}_{\alpha_3}z_{\beta_1}\ket_{G^{(\ell)}}\bra \sigma_{\alpha_2}\sigma^{\prime}_{\alpha_4}z_{\beta_2}\ket_{G^{(\ell)}}\TI{G}{\beta_1\gamma_1}{\ell}\TI{G}{\beta_2\gamma_2}{\ell}\,\NTHF{\gamma_1\alpha_3\gamma_2\alpha_4}{\ell}\, \nonumber\\
&+\o{\frac{1}{n}}\, . \notag
\end{align}
As with the $D$-recursion before, the first term was present in the second-layer $F^{(2)}$ \eqref{eq:F-recursion-second}, while the second term is a direct consequence of having a fluctuating NTK in the previous layer $\ell$. Additionally, we see that at leading order the $F$-type cross correlation\index{cross correlation!NTK-preactivation} doesn't mix at all with any of our other finite-width tensors.

Finally, before moving on to discuss the NTK variance, let us note that the recursions for both $\NTHDwo{\ell}$ and $\NTHFwo{\ell}$ -- combined with
the initial condition $\NTHDwo{1}=\NTHFwo{1}=0$ from the first layer where the NTK is deterministic -- ensure that they each stay of order one. Given the factor of $1/n_\ell$ in the decomposition of the cross correlation\index{cross correlation!NTK-preactivation} into these tensors \eqref{eq:D-F-decomposition-general-layer} and our ``second lesson'' encapsulated by the cross-correlation formula~\eqref{eq:most-general}, this means that any and all cross correlations\index{cross correlation!NTK-preactivation} are suppressed in the \terminate{$1/n$ expansion} and vanish identically in the strict \terminate{infinite-width limit}.

\subsection{NTK Variance}\label{subsec:NTHvariance}
Now let's finally \emph{slay the beast}\index{slay the beast (NTK variance)} that is the NTK variance\index{neural tangent kernel!variance|textbf}.
Similar to the NTK-preactivation cross correlation\index{cross correlation!NTK-preactivation}, the NTK-variance calculation in deeper layers differs from the second-layer calculation due to nontrivial intralayer correlations~\eqref{eq:general-covariance-reprint} in the previous layer
and due to the fluctuating NTK~\eqref{eq:NTK-fluc-decomposition}.

The NTK variance is given by the expected magnitude of the NTK fluctuation
\begin{align}\label{eq:NTHvR_primitive_before}
&\E{\DNTK{i_1i_2}{\alpha_1\alpha_2}{\ell+1}\DNTK{i_3i_4}{\alpha_3\alpha_4}{\ell+1}}=\cov{\Tia{\NTK}{i_1i_2}{\alpha_1\alpha_2}{\ell+1}}{\Tia{\NTK}{i_3i_4}{\alpha_3\alpha_4}{\ell+1}}\, .
\end{align}
To begin our calculation, let us plug the NTK forward equation~\eqref{eq:NTH-recursion-without-expectation} into this defining expression %
and then integrate out\index{integrating out} the weights $W^{(\ell+1)}$, which is easy since they are independent random variables. Although there are many terms, the algebra is mostly straightforward:
\begin{align}\label{eq:NTHvR_primitive}
&\E{\DNTK{i_1i_2}{\alpha_1\alpha_2}{\ell+1}\DNTK{i_3i_4}{\alpha_3\alpha_4}{\ell+1}}\\
=&\delta_{i_1i_2}\delta_{i_3i_4}\frac{1}{n_{\ell}^2}\sum_{j,k=1}^{n_{\ell}}\Bigg\{\le(\lamW{\ell+1}\ri)^2\cov{\s{j}{\alpha_1}{\ell} \s{j}{\alpha_2}{\ell}}{\s{k}{\alpha_3}{\ell} \s{k}{\alpha_4}{\ell}}  \notag \\
&\quad \quad \quad \quad \quad \quad +\le(\lamW{\ell+1}\ri)\CW{\ell+1}\cov{\s{j}{\alpha}{\ell} \s{j}{\alpha_2}{\ell}}{\ds{k}{\alpha_3}{\ell} \ds{k}{\alpha_4}{\ell} \Tia{\NTH}{kk}{\alpha_3\alpha_4}{\ell} } \nonumber\\
&\quad \quad \quad \quad \quad \quad +\le(\lamW{\ell+1}\ri)\CW{\ell+1}\cov{\ds{j}{\alpha}{\ell} \ds{j}{\alpha_2}{\ell}\Tia{\NTH}{jj}{\alpha\alpha_2}{\ell}}{\s{k}{\alpha_3}{\ell} \s{k}{\alpha_4}{\ell}  }\nonumber\\
&\quad \quad \quad \quad \quad \quad +\le(\CW{\ell+1}\ri)^2\cov{\ds{j}{\alpha_1}{\ell} \ds{j}{\alpha_2}{\ell}\Tia{\NTH}{jj}{\alpha_1\alpha_2}{\ell}}{\ds{k}{\alpha_3}{\ell} \ds{k}{\alpha_4}{\ell} \Tia{\NTH}{kk}{\alpha_3\alpha_4}{\ell} }\Bigg\}\nonumber\\
&+\delta_{i_1i_3}\delta_{i_2 i_4}\le(\CW{\ell+1}\ri)^2\frac{1}{n_{\ell}^2}\sum_{j,k=1}^{n_{\ell}}\E{\ds{j}{\alpha_1}{\ell} \ds{k}{\alpha_2}{\ell}\Tia{\NTH}{jk}{\alpha_1\alpha_2}{\ell} \ds{j}{\alpha_3}{\ell} \ds{k}{\alpha_4}{\ell} \Tia{\NTH}{jk}{\alpha_3\alpha_4}{\ell} }\nonumber\\
&+\delta_{i_1i_4}\delta_{i_2 i_3}\le(\CW{\ell+1}\ri)^2\frac{1}{n_{\ell}^2}\sum_{j,k=1}^{n_{\ell}}\E{\ds{j}{\alpha_1}{\ell} \ds{k}{\alpha_2}{\ell}\Tia{\NTH}{jk}{\alpha_1\alpha_2}{\ell} \ds{k}{\alpha_3}{\ell} \ds{j}{\alpha_4}{\ell} \Tia{\NTH}{kj}{\alpha_3\alpha_4}{\ell} }\, .\nonumber
\end{align}
In obtaining these last three terms,
you should have made three distinct pairings for the two pairs of Wick contractions\index{Wick contraction} of the four $W^{(\ell+1)}$'s, one paring %
within the same NTK and two pairings 
across the NTKs.

An inspection of the pattern of \terminate{neural indices} in the \terminate{Kronecker delta}s from~\eqref{eq:NTHvR_primitive} suggests that we should again decompose the NTK variance\index{neural tangent kernel!variance} into two tensors as
\begin{align}\label{eq:NTH-variance-decomposition}\index{tensor decomposition!NTK variance $A$/$B$}
&\E{\DNTK{i_1i_2}{\alpha_1\alpha_2}{\ell}\DNTK{i_3i_4}{\alpha_3\alpha_4}{\ell}}\, \\
\equiv&\frac{1}{n_{\ell-1}}\le[\delta_{i_1i_2}\delta_{i_3i_4} \NTHA{\alpha_1\alpha_2}{\alpha_3\alpha_4}{\ell} + \delta_{i_1i_3}\delta_{i_2i_4} \NTHB{\alpha_1\alpha_3\alpha_2\alpha_4}{\ell} + \delta_{i_1i_4}\delta_{i_2i_3} \NTHB{\alpha_1\alpha_4\alpha_2\alpha_3}{\ell} \ri] \, ,\notag
\end{align}
just as we did for the second layer before in~\eqref{eq:NTH-variance-decomposition-second}.
Here, a factor of $1/n_{\ell-1}$ was pulled out in anticipation that the overall variance will be $O(1/n)$ just as it was for the second layer.
For now you can think of this parameterization as an ansatz; we will soon recursively show that $\NTHA{\alpha_1\alpha_2}{\alpha_3\alpha_4}{\ell}$ and $\NTHB{\alpha_1\alpha_3\alpha_2\alpha_4}{\ell}$ stay of order one as the network width increases.%

Now, let's work out the layer recursions for $\NTHAwo{\ell}$ and $\NTHBwo{\ell}$.

\subsubsection{$B$-recursion}\index{neural tangent kernel!variance!B-recursion@$B$-recursion}
We'll start with $B$-recursion because it's simpler.
Considering 
\eqref{eq:NTHvR_primitive} with the decomposition~\eqref{eq:NTH-variance-decomposition} in mind, we see that $\NTHBwo{\ell+1}$ is given by the following $\ell$-th-layer expectation:
\be\label{eq:Bstart}
\NTHB{\alpha_1\alpha_3\alpha_2\alpha_4}{\ell+1}=\le(\CW{\ell+1}\ri)^2\frac{1}{n_{\ell}}\sum_{j,k=1}^{n_{\ell}}\E{\ds{j}{\alpha_1}{\ell} \ds{k}{\alpha_2}{\ell}\Tia{\NTH}{jk}{\alpha_1\alpha_2}{\ell} \ds{j}{\alpha_3}{\ell} \ds{k}{\alpha_4}{\ell} \Tia{\NTH}{jk}{\alpha_3\alpha_4}{\ell} }\, .
\ee
As should now be familiar, the double summation in~\eqref{eq:Bstart} splits into two types of terms, diagonal ones with $j=k$ and off-diagonal ones with $j\ne k$.

\index{neural tangent kernel!mean}
For the diagonal part, the leading contribution is from the NTK mean 
\begin{align}\label{eq:NTHvariance_piece1}
&\le(\CW{\ell+1}\ri)^2\frac{1}{n_{\ell}}\sum_{j=1}^{n_{\ell}}\E{\ds{j}{\alpha_1}{\ell} \ds{j}{\alpha_2}{\ell} \ds{j}{\alpha_3}{\ell} \ds{j}{\alpha_4}{\ell} }\Ti{\NTKM}{\alpha_1\alpha_2}{\ell}\Ti{\NTKM}{\alpha_3\alpha_4}{\ell} +\o{\frac{1}{n}}\, \\
=&\le(\CW{\ell+1}\ri)^2\bra\sigma^{\prime}_{\alpha_1} \sigma^{\prime}_{\alpha_2}\sigma^{\prime}_{\alpha_3} \sigma^{\prime}_{\alpha_4}  \ket_{G^{(\ell)}}\Ti{\NTKM}{\alpha_1\alpha_2}{\ell}\Ti{\NTKM}{\alpha_3\alpha_4}{\ell} +\o{\frac{1}{n}}\, ,\nonumber
\end{align}
which is analogous to what we found in the second layer~\eqref{eq:B-recursion-second}.\footnote{Again, the subleading $\o{1/n}$ piece includes both a contribution from the non-Gaussian distribution as well as a cross correlation contribution from the previous layer.
}

\index{neural tangent kernel!mean}\index{neural tangent kernel!variance}
For the off-diagonal part of \eqref{eq:Bstart}, the NTK mean vanishes and the leading contribution is from the NTK fluctuation
\begin{align}\label{eq:NTHvariance_piece2_first_line}
&\le(\CW{\ell+1}\ri)^2\frac{1}{n_{\ell}}\sum_{\substack{j,k=1\\j\ne k}}^{n_{\ell}}\E{\ds{j}{\alpha_1}{\ell} \ds{k}{\alpha_2}{\ell} \ds{j}{\alpha_3}{\ell} \ds{k}{\alpha_4}{\ell}\DNTK{jk}{\alpha_1\alpha_2}{\ell}\DNTK{jk}{\alpha_3\alpha_4}{\ell} }\, .
\end{align}
The expectation already is $\o{\Delta^2}$ from the two NTK fluctuations inside it and thus, neglecting higher-order correlations of order $\o{\Delta^3}$, we have
\begin{align}\label{eq:the-most-subtle}
&\E{\ds{j}{\alpha_1}{\ell} \ds{k}{\alpha_2}{\ell} \ds{j}{\alpha_3}{\ell} \ds{k}{\alpha_4}{\ell}\DNTK{jk}{\alpha_1\alpha_2}{\ell}\DNTK{jk}{\alpha_3\alpha_4}{\ell} }\, \\
=&\E{\ds{j}{\alpha_1}{\ell} \ds{k}{\alpha_2}{\ell} \ds{j}{\alpha_3}{\ell} \ds{k}{\alpha_4}{\ell}}\E{\DNTK{jk}{\alpha_1\alpha_2}{\ell}\DNTK{jk}{\alpha_3\alpha_4}{\ell} }+\o{\frac{1}{n^2}}\, ,\notag
\end{align}
where the detailed explanation for such a factorization is given in this footnote.\footnote{In greater detail, you can think of what we are doing here as separating $\ds{j}{\alpha_1}{\ell} \ds{k}{\alpha_2}{\ell} \ds{j}{\alpha_3}{\ell} \ds{k}{\alpha_4}{\ell}$ into a mean $\E{\ds{j}{\alpha_1}{\ell} \ds{k}{\alpha_2}{\ell} \ds{j}{\alpha_3}{\ell} \ds{k}{\alpha_4}{\ell}}$ and fluctuation 
and -- since the expectation already contains two fluctuations $\DNTK{jk}{\alpha_1\alpha_2}{\ell}\DNTK{jk}{\alpha_3\alpha_4}{\ell}$ -- the latter fluctuating piece contributes $\o{\Delta^3}$ and thus can be neglected.

In alternate detail, we can view this expectation~\eqref{eq:the-most-subtle} as a correlator of three random variables $\ds{j}{\alpha_1}{\ell} \ds{k}{\alpha_2}{\ell} \ds{j}{\alpha_3}{\ell} \ds{k}{\alpha_4}{\ell}$, $\DNTK{jk}{\alpha_1\alpha_2}{\ell}$, and $\DNTK{jk}{\alpha_3\alpha_4}{\ell}$, and decompose it into one-point and two-point correlators as
\begin{align}\label{eq:correlator-paradise}
&\E{\ds{j}{\alpha_1}{\ell} \ds{k}{\alpha_2}{\ell} \ds{j}{\alpha_3}{\ell} \ds{k}{\alpha_4}{\ell}\DNTK{jk}{\alpha_1\alpha_2}{\ell}\DNTK{jk}{\alpha_3\alpha_4}{\ell} }  \,   \\
=&\E{\ds{j}{\alpha_1}{\ell} \ds{k}{\alpha_2}{\ell} \ds{j}{\alpha_3}{\ell} \ds{k}{\alpha_4}{\ell}} \E{\DNTK{jk}{\alpha_1\alpha_2}{\ell}}\E{\DNTK{jk}{\alpha_3\alpha_4}{\ell} } \, \notag\\
&+\E{\ds{j}{\alpha_1}{\ell} \ds{k}{\alpha_2}{\ell} \ds{j}{\alpha_3}{\ell} \ds{k}{\alpha_4}{\ell}} \E{\DNTK{jk}{\alpha_1\alpha_2}{\ell}\DNTK{jk}{\alpha_3\alpha_4}{\ell} } \, \notag  \\
&+\E{\ds{j}{\alpha_1}{\ell} \ds{k}{\alpha_2}{\ell} \ds{j}{\alpha_3}{\ell} \ds{k}{\alpha_4}{\ell} \DNTK{jk}{\alpha_1\alpha_2}{\ell}}\E{\DNTK{jk}{\alpha_3\alpha_4}{\ell} }
 \, \notag \\
&+\E{\ds{j}{\alpha_1}{\ell} \ds{k}{\alpha_2}{\ell} \ds{j}{\alpha_3}{\ell} \ds{k}{\alpha_4}{\ell} \DNTK{jk}{\alpha_3\alpha_4}{\ell}}\E{\DNTK{jk}{\alpha_1\alpha_2}{\ell} } + \o{\frac{1}{n^2}}\, \notag ,
\end{align}
where the $\o{1/n^2}$ part contains the connected piece of the decomposition.
Since the NTK fluctuation has mean zero, only the second term survives at this order.}
Then, using the decomposition~\eqref{eq:NTH-variance-decomposition} for the NTK variance\index{neural tangent kernel!variance} and similar logic as \eqref{eq:four-activations-different-deep-connected} to evaluate the four-point correlator of off-diagonal activations, we get
\begin{align}\label{eq:NTHvariance_piece2}
&\le(\CW{\ell+1}\ri)^2\frac{1}{n_{\ell}}\sum_{\substack{j,k=1\\j\ne k}}^{n_{\ell}}\E{\ds{j}{\alpha_1}{\ell} \ds{k}{\alpha_2}{\ell} \ds{j}{\alpha_3}{\ell} \ds{k}{\alpha_4}{\ell}}\E{\DNTK{jk}{\alpha_1\alpha_2}{\ell}\DNTK{jk}{\alpha_3\alpha_4}{\ell} }+\o{\frac{1}{n}}\, \\
=&\le(\CW{\ell+1}\ri)^2\bra\sigma^{\prime}_{\alpha_1} \sigma^{\prime}_{\alpha_3} \ket_{G^{(\ell)}}\bra\sigma^{\prime}_{\alpha_2} \sigma^{\prime}_{\alpha_4}  \ket_{G^{(\ell)}}\frac{n_{\ell}}{n_{\ell-1}}\NTHB{\alpha_1\alpha_3\alpha_2\alpha_4}{\ell}+\o{\frac{1}{n}}\, .\nonumber
\end{align}

Substituting both the diagonal contribution~\eqref{eq:NTHvariance_piece1} and off-diagonal contribution~\eqref{eq:NTHvariance_piece2} back into~\eqref{eq:Bstart}, we get the $B$-recursion:
\begin{align}\label{eq:B-recursion}
\NTHB{\alpha_1\alpha_3\alpha_2\alpha_4}{\ell+1}&= \le(\CW{\ell+1}\ri)^2\Bigg[\bra \dsNL{\alpha_1} \dsNL{\alpha_2}  \dsNL{\alpha_3} \dsNL{\alpha_4}\ket_{G^{(\ell)}}  \Ti{\NTKM}{\alpha_1\alpha_2}{\ell} \Ti{\NTKM}{\alpha_3\alpha_4}{\ell} \\
&\ \ \ \ \ \ \ \ \ \ \ \ \ \ \ \ \  +  \le(\frac{n_\ell}{n_{\ell-1}} \ri) \bra\sigma^{\prime}_{\alpha_1} \sigma^{\prime}_{\alpha_3} \ket_{G^{(\ell)}}\bra\sigma^{\prime}_{\alpha_2} \sigma^{\prime}_{\alpha_4}  \ket_{G^{(\ell)}} \NTHB{\alpha_1\alpha_3\alpha_2\alpha_4}{\ell}\Bigg] +\oninv \, .\nonumber
\end{align}
As promised, we recursively see that $B^{(\ell)}$ is an order-one quantity. Additionally, we note that at leading order this $B$-type NTK variance doesn't mix with any other finite-width tensors.

\subsubsection{$A$-recursion}\index{neural tangent kernel!variance!A-recursion@$A$-recursion}
Let us now determine the $A$-recursion.
Again equating our expression for the NTK variance \eqref{eq:NTHvR_primitive} with the $A/B$-decomposition \eqref{eq:NTH-variance-decomposition}, we see that $A^{(\ell+1)}$ is given by the following $\ell$-th-layer covariances:
\begin{align}\label{eq:A-recursion-mid-point}
\NTHA{\alpha_1\alpha_2}{\alpha_3\alpha_4}{\ell+1}=&\frac{1}{n_{\ell}}\sum_{j,k=1}^{n_{\ell}}\Bigg\{\le(\lamW{\ell+1}\ri)^2\cov{\s{j}{\alpha_1}{\ell} \s{j}{\alpha_2}{\ell}}{\s{k}{\alpha_3}{\ell} \s{k}{\alpha_4}{\ell}}   \\
&\quad \quad \quad \quad +\lamW{\ell+1}\CW{\ell+1}\cov{\s{j}{\alpha}{\ell} \s{j}{\alpha_2}{\ell}}{\ds{k}{\alpha_3}{\ell} \ds{k}{\alpha_4}{\ell} \Tia{\NTH}{kk}{\alpha_3\alpha_4}{\ell} } \nonumber\\
&\quad \quad \quad \quad +\lamW{\ell+1}\CW{\ell+1}\cov{\ds{j}{\alpha}{\ell} \ds{j}{\alpha_2}{\ell}\Tia{\NTH}{jj}{\alpha\alpha_2}{\ell}}{\s{k}{\alpha_3}{\ell} \s{k}{\alpha_4}{\ell}  }\nonumber\\
&\quad \quad \quad \quad +\le(\CW{\ell+1}\ri)^2\cov{\ds{j}{\alpha_1}{\ell} \ds{j}{\alpha_2}{\ell}\Tia{\NTH}{jj}{\alpha_1\alpha_2}{\ell}}{\ds{k}{\alpha_3}{\ell} \ds{k}{\alpha_4}{\ell} \Tia{\NTH}{kk}{\alpha_3\alpha_4}{\ell} }\Bigg\}\, .\nonumber
\end{align}
As we've now seen many times previously, our approach will be to divide up the double summation in \eqref{eq:A-recursion-mid-point} into two types of terms, diagonal terms on a single neuron with $j=k$ and off-diagonal terms on pairs of neurons with $j\ne k$.

\index{neural tangent kernel!mean}
As was the case for the $B$-recursion,
the leading contribution from the diagonal part with $j=k$ comes from the NTK mean and matches what we found for the second layer~\eqref{eq:A-recursion-second}
\be\label{eq:NTHA-piece1}
\bra \Oi{\alpha_1\alpha_2}{\ell+1} \Oi{\alpha_3\alpha_4}{\ell+1} \ket_{G^{(\ell)}}-\bra \Oi{\alpha_1\alpha_2}{\ell+1}\ket_{G^{(\ell)}}\bra \Oi{\alpha_3\alpha_4}{\ell+1} \ket_{G^{(\ell)}}+\o{\frac{1}{n}}\, ,
\ee
where the definition of the auxiliary stochastic tensor $\Oi{\alpha_1\alpha_2}{\ell+1}$ was given in~\eqref{eq:def-omega-without-neural}.\footnote{Once again, the subleading $\o{1/n}$ piece includes both a contribution from the non-Gaussian distribution as well as a cross correlation contribution from the previous layer
\emph{and} now also a contribution from the previous layer's NTK variance.}

\index{neural tangent kernel!mean}
This leaves us with the off-diagonal part of~\eqref{eq:A-recursion-mid-point} with $j\ne k$. Here, there will be leading contributions both from the NTK mean and from the NTK fluctuations. The contributions from the mean are given by
replacing $\Tia{\NTH}{i_1i_2}{\alpha_1\alpha_2}{\ell} \to \delta_{i_1i_2}\Ti{\NTKM}{\alpha_1\alpha_2}{\ell}$:
\begin{align}\label{eq:A-recursion-mean}
&\frac{1}{n_{\ell}}\sum_{\substack{ j,k=1\\j\ne k}}^{n_{\ell}}\Bigg\{\le(\lamW{\ell+1}\ri)^2\cov{\s{j}{\alpha_1}{\ell} \s{j}{\alpha_2}{\ell}}{\s{k}{\alpha_3}{\ell} \s{k}{\alpha_4}{\ell}}   \\
&\quad \quad \quad \quad +\lamW{\ell+1}\CW{\ell+1}\Ti{\NTKM}{\alpha_3\alpha_4}{\ell}\cov{\s{j}{\alpha}{\ell} \s{j}{\alpha_2}{\ell}}{\ds{k}{\alpha_3}{\ell} \ds{k}{\alpha_4}{\ell} } \nonumber\\
&\quad \quad \quad \quad +\lamW{\ell+1}\CW{\ell+1}\Ti{\NTKM}{\alpha_1\alpha_2}{\ell}\cov{\ds{j}{\alpha}{\ell} \ds{j}{\alpha_2}{\ell}}{\s{k}{\alpha_3}{\ell} \s{k}{\alpha_4}{\ell}  }\nonumber\\
&\quad \quad \quad \quad +\le(\CW{\ell+1}\ri)^2\Ti{\NTKM}{\alpha_1\alpha_2}{\ell}\Ti{\NTKM}{\alpha_3\alpha_4}{\ell}\cov{\ds{j}{\alpha_1}{\ell} \ds{j}{\alpha_2}{\ell}}{\ds{k}{\alpha_3}{\ell} \ds{k}{\alpha_4}{\ell}}\Bigg\} .\nonumber
\end{align}
All four of these covariances can be evaluated using the intralayer formula~\eqref{eq:general-covariance-reprint}. After a little bit of algebra, this gives 
\begin{align}\label{eq:NTHA-piece2}%
&\frac{n_{\ell}}{4n_{\ell-1}}  \sum_{\gamma_1, \gamma_2, \gamma_3, \gamma_4} \!\!\!\!
\VU{\gamma_1 \gamma_2}{\gamma_3 \gamma_4}{\ell}
\bra\le(z_{\gamma_1} z_{\gamma_2} -\Ti{G}{\gamma_1 \gamma_2}{\ell} \ri)\Oi{\alpha_1\alpha_2}{\ell+1}\ket_{G^{(\ell)}}
\bra\le(z_{\gamma_3} z_{\gamma_4} -\Ti{G}{\gamma_3 \gamma_4}{\ell} \ri)\Oi{\alpha_3\alpha_4}{\ell+1}\ket_{G^{(\ell)}}\, 
\end{align}
at leading order, where we again made use of the definition of $\Oi{\alpha_1\alpha_2}{\ell+1}$~\eqref{eq:def-omega-without-neural}.

Finally, we're left with the off-diagonal contributions from the NTK fluctuations, which -- if we write them out in excruciating detail -- are given by
\begin{align}\label{eq:A-recursion-rest}
&\frac{1}{n_{\ell}}\sum_{\substack{ j,k=1\\j\ne k}}^{n_{\ell}}\Bigg\{\le(\lamW{\ell+1}\ri)\CW{\ell+1}\cov{\s{j}{\alpha_1}{\ell} \s{j}{\alpha_2}{\ell}}{\ds{k}{\alpha_3}{\ell} \ds{k}{\alpha_4}{\ell} \DNTK{kk}{\alpha_3\alpha_4}{\ell} }\, \\
&\quad \quad \quad \quad +\le(\lamW{\ell+1}\ri)\CW{\ell+1}\cov{\ds{j}{\alpha_1}{\ell} \ds{j}{\alpha_2}{\ell}\DNTK{jj}{\alpha_1\alpha_2}{\ell}}{\s{k}{\alpha_3}{\ell} \s{k}{\alpha_4}{\ell}  }\, \nonumber\\
&\quad \quad \quad \quad +\le(\CW{\ell+1}\ri)^2\Ti{\NTKM}{\alpha_1\alpha_2}{\ell}\cov{\ds{j}{\alpha_1}{\ell} \ds{j}{\alpha_2}{\ell}}{\ds{k}{\alpha_3}{\ell} \ds{k}{\alpha_4}{\ell} \DNTK{kk}{\alpha_3\alpha_4}{\ell} }\, \nonumber\\
&\quad \quad \quad \quad +\le(\CW{\ell+1}\ri)^2\Ti{\NTKM}{\alpha_3\alpha_4}{\ell}\cov{\ds{j}{\alpha_1}{\ell} \ds{j}{\alpha_2}{\ell}\DNTK{jj}{\alpha_1\alpha_2}{\ell}}{\ds{k}{\alpha_3}{\ell} \ds{k}{\alpha_4}{\ell}}\, \nonumber\\
&\quad \quad \quad \quad +\le(\CW{\ell+1}\ri)^2\cov{\ds{j}{\alpha_1}{\ell} \ds{j}{\alpha_2}{\ell}\DNTK{jj}{\alpha_1\alpha_2}{\ell}}{\ds{k}{\alpha_3}{\ell} \ds{k}{\alpha_4}{\ell} \DNTK{kk}{\alpha_3\alpha_4}{\ell} }\Bigg\}\nonumber \, .
\end{align}
The last term involving the two NTK fluctuations can be evaluated similarly to how we evaluated such a term for the $B$-recursion in~\eqref{eq:the-most-subtle},\footnote{The only difference between this and the $B$-version before~\eqref{eq:the-most-subtle} is that here, since we're evaluating a covariance, the term
\be
\E{\ds{j}{\alpha_1}{\ell} \ds{j}{\alpha_2}{\ell}\DNTK{jj}{\alpha_1\alpha_2}{\ell}}\E{\ds{k}{\alpha_3}{\ell} \ds{k}{\alpha_4}{\ell} \DNTK{kk}{\alpha_3\alpha_4}{\ell}}
\ee
is being subtracted. However, this term is of order $\o{1/n^2}$ and can thus be neglected.}
here giving
\be\label{eq:NTHA-piece3}
\le(\CW{\ell+1}\ri)^2\bra\sigma^{\prime}_{\alpha_1}\sigma^{\prime}_{\alpha_2}\ket_{G^{(\ell)}}\bra\sigma^{\prime}_{\alpha_3}\sigma^{\prime}_{\alpha_4}\ket_{G^{(\ell)}}\frac{n_{\ell}}{n_{\ell-1}}\NTHA{\alpha_1\alpha_2}{\alpha_3\alpha_4}{\ell}+\o{\frac{1}{n}}\, .
\ee
The remaining four covariances in \eqref{eq:A-recursion-rest} with only a single NTK fluctuation 
are identical in structure to~\eqref{eq:how-to-unroll}, letting us leave the details of this to you and your roll of parchment.

At this point, let's review all the components of our expression 
for $A^{(\ell+1)}$: we have the diagonal contribution~\eqref{eq:NTHA-piece1}; and off-diagonal contributions from the NTK mean~\eqref{eq:NTHA-piece2}, from the covariance of two NTK fluctuations~\eqref{eq:NTHA-piece3}, and from the four covariances
on your parchment. 
Assembling these components, we get the $A$-recursion:
\begin{align}\label{eq:A-recursion}
&\NTHA{\alpha_1\alpha_2}{\alpha_3\alpha_4}{\ell+1}\, \\
=&\bra \Oi{\alpha_1\alpha_2}{\ell+1} \Oi{\alpha_3\alpha_4}{\ell+1} \ket_{G^{(\ell)}}-\bra \Oi{\alpha_1\alpha_2}{\ell+1}\ket_{G^{(\ell)}}\bra \Oi{\alpha_3\alpha_4}{\ell+1} \ket_{G^{(\ell)}}\, \nonumber\\
&+\frac{n_{\ell}}{4n_{\ell-1}} \! \sum_{\gamma_1, \gamma_2, \gamma_3, \gamma_4} \!\!\!
\VU{\gamma_1 \gamma_2}{\gamma_3 \gamma_4}{\ell}
\bra\Oi{\alpha_1\alpha_2}{\ell+1}\le(z_{\gamma_1} z_{\gamma_2} -\Ti{G}{\gamma_1 \gamma_2}{\ell} \ri)\ket_{G^{(\ell)}}\bra\Oi{\alpha_3\alpha_4}{\ell+1} \le(z_{\gamma_3} z_{\gamma_4} -\Ti{G}{\gamma_3 \gamma_4}{\ell} \ri)\ket_{G^{(\ell)}}\, \nonumber\\
&+\frac{n_{\ell}}{n_{\ell-1}}\le(\CW{\ell+1}\ri)^2\bra\sigma^{\prime}_{\alpha_1}\sigma^{\prime}_{\alpha_2}\ket_{G^{(\ell)}}\bra\sigma^{\prime}_{\alpha_3}\sigma^{\prime}_{\alpha_4}\ket_{G^{(\ell)}}\NTHA{\alpha_1\alpha_2}{\alpha_3\alpha_4}{\ell}\, \, \nonumber\\
&+\frac{n_{\ell}}{n_{\ell-1}}\frac{\CW{\ell+1}}{2}\!\!\!\sum_{\beta_1,\beta_2,\gamma_1,\gamma_2}\!\!\!\!\!\Big[\bra \Oi{\alpha_1\alpha_2}{\ell+1}\le(z_{\beta_1}z_{\beta_2}-\Ti{G}{\beta_1\beta_2}{\ell}\ri)\ket_{G^{(\ell)}}\TI{G}{\beta_1\gamma_1}{\ell}\TI{G}{\beta_2\gamma_2}{\ell}\NTHD{\gamma_1\gamma_2\alpha_3\alpha_4}{\ell}\bra \dsNL{\alpha_3} \dsNL{\alpha_4}\ket_{G^{(\ell)}}\, \nonumber\\
&\quad \quad \quad \quad  \quad \quad  \quad \quad +\bra \Oi{\alpha_3\alpha_4}{\ell+1}\le(z_{\beta_1}z_{\beta_2}-\Ti{G}{\beta_1\beta_2}{\ell}\ri)\ket_{G^{(\ell)}}\TI{G}{\beta_1\gamma_1}{\ell}\TI{G}{\beta_2\gamma_2}{\ell}\NTHD{\gamma_1\gamma_2\alpha_1\alpha_2}{\ell}\bra \dsNL{\alpha_1} \dsNL{\alpha_2}\ket_{G^{(\ell)}}\Big]\, \nonumber\\
&+\o{\frac{1}{n}}\, .\nonumber
\end{align}
As promised, we recursively see that $A^{(\ell)}$ is an order-one quantity.
Additionally, we note with interest that at leading order this $A$-type contribution to the NTK variance at layer $(\ell+1)$ mixes with the \terminate{four-point vertex} $V^{(\ell)}$ and the cross correlation\index{cross correlation!NTK-preactivation} $D^{(\ell)}$ in layer $\ell$, though not with $B^{(\ell)}$ or $F^{(\ell)}$.\index{$1/n$ expansion}

This completes our analysis of all the finite-width effects for the NTK-preactivation joint distribution; both the leading NTK-preactivation cross correlations\index{cross correlation!NTK-preactivation} and the NTK variance\index{neural tangent kernel!variance} scale as $\sim 1/n$ in the large-width expansion and vanish in the \terminate{infinite-width limit}.\footnote{
    Recalling our discussion from footnote~\ref{footnote:self-average} in \S\ref{sec:MLP_distribution}, this means that the NTK \emph{self-averages}\index{self-averaging} in the strict \terminate{infinite-width limit}; in this limit, the particular value of the NTK in any instantiation of the network parameters is fixed and equal to the ensemble mean.\index{neural tangent kernel!mean}\index{neural tangent kernel!variance}
} 
These quantities are sufficient to fully characterize the leading finite-width effects of gradient-based learning.

%% file: Chp9-NTKEFT/9_global.tex
\chapter{Effective Theory of the NTK at Initialization}\label{ch:eft-ntk}%

\epigraph{
In short, we believe that we have answered Minsky and Papert's challenge and \emph{have} found a learning result sufficiently powerful to demonstrate that their pessimism about learning in multilayer machines was misplaced.
}{Rumelhart, Hinton, and Williams \cite{rumelhart1985learning}, acausally
rising to meet the criticism
from \S\ref{ch:eft-mlp}.
\index{Rumelhart, David Everett}\index{Hinton, Geoffrey Everest}\index{Williams, Ronald J.}\index{multilayer perceptron}\index{Minsky, Marvin}\index{Papert, Seymour} }

\noindent{}Since the last chapter was a tempest of equations, algebra, and integration,
let's take some moments\index{moment}
to 
value 
our expectations.\index{expectation value}

\index{connected correlator!}
Our goal in \S\ref{ch:NTKa} was to determine the NTK-preactivation joint distribution  for a given layer $\ell$ at initialization: $p\!\le(z^{(\ell)}, \NTK^{(\ell)}\Big\vert\D\ri)$. The data-dependent couplings\index{data-dependent coupling} and the connected correlators of this distribution \emph{run}\index{running coupling} with depth according to the recursions that we just laboriously derived --~\eqref{eq:NTHmean_final} for the NTK mean\index{neural tangent kernel!mean},~\eqref{eq:D-recursion} and~\eqref{eq:F-recursion} for the NTK-preactivation cross correlations\index{cross correlation!NTK-preactivation}, and~\eqref{eq:B-recursion} and~\eqref{eq:A-recursion} for the NTK variance\index{neural tangent kernel!variance} -- in addition to the recursions derived in~\S\ref{ch:ngp} 
for the \terminate{kernel} \eqref{eq:K-recursion-reprint} and for the \terminate{four-point vertex} \eqref{eq:V-recursion-reprint}.
This \emph{RG-flow}\index{representation group flow} analysis taught us that the NTK is a deterministic object in the first layer (\S\ref{sec:first-layer-deterministic-NTK}),
stochastically fluctuates and cross-correlates in the second layer (\S\ref{sec:second-layer-fluctuating-NTK}),
and then further accumulates fluctuations and cross correlations in deeper layers (\S\ref{sec:deeper-layer-accumulation-NTK}).

Now that we've thoroughly discussed the math,
in this chapter we'll finally be able to consider the physics of this joint distribution.
Building on our discussion of \neo{criticality} and \neo{universality} in~\S\ref{ch:eft-mlp}, we'll first lay the groundwork for a similar analysis of the NTK while highlighting the relevant results from the last chapter (\S\ref{sec:ntk_criticality}).
In particular, our focus will be on understanding how the \terminate{initialization hyperparameters} and the \terminate{training hyperparameters} affect \terminate{gradient descent} at finite width.
We'll once again find that the depth-to-width ratio $L/n$ plays a starring role in controlling finite-width effects, first for the scale-invariant universality class\index{universality class!scale-invariant} (\S\ref{sec:ntk_criticality_scale_invariant}) and then for the $\ker^\star=0$ universality class\index{universality class!K@$K^\star=0$} (\S\ref{sec:ntk_criticality_tanh_univ}).
For both cases, the growing importance of NTK fluctuations and cross correlations\index{cross correlation!NTK-preactivation} with depth makes the finite-width interaction \emph{relevant}\index{relevant (RG flow)} under RG flow of the NTK.\index{representation group flow}\index{neural tangent kernel}

Finally, we'll introduce the infamous \neo{exploding and vanishing gradient problem} of \terminate{deep learning} and see how our notion of \terminate{criticality} completely mitigates this problem (\S\ref{sec:EVGP-WEP}). In this context, we also explain how the bias and weight learning rates should each be scaled with the network depth.

\section{Criticality Analysis of the NTK}\label{sec:ntk_criticality}
Let's set the stage for our \terminate{criticality} analysis.
As we did in our discussion of preactivation criticality in~\S\ref{ch:eft-mlp}, throughout this section we'll set the bias variance $C_b^{(\ell)}$ and the rescaled weight variance $C_W^{(\ell)}$ to be uniform across layers
\be\label{eq:ntk-chapter-initialziation-hyperparameters-dropped-layer-dependence}
C_b^{(\ell)}=C_b\, , \qquad C_W^{(\ell)}=C_W\, .
\ee
Further paralleling 
\S\ref{sec:signal_prop_finite_width}, we will consider MLPs with uniform hidden layer widths
\be\label{eq:ntk-chapter-layer-widths-equalized}
n_1=\ldots=n_{L-1}\equiv n\, ,
\ee
which is a sensible choice in practice
as well as notationally simplifying.

For the \terminate{training hyperparameters}, however, we'll preserve the layer dependence of the bias learning rate $\Lb{\ell}$ and weight learning rate $\LW{\ell}$ for now, as different universality classes will require different treatments. We'll explore the general principle behind these hyperparameter choices in \S\ref{sec:EVGP-WEP}.

Going forward, we'll only focus on the leading contributions from the \terminate{$1/n$ expansion} to the single-input statistics, neglecting the \terminate{subleading corrections} at next-to-leading-order and 
reserving the multi-input analysis for your private amusement.

\subsubsection{Leading-order NTK recursions for a single input}\label{sec:NTHsummary}\index{neural tangent kernel!mean}
Let's start with the NTK mean recursion. Analogously to all other observables, the \terminate{$1/n$ expansion} induces a series expansion on the NTK mean of the form
\begin{align}\label{eq:NTK-mean-expansion}
\Ti{\NTKM}{\alpha_1\alpha_2}{\ell}=&\NTKM_{\alpha_1\alpha_2}^{\le\{0\ri\}(\ell)}+\frac{1}{n_{\ell-1}}\NTKM_{\alpha_1\alpha_2}^{\le\{1\ri\}(\ell)}+\frac{1}{n_{\ell-1}^2}\NTKM_{\alpha_1\alpha_2}^{\le\{2\ri\}(\ell)}+\o{\frac{1}{n^3}}\, .
\end{align}
Just as we defined the \terminate{kernel} $\Ti{K}{\alpha_1\alpha_2}{\ell}$ as the infinite-width limit of the mean metric $\Ti{G}{\alpha_1\alpha_2}{\ell}$ \eqref{eq:definition-of-kernel-first}, let us give the leading $\o{1}$ piece of the NTK mean a special symbol,
\be\label{eq:frozen-NTK}
\Ti{\NTKI}{\alpha_1\alpha_2}{\ell} \equiv \NTKM_{\alpha_1\alpha_2}^{\le\{0\ri\}(\ell)}\, ,
\ee
and a special name: the \textbf{frozen NTK}\index{frozen NTK|textbf}\index{frozen NTK!infinite-width limit of the NTK}\index{neural tangent kernel!frozen|see{frozen NTK}}. The frozen NTK controls the \terminate{training} dynamics in the \terminate{infinite-width limit}, %
which we will investigate in detail next chapter.\footnote{
    Typically in the literature, the \neo{neural tangent kernel} or NTK refers to this deterministic infinite-width NTK mean $\Ti{\NTKI}{\alpha_1\alpha_2}{\ell}$. Since we are principally concerned with understanding finite-width networks, we instead chose to define and refer to the stochastic object $\Tia{\NTK}{i_1i_2}{\alpha_1\alpha_2}{\ell}$ as the NTK. As a concession to the literature, we've used the customary symbol for the NTK, $\NTKI$, to represent the frozen NTK. (As a helpful mnemonic, note that there is an $\NTKM$ frozen inside the $\NTKI$.) Unfortunately, you'll have to wait until~\S\ref{ch:features} to understand the reason why we call the infinite-width NTK \emph{frozen}; there we'll see how finite-width effects \emph{defrost}\index{defrosted NTK|see{neural tangent kernel}}\index{neural tangent kernel!defrosted} the \terminate{training} process and make the NTK move. Here, in this chapter, you can at least see how it gets \emph{agitated} by finite-width fluctuations.
\index{neural tangent kernel!name}\index{neural tangent kernel!agitated}
}

Now, taking the leading piece of the NTK mean recursion~\eqref{eq:NTHmean_final}, we get a recursion solely for the frozen NTK\index{frozen NTK}
\be\label{eq:frozen-NTK-recursion}
\Ti{\NTKI}{\alpha_1\alpha_2}{\ell+1}=\Lb{\ell+1} + \lamW{\ell+1}\bra\sigma_{\alpha_1}\sigma_{\alpha_2}\ket_{K^{(\ell)}}+C_W\bra\sigma^{\prime}_{\alpha_1}\sigma^{\prime}_{\alpha_2}\ket_{K^{(\ell)}}\Ti{\NTKI}{\alpha_1\alpha_2}{\ell}\, .
\ee
Concurrently, as we are neglecting subleading contributions, we have exchanged the Gaussian expectations\index{Gaussian expectation} over the mean metric $G^{(\ell)}$ for ones over the %
\terminate{kernel} $K^{(\ell)}$. 
Finally specializing to a single input, we simply drop the \terminate{sample indices} to get this recursion's final form,
\be\label{eq:frozen-NTK-recursion-single}
\Ti{\NTKI}{}{\ell+1}=\Lb{\ell+1} + \lamW{\ell+1}g\!\le(K^{(\ell)}\ri)+\chi_{\perp}\!\le(K^{(\ell)}\ri)\Ti{\NTKI}{}{\ell}\, ,
\ee
with the initial condition coming directly from our first-layer NTK analysis \eqref{eq:NTHinitial}
\be\label{eq:frozen-ntk-intial}
\Ti{\NTKI}{}{1}=\Lb{1} + \lamW{1} \le(\frac{1}{n_0}\sum_{j=1}^{n_{0}}x_j^2\ri)\, .
\ee
Note that here we have also made use of a helper function and susceptibility\index{perpendicular susceptibility} from \S\ref{ch:eft-mlp}.   

For your convenience, let us also recall and reprint the full set of helper functions --~\eqref{eq:helper_first} and~\eqref{eq:h-function} --  and susceptibilities --~\eqref{eq:chi-parallel}\index{parallel susceptibility} and~\eqref{eq:chi-perp} -- that we first made popular 
in~\S\ref{ch:eft-mlp}: 
\begin{align}
g(K)&=\le\langle \sigma(z)\, \sigma(z)\ri\rangle_{K}\, ,\label{eq:g-function-reprint}\\
h(K)&\equiv \frac{C_W}{4K^2}\le\langle \sigma'(z)\, \sigma'(z)\le(z^2-K\ri)\ri\rangle_{K}=\frac{1}{2}\frac{\td }{\td K}\chi_{\perp}(\ker)\, ,\label{eq:h-function-reprint}\\
\chi_{\parallel}(\ker)&=C_W g'(K)=\frac{C_W}{2\ker^2} \bra \sigma(z)\, \sigma(z)\le(z^2-K\ri)\ket_{\ker}=\frac{C_W}{\ker} \bra z\, \sigma'(z)\, \sigma(z)\ket_{\ker}\, ,\label{eq:chi-parallel-reprint}\\
\chi_{\perp}(K)&=C_W \bra\sigma^\prime(z)\, \sigma^\prime(z) \ket_{\ker}\, .\label{eq:chi-perp-reprint}
\end{align}
As a reminder, to go between the middle and right-hand expression in \eqref{eq:chi-parallel-reprint} you should integrate by parts.\index{integration by parts}

\index{parallel susceptibility}\index{perpendicular susceptibility}
For the remaining recursions, we're going to fast-forward the process as the procedure for converting the multi-input recursions to leading-order single-input recursions surely requires your attention but is somewhat mindless:
\emph{(i)} drop the layer dependence of the \terminate{initialization hyperparameters} as \eqref{eq:ntk-chapter-initialziation-hyperparameters-dropped-layer-dependence}
and uniformize the layer widths as \eqref{eq:ntk-chapter-layer-widths-equalized};
\emph{(ii)} drop \terminate{sample indices} everywhere;
\emph{(iii)} replace the mean metric $G^{(\ell)}$ and the NTK mean $\NTKM^{(\ell)}$\index{neural tangent kernel!mean} with the kernel $K^{(\ell)}$\index{kernel} and the frozen NTK $\NTKI^{(\ell)}$\index{frozen NTK}, respectively;\footnote{Picking nits, we should really make $1/n$ expansions -- similar to~\eqref{eq:definition-of-kernel-first} for the mean metric $G^{(\ell)}$ and~\eqref{eq:NTK-mean-expansion} for the NTK mean $H^{(\ell)}$ -- for the finite-width tensors $A^{(\ell)}$, $B^{(\ell)}$, $D^{(\ell)}$, $F^{(\ell)}$, and also properly make use of the one that we made for $V^{(\ell)}$~\eqref{eq:vertex-decomposition}, denoting the leading-order pieces as $A^{\le\{0\ri\}(\ell)}$ and such, and dropping the subleading pieces. For the interest of notational sanity we won't impose this on you, though our recursions for these tensors should all be understood as referring to these leading-order pieces. (The kernel and the frozen NTK are special in that these infinite-width objects have already been well-studied by the community, and so in this case it's important to differentiate between the finite-width object and the infinite-width piece.)\index{$1/n$ expansion}}
and \emph{(iv)} substitute in for helper functions and susceptibilities~\eqref{eq:g-function-reprint}--\eqref{eq:chi-perp-reprint}.
In particular, this last step has the benefit of letting us recycle our results from \S\ref{ch:eft-mlp} on the deep asymptotic behavior of these functions.

It will also be necessary to recall the single-input leading-order expression for the auxiliary stochastic variable~\eqref{eq:def-omega-without-neural},
\be\label{eq:def-omega-without-neural-single}
\Oi{}{\ell+1} \equiv \lamW{\ell+1} \sigma(z)\sigma(z) +C_W\,\Ti{\NTKI}{}{\ell} \, \sigma'(z)\sigma'(z)  \, ,
\ee
which appears in the recursions for $D^{(\ell)}$~\eqref{eq:D-recursion} and $A^{(\ell)}$~\eqref{eq:A-recursion}; we'll make this substitution the penultimate step \emph{(iii-b)}, if you will. In making these substitutions, please keep in mind that 
the frozen NTK $\Ti{\NTKI}{}{\ell}$ multiplying the second term is not a random variable and hence can be escorted out of any Gaussian expectations\index{Gaussian expectation}.

At this point, you should grab another roll of parchment,
jot down expressions~\eqref{eq:g-function-reprint}--\eqref{eq:def-omega-without-neural-single},
flip back a few pages to locate recursions~\eqref{eq:D-recursion},~\eqref{eq:F-recursion},~\eqref{eq:B-recursion}, and~\eqref{eq:A-recursion},  for $D^{(\ell)}$, $F^{(\ell)}$, $B^{(\ell)}$, and $A^{(\ell)}$, respectively  (or perhaps you kiddos can simply click the equation references in your \terminate{eBook} and copy over the equations to your \terminate{tablet}), and simplify them %
according to the four-(though-sometimes-secretly-five-)step process \emph{(i)}--\emph{(iv)} above. When you're finished, make sure you agree with us:
\begin{align}
\label{eq:D-recursion-single}
\NTHD{}{\ell+1}=&\ \chi_{\perp}^{(\ell)}\chi_{\parallel}^{(\ell)} D^{(\ell)}+\le( \frac{\lamW{\ell+1}}{C_W}\ri)\!  \le[C_W^2\bra \sigma(z)\sigma(z)\sigma(z)\sigma(z) \ket_{\Ti{\ker}{}{\ell}}- \le(C_W g^{(\ell)}\ri)^2+\le(\chi_{\parallel}^{(\ell)}\ri)^2\Ti{\FPV}{}{\ell}\ri]\,  \notag \\
&+\Ti{\NTKI}{}{\ell}\le[C_W^2\bra \sigma(z)\sigma(z) \sigma'(z)\sigma'(z)\ket_{K^{(\ell)}}-C_W g^{(\ell)}\chi_{\perp}^{(\ell)}+2h^{(\ell)}\chi_{\parallel}^{(\ell)} \, V^{(\ell)} \ri]    \, , \\
\NTHF{}{\ell+1}=&\le(\chi_{\parallel}^{(\ell)}\ri)^2\NTHF{}{\ell}+C_W^2\bra\sigma(z)\sigma(z) \sigma'(z)\sigma'(z)\ket_{K^{(\ell)}}\Ti{\NTKI}{}{\ell}\, ,\label{eq:F-recursion-single} \\
\NTHB{}{\ell+1}=&\le(\chi_{\perp}^{(\ell)}\ri)^2 \NTHB{}{\ell}  + C_W^2 \bra  \sigma'(z)\sigma'(z)\sigma'(z)\sigma'(z)\ket_{K^{(\ell)}}\le(\Ti{\NTKI}{}{\ell}\ri)^2  \, ,\label{eq:B-recursion-single} \\
A^{(\ell+1)}=& \le(\chi_{\perp}^{(\ell)}\ri)^2 A^{(\ell)}  + \le(\frac{\LW{\ell+1}}{C_W}\ri)^2 \!\le[C_W^2\bra \sigma(z)\sigma(z)\sigma(z)\sigma(z) \ket_{\Ti{\ker}{}{\ell}}- \le(C_W g^{(\ell)}\ri)^2+\le(\chi_{\parallel}^{(\ell)}\ri)^2\Ti{\FPV}{}{\ell}\ri]
\, \nonumber\\
&+2\le(\frac{\LW{\ell+1}}{C_W}\ri)\Ti{\NTKI}{}{\ell}\le[C_W^2\bra \sigma(z)\sigma(z) \sigma'(z)\sigma'(z)\ket_{K^{(\ell)}}-C_W g^{(\ell)}\chi_{\perp}^{(\ell)}+2h^{(\ell)}\chi_{\parallel}^{(\ell)} \, V^{(\ell)} \ri]\, \nonumber\\
&+2\le(\frac{\LW{\ell+1}}{C_W}\ri)\chi_{\perp}^{(\ell)}\chi_{\parallel}^{(\ell)}D^{(\ell)}+4h^{(\ell)}\chi_{\perp}^{(\ell)}\Ti{\NTKI}{}{\ell}D^{(\ell)}\, \nonumber\\
&+\le(\Ti{\NTKI}{}{\ell}\ri)^2 \le[C_W^2\bra \sigma'(z)\sigma'(z)\sigma'(z)\sigma'(z) \ket_{K^{(\ell)}}-\le(\chi_{\perp}^{(\ell)}\ri)^2+\le(2h^{(\ell)}\ri)^2V^{(\ell)}\ri]\, .\label{eq:A-recursion-single}
\end{align}
For these recursions, the initial conditions (recalling that the first-layer NTK is fully deterministic) all vanish identically as
\be\label{eq:ntk-fluctuations-initial-conditions}
A^{(1)}=B^{(1)}=D^{(1)}=F^{(1)}=0\, .
\ee
Here also, for helper functions and susceptibilities, we used the following simplifying notation 
\be\index{parallel susceptibility}\index{perpendicular susceptibility}\label{eq:notational-squashing}
g^{(\ell)} \equiv g\!\le(K^{(\ell)}\ri) \, , \qquad h^{(\ell)} \equiv h\!\le(K^{(\ell)}\ri) \, , \qquad \Ti{\chi}{\parallel}{\ell} \equiv \chi_\parallel\!\le(K^{(\ell)}\ri) \, , \qquad \Ti{\chi}{\perp}{\ell} \equiv \chi_\perp\!\le(K^{(\ell)}\ri)  \, ,
\ee
making the kernel dependence implicit.

We can further simplify \eqref{eq:D-recursion-single} and~\eqref{eq:A-recursion-single} by recalling the single-input recursion for the \terminate{four-point vertex} \eqref{eq:finite-width-reprinted-vertex}
\begin{align}
\Ti{\FPV}{}{\ell+1}&=\le(\chi_{\parallel}^{(\ell)}\ri)^2\Ti{\FPV}{}{\ell}+C_W^2\le[\bra \sigma(z)\sigma(z)\sigma(z)\sigma(z) \ket_{\Ti{\ker}{}{\ell}}-\le(g^{(\ell)}\ri)^2\ri]\, .
\end{align}
Keep staring at these equations, and you'll see slightly more compact expressions emerge
\begin{align}
\label{eq:D-recursion-single-compacter}
\NTHD{}{\ell+1}=&\ \chi_{\perp}^{(\ell)}\chi_{\parallel}^{(\ell)} D^{(\ell)}+\le( \frac{\lamW{\ell+1}}{C_W}\ri)\!  V^{(\ell+1)}\,  \\
&+\Ti{\NTKI}{}{\ell}\le[C_W^2\bra \sigma(z)\sigma(z) \sigma'(z)\sigma'(z)\ket_{K^{(\ell)}}-C_W g^{(\ell)}\chi_{\perp}^{(\ell)}+2h^{(\ell)}\chi_{\parallel}^{(\ell)} \, V^{(\ell)} \ri] \notag   \, , \\
A^{(\ell+1)}=&\ \le(\chi_{\perp}^{(\ell)}\ri)^2 A^{(\ell)}  - \le(\frac{\LW{\ell+1}}{C_W}\ri)^2 \!V^{(\ell+1)}+2\le(\frac{\LW{\ell+1}}{C_W}\ri)\NTHD{}{\ell+1}+4h^{(\ell)}\chi_{\perp}^{(\ell)}\Ti{\NTKI}{}{\ell}D^{(\ell)} \, \nonumber\\
&+\le(\Ti{\NTKI}{}{\ell}\ri)^2 \le[C_W^2\bra \sigma'(z)\sigma'(z)\sigma'(z)\sigma'(z) \ket_{K^{(\ell)}}-\le(\chi_{\perp}^{(\ell)}\ri)^2+\le(2h^{(\ell)}\ri)^2V^{(\ell)}\ri]\, ,\label{eq:A-recursion-single-compacter}
\end{align}
which you may find makes things simpler when solving these recursions. However, please use these formulae with caution as both $\ell$-th-layer and $(\ell+1)$-th-layer objects appear on their right-hand sides. %

\subsubsection{The relevance of scaling laws}\index{relevant (RG flow)}\index{scaling law}
For the rest of this chapter, we will work through solving the five leading-order single-input NTK recursions~\eqref{eq:frozen-NTK-recursion-single} and~\eqref{eq:D-recursion-single}--\eqref{eq:A-recursion-single}. (Remember that we already solved single-input recursions for the \terminate{kernel} and \terminate{four-point vertex} way back in~\S\ref{ch:eft-mlp}.) In solving these recursions, we will find that each observable 
obeys our \neo{scaling ansatz} \eqref{eq:master-scaling-ansatz}:
\begin{align}\label{eq:master-scaling-ansatz-reprint}
\Ti{\O}{}{\ell} &= \le( \frac{1}{\ell} \ri)^{p_\O} \le[c_{0,0}+  c_{1,1} \le( \frac{\log \ell}{\ell} \ri) + c_{1,0}\le( \frac{ 1}{\ell}\ri) +  c_{2,2} \le(  \frac{\log^2 \ell}{\ell^2} \ri)+  \dots \ri]%
\notag \\
&= \le( \frac{1}{\ell} \ri)^{p_\O} \le[\sum_{s = 0}^\infty \sum_{q=0}^{s} c_{s,q} \le( \frac{\log^q \ell}{\ell^s}  \ri)\ri].
\end{align}
Recall that $p_\O$ is a \neo{critical exponent}, which is universal for a given universality class of activation functions, while the constants $c_{s,q}$ depend on some of the details of the particular activation function under consideration.

\index{neural tangent kernel!variance}\index{frozen NTK}
To properly understand the physics of these observables, recall from \S\ref{sec:signal_prop_finite_width} that we need to consider dimensionless quantities. For the two tensors controlling the NTK variance, we should normalize by the square of the frozen NTK\index{frozen NTK}
\be\label{eq:ntk-variance-scaling-ansatz}
\frac{\Ti{A}{}{\ell}}{n\le( \Ti{\NTKI}{}{\ell} \ri)^2} \sim \frac{1}{n} \le( \frac{1}{\ell} \ri)^{p_A - 2 p_\Theta}+ \dots \, , \qquad \frac{\Ti{B}{}{\ell}}{n\le( \Ti{\NTKI}{}{\ell} \ri)^2} \sim \frac{1}{n} \le( \frac{1}{\ell} \ri)^{p_B - 2 p_\Theta}+ \dots \, ,
\ee
while for the NTK-preactivation cross correlation\index{cross correlation!NTK-preactivation}, we should instead normalize by one factor of the frozen NTK\index{frozen NTK} and one factor of the \terminate{kernel}
\be\label{eq:ntk-cross-corr-scaling-ansatz}
\frac{\Ti{D}{}{\ell}}{ n\Ti{K}{}{\ell}  \Ti{\NTKI}{}{\ell}  } \sim \frac{1}{n} \le( \frac{1}{\ell} \ri)^{p_D - p_\Theta - p_0}+ \dots \, , \qquad \frac{\Ti{F}{}{\ell}}{ n\Ti{K}{}{\ell}  \Ti{\NTKI}{}{\ell}  } \sim \frac{1}{n} \le( \frac{1}{\ell} \ri)^{p_F - p_\Theta - p_0}+ \dots \,,
\ee
where $p_0$ was the \terminate{critical exponent} for the single-input kernel $K^{(\ell)}$.  

By looking at these dimensionless quantities, we'll find \emph{scaling laws}\index{scaling law} that transcend even beyond universality classes. As a particular example, recall that the normalized \terminate{four-point vertex} \eqref{eq:k-star-equals-zero-normalized-four-point-scaling-law},
\be\label{eq:k-star-equals-zero-normalized-four-point-scaling-law-reprint}
\frac{\Ti{\FPV}{}{\ell}}{n \le(\Ti{ \ker}{}{\ell} \ri)^2} \sim  \frac{1}{n}\le(\frac{1}{\ell} \ri)^{p_V-2p_0} + \dots \, ,%
\ee
gave rise to a scaling law~\eqref{eq:vertex-scaling-law}
\be\label{eq:vertex-scaling-law-reprint}
p_V-2p_0 = -1\, ,
\ee
for both scale-invariant and $K^\star=0$ activation functions.\index{universality class!scale-invariant}\index{universality class!K@$K^\star=0$}
This \terminate{scaling law} let us interpret the ratio $\ell/n$ as an \neo{emergent scale} controlling the leading finite-width behavior of the preactivation distribution. \emph{Spoiler alert:}\index{spoiler alert} in much the same way, we'll find scaling laws
\be\label{eq:NTK-scaling-laws}
p_A - 2 p_\Theta=-1\, , \quad p_B- 2 p_\Theta=-1\, , \quad p_D - p_\Theta - p_0=-1\,, \quad p_F - p_\Theta - p_0=-1\, ,
\ee
that also hold for both the scale-invariant and $K^\star=0$ universality classes.\index{universality class!scale-invariant}\index{universality class!K@$K^\star=0$}
Thus, we'll be able to conclude that all the leading finite-width effects of the NTK-preactivation joint distribution are \emph{relevant}\index{relevant (RG flow)} and controlled by the same $\ell/n$ perturbative cutoff\index{cutoff, effective theory}. This means that we can effectively describe the training of realistic deep networks of finite width and nonzero $L/n$.

\subsubsection{Formalities: perpendicular perturbations and the frozen NTK}
\index{frozen NTK}
Before explicitly analyzing universality classes, let us note that the frozen-NTK recursion~\eqref{eq:frozen-NTK-recursion-single} admits a formal solution
\be\label{eq:ntk-mean-k-star-formal}
\NTKI^{(\ell)}=\sum_{\ell'=1}^{\ell}\le\{\le[\Lb{\ell'}+\LW{\ell'}g^{(\ell'-1)}\ri]\le[\prod_{\ell''=\ell'}^{\ell-1}\Ti{\chi}{\perp}{\ell''}\ri] \ri\} \, .
\ee
In words, we see that the solution involves a sum over all the previous layers $1, \dots, \ell$, and that each term in the sum involves an additive contribution $\Lb{\ell'}+\LW{\ell'}g^{(\ell'-1)}$. Such a contribution then gets recursively multiplied by perpendicular susceptibilities up to the $(\ell-1)$-th layer, resulting in an overall multiplicative factor $\prod_{\ell''=\ell'}^{\ell-1}\Ti{\chi}{\perp}{\ell''}$.
To avoid the exponential behavior that's generic with such a factor, we must set $\chi_{\perp}=1$.

It is enlightening to tie this insight to the discussion we had in~\S\ref{sec:bootstrapping} where we performed our general \terminate{criticality} analysis of the kernel recursion. There, we first looked at the single-input kernel and set $\chi_{\parallel}=1$ to avoid exponential behavior in the network outputs. Then, we looked at the two-input kernel and analyzed how the off-diagonal perpendicular perturbations $\Ti{\delta\delta\ker}{[2]}{\ell}$ flow. Turning off the odd perturbations $\Ti{\delta\ker}{[1]}{\ell}$, a brief inspection of the perpendicular recursion~\eqref{K2}
\be\label{eq:ddK-reprint-ntk-eft}
\Ti{\delta\delta \ker}{[2]}{\ell+1}=\chi_{\perp}^{(\ell)}\Ti{\delta\delta \ker}{[2]}{\ell}\, ,
\ee
necessitated the \terminate{criticality} condition $\chi_{\perp}=1$ so as to preserve the difference between nearby inputs as they propagate through the network. 
At the time, we presumed that such a condition would be useful for comparing nearby inputs when learning from data.
Indeed, the same multiplicative factor that appeared in the formal solution for the frozen NTK \eqref{eq:ntk-mean-k-star-formal},
\be\label{eq:chi-perp-factor-solution-ntk}
\prod_{\ell''=\ell'}^{\ell-1}\Ti{\chi}{\perp}{\ell''}=\frac{\Ti{\delta\delta \ker}{[2]}{\ell}}{\Ti{\delta\delta \ker}{[2]}{\ell'}}\, ,
\ee
also appears in a formal solution for $\Ti{\delta\delta \ker}{[2]}{\ell}$. Thus, with both formal solutions \eqref{eq:ntk-mean-k-star-formal} and \eqref{eq:chi-perp-factor-solution-ntk}, we have formalized the connection between preserving $\Ti{\delta\delta \ker}{[2]}{\ell}$ data and
learning from data. %

With the formalities out of the way, let's now analyze our two eminent universality classes, the scale-invariant universality class\index{universality class!scale-invariant} and the $\ker^\star=0$ universality class\index{universality class!K@$K^\star=0$}.

\section{Scale-Invariant Universality Class}\label{sec:ntk_criticality_scale_invariant}
As a reminder, the canonical members of the scale-invariant universality class\index{universality class!scale-invariant} are the $\relu$ and $\linear$ activation functions.
For a general activation function in this universality class,
\be\label{eq:scale-invariant-one-kink-tired}
\sigma(z) = 
    \begin{cases}
   a_+ z \, , & z \ge 0  \, , \\
    a_- z \, , & z < 0 \, ,
    \end{cases}
\ee
recall from \S\ref{sec:scale-invariant-eft} that the helper functions and susceptibilities\index{parallel susceptibility}\index{perpendicular susceptibility} evaluate to
\begin{align}\label{eq:helper-ntk-scale-invariant-g}
g^{(\ell)}&=A_2 K^{(\ell)}\, ,\\
\label{eq:helper-ntk-scale-invariant-h}
h^{(\ell)}&=0\, ,\\
\chi_{\parallel}^{(\ell)}&= \chi\, , \\
\chi_{\perp}^{(\ell)}&= \chi\, ,
\end{align}
with 
$\chi \equiv C_W A_2$. %
By substituting in \eqref{eq:scale-invariant-one-kink-tired} and performing the integrals,
we can just as easily evaluate the three other Gaussian expectations\index{Gaussian expectation} that we'll need
\begin{align}
\label{eq:gaussian-expectation-ntk-scale-invariant-1}
\bra \sigma(z)\sigma(z)\sigma(z)\sigma(z) \ket_{K^{(\ell)}}&=3 A_4 \le(K^{(\ell)}\ri)^2\, ,\\
\label{eq:gaussian-expectation-ntk-scale-invariant-2}
\bra \sigma(z)\sigma(z)\sigma'(z)\sigma'(z) \ket_{K^{(\ell)}}&=A_4 K^{(\ell)}\, ,\\
\label{eq:gaussian-expectation-ntk-scale-invariant-3}
\bra \sigma'(z)\sigma'(z)\sigma'(z)\sigma'(z) \ket_{K^{(\ell)}}&=A_4\, .
\end{align}
Here and right
before, we've also made use of our previous definitions for the constants that naturally arise from these integrations:
\be
A_2\equiv \frac{a_+^2+a_-^2}{2}\, , \qquad A_4\equiv \frac{a_+^4+a_-^4}{2}\, .
\ee

With these recollections, we are reminded of one of this class's principal characteristics: both susceptibilities are independent of the kernel and constant for all layers.
With that in mind, we were able to easily satisfy
\terminate{criticality} for the scale-invariant universality class\index{universality class!scale-invariant} 
by setting the \terminate{initialization hyperparameters} to 
\be\label{eq:scale-invariant-criticality-ntk-reprint}
C_b=0\, ,\qquad  C_W =\frac{1}{A_2} \, .
\ee
With these tunings, both susceptibilities are set to unity $\chi=1$ and the fixed-point value of the kernel is given in terms of the input by the expression \eqref{eq:chapter-5-scale-invariant-kernel-fixed}
\be
\Tif{\ker}{} \equiv \frac{1}{A_2} \le(\frac{1}{n_0}\sum_{j=1}^{n_0}x_j^2\ri)\, ,
\ee
and the \terminate{four-point vertex} at \terminate{criticality} is given by \eqref{eq:scale-invariant-vertex-solution}
\be\label{eq:scale-invariant-vertex-solution-reprint}
\Ti{\FPV}{}{\ell}=\le(\ell-1\ri)\le(\frac{3A_4}{A_2^2}-1\ri) \le(\Tif{\ker}{}\ri)^2 \, .
\ee

\subsubsection{NTK mean (frozen NTK)}\index{neural tangent kernel!mean}\index{frozen NTK}\index{learning rate}
With the above expressions in mind, at \terminate{criticality} the recursion~\eqref{eq:frozen-NTK-recursion-single} for the single-input frozen NTK simplifies to
\be\label{eq:frozen-NTK-recursion-single-scale-invariant}
\Ti{\NTKI}{}{\ell+1}=\Ti{\NTKI}{}{\ell}+\Lb{\ell+1} + \lamW{\ell+1} A_2  \Tif{\ker}{}\, ,
\ee
This recursion, together with the initial condition~\eqref{eq:frozen-ntk-intial}, is easy to solve for a given set of bias and weight learning rates. 

For instance, assuming layer-independent learning rates $\Lb{\ell}=\lambda_b$ and $\lamW{\ell}=\lambda_W$,
we find
\be\label{eq:frozen-ntk-critical-solution-relu}
\Ti{\NTKI}{}{\ell}=\le(\lambda_b+\lambda_W A_2 \Tif{\ker}{}\ri)\ell\, .
\ee
With these uniform learning rates, we see that the frozen NTK\index{frozen NTK} for the scale-invariant universality class\index{universality class!scale-invariant} grows linearly with depth. Since the NTK involves a sum over all the previous layers \eqref{eq:ntk-mean-k-star-formal}, linear growth implies that
these contributions are uniform across the layers;
this is in contrast to the non-critical cases, for which we would have had exponentially different contributions from the different layers of a deep network, as is clear from the formal solution~\eqref{eq:ntk-mean-k-star-formal}. 
Finally, a comparison with the ansatz \eqref{eq:master-scaling-ansatz-reprint} implies that the \terminate{critical exponent} for the frozen NTK\index{frozen NTK} is given by $p_\Theta  = -1$. We will interpret all these points further in \S\ref{sec:EVGP-WEP}.

\subsubsection{NTK variance and NTK-preactivation cross correlation (agitated NTK)}\index{neural tangent kernel!agitated}

Now, let's evaluate our finite-width recursions \eqref{eq:D-recursion-single}--\eqref{eq:A-recursion-single} to find the NTK variance and the NTK-preactivation cross correlations.\footnote{
    You could also choose to evaluate \eqref{eq:D-recursion-single-compacter} and then \eqref{eq:A-recursion-single-compacter} for $D^{(\ell)}$ and $A^{(\ell)}$, respectively; it's about the same level of difficulty and obviously yields the same solution either way.
} First, we can simplify them by substituting in for the helper functions $g^{(\ell)}=A_2 K^{(\ell)}$ \eqref{eq:helper-ntk-scale-invariant-g} and $h^{(\ell)}=0$ \eqref{eq:helper-ntk-scale-invariant-h} as well as making use of our formulae for the three other Gaussian expectations\index{Gaussian expectation}~\eqref{eq:gaussian-expectation-ntk-scale-invariant-1}--\eqref{eq:gaussian-expectation-ntk-scale-invariant-3} involving $A_4$. Then, let us tune the \terminate{initialization hyperparameters} to \terminate{criticality}~\eqref{eq:scale-invariant-criticality-ntk-reprint} by picking $C_W=1/A_2$, which sets both susceptibilities to unity $\Ti{\chi}{\parallel}{\ell}=\Ti{\chi}{\perp}{\ell}=1$ and makes the kernel fixed $K^{(\ell)}=\Tif{\ker}{}$. With these manipulations, we get
\begin{align}
\NTHD{}{\ell+1}=&D^{(\ell)}+\lambda_W A_2  \le[\le(\frac{3A_4}{A_2^2}-1\ri)\le(\Tif{\ker}{}\ri)^2+V^{(\ell)}\ri]+\le(\frac{A_4}{A_2^2}-1\ri)\Tif{\ker}{}\Ti{\NTKI}{}{\ell}\, ,\\
\NTHF{}{\ell+1}=&\NTHF{}{\ell}+\frac{A_4}{A_2^2}\Tif{\ker}{}\Ti{\NTKI}{}{\ell}\, ,\\
\NTHB{}{\ell+1}=& \NTHB{}{\ell}+\frac{A_4}{A_2^2}\le(\Ti{\NTKI}{}{\ell}\ri)^2 \, ,\\
A^{(\ell+1)}=& A^{(\ell)}+\le(\lambda_W A_2\ri)^2 \le[\le(\frac{3A_4}{A_2^2}-1\ri)\le(\Tif{\ker}{}\ri)^2+V^{(\ell)}\ri]\, \\
&+2\lambda_W A_2\le(\frac{A_4}{A_2^2}-1\ri)\Tif{\ker}{}\Ti{\NTKI}{}{\ell}+2\lambda_W A_2 D^{(\ell)}+\le(\frac{A_4}{A_2^2}-1\ri)\le(\Ti{\NTKI}{}{\ell}\ri)^2  \, .\nonumber
\end{align}
Note that we have also assumed layer-independent learning rates as we did just before when working out the NTK mean.

Next, substituting in our solutions for $\Ti{\FPV}{}{\ell}$ \eqref{eq:scale-invariant-vertex-solution-reprint} and $\Ti{\NTKI}{}{\ell}$ \eqref{eq:frozen-ntk-critical-solution-relu},
we can easily solve the recursions for $\NTHD{}{\ell}$, $\NTHF{}{\ell}$, and $\NTHB{}{\ell}$.
Then, with our solution for $\NTHD{}{\ell}$ in hand, we can also solve the recursion for $A^{(\ell)}$. All together, this gives the following solutions
\begin{align}
\Ti{D}{}{\ell}&=\frac{\ell(\ell-1)}{2}\le[\lambda_b \le(\frac{A_4}{A_2^2}-1\ri)\Tif{\ker}{}+\lambda_W A_2 \le(\frac{4A_4}{A_2^2}-2\ri)\le(\Tif{\ker}{}\ri)^2 \ri] \, ,\\
\label{eq:F-solution}
\Ti{F}{}{\ell}&=\frac{\ell(\ell-1)}{2}\le[\frac{A_4}{A_2^2} \le(\lambda_b+\lambda_W A_2 \Tif{\ker}{}\ri)\Tif{\ker}{}\ri] \, ,\\
\Ti{B}{}{\ell}&=\frac{\ell(\ell-1)(2\ell-1)}{6}\le(\frac{A_4}{A_2^2}\ri)\le(\lambda_b+\lambda_W A_2 \Tif{\ker}{}\ri)^2\, ,\\
\Ti{A}{}{\ell}&=\frac{\ell^3}{3}\le[
    \le(\frac{A_4}{A_2^2}-1\ri) \lambda_b^2 + 3\le(\frac{A_4}{A_2^2}-1\ri) \lambda_b \lambda_W A_2 \Tif{\ker}{} + \le(5 A_4 - 3 A_2^2 \ri) \lambda_W^2 \le(\Tif{\ker}{}\ri)^2
\ri] + \ldots \, ,
\end{align}
where for $A^{(\ell)}$ we kept only the leading large-$\ell$ contribution.

From these four solutions, we can read off another four critical exponents\index{critical exponent} for the scale-invariant universality class,\index{universality class!scale-invariant}
\be\label{eq:NTK-agited-exponents-scale-invariant}
p_D = -2\, , \qquad p_F = -2\, , \qquad p_B = -3\, , \qquad p_A = -3\, ,
\ee
which corresponds to the quadratic growth of $\Ti{D}{}{\ell}$ and $\Ti{F}{}{\ell}$ and the cubic growth of $\Ti{B}{}{\ell}$ and $\Ti{A}{}{\ell}$. 
Combined with $p_0=0$ for the kernel and $p_{\Theta}=-1$ for the frozen NTK, we obtain the advertised $\ell/n$-scaling~\eqref{eq:NTK-scaling-laws} of the appropriately           normalized quantities 
\eqref{eq:ntk-variance-scaling-ansatz} and \eqref{eq:ntk-cross-corr-scaling-ansatz}.

\section{\texorpdfstring{$\Tif{\ker}{}=0$}{K*=0} Universality Class}\label{sec:ntk_criticality_tanh_univ}\index{universality class!K@$K^\star=0$}
As a reminder, two notable members of this class are $\tanhA$ and $\sinA$. More generally, the $\ker^\star=0$ universality class contains any activation function with a corresponding kernel that has a nontrivial fixed point at $\Tif{\ker}{}=0$. 

Specifically, recall from~\S\ref{subsec:tanh_univ} that we used the following notation for the Taylor coefficients of an activation function:
\be\label{eq:taylor-expansion-k-star-reprint}
\sigma(z)=\sum_{p=0}^{\infty}\frac{\sigma_{p}}{p!}z^p\,  .
\ee
Then, from an analysis of the single-input kernel recursion we learned that there's nontrivial fixed point at $\Tif{\ker}{}=0$ if and only if the activation function vanishes at the origin with nonzero slope~\eqref{eq:tanh_cri_condition}
\be\label{eq:tanh_cri_condition-reprint-ntk}
\sigma_0=0\,  , \qquad \sigma_1\ne0\, ,
\ee
for which we can satisfy the \terminate{criticality} conditions by tuning the \terminate{initialization hyperparameters} as \eqref{eq:k-star-equals-zero-critical-initialization} 
\be\label{eq:k-star-equals-zero-critical-initialization-reprint-ntk}
C_b=0\,  , \qquad C_W=\frac{1}{\sigma_1^2}\, .
\ee
Going forward, we will assume that the bias variance and rescaled weight variance have been tuned to \terminate{criticality} as \eqref{eq:k-star-equals-zero-critical-initialization-reprint-ntk}. 

\index{parallel susceptibility}\index{perpendicular susceptibility}
Unlike the scale-invariant universality class\index{universality class!scale-invariant}, the \terminate{criticality} analysis for the $K^\star=0$ universality class\index{universality class!K@$K^\star=0$} was perturbative around $K=0$.
For this analysis, we expanded the helper function $g(K)$ and both susceptibilities as \eqref{eq:g0}--\eqref{eq:chi-perp-expansion-K-star-equals-zero}, which 
-- now with \eqref{eq:tanh_cri_condition-reprint-ntk} and \eqref{eq:k-star-equals-zero-critical-initialization-reprint-ntk} in mind --
evaluate to
\begin{align}\label{eq:g0-reprint-ntk}
g(\ker)&=\sigma_1^2\le[ K+a_1 K^2+\o{K^3}\ri]\, , \\
\label{eq:chi-para-reprint-ntk-kstar}
\chi_{\parallel}(\ker)&=1+2 a_1K+\o{K^2}\, ,\\
\label{eq:chi-perp-reprint-ntk-kstar}
\chi_{\perp}(\ker)&=1+b_1 K+\o{K^2}\, ,
\end{align}
where we've also recalled the following combinations of Taylor coefficients
\begin{align}
a_1&\equiv \le(\frac{\sigma_3}{\sigma_1}\ri)+\frac{3}{4}\le(\frac{\sigma_2}{\sigma_1}\ri)^2\ ,\\
b_1&\equiv \le(\frac{\sigma_3}{\sigma_1}\ri)+\le(\frac{\sigma_2}{\sigma_1}\ri)^2\, .%
\end{align}
As a reminder, to get these expressions we first Taylor expanded their definitions \eqref{eq:g-function-reprint},  \eqref{eq:chi-parallel-reprint}, and \eqref{eq:chi-perp-reprint} in $z$,
and then evaluated each series of Gaussian expectations\index{Gaussian expectation} to the desired order. 
Following the same method, we can evaluate the helper function $h(K)$ \eqref{eq:h-function-reprint} as well as the two other Gaussian expectations\index{Gaussian expectation} needed to solve our NTK recursions:
\begin{align}
\label{eq:k-star-h-ntk}
h\!\le(K\ri)&=\frac{b_1}{2}+\o{K^1}\, ,\\
\label{eq:gaussian-expectation-ntk-k-star-2}
\bra \sigma(z)\sigma(z)\sigma'(z)\sigma'(z) \ket_{K}&=\sigma_1^4 \le[K+\o{K^2} \ri]  \, ,\\
\label{eq:gaussian-expectation-k-star-3}
\bra \sigma'(z)\sigma'(z)\sigma'(z)\sigma'(z) \ket_{K}&=  \sigma_1^4 \le[1+\o{K^1} \ri]\, .
\end{align}

Remembering that the \terminate{parallel susceptibility} characterizes the linear response of the kernel perturbations around the fixed point \eqref{eq:chi_parallel_first}, we note from above that the parallel susceptibility at criticality~\eqref{eq:chi-para-reprint-ntk-kstar} is close to one near the nontrivial fixed point at $\Tif{\ker}{}=0$. Consequently, 
we found a power-law large-$\ell$ asymptotic solution for the single-input kernel \eqref{eq:tanh_asymptotic}
\be\label{eq:tanh_asymptotic-reprint-ntk}
\ker^{(\ell)}=\Tif{\ker}{}+\Ti{\Delta\ker}{}{\ell}=\Ti{\Delta\ker}{}{\ell}=\le[\frac{1}{(-a_1)}\ri]\frac{1}{\ell}+  \o{\frac{\log\ell}{\ell^2}}   \, ,
\ee
which slowly but surely approaches the $\Tif{\ker}{}=0$ nontrivial fixed point, justifying our perturbative approach to the deep asymptotics.
As for the single-input \terminate{four-point vertex}, we previously found \eqref{eq:Kstar-equals-zero-vertex-solution}
\begin{align}\label{eq:Kstar-equals-zero-vertex-solution-reprint-ntk}
\Ti{\FPV}{}{\ell}&=\le[\frac{2}{3a_1^2}\ri]\frac{1}{\ell}+  \o{\frac{\log\ell}{\ell^2}}\, , 
\end{align}
and the appropriately normalized quantity \eqref{eq:k-star-equals-zero-normalized-four-point-scaling-law-reprint} has an $\ell/n$ scaling:
\be\label{eq:k-star-equals-zero-normalized-four-point-tanh-univ-reprint}
\frac{\Ti{\FPV}{}{\ell}}{n \le(\Ti{ \ker}{}{\ell} \ri)^2} = \le(\frac{2}{3}\ri)\frac{\ell}{n} + \o{\frac{\log \le(\ell\ri)}{n}} \, .
\ee

\subsubsection{NTK mean (frozen NTK)}\index{neural tangent kernel!mean}\index{frozen NTK}\index{learning rate}\index{universality class!scale-invariant}\index{universality class!K@$K^\star=0$}
Let's start with our generic formal solution~\eqref{eq:ntk-mean-k-star-formal} to the frozen NTK recursion~\eqref{eq:frozen-NTK-recursion-single}. Note that for $K^\star=0$ activation functions the  multiplicative factor~\eqref{eq:chi-perp-factor-solution-ntk} takes the form
\be\label{eq:chi-perp-factor-solution-ntk-specific}
\prod_{\ell''=\ell'}^{\ell-1}\Ti{\chi}{\perp}{\ell''}=\frac{\Ti{\delta\delta \ker}{[2]}{\ell}}{\Ti{\delta\delta \ker}{[2]}{\ell'}}=\le(\frac{\ell'}{\ell}\ri)^{p_{\perp}}\!+\,\ldots\, ,
\ee
when we plug in our large-$\ell$ asymptotic solution for 
$\Ti{\delta\delta\ker}{[2]}{\ell}$~\eqref{eq:perp-asymptotic-solution}. As a reminder, the \terminate{critical exponent} controlling the falloff was given in terms of the Taylor coefficient combinations,  $p_{\perp}=b_1/a_1$, which evaluates to $1$ for the $\tanhA$ and $\sinA$ activation functions.
Plugging this multiplicative factor back into the formal solution 
\eqref{eq:ntk-mean-k-star-formal} along with our expansion for $g(K)$ \eqref{eq:g0-reprint-ntk} evaluated on the asymptotic kernel~\eqref{eq:tanh_asymptotic-reprint-ntk}, we find
\be\label{eq:ntk-mean-k-star-formal-2}
\NTKI^{(\ell)}=\sum_{\ell'=1}^{\ell}\le\{\le[\Lb{\ell'}+\LW{\ell'}\frac{\sigma_1^2}{(-a_1)} \le(\frac{1}{\ell'}\ri) + \dots  \ri] \le[\le(\frac{\ell'}{\ell}\ri)^{p_{\perp}}\!+\,\ldots\,\ri]\ri\} \, .
\ee
Here, 
the factor in the first square bracket is an additive contribution picked up from the $\ell'$-th layer, while the factor in the second square bracket is a multiplicative contribution from recursively passing from the $\ell'$-th layer to the $\ell$-th layer.

\index{universality class!K@$K^\star=0$}\index{effective theory}\index{practical practitioners}
Effective theorists may take issue with two aspects of this solution 
\eqref{eq:ntk-mean-k-star-formal-2} 
if we naively continue to choose layer-independent learning rates $\Lb{\ell}=\lambda_b$ and $\lamW{\ell}=\lambda_W$.
\bi
\item Firstly, notice in the first square bracket that the $\ell'$-dependence of the bias term differs from the $\ell'$-dependence of the weight term by a factor of $\sim (1/\ell')$. This means that the contribution of the weights to the NTK decreases with depth relative to the contribution of the biases.
\item Secondly, notice in the second square bracket that the $\sim\le(\ell'\ri)^{p_\perp}$ behavior means that the NTK is dominated by contributions from deeper layers for $p_\perp > 0$ in comparison to the shallower layers. Remembering that the NTK controls the dynamics of observables \eqref{eq:obsevable-evolution-layer-ell}, this in turn means that the training dynamics %
will be heavily influenced by the model parameters near the output layer.
\ei
Additionally, practical practitioners may now wonder whether these unnatural depth scalings also contribute to the empirical preference for $\relu$ over $\tanhA$ in the \terminate{deep learning} community. (More on this in \S\ref{sec:EVGP-WEP}.)

\index{learning rate}
Having said all that, we can rectify this
imbalance by scaling out the layer dependence as
\be\label{eq:layer-independent-rates}
\Lb{\ell} \equiv \widetilde{\lambda}_b \le( \frac{1}{\ell}\ri)^{p_\perp} \, , \qquad \LW{\ell} \equiv \widetilde{\lambda}_W \le( \frac{1}{\ell}\ri)^{p_\perp-1} \, ,
\ee
where $\widetilde{\lambda}_b$ and $\widetilde{\lambda}_W$ are layer-independent constants.
Substituting this ansatz into our solution \eqref{eq:ntk-mean-k-star-formal-2}, we find
\be\label{eq:frozen-ntk-k-star-solution}
\Ti{\NTKI}{}{\ell}=\le[\widetilde{\lambda}_b+\frac{\widetilde{\lambda}_W\sigma_1^2}{(-a_1)}\ri]\le( \frac{1}{\ell}\ri)^{p_{\perp}-1}+\ldots\, ,
\ee
which manifestly balances the weight and bias contributions.
Thus, we see for the $K^\star=0$ universality class\index{universality class!K@$K^\star=0$} that the \terminate{critical exponent} for the frozen NTK\index{neural tangent kernel!mean}\index{frozen NTK} is given by
\be\label{eq:frozen-ntk-k-star-critical-exponent}
p_{\NTKI}=p_{\perp}-1\, .
\ee
In particular, for both the $\tanhA$ and $\sinA$ activation functions,
$p_{\NTKI}=0$.

\subsubsection{NTK variance and NTK-preactivation cross correlation (agitated NTK)}
\index{neural tangent kernel!agitated}
Now, let's finally deal with the agitated NTK statistics for the $K^\star=0$ universality class\index{universality class!K@$K^\star=0$}.
To aid our computation at \terminate{criticality}, let us make use of the asymptotic behavior of the kernel~\eqref{eq:tanh_asymptotic-reprint-ntk} and record the leading large-$\ell$ asymptotics of the helper functions~\eqref{eq:g0-reprint-ntk} and~\eqref{eq:k-star-h-ntk}, the susceptibilities~\eqref{eq:chi-para-reprint-ntk-kstar} and~\eqref{eq:chi-perp-reprint-ntk-kstar}, and the two other needed Gaussian expectations~\eqref{eq:gaussian-expectation-ntk-k-star-2} and~\eqref{eq:gaussian-expectation-k-star-3}:
\begin{align}
&C_W g^{(\ell)}=\le[\frac{1}{(-a_1)}\ri]\frac{1}{\ell}+\ldots\, , \label{eq:ntk-k-star-asymptotic-begin}\\
&h^{(\ell)}=\frac{b_1}{2}+\ldots\, ,\\
&\chi_{\parallel}^{(\ell)}=1- \frac{2}{\ell}+\ldots\, ,\\
&\chi_{\perp}^{(\ell)}=1-\frac{p_{\perp}}{\ell}+\ldots\, ,\label{eq:ntk-k-star-perp-susceptibility-asymptotic}\\
&C_W^2\bra \sigma(z)\sigma(z)\sigma'(z)\sigma'(z) \ket_{K^{(\ell)}}=\le[\frac{1}{(-a_1)}\ri]\frac{1}{\ell}+\ldots\, ,\\
&C_W^2\bra \sigma'(z)\sigma'(z)\sigma'(z)\sigma'(z) \ket_{K^{(\ell)}}=1+\ldots\, .\label{eq:ntk-k-star-asymptotic-end}
\end{align}
Additionally, going forward we will assume that the bias and weight learning rates have the layer-dependence \eqref{eq:layer-independent-rates} motivated by equal per-layer NTK contribution.

With all this out of the way, it's straightforward to evaluate the large-$\ell$ asymptotics of $\Ti{F}{}{\ell}$ and $\Ti{B}{}{\ell}$. Plugging in the above expressions and the frozen NTK asymptotic solution~\eqref{eq:frozen-ntk-k-star-solution} into their recursions~\eqref{eq:F-recursion-single} and~\eqref{eq:B-recursion-single}, we get
\begin{align}
\NTHF{}{\ell+1}=&\le[1-\frac{4}{\ell}+\ldots\ri]\NTHF{}{\ell}+\le\{\le[\frac{1}{(-a_1)}\ri]\le[\widetilde{\lambda}_b+\frac{\widetilde{\lambda}_W\sigma_1^2}{(-a_1)}\ri]\le(\frac{1}{\ell}\ri)^{p_{\perp}}+\ldots\ri\}\, , \\
\NTHB{}{\ell+1}=&\le[1-\frac{2p_{\perp}}{\ell}+\ldots\ri]\NTHB{}{\ell}+\le\{\le[\widetilde{\lambda}_b+\frac{\widetilde{\lambda}_W\sigma_1^2}{(-a_1)}\ri]^2 \le(\frac{1}{\ell}\ri)^{2p_{\perp}-2}+\ldots\ri\}\, .
\end{align}
Substituting in our \terminate{scaling ansatz} \eqref{eq:master-scaling-ansatz-reprint}, they have the following asymptotic solutions at large $\ell$:
\begin{align}
\Ti{F}{}{\ell}&=\frac{1}{(5-p_{\perp})}\le[\frac{1}{(-a_1)}\ri] \le[\widetilde{\lambda}_b+\frac{\widetilde{\lambda}_W\sigma_1^2}{(-a_1)}\ri]\le(\frac{1}{\ell}\ri)^{p_{\perp}-1}+\ldots\, ,\label{eq:F-k-star-solution}\\
\Ti{B}{}{\ell}&=\frac{1}{3} \le[\widetilde{\lambda}_b+\frac{\widetilde{\lambda}_W\sigma_1^2}{(-a_1)}\ri]^2\le(\frac{1}{\ell}\ri)^{2p_{\perp}-3}+\ldots\, .\label{eq:B-k-star-solution}
\end{align}

Next, for the $\Ti{D}{}{\ell}$ recursion, let's start with the slightly more compact expression~\eqref{eq:D-recursion-single-compacter}.
Plugging in the expressions~\eqref{eq:ntk-k-star-asymptotic-begin}--\eqref{eq:ntk-k-star-asymptotic-end} along with the learning rates~\eqref{eq:layer-independent-rates} and the asymptotic solutions for the four-point vertex~\eqref{eq:Kstar-equals-zero-vertex-solution-reprint-ntk} and the frozen NTK~\eqref{eq:frozen-ntk-k-star-solution}, we get a recursion
\be
\NTHD{}{\ell+1}=\!\!\le[1-\frac{(p_{\perp}+2)}{\ell}+\ldots\ri]\!\NTHD{}{\ell}+\le\{\le[\frac{2}{3(-a_1)}\ri]\!\!\le[-p_{\perp}\widetilde{\lambda}_b-(p_{\perp}-1)\frac{\widetilde{\lambda}_W\sigma_1^2}{(-a_1)}\ri]\!\!\le(\frac{1}{\ell}\ri)^{p_{\perp}}\!\!\!\!\!+\ldots\ri\}\, .
\ee
This recursion can also be easily solved by using our scaling ansatz \eqref{eq:master-scaling-ansatz-reprint}, giving
\be\label{eq:D-k-star-solution}
\NTHD{}{\ell}=\frac{-2}{9(-a_1)}\le[p_{\perp}\widetilde{\lambda}_b+(p_{\perp}-1)\frac{\widetilde{\lambda}_W\sigma_1^2}{(-a_1)}\ri]\le(\frac{1}{\ell}\ri)^{p_{\perp}-1}+\ldots\, .
\ee

Finally, for $\Ti{A}{}{\ell}$ recursion~\eqref{eq:A-recursion-single-compacter}, the by-now-familiar routine of flipping 
back and forth in your book and plugging in the large-$\ell$ asymptotic expressions~\eqref{eq:ntk-k-star-asymptotic-begin}--\eqref{eq:ntk-k-star-asymptotic-end}, the learning rates~\eqref{eq:layer-independent-rates}, and  the asymptotic solutions for the four-point vertex~\eqref{eq:Kstar-equals-zero-vertex-solution-reprint-ntk}, for the frozen NTK~\eqref{eq:frozen-ntk-k-star-solution}, and for $\Ti{D}{}{\ell}$~\eqref{eq:D-k-star-solution}, gives
\be
\Ti{A}{}{\ell+1}=\le[1-\frac{2p_{\perp}}{\ell}+\ldots\ri]\Ti{A}{}{\ell}+\le\{\frac{4}{9}\le[p_{\perp}\widetilde{\lambda}_b+(p_{\perp}-1)\frac{\widetilde{\lambda}_W\sigma_1^2}{(-a_1)}\ri]^2 \le(\frac{1}{\ell}\ri)^{2p_{\perp}-2}+\ldots\ri\}\, ,
\ee
which can be solved using the same large-$\ell$ \terminate{scaling ansatz}~\eqref{eq:master-scaling-ansatz-reprint}, giving
\be\label{eq:A-k-star-solution}
\Ti{A}{}{\ell}=\frac{4}{27}\le[p_{\perp}\widetilde{\lambda}_b+(p_{\perp}-1)\frac{\widetilde{\lambda}_W\sigma_1^2}{(-a_1)}\ri]^2\le(\frac{1}{\ell}\ri)^{2p_{\perp}-3}+\ldots\, .
\ee
With this, we complete our evaluation of the agitated NTK\index{neural tangent kernel!agitated} statistics for the $K^\star=0$ universality class.\index{universality class!K@$K^\star=0$}\footnote{Curiously, for activation functions with $p_\perp=1$ such as $\tanhA$ and $\sinA$, the single-input tensors $D^{(\ell)}$ and $A^{(\ell)}$ are independent of the weight \terminate{learning rate} at leading order.
}

\index{scaling law}
Having solved all the recursions we have, let's collect and recollect the critical exponents.\index{critical exponent} From \eqref{eq:F-k-star-solution} and \eqref{eq:D-k-star-solution} we collect $p_F=p_D=p_{\perp}-1$, while from~\eqref{eq:B-k-star-solution} and~\eqref{eq:A-k-star-solution} we collect $p_B=p_A=2p_{\perp}-3$. Recollecting $p_0=1$ for the kernel and $p_{\Theta}=p_{\perp}-1$ for the frozen NTK, these critical exponents for the $K^\star=0$ universality class\index{universality class!K@$K^\star=0$} again obey $\ell/n$-scaling~\eqref{eq:NTK-scaling-laws} for the normalized quantities
defined in \eqref{eq:ntk-variance-scaling-ansatz} and \eqref{eq:ntk-cross-corr-scaling-ansatz}.
Together with our scale-invariant results~\eqref{eq:NTK-agited-exponents-scale-invariant}, this means the posited relations \eqref{eq:NTK-scaling-laws} do indeed persists across universality classes as scaling laws.\index{scaling law}

\index{neural tangent kernel!variance}
\index{cross correlation!NTK-preactivation}
In summary, we have found that the leading finite-width behavior of the NTK-preactivation joint distribution -- as measured by the NTK variance and NTK-preacitvation cross correlation -- has a \emph{relevant}\index{relevant (RG flow)} $\ell/n$ scaling regardless of activation function, as is natural according to the principles of our \terminate{effective theory}.

\section{Criticality, Exploding and Vanishing Problems, and None of That}\label{sec:EVGP-WEP}\index{exploding and vanishing gradient problem}

\index{exploding and vanishing gradient problem}\index{exploding and vanishing kernel problem}
Having now analyzed the NTK statistics of deep networks, let us culminate our discussion by revisiting our original motivation for \terminate{criticality}: \emph{exploding and vanishing problems}. In particular, let us finally introduce -- and then immediately abolish -- the exploding and vanishing gradient problem.\footnote{This problem was first noticed \cite{hochreiter1991untersuchungen,bengio-evgp-recurrent} in the context of \terminate{training} (the now somewhat deprecated) \emph{recurrent neural networks}\index{recurrent neural network} (RNNs), during the era when neural networks were still \emph{neural networks}\index{neural network} and not yet \neo{deep learning}, that is, at a time when MLPs\index{multilayer perceptron} weren't yet deep enough for this to have been an obvious issue.}

\subsubsection{Traditional view on the exploding and vanishing gradient problem}\index{exploding and vanishing gradient problem}\index{traditionality|seealso{exploding and vanishing gradient problem}}
Traditionally, the \terminate{exploding and vanishing gradient problem} is manifested by considering the behavior of the gradient of the loss for a deep network. Using the \terminate{chain rule} twice, the derivative 
of the loss with respect to a model parameter $\theta^{(\ell)}_{\mu}$ in the $\ell$-th layer -- either a bias $\theta_\mu^{(\ell)} \equiv \bias{j}{\ell}$ or a weight $\theta_\mu^{(\ell)} \equiv \W{jk}{\ell}$ -- takes the form
\be\label{eq:gradient-for-evgp}
\frac{\td \L_\A}{\td \theta_\mu^{(\ell)} }=\sum_{\alpha \in \D} \sum_{i_{L}=1}^{n_{L}}\sum_{i_{\ell}=1}^{n_{\ell}}\epsilon_{i_{L};\alpha} \frac{\td z_{i_{L};\alpha}^{(L)}}{\td z_{i_{\ell};\alpha}^{(\ell)}} \frac{\td  z_{i_{\ell};\alpha}^{(\ell)} }{\td \theta_\mu^{(\ell)}} \, .
\ee
In this gradient, the first factor is the \neo{error factor} \eqref{eq:error-factor-ntk}
\be\label{eq:error-factor-ntk-reprint}
\epsilon_{i;\alpha} \equiv \frac{\partial \L_\A}{\partial z^{(L)}_{i;\alpha}}\, ,
\ee 
the final factor is a \neo{trivial factor}~\eqref{eq:same-layer-derivatives}\index{trivial factor|seealso{exploding and vanishing gradient problem}}
\be\label{eq:same-layer-derivatives-reprint}
\frac{\td \z{i}{\alpha}{\ell}}{\td \bias{j}{\ell}} = \delta_{i j}\, , \qquad \frac{\td \z{i}{\alpha}{\ell}}{\td \W{jk}{\ell}} = \delta_{i j} \, \s{k}{\alpha}{\ell-1} \, ,
\ee
and the middle factor is the \neo{chain-rule factor}
\be\label{eq:backward-pass-iterated}
\frac{\td \z{i_L}{\alpha}{L}}{\td \z{i_\ell}{\alpha}{\ell}}=\sum_{i_{\ell+1}, \ldots, i_{L-1}}\frac{\td \z{i_L}{\alpha}{L}}{\td \z{i_{L-1} }{\alpha}{L-1}} \cdots\frac{\td \z{i_{\ell+1} }{\alpha}{\ell+1}}{\td \z{i_\ell}{\alpha}{\ell}}= \sum_{i_{\ell+1}, \ldots, i_{L-1}} \prod_{\ell'=\ell}^{L-1} \le[ \W{i_{\ell'+1} i_{\ell'} }{\ell'+1}\ds{i_{\ell'} }{\alpha}{\ell'} \ri]\, ,
\ee
which can be derived by iterating
the backward equation \eqref{eq:backward-pass} or equivalently by repeatedly using the \terminate{chain rule} in conjunction with the MLP forward equation~\eqref{eq:forward-pass}.
If this text causes you to experience a large \terminate{error factor} yourself, please flip backward to~\S\ref{sec:NTH-recursions} and review our discussion of the \neo{backpropagation} algorithm.

The point is that without any fine-tuning, the product of matrices from layer $\ell$ to layer $L$ in the chain-rule factor~\eqref{eq:backward-pass-iterated} will generically lead to exponential behavior. Even for networks of moderate depth, this makes it extremely difficult for the shallower-layer parameters to receive a well-behaved gradient and consequentially be properly trained: a vanishing gradient means that such parameters receive no training signal from the data and loss, while an exploding gradient is indicative of an instability in which the loss may increase or even blow up.
This is the \term{exploding and vanishing gradient problem}.
In a sense, this is a backward iteration dual of the already familiar \neo{exploding and vanishing kernel problem} that arises from the forward iteration equation.\footnote{
If you'd like, you can see this \terminate{duality} concretely by considering a deep linear network, for which the statistics of such a product of weights can be worked out exactly exactly as in \S\ref{ch:deep-linear-eft}.\index{deep linear network}}

Of course, not only does the \terminate{chain-rule factor}~\eqref{eq:backward-pass-iterated} need to be well behaved for stable \terminate{training}, but the \terminate{error factor}~\eqref{eq:error-factor-ntk-reprint} and \terminate{trivial factor}~\eqref{eq:same-layer-derivatives-reprint} must be as well. As we'll explain next, these latter factors are directly tied to the \terminate{exploding and vanishing kernel problem}. However, we'll also see that our well-understood notion of \terminate{criticality} is already sufficient to mitigate both exploding and vanishing problems together.

\subsubsection{Critical view on the exploding and vanishing gradient problem}\index{exploding and vanishing gradient problem!relation to criticality}
Critically, let us recall from~\S\ref{sec:criticality_DLN} and then~\S\ref{sec:bootstrapping} our discussion of the \terminate{exploding and vanishing kernel problem}. In those sections, we first motivated \terminate{criticality} as remedying exponential behavior in the \terminate{kernel} $\Ti{\ker}{\alpha_1\alpha_2}{\ell}$. As the $L$-th-layer kernel controls the \emph{typical} values of the network output -- and as the dataset's \terminate{label}s are generically order-one numbers -- we suggested that such an exploding or vanishing kernel would be problematic for \terminate{training}. 
Now that we know a little about \terminate{gradient descent}, we can actually see a more direct manifestation of this instability by considering all the factors that make up the network's gradient \eqref{eq:gradient-for-evgp}.

First, let's see how the \terminate{error factor}~\eqref{eq:error-factor-ntk-reprint} is tied to the \emph{exploding} kernel problem.
For example, the \terminate{error factor} for the MSE loss\index{loss!MSE}~\eqref{eq:MSE-loss} is given by~\eqref{eq:mse-function-approximation-error}
\be\label{eq:mse-function-approximation-error-mlp}
\epsilon_{i;\alpha}=\z{i}{\alpha}{L}-\y{i}{\alpha} \, .
\ee
As you can clearly see, if the kernel explodes, then the typical output -- and hence typical values of the \terminate{error factor} -- will explode as well.\footnote{Getting ahead of ourselves, a precocious reader might wonder whether this matters for the \emph{cross-entropy loss}\index{loss!cross-entropy}, since for that \terminate{loss} the \terminate{error factor} will stay of order one even if the network output explodes. However, in this case the model would then be (exponentially) overconfident on its predictions and such an \terminate{inductive bias} would be difficult to correct via \terminate{training}.
}
To ensure this does not happen, we must set~$\chi_{\parallel}\!\le(\Tif{\ker}{}{}\ri)\leq1$.

Second, notice that the \terminate{trivial factor}~\eqref{eq:same-layer-derivatives-reprint} for the weights is proportional to the activation. For activation functions contained in either of the scale-invariant and $K^\star=0$ universality classes\index{universality class!scale-invariant}\index{universality class!K@$K^\star=0$}, if the \terminate{kernel} -- and consequently the typical preactivation -- is exponentially small, then the activation -- and consequently the \terminate{trivial factor} -- will be exponentially suppressed. Subsequently, the weights in the deeper layers of the network would struggle to train as they only receive an exponentially small update. Thus, in order to avoid this \emph{vanishing} kernel problem, we demand $\chi_{\parallel}\!\le(\Tif{\ker}{}{}\ri)\geq1$.\footnote{Incidentally, for the scale-invariant universality class\index{universality class!scale-invariant}, this same logic provides an additional justification for avoiding the \emph{exploding} kernel problem. That is, since scale-invariant activation functions don't \emph{saturate}\index{saturation (of an activation)}, if $\chi_{\parallel}\!\le(\Tif{\ker}{}{}\ri)>1$, then the activation -- and consequentially the \terminate{trivial factor} -- would explode.}

Combining these two observations,  we see that the \terminate{exploding and vanishing kernel problem} is directly manifested as a subproblem of the \terminate{exploding and vanishing gradient problem} and further see how our \terminate{criticality} condition imposed on the \terminate{parallel susceptibility}, $\chi_{\parallel}\!\le(\Tif{\ker}{}{}\ri)=1$, serves to mitigate it.

\index{frozen NTK}
Moreover, we can shed further light on the vanishing of the \terminate{trivial factor} by considering its embodiment in the NTK. Considering our formal solution for the frozen NTK~\eqref{eq:ntk-mean-k-star-formal} and recalling the original definition \eqref{eq:nth-layer-sum-definition}, we can track the contribution of the weight derivatives as leading to the additive term $\LW{\ell'}g^{(\ell'-1)}$. For an exponentially vanishing kernel, this factor is exponentially suppressed as
\be
\LW{\ell'}g^{(\ell'-1)}\propto K^{(\ell'-1)}\lll 1\, ,
\ee
since $g\!\le(\ker\ri)=A_2 \ker$~\eqref{eq:helper-ntk-scale-invariant-g} for the scale-invariant universality class\index{universality class!scale-invariant} and $g\!\le(\ker\ri)=\sigma_1^2 \ker+\o{K^2}$~\eqref{eq:g0-reprint-ntk} for the $K^\star=0$ universality class\index{universality class!K@$K^\star=0$}. This is another way of seeing that such deeper-layer weights are not contributing to the training dynamics and, in particular, also implies that such weights will have a minimal effect on the updates to \emph{other} parameters.

\index{frozen NTK}\index{neural tangent kernel!mean}\index{gradient descent}\index{forward equation!NTK}
Similarly, we see that the \terminate{chain-rule factor}~\eqref{eq:backward-pass-iterated} is also encoded in the NTK in the multiplicative factor $\prod_{\ell''=\ell'}^{\ell-1}\Ti{\chi}{\perp}{\ell''}$~\eqref{eq:chi-perp-factor-solution-ntk}. In fact, such a factor was secretly always lurking in the NTK forward equation \eqref{eq:NTHchain} or \eqref{eq:NTH-recursion-without-expectation} as the coefficient of the recursive term.
To disclose that secret, note that in the \terminate{infinite-width limit} the expectation of the \terminate{chain-rule factor} factorizes, and we see from \eqref{eq:NTHchain} or \eqref{eq:NTH-recursion-without-expectation} that
\be
\E{\sum_{j_1,j_2}\frac{\td z_{i}^{(\ell+1)}}{\td z_{j_1}^{(\ell)}}\frac{\td z_{i}^{(\ell+1)}}{\td z_{j_2}^{(\ell)}}}\!=\E{\sum_{j_1,j_2}\W{ij_1}{\ell+1}\W{ij_2}{\ell+1}\sigma_{j_1}^{\prime\,(\ell)}\sigma_{j_2}^{\prime\,(\ell)}}\!=C_W\!\bra \sigma'(z)\sigma'(z)\ket_{\ker^{(\ell)}}=\chi_{\perp}^{(\ell)}\, ,
\ee
thus making this connection explicit.

\index{chain-rule factor}
As we discussed in \S\ref{sec:ntk_criticality} under the heading \emph{Formalities: perpendicular perturbations and the frozen NTK}, we need to ensure that these multiplicative chain-rule factors are under control for training to be well behaved, on average. In particular, if $\chi_{\perp}\!\le(\Tif{\ker}{}{}\ri)>1$, then the deeper-layer NTKs will exponentially explode, and the \terminate{training} dynamics will be unstable, potentially leading to growing losses. If instead $\chi_{\perp}\!\le(\Tif{\ker}{}{}\ri)<1$,
then the contribution of the biases and weights in the shallower layers to the deeper-layer NTKs will be exponentially diminished. This means both that such parameters will struggle to move via gradient-descent\index{gradient descent} updates and \emph{also} that they will struggle to influence the evolution of the parameters in the deeper layers. 

All in all, we see that contribution of the chain-rule factor to the \terminate{exploding and vanishing gradient problem} is directly connected to an exponentially growing or decaying multiplicative factor in the NTK and further see how our \terminate{criticality} condition imposed on the \terminate{perpendicular susceptibility}, $\chi_{\perp}\!\le(\Tif{\ker}{}{}\ri)=1$, serves to mitigate it.\footnote{
Having now explained why \neo{criticality} is a complete solution to the \terminate{exploding and vanishing gradient problem},
let us discuss remedies of \neo{traditionality}.

One of the first heuristic solutions -- first discussed in the context of recurrent neural networks\index{recurrent neural network} -- is \terminate{gradient clipping}\index{gradient clipping|seealso{exploding and vanishing gradient problem}} \cite{pascanu2013difficulty}, in which the norm of the gradient is reduced whenever it exceeds a certain threshold. As should be apparent given our discussion, such an ad hoc, distortionary, and hard-to-tune heuristic is completely unnecessary -- and potentially even destructive -- in networks that are at \terminate{criticality}. 

\index{activation function} 
A second heuristic solution was the adoption of the $\relu$. Recall that activation functions such as the $\tanhA$ and the $\sigmoid$ \emph{saturate}\index{saturation (of an activation)} when $|z| \to \infty$. This implies that the derivative of the activation vanishes upon saturation as $\sigma'(z)=0$. 
Such saturation naturally leads to vanishing gradients, as we can easily see from the right-hand side of \eqref{eq:backward-pass-iterated}. It is partially within this context that practitioners adopted the $\relu$ over saturating activation functions such as the $\tanhA$ (see, e.g.~\cite{glorot2011deep}). However, with our deeper and more critical understanding, we now appreciate that \terminate{criticality} is sufficient to mitigate this vanishing gradient problem for any activation function that admits critical \terminate{initialization hyperparameters}, even for saturating ones like $\tanhA$. 
}

\subsubsection{An equivalence principle for learning rates}
\index{frozen NTK}\index{exploding and vanishing gradient problem}\index{forward equation!NTK}
Although originally derived by considering the behavior of the \terminate{kernel} recursion, we just recovered both of our criticality conditions $\chi_{\parallel}\!\le(\Tif{\ker}{}{}\ri)=1$ and $\chi_{\perp}\!\le(\Tif{\ker}{}{}\ri)=1$ by a direct analysis of gradient-descent\index{gradient descent} updates and of the NTK forward equation.
Importantly, the guiding principle we found was that each layer should not make an exponentially different contribution to the NTK.

\index{learning rate equivalence principle|see{equivalence principle}}
However, there's really no reason for us to stop at the exponential level. In fact, we can further demand at the polynomial level that no type of model parameter and no layer dominate over another for \terminate{training}. In other words, rather than requiring \emph{more or less equal} contributions from the parameters in different layers, we demand parametrically \emph{equal} contributions to the NTK for each parameter group from every layer. This gives an \term{equivalence principle} for setting the \terminate{training hyperparameters}, i.e., the bias and weight learning rates. 
In fact, en route to this section, we already found a way to satisfy this equivalence principle for each of our universality classes.

As we retroactively saw in~\S\ref{sec:ntk_criticality_scale_invariant}, the equivalence principle was easily met for activation functions contained in the scale-invariant universality class\index{universality class!scale-invariant} 
by setting the bias and weight learning rates to be layer-independent:
\be\label{eq:super-scale-invariant}
\eta\Lb{\ell}=\frac{\eta\widetilde{\lambda}_b}{L}\, , \qquad \frac{\eta\lamW{\ell}}{n_{\ell-1}}=\frac{\eta\widetilde{\lambda}_W}{Ln_{\ell-1}}\, .
\ee
Here, we have also re-emphasized the rescaling of the weight learning rates by width of the previous layer as we discussed multiple times in~\S\ref{sec:NTH-recursions}. In particular, you should understand the \emph{parametrically equal contributions} provision of equivalence principle as requiring appropriate depth \emph{and} width scalings.
Also here -- with our discussion in~\S\ref{sec:NTH-recursions} under the heading \emph{Scaling in the effective theory} in mind -- note that we rescaled the learning rates by the overall depth $L$ of the network so as to have an order-one NTK in the output layer.
This ensures that we can naturally compare these rescaled learning rates $\eta\widetilde{\lambda}_b$ and $\eta\widetilde{\lambda}_W$ across models of different depths as well as that we have properly scaled order-one changes in observables\index{observable}, e.g.~the loss, after any gradient-descent\index{gradient descent} update.

Meanwhile for the $\ker^\star=0$ universality class\index{universality class!K@$K^\star=0$}, we retroactively saw in~\S\ref{sec:ntk_criticality_tanh_univ} that the \terminate{equivalence principle} requires
\be\label{eq:super-tanh-general}
\eta\Lb{\ell}=\eta\widetilde{\lambda}_b\le(\frac{1}{\ell}\ri)^{p_{\perp}}L^{p_{\perp}-1}\, , \qquad \frac{\eta\lamW{\ell}}{n_{\ell-1}}=\frac{\eta\widetilde{\lambda}_W}{n_{\ell-1}}\le(\frac{L}{\ell}\ri)^{p_{\perp}-1}\, ,
\ee
where here we recall our discussion of having separate depth scalings for the bias and weight learning rates~\eqref{eq:layer-independent-rates} as a way to ensure both uniform contributions in parameter groups and in layers to the asymptotic frozen NTK solution~\eqref{eq:frozen-ntk-k-star-solution}.\footnote{
    Please do not confuse these $\ell$-rescalings with $L$-rescalings. The former modifies the learning rates of a given layer $\ell$ by a factor of $\ell$, and the later modifies the learning rates of any layer in the network by the overall network depth $L$.

Also, with the recent critical discussion of the \terminate{exploding and vanishing kernel problem} in mind, you should see that this relative $\ell$-scaling between the bias and weight learning rates is just a polynomial version of the vanishing kernel problem: the \terminate{equivalence principle} ensures that deeper-layer weights receive polynomially non-vanishing gradients as well as contribute polynomially-equally to the training dynamics of other parameters.
}
In particular, for odd smooth activation functions such as $\tanhA$ and $\sinA$ the \terminate{critical exponent} for perpendicular perturbations is given by $p_{\perp}=1$, and we have a simpler prescription:
\be\label{eq:super-tanh}
\eta\Lb{\ell}=\frac{\eta\widetilde{\lambda}_b}{\ell}\, , \qquad \frac{\eta\lamW{\ell}}{n_{\ell-1}}=\frac{\eta\widetilde{\lambda}_W}{n_{\ell-1}}\, .
\ee
With this, we further wonder whether the empirical preference for $\relu$ over $\tanhA$ is at least partially due to the fact that the $\relu$ learning rates are naturally $\ell$-independent \eqref{eq:super-scale-invariant} while the $\tanhA$ learning rates require nontrivial $\ell$-rescalings \eqref{eq:super-tanh}.

In conclusion, while the optimal values of the order-one training hyperparameters $\eta\widetilde{\lambda}_{b}$ and $\eta\widetilde{\lambda}_{W}$ surely depend on the specifics of a task, we expect that the layer $\ell$, depth $L$, and width  $n_{\ell-1}$ scalings dictated by the \terminate{equivalence principle} will lead to the least variation across network architectures with differing widths $n_\ell$ and depths $L$.

%% file: Chp10-kernel/10_global.tex
\chapter{Kernel Learning}\label{ch:NTHb}
\epigraph{I protest against the use of an infinite quantity as an actual entity; this is never allowed in mathematics. The infinite is only a manner of speaking \dots .}{Carl Friedrich Gauss~\cite{gaussquote}.\index{Gauss, Carl Friedrich}\index{infinity}}

\noindent{}Now that we know essentially everything we can possibly know about the initialization distribution of the preactivations \emph{and the NTK}, it's finally time to learn \emph{with gradients}!

In this chapter, we'll analyze the training of infinite-width neural networks by gradient-descent optimization. Of course, the infinite-width network is really only a manner of speaking, and we cannot actually instantiate one in practice. However, as we saw from our previous finite-width analyses, they can still provide a useful \emph{model} of an actual entity when the depth-to-width ratio 
is sufficiently small.

Thus, the analysis of such networks is important for two reasons. First, this limit can tell us a lot about the correct scaling and tuning of our various hyperparameters; we've already seen this previously as our criticality analysis always begins at infinite width. Second, since our finite-width analysis is perturbative in $1/n$, understanding the infinite-width limit is a prerequisite for us to understand learning with more realistic finite-width networks in \S\ref{ch:features} and \S\ref{ch:eot}. 
With those remarks in mind, let's preview our analysis of gradient-based learning at infinite width.

Just as a new \terminate{biological neural network} begins its journey by taking a small step, so does a freshly-initialized artificial neural network. In \S\ref{sec:small-step} we'll take such a step, observing that the gradient-descent training of infinite-width networks is simply described by the frozen NTK and that the change in the network outputs can be consitently truncated to linear order in the global learning rate.
This simplicity leads us first to observe that the network's output components move independently of each other  (\S\ref{subsec:GD_no_wiring_at_infinity}) and then second to find an absence of representation learning in the hidden layers (\S\ref{subsec:GD_no_RL_at_infinity}).
At this point you might have an uncanny sense of d\'ej\`a vu, as we found the exact same limitations in \S\ref{sec:infinite-posterior} for infinite-width networks that learn via exact Bayesian inference.

After a small step comes a giant leap. In \S\ref{sec:giant-leap} we'll make a large parameter update and find a closed-form solution for a \emph{fully-trained} infinite-width network. For these networks, such a solution memorizes the entire training set, and we'll show that this solution is the same regardless of whether we reach it in one Newton step (\S\ref{subsec:memorization-at-infinity}) or many steps of (stochastic) gradient descent (\S\ref{subsec:algorithmic-independence-at-infinity}), and doesn't depend on the form of the loss that we use (\S\ref{subsec:cross-entropy}). 

In 
fact, in \S\ref{subsec:NTKprediction} we'll see that the prediction of a \emph{particular} fully-trained infinite-width network on unseen test inputs is fixed entirely by the initial network output, the frozen NTK, and the contents of the training set. To analyze this, we evaluate the statistics of the associated \emph{ensemble}, identifying the mean (neural tangent) kernel prediction as well as the covariance of that prediction across different realizations. Recalling our discussion of approximate methods for Bayesian model fitting in \S\ref{subsec:ForIO}, we are now able to make more precise the connection between gradient-based learning and maximum likelihood estimation by discussing the sense in which our distribution over fully-trained infinite-width networks is a generalized posterior distribution.\index{posterior!generalized posterior distribution}\index{generalized posterior distribution|see{posterior distribution}}\index{posterior!generalized posterior distribution|seealso{gradient-based learning}}

In \S\ref{sec:generalization-at-infinity}, we'll put the predictions of these fully-trained infinite-width networks to the test. Here we'll introduce the quantitative measure of training success, the \emph{generalization error}, and decompose it into a \emph{bias} term and a \emph{variance} term. The former term compares the mean predictions of the ensemble on the test inputs to the true function values from the test set, while the latter term measures the instantiation-to-instantiation fluctuations of that prediction across different fully-trained networks in our ensemble.

Naturally, there is a tradeoff between these bias and variance terms, corresponding to our preference for the ensemble to contain networks that are both flexible \emph{and} confident. By explicitly working out the generalization error when a test input is near one of the training samples in \S\ref{subsec:robustness-from-infinite-GD}, we'll see how balancing such a tradeoff gives a prescription for tuning the initialization hyperparmeters, via the principle of criticality, and for tuning the training hyperparameters, according to the learning rate equivalence principle. 

In \S\ref{subsec:star-polation} we'll extend our analysis to situations where a test input is near two training samples. This will let us understand how our fully-trained networks interpolate and extrapolate, letting us comment  on the activation-function-induced inductive bias of the network output in general. In particular, we'll be able to see how nonlinear activation functions are able to nonlinearly interpolate or extrapolate around two training examples.

Finally, in \S\ref{sec:lazy-kernel} we'll take a small step back to give our discussion of infinite-width networks a broader context. In particular, we'll introduce the linear model, one of the simplest models in traditional machine learning, and explain its relationship to another traditional set of algorithms known as kernel methods. This will let us see that infinite-width MLPs are essentially just linear models based on random features and dually let us identify both the infinite-width Bayesian \emph{kernel} and the frozen neural tangent \emph{kernel} with this more traditional notion of a \emph{kernel}.

After discussing the limitations of such kernel methods, you will thoroughly understand the need to go beyond the infinite-width limit so that our effective theory can fully incorporate some of the more exciting properties of practical deep learning models.

\section{A Small Step}\label{sec:small-step}\index{small step}\index{small step|seealso{giant leap}}\index{training dynamics!infinite width}
Now let's take our first step in a journey towards the minimum of the loss. We'll begin by considering how the preactivations change in the first step after initialization at $t=0$. 

Recalling that the  evolution of any neural-network observable $\O\!\le(z^{(\ell)}\ri)$  is governed by the NTK
through the update equation \eqref{eq:obsevable-evolution-layer-ell}, we have for an $\ell$-th-layer preactivation: %
\begin{align}\label{eq:GD-preactivation-reprint}
\dz{i}{\delta}{\ell}&\equiv \z{i}{\delta}{\ell}(t=1)- \z{i}{\delta}{\ell}(t=0) \, \\
&=-\eta\sum_{j=1}^{n_{\ell}}\sum_{\tra\in\A}\frac{d\L_{\A}}{d\z{j}{\tra}{\ell}} %
\Tia{\NTK}{ij}{\tra\delta}{\ell}+\o{\eta^2}\, . \notag
\end{align}
Here, the $\ell$-th-layer NTK $\Tia{\NTK}{ij}{\tra\delta}{\ell}$ and the factor $d\L_{\A}/d\z{j}{\tra}{\ell}$ are both evaluated at initialization; from now on we'll drop the explicit dependence on the number of steps $t$ when a quantity is  being evaluated at initialization $t=0$, unless we want to emphasize it for clarity. Henceforth, 
we'll  use the prefix $\dbar$  to indicate the \emph{update} to a quantity after the first step of gradient descent.

Further, in writing \eqref{eq:GD-preactivation-reprint} we have resurrected the sample-index notation of alpha-with-tilde for the inputs in the \terminate{training set} $\tra \in\A$, and we will soon resurrect beta-with-dot for inputs in the \terminate{test set} $\tea\in\B$; as before, we'll also use delta-with-no-decoration for generic inputs in either set: $\delta\in\D=\A\cup\B$. Thus, to be explicitly clear, the update \eqref{eq:GD-preactivation-reprint} gives the change after the first gradient descent training step in an $\ell$-th-layer preactivation evaluated on a sample from either the test set or the training set.

\index{frozen NTK}
Now, let's specialize to the \terminate{infinite-width limit}. Recall in this limit that the NTK self-averages\index{self-averaging}, such that the NTK for \emph{any} particular realization of the network parameters will be equal to the infinite-width NTK mean, which we have been calling the \emph{frozen NTK}: $\Tia{\NTK}{ij}{\tra\delta}{\ell}=\delta_{ij}\Ti{\NTKI}{\tra\delta}{\ell} + \o{1/n}$. With this in mind, the update equation  at infinite width simplifies to 
\be\label{eq:GD-preactivation-at-infty}
\dz{i}{\delta}{\ell}=-\eta\sum_{\tra\in\A}\Ti{\NTKI}{\delta\tra}{\ell}\frac{d\L_{\A}}{d\z{i}{\tra}{\ell}} + \oninv\, .
\ee
Here, the update does not mix \terminate{neural indices} as the mean of the NTK is diagonal in those indices, while the presence of off-diagonal terms in the frozen NTK would indicate that information from one training sample informs the update of another.
Note importantly that we have purposefully truncated the  $+\, \o{\eta^2}$ part of~\eqref{eq:GD-preactivation-reprint} that contains higher-order corrections to the update from the series expansion in the global learning rate\index{learning rate!global} $\eta$; in \S\ref{ch:features} we'll explicitly analyze these $\o{\eta^2}$ terms and show that they are suppressed by $1/n$.
Thus, in the strict infinite-width limit they identically vanish, making the linear truncation exact.

In this section, we'll take a look at what such an infinite-width small-step update entails for the network outputs with $\ell=L$ (\S\ref{subsec:GD_no_wiring_at_infinity}) and for preactivations in the final hidden layer  with $\ell=L-1$ (\S\ref{subsec:GD_no_RL_at_infinity}).\footnote{After reading the next section, it should be clear that the results here are true even at the minimum of the loss at the end of training. We say this now to head off any potential objections of the form, ``What if there are some number of steps $t$ for which the quantity $\eta t/n$ is of order one?''
}
 These analyses will more or less parallel \S\ref{subsec:absence-FF-Bayes} and \S\ref{subsec:absence-RL-Bayes}, where we considered the posterior distribution for infinite-width networks updated via exact \terminate{Bayesian inference}.

\subsection{No Wiring}\label{subsec:GD_no_wiring_at_infinity}\index{training dynamics!infinite width}
Specializing to  the network output $\z{i}{\delta}{L}$, the update~\eqref{eq:GD-preactivation-at-infty} 
simply becomes
\be\label{eq:GD-output-at-infty}
\dz{i}{\delta}{L}=-\eta\sum_{\tra\in\A}\Ti{\NTKI}{\delta\tra}{L}\epsilon_{i;\tra}\, ,
\ee
where we recall the now-familiar \terminate{error factor} defined in \eqref{eq:grad-loss-def} as
\be\label{eq:error-factor-reprint}
\epsilon_{i;\tra}\equiv \frac{\partial \L_\A}{\partial \z{i}{\tra}{L}}\, .
\ee
We're going to extensively analyze this update to network outputs in \S\ref{sec:giant-leap} and onwards. 
Here, let us just point out that the update to the $i$-th feature $\z{i}{\delta}{L}(t=1)$ depends only on the $i$-th component of the error factor $\epsilon_{i;\tra}$. This mirrors the phenomenon of \emph{no wiring} for the network outputs that we observed in \S\ref{subsec:absence-FF-Bayes} for the exact Bayesian inference at infinite width. 

To be more concrete, for the MSE loss\index{loss!MSE}~\eqref{eq:MSE-loss},
\be\label{eq:MSE-loss-reprint}
\L_{\A}\equiv \frac{1}{2}\sum_{i=1}^{n_L}\sum_{\tra\in\A}\Big( \z{i}{\tra}{L}-\y{i}{\tra}\Big)^2 \, ,
\ee
the error factor is simply given by the difference between the true output and the initial output, $\epsilon_{i;\tra}= \z{i}{\tra}{L}-\y{i}{\tra}$. We thus see that all the output components move independently from each other, and there's no way for correlations between these components to be 
learned.\footnote{As another example, we can take a cross-entropy loss\index{loss!cross-entropy} of the form $\L_{\A}=-\sum_{j,\tra} p_{j;\tra} \log(q_{j;\tra})$; feel free to flip forward and look at~\eqref{eq:loss-cross-entropy}. In this case we have 
a target distribution $p_{i;\tra}$ for which we want to fit the \emph{softmax}\index{softmax distribution} distribution -- cf.~\eqref{eq:softmax} -- of the network outputs $q_{i;\tra}\equiv \exp(\z{i}{\tra}{L})/\big[\sum_{k}\exp(\z{k}{\tra}{L})\big]$. 
After noting that $\partial q_{j;\tra}/\partial \z{i}{\tra}{L}=(\delta_{ij}-q_{i;\tra})q_{j;\tra}$, we find for the error factor
\be\label{eq:error-factor-cross-entropy}
\epsilon_{i;\tra}=\sum_{j}\frac{\partial \L_{\A}}{\partial q_{j;\tra}}\frac{\partial q_{j;\tra}}{\partial \z{i}{\tra}{L}}=-\sum_{j}\frac{p_{j;\tra}}{q_{j;\tra}}q_{j;\tra}(\delta_{ij}-q_{i;\tra})=-p_{i;\tra}+\big(\sum_{j}p_{j;\tra}\big)q_{i;\tra}=q_{i;\tra}-p_{i;\tra}\, .
\ee
Therefore, in this case too we see that the error factor $\epsilon_{i;\tra}$ depends only on the $i$-th component of the softmax\index{softmax distribution} distribution, and no correlation between output components will be generated.\index{error factor!cross-entropy loss}}
In addition, since \eqref{eq:GD-output-at-infty} is a stochastic equation describing the update to any particular network in the ensemble, there is no wiring for any particular realization of a one-step-into-training infinite-width network.
\index{gradient descent!wiring!infinite width}

\subsection{No Representation Learning}\label{subsec:GD_no_RL_at_infinity}\index{training dynamics!infinite width}
We'll have to work a little harder to analyze the update to the preactivations in the penultimate layer $\z{i}{\delta}{L-1}$. To start, we can evaluate the derivative of the loss in the update equation~\eqref{eq:GD-preactivation-at-infty} using the \emph{backward} equation\index{backward equation!MLP} \eqref{eq:backward-pass}:
\be
\frac{d\L_{\A}}{d\z{j}{\tra}{L-1}}=\sum_{i=1}^{n_{L}}\frac{\partial\L_{\A}}{\partial\z{i}{\tra}{L}}\frac{d\z{i}{\tra}{L}}{d\z{j}{\tra}{L-1}}=\sum_{i=1}^{n_{L}}\frac{\partial\L_{\A}}{\partial\z{i}{\tra}{L}}\,\W{ij}{L}\ds{j}{\tra}{L-1}\, .
\ee
Substituting this into the update \eqref{eq:GD-preactivation-at-infty} at $\ell = L-1$, we get a stochastic equation describing the change in the final hidden-layer representation for any particular network:
\be\label{eq:stochastic-penultimate-update}
\dz{j}{\delta}{L-1}=-\eta\sum_{i=1}^{n_{L}}\sum_{\tra\in\A}\Ti{\NTKI}{\delta\tra}{L-1} \frac{\partial\L_{\A}}{\partial\z{i}{\tra}{L}}\, \W{ij}{L}\ds{j}{\tra}{L-1}\, .
\ee 
To make progress here, we're going to have to analyze the distribution over such updates.

First, the mean update is given by
\be\label{eq:expected-move-penultimate-infty-intermediate}
\E{\dz{j}{\delta}{L-1}}=-\eta\sum_{i=1}^{n_{L}}\sum_{\tra\in\A}\Ti{\NTKI}{\delta\tra}{L-1}\E{\frac{\partial\L_{\A}}{\partial\z{i}{\tra}{L}}\,\W{ij}{L}\ds{j}{\tra}{L-1}}\, .
\ee
This expectation involves an \terminate{interlayer correlation} between the error factor $\partial \L_\A/\partial \z{i}{\tra}{L}$ 
from the $L$-th layer and the derivative of the activation $\ds{j}{\tra}{L-1}$ from the $(L-1)$-th layer, in addition to a weight insertion $\W{ij}{L}$. From our previous experience we know that such interlayer expectations are suppressed by a factor of $1/n$, vanishing in the strict infinite-width limit. To be extra careful,
 let's compute the mean explicitly.

\index{training dynamics!infinite width}
To do so, recall our \terminate{generating function} for interlayer correlations\index{interlayer correlation}~\eqref{eq:master-weight-insertion} and specialize to the penultimate layer $\ell=L-1$. (You'll probably want to flip back and remind yourself.) Since we have not previously evaluated the case with a single weight insertion, let's calculate and record it here. Differentiating the generating function once with respect to the source as $\frac{d}{d\JW{ij}}$ and then setting the source to zero, we get
\be\label{eq:one-weight-insertion}
\E{\O\!\le(z^{(L)}\ri) \W{ij}{L} \mathcal{Q}\!\le(z^{(L-1)}\ri)}=\frac{\CW{L}}{n_{L-1}}\sum_{\delta\in\D}\E{\bra\!\!\!\bra\frac{\partial \O}{\partial \z{i}{\delta}{L}} \ket\!\!\!\ket_{\!\!\!\widehat{G}^{(L)}}\!\!  \s{j}{\delta}{L-1} \mathcal{Q}\!\le(z^{(L-1)}\ri)}\, .
\ee
Applying this formula to the above expression for our update \eqref{eq:expected-move-penultimate-infty-intermediate},
we get
\begin{align}\label{eq:infinite-width-no-rl}
\E{\dz{j}{\delta}{L-1}}&=-\eta\,\frac{\CW{L}}{n_{L-1}}\sum_{i=1}^{n_{L}}\sum_{\tra_1\tra_2\in\A}\!\!\Ti{\NTKI}{\delta\tra_1}{L-1}\,\E{\bra\!\!\!\bra \frac{\partial^2 \L_\A}{\partial \z{i}{\tra_1}{L}\partial \z{i}{\tra_2}{L}} \ket\!\!\!\ket_{\!\!\!\widehat{G}^{(L)}}\!\!\s{j}{\tra_2}{L-1}\ds{j}{\tra_1}{L-1}}\, \\
&=-\eta\,\frac{\CW{L}}{n_{L-1}}\sum_{\tra_1,\tra_2\in\A}\!\Ti{\NTKI}{\delta\tra_1}{L-1} \sum_{i=1}^{n_{L}}\bra\!\!\!\bra \frac{\partial^2 \L_\A}{\partial \z{i}{\tra_1}{L}\partial \z{i}{\tra_2}{L}} \ket\!\!\!\ket_{\!\!\!K^{(L)}}\!\! \bra \sigma^{\prime}_{\tra_1}\sigma_{\tra_2} \ket_{K^{(L-1)}}\!+\!\o{\frac{1}{n^2}}\, ,\notag
\end{align}
and see that this expression is manifestly suppressed by $1/n$. Thus, on average across our ensemble, we have found that there's no representation learning in the penultimate layer in the infinite-width limit.

Next, let's consider the variance of the update~\eqref{eq:stochastic-penultimate-update}:
\begin{align}\label{eq:variance-move-penultimate-infty}
 &\E{\dz{j_1}{\delta_1}{L-1} \dz{j_2}{\delta_2}{L-1}} \\
 =&\eta^2 \sum_{i_1,i_2=1}^{n_{L}}\sum_{\tra_1 \tra_2 \in\A}\Ti{\NTKI}{\delta_1\tra_1}{L-1}\Ti{\NTKI}{\delta_2\tra_2}{L-1}\,\E{\frac{\partial\L_{\A}}{\partial\z{i_1}{\tra_1}{L}}\frac{\partial\L_{\A}}{\partial\z{i_2}{\tra_2}{L}}\,\W{i_1j_1}{L} \W{i_2j_2}{L}\ds{j_1}{\tra_1}{L-1}\ds{j_2}{\tra_2}{L-1} }\, \notag \\
 =& \delta_{j_1 j_2} \, \eta^2 \frac{\CW{L} }{n_{L-1}} \sum_{\tra_1 \tra_2 \in\A}\Ti{\NTKI}{\delta_1\tra_1}{L-1}\Ti{\NTKI}{\delta_2\tra_2}{L-1} \sum_{i=1}^{n_L}\bra \!\!\!\bra\frac{\partial\L_{\A}}{\partial\z{i_1}{\tra_1}{L}}\frac{\partial\L_{\A}}{\partial\z{i_2}{\tra_2}{L}}\ket\!\!\!\ket_{\ker^{(L)}} \braket{\dsNL{\tra_1}\dsNL{\tra_2} }{L-1}  + \o{\frac{1}{n^2}}\, \notag .
\end{align}
In the last step, we applied our interlayer correlation\index{interlayer correlation} formula with two weight insertions \eqref{eq:two-weight-insertions-general} and then picked up the leading contribution.\footnote{Note that the second term in \eqref{eq:two-weight-insertions-general} is of order $\o{1/n^2}$ and hence subleading.
} With this, we see that the covariance of the update,
\begin{align}
\cov{\dz{j_1}{\delta_1}{L-1}}{\dz{j_2}{\delta_2}{L-1} } &\equiv\E{\dz{j_1}{\delta_1}{L-1} \dz{j_2}{\delta_2}{L-1}} - \E{\dz{j_1}{\delta_1}{L-1}}\E{\dz{j_2}{\delta_2}{L-1}}=\oninv  \, ,
\end{align}
is manifestly suppressed by $1/n$, 
vanishing in the strict infinite-width limit. Since the distribution of updates to the penultimate-layer preactivations has a vanishing mean and covariance, we conclude that the distributions before and after the learning update are equal: mirroring what we found for exact Bayesian inference in \S\ref{subsec:absence-RL-Bayes}, there's no representation learning for gradient-based learning in the infinite-width limit.\footnote{
    You can check the higher-order connected correlators of the update distribution, if you'd like. However, as we already said before we started these computations, these sorts of interlayer correlations\index{interlayer correlation} are naturally suppressed by factors of $1/n$, and so will be the higher-order connected correlators.
}

\section{A Giant Leap}\label{sec:giant-leap}  \index{giant leap}\index{giant leap|seealso{small step}}\index{training dynamics!infinite width}
\epigraph{That's one \terminate{small step} for [a] machine, one giant leap for AI.}{
Neil AI-Strong\index{Armstrong, Neil}
}

\index{gradient descent}\index{training}
\noindent{}In the last section, we started to understand training for infinite-width networks by taking a small step of \terminate{gradient descent}. 
Of course, what we'd actually like is to understand the behavior of fully-trained networks at the minimum of their losses.
Naturally, we could continue by taking many many small steps until our networks are fully trained, and
indeed this is how networks are typically trained in practice. %

That said, in 
\S\ref{subsec:memorization-at-infinity} we'll first show that we can actually fully train infinite-width networks in \emph{one} theoretical gradient-descent step. That is, we can take a \emph{giant leap} right to the minimum of the loss. We'll then explain in \S\ref{subsec:algorithmic-independence-at-infinity} that the theoretical minimum
we've found by our giant leap is the same minimum we would have found in practice by taking many steps of gradient descent, or even by using \neo{stochastic gradient descent} with decreasing learning rates. This equivalence makes our giant leap a powerful theoretical tool.
After a brief aside about the cross-entropy loss in \S\ref{subsec:cross-entropy},
finally in \S\ref{subsec:NTKprediction} we'll see how our fully-trained infinite-width networks make predictions on previously unseen examples, though a detailed analyses of these test-set predictions will be postponed until the following  section.%

\subsection{Newton's Method}\label{subsec:memorization-at-infinity}\index{training dynamics!infinite width}
Our first goal is to find a single step such that the network outputs equal the true outputs, 
\be\label{eq:fully-trained-condition}
\z{i}{\tra}{L}(t=1) = \y{i}{\tra} \, , 
\ee
for all samples $x_{\tra}$ in the training set $\tra\in\A$. This condition will be our definition of \emph{fully trained}, and it's easy to see that such a condition will minimize the training loss for any of the loss functions that we've described.\index{fully-trained condition!infinite width}
Recalling the gradient-descent update for neural-network outputs~\eqref{eq:GD-output-at-infty} and rearranging terms, we see that our giant-leap update must satisfy
\be\label{eq:what-we-want-rearranged}
\z{i}{\tra_1}{L}-\y{i}{\tra_1}=\eta\sum_{\tra_2 \in\A} \Ti{\NTKIsub}{\tra_1 \tra_2}{L} \frac{\partial \L_{\A}}{\partial \z{i}{\tra_2}{L}}\, 
\ee
for the network to be fully trained. 

As a reminder, our convention is that quantities without an explicit step argument are evaluated at the point of initialization $t=0$; in particular, the constraint \eqref{eq:what-we-want-rearranged} is written solely in terms of the quantities at initialization. Additionally, note that the tilde on $\Ti{\NTKIsub}{\tra_1 \tra_2}{L}$ emphasizes that it's a $\NR\times\NR$-dimensional submatrix of the full frozen NTK matrix $\Ti{\NTKI}{\delta_1 \delta_2}{L}$ evaluated on  pairs of inputs $(x_{\tra_1}, x_{\tra_2})$ in the \terminate{training set} $\A$ \emph{only}. This emphasis will soon prove itself useful, as it did before in \S\ref{ch:bayesian-inference}.

How can we satisfy our giant-leap condition \eqref{eq:what-we-want-rearranged}? Since the left-hand side is exactly the error factor of the \emph{MSE loss}\index{loss!MSE}~\eqref{eq:MSE-loss-reprint}, let's first specialize to the MSE loss. Plugging in~\eqref{eq:MSE-loss-reprint} for $\L_{\A}$, we get a concrete equation to solve: 
\be\label{eq:newton-minimum-implicit}
\z{i}{\tra_1}{L}-\y{i}{\tra_1}=\sum_{\tra_2 \in\A}\eta \Ti{\NTKIsub}{\tra_1 \tra_2}{L}\le(\z{i}{\tra_2}{L}-\y{i}{\tra_2}\ri) \,.
\ee
However, for generic neural networks, the frozen NTK\index{frozen NTK} $\Ti{\NTKI}{\tra_1 \tra_2}{L}$ will have nonzero off-diagonal components mixing different sample indices.
This unfortunately means that the condition \eqref{eq:newton-minimum-implicit} is impossible to satisfy by the tuning of the single global learning rate $\eta$\index{learning rate!global}.

Said another way, the issue is that our global learning rate $\eta$ is a \emph{scalar}, but here we need it to be \neo{tensor} in order to undo the mixing of the sample indices by the frozen NTK\index{frozen NTK}.
To enable this, we need to further generalize gradient descent. In our first extension of the gradient descent algorithm \eqref{eq:gd-update-lambda} -- discussed under the heading \emph{Tensorial Gradient Descent}\index{gradient descent!tensorial} -- we introduced a \emph{learning-rate tensor}\index{learning rate!learning-rate tensor} on \terminate{parameter space},
\be\label{eq:generalization-original}
\eta\to\eta\lambda_{\mu\nu} \, ,
\ee
which let us mediate how each model parameter individually contributes to the gradient-descent update of the others and let us take steps with unequal magnitudes in various directions in parameter space.  The consequence of having such a learning-rate tensor was integrated into the definition of the NTK and then informed our analyses in \S\ref{ch:training}--\S\ref{ch:eft-ntk}.\footnote{Most importantly, this let us scale the effective learning rate differently for the biases and weights; we saw in \S\ref{sec:NTH-recursions} and then in \S\ref{sec:EVGP-WEP} that this was essential for ensuring that both parameter groups get properly trained. Also, please remember that, even when written generally as $\lambda_{\mu\nu}$, our learning-rate tensor is restricted such that it does not mix parameters from different layers.}

\index{training dynamics!infinite width}
Now, we need to further extend this generalization to sample indices as
\be\label{eq:generalization-of-generalization}
\eta\lambda_{\mu\nu}\to\eta \lambda_{\mu\nu}\kappa^{\tra_1\tra_2}\, ,
\ee
where we have introduced a new symmetric matrix $\kappa^{\tra_1\tra_2}$ that we will call the \term{Newton tensor}\index{Newton tensor|seealso{second-order update}}\index{tensor!Newton tensor|see{Newton tensor}}.
This enables us to take an anisotropic step in \terminate{sample space} as well.
Specifically, we extend the parameter update equation \eqref{eq:gd-update-lambda} to
\begin{align}\label{eq:second-order-update}
\dtheta_\mu \equiv\theta_\mu(t=1)-\theta_\mu(t=0) &=  
- \sum_{\nu,\tra_1,\tra_2,i}\eta \lambda_{\mu\nu} \kappa^{\tra_1\tra_2}\frac{\td \z{i}{\tra_1}{L}}{\td \theta_\nu}\frac{\partial \L_{\A}}{\partial \z{i}{\tra_2}{L}} \, ,
\end{align}
which we will call a \term{second-order update}\index{second-order update|seealso{Newton tensor}}.\footnote{The name of this update descends from the fact that similar updates are used to define optimization algorithms that
incorporate information from the second derivative of the loss. Such algorithms are generally called \emph{second-order methods}\index{second-order method (optimization)}\index{second-order method (optimization)|seealso{Newton's method}}. We will show shortly that this new algorithm minimizes the loss just as well (better, actually).}
Plugging this second-order update 
into our expansion for the network outputs, we get
\begin{align}\label{eq:expansion-outputs}
\z{i}{\delta_1}{L}(t=1) &= \z{i}{\delta_1}{L} + \sum_\mu\frac{\td \z{i}{\delta_1}{L}}{d \theta_\mu}\dtheta_\mu + \o{\frac{1}{n}}  \, \\
&=\z{i}{\delta_1}{L} - \eta\sum_{\tra_2, \tra_3 \in\A}  \Ti{\NTKI}{\delta_1 \tra_2}{L} \kappa^{\tra_2 \tra_3}\frac{\partial \L_{\A}}{\partial \z{i}{\tra_3}{L}} + \oninv \, .\notag
\end{align}
Substituting this update into our fully-trained condition~\eqref{eq:fully-trained-condition} and still using the MSE loss\index{loss!MSE}, we get a new constraint
\be\label{eq:newton-minimum-implicit-satisfiable}
\z{i}{\tra_1}{L}-\y{i}{\tra_1}=\sum_{\tra_2,\tra_3 \in\A}\le(\eta \Ti{\NTKIsub}{\tra_1 \tra_2}{L}\kappa^{\tra_2\tra_3}\ri)\le(\z{i}{\tra_3}{L}-\y{i}{\tra_3}\ri) \, .
\ee
We'll satisfy this constraint shortly.

\index{training dynamics!infinite width}
For a different perspective on this new second-order update, rather than modifying the optimization algorithm, we can instead find the same constraint \eqref{eq:newton-minimum-implicit-satisfiable} by adopting a different loss. Consider a \emph{generalized} MSE loss\index{loss!MSE!generalized}
\be\label{eq:loss-MSE-gen}
\Laux{\A}(\theta) =\frac{1}{2}\sum_{i=1}^{n_L} \sum_{\tra_1,\tra_2 \in \A}\kappa^{\tra_1 \tra_2}\le(\z{i}{\tra_1}{L}-\y{i}{\tra_1}\ri)\le(\z{i}{\tra_2}{L}-\y{i}{\tra_2}\ri) \, ,
\ee
where here the \terminate{Newton tensor} $\kappa^{\tra_1 \tra_2}$ acts as a \emph{metric}\index{Newton tensor!as a metric on sample space} on \terminate{sample space}.\footnote{Similarly, we could have taken the perspective that the learning-rate tensor\index{learning rate!learning-rate tensor} $\lambda_{\mu\nu}$ acts as a metric on \neo{parameter space}.

Note also that with this interpretation, the standard MSE loss is just the generalized MSE loss with the Euclidean metric $\kappa^{\tra_1 \tra_2} \to \delta^{\tra_1 \tra_2}$. In some sense, it's more pleasing to write it this way if you're familiar with general relativity\index{general relativity}; writing the \terminate{Newton tensor} with \terminate{sample indices} \emph{raised} allows us to adopt a rule of only summing over sample indices when they come in a raised-lowered pair. Similarly, note that the insertion of the Newton tensor in our \terminate{second-order update} \eqref{eq:second-order-update} follows this pattern as well.
} For this loss the derivative with respect to the network output -- i.e.~the error factor --
is now given by
\be
\frac{\partial \L_{\A}}{\partial \z{i}{\tra_2}{L}}=\sum_{\tra_3\in\A}\kappa^{\tra_2 \tra_3}\le(\z{i}{\tra_3}{L}-\y{i}{\tra_3}\ri)\, .
\ee
Substituting this error factor into our condition for being fully trained~\eqref{eq:what-we-want-rearranged}, we find the same constraint~\eqref{eq:newton-minimum-implicit-satisfiable} using a standard gradient-descent update with our generalized MSE loss \eqref{eq:loss-MSE-gen} as we did just before using our \terminate{second-order update}~\eqref{eq:second-order-update} with the standard MSE loss. Either perspective is a valid way to think about our theoretical optimization and, as we will explain more generally in \S\ref{subsec:algorithmic-independence-at-infinity},  any of these choices of algorithms and losses will lead to the same fully-trained network.

\index{training dynamics!infinite width}
Now, let's find the solution to our giant-leap constraint \eqref{eq:newton-minimum-implicit-satisfiable}. By inspection, this is satisfiable if we can set the term in the first parenthesis to the identity matrix
\be
\sum_{\tra_2 \in\A}\eta \Ti{\NTKIsub}{\tra_1 \tra_2}{L}\kappa^{\tra_2\tra_3} = \delta_{\tra_1}^{\ \tra_3} \, ,
\ee
which we can ensure by setting the product of the global learning rate\index{learning rate!global} and the \terminate{Newton tensor} as
\be\label{eq:newtons-method-update}
\eta \kappa^{\tra_1\tra_2}=\TI{\NTKIsub}{\tra_1 \tra_2}{L} \, .
\ee
Here, the object on the right-hand side is the \emph{inverse} of the $\NR\times\NR$-dimensional submatrix $\Ti{\NTKIsub}{\tra_1 \tra_2}{L}$, which as a reminder is evaluated on pairs of inputs $(x_{\tra_1}, x_{\tra_2})$ in the \terminate{training set} $\A$ \emph{only}, and is defined via the equation
\be\label{eq:training-set-ntk-inverse}
 \sum_{\tra_2\in\A} \TI{\NTKIsub}{\tra_1 \tra_2}{L} \, \Ti{\NTKIsub}{\tra_2 \tra_3}{L} =\delta^{\tra_1}_{\ \tra_3}\, .
\ee
Similarly to our work in \S\ref{sec:infinite-posterior} on infinite-width \terminate{Bayesian inference}, the decoration of these submatrices with tildes is useful in order to clearly distinguish these submatrices from submatrices that also involve the \neo{test set} $\B$. 
Also, as before for the kernel and its inverse, we will always denote the NTK inverse by an object with sample indices \emph{raised}.

The algorithm with the particular choice \eqref{eq:newtons-method-update} is known as \term{Newton's method} (which acausally explains why we called $\kappa^{\tra_1 \tra_2}$ the \neo{Newton tensor}).\footnote{\emph{Newton's method}\label{footnote:newtons-method} is a numerical method for finding a zero of a function. For simplicity of presentation, let's take a single-variable function $g(x)$ and suppose that we want to find a solution to the equation $g\!\le(x_\star\ri)=0$; note that this is equivalent to extremizing a function $L(x)$ whose derivative is $g(x)$, i.e.~$L'(x)=g(x)$.
\terminate{Newton's method} instructs us to start with some guess $x_{0}$ and then iterate as
    \be\label{eq:Newtons-method}
       x_{t+1} = x_{t} - \frac{g\!\le(x_{t}\ri)}{g'\!\le(x_{t}\ri)}= x_{t} - \frac{L'\!\le(x_{t}\ri)}{L''\!\le(x_{t}\ri)}\, .
    \ee 
This algorithm is based on making a linear approximation $g'(x_{t})\approx [g(x_{t+1})-g(x_{t})]/(x_{t+1}-x_{t})$ and then solving for $g(x_{t+1})=0$.
In general, one needs to iterate \eqref{eq:Newtons-method} for several steps in order to get a good approximate solution $x_{\star}$.
When the function is linear as $g(x)=a (x-x_{\star})$, however, we get
 \be
 x_{1}=x_0-\frac{a(x_0-x_{\star})}{a}= x_{\star}\, ,
 \ee
for any starting point $x_0$.
Hence Newton's method can land right on the solution in one step, just like our giant leap \eqref{eq:parameters-update-reprint-generalized} did.

\index{training dynamics!infinite width}
The right-hand side of \eqref{eq:Newtons-method} offers another perspective: Newton's method is gradient descent with a ``loss'' $L(x)$ and a learning rate set as $\eta_t=1/L''(x_{t})$. To see why this is a good choice for the learning rate, let's choose a generic learning rate $x_{t+1}=x_t-\eta_t L'(x_t)$ and Taylor-expand the updated loss $L(x_{t+1})$ to the second order in $\eta_t$:
\be
L\!\le(x_{t+1}\ri) = L\!\le(x_{t}\ri) - \eta_t L'\!\le(x_{t}\ri)^2 + \frac{\eta_t^2}{2} L'\!\le(x_{t}\ri)^2L''\!\le(x_{t}\ri)  + \o{\eta_t^3}\, .
\ee
Optimizing the learning rate, we see that the truncated expression on the right-hand side is minimized when $\eta_t=1/L''(x_{t})$. In particular, for a quadratic function $L(x)=a(x-x_{\star})^2/2$ this truncation is exact, and Newton's method again reaches the minimum in one step.
This also makes it clear why optimization algorithms based on Newton's method\index{Newton's method! as a second-order method} fall in the class of \emph{second-order methods}\index{second-order method (optimization)}: each iteration uses second-order information from the function -- the second derivative $L''(x)$ -- to set the locally optimal learning rate. 
Our giant leap expressed in \eqref{eq:parameters-update-reprint-generalized} and \eqref{eq:network-step-general} is doing exactly that -- successfully -- for the parameter optimization and for the function approximation, respectively.
}
With it, we can simply write down a solution that fully trains the network in one step:
\begin{align}
\theta_\mu^\star &= \theta_\mu(t=0)  - \sum_{\nu,\tra_1,\tra_2,i}\lambda_{\mu\nu} \frac{\td \z{i}{\tra_1}{L}}{\td \theta_\nu} \TI{\NTKIsub}{\tra_1 \tra_2}{L}\le(\z{i}{\tra_2}{L}-\y{i}{\tra_2}\ri) \, . \label{eq:parameters-update-reprint-generalized}
\end{align}
In particular, this is exactly what we'd find by setting the gradient of the loss to zero and solving for the optimal parameters as in~\eqref{eq:gradient-vanishing-mind}.
As we explained back there, such a direct and explicit solution to an optimization problem is only available in special cases, and
it turns out that this is precisely the case at infinite width.\footnote{We'll later show in \S\ref{sec:another-leap} that perturbative solutions are possible at finite width.}

\index{training dynamics!infinite width}
Plugging the Newton's method update in our expansion for the network outputs \eqref{eq:expansion-outputs}, we then find the fully-trained network output for a general input $\delta\in\D$: 
\be
\z{i}{\delta}{L}(t=1)=\z{i}{\delta}{L}- \sum_{\tra_1, \tra_2 \in\A}  \Ti{\NTKI}{\delta \tra_1}{L} \TI{\NTKIsub}{\tra_1 \tra_2}{L}\le(\z{i}{\tra_2}{L}-\y{i}{\tra_2}\ri) \, .\label{eq:network-step-general}
\ee
In particular,
for samples in the training set $\A$, the network output equals the true output $\z{i}{\tra}{L}(t=1)=\y{i}{\tra}$, satisfying our condition for the network to be fully trained~\eqref{eq:fully-trained-condition}.
In other words, our network has perfectly memorized the entire training set.\footnote{Since infinite-width networks have infinite parameters, it shouldn't be surprising that in this limit the network can memorize the finite training set.} 
As such, this solution \eqref{eq:parameters-update-reprint-generalized} also minimizes \emph{any} loss $\L_\A(\theta)$
that is minimized by setting the network outputs to the true outputs $z^{(L)}(x_{\tra};\theta)=\y{i}{\tra}$:
\be\label{eq:loss-minimization-by-Newton}
\theta^\star_{\text{Newton}} = \argmin_\theta\, \L_\A(\theta)\, .
\ee
This means that regardless of whether we used the standard MSE loss \eqref{eq:MSE-loss-reprint} or the generalized MSE loss~\eqref{eq:loss-MSE-gen} (or an entirely different loss as long as it has a minimum at $z^{(L)}(x_{\tra};\theta)=\y{i}{\tra}$), our solution \eqref{eq:network-step-general} will faithfully describe the minimum.\footnote{
    Note that the solution~\eqref{eq:parameters-update-reprint-generalized} depends on the network output at initialization $\z{i}{\delta}{L}$, which ultimately depend on the initialization of the parameters $\theta_{\text{init}}=\theta(t=0)$. For different initializations, we will reach different solutions~\eqref{eq:parameters-update-reprint-generalized}, each of which will minimize the loss given that particular initialization $\theta_{\text{init}}$. We'll have more to say about this in \S\ref{subsec:NTKprediction}.
}

\subsection{Algorithm Independence}\label{subsec:algorithmic-independence-at-infinity}
Now let's discuss a related -- and by now well anticipated -- property of the infinite-width limit: given a particular initialization $\z{i}{\delta}{L}(t=0)$ and the frozen NTK $\Ti{\NTKI}{\delta \tra_1}{L}$, we'll always get to exactly the same minimum \eqref{eq:network-step-general}, whether we get there by one giant leap or we get there by a sequence of many small steps. That is, at infinite width we have \term{algorithm independence}.

\index{learning rate!global!step-dependent}
Let's suppose that we have taken $T-1$ steps towards the minimum with a global learning rate $\eta(t)$, a \terminate{Newton tensor} $\kappa^{\tra_1\tra_2}(t)$, and loss $\L_{\A}(t)$, where these quantities
will in general depend on the step $t$. Different choices of $\eta(t)$, $\kappa^{\tra_1\tra_2}(t)$, and $\L_{\A}(t)$ will lead to different optimization algorithms; included in this class are \terminate{Newton's method}, \terminate{gradient descent}, and \terminate{stochastic gradient descent} (SGD).\footnote{To see how this includes SGD~\eqref{eq:sgd-update}, note that we can either restrict the loss to be a summation over a different \neo{batch} $\mathcal{S}_t\subset \A$ at each step $t$ as $\L_{\A}(t)= \L_{\mathcal{S}_t}$, or equivalently we can choose the Newton tensor $\kappa^{\tra_1\tra_2}(t)$ to project onto the subset $\mathcal{S}_t$.}
Iterating the update~\eqref{eq:expansion-outputs}, the network outputs accumulate the changes as
\be\label{eq:general-gd-steps}
\z{i}{\delta}{L}(T-1)=\z{i}{\delta}{L}(t=0)-\sum_{t=0}^{T-2}\sum_{\tra_1,\tra_2}\Ti{\NTKI}{\delta\tra_1}{L}\,\eta(t)\,\kappa^{\tra_1\tra_2}(t) \, \epsilon_{i;\tra_2}(t) \, ,
\ee
where $\epsilon_{i;\tra_2}(t)\equiv\partial \L_{\A}(t)/\partial \z{i}{\tra_2}{L}$ is the error factor for the training loss $\L_{\A}(t)$ evaluated with respect to the network output $\z{i}{\tra}{L}(t)$ at step $t$.

Let's then suppose that in the next step $t=T$ that we reach the true minimum. We can ensure this by taking a Newton step from $\z{i}{\delta}{L}(T-1)$ such that
\be\label{eq:newton-at-last-intermediate}
\z{i}{\delta}{L}(T)=\z{i}{\delta}{L}(T-1)-\sum_{\tra_1,\tra_2}\Ti{\NTKI}{\delta\tra_1}{L}\TI{\NTKIsub}{\tra_1\tra_2}{L}\le[\z{i}{\tra_2}{L}(T-1)- \y{i}{\tra_2}\ri]\, ,
\ee
where here we set $\eta(T-1)\, \kappa^{\tra_1\tra_2}(T-1)=\TI{\NTKIsub}{\tra_1\tra_2}{L}$ and chose the standard MSE loss at $t=T-1$ with $\epsilon_{i;\tra_2}(T-1)=\z{i}{\tra_2}{L}(T-1)-\y{i}{\tra_2}$.\footnote{Note that if we had already reached a minimum at step $T-1$, then this last Newton's step in \eqref{eq:newton-at-last-intermediate} would give no change to the network outputs: $\z{i}{\delta}{L}(T)=\z{i}{\delta}{L}(T-1)$. Thus, our argument also applies to any algorithm that already reached a minimum with $T-1$ other steps, and we do not have to actually apply the Newton step in practice. We'll address this point again in the next-to-next footnote.
}
Plugging in our expression for the network outputs after the first $T-1$ steps~\eqref{eq:general-gd-steps}, we see
\begin{align}
\z{i}{\delta}{L}(T)=&\z{i}{\delta}{L}(t=0)-\sum_{t=0}^{T-2}\sum_{\tra_1,\tra_2}\Ti{\NTKI}{\delta\tra_1}{L}\, \eta(t)\, \kappa^{\tra_1\tra_2}(t)\, \epsilon_{i;\tra_2}(t)\, \notag\\
&-\sum_{\tra_1,\tra_2}\!\Ti{\NTKI}{\delta\tra_1}{L}\TI{\NTKIsub}{\tra_1\tra_2}{L}\le\{\le[\z{i}{\tra_2}{L}(t=0)\!-\!\!\sum_{t=0}^{T-2}\sum_{\tra_3,\tra_4}\!\Ti{\NTKIsub}{\tra_2\tra_3}{L}\, \eta(t)\, \kappa^{\tra_3\tra_4}(t)\, \epsilon_{i;\tra_4}(t)\ri]\!-\!\y{i}{\tra_2}\!\ri\}\, \notag\\
=&\z{i}{\delta}{L}(t=0)- \sum_{\tra_1, \tra_2 \in\A}  \Ti{\NTKI}{\delta \tra_1}{L} \TI{\NTKIsub}{\tra_1 \tra_2}{L}\le[\z{i}{\tra_2}{L}(t=0)-\y{i}{\tra_2}\ri]\, ,\label{eq:newton-at-last}
\end{align}
where to go from the second to the third line we made use of the defining equation for the NTK submatrix inverse~\eqref{eq:training-set-ntk-inverse}, thus enabling the cancellation. What this result shows is that all the details of the training algorithm in the previous steps $\big\{\eta(t), \kappa^{\tra_1\tra_2}(t), \L_{\A}(t)\big\}_{t=0,\ldots, T-2}$  were erased: the network output $\z{i}{\delta}{L}(T)$ after our final step $t=T$ here in~\eqref{eq:newton-at-last} is exactly the same as the network output reached after one giant leap~\eqref{eq:network-step-general}.\footnote{In \S\ref{subsec:real-GD-at-finite-width}, we'll explicitly analyze the dynamics of another optimization algorithm -- many many steps of vanilla gradient descent \eqref{eq:gd-update-lambda} -- and evaluate its corresponding fully-trained solution. As expected by \terminate{algorithm independence} \eqref{eq:newton-at-last}, in the infinite-width limit this solution agrees completely with other solutions obtained by different training algorithms.}

Thus, in the infinite-width limit the fully-trained solution is determined by \emph{(i)} the frozen NTK\index{frozen NTK} $\Ti{\NTKI}{\delta\tra}{L}$, with its details depending on the \neo{training hyperparameters} in the learning-rate tensor\index{learning rate!learning-rate tensor} $\lambda_{\mu\nu}$,
\emph{(ii)} the initial newtork outputs $\z{i}{\delta}{L}(t=0)$, with its distribution depending on the \neo{initialization hyperparameters}, and \emph{(iii)} the true outputs $\y{i}{\tra}$ for the training set $\A$. It doesn't matter which loss function we used, e.g.~MSE\index{loss!MSE} or cross-entropy\index{loss!cross-entropy}, how many steps we took to get to the minimum, or whether we used gradient descent or SGD\index{stochastic gradient descent}.\footnote{
    Some additional comments that didn't make the cut for the main body:
    \bi
    \item \terminate{Newton's method} is often impractical to implement directly, since we have to compute the inverse of the frozen NTK submatrix evaluated on the entire training set, $\TI{\NTKIsub}{\tra_1 \tra_2}{L}$, similar to how we had to invert the kernel for \terminate{Bayesian inference} in \S\ref{subsec:absence-FF-Bayes}. Unlike the case of exact Bayesian inference where we were considering the feasibility of the learning algorithm, here the point is that Newton's method is a theoretical tool that lets us describe a fully-trained  extremely-wide network, even if the network was trained very practically by a many-step version of (stochastic) gradient descent.
    \item Often theorists will take the limit of a very small step size and approximate the optimization dynamics with an ordinary differential equation (ODE)\index{gradient descent!continuum or ODE limit}. Such an approximation is sometimes misleading, and here we see that it's entirely unnecessary.
    \item For SGD to actually converge to a minimum, you need to decrease the learning rate over the course of training, otherwise the network will fluctuate around, but never actually reach, the minimum. Intuitively, this is because at each step the optimization problem does not include the entire training set.
    \item A curious reader might wonder what happens if one cannot take a final step according to \terminate{Newton's method}, for instance because it's impractical to invert the frozen NTK submatrix. In fact, if you're already close to the minimum at $t=T-1$, i.e.~if you're essentially at the end of training, then the final step to land exactly on the minimum will be extremely small, and the solution \eqref{eq:newton-at-last} is a very good approximation of the network before taking this last theoretical jump.
    \ei
}
Said another way, \neo{algorithm independence} means that these hyperparameters and training set uniquely specify the statistics of fully-trained networks in the infinite-width limit; \terminate{Newton's method} is just a nice theoretical trick to leap right to the solution.
In this way, we can use the giant-leap solution \eqref{eq:network-step-general} to study the outcome of all these different optimization algorithms, which is what we'll do after a brief aside about the cross-entropy loss.\index{loss!cross-entropy}

\subsection{\emph{Aside}: Cross-Entropy Loss}\label{subsec:cross-entropy}
Let's take a brief aside to bring the \emph{cross-entropy loss}\index{loss!cross-entropy} out of the footnotes and into the main body. In general, the cross-entropy loss for some dataset\index{input data} $\D$ is defined as
\be\label{eq:loss-cross-entropy}
\L_{\D}=-\sum_{\delta\in\D}\sum_{i=1}^{n_\text{out}} p\!\le(i|x_\delta\ri) \log\!\le[q\!\le(i|x_\delta\ri)\ri] \, ,
\ee
where $p\!\le(i|x_\delta\ri)$ is a discrete distribution over the components $i$ of the true output
\be\label{eq:softmax-target}
p\!\le(i|x_\delta\ri)\equiv \frac{\exp\!\le[\y{i}{\delta}\ri]}{\sum_{j=1}^{n_\text{out}}\exp\!\le[\y{j}{\delta}\ri]}\, ,
\ee
and $q\!\le(i|x_\delta\ri)$ is similarly a discrete distribution over the components $i$ of the network's output
\be\label{eq:softmax-output}
q\!\le(i|x_\delta\ri)\equiv \frac{\exp\!\le[z_{i;\delta}^{(L)}(t) \ri]}{\sum_{j=1}^{n_\text{out}}\exp\!\le[z_{i;\delta}^{(L)}(t)\ri]}\, .
\ee
As we mentioned when discussing the categorical hypothesis\index{hypothesis (Bayesian inference)!categorical} in the context of Bayesian model fitting\index{Bayesian inference!model fitting} in \S\ref{subsec:ForIO}, the discrete distribution used for \eqref{eq:softmax-target} and \eqref{eq:softmax-output} is sometimes referred to as the \emph{softmax}\index{softmax distribution}~\eqref{eq:softmax}.
The cross-entropy loss~\eqref{eq:loss-cross-entropy}\index{loss!cross-entropy} is a natural measure of the closeness of discrete distributions such as \eqref{eq:softmax-target} and \eqref{eq:softmax-output}.\footnote{\label{foot:KL-in-KL-chapter}The proper measure of closeness of distributions is really the \emph{Kullback–Leibler (KL) divergence}\index{Kullback–Leibler divergence} \eqref{eq:KL-divergence-def}, which we will describe in detail in Appendix~\ref{app:mi-stuff}. However, the KL divergence $KL\le[ p \, ||\, q \ri]$ and the cross-entropy loss\index{loss!cross-entropy}~\eqref{eq:loss-cross-entropy} only differ by a $z^{(L)}$-independent constant, the entropy $\entropy\!\le[p\!\le(i|x_\delta\ri) \ri] = -\sum_{\delta \in \D} \sum_{i=1}^{n_\text{out}} p\!\le(i|x_\delta\ri) \log\!\le[p\!\le(i|x_\delta\ri)\ri]$ to be exact, and thus the use of one versus the other is identical under any gradient-based learning algorithm. Note also the lack of exchange symmetry in either loss between $p$ and $q$. The choice in \eqref{eq:loss-cross-entropy} is purposeful and reflects the fact that an untrained model is on a different footing than the true distribution from which observations arise, analogous to the asymmetry between the prior and posterior in Bayesian inference.\label{footnote:KL}}

\index{regression!linear|see{linear regression}}\index{regression!nearly-linear|see{quadratic regression}}\index{regression!polynomial|see{polynomial regression}}
In particular, cross-entropy loss is the appropriate choice for \neo{classification}, when we want to sort the input $x$ into one of $n_\text{out}$ different classes or categories. Accordingly, the softmax distribution \eqref{eq:softmax-output} transforms the model's output vector with $n_\text{out}$ real components into a discrete probability distribution.
In contrast, the MSE loss\index{loss!comparison of MSE and cross-entropy} is the appropriate choice for \neo{regression}, when the function we want to learn is a vector of real numbers.\footnote{We can think of each loss as descending from a different Bayesian hypothesis, cf.~our discussion of the uncertain hypothesis and the MSE loss \eqref{eq:noisy-conditional-Bayes} \index{hypothesis (Bayesian inference)!uncertain} and the categorical hypothesis and the softmax distribution \eqref{eq:softmax} \index{hypothesis (Bayesian inference)!categorical} in the context of Bayesian model fitting in \S\ref{subsec:ForIO}.} Importantly, when the initialization hyperparameters are tuned to criticality and the training hyperparameters are selected according to the learning rate equivalence principle\index{equivalence principle}, \emph{both losses will be completely well behaved during training}.

When using the cross-entropy loss, typically the true outputs for the training set are given in terms of the softmax values $p\!\le(i|x_{\tra}\ri)$ rather than in terms of continuous vectors $\y{i}{\tra}$. Even more typically, the values $p\!\le(i|x_{\tra}\ri)$ specify a particular \neo{label}, $i=i^\star_{\tra}$, with \neo{absolute certainty}, $p\!\le(i^\star_{\tra}|x_{\tra}\ri) = 1$, while the rest of the components vanish, $p\!\le(i|x_{\tra}\ri) = 0$ for $i\neq i^\star_{\tra}$; this is known  as \emph{hard labeling}\index{label!hard}\index{label!hard|seealso{one-hot encoding}} or \term{one-hot encoding}, and puts the true value of the network output $\y{i^\star_{\tra}}{\tra}$ at infinity. In such case, no finite amount of training will actually reach the minimum, 
and in practice as you approach such a minimum the generalization of the network becomes worse and worse. To remedy this, \textbf{early stopping}\index{regularization!early stopping}\index{early stopping|see{regularization}} of the training algorithm is used as a regularization technique to effectively get finite targets $\y{i^\star_{\tra}}{\tra}$.\footnote{Alternatively we can also explicitly pick a target distribution over the output classes with multiple nonzero components $p\!\le(i|x_{\tra}\ri)$, which is known as \emph{soft labeling}\index{label!soft}. This can be implemented as a \terminate{regularization} technique called \emph{label smoothing}\index{regularization!label smoothing}\index{label smoothing|see{regularization}}, where $p\!\le(i^\star_{\tra}|x_{\tra}\ri) = 1-\epsilon$ and $p\!\le(i\ne i^\star_{\tra}|x_{\tra}\ri) = \epsilon/(n_{\text{out}}-1)$, or as  \neo{knowledge distillation}, mentioned in footnote~\ref{foot:distillation} of \S\ref{ch:bayesian-inference}, when you actually want to learn such a distribution over output classes.}

Now let's specialize to the current context of training neural networks in the infinite-width limit (and assume some kind of regularization is used as described above). In the last section, we noted that any loss that's minimized by setting the network outputs to the true outputs for the training set, $z^{(L)}(x_{\tra};\theta)=\y{i}{\tra}$, is described at the minimum by the Newton's method giant-leap solution~\eqref{eq:loss-minimization-by-Newton}. It's easy to check that the cross-entropy loss \eqref{eq:loss-cross-entropy} is minimized when $ q\!\le(i|x_\delta\ri)  = p\!\le(i|x_\delta\ri)$, and a quick inspection of \eqref{eq:softmax-target} and \eqref{eq:softmax-output} shows that this is obtained by the condition $z^{(L)}(x_{\tra};\theta)=\y{i}{\tra}$.

We do need to make one additional important remark for the cross-entropy loss\index{loss!cross-entropy}. Since in this setting we specify the true output in terms of a softmax\index{softmax distribution}  $p\!\le(i|x_{\tra}\ri)$ rather than an $n_L$-component vector of real numbers $\y{i}{\tra}$, there is an ambiguity in how to set the network output $\z{i}{\tra}{L}$: any component-independent shift $\y{i}{\tra}\to\y{i}{\tra}+c_{\tra}$ keeps the target distribution $p\!\le(i|x_\delta\ri)$ invariant. However, since in this case we care not about the network outputs $\z{i}{\delta}{L}$, but rather their softmax $q\!\le(i|x_\delta\ri)$ \eqref{eq:softmax-output}, this ambiguity doesn't matter in the end. In particular, a shift $\y{i}{\tra}\to\y{i}{\tra}+c_{\tra}$ in the giant-leap solution~\eqref{eq:newton-at-last} shifts all the output components by the same amount for each input $x_\delta$, $\sum_{\tra_1, \tra_2 \in\A} \Ti{\NTKI}{\delta \tra_1}{L} \TI{\NTKIsub}{\tra_1 \tra_2}{L}c_{\tra_2}$, leading to the same softmax $q\!\le(i|x_\delta\ri)$. Thus, we see explicitly that our solution~\eqref{eq:newton-at-last} unambiguously describes networks fully-trained according to the cross-entropy loss\index{loss!cross-entropy}.

\subsection{Kernel Prediction}\label{subsec:NTKprediction}\index{kernel methods!prediction}
After an intelligence -- artificial or otherwise -- undergoes an intense memorization session, often that intelligence is then subjected to \emph{test} with unseen problems in order to probe its actual understanding. In the context of \terminate{machine learning}, we typically evaluate our model's understanding by asking it to make predictions on novel inputs $x_{\tea}$ from the \neo{test set} $\tea \in \B$. 

In the infinite-width limit, the predictions of a fully-trained MLP are governed by the stochastic equation
\be\label{eq:kernel-prediction}
\z{i}{\tea}{L}(T)=\z{i}{\tea}{L}- \sum_{\tra_1, \tra_2 \in\A}  \Ti{\NTKI}{\tea \tra_1}{L} \TI{\NTKIsub}{\tra_1 \tra_2}{L}\le(\z{i}{\tra_2}{L}-\y{i}{\tra_2}\ri)\, ,
\ee
whether we train the model in one step \eqref{eq:network-step-general} or in many steps \eqref{eq:newton-at-last}, and regardless of the choice of loss function or any other details of the \terminate{learning algorithm}.\footnote{
    Note that our analyses of \S\ref{sec:small-step} apply just as much to a small step as they do to a giant leap: the fully-trained infinite-width network has neither wiring in the vectorial components of the network output~(\S\ref{subsec:GD_no_wiring_at_infinity}) nor \terminate{representation learning}~(\S\ref{subsec:GD_no_RL_at_infinity}).
}
For clarity, note that the inverse frozen NTK $\TI{\NTKIsub}{\tra_1 \tra_2}{L}$ is taken with respect to the $\NR$-by-$\NR$ training-set\index{training set} submatrix only, while the frozen NTK $\Ti{\NTKI}{\tea \tra_1}{L}$ is an \emph{off-diagonal} block of the full frozen NTK, connecting an element of the training set to an element of the test set. Note also that the network outputs $\z{i}{\delta}{L}$ on the right-hand side of the equation are evaluated at initialization: once again, 
observables without any step argument should be assumed to be evaluated at initialization, while observables with a step argument $T$ should be assumed to be evaluated at the end of training.

The stochastic equation \eqref{eq:kernel-prediction} describes the predictions of a particular instantiation of a fully-trained neural network. The stochasticity arises from the fact that the prediction \eqref{eq:kernel-prediction} depends on the network outputs at initialization $\z{i}{\delta}{L}$, which themselves depend on the particular realization of the initialized parameters $\theta_{\text{init}} \equiv \theta(t=0)$.
Since we already know that such a network has completely memorized the training set so that $\z{i}{\tra}{L}(T) = \y{i}{\tra}$, the stochasticity here means that any given network in the ensemble can potentially make different predictions on elements of the test set.

With that in mind, let us now compute the full distribution over such test-set predictions for our entire ensemble of fully-trained networks. Inspecting \eqref{eq:kernel-prediction}, we see that the prediction $\z{i}{\tea}{L}(T)$ is a simple linear transformations of the Gaussian-distributed initial outputs $\z{i}{\delta}{L}$ and thus will itself be Gaussian. The mean prediction is simply given by
\be\label{eq:GD-frozen-mean}
\GDGPmean_{i;\tea} \equiv \E{\z{i}{\tea}{L}(T) } = \sum_{\tra_1, \tra_2 \in\A}  \Ti{\NTKI}{\tea \tra_1}{L} \TI{\NTKIsub}{\tra_1 \tra_2}{L}  \y{i}{\tra_2}  \, .
\ee
This expression is entirely analogous to the infinite-width posterior mean\index{posterior!posterior mean} prediction for exact Bayesian inference~\eqref{eq:GP-mean}, with a simple replacement of all types of frozen neural tangent kernels with kernels: $\NTKI^{(L)}\to \ker^{(L)}$. (More on this soon.) Meanwhile, the covariance of the prediction~\eqref{eq:kernel-prediction} is given by
\begin{align}
&\cov{\z{i_1}{\tea_1}{L}(T) }{ \z{i_2}{\tea_2}{L}(T) } \equiv \E{\z{i_1}{\tea_1}{L}(T) \, \z{i_2}{\tea_2}{L}(T)} - \GDGPmean_{i_1;\tea_1}\GDGPmean_{i_2;\tea_2} \notag \\
=& \delta_{i_1 i_2} \Bigg[\Ti{\ker}{\tea_1\tea_2}{L} - \sum_{\tra_1, \tra_2 \in\A}  \Ti{\NTKI}{\tea_2 \tra_1}{L} \TI{\NTKIsub}{\tra_1 \tra_2}{L}  \Ti{\ker}{\tea_1\tra_2}{L} - \sum_{\tra_1, \tra_2 \in\A}  \Ti{\NTKI}{\tea_1 \tra_1}{L} \TI{\NTKIsub}{\tra_1 \tra_2}{L}  \Ti{\ker}{\tea_2\tra_2}{L}  \notag \\
&\qquad+  \sum_{\tra_1, \tra_2, \tra_3, \tra_4 \in\A} \Ti{\NTKI}{\tea_1 \tra_1}{L} \TI{\NTKIsub}{\tra_1 \tra_2}{L} \Ti{\NTKI}{\tea_2 \tra_3}{L} \TI{\NTKIsub}{\tra_3 \tra_4}{L} \Ti{\ker}{\tra_4\tra_2}{L}
\Bigg] \, .
\label{eq:generalized-posterior-variance}
\end{align}
While this expression is somewhat complicated looking, involving both kernels and frozen NTKs, it similarly reduces to the Bayesian infinite-width posterior covariance\index{posterior!posterior covariance}~\eqref{eq:GP-posterior-variance} with the substitution  $\NTKI^{(L)}\to\ker^{(L)}$.\footnote{
    The fact that exact Bayesian inference and gradient descent in general make different predictions is indicative of the fact that -- for general hyperparameter settings -- they are actually different learning algorithms\index{learning algorithm}.
}

This ensemble of network predictions~\eqref{eq:kernel-prediction}, completely specified by its mean~\eqref{eq:GD-frozen-mean} and covariance~\eqref{eq:generalized-posterior-variance}, defines a kind of \textbf{generalized posterior distribution}\index{posterior!generalized posterior distribution|textbf}. This distribution comprises a complete closed-form solution for our ensemble of infinite-width networks at the \terminate{end of training}, regardless of the path that we take to get there.
The mean of the distribution is the prediction of the network \emph{averaged} over instantiations, while the covariance quantifies the instantiation-to-instantiation fluctuations of these predictions.

\index{Bayesian inference!model fitting}\index{maximum likelihood estimation!gradient descent approximation}
Indeed, it is sensible to identify the ensemble of trained networks as a kind of \terminate{posterior} distribution, if you recall  %
our discussion of approximation methods for Bayesian inference in
\S\ref{subsec:ForIO}: minimizing a training loss $\L_\A(\theta)$ gives the maximum likelihood estimation (MLE) of the model parameters~\eqref{eq:mle-estimate-loss}, which we now identify with our fully-trained solution~\eqref{eq:loss-minimization-by-Newton} $\theta_\text{MLE}^\star = \theta_\text{Newton}^\star$. Further recalling the content of footnote~\ref{foot:foretelling-GD} in \S\ref{subsec:ForIO}, for wide networks the minimum of the loss is not unique, 
and the MLE approach will give a family of minima parameterized by the initialization: $\theta^\star_{\text{MLE}}(\theta_{\text{init}})$.\footnote{
    As per that same footnote, we could also try to analyze the MAP\index{maximum a posteriori!gradient descent approximation} estimate~\eqref{eq:map-estimate-loss} in the context of infinite-width gradient-based learning with the addition of a \terminate{regularization} term of the form $\sum_{\mu=1}^P a_{\mu}\theta_\mu^2$ to the loss. If you start this analysis, you'll immediately find that the gradient-descent update $\dtheta_\mu$ includes an additional term $-2 \eta \sum_\nu \lambda_{\mu\nu}\, a_\nu \theta_\nu$, and after some reflection you'll likely also realize the need to define a new stochastic tensor,
    \be
    \Tia{\reg}{i}{\delta}{\ell} \equiv \sum_{\mu, \nu} \lambda_{\mu \nu}\, a_\mu \theta_\mu \frac{\td \z{i}{\delta}{\ell} }{\td \theta_\nu}  \, ,
    \ee
    which has a stochastic iteration given by
    \be
    \Tia{\reg}{i}{\delta}{\ell+1} = a_b^{(\ell+1)} \Lb{\ell+1} \bias{i}{\ell+1} + a_W^{(\ell+1)} \frac{\LW{\ell+1}}{n_\ell} \sum_{j=1}^{n_\ell} \W{ij}{\ell+1} \s{j}{\delta}{\ell} + \sum_{j=1}^{n_\ell} \W{ij}{\ell+1} \ds{j}{\delta}{\ell} \Tia{\reg}{j}{\delta}{\ell} \, ,
    \ee 
    and whose cross-correlation with the preactivations you'll want to compute. Here, you will have defined separate layer-dependent bias and weight regularizations for the coefficients $a_{\mu}$ analogous to what we did for the learning-rate tensor in \eqref{eq:diag_LR}, and you may want to work out the interplay between these \terminate{regularization hyperpameters}\index{hyperparameters!regularization|see{regularization hyperpameters}} and \terminate{initialization hyperparameters} for extra credit.
    \label{footnote:regularization-recursion}
}
This lack of uniqueness ultimately stems from the lingering dependence of the trained network prediction $\z{i}{\tea}{L}(T)$ on the initial function output $\z{i}{\delta}{L}$, which stochastically varies from instantiation to instantiation, cf.~our prediction~\eqref{eq:kernel-prediction}. Considering the ensemble over instantiations of $\theta_{\text{init}}$, we now see exactly how this generalized distribution with the mean \eqref{eq:GD-frozen-mean} and covariance \eqref{eq:generalized-posterior-variance} depends on the \terminate{training hyperparameters} $\Lb{\ell}$ and $\LW{\ell}$, \terminate{initialization hyperparameters} $\Cb{\ell}$ and $\CW{\ell}$, and the training data $\le(x_{\tra}, y_{\tra}\ri)_{\tra\in\A}$.

To be a little pedantic for a paragraph, the covariance in the generalized posterior distribution~\eqref{eq:generalized-posterior-variance} really has a different interpretation than the posterior covariance\index{posterior!posterior covariance}~\eqref{eq:GP-posterior-variance} we computed for exact \terminate{Bayesian inference} at infinite width. In the setting of gradient-based learning, the covariance of the output encodes the variation in the predictions among networks in the ensemble, each corresponding to different parameter settings that still minimize the training loss $\L_\A(\theta)$. In the setting of exact \terminate{Bayesian inference}, 
the covariance encodes our intrinsic uncertainty about unseen data and a small uncertainty can serve as a measure of confidence in our prediction. Thus, these covariances arise for different reasons and are epistemologically quite different in nature.

However, if we can be somewhat pragmatic for a sentence, when you have multiple trained models it's not entirely unreasonable to try to think of this generalized posterior covariance~\eqref{eq:generalized-posterior-variance} as a measure of confidence
as well.

\subsubsection{That One Place Where Gradient Descent = Exact Bayesian Inference}
\index{gradient descent!as Bayesian inference!at infinite width}\index{Bayesian inference!via gradient descent!at infinite width}

Just before, we casually noticed that if we replaced all frozen \emph{neural tangent kernels}\index{frozen NTK} with \emph{kernels}, $\NTKI^{(L)}\to \ker^{(L)}$, then the generalized posterior distribution\index{posterior!generalized posterior distribution} based on~\eqref{eq:kernel-prediction} reduces to the exact Bayesian posterior distribution\index{posterior}~\eqref{eq:posterior-at-infinite-width}. Let us now show how we can actually implement such a substitution in the context of gradient-based learning with a particular choice of \terminate{training hyperparameters}.

To see how to do this, let's put side-by-side the recursion  that defines the output-layer kernel~\eqref{eq:K-recursion-reprint} and the recursion that defines the output-layer frozen NTK\index{frozen NTK}~\eqref{eq:frozen-NTK-recursion}:
\begin{align}\label{eq:kernel-recursion-reminder-reprint}
\Ti{\ker}{\delta_1\delta_2}{L} &= \Cb{L} + \CW{L} \braket{\sigma_{\delta_1}\sigma_{\delta_2}}{L-1} \, , \\
\Ti{\NTKI}{\delta_1\delta_2}{L}&=\Lb{L} + \lamW{L}\bra\sigma_{\delta_1}\sigma_{\delta_2}\ket_{\ker^{(L-1)}}+\CW{L}\bra\sigma^{\prime}_{\delta_1}\sigma^{\prime}_{\delta_2}\ket_{\ker^{(L-1)}}\Ti{\NTKI}{\delta_1\delta_2}{L-1}\, .
\label{eq:NTK-recursion-reminder-reprint}
\end{align}
By inspection, it's immediately clear that setting the final-layer learning rates as
\be\label{eq:last-layer-bayes-learning-rate}
\Lb{L}=\Cb{L}, \qquad \LW{L}=\CW{L}\, ,
\ee
gives us what we want, almost:
\be
\Ti{\NTKI}{\delta_1\delta_2}{L}=\Ti{\ker}{\delta_1\delta_2}{L}+\CW{L}\bra\sigma^{\prime}_{\delta_1}\sigma^{\prime}_{\delta_2}\ket_{\ker^{(L-1)}}\Ti{\NTKI}{\delta_1\delta_2}{L-1}\, .
\ee
To get rid of that pesky last term, we need a way to make the penultimate-layer frozen NTK $\Ti{\NTKI}{\delta_1\delta_2}{L-1}$vanish. To ensure this,  we can simply set all the other \terminate{training hyperparameters} to zero:
\be\label{eq:hidden-layer-bayes-learning-rate}
\Lb{\ell}=0\,, \qquad \LW{\ell}=0\,,\qquad \text{for}\quad \ell <L\, .
\ee
Combined with the initial condition for the NTK recursion \eqref{eq:NTHinitial}, this ensures that $\Ti{\NTKI}{\delta_1\delta_2}{\ell} = 0$ for $\ell <L$, including $\ell=L-1$. Hence, this particular configuration of training hyperparameters sets 
\be
\Ti{\NTKI}{\delta_1\delta_2}{L}=\Ti{\ker}{\delta_1\delta_2}{L} \, ,
\ee
giving us what we wanted, exactly.

In words, this choice of the \terminate{training hyperparameters} \eqref{eq:last-layer-bayes-learning-rate} and \eqref{eq:hidden-layer-bayes-learning-rate} means that we are training only the biases and weights in the last layer.
This establishes that, in the infinite-width limit, exact Bayesian inference is actually a very special case of an ensemble of networks trained with gradient-based learning.

In practice, this means that one could train an ensemble of networks with gradient descent using the \terminate{training hyperparameters} choices \eqref{eq:last-layer-bayes-learning-rate} and \eqref{eq:hidden-layer-bayes-learning-rate} in order to implement a very good approximation of exact \terminate{Bayesian inference}. (The approximation would become exact if you had an infinite number of networks in your ensemble.) On the one hand,
unlike the exact version of Bayesian inference presented in \S\ref{sec:infinite-posterior},
in this case we no longer need to explicitly store or invert the kernel. On the other hand, we may need to fully train a large number of very wide networks for our ensemble to give a good approximation, which may again be expensive in terms of computation and memory.

Interestingly, by explicitly turning off the learning in the hidden layers, 
we are significantly changing the features used to compute our predictions.
In particular, in this case the NTK only has contributions from the biases and weights in the final layer.
Intuitively, what's happening here is that we're taking random features\index{feature} in the penultimate layer $\sigma\!\le(z^{(L-1)}_i\ri)$ and then explicitly training the biases $b^{(L)}$ and weights $W^{(L)}$ to fit the best possible \neo{linear model} of these random features.\footnote{We'll explain in \S\ref{sec:lazy-kernel} that the general version of gradient-based learning in the infinite-width limit is also a \neo{linear model} of random features, but is constructed from a larger set of such features encompassing all the hidden layers.}

\section{Generalization}\label{sec:generalization-at-infinity}\index{generalization}
As remarked before, ultimately we care about how well a model performs on a previously unseen \terminate{test set} $\B$ as compared to the \terminate{training set} $\A$. Some fully-trained networks will \emph{generalize} to these new examples better than others, depending strongly on their initialization\index{initialization hyperparameters} and \terminate{training hyperparameters}. 

To assess this, we can compute the \term{generalization error}: 
\be\label{eq:generalization-error}
\gen \equiv \L_\B - \L_\A\, .
\ee
The generalization error is a quantitative measure of how well a network is really approximating the desired function.\footnote{Without loss of generality\index{loss!of generality}, in the following discussion  we'll implicitly assume that the minimal value of the loss is zero.}
Specifically, if the training loss\index{loss!training loss!relation to overfitting} is small but the test loss\index{loss!test loss!relation to generalization} is large such that there's significant generalization error, then the network isn't really a good model of $f(x)$; instead it's just a lookup table of the values of $f(x)$ when $x$ is taken from the training set $\A$. This is known as \term{overfitting}. In contrast,
if the training and test losses are both small such that there's  little generalization error, then we expect that our model is going beyond simple memorization.\footnote{
    If the generalization error $\gen$ is small but the training loss $\L_\A$ is large, then the model is said to be \neo{underfitting}.
    This situation is not really relevant for very wide networks since, as we've already explained, we can fully train them to achieve zero training loss.
    \index{loss!training loss!relation to underfitting}
}
As such, the generalization error is often considered to be the main quantitative measure of success of a machine learning model.

In the infinite-width limit, we saw in the last section that we can easily set the training loss to zero $\L_\A = 0$ for any particular network. Thus, in the current context the generalization error is completely assessed by the test loss, 
\be
\gen = \L_\B \, ,
\ee
and our current goal is to understand the statistics of the test error in order to characterize how infinite-width networks generalize.

For wide networks we know that there are many configurations of the model parameters that will minimize the training loss and memorize the training data. Some of these configurations might generalize well leading to a small test loss, while some might overfit leading to a large test loss. Thus, the statistics of the generalization error are determined by the statistics of these configurations, which are in turn determined by our initialization and training hyperparameters. 

The mean generalization error captures the \terminate{generalization} properties of the ensemble of networks, and its variance tells us how the instantiation-to-instantiation fluctuations lead some particular networks to generalize better than others.  Understanding these statistics will inform how we should pick our hyperparameters to achieve the best generalization performance on average as well as to ensure that the typical behavior of any fully-trained network is likely to be close to the average.\footnote{
In some applications of \terminate{machine learning}, unseen examples are further divided into two types of datasets, \emph{(i)} a \neo{validation set}, used generally for model selection and often specifically for tuning hyperparameters,
and \emph{(ii)} a \neo{test set}, used to assess the generalization properties of a particular trained model. Despite our liberal usage of the term \neo{test set}, here, as we analytically compute the statistics of the loss on unseen examples and then use them to tune the hyperparameters, what we have is really closer in meaning to a \neo{validation set}, as we are tuning an ensemble of models rather than assessing any particular one.
}

With that in mind, let's first evaluate the MSE test loss\index{loss!MSE}, averaged over an ensemble of fully-trained networks:
\begin{align}\label{eq:bias-variance-decomposition-generalized-mse}
\E{\L_{\B}(T)}=&\E{\frac{1}{2}\sum_{i=1}^{n_L}\sum_{\tea\in\B}\le(\z{i}{\tea}{L}(T)-\y{i}{\tea}\ri)^2}\, \notag\\
=&\E{\frac{1}{2}\sum_{i=1}^{n_L}\sum_{\tea\in\B}\le(\z{i}{\tea}{L}(T)-\GDGPmean_{i;\tea}+\GDGPmean_{i;\tea}-\y{i}{\tea}\ri)^2}\, \notag\\
=&\frac{1}{2}\sum_{\tea\in\B}\le\{\sum_{i=1}^{n_L}\le(\GDGPmean_{i;\tea}-\y{i}{\tea}\ri)^2+\sum_{i=1}^{n_L}\cov{\z{i}{\tea}{L}(T) }{ \z{i}{\tea}{L}(T) }\ri\}\, . 
\end{align}
In the second line, we added and subtracted the infinite-width mean prediction~\eqref{eq:GD-frozen-mean}, and to get to the third line we noted that the cross terms cancel under the expectation.

This decomposition~\eqref{eq:bias-variance-decomposition-generalized-mse} illustrates a type of generalized \term{bias-variance tradeoff}: 
the first term, the \emph{bias}\index{generalization error!bias}, measures the deviation of the mean prediction of the ensemble $\GDGPmean_{i;\tea}$ from the true output $\y{i}{\tea}$; the second term, the \emph{variance}\index{generalization error!variance} -- or specifically the covariance of the generalized posterior distribution\index{posterior!generalized posterior distribution} \eqref{eq:generalized-posterior-variance} -- measures the instantiation-to-instantiation fluctuations of that prediction across different models in our ensemble. The reason why this is called a \emph{tradeoff} is that typically different settings of the hyperparameters
will decrease one term at the cost of increasing the other, making the modeler have to choose between improving one at the cost of the other.\footnote{The reason why we call it \emph{generalized} bias-variance tradeoff\index{bias-variance tradeoff!generalized} is that, in the \emph{standard} bias-variance tradeoff\index{bias-variance tradeoff!generalized!vs.~standard}, the expectation is over different realizations of the training set $\A$ rather than over different initializations of the model parameters $\theta_\mu$. In that typical setting we have only a single model, and the \emph{bias}\index{generalization error!bias} characterizes how well that model can be trained on each different training set -- with a large bias indicative of \neo{underfitting} -- while the \emph{variance}\index{generalization error!variance} characterizes the fluctuations of that model's performance over the different training sets -- with a large variance indicative of \neo{overfitting}. %
}

For more general losses, we can Taylor expand the test loss\index{loss!test loss} around the mean prediction as
\begin{align}
\L_{\B}=&\L_{\B}\!\le(\GDGPmean\ri)+\sum_{i,\tea}\frac{\partial\L_{\B}}{\partial \z{i}{\tea}{L}}\Bigg\vert_{z^{(L)}=\GDGPmean}\!\!\!\!\!\!\!\!\le(\z{i}{\tea}{L}(T)-\GDGPmean_{i;\tea}\ri)\, \\
&+\frac{1}{2}\sum_{i_1,i_2,\tea_1,\tea_2}\frac{\partial^2\L_{\B}}{\partial \z{i_1}{\tea_1}{L}\partial \z{i_2}{\tea_2}{L}}\Bigg\vert_{z^{(L)}=\GDGPmean}\!\!\!\!\!\!\!\!\le(\z{i_1}{\tea_1}{L}(T)-\GDGPmean_{i_1;\tea_1}\ri)\le(\z{i_2}{\tea_2}{L}(T)-\GDGPmean_{i_2;\tea_2}\ri)+\ldots\, ,\notag
\end{align}
where we've denoted the test loss evaluated at the mean prediction as
\be\label{eq:general-loss-bias}
\L_{\B}\!\le(\GDGPmean\ri) \equiv \L_{\B}\!\le(z^{(L)} =\GDGPmean\ri)\, .
\ee 
Performing the expectation over the ensemble and noting that $\E{\z{i}{\tea}{L}(T)-\GDGPmean_{i;\tea}}=0$ by definition~\eqref{eq:GD-frozen-mean}, we get
\be\label{eq:bias-variance-decomposition-generalized-general-loss}
\E{\L_{\B}}=\L_{\B}\!\le(\GDGPmean\ri)+\frac{1}{2}\sum_{i_1,i_2,\tea_1,\tea_2}\frac{\partial^2\L_{\B}}{\partial \z{i_1}{\tea_1}{L}\partial \z{i_2}{\tea_2}{L}}\Bigg\vert_{z^{(L)}=\GDGPmean}\!\!\!\!\!\!\!\!\cov{\z{i_1}{\tea_1}{L}(T) }{ \z{i_2}{\tea_2}{L}(T) }+\ldots\, .
\ee
Here again we find a generalized \terminate{bias-variance decomposition}\index{bias-variance decomposition|seealso{generalization error}}: the first term $\L_{\B}\!\le(\GDGPmean\ri)$ is the bias\index{generalization error!bias}, measuring the deviation of the mean prediction from the true output on the \terminate{test set}, and the second term is the variance\index{generalization error!variance}, measuring the instantiation-to-instantiation uncertainty as the trace of the covariance multiplied by the \terminate{Hessian} of the loss with respect to the network outputs. Thus, for \emph{any} choice of loss function, these bias and variance terms will give a good proxy for the generalization error $\gen$ so long as our models are making predictions that are close to the mean prediction.

Now, let's see how to compute these bias and variance terms in a few different setups.
In most common cases in practice
the loss is \emph{extensive}\index{extensivity!of loss} or additive in samples, i.e.~$\L_{\B}=\sum_{\tea\in\B}\L_{\tea}$, and we can consider the test loss evaluated on one test sample at a time.
Thus, for the purpose of our analysis here, the question is: for a given test input, how many training examples are relevant for making a prediction?

In \S\ref{subsec:robustness-from-infinite-GD}, we'll compute the bias and variance terms of the generalization error \eqref{eq:bias-variance-decomposition-generalized-general-loss} around one training sample using our $\delta$ expansion introduced in \S\ref{ch:signalprop}, giving another lens into hyperparameter tuning and \terminate{criticality} for our two universality classes. However, this view will be somewhat limited by the restriction of our training set to one sample.

In \S\ref{subsec:star-polation}, we'll enlarge our training set to include two samples. Rather than computing the \terminate{generalization error} itself, here we'll be able to explore directly 
a different aspect of generalization: how a fully-trained network either interpolates or extrapolates to make predictions. %

\subsection{Bias-Variance Tradeoff and Criticality}\label{subsec:robustness-from-infinite-GD}
Let us index one training sample by $\tra=+$ and a nearby test sample by $\tea=-$; let us also focus on the output layer $\ell=L$ and temporarily drop the layer index from the frozen NTK\index{frozen NTK}, $\NTKI_{\delta_1\delta_2}\equiv\Ti{\NTKI}{\delta_1\delta_2}{L}$, until later when we need to discuss the depth dependence of various NTK components.

The bias term in the generalization error\index{generalization error!bias} is determined by the deviation of the mean prediction \eqref{eq:GD-frozen-mean} from the true output:
\begin{align}\label{eq:bias-factor-gen-error-not-squared}
\GDGPmean_{i;-}-\y{i}{-}&=\frac{\NTKI_{-+}}{\NTKI_{++}}\y{i}{+}-\y{i}{-} \, \notag \\
&=(\y{i}{+}-\y{i}{-}) + \le(\frac{\NTKI_{-+}}{\NTKI_{++}}-1\ri)\y{i}{+}\, .
\end{align}
In this expression, the first term is the true difference in the function outputs on the two different inputs, $f(x_{+})-f(x_-)$, while the second term is a similar (expected) difference between our predicted output on $x_-$ and the learned true output on $x_+$, $z_-(T) - z_+(T)$.\footnote{
    In general, the bias term $\L_{\B}\!\le(\GDGPmean\ri)$ in the generalization error\index{generalization error!bias}~\eqref{eq:bias-variance-decomposition-generalized-general-loss} depends on the details of the loss. Of course, we can expand $\L_{\B}\!\le(\GDGPmean\ri)$ around the true output $\y{i}{-}$, and the expansion will depend on the difference $\GDGPmean_{i;-}-\y{i}{-}$~\eqref{eq:bias-factor-gen-error-not-squared}. For the MSE loss\index{loss!MSE}, the bias is precisely the square of this difference. For the cross-entropy loss\index{loss!cross-entropy}, it is more natural to expand in terms of the difference $\overline{q}(i\vert x_{-})-p(i\vert x_{-})$, with $\overline{q}\!\le(i|x_\delta\ri)\equiv \exp\!\le[\GDGPmean_{i;\delta}\ri]/\sum_{j=1}^{n_\text{out}}\exp\!\le[\GDGPmean_{j;\delta}\ri]$.
} Note the opposite ordering of $+$ and $-$ in these two terms: if our prediction is exactly correct, these two terms are equal in magnitude and opposite in sign, and the bias term in the generalization error will vanish.\index{generalization error!bias}

With that in mind, the quantity in parenthesis in the second factor of the bias term~\eqref{eq:bias-factor-gen-error-not-squared},
\be\label{eq:robustness-measure}
\frac{\NTKI_{-+}}{\NTKI_{++}}-1\, ,
\ee 
serves as a natural measure of \emph{robustness}\index{generalization error!robustness measure} since it characterizes how sensitively our prediction changes -- i.e.~how $\vert z_-(T) - z_+(T)\vert$ grows -- with corresponding small changes in the input. A model that isn't robust will often be incorrect, making predictions that vary greatly from the network output on nearby training points,
while too robust a model will not have much flexibility in its output. Since a priori we don't know what type of function we are going to approximate, we naturally would want to pick hyperparameters that include a class of networks that are robust, but not overly inflexible.\footnote{A dedicated reader might notice the parallel with \S\ref{subsec:Occam-criticality} where we argued for \terminate{criticality} by considering the evidence for two inputs with differing true outputs, $f(x_{+})-f(x_-)\ne 0$, which called for similar flexibility in choice of function approximators.\index{Bayesian inference!evidence!relation to generalization error}}

Now, since we're considering a test input that's nearby our training input, that should remind you of our $\delta$ expansion from our \terminate{criticality} analysis in \S\ref{sec:bootstrapping}.\footnote{
    The following applies to smooth activation functions and is intended to give the general picture. We will give an analysis particular to nonlinear scale-invariant activation functions later when discussing them in particular.
}
Specifically, we can expand the frozen NTK\index{frozen NTK} in our $\gamma^{[a]}$ basis as we did for the kernel in~\eqref{eq:kernel-expand-gamma},
\be\label{eq:NTK-in-gamma-basis}
\NTKI_{\pm\pm}=\NTKI_{[0]}\pm\NTKI_{[1]}+\NTKI_{[2]}\, ,\qquad \NTKI_{\pm\mp}=\NTKI_{[0]}-\NTKI_{[2]}\, ,
\ee
and make $\delta$ expansions similar to the ones we did for the kernel in \eqref{eq:kernel-expand-1}--\eqref{eq:kernel-expand-3},
\begin{align}\label{eq:frozen-NTK-delta-0}
\NTKI_{[0]}=&\NTKI_{\M\M}+\delta\delta\NTKI_{[0]}+\o{\delta^4}\, ,\\
\NTKI_{[1]}=&\delta\NTKI_{[1]}+\o{\delta^3}\, ,\label{eq:frozen-NTK-delta-1}\\
\NTKI_{[2]}=&\delta\delta\NTKI_{[2]}+\o{\delta^4}\, ,
\label{eq:frozen-NTK-delta-2}
\end{align}
where the expansion is taken around the midpoint frozen NTK\index{frozen NTK!midpoint} $\NTKI_{\M\M}$ evaluated on the midpoint input\index{midpoint input} $\x{i}{0}\equiv (\x{i}{+}+\x{i}{-})/2$.

For simplicity of our presentation, let's now assume that the two inputs have the same norm $\sum_{i=1}^{n_0} \x{i}{+}^2=\sum_{i=1}^{n_0} \x{i}{-}^2$, so that $\ker_{[1]}=0$ and $\NTKI_{[1]}=(\NTKI_{++}-\NTKI_{--})/2=0$.
With this simplification, plugging the decomposition~\eqref{eq:NTK-in-gamma-basis} and then the expansions \eqref{eq:frozen-NTK-delta-0} and \eqref{eq:frozen-NTK-delta-2} into the expression for our robustness measure~\eqref{eq:robustness-measure}\index{generalization error!robustness measure}, we get
\be\label{eq:robustness-measure-plugged-in}
\frac{\NTKI_{-+}}{\NTKI_{++}}-1=\le(\frac{\NTKI_{[0]}-\NTKI_{[2]}}{\NTKI_{[0]}+\NTKI_{[2]}}-1\ri)=-2\frac{\delta\delta\NTKI_{[2]}}{\NTKI_{\M\M}}+\o{\delta^4}\, .
\ee
Thus, we see that the ratio $\delta\delta\NTKI_{[2]}/\NTKI_{\M\M}$ captures the robustness of predictions for nearby test inputs. We'll analyze its depth dependence for two universality classes shortly.\index{generalization error!universality class analysis}

Having covered the bias term in the generalization error\index{generalization error!bias}, let's next consider the variance term\index{generalization error!variance}. The loss-independent piece of the variance
is given by the covariance of the generalized posterior distribution\index{posterior!generalized posterior distribution}~\eqref{eq:generalized-posterior-variance}.
Evaluating \eqref{eq:generalized-posterior-variance} for a single training sample $\tra=+$, using our decompositions for the kernel~\eqref{eq:kernel-expand-gamma} and frozen NTK~\eqref{eq:NTK-in-gamma-basis}, and then using expansions~\eqref{eq:kernel-expand-1},~\eqref{eq:kernel-expand-3},~\eqref{eq:frozen-NTK-delta-0}, and~\eqref{eq:frozen-NTK-delta-2}, we find\index{$\gamma^{[a]}$ basis!kernel}\index{$\delta$ expansion}
\begin{align}\label{eq:generalization-variance-expanded}
&\cov{\z{i}{-}{L}(T) }{ \z{i}{-}{L}(T)}\, \\
=&\ker_{--}-2\frac{\NTKI_{-+}}{\NTKI_{++}}\ker_{-+}+\le(\frac{\NTKI_{-+}}{\NTKI_{++}}\ri)^2\ker_{++}\, \notag\\
=&\ker_{[0]}+\ker_{[2]}-2\le(\frac{\NTKI_{[0]}-\NTKI_{[2]}}{\NTKI_{[0]}+\NTKI_{[2]}}\ri)\le(\ker_{[0]}-\ker_{[2]}\ri)+\le(\frac{\NTKI_{[0]}-\NTKI_{[2]}}{\NTKI_{[0]}+\NTKI_{[2]}}\ri)^2\le(\ker_{[0]}+\ker_{[2]}\ri)\, \notag\\
=&4\delta\delta\ker_{[2]}+\o{\delta^4}\, .\notag
\end{align}
Thus, to leading order the variance term depends only on the perpendicular perturbation of the kernel $\delta\delta\ker_{[2]}$.

At this point, we know everything there is to know about how $\delta\delta\ker_{[2]}$ behaves as a function of depth for our universality classes (cf.~\S\ref{sec:non-scale-invariant-eft} and \S\ref{sec:finite_angle}). On the one hand, we could pick initialization hyperparameters such that $\delta\delta\ker_{[2]}$ grows exponentially with depth. However, with this choice the variance term will grow very quickly, leading to large fluctuations in model predictions between different realizations. On the other hand, we could pick \terminate{initialization hyperparameters} that decay exponentially with depth, leading to a quickly vanishing variance term and very overconfident predictions. However, we will soon see that this overconfidence comes at a cost: an exponentially vanishing perpendicular perturbation $\delta\delta\ker_{[2]}$ implies an exponentially vanishing frozen NTK component $\delta\delta\NTKI_{[2]}$ and thus a vanishing robustness measure~\eqref{eq:robustness-measure-plugged-in} signaling extreme inflexibility. In particular, we will have learned a constant function that's always equal to $y_+$, regardless of the input.

This is precisely the generalized \terminate{bias-variance tradeoff} that we described above: if we try to set the variance to zero by having $\delta\delta\ker_{[2]}$ vanish exponentially, then the vanishing of $\delta\delta\NTKI_{[2]}$ will cause our bias to be larger for generic inputs and consequently the network will not be able to generalize in a nontrivial manner. Vice versa, making the function too flexible with large $\delta\delta\NTKI_{[2]}$ will make the model predictions not only too sensitive to small changes in the input through $\delta\delta\NTKI_{[2]}$, but also will cause large fluctuations in that prediction from realization to realization through $\delta\delta\ker_{[2]}$.

Of course, we know that there's a third option: we could pick our \terminate{criticality} condition $\chi_\perp\!\le(\ker^\star\ri) =1$. This setting of the initialization hyperparameters has the potential to balance the bias-variance tradeoff\index{bias-variance tradeoff!relation to criticality}, leading to the best outcome without a priori knowing anything more about the underling dataset we're trying to model.

What about our other criticality condition $\chi_\parallel\!\le(\ker^\star\ri) =1$? Recall from our discussion of the \terminate{exploding and vanishing gradient problem} in \S\ref{sec:EVGP-WEP} that the \terminate{parallel susceptibility} $\chi_\parallel$ affects the way in which the midpoint frozen NTK $\NTKI_{00}$
receives contributions from different layers. %
As the midpoint frozen NTK $\NTKI_{00}$ enters in the robustness measure as in~\eqref{eq:robustness-measure-plugged-in}, this suggests that it also plays an important role in generalization. In fact, we will see soon in \S\ref{sec:lazy-kernel} that ensuring equal contributions from all layers is another way of saying that we use the greatest set of features available to us in making a prediction. Thus, it stands to reason that also picking the criticality condition $\chi_\parallel\!\le(\ker^\star\ri) =1$ in conjunction with the condition $\chi_\perp\!\le(\ker^\star\ri) =1$ is a natural choice for generalization, in addition to all our other evidence for such a choice.\footnote{Just like in footnote \ref{foot:parallel-criticality} of \S\ref{subsec:Occam-criticality}, additional justification comes from the consideration of  two inputs with unequal norms: $\sum_{i=1}^{n_0} \x{i}{+}^2\ne\sum_{i=1}^{n_0} \x{i}{-}^2$.
In such a case,  the robustness measure~\eqref{eq:robustness-measure-plugged-in} is given by
\be\label{eq:robustness-with-parallel}
\frac{\NTKI_{-+}}{\NTKI_{++}}-1=-\frac{\delta\NTKI_{[1]}}{\NTKI_{\M\M}}-2\frac{\delta\delta\NTKI_{[2]}}{\NTKI_{\M\M}}+\le(\frac{\delta \NTKI_{[1]}}{\NTKI_{\M\M}}\ri)^2+\o{\delta^3}\, ,
\ee
and the covariance is given by
\be
\cov{\z{i}{-}{L}(T) }{ \z{i}{-}{L}(T)}=4\delta\delta\ker_{[2]}-2\delta\ker_{[1]}\frac{\delta \NTKI_{[1]}}{\NTKI_{\M\M}}+\ker_{\M\M}\le(\frac{\delta \NTKI_{[1]}}{\NTKI_{\M\M}}\ri)^2+\o{\delta^3}\, .
\ee
First, we see that the kernel components $\delta\ker_{[1]}$ and $\ker_{\M\M}$ both contribute, necessitating that we set $\chi_{\parallel}=1$ as per our previous discussions. In addition, we will also need to tame the exploding and vanishing problem of $\delta \NTKI_{[1]}$.

For the scale-invariant universality class\index{universality class!scale-invariant}, $\NTKI_{[1]}=(\NTKI_{++}-\NTKI_{--})/2$ has exactly the same depth dependence as the single-input frozen NTK\index{frozen NTK} \eqref{eq:frozen-ntk-critical-solution-relu}. In this case, $\chi_{\parallel}=\chi_{\perp}\equiv \chi$, and all the exponential explosions and vanishments are mitigated by setting $\chi=1$.

For the $\Tif{\ker}{}=0$ universality class, %
we can write a recursion for $\delta\NTKI_{[1]}$ by projecting out the $\gamma^{[1]}$ component of the full frozen NTK recursion \eqref{eq:tired-repetition-of-frozen-yogurt} using \eqref{eq:trace-projection}:
\be\label{eq:deltaNTK-recursion}
\Ti{\delta\NTKI}{[1]}{\ell+1}=\chi_{\perp}^{(\ell)}\Ti{\delta\NTKI}{[1]}{\ell}+\le(\frac{\LW{\ell+1}}{C_W}\chi_{\parallel}^{(\ell)}+\frac{C_W}{\KML}\bra z\sigma'\sigma''\ket_{\KML} \Ti{\NTKI}{\M\M}{\ell}\ri) \Ti{\delta\ker}{[1]}{\ell}\, ,
\ee
i.e.~with a derivation almost isomorphic to the one below for $\delta\delta\NTKI_{[2]}$ \eqref{eq:deltadeltaNTK-recursion}. We in particular see that $\Ti{\delta\ker}{[1]}{\ell}$ contributes to $\Ti{\delta\NTKI}{[1]}{\ell}$ --
which can be thought of as the Bayesian contribution per our last discussion in \S\ref{subsec:NTKprediction} --
and its exploding and vanishing problem is mitigated by setting  $\chi_\parallel\!\le(\ker^\star\ri)=1$: cf.~\eqref{K1}. (At this point you may find it useful to re-read and re-flect on the last paragraph of footnote \ref{foot:parallel-criticality} in \S\ref{subsec:Occam-criticality}.) You can further study the depth dependence of $\Ti{\delta\NTKI}{[1]}{\ell}$ at criticality and find that $\delta\NTKI_{[1]}^{(\ell)}$ decays faster than $\NTKI_{\M\M}^{(\ell)}$ and $\delta\delta\NTKI_{[2]}^{(\ell)}$, thus reducing the problem back to the one studied in the main text.}

Now, returning to the bias part of the generalization error\index{generalization error!bias}, to  complete our analysis we'll need to solve a recursion for the $\delta\delta\NTKI_{[2]}$ component of the frozen NTK recursion~\eqref{eq:frozen-NTK-recursion}, reprinted here in full:
\begin{align}\label{eq:tired-repetition-of-frozen-yogurt} %
\Ti{\NTKI}{\delta_1\delta_2}{\ell+1}&=\Lb{\ell+1} + \lamW{\ell+1}\bra\sigma_{\delta_1}\sigma_{\delta_2}\ket_{\ker^{(\ell)}}+C_W\bra\sigma^{\prime}_{\delta_1}\sigma^{\prime}_{\delta_2}\ket_{\ker^{(\ell)}}\Ti{\NTKI}{\delta_1\delta_2}{\ell}\, .
\end{align}
Let's first work this out for the $K^\star=0$ universality class\index{universality class!K@$K^\star=0$}, and then we'll consider the scale-invariant universality class\index{universality class!scale-invariant} for which we'll need to make use of our finite-angle results from \S\ref{sec:finite_angle}. 
Either way, this should be child's play for us at this point.\footnote{An even more childish play would be studying the Bayesian version of generalization error\index{generalization error!exact Bayesian inference} by setting the \terminate{training hyperparameters} according to \eqref{eq:last-layer-bayes-learning-rate} and \eqref{eq:hidden-layer-bayes-learning-rate}, such that $\Ti{\NTKI}{\delta_1\delta_2}{L}=\Ti{\ker}{\delta_1\delta_2}{L}$. In this case, we know exactly how the bias and variance terms of the generalization error behave. This is a very particular setting of the \terminate{training hyperparameters} and unlikely to be optimal in general (cf.~our discussion of the differences between the frozen NTK and Bayesian kernel in terms of feature functions in \S\ref{sec:lazy-kernel}). Indeed for the scale-invariant universality class, we'll explicitly see around \eqref{eq:weird-asymptotic-behavior} that exact Bayesian inference has inferior asymptotic behavior than the more general gradient-based learning.}

\subsubsection{\texorpdfstring{$\Tif{\ker}{}=0$}{K*=0} Universality Class}\index{universality class!K@$K^\star=0$}
Recall~\eqref{eq:useful-much-later-too} from much much earlier describing the decomposition of the Gaussian expectation of two activations in the $\gamma^{[a]}$ basis.\index{$\gamma^{[a]}$ basis!$\sigma \sigma$} With the parallel perturbation turned off, $\ker_{[1]}^{(\ell)}=0$, this expansion reads 
\begin{align}
\braket{\sigma_{\delta_1} \sigma_{\delta_2}}{\ell}=\!\le[\le\langle \sigma\sigma\ri\rangle_{\KML}+\o{\delta^2}\ri] \gamma^{[0]}_{\delta_1\delta_2}\!+\!\le[\Ti{\delta\delta\ker}{[2]}{\ell}\bra \sigma'\sigma'\ket_{\KML}+\o{\delta^4}\ri] \gamma^{[2]}_{\delta_1\delta_2}\, .
\end{align}
With a replacement $\sigma\to\sigma'$, we have a similar decomposition for the Gaussian expectation of the derivatives of activations:\index{$\gamma^{[a]}$ basis!$\sigma^\prime \sigma^\prime$}
\begin{align}\label{eq:primeprimeprimeprime}
\braket{\sigma'_{\delta_1} \sigma'_{\delta_2}}{\ell}=\!\le[\le\langle \sigma'\sigma'\ri\rangle_{\KML}+\o{\delta^2}\ri] \gamma^{[0]}_{\delta_1\delta_2}\!+\!\le[\Ti{\delta\delta\ker}{[2]}{\ell}\bra \sigma''\sigma''\ket_{\KML}+\o{\delta^4}\ri] \gamma^{[2]}_{\delta_1\delta_2}\, .
\end{align}

\index{$\gamma^{[a]}$ basis!frozen NTK}
Plugging these expansions into the full frozen NTK recursion~\eqref{eq:tired-repetition-of-frozen-yogurt} and 
using the component-wise identities $\gamma^{[0]}_{\delta_1\delta_2}\gamma^{[0]}_{\delta_1\delta_2}=\gamma^{[2]}_{\delta_1\delta_2}\gamma^{[2]}_{\delta_1\delta_2}=\gamma^{[0]}_{\delta_1\delta_2}$ and $\gamma^{[0]}_{\delta_1\delta_2}\gamma^{[2]}_{\delta_1\delta_2}=\gamma^{[2]}_{\delta_1\delta_2}$, we get
\be\label{eq:deltadeltaNTK-recursion}
\Ti{\delta\delta\NTKI}{[2]}{\ell+1}=\chi_{\perp}^{(\ell)}\Ti{\delta\delta\NTKI}{[2]}{\ell}+\le(\frac{\LW{\ell+1}}{C_W}\chi_{\perp}^{(\ell)}+C_W\bra \sigma''\sigma''\ket_{\KML} \Ti{\NTKI}{\M\M}{\ell}\ri) \Ti{\delta\delta\ker}{[2]}{\ell}\, ,
\ee
where we've recalled the definition of the \terminate{perpendicular susceptibility}~\eqref{eq:chi-perp}, $\chi_{\perp}^{(\ell)}=C_W\bra \sigma'\sigma'\ket_{\KML}$.\footnote{More generally there are terms proportional to $\le(\delta\ker_{[1]}\ri)^2$ and $\delta\ker_{[1]} \delta\NTKI_{[1]}$ in this recursion for $\delta\delta\NTKI_{[2]}$; however, when training and test inputs have equal norm, $\delta\ker_{[1]}=0$, these terms vanish. 
}

\index{bias-variance tradeoff!for a universality class!K@$K^\star=0$ activations}
We learned long ago that the perpendicular susceptibility governs the behavior of the perpendicular perturbation~$\Ti{\delta\delta\ker}{[2]}{\ell}$, and we see from~\eqref{eq:deltadeltaNTK-recursion} that it also controls the behavior of the frozen NTK component $\Ti{\delta\delta\NTKI}{[2]}{\ell}$. As we alluded to before, the exponential decay/growth of $\Ti{\delta\delta\ker}{[2]}{\ell}$ and $\Ti{\delta\delta\NTKI}{[2]}{\ell}$ are thusly linked. In particular, trying to eliminate the variance term of the generalization error\index{generalization error!variance} by letting $\Ti{\delta\delta\ker}{[2]}{\ell}$ exponentially decay will also cause $\Ti{\delta\delta\NTKI}{[2]}{\ell}$ to exponentially decay, making the model prediction constant, inflexible, and highly biased.

Given this and our previous discussion on the role of the parallel susceptibility $\chi_{\parallel}$, 
let's now tune to criticality $\chi_{\parallel}\!\le(\ker^\star\ri)=\chi_\perp\!\le(\ker^\star\ri) =1$ and evaluate the depth dependence of $\Ti{\delta\delta\NTKI}{[2]}{\ell}$. For the $K^\star=0$ universality class\index{universality class!K@$K^\star=0$}, \terminate{criticality}~\eqref{eq:k-star-equals-zero-critical-initialization} is found by tuning $C_b=0$ and $C_W=\frac{1}{\sigma_1^2}$. %
With these settings, we recall the large-$\ell$ asymptotic solutions from~\eqref{eq:initial-tanh-wthout-subleading} and~\eqref{eq:perp-asymptotic-solution}
\be\label{eq:perp-asymptotic-solution-reprint}
\KML=\le[\frac{1}{(-a_1)}\ri] \frac{1}{\ell}+\ldots\, , \qquad \delta\delta \ker_{[2]}^{(\ell)}= \frac{\delta^2}{\ell^{p_{\perp}}}+\ldots\,  ,
\ee
where $p_{\perp}\equiv b_1/a_1$, and the activation-function dependent constants $a_1$ and $b_1$ were defined in~\eqref{eq:a1} and~\eqref{eq:b1}, and $\delta^2$ is a constant related to the initial separation of the inputs but isn't fixed by the asymptotic analysis. 
Also recall from the more recent past~\eqref{eq:ntk-k-star-perp-susceptibility-asymptotic} that we can asymptotically expand the perpendicular susceptibility as
\be\label{eq:perp-perp-perp-perp}
\chi_{\perp}^{(\ell)}=1-\frac{p_{\perp}}{\ell}+\ldots\, .
\ee
Similarly, by a simple Gaussian integral we can evaluate the following Gaussian expectation,
\be\label{eq:deri-deri-deri-deri}
\bra \sigma''\sigma''\ket_{\KML}=\sigma_2^2+\o{\KML}=\sigma_2^2+\o{\frac{1}{\ell}}\, ,
\ee
remembering our notation $\sigma_2\equiv \sigma''(0)$.

\index{universality class!K@$K^\star=0$}
Next, we also have to make a choice about the \terminate{training hyperparameters} $\Lb{\ell}$ and $\LW{\ell}$. Indeed, the depth scaling of the \terminate{generalization error} will depend on these hyperparameters, a fact that should not be surprising: we expect the selection of our relative learning rates to affect the performance of our model. Let's first follow the guidance of~\S\ref{sec:EVGP-WEP} where we discussed an \neo{equivalence principle} for learning rates, and set these \terminate{training hyperparameters} according to \eqref{eq:super-tanh-general}, i.e.~\eqref{eq:layer-independent-rates} multiplied by $L^{p_{\perp}-1}$:
\be\label{eq:super-tanh-general-reprint}
\Lb{\ell}=\widetilde{\lambda}_b\le(\frac{1}{\ell}\ri)^{p_{\perp}}L^{p_{\perp}-1}\, , \qquad \lamW{\ell}=\widetilde{\lambda}_W\le(\frac{L}{\ell}\ri)^{p_{\perp}-1}\, .
\ee
With such a choice, we have an asymptotic solution for the midpoint frozen NTK\index{frozen NTK!midpoint}, which is the same solution as in \eqref{eq:frozen-ntk-k-star-solution} up to a multiplication by $L^{p_{\perp}-1}$: 
\be\label{eq:frozen-ntk-k-star-solution-reprint}
\Ti{\NTKI}{\M\M}{\ell}=\le[\widetilde{\lambda}_b+\frac{\widetilde{\lambda}_W\sigma_1^2}{(-a_1)}\ri]\le( \frac{L}{\ell}\ri)^{p_{\perp}-1}+\ldots\, .
\ee
Further plugging these results
\eqref{eq:perp-asymptotic-solution-reprint}--\eqref{eq:frozen-ntk-k-star-solution-reprint}
into \eqref{eq:deltadeltaNTK-recursion}, we get
\begin{align}\label{eq:frozen-ntk-dd-recursion-k-star}
\delta\delta \NTKI_{[2]}^{(\ell+1)}=&\le[1-\frac{p_{\perp}}{\ell}+\ldots\ri]\delta\delta \NTKI_{[2]}^{(\ell)}\, \\
&+ \delta^2\le\{\widetilde{\lambda}_W\sigma_1^2+\frac{\sigma_2^2}{\sigma_1^2}\le[\widetilde{\lambda}_b+\frac{\widetilde{\lambda}_W\sigma_1^2}{(-a_1)}\ri] \ri\}L^{p_{\perp}-1}\le(\frac{1}{\ell}\ri)^{2p_{\perp}-1}+\ldots\, .\notag
\end{align}
With our usual methods, we can solve this recursion in the asymptotically large-$\ell$ limit with
\be\label{eq:k-star-odd-ddTheta-sol}
\delta\delta \NTKI_{[2]}^{(\ell)}=\delta^2\frac{L^{p_{\perp}-1}}{(2-p_{\perp})}\le\{\widetilde{\lambda}_W\sigma_1^2+\frac{\sigma_2^2}{\sigma_1^2}\le[\widetilde{\lambda}_b+\frac{\widetilde{\lambda}_W\sigma_1^2}{(-a_1)}\ri] \ri\} \le(\frac{1}{\ell}\ri)^{2p_{\perp}-2}+\ldots\, .
\ee
Finally, taking the ratio of the midpoint frozen NTK\index{frozen NTK!midpoint} \eqref{eq:frozen-ntk-k-star-solution-reprint} and the perpendicular perturbation \eqref{eq:k-star-odd-ddTheta-sol} and evaluating at the output layer $\ell=L$,
we find the overall network depth dependence for our robustness measure:
\begin{align}\label{eq:robustness-k-star-sol}
\frac{-2\delta\delta \NTKI_{[2]}^{(L)}}{\NTKI_{\M\M}^{(L)}}&=\frac{2\delta^2}{(p_\perp-2)}\le\{
\frac{\widetilde{\lambda}_W\sigma_1^2}{\le[\widetilde{\lambda}_b+(\widetilde{\lambda}_W\sigma_1^2)/(-a_1)\ri]}
+ \frac{\sigma_2^2}{\sigma_1^2} \ri\}L^{1-p_{\perp}}\propto L^{1-p_{\perp}} \, .
\end{align}
This is astonishing: the desire to keep the robustness measure of order one for very deep networks exactly picks out activation functions in this universality class with $p_{\perp}=1$!\footnote{\label{foot:p-perp-less-than-or-equal-to-zero}Since $p_\perp = b_1 / a_1$, cf.~\eqref{eq:perp-asymptotic-solution}, $b_1 \geq a_1$, cf.~\eqref{eq:a1} and \eqref{eq:b1}, and $a_1 <0$,
    cf.~\eqref{eq:initial-tanh-wthout-subleading}, these  overall imply that $p_\perp \leq 1$ for any $K^\star=0$ activation function. In particular, for a non-odd activation function with $\sigma_2\ne 0$, the exponent for perpendicular perturbations is strictly less than one, $p_{\perp}<1$, and thus the bias term in the generalization error \eqref{eq:bias-factor-gen-error-not-squared} will grow with network depth $L$. %
} Such a condition is satisfied by any odd $K^\star=0$ activation function, such as $\tanhA$ and $\sinA$, both of which we've been discussing prominently throughout the book. %

Now, let's zoom out for a moment and reflect on these calculations more broadly. Somewhat miraculously, the theoretically-motivated tuning of all our hyperparameters -- \neo{criticality} for the \terminate{initialization hyperparameters} and the learning rate \neo{equivalence principle} for the \terminate{training hyperparameters} -- has led to the most practically-optimal solution for the generalization error in this one-training-one-test setting\index{generalization error!optimal hyperparameter tuning}, keeping the bias-variance tradeoff in check. Even more importantly, these choices and solutions are robust across many different network widths and depths, making them quite useful for experimentation and the scaling up of models.\footnote{These tunings are more or less still valid even as we relax the infinite-width requirement to allow nonzero aspect ratio, $L/n \ll 1$.
}
Of course, there really was no miracle: our theoretical principles were practically motivated from the start. 

Now that we understand what we \emph{should} do, let's discuss a different choice of training hyperparameters that we \emph{should not} make. Had we not followed the learning rate \terminate{equivalence principle}, perhaps we would have just made both weight and bias learning rates layer independent as $\lambda_b^{(\ell)} = \lambda_b$ and  $\lamW{\ell}=\lambda_W$. Let's see what happens then, specializing to odd activation functions with $\sigma_2=0$ and $p_{\perp}=1$ for simplicity. In this case, our general formal solution for the single-input frozen NTK~\eqref{eq:ntk-mean-k-star-formal-2} reduces to
\be\label{eq:frozen-ntk-k-star-solution-odd-mis-scaled}
\Ti{\NTKI}{\M\M}{\ell}=\le(\frac{\lambda_b}{2} \ri) \ell +\ldots\, ,
\ee
with a linear dependence on the layer $\ell$, while the same calculation as above with $\sigma_2=0$ and $p_{\perp}=1$ in mind gives a layer-independent constant asymptotic solution for $\delta\delta \NTKI_{[2]}^{(\ell)}$:
\be
\delta\delta \NTKI_{[2]}^{(\ell)}=\lambda_W\sigma_1^2\delta^2+\ldots\, .
\ee
Combined, our robustness measure becomes
\be\label{eq:robustness-k-star-sol-bad}
\frac{-2\delta\delta \NTKI_{[2]}^{(L)}}{\NTKI_{\M\M}^{(L)}}=\le[\frac{-4 \lambda_W\sigma_1^2 \delta^2}{ \lambda_b }\ri] \frac{1}{L} +\ldots\, ,
\ee
which is slowly but surely decaying with the overall depth $L$ of the network. (The consideration of more general activation functions with $p_{\perp} \neq 1$ doesn't change this conclusion.) Therefore, this choice of the training hyperparameters is polynomially suboptimal compared to the choice based on our \terminate{equivalence principle}.\footnote{We leave it to the reader to see how disastrous things would be -- in terms of our one-training-one-test generalization error -- if we had decided not to rescale the weight learning rate by the widths of the previous layer as in~\eqref{eq:diag_LR}, leading to an even more extreme violation of the learning rate \terminate{equivalence principle}.\index{criticality!principle of}\index{equivalence principle!connection to generalization}
}

Reflecting back, when we first discussed the learning rate \terminate{equivalence principle} by staring at our formal solution \eqref{eq:ntk-mean-k-star-formal-2}, we were motivated by the desire to ensure equal contributions to the NTK from each layer. Then in \S\ref{sec:EVGP-WEP} we realized that such choices solve a polynomial version of the exploding and vanishing gradient problem\index{exploding and vanishing gradient problem!connection to generalization}. Here we see the downstream consequences of those choices through the lens of generalization error, giving a solid support for the \terminate{equivalence principle} according to our quantitative measure of training success.

\subsubsection{Scale-Invariant Universality Class}\index{universality class!scale-invariant}\index{$\delta$ expansion}
To analyze scale-invariant activation functions, we need to use results from our 
finite-angle analysis in \S\ref{sec:finite_angle}. In particular, the \terminate{Gaussian expectation} $\bra \sigma''\sigma''\ket$ that appeared in the $\delta$ expansion of $\bra \sigma' \sigma' \ket$ in~\eqref{eq:primeprimeprimeprime} is singular for nonlinear scale-invariant functions due to the kink at the origin, and we promised we'd have to recall our finite-angle results when such a singularity occurs. %

Keeping our promise to you, let's recall a bunch of things from that 
section. First, we decomposed the two-input kernel matrix as \eqref{eq:angle-parametrization}\index{kernel!kernel matrix!polar angle parameterization}
\be\label{eq:angle-parametrization-reprint}
\Ti{\ker}{\delta_1\delta_2}{\ell}=
\begin{pmatrix}
\Ti{\ker}{++}{\ell} & \Ti{\ker}{+-}{\ell} \\
\Ti{\ker}{-+}{\ell}  & \Ti{\ker}{--}{\ell} 
\end{pmatrix}=\Kdi{\ell}\begin{pmatrix}
1 & \cos\!\le(\psi^{(\ell)}\ri)\\
\cos\!\le(\psi^{(\ell)}\ri)  & 1 
\end{pmatrix} \, , \qquad \psi^{(\ell)}\in\le[0,\pi\ri]\, ,
\ee
with two dynamical variables being the  diagonal kernel\index{kernel!kernel matrix!diagonal} $\Kdi{\ell}$ and the polar angle $\psi^{(\ell)}$. With this parametrization in mind, let us reprint a bunch of the previous results that we'll need, \eqref{eq:needed-to-recall-in-kernel-learning-chapter-I}, \eqref{eq:needed-to-recall-in-kernel-learning-chapter-II}, \eqref{eq:evaluated-gaussian-expectation}, and \eqref{eq:johnny-b-goode}:
\begin{align}\label{eq:scale-invariant-finite-identity-1}
\bra \sigma_+\sigma_+\ket_{K^{(\ell)}}=&\bra \sigma_-\sigma_-\ket_{K^{(\ell)}}=A_2 \Kdi{\ell}\, ,\\ %
C_W\bra \sigma'_+\sigma'_+\ket_{K^{(\ell)}}=&C_W\bra \sigma'_-\sigma'_-\ket_{K^{(\ell)}}=C_W A_2\equiv\chi\, , \\ %
\bra \sigma_+\sigma_-\ket_{K^{(\ell)}}=&A_2\Kdi{\ell}\le\{\cos\!\le(\psi^{(\ell)}\ri)+\rho\le[\sin\!\le(\psi^{(\ell)}\ri)-\psi^{(\ell)}\cos\!\le(\psi^{(\ell)}\ri)\ri]\ri\}\, ,\\
C_W\bra \sigma'_+\sigma'_-\ket_{K^{(\ell)}}=&\chi (1-\rho \psi^{(\ell)})\, , %
\label{eq:scale-invariant-finite-identity-4}
\end{align}
where $A_2 \equiv (a_+^2+a_{-}^2)/2$, 
$\rho \equiv \frac{1}{\pi} \frac{\le(a_{+}-a_{-}\ri)^2}{\le(a_{+}^2+a_{-}^2\ri)}$, and $a_{+}$ and $a_{-}$ are the two constants that define the particular activation function (though by now you know that by heart).

Let us now make a similar decomposition for the frozen NTK\index{frozen NTK!polar angle parameterization} as 
\be\label{eq:angle-parametrization-for-ntk}
\Ti{\NTKI}{\delta_1\delta_2}{\ell}=
\begin{pmatrix}
\Ti{\NTKI}{++}{\ell} & \Ti{\NTKI}{+-}{\ell} \\
\Ti{\NTKI}{-+}{\ell}  & \Ti{\NTKI}{--}{\ell} 
\end{pmatrix}=\NTKIdi{\ell}\begin{pmatrix}
1 & \cos\!\le(\zeta^{(\ell)}\ri)\\
\cos\!\le(\zeta^{(\ell)}\ri)  & 1 
\end{pmatrix} \, , \qquad \zeta^{(\ell)}\in\le[0,\pi\ri]\, ,
\ee
with a diagonal frozen NTK $\NTKIdi{\ell}$ and another polar angle $\zeta^{(\ell)}$.

Then, plugging this decomposition \eqref{eq:angle-parametrization-for-ntk} and recollected results \eqref{eq:scale-invariant-finite-identity-1}--\eqref{eq:scale-invariant-finite-identity-4} into the frozen NTK recursion \eqref{eq:tired-repetition-of-frozen-yogurt}, we get coupled recursions for the frozen NTK, casted in our finite-angle parameterization: 
\begin{align}
\NTKIdi{\ell+1}=&\chi\NTKIdi{\ell}+\Lb{\ell+1} + \LW{\ell+1} A_2 \Kdi{\ell}\, ,\label{eq:NTKI-diagonal-recursion}\\
\NTKIdi{\ell+1}\cos\!\le(\zeta^{(\ell+1)}\ri)=&\chi (1-\rho \psi^{(\ell)})\NTKIdi{\ell}\cos\!\le(\zeta^{(\ell)}\ri)\, \label{eq:NTKI-offdiagonal-recursion}\\
&+\Lb{\ell+1}\!\!+ \!\LW{\ell+1}\! A_2\Kdi{\ell}\!\!\le\{\cos\!\le(\psi^{(\ell)}\ri)\!+\rho\le[\sin\!\le(\psi^{(\ell)}\ri)\!-\psi^{(\ell)}\cos\!\le(\psi^{(\ell)}\ri)\ri]\ri\}\, .\notag
\end{align}
We see here in the off-diagonal recursion \eqref{eq:NTKI-offdiagonal-recursion} a finite-angle analog of what we saw perturbatively for the $K^\star=0$ universality class\index{universality class!K@$K^\star=0$} in the infinitesimal-angle recursion~\eqref{eq:deltadeltaNTK-recursion}: the polar angle for the kernel $\psi^{(\ell)}$ sources the finite angle for the frozen NTK $\zeta^{(\ell)}$. Said another way, the exponential growth and decay of the kernel angle $\psi^{(\ell)}$ -- at least for small enough angle -- are linked to  the exponential growth and decay of the frozen-NTK\index{frozen NTK} angle $\zeta^{(\ell)}$, which are in turn linked to the generalized bias-variance tradeoff\index{bias-variance tradeoff!for a universality class!scale-invariant activations}.\index{bias-variance tradeoff!generalized}

\index{universality class!scale-invariant}
With that chain of links in mind (as well as parallel discussions of similar issues in almost every other chapter of this book), it's natural that we should set our \terminate{initialization hyperparameters} by tuning to \terminate{criticality}: $\chi=1$. With this choice, we recall the critical solutions from our finite-angle analysis of the kernel in \S\ref{sec:finite_angle}:
\be\label{eq:scale-invariant-polar-angle-reprint}
\Kdi{\ell}=\Kdif\, , \qquad \psi^{(\ell)} = \le(\frac{3}{\rho}\ri) \frac{1}{\ell}+\ldots\, ,
\ee
where $\Kdif$ is exactly constant, set by the first layer.
Additionally, having already made the case for the learning rate equivalence principle when discussing the $K^\star=0$ universality class\index{universality class!K@$K^\star=0$}, 
let's just simplify our discussion here by setting \terminate{training hyperparameters} according to that \terminate{equivalence principle} for scale-invariant activations~\eqref{eq:super-scale-invariant}: $\Lb{\ell} = \widetilde{\lambda}_b /L$ and $\LW{\ell} = \widetilde{\lambda}_W /L$. With this choice, we see that 
\be\label{eq:frozen-ntk-scale-invariant-equivalence-rates}
\NTKIdi{\ell}=\le(\widetilde{\lambda}_b+\widetilde{\lambda}_W A_2 \Kdif\ri)\frac{\ell}{L}\, 
\ee
solves the recursion for the diagonal frozen NTK~\eqref{eq:NTKI-diagonal-recursion} with the initial condition~\eqref{eq:frozen-ntk-intial}. Importantly, here $\ell$ refers to a particular layer of the network, while $L$ is the overall network depth.\footnote{
    The frozen NTK solution~\eqref{eq:frozen-ntk-scale-invariant-equivalence-rates} is identical to our previous single-input solution~\eqref{eq:frozen-ntk-critical-solution-relu}, here we have just rescaled the bias and weight learning rates by the overall depth, $\lambda_b = \widetilde{\lambda}_b /L$ and $\lambda_W = \widetilde{\lambda}_W /L$, as required by the \terminate{equivalence principle}~\eqref{eq:super-scale-invariant}.
}

Plugging our choice of learning rates, our kernel solution \eqref{eq:scale-invariant-polar-angle-reprint}, and NTK solution \eqref{eq:frozen-ntk-scale-invariant-equivalence-rates} into the finite-angle recursion~\eqref{eq:NTKI-offdiagonal-recursion}, we get
after a bit of rearranging
\begin{align}
\cos\!\le(\zeta^{(\ell+1)}\ri)=\le(1-\frac{4}{\ell}+\ldots\ri) \cos\!\le(\zeta^{(\ell)}\ri)+\le(\frac{1}{\ell}+\ldots\ri)\, ,
\end{align}
which we can see easily is solved by an everything-independent constant
\be
\cos\!\le(\zeta^{(\ell)}\ri)=\frac{1}{4}+\ldots\, .
\ee
Thus, our robustness measure~\eqref{eq:robustness-measure} in the bias term of generalization error for nonlinear scale-invariant activation functions is given by a simple order-one number:
\be\label{eq:weird-asymptotic-behavior}
\frac{\Ti{\NTKI}{-+}{L}}{\Ti{\NTKI}{++}{L}}-1=\cos\!\le(\zeta^{(L)}\ri)-1=-\frac{3}{4}+\ldots\, .
\ee

Similarly, given that the nearby-input analysis can break down for nonlinear scale-invariant activations, let's use our finite-angle analysis here to also work out the variance term of the generalization error \eqref{eq:generalization-variance-expanded}; plugging in the asymptotic falloff for the kernel \eqref{eq:finite-angle-spin-0-solution} and \eqref{eq:finite-angle-spin-2-solution} as well as using \eqref{eq:weird-asymptotic-behavior} for the frozen NTK, we get 
\begin{align}\label{eq:scale-invariant-variance-gen-error}
\cov{\z{i}{-}{L}(T) }{ \z{i}{-}{L}(T)} &=\ker_{--}-2\frac{\NTKI_{-+}}{\NTKI_{++}}\ker_{-+}+\le(\frac{\NTKI_{-+}}{\NTKI_{++}}\ri)^2\ker_{++}\, \\
&= \Kdif \le[1- \cos\!\le(\zeta^{(L)}\ri) \ri]^2=\frac{9}{16}\Kdif+\ldots\, . \notag
\end{align}

Unlike the previous case for $K^\star=0$ activations, these asymptotic results for the generalization error, \eqref{eq:weird-asymptotic-behavior} and \eqref{eq:scale-invariant-variance-gen-error}, don't depend on the \terminate{training hyperparameters} $\widetilde{\lambda}_b$ and $\widetilde{\lambda}_W$, nor do they depend on a constant like $\delta^2$ that knows about the separation of the test and training points. (However, just as we discussed for $\psi^{(\ell)}$ in \S\ref{sec:finite_angle}, the depth at which these asymptotic results become valid does depend on $\delta^2$, the input norm, the activation function, and the training hyperparameters.)
Nonetheless, again with the correct tuning of our hyperparameters based on the principles of criticality and equivalence\index{criticality!principle of}\index{equivalence principle!connection to generalization}, we found a constant bias and variance, giving us the best possible tradeoff when training deep networks with nonlinear scale-invariant activations.\footnote{It is worth noting what happens with the special case of exact Bayesian inference where the only nonzero learning rates are in the last layer in order to set $\NTKI^{(L)}=\ker^{(L)}$. In that case, the robustness measure is given by $\cos\!\le(\psi^{(L)}\ri)-1=\o{1/\ell^2}$. Given this decay with depth, we see that the restricted Bayesian case is clearly inferior to an ensemble of networks that are fully-trained via gradient descent with uniform learning rates across layers. %
}

Let us end with the special remark on \terminate{deep linear network}s. For these networks, we use the $\linear$ activation function with $a_{+}=a_{-}$ and hence have $\rho=0$. In particular, we saw in \S\ref{sec:finite_angle} that not only was the diagonal kernel preserved at criticality, but the polar angle was preserved as well: $\Kdi{\ell}=\Kdif$ and $\psi^{(\ell)} = \psi^{\star}$. Noting this, the off-diagonal recursion for the frozen NTK\index{frozen NTK} \eqref{eq:NTKI-offdiagonal-recursion} then becomes
\be
\NTKIdi{\ell+1}\cos\!\le(\zeta^{(\ell+1)}\ri)=\NTKIdi{\ell}\cos\!\le(\zeta^{(\ell)}\ri)+\frac{\widetilde{\lambda}_b}{L}+\frac{\widetilde{\lambda}_W}{L} A_2\Kdif\cos\!\le(\psi^{\star}\ri)\, .
\ee
This recursion is exactly solved by
\be
\NTKIdi{\ell}\cos\!\le(\zeta^{(\ell)}\ri)=\le[\widetilde{\lambda}_b+\widetilde{\lambda}_W A_2\Kdif\cos\!\le(\psi^{\star}\ri)\ri]\frac{(\ell-1)}{L}+\NTKIdi{1}\cos\!\le(\zeta^{(1)}\ri)\, .
\ee
Dividing this result by our solution for the diagonal frozen NTK\index{frozen NTK} \eqref{eq:frozen-ntk-scale-invariant-equivalence-rates} then gives
\be
\cos\!\le(\zeta^{(\ell)}\ri)=\frac{\widetilde{\lambda}_b+\widetilde{\lambda}_W A_2\Kdif\cos\!\le(\psi^{\star}\ri)}{\widetilde{\lambda}_b+\widetilde{\lambda}_W A_2\Kdif}+\ldots\, ,
\ee
which polynomially asymptotes to a constant.
Unlike the case for the nonlinear scale-invariant activation functions, this constant depends on the observables in the first layer in a rather detailed manner, naturally connecting the generalization properties of the network to the input.
The real limitation of the \terminate{deep linear network}s becomes immediately apparent upon considering their (in)ability to interpolate/extrapolate, which we'll analyze next for linearly and nonlinearly activated MLPs.

\subsection{Interpolation and Extrapolation}\label{subsec:star-polation}
Rather than focusing primarily on our evaluation criteria for successful training, the \terminate{generalization error}, in this subsection we will focus more on
the kinds of functions that our trained neural networks actually compute. This analysis will enable us to consider the \emph{inductive bias}\index{inductive bias!of activation functions} of different activation functions and tell us how to relate the properties of those activation functions to the properties of the dataset and function that we're trying to approximate.

In the previous subsection we asked: given the true output $\y{i}{+}$ for an input $\x{i}{+}$, what does a fully-trained MLP in the infinite-width limit predict for the output of a nearby input $\x{i}{-}$? Here, we up the ante and ask: given the true outputs $\y{i}{\pm}$ for \emph{two} inputs $\x{i}{\pm}=\x{i}{\M}\pm \frac{\delta x_i}{2}$, what is the prediction for a one-parameter family\index{one-parameter families} of test inputs,
\be\label{eq:polation-input}
s \x{i}{+}+(1-s)\x{i}{-}=\x{i}{\M}+ \frac{(2s-1)}{2}\delta x_i\equiv \x{i}{(2s-1)}\, ,
\ee
that sit on a line passing through $\x{i}{+}$ and $\x{i}{-}$?
When our parameter $s$ is inside the unit interval $s\in[0,1]$, this is a question about neural-network \term{interpolation}\index{interpolation|seealso{$\ast$-polation}}; for $s$ outside the unit interval, it's a question about \term{extrapolation}\index{extrapolation|seealso{$\ast$-polation}}. For general $s$, let's refer to this collectively as \textbf{$\ast$-polation}.\index{$\ast$-polation|textbf} 

\index{deep linear network!limitations} 
First we'll perform a little exercise in $\ast$-polation with deep linear networks to see what networks with $\linear$ activation functions do. Accordingly, we'll see concretely how deep linear networks approximate a very limited set of functions. Then we'll follow up by assessing smooth nonlinear networks.

\subsubsection{Linear $\ast$-Polation by Deep Linear Networks}\index{$\ast$-polation!deep linear networks}

\index{deep linear network}\index{linear transformations}
There's a very simple way to see how $\ast$-polation works for deep linear networks.
If we recall for a moment (and for one last time) the forward equation for deep linear networks~\eqref{eq:deep-linear-foward-pass},
\be\label{eq:deep-linear-foward-pass-reprint}
\z{i}{\alpha}{\ell+1} = \bias{i}{\ell+1} + \sum_{j=1}^{n_\ell}\W{ij}{\ell+1} \z{j}{\alpha}{\ell} \, ,
\ee
with $\z{j}{\alpha}{0}\equiv\x{j}{\alpha}$,  it's clear that the linear structure in the input \eqref{eq:polation-input} will be preserved from layer to layer. That is, given an $\ell$-th-layer preactivations of the form
\be
\z{i}{(2s-1)}{\ell} = s\z{i}{+}{\ell} + (1-s)\z{i}{-}{\ell}
\ee
that has such a linear structure, we then have for the next layer
\begin{align}
\z{i}{(2s-1)}{\ell+1} &= \bias{i}{\ell+1} + \sum_{j=1}^{n_\ell}\W{ij}{\ell+1}\le[s\z{j}{+}{\ell} + (1-s)\z{j}{-}{\ell}\ri] \, \\
&=s\le( \bias{i}{\ell+1} +    \sum_{j=1}^{n_\ell}\W{ij}{\ell+1}\z{j}{+}{\ell} \ri) + (1-s)\le( \bias{i}{\ell+1} + \sum_{j=1}^{n_\ell}\W{ij}{\ell+1}\z{j}{-}{\ell}\ri) \notag \\
&= s\z{i}{+}{\ell+1} + (1-s)\z{i}{-}{\ell+1} \, \notag ,
\end{align}
which still respects the linear structure. This is just a direct consequence of the fact that deep linear networks compute linear functions of their input.

Therefore, for a test input that's a
linear sum of our two training points \eqref{eq:polation-input}, the network will output the linear sum of the network outputs on the two individual training points:
\be\label{eq:linear-polation-at-init}
\z{i}{(2s-1)}{L}= s\z{i}{+}{L} + (1-s)\z{i}{-}{L} \, .
\ee
This equation holds at initialization as well as at the end of training, which means that any particular fully-trained deep linear network will $\ast$-polate as
\be\label{eq:linear-polation}
\z{i}{(2s-1)}{L}(T) =s \y{i}{+}+(1-s)\y{i}{-}\, ,
\ee
since the fully-trained network output will equal the true output for any element in the training set: $\z{i}{\pm}{L}(T) = \y{i}{\pm}$. With this, we see that fully-trained deep linear network \emph{linearly} $\ast$-polate, no matter what. This is both intuitive and pretty obvious; as deep linear networks perform linear transformations they can only compute linear functions. 

Of course, this is exactly what we said when we studied deep linear networks way back
in \S\ref{ch:deep-linear-eft}.
Here, we explicitly see \emph{why} these networks are limited after training, by showing the (limited) way in which they can use training examples to make predictions. Accordingly, if the function you're trying to approximate is a linear function of the input data, then deep linear networks are a great modeling choice. If the function is nonlinear, we'll have to consider nonlinear activation functions. 
It's not that deep.

\subsubsection{Nonlinear $\ast$-Polation by Smooth Nonlinear Deep Networks}\index{$\ast$-polation!nonlinear networks}
We'll have to work a little harder to see what nonlinear networks do.\footnote{We would have to work even harder to see what nonlinear scale-invariant activation functions do, so here we'll focus on smooth nonlinear activation functions. For such nonlinear scale-invariant activation functions with kinks, since the $\ast$-polated input $\x{i}{(2s-1)}$ does not have the same norm as $\x{i}{\pm}$ for $s\ne0,1$, we would need to extend the finite-angle analysis from \S\ref{sec:finite_angle} to the case of unequal input norms.
This is left as a challenge in pedagogy to future deep-learning book authors.
} In the last section, we saw that the output of a fully-trained network is given by the stochastic kernel prediction equation~\eqref{eq:kernel-prediction}, which we reprint here for convenience:
\be\label{eq:kernel-prediction-reprint}
\z{i}{\tea}{L}(T)=\z{i}{\tea}{L}- \sum_{\tra_1, \tra_2 \in\A}  \Ti{\NTKI}{\tea \tra_1}{L} \TI{\NTKIsub}{\tra_1 \tra_2}{L}\le(\z{i}{\tra_2}{L}-\y{i}{\tra_2}\ri)\, .
\ee
Thus, we see that to study $\ast$-polation\index{$\ast$-polation} more generally we 
will
need to evaluate elements of the frozen NTK between our test and training set, $\Ti{\NTKI}{(2s-1) \pm}{L}$,
and also need to invert the  two-by-two submatrix of the frozen NTK on the training set only, $\TI{\NTKIsub}{\tra_1 \tra_2}{L}$.

This latter inversion can be easily completed with
the standard textbook formula for the inverse of a two-by-two matrix,
\be\label{eq:frozen-NTK-submatrix-inverse}
\NTKIsub^{\tra_1 \tra_2} =
\frac{1}{\NTKI_{++}\NTKI_{--} - \NTKI_{+-}^2}
\begin{pmatrix}
\NTKI_{--} & -\NTKI_{+-} \\
-\NTKI_{+-}  & \NTKI_{++}  \\
\end{pmatrix}\, ,
\ee
where here we've also used the symmetry $\NTKI_{+-}=\NTKI_{-+}$ and further dropped the \terminate{layer indices}. For the rest of this section, we will always assume that these frozen NTKs are evaluated at the output layer.

Next, to compute the off-diagonal elements between the training set and the test set $\NTKI_{(2s-1) \pm}$, we'll need to generalize our $\delta$ expansion a bit.\index{$\delta$ expansion!generalized to $\epsilon_{1,2}$ expansion}
Let's first recall our expressions for the components of the frozen NTK in $\gamma^{[a]}$ basis, \eqref{eq:NTK-in-gamma-basis}, and then plug in the $\delta$ expansion we performed on the two-by-two submatrix $\NTKIsub_{\tra_1\tra_2}$, \eqref{eq:frozen-NTK-delta-0}--\eqref{eq:frozen-NTK-delta-2}, which gives
\begin{align}
\NTKI_{\pm\pm}=&\NTKI_{\M\M}\pm \delta\NTKI_{[1]}+ \le(\delta\delta\NTKI_{[2]}+\delta\delta\NTKI_{[0]}\ri)+\o{\delta^3}\ ,\label{eq:kernel-diagonal-Taylor-in-delta}\\
\NTKI_{\pm\mp}=&\NTKI_{\M\M}+ \le(-\delta\delta\NTKI_{[2]}+\delta\delta\NTKI_{[0]}\ri)+\o{\delta^3}\ ,\label{eq:kernel-offdiagonal-Taylor-in-delta}
\end{align}
for the pair of inputs $\x{i}{\pm}=\x{i}{\M}\pm \frac{\delta x_i}{2}$. Let's now consider a pair of perturbed inputs of a more general form
\be\label{eq:generic-two-perturabations-for-polation}
\x{i}{\epsilon_1}\equiv\x{i}{\M}+\frac{\epsilon_1}{2}\delta x_{i}\, , \qquad \x{i}{\epsilon_2}\equiv\x{i}{\M}+\frac{\epsilon_2}{2}\delta x_{i} \, .
\ee 
Note that picking $\epsilon_{1,2}$ from $\pm1$ reduces them to $\x{i}{\pm}$,
while the new case of the interest, $\NTKI_{(2s-1) \pm}$, corresponds to setting $\epsilon_1=(2s-1)$ and $\epsilon_2=\pm1$.
For a generic pair of inputs \eqref{eq:generic-two-perturabations-for-polation}, the $\delta$ expansion gets modified as
\begin{align}\label{eq:key-extension}
\NTKI_{\epsilon_1\epsilon_2}=&\NTKI_{\M\M}+\le(\frac{\epsilon_1+\epsilon_2}{2}\ri) \delta\NTKI_{[1]}+\le(\frac{\epsilon_1+\epsilon_2}{2}\ri)^2 \le(\delta\delta\NTKI_{[2]}+\delta\delta\NTKI_{[0]}\ri)\, \\
&+\le(\frac{\epsilon_1-\epsilon_2}{2}\ri)^2 \le(-\delta\delta\NTKI_{[2]}+\delta\delta\NTKI_{[0]}\ri)+\o{\epsilon^3\delta^3}\ .\notag
\end{align}
To see why this is the correct expression, note that \emph{(i)} each term has the right scaling with $\epsilon_{1,2}$, \emph{(ii)} for $\epsilon_1=\epsilon_2=\pm1$ we correctly recover the expression for $\NTKI_{\epsilon_1\epsilon_2}=\NTKI_{\pm\pm}$ \eqref{eq:kernel-diagonal-Taylor-in-delta}, \emph{(iii)} for $\epsilon_1=-\epsilon_2=\pm1$, we correctly recover the expression for $\NTKI_{\epsilon_1\epsilon_2}=\NTKI_{\pm\mp}$ \eqref{eq:kernel-offdiagonal-Taylor-in-delta}, and \emph{(iv)} the expression is symmetric under $\epsilon_1\leftrightarrow\epsilon_2$. The frozen NTK\index{frozen NTK!$\epsilon_{1,2}$ expansion|see{$\delta$ expansion}}\index{frozen NTK!$\delta$ expansion|see{$\delta$ expansion}} component $\NTKI_{\epsilon_1\epsilon_2}$ must satisfy these four constraints, and the expression \eqref{eq:key-extension} is the unique formula that satisfies them all.

Applying this formula to evaluate $\NTKI_{(2s-1)\pm}$ and simplifying a bit, we find
\begin{align}\label{eq:test-train-frozen-ntk-polation}
\NTKI_{(2s-1)\pm}=s \NTKI_{\pm+}+(1-s)\NTKI_{\pm-}-2s(1-s)\delta\delta\NTKI_{[0]}+\o{\delta^3}\, .
\end{align}
As we'll see, the key to nonlinear $\ast$-polation\index{$\ast$-polation}, at least for nearby inputs, is in the $\delta\delta\NTKI_{[0]}$ term.\footnote{N.B.~$\delta\delta\NTKI_{[0]}$ is very different from $\delta\delta\NTKI_{[2]}$: the former is the second term in the expansion of the $\gamma^{[0]}$ component  $\NTKI_{[0]}$, cf.~\eqref{eq:frozen-NTK-delta-0}, while the latter is the first term in the expansion of the $\gamma^{[2]}$ component $\NTKI_{[2]}$, cf.~\eqref{eq:frozen-NTK-delta-2}.}
Firstly, it's clear that this term nonlinearly depends on the test input, as evidenced by $s(1-s)$ prefactor.
Indeed, you can go back and check that this term identically vanishes for deep linear networks, i.e., for those networks we simply have $\NTKI_{(2s-1)\pm}=s \NTKI_{\pm+}+(1-s)\NTKI_{\pm-}$.\footnote{To see this quickly, note that both the first-layer metric \eqref{eq:first-layer-metric} and the first-layer NTK \eqref{eq:NTHinitial} are bilinear in the two inputs, and that such bilinear structure is preserved under the recursions for deep linear networks: $\Ti{\ker}{\delta_1\delta_2}{\ell+1}=\Cb{\ell+1} + \CW{\ell+1}\Ti{\ker}{\delta_1\delta_2}{\ell}$,
cf.~\eqref{eq:K-recursion-reprint},
and $ \Ti{\NTKI}{\delta_1\delta_2}{\ell+1}=\Lb{\ell+1} + \lamW{\ell+1}\Ti{\ker}{\delta_1\delta_2}{\ell}+\CW{\ell+1}\Ti{\NTKI}{\delta_1\delta_2}{\ell}$,
cf.~\eqref{eq:frozen-NTK-recursion}.} With that in mind, it also helps to decompose the initial preactivation into linear and nonlinear pieces as
\be\label{eq:init-preactivation-decomposition-for-polation}
\z{i}{(2s-1)}{L}=s\z{i}{+}{L}+(1-s)\z{i}{-}{L}+\le[\z{i}{(2s-1)}{L}-s\z{i}{+}{L}-(1-s)\z{i}{-}{L}\ri]\, .
\ee
Here, the second term vanishes for deep linear networks, as per \eqref{eq:linear-polation-at-init}, and so in general it captures nonlinearity of the network output at initialization.

Plugging \eqref{eq:frozen-NTK-submatrix-inverse}, \eqref{eq:test-train-frozen-ntk-polation}, and \eqref{eq:init-preactivation-decomposition-for-polation} into our kernel prediction formula \eqref{eq:kernel-prediction-reprint}, we see that our fully-trained prediction on the test input $\x{i}{(2s-1)}=s\x{i}{+}+(1-s)\x{i}{-}$ is given by
\begin{align}
&\z{i}{(2s-1)}{L}(T)\, \label{eq:polation-general}\\
=&\le[\z{i}{(2s-1)}{L}-s\z{i}{+}{L}-(1-s)\z{i}{-}{L}\ri]+\le[s \y{i}{+}+(1-s)\y{i}{-}\ri]\, \notag\\
&-s(1-s)\le[\frac{2\delta\delta\NTKI_{[0]}}{\NTKI_{\M\M}\delta\delta\NTKI_{[2]}-\delta\NTKI_{[1]}^2}\ri]\Big[2\delta\delta\NTKI_{[2]}\le(\z{i}{+}{L}+\z{i}{-}{L}+ \y{i}{+}+\y{i}{-}\ri)\, \notag\\
&\quad\quad\quad\quad\quad\quad\quad\quad\quad\quad\quad\quad\quad\quad\quad-\delta\NTKI_{[1]}\le(\z{i}{+}{L}-\z{i}{-}{L}+\y{i}{+}-\y{i}{-}\ri)\Big]+\o{\delta^3}\, .\notag
\end{align}
Comparing with our linear $\ast$-polation formula \eqref{eq:linear-polation}, we see that both the first and last terms are new: nonlinear networks can \emph{nonlinearly} $\ast$-polate!\index{$\ast$-polation!nonlinear networks}
Interestingly, the fully-trained $\ast$-polation for nonlinear activation functions depends on the network output at initialization through the nonlinearity $\z{i}{(2s-1)}{L}-s\z{i}{+}{L}-(1-s)\z{i}{-}{L}$; in contrast, for deep linear networks the $\ast$-polation only depended on the true output of the training examples. %

As a particular illustration of this formula, consider the case when the two training inputs have the same norm. In such a case $\NTKI_{[1]}=0$, and we find a much simpler formula:
\begin{align}\label{eq:polation-equal-norm-inputs}
\z{i}{(2s-1)}{L}(T)=&\le[\z{i}{(2s-1)}{L}-s\z{i}{+}{L}-(1-s)\z{i}{-}{L}\ri]+\le[s \y{i}{+}+(1-s)\y{i}{-}\ri]\, \\
&-4s(1-s)\le(\frac{\delta\delta\NTKI_{[0]}}{\NTKI_{\M\M}}\ri) \le(\z{i}{+}{L}+\z{i}{-}{L}+ \y{i}{+}+\y{i}{-}\ri)+\o{\delta^3}\, .\notag
\end{align}
Averaging over our ensemble, this prediction has a mean
\be\label{eq:equal-norm-polation-mean}
\GDGPmean_{i;(2s-1)}=s \y{i}{+}+(1-s)\y{i}{-}-4s(1-s)\le(\frac{\delta\delta\NTKI_{[0]}}{\NTKI_{\M\M}}\ri)\le(\y{i}{+}+\y{i}{-}\ri)+\o{\delta^3}\, .
\ee
Here the first term in \eqref{eq:polation-equal-norm-inputs} that captured the nonlinearity of the network output at initialization vanished under the expectation, and so the nonlinearity of the $\ast$-polation mean is entirely captured by the  dimensionless ratio $\delta\delta\NTKI_{[0]}/\NTKI_{\M\M}$.

\index{$\ast$-polation!curvature}
So, what kind of a function is our fully-trained infinite-width nonlinear neural network computing? 
To assess this, note that the ratio $\delta\delta\NTKI_{[0]}/\NTKI_{\M\M}$ captures the \emph{curvature} of the $\ast$-polation in the neighborhood of the training points.\footnote{Note that as the two training samples  begin to coincide $x_\pm \to x_\M$, the curvature vanishes quadratically $\delta\delta\NTKI_{[0]}/\NTKI_{\M\M} =\o{\delta^2}$, and the closer the $\ast$-polation will be to a linear $\ast$-polation. Further applying our generalized $\delta$ expansion~\eqref{eq:key-extension} to the kernel, we can show that that the variance of the $\ast$-polation vanishes even more quickly in this coincident limit as
\begin{align}
 \E{\z{i}{2s-1}{L}(T)\z{i}{2s-1}{L}(T)}-\le(\E{\z{i}{2s-1}{L}(T)}\ri)^2=\o{\delta^3}\, .
\end{align}
}\index{$\ast$-polation!curvature}
This curvature encodes a non-universal \emph{inductive bias}\index{inductive bias!of activation functions} of the activation function and architecture indicating how this class of function approximators will generalize to novel data.

For a given task and dataset, some activation functions might produce a more desired type of $\ast$-polation. This can be measured directly via the bias term in the generalization error.\index{generalization error!bias!related to $\ast$-polation} Substituting in our equal norm expression for the mean~\eqref{eq:equal-norm-polation-mean},
\begin{align}\label{eq:polation-bias}
\GDGPmean_{i;(2s-1)}-\y{i}{(2s-1)} =& \le[\y{i}{+}+(1-s)\y{i}{-}-\y{i}{(2s-1)}\ri]  \notag \\
      &-4s(1-s)\le(\frac{\delta\delta\NTKI_{[0]}}{\NTKI_{\M\M}}\ri)\le(\y{i}{+}+\y{i}{-}\ri)+\o{\delta^3} \, ,
\end{align}
we see that this generalization error bias decomposes into a comparison between the nonlinearity in the true output -- given by the first square brackets -- and the network curvature around the midpoint of the true output $\le(\y{i}{+}+\y{i}{-}\ri)/2$.
With this framing, \terminate{deep linear network}s promote a very particular type of inductive bias: only linear functions are computed. %
More generally, we could (but won't here)
compute and solve a recursion for $\delta\delta\NTKI_{[0]}$ for any particular activation function in order to learn more about the kinds of functions computed by deep networks with that activation function.

Finally, note that this analysis doesn't make any particular distinction between \emph{interpolation} and \emph{extrapolation}\index{$\ast$-polation!same for inter- and extra-}, and also that as $s\to 0,1$, the $\ast$-polation \eqref{eq:polation-general} reduces to $\y{i}{\pm}$ with absolute certainty\index{absolute certainty!$\ast$-polation}. In fact, in the neighborhood of $s=0,1$, the $\ast$-polation bias \eqref{eq:polation-bias} has much in common with the prediction bias \eqref{eq:bias-factor-gen-error-not-squared} and \eqref{eq:robustness-measure-plugged-in} that we saw in \S\ref{subsec:robustness-from-infinite-GD} when considering a training set consisting of only one training sample. Importantly, it is the most nearby training point that contributes the most to a test point's prediction.

Taken as a guide to thinking about larger training sets, the \emph{local} nature of these predictions is highly suggestive of some ways to make further progress. On the one hand, we might be able to make theoretical progress on more complicated prediction formulae by weighting the predictions given by nearby training points to a given test point, perhaps using an approximation from \S\ref{subsec:robustness-from-infinite-GD} when there's only one nearby training point and using $\ast$-polation \eqref{eq:polation-general} when there's a nearby pair. On the other hand, we might be able to make practical progress on training-set design -- given the network's inductive biases -- by using this kind of analysis to inform how best to sample training inputs over the data manifold.

\section{Linear Models and Kernel Methods}\label{sec:lazy-kernel} 
Before we back off the infinite-width limit, let's take a section to place what we've done in this chapter into the broader context of \terminate{machine learning}. In the next chapter, such a context will help us understand  the ways in which deep learning at finite width is qualitatively quite different from its infinite-width counterpart.

In particular, in this section we'll explain a \emph{dual}\index{duality!linear model -- kernel methods} way of thinking about the class of models that can be described by a kernel prediction\index{kernel methods!prediction!as a linear model} formula such as \eqref{eq:kernel-prediction}.  On the one hand, kernel predictions can be thought of as being made by $\ast$-polating the training data using the kernel. On the other hand, we can think of them as the output of a trained model that's linear in its parameters. The former perspective has been more natural to us, given that we always consider an ensemble over the model parameters and then integrate them out. So let's begin by explaining the latter \neo{linear model} perspective.\footnote{
The connection between infinite-width networks trained by gradient descent and kernel methods was pointed out in \cite{jacot2018neural} in the context of introducing the NTK. Following that, an extended discussion of such networks as linear models was given in  \cite{brainNTK2019}. %
}

\subsection{Linear Models}\label{subsec:linear-models}

The simplest linear model -- and perhaps the simplest machine learning model -- is just a one-layer (i.e.~zero-hidden-layer) network
\be\label{eq:dumb-linear-model-def}
z_i(x_{\delta}; \theta) =b_i+\sum_{j=1}^{n_0} W_{ij} \x{j}{\delta} \, .
\ee
While this model is linear in both the parameters $\theta = \{b_i,  W_{ij}\}$ and the input $\x{j}{\delta}$, the \emph{linear} in \emph{linear model} takes its name from the dependence on the parameters $\theta$ and not the input $x$. In particular, while the components of the input samples $\x{j}{\delta}$ sometime can serve as a reasonable set of features for function approximation\index{function approximation!for linear models}, in general they do not.
Indeed, considering how much ink we've already spilled on \terminate{representation group flow} and \terminate{representation learning} in the context of deep learning, it's natural to expect that we would need to (pre-)process the input data before it's useful for any machine learning task.

One traditional way to fix this, inherited from statistics, is to engineer better features. Such an approach was necessary when computers were less powerful and models had to be much simpler to optimize.
For instance, in addition to the features $x_j$ perhaps it would also be useful for the model to take into account features $x_j x_k$ that let us consider the dependence of one component upon another. More generally, we might design a fixed set of \index{feature function|textbf}\textbf{feature functions} $\fea_j(x)$ that's meant to work well for the dataset\index{input data} $\D$ and the underlying task at hand.\footnote{
    These type of feature functions are also useful if the input $x$ is something abstract -- such as a document of text -- and thus needs to be transformed into a numerical vector before it can be processed by a parameterized model.\index{feature function!feature engineering for abstract inputs}
}

In this traditional approach, the hope is that all the complicated modeling work goes into the construction of these feature functions $\fea_j(x)$ and, if we do a good enough job, then its associated \term{linear model},
\be\label{eq:linear-model-def}
z_i(x_{\delta}; \theta) =b_i+ \sum_{j=1}^{n_f} W_{ij} \fea_j(x_{\delta})=\sum_{j=0}^{n_f}W_{ij}\fea_j(x_{\delta}) \, ,
\ee
is simple to train, easy to interpret, and performs well on the desired task.
Here, we've
followed a customary notational reductionism, subsuming the bias vector into the weight matrix by setting $\fea_0(x)\equiv1$ and $W_{i0}\equiv b_i$. Thus, the output $z_i(x; \theta)$ of a linear model
depends linearly on the model parameters $\theta$, 
consisting of a combined weight matrix $W_{ij}$ of dimension $n_{\text{out}}\times (n_{f}+1)$. We can still think of this model as a one-layer neural network, but in this case we pre-process each input with the function $\fea_j(x)$ before passing it through the network.

Now let's explain how to learn the optimal values for weight matrix $W_{ij}^\star$ given a training set $\A$. The most common approach is to minimize the MSE loss\index{loss!MSE!for linear models}
\be\label{eq:linear-regression}
\L_{A}(\theta)=\frac{1}{2}\sum_{\tra\in\A}\sum_{i=1}^{n_{\text{out}}}\le[\y{i}{\tra}-z_i(x_{\tra}; \theta)\ri]^2=\frac{1}{2}\sum_{\tra\in\A}\sum_{i=1}^{n_{\text{out}}}\le[\y{i}{\tra}-\sum_{j=0}^{n_f}W_{ij} \fea_j(x_{\tra})\ri]^2\, .
\ee
Supervised learning\index{supervised learning!with linear models|see{linear regression}} with a linear model is known as \term{linear regression}\index{linear regression|seealso{linear model}}, and -- as the MSE loss of a linear model is necessarily quadratic in the model parameters -- this is another case of an analytically-solvable learning problem \eqref{eq:gradient-vanishing-mind}. Taking the derivative of $\L_{\A}$ with respect to the parameters and setting it to zero, we get an implicit equation that determines the optimal weight matrix  $W_{ij}^\star$:
\be\label{eq:linear-model-implicit-expression}
\sum_{k=0}^{n_f}W^{\star}_{ik}\le[\sum_{\tra\in\A}\fea_k(x_{\tra})\fea_j(x_{\tra})\ri]=\sum_{\tra\in\A} \y{i}{\tra}\fea_{j}(x_{\tra})\, .
\ee
To solve this equation, let's define a symmetric $(n_f+1)$-by-$(n_f+1)$ matrix of features,
\be\label{eq:no-good-matrix-name}
M_{ij}\equiv\sum_{\tra\in\A}\fea_i(x_{\tra})\fea_j(x_{\tra})\, ,
\ee
with elements that give a pairwise aggregation of feature functions summed over all the training samples $\tra \in \A$.
Then, applying its inverse to both sides of the implicit expression~\eqref{eq:linear-model-implicit-expression}, we find a solution:
\be\label{eq:linear-regression-optimal}
W^{\star}_{ij}=\sum_{k=0}^{n_f}\sum_{\tra\in\A} \y{i}{\tra}\fea_{k}(x_{\tra}) \le(M^{-1}\ri)_{kj}\, .
\ee
Notice that the solution depends on the training set, linearly for the true function values $\y{i}{\tra}$ and in a more complicated way on the input features $\fea_{k}(x_{\tra})$.\footnote{
    If the number of features $(n_f +1)$ is larger than the size of the training set $\NR$, then the model is \emph{overparameterized}\index{overparameterization}, and $M_{ij}$ is not uniquely invertible. One scheme to specify the solution is to add a regularization\index{regularization!for linear models} term of the form $a \sum_{ij} W_{ij}^2$ to the loss \eqref{eq:linear-regression}, cf.~footnote~\ref{footnote:regularization-recursion} in \S\ref{subsec:NTKprediction} for a related discussion of regularization for infinite-width networks. In this modified regression problem, we can then invert the regularized matrix
    \be\label{eq:inverse-of-M-regularized}
   M_{ij} = 2a\, \delta_{ij} + \sum_{\tra\in\A}\fea_i(x_{\tra})\fea_j(x_{\tra})\, ,
    \ee
and send the regulator to zero, $a \to 0^+$, at the end of our calculations. 
Note that either when the regulator $a$ is kept finite or when we're in the \emph{underparameterized}\index{underparameterization} regime with $(n_f+1) < \NR$, the linear model will no longer reach zero training loss even when fully optimized.\label{footnote:inverse-existing}
}
Finally, we can use this fully-trained linear model with its associated optimal parameters $W^{\star}_{ij}$ to make predictions on novel test-set inputs $x_{\tea}$ as
\be\label{eq:kernel-prediction-kernel-methods-before-duality}
z_i\big(x_{\tea}; \theta^\star\big)=   \sum_{j=0}^{n_f} W_{ij}^\star \fea_j(x_{\tea}) \, , %
\ee
giving us a closed-form solution for our \terminate{linear regression} problem. %
Importantly, after learning is complete we can simply store the optimal parameters $W^{\star}_{ij}$ and forget about the training data.

\subsection{Kernel Methods}\label{subsec:kernel-methods}

While this is all very easy, it's less familiar in our book since we typically do not work explicitly with the parameters. To cast our linear model into a more familiar form, let's consider a \emph{dual}\index{duality} expression for the solution. First, let's substitute our expression for the optimal parameters $W^{\star}_{ij}$, \eqref{eq:linear-regression-optimal}, into our linear regression solution, \eqref{eq:kernel-prediction-kernel-methods-before-duality}, giving
\be\label{eq:kernel-prediction-kernel-methods-before-duality-square}
z_i\big(x_{\tea}; \theta^\star\big) = \sum_{\tra\in \A} \le[\sum_{j,k=0}^{n_f}\fea_{j}(x_{\tea}) \le(M^{-1}\ri)_{jk}\fea_{k}(x_{\tra})\ri]\y{i}{\tra} \, .
\ee
Note that the expression in the square brackets involves the inversion of an $(n_f+1)\times(n_f+1)$-dimensional matrix $M_{ij}$, which was required to obtain the optimal parameters $W^{\star}_{ij}$. 
This works well if the number of features is small, but if the number of feature functions we defined is very large $n_f \gg 1$, then representing and inverting such a matrix might be computationally difficult.

However, it turns out that we actually don't need to do any of that. To see why, let us introduce a new $\ND\times\ND$-dimensional symmetric matrix:
\be\label{eq:kernel-with-features}
\kerm_{\delta_1\delta_2}\equiv\kerm\!\le(x_{\delta_1},x_{\delta_2} \ri) \equiv \sum_{i=0}^{n_{f}} \fea_i\!\le(x_{\delta_1}\ri) \fea_i\!\le(x_{\delta_2}\ri) \, .
\ee
As an inner product of feature functions, $\kerm_{\delta_1\delta_2}$ is a measure of similarity between two inputs $x_{i;\delta_1}$ and $x_{i;\delta_2}$ in \terminate{feature space}. Such a measure of similarity is called a \textbf{kernel}\index{kernel methods!kernel}.\footnote{
For instance, in the case of the simplest linear model \eqref{eq:dumb-linear-model-def}, the kernel is just given by  the inner product between the two inputs
\be\label{eq:linear-kernel-methods}
\kerm_{\delta_1\delta_2} \equiv \sum_{i=1}^{n_0} x_{i;\delta_1} x_{i;\delta_2} \, ,
\ee
which is often called the \emph{linear kernel}\index{kernel methods!kernel!linear}.
}
In a way that should feel very familiar, we'll also denote an $\NR$-by-$\NR$-dimensional submatrix of the kernel evaluated on the training set as $\kermsub_{\tra_1 \tra_2}$ with a tilde. This lets us write its inverse as $\widetilde{\kerm}^{\tra_1 \tra_2}$, which satisfies
\be
\sum_{\tra_2 \in \A} \kermsub^{\tra_1 \tra_2} \kermsub_{\tra_2 \tra_3} = \delta^{\tra_1}_{\ \tra_3} \, .
\ee
Note that given the definition of the kernel \eqref{eq:kernel-with-features}, for this inverse to exist and for this equation to hold we must be in the \emph{overparameterized}\index{overparameterization} regime with $(n_f+1) \ge \NR$.

Now with this, let's see how we might rearrange the factor in the square brackets of our solution \eqref{eq:kernel-prediction-kernel-methods-before-duality-square}. Multiplying it by the submatrix $\kermsub_{\tra \tra_1}$, we can simplify this factor as
\begin{align}\label{eq:kernel-trick}
&\sum_{\tra\in\A}\le[\sum_{j,k=0}^{n_f}\fea_{j}(x_{\tea}) \le(M^{-1}\ri)_{jk}\fea_{k}(x_{\tra})\ri] \kermsub_{\tra\tra_1}\, \\
=&\sum_{\tra\in\A}\sum_{j,k=0}^{n_f}\fea_{j}(x_{\tea}) \le(M^{-1}\ri)_{jk}\fea_{k}(x_{\tra}) \sum_{i=0}^{n_{f}} \fea_i\!\le(x_{\tra}\ri) \fea_i\!\le(x_{\tra_1}\ri)\, \notag\\
=&\sum_{i, j,k=0}^{n_f}\fea_{j}(x_{\tea}) \le(M^{-1}\ri)_{jk}M_{ki}\, \fea_i\!\le(x_{\tra_1}\ri)\, \notag\\
=&\sum_{i=0}^{n_f}\fea_{i}(x_{\tea})\fea_i\!\le(x_{\tra_1}\ri)=\kerm_{\tea\tra_1}\, .\notag
\end{align}
To get this result, in the second line we plugged in the definition of the kernel \eqref{eq:kernel-with-features}, in the third line we performed the sum over $\tra$ using the definition of the feature matrix $M_{ij}$ \eqref{eq:no-good-matrix-name}, and in the last equality of the fourth line we again used the definition of the kernel. Finally, multiplying the first and last expressions by the inverse submatrix $\widetilde{\kerm}^{\tra_1 \tra_2}$, we get a new representation for the factor in the square brackets
\be\label{eq:kernel-trick-result}
\le[\sum_{j,k=0}^{n_f}\fea_{j}(x_{\tea}) \le(M^{-1}\ri)_{jk}\fea_{k}(x_{\tra_2})\ri]=\sum_{\tra_1\in\A}\kerm_{\tea\tra_1}\kermsub^{\tra_1\tra_2}\, ,
\ee
which lets us rewrite the prediction of our linear model \eqref{eq:kernel-prediction-kernel-methods-before-duality} as
\be\label{eq:kernel-prediction-kernel-methods}
z_i\big(x_{\tea}; \theta^\star\big)= \sum_{\tra_1,\tra_2 \in \A} \kerm_{\tea \tra_1} \widetilde{\kerm}^{\tra_1 \tra_2} \y{i}{\tra_2} \, .
\ee
When the prediction of a linear model is computed in this way, it's known as a \emph{kernel machine}\index{kernel machine|see{kernel methods}} or \term{kernel methods}.

Note that in this dual\index{duality} expression of the solution, the optimal parameters $W^{\star}_{ij}$ and the feature functions $\phi_i(x)$ don't appear. Thus, we've successfully exchanged our feature-space quantities, an $(n_f+1)$-dimensional feature vector and the inverse of an $(n_f+1)\times(n_f+1)$-dimensional matrix, for sample-space quantities, an $\NR$-dimensional vector $ \kerm_{\tea \tra_1}$ and the inverse of an $\NR \times \NR$-dimensional matrix $\widetilde{\kerm}_{\tra_1 \tra_2}$.\footnote{In some situations,\label{footnote:kernel-vs-feature-functions} specifying and evaluating the kernel is much simpler than specifying and evaluating the feature functions. For instance, the \emph{Gaussian kernel}\index{kernel methods!kernel!Gaussian}, given by
\be\label{eq:gaussian-kernel-methods}
\kerm_{\delta_1\delta_2} \equiv \exp\!\le[- \frac{1}{2 \sigma^2} \sum_{i=1}^{n_0} \le(x_{i;\delta_1} - x_{i;\delta_2}\ri)^2  \ri]\, ,
\ee
implies an infinite-dimensional feature space, but can be evaluated by simply computing the squared distance between the $n_\text{in}$-dimensional input vectors and then exponentiating the result. (To see why the Gaussian kernel implies an infinite number of feature functions, we can express the squared distance as a sum of three inner products and then Taylor expand the exponential in those inner products; the terms in the Taylor expansion give the feature functions.) 
In this way, we see how computing the kernel can be far easier than representing the features explicitly.

In fact, any algorithm based on a linear kernel \eqref{eq:linear-kernel-methods} can be generalized by swapping the simple kernel for a more complicated kernel like the Gaussian kernel \eqref{eq:gaussian-kernel-methods}.
This is known as the \textbf{kernel trick}\index{kernel trick|see{kernel methods}}\index{kernel methods!kernel trick} and is a way to describe in the language of kernel methods how we generalized our simplest linear model \eqref{eq:dumb-linear-model-def} -- that was linear in the input -- to the more general linear model \eqref{eq:linear-model-def} -- that was nonlinear in the input.
}
This works because in our solution \eqref{eq:kernel-prediction-kernel-methods}, we actually only care about the inner product of the feature functions -- i.e.~the kernel -- and not the values of the features themselves.

By writing the linear model's prediction in terms of the kernel in \eqref{eq:kernel-prediction-kernel-methods}, we can interpret the prediction in terms of direct comparison with previously-seen examples. In particular, this solution computes the similarity of a new test input $x_{\tea}$ with all the training examples with $ \kerm_{\tea \tra_1}$ and then uses that similarity to linearly weight the true function values  from the training set $\y{i}{\tra_2}$ with the sample-space metric $\kermsub^{\tra_1 \tra_2}$.
For this reason, kernel methods are sometimes referred to as \emph{memory-based} methods\index{memory-based method|see{kernel methods}}\index{kernel methods!as a memory-based method} since they involve memorizing the entire training set.\footnote{Often, for a particular kernel method to be tractable, the model's predictions are made \emph{locally}, incorporating information mostly from the training samples nearest to the test sample of interest. Given the $\ast$-polation results of last section, it's not hard to imagine how such methods could be made to work well.

A canonical example of such a local method is \emph{$k$-nearest neighbors}\index{k-nearest neighbors@$k$-nearest neighbors|see{kernel methods}}\index{kernel methods!$k$-nearest neighbors} \cite{fix1952discriminatory,Cover1967NearestNP}, which is a special type of kernel method.  By only considering nearby training points, these kinds of local algorithms can skirt some of the impracticality that we've been pointing out for our exact Bayesian inference predictions as well as for our frozen NTK predictions.
It would be interesting to extend such local algorithms to the finite-width exact Bayesian inference that we discussed in \S\ref{sec:finite-posterior} or the finite-width gradient-based learning prediction that we'll discuss in \S\ref{ch:eot}.
}
This should be contrasted with the parameterized linear model solution \eqref{eq:kernel-prediction-kernel-methods-before-duality}, where we forget the training set samples and instead just explicitly store the optimal parameter values $W^{\star}_{ij}$.

Finally, note that 
there's ``no wiring'' in the prediction: the $z_i$ component of the prediction is entirely determined by the $y_i$ component of the training examples.
This is only implicit in the optimal weight matrix $W^{\star}_{ij}$ \eqref{eq:linear-regression-optimal} in the linear model solution \eqref{eq:kernel-prediction-kernel-methods-before-duality} but is explicit in the kernel method solution \eqref{eq:kernel-prediction-kernel-methods}.
This is one of many ways that such linear models and kernel methods are limited machine learning models.

\subsection{Infinite-Width Networks as Linear Models}\label{subsec:linear-at-infinity}

Surely, the kernel methods' prediction formula \eqref{eq:kernel-prediction-kernel-methods}  should seem awfully familiar to you:  it is precisely the same as the exact Bayesian mean prediction \eqref{eq:GP-mean} if we identify the kernel methods' kernel $k_{\delta_1\delta_2}$ with the Bayesian kernel $\Ti{\ker}{\delta_1\delta_2}{L}$, and it is exactly the same as the (neural tangent) kernel mean prediction~\eqref{eq:GD-frozen-mean} if we identify the kernel methods' kernel $k_{\delta_1\delta_2}$ with the frozen neural tangent kernel\index{frozen NTK} $\Ti{\NTKI}{\delta_1\delta_2}{L}$.\footnote{You may have noticed that our stochastic (neural tangent) kernel prediction\index{kernel methods!prediction} formula \eqref{eq:kernel-prediction} also depended on the network output at initialization and had a nonzero covariance. This is related to our earlier discussion in footnote~\ref{footnote:inverse-existing} that, when we're in the \emph{overparameterized}\index{overparameterization} regime with $(n_f+1)>\NR$, as is especially the case when we have an infinite number of features, the $(n_f+1)$-by-$(n_f+1)$ matrix matrix $M_{ij} \equiv\sum_{\tra\in\A}\fea_i(x_{\tra})\fea_j(x_{\tra})$  \eqref{eq:no-good-matrix-name} does not have a unique inverse. Thus, in this regime, the optimal weight matrix $W^{\star}$ is not unique: if we don't use the regulation trick \eqref{eq:inverse-of-M-regularized} to uniquely pick out one of the solutions, 
the prediction in the dual\index{duality} kernel description 
will have a dependence on the model's initialization.
} 
This finally provides a justification for the names of these objects as well as for the name of the current chapter.

Indeed, there is a very direct connection between these traditional linear models and kernel methods on the one hand and our (neural tangent) kernel learning of infinite-width models on the other hand.
 First, let's discuss the simpler case of the Bayesian kernel.
 As pointed out in \S\ref{subsec:NTKprediction}, this choice corresponds to treating only output-layer biases $b_{i}^{(L)}$ and weights $W_{ij}^{(L)}$ as trainable model parameters, and so the network output at initialization is given by
\be\label{eq:linear-model-BI}
z_i^{(L)}\le(x_{\delta};\theta\ri)=b_{i}^{(L)}+\sum_{j=1}^{n_{L-1}}W_{ij}^{(L)}\sigma^{(L-1)}_{j;\delta}=\sum_{j=0}^{n_{L-1}}W_{ij}^{(L)}\sigma^{(L-1)}_{j;\delta}\, ,
\ee
where on the right-hand side we defined $\sigma^{(L-1)}_{0;\delta}\equiv1$ and $W_{i0}^{(L)}\equiv b_i^{(L)}$. This is almost the same as our linear model~\eqref{eq:linear-model-def} except that the feature functions are \emph{random}: $\widehat{\fea}_{j;\delta} \equiv \sigma^{(L-1)}_{j;\delta}$.\index{feature function!random}
In particular, here we've \emph{hatted} these feature functions to emphasize that  they depend on the parameters in the hidden layers that are sampled from the initialization distribution at the beginning and then \emph{fixed}; this is sometimes called a \term{random feature model}.

In this case, the kernel methods' notion of the kernel is also stochastic\index{kernel methods!stochastic kernel}\index{kernel methods!stochastic kernel|seealso{random feature model}}
\begin{align}\label{eq:stochastic-kernel}
\widehat{k}_{\delta_1\delta_2}=&\Cb{L}\widehat{\fea}_{0;\delta_1}\widehat{\fea}_{0;\delta_2}+\frac{\CW{L}}{n_{L-1}}\sum_{j=1}^{n_{L-1}}\widehat{\fea}_{j;\delta_1}\widehat{\fea}_{j;\delta_2}\, \\
=&\Cb{L}+\CW{L}\le(\frac{1}{n_{L-1}}\sum_{j=1}^{n_{L-1}}\sigma^{(L-1)}_{j;\delta_1}\sigma^{(L-1)}_{j;\delta_2}\ri)\, ,\notag
\end{align}
where in the first line we have re-weighted the terms in the feature sum in the definition of the kernel \eqref{eq:kernel-with-features} by $\Cb{L}$ and $\CW{L}/ n_L$.\footnote{
    A more general definition of the kernel methods' kernel \eqref{eq:kernel-with-features} allows us to weight the contribution of each pair of feature functions as
\be\label{eq:kernel-with-features-weighted}
\kerm_{\delta_1\delta_2} \equiv \sum_{i,j=0}^{n_{f}} c_{ij}\, \fea_i\!\le(x_{\delta_1}\ri) \fea_j\!\le(x_{\delta_2}\ri) \, .
\ee
}
Note that we called this object \eqref{eq:stochastic-kernel} the \emph{stochastic metric}\index{metric!stochastic} \eqref{eq:general-stochastic-metric} when studying the RG flow of preactivations.
Now, taking an expectation over the initialization ensemble, in the infinite-width limit we have
\be
\E{\widehat{k}_{\delta_1\delta_2}}= \Cb{L}+\CW{L} \braket{\sigma_{\delta_1}\sigma_{\delta_2}}{L-1}=K^{(L)}_{\delta_1\delta_2}\, ,
\ee
where in the last equality we used the recursion for the kernel \eqref{eq:kernel-recursion-reminder-reprint}.\footnote{
    Note alternatively, by the central limit theorem, that the stochastic kernel $\widehat{k}_{\delta_1\delta_2}$ will be equal to the kernel $K^{(L)}_{\delta_1\delta_2}$ in the infinite-width limit without explicitly averaging over initializations. This \emph{self-averaging}\index{self-averaging} of the kernel is equivalent to the fact that the connected four-point correlator vanishes at infinite width. Here we see that such self-averaging of the kernel can also be thought of as arising from a sum over an infinite number of random features.
} In this way, we see how we can interpret exact Bayesian inference at infinite width as a simple linear model~\eqref{eq:linear-model-BI} of fixed random features.\index{feature function!random}\index{Bayesian inference!connection to linear models}

Now, let's give a linear model interpretation to gradient-based learning at infinite width\index{frozen NTK!features}. 
Since a linear model is linear in its parameters, we can more generally define the \textbf{random features}\index{feature function!random} by
\be\label{eq:feature-function-stochastic}
 \widehat{\fea}_{i,\mu}(x_{\delta}) \equiv \frac{\td \z{i}{\delta}{L}}{d \theta_\mu}  \, .
\ee
To be clear, this derivative is evaluated at initialization and these features are thus fixed in the infinite-width limit.
Explicitly, for an MLP the random features are given by
\begin{align}
 \widehat{\phi}_{i, \W{k_1 k_2}{\ell}}\!\!\!(x_{\delta}) &=  \le(\sum_{j_{L-1}, \ldots, j_{\ell+1}}\W{i j_{L-1}}{L} \ds{j_{L-1}}{\delta}{L-1} \cdots \W{j_{\ell+1} k_1}{\ell+1} \ds{k_1}{\delta}{\ell}\ri) \s{k_2}{\delta}{\ell-1} \, ,
 \label{eq:NTK-weight-features}
\end{align}
with the bias component given by setting $\s{0}{\delta}{\ell}\equiv1$ and $\W{k 0}{\ell}\equiv\bias{k}{\ell}$. As is apparent from this expression, these features are stochastic, depending on the specific values of the biases and weights at initialization. Note also that the Bayesian linear model \eqref{eq:linear-model-BI} only uses a subset of these features, $ \widehat{\phi}_{i, \W{i j}{L}}=\s{j}{\delta}{L-1}$, and thus is a much more limited and less expressive model.

Note further that these feature functions\index{feature function} $\widehat{\phi}_{i,\mu}(x_{\delta})$ are related to, but not exactly equivalent to,  the previous notion of feature we gave when we discussed \terminate{representation group flow} in \S\ref{sec:marginalization-group-flow}. In that case, our $\ell$-th-layer features corresponds to $\ell$-th-layer preactivations $\z{i}{\delta}{\ell}$ or activations $\s{i}{\delta}{\ell}$. However, here we see that the random feature functions~\eqref{eq:NTK-weight-features}
are proportional to $(\ell-1)$-th-layer activations $\s{i}{\delta}{\ell-1}$ but are also multiplied by objects from deeper layers.\footnote{
Going forward, when referring to the \emph{features} of a network, we will mean the kernel methods' notion of a feature function $\widehat{\fea}_{i,\mu}(x_{\delta})$,
rather than an activation $\sigma_i^{(\ell)}(x_\delta)$.\index{feature!vs.~feature function}\index{kernel methods!feature|see{feature function}}\index{representation learning!as the evolution of feature functions} Accordingly, we will now understand \neo{representation learning} to describe how such feature functions develop a data dependence during training.
}

As should be clear, the stochastic kernel\index{kernel methods!kernel!stochastic} associated with these features,
\begin{align}\label{eq:stochastic-kernel-ntk}
\widehat{\kerm}_{ij;\delta_1\delta_2}\equiv  \sum_{\mu,\nu}\lambda_{\mu\nu} \, \widehat{\phi}_{i,\mu}(x_{\delta_1}) \widehat{\phi}_{j,\nu}(x_{\delta_2})=\sum_{\mu,\nu}\lambda_{\mu\nu}\frac{\td \z{i}{\delta_1}{L}}{d \theta_\mu}\frac{\td \z{j}{\delta_2}{L}}{d \theta_\nu} \equiv \Tia{\NTK}{ij}{\delta_1\delta_2}{L}\, ,
\end{align}
is just the $L$-th-layer stochastic NTK~\eqref{eq:midNTH-definition}.\footnote{
    However please note importantly that at finite width the stochastic neural tangent \emph{kernel} $\widehat{k}_{ij;\delta_1\delta_2}=\Tia{\NTK}{i_1i_2}{\delta_1\delta_2}{L}$ is not fixed during training -- learning useful features from the data -- and randomly varies across initializations; hence  -- despite its name -- it is not actually a kernel.} Here, we have taken advantage of our more general definition of the kernel methods' kernel \eqref{eq:kernel-with-features-weighted} to incorporate the learning-rate tensor $\lambda_{\mu\nu}$ into the expression. Accordingly, at infinite width the NTK is frozen and diagonal in final layer neural indices, giving
\begin{align}
\kerm_{\delta_1\delta_2}\equiv  \sum_{\mu,\nu}\lambda_{\mu\nu} \, \widehat{\phi}_{i,\mu}(x_{\delta_1}) \, \widehat{\phi}_{i,\nu}(x_{\delta_2})=\sum_{\mu,\nu}\lambda_{\mu\nu}\frac{\td \z{i}{\delta_1}{L}}{d \theta_\mu}\frac{\td \z{i}{\delta_2}{L}}{d \theta_\nu}\equiv \Ti{\NTKI}{\delta_1\delta_2}{L} \, .
\end{align}
In this way, we see that at infinite width the fully-trained mean network output is just a linear model based on random features \eqref{eq:feature-function-stochastic}.\index{feature function!random}\index{infinite-width limit!connection to linear models}
In this sense, infinite-width neural networks are rather shallow in terms of model complexity, however deep they may appear.

Looking back, when we discussed the linear model at the beginning of this section, we had to introduce feature functions $\phi_i(x)$, designed using our knowledge of the task and data at hand, as a way to (pre-)process the input. This way, the parametric model that we learn is really simple, i.e.~linear.\footnote{Please don't confuse our discussion of \emph{linear models}\index{linear model} here and our discussion of linear vs.~nonlinear functions as in~\S\ref{subsec:star-polation}.

A linear model is a model that's linear in the model parameters \eqref{eq:linear-model-def} and has a dual\index{duality} kernel description that's linear in the true outputs $y_{i;\tra}$ in the training set $\A$ \eqref{eq:kernel-prediction-kernel-methods}; as we have seen, linear models are very simple and easy to solve analytically. As is clear from the definition~\eqref{eq:linear-model-def}, a linear model in general will be \emph{nonlinear} in the inputs $x$ for general nonlinear feature functions $\phi_i(x)$. Accordingly, linear models can compute nonlinear functions of their input. We saw this explicitly when we worked out how nonlinear $\ast$-polation works for smooth nonlinear networks in~\eqref{eq:polation-general}.  

In contrast, a \terminate{deep linear network} is a neural network that uses a $\linear$ activation function. Such networks compute linear functions of their input and thus may only linearly $\ast$-polate between training points, cf.~\eqref{eq:linear-polation}. However, since they are \emph{not} linear models (for $L>1$), the function they compute depends nonlinearly on the model parameters. Accordingly, their training dynamics can be somewhat complicated, and at finite width they even exhibit representation learning\index{representation learning!for deep linear networks}.\index{linear model!is not a deep linear network} %
}
We then reinterpreted this fit linear model in terms of its associated kernel, which itself has a natural interpretation as measuring similarity between our designed features.

However, for infinite-width networks we didn't \emph{design} the frozen NTK, and its associated features are \emph{random}. Instead, the network is defined by the architecture, hyperparameters, and biases and weights, and the path from those variables to the NTK and kernel prediction is filled with calculations. So the abstraction of the actual neural network seems like a very odd way to design a kernel.

Indeed, we've just learned that infinite-width neural networks can only make predictions that are linear in the true outputs from the training set; they are linear models that can only compute linear combinations of random features.
Of course, deep learning is exciting because it works on problems where classic machine learning methods have failed; it works in cases where we don't know how to design feature functions or kernels, or doing so would be too complicated. 
For neural networks to go beyond the kernel methods, they need to be able to \emph{learn} useful features \emph{from the data}, not just make use of a complicated linear combination of random features.\index{representation learning!vs.~kernel learning} 
Thus, in order for the feature functions \eqref{eq:feature-function-stochastic} to evolve during training 
-- that is, in order to have \emph{representation learning} -- we will need to go beyond kernel methods.

Luckily, finite-width networks are not kernel methods. Please now turn to the next chapter to find out exactly what they are instead.

%% file: Chp11-features/11_global.tex
\chapter{Representation Learning}\label{ch:features}

\epigraph{It can scarcely be denied that the supreme goal of all theory is to make the irreducible basic elements as simple and as few as possible without having to surrender the adequate representation of a single datum of experience.}{Albert Einstein\index{Einstein, Albert} in 
a 1933 lecture, ``On the Method of Theoretical Physics,''  
\cite{einstein-simple}.}
\noindent{}Last chapter,  we understood that
linear models
cannot learn features from data. Thus, the infinite-width limit is too simple to provide an adequate representation of deep learning; in order to include its irreducible basic element -- representation learning -- it is \emph{qualitatively} important to study finite-width networks.

In the first half of this chapter,  we'll analyze the leading correction to the gradient-descent update to the network output by extending its Taylor expansion to second order in the global learning rate $\eta$. After further seeing that a similar contribution arises in the first-order Taylor expansion of the update to the NTK,  we'll then
show that this correction is a finite-width effect.
 This upgrade of the NTK from fixed to dynamical indicates that for finite-width networks, the feature functions that comprise the NTK are themselves learning from the data over the course of training.

Unfortunately, the complete $\o{1/n}$ contribution to the dynamics further includes terms that arise from Taylor expanding the update to the network output to third order in the global learning rate $\eta$, and similarly Taylor expanding the update to the NTK to second order in $\eta$. While it's \emph{necessary} to include these contributions in order to actually compute the distribution of fully-trained finite-width networks, the $\o{\eta^2}$ expansion of the network output and the $\o{\eta}$ expansion of the NTK is \emph{sufficient} to qualitatively investigate the mechanism for representation learning in these models.

With that in mind, in order to separate the pedagogy of representation learning from the messy phenomenological details of the real MLP, we'll spend the second half of this chapter focusing on a simplified model that's equivalent to this $\o{\eta^2}$ truncation and gives a minimal qualitative picture of representation learning. 
These minimal models that we discuss form a valid and potentially useful class of machine learning models that perform representation learning, though annoyingly finite-width MLPs are not in this class.
Listening carefully to these real MLPs, we'll spend all of next chapter (\S\ref{ch:eot}) working out their  $\o{1/n}$ training dynamics in full intricate detail.

To begin, in  \S\ref{sec:dNTK} we'll work out this second-order-in-$\eta$ contribution to the update to preactivations and the first-order-in-$\eta$ contribution to the update to the NTK.
This lets us source all representation learning from a single irreducible basic element, 
the \neo{differential of the neural tangent kernel}\index{differential of the neural tangent kernel|seealso{meta kernel}} (dNTK)\index{dNTK|see{differential of the neural tangent kernel}}: just as the NTK governs the leading-order-in-$\eta$ dynamics of the preactivations, the dNTK governs the leading-order-in-$\eta$ dynamics of the NTK. %

After thusly identifying the dNTK as a driver of representation learning, in \S\ref{sec:dNTK-RG} we'll recursively determine its correlation with the preactivations as a function of network layer $\ell$. In detail, we'll first derive a stochastic forward equation for the dNTK and then evaluate the remaining recursions needed to determine the statistics of the joint preactivation-NTK-dNTK distribution at initialization. As such, this section mirrors the structure of our  \emph{RG-flow}\index{representation group flow} analysis in \S\ref{ch:ngp} and \S\ref{ch:NTKa}.
 Importantly, we'll see that all the statistics involving the dNTK are $\o{1/n}$ and thus only contribute  at finite width.

In \S\ref{sec:dNTK-criticality}, we'll apply the principles of criticality and universality to analyze the new dNTK recursions.
Since all of our hyperparameters have already been fixed by the parallel analysis of the preactivations in \S\ref{ch:eft-mlp} -- fixing the \terminate{initialization hyperparameters} --  and the NTK in \S\ref{ch:eft-ntk} -- fixing the training hyperparameters -- our focus here will be on evaluating the depth and width scaling of the dNTK statistics with these fixed hyperparameters. As you might guess, we'll find across our two universality classes (\S\ref{subsec:dntk_criticality_scale_invariant} and \S\ref{subsec:dntk_criticality_tanh_univ}) that the effect of the dNTK -- and therefore one source of representation learning -- is proportional to our effective theory cutoff\index{cutoff, effective theory}, the depth-to-width ratio $L/n$.

Having now firmly established that the NTK evolves at finite width -- and having worked out an important contribution to its dynamics -- in  \S\ref{sec:nonlinear-model},  we'll take a step back and look for a broader context, mirroring our discussion in \S\ref{sec:lazy-kernel} for infinite-width networks.
To that end, in  \S\ref{subsec:nonlinear-models} we'll introduce a class of \emph{nonlinear models}\index{nonlinear model} -- with a particular focus on the \emph{quadratic model}\index{nonlinear model!quadratic model} -- and thus minimally extend the traditional workhorse of machine learning, the linear model. This quadratic model provides a \emph{minimal model} of representation learning\index{representation learning!minimal model}\index{representation learning!minimal model}, 
independent of any neural-network abstraction. Moreover, these models are simple and completely analyzable, and yet are able to capture the essence of representation learning.

After solving the implied \emph{nearly-linear quadratic regression} problem, in \S\ref{subsec:nearly-kernel-methods} we'll further provide a dual description of the quadratic model solution, which we'll call \neo{nearly-kernel methods}. This will let us identify an object that corresponds to the dNTK in this minimal setting, and show us how to make test-set predictions with a \emph{trained} kernel that learns from the data. 
Overall, we hope this framework will be of further theoretical and practical interest as a new class of nearly-simple machine learning models that learn representations.

At this point, the connection between these nearly-kernel methods and finite-width networks -- at least at order $\eta^2$ -- will be nearly manifest, and in  \S\ref{subsec:nonlinear-at-finite} we'll make it explicit.
By doing so, we'll understand precisely how deep learning is a \emph{non-minimal} model of representation learning. Ultimately, we'll conclude that the power of deep learning is the \emph{deep} -- the inductive bias of the network architecture induced by the layer-to-layer RG flow -- providing a particularly good choice of initial features as a starting point for \emph{learning}. These observations will be quite helpful for us in interpreting our somewhat messy finite-width solution in the following chapter.

\section{Differential of the Neural Tangent Kernel}\label{sec:dNTK}
 \index{differential of the neural tangent kernel}
Recall that in the first step of gradient descent, the change in the $\ell$-th-layer parameters of any particular network is given by \eqref{eq:gd-update-lambda}
\be\label{eq:gd-update-theta-for-one-last-time}
\dtheta^{(\ell)}_{\mu}\equiv\theta^{(\ell)}_{\mu}(t=1)- \theta^{(\ell)}_{\mu}(t=0)=-\eta\sum_{\nu}\lambda_{\mu\nu}^{(\ell)}\le(\sum_{k=1}^{n_{L}}\sum_{\tra\in\A}\frac{\partial\L_{\A}}{\partial \z{k}{\tra}{L}}\frac{d\z{k}{\tra}{L}}{d\theta_{\nu}^{(\ell)}}\ri)\, .
\ee
In this section, it will be helpful to  specify explicitly which layer each parameter comes from.
In particular, here $\theta^{(\ell)}_{\mu}$ denotes either an $\ell$-th-layer bias $\theta_\mu^{(\ell)} \equiv \bias{i}{\ell}$ or an $\ell$-th-layer weight $\theta_\mu^{(\ell)} \equiv \W{ij}{\ell}$, and the $\ell$-th-layer model-parameter indices $\mu, \nu$  run over all the components of the bias vector $\bias{i}{\ell}$ \emph{and} the weight matrix $\W{ij}{\ell}$ in the $\ell$-th layer \emph{only}. Additionally, to emphasize that the learning-rate tensor\index{learning rate!learning-rate tensor} $\lambda_{\mu\nu}^{(\ell)}$ only connects the parameters within a given layer $\ell$, we've decorated it with a layer index for clarity. For now we'll let $\lambda_{\mu\nu}^{(\ell)}$ act arbitrarily within a layer, though ultimately we'll be interested in the case where it's diagonal,
with two \terminate{training hyperparameters} per layer, $\Lb{\ell}$ and $\LW{\ell}$, as usual.

As a further reminder, quantities without any explicit step argument are taken to be evaluated at initialization -- though sometimes we may also explicitly denote $t=0$ for extra emphasis -- and our sample-index notation is \emph{alpha-with-tilde} for the inputs in the training set, $\tra \in\A$, \emph{beta-with-dot} for inputs in the test set, $\tea\in\B$, and \emph{delta-with-no-decoration}
 for inputs that could be in either set, $\delta\in\D=\A\cup\B$. %

Now, to go beyond 
the infinite-width limit,
we'll need to expand the change in $\ell$-th-layer preactivations to \emph{second order} in the parameter update:
\begin{align}\label{eq:preactivation-change-second-order-in-model-parameter}
\dz{i}{\delta}{\ell}\equiv&\z{i}{\delta}{\ell}(t=1)-\z{i}{\delta}{\ell}(t=0)\, \\
=&\sum_{\ell_1=1}^{\ell} \sum_{\mu}\frac{d\z{i}{\delta}{\ell}}{d\theta_{\mu}^{(\ell_1)}}\dtheta_{\mu}^{(\ell_1)}+\frac{1}{2}\sum_{\ell_1, \ell_2 = 1}^\ell \sum_{\mu_1,\mu_2}\frac{d^2\!\z{i}{\delta}{\ell}}{d\theta^{(\ell_1)}_{\mu_1}d\theta_{\mu_2}^{(\ell_2)}}\dtheta_{\mu_1}^{(\ell_1)}\dtheta_{\mu_2}^{(\ell_2)}+\ldots\, .\notag
\end{align}
Note that the $\ell$-th-layer preactivations $\z{i}{\delta}{\ell}$ cannot depend on model parameters $\theta_{\mu}^{(\ell')}$ from layers $\ell'$ that are deeper than the $\ell$-th layer.  Thus, when $\ell' > \ell$, we have $d\z{i}{\delta}{\ell}/d\theta_{\mu}^{(\ell')}=0$, and so we truncated our layer sums in the above expression at $\ell$.

Next, we are going to slightly rewrite the parameter update equation \eqref{eq:gd-update-theta-for-one-last-time} for the parameters $\theta_{\mu}^{(\ell_a)}$ appearing in our preactivation expansion \eqref{eq:preactivation-change-second-order-in-model-parameter}, i.e.~for those parameters in layers $\ell_a\leq\ell$ that contribute. To do so, we'll make use of the chain rule to decompose the derivative of the output-layer preactivations $\z{k}{\tra}{L}$ with respect to the $\ell_a$-th-layer model parameters as
\be\label{eq:chain-rule-rules}
\frac{d\z{k}{\tra}{L}}{d\theta_{\nu}^{(\ell_a)}} =\sum_{j}\frac{\td\z{k}{\tra}{L}}{\td \z{j}{\tra}{\ell}}\frac{d\z{j}{\tra}{\ell}}{d\theta_{\nu}^{(\ell_a)}} \, ,
\ee
for an intermediate layer $\ell$ such that
$\ell_a \leq \ell$.
Using this decomposition, we can rewrite our parameter update \eqref{eq:gd-update-theta-for-one-last-time} as
\be\label{eq:parameter-update-rewritten-error-factor}
\dtheta_{\mu}^{(\ell_a)}=-\eta\sum_{\nu}\lambda_{\mu\nu}^{(\ell_a)}\le(\sum_{j,k,\tra}\frac{\partial\L_{\A}}{\partial \z{k}{\tra}{L}}\frac{\td\z{k}{\tra}{L}}{\td \z{j}{\tra}{\ell}}\frac{d\z{j}{\tra}{\ell}}{d\theta_{\nu}^{(\ell_a)}}\ri)=-\eta\sum_{\nu,j,\tra}\lambda_{\mu\nu}^{(\ell_a)}\,\Tia{\epsilon}{j}{\tra}{\ell}\frac{d\z{j}{\tra}{\ell}}{d\theta_{\nu}^{(\ell_a)}}\, ,
\ee
where in the last equality we
introduced an \emph{$\ell$-th-layer error factor}\index{error factor!$\ell$-th-layer}:
\be\label{eq:ellth-error-chain}
\Tia{\epsilon}{j}{\tra}{\ell}\equiv \sum_{k=1}^{n_{L}}\frac{\partial\L_{\A}}{\partial \z{k}{\tra}{L}}\frac{d\z{k}{\tra}{L}}{d\z{j}{\tra}{\ell}}= \frac{\td\L_{\A}}{\td \z{j}{\tra}{\ell}}\, .
\ee
Substituting this form of the parameter update \eqref{eq:parameter-update-rewritten-error-factor} into the $\ell$-th-layer preactivation update \eqref{eq:preactivation-change-second-order-in-model-parameter},
our second-order expansion becomes
\begin{align}\label{eq:preactivation-change-second-order-in-model-parameter-II}
\dz{i}{\delta}{\ell}=&-\eta\sum_{j,\tra}\le(\sum_{\ell_1=1}^\ell\sum_{\mu,\nu}\lambda_{\mu\nu}^{(\ell_1)}\frac{d\z{i}{\delta}{\ell}}{d\theta_{\mu}^{(\ell_1)}}\frac{d\z{j}{\tra}{\ell}}{d\theta_{\nu}^{(\ell_1)}}\ri)\Tia{\epsilon}{j}{\tra}{\ell}\, \\
&+\frac{\eta^2}{2}\sum_{j_1,j_2,\tra_1,\tra_2}\le(\sum_{\ell_1, \ell_2=1}^\ell\,\, \sum_{\substack{\mu_1,\nu_1, \\ \mu_2,\nu_2} }\lambda_{\mu_1\nu_1}^{(\ell_1)}\lambda_{\mu_2\nu_2}^{(\ell_2)}\frac{d^2\!\z{i}{\delta}{\ell}}{d\theta^{(\ell_1)}_{\mu_1}d\theta^{(\ell_2)}_{\mu_2}}\frac{d\z{j_1}{\tra_1}{\ell}}{d\theta^{(\ell_1)}_{\nu_1}}\frac{d\z{j_2}{\tra_2}{\ell}}{d\theta^{(\ell_2)}_{\nu_2}}\ri)\Tia{\epsilon}{j_1}{\tra_1}{\ell}\Tia{\epsilon}{j_2}{\tra_2}{\ell} \, \notag \\ 
&+\ldots\, ,\notag
\end{align}
which is \emph{quadratic} in such error factors.
Here, it was essential that we treated the parameters in a per-layer manner and that each learning-rate tensor\index{learning rate!learning-rate tensor!layer-independence} $\lambda_{\mu\nu}^{(\ell_a)}$ was restricted to a single layer $\ell_a$; 
had we not done that, our decomposition \eqref{eq:parameter-update-rewritten-error-factor} and update equation \eqref{eq:preactivation-change-second-order-in-model-parameter-II} would have been far more complicated.

Naturally, the object in the first parenthesis of the update equation \eqref{eq:preactivation-change-second-order-in-model-parameter-II} is the stochastic \emph{$\ell$-th-layer NTK}\index{neural tangent kernel!defined in conjunction with dNTK}\eqref{eq:midNTH-definition} 
\be\label{eq:ell-layer-ntk-def-with-layer-indices}
\Tia{\NTK}{i_1i_2}{\delta_1\delta_2}{\ell} \equiv \sum_{\ell_1=1}^\ell\sum_{\mu,\nu}\lambda_{\mu\nu}^{(\ell_1)}\frac{d\z{i_1}{\delta_1}{\ell}}{d\theta_{\mu}^{(\ell_1)}}\frac{d\z{i_2}{\delta_2}{\ell}}{d\theta_{\nu}^{(\ell_1)}} \, ,
\ee
as you know quite well by now,
though in this version of the definition we represent the sum over layers explicitly and the sum over parameter indices $\mu, \nu$ runs per layer.

In contrast, the object in the second parenthesis is new.\footnote{This object first appeared, unnamed, in both \cite{boris-nica} and \cite{dyer2019asymptotics} around the same time.
Here, we'll compute its recursion, determine its scaling with depth, and emphasize its physical importance by highlighting its connection to representation learning.
}
Let's call this object the stochastic \textbf{$\ell$-th-layer differential of the neural tangent kernel} (dNTK)\index{differential of the neural tangent kernel} and symbolize it as
\be\label{eq:dNTK-definition}
\Tia{\dNTK}{i_0i_1i_2}{\delta_0 \delta_1\delta_2}{\ell}\equiv\sum_{\ell_1, \ell_2=1}^\ell\,\, \sum_{\substack{\mu_1,\nu_1, \\ \mu_2,\nu_2} }\lambda_{\mu_1\nu_1}^{(\ell_1)}\lambda_{\mu_2\nu_2}^{(\ell_2)}\frac{d^2\!\z{i_0}{\delta_0}{\ell}}{d\theta^{(\ell_1)}_{\mu_1}d\theta^{(\ell_2)}_{\mu_2}}\frac{d\z{i_1}{\delta_1}{\ell}}{d\theta^{(\ell_1)}_{\nu_1}}\frac{d\z{i_2}{\delta_2}{\ell}}{d\theta^{(\ell_2)}_{\nu_2}} \, .
\ee
Here, the hats on both the NTK and the dNTK remind us that these objects are stochastic, depending on the particular realization of the model parameters at initialization.
Also, from its definition note that the dNTK is symmetric in its second and third paired set of indices $(i_1, \delta_1) \leftrightarrow (i_2, \delta_2)$, while the first neural-sample index $(i_0, \delta_0)$ is distinguished from the other two.

Using both definitions \eqref{eq:ell-layer-ntk-def-with-layer-indices} and \eqref{eq:dNTK-definition},  
our second-order expansion \eqref{eq:preactivation-change-second-order-in-model-parameter-II} can be more compactly written as
\be\label{eq:preactivation-updated-finite-width}
\dz{i}{\delta}{\ell}=-\eta\sum_{j,\tra}\Tia{\NTK}{ij}{\delta\tra}{\ell}\Tia{\epsilon}{j}{\tra}{\ell}+\frac{\eta^2}{2}\sum_{j_1,j_2,\tra_1,\tra_2}\Tia{\dNTK}{i j_1j_2}{\delta\tra_1\tra_2}{\ell}  \Tia{\epsilon}{j_1}{\tra_1}{\ell}\Tia{\epsilon}{j_2}{\tra_2}{\ell}+\ldots\, .
\ee
In other words, we have a power series in error factors. To ultimately understand how the preactivations evolve under gradient descent  at leading order in $1/n$, we'll actually need to extend this expansion to order $\eta^3$, which in turn will require that we introduce a few additional tensors. Rather than worry about that now, we'll put it off to \S\ref{ch:eot}. Regardless of those additional higher-order terms, from \eqref{eq:preactivation-updated-finite-width} we already see that we'll need to know the joint statistics of the preactivations -- encoding the error factors $\Tia{\epsilon}{j}{\tra}{\ell}$ -- the NTK $\Tia{\NTK}{ij}{\delta\tra}{\ell}$, and the dNTK $\Tia{\dNTK}{i j_1j_2}{\delta\tra_1\tra_2}{\ell}$.

Finally, as an explanation for our choice of name and symbol for the dNTK\index{differential of the neural tangent kernel!name},  consider the leading-order update to the $\ell$-th-layer NTK after a step of gradient descent: 
\begin{align}
\dbar\Tia{\NTKM}{i_1i_2}{\delta_1\delta_2}{\ell}\equiv&\Tia{\NTKM}{i_1i_2}{\delta_1\delta_2}{\ell}(t=1)-\Tia{\NTKM}{i_1i_2}{\delta_1\delta_2}{\ell}(t=0)\, \label{eq:dNTK-naming}\\
=&\sum_{\ell_1=1}^\ell \sum_{\mu_1}\frac{d\Tia{\NTKM}{i_1i_2}{\delta_1\delta_2}{\ell}}{d\theta_{\mu_1}^{(\ell_1)}}\dtheta_{\mu_1}^{(\ell_1)}+\ldots\, \nonumber\\
=&-\eta \sum_{\ell_1=1}^\ell \sum_{\mu_1}\le[\frac{d}{d\theta^{(\ell_1)}_{\mu_1}}\le(\sum_{\ell_2=1}^\ell\sum_{\mu_2,\nu_2}\lambda_{\mu_2\nu_2}^{(\ell_2)}\frac{d\z{i_1}{\delta_1}{\ell}}{d\theta_{\mu_2}^{(\ell_2)}}\frac{d\z{i_2}{\delta_2}{\ell}}{d\theta^{(\ell_2)}_{\nu_2}}\ri)\ri]\le[\sum_{\nu_1}\lambda_{\mu_1\nu_1}^{(\ell_1)}\sum_{j,\tra}\frac{d\z{j}{\tra}{\ell}}{d\theta_{\nu_1}^{(\ell_1)}}\Tia{\epsilon}{j}{\tra}{\ell}\ri]+\ldots\, \nonumber\\
=&-\eta\sum_{j,\tra}\le[ \sum_{\ell_1, \ell_2=1}^\ell\,\, \sum_{\substack{\mu_1,\nu_1, \\ \mu_2,\nu_2} }\lambda_{\mu_1\nu_1}^{(\ell_1)}\lambda_{\mu_2\nu_2}^{(\ell_2)}\frac{d^2\!\z{i_1}{\delta_1}{\ell}}{d\theta^{(\ell_1)}_{\mu_1}d\theta_{\mu_2}^{(\ell_2)}}\frac{d\z{i_2}{\delta_2}{\ell}}{d\theta^{(\ell_2)}_{\nu_2}}\frac{d\z{j}{\tra}{\ell}}{d\theta^{(\ell_1)}_{\nu_1}}\ri]\Tia{\epsilon}{j}{\tra}{\ell}\, \nonumber\\
&-\eta\sum_{j,\tra}\le[ \sum_{\ell_1, \ell_2=1}^\ell\,\, \sum_{\substack{\mu_1,\nu_1, \\ \mu_2,\nu_2} } \lambda_{\mu_1\nu_1}^{(\ell_1)}\lambda_{\mu_2\nu_2}^{(\ell_2)}\frac{d\z{i_1}{\delta_1}{\ell}}{d\theta_{\mu_2}^{(\ell_2)}}\frac{d^2\!\z{i_2}{\delta_2}{\ell}}{d\theta_{\mu_1}^{(\ell_1)}d\theta_{\nu_2}^{(\ell_2)}}\frac{d\z{j}{\tra}{\ell}}{d\theta_{\nu_1}^{(\ell_1)}}\ri]\Tia{\epsilon}{j}{\tra}{\ell}+\ldots\, \nonumber\\
=&-\eta\sum_{j,\tra}\le(\Tia{\dNTK}{i_1 i_2j}{\delta_1\delta_2\tra}{\ell}+\Tia{\dNTK}{i_2i_1j}{\delta_2\delta_1\tra}{\ell}\ri)\Tia{\epsilon}{j}{\tra}{\ell}+\ldots\, .\nonumber
\end{align}
Here, in the third line we inserted the definition of NTK \eqref{eq:ell-layer-ntk-def-with-layer-indices} and the parameter update \eqref{eq:parameter-update-rewritten-error-factor} for $\ell_1\leq\ell$, 
and on the final line we used the definition of the dNTK \eqref{eq:dNTK-definition}.
Thus we see that the dNTK -- when multiplied by the global learning rate and contracted with an $\ell$-th-layer error factor --  gives the update to the $\ell$-th-layer NTK after a step of gradient descent.\footnote{
    Please don't confuse our italicized, crossed, and unhatted notation, $\dbar\Tia{\NTKM}{i_1i_2}{\delta_1\delta_2}{\ell}$, representing the first update to the NTK, with our unitalicized, uncrossed, and hatted notation, $\Tia{\dNTK}{i_0i_1i_2}{\delta_0 \delta_1\delta_2}{\ell}$, representing the dNTK. In this chapter we will focus on the statistics of the dNTK, and we will not use this notation when evaluating the NTK dynamics in the following chapter.}

Since we know that the infinite-width NTK is \emph{frozen} $\NTK^{(\ell)} \to \NTKI^{(\ell)}$, the relation between the NTK update and the dNTK implies that the dNTK must be a finite-width effect, vanishing in the strict infinite-width limit $\dNTK^{(\ell)} \to 0$. Similarly, at infinite width we truncated the preactivation updates \eqref{eq:preactivation-updated-finite-width} to be linear in the global learning rate $\eta$, cf.~\eqref{eq:GD-preactivation-at-infty}. In the next section, we will verify all of this by computing the dNTK recursively and showing explicitly that $\dNTK^{(\ell)} = \o{1/n}$.

\section{RG Flow of the dNTK}\label{sec:dNTK-RG}
\index{representation group flow!of the dNTK}
As its title suggests, the structure of this section parallels \S\ref{ch:ngp} -- where we worked out the layer-to-layer representation group (RG) flow of the preactivation distribution $p\!\le(z^{(\ell)}\Big\vert \D\ri)$ -- and \S\ref{ch:NTKa} -- where we worked out the layer-to-layer RG flow of the NTK-preactivation joint distribution $p\!\le(z^{(\ell)}, \NTK^{(\ell)}\Big\vert \D\ri)$. Specifically, we will now work out the effective $\ell$-th-layer joint distribution of the preactivations, the NTK, and the dNTK:
\be
p\Big(z^{(\ell)}, \, \NTK^{(\ell)},\, \dNTK^{(\ell)}\Big\vert \D\Big)\, .
\ee 
This analysis is important for two reasons: \emph{(i)} firstly, understanding the statistics of this $\ell$-th-layer joint distribution at order $1/n$ is a necessary prerequisite for understanding the leading nontrivial finite-width corrections for deep neural networks trained with gradient-based learning; \emph{(ii)} secondly, in \S\ref{sec:nonlinear-model} we will see that a nonvanishing dNTK is sufficient for a network to exhibit representation learning, and  thus by showing that the dNTK is of order $1/n$, we will firmly establish that the leading-order finite-width effective theory is able to describe this essential property of deep learning.

Zeroth, we'll establish the stochastic iteration equation for the dNTK (\S\ref{subsec:dNTK-recursions}). Then, beginning our statistical analysis, first we'll see that the dNTK vanishes identically in the first layer (\S\ref{subsec:first-layer-zero-dNTK}). Second, we'll see that there's a nontrivial cross correlation between the dNTK and the preactivations in the second layer (\S\ref{subsec:second-layer-nonzero-dNTK}). Third and finally, we'll work out a general recursion that controls the accumulation of such dNTK-preactivation cross correlations in deeper layers (\S\ref{subsec:deeper-layer-growing-dNTK}).

\setcounter{subsection}{-1}
\subsection{Forward Equation for the dNTK}\label{subsec:dNTK-recursions}
Just as we needed to derive a stochastic forward iteration equation for the NTK \eqref{eq:NTH-recursion-without-expectation}  in \S\ref{sec:NTH-recursions} before working out recursions for its statistics, here we'll derive such an equation for the dNTK.

Let's start by writing out the definition of the dNTK\index{differential of the neural tangent kernel!iteration equation}\index{differential of the neural tangent kernel!iteration equation|seealso{forward equation}}~\eqref{eq:dNTK-definition} at layer $(\ell+1)$:
\be\label{eq:dNTK-definition-again}
\Tia{\dNTK}{i_0i_1i_2}{\delta_0 \delta_1\delta_2}{\ell+1}\equiv\sum_{\ell_1,\ell_2=1}^{\ell+1}\le[\sum_{\substack{\mu_1,\nu_1, \\ \mu_2,\nu_2} }\lambda_{\mu_1\nu_1}^{(\ell_1)}\lambda_{\mu_2\nu_2}^{(\ell_2)}\frac{d^2\!\z{i_0}{\delta_0}{\ell+1}}{d\theta_{\mu_1}^{(\ell_1)}d\theta_{\mu_2}^{(\ell_2)}}\frac{d\z{i_1}{\delta_1}{\ell+1}}{d\theta_{\nu_1}^{(\ell_1)}}\frac{d\z{i_2}{\delta_2}{\ell+1}}{d\theta_{\nu_2}^{(\ell_2)}}\ri]\, .
\ee
To determine its forward equation, we need to explicitly evaluate the derivatives with respect to the  $(\ell+1)$-th-layer parameters and also rewrite all the $(\ell+1)$-th-layer quantities in terms of the $\ell$-th-layer quantities using the chain rule. Depending on the values of $\ell_1$ and $\ell_2$, there are thus three cases to consider 
for the double summation over layers. 

First, when both layers are maximal $\ell_1=\ell_2=\ell+1$ there is no contribution. Recalling 
for one final time the preactivation forward equation, %
\be\label{eq:mlp-foward-pass-reprint-lol}
\z{i}{\delta}{\ell+1} = \bias{i}{\ell+1} + \sum_{j=1}^{n_\ell}\W{ij}{\ell+1} \s{j}{\delta}{\ell} \, ,
\ee
we see that the $(\ell+1)$-th-layer preactivations are always linear in 
the $(\ell+1)$-th-layer model parameters $\theta^{(\ell+1)}_{\mu}$.
Thus, in this case the second derivative in the dNTK definition \eqref{eq:dNTK-definition-again} will vanish.

Second, when $\ell_1=\ell+1$ and $\ell_2 < \ell +1$, there is a contribution from the $(\ell+1)$-th-layer weights but not from the $(\ell+1)$-th-layer biases. Considering the bias $\theta_{\mu_1}^{(\ell_1)}=\bias{j}{\ell+1}$, the $(\ell+1)$-th-layer derivative gives a Kronecker delta
\be
\frac{\td \z{i}{\delta}{\ell+1}}{\td \bias{j}{\ell+1}}=\delta_{ij}\, ,
\ee
and so the second derivative again vanishes
\be
\frac{\td^2 \z{i}{\delta}{\ell+1}}{\td \bias{j}{\ell+1} d\theta_{\mu_2}^{(\ell_2)}}= 0\, .
\ee
Instead considering the weight matrix $\theta_{\mu_1}^{(\ell_1)}=\W{jk}{\ell+1}$, the $(\ell+1)$-th-layer derivative is not a constant
\begin{align}\label{eq:first-derivative-mundane-other}
\frac{\td \z{i}{\delta}{\ell+1}}{\td \W{jk}{\ell+1}}=&\delta_{ij}\s{k}{\delta}{\ell}\, .
\end{align}
Thus, the second derivative evaluates to something nontrivial
\begin{align}\label{eq:second-derivative-mundane}
\frac{\td^2 \z{i}{\delta}{\ell+1}}{\td \W{jk}{\ell+1}\td \theta_{\mu_2}^{(\ell_2)}}=&\delta_{ij}\ds{k}{\delta}{\ell}\frac{\td \z{k}{\delta}{\ell}}{\td \theta_{\mu_2}^{(\ell_2)}}\, ,
\end{align}
while the remaining first derivative
gives
\be
\frac{\td \z{i}{\delta}{\ell+1}}{\td \theta_{\nu_2}^{(\ell_2)}}=\sum_{k}\W{ik}{\ell+1}\ds{k}{\delta}{\ell}\frac{\td \z{k}{\delta}{\ell}}{\td \theta_{\nu_2}^{(\ell_2)}}\, ,\label{eq:first-derivative-mundane}
\ee
with the use of the chain rule.
Plugging in these three derivative evaluations \eqref{eq:first-derivative-mundane-other}, \eqref{eq:second-derivative-mundane}, and \eqref{eq:first-derivative-mundane} to evaluate terms in the dNTK definition \eqref{eq:dNTK-definition-again} with $\ell_1=\ell+1$ and $\ell_2< \ell+1$, we find
\begin{align}\label{eq:dNTK-mid-one-final-layer-term}
&\sum_{\ell_2=1}^\ell \sum_{j,k}\lambda_{\W{jk}{\ell+1} \W{j k}{\ell+1}}\sum_{\mu_2,\nu_2} \lambda_{\mu_2\nu_2}^{(\ell_2)}\frac{\td^2 \z{i_0}{\delta_0}{\ell+1}}{\td \W{jk}{\ell+1}\td \theta_{\mu_2}^{(\ell_2)}}\frac{\td \z{i_1}{\delta_1}{\ell+1}}{\td \W{jk}{\ell+1}}\frac{\td \z{i_2}{\delta_2}{\ell+1}}{\td \theta_{\nu_2}^{(\ell_2)}}\, \notag \\
=&\frac{\LW{\ell+1}}{n_{\ell}}\sum_{\ell_2=1}^\ell\sum_{j,k}\sum_{\mu_2,\nu_2} \lambda_{\mu_2\nu_2}^{(\ell_2)}\le(\delta_{i_0j}\, \ds{k}{\delta_0}{\ell}\frac{\td \z{k}{\delta_0}{\ell}}{\td \theta_{\mu_2}^{(\ell_2)}}\ri)\le(\delta_{i_1j}\,\s{k}{\delta_1}{\ell}\ri)\le(\sum_{k_2}\W{i_2k_2}{\ell+1}\ds{k_2}{\delta_2}{\ell}\frac{\td \z{k_2}{\delta_2}{\ell}}{\td \theta_{\nu_2}^{(\ell_2)}}\ri)\, \notag\\
=&\frac{\LW{\ell+1}}{n_{\ell}}\delta_{i_0i_1}\sum_{k_0,k_2}\W{i_2k_2}{\ell+1}\ds{k_0}{\delta_0}{\ell}\s{k_0}{\delta_1}{\ell}\ds{k_2}{\delta_2}{\ell}\Tia{\NTK}{k_0k_2}{\delta_0\delta_2}{\ell}\, .
\end{align}
To get this result, on the second line we implemented
our choice of a single intralayer learning rate for the weights \eqref{eq:diag_LR},
\be
\lambda_{\W{j_1k_1}{\ell+1} \W{j_2 k_2}{\ell+1}}=\delta_{j_1 j_2}\delta_{k_1 k_2} \frac{\LW{\ell+1}}{n_{\ell}} \, ,
\ee
importantly rescaled by $n_{\ell}$, and on the third line we used the definition of the stochastic NTK \eqref{eq:ell-layer-ntk-def-with-layer-indices} and relabeled a dummy index. %
By symmetry, there must be a similar contribution when instead  $\ell_2=\ell+1$ and $\ell_1 < \ell+1$. This term is given by \eqref{eq:dNTK-mid-one-final-layer-term} after swapping neural-sample index pairs $(i_1,\delta_1)\leftrightarrow(i_2,\delta_2)$.

Third and finally, when both $\ell_1 < \ell+1$ and $\ell_2 < \ell+1$ both the biases and the weights contribute to the second derivative. When $\theta_\mu^{(\ell_1)}$ and $\theta_\nu^{(\ell_2)}$ are not from the $(\ell+1)$-th layer, we computed their first derivative in~\eqref{eq:first-derivative-mundane}, and their second derivative is given by
\be\label{eq:second-derivative-all-previous}
\frac{\td^2 \z{i}{\delta}{\ell+1}}{\td \theta_{\mu_1}^{(\ell_1)}\td \theta_{\mu_2}^{(\ell_2)}}=\sum_{k}\W{ik}{\ell+1}\dds{k}{\delta}{\ell}\frac{\td \z{k}{\delta}{\ell}}{\td \theta_{\mu_1}^{(\ell_1)}}\frac{\td \z{k}{\delta}{\ell}}{\td \theta_{\mu_2}^{(\ell_2)}}+\sum_{k}\W{ik}{\ell+1}\ds{k}{\delta}{\ell}\frac{\td^2 \z{k}{\delta}{\ell}}{\td \theta_{\mu_1}^{(\ell_1)}\td \theta_{\mu_2}^{(\ell_2)}} \, .
\ee
Multiplying these second derivative terms by the learning-rate tensors $\lambda_{\mu_1\nu_1}^{(\ell_1)}\,\lambda_{\mu_2\nu_2}^{(\ell_2)}$ and by the appropriate first derivatives~\eqref{eq:first-derivative-mundane},  and implementing all the sums over  $\ell_1$, $\ell_2$, $\mu_1$, $\nu_1$, $\mu_2$, $\nu_2$ in the dNTK definition \eqref{eq:dNTK-definition-again}, the first term from  \eqref{eq:second-derivative-all-previous} gives a contribution of
\be\label{eq:dNTK-mid-no-final-layer-term-1}
\sum_{k_0,k_1,k_2}\Ti{W}{i_0 k_0}{\ell+1}\Ti{W}{i_1 k_1}{\ell+1}\Ti{W}{i_2 k_2}{\ell+1}\dds{k_0}{\delta_0}{\ell}\ds{k_1}{\delta_1}{\ell}\ds{k_2}{\delta_2}{\ell}\Tia{\NTK}{k_0k_1}{\delta_0\delta_1}{\ell}\Tia{\NTK}{k_0k_2}{\delta_0\delta_2}{\ell}\, ,
\ee
where we made use of the NTK definition \eqref{eq:ell-layer-ntk-def-with-layer-indices}  twice, 
while the second term from \eqref{eq:second-derivative-all-previous} gives a contribution of
\be\label{eq:dNTK-mid-no-final-layer-term-2}
\sum_{k_0,k_1,k_2}\Ti{W}{i_0 k_0}{\ell+1}\Ti{W}{i_1 k_1}{\ell+1}\Ti{W}{i_2 k_2}{\ell+1}\ds{k_0}{\delta_0}{\ell}\ds{k_1}{\delta_1}{\ell}\ds{k_2}{\delta_2}{\ell}\Tia{\dNTK}{k_0k_1k_2}{\delta_0\delta_1\delta_2}{\ell}\, ,
\ee
where we made use of the dNTK definition \eqref{eq:dNTK-definition} once.

Combining our three types of contributions \eqref{eq:dNTK-mid-one-final-layer-term}, \eqref{eq:dNTK-mid-no-final-layer-term-1}, and \eqref{eq:dNTK-mid-no-final-layer-term-2}, we get a rather involved stochastic iteration equation:
\begin{align}\label{eq:forward-eq-for-dNTK}
\Tia{\dNTK}{i_0i_1i_2}{\delta_0\delta_1\delta_2}{\ell+1}=&\sum_{k_0,k_1,k_2}\Ti{W}{i_0 k_0}{\ell+1}\Ti{W}{i_1 k_1}{\ell+1}\Ti{W}{i_2 k_2}{\ell+1}\ds{k_0}{\delta_0}{\ell}\ds{k_1}{\delta_1}{\ell}\ds{k_2}{\delta_2}{\ell}\Tia{\dNTK}{k_0k_1k_2}{\delta_0\delta_1\delta_2}{\ell}\, \\
&+\sum_{k_0,k_1,k_2}\Ti{W}{i_0 k_0}{\ell+1}\Ti{W}{i_1 k_1}{\ell+1}\Ti{W}{i_2 k_2}{\ell+1}\dds{k_0}{\delta_0}{\ell}\ds{k_1}{\delta_1}{\ell}\ds{k_2}{\delta_2}{\ell}\Tia{\NTK}{k_0k_1}{\delta_0\delta_1}{\ell}\Tia{\NTK}{k_0k_2}{\delta_0\delta_2}{\ell}\, \nonumber\\
&+\frac{\lambda_{W}^{(\ell+1)}}{n_{\ell}}\delta_{i_0i_1}\sum_{k_0,k_2}\Ti{W}{i_2 k_2}{\ell+1}\ds{k_0}{\delta_0}{\ell}\s{k_0}{\delta_1}{\ell}\ds{k_2}{\delta_2}{\ell}\Tia{\NTK}{k_0k_2}{\delta_0\delta_2}{\ell}\, \nonumber\\
&+\frac{\lambda_{W}^{(\ell+1)}}{n_{\ell}}\delta_{i_0i_2}\sum_{k_0,k_1}\Ti{W}{i_1 k_1}{\ell+1}\ds{k_0}{\delta_0}{\ell}\ds{k_1}{\delta_1}{\ell}\s{k_0}{\delta_2}{\ell}\Tia{\NTK}{k_0k_1}{\delta_0\delta_1}{\ell}\, .\nonumber
\end{align}
This is the \textbf{forward equation for the dNTK}\index{forward equation!dNTK}, and we're next going to work out the recursions that determine its statistics.

\subsection{First Layer: Zero dNTK}
\label{subsec:first-layer-zero-dNTK}
Recall from~\S\ref{sec:first-layer-gaussian} and~\S\ref{sec:first-layer-deterministic-NTK} that at initialization the first-layer preactivations,
\be\label{eq:first-layer-preactivation-def-reprint-for-one-last-time}\index{forward equation!MLP preactivations}
\z{i}{\delta}{1} \equiv \bias{i}{1}+\sum_{k=1}^{n_{0}}\W{ik}{1}\x{k}{\delta}\, , %
\ee
are distributed according to a zero-mean \terminate{Gaussian distribution}~\eqref{eq:first-layer-distribution-HS-derivation} and that the NTK $\Tia{\NTK}{i_1i_2}{\delta_1\delta_2}{1}=\delta_{i_1i_2}\Ti{\NTKM}{\delta_1\delta_2}{1}$ is deterministic~\eqref{eq:NTHinitial}. 

As we discussed just before, since the preactivations are linear in the model parameters, their second derivative must vanish. Thus, the dNTK trivially vanishes in the first layer:
\be\label{eq:dNTK-initial}
\Tia{\dNTK}{i_0i_1i_2}{\delta_0\delta_1\delta_2}{1}\equiv\sum_{\substack{\mu_1,\nu_1, \\ \mu_2,\nu_2} }\lambda_{\mu_1\nu_1}^{(1)}\lambda_{\mu_2\nu_2}^{(1)}\frac{d^2\!\z{i_0}{\delta_0}{1}}{d\theta_{\mu_1}^{(1)}d\theta_{\mu_2}^{(1)}}\frac{d\z{i_1}{\delta_1}{1}}{d\theta_{\nu_1}^{(1)}}\frac{d\z{i_2}{\delta_2}{1}}{d\theta_{\nu_2}^{(1)}}=0\, .
\ee
This gives the initial condition for our recursions.

Note that this result should have been expected as the first-layer NTK~\eqref{eq:NTHinitial} is independent of the model parameters, and thus cannot change with any training. As we saw before for the first-layer preactivations and first-layer NTK -- zero-mean Gaussian and fixed, respectively -- this first-layer result for the dNTK will be representative of its infinite-width limit for all layers.

\subsection{Second Layer: Nonzero dNTK}
\label{subsec:second-layer-nonzero-dNTK}
Now, let's analyze the 
dNTK~\eqref{eq:forward-eq-for-dNTK} in the second layer.
Remembering again that the first-layer NTK~\eqref{eq:NTHinitial} is deterministic and diagonal in its neural indices as $\Tia{\NTK}{i_1i_2}{\delta_1\delta_2}{1}=\delta_{i_1i_2}\Ti{\NTKM}{\delta_1\delta_2}{1}$, and remembering for the first time
that the dNTK vanishes in the first layer
$\Tia{\dNTK}{i_0i_1i_2}{\delta_0\delta_1\delta_2}{1}=0$ from  \eqref{eq:dNTK-initial}, the forward equation~\eqref{eq:forward-eq-for-dNTK} in the second layer simplifies to
\begin{align}\label{eq:forward-eq-for-dNTK-second-layer}
\Tia{\dNTK}{i_0i_1i_2}{\delta_0\delta_1\delta_2}{2}=&\Ti{\NTKM}{\delta_0\delta_1}{1}\Ti{\NTKM}{\delta_0\delta_2}{1}\sum_{k=1}^{n_{1}}\Ti{W}{i_0 k}{2}\Ti{W}{i_1 k}{2}\Ti{W}{i_2 k}{2}\dds{k}{\delta_0}{1}\ds{k}{\delta_1}{1}\ds{k}{\delta_2}{1}\, \nonumber\\
&+\delta_{i_0i_1}\frac{\lambda_{W}^{(2)}}{n_1}\Ti{\NTKM}{\delta_0\delta_2}{1}\sum_{k=1}^{n_{1}}\Ti{W}{i_2 k}{2}\ds{k}{\delta_0}{1}\s{k}{\delta_1}{1}\ds{k}{\delta_2}{1}\, \nonumber\\
&+\delta_{i_0i_2}\frac{\lambda_{W}^{(2)}}{n_1}\Ti{\NTKM}{\delta_0\delta_1}{1}\sum_{k=1}^{n_{1}}\Ti{W}{i_1 k}{2}\ds{k}{\delta_0}{1}\ds{k}{\delta_1}{1}\s{k}{\delta_2}{1}\, .
\end{align}
Interestingly, since each term has an odd number of weights, the mean of the dNTK will vanish, and we'll have to look at cross correlations to find leading dNTK statistics that are non-vanishing.

The simplest cross correlation is with a single preactivation.
Considering the product of the second-layer dNTK~\eqref{eq:forward-eq-for-dNTK-second-layer} with second-layer preactivations,
\be\label{eq:second-layer-preactivations-reprint-for-one-last-time}\index{forward equation!MLP preactivations}
\z{i}{\delta}{2}=\bias{i}{2}+\sum_{k=1}^{n_{1}}\W{ik}{2}\s{k}{\delta}{1}\, ,%
\ee
and taking an expectation, we find
\begin{align}\label{eq:second-layer-dNTK-mid}
&\E{\Tia{\dNTK}{i_0i_1i_2}{\delta_0\delta_1\delta_2}{2}\z{i_3}{\delta_3}{2}} \notag \\ %
=&\Ti{\NTKM}{\delta_0\delta_1}{1}\Ti{\NTKM}{\delta_0\delta_2}{1}\le(\frac{\CW{2}}{n_1}\ri)^2\le(\delta_{i_0 i_3}\delta_{i_1i_2}+\delta_{i_0 i_1}\delta_{i_2 i_3}+\delta_{i_0 i_2}\delta_{i_1 i_3}\ri)\sum_{k=1}^{n_1}\E{\dds{k}{\delta_0}{1}\ds{k}{\delta_1}{1}\ds{k}{\delta_2}{1}\s{k}{\delta_3}{1}}\, \notag\\
&+\frac{\lambda_{W}^{(2)}}{n_1}\Ti{\NTKM}{\delta_0\delta_2}{1}\delta_{i_0i_1}\delta_{i_2i_3}\frac{\CW{2}}{n_1} \sum_{k=1}^{n_{1}}\E{\ds{k}{\delta_0}{1}\s{k}{\delta_1}{1}\ds{k}{\delta_2}{1}\s{k}{\delta_3}{1}}\, \notag\\
&+\frac{\lambda_{W}^{(2)}}{n_1}\Ti{\NTKM}{\delta_0\delta_1}{1}\delta_{i_0i_2}\delta_{i_1i_3}\frac{\CW{2}}{n_1} \sum_{k=1}^{n_{1}}\E{\ds{k}{\delta_0}{1}\ds{k}{\delta_1}{1}\s{k}{\delta_2}{1}\s{k}{\delta_3}{1}}\, \notag \\
=&\frac{1}{n_1}\le(\delta_{i_0 i_3}\delta_{i_1i_2}+\delta_{i_0 i_1}\delta_{i_2 i_3}+\delta_{i_0 i_2}\delta_{i_1 i_3}\ri)\CW{2}\Ti{\NTKM}{\delta_0\delta_1}{1}\CW{2}\Ti{\NTKM}{\delta_0\delta_2}{1}\bra \sigma''_{\delta_0}\sigma'_{\delta_1}\sigma'_{\delta_2}\sigma_{\delta_3}\ket_{G^{(1)}}\, \notag\\
&+\frac{1}{n_1}\delta_{i_0i_1}\delta_{i_2i_3}\lambda_{W}^{(2)}\CW{2}\Ti{\NTKM}{\delta_0\delta_2}{1} \bra \sigma'_{\delta_0}\sigma_{\delta_1}\sigma'_{\delta_2}\sigma_{\delta_3}\ket_{G^{(1)}}\, \notag\\
&+\frac{1}{n_1}\delta_{i_0i_2}\delta_{i_1i_3}\lambda_{W}^{(2)}\CW{2}\Ti{\NTKM}{\delta_0\delta_1}{1} \bra \sigma'_{\delta_0}\sigma'_{\delta_1}\sigma_{\delta_2}\sigma_{\delta_3}\ket_{G^{(1)}}\, 
\end{align}
To get this final result, in the first equality we dropped the bias term from~\eqref{eq:second-layer-preactivations-reprint-for-one-last-time}, since it vanishes under the expectation,  and performed various Wick contractions of the weights using
$\E{W^{(2)}_{i_1 j_1}W^{(2)}_{i_2 j_2}} =\delta_{i_1 i_2} \delta_{j_1 j_2}C_{W}^{(2)}/n_{1}$~\eqref{eq:weight-variance-def-naive}.
For the second equality, we remembered that the first-layer preactivation distribution is a zero-mean Gaussian with a two-point correlator that's diagonal in neural indices, $\E{\z{i_1}{\delta_1}{1}\z{i_2}{\delta_2}{1}} = \delta_{i_1i_2}G_{\delta_1\delta_2}^{(1)}$~\eqref{eq:first-layer-distribution-HS-derivation}, and used this to swap full expectations for Gaussian expectations and then performed the sums.

As we did before for the NTK variance \eqref{eq:NTH-variance-decomposition-second} and the NTK-preactivation cross correlation \eqref{eq:F-and-D-decomposition}, it is convenient to decompose this dNTK-preactivation cross correlation~\eqref{eq:second-layer-dNTK-mid} into two tensors with sample indices only:\index{tensor decomposition!dNTK-preactivation $P$/$Q$}
\begin{align}\label{eq:dntk-second-layer-decomposition}
\E{\Tia{\dNTK}{i_0i_1i_2}{\delta_0\delta_1\delta_2}{2}\z{i_3}{\delta_3}{2}}\equiv&\frac{1}{n_1}\le[\delta_{i_0 i_3}\delta_{i_1i_2}\dNTKP{\delta_0\delta_1\delta_2\delta_3}{2}+\delta_{i_0i_1}\delta_{i_2i_3}\dNTKQ{\delta_0\delta_1\delta_2\delta_3}{2}+\delta_{i_0i_2}\delta_{i_1i_3}\dNTKQ{\delta_0\delta_2\delta_1\delta_3}{2}\ri]\, .
\end{align}
Comparing with our explicit formula for the second-layer cross correlation~\eqref{eq:second-layer-dNTK-mid}, we see that these tensors have the following definitions,
\begin{align}
\dNTKP{\delta_0\delta_1\delta_2\delta_3}{2}\equiv&\le( \CW{2}\ri)^2 \Ti{\NTKM}{\delta_0\delta_1}{1}\Ti{\NTKM}{\delta_0\delta_2}{1}\bra \sigma''_{\delta_0}\sigma'_{\delta_1}\sigma'_{\delta_2}\sigma_{\delta_3}\ket_{G^{(1)}}\, ,\label{eq:dNTKP-second-layer}\\
\dNTKQ{\delta_0\delta_1\delta_2\delta_3}{2}\equiv&\le( \CW{2}\ri)^2 \Ti{\NTKM}{\delta_0\delta_1}{1}\Ti{\NTKM}{\delta_0\delta_2}{1}\bra \sigma''_{\delta_0}\sigma'_{\delta_1}\sigma'_{\delta_2}\sigma_{\delta_3}\ket_{G^{(1)}} \label{eq:dNTKQ-second-layer}+\lambda_{W}^{(2)}\CW{2}\Ti{\NTKM}{\delta_0\delta_2}{1} \bra \sigma'_{\delta_0}\sigma_{\delta_1}\sigma'_{\delta_2}\sigma_{\delta_3}\ket_{G^{(1)}}\, ,
\end{align}
and that they are manifestly of order one.\footnote{The cross-correlation tensor $\dNTKP{\delta_0\delta_1\delta_2\delta_3}{2}$~\eqref{eq:dNTKP-second-layer} -- and more generally $\dNTKP{\delta_0\delta_1\delta_2\delta_3}{\ell}$ in deeper layers
-- is only symmetric under the exchange of its middle samples indices $\delta_1\leftrightarrow \delta_2$.
Meanwhile, it's manifestly clear from~\eqref{eq:dNTKQ-second-layer} that the other cross-correlation tensor $\dNTKQ{\delta_0\delta_1\delta_2\delta_3}{2}$ has no symmetry whatsoever.} Overall, the dNTK-preactivation cross correlation \eqref{eq:dntk-second-layer-decomposition} is of order $1/n_1$, vanishing in the strict infinite-width limit $n_1 \to \infty$.

These tensors \eqref{eq:dNTKP-second-layer} and \eqref{eq:dNTKQ-second-layer} -- and their deeper-layer siblings -- control all the leading finite-width correlation between the preactivations and the dNTK, and in fact encapsulate the entire effect of the dNTK at order $1/n$. As we'll show next, any other dNTK-preactivation cross correlators, e.g.~$\E{\Tia{\dNTK}{i_0i_1i_2}{\delta_0\delta_1\delta_2}{\ell}\z{j_3}{\delta_3}{\ell}\z{j_4}{\delta_4}{\ell}\z{j_5}{\delta_5}{\ell}}$, can always be expressed in terms of a combination of $\dNTKP{}{\ell}$ and $\dNTKQ{}{\ell}$ at this order. %

\subsection{Deeper Layers: Growing dNTK}
\label{subsec:deeper-layer-growing-dNTK}
As before with~\S\ref{sec:first-layer-gaussian}$\,\parallel\,$\S\ref{sec:first-layer-deterministic-NTK}$\,\parallel\,$\S\ref{subsec:first-layer-zero-dNTK} and~\S\ref{sec:second-layer-non-gaussian}$\,\parallel\,$\S\ref{sec:second-layer-fluctuating-NTK}$\,\parallel\,$\S\ref{subsec:second-layer-nonzero-dNTK}, this section parallels our other sections analyzing the RG flow in deeper layers (\S\ref{sec:deeper-layer-accumulation}$\,\parallel\,$\S\ref{sec:deeper-layer-accumulation-NTK}). %

To proceed forward, we'll first need to evaluate an interlayer formula with three weight insertions (extending
our work in \S\ref{subsec:interlayer-interlude}), and then we'll immediately put it to use
in order to obtain
recursions for the dNTK-preactivation cross correlation.

\subsubsection{\emph{Inter}lude 2: \emph{Inter}layer Correlations Reloaded}\index{interlayer correlation!for dNTK evaluation}
Since the forward equation for the dNTK \eqref{eq:forward-eq-for-dNTK} has terms with one or three $(\ell+1)$-th-layer weight matrices, we'll need interlayer formulae with one or three weight insertions. 

Let's start by recalling our \terminate{generating function} for interlayer correlations\index{interlayer correlation}~\eqref{eq:master-weight-insertion}: 
\begin{align}\label{eq:master-weight-insertion-reprint}
&\E{\mathcal{O}\!\le(z^{(\ell+1)}\ri)e^{\sum_{i,j} \JW{ij}W^{(\ell+1)}_{ij}}\mathcal{Q}\!\le(z^{(\ell)},  \NTK^{(\ell)}, \dNTK^{(\ell)} \ri)}\, \\
=&\exp\!\le(\frac{\CW{\ell+1}}{2n_\ell}\sum_{i,j}\JW{ij}^2\ri)\! \E{\!\bra\!\!\!\bra\mathcal{O}\Big(z^{(\ell+1)}_{i;\delta}+\frac{\CW{\ell+1}}{n_{\ell}}\sum_{j=1}^{n_{\ell}}\JW{ij}\s{j}{\delta}{\ell}\Big)\ket\!\!\!\ket_{\widehat{G}^{(\ell+1)}}\!\!\!\!\!\!\!\mathcal{Q}\!\le(z^{(\ell)},  \NTK^{(\ell)}, \dNTK^{(\ell)}\ri) }\, .\notag 
\end{align}
In this formula, $\O$ is a generic function of $(\ell+1)$-th-layer preactivations only, and $\mathcal{Q}$ is a function of any of our $\ell$-th-layer objects; in particular, since the original derivation of this formula didn't depend on specifics of $\mathcal{Q}$,
we've also included the $\ell$-th-layer dNTK as part of its argument.

With that in mind, we first wrote down an interlayer formula with one insertion as \eqref{eq:one-weight-insertion} when analyzing (the lack of) representation learning in the infinite-width limit. To save you the need to flip back and refresh your memory, we'll reprint it here after making some minor notational adjustments:
\begin{align}\label{eq:one-weight-insertion-reprint-ish}
&\E{\O\!\le(z^{(\ell+1)}\ri) \W{ij}{\ell+1} \mathcal{Q}\!\le(z^{(\ell)}, \NTK^{(\ell)}, \dNTK^{(\ell)} \ri)}\, \\
=&\frac{\CW{\ell+1}}{n_{\ell}}\sum_{\delta\in\D}\E{\bra\!\!\!\bra\frac{\partial \O}{\partial \z{i}{\delta}{\ell+1}} \ket\!\!\!\ket_{\!\!\!\widehat{G}^{(\ell+1)}}\!\!\!\!  \s{j}{\delta}{\ell}\,\, \mathcal{Q}\!\le(z^{(\ell)}, \NTK^{(\ell)}, \dNTK^{(\ell)} \ri)}\, .\notag
\end{align}
This can be readily rederived by differentiating  the \terminate{generating function} \eqref{eq:master-weight-insertion-reprint} with respect to the source as $\frac{d}{d\JW{ij}}$ once and then setting the source to zero. %

In contrast, three-weight insertion formula will be new.  Thrice-differentiating  the \terminate{generating function} \eqref{eq:master-weight-insertion-reprint} with respect to the source as $\frac{d}{d\JW{i_0j_0}}\frac{d}{d\JW{i_1j_1}}\frac{d}{d\JW{i_2j_2}}$ and then setting the source to zero $\JW{}=0$, we find
\begin{align}\label{eq:three-weight-insertions}
&\E{\O\!\le(z^{(\ell+1)}\ri) \W{i_0j_0}{\ell+1}\W{i_1j_1}{\ell+1}\W{i_2j_2}{\ell+1} \mathcal{Q}\!\le(z^{(\ell)},  \NTK^{(\ell)}, \dNTK^{(\ell)}\ri)}\, \\
=&\le(\frac{\CW{\ell+1}}{n_{\ell}}\ri)^2\delta_{i_0i_1}\delta_{j_0j_1}\sum_{\delta\in\D}\E{\bra\!\!\!\bra\frac{\partial \O}{\partial \z{i_2}{\delta}{\ell+1}} \ket\!\!\!\ket_{\!\!\!\widehat{G}^{(\ell+1)}}\!\!  \s{j_2}{\delta}{\ell}\, \mathcal{Q}\!\le(z^{(\ell)},  \NTK^{(\ell)}, \dNTK^{(\ell)}\ri)}\, \notag\\
&+\le(\frac{\CW{\ell+1}}{n_{\ell}}\ri)^2\delta_{i_0i_2}\delta_{j_0j_2}\sum_{\delta\in\D}\E{\bra\!\!\!\bra\frac{\partial \O}{\partial \z{i_1}{\delta}{\ell+1}} \ket\!\!\!\ket_{\!\!\!\widehat{G}^{(\ell+1)}}\!\!  \s{j_1}{\delta}{\ell}\, \mathcal{Q}\!\le(z^{(\ell)},  \NTK^{(\ell)}, \dNTK^{(\ell)}\ri)}\, \notag\\
&+\le(\frac{\CW{\ell+1}}{n_{\ell}}\ri)^2\delta_{i_1i_2}\delta_{j_1j_2}\sum_{\delta\in\D}\E{\bra\!\!\!\bra\frac{\partial \O}{\partial \z{i_0}{\delta}{\ell+1}} \ket\!\!\!\ket_{\!\!\!\widehat{G}^{(\ell+1)}}\!\!  \s{j_0}{\delta}{\ell}\, \mathcal{Q}\!\le(z^{(\ell)},  \NTK^{(\ell)}, \dNTK^{(\ell)}\ri)}\, \notag\\
&+\le(\frac{\CW{\ell+1}}{n_{\ell}}\ri)^3\!\!\!\! \sum_{\delta_0,\delta_1,\delta_2\in\D}\!\!\!\! \E{\bra\!\!\!\bra\frac{\partial^3 \O}{\partial \z{i_0}{\delta_0}{\ell+1}\partial \z{i_1}{\delta_1}{\ell+1}\partial \z{i_2}{\delta_2}{\ell+1}} \ket\!\!\!\ket_{\!\!\!\widehat{G}^{(\ell+1)}}\!\!\!\! \!\!  \s{j_0}{\delta_0}{\ell} \s{j_1}{\delta_1}{\ell} \s{j_2}{\delta_2}{\ell}\mathcal{Q}\!\le(z^{(\ell)},  \NTK^{(\ell)}, \dNTK^{(\ell)}\ri)}\, .\notag
\end{align}
Intuitively, we can understand this formula as follows: each of the first three terms comes from forming one Wick contraction with two of the weight insertions
and then forming another contraction between the remaining weight
and a weight hidden inside the $z^{(\ell+1)}$ in $\O$,
while the final term comes from all three
weight insertions each forming a contraction with other weights inside the observable $\O$.

\subsubsection{dNTK-Preactivation Cross Correlations}\index{cross correlation!dNTK-preactivation}
Let's first use the interlayer formulae derived above to show that all of the dNTK's contributions to the statistics of the joint distribution $p\Big(z^{(\ell)}, \, \NTK^{(\ell)},\, \dNTK^{(\ell)}\Big\vert \D\Big)$ at order $1/n$  are captured by the cross correlation of the dNTK with a single preactivation, $\E{\Tia{\dNTK}{i_0i_1i_2}{\delta_0\delta_1\delta_2}{\ell}\, \z{i_3}{\delta_3}{\ell+1}}$. 
To that end, we'll examine a dNTK-preactivation cross correlator of a very general form:
\begin{align}\label{eq:cross-dNTK-general}
&\E{\O\!\le(z^{(\ell+1)}\ri)\, \Tia{\dNTK}{i_0i_1i_2}{\delta_0\delta_1\delta_2}{\ell+1}}\, \\
=&\delta_{i_0i_1}\frac{\lambda_{W}^{(\ell+1)}}{n_{\ell}}\sum_{k_0,k_2}\E{\O\!\le(z^{(\ell+1)}\ri)\Ti{W}{i_2 k_2}{\ell+1}\ds{k_0}{\delta_0}{\ell}\s{k_0}{\delta_1}{\ell}\ds{k_2}{\delta_2}{\ell}\Tia{\NTK}{k_0k_2}{\delta_0\delta_2}{\ell}}\, \notag\\
&+\delta_{i_0i_2}\frac{\lambda_{W}^{(\ell+1)}}{n_{\ell}}\sum_{k_0,k_1}\E{\O\!\le(z^{(\ell+1)}\ri)\Ti{W}{i_1 k_1}{\ell+1}\ds{k_0}{\delta_0}{\ell}\ds{k_1}{\delta_1}{\ell}\s{k_0}{\delta_2}{\ell}\Tia{\NTK}{k_0k_1}{\delta_0\delta_1}{\ell}}\, \notag\\
&+\sum_{k_0,k_1,k_2}\E{\O\!\le(z^{(\ell+1)}\ri)\Ti{W}{i_0 k_0}{\ell+1}\Ti{W}{i_1 k_1}{\ell+1}\Ti{W}{i_2 k_2}{\ell+1}\dds{k_0}{\delta_0}{\ell}\ds{k_1}{\delta_1}{\ell}\ds{k_2}{\delta_2}{\ell}\Tia{\NTK}{k_0k_1}{\delta_0\delta_1}{\ell}\Tia{\NTK}{k_0k_2}{\delta_0\delta_2}{\ell}}\, \notag\\
&+\sum_{k_0,k_1,k_2}\E{\O\!\le(z^{(\ell+1)}\ri)\Ti{W}{i_0 k_0}{\ell+1}\Ti{W}{i_1 k_1}{\ell+1}\Ti{W}{i_2 k_2}{\ell+1}\ds{k_0}{\delta_0}{\ell}\ds{k_1}{\delta_1}{\ell}\ds{k_2}{\delta_2}{\ell}\Tia{\dNTK}{k_0k_1k_2}{\delta_0\delta_1\delta_2}{\ell}}\, .\notag
\end{align}
Here, we took the expectation of the dNTK forward equation~\eqref{eq:forward-eq-for-dNTK} multiplied by a generic observable $\O\!\le(z^{(\ell+1)}\ri)$ of $(\ell+1)$-th-layer preactivations. We also ordered the four terms to reflect the order in which we will subsequently evaluate them.

First, let's simplify the first two terms. %
Using our interlayer formula with one weight insertion \eqref{eq:one-weight-insertion-reprint-ish} on the first term in \eqref{eq:cross-dNTK-general}, we get 
\begin{align}\label{eq:dNTK-cross-correlation-contribution-1}
&\frac{\lambda_{W}^{(\ell+1)}}{n_{\ell}}\delta_{i_0i_1}\sum_{k_0,k_2}\E{\O\!\le(z^{(\ell+1)}\ri)\Ti{W}{i_2 k_2}{\ell+1}\ds{k_0}{\delta_0}{\ell}\s{k_0}{\delta_1}{\ell}\ds{k_2}{\delta_2}{\ell}\Tia{\NTK}{k_0k_2}{\delta_0\delta_2}{\ell}}\, \\
=&\frac{\lambda_{W}^{(\ell+1)} \CW{\ell+1}}{n_{\ell}^2}\sum_{k_0,k_2}\sum_{\delta_{}\in\D}\delta_{i_0i_1}\E{\bra\!\!\!\bra\frac{\partial \O}{\partial \z{i_2}{\delta_{}}{\ell+1}} \ket\!\!\!\ket_{\!\!\!\widehat{G}^{(\ell+1)}}\!\!  \s{k_2}{\delta_{}}{\ell} \ds{k_0}{\delta_0}{\ell}\s{k_0}{\delta_1}{\ell}\ds{k_2}{\delta_2}{\ell}\Tia{\NTK}{k_0k_2}{\delta_0\delta_2}{\ell}}\, \notag\\
=&\frac{\lambda_{W}^{(\ell+1)} \CW{\ell+1}}{n_{\ell}^2}\sum_{k,m}\sum_{\delta_{}\in\D}\delta_{i_0i_1}\bra\!\!\!\bra\frac{\partial \O}{\partial \z{i_2}{\delta_{}}{\ell+1}} \ket\!\!\!\ket_{\!\!\!G^{(\ell+1)}}\!\!\E{\s{k}{\delta_1}{\ell}\s{m}{\delta_{}}{\ell}\ds{k}{\delta_0}{\ell}\ds{m}{\delta_2}{\ell}\Tia{\NTK}{km}{\delta_0\delta_2}{\ell}}+\o{\frac{1}{n^2}}\, \notag\\
=&\frac{1}{n_{\ell}}\frac{\lambda_{W}^{(\ell+1)}}{\CW{\ell+1}}\sum_{\delta_{}\in\D}\delta_{i_0i_1}\bra\!\!\!\bra\frac{\partial \O}{\partial \z{i_2}{\delta_{}}{\ell+1}} \ket\!\!\!\ket_{\!\!\!G^{(\ell+1)}}\NTHF{\delta_1\delta_0\delta_3\delta_2}{\ell+1}+\o{\frac{1}{n^2}}\, ,\notag
\end{align}
where in the third line we used the Schwinger-Dyson formula \eqref{eq:SD-again} to expand the metric fluctuation around its mean $G^{(\ell+1)}$, noting that the second term with the fluctuating is subleading, 
and in the last line we identified the remaining expectation as the definition of the NTK-preactivation cross correlation tensor $F^{(\ell+1)}$ from \eqref{eq:F-recursion-almost-there} up to constants.\footnote{
    Note that this is an $(\ell+1)$-th-layer quantity, rather than an $\ell$-th-layer quantity as we typically have on the right-hand side of recursions. %
    If you prefer, you can use the $F$-recursion \eqref{eq:F-recursion} to re-express it in terms of a more complicated collection of $\ell$-th-layer quantities.
}

Similarly, for the second term in \eqref{eq:cross-dNTK-general} we get an identical contribution up to the swapping of neural-sample index pairs as $(i_1,\delta_1)\leftrightarrow(i_2,\delta_2)$.

Next, let's tackle the third term in \eqref{eq:cross-dNTK-general}. Applying our interlayer formula with three weight insertions \eqref{eq:three-weight-insertions}, we get
\begin{align}\label{eq:dNTK-cross-correlation-contribution-2}
&\sum_{k_0,k_1,k_2}\!\!\!\!\E{\O\!\le(z^{(\ell+1)}\ri)\Ti{W}{i_0 k_0}{\ell+1}\Ti{W}{i_1 k_1}{\ell+1}\Ti{W}{i_2 k_2}{\ell+1}\dds{k_0}{\delta_0}{\ell}\ds{k_1}{\delta_1}{\ell}\ds{k_2}{\delta_2}{\ell}\Tia{\NTK}{k_0k_1}{\delta_0\delta_1}{\ell}\Tia{\NTK}{k_0k_2}{\delta_0\delta_2}{\ell}}\, \notag \\
=&\frac{\le(\CW{\ell+1}\ri)^2}{n_{\ell}}\delta_{i_0i_1}\sum_{\delta_{}\in\D}\bra\!\!\!\bra\frac{\partial \O}{\partial \z{i_2}{\delta_{}}{\ell+1}} \ket\!\!\!\ket_{\!\!\!G^{(\ell+1)}}\!\! \le\{\frac{1}{n_{\ell}}\sum_{k,m}\E{  \s{m}{\delta_{}}{\ell} \dds{k}{\delta_0}{\ell}\ds{k}{\delta_1}{\ell}\ds{m}{\delta_2}{\ell}\Tia{\NTK}{kk}{\delta_0\delta_1}{\ell}\Tia{\NTK}{k m}{\delta_0\delta_2}{\ell}}\ri\}\, \notag\\
&+\frac{\le(\CW{\ell+1}\ri)^2}{n_{\ell}}\delta_{i_0i_2}\sum_{\delta_{}\in\D}\bra\!\!\!\bra\frac{\partial \O}{\partial \z{i_1}{\delta_{}}{\ell+1}} \ket\!\!\!\ket_{\!\!\!G^{(\ell+1)}}\!\! \le\{\frac{1}{n_{\ell}}\sum_{k,m}\E{ \s{m}{\delta_{}}{\ell} \dds{k}{\delta_0}{\ell}\ds{k}{\delta_2}{\ell}\ds{m}{\delta_1}{\ell}\Tia{\NTK}{kk}{\delta_0\delta_2}{\ell}\Tia{\NTK}{km}{\delta_0\delta_1}{\ell}}\ri\}\, \notag\\
&+\frac{\le(\CW{\ell+1}\ri)^2}{n_{\ell}}\delta_{i_1i_2}\sum_{\delta_{}\in\D}\bra\!\!\!\bra\frac{\partial \O}{\partial \z{i_0}{\delta_{}}{\ell+1}} \ket\!\!\!\ket_{\!\!\!G^{(\ell+1)}}\!\! \le\{\frac{1}{n_{\ell}}\sum_{k,m}\E{ \s{m}{\delta_{}}{\ell}\dds{m}{\delta_0}{\ell}\ds{k}{\delta_1}{\ell}\ds{k}{\delta_2}{\ell}\Tia{\NTK}{mk}{\delta_0\delta_1}{\ell}\Tia{\NTK}{mk}{\delta_0\delta_2}{\ell}}\ri\}\, \notag\\
&+\o{\frac{1}{n^2}}\, ,
\end{align}
where for each term we again used the Schwinger-Dyson formula \eqref{eq:SD-again} to expand the metric fluctuation around its mean $G^{(\ell+1)}$, picking up the leading contribution from 
mean metric, and then took the Gaussian expectation outside the full $\ell$-th-layer expectation. Note that the final would-be term proportional to $\partial^3\O$ is subleading:
\begin{align}
&\le(\CW{\ell+1}\ri)^3\sum_{\delta_{},\delta_4,\delta_5\in\D}\bra\!\!\!\bra\frac{\partial^3 \O}{\partial \z{i_0}{\delta_{}}{\ell+1}\partial \z{i_1}{\delta_4}{\ell+1}\partial \z{i_2}{\delta_5}{\ell+1}} \ket\!\!\!\ket_{\!\!\!G^{(\ell+1)}}\, \\
&\quad\quad\times\frac{1}{n_{\ell}^3}\sum_{k_0,k_1,k_2}\E{\s{k_0}{\delta_{}}{\ell} \s{k_1}{\delta_4}{\ell} \s{k_2}{\delta_5}{\ell}\dds{k_0}{\delta_0}{\ell}\ds{k_1}{\delta_1}{\ell}\ds{k_2}{\delta_2}{\ell}\Tia{\NTK}{k_0k_1}{\delta_0\delta_1}{\ell}\Tia{\NTK}{k_0k_2}{\delta_0\delta_2}{\ell}}=\o{\frac{1}{n^2}}\, .\notag
\end{align}
 To see why, decompose each stochastic NTK into a mean and fluctuation as $\Tia{\NTK}{i_1i_2}{\delta_1\delta_2}{\ell}=\delta_{i_1i_2}\Ti{\NTKM}{\delta_1\delta_2}{\ell}+\DNTK{i_1i_2}{\delta_1\delta_2}{\ell}$ \eqref{eq:NTK-fluc-decomposition} and evaluate resulting four terms. In each case, you'll find the terms are $\o{1/n^2}$ suppressed due to the  Kronecker deltas constraining the triple sum and/or the additional $1/n$ suppression coming from the fluctuations.

Lastly, let's process the fourth and final term in our general cross correlation \eqref{eq:cross-dNTK-general}. Again applying our interlayer formula with three weight insertions \eqref{eq:three-weight-insertions} and taking only the leading-order pieces,
we get
\begin{align}\label{eq:dNTK-cross-correlation-contribution-3}
&\sum_{k_0,k_1,k_2}\E{\O\!\le(z^{(\ell+1)}\ri)\Ti{W}{i_0 k_0}{\ell+1}\Ti{W}{i_1 k_1}{\ell+1}\Ti{W}{i_2 k_2}{\ell+1}\ds{k_0}{\delta_0}{\ell}\ds{k_1}{\delta_1}{\ell}\ds{k_2}{\delta_2}{\ell}\Tia{\dNTK}{k_0k_1k_2}{\delta_0\delta_1\delta_2}{\ell}}\, \notag \\
=&\frac{\le(\CW{\ell+1}\ri)^2}{n_{\ell}}\sum_{\delta_{}\in\D}\delta_{i_0i_1}\bra\!\!\!\bra\frac{\partial \O}{\partial \z{i_2}{\delta_{}}{\ell+1}} \ket\!\!\!\ket_{\!\!\!G^{(\ell+1)}}\le\{\frac{1}{n_{\ell}}\sum_{k,m}\E{\ds{k}{\delta_0}{\ell}\ds{k}{\delta_1}{\ell}\ds{m}{\delta_2}{\ell}\s{m}{\delta_{}}{\ell} \Tia{\dNTK}{kkm}{\delta_0\delta_1\delta_2}{\ell}}\ri\}\, \notag\\
&+\frac{\le(\CW{\ell+1}\ri)^2}{n_{\ell}}\sum_{\delta_{}\in\D}\delta_{i_0i_2}\bra\!\!\!\bra\frac{\partial \O}{\partial \z{i_1}{\delta_{}}{\ell+1}} \ket\!\!\!\ket_{\!\!\!G^{(\ell+1)}}\le\{\frac{1}{n_{\ell}}\sum_{k,m}\E{\ds{k}{\delta_0}{\ell}\ds{m}{\delta_1}{\ell}\ds{k}{\delta_2}{\ell}\s{m}{\delta_{}}{\ell}\Tia{\dNTK}{kmk}{\delta_0\delta_1\delta_2}{\ell}}\ri\}\, \notag\\
&+\frac{\le(\CW{\ell+1}\ri)^2}{n_{\ell}}\sum_{\delta_{}\in\D}\delta_{i_1i_2}\bra\!\!\!\bra\frac{\partial \O}{\partial \z{i_0}{\delta_{}}{\ell+1}} \ket\!\!\!\ket_{\!\!\!G^{(\ell+1)}}\le\{\frac{1}{n_{\ell}}\sum_{k,m}\E{\ds{m}{\delta_0}{\ell}\ds{k}{\delta_1}{\ell}\ds{k}{\delta_2}{\ell}\s{m}{\delta_{}}{\ell}\Tia{\dNTK}{mkk}{\delta_0\delta_1\delta_2}{\ell}}\ri\}\, \notag\\
&+\o{\frac{1}{n^2}}\, .
\end{align}
Note that the factors in the curly brackets are actually of order one since -- as we saw in the second layer and will recursively show next for general layers $\ell$ -- the leading dNTK-preactivation cross correlation is of order $1/n$. 
Meanwhile, again the final would-be term proportional to $\partial^3\O$ is subleading:
\begin{align}
&\le(\CW{\ell+1}\ri)^3\sum_{\delta_3,\delta_4,\delta_5\in\D}\bra\!\!\!\bra\frac{\partial^3 \O}{\partial \z{i_0}{\delta_3}{\ell+1}\partial \z{i_1}{\delta_4}{\ell+1}\partial \z{i_2}{\delta_5}{\ell+1}} \ket\!\!\!\ket_{\!\!\!\ker^{(\ell+1)}}\, \\
&\quad\quad\times\frac{1}{n_{\ell}^3}\sum_{k_0,k_1,k_2}\E{\le(\ds{k_0}{\delta_0}{\ell}\s{k_0}{\delta_3}{\ell}\ri)\le(\ds{k_1}{\delta_1}{\ell}\s{k_1}{\delta_4}{\ell} \ri)\le(\ds{k_2}{\delta_2}{\ell}\s{k_2}{\delta_5}{\ell}\ri)\Tia{\dNTK}{k_0k_1k_2}{\delta_0\delta_1\delta_2}{\ell}} = \o{\frac{1}{n^2}} \, \notag .
\end{align}
To see why, note that the expectation is another dNTK-preactivation cross correlator and thus is at most of order $1/n$. 
Further, we only get such an order-$1/n$ contribution  when two out of three neural indices $k_0, k_1, k_2$ coincide: this should be clear from the pattern of Kronecker deltas that arise when we evaluate such dNTK-preactivation cross correlations in terms of our 
$P$-$Q$ decomposition; cf.~\eqref{eq:dntk-second-layer-decomposition} for the second layer, or look ahead a paragraph to \eqref{eq:cross-dNTK-general-leading} for deeper layers.
This means that the sum over all three neural indices will be restricted to be only two independent sums and, taking the prefactor of $1/n^3$ into account, the overall contribution of this term will thus go as $\sim(1/n^3)(n^2)(1/n)\sim 1/n^2$.

Substituting all our evaluated contributions \eqref{eq:dNTK-cross-correlation-contribution-1} (twice), \eqref{eq:dNTK-cross-correlation-contribution-2}, and \eqref{eq:dNTK-cross-correlation-contribution-3} into our expression for a general dNTK-preactivation cross correlator \eqref{eq:cross-dNTK-general}, we get
\begin{align}\label{eq:cross-dNTK-general-leading}
&\E{\O\!\le(z^{(\ell+1)}\ri)\, \Tia{\dNTK}{i_0i_1i_2}{\delta_0\delta_1\delta_2}{\ell+1}}\, \\
=&\frac{1}{n_{\ell}}\sum_{\delta_{}\in\D}\Bigg[\delta_{i_1i_2}\bra\!\!\!\bra\frac{\partial \O}{\partial \z{i_0}{\delta_{}}{\ell+1}} \ket\!\!\!\ket_{\!\!\!G^{(\ell+1)}}\!\!\!\!\! \dNTKP{\delta_0\delta_1\delta_2\delta_{}}{\ell+1}\, \notag\\
&\quad\quad\quad\quad+\delta_{i_0i_1}\bra\!\!\!\bra\frac{\partial \O}{\partial \z{i_2}{\delta_{}}{\ell+1}} \ket\!\!\!\ket_{\!\!\!G^{(\ell+1)}}\!\!\!\!\!  \dNTKQ{\delta_0\delta_1\delta_2\delta_{}}{\ell+1}+\delta_{i_0i_2}\bra\!\!\!\bra\frac{\partial \O}{\partial \z{i_1}{\delta_{}}{\ell+1}} \ket\!\!\!\ket_{\!\!\!G^{(\ell+1)}}\!\!\!\!\!  \dNTKQ{\delta_0\delta_2\delta_1\delta_{}}{\ell+1}\Bigg] + \o{\frac{1}{n^2}}, \notag
\end{align}
where we've introduced the $(\ell+1)$-th-layer generalizations of the second-layer tensors $P^{(2)}$~\eqref{eq:dNTKP-second-layer} and $Q^{(2)}$~\eqref{eq:dNTKQ-second-layer}:
\begin{align}
\dNTKP{\delta_0\delta_1\delta_2\delta_3}{\ell+1}\equiv&\le(\CW{\ell+1}\ri)^2 \frac{1}{n_{\ell}}\sum_{k,m=1}^{n_{\ell}}\E{\dds{m}{\delta_0}{\ell}\ds{k}{\delta_1}{\ell}\ds{k}{\delta_2}{\ell} \s{m}{\delta_3}{\ell}\Tia{\NTK}{mk}{\delta_0\delta_1}{\ell}\Tia{\NTK}{mk}{\delta_0\delta_2}{\ell}}\,\label{eq:dNTKP-recursion-implicit} \\
&+\le(\CW{\ell+1}\ri)^2\frac{1}{n_{\ell}}\sum_{k,m=1}^{n_{\ell}}\E{\ds{m}{\delta_0}{\ell}\ds{k}{\delta_1}{\ell}\ds{k}{\delta_2}{\ell}\s{m}{\delta_3}{\ell}\Tia{\dNTK}{mkk}{\delta_0\delta_1\delta_2}{\ell}}+ \oninv\, ,\notag \\
\dNTKQ{\delta_0\delta_1\delta_2\delta_3}{\ell+1}\equiv&\le(\CW{\ell+1}\ri)^2 \frac{1}{n_{\ell}}\sum_{k,m=1}^{n_{\ell}}\E{\dds{k}{\delta_0}{\ell}\ds{k}{\delta_1}{\ell}\ds{m}{\delta_2}{\ell}\s{m}{\delta_3}{\ell}\Tia{\NTK}{kk}{\delta_0\delta_1}{\ell}\Tia{\NTK}{k m}{\delta_0\delta_2}{\ell}}+\frac{\lambda_{W}^{(\ell+1)}}{\CW{\ell+1}}\NTHF{\delta_1\delta_0\delta_3\delta_2}{\ell+1}\, \notag\\
&+\le(\CW{\ell+1}\ri)^2\frac{1}{n_{\ell}}\sum_{k,m=1}^{n_{\ell}}\E{\ds{k}{\delta_0}{\ell}\ds{k}{\delta_1}{\ell}\ds{m}{\delta_2}{\ell}\s{m}{\delta_3}{\ell} \Tia{\dNTK}{kkm}{\delta_0\delta_1\delta_2}{\ell}} + \oninv\, . \label{eq:dNTKQ-recursion-implicit}
\end{align}
To see how these general expressions reduce to the ones we had for $\dNTKP{\delta_0\delta_1\delta_2\delta_3}{2}$  and $ \dNTKQ{\delta_0\delta_1\delta_2\delta_3}{2}$ in the second layer, \eqref{eq:dNTKP-second-layer} and \eqref{eq:dNTKQ-second-layer}, recall that
the first-layer NTK is deterministic and diagonal in neural indices $\Tia{\NTK}{i_1i_2}{\delta_1\delta_2}{1}=\delta_{i_1i_2}\Ti{\NTKM}{\delta_1\delta_2}{1}$, that the first-layer dNTK vanishes $\Tia{\dNTK}{i_0i_1i_2}{\delta_0\delta_1\delta_2}{1}=0$, and that in the second layer we had for the NTK-preactivation cross correlation tensor $\NTHF{\delta_1\delta_0\delta_3\delta_2}{2}=\le(\CW{2}\ri)^2\Ti{\NTKM}{\delta_0\delta_2}{1}\bra\sigma_{\delta_1}\sigma_{\delta_3}\sigma^{\prime}_{\delta_0}\sigma^{\prime}_{\delta_2}\ket_{G^{(1)}}$ \eqref{eq:F-recursion-second}.

Finally, note that by setting our observable to $\O = \z{i_3}{\delta_3}{\ell+1}$, we get
\begin{align}\label{eq:dntk-ell-layer-decomposition}
\E{\Tia{\dNTK}{i_0i_1i_2}{\delta_0\delta_1\delta_2}{\ell+1}\z{i_3}{\delta_3}{\ell+1}}=&\frac{1}{n_{\ell}}\le[\delta_{i_0 i_3}\delta_{i_1i_2}\dNTKP{\delta_0\delta_1\delta_2\delta_3}{\ell+1}+\delta_{i_0i_1}\delta_{i_2i_3}\dNTKQ{\delta_0\delta_1\delta_2\delta_3}{\ell+1}+\delta_{i_0i_2}\delta_{i_1i_3}\dNTKQ{\delta_0\delta_2\delta_1\delta_3}{\ell+1}\ri]\, .
\end{align}
Importantly, this means that at leading non-vanishing order in $1/n$, the tensors in the decomposition\index{tensor decomposition!dNTK-preactivation $P$/$Q$} of our elementary dNTK-preactivation cross correlator with a single preactivation \eqref{eq:dntk-ell-layer-decomposition} completely fix the general dNTK-preactivation cross correlation 
with more complicated observables \eqref{eq:cross-dNTK-general-leading}.\footnote{In other words, at our order in $1/n$
we can always replace the $\ell$-th-layer dNTK inside any expectations by the following differential operator
\begin{align}\label{eq:cross-dNTK-general-leading-operator}
\Tia{\dNTK}{i_0i_1i_2}{\delta_0\delta_1\delta_2}{\ell}\to \frac{1}{n_{\ell-1}}\sum_{\delta_{}\in\D}\Bigg[\delta_{i_1i_2}\dNTKP{\delta_0\delta_1\delta_2\delta_{}}{\ell}\frac{\partial}{\partial \z{i_0}{\delta_{}}{\ell}}+\delta_{i_0i_1}\dNTKQ{\delta_0\delta_1\delta_2\delta_{}}{\ell}\frac{\partial}{\partial \z{i_2}{\delta_{}}{\ell}}+\delta_{i_0i_2} \dNTKQ{\delta_0\delta_2\delta_1\delta_{}}{\ell}\frac{\partial}{\partial \z{i_1}{\delta_{}}{\ell}}\Bigg]\, ,
\end{align}
with non-fluctuating coefficients $P^{(\ell)}$ and $Q^{(\ell)}$. When we use this replacement, we must remember that the derivatives act on all of the $\ell$-th-layer preactivations multiplying the dNTK; i.e.~move the dNTK all the way to the left side of the expectation before making such a replacement.}
Thus, to completely incorporate the leading effects of the dNTK in our analysis we only need to evaluate recursions for $P^{(\ell)}$ and $Q^{(\ell)}$.\footnote{
     Any preactivation-NTK-dNTK cross correlators such as $\E{\DNTK{i_1i_2}{\delta_1\delta_2}{\ell} \Tia{\dNTK}{i_3i_4i_5}{\delta_3\delta_4\delta_5}{\ell}\z{i_6}{\delta_6}{\ell}}$     
      and any higher-order dNTK correlators such as  $\E{ \Tia{\dNTK}{i_0i_1i_2}{\delta_0\delta_1\delta_2}{\ell} \Tia{\dNTK}{i_3i_4i_5}{\delta_3\delta_4\delta_5}{\ell}}$ are subleading. We won't show this explicitly, but you may find additional tranquility in working it out for yourself;
     both examples are relatively simple to work out for the second layer, $L = 2$.
}

\subsubsection{$P$-recursion}\index{differential of the neural tangent kernel!dNTK-preactivation cross correlation|see{cross correlation}}\index{cross correlation!dNTK-preactivation!P-recursion@$P$-recursion}
To find a recursion for $P^{(\ell)}$, we need to evaluate the two expectations in \eqref{eq:dNTKP-recursion-implicit}.

\index{tensor decomposition!NTK mean and fluctuation}
For the first expectation, making a decomposition of the NTK into a mean and fluctuation as $\Tia{\NTK}{i_1i_2}{\delta_1\delta_2}{\ell}=\delta_{i_1i_2}\Ti{\NTKM}{\delta_1\delta_2}{\ell}+\DNTK{i_1i_2}{\delta_1\delta_2}{\ell}$, we get
\begin{align}\label{eq:P-recursion-piece-1}
&\frac{1}{n_{\ell}^2}\sum_{k,m}\E{\dds{m}{\delta_0}{\ell}\ds{k}{\delta_1}{\ell}\ds{k}{\delta_2}{\ell}\s{m}{\delta_3}{\ell}\Tia{\NTK}{mk}{\delta_0\delta_1}{\ell}\Tia{\NTK}{mk}{\delta_0\delta_2}{\ell}}\, \\
=&\frac{1}{n_{\ell}}\bra\sigma^{\prime\prime}_{\delta_0}\sigma^{\prime}_{\delta_1}\sigma^{\prime}_{\delta_2}\sigma_{\delta_3} \ket_{G^{(\ell)}}\Ti{\NTKM}{\delta_0\delta_1}{\ell}\Ti{\NTKM}{\delta_0\delta_2}{\ell}+\frac{1}{n_{\ell-1}}\bra\sigma^{\prime\prime}_{\delta_0} \sigma_{\delta_3} \ket_{G^{(\ell)}}\bra\sigma^{\prime}_{\delta_1} \sigma^{\prime}_{\delta_2}  \ket_{G^{(\ell)}} \NTHB{\delta_0\delta_0\delta_1\delta_2}{\ell}+\o{\frac{1}{n^2}}\, .\notag
\end{align}
In particular, the cross terms consisting of an NTK mean and an NTK fluctuation dropped out because the Kronecker delta from the mean constrained the double sum and the fluctuation gave an additional $1/n$ suppression.
If this explanation was a little too fast, you should review our slower derivation of the $B$-recursion from \eqref{eq:Bstart} to \eqref{eq:B-recursion}, which is identical in form to \eqref{eq:P-recursion-piece-1} above up to where the sample indices and the derivatives go.

For the second expectation in \eqref{eq:dNTKP-recursion-implicit}, we can just apply our general cross-correlation formula \eqref{eq:cross-dNTK-general-leading} for layer $\ell$, letting us simplify it as
\begin{align}
&\frac{1}{n_{\ell}}\sum_{k,m=1}^{n_{\ell}}\E{\ds{m}{\delta_0}{\ell}\ds{k}{\delta_1}{\ell}\ds{k}{\delta_2}{\ell}\s{m}{\delta_3}{\ell}\Tia{\dNTK}{mkk}{\delta_0\delta_1\delta_2}{\ell}}\, \\
=&\frac{1}{n_{\ell}n_{\ell-1}}\!\sum_{k,m=1}^{n_{\ell}}\sum_{\delta_4\in\D}\Bigg[\bra\!\!\!\bra\frac{\partial}{\partial \z{m}{\delta_4}{\ell}}\le(\ds{m}{\delta_0}{\ell}\ds{k}{\delta_1}{\ell}\ds{k}{\delta_2}{\ell}\s{m}{\delta_3}{\ell}\ri)\ket\!\!\!\ket_{G^{(\ell)}}\!\!\!\!\!\!\dNTKP{\delta_0\delta_1\delta_2\delta_4}{\ell}+\delta_{mk}\times\o{1}\!\Bigg]+\o{\frac{1}{n}}\, \notag\\
=&\le(\frac{n_{\ell}}{n_{\ell-1}}\ri)\le[\bra\sigma^{\prime\prime}_{\delta_0}\sigma_{\delta_3}\ket_{G^{(\ell)}}\bra\sigma^{\prime}_{\delta_1}\sigma^{\prime}_{\delta_2}\ket_{G^{(\ell)}}\dNTKP{\delta_0\delta_1\delta_2\delta_0}{\ell}+\bra\sigma^{\prime}_{\delta_0}\sigma^{\prime}_{\delta_3}\ket_{\ker^{(\ell)}}\bra\sigma^{\prime}_{\delta_1}\sigma^{\prime}_{\delta_2}\ket_{\ker^{(\ell)}}\dNTKP{\delta_0\delta_1\delta_2\delta_3}{\ell}\ri]+\o{\frac{1}{n}}\, .\notag
\end{align}
Here in the second line, we've simply written the $Q^{(\ell)}$ terms proportional to $\delta_{mk}$ as $\o{1}$; the details of these terms do not matter because when we perform the double sum with the restriction  $m=k$, they will be subleading. As for the term proportional to $P^{(\ell)}$, the diagonal contribution with $k=m$ is similarly subleading; the leading contribution comes from the $(n_{\ell}^2-n_{\ell})$ off-diagonal pieces with $k\ne m$, for which the derivative acts only on two activations out of the four, and then we can further use Gaussian factorization to write each term as a product of single-neuron Gaussian expectations.

Plugging these two simplified expectations back into our expression for  $P^{(\ell+1)}$ \eqref{eq:dNTKP-recursion-implicit}, we get a recursion:
\begin{align}
\dNTKP{\delta_0\delta_1\delta_2\delta_3}{\ell+1}=&\le(\CW{\ell+1}\ri)^2\!\!\bra\sigma^{\prime\prime}_{\delta_0}\sigma^{\prime}_{\delta_1}\sigma^{\prime}_{\delta_2}\sigma_{\delta_3} \ket_{G^{(\ell)}}\Ti{\NTKM}{\delta_0\delta_1}{\ell}\Ti{\NTKM}{\delta_0\delta_2}{\ell}\, \label{eq:dNTKP-recursion-explicit}\\
&+\le(\frac{n_{\ell}}{n_{\ell-1}}\ri)\le(\CW{\ell+1}\ri)^2\bra\sigma^{\prime}_{\delta_1}\sigma^{\prime}_{\delta_2}\ket_{G^{(\ell)}}\, \notag\\
&\quad \times\!\!\Bigg[\!\!\bra\sigma^{\prime\prime}_{\delta_0}\sigma_{\delta_3}\ket_{\!G^{(\ell)}}\!\dNTKP{\delta_0\delta_1\delta_2\delta_0}{\ell}\!\!+\!\bra\sigma^{\prime}_{\delta_0}\sigma^{\prime}_{\delta_3}\ket_{\!G^{(\ell)}}\!\dNTKP{\delta_0\delta_1\delta_2\delta_3}{\ell}\!\!+\!\bra\sigma^{\prime\prime}_{\delta_0} \sigma_{\delta_3} \ket_{\!G^{(\ell)}}\!\NTHB{\delta_0\delta_0\delta_1\delta_2}{\ell} \!\Bigg]\!\!+\!\o{\frac{1}{n}}\, . \notag
\end{align}
Interestingly, we see that this dNTK tensor $P^{(\ell)}$ mixes with the NTK mean $\NTKM^{(\ell)}$ as well as the NTK-variance tensor $B^{(\ell)}$.
Since $\NTKM^{(\ell)}$ and $B^{(\ell)}$ are of order one, and since $P^{(1)}=0$, this recursion shows that $P^{(\ell)}$ will recursively stay of order one for all layers $\ell$.

\subsubsection{$Q$-recursion}\index{cross correlation!dNTK-preactivation!Q-recursion@$Q$-recursion}
To find a recursion for $Q^{(\ell)}$, we need to work out the two expectations in \eqref{eq:dNTKQ-recursion-implicit}. 

\index{tensor decomposition!NTK mean and fluctuation}
For the  first expectation, again  making a decomposition of the NTK into a mean and fluctuation as $\Tia{\NTK}{i_1i_2}{\delta_1\delta_2}{\ell}=\delta_{i_1i_2}\Ti{\NTKM}{\delta_1\delta_2}{\ell}+\DNTK{i_1i_2}{\delta_1\delta_2}{\ell}$, we get
\begin{align}
&\frac{1}{n_{\ell}^2}\sum_{k,m}\E{\dds{k}{\delta_0}{\ell}\ds{k}{\delta_1}{\ell}\ds{m}{\delta_2}{\ell} \s{m}{\delta_3}{\ell} \Tia{\NTK}{kk}{\delta_0\delta_1}{\ell}\Tia{\NTK}{k m}{\delta_0\delta_2}{\ell}}\, \\
=&\frac{1}{n_{\ell}^2}\sum_{k}\E{\dds{k}{\delta_0}{\ell}\ds{k}{\delta_1}{\ell}\ds{k}{\delta_2}{\ell}\s{k}{\delta_3}{\ell} }\Ti{\NTKM}{\delta_0\delta_1}{\ell}\Ti{\NTKM}{\delta_0\delta_2}{\ell}+\frac{1}{n_{\ell}^2}\sum_{k}\E{  \s{k}{\delta_3}{\ell} \dds{k}{\delta_0}{\ell}\ds{k}{\delta_1}{\ell}\ds{k}{\delta_2}{\ell}\DNTK{kk}{\delta_0\delta_1}{\ell}}\Ti{\NTKM}{\delta_0\delta_2}{\ell}\, \notag\\
&+\frac{1}{n_{\ell}^2}\sum_{k,m}\E{\dds{k}{\delta_0}{\ell}\ds{k}{\delta_1}{\ell}\ds{m}{\delta_2}{\ell}\s{m}{\delta_3}{\ell} \DNTK{k m}{\delta_0\delta_2}{\ell}}\Ti{\NTKM}{\delta_0\delta_1}{\ell}\, \notag\\
&+\frac{1}{n_{\ell}^2}\sum_{k,m}\E{\dds{k}{\delta_0}{\ell}\ds{k}{\delta_1}{\ell}\ds{m}{\delta_2}{\ell}\s{m}{\delta_3}{\ell} \DNTK{kk}{\delta_0\delta_1}{\ell}\DNTK{k m}{\delta_0\delta_2}{\ell}}\, \notag\\
=&\frac{1}{n_{\ell}}\bra\sigma^{\prime\prime}_{\delta_0}\sigma^{\prime}_{\delta_1}\sigma^{\prime}_{\delta_2}\sigma_{\delta_3} \ket_{G^{(\ell)}}\Ti{\NTKM}{\delta_0\delta_1}{\ell}\Ti{\NTKM}{\delta_0\delta_2}{\ell}\, \notag\\
&+\frac{1}{n_{\ell-1}}\sum_{\delta_4,\delta_5, \delta_6, \delta_7} \Ti{\NTKM}{\delta_0\delta_1}{\ell}\bra z_{\delta_4}\sigma^{\prime\prime}_{\delta_0}\sigma^{\prime}_{\delta_1} \ket_{G^{(\ell)}}\bra z_{\delta_5}\sigma^{\prime}_{\delta_2}\sigma_{\delta_3} \ket_{G^{(\ell)}}\TI{G}{\delta_4\delta_6}{\ell}\TI{G}{\delta_5\delta_7}{\ell}\NTHF{\delta_6\delta_0\delta_7\delta_2}{\ell}+\o{\frac{1}{n^2}}\, .\notag
\end{align}
Here, to go from the second expression to the third expression, we had to evaluate four expectations: the first expectation gives a single-neuron Gaussian expectation at leading order; the second expectation is subleading, cf.~\eqref{eq:most-general}; the third expectation can also be evaluated with that same general NTK-preactivation cross-correlation formula \eqref{eq:most-general}, but in this case gives a leading term proportional to $F^{(\ell)}$; and the final expectation vanishes due to the unpaired $m$ neural index, cf.~similar manipulations in \eqref{eq:correlator-paradise} and then the decomposition~\eqref{eq:NTH-variance-decomposition}.

To simplify the second expectation in \eqref{eq:dNTKQ-recursion-implicit}, we can again apply our general cross-correlation formula \eqref{eq:cross-dNTK-general-leading} for layer $\ell$:
\begin{align}
&\frac{1}{n_{\ell}}\sum_{k,m=1}^{n_{\ell}}\E{\ds{k}{\delta_0}{\ell}\ds{k}{\delta_1}{\ell}\ds{m}{\delta_2}{\ell}\s{m}{\delta_3}{\ell} \Tia{\dNTK}{kkm}{\delta_0\delta_1\delta_2}{\ell}}\, \\
=&\frac{1}{n_{\ell}n_{\ell-1}}\sum_{k,m=1}^{n_{\ell}}\sum_{\delta_4\in\D}\Bigg[\bra\!\!\!\bra\frac{\partial}{\partial \z{m}{\delta_4}{\ell}}\le(\ds{k}{\delta_0}{\ell}\ds{k}{\delta_1}{\ell}\ds{m}{\delta_2}{\ell}\s{m}{\delta_3}{\ell}\ri)\ket\!\!\!\ket_{G^{(\ell)}}\dNTKQ{\delta_0\delta_1\delta_2\delta_4}{\ell}\Bigg]+\o{\frac{1}{n}}\, \notag\\
=&\le(\frac{n_{\ell}}{n_{\ell-1}}\ri)\le[\bra\sigma^{\prime\prime}_{\delta_2}\sigma_{\delta_3}\ket_{\ker^{(\ell)}}\bra\sigma^{\prime}_{\delta_0}\sigma^{\prime}_{\delta_1}\ket_{G^{(\ell)}}\dNTKQ{\delta_0\delta_1\delta_2\delta_2}{\ell}+\bra\sigma^{\prime}_{\delta_2}\sigma^{\prime}_{\delta_3}\ket_{G^{(\ell)}}\bra\sigma^{\prime}_{\delta_1}\sigma^{\prime}_{\delta_2}\ket_{G^{(\ell)}}\dNTKQ{\delta_0\delta_1\delta_2\delta_3}{\ell}\ri]+\o{\frac{1}{n}}\, .\notag
\end{align}
Here in the second line, this time we didn't even write the terms proportional to $\delta_{mk}$; just as we saw before when working out the $P$-recursion, the restriction $k=m$ for the double sum will make such terms
subleading.
In the final equality -- again similarly to the $P$-recursion -- we kept the off-diagonal terms with $k\ne m$, took the derivative, used Gaussian factorization, and performed the double sum.

Plugging these two simplified expectations back into our expression for  $Q^{(\ell+1)}$ \eqref{eq:dNTKQ-recursion-implicit}, we get our final recursion of the book:
\begin{align}
\dNTKQ{\delta_0\delta_1\delta_2\delta_3}{\ell+1}=&\le(\CW{\ell+1}\ri)^2\!\!\bra\sigma^{\prime\prime}_{\delta_0}\sigma^{\prime}_{\delta_1}\sigma^{\prime}_{\delta_2}\sigma_{\delta_3} \ket_{G^{(\ell)}}\Ti{\NTKM}{\delta_0\delta_1}{\ell}\Ti{\NTKM}{\delta_0\delta_2}{\ell}+\frac{\lambda_{W}^{(\ell+1)}}{\CW{\ell+1}}\NTHF{\delta_1\delta_0\delta_3\delta_2}{\ell+1}\, \label{eq:dNTKQ-recursion-explicit} \notag \\
&+\!\le(\frac{n_{\ell}}{n_{\ell-1}}\ri)\!\le(\CW{\ell+1}\ri)^2\!\!\bra\sigma^{\prime}_{\delta_0}\sigma^{\prime}_{\delta_1}\ket_{G^{(\ell)}}\!\!\le[\bra\sigma^{\prime\prime}_{\delta_2}\sigma_{\delta_3}\ket_{G^{(\ell)}}\dNTKQ{\delta_0\delta_1\delta_2\delta_2}{\ell}+\bra\sigma^{\prime}_{\delta_2}\sigma^{\prime}_{\delta_3}\ket_{G^{(\ell)}}\dNTKQ{\delta_0\delta_1\delta_2\delta_3}{\ell}\ri]\, \notag\\
&+\!\le(\frac{n_{\ell}}{n_{\ell-1}}\ri)\!\le(\CW{\ell+1}\ri)^2 \!\!\Ti{\NTKM}{\delta_0\delta_1}{\ell}\!\!\!\!\!\!\sum_{\delta_4,\ldots,\delta_7\in\D}\!\! \!\!\!\!\bra z_{\delta_4}\sigma^{\prime\prime}_{\delta_0}\sigma^{\prime}_{\delta_1} \ket_{G^{(\ell)}}\!\!\bra z_{\delta_5}\sigma^{\prime}_{\delta_2}\sigma_{\delta_3} \ket_{G^{(\ell)}}\!\TI{G}{\delta_4\delta_6}{\ell}\TI{G}{\delta_5\delta_7}{\ell}\NTHF{\delta_6\delta_0\delta_7\delta_2}{\ell} \,\notag  \\
&+\o{\frac{1}{n}}\, . 
\end{align}
Interestingly, in this case we see that this dNTK tensor $Q^{(\ell)}$ mixes with the NTK mean $\NTKM^{(\ell)}$ as well as the NTK-preactivation cross-correlation tensor $F^{(\ell)}$.\footnote{Also of possible interest, while the recursions for $F^{(\ell)}$ \eqref{eq:F-recursion} and $B^{(\ell)}$ \eqref{eq:B-recursion} were sourced by the NTK mean $\NTKM^{(\ell)}$, but otherwise didn't mix with any finite-width tensors, the recursion for  NTK-preactivation cross-correlation tensor $D^{(\ell)}$ \eqref{eq:D-recursion} mixed with the four-point vertex $V^{(\ell)}$, and the recursion for NTK-variance tensor $A^{(\ell)}$ \eqref{eq:A-recursion} mixed with both $V^{(\ell)}$ and $D^{(\ell)}$. Thus, at least at this order, it seems like the correlations and fluctuations of the type $F^{(\ell)}$ and $B^{(\ell)}$ are potentially useful for the representation learning that the dNTK induces, while $V^{(\ell)}$, $D^{(\ell)}$ and $A^{(\ell)}$ may be more associated with instantiation-to-instantiation fluctuations.}
Again, since $\NTKM^{(\ell)}$ and $F^{(\ell)}$ are both of order one, and since $Q^{(1)}=0$, this recursion shows that $Q^{(\ell)}$ will also recursively stay of order one for all layers $\ell$.\\

In conclusion, together with the general dNTK-preactivation cross-correlation formula \eqref{eq:cross-dNTK-general-leading}, the $P$-$Q$ recursions, \eqref{eq:dNTKP-recursion-explicit} and \eqref{eq:dNTKQ-recursion-explicit}, show that the leading dNTK-preactivation cross correlation is $1/n$-suppressed. In other words, the effects of the dNTK are visible \emph{at finite width only}.

\section{Effective Theory of the dNTK at Initialization}\label{sec:dNTK-criticality}
This section parallels our previous effective theory work on preactivation statistics and NTK-preactivation joint statistics (\S\ref{ch:eft-mlp}$\,\parallel\,$\S\ref{ch:eft-ntk}). In particular, 
since we already know so many different reasons why criticality is essential, cf.~\S\ref{ch:eft-mlp}, \S\ref{ch:bayesian-inference}, \S\ref{ch:eft-ntk}, and \S\ref{ch:NTHb}, we'll spend less time on the disastrous consequences of not picking critical initialization hyperparameters and more time on finding asymptotic solutions to the $P$- and $Q$-recursions at criticality.\index{criticality}

As we did in our discussion of preactivation criticality in~\S\ref{ch:eft-mlp} and NTK criticality in~\S\ref{ch:eft-ntk}, throughout this section we'll set the bias variance $C_b^{(\ell)}$ and the rescaled weight variance $C_W^{(\ell)}$ to be uniform across layers
\be\label{eq:ntk-chapter-initialziation-hyperparameters-dropped-layer-dependence-reprint}
C_b^{(\ell)}=C_b\, , \qquad C_W^{(\ell)}=C_W\, .
\ee
Further mirroring \S\ref{sec:signal_prop_finite_width} and \S\ref{sec:ntk_criticality}--\S\ref{sec:ntk_criticality_tanh_univ}, we'll consider MLPs with uniform hidden layer widths
\be\label{eq:ntk-chapter-layer-widths-equalized-reprint}
n_1=\ldots=n_{L-1}\equiv n\, .
\ee
Finally, analogous to \S\ref{ch:eft-ntk},  we're only going to focus on single-input statistics, leaving the evaluation of the multi-input recursions as an \terminate{adventure for thrill seekers}.\footnote{\label{foot:thrill-seekers-guide}You'll have to generalize the $\gamma^{[a]}$ into a tensor product $\gamma^{[a]} \otimes \gamma^{[b]}$ and then further decompose such a basis according to the symmetries of the finite-width tensors you'll want to expand.
}

With these choices made, let's write down the leading single-input 
recursions for  $\dNTKP{}{\ell}$ and $\dNTKQ{}{\ell}$. Dropping the sample indices and contributions that are subleading in $1/n$, in particular replacing the mean metric by the kernel $G^{(\ell)} \to \ker^{(\ell)}$ and the NTK mean by the frozen NTK $\NTKM^{(\ell)} \to \NTKI^{(\ell)}$,
the recursions \eqref{eq:dNTKP-recursion-explicit} and \eqref{eq:dNTKQ-recursion-explicit} reduce to 
\begin{align}
\dNTKP{}{\ell+1}=&C_W^2\bra\sigma^{\prime\prime}\sigma^{\prime}\sigma^{\prime}\sigma\ket_{\ker^{(\ell)}}\le( \Ti{\NTKI}{}{\ell} \ri)^2 + C_W \Ti{\chi}{\perp}{\ell}\bra\sigma^{\prime\prime}\sigma\ket_{\ker^{(\ell)}} \NTHB{}{\ell} \, \notag \\
 &+\le[C_W \Ti{\chi}{\perp}{\ell} \bra\sigma^{\prime\prime}\sigma\ket_{\ker^{(\ell)}}+ \le(\Ti{\chi}{\perp}{\ell}\ri)^2 \ri]\dNTKP{}{\ell}  \, , \label{eq:dNTKP-recursion-single-input}\\
\dNTKQ{}{\ell+1}=&C_W^2\bra\sigma^{\prime\prime}\sigma^{\prime}\sigma^{\prime}\sigma\ket_{\ker^{(\ell)}}\le( \Ti{\NTKI}{}{\ell}\ri)^2
+\frac{\lambda_{W}^{(\ell+1)}}{C_W} \NTHF{}{\ell+1}
+ 2 h^{(\ell)} \Ti{\chi}{\parallel}{\ell} \Ti{\NTKI}{}{\ell}\NTHF{}{\ell}
\, \label{eq:dNTKQ-recursion-single-input} \notag \\
&+ \le[C_W \Ti{\chi}{\perp}{\ell} \bra\sigma^{\prime\prime}\sigma\ket_{\ker^{(\ell)}}+\le(\Ti{\chi}{\perp}{\ell}\ri)^2 \ri]\dNTKQ{}{\ell}\, , 
\end{align}
with the initial conditions (cf.~\S\ref{subsec:first-layer-zero-dNTK})
\be\label{eq:dntk-initial-conditions}
\dNTKP{}{1}=\dNTKQ{}{1}=0\, .
\ee
To simplify these expressions, we have recalled our two susceptibilities, the \terminate{parallel susceptibility} \eqref{eq:chi-parallel} and the \terminate{perpendicular susceptibility} \eqref{eq:chi-perp}, 
which are given by
\begin{align}
\Ti{\chi}{\parallel}{\ell}  &\equiv \frac{C_W}{\ker^{(\ell)}} \bra \sigma^\prime \sigma  z \ket_{\ker^{(\ell)}} \, , \\
\Ti{\chi}{\perp}{\ell} 
&\equiv  C_W\bra\sigma^{\prime}\sigma^{\prime}\ket_{\ker^{(\ell)}}\, , 
\end{align}
and we have also recalled our least favorite helper function \eqref{eq:h-function},
\begin{align}
h^{(\ell)}&\equiv \frac{1}{2}\frac{\td }{\td K^{(\ell)}}\chi_{\perp}^{(\ell)} = \frac{C_W}{2K^{(\ell)}} \bra   \sigma^{\prime\prime} \sigma^\prime  z \ket_{K^{(\ell)}} \, ,\label{eq:h-function-reprint-new}
\end{align}
though we've given a new expression for it on the right-hand side, obtained through integration by parts, cf.~\eqref{eq:gaussian-integration-by-parts-formula}.

To solve these recursions at criticality, we need to remember our \terminate{scaling ansatz} for observables \eqref{eq:master-scaling-ansatz},
\begin{align}\label{eq:master-scaling-ansatz-reprint-the-second}
\Ti{\O}{}{\ell} &= \le( \frac{1}{\ell} \ri)^{p_\O} \le[c_{0,0}+  c_{1,1} \le( \frac{\log \ell}{\ell} \ri) + c_{1,0}\le( \frac{ 1}{\ell}\ri) +  c_{2,2} \le(  \frac{\log^2 \ell}{\ell^2} \ri)+  \dots \ri]\, .%
\end{align}
Here, solving the dNTK recursions will yield new critical exponents $p_P$ and $p_Q$ that describe the asymptotic depth scaling of $\dNTKP{}{\ell}$ and $\dNTKQ{}{\ell}$, respectively.

\index{differential of the neural tangent kernel!scaling laws}
Additionally,  in order to understand the relative size of the dNTK-preactivation cross correlation, we will need to identify appropriate dimensionless quantities just as we did before in \S\ref{sec:signal_prop_finite_width} and \S\ref{sec:ntk_criticality}. 
In this case, it will turn out that we should normalize by 
two factors of the (frozen) NTK 
\be\label{eq:scaling-relations-dNTK}
\frac{\dNTKP{}{\ell}}{n \le( \Ti{\NTKI}{}{\ell} \ri)^2 } \sim \frac{1}{n}\le( \frac{1}{\ell} \ri)^{p_P - 2p_\Theta} \,, \qquad \frac{\dNTKQ{}{\ell}}{n \le( \Ti{\NTKI}{}{\ell} \ri)^2} \sim \frac{1}{n}\le( \frac{1}{\ell} \ri)^{p_Q - 2p_\Theta}\, ,
\ee
where on the right-hand side $p_\Theta$ is the \terminate{critical exponent} for the frozen NTK\index{frozen NTK}. 

To see why these are the appropriate quantities to consider, 
recall our discussion of \neo{dimensional analysis} in footnote~\ref{foot:dimensional-analysis} of \S\ref{sec:perturbation} and that our notation of $[z]$ means ``$z$ is measured in units of $[z]$.'' Looking at our second-order update for preactivations in \eqref{eq:preactivation-updated-finite-width}, and remembering that we can only add terms that have the same dimensions, we must have
\be\label{eq:dNTK-dimensional-analysis}
[z]=[\eta]\, [\epsilon] \,[\NTK]=[\eta]^2\,[\epsilon]^2\,[\dNTK] \,,
\ee 
from which it's clear that $[\eta]\,[\epsilon]=[z]\,[\NTK]^{-1}$, and subsequently we see that $P$ and $Q$ have dimensions of  NTK squared:
\be\label{eq:dimensions-of-P-Q}
[P]=[Q] \equiv [z]\,[\dNTK]=[\NTK]^{2} \, . 
\ee
If this still seems a little counterintuitive, the utility of considering these particular ratios \eqref{eq:scaling-relations-dNTK} will become even more apparent when we analyze the stochastic prediction of a fully-trained finite-width network in \S\ref{subsec:prediction-at-finite-width}.

After solving the $P$- and $Q$-recursions for both universality classes and looking at these dimensionless quantities \eqref{eq:scaling-relations-dNTK}, we'll again find \emph{scaling laws}\index{scaling law} that transcend universality class:
\be\label{eq:dNTK-scaling-laws}
p_P - 2 p_\Theta=-1\, , \quad p_Q- 2 p_\Theta=-1\, .
\ee
Specifically, we'll see that these laws hold for both the scale-invariant and $K^\star=0$ universality classes\index{universality class!scale-invariant}\index{universality class!K@$K^\star=0$}.\footnote{To be more specific, for the scale-invariant class, we'll find that $P^{(\ell)}$ identically vanishes and that $Q^{(\ell)}$ solely determines the leading finite-width dNTK-preactivation cross correlation.}
Thus, we'll be able to conclude that all the leading finite-width effects of the  preactivation-NTK-dNTK joint distribution are \emph{relevant}\index{relevant (RG flow)} -- in the sense of RG flow -- controlled by the same $\ell/n$ perturbative cutoff\index{cutoff, effective theory}.

Now that we're fully prepared for what we're going to see, let's actually solve the dNTK recursions for our two important universality classes.

\subsection{Scale-Invariant Universality Class}\label{subsec:dntk_criticality_scale_invariant}\index{universality class!scale-invariant}
First, let's recall some previous results that we'll need in order to evaluate the recursions \eqref{eq:dNTKP-recursion-single-input} and \eqref{eq:dNTKQ-recursion-single-input}. For the scale-invariant universality class, we know from \eqref{eq:needed-to-recall-in-kernel-learning-chapter-II}, \eqref{eq:helper-ntk-scale-invariant-h}, and \eqref{eq:gaussian-expectation-ntk-scale-invariant-2} that
\begin{align}\label{eq:dntk-prior-results-scale}
\Ti{\chi}{\perp}{\ell}=C_WA_2\equiv \chi\, , \qquad h^{(\ell)}=0\, , \qquad \bra\sigma^{\prime}\sigma^{\prime}\sigma\sigma\ket_{\ker^{(\ell)}}=A_4 \ker^{(\ell)}\, ,
\end{align}
where as a reminder $A_2\equiv (a_{+}^2+a_{-}^2)/2$ and $A_4\equiv (a_{+}^4+a_{-}^4)/2$, and the $a_{\pm}$ are the respective slopes of the positive/negative linear pieces of the activation function, cf.~\eqref{eq:scale-invariant-one-kink}.

Next, we see that all the new Gaussian expectations in  the recursions \eqref{eq:dNTKP-recursion-single-input} and \eqref{eq:dNTKQ-recursion-single-input}  involve the second derivative of the activation. For these, we need to be somewhat careful with nonlinear scale-invariant activation functions, since they have an undifferentiable kink at the origin. (For linear activation functions, $\sigma^{\prime\prime}(z)=0$, and there's no subtlety.)
For the first of these new expectations, note that we can integrate it by parts as
\begin{align}\label{eq:scale-invariant-higher-derivative-first}
\bra\sigma^{\prime\prime}\sigma\ket_{\ker}=&\frac{1}{\sqrt{2\pi K}}\int_{-\infty}^{\infty} dz e^{-\frac{z^2}{2K}}\le(\frac{d}{dz}\sigma'\ri)\sigma = \frac{1}{\ker}\bra z \sigma^{\prime}\sigma\ket_{\ker}-\bra \sigma^{\prime}\sigma^{\prime}\ket_{\ker}= 0 \, ,%
\end{align}
where in the final equality we used
\be
\frac{1}{\ker}\bra z \sigma^{\prime}\sigma\ket_{\ker}=\bra \sigma^{\prime}\sigma^{\prime}\ket_{\ker} = A_2 \,.
\ee
Similarly, integrating the other new Gaussian expectation by parts, we find
\be\label{eq:scale-invariant-higher-derivative-second}
\bra\sigma^{\prime\prime}\sigma^{\prime}\sigma^{\prime}\sigma\ket_{\ker}=\frac{1}{\ker}\bra z \sigma^{\prime}\sigma^{\prime}\sigma^{\prime}\sigma\ket_{\ker}-2\bra\sigma^{\prime\prime}\sigma^{\prime}\sigma^{\prime}\sigma\ket_{\ker}-\bra\sigma^{\prime}\sigma^{\prime}\sigma^{\prime}\sigma^{\prime}\ket_{\ker}\, .
\ee
Rearranging this, we easily see that
\be
\bra\sigma^{\prime\prime}\sigma^{\prime}\sigma^{\prime}\sigma\ket_{\ker}=\frac{1}{3\ker}\bra z \sigma^{\prime}\sigma^{\prime}\sigma^{\prime}\sigma\ket_{\ker}-\frac{1}{3}\bra\sigma^{\prime}\sigma^{\prime}\sigma^{\prime}\sigma^{\prime}\ket_{\ker} = 0\, , %
\ee
where in the final equality we used
\be
\frac{1}{\ker}\bra z \sigma^{\prime}\sigma^{\prime}\sigma^{\prime}\sigma\ket_{\ker}= \bra\sigma^{\prime}\sigma^{\prime}\sigma^{\prime}\sigma^{\prime}\ket_{\ker} = A_4 \,.
\ee
Thus, we can safely ignore both of these new expectations.

Substituting in all of \eqref{eq:dntk-prior-results-scale} and ignoring ignorable expectations, our single-input recursions \eqref{eq:dNTKP-recursion-single-input} and \eqref{eq:dNTKQ-recursion-single-input} are extremely simple:
\begin{align}
\dNTKP{}{\ell+1}=&\chi^2\dNTKP{}{\ell}\, , \\
\dNTKQ{}{\ell+1}=&\chi^2\dNTKQ{}{\ell}
+ \frac{\lambda_{W}^{(\ell+1)}}{C_W}\NTHF{}{\ell+1}  \, .\label{eq:dNTKQ-recursion-single-input-relu}
\end{align}
In particular, since our initial condition \eqref{eq:dntk-initial-conditions} is $\dNTKP{}{1}=0$, we see immediately that $\dNTKP{}{\ell}$ vanishes identically for all layers:
\be\label{eq:P-vanish-scale-invariant}
\dNTKP{}{\ell}=0\, .
\ee
In contrast, for $\dNTKQ{}{\ell}$ we see that the susceptibility $\chi$ is going to generically lead to exponential behavior.

\index{criticality}
Now, let's tune to scale-invariant criticality by setting the initialization hyperparameters as $C_b=0$ and $C_W=1/A_2$. As a consequence, this fixes the susceptibility to unity, $\chi=1$, and leaves the kernel fixed for all layers as $\ker^{(\ell)}=\Tif{\ker}{}$. Additionally, let's pick our training hyperparameters according to the learning rate \terminate{equivalence principle}, for which we're instructed to choose layer-independent learning rates \eqref{eq:super-scale-invariant} as 
\be\label{eq:learning-rate-EP-scale-invariant-reprint-for-dNTK}
\Lb{\ell} = \frac{\widetilde{\lambda}_b}{L} \, , \qquad \LW{\ell} = \frac{\widetilde{\lambda}_W }{L}  \,.
\ee
Finally, with these hyperparameter choices, the single-input solution for the frozen NTK\index{frozen NTK} \eqref{eq:frozen-ntk-critical-solution-relu} and the single-input solution for the NTK-preactivation cross correlation $\Ti{F}{}{\ell}$ \eqref{eq:F-solution} are given by
\begin{align}\label{eq:frozen-ntk-critical-solution-relu-reprint} 
\Ti{\NTKI}{}{\ell}&=\le(\widetilde{\lambda}_b+\widetilde{\lambda}_W A_2 \Tif{\ker}{}\ri)\frac{\ell}{L}\, , \\
\Ti{F}{}{\ell}&=\frac{\ell(\ell-1)}{2L}\le[\frac{A_4}{A_2^2} \le(\widetilde{\lambda}_b+\widetilde{\lambda}_W A_2 \Tif{\ker}{}\ri)\Tif{\ker}{}\ri] \, .
\label{eq:F-solution-reprint}
\end{align}

Plugging in the critical initialization hyperparameters, the fixed kernel, the learning rates \eqref{eq:learning-rate-EP-scale-invariant-reprint-for-dNTK}, and the expression for $\Ti{F}{}{\ell}$ \eqref{eq:F-solution-reprint}, the $Q$-recursion \eqref{eq:dNTKQ-recursion-single-input-relu} becomes
\be
\dNTKQ{}{\ell+1}=\dNTKQ{}{\ell}+ \frac{\ell (\ell+1)}{2L^2}\le[\frac{A_4}{A_2 }    \le(\widetilde{\lambda}_b+\widetilde{\lambda}_W A_2 \Tif{\ker}{}\ri) \widetilde{\lambda}_W  \Tif{\ker}{} \ri]\, .
\ee
This simple recursion is exactly solved by
\be\label{eq:Q-solution-scale-invariant}
\dNTKQ{}{\ell}= \frac{\ell (\ell^2-1)}{6L^2}\le[\frac{A_4}{A_2 }    \le(\widetilde{\lambda}_b+\widetilde{\lambda}_W A_2 \Tif{\ker}{}\ri) \widetilde{\lambda}_W  \Tif{\ker}{} \ri]\, ,
\ee
which satisfies the initial condition $\dNTKQ{}{1}=0$. 

With this solution, we can identify the \terminate{critical exponent} associated with the large-$\ell$ behavior of $\dNTKQ{}{\ell}$: $p_Q = -3$. Further, we see that our dimensionless quantity \eqref{eq:scaling-relations-dNTK} will satisfy the scaling law \eqref{eq:dNTK-scaling-laws} as promised,
\be
p_Q - 2p_\Theta = -1\, ,
\ee
where we have also used $p_\Theta=-1$ from the scale-invariant frozen NTK solution reprinted above in \eqref{eq:frozen-ntk-critical-solution-relu-reprint}. More specifically, substituting in the solution for $\Ti{\NTKI}{}{\ell}$ \eqref{eq:frozen-ntk-critical-solution-relu-reprint} and the solution for $\dNTKQ{}{\ell}$ \eqref{eq:Q-solution-scale-invariant} into the dimensionless ratio \eqref{eq:scaling-relations-dNTK}, we find
\be
\frac{\dNTKQ{}{\ell}}{n \le(\Ti{\NTKI}{}{\ell} \ri)^2} = \frac{A_4}{6A_2}\le[ \frac{  \widetilde{\lambda}_W  \Tif{\ker}{}  }{  \widetilde{\lambda}_b+\widetilde{\lambda}_W A_2 \Tif{\ker}{} } \ri]\frac{\ell}{n} + \dots \, .
\ee
Thus, we have verified the $\ell/n$ scaling of the leading dNTK-preactivation cross correlation for scale-invariant activation functions.

\subsection{\texorpdfstring{$\Tif{\ker}{}=0$}{K*=0} Universality Class}\label{subsec:dntk_criticality_tanh_univ}\index{universality class!K@$K^\star=0$}
For the $K^\star=0$ universality class, we'll begin again by recalling some previous results. First, we know from
\eqref{eq:chi-parallel-expansion-K-star-equals-zero}, \eqref{eq:chi-perp-expansion-K-star-equals-zero}, \eqref{eq:gaussian-expectation-k-star-3}, and \eqref{eq:h-function-reprint-new} the following:
\begin{align}\label{eq:recalling-expansion-for-the-sake-of-dntk-for-one-last-time-1}
\chi_{\parallel}(\ker)=&\le(C_W\sigma_1^2\ri)\le[1+2 a_1K+\o{K^2}\ri]\, ,\\
\label{eq:recalling-expansion-for-the-sake-of-dntk-for-one-last-time-2}
\chi_{\perp}(\ker)=&\le(C_W\sigma_1^2\ri)\le[1+b_1 K+O\!\le(K^2\ri)\ri]\, ,\\
\label{eq:recalling-expansion-for-the-sake-of-dntk-for-one-last-time-3}
C_W^2\bra\sigma^{\prime}\sigma^{\prime}\sigma\sigma\ket_{\ker}=&\le(C_W\sigma_1^2\ri)^2\le[\ker+O|1\le(\ker^2\ri)\ri]\,  ,\\
\label{eq:recalling-expansion-for-the-sake-of-dntk-for-one-last-time-4}
h\!\le(K\ri)=&\frac{1}{2}\frac{\td }{\td K}\chi_{\perp}(\ker)=\le(C_W\sigma_1^2\ri)\le[\frac{b_1}{2}+\o{K^1}\ri]\, .
\end{align}
To interpret these results, remember that we Taylor expanded the activation function as %
\be\label{eq:taylor-expansion-k-star-reprint-last}
\sigma(z)=\sum_{p=0}^{\infty}\frac{\sigma_{p}}{p!}z^p\,  ,
\ee
 defined the following combination of Taylor coefficients for convenience
\begin{align}
a_1&\equiv \le(\frac{\sigma_3}{\sigma_1}\ri)+\frac{3}{4}\le(\frac{\sigma_2}{\sigma_1}\ri)^2\ ,\label{eq:a1-recall-last}\\
b_1&\equiv \le(\frac{\sigma_3}{\sigma_1}\ri)+\le(\frac{\sigma_2}{\sigma_1}\ri)^2\, ,\label{eq:b1-recall-last}%
\end{align}
and required that all activation functions in this class satisfy $\sigma_0 = 0$ and $\sigma_1 \neq 0$.
Then, making analogous Taylor expansions and performing Gaussian integrations order by order, we can evaluate the new Gaussian expectations in  the recursions \eqref{eq:dNTKP-recursion-single-input} and \eqref{eq:dNTKQ-recursion-single-input} as
\begin{align}
C_W\bra\sigma^{\prime\prime}\sigma\ket_{\ker}=&\le(C_W\sigma_1^2\ri)\le[(2a_1-b_1)\ker+O\!\le(K^2\ri)\ri]\, ,\label{eq:doing-expansion-for-the-sake-of-dntk-for-one-last-time-3}\\
C_W^2\bra\sigma^{\prime\prime}\sigma^{\prime}\sigma^{\prime}\sigma\ket_{\ker}=&\le(C_W\sigma_1^2\ri)^2\le[(-6a_1+7b_1)\ker+O\!\le(\ker^2\ri)\ri]\, .\label{eq:doing-expansion-for-the-sake-of-dntk-for-one-last-time-4}
\end{align}

Now, let's jump right into $K^\star=0$ criticality  \eqref{eq:k-star-equals-zero-critical-initialization} by tuning the initialization hyperparameters as $C_b=0$ and $C_W =1/\sigma_1^2$. At the same time, let's also tune our training hyperparameters according to the learning rate \terminate{equivalence principle}, which for $K^\star=0$ activation functions is given by \eqref{eq:super-tanh-general},
\be
\Lb{\ell}=\widetilde{\lambda}_b\le(\frac{1}{\ell}\ri)^{p_{\perp}}L^{p_{\perp}-1}\, , \qquad \lamW{\ell}=\widetilde{\lambda}_W\le(\frac{L}{\ell}\ri)^{p_{\perp}-1} \, ,
\ee
where the critical exponent for perpendicular perturbations is defined as $p_{\perp}\equiv b_1/a_1$. With these hyperparameter settings, let's also record the other single-input solutions that we need for the recursions, the kernel $\Ti{\ker}{}{\ell}$ \eqref{eq:initial-tanh-wthout-subleading},  the frozen NTK $\Ti{\NTKI}{}{\ell}$ \eqref{eq:frozen-ntk-k-star-solution}, the NTK-preactivation cross correlation $F^{(\ell)}$ \eqref{eq:F-k-star-solution}, and the NTK variance $B^{(\ell)}$ \eqref{eq:B-k-star-solution}:
\begin{align}
\ker^{(\ell)}=&\le[\frac{1}{(-a_1)}\ri]\frac{1}{\ell}+ \ldots\, ,\label{eq:oh-really-we-have-to-recall-this-again-dots-yes}\\
\label{eq:K-star-frozen-NTK-dNTK-reprint}
\Ti{\NTKI}{}{\ell}=&\le[\widetilde{\lambda}_b+\frac{\widetilde{\lambda}_W\sigma_1^2}{(-a_1)}\ri]\le( \frac{L}{\ell}\ri)^{p_{\perp}-1}+\ldots\, , \\
\Ti{F}{}{\ell}=&\frac{1}{(5-p_{\perp})}\le[\frac{1}{(-a_1)}\ri] \le[\widetilde{\lambda}_b+\frac{\widetilde{\lambda}_W\sigma_1^2}{(-a_1)}\ri]\le(\frac{L}{\ell}\ri)^{p_{\perp}-1}+\ldots\, , \\
\NTHB{}{\ell}=&\frac{L^{2p_{\perp}-2}}{3} \le[\widetilde{\lambda}_b+\frac{\widetilde{\lambda}_W\sigma_1^2}{(-a_1)}\ri]^2\le(\frac{1}{\ell}\ri)^{2p_{\perp}-3}+\ldots\, .
\label{eq:NTKB-final-reprint-for-dNTK}
\end{align}
Finally, plugging all our collected results and tunings \eqref{eq:recalling-expansion-for-the-sake-of-dntk-for-one-last-time-1}--\eqref{eq:recalling-expansion-for-the-sake-of-dntk-for-one-last-time-4} and \eqref{eq:doing-expansion-for-the-sake-of-dntk-for-one-last-time-3}--\eqref{eq:NTKB-final-reprint-for-dNTK}  into the $P$-recursion \eqref{eq:dNTKP-recursion-single-input} and the $Q$-recursion \eqref{eq:dNTKQ-recursion-single-input}, we get
\begin{align}\label{eq:P-recursion-K-star}
\dNTKP{}{\ell+1}=&\le[1-\frac{(p_{\perp}+2)}{\ell}+\ldots\ri]\dNTKP{}{\ell}
+\frac{(p_\perp -2) }{3}\le[\widetilde{\lambda}_b+\frac{\widetilde{\lambda}_W\sigma_1^2}{(-a_1)}\ri]^2\le( \frac{L}{\ell}\ri)^{2p_{\perp}-2}
+ \dots \, ,\\
\dNTKQ{}{\ell+1}=&\le[1-\frac{(p_{\perp}+2)}{\ell}+\ldots\ri]\dNTKQ{}{\ell}\, \label{eq:Q-recursion-K-star}\\
&+\frac{1}{(5-p_\perp)}\le[ (1-p_{\perp})\frac{\widetilde{\lambda}_W \sigma_1^2}{(-a_1)}  -p_{\perp}\widetilde{\lambda}_b\ri]\le[\widetilde{\lambda}_b+\frac{\widetilde{\lambda}_W\sigma_1^2}{(-a_1)}\ri]\le( \frac{L}{\ell}\ri)^{2p_{\perp}-2}+\ldots\, .\notag
\end{align}

Let's tackle the
$P$-recursion first.  Plugging our scaling ansatz \eqref{eq:master-scaling-ansatz-reprint-the-second} into the recursion \eqref{eq:P-recursion-K-star} and matching the terms, we find an asymptotic solution
\begin{align}\label{eq:P-K-star-asymptotic}
\dNTKP{}{\ell}=&-\frac{L^{2p_\perp-2}}{3}\le(\frac{2-p_{\perp}}{5-p_{\perp}} \ri) \le[\widetilde{\lambda}_b+\frac{\widetilde{\lambda}_W\sigma_1^2}{(-a_1)}\ri]^2\le( \frac{1}{\ell}\ri)^{2p_{\perp}-3 }+\ldots\, .
\end{align}
Thus, the \terminate{critical exponent} for this dNTK-preactivation cross correlation is $p_P = 2p_{\perp}-3$, and we obtain our promised \terminate{scaling law} \eqref{eq:dNTK-scaling-laws} 
\be
p_P-2p_\Theta = -1 \, ,
\ee
after substituting in $p_\Theta = p_\perp - 1$ from the $K^\star=0$ frozen NTK solution \eqref{eq:K-star-frozen-NTK-dNTK-reprint}. More specifically, substituting in the solution for $\NTKI^{(\ell)}$ \eqref{eq:K-star-frozen-NTK-dNTK-reprint} and  the solution for $P^{(\ell)}$ \eqref{eq:P-K-star-asymptotic} into the dimensionless ratio \eqref{eq:scaling-relations-dNTK}, we get
\begin{align}
\frac{\dNTKP{}{\ell}}{n \le(\Ti{\NTKI}{}{\ell} \ri)^2}=&-\frac{1}{3}\le(\frac{2-p_{\perp}}{5-p_{\perp}} \ri) \frac{\ell}{n} +\ldots\, ,
\end{align}
which  \emph{(i)} scales as $\ell/n$,  \emph{(ii)} is independent of the training hyperparameters, and \emph{(iii)} is manifestly negative, given $p_{\perp}\leq1$.\footnote{
To recall why $p_{\perp}\leq1$, without flipping back to footnote \ref{foot:p-perp-less-than-or-equal-to-zero} of \S\ref{ch:NTHb}, \emph{(i)} remember that we must have $a_1<0$ in order for the kernel $\ker^{(\ell)}$ to stay positive when asymptotically approaching the $\ker^{\star}=0$ fixed point, cf.~\eqref{eq:oh-really-we-have-to-recall-this-again-dots-yes}, \emph{(ii)} note to yourself that $b_1 \geq a_1$, cf.~\eqref{eq:a1-recall-last} and \eqref{eq:b1-recall-last}, and \emph{(iii)} realize that $a_1<0$ and $b_1 \geq a_1$ together imply that $p_\perp \equiv b_1 / a_1\leq1$.}

Similarly for the $Q$-recursion \eqref{eq:Q-recursion-K-star}, plugging our scaling ansatz \eqref{eq:master-scaling-ansatz-reprint-the-second} into the recursion \eqref{eq:Q-recursion-K-star} and matching the terms one final time, we find
\be
\dNTKQ{}{\ell}=\frac{L^{2p_\perp - 2}}{(5-p_\perp)^2}\le[
1-p_{\perp} - \frac{\widetilde{\lambda}_b}{ \widetilde{\lambda}_b+ \widetilde{\lambda}_W\sigma_1^2 / (-a_1) } \ri]\le[\widetilde{\lambda}_b+\frac{\widetilde{\lambda}_W\sigma_1^2}{(-a_1)}\ri]^2\le( \frac{1}{\ell}\ri)^{2p_{\perp}-3} \, .
\ee
This gives a critical exponent of $p_Q = 2p_\perp -3$ and a dimensionless ratio of
\be\label{eq:Q-ratio}
\frac{\dNTKQ{}{\ell}}{n \le(\Ti{\NTKI}{}{\ell} \ri)^2}=\frac{1}{(5-p_\perp)^2}\le[
1-p_{\perp} - \frac{\widetilde{\lambda}_b}{ \widetilde{\lambda}_b+ \widetilde{\lambda}_W\sigma_1^2 / (-a_1) }  \ri]  \frac{\ell}{n}+\ldots \, .
\ee
Thus, we've now fully verified our \terminate{scaling law} \eqref{eq:dNTK-scaling-laws}: 
\be
p_Q-2p_\Theta = -1 \, .
\ee

\sbreak

In conclusion, all the nonzero dimensionless dNTK-preactivation cross correlations will grow as $\ell/n$ at leading order in the finite-width expansion. As the dNTK leads to a dynamical NTK\index{dynamical NTK}\index{dynamical NTK|seealso{interaction NTK}}\index{dynamical NTK|seealso{effective kernel}}\index{neural tangent kernel!dynamical|see{dynamical NTK}}, this is sufficient to see that deep finite-width networks are representation learners.

In the next section, we're going to set aside our direct investigation of deep MLPs and instead focus on a pedagogical model of representation learning that directly incorporates the effect of a nonzero dNTK. In the following chapter, we'll return to our finite-width networks and complete our goal of computing the effective distribution over wide and deep finite-width MLPs after training. The simplified model presented in the next section will make it easier to understand the results of our later analysis of such fully-trained networks.

\section{Nonlinear Models and Nearly-Kernel Methods}\label{sec:nonlinear-model}
In \S\ref{sec:lazy-kernel}, we placed infinite-width networks into a broader context of \terminate{machine learning} models. There, we discussed how the infinite-width network can be understood as either a \neo{linear model} with fixed random features or, dually, as a \emph{kernel method}\index{kernel methods} with the kernel given by the frozen NTK. 
Now we are ready to find a broader context for finite-width networks.

In this chapter, we've seen that the statistics needed to compute the dynamics of finite-width networks (\S\ref{sec:dNTK-RG} and \S\ref{sec:dNTK-criticality}) are far more complicated than for infinite-width networks, nontrivially incorporating the second derivative of the network function. This additional complexity is irreducibly encapsulated by a single object: the dNTK (\S\ref{sec:dNTK}).
The dNTK enables the NTK to evolve during training, 
leading to nontrivial \neo{representation learning}  from the training data. 

The goal of this section is to abstract this property away from the deep learning\index{deep learning!abstracted} framework and distill it into a \textbf{minimal model of representation learning}\index{representation learning!minimal model|textbf} that captures all of its important features.
Such a model provides a framework for studying the type of representation learning exhibited by deep neural networks, but more succinctly and more broadly.
In other words, we hope that this endeavor will %
extend the standard toolbox of \terminate{machine learning}.

First in \S\ref{subsec:nonlinear-models}, we'll extend \terminate{linear model}s discussed in \S\ref{subsec:linear-models} to \emph{nonlinear models}\index{nonlinear model}.
Then in \S\ref{subsec:nearly-kernel-methods}, we'll give a dual\index{duality!nonlinear model -- nearly-kernel methods} description of these nonlinear models, \neo{nearly-kernel methods}, which will extend the standard \terminate{kernel methods} that we discussed in \S\ref{subsec:kernel-methods}.
Finally in \S\ref{subsec:nonlinear-at-finite}, as we did analogously for infinite-width networks before in \S\ref{subsec:linear-at-infinity}, we'll see how finite-width networks can be understood in terms of this new
framework. %

\subsection{Nonlinear Models}\label{subsec:nonlinear-models}
As a reminder from \S\ref{subsec:linear-models}, a  \neo{linear model} is given by \eqref{eq:linear-model-def}
\be\label{eq:linear-model-def-reprint}
z_i(x_{\delta}; \theta) =\sum_{j=0}^{n_f}W_{ij}\fea_j(x_{\delta}) \, ,
\ee
where the model parameters are given by the weight matrix $\theta=W_{ij}$, the model's features are given by the \neo{feature function} $\fea_{j}(x)$, and we again adopt the old-school convention for incorporating biases, including a constant feature $\fea_{0}(x)\equiv 1$ so that the bias vector is given by $W_{i0}\equiv b_{i}$.
Note that since we're not discussing neural networks in particular, there's no \terminate{neural indices} or \terminate{layer indices} here. Instead, in this equation,
$\delta$ is a \emph{sample index}, running over the $\ND$ samples in the dataset $\delta \in \D$; $i$ is a \emph{vectorial index}\index{vectorial indices}, running over the $n_{\text{out}}$ vectorial component of the model output $z_i$; and $j$ is a \emph{feature index}\index{feature indices}, running over $(n_f+1)$ different features. In this setup, a \terminate{feature function} $\fea_{j}(x)$ is computed on an input sample $x$, and the weight matrix $W_{ij}$ determines the effect of the $j$-th feature on the $i$-th component of the output.
Traditionally, feature functions  $\fea_{j}(x)$ are often \emph{designed} such that the linear model works well for the desired task after optimization or  \neo{linear regression}.

To go beyond this linear paradigm, let's slightly  \emph{deform}\index{deformation!linear model} it to get a \textbf{\emph{nonlinear} model}\index{nonlinear model}:
\be\label{eq:nonlinear-model-def}
z_i(x_{\delta}; \theta) = \sum_{j=0}^{n_f}W_{ij}\fea_j(x_{\delta}) + \frac{\epsilon}{2}\sum_{j_1,j_2=0}^{n_f}W_{ij_1}W_{ij_2}\featwo_{j_1j_2}(x_{\delta}) +\ldots \, .
\ee
Here, $\epsilon \ll 1$ is small parameter that controls the size of the deformation, and more importantly, 
we've introduced another set of feature functions\index{feature function!nonlinear model},  $\featwo_{j_1j_2}(x)$, with \emph{two} feature indices, making the model nonlinear in its weights.
By definition, $\featwo_{j_1j_2}(x)$ is symmetric under the exchange of the feature indices $j_1\leftrightarrow j_2$, and the factor of $1/2$ is there because of the double counting of the double sum. For reasons that will be made clear shortly, we will call each component of $\featwo_{j_1j_2}(x)$ a \textbf{meta feature function}\index{meta feature function|textbf}\index{feature function!meta|see{meta feature function}}, and so there are $(n_f+1)(n_f+2)/2$ different meta feature functions.

To familiarize ourselves with the features of this model, let's see how the model outputs change given a small change in the model parameters $W_{ij}\to W_{ij}+\dbar W_{ij}$:
\begin{align}\label{eq:quadratic-model-small-step}
&z_{i}(x_\delta; \theta+ \dbar \theta)\,\notag \\
=&z_{i}(x_\delta; \theta) +  \sum_{j=0}^{n_f} \dbar W_{ij}\, \le[ \fea_{j}(x_{\delta}) + \epsilon\sum_{j_1=0}^{n_f} W_{ij_1} \featwo_{j_1j}(x_{\delta}) \ri] + \frac{\epsilon}{2}\sum_{j_1,j_2=0}^{n_f} \dbar W_{ij_1}\dbar W_{ij_2}  \featwo_{j_1j_2}(x_{\delta}) +\dots \, \notag \\
=&z_{i}(x_\delta; \theta) + \sum_{j=0}^{n_f} \dbar W_{ij} \,  \feaE_{ij}(x_{\delta};\theta) + \frac{\epsilon}{2}\sum_{j_1,j_2=0}^{n_f} \dbar W_{ij_1}\dbar W_{ij_2}  \featwo_{j_1j_2}(x_{\delta}) +\dots\, ,
\end{align}
where in the last line we summarized the quantity in the square bracket in terms of an \textbf{effective feature function}\index{feature function!effective|textbf}:
\be\label{eq:effective-feature-def}
\feaE_{ij}(x_{\delta};\theta)\equiv \frac{dz_{i}(x_\delta;\theta)}{dW_{ij}}=\fea_{j}(x_{\delta}) + \epsilon\sum_{k=0}^{n_f} W_{ik} \featwo_{kj}(x_{\delta})   \, .
\ee
The utility of this is as follows: the linear response of the model to changing its parameters is as if it \emph{effectively} has a feature function, $\feaE_{ij}(x_{\delta};\theta)$, which itself depends on the value of the model parameters. Moreover, the change in the effective feature function given a small change in the model parameters $W_{ik}\to W_{ik}+\dbar W_{ik}$ is given by
 \be\label{eq:effective features-small-step}
\feaE_{ij}(x_{\delta};\theta+ \dbar \theta) = \feaE_{ij}(x_{\delta};\theta) + \epsilon \sum_{k=0}^{n_f} \dbar W_{ik}\, \featwo_{kj}(x_{\delta})\, .
 \ee
Thus, the effective features $\feaE_{ij}(x_{\delta};\theta)$ are \emph{learnable}, evolving as a linear model\index{linear model!for effective features}, and for these effective features, the meta features $\featwo_{kj}(x_{\delta})$ play the role usually played by the features.\footnote{
    This role of the meta features $\featwo_{kj}(x_{\delta})$ in the linear model for the effective features explains our choice of name. Note also that in this setup, we assume that both the feature function $\fea_{j}(x)$ and the meta feature function $\featwo_{j_1j_2}(x)$ are picked by the model designer, just as the feature function was before for the linear model.
} In this way, we can think of our nonlinear model as having a hierarchical structure, where the features evolve as if they are described by a linear model according to \eqref{eq:effective features-small-step}, while the model's output evolves in a more complicated nonlinear way according to \eqref{eq:quadratic-model-small-step}.

Note that we could extend this hierarchical structure further, e.g.~by further deforming our nonlinear model \eqref{eq:nonlinear-model-def} with a term that's cubic in the parameters and includes a three-indexed \emph{meta-meta feature function}\index{meta-meta feature function} that analogously allows the meta feature functions to learn.\footnote{Such deformations are essentially equivalent to incorporating further terms in the Taylor expansion of an arbitrary general model function $z_i(x_{\delta}; \theta)$, where the linear model is the base model, and the nonlinear model \eqref{eq:nonlinear-model-def} gives the first improvement from the expansion.}
However, as the quadratic term already suffices to provide a minimal model of representation learning\index{representation learning!minimal model}\index{representation learning!minimal model} -- just as a nonzero dNTK sufficed to induce nontrivial representation learning in the MLP --
from here on, we'll explicitly truncate the model \eqref{eq:nonlinear-model-def} at the quadratic order, giving us a \textbf{quadratic model}\index{nonlinear model!quadratic model|textbf}\index{quadratic model|see{nonlinear model}}.\footnote{
To leading order in the $1/n$ expansion\index{$1/n$ expansion},
finite-width networks cannot self-consistently be described by a truncated quadratic model; instead, they fall into a class of  \emph{cubic models}\index{nonlinear model!cubic model}, with evolving meta feature functions. This is one of the ways in which such finite-width networks are \emph{non-minimal} representation learners\index{representation learning!non-minimal model} and is the main reason for discussing quadratic models first before moving on to the more complicated analysis of finite-width networks with their dynamical dNTKs.\index{dynamical dNTK}
}

\index{minimal model!of representation learning|see{representation learning}}
To understand how representation learning\index{representation learning!for quadratic models} works in this model, we should find the optimal values for the weights $W_{ij}^\star$ given a training set $\A$. Mirroring what we did before for linear models, let's minimize the MSE loss, which for our quadratic model is given by
\index{loss!MSE!nonlinear models}
\begin{align}\label{eq:quadratic-regression}
\L_{\A}(\theta)&=\frac{1}{2}\sum_{\tra\in\A}\sum_{i=1}^{n_{\text{out}}}\Big[ \y{i}{\tra} - z_i(x_{\tra}; \theta) \Big]^2  \\
&=\frac{1}{2}\sum_{\tra\in\A}\sum_{i=1}^{n_{\text{out}}}\le[\y{i}{\tra} - \sum_{j=0}^{n_f} W_{ij} \, \fea_{j}(x_{\tra})  - \frac{\epsilon}{2}\sum_{j_1,j_2=0}^{n_f} W_{ij_1}W_{ij_2} \featwo_{j_1j_2}(x_{\tra}) \ri]^2\, . \notag
\end{align}
Supervised learning\index{supervised learning!with quadratic models|see{quadratic regression}} with such a quadratic model \eqref{eq:nonlinear-model-def} doesn't have a particular name, but if it did, we'd all probably agree that its name should be \term{quadratic regression}\index{quadratic regression|seealso{quadratic model}}.
However, unlike linear regression\index{linear regression!vs.~quadratic regression} -- where the MSE loss  \eqref{eq:linear-regression} was quadratic in the model parameters -- quadratic regression -- where the loss is now \emph{quartic} in the parameters -- does not in general yield analytical solutions. Nevertheless, we can perturbatively find a solution to the quadratic regression problem by expanding in our small parameter $\epsilon$, for which the regression is \emph{nearly linear}.\index{nearly-linear model|see{nonlinear model}}\index{nearly-linear regression|see{quadratic regression}}

\subsubsection{Nearly-Linear Quadratic Regression}\index{quadratic regression!nearly-linear}
Taking the derivative of the loss $\L_\A$  \eqref{eq:quadratic-regression} with respect to the model parameter $W_{ij_0}$ and setting it to zero, we find
\be\label{eq:quadratic-optimization-implict-features}
0 = \sum_{\tra} \feaE_{ij_0}(x_{\tra};\theta^\star) \le[z_{i}(x_{\tra};\theta^\star) - \y{i}{\tra} \ri]\,,
\ee
making clear that the effective features \eqref{eq:effective-feature-def} give the linear response of the model.
Next, substituting in for the quadratic model \eqref{eq:linear-model-def-reprint} and the effective feature function \eqref{eq:effective-feature-def} and rearranging we find
\begin{align}\label{eq:nonlinear-model-implicit-expression}
&\sum_{\tra\in\A}\!\le\{\sum_{j_1=0}^{n_f}W^{\star}_{ij_1}\fea_{j_1}(x_{\tra})\fea_{j_0}(x_{\tra})+\epsilon\!\!\!\!\sum_{j_1,j_2=0}^{n_f}\!\!W^{\star}_{ij_1}W^{\star}_{ij_2}\!\le[\fea_{j_1}(x_{\tra})\featwo_{j_2j_0}(x_{\tra})\!+\frac{1}{2}\featwo_{j_1j_2}(x_{\tra})\fea_{j_0}(x_{\tra})\ri]\ri\}\, \notag\\
=&\sum_{\tra\in\A} \y{i}{\tra}\le[\fea_{j_0}(x_{\tra})+\epsilon\sum_{j_1=0}^{n_f}W^{\star}_{ij_1}\featwo_{j_1j_0}(x_{\tra})\ri]+\o{\epsilon^2}\, .
\end{align}
This expression contains the two terms we found for linear regression\index{linear regression!vs.~quadratic regression} in \eqref{eq:linear-model-implicit-expression} as well as three additional terms proportional to the meta features $\featwo_{j_1j_2}(x)$. Additionally, we truncated the $\o{\epsilon^2}$ term since $\epsilon$ is assumed to be parametrically small. 
Importantly, we see clearly that the equation is overall nonlinear, with the two terms quadratic in the weights. 
In the language of \terminate{physics}, the linear equation of \terminate{linear regression} is \emph{free} and exactly solvable, while the nonlinear equation of \terminate{quadratic regression} is \emph{interacting}.\index{interactions!dynamics} Since $\epsilon$ multiplies all the new terms associated with quadratic regression, the nonlinear terms are all small, and so  \eqref{eq:nonlinear-model-implicit-expression} exhibits \emph{weakly-interacting} dynamics.\index{interactions!dynamics!weakly-interacting} This means that we can systematically solve this nonlinear equation via \terminate{perturbation theory}. %

With that in mind, let's decompose the optimal weight matrix into a free linear part and an interacting nonlinear part as
\be\label{eq:quadratic-optimization-decomposition-parameters}
W^{\star}_{ij}\equiv\WF_{ij}+\WI_{ij}\, ,
\ee
with the idea being that the free part $\WF_{ij}$ will solve the free linear regression equation \eqref{eq:linear-model-implicit-expression}, while the interacting part $\WI_{ij}$ will solve the remaining linearized equation after substituting back in the solution for $\WF_{ij}$. Given that the quadratic regression problem \eqref{eq:nonlinear-model-implicit-expression} becomes a linear regression problem \eqref{eq:linear-model-implicit-expression} in the limit of $\epsilon\to0$, we naturally expect that the interacting part of the optimal weights should be proportional to the small parameter $\WI_{ij}=\o{\epsilon}$. 

Let's quickly review our \term{direct optimization} solution to linear regression from \S\ref{subsec:linear-models} in the current context:
defining an $(n_f+1)$-by-$(n_f+1)$ symmetric matrix \eqref{eq:no-good-matrix-name}
\be\label{eq:no-good-matrix-name-reprint}
M_{j_1j_2}\equiv\sum_{\tra\in\A}\fea_{j_1}\!(x_{\tra})\,\fea_{j_2}\!(x_{\tra})\, , %
\ee
the linear part of the quadratic regression problem \eqref{eq:nonlinear-model-implicit-expression} can be written as
\be
\sum_{j_1=0}^{n_f}\WF_{ij_1}M_{j_1j_0}=\sum_{\tra\in\A}\y{i}{\tra}\fea_{j_0}(x_{\tra})\, ,
\ee
which can be solved by the multiplication of the inverse $(M^{-1})_{j_0 j}$,
\be\label{eq:linear-regression-optimal-reprint}
\WF_{ij}=\sum_{j_0=0}^{n_{f}}\sum_{\tra\in\A}\y{i}{\tra}\fea_{j_0}(x_{\tra})\le(M^{-1}\ri)_{j_0j}\, .
\ee
Recall that the inverse $(M^{-1})_{j_0 j}$ will not uniquely exist if we're in the \emph{overparameterized}\index{overparameterization!in quadratic models} regime with more features than training examples, $(n_f+1) > \NR$, but we can use our regularization trick \eqref{eq:inverse-of-M-regularized} to pick out a particular inverse. Going forward, we will assume that we're in this overparameterized regime and that the inverse was picked in this way.

Next, plugging in our decomposition \eqref{eq:quadratic-optimization-decomposition-parameters} into our equation \eqref{eq:nonlinear-model-implicit-expression} and collecting the terms of order $\epsilon$, remembering also that $\WI_{ij}=\o{\epsilon}$, we find for our linearized interacting dynamics,
\begin{align}
\sum_{j_1=0}^{n_f}\WI_{ij_1}M_{j_1j_0}=&\epsilon\sum_{j_1=0}^{n_f}\WF_{ij_1}\sum_{\tra\in\A}\y{i}{\tra}\featwo_{j_1j_0}(x_{\tra})-\epsilon\sum_{j_1,j_2=0}^{n_f}\WF_{ij_1}\WF_{ij_2}\sum_{\tra\in\A}\fea_{j_1}(x_{\tra})\featwo_{j_2j_0}(x_{\tra})\, \notag\\
&-\frac{\epsilon}{2}\sum_{j_1,j_2=0}^{n_f}\WF_{ij_1}\WF_{ij_2}\sum_{\tra\in\A}\fea_{j_0}(x_{\tra})\featwo_{j_1j_2}(x_{\tra}) +\o{\epsilon^2}\, .
\end{align}
Here on the right-hand side, the first two terms actually cancel each other, since the free solution satisfies
\be\label{eq:one-nice-property-of-free-solution}
\sum_{j=0}^{n_f}\WF_{ij}\fea_{j}(x_{\tra})=\y{i}{\tra}\, ,
\ee
i.e.~for overparameterized models, linear part of the optimized model can correctly predict all the training-set examples.\footnote{This is shown in detail in \S\ref{subsec:kernel-methods}. Note that for \emph{underparametrized}\index{underparameterization} models, the solution can still be analyzed, but the details will be different. We're focusing on overparameterized models here since \emph{(i)} deep learning models are typically overparameterized and \emph{(ii)} we suspect that the sort of representation learning our model exhibits is most useful in that regime.
We'll elaborate on this quite a bit more in Epilogue~\ref{epi:overparameterization}.
}
After making that cancellation, we can multiply by the inverse $(M^{-1})_{j_0 j}$ to find a solution:
\begin{align}\label{eq:nearly-linear-regression-optimal}
\WI_{ij}=-\frac{\epsilon}{2}\sum_{j_1,j_2,j_3=0}^{n_f}\WF_{ij_1}\WF_{ij_2}\sum_{\tra\in\A}\le[\featwo_{j_1j_2}(x_{\tra})\fea_{j_3}(x_{\tra})\ri]\le(M^{-1}\ri)_{j_3j}+\o{\epsilon^2}\, .
\end{align}
In particular, the free solution, \eqref{eq:linear-regression-optimal-reprint}, and the interacting solution, \eqref{eq:nearly-linear-regression-optimal}, together solve the nonlinear optimization problem \eqref{eq:nonlinear-model-implicit-expression} to order $\epsilon$.\footnote{When doing nearly-linear quadratic regression practically, it would probably make the most sense to first find the optimal linear parameters $\WF_{ij}$ and then plug them back into \eqref{eq:nearly-linear-regression-optimal} to find the additional nonlinear parameters $\WI_{ij}$.}

Finally, having obtained the solution, we can throw away the training data and simply store the optimal parameters $W^{\star}_{ij}=\WF_{ij}+\WI_{ij}$, making predictions on novel test inputs $x_{\tea}$ as
\begin{align}\label{eq:nearly-kernel-prediction-nearly-kernel-methods-before-duality}
z_i(x_{\tea}; \theta^{\star})=&\sum_{j=0}^{n_f}W^{\star}_{ij}\fea_j(x_{\tea})+ \frac{\epsilon}{2}\sum_{j_1,j_2=0}^{n_f}W^{\star}_{ij_1}W^{\star}_{ij_2}\featwo_{j_1j_2}(x_{\tea})\, \\
=&\sum_{j=0}^{n_f}\WF_{ij}\fea_j(x_{\tea})+\sum_{j=0}^{n_f}\WI_{ij}\fea_j(x_{\tea})  + \frac{\epsilon}{2}\sum_{j_1,j_2=0}^{n_f}\WF_{ij_1}\WF_{ij_2}\featwo_{j_1j_2}(x_{\tea})+\o{\epsilon^2}\, \notag\\
=&\frac{1}{2}\sum_{j=0}^{n_f}W^{\star}_{ij}\le[\fea_j(x_{\tea})+\feaE_{ij}(x_{\tea}; \theta^{\star})\ri]+\o{\epsilon^2}\, .\notag
\end{align}
Here, we've given two different ways to think about the optimal output.
On the first and second lines, we simply have the prediction of the quadratic model\index{nonlinear model!quadratic model} \eqref{eq:nonlinear-model-def} expressed in terms of the fixed features $\fea_j(x_{\tea})$ and the fixed meta features $\featwo_{j_1j_2}(x_{\tea})$. This presentation makes the nonlinearity manifest.
After regrouping the terms and using our definition \eqref{eq:effective-feature-def}, in the last line we instead wrote the model prediction in the form of a linear model, where we see that features in this interpretation are the mean of the fixed unlearned features and the learned effective features. From this perspective, representation learning\index{representation learning!for quadratic models} is manifest: the effective features $\feaE_{ij}(x_{\tea}; \theta^{\star})$ depend on the training data through the optimal parameters, $W^{\star}_{ij}=W^{\star}_{ij}(x_{\tra},y_{\tra})$.

In summary, our nonlinear quadratic model \eqref{eq:nonlinear-model-def} serves as a minimal model of representation learning.
As we will see soon, this captures the mechanism of feature evolution for a nonzero but fixed dNTK.

\subsubsection{\emph{Aside}: Model Comparison of Linear Regression and Quadratic Regression}

Before we move on to the dual sample-space description of the quadratic model, 
let's briefly perform a model comparison between linear regression and quadratic regression.
In particular, let's think about the \neo{model complexity} of these classes of models.\footnote{
    For a further discussion of model complexity,  with a direct focus on overparameterized deep learning models, see Epilogue~\ref{epi:overparameterization}.
}

For both linear and quadratic regression, the number of model parameters is given by the number of elements in the combined weight matrix $W_{ij}$:
\be\label{eq:parameters-in-linear-model}
P \equiv n_{\text{out}} \times (n_f+1)\, .
\ee
Since both models completely memorize the same training set for a fixed and equal number of parameters,
we obviously cannot naively use the \terminate{Occam's razor} heuristic (\S\ref{subsec:bayesian-model-comparison}) for model comparison\index{model comparison!linear model vs.~quadratic model}.
This makes our model comparison\index{model comparison!linear model vs.~quadratic model} somewhat subtle.

On the one hand, there is a sense in which the quadratic model is more complicated, as it computes far more functions of the input per parameter. Specifically, on a per parameter basis, we need to specify a far greater number of underlying functions  for the quadratic model than we do for the linear model:
i.e.~we need
\be\label{eq:quadratic-model-feature-to-parameters}
(n_f+1)+ \le[\frac{1}{2}(n_f +1)(n_f+2)\ri] = 
\o{P^2} \,
\ee
numbers to specify $\fea_{j}(x)$ \emph{and} $\featwo_{j_1j_2}(x)$,
while we need \emph{just}
\be\label{eq:linear-model-feature-to-parameters}
(n_f+1)  = \o{P} \, 
\ee
numbers to specify  $\fea_{j}(x)$.
In particular, the counting of the model functions is dominated by the meta feature functions $\featwo_{j_1j_2}(x)$.
As such, this type of complexity is not really captured by the counting of model parameters, $P$; instead, it is expressed in the structure of the model, with the addition of the meta feature functions $\featwo_{j_1j_2}(x)$.

On the other hand, we can interpret these additional meta feature functions as \emph{constraining} the quadratic model according to an explicit inductive bias for representation learning\index{inductive bias!for representation learning in nonlinear models}.\footnote{
Similarly, we could naively think that the addition of a \neo{regularization}\index{regularization!interpretation of representation learning} term such as $\sum_{\mu=1}^P a_{\mu}\theta_\mu^2$ to the loss as making a model more complex with its extra structure, despite being a well-known remedy for overfitting. Instead, it's probably better to think of this regularization term as an inductive bias for \emph{constraining} the norm squared of the optimized model parameters.\label{footnote:regularization-again}
}
In particular, this additional structure alters the linear model solution \eqref{eq:linear-regression-optimal-reprint} with the addition of $\o{\epsilon}$ tunings $\WI_{ij}$, constrained by the $\o{P^2}$ meta features that are defined before any learning takes place.
Assuming these meta feature functions are \emph{useful}, we might expect that the quadratic model will overfit less and generalize better.\footnote{Interestingly, for the quadratic model, the number of \emph{effective feature functions}\index{feature function!effective}\index{effective feature function|see{feature function}} \eqref{eq:effective-feature-def}, is actually the same as the number of model parameters: $n_{\text{out}} \times (n_f +1) = P$.
Since it's only through these effective features that the meta feature functions enter the model predictions, cf.~\eqref{eq:nearly-kernel-prediction-nearly-kernel-methods-before-duality}, this further underscores that, despite the additional model structure, there aren't actually $\o{P^2}$ independent degrees of freedom that can be applied towards fitting the training data.
} (In fact, that was the whole point of introducing them.)

This latter point is worth a little further discussion. One typical signature of overfitting is that the parameters are extremely \textbf{finely-tuned}\index{fine tuning|textbf}\index{overfitting!by fine tuning the parameters}; 
these tunings are in some sense \emph{unnatural}\index{naturalness}\index{naturalness|seealso{fine tuning}} as they can arise from the extreme flexibility afforded to overparameterized models, enabling models to pass through all the training points \emph{exactly}, to the extreme detriment of the test predictions.\footnote{
   This is very commonly illustrated by using polynomial basis of feature functions for linear regression, which is sometimes called \neo{polynomial regression}. In this case, consider a linear model of a scalar function $f(x)$ with a scalar input $x$:
\be
z(x_{\delta}; \theta) %
=\sum_{j=0}^{n_f}w_{j}\fea_j(x_{\delta}) \equiv w_{0} + w_{1} x_\delta +  w_{2} x_\delta^2 %
+\dots +  w_{n_f} x_\delta^{n_f}   \, .
\ee
If there's any noise at all in the data, when the model is overparameterized, $n_f+1 > \NR$, the plot of this one-dimensional function will make $\sim n_f$ wild turns to go through the $\NR$ training points. (This is particularly evocative if the target function is a simple linear function with noise, 
i.e.~$f(x) = a x + b + \varepsilon$, with $\varepsilon$ a zero-mean Gaussian noise with small variance $\sigma^2_{\varepsilon} \ll 1$.) 
In order to make these turns, the optimal coefficients, $w_{j}^\star$, computed by \eqref{eq:linear-regression-optimal-reprint}, will
be finely-tuned to many significant figures. This kind of fine-tuning problem in model parameters is indicative of the model being unnatural or wrong; in fact, the analog of this problem in high-energy theoretical physics\index{physics} 
is called \neo{naturalness}
(see e.g.~\cite{Giudice:2008bi} for a non-technical discussion).
}
Adding a regularization term on the parameter norm -- i.e.~the one we just discussed in footnote~\ref{footnote:regularization-again} -- combats such tuning: the additional constraints on the optimization problem drive the norm of the parameters towards zero, effectively promoting parameter sparsity.
Here, we see that since the nonlinear contribution to the optimal weights, $\WI_{ij}$, is fixed to be small, $\o{\epsilon}$, it's adding constraints that -- if they're useful -- can combat any fine tunings that may appear in the linear solution, $\WF_{ij}$, and lead to better generalization.

\subsection{Nearly-Kernel Methods}\label{subsec:nearly-kernel-methods}

Now that we have some more parameter-space intuition for the potential advantages of nonlinear models over linear models, 
let's now develop a \emph{dual}\index{duality} sample-space description of quadratic regression where a quadratic-model analog of the dNTK appears naturally. %

Starting with the expression in the second line of the prediction formula  \eqref{eq:nearly-kernel-prediction-nearly-kernel-methods-before-duality} and plugging in the  free solution \eqref{eq:linear-regression-optimal-reprint} and the interacting solution \eqref{eq:nearly-linear-regression-optimal}, we get
\begin{align}\label{eq:nearly-linear-regression-optimal-cleaner}
z_i(x_{\tea}; \theta^{\star})=&\sum_{\tra\in\A}\y{i}{\tra}\le[\sum_{j_1,j_2=0}^{n_{f}}\fea_{j_1}(x_{\tra})\le(M^{-1}\ri)_{j_1j_2}\fea_{j_2}(x_{\tea})\ri]\, \\
&+ \frac{\epsilon}{2}\sum_{\tra_1,\tra_2\in\A}\y{i}{\tra_1}\y{i}{\tra_2}\sum_{j_1,j_2,j_3,j_4=0}^{n_f}\fea_{j_1}(x_{\tra_1})\le(M^{-1}\ri)_{j_1j_3}\fea_{j_2}(x_{\tra_2})\le(M^{-1}\ri)_{j_2j_4}\, \notag\\
&\quad\quad\, \times\!\le[\featwo_{j_3j_4}(x_{\tea})-\!\!\sum_{\tra\in\A}\featwo_{j_3j_4}(x_{\tra})\!\!\!\sum_{j_5,j_6=0}^{n_f}\!\!\fea_{j_5}(x_{\tra})\le(M^{-1}\ri)_{j_5j_6}\!\!\fea_{j_6}(x_{\tea})\ri]+\o{\epsilon^2}\, .\notag
\end{align}
To simplify this expression, recall formula  \eqref{eq:kernel-trick-result} that we derived when discussing \neo{kernel methods},
\be\label{eq:kernel-trick-result-reprint}
\sum_{j_1,j_2=0}^{n_{f}}\fea_{j_1}(x_{\tra})\le(M^{-1}\ri)_{j_1j_2}\fea_{j_2}(x_{\tea})=\kerm_{\tea\tra_1}\kermsub^{\tra_1\tra}\, ,
\ee
where the \emph{kernel}\index{nearly-kernel methods!kernel} was defined in \eqref{eq:kernel-with-features} as 
\be\label{eq:kernel-with-features-reprint}
\kerm_{\delta_1\delta_2}\equiv\kerm\!\le(x_{\delta_1},x_{\delta_2} \ri) \equiv \sum_{j=0}^{n_{f}} \fea_j\!\le(x_{\delta_1}\ri) \fea_j\!\le(x_{\delta_2}\ri) \, ,
\ee
and provided a measure of similarity between two inputs $x_{i;\delta_1}$ and $x_{i;\delta_2}$ in \terminate{feature space}.
Plugging this formula \eqref{eq:kernel-trick-result-reprint} back into our quadratic regression prediction formula \eqref{eq:nearly-linear-regression-optimal-cleaner}, we get
\begin{align}\label{eq:nearly-linear-regression-optimal-cleanerer}
z_i(x_{\tea}; \theta^{\star})=&\sum_{\tra_1,\tra_2\in\A}\kerm_{\tea\tra_1}\kermsub^{\tra_1\tra_2}\y{i}{\tra_2}\, \\
&+ \frac{\epsilon}{2}\sum_{\tra_1,\tra_2\in\A}\y{i}{\tra_1}\y{i}{\tra_2}\sum_{j_1,j_2,j_3,j_4=0}^{n_f}\fea_{j_1}(x_{\tra_1})\le(M^{-1}\ri)_{j_1j_3}\fea_{j_2}(x_{\tra_2})\le(M^{-1}\ri)_{j_2j_4}\, \notag\\
&\quad \quad \quad \quad \quad \quad \quad \quad \times\le[\featwo_{j_3j_4}(x_{\tea})-\sum_{\tra_1,\tra_2\in\A}\kerm_{\tea\tra_1}\kermsub^{\tra_1\tra_2}\featwo_{j_3j_4}(x_{\tra_2})\ri]+\o{\epsilon^2}\, ,\notag
\end{align}
which is already starting to look a little better.

To simplify this expression further, we need to understand an object of the following form:
\be
\sum_{j_1,j_2,j_3,j_4=0}^{n_f}\epsilon\,\fea_{j_1}(x_{\tra_1})\le(M^{-1}\ri)_{j_1j_3}\fea_{j_2}(x_{\tra_2})\le(M^{-1}\ri)_{j_2j_4}\featwo_{j_3j_4}(x_{\delta})\, .
\ee
Taking inspiration from the steps \eqref{eq:kernel-trick} that we took to derive our kernel-method formula \eqref{eq:kernel-trick-result-reprint}, let's act on this object with two training-set kernels:
\begin{align}\label{eq:nearly-kernel-trick}
&\sum_{\tra_1,\tra_2\in\A}\le[\sum_{j_1,j_2,j_3,j_4=0}^{n_f}\epsilon\,\fea_{j_1}(x_{\tra_1})\le(M^{-1}\ri)_{j_1j_3}\fea_{j_2}(x_{\tra_2})\le(M^{-1}\ri)_{j_2j_4}\featwo_{j_3j_4}(x_{\delta})\ri]\kermsub_{\tra_1\tra_3}\kermsub_{\tra_2\tra_4}\, \notag\\
=&\sum_{\tra_1,\tra_2\in\A}\sum_{j_1,\ldots,j_6=0}^{n_f}\epsilon\,\fea_{j_1}(x_{\tra_1})\le(M^{-1}\ri)_{j_1j_3}\fea_{j_2}(x_{\tra_2})\le(M^{-1}\ri)_{j_2j_4}\, \notag\\
&\quad\quad\times\featwo_{j_3j_4}(x_{\delta})\fea_{j_5}(x_{\tra_1})\fea_{j_5}(x_{\tra_3})\fea_{j_6}(x_{\tra_2})\fea_{j_6}(x_{\tra_4})\, \notag\\
=&\sum_{j_1,\ldots,j_6=0}^{n_f}\epsilon\,M_{j_1j_5}\le(M^{-1}\ri)_{j_1j_3}M_{j_2j_6}\le(M^{-1}\ri)_{j_2j_4}\featwo_{j_3j_4}(x_{\delta})\fea_{j_5}(x_{\tra_3})\fea_{j_6}(x_{\tra_4})\, \notag\\
=&\sum_{j_1,j_2=0}^{n_f}\epsilon\,\featwo_{j_1j_2}(x_{\delta})\fea_{j_1}(x_{\tra_3})\fea_{j_2}(x_{\tra_4})\, .
\end{align}
Here on the second line, we used the definition of the kernel \eqref{eq:kernel-with-features-reprint} to swap both kernels for feature functions, on the third line we used the definition of the symmetric matrix $M_{j_1j_2}$, \eqref{eq:no-good-matrix-name-reprint}, to replace two pairs of feature functions, and on the final line we simply canceled these matrices against their inverses. 

This last expression suggests that an important object worth defining is 
\be\label{eq:meta-kernel-definition}
\mkerm_{\delta_0\delta_1\delta_2}\equiv\sum_{j_1,j_2=0}^{n_f}\epsilon\,\featwo_{j_1j_2}(x_{\delta_0})\fea_{j_1}(x_{\delta_1})\fea_{j_2}(x_{\delta_2})\, ,
\ee
which we will call the \textbf{meta kernel}\index{kernel!meta kernel|see{nearly-kernel methods}}\index{meta kernel|see{nearly-kernel methods}}\index{nearly-kernel methods!meta kernel|textbf}.\footnote{
    An alternate name for this object is the \emph{differential of the kernel}, which we would consider symbolizing as $\mkerm_{\delta_0\delta_1\delta_2} \to \text{d}k_{\delta_0\delta_1\delta_2}$. This name-symbol pair highlights the connection we're about to make to finite-width networks, but is perhaps less general in the context of making a broader model of representation learning.\index{nearly-kernel methods!meta kernel!other potential names}
} Analogous to the kernel methods' kernel \eqref{eq:kernel-with-features-reprint}, the meta kernel is a parameter-independent tensor, symmetric under an exchange of its final two sample indices $\delta_1 \leftrightarrow  \delta_2$, and given entirely in terms of the fixed feature and meta feature functions that define the model.  
One way to think about \eqref{eq:meta-kernel-definition} is that for a fixed particular input, $x_{\delta_0}$, the meta kernel computes a different feature-space inner product between the two other inputs, $x_{\delta_{1}}$ and $x_{\delta_{2}}$.
Note also that due to the inclusion of the small parameter $\epsilon$ into the definition of the meta kernel \eqref{eq:meta-kernel-definition}, we should think of $\mkerm_{\delta_0\delta_1\delta_2}$ as being parametrically small too.

With this definition, the relation \eqref{eq:nearly-kernel-trick} can now be succinctly summarized as
\begin{align}
&\sum_{j_1,j_2,j_3,j_4=0}^{n_f}\epsilon\,\fea_{j_1}\!(x_{\tra_1})\le(M^{-1}\ri)_{j_1j_3}\fea_{j_2}(x_{\tra_2})\le(M^{-1}\ri)_{j_2j_4}\featwo_{j_3j_4}(x_{\delta})\, \\
=&\sum_{\tra_3,\tra_4\in\A}\mkerm_{\delta\tra_3\tra_4}\kermsub^{\tra_3\tra_1}\kermsub^{\tra_4\tra_2}\, .\notag
\end{align}
Finally, plugging this simple relation back into \eqref{eq:nearly-linear-regression-optimal-cleanerer}, we get
\begin{align}\label{eq:nearly-linear-regression-optimal-cleanest}
z_i(x_{\tea}; \theta^{\star})=&\sum_{\tra_1,\tra_2\in\A}\kerm_{\tea\tra_1}\kermsub^{\tra_1\tra_2}\y{i}{\tra_2}\, \\
&+ \frac{1}{2}\sum_{\tra_1,\ldots,\tra_4\in\A}\!\le[\mkerm_{\tea\tra_1\tra_2}-\!\!\!\!\sum_{\tra_5,\tra_6\in\A}\kerm_{\tea\tra_5}\kermsub^{\tra_5\tra_6}\mkerm_{\tra_6\tra_1\tra_2}\ri]\!\!\le(\kermsub^{\tra_1\tra_3}\y{i}{\tra_3}\ri)\!\!\le(\kermsub^{\tra_2\tra_4}\y{i}{\tra_4}\ri)\, .\notag
\end{align}
When the prediction of a quadratic model\index{nonlinear model!quadratic model} is computed in this way, we'll hereby make it known as a \emph{nearly-kernel machine}\index{nearly-kernel machine|see{nearly-kernel methods}} or \term{nearly-kernel methods}.\footnote{Unlike kernel methods, this solution actually depends on the details of the \terminate{learning algorithm}. For instance, if we had optimized the quadratic-regression loss \eqref{eq:quadratic-regression} by \neo{gradient descent} rather than by \neo{direct optimization} \eqref{eq:quadratic-optimization-implict-features}, then we would have found instead (for zero initialization $W_{ij}=0$)
\begin{align}\label{eq:nearly-linear-regression-optimal-with-projectors}
z_i(x_{\tea}; \theta^{\star})=&\sum_{\tra_1,\tra_2\in\A}\kerm_{\tea\tra_1}\kermsub^{\tra_1\tra_2}\y{i}{\tra_2}\, \\
&+ \sum_{\tra_1,\ldots,\tra_4\in\A}\!\le[\mkerm_{\tra_1\tea\tra_2}-\!\!\!\!\sum_{\tra_5,\tra_6\in\A}\kerm_{\tea\tra_5}\kermsub^{\tra_5\tra_6}\mkerm_{\tra_1\tra_6\tra_2}\ri] \algodNTKone^{\tra_1\tra_2\tra_3\tra_4} \y{i}{\tra_3}  \y{i}{\tra_4} \, \notag\\
&+ 
\sum_{\tra_1,\ldots,\tra_4\in\A}\!\le[\mkerm_{\tea\tra_1\tra_2}-\!\!\!\!\sum_{\tra_5,\tra_6\in\A}\kerm_{\tea\tra_5}\kermsub^{\tra_5\tra_6}\mkerm_{\tra_6\tra_1\tra_2}\ri] 
\algodNTKtwo^{\tra_1\tra_2\tra_3\tra_4}  \y{i}{\tra_3}  \y{i}{\tra_4}
\, \notag
\end{align}
for our nearly-kernel methods prediction formula, where the \emph{algorithm projectors}\index{algorithm projector} are given by
\begin{align}
\algodNTKone^{\tra_1\tra_2\tra_3\tra_4} \equiv& \kermsub^{\tra_1\tra_3} \kermsub^{\tra_2\tra_4} - \sum_{\tra_5} \kermsub^{\tra_2\tra_5}  \geosumtwo^{\tra_1\tra_5\tra_3\tra_4}
\, ,\\
\algodNTKtwo^{\tra_1\tra_2\tra_3\tra_4} \equiv& \kermsub^{\tra_1\tra_3} \kermsub^{\tra_2\tra_4} - \sum_{\tra_5} \kermsub^{\tra_2\tra_5}  \geosumtwo^{\tra_1\tra_5\tra_3\tra_4} + \frac{\eta}{2}\geosumtwo^{\tra_1\tra_2\tra_3\tra_4} \, ,
\end{align}
with the tensor $\geosumtwo^{\tra_1\tra_2\tra_3\tra_4}$ implicitly satisfying
\be\label{eq:implicit-X-tensor-nearly-kernel-version}
\sum_{\tra_3,\tra_4\in\A}\geosumtwo^{\tra_1\tra_2\tra_3\tra_4}\le(\kermsub_{\tra_3\tra_5}\delta_{\tra_4\tra_6}+\delta_{\tra_3\tra_5}\kermsub_{\tra_4\tra_6}-\eta\kermsub_{\tra_3\tra_5}\kermsub_{\tra_4\tra_6}\ri)=\delta^{\tra_1}_{\ \tra_5}\delta^{\tra_2}_{\ \tra_6}\, ,
\ee
and global learning rate $\eta$.
The origin of this gradient-descent solution should be clear after you traverse through \S\ref{subsec:real-GD-at-finite-width}. Such \neo{algorithm dependence} is to be expected for a nonlinear overparameterized model and is an important characteristic of finite-width networks as well. However, for the rest of the section we will continue to analyze the direct optimization formula, \eqref{eq:nearly-linear-regression-optimal-cleanest}, with $\algodNTKone^{\tra_1\tra_2\tra_3\tra_4}=0$ and $\algodNTKtwo^{\tra_1\tra_2\tra_3\tra_4}=\kermsub^{\tra_1\tra_3} \kermsub^{\tra_2\tra_4}/2$.\label{footnote:algo-dependence-quadratic-regression}
}

Analogous to linear models, we again have two ways of thinking about the solution of our nonlinear quadratic model's predictions: on the one hand, we can use the optimal parameters \eqref{eq:linear-regression-optimal-reprint} and \eqref{eq:nearly-linear-regression-optimal} to make predictions \eqref{eq:nearly-kernel-prediction-nearly-kernel-methods-before-duality}; 
on the other hand, we can make \emph{nearly-kernel predictions}\index{nearly-kernel methods!prediction} using the formula \eqref{eq:nearly-linear-regression-optimal-cleanest}  in which the features, the meta features, and the model parameters do not appear. 
That is, we've successfully traded our feature-space quantities $\fea_{j}(x)$, $\featwo_{j_1j_2}(x)$, and $W^{\star}_{ij}$ for sample-space quantities $ \kerm_{\delta\tra}$, $\mkerm_{\delta_0\tra_1\tra_2}$, and $\y{i}{\tra}$. 
As was the case before for kernel methods, this works because all the feature indices are contracted in our prediction formula \eqref{eq:nearly-kernel-prediction-nearly-kernel-methods-before-duality}, and so only combinations of the form $\kerm_{\delta\tra}$ and $\mkerm_{\delta_0\tra_1\tra_2}$ ever show up in the result and not the value of the features or meta features themselves.\footnote{Just as we discussed for kernel methods in footnote~\ref{footnote:kernel-vs-feature-functions} of \S\ref{ch:NTHb}, in some situations we expect that specifying and evaluating the meta kernel $\mkerm_{\delta_0\delta_1\delta_2}$
is much simpler than specifying and evaluating meta feature function $\featwo_{j_1j_2}(x)$.
Although picking these out of the thin air seems difficult, perhaps there are other inspired ways of determining these functions that don't require an underlying description in terms of neural networks.
It would be interesting to determine the necessary and sufficient conditions for a general three-input function, $\mkerm(x_{\delta_0}, x_{\delta_1}, x_{\delta_2}) \equiv \mkerm_{\delta_0\delta_1\delta_2}$, to be a meta kernel.
} 
This duality between the microscopic feature-space description of the model and a macroscopic sample-space description is another realization of the \terminate{effective theory} approach discussed in \S\ref{sec:ET-approach}, and we will return to comment more broadly on this duality in Epilogue~\ref{epi:overparameterization} after we discuss the dynamics of finite-width networks in \S\ref{ch:eot}.

Finally, as we saw before for kernel methods, the nearly-kernel prediction is computed by direct comparison with previously-seen examples. In this case, it has the same piece linear in the true outputs proportional to $\y{i}{\tra_2}$, and also has a new piece that's quadratic in the true output across different training examples proportional to $\y{i}{\tra_1} \y{i}{\tra_2}$. In this way, nearly-kernel methods are also \emph{memory-based} methods that involve memorizing the entire training set.\index{memory-based method|see{nearly-kernel methods}}\index{nearly-kernel methods!as a memory-based method}

\subsubsection{Trained-Kernel Prediction}\index{nearly-kernel methods!trained kernel|seealso{trained NTK}}
Even though these \terminate{nearly-kernel methods} are \emph{very-nearly} \terminate{kernel methods},
there's a real qualitative difference between them due to the presence of interactions between the parameters.\index{interactions!nearly-kernel methods}
In the feature-space picture described in \S\ref{subsec:nonlinear-models}, this difference manifested itself in terms of the nontrivial feature learning for the effective features $\feaE_{ij}(x,\theta)$, as expressed in the last line of the quadratic model prediction formula \eqref{eq:nearly-kernel-prediction-nearly-kernel-methods-before-duality}. To better understand this from the dual\index{duality} sample-space picture, let's analogously define an \textbf{effective kernel}\index{effective kernel|see{nearly-kernel methods}}\index{nearly-kernel methods!effective kernel}\index{kernel!effective kernel|see{nearly-kernel methods}} 
\begin{align}\label{eq:effective-kernel-def}
\kermE_{ii;\delta_1 \delta_2 }(\theta) \equiv& \sum_{j=0}^{n_f} \feaE_{ij}(x_{\delta_1};\theta) \, \feaE_{ij}(x_{\delta_2};\theta) \, ,
\end{align}
which measures a parameter-dependent similarity between two inputs $x_{\delta_1}$ and $x_{\delta_2}$ using our effective features \eqref{eq:effective-feature-def}. Interestingly, we see that the model actually gives a different effective kernel for each output component $i$.\footnote{Here, the use of two $i$'s in the subscript of the effective kernel to represent the output-component is just our convention; we'll later require a version with off-diagonal components in the slightly-less minimal model~\eqref{eq:stochastic-kernel-ntk-parameter-dependent}.} 
Let's try to understand this a little better by evaluating the
effective kernel\index{nearly-kernel methods!effective kernel} at the \terminate{end of training}:
\begin{align}\label{eq:evolving-kernel-methods-kernel}
\kermE_{ii;\delta_1 \delta_2 }(\theta^{\star}) \equiv& \sum_{j=0}^{n_f} \feaE_{ij}(x_{\delta_1};\theta^{\star}) \, \feaE_{ij}(x_{\delta_2};\theta^{\star}) \, \\
=& \sum_{j=0}^{n_f} \fea_{j}(x_{\delta_1}) \, \fea_{j}(x_{\delta_2})+\!\!\!\!\!\sum_{j_1,j_2=0}^{n_f}\!\!\!\WF_{ij_1}\le[\featwo_{j_1j_2}(x_{\delta_1})\fea_{j_2}(x_{\delta_2})+ \featwo_{j_1j_2}(x_{\delta_2})\fea_{j_2}(x_{\delta_1})\ri]+ \o{\epsilon^2}\, \notag\\
=&\kerm_{\delta_1\delta_2}+\sum_{\tra_1,\tra_2\in\A}(\mkerm_{\delta_1\delta_2\tra_1}+\mkerm_{\delta_2\delta_1\tra_1})\kermsub^{\tra_1\tra_2}\y{i}{\tra_2}+\o{\epsilon^2}\, .
\notag
\end{align}
To get this last result
on the final line we plugged in the free solution \eqref{eq:linear-regression-optimal-reprint}, and then 
secretly used the following relation
\be
\fea_{j_0}(x_{\tra}) \le(M^{-1}\ri)_{j_0j_1}\featwo_{j_1j_2}(x_{\delta_1})\fea_{j_2}(x_{\delta_2})=\sum_{\tra_1\in\A}\mkerm_{\delta_1\delta_2\tra_1}\kermsub^{\tra_1\tra}\, ,
\ee
which can be derived with manipulations analogous to those that we used in \eqref{eq:kernel-trick} and \eqref{eq:nearly-kernel-trick}.\footnote{
    Note that if we had instead optimized the quadratic-regression loss, \eqref{eq:quadratic-regression}, using gradient descent, then the effective kernel\index{nearly-kernel methods!effective kernel} at the end of training, $\kermE_{ii;\delta_1 \delta_2 }(\theta^{\star})$, would have a different expression than the one above, \eqref{eq:evolving-kernel-methods-kernel}, for direct optimization, cf.~our discussion in footnote~\ref{footnote:algo-dependence-quadratic-regression}.
} 
Here, in \eqref{eq:evolving-kernel-methods-kernel} we see that the effective kernel\index{nearly-kernel methods!effective kernel} is shifted from the kernel and includes a contribution proportional to the meta kernel as well as the true training outputs $\y{i}{\tra}$; this is what gives the effective kernel its output-component dependence. 

Finally, let's define one more kernel:
\be\label{eq:learned-unlearned-kernels-averaged}
\kermA_{ii;\delta_1\delta_2}\equiv\frac{1}{2}\le[\kerm_{\delta_1\delta_2}+\kermE_{ii;\delta_1\delta_2}(\theta^{\star})\ri]\, .
\ee
This \textbf{trained kernel}\index{trained kernel|see{nearly-kernel methods}}\index{nearly-kernel methods!trained kernel}\index{kernel!trained kernel|see{nearly-kernel methods}} averages between the simple kernel methods' kernel\index{kernel methods!kernel} from the linear model and the learned nearly-kernel methods' effective kernel.\index{nearly-kernel methods!effective kernel} 
Defining the inverse of the trained-kernel submatrix evaluated on the training set  in the usual way,
\be
\sum_{\tra_2\in\A} \kermAsub_{ii}^{\tra_1\tra_2} \kermAsub_{ii;\tra_2\tra_3} = \delta^{\tra_1}_{\ \tra_3} \, ,
\ee
the utility of this final formulation is that
the nearly-kernel prediction formula \eqref{eq:nearly-linear-regression-optimal-cleanest} can now be compressed as
\be\label{eqtrained-kernel-prediction}
z_i(x_{\tea}; \theta^{\star})=\sum_{\tra_1,\tra_2\in\A}\kermA_{ii;\tea\tra_1}\kermAsub_{ii}^{\tra_1\tra_2}\y{i}{\tra_2}+\o{\epsilon^2}\, ,
\ee
taking the form of a \emph{kernel prediction}\index{kernel methods!prediction}, but with the benefit of nontrivial feature evolution incorporated into the trained kernel.\footnote{
    To verify the formula, use the definition of the trained kernel\index{nearly-kernel methods!trained kernel}, \eqref{eq:learned-unlearned-kernels-averaged}, then expand in the effective kernel\index{nearly-kernel methods!effective kernel} using the Schwinger-Dyson equations\index{Schwinger-Dyson equations} \eqref{eq:stochastic-metric-inversion} to evaluate the matrix inverse. The result should agree with the nearly-kernel prediction formula \eqref{eq:nearly-linear-regression-optimal-cleanest}.
} This is how representation learning manifests itself in \terminate{nearly-kernel methods}.

Finally, note that in our minimal model\index{representation learning!minimal model} of representation learning\index{representation learning!minimal model}, there's no \emph{wiring}\index{nearly-kernel methods!wiring}\index{wiring!in nearly-kernel methods|see{nearly-kernel methods}} or mixing among the $n_{\text{out}}$ different output components: while the prediction $z_{i}(x_{\tea};\theta^{\star})$ is quadratic in the true output $\y{i}{\tra}$ -- most easily seen in  \eqref{eq:nearly-linear-regression-optimal-cleanest} -- it still only involves the $i$-th component. From the perspective of the \textbf{trained-kernel prediction}\index{nearly-kernel methods!trained kernel!prediction|textbf}, \eqref{eqtrained-kernel-prediction}, each output component $i$ has a \emph{different} trained kernel associated with its prediction, but the $i$-th prediction never depends on other true output components $\y{i'}{\tra}$ with $i' \neq i$. 

However, this lack of wiring is by our design; this representation-learning model is really intended to be \emph{minimal}. To enable mixing of the output components, we'll have to slightly generalize the quadratic model. This we'll do next when we explain how finite-width networks can be described in this nearly-kernel methods framework.

\subsection{Finite-Width Networks as Nonlinear Models}\label{subsec:nonlinear-at-finite}\index{differential of the neural tangent kernel!connection to representation learning}
While the discussion so far in this section has been somewhat disconnected from the deep learning framework, much of it should still feel pretty familiar to you. For instance, the  formula for the effective kernel\index{nearly-kernel methods!effective kernel} at the end of training, \eqref{eq:evolving-kernel-methods-kernel}, seems like it could be related to the update to the NTK, \eqref{eq:dNTK-naming}, if we identify the meta kernel $\mkerm_{\delta_0\delta_1\delta_2}$ with the dNTK $\dNTK_{i_0i_1i_2;\delta_0\delta_1\delta_2}$ and also make the previous identifications that we made in \S\ref{subsec:linear-at-infinity} between the kernel methods' kernel and the NTK. Let's now make these connections between finite-width networks and nonlinear models more precise.

To start, for neural networks, let us define an analog of the effective feature function $\feaE_{ij}(x_{\delta};\theta)$ \eqref{eq:effective-feature-def} by
\be\label{eq:feature-function-stochastic-reprint}
 \feaE_{i,\mu}(x_{\delta}; \theta) \equiv \frac{\td \z{i}{\delta}{L}}{d \theta_\mu}  \, .
\ee
Note that for the linear model description of infinite-width networks, the derivative of the model output is a constant, and these features are completely \emph{fixed} throughout training.
In contrast, for quadratic models and finite-width networks, the derivative \eqref{eq:feature-function-stochastic-reprint} is not constant, and so these effective features evolve throughout training as the model parameters move.
As for the function approximator itself, after a small change in the parameters $\theta \to \theta+\dbar\theta$, the network output evolves as 
\begin{align}\label{eq:Taylor-expanding-the-finite-width-network-function}
\z{i}{\delta}{L}(\theta+\dbar\theta)&=\z{i}{\delta}{L}(\theta)+\sum_{\mu=1}^{P} \frac{\td \z{i}{\delta}{L}}{d \theta_\mu} \dbar \theta_{\mu}+\frac{1}{2}\sum_{\mu,\nu=1}^{P} \frac{\td^2 \z{i}{\delta}{L}}{d \theta_\mu d \theta_\nu}\dbar \theta_{\mu}\dbar \theta_{\nu}+\ldots\, ,  \notag \\
&=\z{i}{\delta}{L}(\theta)+\sum_{\mu=1}^{P} \feaE_{i,\mu}(x_{\delta}; \theta)\,\dbar \theta_{\mu}+\frac{\epsilon}{2}\sum_{\mu,\nu=1}^{P} \widehat{\featwo}_{i,\mu\nu}(x_{\delta})\, \dbar \theta_{\mu}\dbar \theta_{\nu}+\ldots\, ,
\end{align}
where we've additionally defined an analog of the meta feature function $ \featwo_{j_1j_2}(x_{\delta})$ for neural networks by
\be\label{eq:nonlinear-feature-function-stochastic}
 \epsilon\widehat{\featwo}_{i,\mu\nu}(x_{\delta}) \equiv \frac{\td^2 \z{i}{\delta}{L}}{d \theta_\mu d \theta_\nu}  \, .
\ee
For this discussion, we truncated the ``$\ldots$'' in \eqref{eq:Taylor-expanding-the-finite-width-network-function} so that the update to the output is exactly quadratic in the small change in the parameters. With this truncation, the update \eqref{eq:Taylor-expanding-the-finite-width-network-function} for a finite-width neural network is identical to the update equation \eqref{eq:quadratic-model-small-step} that we found for our quadratic model\index{nonlinear model!quadratic model} after taking a small step.\footnote{Considering the definition of our quadratic model, \eqref{eq:nonlinear-model-def},  we have included the small parameter $\epsilon$ as part of our identification. For MLPs, this parameter will be set automatically by the architecture, and is given by the effective theory cutoff\index{cutoff, effective theory!nearly-kernel methods}, the depth-to-width ratio of the network: $\epsilon \equiv L/n$. However, for such finite-width networks there are additional terms of order $\epsilon \equiv L/n$ that need to be incorporated in order to have a consistent description, as we will explain soon.}

Let us further note that for the linear model description of infinite-width networks, the meta feature functions \eqref{eq:nonlinear-feature-function-stochastic} vanish identically -- as any linear function has a zero second derivative -- and thus have no effect on the dynamics.
For finite-width networks with a quadratic truncation, these meta features \eqref{eq:nonlinear-feature-function-stochastic} are parametrically small but no longer zero; they are stochastically sampled at initialization and then fixed over the course of training, hence decorated with a hat.
Therefore, at quadratic order we will call these meta feature functions,  $\widehat{\featwo}_{i,\mu\nu}(x)$, as \textbf{random meta features}\index{meta feature function!random}, just as we called the feature functions as \emph{random features} for infinite-width networks.

Having established a connection in the feature space, let us now establish a similar connection in the sample-space dual description. First, associated with the effective feature functions  \eqref{eq:feature-function-stochastic-reprint} is the analog of the effective kernel\index{nearly-kernel methods!effective kernel!in terms of effective feature functions} $\kermE_{ii;\delta_1 \delta_2 }(\theta)$ \eqref{eq:effective-kernel-def}, defined by
\begin{align}\label{eq:stochastic-kernel-ntk-parameter-dependent}
\kermE_{i_1i_2;\delta_1\delta_2}(\theta)=  \sum_{\mu,\nu}\lambda_{\mu\nu}\, \feaE_{i_1,\mu}(x_{\delta_1};\theta) \feaE_{i_2,\nu}(x_{\delta_2};\theta)=\sum_{\mu,\nu}\lambda_{\mu\nu}\frac{\td \z{i_1}{\delta_1}{L}}{d \theta_\mu}\frac{\td \z{i_2}{\delta_2}{L}}{d \theta_\nu} \equiv \Tia{\NTKM}{i_1i_2}{\delta_1\delta_2}{L}\!(\theta)\, .
\end{align}
Here, we used our more general definition of the kernel \eqref{eq:kernel-with-features-weighted} to include the learning-rate tensor, and 
since the effective features \eqref{eq:feature-function-stochastic-reprint} have a parameter dependence, in the final equality we used most general definition of the NTK, \eqref{eq:NTH-definition}, and gave it a $\theta$ argument, $\Tia{\NTKM}{i_1i_2}{\delta_1\delta_2}{L}\!(\theta)$, to indicate its parameter dependence.
In particular, if we evaluated the effective kernel\index{nearly-kernel methods!effective kernel} at initialization $\theta = \theta(t=0)$ in terms of the random features
\be\label{eq:stochastic-feature-at-init-reprint}
\widehat{\fea}_{i,\mu}(x_{\delta})\equiv \feaE_{i,\mu}\Big(x_{\delta};\theta(t=0)\Big)= \frac{\td \z{i}{\delta}{L}}{d \theta_\mu}\Bigg|_{\theta=\theta(t=0) }\, ,
\ee
we'd just have the usual $L$-th-layer stochastic NTK at initialization~\eqref{eq:midNTH-definition}:
\begin{align}\label{eq:stochastic-kernel-ntk-reprint}
 \widehat{\kerm}_{i_1i_2;\delta_1\delta_2}\equiv \kermE_{i_1i_2;\delta_1\delta_2}\!\Big(\theta(t=0)\Big)&=  \sum_{\mu,\nu}\lambda_{\mu\nu}\, \widehat{\fea}_{i_1,\mu}(x_{\delta_1}) \, \widehat{\fea}_{i_2,\nu}(x_{\delta_2}) \, \\
 &=\sum_{\mu,\nu}\lambda_{\mu\nu}\le( \frac{\td \z{i_1}{\delta_1}{L}}{d \theta_\mu}\frac{\td \z{i_2}{\delta_2}{L}}{d \theta_\nu}\ri)\Bigg|_{\theta=\theta(t=0) }\!\!\equiv \Tia{\NTK}{i_1i_2}{\delta_1\delta_2}{L}\, . \notag
 \end{align}
For infinite-width networks, this NTK doesn't evolve during training and is composed of \emph{random features} at initialization \eqref{eq:feature-function-stochastic}.
In contrast, as we saw in \S\ref{sec:dNTK}, for finite-width networks the effective kernel\index{nearly-kernel methods!effective kernel}\index{nearly-kernel methods!effective kernel!relation to dynamical NTK} \eqref{eq:stochastic-kernel-ntk-parameter-dependent} \emph{does} evolve during training, just as the analogous effective kernel \eqref{eq:effective-kernel-def} did for the quadratic model.

Finally, analogously to the meta kernel for the quadratic model \eqref{eq:meta-kernel-definition}, we can form a meta kernel for finite-width networks from the random features  \eqref{eq:stochastic-feature-at-init-reprint} and the random meta features \eqref{eq:nonlinear-feature-function-stochastic} as
\begin{align}\label{eq:stochastic-meta-kernel-definition}
\widehat{\mkerm}_{i_0 i_1 i_2; \delta_0\delta_1\delta_2} &\equiv \sum_{\substack{\mu_1,\nu_1, \\ \mu_2,\nu_2} }\ \epsilon \lambda_{\mu_1\nu_1}\lambda_{\mu_2\nu_2} \widehat{\featwo}_{i_0,\mu_1\mu_2}(x_{\delta_0})\, \widehat{\fea}_{i_1,\nu_1}(x_{\delta_1} ) \,\widehat{\fea}_{i_2,\nu_2}(x_{\delta_2}) \\
&=\sum_{\substack{\mu_1,\nu_1, \\ \mu_2,\nu_2} }\lambda_{\mu_1\nu_1}\lambda_{\mu_2\nu_2}\le( \frac{d^2\!\z{i_0}{\delta_0}{L}}{d\theta_{\mu_1}d\theta_{\mu_2}}\frac{d\z{i_1}{\delta_1}{L}}{d\theta_{\nu_1}}\frac{d\z{i_2}{\delta_2}{L}}{d\theta_{\nu_2}}\ri)\Bigg|_{\theta=\theta(t=0) } \equiv \Tia{\dNTK}{i_0i_1i_2}{\delta_0 \delta_1\delta_2}{L} \, , \notag
\end{align}
where we slightly generalized our earlier definition of the meta kernel \eqref{eq:meta-kernel-definition} with the inclusion of the learning-rate tensors.\footnote{
Our slightly more general definition of the meta kernel here should be understood as analogous to the slightly more general definition of the kernel \eqref{eq:kernel-with-features-weighted}.}
Thus, we've now identified the random meta kernel \eqref{eq:stochastic-meta-kernel-definition}  with the $L$-th-layer stochastic dNTK \eqref{eq:dNTK-definition}.

With all these connections established, there are three notable differences between our minimal quadratic model and finite-width neural networks.

First, as should be clear from the definitions of the random features and random meta features, \eqref{eq:stochastic-feature-at-init-reprint} and \eqref{eq:nonlinear-feature-function-stochastic}, these functions are stochastic rather than designed: they are determined by the details of the neural network architecture and depend on the values of the randomly-sampled parameters at initialization. We might more generally call such a quadratic model of the form \eqref{eq:nonlinear-model-def} with random functions $\widehat{\fea}_j(x)$ and $\widehat{\featwo}_{j_1j_2}(x)$ a \term{random meta feature model}, generalizing the notion of a \neo{random feature model} that we discussed in conjunction with infinite-width networks and linear models in \S\ref{subsec:linear-at-infinity}.

Second, as we discussed at the end of \S\ref{subsec:nearly-kernel-methods}, the quadratic model \eqref{eq:nonlinear-model-def} does not wire together different components of the true outputs from the training set when making nearly-kernel predictions \eqref{eq:nearly-linear-regression-optimal-cleanest} on test-set inputs. In contrast, we will show soon in \S\ref{subsec:prediction-at-finite-width} that the finite-width network predictions do have this wiring property.
This deficiency of the quadratic model was actually by design on our part in an effort to eliminate extra complications when working through our minimal model\index{representation learning!minimal model} of representation learning\index{representation learning!minimal model}. To include wiring in the quadratic model\index{nonlinear model!quadratic model!with wiring}, we can generalize it slightly as 
\be\label{eq:nonlinear-model-def-less-minimal}
z_i(x_{\delta}; \theta) = \sum_{\mu=1}^{P}\theta_{\mu}\widehat{\fea}_{i,\mu}(x_{\delta}) + \frac{\epsilon}{2}\sum_{\mu,\nu=1}^{P}\theta_{\mu}\theta_{\nu}\widehat{\featwo}_{i,\mu\nu}(x_{\delta})\, .
\ee
This slightly-less-minimal model will now allow a parameter $\theta_{\mu}$ to connect to various  different output components, as the feature functions and meta feature functions now also carry vectorial indices specifying an output-component.\footnote{Note that these feature functions may have constraints, cf.~the explicit form of the random feature \eqref{eq:NTK-weight-features}. These constraints end up causing the infinite-width model not to wire, while allowing to wire the predictions of any particular network at finite width. These constraints can be thought of as a type of \emph{weight-tying}.
}

Third, as we've mentioned throughout this chapter, the leading finite-width contributions to the update to the network output include $\o{\eta^3}$ terms. To capture these effects, we need to deform\index{deformation!quadratic model} our quadratic model\index{nonlinear model!quadratic model} \eqref{eq:nonlinear-model-def} into a  \textbf{cubic model}\index{nonlinear model!cubic model|textbf}\index{cubic model|see{nonlinear model}}:
\be\label{eq:cubic-model-def}
z_i(x_{\delta}; \theta) = \sum_{\mu=1}^{P}\theta_{\mu}\widehat{\fea}_{i,\mu}(x_{\delta}) + \frac{1}{2}\sum_{\mu,\nu=1}^{P}\theta_{\mu}\theta_{\nu}\widehat{\featwo}_{i,\mu\nu}(x_{\delta}) + \frac{1}{6}\sum_{\mu,\nu,\rho=1}^{P}\theta_{\mu}\theta_{\nu}\theta_{\rho}\widehat{\feathree}_{i,\mu\nu\rho}(x_{\delta}) \, .
\ee
Here, the random \textbf{meta-meta feature function}\index{meta-meta feature function|textbf}\index{feature function!meta-meta|see{meta-meta feature function}}, are given by the third derivative of the network output,
\be\label{eq:third derative random}
\widehat{\feathree}_{i,\mu\nu\rho}(x_{\delta}) \equiv \frac{\td^3 \z{i}{\delta}{L}}{d \theta_\mu d \theta_\nu d \theta_\rho}  \, ;
\ee 
the addition of this cubic term will enable the meta features to effectively evolve as if they're described by a linear model, while in turn the features will effectively evolve as if they're described by a quadratic model.\footnote{To make this connection precise, we must give the small parameter $\epsilon$ not in the cubic model definition \eqref{eq:cubic-model-def}, but instead in the statistics of the joint distribution, $p(\widehat{\fea}_{i,\mu}, \widehat{\featwo}_{i,\mu\nu},\widehat{\feathree}_{i,\mu\nu\rho})$, that controls the random meta-meta feature model. Schematically, the nontrivial combinations are the following:
\begin{align}
\E{\widehat{\fea}^2 } = \o{1} \, , \qquad 
\E{\widehat{\featwo} \,\widehat{\fea}^2 z} = \o{\epsilon}\, , \qquad
\E{\widehat{\feathree} \,\widehat{\fea}^3} = \o{\epsilon}\,, \qquad
\E{\widehat{\featwo}^2\, \widehat{\fea}^2} = \o{\epsilon}
\,.
\end{align}
In the next chapter, we'll identify these combinations with the NTK, the dNTK, and (soon-to-be-revealed) two ddNTKs, respectively. Importantly, since all of these combinations are the same order in $\epsilon= L/n$, to describe finite-width networks self-consistently, we need to think of them as cubic models.
}
In summary, for finite-width networks of depth $L>1$, this less-minimal model, 
\eqref{eq:cubic-model-def}, is a consistent description, with random features \eqref{eq:stochastic-feature-at-init-reprint},  random meta features \eqref{eq:nonlinear-feature-function-stochastic}, and random meta-meta features \eqref{eq:third derative random}.

\subsubsection{Deep Learning: A Non-Minimal Model of Representation Learning}

Representation learning is a big part of what makes deep learning exciting. 
What our minimal model of representation learning\index{representation learning!minimal model}\index{representation learning!minimal model} has shown us is that
we can actually decouple the analysis of the \emph{learning} from the analysis of the \emph{deep}:
the simple quadratic model
\eqref{eq:nonlinear-model-def} 
exhibits nontrivial representation learning for general choices of feature functions\index{feature function}
$\fea_j(x)$
and meta feature functions\index{meta feature function}
$\featwo_{jk}(x)$,
or dually, of a kernel 
$\kerm_{\delta_1\delta_2}$
and a meta kernel 
$\mkerm_{\delta_0\delta_1\delta_2}$. 
In particular, the
meta kernel is what made learning features from the training data possible, and we hope that this broader class of representation-learning models will be of both theoretical and practical interest in their own right.

Of course, \emph{deep} learning is a non-minimal model of representation learning, and the structure of these kernels and meta kernels \emph{do} matter.
Specifically, for deep neural networks the statistics of these functions encoded in the joint preactivation-NTK-dNTK distribution 
$p\Big(z^{(L)}, \, \NTK^{(L)},\, \dNTK^{(L)}\Big\vert \D\Big)$
are controlled by the \terminate{representation group flow} recursions -- cf.~\S\ref{ch:ngp},~\S\ref{ch:NTKa}, and~\S\ref{sec:dNTK-RG}  -- the details of which are implicitly determined by the underlying architecture and hyperparameters.
In particular, we can understand the importance 
of this RG flow
by remembering there can be a vast improvement from selecting other architectures 
beyond
MLPs when applying function approximation to specific domains or datasets: RG flow \emph{is} the inductive bias of the deep learning architecture.\index{inductive bias!of model architectures}\footnote{Note that the formalism of nonlinear models and nearly-kernel methods that we outlined in this section should also describe these other deep learning architectures so long as they admit an expansion around an infinite-width (or infinite-channel or infinite-head) limit. In particular, everything we learned here about representation learning and the training dynamics can be carried over; the only difference is that we will have have different functions $\fea_{i,\mu}(x)$ and $\featwo_{i,\mu\nu}(x)$, leading to different kernels and meta kernels, $\kerm_{i_1i_2;\delta_1\delta_2}$ and $\mkerm_{i_0i_1i_2;\delta_0\delta_1\delta_2}$, that can be built up from a different set of recursions than the ones that we studied in this book.} Thus, even in the set of models that exhibit nontrivial representation learning, these choices -- the initial features, meta features, and so on -- are still really important.\footnote{
In Appendix~\ref{app:residual}, we'll explore an aspect of this question directly by studying \emph{residual networks}\index{residual network}: these networks let us introduce a parameter that in a single network has an interpretation of trading off more layers of representation group flow against more effective realizations from the ensemble.}

The full power of deep learning is likely due to the deep -- i.e.~the \neo{representation group flow} induced by interactions between neurons in deep models of many iterated layers -- working in conjunction with the  learning -- i.e.~the \neo{representation learning} induced by the nonlinear dynamical interactions present at finite width. The \terminate{principles of deep learning theory} presented in this book are precisely those that will let you analyze both of these irreducible basic elements in full generality.

%% file: ChpInfinity-End/infinity_global.tex
\strangechapter{$\infty$}{The End of Training}
\label{ch:eot}\index{end of training}

\epigraph{The job of a scientist is to listen carefully to nature, not to tell nature how to behave.}{Freeman Dyson, explaining Richard Feynman's approach %
\cite{feynman2006classic}.}

\noindent{}In this chapter, we'll finally finish our leading-order effective-theory analysis 
of finite-width networks and solve their training dynamics under gradient descent.
In contrast to the infinite-width limit, for which the solution is independent of the training algorithm,  the dynamics of such deep networks have a rich phenomenology that captures the different ways in which useful features may develop over the course of training.
The solution to these training dynamics  gives first-principles description of the ensemble of fully-trained finite-width networks, realizing a main goal of the book.

Unfortunately, our job will be disrupted by two facts of nature: \emph{(i)} in order to have a consistent description of training dynamics at order $1/n$, we'll need to incorporate two additional objects that arise in the Taylor expansion
of the update to the network output to third order, the update to the NTK to second order, and the update to the dNTK to first order; and 
\emph{(ii)} due to a lack of smoothness, we won't be able to describe the dynamics of $\relu$ networks nor networks consisting of any of the other nonlinear activation functions from the scale-invariant universality class.\index{universality class!scale-invariant}

As for the first
point,
while the analysis of representation learning in the context of the quadratic model was illuminating,
we've already telegraphed that it was insufficient to capture the particular details of finite-width networks. In particular, 
to leading order in $1/n$, there are two more NTK differentials, which we'll refer to as \neo{ddNTKs}.
Although it's straightforward, working out the stochastic forward equations, recursions, and effective theory for these ddNTKs is somewhat tedious, and no longer has any pedagogical value. As such, we won't provide the details of our derivations -- you've already seen these sets of manipulations three times before in \S\ref{ch:ngp}--\S\ref{ch:eft-mlp}, \S\ref{ch:NTKa}--\S\ref{ch:eft-ntk}, and \S\ref{sec:dNTK-RG}--\S\ref{sec:dNTK-criticality}, for the preactivations, NTK, and dNTK, respectively -- instead we'll simply state the results, leaving the details for you as a kind of post-training test evaluation; after all, this is the end of your training as well.

As for the second
point,
throughout the book we've had to use special methods in order to work out exceptional explanations for any non-smooth activation function such as the $\relu$. In our minds, this extra work was justified by the $\relu$'s current privileged status as one of the most popular activation functions in practice. 
However, we have finally run out of tricks and will have to give up: for a reason that is simple to explain, our Taylor expansion in the global learning rate $\eta$ will break down when applied to the dynamics of networks built with non-smooth activation functions.
Instead, we'll have to follow the direction of the community and begin thinking again about smoothed versions of the $\relu$ -- though only the ones that permit a type of \terminate{criticality} -- such as the $\gelu$ and the $\swish$.

With both those disruptions to our work heard, in \S\ref{sec:ddNTKs} we'll present all the relevant results for the ddNTKs -- we'll define them, we'll give their tensor decomposition, and we'll explain their scaling with width and depth -- while hiding all the irrelevant details at the back of the chapter in \S\ref{sec:gross-ddNTK-things}. 
If you've been paying attention, you'll not be shocked to hear that -- when properly normalized -- the \terminate{ddNTKs} scale as the effective theory cutoff:  $\ell/n$. 
This scaling indicates that we need to consider the joint statistics of the preactivation-NTK-dNTK-ddNTKs in order to understand the leading-order finite-width dynamics of deep MLPs. 
Importantly, these ddNTKs endow the dNTK with its own dynamics; from the parameter-space perspective\index{parameter space} of \S\ref{subsec:nonlinear-models}, this means that the \emph{meta feature functions}\index{meta feature function!dynamical} of the model will now evolve.

With those results stated, in \S\ref{sec:another-leap} we'll return to our regularly scheduled pedagogy and, at long last,
solve the training dynamics at finite width. 
After an initial false start following our infinite-width giant leap, first in  \S\ref{subsec:giant-plus-small} we'll learn how to take a small step following an adjusted  giant leap,
giving us our first finite-width solution.
Then in \S\ref{subsec:real-GD-at-finite-width}, we'll analyze many many steps of vanilla gradient descent, giving us our second finite-width solution. The nonlinear dynamics at finite width ultimately lead to a dependence of the fully-trained solution on the training algorithm, and so the solutions derived in these two subsections actually exhibit meaningful differences. 

In particular, the function approximation of a fully-trained finite-width network can be decomposed into a universal part, independent of the optimization details, and a set of \emph{algorithm projectors}\index{algorithm projector}, whose functional form encodes the entire dependence of the solution on the training algorithm. 
These projectors provide a dual sample-space perspective on the learning algorithm, analogous to the relationship between the model parameters and the different kernels.

Accordingly, in  \S\ref{subsec:prediction-at-finite-width} we'll discuss how these projectors impact the solution, letting us understand the inductive bias of the \emph{training dynamics}\index{training dynamics!inductive bias}\index{inductive bias!of learning algorithms}\index{inductive bias!of learning algorithms|seealso{algorithm projector}} separately from the inductive bias of the \emph{network architecture}. We'll also further analyze the predictions made by such fully-trained networks, considering the growing tradeoff between increased representation learning and increased instantiation-to-instantiation fluctuations with network depth.

While this is the final chapter of the main text, in a small epilogue following this chapter, Epilogue~\ref{epi:overparameterization}, we'll explore 
how to define model complexity for overparameterized networks from our effective theory's macroscopic perspective. Then in two appendices, we'll further touch on some topics that are outside the scope of our main line of inquiry. In Appendix~\ref{app:mi-stuff}, we'll introduce the framework of information theory, which will give us the tools we need in order 
to estimate the \terminate{optimal aspect ratio} that separates \emph{effectively-deep}\index{effectively deep} networks from \emph{overly-deep}\index{overly deep} networks. In Appendix~\ref{app:residual}, we'll apply our effective theory approach to learn about residual networks and see how they can be used to extend the range of effectively-deep networks to greater and greater depths.

\section{Two More Differentials}\label{sec:ddNTKs}
\epigraph{Who ordered that?}{I. I. Rabi, quipping about the $\o{1/n}$ \terminate{ddNTKs}.
\index{Rabi, Isidor Isaac}
}

\noindent{}One last time, let's expand the $\ell$-th-layer preactivations after a parameter update, this time recording terms up to \emph{third order}:
\begin{align}\label{eq:preactivation-change-third-order-in-model-parameter}
\dz{i}{\delta}{\ell}\equiv&\z{i}{\delta}{\ell}(t=1)-\z{i}{\delta}{\ell}(t=0)\, \\
=&\sum_{\ell_1=1}^{\ell} \sum_{\mu}\frac{d\z{i}{\delta}{\ell}}{d\theta_{\mu}^{(\ell_1)}}\dtheta_{\mu}^{(\ell_1)}+\frac{1}{2}\sum_{\ell_1, \ell_2 = 1}^\ell \sum_{\mu_1,\mu_2}\frac{d^2\!\z{i}{\delta}{\ell}}{d\theta^{(\ell_1)}_{\mu_1}d\theta_{\mu_2}^{(\ell_2)}}\dtheta_{\mu_1}^{(\ell_1)}\dtheta_{\mu_2}^{(\ell_2)} \notag\\
&+\frac{1}{6}\sum_{\ell_1, \ell_2, \ell_3 = 1}^\ell \sum_{\mu_1,\mu_2,\mu_3}\frac{d^3\!\z{i}{\delta}{\ell}}{d\theta^{(\ell_1)}_{\mu_1}d\theta_{\mu_2}^{(\ell_2)}d\theta_{\mu_3}^{(\ell_3)}}\dtheta_{\mu_1}^{(\ell_1)}\dtheta_{\mu_2}^{(\ell_2)}\dtheta_{\mu_3}^{(\ell_3)}+\ldots\, .\notag
\end{align}
For gradient descent, also recall that after use of the chain rule \eqref{eq:chain-rule-rules}, the change in the $\ell_a$-th-layer parameters of any particular network is given by 
\eqref{eq:parameter-update-rewritten-error-factor},
\be\label{eq:parameter-update-rewritten-error-factor-reprint}
\dtheta_{\mu}^{(\ell_a)}=-\eta\sum_{\nu}\lambda_{\mu\nu}^{(\ell_a)}\le(\sum_{j,k,\tra}\frac{\partial\L_{\A}}{\partial \z{k}{\tra}{L}}\frac{\td\z{k}{\tra}{L}}{\td \z{j}{\tra}{\ell}}\frac{d\z{j}{\tra}{\ell}}{d\theta_{\nu}^{(\ell_a)}}\ri)=-\eta\sum_{\nu,j,\tra}\lambda_{\mu\nu}^{(\ell_a)}\,\Tia{\epsilon}{j}{\tra}{\ell}\frac{d\z{j}{\tra}{\ell}}{d\theta_{\nu}^{(\ell_a)}}\, ,
\ee
where
we've used our convention from \S\ref{sec:dNTK}
of explicitly specifying which layer each parameter comes from.
Please also recall from there
that the learning-rate tensor\index{learning rate!learning-rate tensor} $\lambda_{\mu\nu}^{(\ell)}$ only connects the parameters within a given layer $\ell$. In the above expression, $\ell$ is an intermediate layer such that $\ell_a \leq \ell$, and we also used our
\emph{$\ell$-th-layer error factor}\index{error factor!$\ell$-th-layer} \eqref{eq:ellth-error-chain}:
\be\label{eq:ellth-error-chain-reprint}
\Tia{\epsilon}{j}{\tra}{\ell}\equiv \sum_{k=1}^{n_{L}}\frac{\partial\L_{\A}}{\partial \z{k}{\tra}{L}}\frac{d\z{k}{\tra}{L}}{d\z{j}{\tra}{\ell}}= \frac{\td\L_{\A}}{\td \z{j}{\tra}{\ell}}\, .
\ee
After substituting the parameter update \eqref{eq:parameter-update-rewritten-error-factor-reprint} back into the preactivation update \eqref{eq:preactivation-change-third-order-in-model-parameter}, you should be able to write it in the form
\begin{align}\label{eq:preactivation-updated-finite-width-refined}
\dz{i}{\delta}{\ell}=&-\eta\sum_{j,\tra}\Tia{\NTK}{ij}{\delta\tra}{\ell}\Tia{\epsilon}{j}{\tra}{\ell}+\frac{\eta^2}{2}\sum_{j_1,j_2,\tra_1,\tra_2}\Tia{\dNTK}{i j_1j_2}{\delta\tra_1\tra_2}{\ell}  \Tia{\epsilon}{j_1}{\tra_1}{\ell}\Tia{\epsilon}{j_2}{\tra_2}{\ell}\, \\
&-\frac{\eta^3}{6}\sum_{j_1,j_2,j_3,\tra_1,\tra_2,\tra_3}\Tia{\ddNTK}{i j_1j_2j_3}{\delta\tra_1\tra_2\tra_3}{\ell}  \Tia{\epsilon}{j_1}{\tra_1}{\ell}\Tia{\epsilon}{j_2}{\tra_2}{\ell}\Tia{\epsilon}{j_3}{\tra_3}{\ell}\, , \notag\\
&+\o{\eta^4}\, \notag
\end{align}
where the first two terms we found in the last chapter \eqref{eq:preactivation-updated-finite-width}, and the cubic term is new, with the \emph{first} of the \term{ddNTKs} defined as
\begin{align}
\Tia{\ddNTK}{i_0i_1i_2i_3}{\delta_0 \delta_1\delta_2\delta_3}{\ell}\equiv&\sum_{\ell_1, \ell_2,\ell_3=1}^\ell\,\, \sum_{\substack{\mu_1,\nu_1, \\ \mu_2,\nu_2, \\ \mu_3,\nu_3} }\lambda_{\mu_1\nu_1}^{(\ell_1)}\lambda_{\mu_2\nu_2}^{(\ell_2)}\lambda_{\mu_3\nu_3}^{(\ell_3)}\frac{d^3\!\z{i_0}{\delta_0}{\ell}}{d\theta^{(\ell_1)}_{\mu_1}d\theta^{(\ell_2)}_{\mu_2}d\theta^{(\ell_3)}_{\mu_3}}\frac{d\z{i_1}{\delta_1}{\ell}}{d\theta^{(\ell_1)}_{\nu_1}}\frac{d\z{i_2}{\delta_2}{\ell}}{d\theta^{(\ell_2)}_{\nu_2}}\frac{d\z{i_3}{\delta_3}{\ell}}{d\theta^{(\ell_3)}_{\nu_3}} \, .\label{eq:ddNTK-definition}
\end{align}
As always, the hat on the ddNTK indicates that it's stochastic, depending on the particular realization of the model parameters at initialization. Also, similar to the dNTK, this ddNTK is totally symmetric in its second, third, and fourth paired set of indices $(i_1, \delta_1) \leftrightarrow (i_2, \delta_2) \leftrightarrow (i_3, \delta_3)$, while the first neural-sample index $(i_0, \delta_0)$ is distinguished from the other three.

By expanding to order $\eta^3$, the update, \eqref{eq:preactivation-updated-finite-width-refined}, is now \emph{cubic} in error factors. Expanding to this order is necessary because the first ddNTK has statistics at initialization that are $\o{1/n}$, and so needs to be included in our analysis. However, any higher-order terms in the update are subleading, so we may replace $\o{\eta^4} = \o{1/n^2}$ in this expression.

Just as we had to expand the update to the NTK to order $\eta$ when we expanded the update to the preactivations to order $\eta^2$, we will now have to expand the update to the NTK to order $\eta^2$ for the dynamics with our cubic update \eqref{eq:preactivation-updated-finite-width-refined} to be consistent:
\begin{align}
\dbar\Tia{\NTKM}{i_1i_2}{\delta_1\delta_2}{\ell}\equiv&\Tia{\NTKM}{i_1i_2}{\delta_1\delta_2}{\ell}(t=1)-\Tia{\NTKM}{i_1i_2}{\delta_1\delta_2}{\ell}(t=0)\, \label{eq:NTK-updated-finite-width-refined}\\
=&\sum_{\ell_1=1}^\ell \sum_{\mu_1}\frac{d\Tia{\NTKM}{i_1i_2}{\delta_1\delta_2}{\ell}}{d\theta_{\mu_1}^{(\ell_1)}}\dtheta_{\mu_1}^{(\ell_1)}+\sum_{\ell_1,\ell_2=1}^\ell \sum_{\mu_1,\mu_2}\frac{d^2\!\Tia{\NTKM}{i_1i_2}{\delta_1\delta_2}{\ell}}{d\theta_{\mu_1}^{(\ell_1)}d\theta_{\mu_2}^{(\ell_2)}}\dtheta_{\mu_1}^{(\ell_1)}\dtheta_{\mu_2}^{(\ell_2)}+\ldots\, \nonumber\\
=&-\eta\sum_{j,\tra}\le(\Tia{\dNTK}{i_1 i_2j}{\delta_1\delta_2\tra}{\ell}+\Tia{\dNTK}{i_2i_1j}{\delta_2\delta_1\tra}{\ell}\ri)\Tia{\epsilon}{j}{\tra}{\ell}\, \notag\\
&+\frac{\eta^2}{2}\sum_{j_1,j_2,\tra_1,\tra_2}\le[\Tia{\ddNTK}{i_1i_2j_1j_2}{\delta_1\delta_2\tra_1\tra_2}{\ell}+\Tia{\ddNTK}{i_2i_1j_1j_2}{\delta_2\delta_1\tra_1\tra_2}{\ell}\ri]\Tia{\epsilon}{j_1}{\tra_1}{\ell}\Tia{\epsilon}{j_2}{\tra_2}{\ell}\, \notag\\
&+ \eta^2 \sum_{j_1,j_2,\tra_1,\tra_2}\Tia{\ddNTKII}{i_1i_2j_1j_2}{\delta_1\delta_2\tra_1\tra_2}{\ell}
\Tia{\epsilon}{j_1}{\tra_1}{\ell}\Tia{\epsilon}{j_2}{\tra_2}{\ell}
+\o{\eta^3}\, .\notag
\end{align}
To go to the final equality, we substituted in our NTK definition \eqref{eq:ell-layer-ntk-def-with-layer-indices} and our parameter update \eqref{eq:parameter-update-rewritten-error-factor-reprint}, computed the derivatives, and then collected the terms. To do so, we identified the \emph{second} of the \term{ddNTKs}, defined as
\begin{align}
\Tia{\ddNTKII}{i_1i_2i_3i_4}{\delta_1 \delta_2\delta_3\delta_4}{\ell}\equiv&\sum_{\ell_1, \ell_2,\ell_3=1}^\ell\,\, \sum_{\substack{\mu_1,\nu_1, \\ \mu_2,\nu_2, \\ \mu_3,\nu_3} }\lambda_{\mu_1\nu_1}^{(\ell_1)}\lambda_{\mu_2\nu_2}^{(\ell_2)}\lambda_{\mu_3\nu_3}^{(\ell_3)}\frac{d^2\!\z{i_1}{\delta_1}{\ell}}{d\theta^{(\ell_1)}_{\mu_1}d\theta^{(\ell_3)}_{\mu_3}}\frac{d^2\z{i_2}{\delta_2}{\ell}}{d\theta^{(\ell_2)}_{\mu_2}d\theta^{(\ell_3)}_{\nu_3}}\frac{d\z{i_3}{\delta_3}{\ell}}{d\theta^{(\ell_1)}_{\nu_1}}\frac{d\z{i_4}{\delta_4}{\ell}}{d\theta^{(\ell_2)}_{\nu_2}} \, .\label{eq:ddNTKII-definition}
\end{align}
The hat on this ddNTK indicates that it's also stochastic at initialization, and we will soon detail that it also has $\o{1/n}$ statistics at leading order. Finally, $\tia{\ddNTKII}{i_1i_2i_3i_4}{\delta_1 \delta_2\delta_3\delta_4}$ has a more constrained symmetry, only symmetric under a joint swap of the paired set of indices as $(i_1, \delta_1)\leftrightarrow (i_2, \delta_2)$ \emph{and} $(i_3, \delta_4) \leftrightarrow  (i_4, \delta_4)$. However, this means that we can also swap indices as
\be\label{eq:symmetry-of-ddNTKII}
\sum_{j_1,j_2,\tra_1,\tra_2}\Tia{\ddNTKII}{i_1i_2j_1j_2}{\delta_1\delta_2\tra_1\tra_2}{\ell}\Tia{\epsilon}{j_1}{\tra_1}{\ell}\Tia{\epsilon}{j_2}{\tra_2}{\ell}=\sum_{j_1,j_2,\tra_1,\tra_2}\Tia{\ddNTKII}{i_2i_1j_1j_2}{\delta_2\delta_1\tra_1\tra_2}{\ell}\Tia{\epsilon}{j_1}{\tra_1}{\ell}\Tia{\epsilon}{j_2}{\tra_2}{\ell}\, ,
\ee
which we used to simplify the final line of the NTK update  \eqref{eq:NTK-updated-finite-width-refined}.

Overall, this NTK update is now \emph{quadratic} in error factors, making its dynamics coupled and nonlinear.
Again, expanding to this order is necessary because both ddNTKs have statistics at initialization that are $\o{1/n}$, and so both need to be included in our analysis.  However, any higher-order terms in the NTK update are subleading, so in \eqref{eq:NTK-updated-finite-width-refined} we may replace $\o{\eta^3} = \o{1/n^2}$.

Finally, consider the leading-order update to the dNTK\index{differential of the neural tangent kernel}:
\begin{align}%
\label{eq:dNTK-updated-finite-width}
\dbar\Tia{\dNTKM}{i_0i_1i_2}{\delta_0\delta_1\delta_2}{\ell}\equiv&\Tia{\dNTKM}{i_0i_1i_2}{\delta_0\delta_1\delta_2}{\ell}(t=1) - \Tia{\dNTKM}{i_0i_1i_2}{\delta_0\delta_1\delta_2}{\ell}(t=0) \, \\
=&\sum_{\ell_1=1}^\ell \sum_{\mu_1}\frac{d\Tia{\dNTKM}{i_0i_1i_2}{\delta_0\delta_1\delta_2}{\ell}}{d\theta_{\mu_1}^{(\ell_1)}}\dtheta_{\mu_1}^{(\ell_1)}+\ldots\, \notag\\
=&-\eta\sum_{j,\tra}\le(\Tia{\ddNTK}{i_0i_1 i_2j}{\delta_0\delta_1\delta_2\tra}{\ell}+\Tia{\ddNTKII}{i_0i_1 i_2j}{\delta_0\delta_1\delta_2\tra}{\ell}+\Tia{\ddNTKII}{i_0i_2 i_1j}{\delta_0\delta_2\delta_1\tra}{\ell}\ri)\Tia{\epsilon}{j}{\tra}{\ell}\,\notag \\
&+\o{\eta^2}\, \notag .
\end{align}
To go to the final equality, we substituted our dNTK definition \eqref{eq:dNTK-definition} and parameter update \eqref{eq:parameter-update-rewritten-error-factor-reprint}, took the derivative, and then collected all the terms using our ddNTK definitions, \eqref{eq:ddNTK-definition} and \eqref{eq:ddNTKII-definition}. The dNTK update is \emph{linear} in error factors and its dynamics will be the simplest. Finally, the higher-order terms in the dNTK update are subleading, so in \eqref{eq:dNTK-updated-finite-width} we may replace $\o{\eta^2} = \o{1/n^2}$. We also would like to apologize for the $\dbar \dNTKM$ notation, and we promise that we won't have to use it again.

The updates \eqref{eq:preactivation-updated-finite-width-refined}, \eqref{eq:NTK-updated-finite-width-refined}, and \eqref{eq:dNTK-updated-finite-width} comprise the complete set of finite-width updates at $\o{1/n}$. Thus, to proceed further in our analysis, we'll need to work out the leading-order statistics of the ddNTKs.

\subsubsection{ddNTK Statistics}\index{representation group flow!of the ddNTKs}
To find the distribution of fully-trained finite-width networks, we need the joint distribution of the network output, the NTK, the dNTK, and the ddNTKs:\index{ddNTKs!statistics}
\be\label{eq:finite-width-joint-earlier}
p\!\le(z^{(L)}, \, \NTK^{(L)},\, \dNTK^{(L)},\, \ddNTK^{(L)} , \,\ddNTKII^{(L)}   \Big\vert \D\ri) \, .
\ee
Rather than working through the details here, we'll do it in our own private notebooks. You already have all the tools you need: you can follow \S\ref{ch:ngp}, \S\ref{ch:NTKa}, and \S\ref{sec:dNTK-RG} for examples of how to use RG flow\index{representation group flow} to work out the recursions; and you can follow \S\ref{ch:eft-mlp}, \S\ref{ch:eft-ntk}, and \S\ref{sec:dNTK-criticality} for examples of how to work out the details of the effective theory at initialization after tuning to \terminate{criticality}. The full details of this distribution \eqref{eq:finite-width-joint-earlier} can be found at the end of the chapter in \S\ref{sec:gross-ddNTK-things}. Here, we'll highlight the results that you need in order to understand our dynamical computations in the following section. 

After working out the stochastic forward equations for both ddNTKs -- feel free to flip forward to \eqref{eq:ddNTK-forward-equation} and \eqref{eq:ddNTK-II-forward-equation} if you're curious about them -- we'll need to find recursions for its statistics. For these tensors, the leading-order statistics come from their means; their cross correlations and their variances are all subleading. When evaluating the mean of each of the ddNTKs,
it will become convenient to decompose them  into the following set of tensors with sample indices only:\index{tensor decomposition!ddNTKs $R$/$S$/$T$/$U$}
\begin{align}\label{eq:decomposition-ddNTK}
\E{\Tia\ddNTK{i_0i_1i_2i_3}{\delta_0\delta_1\delta_2\delta_3}{\ell}} \equiv&\frac{1}{n_{\ell-1}}\le[\delta_{i_0i_1}\delta_{i_2i_3}\ddNTKR{\delta_0\delta_1\delta_2\delta_3}{\ell}+\delta_{i_0i_2}\delta_{i_3i_1}\ddNTKR{\delta_0\delta_2\delta_3\delta_1}{\ell}+\delta_{i_0i_3}\delta_{i_1i_2}\ddNTKR{\delta_0\delta_3\delta_1\delta_2}{\ell}\ri]\, , \\
\E{\Tia\ddNTKII{i_1i_2i_3i_4}{\delta_1\delta_2\delta_3\delta_4}{\ell}}\equiv&\frac{1}{n_{\ell-1}}\le[\delta_{i_1i_2}\delta_{i_3i_4}\ddNTKS{\delta_1\delta_2\delta_3\delta_4}{\ell}+\delta_{i_1i_3}\delta_{i_4i_2}\ddNTKT{\delta_1\delta_3\delta_4\delta_2}{\ell}+\delta_{i_1i_4}\delta_{i_2i_3}\ddNTKU{\delta_1\delta_4\delta_2\delta_3}{\ell}\ri]\, .
\label{eq:decomposition-ddNTK-II}
\end{align}
Thus, there are four new objects whose recursions we need to determine and whose  scaling with depth we'll need to compute.

The ddNTKs both vanish in the first layer, just like the dNTK. Proceeding then from the second layer to deeper layers,
after a bunch of tedious algebra and plenty of flipping around in the book to make various substitutions, you can find recursions for $\ddNTKRS^{(\ell)}$, $\ddNTKSS^{(\ell)}$, $\ddNTKTS^{(\ell)}$, and $\ddNTKUS^{(\ell)}$: \eqref{eq:R-recursion}, \eqref{eq:S-recursion}, \eqref{eq:T-recursion}, and \eqref{eq:U-recursion}, respectively. You can flip forward to take a look at them, but you probably won't want to\dots. Importantly, these recursions in conjunction with the decompositions \eqref{eq:decomposition-ddNTK} and \eqref{eq:decomposition-ddNTK-II} altogether demonstrate the both ddNTKs have nontrivial order-$1/n$ statistics:
\be
\E{\Tia\ddNTK{i_0i_1i_2i_3}{\delta_0\delta_1\delta_2\delta_3}{\ell}} = \oninv\, ,\qquad \E{\Tia\ddNTKII{i_1i_2i_3i_4}{\delta_1\delta_2\delta_3\delta_4}{\ell}} = \oninv \, .
\ee
Thus, as we've been forecasting, we must include them in our finite-width analysis of training dynamics.

Focusing on the single-input statistics, we again make the layer-independent choices $\Cb{\ell}=C_b$, $\CW{\ell}=C_W$, and $n_1=\cdots=n_{L-1}\equiv n$. Ignoring contributions that are subleading in $1/n$, in particular replacing the mean metric by the kernel $G^{(\ell)} \to \ker^{(\ell)}$ and the NTK mean by the frozen NTK $\NTKM^{(\ell)} \to \NTKI^{(\ell)}$,  the four recursions  \eqref{eq:R-recursion}, \eqref{eq:S-recursion}, \eqref{eq:T-recursion}, and \eqref{eq:U-recursion} together reduce to a form that at least fits on a page:
\begin{align}\label{eq:R-recursions-single-input}\index{ddNTKs!statistics!R-recursion@$R$-recursion}
\ddNTKR{}{\ell+1}=&\le(\chi_{\perp}^{(\ell)}\ri)^2\ddNTKR{}{\ell}\, \\
&+\LW{\ell+1}C_W\bra\sigma^{\prime\prime}\sigma^{\prime}\sigma^{\prime}\sigma\ket_{\ker^{(\ell)}}\le(\Ti{\NTKI}{}{\ell}\ri)^2+C_W^2\bra\sigma^{\prime\prime\prime}\sigma^{\prime}\sigma^{\prime}\sigma^{\prime}\ket_{\ker^{(\ell)}}\le(\Ti{\NTKI}{}{\ell}\ri)^3\, \notag\\
&+\chi_{\perp}^{(\ell)}\le(\LW{\ell+1}\bra\sigma^{\prime\prime}\sigma\ket_{\ker^{(\ell)}}+C_W\Ti{\NTKI}{}{\ell}\bra\sigma^{\prime\prime\prime}\sigma^{\prime}\ket_{\ker^{(\ell)}}\ri)\le(\NTHB{}{\ell}+\dNTKP{}{\ell}\ri)\, \notag\\
&+\chi_{\perp}^{(\ell)}\le(\LW{\ell+1}\bra\sigma^{\prime}\sigma^{\prime}\ket_{\ker^{(\ell)}}+C_W\Ti{\NTKI}{}{\ell}\bra\sigma^{\prime\prime}\sigma^{\prime\prime}\ket_{\ker^{(\ell)}}\ri)\dNTKP{}{\ell}\, ,\notag\\
\label{eq:S-recursions-single-input}\index{ddNTKs!statistics!S-recursion@$S$-recursion}
\ddNTKS{}{\ell+1}=&\le(\chi_{\perp}^{(\ell)}\ri)^2\ddNTKS{}{\ell}\, \\
&+C_W\LW{\ell+1}\bra \sigma^{\prime} \sigma^{\prime} \sigma^{\prime}\sigma^{\prime}\ket_{\ker^{(\ell)}}\le(\Ti{\NTKI}{}{\ell}\ri)^2+C_W^2\bra \sigma^{\prime\prime}\sigma^{\prime\prime}\sigma^{\prime}\sigma^{\prime}\ket_{\ker^{(\ell)}}\le(\Ti{\NTKI}{}{\ell}\ri)^3\, \notag\\
&+\chi_{\perp}^{(\ell)}\le[\LW{\ell+1}\bra \sigma^{\prime}\sigma^{\prime}\ket_{\ker^{(\ell)}}+C_W\Ti{\NTKI}{}{\ell}\bra \sigma^{\prime\prime}\sigma^{\prime\prime}\ket_{\ker^{(\ell)}}\ri]\NTHB{}{\ell}\, ,\notag\\
\label{eq:T-recursions-single-input}\index{ddNTKs!statistics!T-recursion@$T$-recursion}
\ddNTKT{}{\ell+1}=&\le(\chi_{\perp}^{(\ell)}\ri)^2\ddNTKT{}{\ell}\, \\
&+2C_W\LW{\ell+1}\bra \sigma^{\prime\prime}\sigma^{\prime}\sigma^{\prime}\sigma\ket_{\ker^{(\ell)}}\le(\Ti{\NTKI}{}{\ell}\ri)^2+C_W^2\bra \sigma^{\prime\prime} \sigma^{\prime\prime} \sigma^{\prime}\sigma^{\prime}\ket_{\ker^{(\ell)}}\le(\Ti{\NTKI}{}{\ell}\ri)^3\, \notag\\
&+\le(\LW{\ell+1}\ri)^2\bra \sigma^{\prime}\sigma^{\prime} \sigma\sigma\ket_{\ker^{(\ell)}}\Ti{\NTKI}{}{\ell}\, \notag\\
&+\le(\LW{\ell+1}\bra z \sigma^{\prime}\sigma\ket_{\ker^{(\ell)}}+C_W\Ti{\NTKI}{}{\ell}\bra z \sigma^{\prime\prime}\sigma^{\prime}\ket_{\ker^{(\ell)}}\ri)^2\frac{\NTHF{}{\ell}}{\le(\ker^{(\ell)}\ri)^2}\, \notag\\
&+2\chi_{\perp}^{(\ell)}\le[\LW{\ell+1}\le(\bra  \sigma^{\prime\prime}\sigma\ket_{\ker^{(\ell)}}+\bra  \sigma^{\prime}\sigma^{\prime}\ket_{\ker^{(\ell)}}\ri)+C_W\Ti{\NTKI}{}{\ell}\le(\bra  \sigma^{\prime\prime\prime}\sigma^{\prime}\ket_{\ker^{(\ell)}}+\bra  \sigma^{\prime\prime}\sigma^{\prime\prime}\ket_{\ker^{(\ell)}}\ri)\ri]\dNTKQ{}{\ell}\, ,\notag\\
\label{eq:U-recursions-single-input}\index{ddNTKs!statistics!U-recursion@$U$-recursion}
\ddNTKU{}{\ell+1}=&\le(\chi_{\perp}^{(\ell)}\ri)^2\ddNTKU{}{\ell}+C_W^2\bra \sigma^{\prime\prime} \sigma^{\prime\prime}\sigma^{\prime}\sigma^{\prime}\ket_{\ker^{(\ell)}}\le(\Ti{\NTKI}{}{\ell}\ri)^3\, .
\end{align}
Here, we also simplified these expressions by recalling the two susceptibilities, \eqref{eq:chi-parallel} and~\eqref{eq:chi-perp}: 
\begin{align}
\Ti{\chi}{\parallel}{\ell}  \equiv \frac{C_W}{\ker^{(\ell)}} \bra \sigma^\prime \sigma  z \ket_{\ker^{(\ell)}} \, , \qquad \Ti{\chi}{\perp}{\ell} 
\equiv  C_W\bra\sigma^{\prime}\sigma^{\prime}\ket_{\ker^{(\ell)}}\, .
\end{align}

\subsubsection{ddNTK Scalings}\index{ddNTKs!scaling laws}

Now, let's tune to criticality, and use our \terminate{scaling ansatz}  \eqref{eq:master-scaling-ansatz} to find the critical exponents $p_\ddNTKRS$, $p_\ddNTKSS$, $p_\ddNTKTS$, and $p_\ddNTKUS$ that describe the asymptotic depth scaling of $\ddNTKR{}{\ell}$, $\ddNTKS{}{\ell}$, $\ddNTKT{}{\ell}$, and $\ddNTKU{}{\ell}$, respectively. 

To understand the relative size of these tensors controlling the means of the ddNTKs, we will again need to identify  appropriate dimensionless ratios. Following the \terminate{dimensional analysis} logic from our dNTK discussion, cf.~\eqref{eq:dNTK-dimensional-analysis}, let's look at our third-order update for preactivations \eqref{eq:preactivation-updated-finite-width-refined}. Remembering again that we can only add terms that have the same dimensions, we see that
\be\label{eq:ddNTK-dimensional-analysis}
[z]=[\eta]\, [\epsilon] \,[\NTK]=[\eta]^2\,[\epsilon]^2\,[\dNTK] =[\eta]^3\,[\epsilon]^3\,[\ddNTK] =[\eta]^3\,[\epsilon]^3\,[\ddNTKII] \,.
\ee 
From the first equality, we see as before that $[\eta]\,[\epsilon]=[z]\,[\NTK]^{-1}$, and so $\ddNTKRS$, $\ddNTKSS$, $\ddNTKTS$, and $\ddNTKUS$ each have dimensions of NTK cubed:
\be\label{eq:dimensions-of-ddNTK-stuff}
[\ddNTKRS]\equiv [\ddNTK]=[\NTK]^{3} \, [z]^{-2} \, , \qquad [\ddNTKSS] = [\ddNTKTS]=[\ddNTKUS] \equiv [\ddNTKII]= [\NTK]^{3} \, [z]^{-2} \, .
\ee
This means that the proper dimensionless combinations for the ddNTKs are
\begin{align}\label{eq:scaling-relations-ddNTKs}
\frac{\ddNTKR{}{\ell} \Ti{\ker}{}{\ell}}{n \le( \Ti{\NTKI}{}{\ell} \ri)^3 } \sim \frac{1}{n}\le( \frac{1}{\ell} \ri)^{p_\ddNTKRS +p_0 - 3p_\Theta} \,, \qquad \frac{\ddNTKS{}{\ell}\Ti{\ker}{}{\ell}}{n \le( \Ti{\NTKI}{}{\ell} \ri)^3} \sim \frac{1}{n}\le( \frac{1}{\ell} \ri)^{p_\ddNTKSS+p_0  - 3p_\Theta}\, , \\
\frac{\ddNTKT{}{\ell} \Ti{\ker}{}{\ell}}{n \le( \Ti{\NTKI}{}{\ell} \ri)^3 } \sim \frac{1}{n}\le( \frac{1}{\ell} \ri)^{p_\ddNTKTS +p_0- 3p_\Theta} \,, \qquad \frac{\ddNTKU{}{\ell}\Ti{\ker}{}{\ell}}{n \le( \Ti{\NTKI}{}{\ell} \ri)^3} \sim \frac{1}{n}\le( \frac{1}{\ell} \ri)^{p_\ddNTKUS +p_0 - 3p_\Theta}\, ,\label{eq:scaling-relations-ddNTKs-2}
\end{align}
where to get a normalized ratio, we multiplied by the kernel $\Ti{\ker}{}{\ell}$ to account for the $[z]^{-2}$ and then divided by the three factors of the frozen NTK $\Ti{\NTKI}{}{\ell}$. Here, $p_0$ is the \terminate{critical exponent} for the kernel, and $p_\Theta$ is the \terminate{critical exponent} for the NTK.

Finally, as a brief aside, let us comment on one aspect of dimensionality that we've been ignoring but will soon become important. In particular, since the network output $z$ is set to the true output $y$, they should really have the same dimensions:
\be\label{eq:dimension-of-labels}
[z] = [y] \, .
\ee
However, while the leading depth scaling of the preactivations is given by the kernel,
\be
\E{z^{(\ell)} z^{(\ell)}} = \ker^{(\ell)} \sim \le( \frac{1}{\ell}\ri)^{p_0} \, ,
\ee
the true output $y$ is fixed and doesn't scale with depth. When comparing the performance of networks of different depths, this suggests that it might be helpful to rescale the final network outputs as
\be
z_{i;\delta} \to z_{i;\delta} \le(\frac{1}{L}\ri)^{-p_0/2} \, ,
\ee
effectively fixing the scaling of the overall network output as $p_0=0$,
or equivalently rescale the training outputs as
\be\label{eq:rescale-true-outputs}
y_{i;\tra} \to y_{i;\tra} \le(\frac{1}{L}\ri)^{p_0/2} \,.
\ee
We will further see how this can affect the predictions of a fully-trained network on its test set in \S\ref{subsec:prediction-at-finite-width}.\footnote{
    For \terminate{regression} tasks, where we want to learn a vector of real numbers, this rescaling is appropriate. For \terminate{classification} tasks, where we want to learn a discrete probability distribution, we should rescale either the network outputs (before any softmax layer) or the raw output targets $y_{i;\delta}$ (again before any softmax layer) as in \eqref{eq:softmax-target}.
}

\subsubsection{$K^\star=0$ Universality Class}

For the $K^\star=0$ universality class\index{universality class!K@$K^\star=0$}, remember that we Taylor expanded the activation function as %
\be\label{eq:taylor-expansion-k-star-reprint-last-actually-through}
\sigma(z)=\sum_{p=0}^{\infty}\frac{\sigma_{p}}{p!}z^p\,  ,
\ee
and defined the following Taylor coefficient for convenience
\begin{align}
a_1&\equiv \le(\frac{\sigma_3}{\sigma_1}\ri)+\frac{3}{4}\le(\frac{\sigma_2}{\sigma_1}\ri)^2\ ,\label{eq:a1-recall-last-actually}
\end{align}
and required that all activation functions in this class satisfy $\sigma_0 = 0$ and $\sigma_1 \neq 0$.

To solve our single input ddNTK recursions, \eqref{eq:R-recursions-single-input}--\eqref{eq:U-recursions-single-input}, you'll have to evaluate a few new Gaussian expectations, taking particular note of that some of them now depend on the third derivative of the activation function. Finally, to tune to $K^\star=0$ \terminate{criticality}  \eqref{eq:k-star-equals-zero-critical-initialization}, we need to set the initialization hyperparameters as $C_b=0$ and $C_W =1/\sigma_1^2$; to implement the learning rate \terminate{equivalence principle}, we need to set our training hyperparameters as
\eqref{eq:super-tanh-general},
\be
\Lb{\ell}=\widetilde{\lambda}_b\le(\frac{1}{\ell}\ri)^{p_{\perp}}L^{p_{\perp}-1}\, , \qquad \lamW{\ell}=\widetilde{\lambda}_W\le(\frac{L}{\ell}\ri)^{p_{\perp}-1} \, .
\ee
For simplicity, let us also focus on odd activation functions, such as $\tanhA$, for which importantly $\sigma_2=0$ and $p_\perp = 1$.

Inspecting the single-input ddNTK recursions \eqref{eq:R-recursions-single-input}--\eqref{eq:U-recursions-single-input}, we see that they depend on Gaussian expectations of preactivations as well as our previous solutions for all the other objects that we've considered: the NTK variance, the NTK-preactivation cross correlation, and the dNTK-preactivation cross correlation. 
Substituting in the solutions for all these quantities as needed -- you'll have to flip around to find them, though most were reprinted in \S\ref{subsec:dntk_criticality_tanh_univ}\index{universality class!K@$K^\star=0$} for the dNTK analysis -- we can find solutions to all the single-input ddNTK recursions:
\begin{align}
\ddNTKR{}{\ell}=&-\frac{\ell^2}{48}\le[3\widetilde{\lambda}_b+4\frac{\widetilde{\lambda}_W\sigma_1^2}{(-a_1)}\ri]\le[\widetilde{\lambda}_b+\frac{\widetilde{\lambda}_W\sigma_1^2}{(-a_1)}\ri]^2 (-a_1)+\ldots\, ,\\
\ddNTKS{}{\ell}=&\frac{\ell^2}{12} \le[\widetilde{\lambda}_b+\frac{\widetilde{\lambda}_W\sigma_1^2}{(-a_1)}\ri]^2\widetilde{\lambda}_W\sigma_1^2+\ldots\, ,\\
\ddNTKT{}{\ell}=&\frac{\ell^2}{32}\le[\widetilde{\lambda}_b+\frac{\widetilde{\lambda}_W\sigma_1^2}{(-a_1)}\ri](-a_1) \widetilde{\lambda}_b^2+\ldots\, ,\\
\ddNTKU{}{\ell}=&\frac{1}{2}\le[\widetilde{\lambda}_b+\frac{\widetilde{\lambda}_W\sigma_1^2}{(-a_1)}\ri]^3(-a_1)+\ldots\, .
\end{align}
From these, we can read off the critical exponents as $p_\ddNTKRS = p_\ddNTKSS = p_\ddNTKTS = -2$, and $p_\ddNTKUS=0$, and we see that the dimensionless ratios are given by 
\begin{align}\label{eq:scaling-relations-ddNTKs-K-star}
\frac{\ddNTKR{}{\ell} \Ti{\ker}{}{\ell}}{n \le( \Ti{\NTKI}{}{\ell} \ri)^3 } &=-\frac{1}{48}\le[\frac{3\widetilde{\lambda}_b +4 \widetilde{\lambda}_W\sigma_1^2/(-a_1) }{\widetilde{\lambda}_b +\widetilde{\lambda}_W\sigma_1^2/(-a_1) }\ri] \frac{\ell}{n}+ \ldots\, , \\
\frac{\ddNTKS{}{\ell}\Ti{\ker}{}{\ell}}{n \le( \Ti{\NTKI}{}{\ell} \ri)^3} &=\frac{1}{12}\le[\frac{ \widetilde{\lambda}_W\sigma_1^2/(-a_1) }{\widetilde{\lambda}_b +\widetilde{\lambda}_W\sigma_1^2/(-a_1) }\ri] \frac{\ell}{n}+ \ldots\, , \\
\frac{\ddNTKT{}{\ell} \Ti{\ker}{}{\ell}}{n \le( \Ti{\NTKI}{}{\ell} \ri)^3 } &=\frac{1}{32}\le[\frac{ \widetilde{\lambda}_b  }{\widetilde{\lambda}_b +\widetilde{\lambda}_W\sigma_1^2/(-a_1) }\ri]^2 \frac{\ell}{n}+ \ldots \, , \\
\frac{\ddNTKU{}{\ell}\Ti{\ker}{}{\ell}}{n \le( \Ti{\NTKI}{}{\ell} \ri)^3} &=\frac{1}{2} \frac{1}{\ell n } + \ldots \, .
\end{align}
This means that $\ddNTKR{}{\ell}$, $\ddNTKS{}{\ell}$, and $\ddNTKT{}{\ell}$ all scale according to our leading effective theory cutoff as
\be
p_\ddNTKRS + p_0 - 3p_\Theta = -1 \, \qquad p_\ddNTKSS + p_0 - 3p_\Theta = -1 \, , \qquad p_\ddNTKTS + p_0 - 3p_\Theta = -1 \, .
\ee
Thus, we see that the leading-order finite-width dynamics of $\ker^\star=0$ activation functions have contributions from the first ddNTK, $\ddNTK$, via $\ddNTKR{}{\ell}$, and from the second ddNTK, $\ddNTKII$,  via $\ddNTKS{}{\ell}$ and $\ddNTKT{}{\ell}$.\footnote{If we relaxed the restriction for the activation function to be odd, we'd find the same scalings for  $\ddNTKR{}{\ell}$, $\ddNTKS{}{\ell}$, and $\ddNTKT{}{\ell}$ -- though with different coefficients -- and we'd find that $\ddNTKU{}{\ell}$ was up by a factor of $\ell$, but still subleading overall.}

\subsubsection{Scale-Invariant Universality Class}\index{universality class!scale-invariant}
Perhaps the most important fact to remember about nonlinear scale-invariant activation functions \eqref{eq:scale-invariant-def-first},
\be\label{eq:scale-invariant-re-reprinted}
\sigma(z) = 
    \begin{cases}
    a_+ z \, , & z \ge 0\, , \\
     a_- z \, , & z < 0  \, ,
    \end{cases}
\ee 
with $a_+ \neq -a_-$, 
is that they are not smooth: their first derivative is a step function centered at the origin, and their second derivative is a \terminate{Dirac delta function}, \eqref{eq:integral-form-delta-function},
\be
\sigma'(z) =
    \begin{cases}
    a_+  \, , & z \ge 0\, , \\
     a_-  \, , & z < 0  \, ,
    \end{cases}
\, , \qquad  
\sigma''(z) = (a_+ - a_-)\delta(z) \, ,
\ee
and higher derivatives will involve  derivatives of the \terminate{Dirac delta function}.
Inspecting again the single-input ddNTK recursions \eqref{eq:R-recursions-single-input}--\eqref{eq:U-recursions-single-input}, the kink in these activation functions
 and the presence of  Gaussian expectations with up to three derivatives of $\sigma(z)$ should scare you, especially if you heeded our warning at the end of \S\ref{sec:finite_angle}. In fact, if you formally try to evaluate some of these expectations, particularly $\bra \sigma'' \sigma'' \ket_{\ker}$ and others related to it via integration by parts, you'll find that they want to blow up, even if you use all the \emph{magic tricks}\index{magic trick} from \terminate{physics} that you might have at your disposal for trying to make sense of divergent integrals.\footnote{
    We were somewhat lucky in \S\ref{subsec:dntk_criticality_scale_invariant} when analyzing the dNTK: all the higher-derivative Gaussian integrals that we needed simplified via integration by parts and gave finite -- in fact, vanishing -- answers, cf.~\eqref{eq:scale-invariant-higher-derivative-first} and \eqref{eq:scale-invariant-higher-derivative-second}.
} 

The divergence of these Gaussian correlators is actually telling us that there is something very wrong with our expansions for the updates to the preactivations \eqref{eq:preactivation-updated-finite-width-refined}, the NTK \eqref{eq:NTK-updated-finite-width-refined}, and the dNTK \eqref{eq:dNTK-updated-finite-width}. In particular, the Taylor expansion in the global learning rate $\eta$ breaks down for these non-smooth activation functions and doesn't accurately describe how a network is updated. As a result, our approach for solving the finite-width dynamics will not work for nonlinear scale-invariant activation functions.

To understand why, let's consider an extremely simple model of a network, a single neuron with a bias:
\be
z(x) = \sigma(x+b) \, .
\ee
Here, the input $x$ and the output $z$ are both scalars, and for the activation function $\sigma(z)$ we'll pick the $\relu$, with $a_+=1$ and $a_-=0$. 
Accordingly, for a particular input $x$ such that $x+b > 0$, the activation fires, and the output is $z(x)=x+b$; for a particular input $x'$ such that $x' + b < 0$, the activation doesn't fire, and the output is $z(x')=0$. 

Now, let's consider a gradient-descent update to the parameters with a training example $x>-b$ such that the activation fires. Then, the bias updates as
\be
\dbar b = - \eta \frac{dz}{db} \epsilon = -\eta \epsilon = \o{\eta}\, ,
\ee
where $\epsilon$ is the error factor of the loss, depending on the true output $y$. Now, if $x + b + \dbar b > 0$, then the change in the output is
\be
\dbar z = -\eta \epsilon = \o{\eta}\, ,
\ee
and would be perfectly described by our Taylor expansion \eqref{eq:preactivation-updated-finite-width-refined}. However, if the training error is large enough
such that $x + b + \dbar b=x + b -\eta \epsilon < 0$, then the activation turns off and
\be
\dbar z = -x -b = \o{1}\, .
\ee
Importantly, this update is independent of our expansion parameter $\eta$, and a Taylor expansion in $\eta$ cannot detect this discontinuity at $\eta=(x+b)/\epsilon$.

Thus, for the $\relu$, any time an activation crosses its firing threshold it can contribute  to the gradient-descent update in an $\eta$-independent way. 
Empirically, if you try to measure the NTK update for a deep MLP consisting of $\relu$ activation functions, you don't find anything consistent with the expansion \eqref{eq:NTK-updated-finite-width-refined}. Instead, for the appropriately normalized quantity, you'll find an $\sim 1/\sqrt{n}$ scaling with width and a linear scaling with depth $\ell$, in contrast to the $\ell/n$ scaling expected from our perturbative formalism.\footnote{
    It's temping to think that this $\sim 1/\sqrt{n}$ scaling arises from
    accumulating the probabilities that one of the $L n$ total hidden-layer $n \gg 1$ activations
 experiences an $\eta$-independent $\o{1}$ change after the gradient-descent step.
} From this we reach the unfortunate conclusion that we'll have to give up on describing the finite-width dynamics of nonlinear scale-invariant activation functions using these methods.

Note that everything we have discussed for nonlinear scale-invariant activation functions with respect to criticality and the infinite-width training dynamics is perfectly fine and experimentally validated, and the presence of a nonzero dNTK at finite width is still indicative of a dynamical NTK\index{dynamical NTK} and representation learning at finite width. 
The problem we just described only affects the analysis we're going to perform in the following section for the  training dynamics of finite-width networks, 
and the breakdown of the Taylor expansion just means that we will be unable to give a quantitative picture of representation learning for these non-smooth activation functions.\footnote{
    However, our analysis applies to any of the smoothed versions of the $\relu$ that permit a type of criticality, cf.~our criticality discussion of the $\swish$ and $\gelu$ in \S\ref{subsec:half_stability}. In particular, the training dynamics we'll work out in the next section describe these networks.
    These dynamical solutions, in conjunction with the output-layer solutions to all their associated recursions, will accurately characterize such fully-trained $\relu$-like networks in practice.
}
So, if you do want to understand this, you'll probably need an entirely new approach.

This leaves us one activation function in the entire scale-invariant universality class\index{universality class!scale-invariant}: the $\linear$ activation function used for deep linear networks. Tuning to criticality, $C_b =0$ and $C_W = 1$, which fixes $\chi=1$, and choosing layer-independent learning rates \eqref{eq:super-scale-invariant} as 
\be\label{eq:learning-rate-EP-scale-invariant-reprint-for-ddNTKs}
\Lb{\ell} = \frac{\widetilde{\lambda}_b}{L} \, , \qquad \LW{\ell} = \frac{\widetilde{\lambda}_W }{L}  \,,
\ee
we can solve the single-input ddNTK recursions \eqref{eq:R-recursions-single-input}--\eqref{eq:U-recursions-single-input}. Note that  for every term in the $\ddNTKUS$-recursion, \eqref{eq:U-recursions-single-input}, and for every term but one in the $\ddNTKRS$-recursion, \eqref{eq:R-recursions-single-input}, the Gaussian expectations of activations involve second derivatives or third derivatives, which vanish for a $\linear$ function. For the one term in the $\ddNTKRS$-recursion that does not vanish, it is also multiplied by $\dNTKP{}{\ell}$, which vanishes for all scale-invariant activations, cf.~\eqref{eq:P-vanish-scale-invariant}. This means that 
\begin{align}
\ddNTKR{}{\ell}=0\, ,\qquad \ddNTKU{}{\ell}=0\, ,
\end{align}
and in particular that the first ddNTK, $\ddNTK$, does not contribute to the deep linear network's dynamics at $\o{1/n}$ since it's entirely determined by $\ddNTKR{}{\ell}$.
However, for the $\ddNTKSS$-recursion, \eqref{eq:S-recursions-single-input}, and the $\ddNTKTS$-recursion, \eqref{eq:T-recursions-single-input}, we find something nontrivial: these recursions can be exactly solved as
\begin{align}
\ddNTKS{}{\ell}=&\frac{\ell^2(\ell^2-1)}{12L^3} (\widetilde{\lambda}_b+\widetilde{\lambda}_W \ker^{\star})^2\widetilde{\lambda}_W \, ,\\
\ddNTKT{}{\ell}=&\frac{\ell^2(\ell^2-1)}{12L^3} (\widetilde{\lambda}_b+\widetilde{\lambda}_W \ker^{\star}) \widetilde{\lambda}_W^2 \ker^{\star}\, ,
\end{align}
from which we see that the critical exponents are $p_\ddNTKSS = p_\ddNTKTS = -4$. 
Finally, the dimensionless ratios are 
\begin{align}\label{eq:scaling-relations-ddNTK-DLN}
\frac{\ddNTKS{}{\ell} \Ti{\ker}{}{\ell}}{n \le( \Ti{\NTKI}{}{\ell} \ri)^3 } &= \frac{1}{12}\le[\frac{\widetilde{\lambda}_W K^\star}{\widetilde{\lambda}_b + \widetilde{\lambda}_W K^\star } \ri] \frac{\ell}{n} + \dots\,, \\
\frac{\ddNTKT{}{\ell}\Ti{\ker}{}{\ell}}{n \le( \Ti{\NTKI}{}{\ell} \ri)^3} &= \frac{1}{12}\le[ \frac{\widetilde{\lambda}_W K^\star }{ \widetilde{\lambda}_b + \widetilde{\lambda}_W K^\star } \ri]^2 \frac{\ell}{n} + \dots \, ,
\end{align}
and we see that these scale according to our leading effective theory cutoff as
\be
p_\ddNTKSS + p_0 - 3p_\Theta = -1 \, , \qquad p_\ddNTKTS + p_0 - 3p_\Theta = -1 \, .
\ee
In conclusion, we see that the second ddNTK, $\ddNTKII$, contributes via $\ddNTKS{}{\ell}$ and $\ddNTKT{}{\ell}$ to the dynamics of deep linear networks at leading order.

\section{Training at Finite Width}\label{sec:another-leap}\index{training dynamics!finite width}

Now that we understand the joint statistics of the preactivations, the NTK, the dNTK, and the ddNTKs, we have nearly all the tools we need in order to evaluate the distribution of \emph{fully-trained} networks at finite width and nonzero depth. To see why, recall our finite-width expansion of the network output evolution \eqref{eq:preactivation-updated-finite-width-refined}
\begin{align}\label{eq:preactivation-updated-finite-width-decomposition}
\z{i}{\delta}{L}(t=1)=\z{i}{\delta}{L}&-\eta\sum_{j,\tra}\Tia{\NTK}{ij}{\delta\tra}{L}\Tia{\epsilon}{j}{\tra}{L}+\frac{\eta^2}{2}\sum_{j_1,j_2,\tra_1,\tra_2}\Tia{\dNTK}{i j_1j_2}{\delta\tra_1\tra_2}{L}  \Tia{\epsilon}{j_1}{\tra_1}{L}\Tia{\epsilon}{j_2}{\tra_2}{L}\\
&-\frac{\eta^3}{6}\sum_{j_1,j_2,j_3,\tra_1,\tra_2,\tra_3}\Tia{\ddNTK}{i j_1j_2j_3}{\delta\tra_1\tra_2\tra_3}{L}  \Tia{\epsilon}{j_1}{\tra_1}{\ell}\Tia{\epsilon}{j_2}{\tra_2}{\ell}\Tia{\epsilon}{j_3}{\tra_3}{\ell} +\o{\frac{1}{n^2}}\, . \notag
\end{align}
Importantly, all the quantities on the right-hand side of the network update \eqref{eq:preactivation-updated-finite-width-decomposition} -- $\z{i}{\delta}{L}$, $\Tia{\NTK}{ij}{\delta\tra}{L}$, $\Tia{\epsilon}{j}{\tra}{L}$, $\Tia{\dNTK}{i j_1j_2}{\delta\tra_1\tra_2}{L}$, and $\Tia{\ddNTK}{i j_1j_2j_3}{\delta\tra_1\tra_2\tra_3}{\ell}$
-- are evaluated at initialization and thus are determined completely by the statistics of the joint preactivation-NTK-dNTK-ddNTKs distribution,
\be\label{eq:finite-width-joint}
p\!\le(z^{(L)}, \, \NTK^{(L)},\, \dNTK^{(L)},\, \ddNTK^{(L)} , \,\ddNTKII^{(L)}   \Big\vert \D\ri) \, ,
\ee
that we spent the majority of this book evaluating in detail. 
Accordingly -- just as we did before at infinite width (\S\ref{ch:NTHb}) -- we could use the joint distribution \eqref{eq:finite-width-joint} to compute the statistics of the fully-trained network outputs after the update \eqref{eq:preactivation-updated-finite-width-decomposition}, \emph{if} we tuned a single step of gradient descent for each realization of a network so that we landed on the minimum of the training loss $\z{i}{\tra}{L}(t=1)=\y{i}{\tra}$.

More generally, \emph{if} we trained the network with $T$ steps of gradient descent such that
\be\label{eq:multi-step-satisfaction}
\z{i}{\tra}{L}(t=T)=\y{i}{\tra}\, ,\qquad \text{for all}\ \tra\in\A\, ,
\ee
\emph{and if} we determined how to express the output of such a fully-trained network for general inputs $\delta\in\D$ as a functional of our statistical variables at initialization,
\be\label{eq:multi-step-satisfied-outputs}
\z{i}{\delta}{L}(T)\equiv \le[\z{i}{\delta}{L}(t=T)\ri]\!\le(z^{(L)}, \, \NTK^{(L)},\, \dNTK^{(L)},\, \ddNTK^{(L)} , \,\ddNTKII^{(L)}  \ri)\, ,
\ee
then we could give an analytical expression for the distribution of \emph{fully-trained network outputs}:
\be\label{eq:finite-width-trained}
p\!\le(z^{(L)}(T) \ri) \, .
\ee
The equation \eqref{eq:multi-step-satisfaction} is our \emph{fully-trained condition}\index{fully-trained condition!finite width}, and the distribution \eqref{eq:finite-width-trained} completely describes the ensemble of finite-width networks at the end of training. The theoretical understanding of this distribution is exactly the goal we set for ourselves at the beginning of this book in \S\ref{ch:introduction}.
The only thing left for us to work out is the functional \eqref{eq:multi-step-satisfied-outputs}; to do so, we first need to figure out what kind of steps to take in order to fully train these finite-width networks.\index{training dynamics!finite width}

Now, recall from \S\ref{subsec:algorithmic-independence-at-infinity} that in the infinite-width limit the fully-trained network solution \eqref{eq:multi-step-satisfied-outputs} had an \neo{algorithm independence}: the distribution at the end of training didn't depend on the details of the optimization algorithm, and thus we could perform our theoretical analysis with any algorithm we wanted. In contrast, at finite width the fully-trained solution~\eqref{eq:multi-step-satisfied-outputs} will have an \term{algorithm dependence}: different fully-trained solutions will make different test-set predictions depending on the details of the particular optimization algorithm used to train the network, even when holding fixed the statistical variables at initialization and their associated initialization and training hyperparameters. Encouragingly, the form of the solutions will nonetheless take a universal form, with the non-universal details of the particular training algorithm captured by six projective tensors: cf.~\eqref{eq:very-general-finite-width-solution-DONT-CHANGE}. Thus, we will be able to very generally study the distribution of fully-trained networks at finite width by working out such solutions.\index{universality!of the fully-trained network solution}

With that in mind, in this section we'll present fully-trained solutions for two different optimization algorithms. First, in \S\ref{subsec:giant-plus-small}
we'll take \emph{two} Newton-like steps in order to satisfy our fully-trained condition \eqref{eq:multi-step-satisfaction}. While practically infeasible for large training sets, this training algorithm is rich in pedagogical value, emphasizing the way in which a finite-width network needs to adapt its representation to the training data in order to minimize its training error. Then, in \S\ref{subsec:real-GD-at-finite-width} we'll analytically solve the dynamics of the vanilla gradient descent at order $1/n$ and obtain a slightly different ensemble of fully-trained finite-width networks. This algorithm is not only practically implementable, but also quite often used to optimize real neural networks, and our corresponding solution is an actual theoretical description of such fully-trained networks. Together, these solutions will help us understand the ways in which the details of the optimization algorithm can affect the corresponding fully-trained solution. Finally, in 
\S\ref{subsec:prediction-at-finite-width}
we'll be able to generally analyze the predictions of these different fully-trained networks on novel examples from the test set.\index{training dynamics!finite width}

Throughout this section, we will declutter the notation a bit by dropping the layer indices, since to understand training we only need to focus on the network output at layer $\ell=L$.

\subsubsection{An Infinite-Width Giant Leap at Finite Width}\index{training dynamics!finite width}
Before we begin, let's first review the \terminate{giant leap} that we took in \S\ref{sec:giant-leap} at infinite width. From the finer-grained perspective of finite width, we'll see that our leap actually missed the minimum, exhibiting training errors of order $1/n$. However, our new eyes on this leap will be instructive, as they will help us see how we can correct for these errors and reduce the finite-width training error even further.

Recall from \S\ref{sec:giant-leap} that, in order to fully train an infinite-width network in a single step, we needed to make a \emph{second-order update} of the form
\begin{align}\label{eq:second-order-update-reprint}
\dtheta_\mu &=  - \sum_{\nu,\tra_1,\tra_2,i}\eta \lambda_{\mu\nu} \kappa^{\tra_1\tra_2}\frac{\td z_{i;\tra_1}}{\td \theta_\nu} (z_{i;\tra_2}-\y{i}{\tra_2})\, ,  %
\end{align}
which we interpreted either 
\emph{(i)} as a \emph{generalized training algorithm}
\eqref{eq:second-order-update} optimizing the standard MSE loss~\eqref{eq:MSE-loss-reprint}, or \emph{(ii)} as a standard (tensorial) gradient-descent step \eqref{eq:gd-update-lambda} optimizing a \emph{generalized MSE loss} \eqref{eq:loss-MSE-gen}. Here, $\kappa^{\tra_1\tra_2}$ was called the \neo{Newton tensor}, and -- in the first understanding of \eqref{eq:second-order-update-reprint} -- could be interpreted as allowing us to take anisotropic steps in training \terminate{sample space}.

\index{training dynamics!finite width}
With this type of parameter update, a finite-width network output will evolves as
\begin{align}\label{eq:preactivation-updated-finite-width-decomposition-kappa}
&z_{i;\delta}(t=1)\, \\
=&z_{i;\delta}-\eta\sum_{\tra_1, \tra_2}\NTKM_{\delta\tra_1}\kappa^{\tra_1\tra_2}\le(z_{i;\tra_2}-\y{i}{\tra_2}\ri) -\eta\sum_{j,\tra_1, \tra_2}\DNTKS_{ij;\delta\tra_1}\kappa^{\tra_1\tra_2}\le(z_{j;\tra_2}-\y{j}{\tra_2}\ri) \, \notag\\
&+\frac{\eta^2}{2}\sum_{j_1,j_2,\tra_1,\tra_2,\tra_3,\tra_4}\!\!\!\!\!\!\dNTK_{i j_1j_2;\delta\tra_1\tra_2} \kappa^{\tra_1\tra_3}\kappa^{\tra_2\tra_4}\! \le(z_{j_1;\tra_3}-\y{j_1}{\tra_3}\ri)\!\le(z_{j_2;\tra_4}-\y{j_2}{\tra_4}\ri) \, \notag \\
&-\frac{\eta^3}{6}\sum_{j_1,j_2,j_3,\tra_1,\ldots, \tra_6}\tia{\ddNTK}{i j_1j_2j_3}{\delta\tra_1\tra_2\tra_3}  
\kappa^{\tra_1\tra_4}\kappa^{\tra_2\tra_5}\kappa^{\tra_3\tra_6} \, \notag \\
&\qquad\qquad\qquad\qquad\quad\times\le(z_{j_1;\tra_4}-\y{j_1}{\tra_4}\ri)\le(z_{j_2;\tra_5}-\y{j_2}{\tra_5}\ri)\le(z_{j_3;\tra_6}-\y{j_3}{\tra_6}\ri)
+\o{\frac{1}{n^2}}\, ,\notag
\end{align}
where here we've made our usual decomposition of the NTK into a mean and fluctuation\index{tensor decomposition!NTK mean and fluctuation} %
\be
\NTK_{ij;\delta\tra}\equiv\delta_{ij}\NTKM_{\delta\tra}+\DNTKS_{ij;\delta\tra} \, .
\ee
In terms of such an update, our fully-trained condition \eqref{eq:multi-step-satisfaction} after a single step $T=1$,
\be
z_{i;\tra}(t=1)=\y{i}{\tra} \,,\qquad \text{for all}\ \tra\in\A\, ,
\ee
can be written as
\begin{align}\label{eq:what-we-want-rearranged-finite-width}
&z_{i;\tra}-\y{i}{\tra}\, \\
=&\eta\sum_{\tra_1, \tra_2}\NTKM_{\tra\tra_1}\kappa^{\tra_1\tra_2}\le(z_{i;\tra_2}-\y{i}{\tra_2}\ri) +\eta\sum_{j,\tra_1, \tra_2}\DNTKS_{ij;\tra\tra_1}\kappa^{\tra_1\tra_2}\le(z_{j;\tra_2}-\y{j}{\tra_2}\ri)  \, \notag\\
&-\frac{\eta^2}{2}\sum_{j_1,j_2,\tra_1,\tra_2,\tra_3,\tra_4}\dNTK_{i j_1j_2;\tra\tra_1\tra_2}\kappa^{\tra_1\tra_3}\kappa^{\tra_2\tra_4}\!\le(z_{j_1;\tra_3}-\y{j_1}{\tra_3}\ri)\le(z_{j_2;\tra_4}-\y{j_2}{\tra_4}\ri) \, \notag \\
&+\frac{\eta^3}{6}\sum_{j_1,j_2,j_3,\tra_1,\ldots, \tra_6}\tia{\ddNTK}{i j_1j_2j_3}{\tra\tra_1\tra_2\tra_3}  
\kappa^{\tra_1\tra_4}\kappa^{\tra_2\tra_5}\kappa^{\tra_3\tra_6}  \, \notag \\
&\qquad\qquad\qquad\qquad\quad\times \le(z_{j_1;\tra_4}-\y{j_1}{\tra_4}\ri)\le(z_{j_2;\tra_5}-\y{j_2}{\tra_5}\ri)\le(z_{j_3;\tra_6}-\y{j_3}{\tra_6}\ri)
+\o{\frac{1}{n^2}}
\, . \notag
\end{align}
This new giant-leap condition \eqref{eq:what-we-want-rearranged-finite-width} perhaps seems a little daunting, and further it's not obvious that there's any particular choice of Newton tensor $\kappa^{\tra_1\tra_2}$ that can land us on the minimum.
Nonetheless, we do expect that our infinite-width solution should be near the true finite-width solution, up to errors of order $\o{1/n}$.\footnote{
    For deep networks, we know that technically the corrections will be of order $\o{L/n}$, corresponding to the cutoff scale of our effective theory description. 
} 

\index{training dynamics!finite width}
With that in mind, as a first step let's try our infinite-width giant leap \eqref{eq:parameters-update-reprint-generalized}
and see where we land. This infinite-width giant leap had an interpretation as \neo{Newton's method} and was given by the second-order update \eqref{eq:second-order-update-reprint}, with a particular choice of the product of the global learning rate and the Newton tensor,
\be\label{eq:newtons-method-update-reprint-finite}
\eta \kappa^{\tra_1\tra_2}=\NTKMsub^{\tra_1 \tra_2} \, ,
\ee
where the \emph{inverse} NTK mean submatrix
$\NTKMsub^{\tra_1 \tra_2}$
was defined implicitly via
\be\label{eq:training-set-full-mean-ntk-inverse}
\sum_{\tra_2\in\A} \NTKMsub^{\tra_1 \tra_2} \, \NTKMsub_{\tra_2 \tra_3} =\delta^{\tra_1}_{\ \tra_3}\, .
\ee
As always, the tilde on
$\NTKMsub_{\tra_1 \tra_2}$
emphasizes that it's an $\NR\times\NR$-dimensional submatrix of the NTK mean evaluated on pairs of training inputs \emph{only}.
By now this distinction should be familiar enough that we will stop belaboring it.

\index{training dynamics!finite width}
More importantly, here we've used the inverse of the full NTK mean
$\NTKMsub_{\tra_1\tra_2}$
rather than the infinite-width frozen NTK\index{frozen NTK}
$\NTKIsub_{\tra_1\tra_2}$.
To explain why, let us recall from \eqref{eq:NTK-mean-expansion} that the NTK mean receives a series of corrections at each order in the $1/n$ expansion\index{$1/n$ expansion}, of which the leading-order piece is the frozen NTK \eqref{eq:frozen-NTK}. Since we're now working at order $1/n$, we should in particular take into account the next-to-leading-order (NLO) $1/n$ correction to the NTK mean
$\NTKM_{\tra_1\tra_2}^{\le\{1\ri\}}$ by working with $\NTKMsub_{\tra_1\tra_2}$ instead of $\NTKIsub_{\tra_1\tra_2}$.\footnote{
    While we won't show it explicitly, we expect that this \emph{NLO NTK mean}\index{neural tangent kernel!mean!next-to-leading-order correction}, $\NTKM_{\tra_1\tra_2}^{\le\{1\ri\}(\ell)}$, will have a solution that scales like $\o{1/n}$ as compared to the frozen NTK; this would be analogous to what we found for the NLO metric\index{metric!next-to-leading-order correction} $\se{\tra_1\tra_2}{\ell}$, cf.~the discussion in \S\ref{sec:signal_prop_finite_width} after \eqref{eq:nlo-metric-subleading}. In particular, if we make a \terminate{$1/n$ expansion} for our training hyperparameters $\lambda_b^{(\ell)}$ and $\lambda_W^{(\ell)}$ as we did for our initialization hyperparameters in \eqref{eq:Cb-expansion} and \eqref{eq:CW-expansion}, then we will have extra freedom in the subleading hyperparameters $\lambda_b^{(\ell)\{1\}}$ and $\lambda_W^{(\ell)\{1\}}$ to eliminate the growing-in-$\ell$ contribution to $\NTKM_{\tra_1\tra_2}^{\le\{1\ri\}(\ell)}$. Overall, this will make the NLO correction to the NTK mean scale as $\o{1/n}$, subleading to the leading finite-width effects which scale as $\o{L/n}$: in the language of RG flow\index{renormalization group flow}\index{representation group flow}, the NLO NTK mean is \emph{marginal}\index{marginal (RG flow)}. In practice, such a contribution is negligible and can thus be neglected for networks of any real depth.

}

\index{training dynamics!finite width}
Substituting our
infinite-width giant-leap Newton tensor \eqref{eq:newtons-method-update-reprint-finite}
into our giant-leap condition \eqref{eq:what-we-want-rearranged-finite-width} at finite width and rearranging, we get
\begin{align}\label{eq:preactivation-updated-finite-width-decomposition-Newton-1-naive}
0=&\sum_{j,\tra_1,\tra_2}\DNTKS_{ij;\tra\tra_1} \NTKMsub^{\tra_1\tra_2}\!\le(z_{j;\tra_1}-\y{j}{\tra_1}\ri)\, \\
&-\frac{1}{2}\sum_{\substack{j_1,j_2,\\ \tra_1,\ldots, \tra_4}}\dNTK_{i j_1j_2;\tra\tra_1\tra_2} \NTKMsub^{\tra_1\tra_3}\NTKMsub^{\tra_2\tra_4}\!\le(z_{j_1;\tra_3}-\y{j_1}{\tra_3}\ri)\!\le(z_{j_2;\tra_4}-\y{j_2}{\tra_4}\ri) \, \notag \\
&+\frac{1}{6}\sum_{\substack{j_1,j_2,j_3, \\ \tra_1,\ldots, \tra_6} }\tia{\ddNTK}{i j_1j_2j_3}{\tra\tra_1\tra_2\tra_3}  
\NTKMsub^{\tra_1\tra_4}\NTKMsub^{\tra_2\tra_5}\NTKMsub^{\tra_3\tra_6}  \notag \\
&\qquad\qquad\qquad\times \le(z_{j_1;\tra_4}-\y{j_1}{\tra_4}\ri)\le(z_{j_2;\tra_5}-\y{j_2}{\tra_5}\ri)\le(z_{j_3;\tra_6}-\y{j_3}{\tra_6}\ri)
+\o{\frac{1}{n^2}}\, . \notag
\end{align}
Thus, we actually missed the minimum: for the network to be fully trained, the right-hand side of \eqref{eq:preactivation-updated-finite-width-decomposition-Newton-1-naive} should have vanished, while here it is clearly nonzero in general.
Taking a step back, it's clear now that what we actually found at infinite width in \S\ref{sec:giant-leap} was
\be
z_{i;\tra}(t=1)-\y{i}{\tra} = \oninv \, .
\ee
In other words, our networks were fully trained only at leading order, and \eqref{eq:preactivation-updated-finite-width-decomposition-Newton-1-naive} gives an explicit expression for the $1/n$ correction.\index{training dynamics!finite width}

Disentangling a little further, there are two such corrections at order $1/n$: the first correction -- the first term on the right-hand side of \eqref{eq:preactivation-updated-finite-width-decomposition-Newton-1-naive} -- arises from the instantiation-to-instantiation fluctuations of the NTK across different realizations of the biases and weights at initialization; the second correction -- comprised of the second and third terms on the right-hand side of \eqref{eq:preactivation-updated-finite-width-decomposition-Newton-1-naive} -- arises from nonlinear changes in the output as we take our step. In particular, this second correction is a \neo{bona fide} manifestation of representation learning\index{representation learning!manifested at finite width} at finite width, accounting for the fact that the network's \emph{effective features}\index{feature function!effective} are evolving. 
If we properly account for these two types of the $1/n$ corrections, we should be able to attain a training error of order $1/n^2$:
\be\label{eq:fully-trained-condition-really}
z_{i;\tra}(T)-\y{i}{\tra} = \o{\frac{1}{n^2}} \, .
\ee
That is, we should be able to improve our effective theory\index{effective theory!representation learning} of fully-trained networks, quantitatively by another multiplicative factor of $1/n$, and qualitatively by properly including representation learning into such an effective description.

Our first approach (\S\ref{subsec:giant-plus-small}) to attain such effectively-zero training error, \eqref{eq:fully-trained-condition-really}, is to continue to engineer theoretical giant leaps so as to account for both the instantiation-to-instantiation fluctuations of the NTK and the effect of the dNTK and the first ddNTK.

Another approach (\S\ref{subsec:real-GD-at-finite-width}) is to simply use the vanilla tensorial gradient descent algorithm as we do in practice; in that case, we will have to not only account for the dynamics of the network output, but also account for the dynamics of the NTK and dNTK. After doing so, we
will see that we can iteratively decrease the training loss to zero after many many such steps.

\index{training dynamics!finite width}
\subsection{A Small Step Following a Giant Leap}\label{subsec:giant-plus-small}
Here we'll train our networks with \emph{two} second-order updates. For the first update, a giant leap, we'll need to further generalize our theoretical optimization algorithm in order to properly account for the instantiation-to-instantiation fluctuations of the NTK. In the second update, a small step, we'll be able to account for the $1/n$ change in representation due to the nonzero dNTK and ddNTK, ultimately landing on the minimum of the loss as \eqref{eq:fully-trained-condition-really}. In particular, 
we can think of these updates as loosely corresponding to distinct phases of training that arise
when implementing gradient-based training of neural networks in practice.

\index{training dynamics!finite width}
\subsubsection{First Update: One Final Generalization of Gradient Decent}
Please flip back to take a closer look at our unsatisfied condition for fully training our networks \eqref{eq:preactivation-updated-finite-width-decomposition-Newton-1-naive}.
Right away, you should notice a serious problem in satisfying this constraint: the NTK 
fluctuation, $\DNTKS_{ij;\tra\tra_1}$, the dNTK, $\dNTK_{i j_1j_2;\tra\tra_1\tra_2}$, and the ddNTK, $\tia{\ddNTK}{i j_1j_2j_3}{\tra\tra_1\tra_2\tra_3}$, all
mix different output components together in an update; i.e.~the $j$-th component of the prediction error
$z_{j;\tra}-\y{j}{\tra}$
at initialization affects the $i$-th component of the output
$z_{i;\delta}$
after the update, even for $i\ne j$. This stands in contrast to what we found for an infinite-width update in \S\ref{subsec:GD_no_wiring_at_infinity}, where there was no such mixing or \emph{wiring}\index{gradient descent!wiring!finite width} of output components. Meanwhile, we did see a similar wiring effect for Bayesian inference at finite width as we discussed in depth in \S\ref{subsec:presence-FF-Bayes}.

While such wiring at finite width is fantastic from a practitioner's standpoint, it makes our theoretical work slightly more complicated. 
In particular, in order to satisfy the training constraint \eqref{eq:preactivation-updated-finite-width-decomposition-Newton-1-naive}, we will need to further generalize our \terminate{second-order update} \eqref{eq:second-order-update-reprint}. Following in the footsteps of our previous two generalizations,
\eqref{eq:gd-update-lambda} and \eqref{eq:generalization-of-generalization}, let's make one final generalization
\be
\eta\to\eta\lambda_{\mu\nu} \to \eta \lambda_{\mu\nu}\kappa^{\tra_1\tra_2} \to\eta \lambda_{\mu\nu}\kappa^{\tra_1\tra_2}_{ij} \,,
\ee
with the final form of our theoretical update given by\index{second-order update!generalized}
\begin{align}\label{eq:second-order-update-per-component}
\dtheta_\mu  &=  - \sum_{\nu,\tra_1,\tra_2,i,j}\eta \lambda_{\mu\nu} \kappa^{\tra_1\tra_2}_{ij}\frac{\td z_{i;\tra_1}}{\td \theta_\nu}\epsilon_{j;\tra_2} \, .
\end{align}
Here, we also introduced a further
\emph{generalized} Newton tensor\index{Newton tensor!generalized} $\kappa^{\tra_1\tra_2}_{ij}$ with output-component indices. This flexibility will allow us to resolve the mixing of the output components in the residual training error from our infinite-width leap \eqref{eq:preactivation-updated-finite-width-decomposition-Newton-1-naive}.\footnote{Similarly to the generalized MSE loss  \eqref{eq:loss-MSE-gen} discussed in \S\ref{subsec:memorization-at-infinity}, we can alternatively think of this further-generalized second-order update \eqref{eq:second-order-update-per-component} optimizing the standard MSE loss as instead arising from a standard first-order (tensorial) gradient descent update \eqref{eq:gd-update-lambda} on a further-generalized MSE loss\index{loss!MSE!generalized},
\be\label{eq:loss-MSE-gen-further}
\Laux{\A}(\theta) \equiv\frac{1}{2}\sum_{i_1,i_2=1}^{n_L} \sum_{\tra_1,\tra_2 \in \A}\kappa^{\tra_1 \tra_2}_{i_1i_2}\le(z_{i_1;\tra_1}-\y{i_1}{\tra_1}\ri)\le(z_{i_2;\tra_2}-\y{i_2}{\tra_2}\ri) \, ,
\ee
as long as the Newton tensor $\kappa^{\tra_1 \tra_2}_{i_1i_2}$ is \emph{also} assumed to be symmetric under the exchange of paired indices $(i_1,\tra_1)\leftrightarrow(i_2,\tra_2)$. This symmetry will in fact be present for both our first update, the giant leap, and our second update, the small step.
With this interpretation \eqref{eq:loss-MSE-gen-further}, our generalized Newton tensor $\kappa^{\tra_1\tra_2}_{ij}$ acts as a metric\index{Newton tensor!as a metric on sample space} on \terminate{sample space}, through its sample indices $\tra_1,\tra_2$, and on output-component space, through its $L$-th layer neural indices $i,j$.}

\index{training dynamics!finite width}
Note importantly that the $\mu,\nu$ indices of $\lambda_{\mu\nu}$ are very different from the $i,j$ indices of $\kappa^{\tra_1\tra_2}_{ij}$: the former each runs over all $P$ parameters, while the latter each only runs over the $n_L$ components of the network output.
In particular, while learning-rate tensor\index{learning rate!learning-rate tensor} $\lambda_{\mu\nu}$ lets us control how the gradient of the $\nu$-th parameter affects the update to the $\mu$-th parameter, the $i,j$ indices of the generalized Newton tensor $\kappa^{\tra_1\tra_2}_{ij}$ instead control how the network's \emph{features}\index{feature} \eqref{eq:feature-function-stochastic}
$\td z_{i;\tra_1}/ \td \theta_\nu$ are combined with the error factor $\epsilon_{j;\tra_2}$
in order to make the update.\footnote{Note that when we developed our interpretation of an infinite-width network as a linear model in \S\ref{sec:lazy-kernel}, we  made a similar distinction between parameter indices $\mu$ and output-component indices $i$ in \eqref{eq:feature-function-stochastic} when defining the random feature functions $\widehat{\fea}_{i,\mu}(x)$.
We also discussed how \emph{wiring} can be incorporated into a non-minimal model of representation learning in \S\ref{subsec:nonlinear-at-finite}.
} Allowing for a $\kappa^{\tra_1\tra_2}_{ij}$ with nonzero off-diagonal components in the $i,j$ indices, we can precisely tune the \emph{wiring}\index{gradient descent!wiring!finite width}\index{wiring!in gradient-based learning|see{gradient descent}} or mixing of output components that occurs in a particular update.

Finally, plugging our final-form second-order update\index{second-order update!generalized}  \eqref{eq:second-order-update-per-component} back into the $1/n$ expansion of the network update 
\eqref{eq:preactivation-updated-finite-width-refined}
and using the NTK, dNTK, and first ddNTK definitions, we can see how the network output changes after making an update with this new optimization algorithm: 
\begin{align}\label{eq:preactivation-updated-finite-width-decomposition-kappa-genearlized}
&z_{i;\delta}(t=1)\, \\
=&z_{i;\delta}-\eta\sum_{j,\tra_1, \tra_2}\NTKM_{\delta\tra_1}\kappa^{\tra_1\tra_2}_{ij}\le(z_{j;\tra_2}-\y{j}{\tra_2}\ri) -\eta\sum_{j,k,\tra_1, \tra_2}\DNTKS_{ij;\delta\tra_1}\kappa^{\tra_1\tra_2}_{jk}\le(z_{k;\tra_2}-\y{k}{\tra_2}\ri) \, \notag\\
&+\frac{\eta^2}{2}\sum_{ \substack{j_1,j_2,k_1,k_2, \\ \tra_1,\tra_2,\tra_3,\tra_4} }\dNTK_{i j_1j_2;\delta\tra_1\tra_2} \kappa^{\tra_1\tra_3}_{j_1k_1}\kappa^{\tra_2\tra_4}_{j_2k_2} \le(z_{k_1;\tra_3}-\y{k_1}{\tra_3}\ri)\le(z_{k_2;\tra_4}-\y{k_2}{\tra_4}\ri) \, \notag \\
&-\frac{\eta^3}{6}\sum_{\substack{j_1,j_2,j_3, k_1, k_2, k_3 \\ \tra_1,\tra_2,\tra_3,\tra_4,\tra_5, \tra_6} }\tia{\ddNTK}{i j_1j_2j_3}{\delta\tra_1\tra_2\tra_3}  
\kappa^{\tra_1\tra_4}_{j_1k_1}\kappa^{\tra_2\tra_5}_{j_2k_2}\kappa^{\tra_3\tra_6}_{j_3k_3}  \, \notag \\
&\qquad\qquad\qquad\qquad\qquad\times \le(z_{k_1;\tra_4}-\y{k_1}{\tra_4}\ri)\le(z_{k_2;\tra_5}-\y{k_2}{\tra_5}\ri)\le(z_{k_3;\tra_6}-\y{k_3}{\tra_6}\ri)
+\o{\frac{1}{n^2}}\, .\notag
\end{align}
Here, we've again used the standard MSE loss\index{loss!MSE}, for which the error factor is given by the residual training error $\epsilon_{j;\tra}=z_{j;\tra}-\y{j}{\tra}$. Now, we'll need to pick our generalized Newton tensor $\kappa^{\tra_1\tra_2}_{ij}$ judiciously in order to make a first update that fully accounts for instantiation-to-instantiation fluctuations of particular networks in our ensemble. %

\index{training dynamics!finite width}
Taking inspiration from our $1/n$-failed infinite-width Newton's step \eqref{eq:newtons-method-update-reprint-finite}, let's take a similar-looking first step according to
\begin{align}\label{eq:newton-step-generalized-first}
\eta\kappa^{\tra_1\tra_2}_{ij}=&\le(\NTK^{-1}\ri)^{\tra_1\tra_2}_{ij}\, \\
=& \delta_{ij} \NTKMsub^{\tra_1 \tra_2} -\sum_{\tra_3,\tra_4\in\A}\NTKMsub^{\tra_1\tra_3}\DNTKS_{ij;\tra_3\tra_4}\NTKMsub^{\tra_4\tra_2} \, \notag\\
&+\sum_{k=1}^{n_L}\sum_{\tra_3,\ldots,\tra_6\in\A}\NTKMsub^{\tra_1\tra_3}\DNTKS_{ik;\tra_3\tra_4}\NTKMsub^{\tra_4\tra_5} \DNTKS_{kj;\tra_5\tra_6}\NTKMsub^{\tra_6\tra_2} +\o{\Delta^3}\, .\notag \,
\end{align}
Here, we've introduced the complete inverse of the stochastic NTK sub-tensor evaluated on the training set, satisfying 
\be\label{eq:stochastic-ntk-mean}
\sum_{j, \tra_2}\le(\NTK^{-1}\ri)^{\tra_1\tra_2}_{ij}\NTK_{jk;\tra_2\tra_3}=\delta_{ik}\delta^{\tra_1}_{\ \tra_3}\, ,
\ee
and in the last equality of \eqref{eq:newton-step-generalized-first}
we've used the Schwinger-Dyson equations\index{Schwinger-Dyson equations} \eqref{eq:stochastic-metric-inversion}, which is a physicist's way of saying that we expanded the inverse of the stochastic NTK around the NTK mean.\footnote{Despite saying that we wouldn't belabor this any further, this is one of those unfortunate situations where we had to decide between decorating the inverse $\le(\NTK^{-1}\ri)^{\tra_1\tra_2}_{ij}$ with either a hat or a tilde, and we went with the hat. Hopefully the tildes on the sample indices, e.g.~$\tra_1,\tra_2$, will remind you that the inverse of this stochastic object is taken only with respect to the training set. Note that at no point will we ever need the inverse of the stochastic NTK evaluated on a general dataset\index{input data} $\D$.}
The main difference between this new giant leap \eqref{eq:newton-step-generalized-first} and our previous infinite-width giant leap \eqref{eq:newtons-method-update-reprint-finite} is that we're now taking into account the instantiation-to-instantiation fluctuations of the NTK across different realizations of the model parameters; in other words, we're implementing a \emph{different} Newton step for each \emph{particular}  network with its associated NTK $\NTK_{i_1i_2;\tra_1\tra_2}$. Accordingly, this step can be thought of as loosely corresponding to the first phase of training for such a particular network.

\index{training dynamics!finite width}
Taking this step, i.e.~plugging this generalized Newton tensor \eqref{eq:newton-step-generalized-first} into our update to the network output \eqref{eq:preactivation-updated-finite-width-decomposition-kappa-genearlized}, we find the training error decreases to
\begin{align}\label{eq:preactivation-updated-finite-width-decomposition-Newton-1}
&z_{i;\tra}(t=1)-\y{i}{\tra}\, \\
=&\frac{1}{2}\!\!\sum_{\substack{j_1,j_2,\\ \tra_1,\ldots, \tra_4}}\!\!\dNTK_{i j_1j_2;\tra\tra_1\tra_2}\NTKMsub^{\tra_1\tra_3} \NTKMsub^{\tra_2\tra_4}\!\le(z_{j_1;\tra_3}-\y{j_1}{\tra_3}\ri)\!\le(z_{j_2;\tra_4}-\y{j_2}{\tra_4}\ri) \, \notag \\
&-\frac{1}{6}\sum_{\substack{j_1,j_2,j_3, \\ \tra_1,\ldots \tra_6} }\tia{\ddNTK}{i j_1j_2j_3}{\delta\tra_1\tra_2\tra_3}  
\NTKMsub^{\tra_1\tra_4}\NTKMsub^{\tra_2\tra_5}\NTKMsub^{\tra_3\tra_6} \, \notag \\
&\qquad\qquad\qquad\times \le(z_{j_1;\tra_4}-\y{j_1}{\tra_4}\ri)\le(z_{j_2;\tra_5}-\y{j_2}{\tra_5}\ri)\le(z_{j_3;\tra_6}-\y{j_3}{\tra_6}\ri)
+\o{\frac{1}{n^2}}\, .\notag
\end{align}
Thus, we've correctly taken care of the first type of the $1/n$ corrections, and
with this step we've reduced the residual prediction error on the training set
from the order-one error of our initial prediction 
\be
z_{i;\tra}(t=0)-\y{i}{\tra} = \o{1} \, ,
\ee
to a much smaller error of
\be\label{eq:one-step-error-factor-suppression}
z_{i;\tra}(t=1)-\y{i}{\tra} = \oninv \, ,
\ee
loosely corresponding to the empirically-large initial decrease of error when training networks in practice.
Moreover, the rest of the error in \eqref{eq:preactivation-updated-finite-width-decomposition-Newton-1} is now entirely due to
the additional finite-width corrections
encoded by the dNTK and first ddNTK, a consequence of representation learning.\footnote{
In particular, this first update \eqref{eq:newton-step-generalized-first} would satisfy the training condition \eqref{eq:fully-trained-condition-really} only if the NTK were constant under gradient descent.
There's actually a name given to this type of phenomenon, \neo{lazy training}, referring to situations when the network function behaves as if it is equal to a linearzation around the initial value of its parameters \cite{chizat2018note}. As we know from our discussion in \S\ref{sec:lazy-kernel}, if the NTK is constant, then the network is a \terminate{linear model}.
}

\index{training dynamics!finite width}
\subsubsection{Second Update: Representation Learning Strikes Back}
To reduce the training error further, we'll need to update our network again to further account for the fact that its representation evolved with the first update.  
In other words, we'll need to make a second gradient-descent update. 
If you'd like, you can imagine that this update 
corresponds to a second phase of training -- following a first phase where the network significantly decreased its training error with the NTK evaluated at initialization --
and so now the model must refine its features\index{feature} in order to further improve its performance.

Accordingly, since our training error is already down to $\sim 1/n$, our second update is going to be a lot smaller than our first. In particular, the update itself will necessarily only be of order $1/n$ so as to precisely cancel the remaining $1/n$ training error; that is, it's actually more of a \emph{small step} than another \emph{giant leap}.\footnote{
As we will see, the overall learning rate is essentially the same for both updates; the second update is only smaller because the gradient of the loss after the first update is itself much smaller when near a minimum.
}

To determine which step we need to take, let's write the network output after a second update as
\begin{align}\label{eq:small-step-to-the-bottom}
&z_{i;\delta}(t=2) \\
=&z_{i;\delta}(t=1)-\sum_{j,k,\tra_1, \tra_2}\NTKM_{i j;\delta\tra_1}\!(t=1) \, \eta\kappa^{\tra_1\tra_2}_{jk}(t=1)   \le[z_{k;\tra_2}(t=1)  -\y{k}{\tra_2}\ri]  + \o{\frac{1}{n^2}}\, \notag \\
=&z_{i;\delta}(t=1)-\sum_{j,\tra_1, \tra_2}\NTKM_{\delta\tra_1} \, \eta\kappa^{\tra_1\tra_2}_{ij}(t=1)   \le[z_{j;\tra_2}(t=1)  -\y{j}{\tra_2}\ri]  + \o{\frac{1}{n^2}}\, , \notag
\end{align}
where we left the second update's product of global learning rate and generalized Newton tensor $\eta \kappa^{\tra_1\tra_2}_{ij}(t=1)$ unspecified for now. Note here that in the first equality we dropped the d(d)NTK terms from the update: since after our first update \eqref{eq:preactivation-updated-finite-width-decomposition-Newton-1} the training error has already decreased to $\o{1/n}$, the would-be dNTK term is very subleading $\sim \dNTK \times [\epsilon(t=1)]^2=\o{1/n^3}$, and
the would-be ddNTK term is ridiculously subleading $\sim \ddNTK \times [\epsilon(t=1)]^3=\o{1/n^4}$. Similarly, on the third line we replaced the NTK after the first step
$\NTKM_{ij;\delta\tra_1}\!(t=1)$ by the NTK mean 
at initialization $\delta_{ij}\NTKM_{\delta\tra_1}$. Given the $1/n$-suppressed training error after the first step, this substitution can be justified for these two reasons in conjunction: \emph{(i)} the update to the NTK  $\NTK_{ij;\delta\tra}(t=1)-\NTK_{ij;\delta\tra}(t=0)$ is itself suppressed by $1/n$, 
cf.~\eqref{eq:NTK-updated-finite-width-refined}, 
so we may use the version from before the update, and \emph{(ii)} the NTK fluctuation is also suppressed compared to its mean, so we may then swap the stochastic NTK at initialization for its mean.
This means that if we make the following choice for our \emph{small step} second update
\begin{align}\label{eq:newton-step-generalized-second}
\eta \kappa^{\tra_1\tra_2}_{ij}(t=1) =& \delta_{ij} \NTKMsub^{\tra_1 \tra_2} + \oninv \, ,
\end{align}
then it's easy to see from  \eqref{eq:small-step-to-the-bottom} that our network will now be fully trained  as \eqref{eq:fully-trained-condition-really}:
\be\label{eq:two-step-twice-reduced-error}
z_{i;\tra}(t=2) - \y{i}{\tra} = \o{\frac{1}{n^2}}\, .
\ee
Thus, with our second update we were able to reduce the residual training error by an additional factor of $1/n$ as compared to the error after the first update \eqref{eq:preactivation-updated-finite-width-decomposition-Newton-1}.\footnote{
This suggests that additional updates
could continue to reduce the training error by additional factors of $1/n$. However, these further refinements -- determined in terms of the higher-order corrections to our effective theory description -- will be qualitatively the same as our leading finite-width description at $\o{L/n}$. In other words, improving an infinite-width description to finite-width description incorporates representation learning, while more precise finite-width descriptions just allow the model to make further refinements to its features. Practically speaking, we expect our leading finite-width description to be very accurate for networks with reasonable values of the aspect ratio $L/n$.
}

\index{training dynamics!finite width}
For general inputs $\delta\in\D$, plugging our choice of learning rate and Newton tensor \eqref{eq:newton-step-generalized-second} back into our second update \eqref{eq:small-step-to-the-bottom} and further re-expressing the network output after the first step $z_{i;\delta}(t=1)$ by \eqref{eq:preactivation-updated-finite-width-decomposition-kappa-genearlized} with \eqref{eq:newton-step-generalized-first}, we get
\begin{align}\label{eq:finite-width-network-output-general-data}
&z_{i;\delta}(t=2)\, \\
=&z_{i;\delta}-\sum_{\tra_1,\tra_2\in\A}\NTKM_{\delta\tra_1}\NTKMsub^{\tra_1\tra_2}\!\le(z_{i;\tra_2}-\y{i}{\tra_2}\ri)\, \notag \\
&+\sum_{j=1}^{n_{L}}\sum_{\tra_1,\tra_2\in\A}\le[\DNTKS_{ij;\delta\tra_1}-\sum_{\tra_3,\tra_4\in\A}\NTKM_{\delta\tra_3}\NTKMsub^{\tra_3\tra_4}\DNTKS_{ij;\tra_4\tra_1}\ri]\NTKMsub^{\tra_1\tra_2}\!\le(z_{j;\tra_2}-\y{j}{\tra_2}\ri)\, \notag\\
&-\sum_{j,k=1}^{n_{L}}\sum_{\tra_1,\ldots,\tra_4\in\A}\le[\DNTKS_{ij;\delta\tra_1}-\sum_{\tra_5,\tra_6\in\A}\NTKM_{\delta\tra_5}\NTKMsub^{\tra_5\tra_6}\DNTKS_{ij;\tra_6\tra_1}\ri]\, \notag\\
&\quad\quad\quad\quad\quad\quad\quad\quad\times\NTKMsub^{\tra_1\tra_2}\DNTKS_{jk;\tra_2\tra_3}\NTKMsub^{\tra_3\tra_4}\!\le(z_{k;\tra_4}-\y{k}{\tra_4}\ri)\, \notag\\
&+\frac{1}{2}\sum_{j_1,j_2=1}^{n_{L}}\sum_{\tra_1,\ldots,\tra_4\in\A}\le[\dNTK_{i j_1j_2;\delta\tra_1\tra_2}-\sum_{\tra_5,\tra_6\in\A}\NTKM_{\delta\tra_5}\NTKMsub^{\tra_5\tra_6}\dNTK_{i j_1j_2;\tra_6\tra_1\tra_2}\ri]\, \notag\\
&\quad\quad\quad\quad\quad\quad\quad\quad\times\NTKMsub^{\tra_1\tra_3}\NTKMsub^{\tra_2\tra_4} \!\le(z_{j_1;\tra_3}-\y{j_1}{\tra_3}\ri)\!\le(z_{j_2;\tra_4}-\y{j_2}{\tra_4}\ri) \, \notag \\ 
&-\frac{1}{6}\sum_{j_1,j_2,j_3 =1 }\sum_{\tra_1,\ldots,\tra_6\in\A}\le[\tia{\ddNTK}{i j_1j_2j_3}{\delta\tra_1\tra_2\tra_3}
- \sum_{\tra_7, \tra_8 \in \A}\NTKM_{\delta\tra_7}\NTKMsub^{\tra_7\tra_8}\tia{\ddNTK}{i j_1j_2j_3}{\tra_8\tra_1\tra_2\tra_3} \ri]  
 \, \notag \\
&\quad\quad\quad\quad\quad\quad\quad\quad\times \NTKMsub^{\tra_1\tra_4}\NTKMsub^{\tra_2\tra_5}\NTKMsub^{\tra_3\tra_6} \le(z_{j_1;\tra_4}-\y{j_1}{\tra_4}\ri)\le(z_{j_2;\tra_5}-\y{j_2}{\tra_5}\ri)\le(z_{j_3;\tra_6}-\y{j_3}{\tra_6}\ri) \, \notag \\
&+\o{\frac{1}{n^2}}\, .\notag
\end{align}
Here, we see that the expressions in the square brackets vanish identically for all the training inputs $\delta=\tra\in\A$: our network is thus fully trained.
In particular, since all of the variables on the right-hand side of \eqref{eq:finite-width-network-output-general-data} are at initialization, this solution realizes our goal \eqref{eq:multi-step-satisfied-outputs} of expressing the output of a fully-trained network as a functional of such variables at initialization. Accordingly, the statistics of the fully-trained distribution \eqref{eq:finite-width-trained} can now be worked out from the joint preactivation-NTK-dNTK-ddNTKs distribution \eqref{eq:finite-width-joint} at initialization.\footnote{Alternatively, we could have reached this same solution in a \emph{single update} if we had instead made a  \emph{finely tuned} giant leap, picking the generalized Newton tensor as
\begin{align}\label{eq:newton-step-generalized}
\eta \kappa^{\tra_1\tra_2}_{ij} =& \delta_{ij} \NTKMsub^{\tra_1 \tra_2} -\sum_{\tra_3,\tra_4\in\A}\NTKMsub^{\tra_1\tra_3}\DNTKS_{ij;\tra_3\tra_4}\NTKMsub^{\tra_4\tra_2} \, \\
&+\sum_{k=1}^{n_L}\sum_{\tra_3,\ldots,\tra_6\in\A}\NTKMsub^{\tra_1\tra_3}\DNTKS_{ik;\tra_3\tra_4}\NTKMsub^{\tra_4\tra_5} \DNTKS_{kj;\tra_5\tra_6}\NTKMsub^{\tra_6\tra_2} \, \notag \\
&+\frac{1}{2}\sum_{k=1}^{n_L}\sum_{\tra_3,\ldots,\tra_6 \in \A} \NTKMsub^{\tra_1\tra_3}\NTKMsub^{\tra_2\tra_4}\NTKMsub^{\tra_5\tra_6}\dNTK_{i j k;\tra_3\tra_4\tra_5}\le(z_{k;\tra_6}-\y{k}{\tra_6} \ri)   \, \notag \\
&-\frac{1}{6}\sum_{k_1,k_2=1}^{n_L}\sum_{\tra_3,\ldots,\tra_8 \in \A} \NTKMsub^{\tra_1\tra_3}\NTKMsub^{\tra_2\tra_4}\NTKMsub^{\tra_5\tra_7}\NTKMsub^{\tra_6\tra_8}\tia{\ddNTK}{i j k_1k_2}{\tra_3\tra_4\tra_5\tra_6}  
  \, \notag \\
&\qquad\qquad\qquad\qquad\qquad\times\le(z_{k_1;\tra_7}-\y{k_1}{\tra_7}\ri)\le(z_{k_2;\tra_8}-\y{k_2}{\tra_8}\ri)
\notag \, .
\end{align}
In essence, the first three terms of \eqref{eq:newton-step-generalized}  come from inverting the stochastic NTK,
corresponding to our first-update giant leap \eqref{eq:newton-step-generalized-first} and 
accounting for the instantiation-to-instantiation fluctuations in the NTK. In contrast, the final two terms correspond to our second-update small step \eqref{eq:two-step-twice-reduced-error} and accounts for the dNTK-ddNTK-induced representation learning.
Note that this generalized Newton tensor \eqref{eq:newton-step-generalized} is \emph{asymmetric} under the exchange of paired indices $(i,\tra_1)\leftrightarrow(j,\tra_2)$ due to the last term, and thus this finely-tuned update doesn't admit an alternative interpretation of optimization with a further generalized loss \eqref{eq:loss-MSE-gen-further}.

This single update is analogous to the \neo{direct optimization} solution \eqref{eq:nearly-linear-regression-optimal-cleanest} for \terminate{quadratic regression} in \S\ref{sec:nonlinear-model}.
The very finely-tuned nature of this single-update algorithm \eqref{eq:newton-step-generalized} suggests that it's easier to make the fine adjustments required to reach a solution by taking many simpler steps rather than fewer complicated steps.
We will see the other side of this next in \S\ref{subsec:real-GD-at-finite-width} when we study training by many many steps of vanilla gradient descent.
}
\index{training dynamics!finite width}

\index{training dynamics!finite width}
While this is all very exciting, let us also caution you that these generalized second-order updates are probably best thought of as giving a simple theoretical model of a training algorithm -- designed to let us understand the training process analytically -- and by no means are we suggesting that they provide a good or useful option for practical optimization.
Instead, the most practical algorithm for optimization is vanilla first-order gradient descent. As we've already pointed out that the output of a fully-trained network  \emph{does} depend on the details of the algorithm used for optimization,
 now we really have no choice left other than to explicitly analyze many many steps of gradient descent.

 \subsection{Many Many Steps of Gradient Descent}\label{subsec:real-GD-at-finite-width}\index{training dynamics!finite width}
In this extended subsection, we're going to study tensorial gradient descent~\eqref{eq:gd-update-lambda} and optimize finite-width neural networks according to the MSE loss with a constant global learning rate $\eta$.\footnote{
    Don't let the word \emph{tensorial} scare you here; this just means that we will allow for our training hyperparameters $\lambda_b^{(\ell)}$ and $\lambda_W^{(\ell)}$ as part of the definition of the NTK. We hope it's already clear why including these hyperparameters is a good idea -- if not, please flip back to \S\ref{sec:EVGP-WEP} and reread the paragraphs on the learning rate \terminate{equivalence principle} -- and they are actually simple to include practically as part of any optimization algorithm. 

    Moreover, as they are just part of the definition of the NTK, they have absolutely no consequence on the dynamics presented here; i.e. our solution also covers \emph{non-tensorial} gradient descent, though in such a case you'd have different asymptotic solutions for the statistics of the NTK, dNTK, and ddNTKs. This is also why we think of the training hyperparameters $\lambda_b^{(\ell)}$ and $\lambda_W^{(\ell)}$ as being independent from the details of the optimization algorithm itself.\index{training hyperparameters!independent from the optimization algorithm}
}
With\index{training dynamics!finite width} this progenitor-of-all-other-gradient-based-learning-algorithm algorithm, the network output will evolve as
 \eqref{eq:preactivation-updated-finite-width-refined}
\begin{align}\label{eq:preactivation-update-redux}
z_{i;\delta}(t+1)=&z_{i;\delta}(t)-\eta\sum_{j,\tra}\NTKM_{ij;\delta\tra}(t)\le[z_{j;\tra}(t)-\y{j}{\tra}\ri]\, \\
&+\frac{\eta^2}{2}\sum_{j_1,j_2,\tra_1,\tra_2}\dNTKM_{i j_1j_2;\delta\tra_1\tra_2}(t)\le[z_{j_1;\tra_1}(t)-\y{j_1}{\tra_1}\ri]\le[z_{j_2;\tra_2}(t)-\y{j_2}{\tra_2}\ri]\, \notag \\
&-\frac{\eta^3}{6}\sum_{j_1,j_2,j_3,\tra_1,\tra_2,\tra_3}\tia{\ddNTK}{i j_1j_2j_3}{\delta\tra_1\tra_2\tra_3} \notag \,\\
&\qquad\qquad\qquad\qquad\times\le[z_{j_1;\tra_1}(t)-\y{j_1}{\tra_1}\ri]\le[z_{j_2;\tra_2}(t)-\y{j_2}{\tra_2}\ri]\le[z_{j_3;\tra_3}(t)-\y{j_3}{\tra_3}\ri]
\,  \notag,
\end{align}
in conjunction the NTK will evolve as \eqref{eq:NTK-updated-finite-width-refined}
\begin{align}\label{eq:NTK-update-redux}
&\NTKM_{i_1i_2;\delta_1\delta_2}(t+1) \, \\
=&\NTKM_{i_1i_2;\delta_1\delta_2}(t)-\eta\sum_{j,\tra}\Big(\dNTKM_{i_1 i_2j;\delta_1\delta_2\tra}(t)+\dNTKM_{i_2 i_1j;\delta_2\delta_1\tra}(t)\Big)\le[z_{j;\tra}(t)-\y{j}{\tra}\ri]\, \notag\\
&+\frac{\eta^2}{2}\sum_{j_1,j_2,\tra_1,\tra_2}\le(\tia{\ddNTK}{i_1i_2j_1j_2}{\delta_1\delta_2\tra_1\tra_2}+\tia{\ddNTK}{i_2i_1j_1j_2}{\delta_2\delta_1\tra_1\tra_2} + 2\tia{\ddNTKII}{i_1i_2j_1j_2}{\delta_1\delta_2\tra_1\tra_2} \ri) \, \notag\\ 
&\qquad\qquad\qquad\qquad\times\le[z_{j_1;\tra_1}(t)-\y{j_1}{\tra_1}\ri]\le[z_{j_2;\tra_2}(t)-\y{j_2}{\tra_2}\ri] \notag
\, , 
\end{align}
and in further conjunction the dNTK will evolve as \eqref{eq:dNTK-updated-finite-width}
\begin{align}\label{eq:dNTK-update-redux}
&\tia{\dNTKM}{i_0i_1i_2}{\delta_0\delta_1\delta_2}(t+1) \, \\
=&\tia{\dNTKM}{i_0i_1i_2}{\delta_0\delta_1\delta_2}(t) \, \notag \\
&-\eta\sum_{j,\tra}\le(\tia{\ddNTK}{i_0i_1 i_2j}{\delta_0\delta_1\delta_2\tra}+\tia{\ddNTKII}{i_0i_1 i_2j}{\delta_0\delta_1\delta_2\tra}+\tia{\ddNTKII}{i_0i_2 i_1j}{\delta_0\delta_2\delta_1\tra}\ri)\le[z_{j;\tra}(t)-\y{j}{\tra}\ri]\,\notag .
\end{align}
In writing these dynamical equations, we've stopped explicitly denoting that our equations have $\o{1/n^2}$ errors, and we've also expressed the fact that the ddNTKs\index{ddNTKs!step-independence} are $t$-independent at order $1/n$, using their values at initialization, $\ddNTK_{i_1 i_2i_3i_4;\delta_1\delta_2\delta_3\delta_4}$ and $\ddNTKII_{i_1 i_2i_3i_4;\delta_1\delta_2\delta_3\delta_4}$, with hats to remind us of their stochasticity.
These joint updates \eqref{eq:preactivation-update-redux}, \eqref{eq:NTK-update-redux}, and \eqref{eq:dNTK-update-redux} are coupled \emph{difference equations}\index{difference equation!nonlinear}\index{difference equation|seealso{training dynamics}}, and the equations for the network output and the dynamical NTK\index{dynamical NTK} are both \emph{nonlinear} in the output $z_{i;\delta}$. 

\index{training dynamics!finite width}
Although this seems daunting, we are now going to solve these equations in a closed form. First, we'll analyze the dynamics while neglecting the finite-width effects of the dNTK and ddNTKs, in which the problem will reduce to a single \emph{linear} difference equation.\index{difference equation!linear} Then, we'll use perturbation theory to incorporate the effect of all the NTK differentials and allow for a dynamically evolving NTK and dNTK.

\subsubsection{Free Theory: Step-Independent NTK}\index{training dynamics!finite width}
Let's begin by setting the dNTK and ddNTKs to zero. Since their leading statistics are $1/n$-suppressed, we'll be able to use perturbation theory to reincorporate all their effects later.
In this limit the NTK update equation \eqref{eq:NTK-update-redux} is trivial, solved by the \emph{free} or \emph{step-independent} NTK:\index{neural tangent kernel!step-independent}
\be\label{eq:step-independent-ntk-solution}
\NTKM_{i_1i_1;\delta_1\delta_2}(t)=\NTKM_{i_1i_2;\delta_1\delta_2}(t=0) \equiv \NTK_{i_1i_2;\delta_1\delta_2} \, .
\ee
Unsurprisingly, this just means that when the dNTK and ddNTKs vanish, the NTK doesn't update from its initialization.

\index{training dynamics!finite width}
Plugging this solution into the preactivation update equation  \eqref{eq:preactivation-update-redux} and turning off the dNTK and  ddNTKs, the remaining dynamical equation simplifies to
\be\label{eq:preactivation-update-redux-free}
z_{i;\delta}(t+1)=z_{i;\delta}(t)-\eta\sum_{j,\tra}\NTK_{ij;\delta\tra}[z_{j;\tra}(t)-\y{j}{\tra}]\, .
\ee
Thus, the residual training error, $z_{\tra}(t)-y_{\tra}$, sources the updates to the network output $z_{\delta}(t)$ for general inputs $\delta \in \D$.
Moreover, when restricted to the inputs from the training set $\tra\in\A$, we can rewrite this \terminate{difference equation} \eqref{eq:preactivation-update-redux-free} as
\be\label{eq:gd-as-matrix-multiplication}
z_{i;\tra}(t+1)-\y{i}{\tra}=\sum_{j,\tra_1}\le(\Iden_{ij;\tra\tra_1}-\eta\NTK_{ij;\tra\tra_1}\ri)[z_{j;\tra_1}\!(t)-\y{j}{\tra_1}]\, ,
\ee
where we've defined the \neo{identity operator}
\be
\Iden_{i_1i_2;\tra_1\tra_2}\equiv \delta_{i_1 i_2}\delta_{\tra_1\tra_2}\, .
\ee
In this form, \eqref{eq:gd-as-matrix-multiplication} is a first-order homogeneous linear difference equation\index{difference equation!linear!homogeneous} for the residual training error, $z_{\tra}(t)-y_{\tra}$, which is just a fancy way of saying that this is going to be a \terminate{piece of cake}\index{piece of cake|seealso{free dynamics}}. 

In particular, the update to the prediction error is just a simple multiplication by a constant matrix, and the solution is given by an exponential:
\be\label{eq:free-solution-training}
\zF{i}{\tra}(t)-\y{i}{\tra}=\sum_{j,\tra_1}\Unit_{ij;\tra\tra_1}\!(t)\le(z_{j;\tra_1}-\y{j}{\tra_1}\ri) \, .
\ee
Here, on the left-hand side, we've labeled the solution with an ``F'' to indicate it's the \emph{free solution}, with the nonlinear effects from the dNTK and ddNTKs turned off; on the right-hand side, we have the residual training error at initialization, and the \term{step-evolution operator} is defined as an iterative product of $t$ steps:
\begin{align}\label{eq:unitary-operator-at-least-in-continuum-limit-with-imaginary-time}
\Unit_{i_t i_0;\tra_t\tra_0}(t)\equiv&\le[\le(\Iden-\eta\NTK\ri)^t\ri]_{i_ti_0;\tra_t\tra_0}\, \\
=&\sum_{\substack{i_1,\ldots i_{t-1}\\ \tra_1,\ldots,\tra_{t-1}}}\le(\Iden_{i_t i_{t-1};\tra_t\tra_{t-1}}-\eta\NTK_{i_t i_{t-1};\tra_{t}\tra_{t-1}}\ri)\cdots\le(\Iden_{i_1 i_0;\tra_1,\tra_0}-\eta\NTK_{i_1 i_0;\tra_1\tra_0}\ri)\, .\notag
\end{align}
For any positive-definite NTK and sufficiently small global learning rate $\eta$, this operator will exponentially decay to zero, $\Unit(t)\to 0$, as the number of steps becomes large, $t\to\infty$.\footnote{
    In particular, for this limit to converge we just need $ ||\Iden-\eta \NTK||_\infty < 1$, i.e.~the largest eigenvalue of the operator $\Iden-\eta \NTK$  must be less than one. With our attention to the principles of criticality and equivalence,\index{equivalence principle}\index{criticality!principle of} our choices of initialization and training hyperparameters were made so that the NTK is always of order one, and thus it's very easy for our networks to satisfy this constraint. \label{footnote:convergence-dynamics}
} 
Thus, the residual training error \eqref{eq:free-solution-training} will vanish exponentially quickly, 
\be\label{eq:free-training-converges}
\lim_{t\to\infty} \zF{i}{\tra}(t) = \y{i}{\tra} \, ,
\ee
with the step scale for the decay of the individual components set by the step-independent NTK.\footnote{If we wanted to study the ODE or \neo{continuum limit}\index{continuum limit|see{gradient descent}}\index{gradient descent!continuum or ODE limit} of the dynamics, we could take the global learning rate to zero, $\eta\to 0$, while holding the product, $\tau\equiv \eta t$, fixed. In such a limit, the step-evolution operator becomes simply $\Unit(t)\to \exp(-\NTK\tau)$. While such a limit is mostly unnecessary for any theoretical purpose -- it's just as easy to study the discrete dynamics that actually describe the practical optimization algorithm -- it does provide more substance to objection in footnote~\ref{footnote:ntk-name} of \S\ref{sec:gd} of the name \neo{neural tangent kernel} for the stochastic operator $\NTK$.
\index{neural tangent kernel!name}
In particular, this continuum limit makes clear that the NTK is really best thought of as a \neo{Hamiltonian}, as it generates the evolution of observables, and the step-evolution operator $\Unit(t)$ is like a unitary time-evolution operator, albeit in $i$maginary time.
\index{imaginary time}
More precisely, in the limit where the dNTK and ddNTKs are set to zero, the NTK is akin to a \emph{free} Hamiltonian, with exactly solvable dynamics; with a nonzero dNTK or ddNTKs, the Hamiltonian includes nontrivial \emph{interactions}\index{interactions!dynamics} and can be analyzed via time-dependent \terminate{perturbation theory}.\label{eq:footnote-continuum-limit}
}

\index{dynamics|see{training dynamics}}\index{training dynamics!finite width}
Having now solved the free dynamics on the training set, we can plug this solution~\eqref{eq:free-solution-training} back into the \terminate{difference equation} \eqref{eq:preactivation-update-redux-free} for general inputs $\delta \in \D$. With the source known explicitly, we can easily write down a solution that satisfies the initial condition $\zF{i}{\delta}(t=0)=z_{i;\delta}$:
\begin{align}\label{eq:free-solution-general}
\zF{i}{\delta}(t)=&z_{i;\delta}-\sum_{j,\tra}\NTK_{ij;\delta\tra}\le\{\eta\sum_{s=0}^{t-1}\le[\zF{j}{\tra}(s)-\y{j}{\tra}\ri]\ri\}\, \\
=&z_{i;\delta}-\sum_{j,\tra_1,\tra_2}\NTK_{ij;\delta\tra_1}a_{j;\tra}(t)\, .\notag
\end{align}
Here we've defined a dynamical helper function, with an explicit representation given by
\begin{align}\label{eq:dynamical-helper-a}
a_{j;\tra}(t)\equiv&\eta\sum_{s=0}^{t-1}\le[\zF{j}{\tra}(s)-\y{j}{\tra}\ri]=\eta\sum_{s=0}^{t-1}\le[\sum_{k,\tra_1}\Unit_{jk;\tra\tra_1}\!(t)\le(z_{k;\tra_1}-\y{k}{\tra_1}\ri)\ri] \, \\
=&\eta\sum_{k,\tra_1} \le\{ \sum_{s=0}^{t-1}\!\le[ \le(\Iden-\eta\NTK\ri)^{s}\ri]_{j k;\tra\tra_1}\ri\}\le(z_{k;\tra_1}-\y{k}{\tra_1}\ri)\, \notag\\
=&\eta\sum_{k,m,\tra_1,\tra_2}\le\{\le[\Iden-\le(\Iden-\eta\NTK\ri)\ri]^{-1}\ri\}_{jm}^{\tra\tra_2}\le[\Iden-\le(\Iden-\eta\NTK\ri)^t\ri]_{mk;\tra_2\tra_1}\le(z_{k;\tra_1}-\y{k}{\tra_1}\ri)\, \notag\\
=&\sum_{m,\tra_2}\le(\NTK^{-1}\ri)_{jm}^{\tra\tra_2}\le\{z_{m;\tra_2}-\y{m}{\tra_2}-\le[\zF{m}{\tra_2}(t)-\y{m}{\tra_2}\ri]\ri\}\,\notag \, .
\end{align}
In this expression, to get to the third line we used the standard formula for evaluating geometric sums, $1+x+ x^2 + \dots + x^{t-1} = (1-x^t)/(1-x)$, and to get to the final line we evaluated the inverse and substituted in for the \terminate{step-evolution operator} \eqref{eq:unitary-operator-at-least-in-continuum-limit-with-imaginary-time}. Recall also that $\NTK^{-1}$,  defined by 
\eqref{eq:stochastic-ntk-mean}, 
is the inverse of the stochastic NTK submatrix evaluated on the training set for a particular realization of the parameters.\footnote{Unfortunately, in the absence of a Newton tensor, the raised/lowered sample indices in \eqref{eq:dynamical-helper-a} do not align well. (In fact, the problem really began in the update equation \eqref{eq:preactivation-update-redux-free} with the doubled-lowered $\tra$ index.) If you'd like, you can fix this by judicious use of identity matrices such as $\delta_{\tra_1 \tra_2}$ and $\delta^{\tra_3 \tra_4}$.
However, such usage over-clutters the presentation and so is probably not worth it, despite the additional clarity and \neo{type safety} that such index alignment provides.
}

\index{training dynamics!finite width}
In particular, for sufficiently small $\eta$, as the number of steps becomes large, $t\to\infty$, the residual error $\zF{m}{\tra_2}(t)-\y{m}{\tra_2}$ exponentially vanishes \eqref{eq:free-training-converges}.
Thus, the prediction for a general input $\delta \in \D$ \eqref{eq:free-solution-general} will exponentially converge to
\be\label{eq:free-solution-general-end-of-time}
\zF{i}{\delta}(t=\infty)=z_{i;\delta}-\sum_{j,k,\tra_1,\tra_2}\NTK_{ij;\delta\tra_1}\le(\NTK^{-1}\ri)_{jk}^{\tra_1\tra_2}\le(z_{k;\tra_2}-\y{k}{\tra_2}\ri)\, .
\ee
Although we took the limit of a large number of steps here, the exponential convergence means that for any number of steps $T$ such that $T \gtrsim (\eta \NTK)^{-1}$, the prediction error will be exponentially close to its final $T\to\infty$ value.

In fact, this solution \eqref{eq:free-solution-general-end-of-time} precisely matches the solution we  would have gotten in \S\ref{subsec:giant-plus-small} after the first update \eqref{eq:second-order-update-per-component} with the Newton tensor \eqref{eq:newton-step-generalized-first}, so long as the dNTK and ddNTKs are turned off.
This means that the \neo{algorithm dependence} that emerges at finite width is solely due to the presence of the NTK differentials and the resulting nonlinear dynamics. %

\index{training dynamics!finite width}
\subsubsection{Interacting Theory: Dynamical NTK and dNTK}\index{interactions!dynamics}
Now, let's incorporate the nonzero dNTK and ddNTKs into our analysis. To do so, we need to decompose both the dynamical network output and the dynamical NTK\index{dynamical NTK} as
\begin{align}\label{eq:free-interaction-decomposition-it-is-physics}
z_{i;\delta}(t)\equiv& \zF{i}{\delta}(t)+\zI{i}{\delta}(t)\, , \\
\NTKM_{ij ;\delta\tra}(t) \equiv& \NTK_{ij;\delta\tra} + \NTKM^{\text{I}}_{ij ;\delta\tra}(t)  \, .
\label{eq:free-interaction-decomposition-for-NTK} 
\end{align}
Here, for the network output, $\zF{i}{\delta}(t)$ is the \emph{free} part, satisfying the linear difference equation\index{difference equation!linear} \eqref{eq:preactivation-update-redux-free} with a solution given by \eqref{eq:free-solution-general}, and $\zI{i}{\delta}(t)$ is the \emph{interacting}\index{interactions!dynamics} part, encapsulating the corrections due to all the nonzero NTK differentials. Similarly, for the NTK, $\NTK_{ij;\delta\tra}$ is the step-independent or \emph{free} part, fixed at initialization, and $\NTKM^{\text{I}}_{ij ;\delta\tra}(t)$  is the dynamical step-dependent or \emph{interaction} NTK\index{interaction NTK}\index{neural tangent kernel!interaction|see{interaction NTK}}.
Since both $\zI{i}{\delta}(t)$ and $\NTKM^{\text{I}}_{ij ;\delta\tra}(t)$ are absent in the free limit,  $\dNTK,\, \ddNTK,\, \ddNTKII \to 0$, at leading order we expect them to be a linear combination of these objects, schematically
\begin{align}\label{eq:linear-combo}
z^{\text{I}}(t)  &= [\text{thing}\ 0](t)~ \dNTK + [\text{thing}\ 1](t)~ \ddNTK  + [\text{thing}\ 2](t)~ \ddNTKII  \, , \\
\NTKM^{\text{I}}(t)  &=  [\widetilde{\text{thing}}\ 0](t)~ \dNTK + [\widetilde{\text{thing}}\ 1](t)~ \ddNTK  + [\widetilde{\text{thing}}\ 2](t)~ \ddNTKII\, ,
\end{align}
where various tensorial things will have various time dependencies.
This  means in turn that any product of  $\zI{i}{\delta}(t)$ or $\NTKM^{\text{I}}_{ij ;\delta\tra}(t)$ with one of $ \dNTK, \ddNTK, \ddNTKII $ can be neglected to leading order, which is the main reason why we'll be able to systematically solve these nonlinear dynamics by perturbation theory.
Finally, the initial condition for the interacting parts of the network output and the NTK must satisfy
\begin{align}\label{eq:interacting-piece-time-initial-condition}
\zI{i}{\delta}(t=0)=0\, ,\\
\NTKM^{\text{I}}_{ij ;\delta\tra}(t=0) =0 \, ,
\label{eq:interacting-piece-NTK-time-initial-condition}
\end{align}
since the free part of the network output already satisfies $\zF{i}{\delta}(t=0)=z_{i;\delta}$ at initialization, and the free part of the NTK is step-independent and thus \emph{is} the NTK at initialization.
With all those in mind, our goal now is to find a solution for $\zI{i}{\delta}(t)$ such that 
the full network output, $z_{i;\delta}(t)$, 
the \terminate{interaction NTK}, $\NTKM^{\text{I}}_{ij ;\delta\tra}(t)$, and the dynamical dNTK\index{dynamical dNTK}\index{differential of the neural tangent kernel!dynamical|see{dynamical dNTK}}, $\tia{\dNTKM}{i_0i_1i_2}{\delta_0\delta_1\delta_2}(t)$, all together satisfy the coupled nonlinear dynamics \eqref{eq:preactivation-update-redux}, \eqref{eq:NTK-update-redux}, and \eqref{eq:dNTK-update-redux} to leading order in $1/n$.

\index{training dynamics!finite width}
First, let's work out the dynamics of the dNTK. From \eqref{eq:dNTK-update-redux} we see that the updates to the dNTK are sourced by the residual training error $z_{i;\tra}(t)-y_{i;\tra}$. Using our decomposition for the network output, \eqref{eq:free-interaction-decomposition-it-is-physics} and the initial condition,
\be
\tia{\dNTKM}{i_0i_1i_2}{\delta_0\delta_1\delta_2}(t=0) =\tia{\dNTK}{i_0i_1i_2}{\delta_0\delta_1\delta_2} \, ,
\ee
after iterating the dynamics, we have
\begin{align}\label{eq:dNTK-time-dependent-solution-semiformal}
&\tia{\dNTKM}{i_0i_1i_2}{\delta_0\delta_1\delta_2}(t) \,  \\
=&\tia{\dNTK}{i_0i_1i_2}{\delta_0\delta_1\delta_2}-\sum_{j,\tra}\le(\tia{\ddNTK}{i_0i_1 i_2j}{\delta_0\delta_1\delta_2\tra}+\tia{\ddNTKII}{i_0i_1 i_2j}{\delta_0\delta_1\delta_2\tra}+\tia{\ddNTKII}{i_0i_2 i_1j}{\delta_0\delta_2\delta_1\tra}\ri) \, \notag \\
&\qquad\qquad\times \le\{\eta\sum_{s=0}^{t-1}\le[\zF{j}{\tra}(s)+\zI{j}{\tra}(s)-\y{j}{\tra}\ri]\ri\} \, \notag \\
=&\tia{\dNTK}{i_0i_1i_2}{\delta_0\delta_1\delta_2}-\!\!\sum_{j,\tra}\!\le(\tia{\ddNTK}{i_0i_1 i_2j}{\delta_0\delta_1\delta_2\tra}\!+\!\tia{\ddNTKII}{i_0i_1 i_2j}{\delta_0\delta_1\delta_2\tra}\!+\!\tia{\ddNTKII}{i_0i_2 i_1j}{\delta_0\delta_2\delta_1\tra}\!\ri)\! a_{j;\tra}(t)\, . \notag
\end{align}
Here in the last step, we first dropped the product of $\zI{i}{\delta}(t)$ with $\ddNTK$ as explained earlier, and then substituted in our dynamical helper function \eqref{eq:dynamical-helper-a}.
Now, plugging in our final expression
from \eqref{eq:dynamical-helper-a} and neglecting the fluctuation part of the NTK inverse as subleading, we find an expression for the %
dynamical dNTK:
\begin{align}\label{eq:dNTK-time-dependent-solution}
&\tia{\dNTK}{i_0i_1i_2}{\delta_0\delta_1\delta_2}(t) \, \notag \\
=&\tia{\dNTK}{i_0i_1i_2}{\delta_0\delta_1\delta_2}-\sum_{k,\tra_1,\tra_2}\le(\tia{\ddNTK}{i_0i_1 i_2j}{\delta_0\delta_1\delta_2\tra_1}+\tia{\ddNTKII}{i_0i_1 i_2j}{\delta_0\delta_1\delta_2\tra_1}+\tia{\ddNTKII}{i_0i_2 i_1j}{\delta_0\delta_2\delta_1\tra_1}\ri) \, \notag \\
&\qquad\qquad\times\NTKMsub^{\tra_1\tra_2}\le\{z_{j;\tra_2}-\y{j}{\tra_2}-\le[\zF{j}{\tra_2}(t)-\y{j}{\tra_2}\ri]\ri\}\,  . %
\end{align}
In particular, the quantity in the curly brackets represents the difference in training errors between initialization and step $t$: the larger this difference -- i.e.~the more the residual training error decreases -- the greater the evolution of the dNTK, and the more the meta feature functions evolve, undergoing their own form of \emph{meta representation learning}\index{representation learning!meta representation learning}\index{meta representation learning|see{representation learning}} %
over the course of training. %

\index{neural tangent kernel!dynamics}\index{training dynamics!finite width}
Next, using our decompositions for the network output and NTK, \eqref{eq:free-interaction-decomposition-it-is-physics} and \eqref{eq:free-interaction-decomposition-for-NTK}, we can rewrite the NTK dynamics \eqref{eq:NTK-update-redux} as a \terminate{difference equation} for the \terminate{interaction NTK}:
\begin{align}\label{eq:interacting-NTK-dynamics-unsimplified}
&\NTKM^{\text{I}}_{i_1i_2;\delta_1\delta_2}(t+1) \, \\
=&\NTKM^{\text{I}}_{i_1i_2;\delta_1\delta_2}(t)-\eta\sum_{j,\tra}\Big(\dNTKM_{i_1 i_2j;\delta_1\delta_2\tra}(t)+\dNTKM_{i_2 i_1j;\delta_2\delta_1\tra}(t)\Big)\le[\zF{j}{\tra}(t)-\y{j}{\tra}\ri]\, \notag\\
&+\frac{\eta^2}{2}\sum_{j_1,j_2,\tra_1,\tra_2}\le(\tia{\ddNTK}{i_1i_2j_1j_2}{\delta_1\delta_2\tra_1\tra_2}+\tia{\ddNTK}{i_2i_1j_1j_2}{\delta_2\delta_1\tra_1\tra_2} + 2\tia{\ddNTKII}{i_1i_2j_1j_2}{\delta_1\delta_2\tra_1\tra_2} \ri) \, \notag\\ 
&\qquad\qquad\qquad\qquad\times\le[\zF{j_1}{\tra_1}(t)-\y{j_1}{\tra_1}\ri]\le[\zF{j_2}{\tra_2}(t)-\y{j_2}{\tra_2}\ri] \, .\notag %
\end{align}
Here, once again, we've dropped the interacting part of the network output on the right-hand side, as it is always multiplied by either the dNTK or the ddNTKs and thus will be subleading in $1/n$. Denoting the free part of the residual training error \eqref{eq:free-solution-training} as
\be\label{eq:free-residual-training-error}
\eF{j}{\tra}(t)\equiv\zF{j}{\tra}(t)-\y{j}{\tra}\, ,
\ee
for notational convenience, and substituting in our solution for the dynamical  dNTK  \eqref{eq:dNTK-time-dependent-solution},  we get
\begin{align}\label{eq:interacting-NTK-dynamics-unsimplified-substituted}
&\NTKM^{\text{I}}_{i_1i_2;\delta_1\delta_2}(t+1)-\NTKM^{\text{I}}_{i_1i_2;\delta_1\delta_2}(t) \, \\
=&-\eta\sum_{j,\tra}\Big(\dNTK_{i_1 i_2j;\delta_1\delta_2\tra}+\dNTK_{i_2 i_1j;\delta_2\delta_1\tra}\Big)\eF{j}{\tra}(t)\, \notag\\
&+ \eta\sum_{j,k,\tra,\tra_1,\tra_2}\Big(
\tia{\ddNTK}{i_1i_2 j k}{\delta_1\delta_2\tra\tra_1}+
\tia{\ddNTKII}{i_1i_2 j k}{\delta_1\delta_2\tra\tra_1}+
\tia{\ddNTKII}{i_1ji_2  k}{\delta_1\tra\delta_2\tra_1}
\, \notag \\
&\qquad\qquad\quad\ 
+\tia{\ddNTK}{i_2i_1 j k}{\delta_2\delta_1\tra\tra_1}
+\tia{\ddNTKII}{i_2i_1 j k}{\delta_2\delta_1\tra\tra_1}
+\tia{\ddNTKII}{i_2ji_1  k}{\delta_2\tra\delta_1\tra_1}
\Big) \, \notag \\
&\qquad\qquad\qquad\qquad\times\NTKMsub^{\tra_1\tra_2}\eF{j}{\tra}(t)\le[ z_{k;\tra_2}-\y{k}{\tra_2}- \eF{k}{\tra_2}(t)\ri] \notag \, \\
&+\frac{\eta^2}{2}\sum_{j_1,j_2,\tra_1,\tra_2}\le(\tia{\ddNTK}{i_1i_2j_1j_2}{\delta_1\delta_2\tra_1\tra_2}+\tia{\ddNTK}{i_2i_1j_1j_2}{\delta_2\delta_1\tra_1\tra_2} + 2\tia{\ddNTKII}{i_1i_2j_1j_2}{\delta_1\delta_2\tra_1\tra_2} \ri) \, \notag\\ 
&\qquad\qquad\qquad\qquad\times\eF{j_1}{\tra_1}(t)\, \eF{j_2}{\tra_2}(t) \notag
\, .
\end{align}
In particular, the step dependence on the right-hand side is expressed entirely in terms of the free residual training error $\eF{j}{\tra}(t)$, and each term is either 
linear or quadratic in $\eF{j}{\tra}(t)$.
Thus, in order to solve this \terminate{difference equation} and get $\NTKM^{\text{I}}_{i_1i_2;\delta_1\delta_2}(t)$, we'll just have to compute sums over these terms.

\index{training dynamics!finite width}
One of those sums -- the one that's linear in the free residual training error -- is the dynamical helper function $a_{j;\tra}(t)\equiv\eta\sum_{s=0}^{t-1}\eF{j}{\tra}(t)$ that we evaluated in \eqref{eq:dynamical-helper-a}.
The other type of sum is quadratic in the free residual training error, which will define a second dynamical helper function:
\begin{align}\label{eq:dynamical-helper-b}
 &b_{j_1j_2;\tra_1\tra_2}(t)\, \\
\equiv&\eta\sum_{s=0}^{t-1}\eF{j_1}{\tra_1}(t)\eF{j_2}{\tra_2}(t)\, \notag\\
=&\eta\!\!\sum_{k_1,k_2,\tra_3,\tra_4}\sum_{s=0}^{t-1}\!\le[ \le(\Iden-\eta\NTK\ri)^{s}\ri]_{j_1k_1;\tra_1\tra_3}\!\!\le[ \le(\Iden-\eta\NTK\ri)^{s}\ri]_{j_2k_2;\tra_2\tra_4}\!\!\le(z_{k_1;\tra_3}-\y{k_1}{\tra_3}\ri)\le(z_{k_2;\tra_4}-\y{k_2}{\tra_4}\ri)\, \notag\\
=&\sum_{\tra_3,\tra_4}\!\!\geosumtwo^{\tra_1\tra_2\tra_3\tra_4}\le[\le(z_{j_1;\tra_3}-\y{j_1}{\tra_3}\ri)\le(z_{j_2;\tra_4}-\y{j_2}{\tra_4}\ri) -\eF{j_1}{\tra_3}(t)\eF{j_2}{\tra_4}(t)\ri]+\o{\frac{1}{n}}\, .\notag
\end{align}
To evaluate this sum, we again used our expression for the free residual training error, \eqref{eq:free-solution-training}, in conjunction with the definition of the \terminate{step-evolution operator}, \eqref{eq:unitary-operator-at-least-in-continuum-limit-with-imaginary-time}. Then, we replaced the stochastic NTK $\tia{\NTK}{ij}{\tra_1\tra_2}$ of the training set by its mean $\delta_{ij}\NTKM_{\tra_1\tra_2}$ at the cost of subleading corrections, and formally performed the geometric sum as in \eqref{eq:dynamical-helper-a}. This last operation yielded an \neo{inverting tensor}
$\geosumtwo^{\tra_1\tra_2\tra_3\tra_4}$ implicitly defined by
\begin{align}
\delta^{\tra_1}_{\ \tra_5}\delta^{\tra_2}_{\ \tra_6}=&\sum_{\tra_3,\tra_4\in\A}\geosumtwo^{\tra_1\tra_2\tra_3\tra_4}\frac{1}{\eta}\le[\delta_{\tra_3\tra_5}\delta_{\tra_4\tra_6}-(\delta_{\tra_3\tra_5}-\eta\NTKMsub_{\tra_3\tra_5})(\delta_{\tra_4\tra_6}-\eta\NTKMsub_{\tra_4\tra_6})\ri]\, \notag\\
=&\sum_{\tra_3,\tra_4\in\A}\geosumtwo^{\tra_1\tra_2\tra_3\tra_4}\le(\NTKMsub_{\tra_3\tra_5}\delta_{\tra_4\tra_6}+\delta_{\tra_3\tra_5}\NTKMsub_{\tra_4\tra_6}-\eta\NTKMsub_{\tra_3\tra_5}\NTKMsub_{\tra_4\tra_6}\ri)\, .\label{eq:implicit-X-tensor}
\end{align}
This tensor is a generalization of the familiar inverting matrix $\geosumone^{\tra_1\tra_2}\equiv \NTKMsub^{\tra_1\tra_2}$ for geometric sums over matrices that satisfies
\be
\delta^{\tra_1}_{\ \tra_3}=\sum_{\tra_2\in\A}\geosumone^{\tra_1\tra_2}\frac{1}{\eta}\le[\delta_{\tra_2\tra_3}-(\delta_{\tra_2\tra_3}-\eta\NTKMsub_{\tra_2\tra_3})\ri]=\sum_{\tra_2\in\A}\geosumone^{\tra_1\tra_2}\NTKMsub_{\tra_2\tra_3}\, ,
\ee
and appeared in the last expression of the dynamical helper function $a_{j;\tra}(t)$ \eqref{eq:dynamical-helper-a}.
While we are on fire like an activated neuron, let's also define a final dynamical helper function for sums that are cubic in the free residual training error:
\begin{align}\label{eq:dynamical-helper-c}
 &c_{j_1j_2j_3;\tra_1\tra_2\tra_3}(t)\, \\
\equiv&\eta\sum_{s=0}^{t}\eF{j_1}{\tra_1}(t)\eF{j_2}{\tra_2}(t)\eF{j_3}{\tra_3}(t)\, \notag\\
=&\sum_{\tra_4,\tra_5,\tra_6}\!\!\geosumthree^{\tra_1\tra_2\tra_3\tra_4\tra_5\tra_6}\Big[ \le(z_{k_1;\tra_4}-\y{k_1}{\tra_4}\ri)\le(z_{k_2;\tra_5}-\y{k_2}{\tra_5}\ri)\le(z_{k_3;\tra_6}-\y{k_3}{\tra_6}\ri)\, \notag\\
&\quad\quad\quad\quad\quad\quad\quad\quad\quad\quad\quad\quad\quad\quad\quad\quad\quad\quad\quad-\eF{j_1}{\tra_4}(t)\,\eF{j_2}{\tra_5}(t)\,\eF{j_3}{\tra_6}(t)\Big]+\!\o{\frac{1}{n}}\, .\notag
\end{align}
In this case, the \terminate{inverting tensor} $\geosumthree^{\tra_1\tra_2\tra_3\tra_4\tra_5\tra_6}$ is implicitly defined by
\begin{align}\label{eq:implicit-X3-tensor}
&\delta^{\tra_1}_{\ \tra_7}\delta^{\tra_2}_{\ \tra_8}\delta^{\tra_3}_{\ \tra_9}\, \\
 =&\sum_{\tra_4,\tra_5,\tra_6}\geosumthree^{\tra_1\tra_2\tra_3\tra_4\tra_5\tra_6}  \bigg[ \Big(\NTKMsub_{\tra_4\tra_7}\delta_{\tra_5\tra_8}\delta_{\tra_6\tra_9} +\delta_{\tra_4\tra_7}\NTKMsub_{\tra_5\tra_8}\delta_{\tra_6\tra_9} +  \delta_{\tra_4\tra_7}\delta_{\tra_5\tra_8}\NTKMsub_{\tra_6\tra_9} \Big)\notag \, \\
&\qquad\qquad\qquad\qquad\quad-\eta\Big(\NTKMsub_{\tra_4\tra_7}\NTKMsub_{\tra_5\tra_8}\delta_{\tra_6\tra_9} + \NTKMsub_{\tra_4\tra_7}\delta_{\tra_5\tra_8}\NTKMsub_{\tra_6\tra_9} +  \delta_{\tra_4\tra_7}\NTKMsub_{\tra_5\tra_8}\NTKMsub_{\tra_6\tra_9} \Big) \notag \\
&\qquad\qquad\qquad\qquad\quad+ \eta^2 \Big( \NTKMsub_{\tra_4\tra_7}\NTKMsub_{\tra_5\tra_8}\NTKMsub_{\tra_6\tra_9} \Big) \bigg] \, .\notag
\end{align}
Essentially, these three dynamical helper functions encode the step dependence of various geometric sums of the free residual training error $\eF{j}{\tra}(t)$.
Note that we have\index{training dynamics!finite width}
\begin{align}
a_{j_1;\tra_1}(t=0) &= 0 \, ,\label{eq:a-init-vanish}\\
b_{j_1j_2;\tra_1\tra_2}(t=0) &=0 \, ,\label{eq:b-init-vanish}\\
c_{j_1j_2j_3;\tra_1\tra_2\tra_3}(t=0) &=0 \, ,\label{eq:c-init-vanish}
\end{align}
and so they all vanish at initialization, while at the end of training we have
\begin{align}
a_{j_1;\tra_1}(\infty) =&\!\sum_{\tra_2}\geosumone^{\tra_1\tra_2}\le(z_{j_1;\tra_2}-\y{j_1}{\tra_2}\ri)+\o{\frac{1}{n}} \, ,\label{eq:a-end}\\
b_{j_1j_2;\tra_1\tra_2}(\infty) =&\!\!\sum_{\tra_3,\tra_4}\geosumtwo^{\tra_1\tra_2\tra_3\tra_4}\le(z_{j_1;\tra_3}-\y{j_1}{\tra_3}\ri)\le(z_{j_2;\tra_4}-\y{j_2}{\tra_4}\ri)+\o{\frac{1}{n}} \, ,\label{eq:b-end}\\
c_{j_1j_2j_3;\tra_1\tra_2\tra_3}(\infty) =&\!\!\!\sum_{\tra_4,\tra_5,\tra_6}\geosumthree^{\tra_1\tra_2\tra_3\tra_4\tra_5\tra_6}\le(z_{j_1;\tra_4}-\y{j_1}{\tra_4}\ri)\le(z_{j_2;\tra_5}-\y{j_2}{\tra_5}\ri)\le(z_{j_3;\tra_6}-\y{j_3}{\tra_6}\ri)\,\notag\\
&+\o{\frac{1}{n}}\, ,\label{eq:c-end}
\end{align}
since the free residual training error  $\eF{j}{\tra}(t)$ vanishes exponentially quickly, cf.~\eqref{eq:free-training-converges}.

\index{training dynamics!finite width}
With the help of the first two of these dynamical helper functions, we can semi-compactly write the solution to the \terminate{difference equation} for the \terminate{interaction NTK} \eqref{eq:interacting-NTK-dynamics-unsimplified-substituted} as
\begin{align}\label{eq:interacting-NTK-solution}
&\NTKM^{\text{I}}_{i_1i_2;\delta_1\delta_2}(t) \, \\
=&-\sum_{j,\tra}\Big(\dNTK_{i_1 i_2j;\delta_1\delta_2\tra}+\dNTK_{i_2 i_1j;\delta_2\delta_1\tra}\Big)
a_{j;\tra}(t)
\, \notag\\
&+ \sum_{j,k,\tra_1,\tra_2,\tra_3}\!\!\Big(
\tia{\ddNTK}{i_1i_2 j k}{\delta_1\delta_2\tra_1\tra_2}+
\tia{\ddNTKII}{i_1i_2 j k}{\delta_1\delta_2\tra_1\tra_2}+
\tia{\ddNTKII}{i_1ji_2  k}{\delta_1\tra_1\delta_2\tra_2}
\, \notag \\
&\qquad\qquad\ \ 
+\tia{\ddNTK}{i_2i_1 j k}{\delta_2\delta_1\tra_1\tra_2}
+\tia{\ddNTKII}{i_2i_1 j k}{\delta_2\delta_1\tra_1\tra_2}
+\tia{\ddNTKII}{i_2ji_1  k}{\delta_2\tra_1\delta_1\tra_2}
\Big) \, \notag \\
&\qquad\qquad\qquad\qquad\times\NTKMsub^{\tra_2\tra_3}\Big[ 
\le( z_{k;\tra_3}-\y{k}{\tra_3} \ri) a_{j;\tra_1}(t) 
-
 b_{jk;\tra_1\tra_3}(t)
\Big]
\, \notag \\
&+\frac{\eta}{2}\!\!\sum_{j_1,j_2,\tra_1,\tra_2}\!\!\!\le(\tia{\ddNTK}{i_1i_2j_1j_2}{\delta_1\delta_2\tra_1\tra_2}+\tia{\ddNTK}{i_2i_1j_1j_2}{\delta_2\delta_1\tra_1\tra_2} + 2\tia{\ddNTKII}{i_1i_2j_1j_2}{\delta_1\delta_2\tra_1\tra_2} \ri)\! b_{j_1j_2;\tra_1\tra_2}(t) \, \notag .
\end{align}
As a quick sanity check, the vanishing initial condition for the \terminate{interaction NTK},  \eqref{eq:interacting-piece-NTK-time-initial-condition}, is satisfied due to the vanishing of the helper functions at initialization, \eqref{eq:a-init-vanish} and \eqref{eq:b-init-vanish}.
 We can also plug in \eqref{eq:a-end} and \eqref{eq:b-end} to evaluate the change in NTK at the end of training. In particular, we see that the larger the change in the NTK, the more the feature functions evolve. Thus, more initial error entails more \terminate{representation learning} %
over the course of training. 

\index{training dynamics!finite width}
Lastly, we need to determine the step dependence of the interacting part of the network output, $\zI{i}{\delta}(t)$.
Inserting our free-interacting decomposition, \eqref{eq:free-interaction-decomposition-it-is-physics}, into our dynamics for the network output, \eqref{eq:preactivation-update-redux}, and using the fact that the free part
satisfies the step-independent evolution equation \eqref{eq:gd-as-matrix-multiplication}, we can 
reorganize terms to find a dynamical equation for the interacting part only:
\be\label{eq:gd-interacting-piece}
\zI{i}{\delta}(t+1)=\zI{i}{\delta}(t)-\sum_{j,\tra}\eta\,\NTK_{ij;\delta\tra}\,\zI{j}{\tra}(t)+\eta \,\force_{i;\delta}(t)\, .
\ee
Here, we've defined
a \neo{damping force}:
\begin{align}\label{eq:damping-force-definition}
\force_{i;\delta}(t)\equiv&-\sum_{j,\tra}\NTKM^{\text{I}}_{ij ;\delta\tra}(t)\, \eF{j}{\tra}(t)+\frac{\eta}{2}\sum_{j_1,j_2,\tra_1,\tra_2}\dNTKM_{i j_1j_2;\delta\tra_1\tra_2}(t)\,\eF{j_1}{\tra_1}(t)\,\eF{j_2}{\tra_2}(t)\, \\
&-\frac{\eta^2}{6}\sum_{j_1,j_2,j_3,\tra_1,\tra_2,\tra_3}\tia{\ddNTK}{i j_1j_2j_3}{\delta\tra_1\tra_2\tra_3}\,\eF{j_1}{\tra_1}(t)\,\eF{j_2}{\tra_2}(t)\,\eF{j_3}{\tra_3}(t)\, . \notag
\end{align}
Since we have solutions for the \terminate{interaction NTK}\index{training dynamics!finite width} 
and  the dNTK 
dynamics
in terms of the free residual training error solution $\eF{j}{\tra}(t)$, \eqref{eq:interacting-NTK-solution} and \eqref{eq:dNTK-time-dependent-solution}, this damping force is an explicitly known function of the step $t$.

\index{training dynamics!finite width}
Let us now try to implicitly express the solution to this difference equation \eqref{eq:gd-interacting-piece} as a sum over steps.
First for inputs in the training set $\tra \in \A$, the dynamics of the interacting part of the output, \eqref{eq:gd-interacting-piece}, reduce to a first-order inhomogeneous linear difference equation\index{difference equation!linear!inhomogeneous}, with the \terminate{damping force} ruining the homogeneity:
\be
\zI{i}{\tra}(t+1)=\sum_{j,\tra_1}\le(\Iden_{ij;\tra\tra_1}-\eta\NTK_{ij;\tra\tra_1}\ri)\zI{j}{\tra_1}\!(t)+\eta\, \force_{i;\tra}(t)\, .
\ee
We can formally solve this equation as a convolution of the damping force with the free step-evolution operator \eqref{eq:unitary-operator-at-least-in-continuum-limit-with-imaginary-time},
\be\label{eq:interacting-solution-training-formal}
\zI{i}{\tra}(t)=\eta\sum_{s=0}^{t-1}\sum_{j,\tra_1} \Unit_{ij;\tra\tra_1}\!(t-1-s)\,\force_{j;\tra_1}\!(s)\, ,
\ee
which satisfies the initial condition $\zI{i}{\tra}(t=0)=0$. 
Plugging this result back into the dynamical equation for general inputs $\delta\in\D$, \eqref{eq:gd-interacting-piece}, we can then find a solution for the interacting part of the network output for such general inputs:
\be\label{eq:interacting-solution-general-formal}
\zI{i}{\delta}(t)=\eta\sum_{s=0}^{t-1}\le[\force_{i;\delta}(s)-\sum_{j,\tra}\NTK_{ij;\delta\tra}\, \zI{j}{\tra}(s)\ri]\, .
\ee

\index{training dynamics!finite width}
In order to further simplify these expressions, we need to work out one convoluted sum:
\begin{align}
\eta\sum_{s=0}^{t-1}\zI{j}{\tra}(s)=&\eta^2\sum_{s=0}^{t-1}\sum_{\tilde{s}=0}^{s-1}\sum_{k,\tra_1} \Unit_{jk;\tra\tra_1}\!(s-1-\tilde{s})\,\force_{k;\tra_1}\!(\tilde{s})\, \\
=&\eta \sum_{u=0}^{t-2}\sum_{k,\tra_1} \le[\eta \sum_{\tilde{u}=0}^{t-u-2}\Unit_{jk;\tra\tra_1}\!(\tilde{u})\ri]\force_{k;\tra_1}\!(u)\, \notag\\
=&\eta\sum_{u=0}^{t-2}\sum_{k,m,\tra_1,\tra_2}\le(\NTK^{-1}\ri)_{jm}^{\tra\tra_2}\le[\Iden-\Unit(t-u-1)\ri]_{mk;\tra_2\tra_1}\force_{k;\tra_1}(u)\, \notag\\
=&\sum_{m,\tra_2}\le(\NTK^{-1}\ri)_{jm}^{\tra\tra_2}\le\{\le[\eta\sum_{u=0}^{t-2}\force_{m;\tra_2}(u)\ri]+ \le[ \eta\force_{m;\tra_2}(t-1) -\zI{m}{\tra_2}(t)\ri]   \ri\}\, \notag\\
=&\sum_{m,\tra_2}\le(\NTK^{-1}\ri)_{jm}^{\tra\tra_2}\le\{\le[\eta\sum_{u=0}^{t-1}\force_{m;\tra_2}(u)\ri]-\zI{m}{\tra_2}(t)\ri\}\, ,\notag
\end{align}
Step by step, on the first line we used the expression for the formal solution \eqref{eq:interacting-solution-training-formal}; on the second line we first reversed the order of the sums as $\sum_{s=0}^{t-1} \sum_{\tilde{s}=0}^{s-1} = \sum_{\tilde{s}=0}^{t-2} \sum_{s=\tilde{s}+1}^{t-1}$, and then we rewrote these sums in terms of new variables, $u\equiv \tilde{s}$ and $\tilde{u}\equiv s-1-\tilde{s}$; on the third line, we performed the geometric sum exactly as in \eqref{eq:dynamical-helper-a};
on the fourth line, we used our formal solution \eqref{eq:interacting-solution-training-formal} at step $t$; and on the final line, we combined terms to extend the limits of the sum over the \terminate{damping force}. Plugging this evaluation back into our formal solution for $\zI{i}{\delta}(t)$,
\eqref{eq:interacting-solution-general-formal}, we get a slightly less formal solution:
\begin{align}\label{eq:interacting-solution-general}
\zI{i}{\delta}(t)=&\eta\sum_{s=0}^{t-1}\le[\force_{i;\delta}(s)-\sum_{\tra_1,\tra_2}\NTKM_{\delta\tra_1}\NTKMsub^{\tra_1\tra_2}\force_{i;\tra_2}(s)\ri]+\sum_{\tra_1,\tra_2}\NTKM_{\delta\tra_1}\NTKMsub^{\tra_1\tra_2}\zI{i}{\tra_2}(t) + \o{\frac{1}{n^2}}\, .
\end{align}
In this result, we've replaced the stochastic NTK by its mean, $\NTK_{ij;\delta_1\delta_2}\to\delta_{ij}\NTKM_{\delta_1,\delta_2}$, as every term is otherwise already proportional to the dNTK or ddNTKs. As a quick sanity check, note that for inputs in the training set, $\delta \to \tra \in \A$, this general expression reduces to an identity on $\zI{i}{\tra}(t)$.

\index{training dynamics!finite width}
Ultimately, what we care about is the
interacting
solution at the end of training, $t\to \infty$. As before, let's assume that the product of $\eta$ with the stochastic NTK is sufficiently small such that the free step-evolution operator,
\be\label{eq:step-evolution-operator-coverge}
\lim_{t\to\infty} \Unit(t) \propto \exp\!\le(- \eta \NTK t \ri) \, ,
\ee
exponentially decays to zero.\footnote{
    Note that since the \terminate{step-evolution operator} $U(t)$ is constructed in \eqref{eq:unitary-operator-at-least-in-continuum-limit-with-imaginary-time} from the \emph{step-independent} NTK\index{neural tangent kernel!step-independent}, $\NTK_{ij;\delta\tra}$, the condition for this convergence is the same as the free analysis discussed in footnote~\ref{footnote:convergence-dynamics}. %
}
Then, for training inputs, the interacting part of the network outputs, $\zI{i}{\tra}(t)$, will exponentially converge to zero
\be\label{eq:interacting-training-converge}
\lim_{t\to\infty}\zI{i}{\tra}(t) =0\, .
\ee
To see why 
this
holds, note that the dampening force \eqref{eq:damping-force-definition} decays exponentially as
\be
\lim_{s\to\infty} \force(s) \propto \exp\!\le(- \eta\NTK s \ri) \, ,
\ee
since its leading behavior is linearly proportional to the free residual training error $\eF{j}{\tra}(t)$.
Combined with \eqref{eq:step-evolution-operator-coverge}, this means that the interacting solution \eqref{eq:interacting-solution-training-formal} converges as
\begin{align}\label{eq}
\lim_{t\to\infty}\zI{i}{\tra}(t)&=\eta\lim_{t\to\infty}\sum_{s=0}^{t-1}\sum_{j,\tra_1} \Unit_{ij;\tra\tra_1}\!(t-1-s)\,\force_{j;\tra_1}\!(s) \notag \\
&\propto \lim_{ t\to\infty}\le\{  \eta t \,  \exp\!\le[- (t-1) \eta \NTK  \ri]\ri\}=0 \, ,
\end{align}
which is slightly slower than the convergence of the free solution $\zF{i}{\tra}(t) \propto \exp(-\NTK t)$.
Thus, overall the training algorithm converges:
\be\label{eq:gd-training-converge}
\lim_{t\to\infty} z_{i;\tra}(t) - \y{i}{\tra} =  \lim_{t\to\infty} \le[ \zF{i}{\tra}(t)+\zI{i}{\tra}(t)\ri]- \y{i}{\tra}  = 0\, ,
\ee
where here we also used the free solution \eqref{eq:free-training-converges}.\footnote{
    Remember that in this section we have stopped explicitly denoting that there are $\o{1/n^2}$ corrections. Recalling our fully-trained condition \eqref{eq:fully-trained-condition-really}, this result \eqref{eq:gd-training-converge} should be understood to be true up to such corrections, cf.~\eqref{eq:two-step-twice-reduced-error} for our two-step solution where the situation was analogous.
} 
Incidentally, and of possible broader interest, this altogether shows that gradient descent converges exponentially quickly to a zero-error minimum for realistic deep neural networks of finite width and nonzero depth, up to errors that are at most quadratic in our effective theory cutoff $L/n$.

\index{training dynamics!finite width}

For general inputs, the interacting part of the output, $\zI{i}{\delta}(t)$, is given by the expression \eqref{eq:interacting-solution-general}. With the convergence on the training set in mind \eqref{eq:interacting-training-converge}, the expression in the end-of-training limit reduces to
\be\label{eq:interacting-solution-in-limit}
\lim_{t\to\infty}\zI{i}{\delta}(t)=\le[\eta\sum_{s=0}^{\infty}\force_{i;\delta}(s)\ri]-\sum_{\tra_1,\tra_2}\NTKM_{\delta\tra_1}\NTKMsub^{\tra_1\tra_2}\le[\eta\sum_{s=0}^{\infty}\force_{i;\tra_2}(s)\ri]\, .
\ee
Thus, all that remains is for us to perform an infinite sum over the \terminate{damping force}:
\begin{align}\label{eq:sums-to-evaluate-to-win}
\eta\sum_{s=0}^{\infty}\force_{i;\delta}(s)\equiv&- \sum_{j,\tra_1} \Bigg[ \eta\sum_{s=0}^{\infty} \NTKM^{\text{I}}_{ij ;\delta\tra_1}(s)\,\eF{j}{\tra_1}(s) \Bigg]\, \\
&+\frac{\eta}{2} \sum_{j_1,j_2,\tra_1,\tra_2} \Bigg[\eta\sum_{s=0}^{\infty}\dNTKM_{i j_1j_2;\delta\tra_1\tra_2}(s)\,\eF{j_1}{\tra_1}(s)\,\eF{j_2}{\tra_2}(s)\Bigg]\, \notag \\
&-\frac{\eta^2}{6}\sum_{j_1,j_2,j_3,\tra_1,\tra_2,\tra_3}  \Bigg[\eta\sum_{s=0}^{\infty}\tia{\ddNTK}{i j_1j_2j_3}{\delta\tra_1\tra_2\tra_3}\,\eF{j_1}{\tra_1}(s)\,\eF{j_2}{\tra_2}(s)\, \eF{j_3}{\tra_3}(s) \Bigg]
\, . \notag
\end{align}
The third sum is exactly the end-of-training limit of the third dynamical helper function $c_{j_1j_2j_3;\tra_1\tra_2\tra_3}(t)$ that we already evaluated in \eqref{eq:c-end}, giving
\begin{align}\label{eq:ddNTK-time-dependent-solution-sum}
&\eta\sum_{s=0}^{\infty}\tia{\ddNTK}{i j_1j_2j_3}{\delta\tra_1\tra_2\tra_3}\,\eF{j_1}{\tra_1}(s)\,\eF{j_2}{\tra_2}(s) \, \eF{j_3}{\tra_3}(s)\, \\
=&\!\!\sum_{\tra_4,\tra_5,\tra_6}\!\!\!\!\!\tia{\ddNTK}{i j_1j_2j_3}{\delta\tra_1\tra_2\tra_3}\geosumthree^{\tra_1\tra_2\tra_3\tra_4\tra_5\tra_6}\le(z_{j_1;\tra_4}-\y{j_1}{\tra_4}\ri)\le(z_{j_2;\tra_5}-\y{j_2}{\tra_5}\ri)\le(z_{j_3;\tra_6}-\y{j_3}{\tra_6}\ri)\, .\notag
\end{align}
Thus, we are left with two more sums to evaluate. Let's proceed slowly: everything is simple, but there are a lot of terms to get right.

\index{training dynamics!finite width}
To start,  we can evaluate the second sum in \eqref{eq:sums-to-evaluate-to-win} as
\begin{align}\label{eq:dNTK-time-dependent-solution-sum-pre}
&\eta\sum_{s=0}^{\infty}\,\tia{\dNTK}{i j_1j_2}{\delta\tra_1\tra_2}(s)\, \eF{j_1}{\tra_1}(s)\, \eF{j_2}{\tra_2}(s)\, \\
=&\tia{\dNTK}{i j_1j_2}{\delta\tra_1\tra_2}\sum_{s=0}^{\infty} \bigg[ \eta \, \eF{j_1}{\tra_1}(s)\, \eF{j_2}{\tra_2}(s) \bigg] \, \notag \\
&-\!\!\!\sum_{k,\tra_3,\tra_4}\!\!\!\le(\tia{\ddNTK}{i j_1j_2k}{\delta\tra_1\tra_2\tra_3}+2\tia{\ddNTKII}{i j_1j_2k}{\delta\tra_1\tra_2\tra_3}\ri) \notag \, \\ 
&\qquad\qquad\times\NTKMsub^{\tra_3\tra_4}\sum_{s=0}^{\infty} \bigg\{ \eta\,\eF{j_1}{\tra_1}(s)\,\eF{j_2}{\tra_2}(s)\le[z_{k;\tra_4}-\y{k}{\tra_4}-\eF{k}{\tra_4}(s)\ri] \bigg\}\, ,\notag 
\end{align}
where we substituted in our dynamical dNTK\index{dynamical dNTK} solution, \eqref{eq:dNTK-time-dependent-solution} and used the fact that the overall expression is symmetric under $(\tra_1, j_1) \leftrightarrow (\tra_2, j_2)$ to combine the two $\ddNTKMII$ terms. Then, using our already evaluated sums,  \eqref{eq:b-end} and \eqref{eq:c-end}, we get
\begin{align}\label{eq:dNTK-time-dependent-solution-sum}
&\eta\sum_{s=0}^{\infty}\,\tia{\dNTK}{i j_1j_2}{\delta\tra_1\tra_2}(s)\, \eF{j_1}{\tra_1}(s)\, \eF{j_2}{\tra_2}(s)\, \\
=& \sum_{\tra_3,\tra_4 } \tia{\dNTK}{i j_1j_2}{\delta\tra_1\tra_2} \geosumtwo^{\tra_1\tra_2\tra_3\tra_4} \le(z_{j_1;\tra_3}-\y{j_1}{\tra_3}\ri)\le(z_{j_2;\tra_4}-\y{j_2}{\tra_4}\ri) \notag\, \\
&-\!\!\!\sum_{k,\tra_3,\ldots,\tra_6}\!\!\!\le(\tia{\ddNTK}{i j_1j_2k}{\delta\tra_1\tra_2\tra_3}+2\tia{\ddNTKII}{i j_1j_2k}{\delta\tra_1\tra_2\tra_3}
\ri) \notag \, \\ 
&\qquad\qquad\times\dampsumDNTK^{\tra_1\tra_2\tra_3\tra_4\tra_5\tra_6}\le(z_{j_1;\tra_4}-\y{j_1}{\tra_4}\ri)\le(z_{j_2;\tra_5}-\y{j_2}{\tra_5}\ri)\le(z_{k;\tra_6}-\y{k}{\tra_6}\ri) \, . \notag
\end{align}
where we introduced a shorthand
\be\label{eq:damping-force-sum-dNTK}
\dampsumDNTK^{\tra_1\tra_2\tra_3\tra_4\tra_5\tra_6} \equiv \geosumtwo^{\tra_1\tra_2\tra_4\tra_5}\NTKMsub^{\tra_3\tra_6} -\sum_{\tra_7}\NTKMsub^{\tra_3\tra_7} \geosumthree^{\tra_1\tra_2\tra_7\tra_4\tra_5\tra_6}  \, ,
\ee
to ease the collection of terms later.

\index{training dynamics!finite width}
To finish,  let us write the first sum in \eqref{eq:sums-to-evaluate-to-win} as
\begin{align}\label{eq:interacting-NTK-solution-sum-pre}
&\eta\sum_{s=0}^{\infty}\, \NTKM^{\text{I}}_{ij;\delta\tra_1}(s)\,\eF{j}{\tra_1}(s)\, \\
=&-\sum_{k,\tra_2}\Big(\dNTK_{ijk;\delta\tra_1\tra_2}+\dNTK_{ji k;\tra_1\delta\tra_2}\Big)\sum_{s=0}^{\infty}\bigg[\eta\, \eF{j}{\tra_1}(s)\, a_{k;\tra_2}(s)\bigg] \notag \, \\
&+ \sum_{k_1,k_2,\tra_2,\tra_3,\tra_4}\Big(
\tia{\ddNTK}{ij k_1 k_2}{\delta\tra_1\tra_2\tra_3}+
\tia{\ddNTKII}{ij  k_1 k_2}{\delta\tra_1\tra_2\tra_3}+
\tia{\ddNTKII}{i k_1 j  k_2}{\delta\tra_2\tra_1\tra_3}
\, \notag \\
&\qquad\qquad\quad\ \ \ +
\tia{\ddNTK}{ji  k_1 k_2}{\tra_1\delta\tra_2\tra_3}+ 
\tia{\ddNTKII}{ji  k_1 k_2}{\tra_1\delta\tra_2\tra_3}+
\tia{\ddNTKII}{j k_1 i  k_2}{\tra_1\tra_2\delta\tra_3}
\Big)\, \notag \\
&\qquad\qquad\qquad\qquad\times\NTKMsub^{\tra_3\tra_4} \sum_{s=0}^{\infty}\bigg\{ \eta\, \eF{j}{\tra_1}(s)\Big[ 
\le( z_{k_2;\tra_4}-\y{k_2}{\tra_4} \ri) a_{k_1;\tra_2}(s) 
-
 b_{k_1k_2;\tra_2\tra_4}(s)
\Big]\bigg\}
\, \notag \\
&+\frac{\eta}{2}\sum_{k_1,k_2,\tra_2,\tra_3}\!\!\!\le(\tia{\ddNTK}{ij k_1k_2}{\delta\tra_1\tra_2\tra_3}+\tia{\ddNTK}{ji k_1k_2}{\tra_1\delta\tra_2\tra_3} + 2\tia{\ddNTKII}{ij k_1k_2}{\delta\tra_1\tra_2\tra_3} \ri) \, \notag\\
&\qquad\qquad\qquad\qquad\times \sum_{s=0}^{\infty} \bigg[ \eta\, \eF{j}{\tra_1}(s)\, b_{k_1k_2;\tra_2\tra_3}(s) \bigg] \, ,\notag 
\end{align}
where we substituted in our solution for the \terminate{interaction NTK}, \eqref{eq:interacting-NTK-solution}. Then, substituting for the helper functions with \eqref{eq:dynamical-helper-a} and \eqref{eq:dynamical-helper-b}, performing the \emph{additional} sums over these terms, and then using the end-of-training limits
 \eqref{eq:a-end}--\eqref{eq:c-end}, we get
\begin{align}\label{eq:interacting-NTK-solution-sum}
&\eta\sum_{s=0}^{\infty}\, \NTKM^{\text{I}}_{ij;\delta\tra_1}(s)\eF{j}{\tra_1}(s)\, \\
=&-\sum_{k,\tra_2,\tra_3,\tra_4}\Big(\dNTK_{ijk;\delta\tra_1\tra_2}+\dNTK_{ji k;\tra_1\delta\tra_2}\Big)\dampsumNTKminus^{\tra_1\tra_2\tra_3\tra_4}
\le(z_{j;\tra_3}-\y{j}{\tra_3}\ri)\le(z_{k;\tra_4}-\y{k}{\tra_4}\ri)\, \notag\\
&+ \sum_{k_1,k_2,\tra_2,\tra_3,\tra_4,\tra_5,\tra_6}\Big(
\tia{\ddNTK}{ij k_1 k_2}{\delta\tra_1\tra_2\tra_3}+
\tia{\ddNTKII}{ij  k_1 k_2}{\delta\tra_1\tra_2\tra_3}+
\tia{\ddNTKII}{i k_1 j  k_2}{\delta\tra_2\tra_1\tra_3}
\, \notag \\
&\qquad\qquad\qquad\quad\ \ \ +
\tia{\ddNTK}{ji  k_1 k_2}{\tra_1\delta\tra_2\tra_3}+ 
\tia{\ddNTKII}{ji  k_1 k_2}{\tra_1\delta\tra_2\tra_3}+
\tia{\ddNTKII}{j k_1 i  k_2}{\tra_1\tra_2\delta\tra_3}
\Big) \, \notag \\
&\qquad\qquad\qquad\qquad\times\dampsumNTKone^{\tra_1\tra_2\tra_3\tra_4\tra_5\tra_6}\le(z_{j;\tra_4}-\y{j}{\tra_4}\ri) \le(z_{k_1;\tra_5}-\y{k_1}{\tra_5}\ri)\le(z_{k_2;\tra_6}-\y{k_2}{\tra_6}\ri) \,  \notag\\
&+\frac{\eta}{2}\sum_{k_1,k_2,\tra_2,\tra_3,\tra_4,\tra_5,\tra_6}\!\!\!\le(\tia{\ddNTK}{ij k_1k_2}{\delta\tra_1\tra_2\tra_3}+\tia{\ddNTK}{ji k_1k_2}{\tra_1\delta\tra_2\tra_3} + 2\tia{\ddNTKII}{ij k_1k_2}{\delta\tra_1\tra_2\tra_3} \ri) \, \notag\\
&\qquad\qquad\qquad\qquad\times\dampsumNTKtwo^{\tra_1\tra_2\tra_3\tra_4\tra_5\tra_6}\le(z_{j;\tra_4}-\y{j}{\tra_4}\ri)\le(z_{k_1;\tra_5}-\y{k_1}{\tra_5}\ri)\le(z_{k_2;\tra_6}-\y{k_2}{\tra_6}\ri) \, , \notag
\end{align}
where we introduced additional shorthands
\begin{align}
\dampsumNTKminus^{\tra_1\tra_2\tra_3\tra_4} \equiv &\NTKMsub^{\tra_1\tra_3}\NTKMsub^{\tra_2\tra_4} -\sum_{\tra_5}\NTKMsub^{\tra_2\tra_5} \geosumtwo^{\tra_1\tra_5\tra_3\tra_4}  \, ,\label{eq:damping-force-sum-NTK-minus}\, \\
\dampsumNTKone^{\tra_1\tra_2\tra_3\tra_4\tra_5\tra_6} \equiv& \NTKMsub^{\tra_1\tra_4} \NTKMsub^{\tra_2\tra_5}\NTKMsub^{\tra_3\tra_6} - \sum_{\tra_7} \NTKMsub^{\tra_2\tra_7} \geosumtwo^{\tra_1\tra_7\tra_4\tra_5} \NTKMsub^{\tra_3\tra_6} \,  \label{eq:damping-force-sum-NTK-one}\\ 
&-\sum_{\tra_7} \NTKMsub^{\tra_1\tra_4}\NTKMsub^{\tra_3\tra_7}\geosumtwo^{\tra_2\tra_7\tra_5\tra_6}  + \sum_{\tra_7,\tra_8,\tra_9}\NTKMsub^{\tra_3\tra_9}\geosumtwo^{\tra_2\tra_9\tra_7\tra_8}\geosumthree^{\tra_1\tra_7\tra_8\tra_4\tra_5\tra_6} \,, \notag \\
\dampsumNTKtwo^{\tra_1\tra_2\tra_3\tra_4\tra_5\tra_6} \equiv&  \NTKMsub^{\tra_1\tra_4} \geosumtwo^{\tra_2\tra_3\tra_5\tra_6} - \sum_{\tra_7,\tra_8}\geosumtwo^{\tra_2\tra_3\tra_7\tra_8}\geosumthree^{\tra_1\tra_7\tra_8\tra_4\tra_5\tra_6}\, . \label{eq:damping-force-sum-NTK-two}
\end{align}

\index{training dynamics!finite width}
Now, it's time to frantically flip through pages and collect everything we computed.
Plugging the sums \eqref{eq:interacting-NTK-solution-sum}, \eqref{eq:dNTK-time-dependent-solution-sum}, and \eqref{eq:ddNTK-time-dependent-solution-sum} back into our expression for the sum over the \terminate{damping force}, \eqref{eq:sums-to-evaluate-to-win}, plugging that back into our expression for the interaction part of the network output \eqref{eq:interacting-solution-in-limit}, and then combining with the free part of the network output, \eqref{eq:free-solution-general-end-of-time},
we obtain our fully-trained solution for finite-width networks trained by gradient descent:\index{training dynamics!finite width}
\begin{align}\label{eq:very-general-finite-width-solution}
&z_{i;\delta}(t=\infty)\, \\
\equiv&\zF{i}{\delta}(t=\infty)+\zI{i}{\delta}(t=\infty)\, \notag \\
=&z_{i;\delta}-\sum_{j,k,\tra_1,\tra_2}\NTK_{ij;\delta\tra_1}\le(\NTK^{-1}\ri)_{jk}^{\tra_1\tra_2}\le(z_{k;\tra_2}-\y{k}{\tra_2}\ri)\, \notag\\
&+\sum_{j_1,j_2,\tra_1,\tra_2,\tra_3,\tra_4}\le[\dNTK_{j_1 i j_2;\tra_1\delta\tra_2}-\sum_{\tra_5,\tra_6}\NTKM_{\delta\tra_5}\NTKMsub^{\tra_5\tra_6}\dNTK_{j_1 i j_2;\tra_1\tra_6\tra_2}\ri]\, \notag\\
&\quad\quad\quad\quad\quad\quad\quad\quad\quad\times \algodNTKone^{\tra_1\tra_2\tra_3\tra_4}\le(z_{j_1;\tra_3}-\y{j_1}{\tra_3}\ri)\le(z_{j_2;\tra_4}-\y{j_2}{\tra_4}\ri)\, \notag \\
&+\sum_{j_1,j_2,\tra_1,\tra_2,\tra_3,\tra_4}\le[\dNTK_{i j_1j_2;\delta\tra_1\tra_2}-\sum_{\tra_5,\tra_6}\NTKM_{\delta\tra_5}\NTKMsub^{\tra_5\tra_6}\dNTK_{i j_1j_2;\tra_6\tra_1\tra_2}\ri]\, \notag\\
&\quad\quad\quad\quad\quad\quad\quad\quad\quad\times \algodNTKtwo^{\tra_1\tra_2\tra_3\tra_4}\le(z_{j_1;\tra_3}-\y{j_1}{\tra_3}\ri)\le(z_{j_2;\tra_4}-\y{j_2}{\tra_4}\ri)\, \notag\\
&+\sum_{\substack{ j_1,j_2,j_3, \\ \tra_1,\tra_2,\tra_3,\tra_4,\tra_5,\tra_6} }\!\! \le[ \tia{\ddNTK}{j_1i j_2j_3}{\tra_1\delta\tra_2\tra_3}-\sum_{\tra_7,\tra_8}\NTKM_{\delta\tra_7}\NTKMsub^{\tra_7\tra_8}\tia{\ddNTK}{j_1ij_2j_3}{\tra_1\tra_8\tra_2\tra_3}\ri] 
\notag \, \\
&\quad\quad\quad\quad\quad\quad\quad\quad\quad\times \algoddNTKIone^{\tra_1\tra_2\tra_3\tra_4\tra_5\tra_6}\le(z_{j_1;\tra_4}-\y{j_1}{\tra_4}\ri)\le(z_{j_2;\tra_5}-\y{j_2}{\tra_5}\ri)\le(z_{j_3;\tra_6}-\y{j_3}{\tra_6}\ri)\, \notag \\
&+\sum_{\substack{ j_1,j_2,j_3, \\ \tra_1,\tra_2,\tra_3,\tra_4,\tra_5,\tra_6} }\!\! \le[ \tia{\ddNTK}{i j_1j_2j_3}{\delta\tra_1\tra_2\tra_3}-\sum_{\tra_7,\tra_8}\NTKM_{\delta\tra_7}\NTKMsub^{\tra_7\tra_8}\tia{\ddNTK}{i j_1j_2j_3}{\tra_8\tra_1\tra_2\tra_3}\ri] 
\notag \, \\
&\quad\quad\quad\quad\quad\quad\quad\quad\quad\times \algoddNTKItwo^{\tra_1\tra_2\tra_3\tra_4\tra_5\tra_6}\le(z_{j_1;\tra_4}-\y{j_1}{\tra_4}\ri)\le(z_{j_2;\tra_5}-\y{j_2}{\tra_5}\ri)\le(z_{j_3;\tra_6}-\y{j_3}{\tra_6}\ri)\, \notag \\
&+\sum_{\substack{ j_1,j_2,j_3, \\ \tra_1,\tra_2,\tra_3,\tra_4,\tra_5,\tra_6} }\!\! \le[ \tia{\ddNTKII}{j_1 j_2ij_3}{\tra_1\tra_2\delta\tra_3}-\sum_{\tra_7,\tra_8}\NTKM_{\delta\tra_7}\NTKMsub^{\tra_7\tra_8}\tia{\ddNTKII}{j_1j_2ij_3}{\tra_1\tra_2\tra_8\tra_3}\ri] 
\notag \, \\
&\quad\quad\quad\quad\quad\quad\quad\quad\quad\times \algoddNTKIIone^{\tra_1\tra_2\tra_3\tra_4\tra_5\tra_6}\le(z_{j_1;\tra_4}-\y{j_1}{\tra_4}\ri)\le(z_{j_2;\tra_5}-\y{j_2}{\tra_5}\ri)\le(z_{j_3;\tra_6}-\y{j_3}{\tra_6}\ri)\, \notag \\
&+\sum_{\substack{ j_1,j_2,j_3, \\ \tra_1,\tra_2,\tra_3,\tra_4,\tra_5,\tra_6} }\!\! \le[ \tia{\ddNTKII}{i j_1j_2j_3}{\delta\tra_1\tra_2\tra_3}-\sum_{\tra_7,\tra_8}\NTKM_{\delta\tra_7}\NTKMsub^{\tra_7\tra_8}\tia{\ddNTKII}{i j_1j_2j_3}{\tra_8\tra_1\tra_2\tra_3}\ri] 
\notag \, \\
&\quad\quad\quad\quad\quad\quad\quad\quad\quad\times \algoddNTKIItwo^{\tra_1\tra_2\tra_3\tra_4\tra_5\tra_6}\le(z_{j_1;\tra_4}-\y{j_1}{\tra_4}\ri)\le(z_{j_2;\tra_5}-\y{j_2}{\tra_5}\ri)\le(z_{j_3;\tra_6}-\y{j_3}{\tra_6}\ri)\, \notag \\
&+\o{\frac{1}{n^2}} \, \notag .
\end{align}
Here we defined our final tensors with our final alphabet letter (and various
subscripts),
\begin{align}
\algodNTKone^{\tra_1\tra_2\tra_3\tra_4}\equiv&\dampsumNTKminus^{\tra_1\tra_2\tra_3\tra_4}\, ,\label{eq:implicit-ZA-tensor} \\
\algodNTKtwo^{\tra_1\tra_2\tra_3\tra_4}\equiv&
\dampsumNTKminus^{\tra_1\tra_2\tra_3\tra_4}
+\frac{\eta}{2}\geosumtwo^{\tra_1\tra_2\tra_3\tra_4}\, ,\label{eq:implicit-ZB-tensor}\\
\algoddNTKIone^{\tra_1\tra_2\tra_3\tra_4\tra_5\tra_6}\equiv&-\dampsumNTKone^{\tra_1\tra_2\tra_3\tra_4\tra_5\tra_6} -\frac{\eta}{2} \dampsumNTKtwo^{\tra_1\tra_2\tra_3\tra_4\tra_5\tra_6}
\, ,\label{eq:implicit-ZA-tensor-I}\\
\algoddNTKItwo^{\tra_1\tra_2\tra_3\tra_4\tra_5\tra_6}\equiv&-\dampsumNTKone^{\tra_1\tra_2\tra_3\tra_4\tra_5\tra_6}- \frac{\eta}{2}\dampsumNTKtwo^{\tra_1\tra_2\tra_3\tra_4\tra_5\tra_6}\, \notag\\
&- \frac{\eta}{2}\dampsumDNTK^{\tra_1\tra_2\tra_3\tra_4\tra_5\tra_6} - \frac{\eta^2}{6}\geosumthree^{\tra_1\tra_2\tra_3\tra_4\tra_5\tra_6}
\, ,\label{eq:implicit-ZB-tensor-I}\\
\algoddNTKIIone^{\tra_1\tra_2\tra_3\tra_4\tra_5\tra_6}\equiv&-\dampsumNTKone^{\tra_1\tra_2\tra_3\tra_4\tra_5\tra_6}
\, ,\label{eq:implicit-ZA-tensor-II}\\
\algoddNTKIItwo^{\tra_1\tra_2\tra_3\tra_4\tra_5\tra_6}\equiv&-\dampsumNTKone^{\tra_1\tra_2\tra_3\tra_4\tra_5\tra_6}-\dampsumNTKone^{\tra_2\tra_1\tra_3\tra_5\tra_4\tra_6}-\dampsumNTKone^{\tra_1\tra_3\tra_2\tra_4\tra_6\tra_5}\, \notag \\
&-\eta\dampsumNTKtwo^{\tra_1\tra_2\tra_3\tra_4\tra_5\tra_6} - \eta \dampsumDNTK^{\tra_1\tra_2\tra_3\tra_4\tra_5\tra_6}
\,, \label{eq:implicit-ZB-tensor-II}
\end{align}
making use of our previous\index{training dynamics!finite width} shorthand tensors \eqref{eq:damping-force-sum-dNTK} and \eqref{eq:damping-force-sum-NTK-minus}--\eqref{eq:damping-force-sum-NTK-two}.\footnote{Note that for the last of these tensors, \eqref{eq:implicit-ZB-tensor-II}, in order to coax the various contributions in our solution \eqref{eq:very-general-finite-width-solution} into the proper form, we used the symmetry of $\tia{\ddNTKII}{i j_1j_2j_3}{\delta\tra_1\tra_2\tra_3}$ and relabeled various dummy sample indices.}
These \textbf{algorithm projectors}\index{algorithm projector|textbf}\index{training dynamics!finite width}, \eqref{eq:implicit-ZA-tensor}--\eqref{eq:implicit-ZB-tensor-II}, serve to project the initial training error onto two different combinations of the dNTK, two different combinations of the first ddNTK, and two other different combinations of the second ddNTK, all according to the details of the gradient descent algorithm. As a final sanity check, note that as for inputs in the training set, $\delta \to \tra \in \A$, the quantities in the square brackets in the finite-width solution  \eqref{eq:very-general-finite-width-solution} each vanish, and we recover our fully-trained condition \eqref{eq:fully-trained-condition-really}.

Before we retire this subsection, let us elaborate on the algorithm dependence. First, note that our two-update solution \eqref{eq:finite-width-network-output-general-data} has the same form as the gradient-descent solution \eqref{eq:very-general-finite-width-solution}, but with different algorithm projectors:
\begin{align}\label{eq:ZB-two-step}
\algodNTKtwo^{\tra_1\tra_2\tra_3\tra_4}\equiv \frac{1}{2}\NTKMsub^{\tra_1\tra_3}\NTKMsub^{\tra_2\tra_4}\, ,\qquad \algoddNTKItwo^{\tra_1\tra_2\tra_3\tra_4\tra_5\tra_6}\equiv-\frac{1}{6}\NTKMsub^{\tra_1\tra_4}\NTKMsub^{\tra_2\tra_5}\NTKMsub^{\tra_3\tra_6} \, , 
\end{align}
and all the others vanishing.\index{training dynamics!finite width}
Clearly these algorithms have very different inductive biases\index{inductive bias!of learning algorithms}!
Second, we can study the ODE limit\index{gradient descent!continuum or ODE limit} of the dynamics by taking $\eta\to0$, cf.~footnote~\ref{eq:footnote-continuum-limit}: in this case,  we see that the ODE dynamics have a solution given by
\begin{align}\label{eq:ODE-YZ}
&\algodNTKone^{\tra_1\tra_2\tra_3\tra_4}=\algodNTKtwo^{\tra_1\tra_2\tra_3\tra_4}\equiv\dampsumNTKminus^{\tra_1\tra_2\tra_3\tra_4}\, , \\
\label{eq:ODE-YZ-ddNTK-1}
&\algoddNTKIone^{\tra_1\tra_2\tra_3\tra_4\tra_5\tra_6} = \algoddNTKItwo^{\tra_1\tra_2\tra_3\tra_4\tra_5\tra_6}= \algoddNTKIIone^{\tra_1\tra_2\tra_3\tra_4\tra_5\tra_6} \equiv -\dampsumNTKone^{\tra_1\tra_2\tra_3\tra_4\tra_5\tra_6} \, \\
&\algoddNTKIItwo^{\tra_1\tra_2\tra_3\tra_4\tra_5\tra_6}\equiv-\dampsumNTKone^{\tra_1\tra_2\tra_3\tra_4\tra_5\tra_6}-\dampsumNTKone^{\tra_2\tra_1\tra_3\tra_5\tra_4\tra_6}-\dampsumNTKone^{\tra_1\tra_3\tra_2\tra_4\tra_6\tra_5}
 \, ,\label{eq:ODE-YZ-ddNTK-II-2}
\end{align}
where $\dampsumNTKminus^{\tra_1\tra_2\tra_3\tra_4} $ and $\dampsumNTKone^{\tra_1\tra_2\tra_3\tra_4\tra_5\tra_6}$ are given by \eqref{eq:damping-force-sum-NTK-minus} and \eqref{eq:damping-force-sum-NTK-one}, respectively, and the inverting tensor $\geosumtwo^{\tra_1\tra_2\tra_3\tra_4}$, \eqref{eq:implicit-X-tensor}, now satisfies
\be\label{eq:ODE-X}
\sum_{\tra_3,\tra_4\in\A}\geosumtwo^{\tra_1\tra_2\tra_3\tra_4}\le(\NTKMsub_{\tra_3\tra_5}\delta_{\tra_4\tra_6}+\delta_{\tra_3\tra_5}\NTKMsub_{\tra_4\tra_6}\ri)=\delta^{\tra_1}_{\ \tra_5}\delta^{\tra_2}_{\ \tra_6}\, ,
\ee
and the other inverting tensor $\geosumtwo^{\tra_1\tra_2\tra_3\tra_4\tra_5\tra_6}$, \eqref{eq:implicit-X3-tensor}, now satisfies\index{training dynamics!finite width}
\begin{align}\label{eq:ODE-X-III}
&\delta^{\tra_1}_{\ \tra_7}\delta^{\tra_2}_{\ \tra_8}\delta^{\tra_3}_{\ \tra_9}\,  \\
 =&\sum_{\tra_4,\tra_5,\tra_6}\geosumthree^{\tra_1\tra_2\tra_3\tra_4\tra_5\tra_6}  \Big(\NTKMsub_{\tra_4\tra_7}\delta_{\tra_5\tra_8}\delta_{\tra_6\tra_9} +\delta_{\tra_4\tra_7}\NTKMsub_{\tra_5\tra_8}\delta_{\tra_6\tra_9} +  \delta_{\tra_4\tra_7}\delta_{\tra_5\tra_8}\NTKMsub_{\tra_6\tra_9} \Big)\notag \,  \, . 
\end{align}
This entirely captures the difference between gradient flow and gradient descent for these fully-trained networks.
In general, we conjecture that for finite-width networks, at leading order the fully-trained solution takes the universal\index{universality!of the fully-trained network solution} form of \eqref{eq:very-general-finite-width-solution}, with all of the \neo{algorithm dependence} encoded by the  six \emph{algorithm projectors}\index{algorithm projector}: $\algodNTKone^{\tra_1\tra_2\tra_3\tra_4}$, $\algodNTKtwo^{\tra_1\tra_2\tra_3\tra_4}$, $\algoddNTKIone^{\tra_1\tra_2\tra_3\tra_4\tra_5\tra_6}$, $\algoddNTKItwo^{\tra_1\tra_2\tra_3\tra_4\tra_5\tra_6}$, $\algoddNTKIIone^{\tra_1\tra_2\tra_3\tra_4\tra_5\tra_6}$, and $\algoddNTKIItwo^{\tra_1\tra_2\tra_3\tra_4\tra_5\tra_6}$.\footnote{
More precisely, we conjecture that our conjecture, as stated, holds for the MSE loss\index{loss!algorithm dependence at finite width}. For the cross-entropy loss\index{loss!algorithm dependence at finite width} the solution will take a slightly different form, but with a similar partition into an algorithm-independent part and an algorithm-dependent part described by similar algorithm projectors.
}
\index{training dynamics!finite width}

Finally, let's understand what is meant by the remaining error in our solution \eqref{eq:very-general-finite-width-solution}. It is not really the error of an actual network that is instantiated and then fully trained through many many gradient-descent steps, but instead it is the error in our effective description of such a particular fully-trained network. Of course, our \terminate{effective theory} formalism can compute higher-order corrections, if they're of interest. However, the leading-order finite-width corrections should really be sufficient for most cases: for instance, for a network of depth $L=10$ layers and of hidden-layer width $n=100$ neurons each, our effective description will only be off by $\sim(L/n)^2= 1\%$.

\index{training dynamics!finite width}
Furthermore, for any theoretical analysis, the main qualitative difference in the solution appears when going from infinite width to finite width, as we go from a free theory to an interacting theory and from linear dynamics to nonlinear dynamics. Thus, the effective theory that gave us the solution \eqref{eq:very-general-finite-width-solution} really is ``as simple \ldots as possible'' while still providing an extremely accurate description of real deep learning models.

\subsection{Prediction at Finite Width}\label{subsec:prediction-at-finite-width}

Having solved the training dynamics in two different ways, we can now rather generally study the predictions of our networks on novel inputs $x_{\tea}$ from the test set $\tea \in \B$. 

\index{finite-width prediction!gradient descent}\index{finite-width prediction!gradient descent|seealso{T-shirt equation}}\index{universality!of the fully-trained network solution}
At finite width, the predictions of a fully-trained network are universally governed by the stochastic equation
\begin{align}\label{eq:very-general-finite-width-solution-DONT-CHANGE}
&z_{i;\tea}(t=T)\, \\
=&z_{i;\tea}-\sum_{\tra_1,\tra_2}\NTKM_{\tea\tra_1}\NTKMsub^{\tra_1\tra_2}\!\le(z_{i;\tra_2}-\y{i}{\tra_2}\ri)\,  \notag\\
&+\sum_{j, \tra_1,\tra_2}\le[\DNTKS_{ij;\tea\tra_1}-\sum_{\tra_3,\tra_4\in\A}\NTKM_{\tea\tra_3}\NTKMsub^{\tra_3\tra_4}\DNTKS_{ij;\tra_4\tra_1}\ri]\NTKMsub^{\tra_1\tra_2}\!\le(z_{j;\tra_2}-\y{j}{\tra_2}\ri)\, \notag\\
&-\sum_{\substack{j,k\\ \tra_1,\ldots,\tra_4}}\le[\DNTKS_{ij;\tea\tra_1}-\sum_{\tra_5,\tra_6}\NTKM_{\tea\tra_5}\NTKMsub^{\tra_5\tra_6}\DNTKS_{ij;\tra_6\tra_1}\ri]\NTKMsub^{\tra_1\tra_2}\DNTKS_{jk;\tra_2\tra_3}\NTKMsub^{\tra_3\tra_4}\!\le(z_{k;\tra_4}-\y{k}{\tra_4}\ri)\, \notag\\
&+\sum_{j_1,j_2,\tra_1,\tra_2,\tra_3,\tra_4}\le[\dNTK_{j_1 i j_2;\tra_1\tea\tra_2}-\sum_{\tra_5,\tra_6}\NTKM_{\tea\tra_5}\NTKMsub^{\tra_5\tra_6}\dNTK_{j_1 i j_2;\tra_1\tra_6\tra_2}\ri]\, \notag\\
&\quad\quad\quad\quad\quad\quad\quad\quad\quad\quad\times \algodNTKone^{\tra_1\tra_2\tra_3\tra_4}\le(z_{j_1;\tra_3}-\y{j_1}{\tra_3}\ri)\le(z_{j_2;\tra_4}-\y{j_2}{\tra_4}\ri)\, \notag \\
&+\sum_{j_1,j_2,\tra_1,\tra_2,\tra_3,\tra_4}\le[\dNTK_{i j_1j_2;\tea\tra_1\tra_2}-\sum_{\tra_5,\tra_6}\NTKM_{\tea\tra_5}\NTKMsub^{\tra_5\tra_6}\dNTK_{i j_1j_2;\tra_6\tra_1\tra_2}\ri]\, \notag\\
&\quad\quad\quad\quad\quad\quad\quad\quad\quad\quad\times \algodNTKtwo^{\tra_1\tra_2\tra_3\tra_4}\le(z_{j_1;\tra_3}-\y{j_1}{\tra_3}\ri)\le(z_{j_2;\tra_4}-\y{j_2}{\tra_4}\ri)\, \notag\\
&+\sum_{\substack{ j_1,j_2,j_3, \\ \tra_1,\tra_2,\tra_3,\tra_4,\tra_5,\tra_6} }\!\! \le[ \tia{\ddNTK}{j_1i j_2j_3}{\tra_1\tea\tra_2\tra_3}-\sum_{\tra_7,\tra_8}\NTKM_{\tea\tra_7}\NTKMsub^{\tra_7\tra_8}\tia{\ddNTK}{j_1ij_2j_3}{\tra_1\tra_8\tra_2\tra_3}\ri] 
\notag \, \\
&\quad\quad\quad\quad\quad\quad\quad\quad\quad\quad\times \algoddNTKIone^{\tra_1\tra_2\tra_3\tra_4\tra_5\tra_6}\le(z_{j_1;\tra_4}-\y{j_1}{\tra_4}\ri)\le(z_{j_2;\tra_5}-\y{j_2}{\tra_5}\ri)\le(z_{j_3;\tra_6}-\y{j_3}{\tra_6}\ri)\, \notag \\
&+\sum_{\substack{ j_1,j_2,j_3, \\ \tra_1,\tra_2,\tra_3,\tra_4,\tra_5,\tra_6} }\!\! \le[ \tia{\ddNTK}{i j_1j_2j_3}{\tea\tra_1\tra_2\tra_3}-\sum_{\tra_7,\tra_8}\NTKM_{\tea\tra_7}\NTKMsub^{\tra_7\tra_8}\tia{\ddNTK}{i j_1j_2j_3}{\tra_8\tra_1\tra_2\tra_3}\ri] 
\notag \, \\
&\quad\quad\quad\quad\quad\quad\quad\quad\quad\quad\times \algoddNTKItwo^{\tra_1\tra_2\tra_3\tra_4\tra_5\tra_6}\le(z_{j_1;\tra_4}-\y{j_1}{\tra_4}\ri)\le(z_{j_2;\tra_5}-\y{j_2}{\tra_5}\ri)\le(z_{j_3;\tra_6}-\y{j_3}{\tra_6}\ri)\, \notag \\
&+\sum_{\substack{ j_1,j_2,j_3, \\ \tra_1,\tra_2,\tra_3,\tra_4,\tra_5,\tra_6} }\!\! \le[ \tia{\ddNTKII}{j_1 j_2ij_3}{\tra_1\tra_2\tea\tra_3}-\sum_{\tra_7,\tra_8}\NTKM_{\tea\tra_7}\NTKMsub^{\tra_7\tra_8}\tia{\ddNTKII}{j_1j_2ij_3}{\tra_1\tra_2\tra_8\tra_3}\ri] 
\notag \, \\
&\quad\quad\quad\quad\quad\quad\quad\quad\quad\quad\times \algoddNTKIIone^{\tra_1\tra_2\tra_3\tra_4\tra_5\tra_6}\le(z_{j_1;\tra_4}-\y{j_1}{\tra_4}\ri)\le(z_{j_2;\tra_5}-\y{j_2}{\tra_5}\ri)\le(z_{j_3;\tra_6}-\y{j_3}{\tra_6}\ri)\, \notag \\
&+\sum_{\substack{ j_1,j_2,j_3, \\ \tra_1,\tra_2,\tra_3,\tra_4,\tra_5,\tra_6} }\!\! \le[ \tia{\ddNTKII}{i j_1j_2j_3}{\tea\tra_1\tra_2\tra_3}-\sum_{\tra_7,\tra_8}\NTKM_{\tea\tra_7}\NTKMsub^{\tra_7\tra_8}\tia{\ddNTKII}{i j_1j_2j_3}{\tra_8\tra_1\tra_2\tra_3}\ri] 
\notag \, \\
&\quad\quad\quad\quad\quad\quad\quad\quad\quad\quad\times \algoddNTKIItwo^{\tra_1\tra_2\tra_3\tra_4\tra_5\tra_6}\le(z_{j_1;\tra_4}-\y{j_1}{\tra_4}\ri)\le(z_{j_2;\tra_5}-\y{j_2}{\tra_5}\ri)\le(z_{j_3;\tra_6}-\y{j_3}{\tra_6}\ri)\, \notag \\
&+\o{\frac{1}{n^2}} \, \notag .\end{align}
This formula could boarder-line be fit on a T-shirt\index{T-shirt equation}.\footnote{
To better facilitate such \terminate{brand awareness}, first recall that for \terminate{nearly-kernel methods} 
we were able to compress the model predictions in terms of a \emph{trained kernel} \eqref{eqtrained-kernel-prediction}\index{nearly-kernel methods!trained kernel}; similarly, we can compress the predictions of finite-width networks \eqref{eq:very-general-finite-width-solution-DONT-CHANGE} in terms of a \textbf{trained NTK}\index{trained NTK|see{neural tangent kernel}}\index{neural tangent kernel!trained|textbf}\index{neural tangent kernel!trained|seealso{trained kernel}}, $\NTKMA_{ij;\delta \tra}$, as
\begin{align}\label{eq:trained-NTK-prediction}
z_{i;\tea}(t=T) =z_{i;\tea} - \sum_{j,k,\tra_1,\tra_2}\NTKMA_{ij;\tea\tra_1}\NTKMAsub_{jk}^{\tra_1\tra_2}\le(z_{k;\tra_2}-\y{k}{\tra_2}\ri)+\o{\frac{1}{n^2}}\, ,
\end{align}
taking the form of a \emph{(neural tangent) kernel prediction}\index{kernel methods!prediction} \eqref{eq:kernel-prediction}.
To see how this works,
let us decompose the trained NTK into free and training-dependent terms:
\be\label{eq:trained-NTK}
\NTKMA_{ij;\delta \tra} \equiv \NTK_{ij;\delta \tra} + \NTKdd_{ij;\delta \tra}\, .
\ee
Please don't confuse this decomposition with our earlier decomposition \eqref{eq:free-interaction-decomposition-for-NTK}: that former one was convenient for solving the training dynamics, while this new one is useful for determining the trained NTK in \eqref{eq:trained-NTK-prediction}.
Considering the inverse of the trained NTK restricted to the training set only,
\be\label{eq:trained-NTK-inverse}
\NTKMAsub_{jk}^{\tra_1\tra_2} \equiv \le(\NTK^{-1}\ri)_{jk}^{\tra_1\tra_2} - \sum_{\tra_3, \tra_4} \NTKdd_{jk;\tra_3 \tra_4}\NTKMsub^{\tra_1\tra_3}\NTKMsub^{\tra_2\tra_4} + \o{\frac{1}{n^2}} \,,
\ee
and plugging it along with the decomposition \eqref{eq:trained-NTK}
into the formula \eqref{eq:trained-NTK-prediction}, we get
\begin{align}\label{eq:trained-NTK-prediction-expanded}
z_{i;\tea}(t=T) =&z_{i;\tea}-\sum_{j,k,\tra_1,\tra_2}\NTK_{ij;\tea\tra_1}\le(\NTK^{-1}\ri)_{jk}^{\tra_1\tra_2}\le(z_{k;\tra_2}-\y{k}{\tra_2}\ri)\, \\
&-\sum_{j,\tra_1,\tra_2}\le[\NTKdd_{ij;\tea \tra_1}-\sum_{\tra_3,\tra_4}\NTKM_{\tea\tra_3}\NTKMsub^{\tra_3\tra_4}\NTKdd_{ij;\tra_4 \tra_1}\ri]\NTKMsub^{\tra_1\tra_2}(z_{j;\tra_2}-\y{j}{\tra_2}) + \o{\frac{1}{n^2}}\, .\notag
\end{align}
The terms on the first line of the right-hand side of this expression give the free
contribution to the solution,
\eqref{eq:free-solution-general-end-of-time}, while the terms on the second line give the interacting contribution, \eqref{eq:interacting-solution-in-limit}, encapsulating the effect of nontrivial representation learning at finite width. 
The specific form of $\NTKMA_{ij;\delta \tra}$ can be found by matching the terms on the right-hand sides of \eqref{eq:trained-NTK-prediction-expanded} and \eqref{eq:very-general-finite-width-solution-DONT-CHANGE}. You can also express the training-dependent part, $\NTKdd_{ij;\delta \tra}$, implicitly in terms of the \terminate{damping force} \eqref{eq:damping-force-definition} as
\be\label{eq:ntk-data-dependent-in-terms-of-force}
\eta\sum_{s=0}^{\infty}\force_{i;\delta}(s)= -\sum_{j, \tra_1, \tra_2} \NTKdd_{ij;\delta \tra_1} \, \NTKMsub^{\tra_1\tra_2}\le(z_{j;\tra_2}-\y{j}{\tra_2}\ri) \,,
\ee
as is clear by comparing \eqref{eq:trained-NTK-prediction-expanded} with \eqref{eq:interacting-solution-in-limit}. This implicit expression also makes it clear that the form of $\NTKdd_{ij;\delta \tra}$ isn't unique and can always be adjusted by the addition of a term orthogonal to $\sum_{\tra_2}\NTKMsub^{\tra_1\tra_2}\le(z_{j;\tra_2}-\y{j}{\tra_2}\ri)$. In any event, with the trained NTK you can now fit the finite-width prediction formula \eqref{eq:trained-NTK-prediction} on any newborn AI's onesie.
}
For this expression, we've expanded the complete inverse of the stochastic NTK sub-tensor as per \eqref{eq:newton-step-generalized-first}; in particular, the second line is more or less the infinite-width kernel prediction \eqref{eq:kernel-prediction}, and the terms on the third and fourth lines are finite-width corrections due to NTK fluctuations. 
Further, the algorithm projectors\index{algorithm projector} \eqref{eq:implicit-ZA-tensor}--\eqref{eq:implicit-ZB-tensor-II}
contain all the finite-width \terminate{algorithm dependence} of the solution, and thus this general solution \eqref{eq:very-general-finite-width-solution-DONT-CHANGE} can describe both our explicit solutions,
whether we train in two steps \eqref{eq:finite-width-network-output-general-data}, in many many steps \eqref{eq:very-general-finite-width-solution}, or with any other choice of optimization algorithm that uses the MSE loss.

Furthermore, this solution \eqref{eq:very-general-finite-width-solution-DONT-CHANGE} describes the predictions of a \emph{particular} network from our ensemble:
 the instantiation-to-instantiation difference is encoded in the particular initial network output,
 $z$, the NTK fluctuation, $\DNTKS$, the dNTK, $\dNTK$, the first ddNTK, $\ddNTK$, and the second ddNTK, $\ddNTKII$.
Since we know explicitly the statistics of these variables at initialization, we can also analyze the statistics of the fully-trained distribution in full.
With an eye towards a discussion of the depth dependence of these statistics, we'll now revive the layer indices.

First and foremost, the mean prediction is given by
\begin{align}\label{eq:very-general-finite-width-solution-mean-2}
m_{i;\tea}\equiv&\E{z^{(L)}_{i;\tea}(T)}\, \\
=&m_{i;\tea}^{\text{NTK}}+\frac{1}{n_{L-1}}\le(m_{i;\tea}^{\Delta\text{NTK}}+m_{i;\tea}^{\text{dNTK}}+m_{i;\tea}^{\text{ddNTK-I}}+m_{i;\tea}^{\text{ddNTK-II}}\ri)\, \notag\\
&-\frac{1}{n_{L-1}}\sum_{\tra_1,\tra_2}\NTKM^{(L)}_{\tea\tra_1}\NTKMsub_{(L)}^{\tra_1\tra_2}\le(m_{i;\tra_2}^{\Delta\text{NTK}}+m_{i;\tra_2}^{\text{dNTK}}+m_{i;\tra_2}^{\text{ddNTK-I}}+m_{i;\tra_2}^{\text{ddNTK-II}}\ri)+\o{\frac{1}{n^2}}\, ,\notag
\end{align}
where the first term is the (neural tangent) kernel prediction
\be\label{eq:infinite-width-ish-contribution-to-the-bias}
m_{i;\tea}^{\text{NTK}}\equiv\sum_{\tra_1,\tra_2}\NTKM^{(L)}_{\tea\tra_1}\NTKMsub_{(L)}^{\tra_1\tra_2}\y{i}{\tra_2}\, ,
\ee
and the four other kinds of terms come from the leading-order finite-width correction. Specifically, \emph{(i)} the fluctuation of the NTK gives
\be\label{eq:mean-prediction-DNTK-contribution}
m_{i;\delta}^{\Delta\text{NTK}}\equiv\sum_{\tra_1,\ldots,\tra_4}\le(\NTHA{\delta\tra_1}{\tra_2\tra_3}{L}+\NTHB{\delta\tra_2\tra_1\tra_3}{L}+n_{L}\NTHB{\delta\tra_3\tra_1\tra_2}{L}\ri)\NTKMsub_{(L)}^{\tra_1\tra_2}\NTKMsub_{(L)}^{\tra_3\tra_4} \y{i}{\tra_4}\, \,
\ee
where we used \eqref{eq:NTH-variance-decomposition} to evaluate the NTK variance in terms of our decomposition\index{tensor decomposition!NTK variance $A$/$B$} into tensors $A^{(L)}$ and $B^{(L)}$; \emph{(ii)} the dNTK gives
\begin{align}\label{eq:mean-prediction-dNTK-contribution}
m_{i;\delta}^{\text{dNTK}}\equiv&-\sum_{\tra_1,\ldots,\tra_4}\Bigg[2\le(\dNTKP{\delta\tra_1\tra_2\tra_3}{L}+\dNTKQ{\delta\tra_1\tra_2\tra_3}{L}+n_{L}\dNTKQ{\delta\tra_2\tra_1\tra_3}{L}\ri)\algodNTKtwo^{\tra_1\tra_2\tra_3\tra_4}\, \\
&\quad\quad\quad\quad\ \ +\Big(n_{L}\dNTKP{\tra_1\delta\tra_2\tra_3}{L}+\dNTKQ{\tra_1\delta\tra_2\tra_3}{L}+\dNTKQ{\tra_1\tra_2\delta\tra_3}{L}\Big)\algodNTKone^{\tra_1\tra_2\tra_3\tra_4}\, \notag\\
&\quad\quad\quad\quad\ \ +\Big(\dNTKP{\tra_1\delta\tra_2\tra_3}{L}+n_{L}\dNTKQ{\tra_1\delta\tra_2\tra_3}{L}+\dNTKQ{\tra_1\tra_2\delta\tra_3}{L}\Big)\algodNTKone^{\tra_1\tra_2\tra_4\tra_3}\Bigg] \y{i}{\tra_4}\, ,\notag
\end{align}
where we used \eqref{eq:cross-dNTK-general-leading} to evaluate the dNTK-preactivation cross correlators in terms of our decomposition into tensors $P^{(L)}$ and $Q^{(L)}$; \emph{(iii)} the first ddNTK gives\index{tensor decomposition!dNTK-preactivation $P$/$Q$}
\begin{align}\label{eq:mean-prediction-ddNTK-I-contribution}
&m_{i;\delta}^{\text{ddNTK-I}}\, \\
\equiv&-\sum_{\tra_1,\ldots,\tra_6}\Big[\ddNTKR{\delta\tra_1\tra_2\tra_3}{L}\Big(\algoddNTKItwo^{\tra_1\tra_2\tra_3\tra_4\tra_5\tra_6}+\algoddNTKItwo^{\tra_2\tra_3\tra_1\tra_5\tra_6\tra_4}+\algoddNTKItwo^{\tra_3\tra_1\tra_2\tra_6\tra_4\tra_5}\Big)\, \notag\\
&\qquad\quad\ \ +\ddNTKR{\tra_1\delta\tra_2\tra_3}{L}\algoddNTKIone^{\tra_1\tra_2\tra_3\tra_4\tra_5\tra_6}+\ddNTKR{\tra_1\tra_2\tra_3\delta}{L}\algoddNTKIone^{\tra_1\tra_2\tra_3\tra_5\tra_6\tra_4}+\ddNTKR{\tra_1\tra_3\delta\tra_2}{L}\algoddNTKIone^{\tra_1\tra_2\tra_3\tra_6\tra_4\tra_5}\Big] \notag\\
&\qquad\qquad\qquad\qquad\qquad\ \times\Big[\y{i}{\tra_4}\Big(\sum_j\y{j}{\tra_5}\y{j}{\tra_6}+n_L \Ti{\ker}{\tra_5\tra_6}{L}\Big)+\y{i}{\tra_5}\Ti{\ker}{\tra_6\tra_4}{L}+\y{i}{\tra_6}\Ti{\ker}{\tra_4\tra_5}{L}\Big]\, \notag
\end{align}
where\index{tensor decomposition!ddNTKs $R$/$S$/$T$/$U$} we used \eqref{eq:decomposition-ddNTK} to evaluate the first ddNTK mean in terms of decomposition into the tensor $\ddNTKRS^{(L)}$; and \emph{(iv)} the second ddNTK gives
\begin{align}\label{eq:mean-prediction-ddNTK-II-contribution}
&m_{i;\delta}^{\text{ddNTK-II}}\, \\
\equiv&-\sum_{\tra_1,\ldots,\tra_6}\Big[\ddNTKS{\delta\tra_1\tra_2\tra_3}{L}\algoddNTKIItwo^{\tra_1\tra_2\tra_3\tra_4\tra_5\tra_6}+\ddNTKT{\delta\tra_1\tra_2\tra_3}{L}\algoddNTKIItwo^{\tra_2\tra_3\tra_1\tra_5\tra_6\tra_4}+\ddNTKU{\delta\tra_1\tra_2\tra_3}{L}\algoddNTKIItwo^{\tra_3\tra_1\tra_2\tra_6\tra_4\tra_5}\, \notag\\
&\qquad\quad\ \ +\ddNTKS{\tra_1\tra_2\delta\tra_3}{L}\algoddNTKIIone^{\tra_1\tra_2\tra_3\tra_5\tra_6\tra_4}+\ddNTKT{\tra_1\delta\tra_3\tra_2}{L}\algoddNTKIIone^{\tra_1\tra_2\tra_3\tra_4\tra_5\tra_6}+\ddNTKU{\tra_1\tra_3\tra_2\delta}{L}\algoddNTKIIone^{\tra_1\tra_2\tra_3\tra_6\tra_4\tra_5}\Big] \notag\\
&\qquad\qquad\qquad\qquad\qquad\ \times\Big[\y{i}{\tra_4}\Big(\sum_j\y{j}{\tra_5}\y{j}{\tra_6}+n_L \Ti{\ker}{\tra_5\tra_6}{L}\Big)+\y{i}{\tra_5}\Ti{\ker}{\tra_6\tra_4}{L}+\y{i}{\tra_6}\Ti{\ker}{\tra_4\tra_5}{L}\Big]\, \notag
\end{align}
where we used \eqref{eq:decomposition-ddNTK-II} to evaluate the second ddNTK mean in terms of decomposition into the tensors $\ddNTKSS^{(L)}$, $\ddNTKTS^{(L)}$, and $\ddNTKUS^{(L)}$.

Interestingly, we see that not only does the mean prediction depend on the
NTK differentials -- as we expect, given the nontrivial representation learning at finite width -- but also it depends on the NTK variance as well, cf.~\eqref{eq:mean-prediction-DNTK-contribution}. 
This is natural as each network fits the training data with its own \emph{particular} NTK, and so the resulting fully-trained particular output \eqref{eq:very-general-finite-width-solution-DONT-CHANGE} depends on the NTK fluctuation, as we've already emphasized. In general, it is really important to understand the tradeoffs between both contributions, and in the next subsubsection we will return to comment on this interplay between random fluctuations and directed representation learning in the overall ensemble.

\index{gradient descent!wiring!finite width}\index{ddNTKs!contribution to finite-width prediction}
In addition, looking at the terms that are cubic in $\y{i}{\tra}$ in the contributions from the ddNTKs, \eqref{eq:mean-prediction-ddNTK-I-contribution} and \eqref{eq:mean-prediction-ddNTK-II-contribution}, we see that the $j$-th component of the observed outputs influences the $i$-th component of the mean prediction for $i\ne j$. Just as we saw for the mean prediction of Bayesian inference, \eqref{eq:mean-posterior-prediction-by-exact-Bayesian-at-finite-width}, this is one consequence of the wiring property of finite-width neural networks.
To see another manifestation of this wiring property, let's consider the covariance:
\be\label{eq:definition-of-covariance-fully-trained-reprint-prolly}
\cov{z_{i_1;\tea_1}^{(L)}(T)}{z_{i_2;\tea_2}^{(L)}(T)}\equiv \E{z_{i_1;\tea_1}^{(L)}(T)~z_{i_2;\tea_2}^{(L)}(T)}-\E{z_{i_1;\tea_1}^{(L)}(T)}\E{z_{i_2;\tea_2}^{(L)}(T)}\, .
\ee
While we won't print this quantity in full -- the full expression doesn't really play nicely with the constraints of the page -- you can easily extract insight by considering some specific contributions.
 For instance, we can see the imprint of output-component wiring by looking at the following contribution to the covariance,
\begin{align}\label{eq:finite-width-fully-trained-cov-wiring-term}
&\sum_{\substack{j_1,j_2\\ \tra_1,\ldots,\tra_4}}\E{\z{i_2}{\tea_2}{L}\Tia{\dNTK}{i_1 j_1j_2}{\tea_1\tra_1\tra_2}{L}} \algodNTKtwo^{\tra_1\tra_2\tra_3\tra_4}\E{\le(\z{j_1}{\tra_3}{L}-\y{j_1}{\tra_3}\ri)\le(\z{j_2}{\tra_4}{L}-\y{j_2}{\tra_4}\ri)}\, \notag\\
=&\frac{1}{n_{L-1}}\delta_{i_1i_2}\sum_{\tra_1,\ldots,\tra_4}\Bigg[\le(n_L \dNTKP{\tea_1\tra_1\tra_2\tea_2}{L}+ \dNTKQ{\tea_1\tra_1\tra_2\tea_2}{L} + \dNTKQ{\tea_1\tra_2\tra_1\tea_2}{L}\ri)\Ti{G}{\tra_3\tra_4}{L}\, \notag\\
&\quad\quad\quad\quad\quad\quad\quad\quad\quad\quad\quad\quad\quad\quad\quad\quad+\dNTKP{\tea_1\tra_1\tra_2\tea_2}{L}\Big(\sum_{j}\y{j}{\tra_3}\y{j}{\tra_4}\Big)\Bigg]\algodNTKtwo^{\tra_1\tra_2\tra_3\tra_4}\, \notag\\
&+\frac{1}{n_{L-1}}\sum_{\tra_1,\ldots,\tra_4}\dNTKQ{\tea_1\tra_1\tra_2\tea_2}{L} \algodNTKtwo^{\tra_1\tra_2\tra_3\tra_4}\le(\y{i_1}{\tra_3}\y{i_2}{\tra_4}+\y{i_1}{\tra_4}\y{i_2}{\tra_3}\ri)\, ,
\end{align}
which comes from the cross correlation between a factor of $\z{i_2}{\tea_2}{L}$ from $z_{i_2;\tea_2}^{(L)}(T)$ and one of the dNTK terms from $z_{i_1;\tea_1}^{(L)}(T)$ that involves the algorithm projector $ \algodNTKtwo^{\tra_1\tra_2\tra_3\tra_4}$.
In particular, we see that wiring is exhibited in the last term on the final line when the true outputs have nonzero components for both $i_1$ and $i_2$.
In this case, this correlation \eqref{eq:finite-width-fully-trained-cov-wiring-term} implies that the test-set predictions for $z_{i_1;\tea_1}^{(L)}(T)$ can be shifted given $z_{i_2;\tea_2}^{(L)}(T)$, for $i_1 \neq i_2$: cf.~our related discussion of \terminate{Hebbian learning} in terms of the \emph{fire-together} inductive bias in the finite-width prior in \S\ref{subsec:Hebbian} and then our following discussion of how that leads to the posterior distribution's \emph{wiring together} in \S\ref{subsec:presence-FF-Bayes}. Here, the presence of this wiring in the covariance of the solution means that this contribution to wiring occurs differently for  each network in the ensemble: the fluctuations from instantiation to instantiation of the  parameters at initialization break the permutation symmetry among the output neurons. This type of wiring suggests that each network is able to use the fluctuations between the different initial output components to its advantage in order to correlate such components together over the course of learning.

More broadly, unlike the kernel prediction\index{kernel methods!prediction} at infinite width \eqref{eq:kernel-prediction}, the prediction at finite width \eqref{eq:very-general-finite-width-solution-DONT-CHANGE} now has non-Gaussian statistics. In particular, since finite-width prediction 
is a nontrivial functional of the network output, the NTK, the dNTK, and the ddNTKs -- all at initialization --  
and since we know that the joint distribution of those quantities $p\Big(z^{(L)}, \, \NTK^{(L)},\, \dNTK^{(L)}, \, \ddNTK^{(L)},\, \ddNTKII^{(L)}\Big\vert \D\Big)$ 
is a \neo{nearly-Gaussian distribution}, then so is the fully-trained distribution $p\Big(z^{(L)}(T) \, \Big\vert \D\Big)$. In particular, there are nontrivial higher-point connected correlators; the explicit expressions of such correlators are challenging to display in any media format, though all the information needed to do so is contained in \eqref{eq:very-general-finite-width-solution-DONT-CHANGE}, and it's not that hard to zero in on any particular term of interest. The information carried in such correlators probably contains useful insight into some of the behavior of fully-trained networks and is likely worth further consideration.

\subsubsection{Generalization at Finite Width}\index{generalization error!finite-width}
Having discussed many of the qualitative differences between the finite-width prediction \eqref{eq:very-general-finite-width-solution-DONT-CHANGE} and the infinite-width kernel prediction \eqref{eq:kernel-prediction}, now let's  make some quantitative statements. In order to get a high-level understanding of how generalization is modified at finite width as compared to our extensive infinite-width analysis in \S\ref{sec:generalization-at-infinity},  we need to determine the size of the relative importance of the finite-width corrections to our predictions, namely how these corrections depend on the widths of the hidden layers $n_\ell$ and the depth of the network $L$. In particular, we want to understand corrections to our generalized \terminate{bias-variance tradeoff} \eqref{eq:bias-variance-decomposition-generalized-mse} at finite width.

\index{generalization error!bias}
Let's start by considering the bias term, $m_{i;\tea} - \y{i}{\tea}$, for which we just need to look at the mean prediction $m_{i;\tea}$ in  \eqref{eq:very-general-finite-width-solution-mean-2}. In particular, we need to compare the first term, \eqref{eq:infinite-width-ish-contribution-to-the-bias},
which more or less corresponds to the infinite-width kernel prediction $\GDGPmean_{i;\tea}$ -- up to the subleading and unimportant correction to the NTK mean --
against the other set of the finite-width contributions to the mean, \eqref{eq:mean-prediction-DNTK-contribution}--\eqref{eq:mean-prediction-ddNTK-II-contribution}. 

Now, among the finite-width contributions in \eqref{eq:very-general-finite-width-solution-mean-2}, 
the terms inside the first set of large parentheses have a sample index corresponding to a test input, $\tea$, in each of the tensors $A^{(L)}$, $B^{(L)}$, $P^{(L)}$, $Q^{(L)}$, $\ddNTKRS^{(L)}$, $\ddNTKSS^{(L)}$, $\ddNTKTS^{(L)}$, and $\ddNTKUS^{(L)}$. 
Thus, even for a training set with one training sample as we studied in \S\ref{subsec:robustness-from-infinite-GD}, the particular details of these terms require an understanding of asymptotic multiple-input solutions of the recursions for the NTK variance, the preactivation-dNTK cross correlation, and the ddNTK mean; such an analysis 
is kind of annoying and was previously left as an \terminate{adventure for thrill seekers}, with a brief \terminate{instruction manual} for those interested buried in footnote \ref{foot:thrill-seekers-guide} of \S\ref{sec:dNTK-criticality}.
Unfortunately, we have no plans here to update that manual any further, and we don't expect to miss much by not doing so.\footnote{In particular, we expect that this first set of terms will behave qualitatively similar to the second set of terms that we will discuss in the next paragraph, though the particular order-one-level details may differ.}

In contrast, the terms inside the second set of large parentheses in \eqref{eq:very-general-finite-width-solution-mean-2}
can be analyzed with the results we already have in hand.
Thus, to nonlinearly gain a large amount of intuition with a small amount of effort, let's compare the infinite-width term \eqref{eq:infinite-width-ish-contribution-to-the-bias} against only this last set of terms. In particular, since both of these terms are preceded by a common prefactor $\sum_{\tra_1,\tra_2}\NTKM^{(L)}_{\tea\tra_1}\NTKMsub_{(L)}^{\tra_1\tra_2}$
-- whose effect was analyzed in \S\ref{sec:generalization-at-infinity} --
we simply need to understand the depth and width dependence of the tensors inside the second set of brackets in \eqref{eq:very-general-finite-width-solution-mean-2}. Here we'll evaluate these tensors for a single training set input $x_{\tra}=x$, dropping the training sample indices for the rest of this analysis for notational simplicity. 

To see the physics, it's simplest to look at the  ODE limit\index{gradient descent!continuum or ODE limit} of many many steps of gradient descent, $\eta\searrow 0$,
 for which the algorithm projectors\index{algorithm projector} take particularly simple form, 
\begin{align}\label{eq:ode-algo-projector-single-input}
\algodNTKone=\algodNTKtwo=\frac{1}{2\le(\NTKMsub^{(L)}\ri)^2}\, , \quad \algoddNTKIone = \algoddNTKItwo= \algoddNTKIIone= \frac{-1}{6\le(\NTKMsub^{(L)}\ri)^3}\, , \quad \algoddNTKIItwo=\frac{-1}{2\le(\NTKMsub^{(L)}\ri)^3}
 \, ,
\end{align}
cf.~\eqref{eq:damping-force-sum-NTK-minus},  \eqref{eq:damping-force-sum-NTK-one}, and \eqref{eq:ODE-YZ}--\eqref{eq:ODE-X-III} to derive these expressions.\footnote{Backing off the ODE limit\index{gradient descent!continuum or ODE limit}, for many many steps of gradient descent with a finite learning rate $\eta$,
the single-input dNTK algorithm projectors, \eqref{eq:implicit-ZA-tensor} and \eqref{eq:implicit-ZB-tensor}, are instead given by 
\begin{align}
\algodNTKone=\frac{1}{\big(\NTKMsub^{(L)}\big)^2}\le(\frac{1-\eta\NTKMsub^{(L)}}{2-\eta\NTKMsub^{(L)}}\ri)\, , \qquad \algodNTKtwo=\frac{1}{2\big(\NTKMsub^{(L)}\big)^2}\, .
\end{align}
In conjunction with similar single-input limits of the ddNTK algorithm projectors, this will mean that the same set of ratios, \eqref{eq:all-the-ratios},  determine the generalization error bias term, though now with an additional $\eta$ dependence.
In particular, the projector $\algodNTKone$ diverges as the global learning rate approaches from below $\eta\nearrow 2/\NTKMsub^{(L)}$. 
This divergence is expected: as we noted in footnote~\ref{footnote:convergence-dynamics}, we need $ ||\Iden-\eta \NTK||_\infty < 1$ for the training dynamics to converge after many many steps $t\to\infty$.  If you check, you'll also see that some of the ddNTK algorithm projectors have the same divergence.} 
Substituting these projectors \eqref{eq:ode-algo-projector-single-input} into the dNTK and ddNTK contributions to the mean prediction, \eqref{eq:mean-prediction-dNTK-contribution}--\eqref{eq:mean-prediction-ddNTK-II-contribution}, and then considering  the $1/n$ part of the mean prediction $m_{i;\tea}$ \eqref{eq:very-general-finite-width-solution-mean-2}, we see that all the individual terms are proportional to one of the following dimensionless ratios:
\begin{align}\label{eq:all-the-ratios}
\frac{A^{(L)}}{n_{L-1}\le(\NTKMsub^{(L)}\ri)^2}\, ,\qquad\frac{B^{(L)}}{n_{L-1}\le(\NTKMsub^{(L)}\ri)^2}\, ,\qquad\frac{P^{(L)}}{n_{L-1}\le(\NTKMsub^{(L)}\ri)^2}\, ,\qquad\frac{Q^{(L)}}{n_{L-1}\le(\NTKMsub^{(L)}\ri)^2}\,  , \\[1em]
\frac{\ddNTKRS^{(L)}\ker^{(L)}}{n_{L-1}\le(\NTKMsub^{(L)}\ri)^3}\, ,\qquad\frac{\ddNTKSS^{(L)}\ker^{(L)}}{n_{L-1}\le(\NTKMsub^{(L)}\ri)^3}\, ,\qquad\frac{\ddNTKTS^{(L)}\ker^{(L)}}{n_{L-1}\le(\NTKMsub^{(L)}\ri)^3}\, ,\qquad\frac{\ddNTKUS^{(L)}\ker^{(L)}}{n_{L-1}\le(\NTKMsub^{(L)}\ri)^3}\,  . \notag
\end{align}
Thus, these eight ratios
determine the size of the finite-width effects as compared to the leading infinite-width term.\footnote{Incidentally, the form of the final two ratios on the first line of \eqref{eq:all-the-ratios} is the ultimate justification for why in \eqref{eq:scaling-relations-dNTK} we divided the dNTK-preactivation cross-correlation tensors by two factors of the NTK mean, cf.~our discussion of dimensional analysis around~\eqref{eq:dimensions-of-P-Q}. Similarly the form of the four ratios on the second line of \eqref{eq:all-the-ratios}  is the ultimate justification for the ratios \eqref{eq:scaling-relations-ddNTKs} and \eqref{eq:scaling-relations-ddNTKs-2}. For this latter set of ratios, we should have really written $\ker^{(L)}+n_L^{-1}\sum_j y_j^2$ instead of just $\ker^{(L)}$, especially when $\ker^{(L)}$ behaves as a nontrivial power law in $L$; in such cases, we should really rescale the target output $\y{i}{\tra}$ by the power-law factor $L^{-p_0/2}$ as discussed around \eqref{eq:rescale-true-outputs}.
}

Importantly, recalling our scaling laws\index{scaling law} for the NTK variance, \eqref{eq:NTK-scaling-laws}, for the dNTK-preactivation cross correlation, \eqref{eq:dNTK-scaling-laws}, and for the dNTK-preactivation cross correlation, \eqref{eq:scaling-relations-ddNTKs} and \eqref{eq:scaling-relations-ddNTKs-2}, we see that each of these dimensionless ratios will scale like the depth-to-width ratio $L/n$ (except for the subdominant $\ddNTKUS^{(L)}$ contribution).
Thus, overall we should find for the finite-width corrections
\be
m_{i;\tea} - \GDGPmean_{i;\tea} = \o{\frac{L}{n}} \, ,
\ee
where the exact order-one constant is not important. What is important is that we confirmed our expectation for the overall scaling of this correction, going as the \emph{cutoff}\index{cutoff, effective theory} of our effective theory: $r\equiv L/n$.
Similarly, we could do an analysis for the variance term of the generalization error by looking at the covariance \eqref{eq:definition-of-covariance-fully-trained-reprint-prolly}, which will be a whole lot of work with very little payoff; such an analysis would merely confirm that the leading finite-width corrections will again be of order $\o{L/n}$.\index{generalization error!variance}

\index{fluctuations!vs.~representation learning}\index{representation learning!vs.~fluctuations}
In general, given that the aspect ratio $L/n$ controls both the \terminate{fluctuations} in the ensemble \emph{and} \terminate{representation learning}, the optimal value of the ratio is likely nonzero but also small. In particular, representation learning is enhanced by the depth, but networks with too large a value of $L/n$ will both have an effect on the mean prediction of the ensemble, but perhaps even more importantly lead to exponentially-large problems when working with only a \emph{particular} network: for large enough $L/n$ our principle of typicality\index{typicality!principle of} can break down, and so the generalization error can begin to exhibit exponential behavior.\footnote{In our discussion of \neo{fluctuations} in  \S\ref{sec:signal_prop_finite_width}, we explained that too large a value of $L/n$ may lead to a difficult fine-tuning problem in terms of getting a particular networks to behave critically.

On the one hand, the contribution of such fluctuations to the bias part of the generalization error is one way to see how the downstream effect of such fluctuations may lead to problems after training. On the other hand, the fact that fluctuations can ruin criticality for a particular network is another problem. The former problem affects the ensemble as a whole, while the latter problem affects individual networks.}

This further explains the success of our effective theory description at $\o{L/n}$: a description with vanishing $L/n$, i.e.~the \terminate{infinite-width limit}, is too simple to model the properties of deep neural networks in practice; a description for larger values of $L/n$, i.e.~a \emph{small}-width or \emph{overly}-deep regime that includes many higher-order corrections, describes networks that are unlikely to be trainable; a description with small but nonzero $L/n$, i.e.~the leading finite-width effective theory accurate to $\o{L/n}$, is as simple as it can be and still accurately describe the networks that work well in practice.\footnote{
    To better understand the scale that separates the trainable regime from the overly-deep regime, see Appendix~\ref{app:mi-stuff}.
    For a discussion of how to extend this trainable regime to greater depths for fixed width, see our discussion of residual networks\index{residual network} in Appendix~\ref{app:residual}.
}

In summary, we've seen that the leading finite-width corrections all scale
 exactly according to our long-standing expectations, and we've discussed at a high level the potential tradeoff of depth versus width. In principle, we could go further in our analysis, evaluating the multi-input recursions for nearby inputs, evaluating all the terms in the covariance, and finding all the $\o{L/n}$ contributions to the generalization error with specific coefficients.\footnote{You can also more easily evaluate the multi-input version of the relevant recursions numerically with this scaling in mind in order to determine the overall coefficient for a particular activation function of interest. This would be one way to analyze these solutions for the $\relu$-like $\gelu$ and $\swish$ networks.} If we did this, what could it tell us? In particular, can we theoretically optimize the aspect ratio $L/n$ for a particular activation function without any experimentation?

\index{optimal aspect ratio}
Unfortunately at the order we're working, we won't be able to optimize the aspect ratio $L/n$ using the prediction formula \eqref{eq:very-general-finite-width-solution-DONT-CHANGE} from the end of training: the linear dependence of the generalization on the ratio means that its derivative is independent of the ratio; instead, we'd need to compute the higher-order corrections of order $\o{L^2/n^2}$ to optimize the ratio -- which is hard, though at least straightforward to do in the effective theory framework we've developed.\footnote{
    In \S\ref{sec:information-beyond-infinity}, we'll compute such higher-order effects  from an \emph{information-theoretic}\index{information theory} perspective. This analysis will give a heuristic prescription for optimizing the network's depth-to-width ratio $r$: we consider an auxiliary \neo{unsupervised learning} objective thought to be beneficial for building representations and optimize the aspect ratio in terms of \emph{that} criterion rather than the generalization error. In this setting, we're able to determine optimal aspect ratios for different activation functions by trading off leading-order effects against higher-order effects. This is somewhat similar in spirit to the way in which \terminate{Newton's method} trades off the leading decrease of the loss against the subleading loss increase in order to optimize the overall learning rate, cf.~our discussion of Newton's method in footnote~\ref{footnote:newtons-method} of~\S\ref{sec:giant-leap}.
}
At order $\o{L/n}$, the best we could hope to see is whether nonzero $L/n$ improves generalization or not by looking at the sign of its overall coefficient.
Rather than going through that hassle, we'll instead get a little more mileage out of the mean prediction $m_{i;\tea}$ by trying to understand how the \neo{algorithm dependence} at finite width leads to additional tradeoffs within the bias term of the generalization error.

\subsubsection{Inductive Bias of the Training Algorithm}\index{universality!of the fully-trained network solution}\index{ODE limit|see{gradient descent}}
As multiply mentioned, one of the key differences between the infinite-width and finite-width networks optimized via gradient-based learning is the \neo{algorithm dependence} of the fully-trained solution for finite-width networks. In particular, the solution -- at least for MSE losses -- takes a universal form, \eqref{eq:very-general-finite-width-solution-DONT-CHANGE}, with all of the dependence of the solution on the training algorithm encoded in  the \emph{algorithm projectors}\index{algorithm projector},
$\algodNTKone$, $\algodNTKtwo$, $\algoddNTKIone$, $\algoddNTKItwo$, $\algoddNTKIIone$, and $\algoddNTKIItwo$, 
whose functional forms we've established explicitly for the two second-order updates in \eqref{eq:ZB-two-step}, for many many steps of gradient descent in \eqref{eq:implicit-ZA-tensor}--\eqref{eq:implicit-ZB-tensor-II}, and for the ODE limit\index{gradient descent!continuum or ODE limit} of gradient descent in \eqref{eq:ODE-YZ}--\eqref{eq:ODE-YZ-ddNTK-II-2}. Through this universal projection, we now have a theoretical means of isolating the \emph{inductive bias of the training algorithm} from all the other inductive biases that are present in a fully-trained neural network. This provides a natural way of theoretically evaluating the relative merits of different training algorithms.\index{inductive bias!of learning algorithms|textbf}

One aspect that is apparent just from staring at the stochastic prediction \eqref{eq:very-general-finite-width-solution-DONT-CHANGE} is that the algorithm projectors only induce projections on the NTK-differential part of the finite-width prediction and cannot affect the NTK-fluctuation contribution.
More concretely, let's continue our discussion of the bias term in the generalization error\index{generalization error!bias}. In particular, the bias is given by the following difference:
\be
m_{i;\tea} - \y{i}{\tea} \, .
\ee
On the one hand, it's clear that the NTK-variance contribution to the mean prediction, \eqref{eq:mean-prediction-DNTK-contribution}, 
encoded in $A^{(L)}$ and $B^{(L)}$ is irreducible, entirely independent of the training algorithm; 
this irreducible NTK-variance contribution depends on the training data in a fixed way and arises due to instantiation-to-instantiation fluctuations in the NTK across different realizations of the parameters.
On the other hand, the projectors always act on the NTK-differential tensors $P^{(L)}$, $Q^{(L)}$, $\ddNTKRS^{(L)}$, $\ddNTKSS^{(L)}$, $\ddNTKTS^{(L)}$, and $\ddNTKUS^{(L)}$; this  means that the dNTK's and ddNTKs' effect on the network's predictions is adjustable by the training algorithm and can be tuned differently for different datasets and tasks.

To reiterate our previous discussion of the tradeoff, on the one hand the NTK-fluctuation contribution is likely harmful as it leads to a breakdown of criticality for any \emph{particular} network.
On the other hand, the latter NTK-differential contribution is the ultimate source of the nontrivial representation learning at finite width and so it would be nice to make this contribution large.
With these algorithm projectors, we now have direct means to enhance the latter benefit while keeping fixed the former cost.

One way of understanding these algorithm projectors\index{algorithm projector} is as the sample-space \emph{dual description} of a \terminate{learning algorithm},\index{duality!learning algorithm -- algorithm projectors}\index{learning algorithm!dual to algorithm projector}
analogous to the relationship between the feature functions $\fea_j(x)$ and the kernel $\kerm(x_{\delta_1},x_{\delta_2})$ that we explored in \S\ref{sec:lazy-kernel} and \S\ref{sec:nonlinear-model}.\index{feature function}\index{kernel methods!kernel} 
From the parameter-space \terminate{microscopic perspective}, a learning algorithm, such as \terminate{gradient descent}, explicitly operates on the parameters.  At the end of training, all the details of the algorithm are left implicit in the trained parameters $\theta^\star$, making it difficult to understand its effect on the model's predictions. From this perspective, the algorithm is simple to define  (\S\ref{sec:gd}), but decidedly difficult to analyze for general interacting machine-learning models (\S\ref{subsec:real-GD-at-finite-width}). However, if you can analyze or \emph{solve} a model to find its sample-space dual description,
then the algorithm projectors make the influence of the learning algorithm explicit.\footnote{This is the sense in which we meant that the conditioning on the learning algorithm would be \emph{simple} for the trained ensemble in \eqref{eq:lofty-goal-refined}.}

Analogously to the (nearly-)kernel perspective on (nearly-)linear models, 
the appearance of the algorithm projectors\index{algorithm projector} in the generalization error\index{generalization error!finite-width} means that we can begin to think about \neo{engineering} them directly, either by picking them to be directly used with a prediction formula such as \eqref{eq:very-general-finite-width-solution-DONT-CHANGE}, or alternatively by attempting to find the parameter-space dual of an algorithm projector.
While we expect this latter task to be difficult -- just as it's often hard to find the feature functions that correspond to a particular kernel -- it could be very worthwhile to explore as a means of engineering the inductive bias of the training algorithm\index{inductive bias!of learning algorithms} directly.\footnote{
    This can also be seen as another advantage of nonlinear or interacting machine learning models. For linear models, the solution is always independent of the details of the learning algorithm (\S\ref{subsec:algorithmic-independence-at-infinity}).\index{algorithm independence}
}

In particular, we expect that this could significantly improve generalization by designing algorithms that are better tailored to the details of an architecture or on the properties of the underlying dataset or task.\footnote{
Since the algorithm projectors have sample indices, the \emph{optimal} choice of the algorithm will in general depend on the details and structure of the training set in addition to the definition of the model.
}
 This design process could in principle proceed as follows: %
\emph{(i)} engineer a desired functional form for the algorithm projectors and then \emph{(ii)} determine the parameter-space training algorithm that leads to the projectors having those desired properties or specific functional form. 
While further details of such \term{inverse algorithm design}\index{inverse algorithm design|seealso{algorithm projector}} is outside the scope of this book, we think that this line of \emph{dual} analysis has the potential to unlock a much deeper understanding of the relationship among the inductive bias of the training algorithm, the inductive bias of the network architecture, and the ultimate success of the model.

\subsubsection{There's No Place Where Gradient Descent = Exact Bayesian Inference}\index{Bayesian inference!via gradient descent!but not at finite width}\index{gradient descent!as Bayesian inference!but not at finite width}
Finally, a curious reader might wonder whether the connection that we detailed in \S\ref{subsec:NTKprediction}  between gradient descent optimization and exact Bayesian inference for the infinite-width limit  persists for more realistic networks at finite width. In short, the answer is no.

In long, 
recall the training hyperparameter settings \eqref{eq:last-layer-bayes-learning-rate} and \eqref{eq:hidden-layer-bayes-learning-rate} that matched gradient-based learning with Bayesian inference at infinite width. In particular, the latter condition \eqref{eq:hidden-layer-bayes-learning-rate} required turning off any learning in the hidden layers:
\be\label{eq:hidden-layer-bayes-learning-rate-reprint}
\Lb{\ell}=0\,, \qquad \LW{\ell}=0\,,\qquad \text{for}\quad \ell <L\, .
\ee
Importantly, this makes the hidden-layer NTK vanish exactly: $\Tia{\NTK}{i_1i_2}{\delta_1\delta_2}{\ell} = 0$, for $\ell <L$ at any finite width, cf.~\eqref{eq:NTH-recursion-without-expectation}.
Next, flip back and look at the $P$- and $Q$-recursions for the dNTK-preactivation cross-correlation tensors, \eqref{eq:dNTKP-recursion-explicit} and \eqref{eq:dNTKQ-recursion-explicit}, and then you'll probably also want to flip further back and visit the $F$- and $B$-recursions, \eqref{eq:F-recursion} and \eqref{eq:B-recursion}, respectively.
(We'll be  waiting here for you when you return.) Immediately, you should notice a problem: if the hidden-layer NTK mean vanishes, then at this order $B^{(\ell)}=0$, $F^{(\ell)}=0$, and all this together implies that $P^{(L)}=Q^{(L)}=0$. Thus, all the effects of the dNTK are turned off. Similarly, if you flip forth and look at the $R$-, $S$-, $T$-, and $U$-recursions, \eqref{eq:R-recursion} and \eqref{eq:S-recursion}--\eqref{eq:U-recursion}, then you'll notice that all the effects of the ddNTKs are turned off as well.
Thus, the mean prediction of our finite-width gradient-based-learning ensemble \eqref{eq:very-general-finite-width-solution-mean-2} \emph{cannot} match the exact Bayesian posterior mean\index{posterior!posterior mean} at finite width \eqref{eq:mean-posterior-prediction-by-exact-Bayesian-at-finite-width}.

Note that this mismatch is obvious in hindsight; by only training the last layer \eqref{eq:hidden-layer-bayes-learning-rate-reprint}, we get a \terminate{linear model} \eqref{eq:linear-model-BI}.
In other words, since the hyperparameter choices of \eqref{eq:hidden-layer-bayes-learning-rate-reprint} lead to a model with random features that are fixed over the course of training, there's no representation learning possible for these settings.
In contrast, we found nontrivial representation learning when studying exact Bayesian inference at finite width in \S\ref{subsec:presence-RL-Bayes}.

Ultimately, 
for Bayesian inference we only care about the preactivation distribution $p\!\le( z^{(\ell)} \Big\vert \D \ri)$, while for gradient-based learning we need to consider the joint preactivation-NTK-dNTK-ddNTKs distribution $p\!\le( z^{(L)}, \NTK^{(L)}, \dNTK^{(L)}, \ddNTK^{(L)}, \ddNTKII^{(L)} \Big\vert \D \ri)$, 
which
incorporates the statistics of the derivatives of the preactivations at initialization: such derivatives of the model output are invisible to the exact Bayesian inferencer.

\section{RG Flow of the ddNTKs: The Full Expressions}\label{sec:gross-ddNTK-things}\index{ddNTKs!full expressions}
These expressions were kind of horrible, so we decided to hide them here at the end of the chapter. As such, they are only really needed for three reasons: \emph{(i)} to explicitly check the absence of any NTK differentials for the hyperparameter setup \eqref{eq:hidden-layer-bayes-learning-rate-reprint} and thus confirm that the connection between gradient descent and Bayesian inference doesn't persist at finite width, \emph{(ii)} to check the details of the depth-to-width scaling of the ddNTKs that we discussed in \S\ref{sec:ddNTKs}, and \emph{(iii)} to more generally evaluate the ddNTKs' contributions to
the ensemble's mean prediction, \eqref{eq:mean-prediction-ddNTK-I-contribution} and \eqref{eq:mean-prediction-ddNTK-II-contribution}, for multiple inputs  -- analytically or numerically --  or to compute other higher-order statistics of our stochastic prediction \eqref{eq:very-general-finite-width-solution-DONT-CHANGE}.

\subsubsection{$\ddNTK$ Stochastic Forward Equation}\index{forward equation!ddNTKs}
\begin{align}\label{eq:ddNTK-forward-equation}
&\Tia{\ddNTK}{i_0i_1i_2i_3}{\delta_0\delta_1\delta_2\delta_3}{\ell+1}\, \\
=&\ \ \ \delta_{i_0i_1}\frac{\LW{\ell+1}}{n_{\ell}}\!\!\sum_{j_0,j_1,j_2,j_3=1}^{n_{\ell}}\!\!\delta_{j_0j_1}\W{i_2j_2}{\ell+1}\W{i_3j_3}{\ell+1}\s{j_1}{\delta_1}{\ell}\ds{j_2}{\delta_2}{\ell}\ds{j_3}{\delta_3}{\ell} \, \notag \\
&\qquad\qquad\quad\times\Big[\dds{j_0}{\delta_0}{\ell}\Tia{\NTK}{j_0j_2}{\delta_0\delta_2}{\ell}\Tia{\NTK}{j_0j_3}{\delta_0\delta_3}{\ell}+\ds{j_0}{\delta_0}{\ell}\Tia{\dNTK}{j_0j_2j_3}{\delta_0\delta_2\delta_3}{\ell}\Big]\, \notag\\
&+\delta_{i_0i_2}\frac{\LW{\ell+1}}{n_{\ell}}\!\!\sum_{j_0,j_1,j_2,j_3=1}^{n_{\ell}}\!\!\delta_{j_0j_2}\W{i_3j_3}{\ell+1}\W{i_1j_1}{\ell+1}\s{j_2}{\delta_2}{\ell}\ds{j_3}{\delta_3}{\ell}\ds{j_1}{\delta_1}{\ell} \, \notag \\
&\qquad\qquad\quad\times\Big[\dds{j_0}{\delta_0}{\ell}\Tia{\NTK}{j_0j_3}{\delta_0\delta_3}{\ell}\Tia{\NTK}{j_0j_1}{\delta_0\delta_1}{\ell}+\ds{j_0}{\delta_0}{\ell}\Tia{\dNTK}{j_0j_3j_1}{\delta_0\delta_3\delta_1}{\ell}\Big]\, \notag\\
&+\delta_{i_0i_3}\frac{\LW{\ell+1}}{n_{\ell}}\!\!\sum_{j_0,j_1,j_2,j_3=1}^{n_{\ell}}\!\!\delta_{j_0j_3}\W{i_1j_1}{\ell+1}\W{i_2j_2}{\ell+1}\s{j_3}{\delta_3}{\ell}\ds{j_1}{\delta_1}{\ell}\ds{j_2}{\delta_2}{\ell}\, \notag \\
&\qquad\qquad\quad\times\Big[\dds{j_0}{\delta_0}{\ell}\Tia{\NTK}{j_0j_1}{\delta_0\delta_1}{\ell}\Tia{\NTK}{j_0j_2}{\delta_0\delta_2}{\ell}+\ds{j_0}{\delta_0}{\ell}\Tia{\dNTK}{j_0j_1j_2}{\delta_0\delta_1\delta_2}{\ell}\Big]\, \notag\\
&+\!\!\sum_{j_0,j_1,j_2,j_3=1}^{n_{\ell}}\!\!\W{i_0j_0}{\ell+1}\W{i_1j_1}{\ell+1}\W{i_2j_2}{\ell+1}\W{i_3j_3}{\ell+1}\ds{j_1}{\delta_1}{\ell}\ds{j_2}{\delta_2}{\ell}\ds{j_3}{\delta_3}{\ell}\, \notag\\
&\qquad\qquad\quad\times\Big[\ds{j_0}{\delta_0}{\ell}\Tia{\ddNTK}{j_0j_1j_2j_3}{\delta_0\delta_1\delta_2\delta_3}{\ell}
+\dds{j_0}{\delta_0}{\ell}\Tia{\dNTK}{j_0j_1j_2}{\delta_0\delta_1\delta_2}{\ell}\Tia{\NTK}{j_0j_3}{\delta_0\delta_3}{\ell}
\, \notag\\
&\qquad\qquad\quad\qquad
+\dds{j_0}{\delta_0}{\ell}\Tia{\dNTK}{j_0j_2j_3}{\delta_0\delta_2\delta_3}{\ell}\Tia{\NTK}{j_0j_1}{\delta_0\delta_1}{\ell}
+\dds{j_0}{\delta_0}{\ell}\Tia{\dNTK}{j_0j_3j_1}{\delta_0\delta_3\delta_1}{\ell}\Tia{\NTK}{j_0j_2}{\delta_0\delta_2}{\ell}
\, \notag\\
&\qquad\qquad\qquad\qquad
+\sigma^{\prime\prime\prime(\ell)}_{j_0\delta_0}\Tia{\NTK}{j_0j_1}{\delta_0\delta_1}{\ell}\Tia{\NTK}{j_0j_2}{\delta_0\delta_2}{\ell}\Tia{\NTK}{j_0j_3}{\delta_0\delta_3}{\ell}
\Big]\, ,\notag
\end{align}

\subsubsection{$\ddNTKII$ Stochastic Forward Equation}\index{forward equation!ddNTKs}

\begin{align}\label{eq:ddNTK-II-forward-equation}
&\Tia{\ddNTKII}{i_1i_2i_3i_4}{\delta_1\delta_2\delta_3\delta_4}{\ell+1}\, \\
=&\delta_{i_1i_3}\delta_{i_2i_4}\le(\frac{\LW{\ell+1}}{n_{\ell}}\ri)^2\!\!\sum_{j,k=1}^{n_{\ell}}\!\!\ds{j}{\delta_1}{\ell}\ds{k}{\delta_2}{\ell}\s{j}{\delta_3}{\ell}\s{k}{\delta_4}{\ell}\Tia{\NTK}{jk}{\delta_1\delta_2}{\ell}\, \notag\\
&+\delta_{i_1i_2}\frac{\LW{\ell+1}}{n_{\ell}}\!\!\sum_{j_1,\ldots,j_4=1}^{n_{\ell}}\!\!\delta_{j_1j_2}\W{i_3j_3}{\ell+1}\W{i_4j_4}{\ell+1}\ds{j_1}{\delta_1}{\ell}\ds{j_2}{\delta_2}{\ell}\ds{j_3}{\delta_3}{\ell}\ds{j_4}{\delta_4}{\ell}\Tia{\NTK}{j_1j_3}{\delta_1\delta_3}{\ell}\Tia{\NTK}{j_2j_4}{\delta_2\delta_4}{\ell}\, \notag\\
&+\delta_{i_1i_3}\frac{\LW{\ell+1}}{n_{\ell}}\!\!\sum_{j_1,\ldots,j_4=1}^{n_{\ell}}\!\!\delta_{j_1j_3}\W{i_2j_2}{\ell+1}\W{i_4j_4}{\ell+1}\s{j_3}{\delta_3}{\ell}\ds{j_4}{\delta_4}{\ell}\ds{j_1}{\delta_1}{\ell} \, \notag \\
&\qquad\qquad\quad\times\Big[\dds{j_2}{\delta_2}{\ell}\Tia{\NTK}{j_2j_1}{\delta_2\delta_1}{\ell}\Tia{\NTK}{j_2j_4}{\delta_2\delta_4}{\ell} +\ds{j_2}{\delta_2}{\ell}\Tia{\dNTK}{j_2j_1j_4}{\delta_2\delta_1\delta_4}{\ell}\Big]\, \notag\\
&+\delta_{i_2i_4}\frac{\LW{\ell+1}}{n_{\ell}}\!\!\sum_{j_1,\ldots,j_4=1}^{n_{\ell}}\!\!\delta_{j_2j_4}\W{i_1j_1}{\ell+1}\W{i_3j_3}{\ell+1}\s{j_4}{\delta_4}{\ell}\ds{j_3}{\delta_3}{\ell}\ds{j_2}{\delta_2}{\ell}\, \notag \\
&\qquad\qquad\quad\times\Big[\dds{j_1}{\delta_1}{\ell}\Tia{\NTK}{j_1j_2}{\delta_1\delta_2}{\ell}\Tia{\NTK}{j_1j_3}{\delta_1\delta_3}{\ell}+\ds{j_1}{\delta_1}{\ell}\Tia{\dNTK}{j_1j_2j_3}{\delta_1\delta_2\delta_3}{\ell}\Big]\, \notag\\
&+\!\!\sum_{j_1,j_2,j_3,j_4=1}^{n_{\ell}}\!\!\W{i_1j_1}{\ell+1}\W{i_2j_2}{\ell+1}\W{i_3j_3}{\ell+1}\W{i_4j_4}{\ell+1}\ds{j_3}{\delta_3}{\ell}\ds{j_4}{\delta_4}{\ell}\, \notag\\
&\qquad\qquad\quad\times\Big[\ds{j_1}{\delta_1}{\ell}\ds{j_2}{\delta_2}{\ell}\Tia{\ddNTKII}{j_1j_2j_3j_4}{\delta_1\delta_2\delta_3\delta_4}{\ell}+\dds{j_1}{\delta_1}{\ell}\dds{j_2}{\delta_2}{\ell}\Tia{\NTK}{j_1j_2}{\delta_1\delta_2}{\ell}\Tia{\NTK}{j_1j_3}{\delta_1\delta_3}{\ell}\Tia{\NTK}{j_2j_4}{\delta_2\delta_4}{\ell}\, \notag\\
&\qquad\qquad\qquad\quad+\ds{j_1}{\delta_1}{\ell}\dds{j_2}{\delta_2}{\ell}\Tia{\NTK}{j_2j_4}{\delta_2\delta_4}{\ell}\Tia{\dNTK}{j_1j_2j_3}{\delta_1\delta_2\delta_3}{\ell}+\ds{j_2}{\delta_2}{\ell}\dds{j_1}{\delta_1}{\ell}\Tia{\NTK}{j_1j_3}{\delta_1\delta_3}{\ell}\Tia{\dNTK}{j_2j_1j_4}{\delta_2\delta_1\delta_4}{\ell}\Big]\, ,\notag
\end{align}

\subsubsection{$\ddNTK$ Recursion}
The mean of the first ddNTK decomposes as
\begin{align}
&\E{\Tia\ddNTK{i_0i_1i_2i_3}{\delta_0\delta_1\delta_2\delta_3}{\ell}}\, \\
=&\frac{1}{n_{\ell-1}}\le[\delta_{i_0i_1}\delta_{i_2i_3}\ddNTKR{\delta_0\delta_1\delta_2\delta_3}{\ell}+\delta_{i_0i_2}\delta_{i_3i_1}\ddNTKR{\delta_0\delta_2\delta_3\delta_1}{\ell}+\delta_{i_0i_3}\delta_{i_1i_2}\ddNTKR{\delta_0\delta_3\delta_1\delta_2}{\ell}\ri]\, .\notag
\end{align}
The tensor $\ddNTKR{}{\ell}$ satisfies the following layer-to-layer recursion:
\begin{align}\label{eq:R-recursion}\index{ddNTKs!statistics!R-recursion@$R$-recursion}
&\ddNTKR{\delta_0\delta_1\delta_2\delta_3}{\ell+1}\, \\
=&\LW{\ell+1}\CW{\ell+1}\bra\sigma^{\prime\prime}_{\delta_0}\sigma_{\delta_1}\sigma^{\prime}_{\delta_2}\sigma^{\prime}_{\delta_3}\ket_{G^{(\ell)}}\Ti{\NTKM}{\delta_0\delta_2}{\ell}\Ti{\NTKM}{\delta_0\delta_3}{\ell}\, \notag\\
&+\le(\frac{n_{\ell}}{n_{\ell-1}}\ri)\le(\CW{\ell+1}\bra\sigma^{\prime}_{\delta_2}\sigma^{\prime}_{\delta_3}\ket_{G^{(\ell)}}\ri)\le(\LW{\ell+1}\bra\sigma^{\prime\prime}_{\delta_0}\sigma_{\delta_1}\ket_{G^{(\ell)}}\ri)\NTHB{\delta_0\delta_0\delta_2\delta_3}{\ell}\, \notag\\
&+\le(\frac{n_{\ell}}{n_{\ell-1}}\ri)\le(\CW{\ell+1}\bra\sigma^{\prime}_{\delta_2}\sigma^{\prime}_{\delta_3}\ket_{G^{(\ell)}}\ri)\le[\LW{\ell+1}\le(\bra\sigma^{\prime\prime}_{\delta_0}\sigma_{\delta_1}\ket_{G^{(\ell)}}\dNTKP{\delta_0\delta_2\delta_3\delta_0}{\ell}+\bra\sigma^{\prime}_{\delta_0}\sigma^{\prime}_{\delta_1}\ket_{G^{(\ell)}}\dNTKP{\delta_0\delta_2\delta_3\delta_1}{\ell}\ri)\ri]\, \notag\\
&+\le(\CW{\ell+1}\ri)^2\bra\sigma^{\prime\prime\prime}_{\delta_0}\sigma^{\prime}_{\delta_1}\sigma^{\prime}_{\delta_2}\sigma^{\prime}_{\delta_3}\ket_{G^{(\ell)}}\Ti{\NTKM}{\delta_0\delta_1}{\ell}\Ti{\NTKM}{\delta_0\delta_2}{\ell}\Ti{\NTKM}{\delta_0\delta_3}{\ell}\, \notag\\
&+\le(\frac{n_{\ell}}{n_{\ell-1}}\ri)\le(\CW{\ell+1}\bra\sigma^{\prime}_{\delta_2}\sigma^{\prime}_{\delta_3}\ket_{G^{(\ell)}}\ri)\le(\CW{\ell+1}\bra\sigma^{\prime\prime\prime}_{\delta_0}\sigma^{\prime}_{\delta_1}\ket_{G^{(\ell)}}\ri)\NTHB{\delta_0\delta_0\delta_2\delta_3}{\ell}\Ti{\NTKM}{\delta_0\delta_1}{\ell}\, \notag\\
&+\le(\frac{n_{\ell}}{n_{\ell-1}}\ri)\le(\CW{\ell+1}\bra\sigma^{\prime}_{\delta_2}\sigma^{\prime}_{\delta_3}\ket_{G^{(\ell)}}\ri)\le[\CW{\ell+1}\le(\bra\sigma^{\prime\prime\prime}_{\delta_0}\sigma^{\prime}_{\delta_1}\ket_{G^{(\ell)}}\dNTKP{\delta_0\delta_2\delta_3\delta_0}{\ell}+\bra\sigma^{\prime\prime}_{\delta_0}\sigma^{\prime\prime}_{\delta_1}\ket_{G^{(\ell)}}\dNTKP{\delta_0\delta_2\delta_3\delta_1}{\ell}\ri)\ri]\Ti{\NTKM}{\delta_0\delta_1}{\ell}\, \notag\\
&+\le(\frac{n_{\ell}}{n_{\ell-1}}\ri)\le(\CW{\ell+1}\bra\sigma^{\prime}_{\delta_2}\sigma^{\prime}_{\delta_3}\ket_{G^{(\ell)}}\ri)\le(\CW{\ell+1}\bra\sigma^{\prime}_{\delta_0}\sigma^{\prime}_{\delta_1}\ket_{G^{(\ell)}}\ri)\ddNTKR{\delta_0\delta_1\delta_2\delta_3}{\ell}\, +\o{\frac{1}{n}}\, .\notag
\end{align}

\subsubsection{$\ddNTKII$ Recursions}
The mean of the second ddNTK decomposes as
\begin{align}
&\E{\Tia\ddNTKII{i_1i_2i_3i_4}{\delta_1\delta_2\delta_3\delta_4}{\ell}}\, \\
=&\frac{1}{n_{\ell-1}}\le[\delta_{i_1i_2}\delta_{i_3i_4}\ddNTKS{\delta_1\delta_2\delta_3\delta_4}{\ell}+\delta_{i_1i_3}\delta_{i_4i_2}\ddNTKT{\delta_1\delta_3\delta_4\delta_2}{\ell}+\delta_{i_1i_4}\delta_{i_2i_3}\ddNTKU{\delta_1\delta_4\delta_2\delta_3}{\ell}\ri]\, .\notag
\end{align}
The tensor $\ddNTKS{}{\ell}$ satisfies the following layer-to-layer recursion:
\begin{align}\label{eq:S-recursion}\index{ddNTKs!statistics!S-recursion@$S$-recursion}
&\ddNTKS{\delta_1\delta_2\delta_3\delta_4}{\ell+1}\, \\
=&\CW{\ell+1}\LW{\ell+1}\bra \sigma^{\prime}_{\delta_1} \sigma^{\prime}_{\delta_2} \sigma^{\prime}_{\delta_3}\sigma^{\prime}_{\delta_4}\ket_{G^{(\ell)}}\Ti{\NTKM}{\delta_1\delta_3}{\ell}\Ti{\NTKM}{\delta_2\delta_4}{\ell}\, \notag\\
&+\le(\frac{n_{\ell}}{n_{\ell-1}}\ri)\le(\CW{\ell+1}\bra \sigma^{\prime}_{\delta_3}\sigma^{\prime}_{\delta_4}\ket_{G^{(\ell)}}\ri)\le[\LW{\ell+1}\bra \sigma^{\prime}_{\delta_1}\sigma^{\prime}_{\delta_2}\ket_{G^{(\ell)}}+\CW{\ell+1}\Ti{\NTKM}{\delta_1\delta_2}{\ell}\bra \sigma^{\prime\prime}_{\delta_1}\sigma^{\prime\prime}_{\delta_2}\ket_{G^{(\ell)}}\ri]\NTHB{\delta_1\delta_2\delta_3\delta_4}{\ell}\, \notag\\
&+\le(\CW{\ell+1}\ri)^2\bra \sigma^{\prime\prime}_{\delta_1} \sigma^{\prime\prime}_{\delta_2} \sigma^{\prime}_{\delta_3}\sigma^{\prime}_{\delta_4}\ket_{G^{(\ell)}}\Ti{\NTKM}{\delta_1\delta_2}{\ell}\Ti{\NTKM}{\delta_1\delta_3}{\ell}\Ti{\NTKM}{\delta_2\delta_4}{\ell}\, \notag\\
&+\le(\frac{n_{\ell}}{n_{\ell-1}}\ri)\le(\CW{\ell+1}\bra \sigma^{\prime}_{\delta_1}\sigma^{\prime}_{\delta_2}\ket_{G^{(\ell)}}\ri)\le(\CW{\ell+1}\bra \sigma^{\prime}_{\delta_3}\sigma^{\prime}_{\delta_4}\ket_{G^{(\ell)}}\ri)\ddNTKS{\delta_1\delta_2\delta_3\delta_4}{\ell}+\o{\frac{1}{n}}\, .\notag
\end{align}
The tensor $\ddNTKT{}{\ell}$ satisfies the following layer-to-layer recursion:
\begin{align}\label{eq:T-recursion}\index{ddNTKs!statistics!T-recursion@$T$-recursion}
&\ddNTKT{\delta_1\delta_3\delta_4\delta_2}{\ell+1}\, \\
=&\le(\LW{\ell+1}\ri)^2\bra \sigma^{\prime}_{\delta_1} \sigma^{\prime}_{\delta_2} \sigma_{\delta_3}\sigma_{\delta_4}\ket_{G^{(\ell)}}\Ti{\NTKM}{\delta_1\delta_2}{\ell}\, \notag\\
&+\le(\frac{n_{\ell}}{n_{\ell-1}}\ri)\le(\LW{\ell+1}\ri)^2\sum_{\delta_5,\ldots,\delta_8\in\D}\bra z_{\delta_5}\sigma^{\prime}_{\delta_1} \sigma_{\delta_3}\ket_{G^{(\ell)}}\bra z_{\delta_6} \sigma^{\prime}_{\delta_2}\sigma_{\delta_4}\ket_{G^{(\ell)}}\TI{G}{\delta_5\delta_7}{\ell}\TI{G}{\delta_6\delta_8}{\ell}\NTHF{\delta_7\delta_1\delta_8\delta_2}{\ell}\, \notag\\
&+\CW{\ell+1}\LW{\ell+1}\bra \sigma^{\prime}_{\delta_1} \sigma^{\prime\prime}_{\delta_2} \sigma_{\delta_3}\sigma^{\prime}_{\delta_4}\ket_{G^{(\ell)}}\Ti{\NTKM}{\delta_2\delta_1}{\ell}\Ti{\NTKM}{\delta_2\delta_4}{\ell}\, \notag\\
&+\le(\frac{n_{\ell}}{n_{\ell-1}}\ri)\CW{\ell+1}\LW{\ell+1}\Ti{\NTKM}{\delta_2\delta_4}{\ell}\sum_{\delta_5,\ldots,\delta_8\in\D}\bra z_{\delta_5}\sigma^{\prime}_{\delta_1} \sigma_{\delta_3}\ket_{G^{(\ell)}}\bra z_{\delta_6} \sigma^{\prime\prime}_{\delta_2}\sigma^{\prime}_{\delta_4}\ket_{G^{(\ell)}}\TI{G}{\delta_5\delta_7}{\ell}\TI{G}{\delta_6\delta_8}{\ell}\NTHF{\delta_7\delta_1\delta_8\delta_2}{\ell}\, \notag\\
&+\le(\frac{n_{\ell}}{n_{\ell-1}}\ri)\CW{\ell+1}\LW{\ell+1}\bra  \sigma^{\prime}_{\delta_2}\sigma^{\prime}_{\delta_4}\ket_{G^{(\ell)}}\le(\bra  \sigma^{\prime\prime}_{\delta_1}\sigma_{\delta_3}\ket_{G^{(\ell)}}\dNTKQ{\delta_2\delta_4\delta_1\delta_1}{\ell}+\bra  \sigma^{\prime}_{\delta_1}\sigma^{\prime}_{\delta_3}\ket_{G^{(\ell)}}\dNTKQ{\delta_2\delta_4\delta_1\delta_3}{\ell}\ri)\, \notag\\
&+\CW{\ell+1}\LW{\ell+1}\bra \sigma^{\prime}_{\delta_2} \sigma^{\prime\prime}_{\delta_1} \sigma_{\delta_4}\sigma^{\prime}_{\delta_3}\ket_{G^{(\ell)}}\Ti{\NTKM}{\delta_1\delta_2}{\ell}\Ti{\NTKM}{\delta_1\delta_3}{\ell}\, \notag\\
&+\le(\frac{n_{\ell}}{n_{\ell-1}}\ri)\CW{\ell+1}\LW{\ell+1}\Ti{\NTKM}{\delta_1\delta_3}{\ell}\sum_{\delta_5,\ldots,\delta_8\in\D}\bra z_{\delta_5}\sigma^{\prime}_{\delta_2} \sigma_{\delta_4}\ket_{G^{(\ell)}}\bra z_{\delta_6} \sigma^{\prime\prime}_{\delta_1}\sigma^{\prime}_{\delta_3}\ket_{G^{(\ell)}}\TI{G}{\delta_5\delta_7}{\ell}\TI{G}{\delta_6\delta_8}{\ell}\NTHF{\delta_7\delta_2\delta_8\delta_1}{\ell}\, \notag\\
&+\le(\frac{n_{\ell}}{n_{\ell-1}}\ri)\CW{\ell+1}\LW{\ell+1}\bra  \sigma^{\prime}_{\delta_1}\sigma^{\prime}_{\delta_3}\ket_{G^{(\ell)}}\le(\bra  \sigma^{\prime\prime}_{\delta_2}\sigma_{\delta_4}\ket_{G^{(\ell)}}\dNTKQ{\delta_1\delta_3\delta_2\delta_2}{\ell}+\bra  \sigma^{\prime}_{\delta_2}\sigma^{\prime}_{\delta_4}\ket_{G^{(\ell)}}\dNTKQ{\delta_1\delta_3\delta_2\delta_4}{\ell}\ri)\, \notag\\
&+\le(\frac{n_{\ell}}{n_{\ell-1}}\ri)\le(\CW{\ell+1}\ri)^2\bra \sigma^{\prime}_{\delta_1} \sigma^{\prime}_{\delta_3} \ket_{G^{(\ell)}}\bra \sigma^{\prime}_{\delta_2} \sigma^{\prime}_{\delta_4} \ket_{G^{(\ell)}}\ddNTKT{\delta_1\delta_3\delta_4\delta_2}{\ell}\, \notag\\
&+\le(\CW{\ell+1}\ri)^2\bra \sigma^{\prime\prime}_{\delta_1} \sigma^{\prime\prime}_{\delta_2} \sigma^{\prime}_{\delta_3}\sigma^{\prime}_{\delta_4}\ket_{G^{(\ell)}}\Ti{\NTKM}{\delta_1\delta_2}{\ell}\Ti{\NTKM}{\delta_1\delta_3}{\ell}\Ti{\NTKM}{\delta_2\delta_4}{\ell}\, \notag\\
&+\le(\frac{n_{\ell}}{n_{\ell-1}}\ri)\le(\CW{\ell+1}\ri)^2\Ti{\NTKM}{\delta_1\delta_3}{\ell}\Ti{\NTKM}{\delta_2\delta_4}{\ell}\sum_{\delta_5,\ldots,\delta_8\in\D}\bra z_{\delta_5}\sigma^{\prime\prime}_{\delta_1} \sigma^{\prime}_{\delta_3}\ket_{G^{(\ell)}}\bra z_{\delta_6} \sigma^{\prime\prime}_{\delta_2}\sigma^{\prime}_{\delta_4}\ket_{G^{(\ell)}}\TI{G}{\delta_5\delta_7}{\ell}\TI{G}{\delta_6\delta_8}{\ell}\NTHF{\delta_7\delta_1\delta_8\delta_2}{\ell}\, \notag\\
&+\le(\frac{n_{\ell}}{n_{\ell-1}}\ri)\le(\CW{\ell+1}\ri)^2\Ti{\NTKM}{\delta_2\delta_4}{\ell}\bra  \sigma^{\prime}_{\delta_1}\sigma^{\prime}_{\delta_3}\ket_{G^{(\ell)}}\le(\bra  \sigma^{\prime\prime\prime}_{\delta_2}\sigma^{\prime}_{\delta_4}\ket_{G^{(\ell)}}\dNTKQ{\delta_1\delta_3\delta_2\delta_2}{\ell}+\bra  \sigma^{\prime\prime}_{\delta_2}\sigma^{\prime\prime}_{\delta_4}\ket_{G^{(\ell)}}\dNTKQ{\delta_1\delta_3\delta_2\delta_4}{\ell}\ri)\, \notag\\
&+\le(\frac{n_{\ell}}{n_{\ell-1}}\ri)\le(\CW{\ell+1}\ri)^2\Ti{\NTKM}{\delta_1\delta_3}{\ell}\bra  \sigma^{\prime}_{\delta_2}\sigma^{\prime}_{\delta_4}\ket_{G^{(\ell)}}\le(\bra  \sigma^{\prime\prime\prime}_{\delta_1}\sigma^{\prime}_{\delta_3}\ket_{G^{(\ell)}}\dNTKQ{\delta_2\delta_4\delta_1\delta_1}{\ell}+\bra  \sigma^{\prime\prime}_{\delta_1}\sigma^{\prime\prime}_{\delta_3}\ket_{G^{(\ell)}}\dNTKQ{\delta_2\delta_4\delta_1\delta_3}{\ell}\ri)\, \notag\\
&+\o{\frac{1}{n}}\, .\notag
\end{align}
The tensor $\ddNTKU{}{\ell}$ satisfies the following layer-to-layer recursion:
\begin{align}\label{eq:U-recursion}\index{ddNTKs!statistics!U-recursion@$U$-recursion}
\ddNTKU{\delta_1\delta_4\delta_2\delta_3}{\ell+1}=&\le(\CW{\ell+1}\ri)^2\bra \sigma^{\prime\prime}_{\delta_1} \sigma^{\prime\prime}_{\delta_2} \sigma^{\prime}_{\delta_3}\sigma^{\prime}_{\delta_4}\ket_{G^{(\ell)}}\Ti{\NTKM}{\delta_1\delta_2}{\ell}\Ti{\NTKM}{\delta_1\delta_3}{\ell}\Ti{\NTKM}{\delta_2\delta_4}{\ell}\,\\
&+\le(\frac{n_{\ell}}{n_{\ell-1}}\ri)\le(\CW{\ell+1}\bra \sigma^{\prime}_{\delta_1}\sigma^{\prime}_{\delta_4}\ket_{G^{(\ell)}}\ri)\le(\CW{\ell+1}\bra \sigma^{\prime}_{\delta_2}\sigma^{\prime}_{\delta_3}\ket_{G^{(\ell)}}\ri)\ddNTKU{\delta_1\delta_4\delta_2\delta_3}{\ell}+\o{\frac{1}{n}}\, .\notag
\end{align}

%% file: ChpEpsilon-epilogue/epsilon_global.tex
\epiloguechapter[Epilogue: Model Complexity from the Macroscopic Perspective]{Model Complexity from the Macroscopic Perspective}\label{epi:overparameterization}
\index{duality!microscopic-macroscopic} 
\epigraph{According to the \terminate{hype} of 1987, neural networks\index{neural network} were meant to be intelligent models that discovered features\index{feature} and patterns in data. Gaussian processes in contrast are simply smoothing devices. How can Gaussian processes possibly replace neural networks?\index{neural network} Were neural networks over-hyped, or have we underestimated the power of smoothing methods?}{David MacKay \cite{mackay2003information}.\index{MacKay, David}}

\noindent{}Throughout this book, we've focused on deep-learning models that are very wide but also deep. Our reason for this focus is that such large neural networks with many many \terminate{model parameters} work extremely well in practice and thus form 
the 
foundation
of the modern approach to artificial intelligence.

The success of these \textbf{overparameterized}\index{overparameterization|textbf} models with far more parameters than training data has led many to simply conjecture that ``more is better'' when it comes to \terminate{deep learning}.
In a 
more refined sense, there's mounting empirical evidence that a \textbf{scaling hypothesis}\index{scaling hypothesis} can accurately capture the behavior of deep neural networks, and its associated \emph{scaling laws}\index{scaling law} overall point towards the optimality of the overparameterized regime.\footnote{
    See e.g.~\cite{kaplan2020scaling} for an empirical study of scaling laws\index{scaling law} in deep learning language models\index{natural language processing} based on the \terminate{transformer} architecture. Empirically, it's observed that overparameterization is good; the optimal growth of the number of training samples $\NR$  scales sublinearly with the growth in parameters $P$, though importantly they should still scale together with a power law: $\NR \propto P^\alpha$ for $0 <\alpha < 1$. 
}
The simplicity of these empirical laws recalls an earlier period in statistical physics, when
a similar scaling hypothesis was
conjectured
to govern the behavior of certain complicated systems
in statistical mechanics\index{statistical physics}.\footnote{
    In physics, such scaling laws are an example of the phenomenon of \neo{universality}, the fact that when a system has many elementary components, it can often be described by a very simple \neo{effective theory} that's independent of the microscopic details of the underlying system \cite{Kadanoff:1971pc}. The framework of \emph{renormalization group}\index{renormalization group flow} then offers an explanation for how this universality arises by characterizing the flow from the microscopic to the macroscopic \cite{PhysRevB.4.3174,PhysRevB.4.3184}. This perhaps suggests that
    the 
    analogous
    notion of \neo{representation group flow} (cf.~\S\ref{sec:marginalization-group-flow})
    may be able
    to explain the neural scaling laws\index{scaling law} of \cite{kaplan2020scaling}.
}

However, the practical success of overparameterized models in deep learning appears to be in tension with orthodox \terminate{machine learning} and classic statistical theory\index{statistics (branch of mathematics)}. 
Heuristically, the \neo{Occam's razor} principle of sparsity
posits that we should favor the simplest hypothesis that explains our observations:
in the context of  
\terminate{machine learning}, this is usually interpreted to mean that we should prefer models with fewer parameters when comparing models 
performing the same tasks.
More quantitatively, we expect that models with fewer parameters will have smaller \emph{generalization errors}\index{generalization error},
$\gen \equiv \L_\B - \L_\A$, %
and will be less prone to overfit\index{overfitting} their \terminate{training set} $\A$.
Vice versa, we should naively expect that overparameterized models 
\emph{will} overfit\index{overfitting} their training data and generalize poorly. Thus, this orthodoxy is in direct conflict with the empirical success of overparameterized neural networks and 
is a big
theoretical puzzle in understanding modern deep learning.\footnote{
    See, e.g., the extensive discussion in \cite{zhang2016understanding} on the difficulty of trying to understand why large neural networks generalize so well according to traditional measures of model complexity.
}

In this 
brief
\terminate{epilogue}, we're going to
offer a resolution of this puzzle.
The crux of the matter hinges on the notion of \term{model complexity}. 
On the one hand, our orthodox discussion of
generalization above took a \term{microscopic perspective} -- focusing on how a network works in terms of its explicit low-level components -- and wanted to 
identify model complexity with model parameters.
On the other hand, in this book 
we integrated out\index{integrating out} the model parameters and
developed 
a \term{macroscopic perspective}  -- providing an effective theory description of 
the predictions of
realistic fully-trained networks -- for which this notion of model complexity is completely reversed.

Indeed,
we 
motivated our effective theory approach in \S\ref{ch:introduction} on the basis that we would be able to find simplicity in the limit of a large number of model parameters, and from \S\ref{ch:tools} to \S\ref{ch:eot}  we've now seen how this hypothesis has been borne out again and again for 
realistic large-but-finite-width networks.
Having finished all the technical calculations (outside of the appendices), we can see now
that it's the depth-to-width aspect ratio,
\be
r \equiv L/n \, ,
\ee
that controls the model complexity of overparameterized neural networks.
To understand why, recall that 
this ratio emerged from our calculations as the expansion parameter or \emph{cutoff} of our effective theory and determined how we could \emph{truncate}
the series expansion of the 
fully-trained distribution
while still approximating the true behavior of networks with minimal error. 
This means that
it's the number of data-dependent couplings\index{data-dependent coupling} of the truncated nearly-Gaussian distribution -- and \emph{not} the 
number of 
model parameters -- that ultimately define the model complexity in deep learning. From this macroscopic perspective, there's absolutely no conflict between the 
sparse 
intuition of \terminate{Occam's razor} in theory and the 
simplicity of 
the \terminate{scaling hypothesis} in practice.

To see how this works in even greater detail, let us
recall the three main problems 
we discussed at the beginning of the book 
in \S\ref{sec:why-it-works}, 
and then review how 
the principle of sparsity\index{sparsity, principle of}
enabled us to solve them.\footnote{In this epilogue, we'll drop layer indices on our variables to ease the notation, as everything is evaluated at the output layer; we'll also sometimes drop the neural indices when they are unimportant.}
Taylor expanding the trained network output around the network's initialization~\eqref{eq:proto-dynamics}, 
\begin{align}\label{eq:proto-dynamics-reprint}
z(x_{\delta};\theta^\star) =&z(x_{\delta};\theta) +\sum_{\mu=1}^P \dthetaI_\mu \frac{dz_{\delta}}{d\theta_\mu } +\frac{1}{2} \sum_{\mu, \nu=1}^P\dthetaI_\mu \dthetaI_\nu   \frac{d^2z_{\delta} }{d\theta_\mu d\theta_\nu}+\ldots\,,  
\end{align}
we illustrated \textbf{Problem 1}, \eqref{eq:grand-probelm-1}, that we might have to 
compute
an infinite number of terms,
\be\label{eq:grand-probelm-1-reprint}
z\, ,\quad \frac{dz}{d\theta }\, ,\quad \frac{d^2z }{d\theta^2}\, ,\quad \frac{d^3z }{d\theta^3}\,,\quad \frac{d^4z }{d\theta^4}\,,\quad \dots \, , %
\ee
\textbf{Problem 2}, \eqref{eq:grand-probelm-2}, that we have to determine the map from the \terminate{initialization distribution} over the model parameters to the induced initialization distribution over the network output and its derivatives,
\be\label{eq:grand-probelm-2-reprint}
p(\theta)\to p\!\le(z, \frac{dz}{d\theta }, \frac{d^2z }{d\theta^2},\, \dots  \ri) \, ,
\ee
and \textbf{Problem 3},\eqref{eq:grand-probelm-3}, that we have to solve the training dynamics, 
which can depend on \emph{everything},
\be\label{eq:grand-probelm-3-reprint}
\theta^\star \equiv\le[\theta^{\star}\ri]\!\le(\theta,\, z,\, \frac{dz}{d\theta},\, \frac{d^2 z}{d\theta^2},\, \ldots;\, \text{learning algorithm};\, \text{training data}\ri)\, .
\ee
Let us now give our detailed solutions to this problem set -- expanding on the schematic explanations that we gave in \S\ref{sec:why-it-works} -- and then carefully examine them through the lens of \neo{model complexity}. We'll begin first with the simple infinite-width limit and then discuss the nearly-simple-but-realistic $1/n$ truncation.

\subsubsection{Sparsity at Infinite Width}
In the infinite-width limit, we can now understand our solutions as follows:
\bi
\item Addressing \textbf{Problem 1}, \eqref{eq:example-of-sparsity-1}, all the higher derivative terms vanish, and we only need to keep track of two statistical variables:
\be\label{eq:example-of-sparsity-1-reprint}
z\, , \quad \frac{dz}{d\theta} \quad \implies \quad z_{\delta} \, , \quad \NTK_{\delta_1\delta_2}\, .
\ee
Note that this first derivative 
gives
the \emph{random features}\index{feature function!random} of the \terminate{linear model} description, cf.~\eqref{eq:feature-function-stochastic}, and the kernel
associated with these features is 
just 
the
NTK.
\item Addressing \textbf{Problem 2}, we found that the network output and its first derivative are statistically independent, each governed by the simple distribution \eqref{eq:example-of-sparsity-2}:
\be\label{eq:example-of-sparsity-2-reprint}
\lim_{n \to \infty}  p\!\le(z, \frac{dz}{d\theta}, \frac{d^2z}{d\theta^2},\, \dots  \ri)=p\!\le(z_{\delta}\ri) \delta\!\le( \sum_{\mu,\nu} \lambda_{\mu\nu} \frac{dz_{\delta_1}}{d\theta_\mu}\frac{dz_{\delta_2}}{d\theta_\nu} -\NTKI_{\delta_1 \delta_2} \ri) \, ,
\ee 
where on the right-hand side, $p(z_{\delta})$ is a zero-mean Gaussian distribution
with its variance given by the kernel $\ker_{\delta_1\delta_2}$~\eqref{eq:definition-of-kernel-first}, and the second factor is a \terminate{Dirac delta function} distribution 
that fixes the contraction of first derivatives that make up the  NTK $\NTK_{\delta_1\delta_2}$ 
to be deterministically given by the
frozen NTK $\NTKI_{\delta_1\delta_2}$ \eqref{eq:frozen-NTK}.
\item Addressing \textbf{Problem 3},
we obtained a solution for the trained model parameters $\theta^\star$ in a \emph{closed form} \eqref{eq:example-of-sparsity-3}:
\be\label{eq:example-of-sparsity-3-reprint}
\lim_{n \to \infty}\theta_\mu^\star = \theta_\mu(t=0)  - \sum_{\nu,\tra_1,\tra_2,i}\lambda_{\mu\nu} \frac{\td z_{i;\tra_1}}{\td \theta_\nu} \NTKIsub^{\tra_1 \tra_2}\le(z_{i;\tra_2}-y_{i;\tra_2}\ri) \, ,
\ee
with the associated fully-trained network outputs $z_{\delta}(T)=z(x_{\delta};\theta^{\star})$ given in \eqref{eq:network-step-general}. 
We further showed in \S\ref{subsec:algorithmic-independence-at-infinity} that this fully-trained solution is independent of the algorithm used to train the network. 
\ei
Combining these insights, we found that the fully-trained distribution,
\be\label{eq:infinite-width-correct-conditioning}
\lim_{n \to \infty}p\Big(z(T)\Big) \equiv p\!\le(z(T)\Big\vert y_{\tra}, \ker_{\delta_1 \delta_2}, \NTKI_{\delta_1 \delta_2} \ri) \, ,
\ee
is a \neo{Gaussian distribution}; the reason for writing it as a conditional distribution in this way is that the mean, \eqref{eq:GD-frozen-mean}, is only a function of vector of training set labels, $y_{\tra}$, and the frozen NTK matrix, $\NTKI^{(L)}_{\delta_1 \delta_2}$, while the variance, \eqref{eq:generalized-posterior-variance}, is only a function of  the kernel matrix, $\ker^{(L)}_{\delta_1 \delta_2}$, and the frozen NTK matrix, $\NTKI^{(L)}_{\delta_1 \delta_2}$. 
In other words,
the distribution of 
predictions on a test sample, $x_{\tea}$,  
will depend on the test sample together with all of the training data, $\D = \{\tea\} \cup \A$, and 
the shape of that
data dependence is governed  by the specific functional forms of the data-dependent couplings\index{data-dependent coupling}, i.e.~the output-layer kernel  $\ker(x_{\delta_1}, x_{\delta_2})$  and output-layer frozen NTK 
$\NTKI(x_{\delta_1}, x_{\delta_2})$.
Thus, the infinite-width solution \eqref{eq:infinite-width-correct-conditioning} allows for a very sparse description, depending only on a few objects in a simple way.

\subsubsection{Near-Sparsity at Finite Width}\index{sparsity, principle of!near-sparsity at finite width}\index{near-sparsity|see{sparsity, principle of}}
Similarly, at large-but-finite width, we can now understand our solutions as follows:
\bi
\item Addressing \textbf{Problem 1}, \eqref{eq:example-of-sparsity-finite-n-1}, all derivatives $d^k f/d\theta^k$ for $k\geq 4$ are $\o{1/n^2}$, and so we only need to keep track of the statistical variables up to the third derivative:
\be\label{eq:example-of-sparsity-finite-n-1-reprint}
z\, , \quad \frac{dz}{d\theta}\, , \quad \frac{d^2z}{d\theta^2}\,, \quad \frac{d^3z}{d\theta^3} \quad \implies \quad z_\delta \, ,  \NTK_{\delta_1 \delta_2} \, ,  \dNTK_{\delta_0 \delta_1 \delta_2}\, ,  \ddNTK_{\delta_0 \delta_1 \delta_2 \delta_3}\, ,  \ddNTKII_{\delta_1 \delta_2 \delta_3 \delta_4} \, .
\ee
Here, we note that the NTK, dNTK, and ddNTKs capture all the terms up to the third derivative in our Taylor expansion \eqref{eq:proto-dynamics-reprint} for a gradient-based learning update, cf.~\eqref{eq:preactivation-updated-finite-width-refined}, \eqref{eq:NTK-updated-finite-width-refined}, and \eqref{eq:dNTK-updated-finite-width}.
\item Addressing \textbf{Problem 2}, \eqref{eq:example-of-sparsity-finite-n-2}, we evaluated the distribution of all these statistical variables, and found that its joint distribution is nearly-Gaussian:
\be\label{eq:example-of-sparsity-finite-n-2-reprint}
 p\!\le(z, \frac{dz}{d\theta}, \frac{d^2z}{d\theta^2},
 \, \dots 
 \ri) = p\!\le(z, \, \NTK,\, \dNTK,\, \ddNTK, \,\ddNTKII \ri) + \o{\frac{1}{n^2}}\, .
\ee
\item Addressing \textbf{Problem 3}, \eqref{eq:example-of-sparsity-finite-n-3}, we were able to use \terminate{perturbation theory} to solve the  nonlinear training dynamics and evaluate the predictions of fully-trained networks at finite width:
\be\label{eq:example-of-sparsity-finite-n-3-reprint}
z(x_{\delta};\theta^\star)=\le[z(x_{\delta};\theta^\star)\ri]\le(z, \, \NTK,\, \dNTK,\, \ddNTK, \,\ddNTKII;\,  \text{algorithm projectors}\ri)\, .
\ee
Here, the details of the \neo{algorithm dependence} of the  
prediction is manifest, captured entirely by a handful of algorithm projectors, cf.~\eqref{eq:implicit-ZA-tensor}--\eqref{eq:implicit-ZB-tensor-II}.
\ei
Combining these insights, we found that the fully-trained distribution,
\be\label{eq:finite-width-correct-conditioning}
p\Big(z(T) \Big) \equiv p\!\le(z(T) \Big\vert y, G, \NTKM,V,A, B, D, F, P, Q, \ddNTKRS, \ddNTKSS, \ddNTKTS, \ddNTKUS\ri)
+ \o{\frac{1}{n^2}}
\, ,
\ee
is a \neo{nearly-Gaussian distribution}; its statistics are entirely described by the conditional variables listed here, though we've  suppressed the sample indices in order to fit them all on one line.\footnote{
    Note that we've also suppressed the algorithm projectors\index{algorithm projector} in this conditioning, presuming that we are considering a fixed \terminate{learning algorithm}: once an algorithm is fixed, the projectors
    can only be 
    fixed functions of any training set tensors that contain $\o{1}$ terms -- i.e.~the metric submatrix and the NTK mean submatrix, both evaluated on the training set only, $\widetilde{G}_{\tra_1\tra_2}$ and  $\widetilde{\NTKM}_{\tra_1\tra_2}$ -- and
    of the global learning rate\index{learning rate!global}, $\eta$,
    see e.g.~\eqref{eq:implicit-ZA-tensor}--\eqref{eq:implicit-ZB-tensor-II} for \terminate{gradient descent}.
    Thus, the algorithm dependence is taken care of entirely by the tensors we're conditioning on already.
}
In addition to the metric $G$ and the NTK mean $H$, in this conditioning we're accounting for the finite-width data-dependent couplings\index{data-dependent coupling} arising from our decompositions\index{tensor decomposition!giving data-dependent couplings} of the four-point connected correlator of preactivations $\E{zzzz}_{\text{connected}}$, \eqref{eq:C4_MLPH},  the NTK-preactivation cross correlator $\E{\DNTKS z z}$, \eqref{eq:D-F-decomposition-general-layer}, the NTK variance $\E{\DNTKS^2}$, \eqref{eq:NTH-variance-decomposition},
the dNTK-preactivation cross correlator $\E{\dNTK z}$, \eqref{eq:dntk-ell-layer-decomposition},
and the means of the ddNTKs $\E{\ddNTK}$ and $\E{\ddNTKII}$, \eqref{eq:decomposition-ddNTK} and \eqref{eq:decomposition-ddNTK-II}.
Importantly, all of these finite-width tensors at $\o{1/n}$ are functions of exactly four input samples each, e.g.~for the four-point vertex we have $V_{(\delta_1 \delta_2)(\delta_3\delta_4)} \equiv V\!\le(x_{\delta_1},  x_{\delta_2},x_{\delta_3},x_{\delta_4}\ri)$, and the specific functional forms of these data-dependent couplings\index{data-dependent coupling} determine the 
overall data dependence of the distribution.
Thus, though 
slightly more complicated than the infinite-width description \eqref{eq:infinite-width-correct-conditioning}, the solution truncated to $\o{1/n}$, \eqref{eq:finite-width-correct-conditioning}, is 
a nearly-sparse\index{sparsity, principle of!near-sparsity at finite width} description, depending only on two-hands-full of objects in a nearly-simple way.

\subsubsection{Model Complexity of Fully-Trained Neural Networks}
These effective theory results, \eqref{eq:infinite-width-correct-conditioning} and \eqref{eq:finite-width-correct-conditioning}, should make it clear that for overparameterized\index{overparameterization} neural networks, 
it is no longer appropriate to identify
the number of \terminate{model parameters} with the
\terminate{model complexity}.
Consider a fixed combined training and test dataset\index{input data} of size $\ND$:
\bi
\item For the %
\terminate{Gaussian distribution}, \eqref{eq:infinite-width-correct-conditioning}, that describes an ensemble of fully-trained infinite-width networks, we only need
\be
n_{\text{out}}\NR+\le[\frac{\ND(\ND+1)}{2}\ri] + \le[\frac{\ND(\ND+1)}{2}\ri] =\o{\ND^2} \, 
\ee
numbers in order to completely specify the distribution, with each term corresponding to the numbers needed to enumerate $y_{i;\tra}$, $\ker_{\delta_1 \delta_2}$, and $\NTKI_{\delta_1 \delta_2}$, respectively.
\item For the \terminate{nearly-Gaussian distribution}, \eqref{eq:finite-width-correct-conditioning}, that describes an ensemble of 
fully-trained
finite-width networks with small-but-nonzero aspect ratios, $0 < r \ll 1$, we now need 
\be
\o{\ND^4} \,
\ee
numbers in order to completely specify the distribution, with the counting dominated by the finite-width tensors, each of which having exactly four sample indices. 
\ei\index{macroscopic perspective}\index{microscopic perspective}
Thus, while each infinite-width network has an \emph{infinite} number of microscopic model parameters, its macroscopic data-dependent couplings are only \emph{quadratic} in samples. 
Meanwhile, our finite-width networks have less model parameters than our infinite-width network -- i.e.~finite
$<$ infinite -- but their macroscopic effective description is 
more complicated!
What we have found here is the manifestation of the \textbf{microscopic-macroscopic duality}\index{duality!microscopic-macroscopic}\index{duality|textbf}: 
under this duality,
complexity in \neo{parameter space} is transformed into simplicity in \neo{sample space}, 
and density in \neo{model parameters} is exchanged for sparsity in \emph{data-dependent couplings}\index{data-dependent coupling}. In the overparameterized regime\index{overparameterization},
this duality indicates that 
we 
really should
identify the \terminate{model complexity} with 
the data-dependent couplings rather than 
the model parameters.

To further elaborate on this general point, we could imagine carrying out our finite-width \terminate{$1/n$ expansion},
 \eqref{eq:finite-width-limit-distribution-truncated}, to higher orders as
\be\label{eq:finite-width-limit-distribution-reprint}
p\Big(z(T)\Big)=p^{\{0\}}\Big(z(T)\Big) + \frac{p^{\{1\}}\Big(z(T)\Big)}{n} + \frac{p^{\{2\}}\Big(z(T)\Big)}{n^2} + \o{ \frac{1}{n^3} }\, .
\ee
To begin,
now the preactivation distribution at initialization,
\be\label{eq:marginal-distribution-action-ansatz-reprint}
p\!\le(z \Big\vert\D\ri) = \frac{1}{Z}  e^{-\ac(z) } + \o{\frac{1}{n^3}} \, ,
\ee
will be effectively described in terms of a \emph{sextic action}\index{action!sextic},
\begin{align}\label{eq:general-L-action-sixth}
\ac(z)
\equiv&\frac{1}{2}\sum_{i=1}^{n_{L}}\sum_{\delta_1,\delta_2\in\D} g^{\delta_1\delta_2} \zNL{i}{\delta_1}\zNL{i}{\delta_2}\, \\
&-\frac{1}{8}\sum_{i_1,i_2=1}^{n_{L}}\sum_{\delta_1,\ldots,\delta_4\in\D}v^{(\delta_1\delta_2)(\delta_3\delta_4)} \zNL{i_1}{\delta_1}\zNL{i_1}{\delta_2}\, \zNL{i_2}{\delta_3}\zNL{i_2}{\delta_4}
\nonumber \\
&+\frac{1}{24}\sum_{i_1,i_2,i_3=1}^{n_{L}}\sum_{\delta_1,\ldots,\delta_6\in\D}\SPC^{(\delta_1\delta_2)(\delta_3\delta_4)(\delta_5\delta_6)} \zNL{i_1}{\delta_1}\zNL{i_1}{\delta_2}\, \zNL{i_2}{\delta_3}\zNL{i_2}{\delta_4}\,\zNL{i_3}{\delta_5}\zNL{i_3}{\delta_6}\, . \notag
\end{align}
In this action, 
the \emph{sextic coupling}\index{coupling!sextic} scales as
\be
\SPC^{(\delta_1\delta_2)(\delta_3\delta_4)(\delta_5\delta_6)} = \o{\frac{1}{n^2}} \, ,
\ee 
and leads to a nontrivial  connected six-point correlator characterized by a \neo{six-point vertex}\index{six-point vertex|seealso{data-dependent coupling}}: $\SPV_{(\delta_1\delta_2)(\delta_3\delta_4)(\delta_5\delta_6)}$.\footnote{We computed this vertex in the second layer
in footnote \ref{foot:second-layer-hierarchy} of \S\ref{sec:second-layer-non-gaussian}, and  we've further taught you everything that you need to know in order to extend the computation of such higher-point correlators to deeper layers, and then analyze their scaling at criticality.
As a \terminate{checkpoint} for your algebra, the single-input recursion for the six-point vertex specialized to the $\relu$ activation function is
\be
\SPV^{(\ell+1)}= \frac{1}{8}\left[C_W^{(\ell+1)}\right]^3 \le[\SPV^{(\ell)} + 
30 V^{(\ell)} \ker^{(\ell)}
+44 \le(\ker^{(\ell)}\ri)^3
\ri]\, ,
\ee
which at criticality has a solution
\be
\frac{\SPV^{(\ell)}}{n^2 \le(K^{(\ell)}\ri)^3} = 75 \frac{\ell^2}{n^2} + \dots \,.
\ee
} At criticality, this connected correlator scales as
\be
\frac{1}{n^2} \SPV_{(\delta_1\delta_2)(\delta_3\delta_4)(\delta_5\delta_6)} \propto \o{\frac{L^2}{n^2}} \, ,
\ee
consistent with the expectations of our \terminate{effective theory} cutoff\index{cutoff, effective theory}.
Thus, we expect that the refined description, \eqref{eq:finite-width-limit-distribution-reprint}, is accurate to order $L^2/n^2$, but at the cost of a significant increase in the \terminate{model complexity}:
the counting will be dominated by the $1/n^2$ finite-width tensors -- each of which has six sample indices -- and so we now
require 
\be
\o{\ND^6}
\ee 
numbers in order to specify all the data-dependent couplings\index{data-dependent coupling} of the distribution.
In general, to achieve an accuracy of order $L^k/n^k$, we expect that a
macroscopic
description
\be\label{eq:finite-width-limit-distribution-reprint-more-terms}
p\Big(z(T)\Big) =\sum_{m=0}^{k} \frac{p^{\{m\}}\Big(z(T)\Big)}{n^m}+ \o{\frac{L^{k+1}}{n^{k+1}}}\, ,
\ee
will have its \terminate{model complexity} dominated by data-dependent couplings\index{data-dependent coupling} with $2k$-sample indices, requiring 
\be
\o{\ND^{2k}}
\ee
numbers.
In this way, the \textbf{1/\emph{n} expansion}\index{$1/n$ expansion|textbf} gives a sequence of effective theories with increasing accuracy at the cost of increasing complexity.\footnote{
    Note here that this counting is for the union of the \terminate{training set} and the \terminate{test set}, $\D = \A \cup \B$, rather than just for the training set $\A$; in other words, the stochastic predictions made on a test sample, $x_{\tea}$, will necessarily depend on that test sample.
In particular, every time you want to make a prediction on a new sample that you haven't predicted before, $\ND$ increases in size, and so does your description of the joint distribution -- \eqref{eq:infinite-width-correct-conditioning} and \eqref{eq:finite-width-correct-conditioning} -- over the entire dataset\index{input data} $\D$. Statistical models that have this property are sometimes called \textbf{non-parametric models}\index{non-parametric model|textbf}, since the ``parameters'' of the full distribution -- i.e.~the \emph{data-dependent couplings}\index{data-dependent coupling} --  depend on data points that you may not have even seen yet.

From the \terminate{macroscopic perspective}, the description of any \emph{single} prediction scales with the size of the training set as $O\big((\NR + 1)^p\big) \approx \o{\NR^p}$; in principle you can just plug $x_{\tea}$ into a prediction formula such as \eqref{eq:very-general-finite-width-solution-DONT-CHANGE}.
From the \terminate{microscopic perspective}, if we train a model and find a solution $\theta^\star \equiv \theta^\star(\A)$ for the model parameters given a training set $\A$, then in practice we can then forget about that training set and simply make predictions as $z(x_{\tea}; \theta^\star)$ --
paying only the computation complexity cost of using the model to make a prediction. %
Thus, in deep learning we can think of this non-parametric growth of macroscopic model complexity with the size of the test set as similar in nature to the microscopic $\o{P}$ complexity of the forward pass needed to make a new prediction.

Potentially it may have been useful to tell you -- at the very least in order to understand this \terminate{epilogue}'s \terminate{epigraph} --  that a non-parametric model based on a Gaussian distribution\index{Gaussian distribution!as a Gaussian process}
 is called a \textbf{Gaussian process}\index{Gaussian process|see{Gaussian distribution}}, and accordingly people sometimes say that neural networks in the \terminate{infinite-width limit} are Gaussian processes. Similarly, if you wanted to talk about the distribution over finite-width networks in the context of non-parametric statistics, you might call it a \textbf{nearly-Gaussian process}\index{nearly-Gaussian process|see{nearly-Gaussian distribution}}.\index{nearly-Gaussian distribution!as a nearly-Gaussian process} These processes are distributions over functions $z(x)$, where e.g.~$z$ is the trained model output function.
However, the reason why we haven't brought this up before is that we find this terminology
unnecessary: for any fixed dataset\index{input data} $\D$, we don't have to worry about distributions over \emph{functions} and instead can just think about a joint distribution over the finite \emph{sets} of outputs evaluated on the dataset.
}

Importantly, for any particular network architecture that we want to describe, as the depth-to-width ratio $r\equiv L/n$ increases, we'll in principle need to include more and more of these higher-order terms, making our macroscopic \terminate{effective theory} more and more complex:
\bi
\item In the strict limit  $r \to 0$, the interactions between neurons turn off, and the sparse $\o{\ND^2}$ \emph{Gaussian}\index{Gaussian distribution} description of the \terminate{infinite-width limit}, \eqref{eq:infinite-width-correct-conditioning},  will be accurate. Such networks are not really deep, as $L/n=0$, and they do not learn representations (\S\ref{subsec:linear-at-infinity}).\index{representation learning}
\item In the regime $0 < r \ll 1$, there are small nontrivial \terminate{interactions} between neurons, and the nearly-sparse\index{sparsity, principle of!near-sparsity at finite width} $\o{\ND^4}$ \emph{nearly-Gaussian}\index{nearly-Gaussian distribution} description of the finite-width effective theory truncated at order $1/n$,  \eqref{eq:finite-width-correct-conditioning},  will be accurate. Such networks are wide while at the same time having nontrivial depth, $L/n\neq 0$,  and they do learn representations (\S\ref{subsec:nonlinear-at-finite}).\index{representation learning}
\item For larger $r$, the neurons are strongly-coupled\index{interactions!strong coupling}, and a more generic $O\big(\ND^{2k}\big)$ \emph{non-Gaussian}\index{non-Gaussian distribution} description, \eqref{eq:finite-width-limit-distribution-reprint-more-terms}, would in principle be necessary. However, in this case the 
the \terminate{macroscopic perspective} leads to an \emph{ineffective} description that is not tractable, and relatedly, we do not expect such networks to be practically useful for machine learning tasks (\S\ref{sec:solution_DLN}).\footnote{
In \S\ref{sec:solution_DLN}, we were able to access this regime in the special case of \emph{deep linear networks}\index{deep linear network}. There, we saw that 
higher-point connected correlators\index{connected correlator!} can grow uncontrollably even when the network is tuned to \terminate{criticality}, and there's no reason to expect that this would be any more favorable for nonlinear activation functions.
Moreover, as we discussed in \S\ref{sec:signal_prop_finite_width}, the growth of these higher-order correlators in networks for all choices of activation functions will lead to large  \terminate{fluctuations} from instantiation-to-instantiation, meaning that the ensemble description \eqref{eq:finite-width-limit-distribution-reprint-more-terms} can no longer be trusted for any \emph{particular} network.
Altogether, this suggests that networks of an aspect ratio $r$ that require large sample-space complexity, $O\big(\ND^{2k}\big)$ with large $k$, will generically exhibit strongly-coupled \terminate{chaos}; we do not expect such networks  to be 
effectively describable or practically trainable.

Even if we could find a workable network with such a large complexity, would we ever need it?
As a \neo{gedanken model},
let's consider an ineffective description with almost no truncation, say $k\sim \ND$, which comes with an exponential number of data-dependent couplings, $O\big(\ND^{\ND}\big)$.
Such an exponentially complex description would  only really be appropriate when we have \neo{unstructured data}: e.g.~if we have a binary labeling $f(x) = \{0,1\}$ for which each label is chosen \emph{randomly}, then the number of possible functions for $\ND$ uncorrelated data points is $2^{\ND}$; each time we want to incorporate a new input-output pair into our dataset, we'd have to \emph{double} the complexity of our description.
Such unstructured data is at odds with one of the main purposes of \terminate{machine learning}: recognizing patterns in the data -- i.e.~correlations -- which allow for the learning of representations and the finding of sparse descriptions. In fact, as there's no efficient way to learn such unstructured datasets -- see e.g.~the \neo{no-free-lunch theorem} of \cite{wolpert1996lack,wolpert1997no} -- in practice we cannot possibly require these ineffective descriptions
for any realistic machine learning scenarios.
\label{footnote:epilogue-chaos}
}
\ei
In this way, our effective theory cutoff scale $r$ governs \terminate{model complexity} of the statistics needed to faithfully describe the behavior of different neural networks. The simplicity of the \neo{macroscopic perspective} only emerges for small values of the cutoff $r$.

With that in mind, the practical success of \terminate{deep learning} in the overparameterized\index{overparameterization} regime and the empirical accuracy of a simple \terminate{scaling hypothesis} is really telling us that useful neural networks\index{neural network} should be \emph{sparse}  -- hence the preference for larger and larger models -- but not too sparse -- so that they are also \emph{deep}.
Thus, from the \terminate{macroscopic perspective}, a \textbf{nearly-sparse}\index{sparsity, principle of} \terminate{model complexity} is perhaps the most important
\terminate{inductive bias}\index{inductive bias!of sparsity in deep learning} of \terminate{deep learning}.

\sbreak

For 
an \emph{information-theoretic}\index{information theory} estimate of the depth-to-width ratio $r^\star$
for which the wide attraction of simplicity and the deep need of complexity are balanced to the end of \emph{near-sparsity}\index{sparsity, principle of!near-sparsity at finite width},
please 
feel free to flip the page and make your way through Appendix \ref{app:mi-stuff}.

%% file: AppA-thermo/A_global.tex
\chapter{Information in Deep Learning}\label{app:mi-stuff}
\epigraph{What can we demand from any physical theory? \dots
nature
[is]
a difficult problem, but not a mystery for the human
mind. 
}{Ludwig Boltzmann\index{Boltzmann, Ludwig} 
\cite{boltzman-quote}
}

\index{macroscopic perspective}\index{microscopic perspective}
\noindent{}In our \emph{Initialization}\index{initialization (of you)}, \S\ref{ch:introduction}, we introduced our effective theory approach to understanding neural networks via the lens of theoretical physics. In particular, we discussed how thermodynamics was used to clarify the behavior of artificial machines such as the steam engine, and then we described how statistical mechanics was developed to explain how these macroscopic laws  arise from the statistical behavior of many microscopic elements. With this perspective,
we suggested that a similar framework might be applied to the difficult problem of deep learning theory, which we have now  demystified  from \emph{Pretraining}\index{pretraining} all the way to the \emph{End of Training}\index{end of training}, \S\ref{ch:tools} to \S\ref{ch:eot}.

In this first appendix, we'll make the connection between deep learning and these fields of physics even more detailed. To do so, we'll reformulate a few of our previous results in 
terms of 
\term{information theory}. Initially formalized by Shannon\index{Shannon, Claude} as a way of quantitatively understanding digital communication, information theory  was developed about half a century after statistical mechanics 
and is the statistical microscopic theory 
most fitting for the digital \terminate{Information Age}.

Although they a priori consider very different subject matter, both statistical mechanics and information theory share a joint language and ultimate focus on 
the same main fundamental concept: \neo{entropy}.
A particularly nice organization of entropies defines the \neo{mutual information}, a positive quantity that can be used to characterize how much the measurement of one observable can inform us about another.
This gives a nice way to quantify the overall statistical dependence due to the \terminate{interactions} of random variables. As such, the first section of this appendix (\S\ref{sec:information-theory}) will give a self-contained introduction to entropy and mutual information, 
making sure to point out
the connections between the very physical and analog setting of statistical mechanics and the very abstract and 
digital world of information theory.

With these new tools, we will be able to further 
understand information in deep learning. In particular, for infinite-width neural networks (\S\ref{sec:information-infinite}) we'll find new perspectives on the non-interaction of neurons within a layer and on the necessity of \terminate{criticality} for 
preserving the information essential for distinguishing between input samples in deep networks.
Then, at finite width (\S\ref{sec:information-beyond-infinity}) we'll use a \neo{variational principle}\index{variational principle|seealso{maximum entropy, principle}} to demonstrate how the \emph{principle of maximum entropy}\index{maximum entropy, principle} for nearly-Gaussian distributions\index{nearly-Gaussian distribution} enables us to 
compute entropies and informations up to order $1/n^3$ while only needing to know the effective $\ell$-th-layer preactivation distribution 
truncated at order $1/n$. 

At order $1/n^2$, this will let us see a nonzero \terminate{mutual information} between groups of neurons in a layer that grows quadratically with depth, further quantifying the interaction of neurons at finite width and providing an information-theoretic interpretation and generalization of our \terminate{Hebbian learning} result from \S\ref{subsec:Hebbian}. 

At order $1/n^3$, we will see how to pick a depth-to-width ratio, $r\equiv L/n$, that maximizes the mutual information as a functional of
the activation function. This \terminate{optimal aspect ratio}, $r^\star$, arises from an unsupervised learning objective\index{unsupervised learning}, and defines an activation-function-dependent scale that separates \emph{effectively-deep}\index{effectively deep} networks -- that perform well -- from \emph{overly-deep}\index{overly deep} networks -- that are no longer trainable. 

Also at order $1/n^3$, a generalization of the \terminate{mutual information} for three groups of neurons will show that the information in a finite-width layer is stored redundantly\index{redundancy (information theory)} between neurons. 
This analysis 
provides a new perspective on coarse-graining mechanism of $\emph{RG flow}$\index{representation group flow}
by enabling us to understand how the information from inputs gets represented by, and shared among, the deeper-layer neurons.

Finally, note that while most of the computations presented here focus on the prior distribution\index{prior} of preactivations as a means of investigating the \terminate{inductive bias} of the network architecture and activation function, these information-theoretic tools naturally extend to the various joint distributions we've considered throughout the book as well as the Bayesian posterior distribution and the complicated distributions of fully-trained networks. This leaves many more things to be computed, and so we hope that this chapter provides a useful introduction to a new toolkit that can be used for furthering your understanding of deep learning. 
In our second and final appendix, \S\ref{app:residual}, we'll give a \emph{residual} example 
to demonstrate this
in the setting of \emph{residual networks}\index{residual network}.

\section{Entropy and Mutual Information}\label{sec:information-theory}
In this section we give a very brief overview of the concepts of entropy and \terminate{mutual information} that play essential roles both in %
\emph{statistical mechanics}\index{statistical physics} \cite{boltzmann}
and 
\neo{information theory} \cite{Shannon-1,Shannon-2}.

Let's start with a discrete random variable $x\in\outcomes$ 
governed by the \terminate{probability distribution} $p(x)$.
For this discussion, it is nice to think of $x$ as a particular observable\index{observable} outcome, and $\outcomes$ as the set of possible outcomes.
The \term{entropy} of a \terminate{probability distribution} is given by 
\be\label{eq:entropy-shannon-formula-sum}
\entropy\!\le[p(x)\ri]
\equiv - \sum_{x \in \outcomes} p(x) \log p(x)  
\, ,
\ee
which is a \neo{functional} of the distribution, taking a
distribution as an argument and outputting a number. Thus, we should think of the entropy as a property of an entire probability distribution.

To gain some intuition for why this quantity could be useful, let's consider its two extremes. First, when the distribution is perfectly \emph{ordered} -- that is, $p(x)=\delta_{xx'}$ such that $p(x')=1$ with \neo{absolute certainty} for a particular outcome, $x'\in \outcomes$, and zero for all others 
-- then the entropy is minimal, given by 
\be\label{eq:entropy-minimal}
\entropy\!\le[p(x)\ri]
= - \sum_{x \in \outcomes} \delta_{x x'} \log \!\le(\delta_{x x'}\ri)
=0 \, ,
\ee 
since $x'$ contributes $1 \log(1)=0$ and the other values of $x$ contribute $0\log(0)=0$.
Second, when the distribution is completely \emph{disordered} -- that is,  $p(x)=1/\vert\outcomes\vert$, such that the possible outcomes $x$ are distributed uniformly and no outcome is more likely than any other -- then entropy is
maximal, given by
\be\label{eq:entropy-maximal}
\entropy\!\le[p(x)\ri]
= - \sum_{x \in \outcomes} \frac{1}{\vert\outcomes\vert} \log \!\le(\frac{1}{\vert\outcomes\vert} \ri)
=\log(\vert\outcomes\vert) \, .
\ee
For this reason, in 
\terminate{physics}
the \terminate{entropy} is often regarded as a measure of the \textbf{disorder}\index{entropy!as a measure of disorder}\index{disorder|see{entropy}} of a system characterized by a random variable $x$.\footnote{Since we're currently wearing our physicist hats, one remark is in (dis)order for the units of the entropy. As per our discussion of  \neo{dimensional analysis} in footnote~\ref{foot:dimensional-analysis} of \S\ref{sec:perturbation}, the logarithm of a probability has the same units as the \neo{action}, and for the same reason must be dimensionless. Nevertheless, by changing the base of the logarithm, we can change the multiplicative constant in front of \eqref{eq:entropy-shannon-formula-sum}, and thus change the meaning of the entropy:
\bi
\item With our physicist\index{physics} hats\index{hat (occupational)} still on, we would use the \terminate{natural logarithm} with base $e$, which measures entropy in units with a very silly name called \emph{nats}\index{nat@\texttt{nat} (unit of entropy)}. (Some physicists also multiply the expression~\eqref{eq:entropy-shannon-formula-sum} by the \terminate{Boltzmann constant} $k_{\text{B}}$, as is natural in macroscopic applications of the entropy in \neo{thermodynamics}, which gives the entropy the unit of \emph{joules per kelvin}\index{joules per kelvin@\texttt{joules per kelvin} (unit of entropy)}.)
\item With our computer scientist hats on, which we put on in anticipation of the next paragraph in the main text, we would use the logarithm with base $2$, which measures units in \emph{binary digits} or the hopefully familiar \textbf{bits}\index{bit@\texttt{bit} (unit of entropy)}. This is most natural in an information theory context, for which the entropy~\eqref{eq:entropy-shannon-formula-sum} is sometimes called the \emph{Shannon entropy}.\index{Shannon entropy|see{entropy}}\index{entropy!Shannon entropy} The reason bits are also used as the units for computer memory is due to the counting-of-states intuition for the entropy: a \terminate{hard drive} that can store $10^9$ $\texttt{bits}$ has $2^{10^9}$ unique states, with a priori equal plausibility for any particular arrangement of those bits, i.e.~for any particular state of the system.
\ei
In this chapter, we'll use the natural base $e$, which is also very natural when studying the entropy of Gaussian distributions\index{Gaussian distribution} and nearly-Gaussian distributions\index{nearly-Gaussian distribution}.\label{footnote-entropy-dimensions-nats-vs-bits}
}
In this maximal case when each outcome is equally likely, the entropy can also be interpreted as a way of counting the number of states of a system -- i.e.~the number of distinct outcomes in $\outcomes$ -- since it's equal to the logarithm of such a count.\footnote{This connection between the uniform distribution over a finite number of states and the maximum of the entropy is an example of the \emph{principle of maximum entropy}\index{maximum entropy, principle} and descends from what's called Laplace's \emph{principle of indifference}\index{Laplace's principle of indifference}: without any other information, all the a priori probabilities 
for a system to be in any particular state should be equal.
(As we will discuss in \S\ref{sec:information-beyond-infinity}, when we do have some information about the state, then the entropy is no longer maximized by a uniform distribution.)

\index{macroscopic perspective}\index{microscopic perspective}
Note that this indifference principle is applied to macroscopic thermodynamic states of (nearly-)equal energy as a principal assumption of microscopic statistical mechanics, and the associated entropy, \eqref{eq:entropy-maximal}, is sometimes called the \emph{Boltzmann entropy}.\index{Boltzmann entropy|see{entropy}}\index{entropy!Boltzmann entropy}
In contrast, the fully-general formula \eqref{eq:entropy-shannon-formula-sum}   is sometimes called the \emph{Gibbs entropy} in the context of statistical mechanics\index{statistical physics}.\index{Gibbs entropy|see{entropy}}\index{entropy!Gibbs entropy} %

Finally, for a Bayesian, this principle of indifference offers a natural way to pick a set of prior beliefs, and a prior distribution that respects this indifference principle is sometimes called a \emph{non-informative prior}\index{non-informative prior|see{prior}}\index{prior!non-information prior}\index{prior!non-information prior|seealso{Laplace's principle of indifference}}. While motivated in part by the \terminate{Occam's razor} heuristic, a Bayesian's \emph{subjective} adoption of such a prior  is very different from the automatic and objective embodiment of \terminate{Occam's razor} by the Bayes' factor\index{Bayesian inference!model comparison!Bayes' factor}, cf.~our discussion of Bayesian model comparison\index{Bayesian inference!model comparison} in~\S\ref{subsec:bayesian-model-comparison}.
}

To gain even more intuition, let's consider a perspective from \terminate{information theory}.
In the \terminate{entropy} formula \eqref{eq:entropy-shannon-formula-sum}, the quantity inside the expectation
is called the \textbf{surprisal}\index{surprisal (information theory)|textbf}: 
\be\label{eq:surprisal-def}
\surprise(x) \equiv - \log p(x)=\log\!\le[\frac{1}{p(x)}\ri] \,;
\ee
since for a particular outcome, $x$, the distribution $p(x)$ quantifies the frequency or plausibility of observing $x$, and since the logarithm is monotonic in its argument, the surprisal
grows with the a priori rareness of the outcome $x$, and hence quantifies how surprising or informative actually observing a particular $x$ is.
As such, the  surprisal is also a quantitative measure of how much new \term{information}\index{information|seealso{surprisal}} is gained after making such an observation. 
Averaging the surprisal over all possible outcomes, $\outcomes$, gives the \terminate{entropy}~\eqref{eq:entropy-shannon-formula-sum}, which can thus be understood as the expected surprise or amount of information to be gained from making an observation: 
\be\label{eq:entropy-as-expectation-of-surprisal}
\entropy\!\le[p(x)\ri] \equiv \E{\surprise(x)}\, .
\ee

In addition to admitting various nice interpretations, the \terminate{entropy} also obeys a few nice mathematical properties.
To start, it is manifestly nonnegative:
\be\label{eq:entropy-positivity}
\entropy\!\le[p(x)\ri]
\geq 0\, .
\ee
You can see this by noting that the allowed range of a probability, $0 \leq p(x) \leq 1$, implies that the nonnegativity of the quantity, $-p(x)\log p(x) \geq 0$, which in turn implies that the entropy \eqref{eq:entropy-shannon-formula-sum} is a sum of nonnegative terms. Moreover, the \terminate{entropy} takes its minimum value and vanishes, \eqref{eq:entropy-minimal}, if and only if the distribution is perfectly ordered, given by a \terminate{Kronecker delta} for one particular outcome.

\index{statistical independence}\index{statistical dependence}
Another important property of the \terminate{entropy} is its  \emph{additivity}\index{entropy!additivity} for two random variables $x \in \mathcal{X}$ and $y \in \mathcal{Y}$ that are described by a factorized joint distribution, $p(x,y)=p(x)\,p(y)$, and thus are statistically independent~\eqref{eq:independence-random-variables}:
\begin{align}
\entropy\!\le[p(x,y)\ri] &= - \sum_{x \in \mathcal{X}, y \in \mathcal{Y}} p(x, y) \log p(x, y) \notag \\
&=  - \sum_{x \in \mathcal{X}, y \in \mathcal{Y}} p(x) p(y) \le[ \log p(x) + \log p(y)\ri] \notag \\
&= - \sum_{x \in \mathcal{X}} p(x) \log p(x) - \sum_{y \in \mathcal{Y}} p(y) \log p(y) \notag \\
&= \entropy\!\le[p(x)\ri] +\entropy\!\le[p(y)\ri] \, .
\label{eq:entropy-additivity}
\end{align}
Intuitively, if two observables are independent, then the expected amount of \terminate{information} learned from making an observation of each is just the sum of the \terminate{information} learned from making each individual observation.
For macroscopic systems with small probabilities for individual outcomes, this makes the entropy a practically useful quantity to work with: while the probability of independent outcomes multiply and create smaller and smaller numbers, the surprisals, and thus the entropies, simply add.

The additivity property \eqref{eq:entropy-additivity} has a very physical interpretation: the entropy is typically an \textbf{extensive}\index{extensivity!of entropy|textbf} quantity, meaning that it scales with the number of \term{degrees of freedom} or microscopic size of the system.
This should make sense given the counting-of-states interpretation of the entropy: if a variable $x$ describes the potential contents of a $1~\texttt{TB}$ hard drive, and the variable $y$ independently describes the potential contents of another $1~\texttt{TB}$ hard drive, then the total storage capacity of the hard drives together is additive and equal to $2~\texttt{TB}$. As the number of states available to the combined ``hard drive'' system is the product of the states of the individual ``hard drive'' systems, their entropies are additive.

However, if the hard drives are not independent and \emph{constrained} so that one hard drive mirrors the other, then their total storage capacity is \emph{subadditive} and instead would be equal to $1~\texttt{TB}$.\footnote{
    This configuration has a practical realization called \emph{RAID 1}, where ``\terminate{RAID}\index{RAID|seealso{Redundant Array of Inexpensive Disks}}'' stands for \neo{Redundant Array of Inexpensive Disks}\index{Redundant Array of Inexpensive Disks|seealso{RAID}}, and allows for fault tolerance by creating data redundancy between the two hard drives.
}
In this case, the system had only half as many degrees of freedom as we naively thought it had due to the strict constraint creating strong correlations between the contents of the hard drives.

\index{statistical dependence}
Thus, more generally for two\index{observable} statistically \emph{dependent} observables 
constrained by
a nonzero interaction between them, the \terminate{entropy} 
obeys a property called \emph{subadditivity}\index{entropy!additivity!subadditivity}:
\be\label{eq:entropy-subadditivity}
\entropy\!\le[p(x,y)\ri] <  \entropy\!\le[p(x)\ri] +\entropy\!\le[p(y)\ri] \, .
\ee
In words, this means that the entropy of a joint distribution $p(x,y)$ will be always less than or equal to the sum of the entropies of the marginal distributions $p(x)$ and $p(y)$.
This is also physically intuitive: if two random variables are statistically dependent, then an observation of $x$ conveys information about the likely outcome of making an observation of $y$, and so an observation of $y$ doesn't convey as much information as it would have if we didn't already know $x$.\footnote{For the mathematically curious, we can turn this intuition into a quick proof. The usual route is to first prove the \neo{Jensen inequality} and then apply it to an auxiliary object called the  \textbf{Kullback–Leibler (KL) divergence}\index{Kullback–Leibler divergence}\index{KL divergence|see{Kullback–Leibler divergence}}.

First let's state and then prove the \terminate{Jensen inequality}. Consider a discrete probability distribution over $N$ elements $a_{\mu=1,\ldots,N}$ 
with corresponding probabilities $p_\mu$
such that $\sum_{\mu=1}^{N} p_\mu=1$. Next, consider a convex functions\index{convex function} $f(a)$, i.e.~a function that satisfies
\be\label{eq:convexity-def}
f\big(\lambda a_1 + (1-\lambda)a_2\big)\geq \lambda f(a_1) +(1-\lambda)f(a_2)\, ,
\ee 
for any $\lambda\in[0,1]$ and for any numbers $a_{1}$ and $a_2$ in the domain of the function.
The \terminate{Jensen inequality} states that the expectation of such a function is greater than or equal to the function applied to the mean:
\be\label{eq:jensen-def}
\E{f(a)} \geq f\!\le(\E{a}\ri) \,.
\ee
This can be proved by induction on $N$ as follows. First, note that \eqref{eq:jensen-def} holds trivially when $N=1$. Then, see that
\begin{align}
\E{f(a)} \equiv\sum_{\mu=1}^{N+1} p_{\mu} f(a_{\mu})=&p_{N+1} f(a_{N+1})+\le(1-p_{N+1} \ri)\sum_{\mu=1}^{N} \frac{p_{\mu}}{1-p_{N+1} } f(a_{\mu})\, \\
\geq&p_{N+1} f(a_{N+1}) +\le(1-p_{N+1} \ri) f\!\le(\sum_{\mu=1}^{N} \frac{p_{\mu}}{1-p_{N+1}}a_{\mu}\ri)\, \notag\\
\geq&f\!\le(p_{N+1}a_{N+1}+\le(1-p_{N+1}\ri)\sum_{\mu=1}^{N} \frac{p_{\mu}}{1-p_{N+1}}a_{\mu}\ri)=f\!\le(\E{a}\ri) \, ,\notag
\end{align}
where in going from the first line to the second line we used the \terminate{Jensen inequality} \eqref{eq:jensen-def} for $N$ elements, and in going from the second line to the third line we used the convexity of the function \eqref{eq:convexity-def}.

As the next step, let us introduce the KL divergence\index{Kullback–Leibler divergence} (sometimes known as the \emph{relative entropy}\index{relative entropy|see{Kullback–Leibler divergence}}) between two \emph{different} probability distributions $p(x)$ and $q(x)$ for a discrete variable, $x \in \outcomes$,
\be\label{eq:KL-divergence-def}
KL\le[ p(x) \, ||\, q(x) \ri] \equiv\sum_{x \in \outcomes} p(x) \log \!\le[\frac{p(x)}{q(x)} \ri] 
= -\entropy\!\le[p(x)\ri]+\entropy\!\le[p(x), q(x)\ri]
\, ,
\ee
which is a multi-function \terminate{functional} of both distributions, $p(x)$ and $q(x)$, and importantly is \emph{not} symmetric in its function arguments. Here $\entropy\!\le[p(x), q(x)\ri]\equiv - \sum_{x \in \outcomes} p(x) \log q(x) $ is an asymmetric quantity known as the \neo{cross entropy}. As first mentioned in footnote~\ref{footnote:KL} of \S\ref{subsec:cross-entropy}, the KL divergence is a asymmetric measure of the closeness of two different distributions and is closely related to the \emph{cross-entropy loss}\index{loss!cross-entropy} \eqref{eq:loss-cross-entropy}.

Finally, let's show that the KL divergence is nonnegative by applying the \terminate{Jensen inequality} to the convex function $f(a)=-\log(a)$ with discrete elements $a_\mu=q(x)/p(x)$:
\begin{align}\label{eq:KL-proof}
KL\le[ p(x) \, ||\, q(x) \ri]\equiv\sum_{x\in \outcomes} p(x) \log\!\le[\frac{p(x)}{q(x)} \ri]\geq - \log\!\le[\sum_{x\in \outcomes} p(x)\frac{q(x)}{p(x)} \ri]
=-\log(1)=0\, .
\end{align}
To complete our proof, note that the positivity of the KL divergence \eqref{eq:KL-proof} implies the subadditivty\index{entropy!additivity!subadditivity}  of the entropy \eqref{eq:entropy-subadditivity} for a choice of distributions as $KL\le[ p(x,y) \, ||\, p(x)\,p(y) \ri]$.
} %

\index{entropy!additivity!subadditivity}\index{entropy!additivity}\index{statistical independence}\index{interactions}
Turning this argument upside down and shuffling~\eqref{eq:entropy-subadditivity} leftside right, let us define the \term{mutual information} between two random variables as:
\begin{align}\label{eq:mutual-information}
\MI\le[p(x,y)\ri] &\equiv \entropy\!\le[p(x)\ri] +\entropy\!\le[p(y)\ri]-\entropy\!\le[p(x,y)\ri]\, \\
&= \sum_{x \in \mathcal{X}, y \in \mathcal{Y}} p(x,y) \log \!\le[ \frac{p(x,y)}{p(x)\, p(y)} \ri] \, . \notag
\end{align}
This is a \terminate{functional} of a joint probability distribution and gives an average measure of how much information an observation of $x\in\mathcal{X}$ conveys about an observation of $y\in\mathcal{Y}$, and vice versa.
Rearranged in this way, we see that the subadditivity\index{entropy!additivity!subadditivity} of the \terminate{entropy}~\eqref{eq:entropy-subadditivity} implies the nonnegativity of the mutual information,
\be\label{eq:positivity-of-mutual-information}
\MI\le[p(x,y)\ri]   \ge 0\, ,
\ee
with equality holding when and only when the sets of observable outcomes, $\mathcal{X}$ and $\mathcal{Y}$, are statistically independent.\footnote{Note that the mutual information \eqref{eq:mutual-information} can also be thought of as the KL divergence\index{Kullback–Leibler divergence} \eqref{eq:KL-divergence-def} between the joint distribution $p(x,y)$ and the product of the marginal distributions $p(x)\,p(y)$.
That is, the mutual information $\MI\le[p(x,y)\ri]$ tells us how close a joint distribution is to being a product of independent distributions, with a nonzero mutual information signaling statistical dependence.
}
Thus, the mutual information of a joint distribution is telling us something about the \neo{interactions} that create the nontrivial correlations that are a signature of statistical dependence.
We'll compute explicitly the \terminate{mutual information} of preactivations for infinite-width neural networks in \S\ref{sec:information-infinite} and for finite-width neural networks in \S\ref{sec:information-beyond-infinity}: as you might imagine, 
the 
interaction\index{interactions} of neurons at finite width will lead to a nonzero mutual information. 

As the last preparation before such computations,  we will need to extend the definition of the \terminate{entropy}~\eqref{eq:entropy-shannon-formula-sum} from discrete outcomes to continuous random variables as
\be\label{eq:entropy-shannon-formula-integral}
\entropy\!\le[ p(x)\ri] \equiv - \int \!dx \ p(x)\, \log p(x)  = \E{s(x)} \, .
\ee
While the entropy is still the expectation of the surprisal\index{surprisal (information theory)} \eqref{eq:surprisal-def}, to take that expectation we now have to evaluate an integral rather than compute a sum.
As passing to the continuum actually involves a physically interesting subtlety, let's take a few paragraphs to unravel this definition \eqref{eq:entropy-shannon-formula-integral}.

First, let's understand this subtlety mathematically. Since our random variable is continuous, we can make a smooth and monotonic change of coordinates in our continuous \neo{outcome space} $\outcomes$ from $x$ to $\widetilde{x}\equiv\widetilde{x}(x)$. Since such coordinates are arbitrary, consistency requires that the probability of observing $x$ in the interval between $[x_1,x_2]$ be the same as the probability of observing $\widetilde{x}$ in $[\widetilde{x}(x_1),\widetilde{x}(x_2)]$.
In equations, this means that
\be
p(x_1 < x < x_2) \equiv \int_{x_1}^{x_2} dx\ p(x) \, 
\ee
must equal
\begin{align}
p(\widetilde{x}_1 < \widetilde x < \widetilde{x}_2)\equiv\int_{\widetilde{x}(x_1)}^{\widetilde{x}(x_2)} d\widetilde{x}\ p(\widetilde{x})=\int_{x_1}^{x_2} dx\  \frac{d\widetilde{x}}{dx}\ p\Big(\widetilde{x}(x)\Big)\, ,
\end{align}
where in the last step we used the standard transformation property of the integral measure under a change of coordinates.
Thus, the two distributions $p(\widetilde{x})$ and $p(x)$ must be related as
\be\label{eq:probability-coordinate-change}
p\Big(\widetilde{x}(x)\Big)\equiv\frac{dx}{d\widetilde{x}}\ p(x)\, ,
\ee
which is the well-known coordinate transformation formula for a probability density of a single variable. What about the expected surprisal? 
Under this same coordinate transformation, the entropy \eqref{eq:entropy-shannon-formula-integral} is given by
\begin{align}\label{eq:entropy-shift-coordinate}
\entropy\!\le[ p(\widetilde{x})\ri]=&-\int\! d\widetilde{x}\ p(\widetilde{x})\, \log p(\widetilde{x}) \, \\
=& -\int dx \, \frac{d\widetilde{x}}{dx}\ \frac{dx}{d\widetilde{x}}\ p(x)\, \le[\log p(x) +\log\!\le(\frac{dx}{d\widetilde{x}}\ri) \ri]\, \notag\\
=& \entropy\!\le[ p(x)\ri]+\int dx \ p(x)\, \log\!\le(\frac{d\widetilde{x}}{dx}\ri)\, ,\notag
\end{align}
where in the second line we again used the standard transformation property of the integral measure under a change of coordinates and we also used \eqref{eq:probability-coordinate-change} to express the probability distribution in the old coordinates. 
So with a general (monotonic) change of coordinates, we can make the entropy take any value we'd like by a judicious choice of the \terminate{Jacobian} of the transformation, $\td \widetilde{x} / \td x$, even a negative one!

Now, let's understand the physical meaning of this subtlety. Let's consider a coordinate change that's just a multiplicative factor, $\widetilde{x}=c x$, which is like changing the base unit that we use to measure the observable $x$ from $\texttt{inches}$ to $\texttt{meters}$. In this case, the surprisal of each particular outcome shifts by the same constant, $\Delta s(x) = \log c$, and thus so does the entropy. Thus, for continuous quantities the entropy is additively sensitive to the choice of units with which we measure our observable quantities, and we really should specify these units along with the definition of the entropy \eqref{eq:entropy-shannon-formula-integral}.\footnote{For any \emph{dimensionful}\index{dimensional analysis} observable, the probability \emph{density}\index{probability distribution!as a density} $p(x)$ must also be dimensionful so that the probability of observing $x$ in the interval $[x_1,x_2]$,  $p(x_1 < x < x_2) \equiv \int_{x_1}^{x_2} dx\ p(x)$, is properly dimensionless. Yet, putting such a dimensionful object $p(x)$ into an argument of the logarithm as $\log p(x)$ is illegitimate, as we discussed in footnote~\ref{foot:dimensional-analysis} of \S\ref{ch:tools}. This is another way of seeing that for continuous probability distributions, i.e.~probability densities, we need to specify the measuring units to properly define the entropy. 

For the curious and potentially worried reader, it should be noted here that the \emph{Bayes' factor}\index{Bayesian inference!model comparison!Bayes' factor}~\eqref{eq:bayes-factor} that contained the observation dependence of our Bayesian model comparison\index{Bayesian inference!model comparison} is invariant under a coordinate transformation of our observations $y_{\A}$. What ultimately mattered there was the relative magnitude of the \emph{evidence}\index{Bayesian inference!evidence} of different hypotheses, and not the absolute value of the individual evidences.

Also, please do not confuse this issue of the units for a probability density, which change the entropy by an additive constant, with the units of entropy that we discussed in footnote~\ref{footnote-entropy-dimensions-nats-vs-bits}, which change the entropy by a multiplicative constant. Note that for discrete probability distributions, the probability distribution gives a simple probability and is already dimensionless, cf.~our brief discussion in footnote~\ref{footnote-entropy-dimensions-nats-vs-bits}.
}

Perhaps the most sensible choice is to set the measuring units according to the smallest possible measurements of $x$ that we can physically distinguish -- in other words, according to the precision limit set by the constraints of the physical world. 
This precision limit $\epsilon$ is sometimes called a \emph{cutoff}, and in practice means that we only care about the discrete probability of finding $x$ between $[x_0, x_0+\epsilon]$.\footnote{
    Please don't confuse our \emph{measurement precision cutoff}\index{measurement precision cutoff|see{cutoff}}\index{cutoff, effective theory!vs.~measurement precision} here with the \emph{perturbative cutoff}\index{perturbative cutoff|see{cutoff}} of the \terminate{effective theory}, the depth-to-width ratio $r \equiv L/n$. While in both cases we can think of them as important scales, they have very different physical meanings: in the former case, the precision cutoff gives the minimum distinguishable difference between measurements of two observables, $\epsilon \equiv \min(|z_2-z_1|)$; in the latter case, the cutoff of the effective theory, $L/n$, sets the scale at which finite-width corrections need to be taken into account in the preactivation distribution $p(z)$.     
}
Such a discretization of the \terminate{outcome space} will then ensure that the \terminate{entropy} is always positive, \eqref{eq:entropy-positivity}, since we're now dealing with discrete probabilities again.\footnote{
    In the context of deep learning, a natural choice is the precision limit of the floating-point representation of the network's variables. Since type \texttt{float}\index{float@\texttt{float}|see{type (data)}}\index{type (data)!floating-point precision} eponymously has a precision that's relative to the value being stored, one could perhaps choose the minimum precision in the relevant range.
}

If this physical sensitivity of the \terminate{entropy} to the somewhat arbitrary cutoff\index{cutoff, effective theory} bothers you, then perhaps you should consider the continuous analog of \terminate{mutual information},  
\begin{align}\label{eq:mutual-information-continuous}
\MI\le[p(x,y)\ri] &\equiv \entropy\!\le[p(x)\ri] +\entropy\!\le[p(y)\ri]-\entropy\!\le[p(x,y)\ri]\, \\
&= \int \! dx dy \ p(x,y) \, \log \!\le[ \frac{p(x,y)}{p(x)\, p(y)} \ri] \, , \notag
\end{align}
where the definition in terms of the entropy is the same as in the discrete case, \eqref{eq:mutual-information}.
In particular,  mutual information is insensitive to the choice of the measuring coordinates. To see why, let's consider  two continuous random variables $x$ and $y$ and independent monotonic coordinate transformations,
\be\label{eq:MI-invariant-transformations}
p\Big(\widetilde{x}(x)\Big)=\frac{dx}{d\widetilde{x}}\ p(x)\, ,\qquad p\Big(\widetilde{y}(y)\Big)=\frac{dy}{d\widetilde{y}}\ p(y)\, , \qquad p\Big(\widetilde{x}(x), \widetilde{y}(y)\Big)=\frac{dx}{d\widetilde{x}}\frac{dy}{d\widetilde{y}}\ p(x, y)\, ,
\ee
where to get this we applied a similar consistency-of-probabilities argument to the one that we gave above.
With this in mind, we can now show that the \terminate{mutual information} \eqref{eq:mutual-information-continuous}  stays invariant under these coordinate transformations:
\begin{align}\label{eq:mutual-information-invariant}
\MI\le[p(\widetilde{x}, \widetilde{y})\ri] &\equiv \entropy\!\le[p(\widetilde{x})\ri] +\entropy\!\le[p(\widetilde{y})\ri]-\entropy\!\le[p(\widetilde{x},\widetilde{y})\ri]\, \\
&= \int \! d\widetilde{x} d\widetilde{y} \ p(\widetilde{x},\widetilde{y}) \,\log\!\le[ \frac{p(\widetilde{x},\widetilde{y})}{p(\widetilde{x})\, p(\widetilde{y})} \ri] \, \notag\\
&=\int\!  dx dy\ p(x,y) \,\log\!\le[ \frac{p(x,y)}{p(x)\, p(y)} \ri]\equiv\MI\le[p(x,y)\ri]\, .\notag
\end{align}
In going from the second line to the third line, the coordinate transformation factors $ dx / d\widetilde{x}$ and $dy/d\widetilde{y}$ completely cancelled inside the logarithm, and the transformation of the measure cancelled the coordinate transformation  factors outside the logarithm.
Thus, the \terminate{mutual information} is completely well defined for continuous random variables, independent of the cutoff\index{cutoff, effective theory} $\epsilon$. For this reason, with a consistent and fixed choice of units, it's completely valid to compute the entropy as an intermediate step in the computation of the mutual information, and we will make use of this fact in the following sections.

Finally, note that the notion of the \terminate{mutual information} can be extended to more than two random variables. 
For instance, in \S\ref{sec:information-beyond-infinity} we will consider the \term{tripartite information} among three random variables $x\in\mathcal{X}$, $y\in\mathcal{Y}$, and $z\in\mathcal{Z}$:
\begin{align}\label{eq:TI-definition}
&\MI_{3}\le[p(x,y,z)\ri]\, \\
\equiv&\MI\le[p(x, y)\ri]-\MI\le[p(x, y| z)\ri] \, \notag\\
= & \entropy\!\le[p(x)\ri]+  \entropy\!\le[p(y)\ri] + \entropy\!\le[p(z)\ri] -  \entropy\!\le[p(x,y)\ri]-  \entropy\!\le[p(y,z)\ri]  -   \entropy\!\le[p(z,x)\ri] +   \entropy\!\le[p(x,y,z)\ri]\, \notag\\
=&\sum_{x \in \mathcal{X}, y \in \mathcal{Y}, z\in\mathcal{Z}} p(x,y,z)\, \log \!\le[ \frac{p(x,y) \, p(y,z)\, p(z,x)}{p(x)\, p(y)\, p(z)\, p(x,y,z)} \ri] \, .\notag
\end{align}
Here, $\MI\le[p(x, y| z)\ri]$ is the mutual information of the joint distribution between $x$ and $y$ conditioned on $z$, $p(x, y| z)$, and the final expression for $\MI_{3}\le[p(x,y,z)\ri]$ makes it clear that \emph{(a)} it is fully symmetric in its three arguments, and that \emph{(b)} its continuous analog that's in your imagination is cutoff independent and invariant under similar coordinate transformations as those in \eqref{eq:MI-invariant-transformations}.

What is not immediately obvious is that tripartite information can be either positive or negative. 
From the first expression in \eqref{eq:TI-definition}, we can gain some intuition for the meaning of the tripartite information: it is a measure of whether knowledge of a third random variable, $z$ in this way of writing the expression,
 increases or decreases the mutual information between the other two variables. When positive, it indicates that $z$ contains information about $x$ and $y$, and so knowing $z$ decreases the amount of information you'd learn about $x$ by measuring $y$; the information is stored \emph{redundantly}\index{redundancy (information theory)} between these three variables. When negative, it indicates that the information is distributed across $x$, $y$, and $z$ in such a way that you'd learn less about $x$ by measuring $y$ than you would with first knowing $z$; in this case, the information is stored \emph{synergistically}\index{synergy (information theory)} between these three variables.\footnote{
An extreme example of positive tripartite information occurs when $x$, $y$, and $z$ are completely dependent and exact copies of each other. In this redundant case, $\MI\le[p(x, y)\ri]>0$, since knowledge of  $y$ tells you everything about $x$, but $\MI\le[p(x, y| z)\ri]=0$, since conditioning on $z$ means that there's nothing left for you to learn about $x$ by also observing $y$. An example of such a situation would be three copies of the same book.

An extreme example of negative tripartite information occurs when the joint distribution between $x$ and $y$ factorizes, $p(x,y)=p(x) \, p(y)$, but joint distribution conditioned on $z$ does not, $p(x, y|z) \neq p(x|z)\,p(y|z)$.  In this synergistic case,
$\MI\le[p(x, y)\ri]=0$, since without $z$ the variables are statistically independent, but $\MI\le[p(x, y| z)\ri]>0$, since there are correlations that are mediated by $z$. An example of such a situation could be a code that distributes a key among three different parties: knowledge of any two parts would give absolutely no information about the key, but with all three parts together the key can be reconstructed.}

\section{Information at Infinite Width: Criticality}\label{sec:information-infinite} 
With that informative introduction out of the way, let's us now make these abstract definitions concrete by using them to better understand the neural-network prior distribution\index{prior}.

To begin, let's focus on $m$ preactivations $\z{i}{\alpha}{\ell}$ from the $\ell$-th layer of an infinite-width neural network. Depending on when you're coming to this appendix from the main text, it's probably ingrained in your mind already that such preactivations are distributed according to a zero-mean Gaussian distribution\index{Gaussian distribution!entropy}  
\be\label{eq:GP-for-once-more}
p\!\le(z_{1},\ldots,z_{m}\Big\vert \D\ri) = \frac{1}{\sqrt{\dete{2\pi \ker}^{m}}} \exp\!\le(-\frac{1}{2}\sum_{i=1}^{m}\sum_{\alpha_1,\alpha_2\in\D}\ker^{\alpha_1 \alpha_2}z_{i;\alpha_1}z_{i;\alpha_2}\ri)\, ,
\ee
where in this expression, and in this section, we will drop \terminate{layer indices} everywhere to declutter expressions. 
The \terminate{entropy}~\eqref{eq:entropy-shannon-formula-integral} 
of this distribution is then given by
\begin{align}\label{eq:entropy-preactivation-infinite-width-multi}
\entropy\!\le[p\!\le(z_{1},\ldots,z_{m}\Big\vert \D\ri)\ri]=&\bra\!\!\!\bra\frac{1}{2}\sum_{i=1}^{m}\sum_{\alpha_1,\alpha_2\in\D}\ker^{\alpha_1 \alpha_2}z_{i;\alpha_1}z_{i;\alpha_2}\ket\!\!\!\ket_{\ker}+\log\!\le(\sqrt{\dete{2\pi \ker}^{m}}\ri)\, \\
=&\frac{m}{2} \le(\ND+\log \dete{2\pi \ker} \ri)=\frac{m}{2} \log\dete{2e\pi \ker}\, ,\notag
\end{align}
where as a reminder $\dete{ 2\pi e \ker }$ is the determinant of the $\ND$-by-$\ND$ matrix $2\pi e\ker_{\alpha_1\alpha_2}$.\footnote{In this and the next section, for convenience we will ignore the ambiguity in the definition of the continuous \terminate{entropy} and use the formula \eqref{eq:entropy-shannon-formula-integral} naively. This is permissible since we ultimately are interested in cutoff-independent quantities, such as the \terminate{mutual information}, and when interpreting an entropy such as \eqref{eq:entropy-preactivation-infinite-width-multi} we will never care about its absolute value.
}
From this we conclude that entropy is \emph{extensive}\index{extensivity!of entropy}, proportional to the number of neurons $m$ in the joint distribution \eqref{eq:GP-for-once-more}. This shows how the entropy can count the \neo{degrees of freedom} in a deep learning context -- in this case, by counting the neurons -- and the exact additivity in the number of neurons signals to us that the individual neurons in an infinite-width layer are non-interacting and statistically independent.\index{statistical independence}

To confirm this directly, let's work out the \terminate{mutual information} between two sets of neurons, $\mathcal{M}_1=\le\{1,\ldots,m_1\ri\}$ and $\mathcal{M}_2=\le\{m_1+1,\ldots,m_1+m_2\ri\}$, in the same layer $\ell$. In excruciating detail for its triviality, 
we have
\begin{align}\label{eq:MIT-zeroth-order}
&\MI\le[p\!\le(\mathcal{M}_1, \mathcal{M}_2\Big\vert \D\ri)\ri]\, \\
=&\entropy\!\le[p\!\le(z_{1},\ldots,z_{m_1}\Big\vert \D\ri)\ri]+\entropy\!\le[p\!\le(z_{m_1+1},\ldots,z_{m_1+m_2}\Big\vert \D\ri)\ri]-\entropy\!\le[p\!\le(z_{1},\ldots,z_{m_1+m_2}\Big\vert \D\ri)\ri]\, \notag\\
=&\le[m_1+m_2-(m_1+m_2)\ri]\frac{1}{2}\log\dete{2e\pi \ker}=0\, .\notag
\end{align}
where to go to the last line we used \eqref{eq:entropy-preactivation-infinite-width-multi} three different ways.
This zero mutual information confirms that at infinite width 
learning the activities of some neurons in a layer conveys no information about the activities of any of the other neurons.
To find a finite result, we'll have to back off the \terminate{infinite-width limit} (\S\ref{sec:information-beyond-infinity}).

That said, for a fixed neuron we do expect nontrivial correlations between different inputs and therefore also a finite mutual information.
To investigate this, let's take two inputs $\x{i}{+}$ and $\x{i}{-}$ and compute the \terminate{mutual information} between the preactivations  for a particular neuron, $z_{1;+}$ and $z_{1;-}$.
Plugging the entropy~\eqref{eq:entropy-preactivation-infinite-width-multi} into the definition of the mutual information~\eqref{eq:mutual-information}, we find
\begin{align}\label{eq:MI-one-neuron-many-samples}
\MI\le[p\!\le(z_{1;+},z_{1;-}| x_{\pm}\ri)\ri]=&\frac{1}{2}\le[\log(\ker_{++})+\log(\ker_{--})-\log\!\le(\ker_{++}\ker_{--}-\ker_{+-}^2\ri)\ri]\, \\
=&\frac{1}{2}\log\!\le(\frac{\ker_{++}\ker_{--}}{\ker_{++}\ker_{--}-\ker_{+-}^2}\ri)\, .\notag
\end{align}
Focusing on inputs $\x{i}{\pm}$ with equal norms such that $\ker_{++}=\ker_{--}=\ker_{[0]}+\ker_{[2]}$ and $\ker_{+-}=\ker_{[0]}-\ker_{[2]}$, cf.~\eqref{eq:K0-decomposition} and \eqref{eq:K2-decomposition}, we can rewrite this mutual information as\index{$\gamma^{[a]}$ basis!kernel}
\be
\MI\le[p\!\le(z_{1;+},z_{1;-}\ri)\ri]=\frac{1}{2}\log\!\le[\frac{\le(1+\frac{\ker_{[2]}}{\ker_{[0]}}\ri)^2}{4\frac{\ker_{[2]}}{\ker_{[0]}}}\ri]\, ,
\ee
which is parameterized  entirely by the dimensionless ratio $\ker_{[2]}/\ker_{[0]}$ that for nearby inputs $\x{i}{\pm} = \x{i}{\M} \pm \delta x_i /2$ characterizes their relative angle after passing through the network to the $\ell$-th layer.

There are two interesting limits here. First, when $\ker_{[2]}/\ker_{[0]}\rightarrow 0$, the mutual information becomes large.
This follows because as the relative angle between the preactivations vanishes, they become close to each other: knowing the preactivation for one input tells us about the value of the preactivation for  the other input.  In a \terminate{classification} problem, this is a great prior if the inputs are from the same class, but would make learning really difficult if they're from different classes. Second, when $\ker_{[0]}=\ker_{[2]}$, the mutual information vanishes. This follows because in this limit the off-diagonal components of the kernel, $\ker_{+-}=\ker_{[0]}-\ker_{[2]}$, vanishes: the preactivations become statistically independent. In a \terminate{classification} problem, this may be a good prior if the inputs are from different classes -- so long as the details of the classes don't correlate in some way -- but would make learning really difficult if the inputs are from the same class. Altogether, this suggests that as a prior we don't want $\ker_{[2]}/\ker_{[0]}$ to be too big or too small, which can be best ensured by setting both $\chi_\parallel =1$ and $\chi_\perp =1$, cf.~\eqref{eq:criticality-conditions}. This gives an information-theoretic\index{information theory!perspective on criticality} perspective on \neo{criticality}.

Finally, while we have focused here on the prior distribution for  networks at initialization, we could also study these same quantities after learning using the Bayesian posterior \eqref{eq:posterior-at-infinite-width} or the fully-trained distribution of gradient-based learning \eqref{eq:kernel-prediction}.
Since the entropy of a Gaussian distribution\index{Gaussian distribution!entropy} is independent of its mean
 -- it can be eliminated by a change of dummy integration variables -- 
the mutual information  for either trained infinite-width network is given by this same expression, \eqref{eq:MI-one-neuron-many-samples}, but with the kernel replaced by the covariance of the Bayesian posterior distribution~\eqref{eq:GP-posterior-variance} or the covariance of the generalized posterior distribution\index{posterior!generalized posterior distribution}~\eqref{eq:generalized-posterior-variance}. 
In the latter case, the mutual information will involve both the kernel and the frozen NTK, and its analysis would yield  similar results to those that we found in \S\ref{subsec:robustness-from-infinite-GD} when we investigated the \terminate{bias-variance tradeoff}. This would give an information-theoretic perspective on \terminate{generalization}\index{$\ast$-polation}.\footnote{
    Analogously, studying the \terminate{tripartite information} generalization of \eqref{eq:MI-one-neuron-many-samples} for  either type of posterior distribution would give an alternative perspective on the $\ast$-polation results of \S\ref{subsec:star-polation}.
}\index{information theory!perspective on generalization}\index{information theory!perspective on $\ast$-polation}

\section{Information at Finite Width: Optimal Aspect Ratio}\label{sec:information-beyond-infinity}
In this section, we'll see how finite-width networks have a prior for nonzero mutual information between different neurons.
Assuming that nonzero mutual information is desirable by intuition -- and by analogy to an \emph{unsupervised} learning\index{unsupervised learning} objective -- we can use this computation to optimize the depth-to-width ratio for finite-width MLPs at criticality. This \neo{optimal aspect ratio}, $r^\star$, defines the scale that separates effectively-deep networks -- describable by our  effective theory approach -- from overly-deep networks -- not describable due to strong interactions and not trainable due to large fluctuations.\footnote{Note that in this section, we'll need the expressions for the running couplings that we derived in~\S\ref{sec:sum-rule} when finding the effective distribution of $m$ neurons in a wide-but-finite layer. As such, it may be helpful to reread that section before proceeding.}

In general, the \terminate{entropy} and \terminate{mutual information} of any interacting theory\index{interacting theory!entropy and mutual information} are really difficult to compute. However, when the interactions are \emph{nearly-Gaussian}\index{nearly-Gaussian distribution}, we can use \terminate{perturbation theory}. To do so, there is actually a neat \emph{variational principle} that we can use to organize our calculations, so let us explain that first.\index{nearly-Gaussian distribution!entropy}

As in the last section, we'll focus on the distribution of $m$ preactivations in layer $\ell$, drop (almost) all the \terminate{layer indices}, and focus exclusively on a single input $x$.
As we have been doing since the beginning of time (\S\ref{sec:perturbation}), let's express the probability distribution in terms of an \terminate{action} as
\be\label{eq:probability-action-appendix}
p(z_1,\ldots, z_{m}\vert x)=\frac{e^{-\ac\le(z_1,\ldots, z_{m}\ri)}}{Z} \, ,
\ee
which must be normalized by the \neo{partition function} 
\be
Z=\int\le[\prod_{i=1}^{m}dz_i\ri] e^{-\ac\le(z_1,\ldots, z_{m}\ri)}\, .
\ee
In terms of these 
quantities, the entropy~\eqref{eq:entropy-shannon-formula-integral} can be expressed as
\be\label{eq:entropy-action-representation}
\entropy\!\le[p\!\le(z_1,\ldots, z_{m}\vert x\ri)\ri]=\log(Z)+\frac{1}{Z} \int\le[\prod_{i=1}^{m}dz_i\ri] e^{-\ac\le(z_1,\ldots, z_{m}\ri)} \ac(z_1,\ldots, z_{m})\, .
\ee
Just to make sure, please don't get confused between the action $\ac(z)$, which is a function of the preactivations, and the entropy $\entropy\!\le[p(z)\ri]$, which is a functional of the probability distribution.\footnote{
    For choices of units of the preactivations $z$ for which the partition function is unity, $Z=1$, the action is simply the surprisal\index{surprisal (information theory)} \eqref{eq:surprisal-def}, and thus the entropy is the expectation of the action: $\entropy\!\le[p(z)\ri] = \E{\ac(z)}$ . Note that in \terminate{thermodynamics}, the constant $-\log Z$ is sometimes called the \neo{free energy}; this discussion should make clear that only its relative value for two different distributions is physical.
}

To proceed, let's adopt a \term{variational ansatz}: we'll divide the action into two parts as
\be\label{eq:variational-action}
\ac(z_1,\ldots, z_{m})=\acfree(z_1,\ldots, z_{m})+\acvar(z_1,\ldots, z_{m})\, ,
\ee
with the idea being that the second term, $\acvar$, encodes the part of the distribution that is perturbatively small.
Specifically, the \term{variational principle} instructs us to choose the first term in \eqref{eq:variational-action}, $\acfree(z)$,  that gives no variations of the entropy with respect to $\acvar(z)$:
\be\label{eq:variational-principle-equation}
0=\frac{\delta \entropy\!\le[p(z) \ri]}{\delta \acvar(z)}\bigg\vert_{\acvar(z)=0}  \, .
\ee
We'll satisfy this shortly.
Additionally, $\acvar(z)$ should not be completely arbitrary, but instead be constructed to properly reflect the statistics of the preactivation distribution $p(z_1,\ldots, z_{m}\vert x)$.
The first such constraint comes from demanding that the two-point correlator, when computed with the variational action, is determined by the \emph{exact} single-input metric $G$:
\be\label{eq:variational-constraint}
\E{z_{i_1}z_{i_2}}\equiv \delta_{i_1i_2} G \,.%
\ee
The second constraint comes from demanding that the connected four-point correlator, when computed with the variational action, is determined by the \emph{exact} single-input \terminate{four-point vertex}:
\begin{align}\label{eq:variational-constraint-2}
\E{z_{i_1}z_{i_2}z_{i_3}z_{i_4}}\vert_{\text{connected}} = \frac{1}{n} \le(\delta_{i_1i_2}\delta_{i_3 i_4} +\delta_{i_1i_3}\delta_{i_2 i_4}+\delta_{i_1i_4}\delta_{i_2 i_3} \ri) V\, .
\end{align}
Together, the constraints \eqref{eq:variational-constraint} and \eqref{eq:variational-constraint-2} will fix the couplings of the variation action $\acvar$ so that the full action \eqref{eq:variational-action} correctly specifies the $m$-neuron preactivation distribution \eqref{eq:probability-action-appendix}.\footnote{In principle there are additional constraints coming from the statistics of higher-point correlators, but their contribution is  subleading to both the leading and next-to-leading orders that we will work.}

To understand why we're doing this, note that our \terminate{variational principle} is ultimately just a realization of the \textbf{principle of maximum entropy}\index{maximum entropy, principle} for nearly-Gaussian distributions\index{nearly-Gaussian distribution}.\footnote{For those readers that enjoy our historical asides, the maximum entropy principle was discovered by Jaynes\index{Jaynes, Edwin T.} 
    and provides a link between \neo{information theory} on the one hand and \emph{statistical mechanics}\index{statistical physics} on the other hand \cite{PhysRev.106.620,PhysRev.108.171}.
As an example of this, consider a central problem in statistical mechanics: find the probability
distribution $p_i$ for the fundamental \emph{microstates}\index{microstate (statistical mechanics)} $i$ of a system that has a macroscopic average energy $\overline{E}\equiv \E{E_i}=\sum_{i} p_i E_i$. An application of the principle of maximum entropy then correctly picks out the  \neo{Boltzmann distribution} (or sometimes, the \emph{Gibbs distribution}\index{Gibbs distribution|see{Boltzmann distribution}})  of statistical mechanics\index{statistical physics}
\be\label{eq:boltzmann-distribution}
p_i \propto e^{- \beta E_i} \, ,
\ee
if we optimize the entropy \eqref{eq:entropy-shannon-formula-sum} subject to the observable constraint for the energy, $\sum_{i} p_i E_i=\overline{E}$, and the normalization condition for the distribution, $\sum_{i} p_i=1$.
Here, $\beta$ is a Lagrange multiplier that depends on the energy $\overline{E}$ and has a physical interpretation as the inverse temperature, $\beta=1/(k_{\text{B}}T)$, with the aforementioned \terminate{Boltzmann constant} $k_{\text{B}}$ and the familiar-to-everyone \terminate{temperature} $T$. This example also shows how statistical mechanics links the details of the fundamental microstates $i$ to the macroscopic thermodynamic variables such as $\overline{E}$ and $T$.
\label{foot:gibbs}
} In particular, first note that the Gaussian distribution itself can be derived as a distribution that maximizes the entropy \eqref{eq:entropy-shannon-formula-integral}, subject to the constraints of fixed first and second moment.\footnote{
    For those of you keeping track: the maximum entropy distribution with no information fixed is the \terminate{uniform distribution}, cf.~\eqref{eq:entropy-maximal}; the maximum entropy distribution with a fixed first moment is the \terminate{Boltzmann distribution}, cf.~\eqref{eq:boltzmann-distribution}; and the maximum entropy distribution with a fixed first and second moment is the \terminate{Gaussian distribution}, cf.~(nearly-)everywhere.
} As we will see in a moment, in \eqref{eq:variational-principle-equation} we are maximizing the entropy of the distribution with respect to the deformation\index{deformation!Gaussian distribution} of the action away from Gaussianity, $\acvar$, subject to the constraint of fixing the higher-order cumulants\index{cumulant} order by order in perturbation theory. In general, the maximum entropy principle is an appropriate procedure when we have fixed observable information for which we want to find an underlying distribution. Here, we actually know the distribution that produces $G$ and $V$, \eqref{eq:probability-action-appendix}, but we can still use this principle as convenient tool for computing the entropy.\footnote{
    In particular, this procedure will organize our perturbative computation of the entropy and ensure that we only need to compute Gaussian expectations of the form \eqref{eq:multi-neuron-expectation-reprint}.
}

Now, to satisfy the variational principle
 \eqref{eq:variational-principle-equation}, let's choose
\be\label{eq:variational-free-choice}
\acfree(z_1,\ldots, z_{m})=\frac{1}{2G}\sum_{i=1}^m z_i^2\, .
\ee
Importantly, $G$ is 
not just the inverse of the quadratic coefficient in the action $\ac(z)$, but instead is the \emph{exact} two-point correlator that we would actually measure, \eqref{eq:variational-constraint},
incorporating the full series of corrections due to the interactions, cf.~\eqref{eq:self-energy-decomposition}.\footnote{
    This will let us express the entropy in terms of these measurable quantities and in no way will we need to actually compute any of the corrections in this series.
} 
To see why such a choice satisfies the variational principle, let us start by rewriting expectations with respect to the full distribution \eqref{eq:probability-action-appendix} in terms of simpler \index{Gaussian expectation}\emph{Gaussian} expectations taken with respect to a zero-mean Gaussian distribution with the same variance \eqref{eq:variational-constraint}: 
\be
\bra\!\bra z_{i_1}z_{i_2}\ket\!\ket_{G} = \delta_{i_1i_2} G \, .
\ee
Here, please recall 
our notation $\bra\!\bra \cdot \ket\!\ket_G$ for a \terminate{Gaussian expectation} of a multi-neuron function with variance $\delta_{i_1i_2} G$, 
\be\label{eq:multi-neuron-expectation-reprint}
\bra\!\bra f(z_1, \ldots, z_m) \ket\!\ket_G \equiv \frac{1}{Z_{G}} \int\le[\prod_{i=1}^{m}dz_i\ri] e^{-\frac{1}{2G}\sum_{i=1}^{m}z_{i}^2} f(z_1, \ldots, z_m) \,,
\ee
and note also that such a Gaussian distribution will require a different partition function
\be
Z_{G}\equiv \int\le[\prod_{i=1}^{m}dz_i\ri] e^{-\frac{1}{2G}\sum_{i=1}^{m}z_{i}^2}=\le(2\pi G\ri)^{\frac{m}{2}}\, ;
\ee
importantly, $Z \neq Z_G$. Next, let us
rewrite the entropy~\eqref{eq:entropy-action-representation} in terms of these simpler expectations as
\begin{align}\label{eq:entropy-action-variational-representation}
\entropy\!\le[p\!\le(z_1,\ldots, z_{m}\vert x\ri)\ri]=&\log Z +\E{\frac{1}{2G}\sum_{i=1}^{m}z_{i}^2}+\frac{1}{Z} \int\le[\prod_{i=1}^{m}dz_i\ri] e^{-\frac{1}{2G}\sum_{i=1}^{m}z_{i}^2}e^{-\acvar} \acvar\, \notag\\
=&\frac{m}{2}\log(2\pi G)+\log\!\le(\frac{Z}{Z_G}\ri)+\frac{m}{2}+\le(\frac{Z}{Z_G}\ri)^{-1}\brabra e^{-\acvar} \acvar\ketket_{G} \, ,
\end{align}
where in the first equality we just plugged in \eqref{eq:variational-action}, and in the second equality 
we rewrote all the full expectations in terms of the simpler Gaussian expectations using~\eqref{eq:multi-neuron-expectation-reprint}.
Then, by Taylor-expanding in $\acvar$, we can evaluate the ratio of partition functions as
\begin{align}\label{eq:variational-evaluation}
&\frac{Z}{Z_{G}}=\frac{1}{Z_{G}} \int\le[\prod_{i=1}^{m}dz_i\ri] e^{-\frac{1}{2G}\sum_{i=1}^{m}z_{i}^2}\le(e^{-\acvar} \ri)=\brabra e^{-\acvar}\ketket_{G}\, \\
=&1-\bra\!\bra \acvar\ket\!\ket_{G}+\frac{1}{2}\brabra \acvar^2\ketket_{G}-\frac{1}{6}\brabra \acvar^3\ketket_{G}+\o{\acvar^4}\, ,\notag
\end{align}
and similarly we can evaluate the needed Gaussian expectation as
\be
\brabra e^{-\acvar} \acvar\ketket_{G}=\bra\!\bra \acvar\ket\!\ket_{G}-\brabra \acvar^2\ketket_{G}+\frac{1}{2}\brabra \acvar^3\ketket_{G}+\o{\acvar^4}\, .
\ee
Plugging these back into the variational expression for the entropy~\eqref{eq:entropy-action-variational-representation} and organizing terms, for which you might find $\log(1+x)=x-\frac{x^2}{2}+\frac{x^3}{3}+\ldots$ and $1/(1+x)=1-x+x^2-x^3+\ldots$ helpful,  we get
\begin{align}\label{eq:entropy-action-variational-representation-practical}
\entropy\!\le[p\!\le(z_1,\ldots, z_{m}\vert x\ri)\ri]=&\frac{m}{2}\log(2\pi e G)-\frac{1}{2}\le[\brabra \acvar^2\ketket_{G}-\bra\!\bra \acvar\ket\!\ket_{G}^2\ri]\, \\
&+\frac{1}{3}\le[\brabra \acvar^3\ketket_{G}-3\brabra \acvar^2\ketket_{G}\bra\!\bra \acvar\ket\!\ket_{G}+2\bra\!\bra \acvar\ket\!\ket_{G}^3\ri]+\o{\acvar^4} \, .\notag
\end{align}
First, note that the first term is exactly the  \terminate{entropy} for a multivariate Gaussian distribution with a covariance $\delta_{i_1i_2}G$, cf.~\eqref{eq:entropy-preactivation-infinite-width-multi}.  Second, and most importantly, note that the would-be linear term proportional to $\bra\!\bra \acvar\ket\!\ket_{G}$ exactly cancelled out. In other words, our ansatz\index{variational ansatz} for the decomposition of the action \eqref{eq:variational-action}  with the choice \eqref{eq:variational-free-choice} automatically satisfies the \terminate{variational principle} \eqref{eq:variational-principle-equation}.

Finally, let us note in passing that the leading correction coming from the quadratic term is definitively negative.\footnote{To see this, notice that the expression in the square brackets of the quadratic term is the variance of the variational part of the action, and thus is positive: $\bra\!\bra \acvar^2\ket\!\ket_{G}-\bra\!\bra \acvar\ket\!\ket_{G}^2=\langle\!\langle\le(\acvar-\bra\!\bra \acvar\ket\!\ket_{G}\ri)^2\rangle\!\rangle_{G}\geq 0$.
}
This negativity is actually necessary for mathematical consistency: as the \terminate{Gaussian distribution} maximizes the entropy of any set of random variables with known and fixed first and second moments, any deformation\index{deformation!Gaussian distribution} of a distribution away from Gaussianity, while also respecting such moment constraints, must necessarily have less entropy.
In the current case, our variational ansatz \eqref{eq:variational-action} gives a nearly-Gaussian deformation of a zero-mean Gaussian distribution.\footnote{
    N.B.~the terms in the second set of the square brackets in \eqref{eq:entropy-action-variational-representation-practical} are necessary for computing the next-to-leading-order correction.
}

Now that we're done passing notes,
let's satisfy our constraints, \eqref{eq:variational-constraint} and \eqref{eq:variational-constraint-2}, and then use our variational expression \eqref{eq:entropy-action-variational-representation-practical} to compute the \terminate{entropy} and \terminate{mutual information} of the preactivations in a finite-width network.

\subsubsection{Leading-Order Correction: Nonzero Mutual Information}
At leading order, we've already worked out how to relate the couplings of the variational action $\acvar$ to the single-input metric $G$ and the single-input four-point correlator $V$.
Recall from our discussion of the running couplings\index{running coupling} when integrating out neurons in~\S\ref{sec:sum-rule} that the leading correction to the action was given by
\be\label{eq:variational-action-first-order}
\acvar\!\le(z_1,\ldots, z_{m}\ri)=\frac{1}{2}\le(g_{m}-\frac{1}{G}\ri)\sum_{i=1}^m z_i^2-\frac{v_m}{8}\sum_{i,j=1}^m z_i^2z_j^2+\o{\frac{1}{n^2}} \, ,
\ee
where 
the running quadratic coupling\index{running coupling!quadratic} was given by ~\eqref{eq:quadratic-reprint-m-emphasis}, 
\be\label{eq:quadratic-reprint-m-emphasis-reprint}
g_{m}=\frac{1}{G}+\le(\frac{m+2}{2}\ri)\frac{V}{n G^3 } +\o{\frac{1}{n^2}} \, ,
\ee
and the running quartic coupling\index{running coupling!quartic} was given by~\eqref{eq:quartic-single-input-coupling-for-vertex},
\be\label{eq:quartic-single-input-coupling-for-vertex-reprint}
v_{m}= \frac{V}{n G^4 }+\o{\frac{1}{n^2}}\, .
\ee
To get the expression \eqref{eq:variational-action-first-order}, look at our expression for the distribution of $m$ preactivations, \eqref{eq:m-neuron-action}, and then rearrange \eqref{eq:variational-action}  with \eqref{eq:variational-free-choice} to solve for $\acvar$.
If you don't recall how to get \eqref{eq:quadratic-reprint-m-emphasis-reprint} and \eqref{eq:quartic-single-input-coupling-for-vertex-reprint},  feel free to flip back and reread the last subsubsection of \S\ref{sec:sum-rule}, or feel free to flip forward to the next subsection where we'll have to derive these expressions again to higher order in the \terminate{$1/n$ expansion}, cf.~\eqref{eq:quadratic-reprint-m-emphasis-refined} and~\eqref{eq:quartic-single-input-coupling-for-vertex-refined}.

Given that this leading-order variational correction to the action $\acvar$~\eqref{eq:variational-action-first-order} now satisfies the constraints \eqref{eq:variational-constraint} and \eqref{eq:variational-constraint-2}, we can now evaluate the leading correction to the \terminate{entropy}~\eqref{eq:entropy-action-variational-representation-practical}:%
\begin{align}\label{eq:first-and-last-unrolling}
&\brabra \acvar^2\ketket_{G}-\bra\!\bra \acvar\ket\!\ket_{G}^2\, \\
=&\frac{1}{4}\le(g_{m}-\frac{1}{G}\ri)^2\sum_{i_1,i_2=1}^m\le[\brabra z_{i_1}^2 z_{i_2}^2 \ketket_{G}-\brabra z_{i_1}^2\ketket_{G}\brabra z_{i_2}^2 \ketket_{G}\ri]\notag\\
&-\frac{1}{8}\le(g_{m}-\frac{1}{G}\ri)v_m\sum_{i_1,i_2,i_3=1}^m \le[\brabra z_{i_1}^2 z_{i_2}^2 z_{i_3}^2 \ketket_{G}-\brabra z_{i_1}^2 z_{i_2}^2\ketket_{G}\brabra z_{i_3}^2 \ketket_{G}\ri]\, \notag\\
&+\frac{1}{64}v_m^2 \sum_{i_1,i_2,i_3,i_4=1}^m\le[\brabra z_{i_1}^2 z_{i_2}^2 z_{i_3}^2z_{i_4}^2 \ketket_{G}-\brabra z_{i_1}^2 z_{i_2}^2\ketket_{G}\brabra z_{i_3}^2z_{i_4}^2 \ketket_{G}\ri]+\o{\frac{1}{n^3}}\, \notag\\
=&\frac{m}{2}\le(g_{m}-\frac{1}{G}\ri)^2G^2\!-\frac{m(m+2)}{2}\le(g_{m}-\frac{1}{G}\ri)v_mG^3\!+\frac{m(m+2)(m+3)}{8}v_m^2 G^4\!+\o{\frac{1}{n^3}}\, .\notag
\end{align}
Here, in going from the second line to the last line, you may find this formula for these Gaussian expectations useful:
\be\label{eq:combinatorial-2m-resurrection}
\sum_{i_1,\ldots,i_{k}=1}^m\brabra z_{i_1}^2 \cdots z_{i_k}^2 \ketket_{G}=m(m+2)\cdots [m+2(k-1)] \, G^k \, ,
\ee
which is akin to~\eqref{eq:combinatorial-2m}
and will be  used again in the next subsubsection repeatedly, up to $k=6$.
To complete the computation, plug in the quadratic coupling, \eqref{eq:quadratic-reprint-m-emphasis-reprint}, and the quartic coupling, \eqref{eq:quartic-single-input-coupling-for-vertex-reprint}, into~\eqref{eq:first-and-last-unrolling}, giving 
\be
\brabra \acvar^2\ketket_{G}-\bra\!\bra \acvar\ket\!\ket_{G}^2
=\frac{m(m+2)}{8}\le(\frac{V}{n G^2 }\ri)^2+\o{\frac{1}{n^3}}\, ,
\ee
for the variance of the variational action.
Therefore, the entropy~\eqref{eq:entropy-action-variational-representation-practical} is given by
\be\label{eq:entropy-second-order}
\entropy\!\le[p\!\le(z_1,\ldots, z_{m}\vert x\ri)\ri]=\frac{m}{2}\log(2\pi e G)-\frac{\le(m^2+2m\ri)}{16}\le(\frac{V}{n G^2 }\ri)^2+\o{\frac{1}{n^3}}\, ,
\ee
exhibiting a nontrivial correction at finite width.

Let us reflect on this formula by making some comments. First, note that the correction is definitely negative, as we pointed out before: recall that the Gaussian distribution maximizes the entropy of a set of random variables with a fixed covariance, and since our first variational constraint \eqref{eq:variational-constraint} fixes the two-point correlator of the preactivations, the entropy for the nearly-Gaussian distribution \eqref{eq:probability-action-appendix} must be less than the entropy of a Gaussian distribution with the same variance. Second, note that unlike all our previous results, the leading correction here is  \emph{second order} in the inverse layer width as $\sim V^2/n^2$ and  \emph{not} $\sim V/n$. Indeed, since the quartic coupling\index{coupling!quartic} $v_m$ in a generic nearly-Gaussian action can take either sign -- corresponding to a distribution with either fat tails or thin tails -- the leading correction to the entropy must be proportional to the minus the \emph{square} of the quartic coupling, $v_m^2 \propto V^2/n^2$, in order to guarantee that the entropy decreases.\footnote{This same argument excludes a contribution of the form $\o{\SPC_m} = \o{\SPV/n^2}$ coming from the sextic coupling\index{coupling!sextic}, cf.~\eqref{eq:variational-action-second-order}, and similarly excludes any linear contributions from other higher-order couplings.
 }
Finally, we see that this correction breaks the perfect additivity\index{entropy!additivity} for the neurons that we found in the infinite-width limit \eqref{eq:entropy-preactivation-infinite-width-multi}. In particular, although the decrease in the entropy is perturbatively small with an order $1/n^2$ scaling, it also depends \emph{quadratically} on $m$. This 
nonlinearity
in the number of neurons signals the presence of nontrivial interactions at finite width.

Accordingly, we can characterize this \terminate{statistical dependence} by computing \terminate{mutual information} between two non-overlapping sets of neurons, $\mathcal{M}_1=\le\{1,\ldots,m_1\ri\}$ and $\mathcal{M}_2=\le\{m_1+1,\ldots,m_1+m_2\ri\}$, %
\begin{align}\label{eq:MIT-second-order}
\MI\le[p\!\le( \mathcal{M}_1 , \mathcal{M}_2 \vert x\ri)\ri]\equiv&\entropy\!\le[p\!\le(\mathcal{M}_1 \vert x\ri)\ri]+\entropy\!\le[p\!\le(\mathcal{M}_2 \vert x\ri)\ri]-\entropy\!\le[p\!\le(\mathcal{M}_1 , \mathcal{M}_2 \vert x\ri)\ri]\, \\
=&\frac{m_1m_2}{8}\le[\frac{V^{(\ell)}}{n_{\ell-1} \le(G^{(\ell)}\ri)^2 }\ri]^2+\o{\frac{1}{n^3}}\, ,\notag
\end{align}
where we used our entropy formula \eqref{eq:entropy-second-order} in three different ways and also restored layer indices 
to better interpret this formula.
This nonzero mutual information signifies that -- at finite width, \emph{only} --  for a given layer $\ell$, observing the activities of a group of neurons $\mathcal{M}_1$ will convey information about the activities of another group of neurons $\mathcal{M}_2$. This can be thought of as an information-theoretic reformulation and generalization of the \neo{Hebbian learning} principle that we saw for the conditional variance \eqref{eq:conditional-variance} in \S\ref{subsec:Hebbian}: there we more simply saw that the variance of one neuron $z_2^{(\ell)}$, conditioned on an atypical observation of a second neuron, $z_1^{(\ell)} = \check{z}_1^{(\ell)}$, will itself be atypical; here, we can directly characterize how much one group of neurons can know about another non-overlapping group.\footnote{
    More generally, it would be interesting to work out the mutual information for multiple inputs in order to understand its data dependence. In that case, we expect it to be mediated by the multi-input four-point vertex and depend on the details of different groupings of four samples from the dataset\index{input data} $\D$.
}

Finally, remembering our \neo{scaling law} for the normalized vertex \eqref{eq:k-star-equals-zero-normalized-four-point-scaling-law}, we see that the \terminate{mutual information} \eqref{eq:MIT-second-order} scales with the depth $\ell$ of the hidden layer \emph{squared}: 
\be\label{eq:MIT-second-order-depth}
\MI\le[p\!\le( \mathcal{M}_1 , \mathcal{M}_2 \vert x\ri)\ri] \propto \ell^2/n^2 \, . 
\ee
In terms of \emph{RG flow}\index{representation group flow}, this means that the mutual information is \emph{relevant}\index{relevant (RG flow)}, suggesting that a growing mutual information is helpful when coarse-graining representations: as the fine-grained features are marginalized over, deeper hidden layers will have a growing set of correlations between groups of neurons. In other words, by increasing the size of the nonlinear subadditive term in the entropy, this reduces the number of independently available degrees of freedom in these deeper layers.\footnote{
    It would be interesting to try to interpret this in terms of the \neo{optimal brain damage} of \cite{brain-damage} or the \neo{lottery ticket hypothesis} of \cite{frankle2018the}.
}

\subsubsection{NLO Correction: Optimal Aspect Ratio and Tripartite Information}
Our leading-order result for the finite-width mutual information, \eqref{eq:MIT-second-order-depth}, naively grows in depth without bounds, suggesting that deeper is always better. Of course, if the depth becomes too large, this naive answer breaks down as the higher-order terms in the perturbation series start to become important. To understand the mutual information at even greater depths, we'll need to compute the \emph{next-to-leading-order} (NLO) correction.\index{entropy!next-to-leading-order correction}\index{mutual information!next-to-leading-order correction}

To push our calculations to the next level, we need to ensure that our two variational constraints, \eqref{eq:variational-constraint} and \eqref{eq:variational-constraint-2}, are satisfied to next-to-leading-order, $\o{1/n^2}$. In principle, this means that we will need to also include an $\o{1/n^2}$ \emph{sextic} term in our variational action $\acvar$ as
\be\label{eq:variational-action-second-order}
\acvar\!\le(z_1,\ldots, z_{m}\ri)=\frac{1}{2}\le(g_{m}-\frac{1}{G}\ri)\sum_{i=1}^m z_i^2-\frac{1}{8}v_m\sum_{i,j=1}^m z_i^2z_j^2+\frac{1}{24}\SPC_m\sum_{i,j,k=1}^m z_i^2z_j^2z_k^2+\o{\frac{1}{n^3}}\, .
\ee
Such a sextic term was originally introduced in \eqref{eq:general-L-action-sixth}; here, we've specialized it to focus on $m$ preactivations in a layer $\ell$, with $\SPC_m$ the running sextic coupling\index{running coupling!sextic}.

Now, let's explain how to explicitly satisfy the variational constraints.
First, just as we did before for the \terminate{entropy} in \eqref{eq:entropy-action-variational-representation-practical}, 
note that for a general observable we can express its full expectation in terms of simpler Gaussian expectations as
\begin{align}\label{eq:observable-with-variation}
\E{\O}\equiv&\frac{1}{Z}\int\le[\prod_{i=1}^{m}dz_i\ri] e^{-\ac}\O=\le(\frac{Z}{Z_G}\ri)^{-1}\brabra e^{-\acvar}\O\ketket_{G}\, \\
=&\bra\!\bra \O\ket\!\ket_{G}-\Big[\bra\!\bra \O\acvar\ket\!\ket_{G}-\bra\!\bra \O\ket\!\ket_{G}\bra\!\bra \acvar\ket\!\ket_{G}\Big]\, \notag\\
&+\frac{1}{2}\le[\brabra \O\acvar^2\ketket_{G}-2\bra\!\bra \O\acvar\ket\!\ket_{G}\bra\!\bra\acvar\ket\!\ket_{G}-\bra\!\bra \O\ket\!\ket_{G}\brabra \acvar^2\ketket_{G}+2\bra\!\bra \O\ket\!\ket_{G}\bra\!\bra\acvar\ket\!\ket_{G}^2\ri]+\o{\frac{1}{n^3}}\,, \notag
\end{align}
where in the last equality, we expanded $\brabra e^{-\acvar}\O\ketket_{G}$ in $\acvar$ and 
also
used the formula that we already evaluated  in \eqref{eq:variational-evaluation} for the ratio of partition functions. %
Physically, the first square brackets says that in an interacting theory\index{interacting theory!variational method}, the leading variational correction to an observable  is given by the correlation of the observable with the variational part of the action. 

To satisfy our first constraint for the metric, \eqref{eq:variational-constraint}, we can use this general expression \eqref{eq:observable-with-variation} with the observable 
\be
\O\equiv \frac{1}{m}\sum_{i=1}^m z_i^2 \,,
\ee 
for which we can use our constraint \eqref{eq:variational-constraint} to easily see that $\E{\O}=G$. Evaluating this same expectation using the variational sextic action, \eqref{eq:variational-action-second-order}, we find
\begin{align}\label{eq:metric-shifts-nlo-variational}
G=&\frac{1}{m}\sum_{i=1}^m\E{z_i^2}\, \\
=&G
-\le(g_m-\frac{1}{G}\ri)G^2
+\frac{(m+2)}{2}v_m G^3
+\le(g_m-\frac{1}{G}\ri)^2\!G^3
-\frac{(m+2)(m+4)}{4}\SPC_m G^4
\, \notag\\
&-\frac{3(m+2)}{2}\le(g_m-\frac{1}{G}\ri)v_m G^4
+\frac{(m+2)(m+3)}{2}v_m^2 G^5
+\o{\frac{1}{n^3}}\, .\notag
\end{align}
In evaluating this, you  might again find the formula \eqref{eq:combinatorial-2m-resurrection} helpful. This expression, \eqref{eq:metric-shifts-nlo-variational},  shows how the interacting theory modifies the two-point correlator.
Rearranging terms to solve for $g_m$ order by order, we find an NLO version of~\eqref{eq:quadratic-reprint-m-emphasis-reprint}, which determines the variational quadratic coupling in terms of the metric and the other higher-order couplings: 
\be\label{eq:quadratic-reprint-m-emphasis-refined}
g_m = \frac{1}{G}+\frac{(m+2)}{2}v_mG-\frac{(m+2)(m+4)}{4}\SPC_m G^2+\frac{(m+2)}{2} v_m^2 G^3 +\o{\frac{1}{n^3}}\, .
\ee

Next, to satisfy the second constraint for the four-point vertex, \eqref{eq:variational-constraint-2}, we can use our general expression \eqref{eq:observable-with-variation} with the observable 
\be
\O \equiv \frac{1}{m(m+2)}\sum_{i,j=1}^m z_i^2z_j^2 \, ,
\ee
for which we can use both our constraints \eqref{eq:variational-constraint} and \eqref{eq:variational-constraint-2} to show that $\E{\O}= G^2+V/n$.
Now evaluating $\O$ according to the variational sextic action, \eqref{eq:variational-action-second-order}, we get
\begin{align}
G^2+\frac{V}{n}=&\frac{1}{m(m+2)}\sum_{i,j=1}^m\E{z_i^2z_j^2}\, \\
=&G^2
-2\le(g_m-\frac{1}{G}\ri)G^3
+3\le(g_m-\frac{1}{G}\ri)^2\!G^4
+(m+3)v_m G^4
-\frac{(m+4)^2}{2}\SPC_m G^5
\, \notag\\
&-4(m+3)\le(g_m-\frac{1}{G}\ri)v_m G^5
+\frac{(5m^2+34m+60)}{4}v_m^2 G^6
+\o{\frac{1}{n^3}}\, \notag\\
=&G^2+v_mG^4-(m+4)\SPC_m G^5+\frac{(m+8)}{2}v_m^2G^6+\o{\frac{1}{n^3}}\, ,
\end{align}
where to get to the second line we again made heavy use of our formula \eqref{eq:combinatorial-2m-resurrection}, and to get to the final line we plugged in \eqref{eq:quadratic-reprint-m-emphasis-refined} for the quadratic coupling. 
Rearranging terms to solve for $v_m$ order by order, we find an NLO version of~\eqref{eq:quartic-single-input-coupling-for-vertex-reprint}, which determines the variational quartic coupling in terms of the two constraints and the sextic coupling: 
\be\label{eq:quartic-single-input-coupling-for-vertex-refined}
v_m =\frac{V}{nG^4}-\frac{(m+8)}{2}\le(\frac{V}{nG^3}\ri)^2+(m+4)\SPC_m G+\o{\frac{1}{n^3}}\, .
\ee

We now finally have all the pieces we need to evaluate the NLO correction to the \terminate{entropy} \eqref{eq:entropy-action-variational-representation-practical}.\footnote{
    In principle,  at this order we would need to continue and find an expression for $u_m$ in terms of an additional six-point vertex constraint, but as we will soon see, $u_m$ doesn't factor into any of our expressions for the mutual information. %
    This is the reason why we are able to analyze these next-to-leading-order corrections without otherwise evaluating  additional MLP recursions.
}
First, let's  reevaluate the variance of the variational action to NLO by
repeated use of the formula~\eqref{eq:combinatorial-2m-resurrection},
\begin{align}\label{eq:not-too-bad-1}
&\brabra \acvar^2\ketket_{G}-\bra\!\bra \acvar\ket\!\ket_{G}^2\, \\
=&\frac{m}{2}\le(g_{m}-\frac{1}{G}\ri)^2G^2\!-\frac{m(m+2)}{2}\le(g_{m}-\frac{1}{G}\ri)v_mG^3\!+\frac{m(m+2)(m+3)}{8}v_m^2 G^4\, \notag\\
&+\frac{m(m+2)(m+4)}{4}\le(g_{m}-\frac{1}{G}\ri)\SPC_mG^4-\frac{m(m+2)(m+4)^2}{8}v_m \SPC_m G^5+\o{\frac{1}{n^4}}\, \notag\\
=&\frac{m(m+2)}{8}\le[\le(v_mG^2\ri)^2-2(m+4)\le(v_mG^2\ri)\le(\SPC_mG^3\ri)\ri]+\o{\frac{1}{n^4}}\, \notag\\
=&\frac{m(m+2)}{8}\le[\le(\frac{V}{nG^2}\ri)^2-(m+8)\le(\frac{V}{nG^2}\ri)^3\ri]+\o{\frac{1}{n^4}}\, .\notag
\end{align}
Here, to get to the penultimate line we used our NLO expression for the quadratic coupling~\eqref{eq:quadratic-reprint-m-emphasis-refined}, and then to get to the final line we used our NLO expression for the quartic coupling~\eqref{eq:quartic-single-input-coupling-for-vertex-refined}.
Second, let's similarly evaluate the subleading term in our expression \eqref{eq:entropy-action-variational-representation-practical} for the  \terminate{entropy}:
\begin{align}\label{eq:not-too-bad-2}
&\brabra \acvar^3\ketket_{G}-3\brabra \acvar^2\ketket_{G}\bra\!\bra \acvar\ket\!\ket_{G}+2\bra\!\bra \acvar\ket\!\ket_{G}^3\, \\
=&m\le(g_{m}-\frac{1}{G}\ri)^3G^3-\frac{9m(m+2)}{4}\le(g_{m}-\frac{1}{G}\ri)^2v_mG^4\, \notag\\
&+\frac{3m(m+2)(m+3)}{2}\le(g_{m}-\frac{1}{G}\ri)v_m^2G^5-\frac{m(m+2)(5m^2+34m+60)}{16}v_m^3 G^6+\o{\frac{1}{n^4}}\, \notag\\
=&-\frac{m(m+2)(m+8)}{8}\le(v_mG^2\ri)^3+\o{\frac{1}{n^4}}\, \notag\\
=&-\frac{m(m+2)(m+8)}{8}\le(\frac{V}{nG^2}\ri)^3+\o{\frac{1}{n^4}}\, .\notag
\end{align}
Putting \eqref{eq:not-too-bad-1} and \eqref{eq:not-too-bad-2} back into our variational expression for the \terminate{entropy} \eqref{eq:entropy-action-variational-representation-practical}, we finally arrive at
\begin{align}\label{eq:entropy-third-order}
&\entropy\!\le[p\!\le(z_1,\ldots, z_{m}\vert x\ri)\ri]\\
=&\frac{m}{2}\log(2\pi e G)-\frac{\le(m^2+2m\ri)}{16}\le(\frac{V}{n G^2 }\ri)^2+\frac{\le(m^3+10m^2+16m\ri)}{48}\le(\frac{V}{n G^2 }\ri)^3+\o{\frac{1}{n^4}}\, .\notag
\end{align}
This result 
in turn lets us compute the NLO correction to the mutual information between two sets of neurons, $\mathcal{M}_1=\le\{1,\ldots,m_1\ri\}$ and $\mathcal{M}_2=\le\{m_1+1,\ldots,m_1+m_2\ri\}$:
\begin{align}\label{eq:MIT-third-order}
&\MI\le[p\!\le( \mathcal{M}_1 , \mathcal{M}_2 \vert x\ri)\ri] \, \\
\equiv&\entropy\!\le[p\!\le(\mathcal{M}_1 \vert x\ri)\ri]+\entropy\!\le[p\!\le(\mathcal{M}_2 \vert x\ri)\ri]-\entropy\!\le[p\!\le(\mathcal{M}_1 , \mathcal{M}_2 \vert x\ri)\ri]\, \notag \\
=&\frac{m_1m_2}{8}\le[\frac{ V^{(\ell)} }{ n_{\ell-1} \le(G^{(\ell)}\ri)^2} \ri]^2 -\frac{m_1 m_2(20+3m_1+3m_2)}{48}\le[\frac{ V^{(\ell)} }{ n_{\ell-1} \le(G^{(\ell)}\ri)^2} \ri]^3 +\o{\frac{1}{n^4}} \, \notag,
\end{align}
where once again we have restored  layer indices on the metric, the four-point vertex, and the layer width.
Note that as promised, the sextic coupling $\SPC_m$ dropped out in the end.\footnote{
This is another realization of the principle of maximum entropy\index{maximum entropy, principle} for nearly-Gaussian distributions. 
In particular, the entropy for a zero-mean distribution that satisfies constraints 
fixing its two-point correlator and its connected four-point correlator -- and otherwise leaves unfixed its connected six-point correlator -- 
is given to order $1/n^3$ by our expression \eqref{eq:entropy-third-order}. Thus, if we did add a third constraint that fixed the connected six-point correlator, with $\SPV = \o{1/n^2}$, then any terms of the form $\o{v_m \SPC_m } = \o{\SPV V/n^3}$ cannot appear in \eqref{eq:entropy-third-order}, as they could \emph{increase} the entropy depending on the sign of $\SPV V$.
}

Excitingly, these two terms have opposite signs.
To explain our excitement, let us plug in our scaling solution for the normalized \terminate{four-point vertex}  \eqref{eq:k-star-equals-zero-normalized-four-point-scaling-law} evaluated at the final layer $\ell=L$:
\be\label{eq:scaling-solution-reprinted-for-MIT}
\frac{ V^{(L)} }{ n_{L-1} \le(G^{(L)}\ri)^2} \equiv \nu r \, .
\ee
Here, $r \equiv L/n$ is the overall depth-to-width aspect ratio of the network, and $\nu$ is an activation-function dependent constant: for the $K^\star=0$ universality\index{universality class!K@$K^\star=0$}, we have~\eqref{eq:k-star-equals-zero-normalized-four-point-tanh-univ}
\be\label{eq:tanh-nu}
\nu = \frac{2}{3}\, ,
\ee
independent of the details of the activation function itself; for scale-invariant activation functions\index{universality class!scale-invariant}, we have~\eqref{eq:k-star-equals-zero-normalized-four-point-relu-univ}
\be\label{eq:scale-invariant-nu}
\nu = \le(\frac{3A_4}{A_2^2} -1 \ri) \,,
\ee
with $A_2\equiv (a_+^2+a_-^2)/2$ and $A_4\equiv (a_+^4+a_-^4)/2$.\footnote{This gives $\nu=2$ for $\linear$ activations and $\nu=5$ for the $\relu$.} Plugging this scaling solution~\eqref{eq:scaling-solution-reprinted-for-MIT} into our NLO expression for the mutual informtion \eqref{eq:MIT-third-order}, we get
\begin{align}\label{eq:MIT-third-order-tangible}
\MI\le[p\!\le( \mathcal{M}_1 , \mathcal{M}_2 \vert x\ri)\ri]=&\frac{m_1m_2}{8}\nu^2r^2-\frac{m_1 m_2(20+3m_1+3m_2)}{48}\nu^3r^3+\o{r^4} \, .
\end{align}

Now, let us explain our excitement.
At one extreme, in the \terminate{infinite-width limit} $r\to 0$, this mutual information vanishes, as we already knew from \eqref{eq:MIT-zeroth-order}. Thus, for small aspect ratios $r \ll 1$, the first leading-order term dominates and the mutual information increases as depth increases. As the depth continues to increase, then the second term begins to dominate, \emph{decreasing} the mutual information. But we know by the \emph{subadditivity} of the entropy\index{entropy!additivity!subadditivity}, \eqref{eq:positivity-of-mutual-information}, that the \terminate{mutual information} is always bounded from below by zero,
and so for large enough aspect ratios $r$, this decreasing will be balanced by additional higher-order corrections. Taken altogether, we expect that at some nonzero but not too large $r$, the mutual information will reach a local maximum.
Indeed, maximizing \eqref{eq:MIT-third-order-tangible}, we find an \term{optimal aspect ratio} for the network:
\be\label{eq:Banks-Zaks}
r^{\star} = \le( \frac{4}{20+3 m_1 + 3m_2}\ri)\frac{1}{\nu} \, ,
\ee
with $\nu$ containing all of the details of the activation function.\footnote{Don't worry about the higher-order contributions $\o{r^4}$ that we neglected in obtaining the solution~\eqref{eq:Banks-Zaks}: for a particular choice of activation function, i.e.~for a choice of $\nu$, the estimate of the optimal value \eqref{eq:Banks-Zaks} is \emph{a posteriori} justified so long as the product $\nu r^{\star}$ is perturbatively small for a particular  $\nu$ and a particular grouping of neurons $(m_1, m_2)$.  FYI\index{for your information}, this argument is analogous 
the one given to justify the two-loop \emph{Banks-Zaks fixed point}\index{Banks-Zaks fixed point|see{fixed point}}\index{fixed point!Banks-Zaks}\index{fixed point!Banks-Zaks|seealso{optimal aspect ratio}} of the renormalization group \index{renormalization group flow} in \terminate{non-Abelian gauge theory}\index{non-Abelian gauge theory|seealso{Banks-Zaks fixed point}} \cite{banks1982phase}.}

Although it's not immediately obvious, maximizing a \terminate{mutual information} such as \eqref{eq:scaling-solution-reprinted-for-MIT} is closely related to well-known \term{unsupervised learning} objectives.\footnote{In particular, the \neo{InfoMax principle}\index{InfoMax principle|seealso{unsupervised learning}} \cite{linsker1988self} 
recommends
maximizing the \terminate{mutual information} between the input $x$ and a \terminate{representation} $z(x)$. A related notion involves maximizing the mutual information between different representations, $z_1(x)$ and $z_2(x)$, for the same input $x$  \cite{becker1992self}. %
This latter notion can be shown 
to
lower bound the InfoMax objective and thus motivates our analysis here.\label{footnote:info-max}} 
In contrast to a \emph{supervised} learning\index{supervised learning} setting where the goal is to predict the true output $y\equiv f(x)$ for any input sample $x$, in an \emph{unsupervised} learning setting the goal is to learn representations for a collection of input samples by observing patterns in the data.
For human-generated datasets, this has the advantage of eliminating the tedious task of labeling all the samples.
It should also be no surprise, given the benefit of \terminate{representation learning} (\S\ref{ch:features}), that models (pre)trained\index{pretraining}\index{unsupervised learning!as pretraining} with unsupervised learning algorithms  can often be 
efficiently fine-tuned on subsequent supervised learning tasks.

In the current context, rather than doing any actual learning,
we are understanding, a priori, which choice of the architecture and activation function can lead to a larger mutual information between neurons in deeper layers.\footnote{
    Similarly, we may think of our criticality analysis\index{criticality!as unsupervised learning} as a type of unsupervised learning or pretraining\index{pretraining} criteria in which we are understanding, a priori, which choice of \terminate{initialization hyperparameters} leads to an \terminate{ensemble} that generalizes most robustly after training, cf.~\S\ref{sec:generalization-at-infinity}.
} 
In particular, comparing the values of $\nu$ in \eqref{eq:tanh-nu} and \eqref{eq:scale-invariant-nu} for different choices of activation function, we see in principle that
networks built from $\tanhA$ activation function should be deeper than networks built from $\relu$ activation functions to have the same mutual information in a layer.

With this objective in mind, we should continue maximizing \eqref{eq:MIT-third-order-tangible} across all the possible partitions $(m_1, m_2)$. This picks out
a very natural choice of a partition spanning the whole layer and of equal sizes: $m_1=m_2=n_L/2$.
With this further maximization, we get
\be\label{eq:optimal-aspect-ratio-natural-choice}
r^{\star}= \le( \frac{4}{20+3n_L}\ri)\frac{1}{\nu} \, ,
\ee
for the optimal aspect ratio and
\be\label{eq:optimal-MI-value}
\MI\le[p\!\le( \mathcal{M}_1 , \mathcal{M}_2 \vert x\ri)\ri]= \frac{1}{6}\le(\frac{n_L}{20+3n_L}\ri)^2 \, ,
\ee
for the associated value of the maximized mutual information. In particular, this corresponds to a lower bound on the \emph{InfoMax}\index{InfoMax principle} objective that we discussed in footnote~\ref{footnote:info-max}.\footnote{
    While this value \eqref{eq:optimal-MI-value} may seem small, remember that our analysis is based entirely on the prior distribution  before any learning has taken place. 
}

As a final comment on \eqref{eq:optimal-aspect-ratio-natural-choice}, we can think of this optimal aspect ratio as defining a scale that separates the effectively-deep\index{effectively deep} regime for which our effective theory is valid from the overly-deep\index{overly deep} regime where the theory is strongly-coupled and networks are no longer trainable. We will push this interpretation further in the next appendix when we discuss residual networks\index{residual network}.

For a final computation, let's look at the \neo{tripartite information}~\eqref{eq:TI-definition} for mutually-exclusive subsystems $\mathcal{M}_1$, $\mathcal{M}_2$, and $\mathcal{M}_3$ of sizes $(m_1, m_2, m_3)$ neurons, respectively. Plugging in \eqref{eq:entropy-third-order} for various different combinations of the sizes, we find
\be\label{eq:TIM-third-order}
\MI_{3}\le[p\!\le( \mathcal{M}_1 , \mathcal{M}_2, \mathcal{M}_3 \vert x\ri)\ri]= \frac{m_1 m_2 m_3}{8} \le[\frac{ V^{(\ell)} }{ n_{\ell-1} \le(G^{(\ell)}\ri)^2} \ri]^3 +
 \o{\frac{1}{n^4}}\, ,
\ee
which, as per \eqref{eq:k-star-equals-zero-normalized-four-point-scaling-law}, scales cubically with depth, $\MI_{3}\le[p\!\le( \mathcal{M}_1 , \mathcal{M}_2, \mathcal{M}_3 \vert x\ri)\ri] \propto \ell^3/n_{\ell-1}^3$, and thus indicates that a nonzero result only first appears at $\o{1/n^3}$.\footnote{
    In hindsight, it's obvious that we needed the entropy to have at least a cubic dependence on the 
    sizes $m_i$ to find a nonzero answer for the \terminate{tripartite information}, just as we needed the entropy to have at least a quadratic dependence on the sizes to find a nonzero answer for the mutual information \eqref{eq:MIT-second-order}.
}
Importantly, we see that the tripartite information is exclusively \emph{positive}, meaning that any three groups of neurons in a layer will form \emph{redundant}\index{redundancy (information theory)} representations under RG flow: knowing the activities of any one group of neurons $\mathcal{M}_1$ means you would learn less information about a second group of neurons $\mathcal{M}_2$ from observing a third group $\mathcal{M}_3$ than you otherwise would have learned had you not already known $\mathcal{M}_1$. It would be interesting to try to understand further how this property relates to the coarse-graining mechanism of the \terminate{representation group flow} to the deeper layers.\footnote{It would also be interesting to try and interpret this in terms of the \neo{optimal brain damage} of \cite{brain-damage} or the \neo{lottery ticket hypothesis} of \cite{frankle2018the}.}

%% file: AppB-Residual/B_global.tex
\chapter{Residual Learning}
\label{app:residual}

\epigraph{No Doc\index{Brown, Emmett Lathrop ``Doc''}, not me, the other me! The one that's up on stage playing \neo{Johnny B. Goode}!}{Marty McFly \cite{BttFP2}.\index{McFly, Martin Seamus ``Marty''}}

\noindent{}In this final appendix, we'll analyze neural networks with residual connections\index{residual connection}, generally called  \textbf{residual networks}\index{residual network|textbf}.
These networks
were originally introduced in order to enable the training of deeper and deeper networks: traditionally deep networks suffer from the \neo{exploding and vanishing gradient problem}, but even in networks where various tricks of the trade are used to ensure the propagation of forward and backward signals, 
overly-deep\index{overly deep} 
networks are empirically found to have \emph{higher training and test errors} than their shallower-network counterparts.

From the \terminate{microscopic perspective}, 
 the increase of the 
\terminate{generalization error} is intuitive for a deeper model with more \terminate{model parameters}, but the increase in the training error is not: the additional parameters should naively enable \emph{better} fits to the training data.
At the very least, one might hope that the additional layers could 
-- in principle --
approximate the identity map and do \emph{no worse}.
Yet the empirical evidence mentioned above suggests that it's difficult for optimization algorithms 
to tune the
hidden layers of a deep network %
to such a nearly-identity map.
This is called \textbf{degradation}\index{degradation problem|textbf}\index{degradation problem|seealso{residual network}}\index{degradation problem|seealso{overly deep}}, and in principle 
is
a major limiting factor for developing larger scale deep-learning models.

From the \terminate{macroscopic perspective} of our effective theory, we can offer a \emph{dual}\index{duality!microscopic-macroscopic} explanation for this degradation problem.
As a precursor to our explanation, first recall that, rather than using any heuristic approach to solve the \terminate{exploding and vanishing gradient problem}, in \S\ref{sec:EVGP-WEP}
we analytically solved its \emph{exponential} manifestation by means of \terminate{criticality} and then solved its \emph{polynomial} manifestation by means of the learning-rate \terminate{equivalence principle}.
In \S\ref{sec:generalization-at-infinity} we further confirmed that the associated tunings of the \terminate{initialization hyperparameters}, $\Cb{\ell}$ and $\CW{\ell}$, and of the \terminate{training hyperparameters}, $\Lb{\ell}$ and $\LW{\ell}$,  lead to the most robust generalization performance for MLPs. 

Now, \emph{even if} these hyperparameters are  properly tuned, we would expect that overly-deep networks 
will suffer horribly from instantiation-to-instantiation \terminate{fluctuations}, leading to the breakdown of \terminate{criticality} for any \emph{particular} network. This problem was first discussed in \S\ref{sec:solution_DLN} for extremely deep (linear) networks, then more generally  in \S\ref{sec:signal_prop_finite_width}, and finally in footnote~\ref{footnote:epilogue-chaos} of Epilogue~\ref{epi:overparameterization}. Thus, combined with our discussion  
of the MLP's depth-to-width ratio $r\equiv L/n$ in 
\S\ref{sec:information-beyond-infinity},
perhaps we %
can understand the training loss degradation\index{degradation problem} problem in terms of the network leaving the regime of optimality and proceeding towards the regime of \terminate{chaos}.

\begin{figure}
\begin{center}
 \includegraphics[width=1\linewidth]{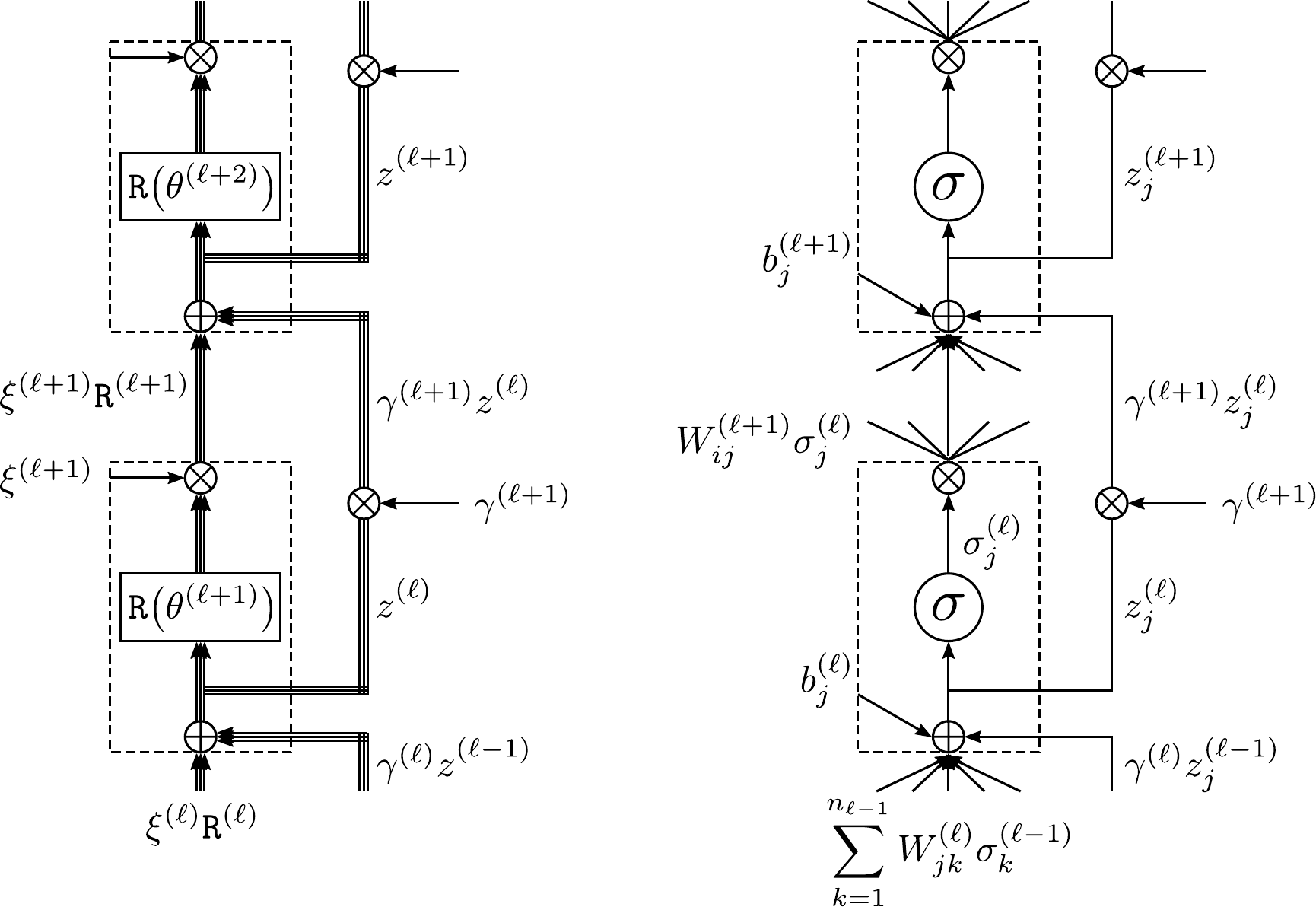}
\caption{
\textbf{Left:} two residual blocks\index{residual block} from adjacent layers for a very general \neo{residual network}. This detailed structure depicts how  each layer
\emph{(i)} adds a weighted block and the weighted preactivation to produce the next layer's preactivation,
\emph{(ii)} copies the preactivation to skip to the next layer,
and \emph{(iii)} generates the next layer's block.
\textbf{Right:} two neurons from adjacent layers for a 
\emph{residual MLP}.\index{multilayer perceptron!with residual connections} 
This detailed structure depicts how each layer \emph{(i)} adds the bias, 
the weighted activation, and the weighted preactivation 
to produce the next layer's preactivation, 
\emph{(ii)} copies the preactivation to skip to the next layer,
\emph{(iii)} generates the activation 
from the preactivation, and \emph{(iv)} multiplies the activation by the next-layer weight.
}
\label{fig:residual-example}
\end{center}
\end{figure}

If you'll please move the microscopic explanation back to the front of your mind, we can explain an ingenious solution to 
degradation
by He \emph{et al.}~\cite{he2016deep}.
Rather than trying to find a better learning algorithm, we can instead modify the deep network architecture so that the hidden layers 
only have to learn a \neo{residual function}: in place of a generic nonlinear $(\ell+1)$-th layer
\be\label{eq:zeroth-residual-connection}
z^{(\ell+1)}=\layer\!\le(z^{(\ell)}; \theta^{(\ell+1)}\ri)\, ,
\ee
we design the layer as
\be\label{eq:first-residual-connection}
z^{(\ell+1)}=\block\!\le(z^{(\ell)}; \theta^{(\ell+1)}\ri) + z^{(\ell)}\, ,
\ee
such that the \term{residual block}, $\block\!\le(z^{(\ell)}; \theta^{(\ell+1)}\ri)$,
is the residual of the function
that we want our layer, \eqref{eq:zeroth-residual-connection}, to implement.
The basic structure of this generic residual layer is depicted in the left panel of Figure~\ref{fig:residual-example} and will be further explained later on.

From a \terminate{microscopic perspective}, 
these residual connections make learning a nearly-identity map much easier. %
Indeed, it is 
far easier to set the residual block $\block\!\le(z^{(\ell)}; \theta^{(\ell+1)}\ri)$ to near-zero than it is to coax a generic nonlinear function $\layer\!\le(z^{(\ell)}; \theta^{(\ell+1)}\ri)$ to approximate the identity.
In particular, since standard building blocks of the neural network often have the property that they vanish when their parameters are set to zero -- cf.~\eqref{eq:mlp-definition} for the \emph{MLP-layer block} -- and since typical initialization distributions\index{initialization distribution} have zero mean -- cf.~\eqref{eq:full-bias-initialization} and \eqref{eq:full-weights-initialization} -- residual networks make it fairly easy to find a solution with $\block\!\le(z^{(\ell)}; \theta^{\star}\ri)\approx 0$ %
such that the addition of the hidden layer doesn't necessarily degrade\index{degradation problem} the performance of the network.

More generally, we hope that the preactivation will actually play two roles,  one of \emph{coarse-graining} the input signal according to \terminate{representation group flow} (\S\ref{sec:marginalization-group-flow}) and the other of propagating an \emph{undegraded}\index{degradation problem} copy of the input signal. This is plausibly quite helpful as it allows us to train deeper models with more parameters, and it has indeed been empirically demonstrated that such deeper residual networks lead to significant performance gains on the test set.

\index{overly deep}\index{effectively deep!range extended by residual connections}
One of the goals of this appendix is to provide a dual macroscopic explanation for why the residual connections\index{residual connection} let us train overly-deep networks. Our macroscopic explanation above for the origin of the degradation\index{degradation problem} problem  -- combined with the empirical success of very deep residual networks -- suggests that the inclusion of residual connections\index{residual connection} shifts the optimal aspect ratio $r^\star$ from its MLP value, \eqref{eq:Banks-Zaks}, to higher values. This would extend the range of effectively-deep networks and thus explain why residual connections let us train deeper networks.
To test this hypothesis, we will need to 
carry out our
effective theory analysis for residual networks\index{residual network}.

Despite the long prelude, this appendix will be relatively brief and only involve some simple calculations.
These 
exercises will serve two purposes.
First, given the overwhelming ubiquity of \textbf{residual connections}\index{residual connection|textbf}\index{residual connection!other names} -- sometimes called \emph{skip connections}\index{skip connection|see{residual connection}} or \emph{shortcuts}\index{shortcuts|see{residual connection}} -- in modern deep learning architectures, it is practically useful to explain how the critical initialization hyperparameters\index{initialization hyperparameters!critical} \emph{shift} when residual connections are included.
Second, this will showcase how our effective theory formalism can easily be applied to neural network architectures other than vanilla MLPs\index{multilayer perceptron!vanilla}.

To those ends -- and to the end of the book -- in \S\ref{sec:residual-mlp} we'll begin by briefly introducing the perhaps simplest residual network, a \emph{multilayer perceptron with residual connections}. In \S\ref{sec:residual-criticality}, we'll study the infinite-width limit of this model,
performing a criticality analysis 
in order to understand how the \emph{residual hyperparameters} 
interplay with 
the critical initialization hyperparameters. 

Then, 
in \S\ref{sec:residual-MI} 
we'll study the residual MLP at finite width. Using our auxiliary \terminate{unsupervised learning} objective from the last appendix, we'll see how the inclusion of residual connections can shift the \neo{optimal aspect ratio}  of a network $r^\star$ 
to large values. 
We will also learn that for networks with aspect ratios below a certain threshold, residual connections are \emph{always} harmful according to this criterion.
Altogether, this provides a new effective-theory \terminate{macroscopic perspective} on how residual connections solve the degradation\index{degradation problem} problem described above, and further lets us understand the 
tradeoff of propagating signals through the \terminate{residual block}  
-- in this case, a nonlinear MLP-layer block -- against propagation through identity block -- skipping signals to deeper layers. %

Finally, in \S\ref{sec:residual-emprical} we'll give a hybrid theoretical-empirical recipe 
applying the analyses in the previous two sections to
\emph{general} residual networks with arbitrarily complicated residual blocks $\block\!\le(z^{(\ell)}; \theta^{(\ell+1)}\ri)$. We hope this may have broad application to the many deep learning architectures that implement residual connections.\footnote{
    Residual networks were first described by He \emph{et al.}~\cite{he2016deep}. Typically, when referring to a residual network as a \neo{ResNet},  the base block is composed of convolutional layers\index{convolutional neural network},
 cf.~\eqref{eq:conv-layer}, that are further augmented with very popular heuristic for mitigating the exploding and vanishing gradient problem \cite{batch-norm}. 
    While original domain of ResNets was \terminate{computer vision} tasks, they have now been applied to other domains; more broadly, \emph{residual connections} are components in a wide variety of modern deep learning architectures,  including importantly the \terminate{transformer}-based language models\index{language model} that have 
    been revolutionizing
    \terminate{natural language processing}~\cite{attention2017}. \index{language model}
} 
After this, you will have finished all your residual learning from this book.

\section{Residual Multilayer Perceptrons}\label{sec:residual-mlp}
A \textbf{multilayer perceptron with residual connections}\index{multilayer perceptron!with residual connections} 
can be defined by the forward equation\index{forward equation!residual MLP preactivations},
\be\label{eq:ResMLP}
\z{i}{\delta}{\ell+1}=\xi^{(\ell+1)}\! \le[\bias{i}{\ell+1}+\sum_{j=1}^{n_{\ell}}\W{ij}{\ell+1}\s{j}{\delta}{\ell}\ri]+\gamma^{(\ell+1)} \z{i}{\delta}{\ell}\, .
\ee
Compared with our schematic description, \eqref{eq:first-residual-connection},
here we've picked just a standard MLP layer as our \terminate{residual block}, $\block\!\le(z^{(\ell)}; \theta^{(\ell+1)}\ri)$, and we've allowed for \emph{scalar} \term{residual hyperparameters}, $\xi^{(\ell)}$ and $\gamma^{(\ell)}$, that control the relative magnitudes of the residual-block term vs.~the identity term. 
Here we need to take 
\be
n_{\ell}=n_{\ell+1}\equiv n \, 
\ee 
so that we can add the weighted and biased activation back to the preactivation before the activation, and 
such residual connections are thus only included for the network's hidden layers.\footnote{More generally, if we wanted $n_{\ell+1}\ne n_{\ell}$ we could instead use an iteration equation such as
\be
\z{i}{\delta}{\ell+1}=\sum_{j=1}^{n_{\ell+1}}\xi^{(\ell+1)}_{ij} \le[\bias{j}{\ell+1}+\sum_{k=1}^{n_{\ell}}\W{jk}{\ell+1}\s{k}{\delta}{\ell}\ri]+\sum_{j=1}^{n_{\ell}}\gamma^{(\ell+1)}_{ij} \z{j}{\delta}{\ell}\, ,
\ee
where the residual hyperparameters are now matrices: $\xi^{(\ell+1)}_{ij}$ is an $n_{\ell+1}$-by-$n_{\ell+1}$ matrix and $\gamma^{(\ell+1)}_{ij}$ is an $n_{\ell+1}$-by-$n_{\ell}$ matrix, and in general both matrix could vary from layer to layer.}
A graph of two neurons in adjacent layers from 
a residual MLP
is shown in the right panel of Figure~\ref{fig:residual-example}, 
and if you need to recall the overall global structure of an MLP, please refer back to Figure~\ref{fig:ff-example}.

Although nice for the symmetry, one of the residual hyperparameters is redundant: you can show that adjusting $\xi^{(\ell)}$ has the same effect on the ensemble of residual MLPs as rescaling the \terminate{initialization hyperparameters}, $\Cb{\ell}$ and $\CW{\ell}$, and the \terminate{training hyperparameters}, $\Lb{\ell}$ and $\LW{\ell}$. As such, we'll henceforth set
\be
\xi^{(\ell)}=1\, ,
\ee
without loss of generality, leaving us only one (potentially layer-dependent) residual hyperparameter: $\gamma^{(\ell)}$. Note importantly that in the limit $\gamma^{(\ell)}\to 0$ we recover a vanilla MLP\index{multilayer perceptron!vanilla} without any residual connections\index{residual connection}. A customary choice for this hyperparameter is $\gamma^{(\ell)}=1$, cf.~\eqref{eq:first-residual-connection},
but let us now show that this is suboptimal because it \emph{breaks} criticality.\footnote{
    In particular, this choice is problematic without any additional heuristics, e.g.~\cite{batch-norm}, for otherwise mitigating the \terminate{exploding and vanishing gradient problem}. The following analysis thus offers an explanation for why deep $\gamma^{(\ell)}=1$ residual networks\index{residual network}  without such heuristics will always perform poorly.
}

\section{Residual Infinite Width: Criticality Analysis}\label{sec:residual-criticality}
To understand how to set the residual hyperparameter $\gamma^{(\ell)}$ and preserve criticality for the residual MLP\index{multilayer perceptron!with residual connections},
let's work out a recursion for the two-point correlator:
\be
\E{\z{i_1}{\delta_1}{\ell}\z{i_2}{\delta_2}{\ell}}=\delta_{i_1i_2} G_{\delta_1\delta_2}^{(\ell)} \, .
\ee
Using the forward equation \eqref{eq:ResMLP}, we find
\be\label{eq:mean-metric-residual-MLP}
G_{\delta_1\delta_2}^{(\ell+1)}=C_b+C_W\frac{1}{n}\sum_{j=1}^{n}\E{\s{j}{\delta_1}{\ell}\s{j}{\delta_2}{\ell}}+\gamma^2 G^{(\ell)}_{\delta_1\delta_2}\, .
\ee
Here, mirroring all our previous discussions of criticality that are now too many to enumerate with references,
throughout this section we'll set the bias variance, $C_b^{(\ell)}$, the rescaled weight variance, $C_W^{(\ell)}$, and the residual hyperparameter, $\gamma^{(\ell)}$, to be uniform across layers,
\be\label{eq:everything-is-as-it-always-is-for-each-layer-is-the-same-amen}
C_b^{(\ell)}=C_b\, , \qquad C_W^{(\ell)}=C_W\, , \qquad \gamma^{(\ell)}=\gamma\, .
\ee
Compared to the metric recursion for a vanilla MLP\index{multilayer perceptron!vanilla} \eqref{eq:mean-metric-any-layer},
\be\label{eq:mean-metric-any-layer-reprint}
\Ti{G}{\delta_1 \delta_2}{\ell+1} = C_b +C_W \frac{1}{n}\sum_{j=1}^{n}\E{\s{j}{\delta_1}{\ell}\s{j}{\delta_2}{\ell}}\, ,
\ee
the 
$(\ell+1)$-th-layer metric for the residual MLP
is shifted by the (rescaled) identity term from the forward equation, leading to an additional contribution of the $\ell$-th-layer metric as the (rescaled) final term of \eqref{eq:mean-metric-residual-MLP}.

In particular, the infinite-width limit of the residual MLP\index{infinite-width limit!of residual MLPs} leads to the following recursion for its kernel,
\be\label{eq:kernel-recursion-res}
\ker_{\delta_1\delta_2}^{(\ell+1)}=C_b+C_W\bra\sigma_{\delta_1}\sigma_{\delta_2}\ket_{\ker^{(\ell)}}+\gamma^2 \ker^{(\ell)}_{\delta_1\delta_2}\, .
\ee
Then, writing the kernel in our $\gamma^{[a]}$ basis~\eqref{eq:kernel-expand-gamma}, using our $\delta$ expansion, \eqref{eq:kernel-expand-1}--\eqref{eq:kernel-expand-3}, we can follow the two-input perturbative analysis of \S\ref{sec:bootstrapping} to derive component recursions
\begin{align}
\Ti{\ker}{\M\M}{\ell+1}&=C_b+C_W\, g\!\le(\KML\ri)+\gamma^2\KML\, ,  \label{K0-Res}\\
\Ti{\delta \ker}{[1]}{\ell+1}&=\chi_{\parallel}\!\le(\Ti{\ker}{\M\M}{\ell}\ri)\Ti{\delta \ker}{[1]}{\ell}\, ,\label{K1-Res}\\
\Ti{\delta\delta \ker}{[2]}{\ell+1}&=\chi_{\perp}\!\le(\Ti{\ker}{\M\M}{\ell}\ri)\Ti{\delta\delta \ker}{[2]}{\ell}+h\!\le(\KML\ri)\le(\Ti{\delta \ker}{[1]}{\ell}\ri)^2\, , \label{K2-Res}
\end{align}
which are modified from their vanilla expressions \eqref{K0}--\eqref{K2}.\footnote{
    Note that, as per this first recursion \eqref{K0-Res}, a  perturbation to a single-input, $\KML=\Tif{\ker}{\M\M}+\Delta \KML$, is governed by the \emph{shifted} parallel susceptibility \eqref{eq:chi-parallel-Res}, cf.~\eqref{eq:single_expand}.
}
Here, 
the \neo{parallel susceptibility},
\be\label{eq:chi-parallel-Res}
\chi_{\parallel}(\ker)\equiv 
\gamma^2 + \frac{C_W}{2\ker^2} \bra \sigma(z)\, \sigma(z)\le(z^2-K\ri)\ket_{\ker}
\, ,
\ee
and the \neo{perpendicular susceptibility},
\be\label{eq:chi-perp-Res}
\chi_{\perp}(\ker)\equiv \gamma^2  + C_W \bra\sigma^\prime(z)\, \sigma^\prime(z) \ket_{\ker}\, ,
\ee
are each shifted from their previous values, \eqref{eq:chi-parallel} and \eqref{eq:chi-perp}, by a constant $\gamma^2$ term, while
the helper functions, $g\!\le(K\ri)$ and $h\!\le(K\ri)$, \eqref{eq:g-function} and \eqref{eq:h-function}, are the same as before:
\begin{align}\label{eq:g-function-appendix-B-reprint}
g\!\le(K\ri)&\equiv \le\langle \sigma(z) \, \sigma(z)\ri\rangle_{K}\, , \\
h\!\le(K\ri)&\equiv 
\frac{1}{2}\frac{\td }{\td K}\chi_{\perp}(\ker)\, .
\label{eq:h-function-appendix-B-reprint}
\end{align}

As should be already clear from this simple analysis, inclusion of residual connections\index{residual connection} ($\gamma \neq 0$) will shift the critical initialization hyperparameters\index{initialization hyperparameters!critical!for residual networks}. Fixing the
criticality conditions
\be
\chi_{\parallel}\!\le(\ker^{\star}\ri)= 1 \, , \qquad \chi_{\perp}\!\le(\ker^{\star}\ri)=1 \,,
\ee
 we can find the new critical initializations for our universality classes:
specifically, for the scale-invariant universality class\index{universality class!scale-invariant}, the critical initialization hyperparameters, \eqref{eq:criticality_scale_invariant}, are modified as 
\be\label{eq:criticality_scale_invariant-Res}
C_b=0\, , \qquad C_W= C_W(\gamma) \equiv \frac{1}{A_2}\le( 1-\gamma^2\ri)\, ,
\ee
where $A_2\equiv (a_{+}^2+a_{-}^2)/2$, cf.~\eqref{eq:scale-invariant-one-kink}; analogously,  for the the $K^{\star}=0$ universality class\index{universality class!K@$K^\star=0$}, the critical initialization hyperparameters \eqref{eq:k-star-equals-zero-critical-initialization} are modified as
\be\label{eq:criticality_scale_invariant-K-star}
C_b=0\, , \qquad C_W= C_W(\gamma) \equiv \frac{1}{\sigma_1^2}\le( 1-\gamma^2\ri)\, ,
\ee
where $\sigma_1\equiv\sigma'(0)$.

In \S\ref{sec:MLP_distribution}, we discussed that the \neo{zero initialization}, $\bias{i}{\ell}=\W{ij}{\ell}=0$, fails to break the \terminate{permutation symmetry} among the $n$ neurons in a hidden layer. In conjunction with this reasoning, we now see that the criticality conditions for either class,  \eqref{eq:criticality_scale_invariant-Res} and \eqref{eq:criticality_scale_invariant-K-star}, are unsatisfiable for the customary choice $\gamma=1$ of the residual hyperparameter.
More broadly, for each universality class there is a \emph{one-parameter family}\index{one-parameter families} of critical rescaled weight variances: for each $0 < \gamma^2<1$, there is an associated critical value of $C_W=C_W(\gamma)$.
Thus, in order to directly mitigate the \terminate{exploding and vanishing gradient problem}, for a particular choice of $0<\gamma^2<1$ and activation function, we must pick $C_W$ according to the appropriate $C_W(\gamma)$.

\section{Residual Finite Width: Optimal Aspect Ratio}\label{sec:residual-MI}

From our infinite-width analysis, we just saw that residual networks\index{residual network} have a one-parameter family\index{one-parameter families} of critical solutions: $C_W(\gamma)$. As per the kernel recursion \eqref{eq:kernel-recursion-res}, these solutions trade off the degree to which the identity branch vs.~the MLP-layer branch contributes to the next layer's preactivations. In the strict \terminate{infinite-width limit}, criticality ensures that the kernel is preserved for any choice $C_W(\gamma)$ in the range $0 < \gamma^2 <1$, and all \terminate{fluctuations} are suppressed; thus, we're in principle completely indifferent between any of these critical tunings.

However, as should now be a familiar point of discussion, networks in the infinite-width limit effectively have a depth-to-width ratio $r\equiv L/n \to 0$. This means that infinite-width analysis cannot really get at the question of \textbf{degradation}\index{degradation problem|textbf}: why do extremely deep networks have  \emph{higher training errors} than otherwise equivalent shallower networks. 
To compare wide networks of different depths, we need to consider finite-width networks with different aspect ratios $r > 0$.\footnote{
In particular, at finite width the \neo{representation group flow} through the MLP-layer blocks leads to two competing finite-width effects: \emph{(i)}
 the relevant\index{relevant (RG flow)} and thus \emph{growing} dNTK and ddNTKs lead to nontrivial representation learning during training,
and \emph{(ii)} the relevant\index{relevant (RG flow)} and \emph{growing} four-point vertex, NTK variance, and NTK-preactivation cross correlation lead to fluctuations in the ensemble from instantiation to instantiation.
As the residual connections\index{residual connection} are supposed to mitigate this second harmful effect by xeroxing the undegraded input via the identity branch, we naturally expect that they will have a meaningful physical effect on this competition.
In other words, we expect that residual networks\index{residual network} of different residual hyperparameters $\gamma$ and  different aspect ratios $r$ will lead to very different test-set generalization.
}

Our main tool of analysis in this section will be a computation of the \emph{mutual information at initialization}\index{mutual information} between non-overlapping representations in deep layers of a residual MLP.\index{multilayer perceptron!with residual connections} 
As per our auxiliary \terminate{unsupervised learning} criterion from the last appendix, cf.~footnote~\ref{footnote:info-max} of
\S\ref{sec:information-beyond-infinity},
this mutual information gives a natural way to estimate the \terminate{optimal aspect ratio} $r^\star$ of a finite-width network.
Given that residual networks  without an \terminate{exploding and vanishing gradient problem} solve the degradation\index{degradation problem} problem, a natural theoretical prediction for our residual MLPs is that
the inclusion of residual connections\index{residual connection}, $\gamma > 0$, and then tuning to criticality as $C_W(\gamma)$ will together \emph{shift their optimal aspect ratios to larger values}.

To evaluate 
the optimal aspect ratio via our formula \eqref{eq:optimal-aspect-ratio-natural-choice}, 
we just need to compute the coefficient $\nu$ of normalized, output-layer, single-input four-point vertex 
for the residual MLP\index{multilayer perceptron!with residual connections}:
\be\label{eq:scaling-solution-reprinted-for-MIT-reprint}
\frac{ V^{(L)} }{ n \le(G^{(L)}\ri)^2} \equiv \nu r \, .
\ee
Note that we've already derived the recursion for the leading-order single-input metric, i.e.~the kernel $\Ti{\ker}{}{L}\equiv G_{}^{\le\{0\ri\}(L)}$, in \eqref{K0-Res}. To compute the four-point vertex, first recall that the multi-input four-point vertex
defines the four-point connected correlator \eqref{eq:C4_MLPH} as
\begin{align}\label{eq:C4_MLPRes}
&\E{\z{i_1}{\delta_1}{\ell}\z{i_2}{\delta_2}{\ell}\z{i_3}{\delta_3}{\ell}\z{i_4}{\delta_4}{\ell}}\Big\vert_{\text{connected}}\, \\
=&\frac{1}{n}\le[\delta_{i_1i_2}\delta_{i_3 i_4}V^{(\ell)}_{(\delta_1\delta_2)(\delta_3\delta_4)}+\delta_{i_1i_3}\delta_{i_2 i_4}V^{(\ell)}_{(\delta_1\delta_3)(\delta_2\delta_4)}+\delta_{i_1i_4}\delta_{i_2 i_3}V^{(\ell)}_{(\delta_1\delta_4)(\delta_2\delta_3)} \ri]\, .\nonumber
\end{align}
Using the forward equation \eqref{eq:ResMLP} and following our analysis from \S\ref{sec:deeper-layer-accumulation}, we see that
the recursion for the residual-MLP's four-point vertex\index{four-point vertex!residual MLPs} is shifted by two additional terms proportional to $\gamma^2$ and $\gamma^4$ as compared to the leading-order recursion for the vanilla MLP \eqref{eq:V-recursion-tree},
\begin{align}\label{eq:V-recursion-tree-Res}
&V^{(\ell+1)}_{(\delta_1\delta_2)(\delta_3\delta_4)}\, \\
=&C_W^2\le[\bra\sigma_{\delta_1} \sigma_{\delta_2} \sigma_{\delta_3} \sigma_{\delta_4}\ket_{\ker^{(\ell)}}  - \bra \sigma_{\delta_1} \sigma_{\delta_2}\ket_{\ker^{(\ell)}} \bra \sigma_{\delta_3} \sigma_{\delta_4}\ket_{\ker^{(\ell)}} \ri]\, \notag\\
&+\frac{C_W^2}{4}\!\!\!\sum_{\delta_5,\ldots,\delta_8\in\D}\!\!\!\!\TI{V}{(\delta_5\delta_6)(\delta_7\delta_8)}{\ell}\bra\sigma_{\delta_1}\sigma_{\delta_2} \le(z_{\delta_5} z_{\delta_6}-\ker_{\delta_5\delta_6}\ri)\ket_{\ker^{(\ell)}}\bra \sigma_{\delta_3}\sigma_{\delta_4} \le(z_{\delta_7} z_{\delta_8}-\ker_{\delta_7\delta_8}\ri)\ket_{\ker^{(\ell)}}\,  \notag\\
&+C_W\gamma^2  \Big[\bra\sigma_{\delta_1}\sigma_{\delta_2} \le(z_{\delta_3} z_{\delta_4}-\ker_{\delta_3\delta_4}\ri)\ket_{\ker^{(\ell)}}+\bra\sigma_{\delta_3}\sigma_{\delta_4} \le(z_{\delta_1} z_{\delta_2}-\ker_{\delta_1\delta_2}\ri)\ket_{\ker^{(\ell)}}\, \notag\\
&\quad\quad\quad\quad+\frac{1}{2}\sum_{\delta_5,\ldots,\delta_8\in\D}\!\!\!\!\TI{V}{(\delta_5\delta_6)(\delta_7\delta_8)}{\ell}\bra\sigma_{\delta_1}\sigma_{\delta_2} \le(z_{\delta_5} z_{\delta_6}-\ker_{\delta_5\delta_6}\ri)\ket_{\ker^{(\ell)}}\ker_{\delta_7\delta_3}^{(\ell)}\ker_{\delta_8\delta_4}^{(\ell)}\notag\\
&\quad\quad\quad\quad+\frac{1}{2}\sum_{\delta_5,\ldots,\delta_8\in\D}\!\!\!\!\TI{V}{(\delta_5\delta_6)(\delta_7\delta_8)}{\ell}\bra\sigma_{\delta_3}\sigma_{\delta_4} \le(z_{\delta_5} z_{\delta_6}-\ker_{\delta_5\delta_6}\ri)\ket_{\ker^{(\ell)}}\ker_{\delta_7\delta_1}^{(\ell)}\ker_{\delta_8\delta_2}^{(\ell)}\Big]\notag\\
&+\gamma^4 \,V^{(\ell)}_{(\delta_1\delta_2)(\delta_3\delta_4)}\, ,
\end{align}
where in parsing this expression, please remember that the raised indices of the \terminate{four-point vertex} are our shorthand for contraction with the $\ell$-th-layer inverse kernel, cf.~\eqref{eq:vertex-UUUU-dddd}.
Here, in working out the middle terms proportional to $\gamma^2$, you might find the \emph{intra}layer formula~\eqref{eq:general-covariance} to be useful.

Specializing now to a single input, 
we get
\begin{align}\label{eq:V-recursion-tree-Res-single}
V^{(\ell+1)}=&C_W^2\le[\bra \sigma^4(z) \ket_{\Ti{\ker}{}{\ell}}-\bra \sigma^2(z) \ket_{\Ti{\ker}{}{\ell}}^2\ri]
+\le(\chi^{(\ell)}_{\parallel}\ri)^2\Ti{\FPV}{}{\ell}
+4\gamma^2
\le(\chi_{\parallel}^{(\ell)}-\gamma^2\ri)
\le(\ker^{(\ell)}\ri)^2
\, ,
\end{align}
where for convenience we have abbreviated  $\chi_{\parallel}^{(\ell)}\equiv \chi_{\parallel}\!\le(\ker^{(\ell)}\ri)$ as we often do.
Here, this parallel susceptibility 
is the shifted one appropriate for residual MLPs as defined in \eqref{eq:chi-parallel-Res} with the $\gamma^2$ term; 
otherwise, %
 we notice the addition of the  $\gamma$-dependent final term compared with the single-input recursion for vanilla MLPs \eqref{eq:finite-width-reprinted-vertex}. At criticality with $\chi_{\parallel}^{(\ell)}=1$, this final term will make a nontrivial difference on the deep asymptotic behavior of the four-point vertex.
Now, let's solve this recursion at criticality for our two universality classes, with the usual initial condition $V^{(1)}=0$ (cf.~the title of \S\ref{sec:first-layer-gaussian}).

\subsubsection{Scale-Invariant Universality Class}
For the scale-invariant universality class\index{universality class!scale-invariant}, we tune the rescaled weight variance to criticality with \eqref{eq:criticality_scale_invariant-Res} so that $\chi_{\parallel}^{(\ell)}=1$.  
Then, the single-input kernel is exactly fixed, $\ker^{(\ell)}=\ker^{\star}$, 
and we also need to remember \eqref{eq:acctivation-function-gaussian-cov},
\be
\le[\bra \sigma^4\ket_{\ker^{\star}}-\bra \sigma^2\ket_{\ker^{\star}}^2\ri]=(3A_4-A_2^2)\le(\ker^{\star}\ri)^2 \, ,
\ee 
with $A_2\equiv (a_{+}^2+a_{-}^2)/2$ and $A_4\equiv (a_{+}^4+a_{-}^4)/2$. Substituting all of this into \eqref{eq:V-recursion-tree-Res-single}, the recursion becomes
\be\label{eq:V-recursion-tree-Res-single-scale-invariant}
V^{(\ell+1)}=\Ti{\FPV}{}{\ell}+(1-\gamma^2)\le[(1-\gamma^2)\le(\frac{3A_4}{A_2^2}-1\ri)+4\gamma^2\ri]\le(\ker^{\star}\ri)^2\, ,
\ee
which gives a 
simple additive solution 
of the form \eqref{eq:scaling-solution-reprinted-for-MIT-reprint}, with
\be\label{eq:gamma-nu-scale-invariant}
\nu=\nu(\gamma)\equiv (1-\gamma^2)\le[(1-\gamma^2)\le(\frac{3A_4}{A_2^2}-1\ri)+4\gamma^2\ri]\, .
\ee
As a quick check, we see that $\nu$ reduces to our previous result for vanilla MLPs, \eqref{eq:scale-invariant-nu}, 
as $\gamma\to0$. More importantly,  $\nu(\gamma)$ is strictly positive in the allowed range $0 < \gamma^2 < 1$ and monotonically decreases to $\nu(1)=0$ with increasing $\gamma$.\footnote{To see this, you'll need to use the fact that
$A_4 \geq A_2^2$ for all scale-invariant activation functions.
}

\subsubsection{$K^\star=0$ Universality Class}
For the $K^{\star}=0$ universality class\index{universality class!K@$K^\star=0$},  we tune the rescaled weight variance to criticality with \eqref{eq:criticality_scale_invariant-K-star}.  In this case, the single-input asymptotic behavior of the kernel is modified. Evaluating the residual-MLP kernel recursion \eqref{K0-Res} at criticality and recalling our expansion \eqref{eq:g0}, 
\be
g(\ker)
= \sigma_1^2\le[K+a_1 K^2+\o{K^3}\ri] \,,
\ee
the single-input kernel recursion becomes
\be
\Ti{\ker}{}{\ell+1}= \Ti{\ker}{}{\ell} + a_1(1-\gamma^2)\le(\Ti{\ker}{}{\ell}\ri)^2+\ldots\,  ,\label{K0-Res-K-star}
\ee
which has deep asymptotic behavior of
\begin{align}
\ker^{(\ell)}=&\le[\frac{1}{(-a_1)(1-\gamma^2)}\ri]\frac{1}{\ell}+\ldots\, .
\end{align}
To evaluate the four-point vertex recursion, we need 
\begin{align}\label{eq:chi-parallel-agebriac-reprint}
\chi_{\parallel}(K)&=1 + (1-\gamma^2) \le[2a_1\ker+\o{\ker^2}\ri]\, ,\\
\label{eq:first-part-of-vertex-algebriac-reprint}
\le[\bra \sigma^4\ket_{\ker^{\star}}-\bra \sigma^2\ket_{\ker^{\star}}^2\ri]&= \sigma_1^4 \le[ 2 \ker^2+\o{\ker^3}\ri]\, , 
\end{align}  
where the first expression is shifted from \eqref{eq:chi-parallel-agebriac}, as per \eqref{eq:chi-parallel-Res} and \eqref{eq:criticality_scale_invariant-K-star}, but the second expression is the same as before \eqref{eq:first-part-of-vertex-algebriac}. Plugging in these expressions and matching terms, we find
\be\label{eq:gamma-nu-k-star}
\nu= \nu(\gamma) \equiv \frac{2}{3}(1-\gamma^4)\, .
\ee
This also reduces to our previous result for vanilla $K^\star=0$ MLPs, \eqref{eq:tanh-nu}, 
as $\gamma\to0$, and also is strictly positive in the allowed range $0 < \gamma^2 < 1$, monotonically decreasing to $\nu(1)=0$ with increasing $\gamma$. 
Thus, $\nu(\gamma)$ for $\ker^\star=0$ universality class is qualitatively comparable to
the scale-invariant universality class.

\subsubsection{Physics of the Optimal Aspect Ratio}\index{optimal aspect ratio}
As we saw at the end of \S\ref{sec:information-beyond-infinity}, the maximization of the $1/n^3$ mutual information \eqref{eq:MIT-third-order} according to our \terminate{unsupervised learning} criterion led to a natural estimate  of the optimal aspect ratio of a network \eqref{eq:optimal-aspect-ratio-natural-choice}. For residual MLPs, this gives
\be\label{eq:Banks-Zaks-almost-reprint}
r^{\star}(\gamma) \equiv \le( \frac{4}{20+3 n_L}\ri)\frac{1}{\nu(\gamma)} \, ,
\ee
with $\nu(\gamma)$ given by \eqref{eq:gamma-nu-scale-invariant} for the scale-invariant universality class and \eqref{eq:gamma-nu-k-star} for the $K^\star=0$ universality class. Thus, we see that the monotonic decrease of $\nu(\gamma)$ for either class leads to a monotonic increase of $r^{\star}(\gamma)$ as $\gamma$ increases: beginning from vanilla MLPs at $\gamma=0$, residual MLPs will prefer larger and larger aspect ratios.
Thus, we have realized our theoretical prediction at the beginning of the section, finding  support in our effective theory formalism for the hypothesis that residual connections\index{residual connection} can solve the \terminate{degradation problem}.

Let us offer a further interpretation of this mechanism.
In \cite{resnet-as-ensemble}, it was suggested that %
the benefit of 
residual connections\index{residual connection} 
in extremely
deep networks 
is that 
they
let the global network architecture 
behave as an \neo{ensemble} of many shallower networks:
in this interpretation, each \emph{particular} ``shallow network'' is given by following the path of a signal from the input to the output of the single residual network.
What was discovered is that gradients are dominated by paths skipping over most of the network
 and only passing through a small fraction of the 
hidden-layer residual blocks\index{residual block}. This actually gives an interesting way to look at our result \eqref{eq:Banks-Zaks-almost-reprint}: if our maximization of mutual information \eqref{eq:MIT-third-order} is weighing the helpful effect of depth %
as an \terminate{inductive bias} for
\terminate{neural association} (\S\ref{subsec:Hebbian}) against the harmful effect of depth due to growing fluctuations (\S\ref{sec:signal_prop_finite_width}), then such ensembling\index{ensemble} would be a natural way to suppress the fluctuations while preserving the neural association. Then, using residual connections to extend the depth of \emph{trainable} networks should, at least to a point, also lead to better test set performance.

Note finally that we can actually interpret \eqref{eq:Banks-Zaks-almost-reprint} in another way: rather than estimating the \terminate{optimal aspect ratio} $r^\star(\gamma)$ for a fixed residual hyperparameter\index{residual hyperparameters} $\gamma$, \emph{instead} we can think of it as estimating the optimal residual hyperparameter\index{residual hyperparameters!optimal} $\gamma^\star(r)$ for a fixed aspect aspect ratio $r$. 
Taking this latter perspective, we learn something interesting. On the one hand, for very shallow network with $r \ll 1$,  our criterion \eqref{eq:Banks-Zaks-almost-reprint} is unsatisfiable since $\nu(\gamma)$ monotonically decreases from its maximal value at $\gamma=0$.  Such networks are shallower than optimal according to our original criterion \eqref{eq:optimal-aspect-ratio-natural-choice} for vanilla MLPs, $r <r^\star(\gamma=0)$, and their mutual information will be greatest \emph{without}  the residual connections\index{residual connection} $\gamma^\star(r)=0$.
On the other hand, for very deep networks with $r >r^\star(\gamma=0)$, the optimal residual hyperparameter $\gamma^\star(r)$ monotonically asymptotes to $1$ with growing $r$.\footnote{
    When setting $\gamma^\star$ it's important to always remember to also adjust the rescaled weight variance $C_W(\gamma^\star)$  according to \eqref{eq:criticality_scale_invariant-Res} or \eqref{eq:criticality_scale_invariant-K-star} in order to maintain criticality.
In the limit of $r \gg r^\star(\gamma=0)$, the optimal residual hyperparameter asymptotes to one as $1-[\gamma^\star(r)]^2\sim 1/r$ and the critical initialization gets smaller as $C_W\big(\gamma=\gamma^\star(r)\big)\sim 1/r$.
} 
Altogether, this suggests that we should only turn on the residual connections for networks with aspect ratios $r$ that are greater than the threshold $r^\star(\gamma=0)$ set by the \terminate{optimal aspect ratio} of vanilla MLPs.

Incidentally, if you are worried about the validity of perturbation theory for large $r$, note that the real physical expansion parameter is given by the combination 
\be
\frac{ V^{(L)} }{ n \le(G^{(L)}\ri)^2}= \nu\Big(\gamma=\gamma^\star(r)\Big) \times r \,, 
\ee
which stays finite even as the ratio $r$ asymptotes to infinity. In this sense, we can use $\gamma^\star(r)$ to arbitrarily extend the regime of effectively-deep\index{effectively deep} networks that can be described by our effective theory.

\section{Residual Building Blocks}\label{sec:residual-emprical}
In this final section, we will explain a hybrid theoretical-empirical method for tuning a very general residual network to criticality.
This method may be implemented practically 
in order to tune
the hyperparameters of residual networks beyond the multilayer perceptron\index{multilayer perceptron!beyond} architecture. 
We hope this discussion provides a \neo{blueprint} for how a few simple measurements on small networks can then be scaled up to efficiently 
design much larger models according to our effective theory approach\index{effective theory}.

A \emph{general} \terminate{residual network}\index{residual network!general} can be defined by replacing a simple MLP-layer block\index{multilayer perceptron!with residual connections}~\eqref{eq:ResMLP} 
by a generic nonlinear \term{residual block}:
\be\label{eq:generic-residual-block}
\bias{i}{\ell+1}+\sum_{j=1}^{n}\W{ij}{\ell+1}\s{j}{\delta}{\ell}\, \to\  \block_i\!\le(z^{(\ell)}_{\delta}; \theta^{(\ell+1)}\ri)\equiv \block_{i;\delta}^{(\ell+1)}\, .
\ee
Here, the $\ell$-th-layer residual block $\block_{i;\delta}^{(\ell)}$\index{forward equation!general residual network} is shaped by model parameters\index{model parameters!residual network} $\theta^{(\ell)}$, and we should pick $\block_{i;\delta}^{(\ell)}$ to be a square matrix in its neural indices, $n_{\ell}=n_{\ell+1}\equiv n$, so that we can add the output of the residual block back to the preactivation.
This leads to the following forward equation,
\be\label{eq:generic-residual-block-layer}
\z{i}{\delta}{\ell+1}=\xi^{(\ell+1)} \block_i\!\le(z^{(\ell)}_{\delta}; \theta^{(\ell+1)}\ri)+\gamma^{(\ell+1)} \z{i}{\delta}{\ell}\, ,
\ee
where
we've 
restored the second residual hyperparameter\index{residual hyperparameters}, $\xi^{(\ell)}$,  in order to let us scale the overall magnitude of the residual-block term. %
This iteration equation   \eqref{eq:generic-residual-block-layer} is sufficiently generic to schematically capture 
many popular deep learning architectures, including the \terminate{computer vision} workhorse architecture, the \emph{residual convolutional network} or \term{ResNet}, and the multi-headed self-attention-seeking language-modeling\index{natural language processing} \term{transformer}.\index{language model}
A graph of two residual blocks in adjacent layers from 
a general residual network described by \eqref{eq:generic-residual-block-layer}
is shown in the left panel of Figure~\ref{fig:residual-example}.

In practice, certain heuristics and ad hoc methods are used in these general architectures to try and mitigate the exploding and vanishing gradient problem\index{exploding and vanishing gradient problem!for residual networks}. However, as we know from \S\ref{sec:EVGP-WEP}, \neo{criticality} is a much more motivated solution. In \S\ref{sec:residual-criticality}, we saw that for residual MLPs\index{multilayer perceptron!with residual connections}, the residual hyperparameters can be used to control criticality for the network. Now, let's see how we can implement a form of criticality for our general residual network \eqref{eq:generic-residual-block-layer}.

Broadly speaking, we now need to solve two problems of different nature and difficulties:
\bi
\item First, we need to ensure that signals can easily propagate through the residual block, especially for blocks that are \emph{deep} in some sense; for instance, if a block individually consists of $L_\block$ MLP layers, i.e.~\emph{an MLP-$L_\block$-layer block}, it will then have an internal version of the \terminate{exploding and vanishing gradient problem}. Theorists should make every effort to critically analyze such blocks  of practical interest, but if we have to treat the residual block $\block_{i;\delta}^{(\ell)}$ as a black  box for one reason or another, then 
this will require some \neo{engineering}:
 you need to measure the two-point block-block correlator at the output of a block
\be\label{eq:block-block-correlator}
\E{\block_{i_1;\delta_1}^{(\ell+1)}\block_{i_2;\delta_2}^{(\ell+1)}}\, ,
\ee
and then compare it with the two-point correlator of the input to the block
\be\label{eq:residual-two-point-general-correlator}
\E{\z{i_1}{\delta_1}{\ell}\z{i_2}{\delta_2}{\ell}}\, .
\ee
To be brief and concrete here, we'll focus on their diagonal components with $i_1=i_2\equiv i$ and $\delta_1=\delta_2\equiv\delta$. In particular, let us take an average over neural indices $i=1,\ldots,n$ as well as over the sample indices $\delta\in\D$, so that we can compare two scalar quantities
\be
\overline{G}_{\block\block}^{(\ell+1)}\equiv \frac{1}{\vert\D\vert n}\sum_{i=1}^{n}\sum_{\delta\in\D}\E{\block_{i;\delta}^{(\ell+1)}\block_{i;\delta}^{(\ell+1)}}\, ,\qquad \overline{G}_{zz}^{(\ell)}\equiv \frac{1}{\vert\D\vert n}\sum_{i=1}^{n}\sum_{\delta\in\D}\E{\z{i}{\delta}{\ell}\z{i}{\delta}{\ell}}
\ee
rather than two matrices, \eqref{eq:block-block-correlator} and \eqref{eq:residual-two-point-general-correlator}\index{$\gamma^{[a]}$ basis!kernel!for hybrid approach}.\footnote{
    More generally, we should also account for off-diagonal components by averaging over pairs of inputs to estimate the appropriate analogs of $\Ti{\ker}{[2]}{\ell}$, cf.~\eqref{eq:K2-decomposition}. We'd then also want to ensure that the analog of the recursion for this component, e.g.~\eqref{K2-Res}, is preserved. %
    As our discussion in these bullets is somewhat schematic, we will leave these details to the \terminate{PyTorch} documentation.
}
After setting up these measurements, we need to adjust the \neo{initialization hyperparameters} for the block such that these 
quantities
have similar magnitudes. For instance, if internally the block is parameterized by many iterative layers -- like our MLP-$L_\block$-layer block -- then we need to make sure that the difference in magnitudes is no worse than a \emph{polynomial} in this depth hyperparameter $L_\block$; importantly, we want to ensure that there's no exponential growth or decay in $\overline{G}_{\block\block}^{(\ell+1)}/\overline{G}_{zz}^{(\ell)}$.\footnote{
    One reasonable heuristic solution to this problem, used by the \neo{transformer} architecture, could be \neo{layer normalization} \cite{ba2016layer}. However, more ideally, the block is not treated as a black box, and instead we use something about the structure of the block itself to find the \terminate{criticality} conditions.
} 
Note that while you're setting up measurements, it will also be helpful to measure the cross correlator
\be
\overline{G}_{\block z}^{(\ell+0.5)}\equiv \frac{1}{\vert\D\vert n}\sum_{i=1}^{n}\sum_{\delta\in\D} \E{\block_{i;\delta}^{(\ell+1)}z_{i;\delta}^{(\ell)}} \, ,
\ee
which will be needed in the next bullet point.
\item Now, using the forward equation \eqref{eq:generic-residual-block-layer}, we can write a recursion for the diagonal component of the two-point correlator as
\be
\overline{G}_{zz}^{(\ell+1)}=\gamma^2\, \overline{G}_{zz}^{(\ell)}+2\gamma\xi\, \overline{G}_{\block z}^{(\ell+0.5)} +\xi^2\, \overline{G}_{\block\block}^{(\ell+1)} \, ,
\ee
where we've suppressed the layer indices in the residual hyperparameters to declutter the expression.\footnote{
    Note that unlike the MLP-single-layer block we discussed in \S\ref{sec:residual-criticality}, we cannot in general simply scale away $\xi$ by a rescaling of $C_W$: for instance, if we have a deep MLP-$L_\block$-layer block with $L_\block\gg1$ built with $\relu$ activations, then it is probably easier to set $C_W=2$ 
    and adjust $\xi$ at the end. %
    Thus, in general we should think of the initialization hyperparameters of the block as being fixed \emph{first} -- to ensure criticality of the block internally -- and \emph{then} for each $\gamma$, we tune $\xi (\gamma)$ according to \eqref{eq:criticality-residual-general} to ensure criticality of the whole network.
}
To preserve this component of the two-point correlator of preactivations, we set the right-hand side of this equation equal to $\ell$-th-layer correlator. %
Rearranging, we thus want
\be\label{eq:criticality-residual-general}
\le(1-\gamma^2\ri)\overline{G}_{zz}^{(\ell)} =\xi^2\, \overline{G}_{\block\block}^{(\ell+1)}+2\gamma\xi\, \overline{G}_{\block z}^{(\ell+0.5)}  \, .
\ee
Since we've supposedly measured all these quantities, this equation should give simple \emph{analytical} solutions for the residual hyperparameters.\footnote{
Accordingly, each critical solution $\le(\gamma, \xi \ri)$ to the equation \eqref{eq:criticality-residual-general} 
will then yield architectures with different optimal aspect ratios\index{optimal aspect ratio} $r^\star(\gamma, \xi)$. The \terminate{optimal aspect ratio} for a particular $(\gamma, \xi)$  can be analogously estimated for general residual networks with a hybrid approach by combining a theoretical analysis as we did in \S\ref{sec:residual-MI} with  measurements of an appropriate combination of four-point connected correlators. 
}
\ei

Overall, this %
\terminate{hybrid approach} realizes one of our goals of using experimentally measurable observables\index{observable} as input to an effective theory analysis. We hope that similar ways of thinking will lead to powerful ways of designing and tuning deep-learning models in the future.